# Evaluating Loss Functions for Graph Neural Networks: Towards Pretraining and Generalization

KHUSHNOOD ABBAS*, School of Computer Science and Technology, Zhoukou Normal University, China

RUIZHE HOU, School of Automation Science and Engineering,South China University of Technology, China

ZHOU WENGANG, DONG SHI, NIU LING, School of Computer Science and Technology, Zhoukou Normal University, China

SATYAKI NAN, College of Business and Computing, Georgia Southwestern State University, USA

ALIREZA ABBASI, School of Engineering and IT, University of New South Wales (UNSW Canberra), Australia

Graph Neural Networks (GNNs) became useful for learning on non-Euclidean data. However, their best performance depends on choosing the right model architecture and the training objective, also called the loss function. Researchers have studied these parts separately, but a large-scale evaluation has not looked at how GNN models and many loss functions work together across different tasks. To fix this, we ran a thorough study - it included seven well-known GNN architectures. We also used a large group of 30 single plus mixed loss functions. The study looked at both inductive and transductive settings. Our evaluation spanned three distinct real-world datasets, assessing performance in both inductive and transductive settings using 21 comprehensive evaluation metrics. From these extensive results (detailed in supplementary information 1 & 2), we meticulously analyzed the top ten model-loss combinations for each metric based on their average rank. Our findings reveal that, especially for the inductive case: 1) Hybrid loss functions generally yield superior and more robust performance compared to single loss functions, indicating the benefit of multi-objective optimization. 2) The GIN architecture always showed the highest-level average performance, especially with Cross-Entropy loss. 3) Although some combinations had overall lower average ranks, models such as GAT, particularly with certain hybrid losses, demonstrated incredible specialized strengths, maximizing the most top-1 results among the individual metrics, emphasizing subtle strengths for particular task demands. 4) On the other hand, the MPNN architecture typically lagged behind the scenarios it was tested against. This study provides crucial empirical rationale supporting the selection of GNN structure and loss function, pointing out that optimal performance is not monolithic but arises from synergistic interaction, which suggests that a tailor-made approach, considering both general effectiveness and specialist strengths, is critical to the successful application of GNNs.

Authors' Contact Information: Khushnood Abbas, Khushnood.abbas@zknu.edu.cn, School of Computer Science and Technology, Zhoukou Normal University, Zhoukou, Henan, China; Ruizhe Hou, auruizhe@mail.scut.edu.cn, School of Automation Science and Engineering,South China University of Technology, Guangzhou, Henan, China; Zhou Wengang, Dong Shi, Niu Ling, School of Computer Science and Technology, Zhoukou Normal University, Zhoukou, Henan, China; Satyaki Nan, satyaki.nan@gsw.edu, College of Business and Computing, Georgia Southwestern State University, Americus, GA, USA; Alireza Abbasi, a.abbasi@unsw.edu.au, School of Engineering and IT, University of New South Wales (UNSW Canberra), Canberra, ACT, Australia.

## 1 Introduction to Unsupervised Node Representation Learning

Unsupervised graph representation learning (UGRL) is an important area in machine learning that
focuses on transforming complex, high-dimensional, and often sparse graph data into compact,
dense vector representations [23]. The main goal is to capture and summarize information from
graphs in a way that can be broadly useful for different downstream tasks—without relying on
labeled data during training [23]. In essence, UGRL aims to map core elements of a graph—like
nodes, edges, or even entire substructures—into a lower-dimensional space, while still preserving
the key relationships and patterns that exist in the original graph [17]. For node embeddings
specifically, the objective is to learn a mapping function $f : v_i \rightarrow \mathbb{R}^d$ that projects each node $v_i$
into a low-dimensional vector of size $d$, where $d \ll |V|$, the total number of nodes. The learned
embeddings should reflect the similarity relationships between nodes as they appear in the original
graph structure [4, 17]. A defining characteristic of unsupervised methods is their ability to generate
general-purpose embeddings, meaning they are not tailored or optimized for any single downstream
task. This contrasts with semi-supervised approaches, which are typically trained with specific
applications in mind and therefore produce task-specific embeddings [4, 28].

The utility of low-dimensional node embeddings is substantial, as they provide powerful feature
representations that help bridge the gap between traditional machine learning algorithms and the
complex, interconnected nature of graph-structured data [28]. These embeddings play a crucial
role in a wide range of predictive tasks, including node classification, link prediction (both missing
and future connections), community detection, and anomaly detection [14, 17, 22, 26].One of the
key advantages of graph embeddings is their ability to reduce the complexity of graph mining
tasks by transforming them into more manageable problems in a continuous vector space. This
transformation enables more efficient application of artificial intelligence and machine learning
techniques on graph data [22]. In recent years, the field of unsupervised graph representation
learning has seen rapid growth and innovation, producing strong results across a wide range of
graph analysis tasks [26]. Current approaches in this area can be broadly grouped into several
main categories: random walk-based methods, matrix factorization techniques, deep learning-based
frameworks (such as autoencoders and Graph Neural Networks), and, more recently, contrastive
learning-based methods [17, 22, 28]. Each of these paradigms uses distinct techniques and objective
functions tailored to their specific learning goals.

A notable theme in the evolution of unsupervised graph representation learning is the inherent
tension between the theoretical goal of producing truly *downstream task-agnostic* embeddings
and the practical realities of model design [23]. While the central objective is to generate repre-
sentations that are independent of any specific task, many existing methods—either implicitly or
explicitly—introduce inductive biases that enhance performance for particular types of applications.
For instance, the PairE method was explicitly developed to address limitations in earlier techniques
that performed well on node-centric tasks but struggled with edge classification. This design reflects
a deliberate effort to broaden the applicability of embeddings across different task types [23]. It
illustrates a common trade-off in practice: rather than aiming for an idealized, fully general-purpose
embedding, many approaches incorporate cost functions that balance task-independence with improved performance across a spectrum of commonly encountered downstream tasks. For researchers and practitioners, this highlights the importance of selecting unsupervised methods with an awareness of both the graph characteristics and the nature of the intended applications. In reality, even methods described as "task-agnostic" can vary significantly in effectiveness across different tasks, depending on the structural or feature-level signals that their objective functions prioritize capturing.

## 2 Categories of Unsupervised Node Embedding Methods

Unsupervised node embedding methods can be broadly classified into several categories, each employing distinct mechanisms and associated cost functions to learn low-dimensional representations of nodes in a graph.

### 2.1 Random Walk-based Methods

These approaches draw inspiration from the successes observed in natural language processing (NLP) [23]. They operate by simulating random walks across the graph, which generate sequences of nodes similar to "sentences" in text [28]. Subsequently, models like the Skip-Gram model, originally developed for word embeddings, are applied to these sequences to learn node representations [4]. The fundamental objective is to maximize the likelihood of observing neighboring nodes that appear within these generated random walks, given the embedding of the central node [4, 12]. Prominent examples within this category include DeepWalk, Node2vec, and LINE [12, 30, 34]. These methods effectively transform the complex, non-Euclidean structure of a graph into a linear sequence format, enabling the application of well-established and computationally efficient sequence modeling techniques [19, 31, 48]. This transformation is a pivotal conceptual leap, simplifying the problem and making it amenable to existing machine learning tools. The effectiveness of such graph representation learning methods often stems not only from novel algorithms but also from innovative ways of re-framing the input data. However, this approach also introduces the challenge of understanding how well these "linearized" sequences truly capture the full, multi-relational complexity of the original graph, as certain structural nuances might be lost in the linear projection.

### 2.2 Matrix Factorization-based Methods

Matrix factorization methods learn low-dimensional node embeddings by decomposing matrices that capture node-to-node similarity or proximity within a graph [31]. These input matrices may include the graph Laplacian, incidence matrices, the adjacency matrix $A$, or its polynomial expansions [31]. The core idea is to factorize one of these matrices to extract meaningful vector representations for each node. Representative algorithms in this category include *GraRep*, *HOPE*, *NetMF*, and *M-NMF* [17].

Although random walk-based and matrix factorization methods are often presented as distinct classes, deeper theoretical analysis reveals a strong connection between them. Research has shown that several random walk-based approaches—such as *DeepWalk* and *LINE*—implicitly perform matrix factorizations [31, 50]. This insight uncovers a fundamental mathematical equivalence between these two paradigms. Despite their operational mechanisms differing—one relying on stochastic processes and the other on direct algebraic decomposition—they converge on optimizing similar underlying objectives related to graph proximity. This theoretical unification simplifies the conceptual landscape of unsupervised graph embedding, providing a more cohesive understanding of the diverse methods. It also opens avenues for future research, potentially leading to novel hybrid approaches that strategically combine the strengths of both explicit matrix factorization (e.g., direct control over proximity measures) and random walk-based methods (e.g., scalability through sampling) [50].

## 2.3 Deep Neural Network-based Methods (Autoencoders, GNNs)

This category harnesses the power of deep learning architectures, such as autoencoders [21], Siamese Graph networks [1, 24], or Graph convolutional Neural Networks (CNNs), to learn node representations directly from the graph's inherent link structure in an unsupervised fashion [1]. Graph Neural Networks (GNNs) stand out as a particularly effective subcategory, where node embeddings are iteratively refined by aggregating information from a node's immediate neighbors and, through multiple layers, from multi-hop neighbors. Notable examples include SDNE, which employs a non-GNN autoencoder, as well as Graph Autoencoders (GAE) and Variational Graph Autoencoders (VGAE) [17, 47].

The evolution from shallow to deep architectures for graph representation learning marks a significant advancement in capturing non-linearity. Early unsupervised methods, including some matrix factorization techniques and simpler deep learning models, often operated under linear assumptions or possessed limited model capacities [23]. The advent and widespread adoption of deep neural networks, particularly GNNs, allowed for the capture of complex, non-linear structural information inherent in graph data, which shallow models struggled to represent [3]. This transition mirrors a broader trend in machine learning, where deep learning has proven superior in extracting hierarchical and abstract features from complex data. The increased model capacity provided by deep learning methods enables the design of more intricate and expressive cost functions that can capture a wider array of graph properties. However, this advancement also introduces new challenges related to increased computational demands, the complexity of optimizing these models, and a reduction in the interpretability of the learned representations.

## 2.4 Contrastive Learning Methods

Contrastive learning has emerged as a dominant and highly effective paradigm within self-supervised learning (SSL) for graphs [16, 41]. Its central principle is to learn discriminative representations by contrasting data samples based on their semantic similarity [41, 53]. Specifically, the goal is to maximize agreement between augmented views of the same graph instance—termed *positive pairs*—while minimizing agreement with views derived from different instances, known as *negative pairs* [16, 53]. Common implementations include the InfoNCE loss, as well as more recent methods that incorporate optimal transport distances to quantify similarity more effectively [41, 46]. This contrastive framework generalizes traditional proximity-based objectives found in earlier unsupervised graph embedding methods, such as DeepWalk and LINE, which implicitly embed similar nodes—based on co-occurrence in random walks or direct edge connections—closely in the latent space. Contrastive learning formalizes this idea by explicitly defining positive and negative pairs, often using graph augmentations to capture deeper semantic relationships beyond raw structural proximity [41, 53]. The corresponding loss functions are designed to bring positive pairs closer in representation space while pushing negative pairs apart, offering a more flexible and robust way to enforce similarity constraints. This paradigm shift enables the model to learn more generalizable representations by leveraging a wider spectrum of self-supervised signals. Rather than relying on fixed, predefined notions of similarity, contrastive learning adapts to the data, allowing the model to infer what constitutes meaningful similarity in a task-specific and data-driven manner.

## 3 Detailed Analysis of Cost Functions

In this section, we examine how the efficacy of unsupervised node representation learning hinges significantly on the design and optimization of its cost functions. These functions dictate how the model learns to capture and preserve the intricate structural and semantic properties of graphs in a low-dimensional embedding space. This section also delves into the specific cost functions employed by various prominent unsupervised node embedding methods.

## 3.1 Random Walk-based Methods: Maximizing Contextual Likelihood

The foundational principle of random walk-based methods is to learn node representations such that nodes frequently co-occurring within a defined "context"—typically generated by random walks on the graph—exhibit similar embeddings in the low-dimensional space. This objective is mathematically framed as maximizing the likelihood of observing a node's network neighborhood given its feature representation.

*3.1.1 DeepWalk: Skip-Gram Objective with Negative Sampling.* DeepWalk initiates the embedding process by generating multiple fixed-length random walks starting from each node in the graph [12, 30]. It then adapts the Skip-Gram model, a core component of the Word2vec framework, to learn low-dimensional node embeddings from these generated sequences. The learning objective is to maximize the likelihood of observing nodes within a fixed-size window surrounding a target node in each random walk, conditioned on the embedding of the target node [12, 30]. The mathematical formulation for a target node $u$ and its observed neighborhood $N_S(u)$ is expressed as maximizing the following log-probability [12]:

$$L = \sum_{u \in V} \sum_{n_i \in N_S(u)} \left[ \log \sigma(\mathbf{h}_{n_i}^\top \mathbf{h}_u) + k \cdot \mathbb{E}_{v_j \sim P_n(v)} [\log \sigma(-\mathbf{h}_{v_j}^\top \mathbf{h}_u)] \right] \tag{1}$$

Here, $\mathbf{h}_u$ and $\mathbf{h}_{n_i}$ denote the learned embeddings of node $u$ and its neighbor $n_i$, respectively. The sigmoid function, $\sigma(x) = 1/(1 + \exp(-x))$, maps dot products to probabilities. The parameter $k$ signifies the number of negative samples drawn for each positive pair. $P_n(v)$ is the noise distribution utilized for negative sampling, which is empirically often set to $P_n(v) \propto d_v^{3/4}$, where $d_v$ is the degree of node $v$ [12, 31, 44]. The first term in this objective function encourages a high similarity (large dot product) between the target node and its positive (contextual) neighbors. Conversely, the second term penalizes high similarity (minimizes dot product) between the target node and randomly sampled negative (non-contextual) nodes.

This loss function is critical because it effectively transforms the complex problem of learning graph proximity into a series of computationally efficient binary classification problems, where the model distinguishes between true context nodes and randomly sampled noise nodes. This formulation, particularly with the integration of negative sampling, is crucial for scalability, as directly computing the softmax over all nodes in large graphs would be computationally prohibitive [11, 12, 49].

The pervasive success of negative sampling underscores a fundamental engineering principle in large-scale machine learning: it often involves sacrificing a theoretically exact objective for a computationally tractable and empirically effective approximation. The canonical Skip-Gram objective function includes a normalization term in its softmax denominator that requires summing over all nodes in the vocabulary (or graph), which becomes computationally infeasible for very large networks [12, 49]. Negative sampling directly addresses this bottleneck by replacing the full summation with a small, fixed number of randomly sampled "negative" examples [11, 49]. This approximation significantly reduces the computational cost, making the optimization feasible for large graphs [12]. Its widespread adoption across various representation learning models, beyond just DeepWalk, highlights its fundamental importance as an algorithmic innovation for achieving scalability in graph embedding.

*3.1.2 Node2vec: Biased Random Walks and Objective Function.* Node2vec extends DeepWalk by introducing a more flexible and biased random walk procedure [11, 12, 17]. This procedure is meticulously designed to allow for a smooth interpolation between Breadth-First Search (BFS)-like and Depth-First Search (DFS)-like exploration strategies [11, 12, 17]. The overarching objective remains to maximize the likelihood of preserving network neighborhoods, but with a significantly richer and more adaptable definition of what constitutes a "neighborhood" [12]. The underlying objective function for learning embeddings is the same Skip-Gram likelihood used in DeepWalk, but the critical difference lies in *how the neighborhoods $N_S(u)$ are generated* [12].

The transition probability $P(c_i = x|c_{i-1} = v)$ for the random walk is controlled by two key parameters: $p$, the return parameter, and $q$, the in-out parameter [11, 12]. The return parameter, $p$, governs the likelihood of the walk immediately revisiting a node it just came from. A high value of $p$ (e.g., $p > \max(q, 1)$) makes it less likely to sample an already-visited node in the next two steps, thereby encouraging a more moderate exploration of the graph and avoiding redundant 2-hop paths. Conversely, a low value of $p$ (e.g., $p < \min(q, 1)$) biases the walk to backtrack, keeping the exploration more "local" to the starting node $u$. The in-out parameter, $q$, allows the search to differentiate between "inward" (local) and "outward" (global) nodes relative to the previous node in the walk. If $q > 1$, the random walk is biased towards nodes closer to the previous node $t$, leading to a local view of the graph that approximates BFS behavior, where samples primarily comprise nodes within a small locality. If $q < 1$, the walk is more inclined to visit nodes further away from node $t$, reflecting a DFS-like behavior that encourages outward exploration [11, 12].

Node2vec's significant contribution is not a novel cost function, but rather a sophisticated *sampling strategy* for generating the positive pairs that feed into the existing Skip-Gram objective [12]. *This demonstrates that the effectiveness of a representation learning method is not solely determined by the mathematical form of its loss function. Instead, it is profoundly influenced by how the input data, specifically positive and negative pairs, are constructed to reflect desired graph properties.* For instance, emphasizing homophily through BFS-like walks or structural equivalence through DFS-like walks directly impacts the learned representations. The sampling process implicitly defines the "neighborhood" that the cost function then attempts to preserve. This highlights a critical principle in self-supervised learning: the design of the "pretext task"—the mechanism by which self-supervisory signals are generated, often through data augmentation or sampling—is as important as the loss function itself in guiding the model to learn specific, useful representations. Future advancements might involve more sophisticated, adaptive sampling strategies that dynamically adjust based on the evolving characteristics or specific requirements of the graph.

*3.1.3 LINE: First-Order and Second-Order Proximity Loss Functions.* LINE (Large-scale Information Network Embedding) is designed to preserve two complementary notions of node proximity in graph-structured data. First-order proximity captures the observed connections between directly linked nodes, while second-order proximity models similarity in shared neighborhood distributions, effectively capturing structural equivalence [17, 34? ]. One of LINE's notable strengths lies in its versatility as it is applicable to a wide range of information network types, including undirected, directed, and weighted graphs, making it well-suited for diverse real-world information networks [34].

The first-order proximity loss ($L_1$) focuses on preserving the strength of direct connections between nodes. It models the probability of an edge existing between two nodes ($i, j$) using a sigmoid function applied to the dot product of their embeddings. The objective is to maximize this probability for existing edges:

$$L_1 = - \sum_{(i,j) \in E} w_{ij} \log p_1(v_i, v_j) \quad \text{where} \quad p_1(v_i, v_j) = \frac{1}{1 + \exp(-\mathbf{h}_i^\top \mathbf{h}_j)} \tag{2}$$

Here, $w_{ij}$ represents the weight of the edge between nodes $i$ and $j$, and $\mathbf{h}_i, \mathbf{h}_j$ are their respective embeddings. This objective encourages embeddings of directly connected nodes to be similar.

The second-order proximity loss ($L_2$) addresses the sparsity of direct connections by focusing on shared neighborhood structures. It models the probability of a node $j$ being a "context" of node $i$ (i.e., sharing common neighbors) using a separate "context embedding" for node $j$. The objective aims to make nodes with similar neighborhood distributions have similar embeddings [34]:

$$L_2 = -\sum_{i \in V} \sum_{j \in N(i)} \frac{w_{ij}}{\sum_{k \in N(i)} w_{ik}} \log p_2(v_j|v_i) \qquad (3)$$

where $p_2(v_j|v_i) = \frac{\exp(\mathbf{h}_j^\top \mathbf{h}_i')}{\sum_{k \in V} \exp(\mathbf{h}_k^\top \mathbf{h}_i')}$. In this formulation, $\mathbf{h}_i$ is the node embedding for $i$, and $\mathbf{h}_i'$ is the context embedding for $i$. The computational cost of the denominator is typically mitigated using negative sampling.

Explicitly addressing both first-order (local) and second-order (global) proximities is crucial for capturing a comprehensive view of the graph's structure, especially in sparse networks where direct links alone might provide insufficient information for robust representation learning [34]. This dual objective allows LINE to learn richer embeddings that are more informative for a variety of downstream tasks. LINE's design explicitly incorporates two distinct loss components ($L_1$ and $L_2$) to capture different granularities of graph proximity: direct connectivity (local) and shared neighborhood patterns (global) [34]. *This is a clear instance of multi-objective optimization within a single framework, acknowledging that different structural properties of a graph require unique inductive biases in the loss function to be effectively preserved. By combining these, LINE aims for a more comprehensive representation than methods focusing on a single type of proximity.* This approach demonstrates the power of designing complex loss functions that integrate multiple, complementary objectives. It allows for the generation of richer embeddings that are more robust to graph sparsity and can generalize better across a wider range of downstream tasks, as different tasks may rely on different levels of structural information (e.g., local versus global context). This sets a precedent for developing more sophisticated loss functions that capture diverse aspects of graph data.

## 3.2 Matrix Factorization-based Methods: Proximity Preservation

The fundamental principle of matrix factorization methods is to identify low-dimensional embeddings by decomposing a matrix that encapsulates some form of node-to-node similarity or proximity within the graph. The associated cost function typically quantifies the discrepancy between this original similarity matrix and the similarity that is reconstructed from the learned low-dimensional embeddings. The optimization aims to minimize this reconstruction error, thereby preserving the inherent graph properties. While various matrix factorization methods exist, such as GraRep, HOPE, NetMF, and M-NMF, the literature often focuses on their general approach of factorizing graph-derived matrices (e.g., adjacency or Laplacian) rather than providing explicit, universally applicable cost function formulations for all of them [31]. However, the core idea remains consistent: minimizing a loss that reflects how well the low-dimensional embeddings can reconstruct the original graph's proximity information.

### 3.2.1 Laplacian Eigenmaps: Graph Laplacian and Eigenvalue Problem.
Laplacian Eigenmaps (LE) is a dimensionality reduction technique deeply rooted in spectral graph theory and manifold learning. Its primary objective is to preserve local neighborhood information [2, 51, 52]. It achieves this by ensuring that if two nodes are close in the original high-dimensional space (typically indicated by an existing edge between them), their corresponding embeddings remain close in the learned low-dimensional space [2, 51, 52]. The method seeks a mapping $f : V \rightarrow \mathbb{R}^m$ that minimizes the following objective function:

$$L = \frac{1}{2} \sum_{i,j} W_{ij} ||\mathbf{f}_i - \mathbf{f}_j||^2 \qquad (4)$$

Here, $W_{ij}$ represents the weight of the edge connecting nodes $i$ and $j$. These weights can be binary (1 if connected, 0 otherwise) or can be derived from a similarity measure, such as Gaussian weights based on the distance between nodes [52]. This objective function inherently imposes a heavy penalty if there are large distances between the embeddings of nodes that are connected by a high-weight edge [2, 52].

This minimization problem can be elegantly reformulated as a generalized eigenvalue problem: $L\mathbf{f} = \lambda D\mathbf{f}$. In this equation, $L = D - W$ is the graph Laplacian matrix, where $D$ is the diagonal degree matrix with entries $D_{ii} = \sum_j W_{ij}$ [2]. The optimal low-dimensional embeddings for the nodes are then obtained from the eigenvectors corresponding to the smallest non-zero eigenvalues of the graph Laplacian [2].

Laplacian Eigenmaps provides a theoretically sound framework for preserving local geometry, drawing its justification from the role of the Laplace-Beltrami operator in manifold learning [2, 52]. Its locality-preserving characteristic makes it relatively insensitive to outliers and noise, and it is not prone to "short-circuiting" issues often seen in global dimensionality reduction methods because it only considers local distances [2].

### 3.3 Deep Neural Network-based Methods: Reconstruction and Generative Models

Deep neural network-based methods leverage the power of deep architectures to learn complex, non-linear relationships in graph data. A common approach within this category is the use of autoencoders, which learn representations by attempting to reconstruct their input.

*3.3.1 SDNE: First-Order and Second-Order Proximity Loss Functions.* SDNE (Structural Deep Network Embedding) is a deep neural model designed to capture both first-order and second-order proximities within a graph [6]. It employs an autoencoder architecture that aims to reconstruct the graph's adjacency matrix from learned node embeddings [17]. The total loss function of SDNE is a joint optimization of two main components:

(1) **Preserving Second-Order Proximity ($L_1$):** This part of the loss function focuses on reconstructing the adjacency matrix, with a particular emphasis on penalizing errors in reconstructing existing connections more heavily than non-connections. It is formulated as:

$$L_1 = \sum_{v_i \in V} ||(\mathbf{x}_i - \mathbf{x}'_i) \odot \mathbf{b}_i||^2 \qquad (5)$$

Here, $\mathbf{x}_i$ represents the row corresponding to node $v_i$ in the graph's adjacency matrix, and $\mathbf{x}'_i$ is its reconstruction. $\mathbf{b}_i$ is a vector where $b_{ij} = \beta > 1$ if an edge exists between $v_i$ and $v_j$ ($a_{ij} = 1$), and $b_{ij} = 1$ if no edge exists ($a_{ij} = 0$) [17]. This weighting scheme ensures that the model prioritizes accurate reconstruction of existing edges, which is crucial for sparse graphs where zeros are abundant [17].

(2) **Capturing First-Order Proximity ($L_2$):** This component encourages connected nodes to have similar embeddings. It is inspired by Laplacian Eigenmaps and penalizes large differences between the embedding vectors of directly connected nodes:

$$L_2 = \sum_{(v_i, v_j) \in E} a_{ij} ||\mathbf{z}_i - \mathbf{z}_j||^2 \qquad (6)$$

In this formula, $\mathbf{z}_i$ and $\mathbf{z}_j$ are the embedding vectors for nodes $v_i$ and $v_j$, respectively, and $a_{ij}$ is the element in the adjacency matrix. This loss component ensures that nodes connected by an edge are mapped to nearby points in the embedding space [17].

By jointly optimizing both $L_1$ and $L_2$, SDNE generates node embedding vectors that capture both the local (first-order) and global (second-order) structural properties of the graph [6, 17].

### 3.3.2 Graph Autoencoders (GAE).

Graph Autoencoders (GAEs), proposed by Kipf and Welling (2016), are prominent unsupervised models that predict link probabilities by computing the inner products of node representations learned through a Message Passing Neural Network (MPNN) [10, 27]. The core objective of GAEs is to reconstruct and preserve the graph topology by mapping nodes into a latent space [10].

The GAE architecture typically consists of a GNN encoder that generates node embeddings and a simple decoder that reconstructs the adjacency matrix using an inner product [10, 15, 27]. The reconstruction loss function, often based on mean squared error or binary cross-entropy, aims to minimize the discrepancy between the original adjacency matrix $A$ and the reconstructed adjacency matrix $\hat{A}$ [10]. The reconstructed adjacency matrix $\hat{A}$ is typically obtained by applying a logistic sigmoid function to the inner product of the node embeddings: $\hat{A} = \sigma(ZZ^\top)$, where $Z$ is the matrix of node embeddings [10, 21].

GAEs learn representations by recovering missing information from incomplete input graphs, following a corruption-reconstruction framework [25]. While effective for link prediction and other tasks, existing GAEs often focus primarily on reconstructing low-frequency information in graphs, potentially overlooking valuable high-frequency signals [25]. This is because their optimization tends to prioritize larger discrepancies, which are more pronounced in low-frequency components [25].

### 3.3.3 Variational Graph Autoencoders (VGAE).

Variational Graph Autoencoders (VGAEs), also introduced by Kipf and Welling (2016), extend the GAE framework by incorporating a probabilistic approach based on the Variational Autoencoder (VAE) [29, 32, 43]. VGAEs aim to learn interpretable latent representations for undirected graphs by modeling the latent variables probabilistically [21].

The VGAE model typically employs two GNN encoders to learn the mean ($\mu$) and variance ($\sigma^2$) of the embedding vectors, assuming a Gaussian distribution for the latent variables [7, 21, 29, 32]. The generative model then reconstructs the adjacency matrix through an inner product between these latent variables, similar to GAE: $p(A_{ij} = 1 | \mathbf{z}_i, \mathbf{z}_j) = \sigma(\mathbf{z}_i^\top \mathbf{z}_j)$ [21].

The learning objective for VGAE is the variational lower bound (Evidence Lower Bound or ELBO), which consists of two main terms [21]:

$$L = \mathbb{E}_{q(\mathbf{Z}|\mathbf{X},\mathbf{A})}[\log p(\mathbf{A}|\mathbf{Z})] - \mathrm{KL}[q(\mathbf{Z}|\mathbf{X},\mathbf{A})||p(\mathbf{Z})] \tag{7}$$

(1) **Reconstruction Term ($\mathbb{E}_{q(\mathbf{Z}|\mathbf{X},\mathbf{A})}[\log p(\mathbf{A}|\mathbf{Z})]$):** This term measures how well the decoder can reconstruct the original graph's adjacency matrix from the sampled latent embeddings. It is typically a binary cross-entropy loss that encourages the model to assign high probabilities to existing edges and low probabilities to non-existent ones.

(2) **KL Divergence Term (KL$[q(\mathbf{Z}|\mathbf{X},\mathbf{A})||p(\mathbf{Z})]$):** This term acts as a regularizer, forcing the approximate posterior distribution $q(\mathbf{Z}|\mathbf{X},\mathbf{A})$ (learned by the encoder) to be close to a predefined prior distribution $p(\mathbf{Z})$ (typically a standard Gaussian distribution) [21, 29]. This regularization helps in learning a smooth and continuous latent space, enabling better generalization and sampling.

VGAEs are effective for tasks like link prediction, but they can face challenges such as posterior collapse, where the model might prioritize reconstruction over learning a meaningful latent distribution, especially when initialized poorly [5].

*3.3.4 Masked Graph Autoencoders (MGAE).* Masked Graph Autoencoders (MGAE) represent a novel framework for unsupervised graph representation learning, drawing inspiration from self-supervised learning techniques like masked autoencoding in other domains [33]. Unlike traditional GAEs that reconstruct the entire graph, MGAE focuses on reconstructing *masked* edges, rather than the observed ones, during training [33]. This forces the GNN encoder to become more robust to network noise and to encode graph information more effectively [33].

MGAE operates by randomly masking a large proportion of edges (e.g., 70% masking ratio) in the input graph structure [33]. The GNN encoder then processes this partially masked graph, and the decoder is specifically designed to reconstruct only the missing (masked) edges [33].

The standard graph-based loss function used to train the MGAE model is defined as [33]:

$$L = - \sum_{(v,u) \in E_{mask}} \log \frac{\exp(y_{vu})}{\sum_{z \in V} \exp(y_{vz})} \tag{8}$$

Where $y_{v,u} = \text{MLP}(h_{e_{v,u}})$ is the reconstructed score for the edge $e_{v,u}$, and $E_{mask}$ represents the set of masked edges that the model aims to reconstruct. The summation in the denominator, $\sum_{z \in V} \exp(y_{vz})$, is computationally expensive for large graphs, so negative sampling is typically employed to accelerate the optimization process [33]. This approach allows MGAE to achieve strong performance while processing only a fraction of the original graph structure during encoding [33].

### 3.4 Contrastive Learning Methods in GNNS: Maximizing Agreement and Discrepancy

Contrastive learning has emerged as a powerful self-supervised paradigm for graph representation learning, focusing on learning discriminative embeddings by contrasting positive and negative sample pairs [41, 53].

*3.4.1 General Principles and InfoNCE Loss.* The core principle of contrastive learning is to learn representations by encouraging semantically similar instances to occupy nearby positions in the embedding space, while dissimilar instances are pushed farther apart [53]. This objective is typically operationalized through the construction of instance pairs: *positive pairs*, which consist of different augmented views of the same data point, and are intended to be close in the embedding space; and *negative pairs*, which are drawn from distinct data instances and are expected to be dissimilar [53].

The InfoNCE loss is a widely used cost function in contrastive learning, particularly in graph-based applications [41, 46, 53]. For a given anchor sample (or view) $x$, and a positive sample (another view of the same instance) $y$, the InfoNCE loss aims to maximize the similarity between $x$ and $y$ relative to a set of negative samples $\{x_i^-\}$ [46]:

$$L_{InfoNCE} = - \log \frac{\exp(\text{sim}(f(x), f(y))/\tau)}{\exp(\text{sim}(f(x), f(y))/\tau) + \sum_i \exp(\text{sim}(f(x), f(x_i^-))/\tau)} \tag{9}$$

Here, $f(\cdot)$ is the encoder network that transforms raw inputs into vector representations, $\text{sim}(\cdot, \cdot)$ is a similarity metric (often cosine similarity), and $\tau$ is a temperature parameter that controls the sharpness of the distribution [46, 53]. This loss function encourages alignment (positive samples are brought closer) and uniformity (representations are evenly distributed by pushing negative samples apart) [46]. The success of InfoNCE is strongly influenced by the quantity and quality of negative samples [46].

*3.4.2 Subgraph Gaussian Embedding Contrast (SGEC).* Subgraph Gaussian Embedding Contrast (SGEC) is a novel method that introduces a subgraph Gaussian embedding (SGE) module to adaptively map subgraphs to a structured Gaussian space [41]. This approach aims to preserve graph characteristics while controlling the distribution of generated subgraphs [41].

SGEC starts by sampling BFS-induced subgraphs. Node representations and topology information within these subgraphs are then embedded into a latent space that is regularized towards a Gaussian distribution using Kullback–Leibler (KL) divergence [41]. The embedded features $\tilde{x}_i$ are generated using the reparameterization trick: $\tilde{x}_i = \mu_i + \exp(\log(\sigma_i)) \odot \epsilon$, where $\epsilon \sim N(0, I)$ is Gaussian noise [41]. A KL divergence term regularizes the embedding distribution towards a Gaussian prior, preventing mode collapse [41].

For contrastive learning, SGEC integrates optimal transport distances into the InfoNCE loss formulation, addressing the complexities of graph-based data [41]. The complete contrastive loss $L_{contrast}$ is a sum of two components:

(1) **Wasserstein Distance ($L_W$):** This component captures feature distribution representation within subgraphs. It measures the minimum cost of transforming one feature distribution into another [41].

$$L_W = \alpha \left( - \sum_{i \in S} \log \frac{\exp(-W(X_i, \tilde{X}_i)/\tau)}{\sum_{j \in S, j \neq i} (\exp(-W(X_i, \tilde{X}_j)/\tau) + \exp(-W(X_i, X_j)/\tau))} \right) \quad (10)$$

where $W(X_i, \tilde{X}_i)$ is the Wasserstein distance between the feature matrices of the original subgraph $X_i$ and its embedded version $\tilde{X}_i$ [41].

(2) **Gromov-Wasserstein Distance ($L_{GW}$):** This component captures structural discrepancies, providing a topology-aware similarity measure. It measures the dissimilarity between two metric spaces (subgraphs) while considering their internal structures [41].

$$L_{GW} = (1 - \alpha) \left( - \sum_{i \in S} \log \frac{\exp(-GW(A_i, X_i, A_i, \tilde{X}_i)/\tau)}{\sum_{j \in S, j \neq i} (\exp(-GW(A_i, X_i, A_j, \tilde{X}_j)/\tau) + \exp(-GW(A_i, X_i, A_j, X_j)/\tau))} \right) \quad (11)$$

where $GW(A_i, X_i, A_i, \tilde{X}_i)$ is the Gromov-Wasserstein distance, considering adjacency matrices $A$ and feature matrices $X$ [41].

The final loss $L$ of the SGEC model combines both the contrastive and regularization components, balanced by a hyperparameter $\beta$: $L = L_{contrast} + \beta \text{KL}(q(\tilde{X}|X, A)||p(\tilde{X}))$ [41]. This comprehensive approach leverages the strengths of Gaussian embeddings and optimal transport to learn robust graph representations.

### 3.4.3 Discrepancy-based Self-supervised Learning (D-SLA).

Discrepancy-based Self-supervised Learning (D-SLA) introduces a novel perspective from traditional contrastive learning paradigms by focusing on modeling the *discrepancy* between graphs, rather than maximizing their similarity [18]. Instead of aligning representations of different augmented views, D-SLA encourages the model to distinguish the original graph from its perturbed variants. This approach compels the model to capture even subtle structural differences that may have significant implications for global graph properties [18].

D-SLA's objective functions are designed to achieve this:

(1) **Graph Discrimination Loss (LGD):** This objective trains the model to distinguish the original graph from its perturbed counterparts.

$$L_{GD} = - \log \left( \frac{e^{S_0}}{e^{S_0} + \sum_{i \geq 1} e^{S_i}} \right) \quad (12)$$

where $S_k$ is a score for graph $G_k$ (original $G_0$ or perturbed $G_i$), obtained from a learnable score network [18]. Perturbed graphs are generated through edge manipulation (addition/deletion) and node attribute masking, making the discrimination challenging and forcing the model to learn deeper discrepancies [18].

(2) **Edit Distance Learning (Ledit):** This objective quantifies *how* dissimilar graphs are by leveraging the graph edit distance. The exact number of edge additions/deletions during perturbation directly provides this distance, making it computationally efficient [18].

$$L_{edit} = \sum_{i,j} \left( \frac{d_i}{e_i} - \frac{d_j}{e_j} \right)^2 \tag{13}$$

Here, $e_i$ is the graph edit distance between $G_0$ and $G_i$, and $d_i$ is the embedding-level distance (L2-norm) between their representations. This term enforces that embedding differences are proportional to actual graph edit distances, ensuring that graphs with larger edit distances are farther apart in the embedding space [18].

(3) **Relative Discrepancy Learning (Lmargin):** This objective extends the learning to differentiate the target graph from *completely different* graphs (negative graphs from the same batch). It uses a triplet margin loss:

$$L_{margin} = \sum_{i,j} \max(0, \alpha + d_i - d'_j) \tag{14}$$

where $d_i$ is the distance between the original graph and its perturbed versions, and $d'_j$ is the distance between the original graph and a negative graph from the batch. $\alpha > 0$ is a margin hyperparameter [18]. This ensures that negative graphs are embedded further away from the original graph than its perturbed versions, preventing collapse of semantically dissimilar graphs [18].

The complete D-SLA learning objective combines these three components: $L = L_{GD} + \lambda_1 L_{edit} + \lambda_2 L_{margin}$, where $\lambda_1$ and $\lambda_2$ are scaling weights [18]. This framework allows D-SLA to capture subtle differences, discriminate between perturbed graphs, preserve exact discrepancy amounts, and capture relative distances, offering a robust approach to unsupervised graph representation learning.

## 4 Empirical Methodology

In this section, we present our approach to rendering the unsupervised model in inductive settings. To achieve this, we introduce a Universal Feature Encoder designed to map variable-length input vectors to fixed-length representations. By incorporating this encoder, the model can be trained on one graph dataset and subsequently applied to generate embeddings for unseen or different graph datasets, thereby enabling effective inductive generalization.

### 4.1 Universal feature encoder

Let $\mathbf{X} \in \mathbb{R}^{B \times d_{in}}$ denote the input feature matrix, where $B$ is the batch size, $d_{in}$ is the (variable) input dimension, and $d_h$, $d_{out}$ are the fixed hidden and output dimensions, respectively.

*4.1.1 Dynamic Linear Projection.* A linear transformation is applied to map input features to the hidden space:

$$\mathbf{H} = \mathbf{X}\mathbf{W}_p + \mathbf{b}_p, \quad \text{where } \mathbf{W}_p \in \mathbb{R}^{d_{in} \times d_h}, \ \mathbf{b}_p \in \mathbb{R}^{d_h} \tag{15}$$

*4.1.2 Layer Normalization with ReLU.* Each row $\mathbf{h}_i$ of $\mathbf{H}$ is normalized using layer normalization:

$$\hat{\mathbf{h}}_i = \frac{\mathbf{h}_i - \mu_i}{\sqrt{\sigma_i^2 + \epsilon}}, \quad \text{where}$$

$$\mu_i = \frac{1}{d_h} \sum_{j=1}^{d_h} h_{ij}, \qquad \sigma_i^2 = \frac{1}{d_h} \sum_{j=1}^{d_h} (h_{ij} - \mu_i)^2$$

Then apply the ReLU activation:

$$\mathbf{Z} = \text{ReLU}(\hat{\mathbf{H}})$$

### 4.1.3 Adaptive Average Pooling. First, reshape for 1D pooling:

$$\mathbf{Z} \in \mathbb{R}^{B \times d_h} \quad \rightarrow \quad \mathbf{Z}_{\text{reshaped}} \in \mathbb{R}^{B \times 1 \times d_h}$$

Apply 1D adaptive average pooling to obtain:

$$\mathbf{X}_{\text{tmp}} = \text{AdaptiveAvgPool1D}(\mathbf{Z}_{\text{reshaped}}) \in \mathbb{R}^{B \times 1 \times d_{\text{out}}}$$

Finally, remove the singleton dimension:

$$\mathbf{X}_{\text{final}} = \mathbf{X}_{\text{tmp}}[:, 0, :] \in \mathbb{R}^{B \times d_{\text{out}}}$$

**End-to-End Transformation Summary:**

$$\mathbf{X}_{\text{final}} = \text{AdaptiveAvgPool1D}\big(\text{ReLU}\big(\text{LayerNorm}(\mathbf{X}\mathbf{W}_p + \mathbf{b}_p)\big)\big) \tag{16}$$

## 4.2 Evaluation metrics for unsupervised node representations

Unsupervised graph node representation learning frameworks require specialized evaluation metrics due to the unique challenges and objectives inherent in learning from graph-structured data without explicit supervision. These frameworks aim to learn meaningful node embeddings that can be effectively applied to various downstream tasks. The following points highlight why distinct evaluation metrics are necessary:

(1) **Absence of Labeled Data:** Since no labeled data is available during training, the quality of learned embeddings must be assessed without relying on ground truth labels. Metrics that evaluate how well similar nodes are clustered together and dissimilar nodes are separated, such as cluster cohesion and separation measures, are essential.

(2) **Graph Topology Preservation:** A core goal of unsupervised graph representation learning is to capture the underlying structure and relationships within the graph. Metrics that quantify how well the embeddings preserve graph topology—through node classification accuracy, link prediction performance, or community detection quality—are critical.

(3) **Robustness and Generalization:** To evaluate the stability and applicability of learned embeddings across varying conditions, it is important to test their robustness to noise or perturbations in the graph. This helps assess how well the framework generalizes beyond the training data.

(4) **Interpretability:** Understanding and explaining the learned representations is crucial for trust and insight. Visualization techniques, such as network graphs, enable interpretation of embeddings in relation to graph structure, facilitating transparency and model validation.

Considering all scenarios we have considered 21 evaluation metrics for unsupervised embedding evaluation [8, 8, 35, 36, 38, 39]. Which are as follows:

**Node Classification Metrics:**

(1) node_cls_accuracy: Overall proportion of correctly predicted node classes.

$$\text{Accuracy} = \frac{TP + TN}{TP + TN + FP + FN} \tag{17}$$

(2) node_cls_precision: Correct positive predictions over total predicted positives.

$$\text{Precision} = \frac{TP}{TP + FP} \tag{18}$$

(3) `node_cls_recall` (`sensitivity`): Correct positive predictions over actual positives.

$$\text{Recall} = \frac{TP}{TP + FN} \tag{19}$$

(4) `node_cls_f1`: Harmonic mean of precision and recall.

$$F_1 = 2 \cdot \frac{\text{Precision} \cdot \text{Recall}}{\text{Precision} + \text{Recall}} \tag{20}$$

**Link Prediction Metrics (Binary Classification):**
Link prediction is one of the important factor in evaluating unsupervised node embedding.
(1) `LP_accuracy`: Fraction of correctly predicted links/non-links.
(2) `LP_precision`, `LP_recall`, `LP_f1`: Same definitions as node classification.
(3) `LP_auroc`: Area under ROC curve; measures true vs. false positive rate trade-off.
(4) `LP_aupr`: Area under Precision-Recall curve.
(5) `LP_specificity`: True negative rate.

$$\text{Specificity} = \frac{TN}{TN + FP} \tag{21}$$

**Embedding–Adjacency Alignment:**

(1) Cosine Similarity–Adjacency Correlation (`cosine_adj_corr`): Pearson correlation between cosine similarity matrix of embeddings and adjacency matrix.

$$\frac{\text{cov}(\text{CosSim}(X), A)}{\sigma_{\text{CosSim}} \cdot \sigma_A} \tag{22}$$

(2) Dot-Product–Adjacency Correlation(`dot_adj_corr`): Correlation between dot product similarity and adjacency.

$$\rho = \frac{\text{cov}(XX^\top, A)}{\sigma_{XX^\top} \cdot \sigma_A} \tag{23}$$

(3) Inverted Distance-Adjacency Correlation (`euclidean_adj_corr`): Negative correlation between Euclidean distance and adjacency.

$$\rho = \text{corr}(-\|x_i - x_j\|_2^2, A_{ij}) \tag{24}$$

(4) Edge Reconstruction BCE Loss (`graph_reconstruction_bce_loss`): Binary cross-entropy loss between predicted and actual adjacency.

$$\text{BCE} = -\frac{1}{|E|} \sum_{(i,j)} \left[ A_{ij} \log \hat{A}_{ij} + (1 - A_{ij}) \log(1 - \hat{A}_{ij}) \right] \tag{25}$$

**Clustering Quality Metrics** This is also important because unsupervised learning is based on clustering or geometry of the data.
(1) `silhouette`: Mean silhouette coefficient of samples:

$$s(i) = \frac{b(i) - a(i)}{\max\{a(i), b(i)\}} \quad \in [-1, 1] \tag{26}$$

(2) `calinski_harabasz`: Ratio of between-cluster to within-cluster dispersion:

$$\text{CH} = \frac{\text{Tr}(B_k)}{\text{Tr}(W_k)} \cdot \frac{n - k}{k - 1} \tag{27}$$

(3) Neighborhood Overlap Score (`knn_consistency`): Average proportion of shared neighbors between embedding and graph space.

$$\frac{1}{n} \sum_i \frac{|N_k^{\text{graph}}(i) \cap N_k^{\text{embed}}(i)|}{k} \tag{28}$$

**Semantic Coherence and Ranking:** For measuring the semantic coherence of embedding we have used the following metrics.

(1) `coherence`: Semantic similarity within cluster, usually using PMI or word co-occurrence.

$$\text{Coherence}(C) = \sum_{i<j} \text{PMI}(w_i, w_j) \tag{29}$$

(2) `selfCluster`: Internal measure of clusterability using self-supervision loss or contrastive consistency.

(3) `Rankme`: Measures the effective rank (entropy of singular values) of the embedding matrix.

$$\text{RankMe}(X) = \exp\left(-\sum_i p_i \log p_i\right), \quad p_i = \frac{\sigma_i}{\sum_j \sigma_j} \tag{30}$$

## 4.3 Classical GNN Architectures

To asses the learning of the model further we wanted to analyse if all GNN gives similar embedding or the embedding quality depends on the combinations of the loss function and Graph Neural networks. There fore we have used 6+1 different GNNs in our study.

### 4.3.1 Graph Convolutional Network (GCN) [20].
GCN is a foundational spectral-based graph neural network that performs convolution-like operations on graphs. It utilizes a normalized adjacency matrix to propagate features across graph neighborhoods. The layer-wise propagation rule is defined as:

$$H^{(l+1)} = \sigma\left(\hat{D}^{-1/2} \hat{A} \hat{D}^{-1/2} H^{(l)} W^{(l)}\right) \tag{31}$$

where $\hat{A} = A + I$ is the adjacency matrix with added self-loops, $\hat{D}$ is the corresponding degree matrix, $W^{(l)}$ is the trainable weight matrix, and $\sigma$ denotes a non-linear activation function.

### 4.3.2 Graph Attention Network (GAT) [37].
GAT introduces an attention mechanism to assign learnable importance weights to neighboring nodes during feature aggregation. This allows the network to focus on the most relevant parts of a node's neighborhood:

$$h_i' = \sigma\left(\sum_{j \in \mathcal{N}(i)} \alpha_{ij} W h_j\right) \tag{32}$$

$$\alpha_{ij} = \text{softmax}_j\left(\text{LeakyReLU}(a^\top[W h_i \| W h_j])\right) \tag{33}$$

Here, $\alpha_{ij}$ denotes the attention coefficient computed between nodes $i$ and $j$, and $\|$ represents concatenation.

### 4.3.3 GraphSAGE (Sample and Aggregate) [13].
GraphSAGE extends GCN by enabling inductive learning. It samples a fixed-size neighborhood and applies an aggregation function to produce node embeddings, making it suitable for large or dynamic graphs:

$$h_i^{(l+1)} = \sigma\left(W^{(l)} \cdot \text{AGGREGATE}^{(l)}\left(\{h_i^{(l)}\} \cup \{h_j^{(l)}, j \in \mathcal{N}(i)\}\right)\right) \tag{34}$$

Different aggregation functions such as mean, LSTM, or pooling can be used.

*4.3.4 Graph Isomorphism Network (GIN) [42].* GIN is designed to achieve high representational power by mimicking the Weisfeiler-Lehman test for graph isomorphism. It updates node features through a learnable weighted sum of neighbors:

$$\boldsymbol{h}_v^{(k)} = \text{MLP}^{(k)}\left((1 + \epsilon^{(k)}) \cdot \boldsymbol{h}_v^{(k-1)} + \sum_{u \in \mathcal{N}(v)} \boldsymbol{h}_u^{(k-1)}\right) \tag{35}$$

where $\epsilon^{(k)}$ is either a learnable parameter or a fixed scalar.

*4.3.5 Position-Aware Graph Neural Network (PAGNN) [45].* PAGNN integrates node positional encodings, often derived from Laplacian eigenvectors, into the attention mechanism. This allows the network to be sensitive to node positions or roles within the graph:

$$\alpha_{ij} = \text{Attention}\left(\boldsymbol{h}_i, \boldsymbol{h}_j, \boldsymbol{p}_i - \boldsymbol{p}_j\right) \tag{36}$$

Here, $\boldsymbol{p}_i$ encodes the structural position of node $i$, enhancing message passing with spatial awareness.

*4.3.6 Message Passing Neural Network (MPNN) [9].* MPNN is a general framework that encapsulates various GNN architectures. It separates message computation and node update phases, supporting edge features and more flexible interactions:

$$\boldsymbol{m}_v^{(t)} = \sum_{u \in \mathcal{N}(v)} M_t(\boldsymbol{h}_v^{(t)}, \boldsymbol{h}_u^{(t)}, \boldsymbol{e}_{uv}) \tag{37}$$

$$\boldsymbol{h}_v^{(t+1)} = U_t(\boldsymbol{h}_v^{(t)}, \boldsymbol{m}_v^{(t)}) \tag{38}$$

where $M_t$ is the message function, $U_t$ is the update function, and $\boldsymbol{e}_{uv}$ denotes edge features.

*4.3.7 ALL: Model Fusion Strategy.* The ALL model combines embeddings from multiple GNN architectures to enhance robustness and leverage complementary strengths. Feature fusion can be implemented via concatenation:

$$\boldsymbol{h}_v^{\text{ALL}} = \text{Concat}(\boldsymbol{h}_v^{\text{GCN}}, \boldsymbol{h}_v^{\text{GAT}}, \boldsymbol{h}_v^{\text{SAGE}}, \ldots) \tag{39}$$

or via summation:

$$\boldsymbol{h}_v^{\text{ALL}} = \boldsymbol{h}_v^{\text{GCN}} + \boldsymbol{h}_v^{\text{GAT}} + \boldsymbol{h}_v^{\text{SAGE}} + \ldots \tag{40}$$

This fusion strategy aims to capture diverse structural and semantic graph features.

## 4.4 Datasets used

We have used 3 publicly available datasets, described as follows:

- **Cora**:
  - 2708 scientific publications across 7 classes
  - 5429 citation links
  - Each publication is a 1433-dimensional bag-of-words vector
- **CiteSeer**:
  - 3312 documents classified into 6 classes
  - 4732 citation links
  - Each document represented by a 3703-dimensional feature vector
- **Bitcoin Transaction network**:

– The Elliptic dataset is a labeled graph-based representation of Bitcoin transactions, designed to facilitate the analysis of illicit financial activity in cryptocurrency networks. Each node in the graph corresponds to an individual Bitcoin transaction, and directed edges denote the flow of Bitcoin between transactions. The dataset categorizes transactions based on the type of entity controlling the input addresses, with labels indicating whether a transaction is *licit* (e.g., exchanges, wallet providers, miners, and other legal services) or *illicit* (e.g., scams, malware, ransomware, Ponzi schemes, and terrorist organizations).

– The graph comprises 203,769 nodes and 234,355 edges, offering a focused view of transaction behavior in contrast to the full Bitcoin network, which contains hundreds of millions of nodes and edges. Out of all transactions, 4,545 (2%) are labeled as illicit, 42,019 (21%) as licit, and the remaining majority (77%) are unlabeled or unknown [40].

– Each transaction is associated with a 166-dimensional feature vector derived entirely from publicly available blockchain data. For our study we have considered only subgraph of 5000 nodes from this graph.

## 4.5 Loss functions used

We have considered 5 base loss and further its higher order combinations.

*4.5.1 Pointwise Mutual Information Loss (*PMI_L*).* Let $N$ be the number of nodes in the graph, and let $\mathbf{z}_i \in \mathbb{R}^d$ denote the learned embedding of node $i$. Define $\text{PMI}_{ij}$ as the pointwise mutual information between node $i$ and node $j$, computed from structural statistics of the graph. Let $\text{CosSim}(\mathbf{z}_i, \mathbf{z}_j)$ denote the cosine similarity between embeddings $\mathbf{z}_i$ and $\mathbf{z}_j$.

The loss aims to align embedding similarity with structural PMI and is defined as:

$$\mathcal{L}_{\text{PMI}} = -\frac{1}{N^2} \sum_{i=1}^{N} \sum_{j=1}^{N} \text{PMI}_{ij} \cdot \text{CosSim}(\mathbf{z}_i, \mathbf{z}_j) \tag{41}$$

This formulation promotes high cosine similarity for node pairs with high PMI, and discourages it for those with low or negative PMI.

*4.5.2 Margin-based Contrastive Loss (*Contr_L*).* Given a graph with edge set $E$, for each positive pair $(u, v) \in E$, we sample a negative node $k_u$ not connected to $u$. The embeddings $\mathbf{z}_u, \mathbf{z}_v, \mathbf{z}_{k_u} \in \mathbb{R}^d$ correspond to the anchor, positive, and negative nodes, respectively. Let $M > 0$ be the margin hyperparameter.

The loss encourages the anchor to be closer to the positive than the negative:

$$\mathcal{L}_{\text{Contrastive}} = \frac{1}{|E|} \sum_{(u,v) \in E} \max\left(0, M - \text{CosSim}(\mathbf{z}_u, \mathbf{z}_v) + \text{CosSim}(\mathbf{z}_u, \mathbf{z}_{k_u})\right) \tag{42}$$

*4.5.3 Cross-Entropy-based Denoising Loss (*CrossE_L*).* Let $\mathbf{E} \in \mathbb{R}^{N \times D}$ be the clean input embedding matrix and $\hat{\mathbf{E}}$ be the reconstructed output from a denoising function. The loss minimizes the mean squared reconstruction error:

$$\mathcal{L}_{\text{DAE}} = \frac{1}{ND} \sum_{i=1}^{N} \sum_{j=1}^{D} \left(\mathbf{E}_{ij} - \hat{\mathbf{E}}_{ij}\right)^2 \tag{43}$$

Here, $\hat{\mathbf{E}} = \text{Denoiser}(\mathbf{E} + \boldsymbol{\epsilon})$, where $\boldsymbol{\epsilon} \sim \mathcal{N}(0, \sigma^2)$ is Gaussian noise.

*4.5.4 PageRank-based Contrastive Loss (*PR_L*).* Let $A \subseteq \{1, \ldots, N\}$ denote the set of anchor nodes. For each anchor $u \in A$, select:

- $P_u = \arg \min_{v \neq u} |PR_u - PR_v|$ (positive: similar PageRank)
- $N_u = \arg \max_{w \neq u, w \neq P_u} |PR_u - PR_w|$ (negative: dissimilar PageRank)

With margin $M > 0$, the loss is:

$$\mathcal{L}_{\text{PR}} = \frac{1}{|A|} \sum_{u \in A} \max \left(0, M - \text{CosSim}(\mathbf{z}_u, \mathbf{z}_{P_u}) + \text{CosSim}(\mathbf{z}_u, \mathbf{z}_{N_u})\right) \tag{44}$$

This encourages embeddings to reflect global ranking similarity.

*4.5.5 Triplet Loss (*Triplet_L*).* Similar to contrastive loss, the triplet loss enforces that an anchor $u$ is closer to its positive neighbor $v$ than to a randomly sampled negative node $k_u$:

$$\mathcal{L}_{\text{Triplet}} = \frac{1}{|E|} \sum_{(u,v) \in E} \max \left(0, M - \text{CosSim}(\mathbf{z}_u, \mathbf{z}_v) + \text{CosSim}(\mathbf{z}_u, \mathbf{z}_{k_u})\right) \tag{45}$$

This loss operates on edge-based triplets, using hinge-style ranking constraints to separate positive and negative pairs.

*4.5.6 Other Higher order Loss.* Further, we have considered all possible higher-order combinations. For example, second-order combinations include *Contr_L + Triplet_L*, *Contr_L + PR_L*, and so on. Third-order combinations include *Contr_L + Triplet_L + PMI_L*, *Contr_L + Triplet_L + PR_L*, etc. The hybrid loss functions are optimized as follows. Let the base loss functions be denoted as: $\mathcal{L}_1, \mathcal{L}_2, \mathcal{L}_3, \mathcal{L}_4, \mathcal{L}_5$.

Define a set of raw, learnable parameters:

$$\boldsymbol{\theta} = \left\{\theta_i, \theta_{ij}, \theta_{ijk} \mid 1 \leq i < j < k \leq 5\right\}$$

These parameters are passed through a sigmoid activation to constrain their corresponding weights in the interval $[0, 1]$ :

$$w_i = \sigma(\theta_i) = \frac{1}{1 + e^{-\theta_i}}, \quad w_{ij} = \sigma(\theta_{ij}), \quad w_{ijk} = \sigma(\theta_{ijk})$$

**Hybrid Loss with Learnable Weights**

The hybrid loss is constructed as a weighted sum of:

- Individual (first-order) losses
- Pairwise (second-order) products of losses
- Triple (third-order) products of losses

The full loss is given by:

$$\mathcal{L}_{\text{hybrid}} = \sum_{i=1}^{5} w_i \mathcal{L}_i \tag{46}$$

$$\mathcal{L}_{\text{hybrid}} = \sum_{1 \leq i < j \leq 5} w_i \cdot \mathcal{L}_i + w_j \cdot \mathcal{L}_j \tag{47}$$

$$\mathcal{L}_{\text{hybrid}} = \sum_{1 \leq i < j < k \cdots \leq 5} w_{ijk} \cdot \mathcal{L}_i + w_j \cdot \mathcal{L}_j + w_k \cdot \mathcal{L}_k + \cdots \tag{48}$$

Similarly, 4th and 5th orders were also considered. This gives us a large space of exploration and research to understand the emergent behavior of different loss functions

## 5 Experimental Results

We have considered two settings, one for transductive and one for inductive, for evaluating the model's capability. Average Rank (last column in each table) is the mean of global ranks across all metrics (lower is better). We have kept all the experimental conditions the same for all data sets and results. More specifically, we have trained all models for 500 epochs with an early stopping patience level is 10, embedding dimension 128. As most of the evaluation metrics lie in the compact domain [0,1] to avoid more information loss, we have multiplied the results by 100. It also reduces the size while giving better information. In the table caption arrow sign indicates if the metric is higher the better (↑) or lower the better (↓).

### 5.1 Inductive Results

In this setting, we have trained the model on one dataset and tested it on other datasets. Further, the generated embeddings were evaluated and reported for the following metrics.

**Pretrained on Cora and Citeseer datasets:** In these results, we have trained the model on the Cora and Citeseer dataset, and applied the model to generate node embeddings on an unknown dataset. The results are represented as *Data used for pretraining ↓ Applied Data*.

### 5.2 Analysis based on top 3 performance on each metric

We aimed to determine which model performs best overall and under what conditions specific combinations of models and loss functions should be used.

*5.2.1 Transductive Case.* To understand over all effect of the model and loss in transductive settings by analyzing the Tables in Supplementary Information 1 & 2 [1,2,3,4,5,6,7,8,9,10,11,12,13,14,15,16,17,18, 19,20,21]. Table 1 presents a comparative analysis of unsupervised GNN models trained with various loss functions. The ranking is based on inclusion in the top-3 positions across 21 evaluation metrics, capturing three key indicators: average rank, coverage, and top-1 wins.

**1. Performance Leaders (Low AvgRank):**
The *GCN + CrossE_L* combination achieves the best average rank (1.00) with perfect coverage (1) and a top-1 win, indicating strong and consistent performance across selected metrics. Similarly, *GAT + CrossE_L* (AvgRank 1.70) and *SAGE + CrossE_L* (AvgRank 2.00) also demonstrate competitive results, suggesting that Cross Entropy Loss performs reliably when applied individually to classical GNN models.

**2. Effective Hybrid Losses:**
The combinations *GAT + CrossE_L + Triplet_L* and *GAT + Contr_l + CrossE_L + Triplet_L* achieve moderate average ranks (2.27 and 3.74, respectively), but stand out with the highest coverage (9). This reflects their consistent presence in the top-3 across many metrics. Notably, Triplet-based hybrid losses appear to generalize well across evaluation criteria, making them strong candidates for robust unsupervised learning.

**3. Specialized High Performers (Top-1 Wins):**
The *GAT + Triplet_L* combination achieves the highest number of top-1 wins (9) and the highest coverage (14), despite a less favorable average rank (6.97). This suggests that while its overall performance may be inconsistent, it excels on specific metrics. Similarly, *MPNN + Contr_l* and *GIN + CrossE_L* achieve multiple top-1 wins, highlighting their targeted effectiveness with particular loss functions.

**4. Less Effective or Noisy Configurations:**
Several combinations involving *Contr_l + PMI_L* or *PR_L* (e.g., *ALL + Contr_l*, *PAGNN + Contr_l + CrossE_L + PMI_L*) show high average ranks (ranging from 14.30 to 36.00) with low or zero top-1 wins. This suggests that adding more loss terms does not necessarily improve performance and may dilute the learning signal. In particular, the *ALL + Contr_l* configuration (AvgRank 29.30) significantly underperforms despite the theoretical advantage of combining multiple losses.

Table 1.  Summary of Model Performance with Average Rank, Coverage, and Top-1 Wins based on transductive results.

| Model | Loss | AvgRank | Coverage | Top1Wins |
|-------|------|---------|----------|----------|
| GCN | CrossE_L | 1.00 | 1 | 1 |
| GAT | CrossE_L | 1.70 | 1 | 1 |
| SAGE | CrossE_L | 2.00 | 1 | 0 |
| GAT | CrossE_L + Triplet_L | 2.27 | 9 | 3 |
| ALL | Contr_l + CrossE_L + PMI_L + PR_L | 2.30 | 1 | 1 |
| GIN | CrossE_L | 2.33 | 3 | 1 |
| GAT | PMI_L | 3.70 | 1 | 1 |
| GAT | Contr_l + CrossE_L + Triplet_L | 3.74 | 9 | 0 |
| GAT | CrossE_L + PMI_L + Triplet_L | 4.00 | 1 | 0 |
| GAT | CrossE_L + PMI_L | 4.70 | 1 | 0 |
| GIN | PR_L | 5.00 | 1 | 0 |
| GAT | Contr_l | 5.00 | 1 | 1 |
| PAGNN | PR_L | 5.00 | 1 | 0 |
| GIN | Contr_l + PR_L | 5.70 | 1 | 0 |
| GAT | Triplet_L | 6.97 | 14 | 9 |
| GAT | Contr_l + Triplet_L | 7.30 | 1 | 0 |
| PAGNN | Contr_l + CrossE_L + PR_L | 9.00 | 1 | 0 |
| GAT | PMI_L + Triplet_L | 11.85 | 2 | 0 |
| GAT | Contr_l + PMI_L | 12.00 | 2 | 0 |
| ALL | Contr_l + CrossE_L + PR_L + Triplet_L | 14.30 | 1 | 0 |
| PAGNN | Contr_l + CrossE_L + PMI_L | 14.30 | 1 | 0 |
| ALL | Contr_l + PMI_L | 14.70 | 1 | 0 |
| GAT | Contr_l + CrossE_L + PR_L + Triplet_L | 15.00 | 1 | 0 |
| MPNN | Contr_l | 15.43 | 3 | 2 |
| GAT | Contr_l + PR_L + Triplet_L | 17.50 | 2 | 0 |
| GAT | PMI_L + PR_L + Triplet_L | 18.30 | 1 | 1 |
| ALL | Contr_l | 29.30 | 1 | 0 |
| MPNN | Contr_l + CrossE_L | 29.30 | 1 | 0 |
| GAT | PMI_L + PR_L | 36.00 | 1 | 1 |
| SAGE | Triplet_L | 37.30 | 1 | 0 |

*5.2.2  Inductive Case.* To answer these questions we have filtered only top 3 results based on average rank from inductive results. We have following findings based on tables in Supplementary Information 1 & 2 [22,23,24,25,26,27,28,29,30,31,32,33,34,35,36,37,38,39,40,41,42]. We have analyzed many aspects because it's inductive settings where model is used as a pretrained on different datasets.

Table 2. The overall statistics of the model and loss function considering top 3 for each evaluation metric for inductive results.

| Model | Loss | AvgRank | Coverage | Top1Wins |
|-------|------|---------|----------|----------|
| GIN | CrossE_L | 4.95 | 2 | 2 |
| GCN | Contr_l + PMI_L | 5.20 | 1 | 1 |
| SAGE | CrossE_L + PMI_L + PR_L | 6.10 | 1 | 0 |
| GIN | CrossE_L + PMI_L + PR_L | 6.40 | 1 | 0 |
| GIN | Contr_l + PR_L | 7.10 | 1 | 0 |
| GAT | CrossE_L + PMI_L + Triplet_L | 7.20 | 1 | 0 |
| GCN | PMI_L | 8.00 | 1 | 0 |
| GAT | PMI_L | 8.04 | 9 | 3 |
| GCN | CrossE_L + PMI_L + Triplet_L | 8.10 | 1 | 0 |
| GAT | Triplet_L | 9.27 | 6 | 0 |
| GAT | Contr_l + PMI_L | 10.27 | 3 | 0 |
| GAT | CrossE_L + PMI_L | 10.28 | 5 | 2 |
| GAT | Contr_l + CrossE_L + PMI_L | 11.60 | 1 | 0 |
| GAT | CrossE_L + Triplet_L | 12.80 | 11 | 9 |
| GAT | Contr_l + CrossE_L + PMI_L + PR_L + Triplet_L | 14.00 | 1 | 0 |
| MPNN | CrossE_L + PR_L | 14.20 | 1 | 1 |
| MPNN | Contr_l + PMI_L | 16.40 | 1 | 1 |
| GAT | Contr_l + CrossE_L + PMI_L + Triplet_L | 17.40 | 4 | 0 |
| MPNN | PR_L + Triplet_L | 19.00 | 1 | 0 |
| MPNN | PR_L | 20.00 | 1 | 0 |
| MPNN | CrossE_L + PMI_L + PR_L | 31.20 | 1 | 0 |
| MPNN | CrossE_L + PMI_L + Triplet_L | 35.30 | 2 | 0 |
| GAT | PMI_L + Triplet_L | 37.20 | 1 | 0 |
| MPNN | CrossE_L + Triplet_L | 53.60 | 1 | 1 |
| SAGE | Triplet_L | 55.80 | 1 | 0 |

**Average Rank per Model (Lower is Better):** This bar plot (Figure 1) displays the mean of average ranks for each model across all metrics and loss combinations. A lower average rank indicates a more consistent and generally better-performing model across diverse evaluation settings. Models are sorted in ascending order of their mean rank.

**Average Rank per Loss Function (Lower is Better):** Similar to the previous plot, this figure (Figure 2) shows the average rank for each loss function, aggregated over all models and tasks. This helps identify which loss functions contribute to more stable and superior performance in GNN training across various graph datasets and evaluation settings.

**Heatmap of Average Rank for (Model, Loss) across Metrics:** This heatmap (Figure 3) visualizes the average rank for each model-loss pair (rows) across all considered evaluation metrics (columns). The color gradient from green (lower rank, better) to red (higher rank, worse) highlights how different combinations perform on specific metrics. It helps identify task-specific strengths and weaknesses.

**Count of Top-1 Ranked (Model, Loss) Combinations:** This bar chart (Figure 4) reports how many times each model-loss combination achieved the best performance (rank 1) for any evaluation metric. A higher count indicates that a combination is frequently the top-performing setup, even if it may not have the best average rank overall. The x-axis represents the number of top-1 wins, indicating how often each model-loss combination achieved the best performance. The y-axis lists the different GNN models (GAT, GIN, GCN, MPNN), and the bars show the count of top-1 wins for each model when paired with specific loss functions (e.g., CrossE_L + Triplet_L, PMI_L, CrossE_L + PMI_L, etc.), highlighting which combinations were most effective. GAT along *CrossE_L+Triplet_L* with have been found the maximum number of times in top-1.

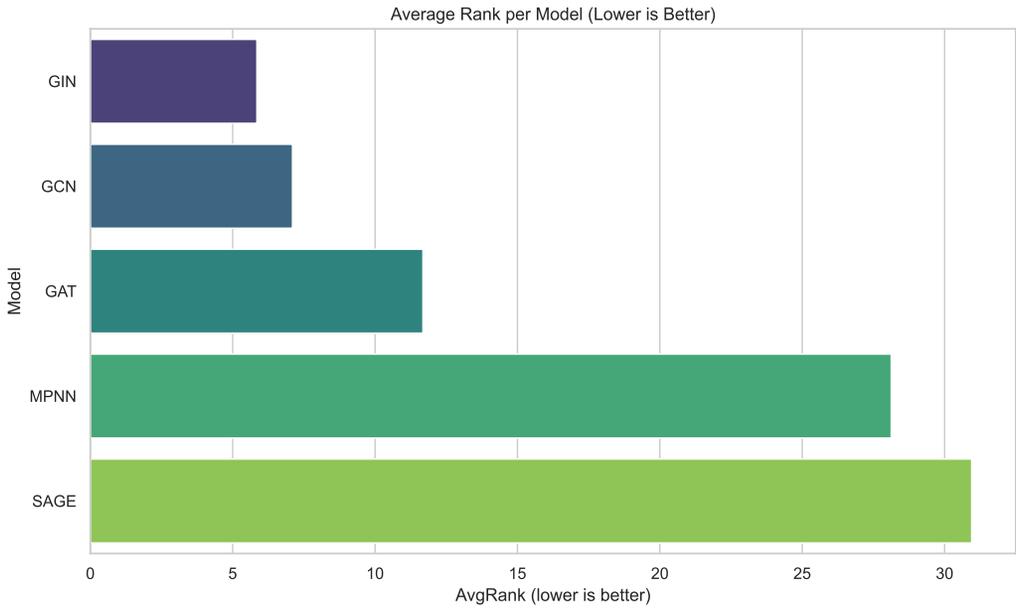

Fig. 1. This figure shows Average Rank per Model

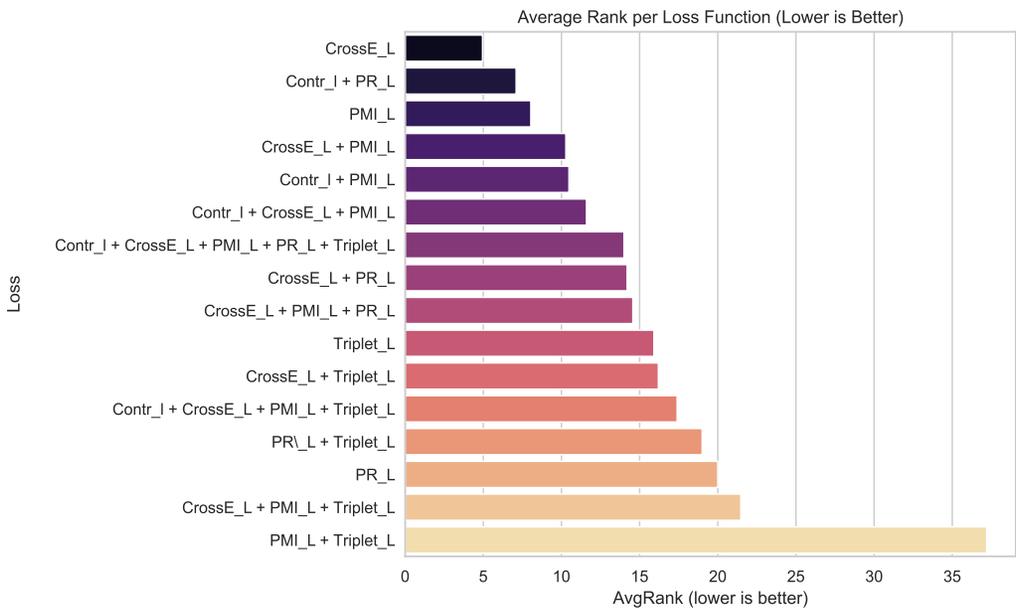

Fig. 2. This figure shows Average Rank per Loss Function. From inductive case top 3 results only.

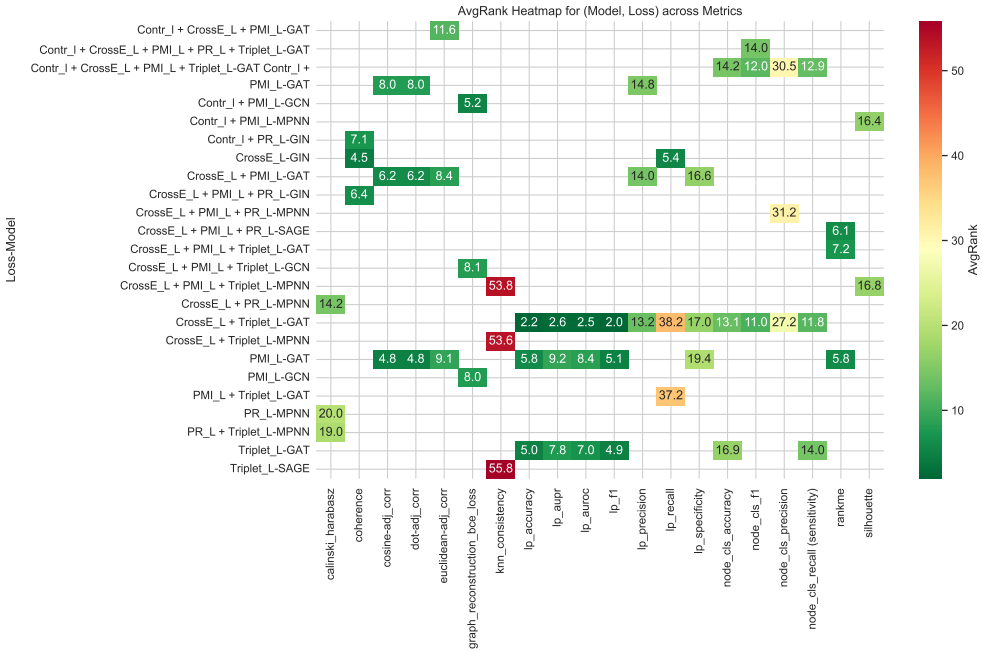

Fig. 3. This is heatmap of model+loss function across metrics in inductive settings.

**Histogram of Appearance Counts (Model and Loss Function):** These two histograms (Figure 5, 6) report the frequency with which each model and loss function appeared in the top-k results across evaluation metrics. This gives a sense of which methods were commonly evaluated and ensures fair comparison coverage.

**Summary Table of Aggregated Results:** The summary Table 2 includes the following statistics for each model-loss pair:

- **AvgRank:** Mean rank across all metrics (lower is better).
- **Coverage:** Number of unique metrics in which the model-loss pair was found in top 3 rankings.
- **Top1Wins:** Number of metrics for which the combination achieved rank 1.

The results in Table 2 reveal key insights into the performance of model−loss function combinations in inductive settings. The GIN model with CrossEntropy loss (CrossE_L) achieves the lowest average rank (4.95) and secures 2 Top1Wins, marking it as the most consistent overall performer. Similarly, GCN with Contr_L + PMI_L ranks second (AvgRank 5.20), reaffirming the effectiveness of pairing simple architectures with hybrid losses. Notably, while GAT models do not dominate in average rank, they exhibit high Coverage and Top1Wins, especially with CrossE_L + Triplet_L (Coverage = 11, Top1Wins = 9), and PMI_L alone (Coverage = 9, Top1Wins = 3). This indicates GAT's strength in excelling at specific metrics, despite overall inconsistency. The prevalence of hybrid losses (e.g., CrossE_L + PMI_L + PR_L) among top-ranked combinations underscores their benefit in capturing diverse graph properties. In contrast, MPNN consistently underperforms, occupying the bottom ranks, with high average ranks and minimal Top1Wins, suggesting limited generalization in inductive contexts. In summary, the analysis highlights GIN and GCN as the most robust architectures overall, while GAT demonstrates targeted excellence. Hybrid loss design emerges as a critical factor in enhancing inductive performance.

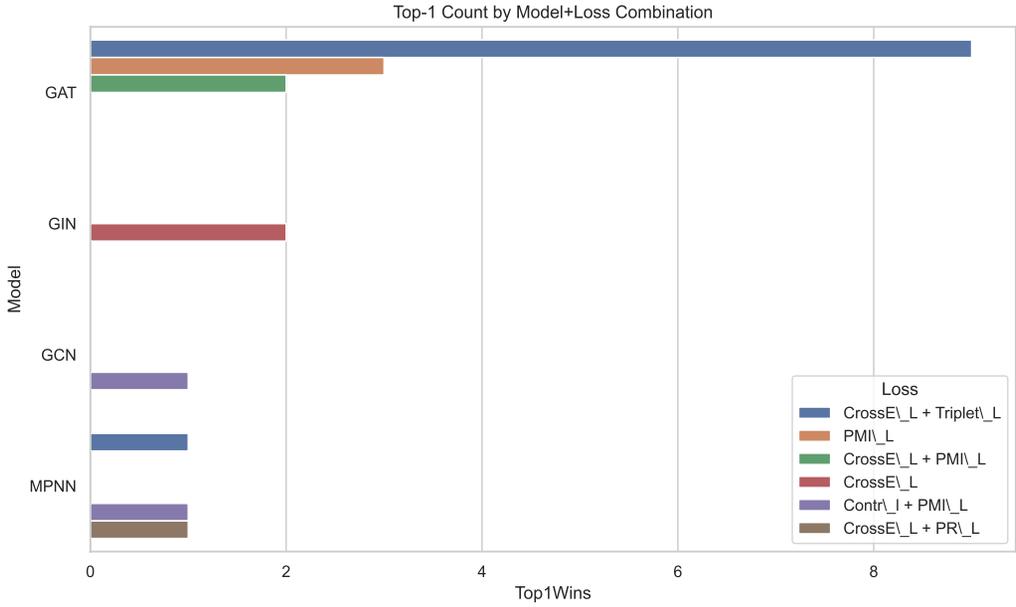

Fig. 4. This figure shows top-1 model combinations. It shows how many times a model and loss function together ranked-1.

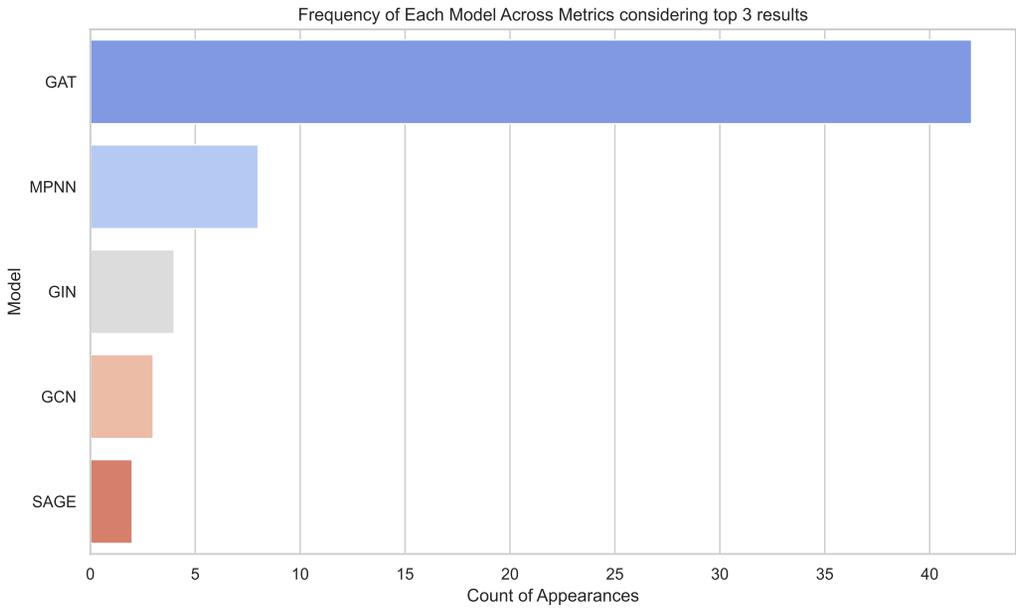

Fig. 5. Model appearance in top 3 rankings across all metrics.

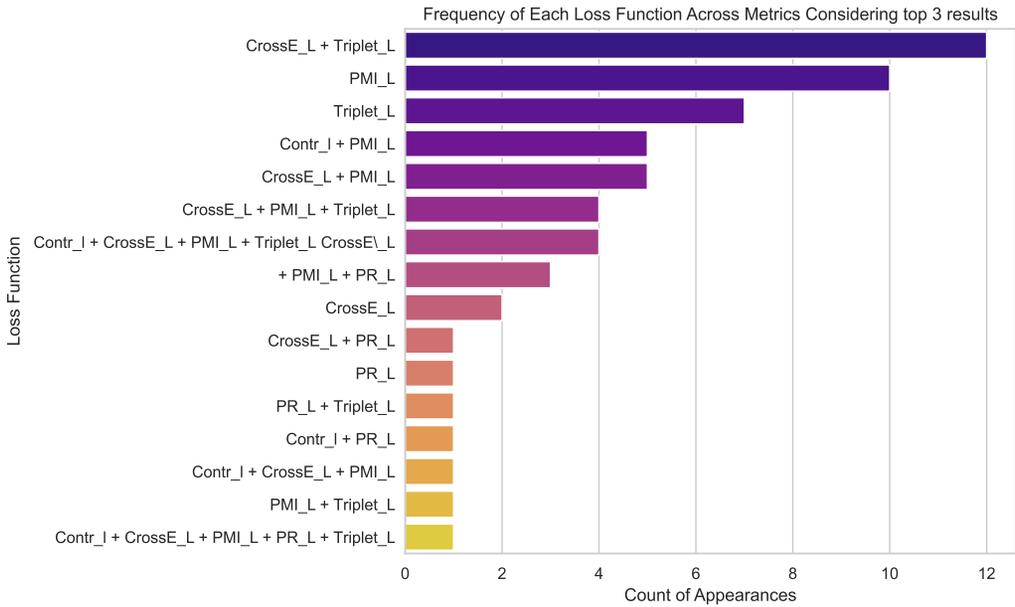

Fig. 6.  Loss appearance in top 3 rankings across all metrics.

*5.2.3   Inductive and transductive together.* **Models in both the cases for top 3 cases:** The figure.,7 presents a comparison of the average ranks of various Graph Neural Network (GNN) models across two different settings: inductive and transductive. The x-axis lists the GNN models (GIN, GCN, PAGNN, GAT, ALL, SAGE, MPNN), while the y-axis represents the average rank, with lower values indicating better performance. The figure shows that the performance of these models varies between the inductive and transductive settings, with some models achieving lower ranks (better performance) in one setting compared to the other.

**Inductive vs transductive : (Model, loss) plot for top 3 cases:** The figure 8 illustrates the average ranks of various models across different settings, specifically inductive and transductive. The y-axis indicating the number of top-1 models higher is better. The model+loss function combination is presented in this x-y plain giving deep insights.

## 6   Discussion

The comprehensive analysis, spanning six distinct GNN architectures plus an additional combined model, over 30 loss functions, and rigorously evaluated across three diverse datasets in both inductive and transductive settings using 21 distinct evaluation metrics, provides profound insights into the interplay of model architecture and loss function design. The summarized results in Table 1, 2, and Figure 8 which aggregate the top three performers for each of the 21 metrics based on their average rank, reveal several critical conclusions:

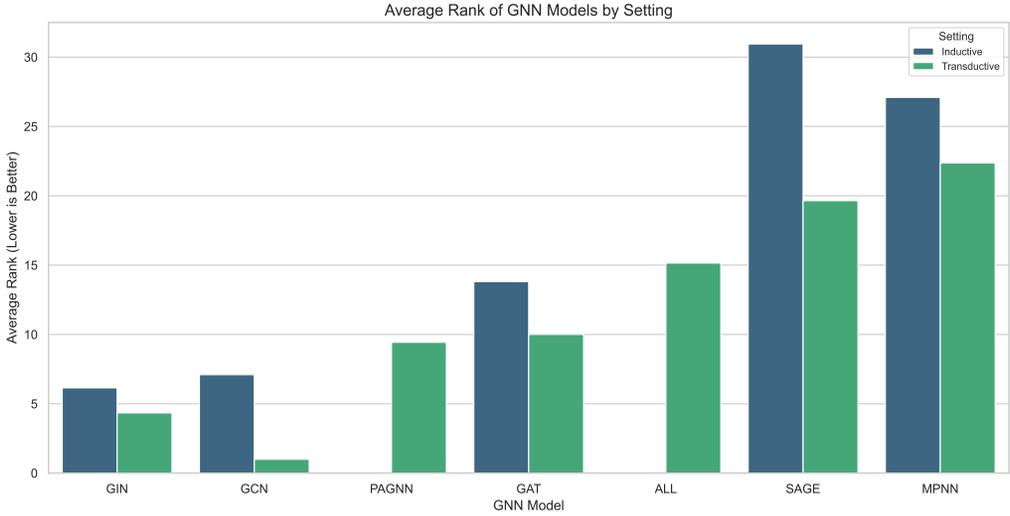

Fig. 7. This figure shows Average Rank achieved by models in inductive vs transductive settings.

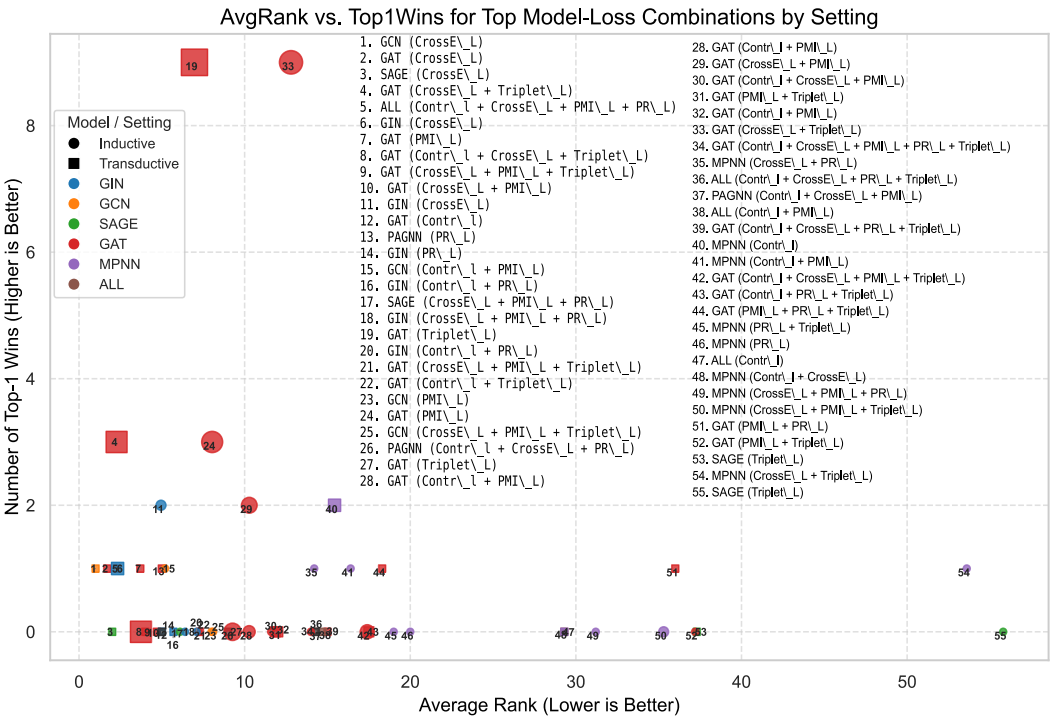

Fig. 8. This figure shows Average Rank achieved by models+loss function in inductive vs transductive settings.

## 6.1 Overall Performance and the Efficacy of Hybrid Loss Functions(inductive)

Since the focus of this work is on training GNNs for use as pretrained models, our conclusions are drawn exclusively from inductive settings. The results clearly demonstrate the superior overall performance of specific model–loss function pairings. Notably, the GIN architecture paired with Cross-Entropy loss (CrossE_L) stands out as the most consistent top performer, achieving the lowest average rank (4.95). This suggests that across a broad spectrum of tasks and evaluation metrics, GIN's aggregation mechanism combined with a fundamental classification loss provides a robust and effective approach. Additionally, the strong performance of GCN with a hybrid loss combining Contrastive Loss (Contr_L) and PMI Loss (PMI_L) (average rank 5.20) highlights that simpler architectures, when thoughtfully paired with suitable loss functions, can also deliver highly competitive results.

Furthermore, the prominence of hybrid loss functions among the best-performing combinations underscores their vital role in enhancing GNN performance. While single loss functions serve as a foundation, they often lack the synergistic effects achieved by integrating multiple objectives. For example, the second-best GIN variant (GIN + CrossE_L + PMI_L + PR_L, average rank 6.40) and the leading SAGE model (SAGE + CrossE_L + PMI_L + PR_L, average rank 6.10) both leverage a blend of cross-entropy, pointwise mutual information, and neighborhood preservation losses. This demonstrates that optimizing across multiple complementary criteria—promoting accurate classification, capturing statistical dependencies, and maintaining structural fidelity—yields more balanced and effective node representations in inductive scenarios.

## 6.2 Nuanced Strengths of GAT and Architectural Comparisons(inductive settings)

Thirdly, the 'Coverage' and 'Top1Wins' metrics offer a more nuanced perspective on model performance that extends beyond average ranking alone. While combinations involving GIN and GCN frequently achieve the lowest average ranks, the GAT architecture exhibits a notable capacity for targeted dominance. For instance, GAT with CrossE_L + Triplet_L records 11 appearances in the top 3 and secures 9 first-place finishes, despite a comparatively higher average rank of 12.80. Similarly, GAT combined with a simple PMI_L loss achieves 9 top-3 coverages and 3 top-1 wins with an average rank of 8.04. These findings suggest that GAT, while not universally superior across all metrics, demonstrates distinct advantages in specific evaluation dimensions. This highlights its potential suitability for applications emphasizing particular criteria—such as recall, precision, or structural sensitivity—where tailored hybrid loss functions may amplify its strengths.

Lastly, the results reveal a clear stratification among GNN architectures in this comprehensive inductive evaluation. GIN and GCN consistently emerge as top performers, underscoring their robust and generalizable representation capabilities. GraphSAGE also exhibits competitive performance, particularly when enhanced with hybrid loss functions. In contrast, the MPNN architecture consistently underperforms, indicating challenges in adapting effectively across the diverse tasks and evaluation metrics considered.

## 7 Conclusion

In conclusion, this rigorous evaluation, synthesizing performance across a multitude of settings and metrics, provides invaluable guidance. It underscores that optimal GNN performance is not solely a function of model complexity but rather a delicate and synergistic interplay between the graph neural network architecture and the carefully designed, often hybrid, loss functions. Researchers and practitioners should consider the specific requirements of their tasks and the strengths revealed by metrics like 'Coverage' and 'Top1Wins' when selecting and tuning models, rather than relying solely on aggregated performance metrics.

## Acknowledgments

This research was supported by the Key Scientific and Technological Research Projects in Henan Province under Grant No.202102210379, and by the Super-Science Project of Zhoukou Normal

University under Grant No. ZKNUC2018019. As English is not the authors' first language, we used language assistance tools such as ChatGPT (https://chatgpt.com/), Google Gemini etc. to help refine the writing, correct grammatical errors, and improve the overall clarity and presentation of the manuscript.

# Supplementary Information 1 for: *Evaluating Loss Functions for Graph Neural Networks: Towards Pretraining and Generalization*

KHUSHNOOD ABBAS*, School of Computer Science and Technology, Zhoukou Normal University, China

RUIZHE HOU, School of Automation Science and Engineering,South China University of Technology, China

ZHOU WENGANG, DONG SHI, NIU LING, School of Computer Science and Technology, Zhoukou Normal University, China

SATYAKI NAN, College of Business and Computing, Georgia Southwestern State University, USA

ALIREZA ABBASI, School of Engineering and IT, University of New South Wales (UNSW Canberra), Australia

**Overview**

This document provides supplementary information accompanying the manuscript titled "*Evaluating Loss Functions for Graph Neural Networks: Towards Pretraining and Generalization*". In this file the results are filtered based on top 10 average results for each metrics in inductive and transductive settings. There is another file Supplementary Information 2 which includes full tabular results.

## 1 Supplementary Tables for top 10 average results

### 1.1 Transductive settings

*1.1.1 Assessment of Learned Representations through Node Classification.* In this task, we first train the Graph Neural Network (GNN) models using various unsupervised loss functions. The resulting node embeddings are then used as input to a Multi-Layer Perceptron (MLP) for a supervised downstream task. The table below presents the results based on node classification accuracy.

Table 1. Node Cls Accuracy Performance (↑): This table presents the top 10 models ranked by their average performance in terms of node cls accuracy. For the complete list of models, refer to Supplementary Information 2 Table 1. Top-ranked results are highlighted in red, second-ranked in blue, and third-ranked in green.

| Loss Type | Model | CORA | | Citeseer | | Bitcoin Fraud Transaction | | Average Rank |
|---|---|---|---|---|---|---|---|---|
| Contr_l | GAT | 81.66 | ± 0.38 | 68.23 | ± 1.30 | 74.26 | ± 0.73 | 16.7 |

Continued on next page

Authors' Contact Information: Khushnood Abbas, Khushnood.abbas@zknu.edu.cn, School of Computer Science and Technology, Zhoukou Normal University, Zhoukou, Henan, China; Ruizhe Hou, auruizhe@mail.scut.edu.cn, School of Automation Science and Engineering,South China University of Technology, Guangzhou, Henan, China; Zhou Wengang, Dong Shi, Niu Ling, School of Computer Science and Technology, Zhoukou Normal University, Zhoukou, Henan, China; Satyaki Nan, satyaki.nan@gsw.edu, College of Business and Computing, Georgia Southwestern State University, Americus, GA, USA; Alireza Abbasi, a.abbasi@unsw.edu.au, School of Engineering and IT, University of New South Wales (UNSW Canberra), Canberra, ACT, Australia.

Node Cls Accuracy Continued (↑)

| Loss Type | Model | CORA | | Citeseer | | Bitcoin Fraud Transaction | | Average Rank |
|---|---|---|---|---|---|---|---|---|
| Contr_l + CrossE_L + PMI_L + PR_L + Triplet_L | GAT | 79.08 ± 0.79 | | 64.17 ± 0.90 | | 75.26 ± 0.67 | | 17.0 |
| Contr_l + CrossE_L + PR_L + Triplet_L | GAT | 79.30 ± 0.92 | | 64.11 ± 2.01 | | 75.38 ± 0.63 | | 15.0 |
| Contr_l + PMI_L | GAT | 79.67 ± 0.91 | | 63.90 ± 0.87 | | 74.92 ± 1.28 | | 22.0 |
| Contr_l + PMI_L + PR_L + Triplet_L | GAT | 79.41 ± 1.29 | | 63.27 ± 1.44 | | 75.30 ± 0.33 | | 22.0 |
| Contr_l + PR_L + Triplet_L | GAT | 79.78 ± 2.70 | | 64.08 ± 1.37 | | 75.38 ± 0.63 | | 13.7 |
| CrossE_L + Triplet_L | GAT | 80.89 ± 0.67 | | 65.20 ± 0.83 | | 74.72 ± 1.00 | | 16.7 |
| CrossE_L + Triplet_L | MPNN | 79.89 ± 0.95 | | 64.35 ± 1.55 | | 75.00 ± 0.74 | | 15.0 |
| PMI_L + Triplet_L | GAT | 79.78 ± 1.97 | | 64.05 ± 1.11 | | 74.96 ± 0.67 | | 19.0 |
| Triplet_L | GAT | 81.70 ± 0.86 | | 65.92 ± 0.51 | | 75.28 ± 0.72 | | 8.0 |

Table 2. Node Cls F1 Performance (↑): This table presents the top 10 models ranked by their average performance in terms of node cls f1. For the complete list of models, refer to Supplementary Information 2 Table 2. Top-ranked results are highlighted in red, second-ranked in blue, and third-ranked in green.

| Loss Type | Model | CORA | | Citeseer | | Bitcoin Fraud Transaction | | Average Rank |
|---|---|---|---|---|---|---|---|---|
| Contr_l | GAT | 79.31 ± 0.32 | | 60.52 ± 1.31 | | 39.10 ± 1.00 | | 15.3 |
| Contr_l + CrossE_L + PMI_L | GAT | 77.60 ± 2.42 | | 56.54 ± 1.02 | | 40.62 ± 1.30 | | 16.0 |

Node Cls F1 Continued (↑)

| Loss Type | Model | CORA | | Citeseer | | Bitcoin Fraud Transaction | | Average Rank |
|---|---|---|---|---|---|---|---|---|
| Contr_l + CrossE_L + PMI_L + PR_L + Triplet_L | GAT | 76.03 ± 0.72 | | 56.78 ± 0.88 | | 41.23 ± 1.15 | | 17.3 |
| Contr_l + PMI_L | GAT | 77.79 ± 1.27 | | 56.79 ± 1.06 | | 41.12 ± 1.26 | | 11.3 |
| Contr_l + PR_L + Triplet_L | GAT | 77.61 ± 2.82 | | 56.27 ± 1.84 | | 40.98 ± 0.55 | | 14.7 |
| CrossE_L + PMI_L + Triplet_L | GAT | 76.87 ± 1.04 | | 56.91 ± 1.00 | | 40.95 ± 0.81 | | 15.7 |
| CrossE_L + Triplet_L | GAT | 79.23 ± 1.28 | | 57.76 ± 1.09 | | 40.27 ± 1.69 | | 13.0 |
| PMI_L | GAT | 77.20 ± 1.98 | | 57.06 ± 2.63 | | 40.94 ± 0.89 | | 13.3 |
| PMI_L + Triplet_L | GAT | 76.97 ± 2.22 | | 57.31 ± 1.37 | | 41.22 ± 0.77 | | 11.7 |
| Triplet_L | GAT | 79.11 ± 1.25 | | 58.32 ± 1.37 | | 41.44 ± 1.13 | | 6.3 |

Table 3. Node Cls Precision Performance (↑): This table presents the top 10 models ranked by their average performance in terms of node cls precision. For the complete list of models, refer to Supplementary Information 2 Table 3. Top-ranked results are highlighted in red, second-ranked in blue, and third-ranked in green.

| Loss Type | Model | CORA | | Citeseer | | Bitcoin Fraud Transaction | | Average Rank |
|---|---|---|---|---|---|---|---|---|
| Contr_l | GAT | 79.27 ± 0.31 | | 64.64 ± 3.99 | | 47.04 ± 1.41 | | 22.7 |
| Contr_l + CrossE_L + PMI_L + PR_L + Triplet_L | GAT | 77.08 ± 1.06 | | 60.38 ± 3.72 | | 47.90 ± 1.19 | | 29.3 |
| Contr_l + CrossE_L + Triplet_L | GAT | 79.39 ± 1.29 | | 63.30 ± 3.66 | | 46.37 ± 1.62 | | 27.7 |

Node Cls Precision Continued (↑)

| Loss Type | Model | CORA | | Citeseer | | Bitcoin Fraud Transaction | | Average Rank |
|---|---|---|---|---|---|---|---|---|
| Contr_l + PMI_L | GAT | 79.03 ± 1.14 | | 61.66 ± 4.93 | | 47.14 ± 2.04 | | 25.3 |
| Contr_l + PR_L + Triplet_L | GAT | 78.96 ± 1.49 | | 60.95 ± 3.80 | | 48.40 ± 1.25 | | 21.3 |
| CrossE_L + Triplet_L | MPNN | 77.67 ± 1.21 | | 60.05 ± 3.42 | | 47.98 ± 1.00 | | 28.0 |
| PMI_L | GAT | 79.25 ± 1.81 | | 62.25 ± 5.03 | | 47.04 ± 1.53 | | 25.0 |
| PMI_L + PR_L + Triplet_L | GAT | 77.72 ± 2.84 | | 61.48 ± 4.65 | | 48.93 ± 1.00 | | 18.3 |
| PMI_L + Triplet_L | GAT | 78.35 ± 2.88 | | 61.81 ± 3.83 | | 47.15 ± 1.00 | | 25.7 |
| Triplet_L | GAT | 79.99 ± 1.42 | | 61.21 ± 1.97 | | 47.68 ± 0.86 | | 20.7 |

Table 4. Node Cls Recall (Sensitivity) Performance (↑): This table presents the top 10 models ranked by their average performance in terms of node cls recall (sensitivity). For the complete list of models, refer to Supplementary Information 2 Table 4. Top-ranked results are highlighted in red, second-ranked in blue, and third-ranked in green.

| Loss Type | Model | CORA | | Citeseer | | Bitcoin Fraud Transaction | | Average Rank |
|---|---|---|---|---|---|---|---|---|
| Contr_l | GAT | 79.77 ± 0.62 | | 61.90 ± 1.07 | | 39.20 ± 0.72 | | 15.3 |
| Contr_l + CrossE_L + PMI_L + PR_L + Triplet_L | GAT | 75.39 ± 0.79 | | 58.24 ± 0.88 | | 40.77 ± 0.86 | | 17.7 |
| Contr_l + CrossE_L + PR_L + Triplet_L | GAT | 75.85 ± 1.76 | | 57.92 ± 2.07 | | 40.59 ± 0.98 | | 19.3 |
| Contr_l + PMI_L | GAT | 77.02 ± 1.65 | | 58.14 ± 0.87 | | 40.65 ± 1.01 | | 12.7 |

Continued on next page

Node Cls Recall (Sensitivity) Continued (↑)

| Loss Type | Model | CORA | | Citeseer | | Bitcoin Fraud Transaction | | Average Rank |
|---|---|---|---|---|---|---|---|---|
| Contr_l + PR_L + Triplet_L | GAT | 77.04 ± 3.90 | | 58.03 ± 1.41 | | 40.57 ± 0.43 | | 14.0 |
| CrossE_L + PMI_L + Triplet_L | GAT | 76.02 ± 1.15 | | 58.17 ± 0.84 | | 40.51 ± 0.63 | | 17.7 |
| CrossE_L + Triplet_L | GAT | 78.98 ± 1.54 | | 59.20 ± 0.89 | | 40.04 ± 1.23 | | 13.7 |
| PMI_L | GAT | 75.80 ± 2.09 | | 58.24 ± 1.94 | | 40.50 ± 0.71 | | 19.3 |
| PMI_L + Triplet_L | GAT | 76.25 ± 1.95 | | 58.42 ± 1.14 | | 40.72 ± 0.61 | | 12.0 |
| Triplet_L | GAT | 78.70 ± 1.63 | | 59.75 ± 0.82 | | 40.92 ± 0.89 | | 6.3 |

*1.1.2 Assessment of Learned Representations through Link Prediction (LP).* In this task, The resulting node embedding is used for link prediction task.

Table 5. Lp Accuracy Performance (↑): This table presents the top 10 models ranked by their average performance in terms of lp accuracy. For the complete list of models, refer to Supplementary Information 2 Table 5. Top-ranked results are highlighted in red, second-ranked in blue, and third-ranked in green.

| Loss Type | Model | CORA | | Citeseer | | Bitcoin Fraud Transaction | | Average Rank |
|---|---|---|---|---|---|---|---|---|
| Contr_l | GAT | 96.28 ± 0.16 | | 99.07 ± 0.06 | | 91.21 ± 0.36 | | 10.7 |
| Contr_l + CrossE_L | GAT | 96.04 ± 0.22 | | 99.04 ± 0.08 | | 92.47 ± 1.76 | | 12.3 |
| Contr_l + CrossE_L + Triplet_L | GAT | 97.40 ± 0.37 | | 99.32 ± 0.10 | | 94.61 ± 0.85 | | 2.7 |
| Contr_l + Triplet_L | GAT | 97.22 ± 0.38 | | 99.28 ± 0.11 | | 94.81 ± 0.97 | | 3.7 |

Continued on next page

Lp Accuracy Continued (↑)

| Loss Type | Model | CORA | | Citeseer | | Bitcoin Fraud Transaction | | Average Rank |
|-----------|-------|------|---|----------|---|---------------------------|---|--------------|
| CrossE_L + PMI_L + Triplet_L | GAT | 97.12 ± 0.23 | | 98.94 ± 0.10 | | 89.82 ± 0.90 | | 12.3 |
| CrossE_L + Triplet_L | GAT | 98.26 ± 0.26 | | 99.51 ± 0.08 | | 93.34 ± 1.24 | | 2.3 |
| CrossE_L + Triplet_L | GCN | 97.14 ± 0.38 | | 98.90 ± 0.19 | | 90.80 ± 0.44 | | 10.0 |
| CrossE_L + Triplet_L | MPNN | 96.45 ± 0.46 | | 98.95 ± 0.29 | | 90.86 ± 0.60 | | 10.7 |
| Triplet_L | GAT | 98.71 ± 0.25 | | 99.63 ± 0.06 | | 92.65 ± 0.58 | | 2.0 |
| Triplet_L | GCN | 97.12 ± 0.14 | | 98.95 ± 0.17 | | 90.67 ± 0.47 | | 10.7 |

Table 6. Lp Aupr Performance (↑): This table presents the top 10 models ranked by their average performance in terms of lp aupr. For the complete list of models, refer to Supplementary Information 2 Table 6. Top-ranked results are highlighted in red, second-ranked in blue, and third-ranked in green.

| Loss Type | Model | CORA | | Citeseer | | Bitcoin Fraud Transaction | | Average Rank |
|-----------|-------|------|---|----------|---|---------------------------|---|--------------|
| Contr_l + CrossE_L | GAT | 99.23 ± 0.06 | | 99.86 ± 0.03 | | 94.85 ± 1.78 | | 11.3 |
| Contr_l + CrossE_L + Triplet_L | GAT | 99.55 ± 0.08 | | 99.90 ± 0.02 | | 96.81 ± 0.75 | | 4.7 |
| Contr_l + CrossE_L + Triplet_L | GCN | 99.32 ± 0.06 | | 99.79 ± 0.05 | | 93.52 ± 0.34 | | 15.7 |
| Contr_l + Triplet_L | GAT | 99.53 ± 0.06 | | 99.89 ± 0.02 | | 97.04 ± 0.67 | | 6.0 |
| Contr_l + Triplet_L | GCN | 99.34 ± 0.03 | | 99.84 ± 0.04 | | 93.61 ± 0.36 | | 14.3 |

Lp Aupr Continued (↑)

| Loss Type | Model | CORA | | Citeseer | | Bitcoin Fraud Transaction | | Average Rank |
|---|---|---|---|---|---|---|---|---|
| CrossE_L + PMI_L + Triplet_L | GAT | 99.57 ± 0.06 | | 99.85 ± 0.04 | | 92.56 ± 0.91 | | 12.0 |
| CrossE_L + Triplet_L | GAT | 99.74 ± 0.06 | | 99.92 ± 0.02 | | 96.34 ± 1.05 | | 2.3 |
| CrossE_L + Triplet_L | GCN | 99.56 ± 0.08 | | 99.86 ± 0.05 | | 94.78 ± 0.16 | | 7.3 |
| Triplet_L | GAT | 99.80 ± 0.04 | | 99.93 ± 0.02 | | 95.61 ± 0.37 | | 2.0 |
| Triplet_L | GCN | 99.55 ± 0.04 | | 99.87 ± 0.04 | | 94.47 ± 0.62 | | 7.7 |

Table 7. Lp Auroc Performance (↑): This table presents the top 10 models ranked by their average performance in terms of lp auroc. For the complete list of models, refer to Supplementary Information 2 Table 7. Top-ranked results are highlighted in red, second-ranked in blue, and third-ranked in green.

| Loss Type | Model | CORA | | Citeseer | | Bitcoin Fraud Transaction | | Average Rank |
|---|---|---|---|---|---|---|---|---|
| Contr_l + CrossE_L | GAT | 99.24 ± 0.08 | | 99.90 ± 0.02 | | 96.75 ± 1.37 | | 10.7 |
| Contr_l + CrossE_L + Triplet_L | GAT | 99.58 ± 0.09 | | 99.93 ± 0.02 | | 98.25 ± 0.53 | | 3.0 |
| Contr_l + Triplet_L | GAT | 99.55 ± 0.07 | | 99.92 ± 0.01 | | 98.40 ± 0.51 | | 4.0 |
| CrossE_L + PMI_L + Triplet_L | GAT | 99.55 ± 0.07 | | 99.89 ± 0.03 | | 95.07 ± 0.51 | | 12.7 |
| CrossE_L + Triplet_L | GAT | 99.76 ± 0.06 | | 99.95 ± 0.01 | | 97.78 ± 0.64 | | 2.0 |
| CrossE_L + Triplet_L | GCN | 99.54 ± 0.09 | | 99.89 ± 0.03 | | 96.80 ± 0.11 | | 8.3 |

Lp Auroc Continued (↑)

| Loss Type | Model | CORA | | Citeseer | | Bitcoin Fraud Transaction | | Average Rank |
|---|---|---|---|---|---|---|---|---|
| CrossE_L + Triplet_L | SAGE | 99.62 ± 0.06 | | 99.89 ± 0.03 | | 95.34 ± 0.45 | | 10.0 |
| Triplet_L | GAT | 99.83 ± 0.03 | | 99.95 ± 0.01 | | 97.31 ± 0.24 | | 2.3 |
| Triplet_L | GCN | 99.54 ± 0.04 | | 99.90 ± 0.03 | | 96.56 ± 0.41 | | 8.3 |
| Triplet_L | SAGE | 99.56 ± 0.09 | | 99.89 ± 0.04 | | 95.29 ± 0.29 | | 12.0 |

Table 8. Lp F1 Performance (↑): This table presents the top 10 models ranked by their average performance in terms of lp f1. For the complete list of models, refer to Supplementary Information 2 Table 8. Top-ranked results are highlighted in red, second-ranked in blue, and third-ranked in green.

| Loss Type | Model | CORA | | Citeseer | | Bitcoin Fraud Transaction | | Average Rank |
|---|---|---|---|---|---|---|---|---|
| Contr_l | GAT | 96.41 ± 0.15 | | 99.02 ± 0.06 | | 87.01 ± 0.50 | | 10.0 |
| Contr_l + CrossE_L | GAT | 96.16 ± 0.22 | | 98.99 ± 0.08 | | 88.98 ± 2.69 | | 12.0 |
| Contr_l + CrossE_L + Triplet_L | GAT | 97.48 ± 0.37 | | 99.29 ± 0.11 | | 92.17 ± 1.28 | | 2.7 |
| Contr_l + Triplet_L | GAT | 97.31 ± 0.37 | | 99.25 ± 0.12 | | 92.47 ± 1.39 | | 3.7 |
| CrossE_L + PMI_L + Triplet_L | GAT | 97.20 ± 0.22 | | 98.89 ± 0.10 | | 84.90 ± 1.18 | | 12.7 |
| CrossE_L + Triplet_L | GAT | 98.31 ± 0.25 | | 99.49 ± 0.08 | | 90.14 ± 1.99 | | 2.3 |
| CrossE_L + Triplet_L | GCN | 97.21 ± 0.36 | | 98.85 ± 0.20 | | 86.69 ± 0.45 | | 10.0 |

Lp F1 Continued (↑)

| Loss Type | Model | CORA | | Citeseer | | Bitcoin Fraud Transaction | | Average Rank |
|---|---|---|---|---|---|---|---|---|
| CrossE_L + Triplet_L | MPNN | 96.58 | ± | 98.91 | ± | 86.80 | ± | 10.7 |
| | | 0.43 | | 0.30 | | 0.87 | | |
| Triplet_L | GAT | 98.76 | ± | 99.61 | ± | 89.21 | ± | 2.0 |
| | | 0.24 | | 0.06 | | 0.89 | | |
| Triplet_L | GCN | 97.20 | ± | 98.90 | ± | 86.27 | ± | 10.7 |
| | | 0.15 | | 0.18 | | 0.94 | | |

Table 9. Lp Precision Performance (↑): This table presents the top 10 models ranked by their average performance in terms of lp precision. For the complete list of models, refer to Supplementary Information 2 Table 9. Top-ranked results are highlighted in red, second-ranked in blue, and third-ranked in green.

| Loss Type | Model | CORA | | Citeseer | | Bitcoin Fraud Transaction | | Average Rank |
|---|---|---|---|---|---|---|---|---|
| Contr_l + CrossE_L | GAT | 95.81 | ± | 98.70 | ± | 90.09 | ± | 18.3 |
| | | 0.48 | | 0.16 | | 1.92 | | |
| Contr_l + CrossE_L + PMI_L + Triplet_L | GAT | 96.99 | ± | 98.70 | ± | 85.76 | ± | 15.0 |
| | | 0.58 | | 0.06 | | 1.76 | | |
| Contr_l + CrossE_L + Triplet_L | GAT | 96.96 | ± | 99.03 | ± | 92.60 | ± | 5.3 |
| | | 0.32 | | 0.11 | | 1.17 | | |
| Contr_l + Triplet_L | GAT | 96.72 | ± | 98.89 | ± | 93.04 | ± | 7.3 |
| | | 0.53 | | 0.16 | | 1.57 | | |
| CrossE_L + PMI_L + Triplet_L | GAT | 97.23 | ± | 98.68 | ± | 87.38 | ± | 8.7 |
| | | 0.43 | | 0.07 | | 2.34 | | |
| CrossE_L + Triplet_L | GAT | 97.86 | ± | 99.19 | ± | 92.20 | ± | 2.3 |
| | | 0.31 | | 0.13 | | 0.95 | | |
| CrossE_L + Triplet_L | GCN | 97.18 | ± | 98.59 | ± | 86.97 | ± | 12.3 |
| | | 0.56 | | 0.45 | | 1.92 | | |
| PMI_L + Triplet_L | GAT | 97.04 | ± | 98.72 | ± | 85.19 | ± | 17.3 |
| | | 0.41 | | 0.18 | | 1.43 | | |

Lp Precision Continued (↑)

| Loss Type | Model | CORA | | Citeseer | | Bitcoin Fraud Transaction | | Average Rank |
|---|---|---|---|---|---|---|---|---|
| Triplet_L | GAT | 98.24 | ± 0.41 | 99.38 | ± 0.08 | 90.73 | ± 0.99 | 2.0 |
| Triplet_L | GCN | 96.95 | ± 0.24 | 98.80 | ± 0.07 | 87.97 | ± 1.08 | 8.3 |

Table 10. Lp Recall Performance (↑): This table presents the top 10 models ranked by their average performance in terms of lp recall. For the complete list of models, refer to Supplementary Information 2 Table 10. Top-ranked results are highlighted in red, second-ranked in blue, and third-ranked in green.

| Loss Type | Model | CORA | | Citeseer | | Bitcoin Fraud Transaction | | Average Rank |
|---|---|---|---|---|---|---|---|---|
| Contr_l + CrossE_L | GAT | 96.52 | ± 0.60 | 99.29 | ± 0.13 | 87.91 | ± 3.47 | 28.0 |
| Contr_l + CrossE_L + Triplet_L | GAT | 98.00 | ± 0.69 | 99.55 | ± 0.19 | 91.77 | ± 1.91 | 12.0 |
| Contr_l + Triplet_L | GAT | 97.90 | ± 0.41 | 99.61 | ± 0.19 | 91.90 | ± 1.37 | 12.0 |
| CrossE_L | GCN | 100.00 | ± 0.00 | 100.00 | ± 0.00 | 100.00 | ± 0.00 | 1.0 |
| CrossE_L | GIN | 100.00 | ± 0.00 | 100.00 | ± 0.00 | 100.00 | ± 0.00 | 2.0 |
| CrossE_L + Triplet_L | GAT | 98.77 | ± 0.28 | 99.79 | ± 0.06 | 88.25 | ± 3.68 | 16.7 |
| CrossE_L + Triplet_L | MPNN | 97.70 | ± 0.40 | 99.41 | ± 0.35 | 86.78 | ± 0.87 | 22.7 |
| CrossE_L + Triplet_L | SAGE | 97.99 | ± 0.26 | 99.06 | ± 0.29 | 86.08 | ± 1.49 | 27.7 |
| PR_L | PAGNN | 99.81 | ± 0.05 | 99.76 | ± 0.10 | 98.13 | ± 0.22 | 5.0 |

Continued on next page

Lp Recall Continued (↑)

| Loss Type | Model | CORA | | Citeseer | | Bitcoin Fraud Transaction | | Average Rank |
|-----------|-------|------|---|----------|---|----------|---|--------------|
| Triplet_L | GAT | 99.27 | ± | 99.84 | ± | 87.76 | ± | 17.3 |
|           |     | 0.15 |   | 0.05 |   | 1.52 |   | |

Table 11. Lp Specificity Performance (↑): This table presents the top 10 models ranked by their average performance in terms of lp specificity. For the complete list of models, refer to Supplementary Information 2 Table 11. Top-ranked results are highlighted in <span style="color:red">red</span>, second-ranked in <span style="color:blue">blue</span>, and third-ranked in <span style="color:green">green</span>.

| Loss Type | Model | CORA | | Citeseer | | Bitcoin Fraud Transaction | | Average Rank |
|-----------|-------|------|---|----------|---|----------|---|--------------|
| Contr_l + CrossE_L + PMI_L + Triplet_L | GAT | 96.81 | ± | 98.81 | ± | 92.87 | ± | 15.7 |
|  |  | 0.63 |  | 0.05 |  | 1.11 |  | |
| Contr_l + CrossE_L + Triplet_L | GAT | 96.76 | ± | 99.11 | ± | 96.11 | ± | 6.0 |
|  |  | 0.35 |  | 0.11 |  | 0.64 |  | |
| Contr_l + PR_L + Triplet_L | GAT | 96.34 | ± | 98.59 | ± | 93.66 | ± | 18.7 |
|  |  | 1.21 |  | 0.57 |  | 1.09 |  | |
| Contr_l + Triplet_L | GAT | 96.50 | ± | 98.98 | ± | 96.35 | ± | 7.7 |
|  |  | 0.59 |  | 0.15 |  | 0.85 |  | |
| CrossE_L + PMI_L + Triplet_L | GAT | 97.07 | ± | 98.79 | ± | 93.65 | ± | 10.0 |
|  |  | 0.47 |  | 0.06 |  | 1.35 |  | |
| CrossE_L + Triplet_L | GAT | 97.72 | ± | 99.26 | ± | 96.04 | ± | 2.3 |
|  |  | 0.33 |  | 0.12 |  | 0.52 |  | |
| CrossE_L + Triplet_L | GCN | 97.02 | ± | 98.71 | ± | 93.10 | ± | 14.3 |
|  |  | 0.61 |  | 0.42 |  | 1.24 |  | |
| PMI_L + Triplet_L | GAT | 96.86 | ± | 98.83 | ± | 92.41 | ± | 18.3 |
|  |  | 0.45 |  | 0.17 |  | 0.84 |  | |
| Triplet_L | GAT | 98.13 | ± | 99.43 | ± | 95.24 | ± | 3.3 |
|  |  | 0.44 |  | 0.07 |  | 0.57 |  | |
| Triplet_L | GCN | 96.77 | ± | 98.90 | ± | 93.84 | ± | 9.7 |
|  |  | 0.27 |  | 0.07 |  | 0.76 |  | |

### 1.1.3 Embedding–Adjacency Alignment.

Table 12. Cosine-Adj Corr Performance (↑): This table presents the top 10 models ranked by their average performance in terms of cosine-adj corr. For the complete list of models,refer to Supplementary Information 2 Table 12. Top-ranked results are highlighted in red, second-ranked in blue, and third-ranked in green.

| Loss Type | Model | CORA | | Citeseer | | Bitcoin Fraud Transaction | | Average Rank |
|---|---|---|---|---|---|---|---|---|
| Contr_l + CrossE_L | GAT | 11.93 ± 0.71 | | 12.89 ± 0.14 | | 6.18 ± 0.51 | | 15.0 |
| Contr_l + CrossE_L + PMI_L + Triplet_L | GAT | 13.23 ± 0.91 | | 12.32 ± 0.28 | | 4.97 ± 0.09 | | 16.0 |
| Contr_l + CrossE_L + Triplet_L | GAT | 13.67 ± 0.83 | | 12.92 ± 0.53 | | 7.01 ± 0.24 | | 3.0 |
| Contr_l + Triplet_L | GAT | 13.02 ± 0.70 | | 12.76 ± 0.61 | | 6.95 ± 0.16 | | 6.7 |
| CrossE_L + PMI_L + Triplet_L | GAT | 13.31 ± 0.35 | | 12.57 ± 0.35 | | 5.36 ± 0.26 | | 10.0 |
| CrossE_L + Triplet_L | GAT | 14.66 ± 0.66 | | 13.65 ± 0.13 | | 6.35 ± 0.49 | | 2.3 |
| CrossE_L + Triplet_L | GCN | 13.42 ± 0.48 | | 12.20 ± 0.46 | | 6.01 ± 0.06 | | 9.0 |
| PMI_L + Triplet_L | GAT | 13.38 ± 0.38 | | 12.67 ± 0.40 | | 5.10 ± 0.33 | | 12.7 |
| Triplet_L | GAT | 15.89 ± 0.65 | | 14.08 ± 0.16 | | 6.20 ± 0.26 | | 2.0 |
| Triplet_L | GCN | 13.43 ± 0.30 | | 12.32 ± 0.39 | | 5.71 ± 0.30 | | 9.3 |

Table 13. Dot-Adj Corr Performance (↑): This table presents the top 10 models ranked by their average performance in terms of dot-adj corr. For the complete list of models,refer to Supplementary Information 2 Table 13. Top-ranked results are highlighted in red, second-ranked in blue, and third-ranked in green.

| Loss Type | Model | CORA | | Citeseer | | Bitcoin Fraud Transaction | | Average Rank |
|---|---|---|---|---|---|---|---|---|
| Contr_l + CrossE_L | GAT | 11.93 ± 0.71 | | 12.89 ± 0.14 | | 6.18 ± 0.51 | | 15.0 |

Continued on next page

Dot-Adj Corr Continued (↑)

| Loss Type | Model | CORA | | | Citeseer | | | Bitcoin Fraud Transaction | | | Average Rank |
|-----------|-------|------|---|------|----------|---|------|---------------------------|---|------|--------------|
| Contr_l + CrossE_L + PMI_L + Triplet_L | GAT | 13.23 | ± | 0.91 | 12.32 | ± | 0.28 | 4.97 | ± | 0.09 | 16.0 |
| Contr_l + CrossE_L + Triplet_L | GAT | 13.67 | ± | 0.83 | 12.92 | ± | 0.53 | 7.01 | ± | 0.24 | 3.0 |
| Contr_l + Triplet_L | GAT | 13.02 | ± | 0.70 | 12.76 | ± | 0.61 | 6.95 | ± | 0.16 | 6.7 |
| CrossE_L + PMI_L + Triplet_L | GAT | 13.31 | ± | 0.35 | 12.57 | ± | 0.35 | 5.36 | ± | 0.26 | 10.0 |
| CrossE_L + Triplet_L | GAT | 14.66 | ± | 0.66 | 13.65 | ± | 0.13 | 6.35 | ± | 0.49 | 2.3 |
| CrossE_L + Triplet_L | GCN | 13.42 | ± | 0.48 | 12.20 | ± | 0.46 | 6.01 | ± | 0.06 | 9.0 |
| PMI_L + Triplet_L | GAT | 13.38 | ± | 0.38 | 12.67 | ± | 0.40 | 5.10 | ± | 0.33 | 12.7 |
| Triplet_L | GAT | 15.89 | ± | 0.65 | 14.08 | ± | 0.16 | 6.20 | ± | 0.26 | 2.0 |
| Triplet_L | GCN | 13.43 | ± | 0.30 | 12.32 | ± | 0.39 | 5.71 | ± | 0.30 | 9.3 |

Table 14. Euclidean-Adj Corr Performance (↑): This table presents the top 10 models ranked by their average performance in terms of euclidean-adj corr. For the complete list of models, refer to Supplementary Information 2 Table 14. Top-ranked results are highlighted in red, second-ranked in blue, and third-ranked in green.

| Loss Type | Model | CORA | | | Citeseer | | | Bitcoin Fraud Transaction | | | Average Rank |
|-----------|-------|------|---|------|----------|---|------|---------------------------|---|------|--------------|
| Contr_l + CrossE_L | GAT | 16.33 | ± | 0.79 | 18.93 | ± | 0.15 | 8.38 | ± | 0.76 | 15.0 |
| Contr_l + CrossE_L + Triplet_L | GAT | 18.84 | ± | 1.06 | 19.18 | ± | 0.78 | 9.48 | ± | 0.38 | 3.3 |

Continued on next page

Euclidean-Adj Corr Continued ($\uparrow$)

| Loss Type | Model | CORA | | Citeseer | | Bitcoin Fraud Transaction | | Average Rank |
|---|---|---|---|---|---|---|---|---|
| Contr_l + Triplet_L | GAT | 18.08 | ± | 18.97 | ± | 9.41 | ± | 7.0 |
| | | 0.96 | | 0.89 | | 0.25 | | |
| CrossE_L + PMI_L + Triplet_L | GAT | 18.47 | ± | 18.45 | ± | 7.05 | ± | 12.0 |
| | | 0.49 | | 0.50 | | 0.33 | | |
| CrossE_L + PMI_L + Triplet_L | GCN | 17.94 | ± | 17.58 | ± | 7.31 | ± | 15.3 |
| | | 0.86 | | 0.33 | | 0.35 | | |
| CrossE_L + Triplet_L | GAT | 20.27 | ± | 20.17 | ± | 8.43 | ± | 2.3 |
| | | 0.87 | | 0.22 | | 0.68 | | |
| CrossE_L + Triplet_L | GCN | 18.86 | ± | 18.25 | ± | 8.28 | ± | 7.7 |
| | | 0.62 | | 0.66 | | 0.08 | | |
| PMI_L + Triplet_L | GAT | 18.55 | ± | 18.60 | ± | 6.69 | ± | 14.3 |
| | | 0.50 | | 0.57 | | 0.44 | | |
| Triplet_L | GAT | 21.86 | ± | 20.80 | ± | 8.18 | ± | 2.7 |
| | | 0.84 | | 0.23 | | 0.35 | | |
| Triplet_L | GCN | 18.88 | ± | 18.44 | ± | 7.88 | ± | 8.3 |
| | | 0.39 | | 0.57 | | 0.37 | | |

Table 15. Graph Reconstruction Bce Loss Performance ($\downarrow$): This table presents the top 10 models ranked by their average performance in terms of graph reconstruction bce loss. For the complete list of models, refer to Supplementary Information 2 Table 15. Top-ranked results are highlighted in red, second-ranked in blue, and third-ranked in green.

| Loss Type | Model | CORA | | Citeseer | | Bitcoin Fraud Transaction | | Average Rank |
|---|---|---|---|---|---|---|---|---|
| Contr_l | ALL | 54.22 | ± | 53.77 | ± | 57.29 | ± | 13.0 |
| | | 0.20 | | 0.15 | | 0.10 | | |
| Contr_l | GAT | 54.02 | ± | 53.57 | ± | 56.86 | ± | 5.0 |
| | | 0.30 | | 0.13 | | 0.08 | | |
| Contr_l | GCN | 54.14 | ± | 53.55 | ± | 57.72 | ± | 8.3 |
| | | 0.13 | | 0.32 | | 0.32 | | |

Graph Reconstruction Bce Loss Continued (↓)

| Loss Type | Model | CORA | | Citeseer | | Bitcoin Fraud Transaction | | Average Rank |
|---|---|---|---|---|---|---|---|---|
| Contr_l | MPNN | 54.16 | ± | 53.27 | ± | 57.67 | ± | 5.3 |
| | | 0.32 | | 0.43 | | 0.11 | | |
| Contr_l + CrossE_L | GAT | 54.12 | ± | 53.90 | ± | 57.20 | ± | 12.3 |
| | | 0.45 | | 0.10 | | 0.28 | | |
| Contr_l + CrossE_L | GCN | 54.46 | ± | 53.41 | ± | 57.72 | ± | 11.3 |
| | | 0.25 | | 0.13 | | 0.09 | | |
| Contr_l + CrossE_L + Triplet_L | GAT | 54.47 | ± | 53.69 | ± | 57.16 | ± | 13.3 |
| | | 0.48 | | 0.32 | | 0.17 | | |
| Contr_l + Triplet_L | GAT | 54.23 | ± | 53.54 | ± | 57.35 | ± | 7.3 |
| | | 0.26 | | 0.35 | | 0.20 | | |
| Contr_l + Triplet_L | MPNN | 54.32 | ± | 53.54 | ± | 57.85 | ± | 11.7 |
| | | 0.19 | | 0.40 | | 0.31 | | |
| Triplet_L | ALL | 54.20 | ± | 53.54 | ± | 57.61 | ± | 7.7 |
| | | 0.30 | | 0.33 | | 0.13 | | |

### 1.1.4 Clustering Quality Metrics.

Table 16. Silhouette Performance (↑): This table presents the top 10 models ranked by their average performance in terms of silhouette. For the complete list of models, refer to Supplementary Information 2 Table 16. Top-ranked results are highlighted in red, second-ranked in blue, and third-ranked in green.

| Loss Type | Model | CORA | | Citeseer | | Bitcoin Fraud Transaction | | Average Rank |
|---|---|---|---|---|---|---|---|---|
| Contr_l | ALL | 13.75 | ± | 5.39 | ± | -2.96 | ± | 29.3 |
| | | 1.88 | | 0.84 | | 0.72 | | |
| Contr_l | GAT | 12.59 | ± | 4.15 | ± | -1.39 | ± | 29.7 |
| | | 0.76 | | 0.19 | | 0.12 | | |
| Contr_l | GIN | 15.02 | ± | 4.83 | ± | -3.61 | ± | 32.7 |
| | | 1.24 | | 0.72 | | 1.11 | | |
| Contr_l | MPNN | 13.47 | ± | 5.35 | ± | -2.63 | ± | 29.0 |
| | | 1.41 | | 0.98 | | 0.43 | | |

Continued on next page

Silhouette Continued (↑)

| Loss Type | Model | CORA | | Citeseer | | Bitcoin Fraud Transaction | | Average Rank |
|---|---|---|---|---|---|---|---|---|
| Contr_l + CrossE_L | GAT | 11.94 ± 0.95 | | 4.18 ± 0.23 | | -1.56 ± 0.20 | | 31.0 |
| Contr_l + CrossE_L | GIN | 14.17 ± 0.88 | | 4.65 ± 0.45 | | -3.05 ± 2.12 | | 31.3 |
| Contr_l + CrossE_L | MPNN | 14.34 ± 2.80 | | 5.85 ± 0.74 | | -3.21 ± 1.01 | | 29.3 |
| Contr_l + CrossE_L + Triplet_L | ALL | 12.83 ± 1.09 | | 4.52 ± 0.67 | | -2.50 ± 0.83 | | 30.7 |
| Contr_l + CrossE_L + Triplet_L | MPNN | 12.70 ± 1.07 | | 4.04 ± 0.28 | | -2.13 ± 0.60 | | 33.0 |
| Contr_l + Triplet_L | ALL | 12.97 ± 0.86 | | 4.12 ± 0.43 | | -2.12 ± 0.41 | | 31.3 |

Table 17. Calinski Harabasz Performance (↑): This table presents the top 10 models ranked by their average performance in terms of calinski harabasz. For the complete list of models, refer to Supplementary Information 2 Table 17. Top-ranked results are highlighted in red, second-ranked in blue, and third-ranked in green.

| Loss Type | Model | CORA | Citeseer | Bitcoin Fraud Transaction | Average Rank |
|---|---|---|---|---|---|
| Contr_l | ALL | 33999.80 ± 1554.88 | 20866.42 ± 2637.95 | 1599.91 ± 497.12 | 20.0 |
| Contr_l | MPNN | 28655.93 ± 1127.06 | 15089.31 ± 1261.35 | 2573.91 ± 972.52 | 12.0 |
| Contr_l + CrossE_L | ALL | 34582.58 ± 2397.13 | 19808.53 ± 3260.27 | 1265.13 ± 475.16 | 28.3 |
| Contr_l + CrossE_L | GIN | 24083.73 ± 754.84 | 12194.74 ± 1561.17 | 2492.66 ± 1152.14 | 26.7 |
| Contr_l + CrossE_L | MPNN | 29319.66 ± 1433.96 | 15618.32 ± 630.39 | 1682.53 ± 535.11 | 23.3 |

Continued on next page

Calinski Harabasz Continued (↑)

| Loss Type | Model | CORA | Citeseer | Bitcoin Fraud Transaction | Average Rank |
|-----------|-------|------|----------|---------------------------|--------------|
| Contr_l + CrossE_L + PMI_L + Triplet_L | ALL | 28131.10 ± 1789.24 | 15135.16 ± 2657.31 | 2327.37 ± 594.73 | 15.7 |
| Contr_l + CrossE_L + PR_L + Triplet_L | ALL | 26190.76 ± 3829.57 | 15076.12 ± 1234.50 | 2743.79 ± 1095.69 | 14.3 |
| Contr_l + PMI_L | ALL | 27155.80 ± 5178.65 | 15382.98 ± 2055.08 | 2467.89 ± 805.42 | 14.7 |
| Contr_l + PR_L + Triplet_L | ALL | 24655.94 ± 6414.50 | 13289.51 ± 1725.12 | 3197.13 ± 830.95 | 19.7 |
| PMI_L + Triplet_L | ALL | 23923.04 ± 1830.79 | 12575.87 ± 397.09 | 2306.15 ± 163.35 | 27.7 |

Table 18. Knn Consistency Performance (↑): This table presents the top 10 models ranked by their average performance in terms of knn consistency. For the complete list of models, refer to Supplementary Information 2 Table 18. Top-ranked results are highlighted in red, second-ranked in blue, and third-ranked in green.

| Loss Type | Model | CORA | Citeseer | Bitcoin Fraud Transaction | Average Rank |
|-----------|-------|------|----------|---------------------------|--------------|
| Contr_l + CrossE_L + PMI_L + PR_L | GAT | 84.44 ± 0.32 | 70.52 ± 0.27 | 75.36 ± 0.55 | 43.3 |
| Contr_l + CrossE_L + PMI_L + Triplet_L | GAT | 84.57 ± 0.17 | 70.64 ± 0.38 | 75.10 ± 0.23 | 44.0 |
| Contr_l + PMI_L + PR_L | GAT | 84.62 ± 0.35 | 70.25 ± 0.53 | 75.47 ± 0.42 | 42.3 |
| CrossE_L + PMI_L + Triplet_L | GAT | 84.65 ± 0.29 | 70.70 ± 0.28 | 75.10 ± 0.25 | 43.7 |
| CrossE_L + PR_L + Triplet_L | SAGE | 83.12 ± 1.20 | 69.32 ± 0.88 | 77.19 ± 0.82 | 42.3 |
| CrossE_L + Triplet_L | GAT | 85.07 ± 0.16 | 71.07 ± 0.33 | 74.90 ± 0.09 | 39.7 |

Continued on next page

Knn Consistency Continued (↑)

| Loss Type | Model | CORA | | Citeseer | | Bitcoin Fraud Transaction | | Average Rank |
|-----------|-------|------|---|----------|---|---------------------------|---|--------------|
| CrossE_L + Triplet_L | SAGE | 84.79 ± 0.21 | | 70.55 ± 0.57 | | 75.25 ± 0.19 | | 40.7 |
| PMI_L + PR_L | GAT | 84.21 ± 0.23 | | 69.64 ± 0.96 | | 75.94 ± 0.31 | | 36.0 |
| Triplet_L | GAT | 84.93 ± 0.14 | | 71.11 ± 0.35 | | 75.13 ± 0.15 | | 36.0 |
| Triplet_L | SAGE | 84.69 ± 0.21 | | 70.64 ± 0.39 | | 75.38 ± 0.17 | | 37.3 |

*1.1.5 Semantic Coherence and Ranking:*

Table 19. Coherence Performance (↑): This table presents the top 10 models ranked by their average performance in terms of coherence. For the complete list of models, refer to Supplementary Information 2 Table 19. Top-ranked results are highlighted in red, second-ranked in blue, and third-ranked in green.

| Loss Type | Model | CORA | | Citeseer | | Bitcoin Fraud Transaction | | Average Rank |
|-----------|-------|------|---|----------|---|---------------------------|---|--------------|
| Contr_l + CrossE_L + PR_L | GIN | 99.40 ± 1.35 | | 100.00 ± 0.00 | | 98.65 ± 3.02 | | 8.0 |
| Contr_l + CrossE_L + PR_L + Triplet_L | GIN | 100.00 ± 0.00 | | 69.58 ± 21.85 | | 99.96 ± 0.10 | | 13.3 |
| Contr_l + PR_L | GIN | 100.00 ± 0.00 | | 84.12 ± 12.29 | | 100.00 ± 0.00 | | 5.7 |
| Contr_l + PR_L + Triplet_L | GIN | 99.75 ± 0.57 | | 88.51 ± 22.65 | | 95.42 ± 10.25 | | 12.0 |
| CrossE_L | GCN | 97.61 ± 2.49 | | 99.94 ± 0.14 | | 99.46 ± 0.44 | | 8.3 |
| CrossE_L | GIN | 100.00 ± 0.01 | | 100.00 ± 0.00 | | 100.00 ± 0.00 | | 2.7 |
| CrossE_L + PR_L | GIN | 98.24 ± 3.92 | | 95.73 ± 7.25 | | 99.70 ± 0.67 | | 7.0 |

Coherence Continued (↑)

| Loss Type | Model | CORA | | Citeseer | | Bitcoin Fraud Transaction | | Average Rank |
|---|---|---|---|---|---|---|---|---|
| CrossE_L + PR_L + Triplet_L | GIN | 91.56 11.62 | ± | 95.32 6.71 | ± | 99.99 0.01 | | 7.7 |
| PR_L | GIN | 100.00 0.00 | ± | 95.16 7.14 | ± | 100.00 0.00 | | 5.0 |
| PR_L + Triplet_L | GIN | 99.95 0.10 | ± | 81.82 16.55 | ± | 100.00 0.00 | | 9.7 |

Table 20. Selfcluster Performance (↑): This table presents the top 10 models ranked by their average performance in terms of selfcluster. For the complete list of models, refer to Supplementary Information 2 Table 20. Top-ranked results are highlighted in red, second-ranked in blue, and third-ranked in green.

| Loss Type | Model | CORA | | Citeseer | | Bitcoin Fraud Transaction | | Average Rank |
|---|---|---|---|---|---|---|---|---|
| Contr_l + CrossE_L + PMI_L | PAGNN | -0.80 0.00 | ± | -0.79 0.00 | ± | -0.79 0.00 | | 14.3 |
| Contr_l + CrossE_L + PMI_L + PR_L | ALL | -0.79 0.01 | ± | -0.79 0.01 | ± | -0.79 0.00 | ± | 2.3 |
| Contr_l + CrossE_L + PMI_L + PR_L | PAGNN | -0.80 0.00 | ± | -0.79 0.01 | ± | -0.79 0.00 | | 16.7 |
| Contr_l + CrossE_L + PMI_L + Triplet_L | PAGNN | -0.80 0.00 | ± | -0.79 0.00 | ± | -0.79 0.00 | | 19.0 |
| Contr_l + CrossE_L + PR_L | GAT | -0.80 0.01 | ± | -0.79 0.00 | ± | -0.79 0.00 | | 20.3 |
| Contr_l + CrossE_L + PR_L | PAGNN | -0.79 0.00 | ± | -0.79 0.00 | ± | -0.79 0.00 | | 9.0 |
| Contr_l + PR_L | ALL | -0.79 0.00 | ± | -0.79 0.01 | ± | -0.79 0.00 | | 15.0 |
| Contr_l + PR_L | GIN | -0.79 0.00 | ± | -0.79 0.01 | ± | -0.79 0.00 | | 16.7 |

Selfcluster Continued (↑)

| Loss Type | Model | CORA | | Citeseer | | Bitcoin Fraud Transaction | | Average Rank |
|---|---|---|---|---|---|---|---|---|
| Contr_l + PR_L | MPNN | -0.79 | ± | -0.79 | ± | -0.79 | ± | 17.7 |
| | | 0.00 | | 0.00 | | 0.00 | | |
| Contr_l + PR_L | PAGNN | -0.79 | ± | -0.79 | ± | -0.79 | ± | 18.7 |
| | | 0.00 | | 0.00 | | 0.00 | | |

Table 21. Rankme Performance (↑): This table presents the top 10 models ranked by their average performance in terms of rankme. For the complete list of models, refer to Supplementary Information 2 Table 21. Top-ranked results are highlighted in red, second-ranked in blue, and third-ranked in green.

| Loss Type | Model | CORA | | Citeseer | | Bitcoin Fraud Transaction | | Average Rank |
|---|---|---|---|---|---|---|---|---|
| Contr_l + CrossE_L + PMI_L | GAT | 449.49 | ± | 457.64 | ± | 435.66 | ± | 7.0 |
| | | 1.34 | | 0.96 | | 1.17 | | |
| Contr_l + CrossE_L + PMI_L + PR_L + Triplet_L | GAT | 447.12 | ± | 453.59 | ± | 435.61 | ± | 10.7 |
| | | 2.17 | | 5.24 | | 4.92 | | |
| Contr_l + CrossE_L + PMI_L + Triplet_L | GAT | 449.78 | ± | 457.25 | ± | 439.97 | ± | 5.3 |
| | | 1.96 | | 0.46 | | 1.07 | | |
| Contr_l + PMI_L | GAT | 449.80 | ± | 457.47 | ± | 437.18 | ± | 5.0 |
| | | 0.97 | | 0.71 | | 1.43 | | |
| CrossE_L + PMI_L | GAT | 449.94 | ± | 458.30 | ± | 436.27 | ± | 4.7 |
| | | 0.74 | | 0.32 | | 1.44 | | |
| CrossE_L + PMI_L + PR_L + Triplet_L | GAT | 446.64 | ± | 452.39 | ± | 434.67 | ± | 13.0 |
| | | 1.62 | | 3.50 | | 3.88 | | |
| CrossE_L + PMI_L + Triplet_L | GAT | 449.59 | ± | 457.62 | ± | 441.40 | ± | 4.0 |
| | | 0.25 | | 0.51 | | 0.70 | | |
| PMI_L | GAT | 451.40 | ± | 458.38 | ± | 436.68 | ± | 3.7 |
| | | 1.45 | | 0.87 | | 1.67 | | |
| PMI_L + Triplet_L | GAT | 449.19 | ± | 457.69 | ± | 440.38 | ± | 4.7 |
| | | 0.62 | | 0.43 | | 0.66 | | |

Rankme Continued (↑)

| Loss Type | Model | CORA | Citeseer | Bitcoin Fraud Transaction | Average Rank |
|-----------|-------|------|----------|---------------------------|--------------|
| Triplet_L | GAT | 444.68 ± 0.94 | 453.75 ± 0.48 | 442.15 ± 0.94 | 9.7 |

## 1.2 Inductive results tables

**Pretrained on Cora and Citeseer datasets:** In these results we have trained the model on Cora and Citeseer dataset, and applied the model to generate node embedding on unknown dataset. The results represented as *Data used for pretraining ↓ Applied Data.*

### 1.2.1 *Assessment of Learned Representations through Node Classification.* In this setting the embedding is generated based on pretrained model. Further the resulting node embeddings are then used as input to a Multi-Layer Perceptron (MLP) for a supervised downstream task. The table below presents the results based on node classification accuracy.

Table 22. Node Cls Accuracy Performance (↑: This table presents the top 10 models ranked by their average performance in terms of node cls accuracy. For the complete list of models,refer to Supplementary Information 2 Table 22. Top-ranked results are highlighted in red, second-ranked in blue, and third-ranked in green.

| Loss Type | Model | Cora ↓ Citeseer | Cora ↓ Bitcoin | Citeseer ↓ Cora | Citeseer ↓ Bitcoin | Average Rank |
|-----------|-------|-----------------|----------------|-----------------|--------------------|--------------|
| Contr_l + CrossE_L + PMI_L | GAT | 62.61 ± 2.23 | 75.86 ± 0.58 | 76.13 ± 1.64 | 75.28 ± 0.70 | 22.125 |
| Contr_l + CrossE_L + PMI_L + PR_L + Triplet_L | GAT | 63.00 ± 1.56 | 75.62 ± 0.36 | 76.97 ± 1.37 | 75.46 ± 0.44 | 19.375 |
| Contr_l + CrossE_L + PMI_L + Triplet_L | GAT | 64.02 ± 1.41 | 75.58 ± 0.37 | 77.68 ± 1.56 | 75.52 ± 0.54 | 14.2 |
| Contr_l + Triplet_L | GAT | 63.75 ± 1.46 | 75.14 ± 0.50 | 77.38 ± 1.06 | 75.66 ± 0.61 | 20.375 |
| CrossE_L + PMI_L + PR_L | MPNN | 62.04 ± 1.40 | 76.18 ± 1.01 | 76.53 ± 1.22 | 76.04 ± 0.80 | 17.75 |
| CrossE_L + PMI_L + Triplet_L | GAT | 63.42 ± 2.18 | 75.30 ± 0.53 | 76.83 ± 0.43 | 75.60 ± 0.63 | 21.375 |

Node Cls Accuracy (continued) (↑)

| Loss Type | Model | Cora ↓ Citeseer | Cora ↓ Bitcoin | Citeseer ↓ Cora | Citeseer ↓ Bitcoin | Average Rank |
|---|---|---|---|---|---|---|
| CrossE_L + Triplet_L | GAT | 63.36 ± 1.98 | 75.84 ± 0.31 | 77.38 ± 1.15 | 76.04 ± 0.27 | 13.1 |
| PMI_L | GAT | 63.16 ± 2.77 | 75.70 ± 0.59 | 76.01 ± 1.47 | 75.46 ± 0.30 | 21.75 |
| PMI_L + Triplet_L | MPNN | 61.29 ± 2.12 | 75.80 ± 0.72 | 77.71 ± 1.59 | 75.48 ± 0.87 | 20.5 |
| Triplet_L | GAT | 63.99 ± 1.05 | 75.50 ± 0.78 | 78.49 ± 1.53 | 75.36 ± 0.48 | 16.9 |

Table 23. Node Cls F1 Performance (↑): This table presents the top 10 models ranked by their average performance in terms of node cls f1. For the complete list of models, refer to Supplementary Information 2 Table 23. Top-ranked results are highlighted in red, second-ranked in blue, and third-ranked in green.

| Loss Type | Model | Cora ↓ Citeseer | Cora ↓ Bitcoin | Citeseer ↓ Cora | Citeseer ↓ Bitcoin | Average Rank |
|---|---|---|---|---|---|---|
| Contr_l + CrossE_L + PMI_L | GAT | 55.07 ± 1.99 | 42.26 ± 0.73 | 73.20 ± 1.66 | 41.31 ± 0.74 | 16.5 |
| Contr_l + CrossE_L + PMI_L + PR_L + Triplet_L | GAT | 54.78 ± 1.65 | 41.61 ± 0.57 | 74.45 ± 1.70 | 41.74 ± 0.76 | 14.0 |
| Contr_l + CrossE_L + PMI_L + Triplet_L | GAT | 56.40 ± 1.83 | 41.39 ± 0.60 | 75.16 ± 1.56 | 41.49 ± 0.65 | 12.0 |
| Contr_l + Triplet_L | GAT | 56.10 ± 1.42 | 40.56 ± 0.84 | 74.32 ± 1.42 | 41.49 ± 1.08 | 19.25 |
| CrossE_L + PMI_L + PR_L | MPNN | 54.65 ± 2.44 | 41.72 ± 1.52 | 71.92 ± 1.34 | 41.49 ± 1.21 | 21.125 |
| CrossE_L + PMI_L + PR_L + Triplet_L | GAT | 56.44 ± 2.71 | 41.36 ± 1.14 | 72.74 ± 2.13 | 40.50 ± 1.14 | 20.875 |
| CrossE_L + PMI_L + Triplet_L | GAT | 55.80 ± 3.11 | 41.29 ± 0.77 | 74.00 ± 0.84 | 41.48 ± 0.95 | 17.25 |

Node Cls F1 (continued) (↑)

| Loss Type | Model | Cora ↓ Citeseer | | Cora ↓ Bitcoin | | Citeseer ↓ Cora | | Citeseer ↓ Bitcoin | | Average Rank |
|---|---|---|---|---|---|---|---|---|---|---|
| CrossE_L + Triplet_L | GAT | 55.21 ± 2.04 | | 41.79 ± 1.06 | | 75.24 ± 2.04 | | 41.76 ± 0.32 | | 11.0 |
| PMI_L | GAT | 55.82 ± 2.24 | | 41.65 ± 0.74 | | 73.12 ± 2.28 | | 41.59 ± 0.50 | | 15.25 |
| Triplet_L | GAT | 55.98 ± 1.52 | | 41.28 ± 1.31 | | 75.77 ± 2.12 | | 41.42 ± 0.63 | | 14.5 |

Table 24. Node Cls Precision Performance (↑): This table presents the top 10 models ranked by their average performance in terms of node cls precision. For the complete list of models, refer to Supplementary Information 2 Table 24. Top-ranked results are highlighted in <span style="color:red">red</span>, second-ranked in <span style="color:blue">blue</span>, and third-ranked in <span style="color:green">green</span>.

| Loss Type | Model | Cora ↓ Citeseer | | Cora ↓ Bitcoin | | Citeseer ↓ Cora | | Citeseer ↓ Bitcoin | | Average Rank |
|---|---|---|---|---|---|---|---|---|---|---|
| Contr_l + CrossE_L + PMI_L + Triplet_L | GAT | 58.78 ± 3.73 | | 48.54 ± 0.81 | | 78.42 ± 1.47 | | 48.35 ± 1.23 | | 30.5 |
| Contr_l + PMI_L + PR_L | GAT | 61.04 ± 1.52 | | 48.31 ± 0.94 | | 76.83 ± 2.15 | | 48.68 ± 0.58 | | 31.375 |
| CrossE_L + PMI_L + PR_L | MPNN | 58.08 ± 5.72 | | 49.77 ± 1.31 | | 75.28 ± 1.32 | | 49.65 ± 1.05 | | 31.2 |
| CrossE_L + PMI_L + PR_L + Triplet_L | MPNN | 58.83 ± 3.14 | | 48.52 ± 1.12 | | 75.83 ± 1.11 | | 48.70 ± 0.96 | | 35.125 |
| CrossE_L + PMI_L + Triplet_L | MPNN | 59.49 ± 4.51 | | 49.81 ± 1.65 | | 76.39 ± 1.40 | | 47.53 ± 1.31 | | 33.25 |
| CrossE_L + Triplet_L | GAT | 58.39 ± 6.90 | | 48.79 ± 0.45 | | 77.58 ± 0.90 | | 49.39 ± 0.54 | | 27.2 |
| PMI_L | GAT | 62.23 ± 2.12 | | 48.60 ± 1.07 | | 76.52 ± 2.19 | | 48.03 ± 0.44 | | 32.25 |
| PMI_L + Triplet_L | MPNN | 55.83 ± 3.58 | | 49.56 ± 0.82 | | 76.23 ± 1.34 | | 48.90 ± 1.15 | | 34.75 |

Node Cls Precision (continued) (↑)

| Loss Type | Model | Cora ↓ Citeseer | | Cora ↓ Bitcoin | | Citeseer ↓ Cora | | Citeseer ↓ Bitcoin | | Average Rank |
|---|---|---|---|---|---|---|---|---|---|---|
| Triplet_L | GAT | 62.04 | ± 5.31 | 48.34 | ± 0.79 | 78.04 | ± 1.86 | 47.91 | ± 0.68 | 33.25 |
| Triplet_L | MPNN | 60.16 | ± 5.33 | 48.34 | ± 1.45 | 76.96 | ± 2.64 | 47.87 | ± 1.02 | 35.625 |

Table 25. Node Cls Recall (Sensitivity) Performance (↑): This table presents the top 10 models ranked by their average performance in terms of node cls recall (sensitivity). For the complete list of models, refer to Supplementary Information 2 Table 25. Top-ranked results are highlighted in red, second-ranked in blue, and third-ranked in green.

| Loss Type | Model | Cora ↓ Citeseer | | Cora ↓ Bitcoin | | Citeseer ↓ Cora | | Citeseer ↓ Bitcoin | | Average Rank |
|---|---|---|---|---|---|---|---|---|---|---|
| Contr_l + CrossE_L + PMI_L | GAT | 56.58 | ± 2.06 | 41.56 | ± 0.60 | 71.50 | ± 1.85 | 40.80 | ± 0.58 | 17.0 |
| Contr_l + CrossE_L + PMI_L + PR_L + Triplet_L | GAT | 56.56 | ± 1.60 | 41.05 | ± 0.44 | 72.64 | ± 1.95 | 41.15 | ± 0.59 | 14.625 |
| Contr_l + CrossE_L + PMI_L + Triplet_L | GAT | 57.89 | ± 1.41 | 40.89 | ± 0.46 | 73.12 | ± 1.60 | 40.96 | ± 0.49 | 12.9 |
| Contr_l + Triplet_L | GAT | 57.60 | ± 1.28 | 40.26 | ± 0.62 | 72.83 | ± 1.82 | 40.97 | ± 0.83 | 19.125 |
| CrossE_L + PMI_L + PR_L | MPNN | 56.08 | ± 1.65 | 41.17 | ± 1.15 | 70.23 | ± 2.03 | 40.99 | ± 0.92 | 21.25 |
| CrossE_L + PMI_L + PR_L + Triplet_L | GAT | 57.92 | ± 2.28 | 40.86 | ± 0.86 | 70.50 | ± 1.84 | 40.19 | ± 0.85 | 23.0 |
| CrossE_L + PMI_L + Triplet_L | GAT | 57.37 | ± 2.52 | 40.80 | ± 0.59 | 72.12 | ± 1.13 | 40.95 | ± 0.73 | 17.25 |
| CrossE_L + Triplet_L | GAT | 56.95 | ± 1.89 | 41.19 | ± 0.36 | 73.72 | ± 1.61 | 41.17 | ± 0.24 | 11.8 |
| PMI_L | GAT | 57.28 | ± 2.38 | 41.08 | ± 0.58 | 71.09 | ± 2.30 | 41.03 | ± 0.39 | 16.875 |

Node Cls Recall (Sensitivity) (continued) (↑)

| Loss Type | Model | Cora ↓ Citeseer | ± | Cora ↓ Bitcoin | ± | Citeseer ↓ Cora | ± | Citeseer ↓ Bitcoin | ± | Average Rank |
|-----------|-------|----------------|---|----------------|---|-----------------|---|--------------------|---|--------------|
| Triplet_L | GAT | 57.56 <br> 1.30 | ± | 40.81 <br> 1.01 | ± | 74.23 <br> 2.21 | ± | 40.90 <br> 0.49 | | 14.0 |

*1.2.2 Assessment of Learned Representations through Link Prediction (LP).* In this task, The resulting node embedding is used for link prediction task.

Table 26. Lp Accuracy Performance (↑): This table presents the top 10 models ranked by their average performance in terms of lp accuracy. For the complete list of models, refer to Supplementary Information 2 Table 26. Top-ranked results are highlighted in red, second-ranked in blue, and third-ranked in green.

| Loss Type | Model | Cora ↓ Citeseer | ± | Cora ↓ Bitcoin | ± | Citeseer ↓ Cora | ± | Citeseer ↓ Bitcoin | ± | Average Rank |
|-----------|-------|----------------|---|----------------|---|-----------------|---|--------------------|---|--------------|
| Contr_l + CrossE_L + PMI_L | GAT | 99.01 <br> 0.19 | ± | 89.18 <br> 0.75 | ± | 97.58 <br> 0.23 | ± | 89.44 <br> 0.36 | ± | 10.625 |
| Contr_l + CrossE_L + PMI_L + Triplet_L | GAT | 99.04 <br> 0.07 | ± | 89.50 <br> 0.35 | ± | 97.33 <br> 0.13 | ± | 89.27 <br> 0.53 | ± | 9.25 |
| Contr_l + CrossE_L + Triplet_L | GAT | 98.95 <br> 0.19 | ± | 88.78 <br> 0.77 | ± | 97.61 <br> 0.28 | ± | 89.22 <br> 0.42 | ± | 17.25 |
| Contr_l + PMI_L | GAT | 99.02 <br> 0.19 | ± | 89.71 <br> 0.57 | ± | 97.45 <br> 0.20 | ± | 89.33 <br> 0.84 | ± | 7.5 |
| CrossE_L + PMI_L | GAT | 99.03 <br> 0.14 | ± | 89.19 <br> 0.10 | ± | 97.56 <br> 0.24 | ± | 89.18 <br> 0.18 | ± | 12.375 |
| CrossE_L + PMI_L + Triplet_L | GAT | 99.01 <br> 0.10 | ± | 89.24 <br> 1.10 | ± | 97.48 <br> 0.23 | ± | 89.42 <br> 0.34 | ± | 10.5 |
| CrossE_L + Triplet_L | GAT | 99.16 <br> 0.10 | ± | 89.78 <br> 0.40 | ± | 97.67 <br> 0.28 | ± | 89.79 <br> 0.74 | ± | 2.2 |
| PMI_L | GAT | 99.14 <br> 0.13 | ± | 89.44 <br> 0.95 | ± | 97.53 <br> 0.31 | ± | 89.74 <br> 0.39 | ± | 5.8 |
| PMI_L + Triplet_L | GAT | 99.05 <br> 0.09 | ± | 89.20 <br> 0.79 | ± | 97.51 <br> 0.16 | ± | 89.48 <br> 0.35 | ± | 9.25 |

Continued on next page

Lp Accuracy (continued) (↑)

| Loss Type | Model | Cora ↓ Citeseer | | Cora ↓ Bitcoin | | Citeseer ↓ Cora | | Citeseer ↓ Bitcoin | | Average Rank |
|-----------|-------|-----------------|---|----------------|---|-----------------|---|--------------------|---|--------------|
| Triplet_L | GAT | 99.24 0.12 | ± | 89.25 0.46 | ± | 97.83 0.26 | ± | 89.52 0.79 | ± | 5.0 |

Table 27. Lp Aupr Performance (↑): This table presents the top 10 models ranked by their average performance in terms of lp aupr. For the complete list of models, refer to Supplementary Information 2 Table 27. Top-ranked results are highlighted in red, second-ranked in blue, and third-ranked in green.

| Loss Type | Model | Cora ↓ Citeseer | | Cora ↓ Bitcoin | | Citeseer ↓ Cora | | Citeseer ↓ Bitcoin | | Average Rank |
|-----------|-------|-----------------|---|----------------|---|-----------------|---|--------------------|---|--------------|
| Contr_l + CrossE_L + PMI_L | GAT | 99.91 0.04 | ± | 92.42 0.61 | ± | 99.66 0.05 | ± | 92.32 0.35 | ± | 11.25 |
| Contr_l + CrossE_L + PMI_L + Triplet_L | GAT | 99.91 0.02 | ± | 92.69 0.23 | ± | 99.63 0.03 | ± | 92.45 0.40 | ± | 9.5 |
| Contr_l + CrossE_L + Triplet_L | GAT | 99.91 0.02 | ± | 91.88 0.80 | ± | 99.67 0.04 | ± | 92.31 0.45 | ± | 17.125 |
| Contr_l + PMI_L | GAT | 99.90 0.03 | ± | 92.83 0.80 | ± | 99.65 0.03 | ± | 92.27 0.99 | ± | 12.0 |
| CrossE_L + PMI_L + Triplet_L | GAT | 99.90 0.01 | ± | 92.04 1.13 | ± | 99.65 0.04 | ± | 92.43 0.55 | ± | 17.5 |
| CrossE_L + Triplet_L | GAT | 99.92 0.02 | ± | 92.86 0.62 | ± | 99.70 0.02 | ± | 92.72 0.83 | ± | 2.6 |
| PMI_L | GAT | 99.90 0.02 | ± | 92.41 1.02 | ± | 99.69 0.07 | ± | 92.70 0.48 | ± | 9.2 |
| PMI_L + Triplet_L | GAT | 99.90 0.02 | ± | 92.23 0.86 | ± | 99.66 0.03 | ± | 92.77 0.65 | ± | 11.75 |
| Triplet_L | GAT | 99.93 0.02 | ± | 92.30 0.78 | ± | 99.71 0.04 | ± | 92.51 1.08 | ± | 7.8 |
| Triplet_L | GCN | 99.86 0.02 | ± | 92.71 0.77 | ± | 99.43 0.12 | ± | 92.48 0.53 | ± | 16.0 |

Table 28. Lp Auroc Performance (↑): This table presents the top 10 models ranked by their average performance in terms of lp auroc. For the complete list of models,refer to Supplementary Information 2 Table 28. Top-ranked results are highlighted in <span style="color:red">red</span>, second-ranked in <span style="color:blue">blue</span>, and third-ranked in <span style="color:green">green</span>.

| Loss Type | Model | Cora ↓ Citeseer | | Cora ↓ Bitcoin | | Citeseer ↓ Cora | | Citeseer ↓ Bitcoin | | Average Rank |
|---|---|---|---|---|---|---|---|---|---|---|
| Contr_l + CrossE_L + PMI_L | GAT | 99.92 0.03 | ± | 94.88 0.35 | ± | 99.66 0.06 | ± | 94.84 0.33 | ± | 11.0 |
| Contr_l + CrossE_L + PMI_L + Triplet_L | GAT | 99.92 0.01 | ± | 95.15 0.20 | ± | 99.63 0.03 | ± | 94.85 0.32 | ± | 8.875 |
| Contr_l + CrossE_L + Triplet_L | GAT | 99.92 0.02 | ± | 94.43 0.67 | ± | 99.68 0.05 | ± | 94.80 0.37 | ± | 15.75 |
| Contr_l + PMI_L | GAT | 99.92 0.02 | ± | 95.19 0.70 | ± | 99.65 0.02 | ± | 94.73 0.81 | ± | 9.25 |
| CrossE_L + PMI_L | GAT | 99.93 0.02 | ± | 94.66 0.32 | ± | 99.68 0.03 | ± | 94.40 0.21 | ± | 15.75 |
| CrossE_L + PMI_L + Triplet_L | GAT | 99.92 0.01 | ± | 94.63 0.76 | ± | 99.65 0.04 | ± | 95.01 0.28 | ± | 13.25 |
| CrossE_L + Triplet_L | GAT | 99.93 0.02 | ± | 95.30 0.47 | ± | 99.71 0.02 | ± | 95.13 0.61 | ± | <span style="background-color:#f08080">2.5</span> |
| PMI_L | GAT | 99.92 0.02 | ± | 94.89 0.72 | ± | 99.68 0.07 | ± | 95.10 0.39 | ± | <span style="background-color:#90ee90">8.4</span> |
| PMI_L + Triplet_L | GAT | 99.92 0.02 | ± | 94.79 0.70 | ± | 99.66 0.02 | ± | 94.98 0.64 | ± | 10.625 |
| Triplet_L | GAT | 99.94 0.02 | ± | 94.84 0.60 | ± | 99.72 0.03 | ± | 94.93 0.78 | ± | <span style="background-color:#6a5acd">7.0</span> |

Table 29. Lp F1 Performance (↑): This table presents the top 10 models ranked by their average performance in terms of lp f1. For the complete list of models, refer to Supplementary Information 2 Table 29. Top-ranked results are highlighted in red, second-ranked in blue, and third-ranked in green.

| Loss Type | Model | Cora ↓ Citeseer | | Cora ↓ Bitcoin | | Citeseer ↓ Cora | | Citeseer ↓ Bitcoin | | Average Rank |
|---|---|---|---|---|---|---|---|---|---|---|
| Contr_l + CrossE_L + PMI_L | GAT | 98.96 0.20 | ± | 84.22 0.94 | ± | 97.65 0.23 | ± | 84.55 0.37 | ± | 9.625 |
| Contr_l + CrossE_L + PMI_L + Triplet_L | GAT | 99.00 0.07 | ± | 84.77 0.40 | ± | 97.39 0.13 | ± | 84.34 0.60 | ± | 8.0 |
| Contr_l + CrossE_L + Triplet_L | GAT | 98.90 0.19 | ± | 83.69 1.10 | ± | 97.67 0.27 | ± | 84.20 0.51 | ± | 16.5 |
| Contr_l + PMI_L | GAT | 98.97 0.20 | ± | 84.91 0.98 | ± | 97.53 0.19 | ± | 84.30 1.09 | ± | 7.75 |
| CrossE_L + PMI_L | GAT | 98.99 0.14 | ± | 83.99 0.22 | ± | 97.63 0.23 | ± | 83.94 0.49 | ± | 13.75 |
| CrossE_L + PMI_L + Triplet_L | GAT | 98.96 0.10 | ± | 84.17 1.58 | ± | 97.55 0.23 | ± | 84.56 0.59 | ± | 10.875 |
| CrossE_L + Triplet_L | GAT | 99.12 0.11 | ± | 84.97 0.52 | ± | 97.74 0.28 | ± | 85.03 0.86 | ± | 2.0 |
| PMI_L | GAT | 99.10 0.14 | ± | 84.51 1.35 | ± | 97.61 0.29 | ± | 84.92 0.55 | ± | 5.1 |
| PMI_L + Triplet_L | GAT | 99.01 0.09 | ± | 84.31 1.12 | ± | 97.58 0.16 | ± | 84.67 0.63 | ± | 7.375 |
| Triplet_L | GAT | 99.20 0.12 | ± | 84.31 0.77 | ± | 97.89 0.25 | ± | 84.64 1.18 | ± | 4.9 |

Table 30. Lp Precision Performance (↑): This table presents the top 10 models ranked by their average performance in terms of lp precision. For the complete list of models,refer to Supplementary Information 2 Table 30. Top-ranked results are highlighted in red, second-ranked in blue, and third-ranked in green.

| Loss Type | Model | Cora ↓ Citeseer | Cora ↓ Bitcoin | Citeseer ↓ Cora | Citeseer ↓ Bitcoin | Average Rank |
|-----------|-------|-----------------|----------------|-----------------|--------------------|--------------|
| Contr_l + CrossE_L + PMI_L | GAT | 98.82 ± 0.09 | 85.17 ± 1.94 | 97.53 ± 0.33 | 85.78 ± 1.47 | 25.375 |
| Contr_l + CrossE_L + PMI_L + Triplet_L | GAT | 98.85 ± 0.10 | 85.27 ± 1.19 | 97.56 ± 0.30 | 85.36 ± 1.81 | 25.5 |
| Contr_l + PMI_L | GAT | 98.96 ± 0.17 | 86.28 ± 1.15 | 97.06 ± 0.24 | 86.05 ± 1.99 | 14.8 |
| CrossE_L + PMI_L | GAT | 98.84 ± 0.24 | 86.29 ± 0.94 | 97.53 ± 0.20 | 86.34 ± 1.07 | 14.0 |
| CrossE_L + PMI_L + PR_L + Triplet_L | GAT | 98.84 ± 0.16 | 86.67 ± 1.11 | 96.97 ± 0.89 | 84.75 ± 0.41 | 24.625 |
| CrossE_L + PMI_L + Triplet_L | GAT | 98.94 ± 0.28 | 85.85 ± 1.87 | 97.36 ± 0.31 | 85.49 ± 0.94 | 20.375 |
| CrossE_L + Triplet_L | GAT | 98.82 ± 0.14 | 86.62 ± 1.26 | 97.32 ± 0.34 | 86.55 ± 2.32 | 13.2 |
| PMI_L | GAT | 98.93 ± 0.15 | 85.96 ± 1.67 | 97.30 ± 0.66 | 86.53 ± 1.26 | 16.0 |
| Triplet_L | GAT | 99.02 ± 0.24 | 85.24 ± 0.89 | 97.64 ± 0.61 | 85.94 ± 1.41 | 19.5 |
| Triplet_L | GCN | 98.57 ± 0.10 | 86.05 ± 1.68 | 96.35 ± 0.71 | 85.83 ± 1.55 | 26.75 |

Table 31. Lp Recall Performance (↑): This table presents the top 10 models ranked by their average performance in terms of lp recall. For the complete list of models, refer to Supplementary Information 2 Table 31. Top-ranked results are highlighted in red, second-ranked in blue, and third-ranked in green.

| Loss Type | Model | Cora ↓ Citeseer | | Cora ↓ Bitcoin | | Citeseer ↓ Cora | | Citeseer ↓ Bitcoin | | Average Rank |
|---|---|---|---|---|---|---|---|---|---|---|
| Contr_l + CrossE_L + PMI_L | GAT | 99.10 ± 0.31 | | 83.31 ± 0.93 | | 97.77 ± 0.49 | | 83.39 ± 1.04 | | 43.25 |
| Contr_l + CrossE_L + PMI_L + Triplet_L | GAT | 99.15 ± 0.18 | | 84.29 ± 0.77 | | 97.24 ± 0.49 | | 83.39 ± 1.24 | | 40.125 |
| Contr_l + CrossE_L + Triplet_L | GAT | 98.92 ± 0.34 | | 83.09 ± 1.60 | | 97.77 ± 0.47 | | 82.88 ± 1.14 | | 48.875 |
| Contr_l + PMI_L | GAT | 98.98 ± 0.32 | | 83.63 ± 2.14 | | 98.00 ± 0.27 | | 82.63 ± 0.43 | | 45.625 |
| CrossE_L | GIN | 100.00 ± 0.00 | | 100.00 ± 0.00 | | 100.00 ± 0.00 | | 100.00 ± 0.00 | | 5.4 |
| CrossE_L + PMI_L + Triplet_L | GAT | 98.98 ± 0.30 | | 82.55 ± 1.39 | | 97.74 ± 0.44 | | 83.67 ± 1.57 | | 46.5 |
| CrossE_L + Triplet_L | GAT | 99.42 ± 0.15 | | 83.40 ± 0.98 | | 98.17 ± 0.49 | | 83.60 ± 0.79 | | 38.2 |
| PMI_L | GAT | 99.26 ± 0.18 | | 83.11 ± 1.20 | | 97.92 ± 0.24 | | 83.38 ± 1.24 | | 42.25 |
| PMI_L + Triplet_L | GAT | 99.17 ± 0.20 | | 83.73 ± 1.08 | | 97.78 ± 0.44 | | 83.87 ± 1.29 | | 37.2 |
| Triplet_L | GAT | 99.38 ± 0.10 | | 83.43 ± 1.59 | | 98.14 ± 0.15 | | 83.40 ± 1.68 | | 39.25 |

Table 32. Lp Specificity Performance (↑): This table presents the top 10 models ranked by their average performance in terms of lp specificity. For the complete list of models, refer to Supplementary Information 2 Table 32. Top-ranked results are highlighted in <span style="color:red">red</span>, second-ranked in <span style="color:blue">blue</span>, and third-ranked in <span style="color:green">green</span>.

| Loss Type | Model | Cora ↓ Citeseer | | Cora ↓ Bitcoin | | Citeseer ↓ Cora | | Citeseer ↓ Bitcoin | | Average Rank |
|---|---|---|---|---|---|---|---|---|---|---|
| Contr_l | GAT | 98.81 | ± 0.20 | 91.87 | ± 0.71 | 97.47 | ± 0.36 | 93.70 | ± 0.75 | 30.25 |
| Contr_l + CrossE_L | GCN | 98.75 | ± 0.20 | 92.58 | ± 1.07 | 96.15 | ± 0.31 | 93.42 | ± 0.53 | 28.25 |
| Contr_l + CrossE_L + PMI_L + Triplet_L | GCN | 98.38 | ± 0.34 | 93.30 | ± 0.55 | 96.02 | ± 0.31 | 92.67 | ± 0.58 | 30.125 |
| Contr_l + PMI_L | GAT | 99.05 | ± 0.16 | 92.94 | ± 0.78 | 96.87 | ± 0.26 | 92.88 | ± 1.14 | 19.75 |
| CrossE_L + PMI_L | GAT | 98.94 | ± 0.22 | 93.10 | ± 0.64 | 97.39 | ± 0.21 | 93.13 | ± 0.78 | <span style="background-color:#f08080">16.6</span> |
| CrossE_L + PMI_L + PR_L + Triplet_L | GAT | 98.94 | ± 0.15 | 93.28 | ± 0.72 | 96.82 | ± 0.92 | 92.13 | ± 0.24 | 27.625 |
| CrossE_L + PMI_L + Triplet_L | GAT | 99.04 | ± 0.26 | 92.78 | ± 1.00 | 97.21 | ± 0.34 | 92.46 | ± 0.68 | 25.125 |
| CrossE_L + Triplet_L | GAT | 98.92 | ± 0.13 | 93.16 | ± 0.78 | 97.14 | ± 0.37 | 93.08 | ± 1.41 | <span style="background-color:#6a6ae0">17.0</span> |
| PMI_L | GAT | 99.02 | ± 0.14 | 92.80 | ± 0.91 | 97.12 | ± 0.72 | 93.11 | ± 0.83 | <span style="background-color:#7ee07e">19.4</span> |
| Triplet_L | GAT | 99.11 | ± 0.22 | 92.34 | ± 0.60 | 97.49 | ± 0.67 | 92.76 | ± 0.86 | 24.375 |

*1.2.3   Embedding–Adjacency Alignment.*

Table 33. Cosine-Adj Corr Performance (↑): This table presents the top 10 models ranked by their average performance in terms of cosine-adj corr. For the complete list of models, refer to Supplementary Information 2 Table 33. Top-ranked results are highlighted in <span style="color:red">red</span>, second-ranked in <span style="color:blue">blue</span>, and third-ranked in <span style="color:green">green</span>.

| Loss Type | Model | Cora ↓ Citeseer | | Cora ↓ Bitcoin | | Citeseer ↓ Cora | | Citeseer ↓ Bitcoin | | Average Rank |
|---|---|---|---|---|---|---|---|---|---|---|
| Contr_l + CrossE_L + PMI_L | GAT | 10.86 0.42 | ± | 4.68 0.24 | ± | 10.32 0.77 | ± | 4.76 0.30 | ± | 8.5 |
| Contr_l + CrossE_L + PMI_L + Triplet_L | GAT | 10.40 0.46 | ± | 4.75 0.25 | ± | 10.17 0.62 | ± | 4.61 0.19 | ± | 10.375 |
| Contr_l + PMI_L | GAT | 10.64 1.14 | ± | 4.72 0.33 | ± | 10.67 0.52 | ± | 4.68 0.31 | ± | 8.0 |
| Contr_l + PMI_L | GCN | 9.42 1.08 | ± | 4.63 0.35 | ± | 9.20 0.82 | ± | 4.77 0.45 | ± | 17.125 |
| CrossE_L + PMI_L | GAT | 10.67 0.70 | ± | 4.72 0.10 | ± | 10.42 0.79 | ± | 4.80 0.17 | ± | 6.2 |
| CrossE_L + PMI_L + Triplet_L | GAT | 10.41 0.63 | ± | 4.56 0.39 | ± | 10.29 0.53 | ± | 4.80 0.14 | ± | 11.875 |
| CrossE_L + Triplet_L | GAT | 10.17 0.50 | ± | 4.52 0.19 | ± | 10.11 0.57 | ± | 4.59 0.27 | ± | 16.75 |
| PMI_L | GAT | 10.59 0.68 | ± | 4.83 0.31 | ± | 10.12 0.38 | ± | 4.89 0.27 | ± | 4.8 |
| PMI_L + Triplet_L | GAT | 10.48 0.62 | ± | 4.66 0.23 | ± | 10.84 0.37 | ± | 4.78 0.42 | ± | 9.0 |
| Triplet_L | GCN | 9.53 0.41 | ± | 4.87 0.49 | ± | 9.03 1.02 | ± | 4.80 0.45 | ± | 13.0 |

Table 34. Dot-Adj Corr Performance (↑): This table presents the top 10 models ranked by their average performance in terms of dot-adj corr. For the complete list of models, refer to Supplementary Information 2 Table 34. Top-ranked results are highlighted in <span style="color:red">red</span>, second-ranked in <span style="color:blue">blue</span>, and third-ranked in <span style="color:green">green</span>.

| Loss Type | Model | Cora ↓ Citeseer | | Cora ↓ Bitcoin | | Citeseer ↓ Cora | | Citeseer ↓ Bitcoin | | Average Rank |
|---|---|---|---|---|---|---|---|---|---|---|
| Contr_l + CrossE_L + PMI_L | GAT | 10.86 0.42 | ± | 4.68 0.24 | ± | 10.32 0.77 | ± | 4.76 0.30 | ± | 8.5 |
| Contr_l + CrossE_L + PMI_L + Triplet_L | GAT | 10.40 0.46 | ± | 4.75 0.25 | ± | 10.17 0.62 | ± | 4.61 0.19 | ± | 10.375 |
| Contr_l + PMI_L | GAT | 10.64 1.14 | ± | 4.72 0.33 | ± | 10.67 0.52 | ± | 4.68 0.31 | ± | 8.0 |
| Contr_l + PMI_L | GCN | 9.42 1.08 | ± | 4.63 0.35 | ± | 9.20 0.82 | ± | 4.77 0.45 | ± | 17.125 |
| CrossE_L + PMI_L | GAT | 10.67 0.70 | ± | 4.72 0.10 | ± | 10.42 0.79 | ± | 4.80 0.17 | ± | 6.2 |
| CrossE_L + PMI_L + Triplet_L | GAT | 10.41 0.63 | ± | 4.56 0.39 | ± | 10.29 0.53 | ± | 4.80 0.14 | ± | 11.875 |
| CrossE_L + Triplet_L | GAT | 10.17 0.51 | ± | 4.52 0.19 | ± | 10.11 0.57 | ± | 4.59 0.27 | ± | 16.75 |
| PMI_L | GAT | 10.59 0.68 | ± | 4.83 0.31 | ± | 10.12 0.38 | ± | 4.89 0.27 | ± | 4.8 |
| PMI_L + Triplet_L | GAT | 10.48 0.62 | ± | 4.66 0.23 | ± | 10.84 0.37 | ± | 4.78 0.42 | ± | 9.0 |
| Triplet_L | GCN | 9.53 0.41 | ± | 4.87 0.49 | ± | 9.03 1.02 | ± | 4.80 0.45 | ± | 13.0 |

Table 35. Euclidean-Adj Corr Performance (↑): This table presents the top 10 models ranked by their average performance in terms of euclidean-adj corr. For the complete list of models, refer to Supplementary Information 2 Table 35. Top-ranked results are highlighted in red, second-ranked in blue, and third-ranked in green.

| Loss Type | Model | Cora ↓ Citeseer | | Cora ↓ Bitcoin | | Citeseer ↓ Cora | | Citeseer ↓ Bitcoin | | Average Rank |
|---|---|---|---|---|---|---|---|---|---|---|
| Contr_l + CrossE_L + PMI_L | GAT | 16.20 ± 0.56 | | 6.21 ± 0.31 | | 14.66 ± 1.02 | | 6.29 ± 0.35 | | 11.6 |
| Contr_l + CrossE_L + PMI_L + Triplet_L | GAT | 15.56 ± 0.63 | | 6.30 ± 0.31 | | 14.44 ± 0.80 | | 6.10 ± 0.25 | | 14.75 |
| Contr_l + PMI_L | GAT | 15.91 ± 1.60 | | 6.23 ± 0.43 | | 15.13 ± 0.65 | | 6.19 ± 0.41 | | 11.875 |
| Contr_l + PMI_L | GCN | 14.22 ± 1.56 | | 6.39 ± 0.49 | | 13.23 ± 1.12 | | 6.59 ± 0.62 | | 15.5 |
| CrossE_L + PMI_L | GAT | 15.94 ± 0.98 | | 6.29 ± 0.13 | | 14.84 ± 1.03 | | 6.38 ± 0.22 | | 8.4 |
| CrossE_L + PMI_L + Triplet_L | GAT | 15.57 ± 0.93 | | 6.06 ± 0.52 | | 14.60 ± 0.72 | | 6.33 ± 0.20 | | 15.375 |
| CrossE_L + PMI_L + Triplet_L | GCN | 13.61 ± 0.98 | | 6.88 ± 0.30 | | 13.42 ± 0.69 | | 6.26 ± 0.41 | | 17.25 |
| PMI_L | GAT | 15.81 ± 0.96 | | 6.40 ± 0.41 | | 14.41 ± 0.53 | | 6.45 ± 0.37 | | 9.1 |
| PMI_L + Triplet_L | GAT | 15.67 ± 0.91 | | 6.17 ± 0.31 | | 15.35 ± 0.47 | | 6.32 ± 0.52 | | 12.5 |
| Triplet_L | GCN | 14.48 ± 0.57 | | 6.73 ± 0.67 | | 13.12 ± 1.36 | | 6.65 ± 0.60 | | 11.875 |

Table 36. Graph Reconstruction Bce Loss Performance (↓): This table presents the top 10 models ranked by their average performance in terms of graph reconstruction bce loss. For the complete list of models, refer to Supplementary Information 2 Table 36. Top-ranked results are highlighted in <span style="color:red">red</span>, second-ranked in <span style="color:blue">blue</span>, and third-ranked in <span style="color:green">green</span>.

| Loss Type | Model | Cora ↓ Citeseer | | Cora ↓ Bitcoin | | Citeseer ↓ Cora | | Citeseer ↓ Bitcoin | | Average Rank |
|---|---|---|---|---|---|---|---|---|---|---|
| Contr_l + CrossE_L | GCN | 60.50 0.53 | ± | 61.52 0.73 | ± | 63.62 1.69 | ± | 61.90 0.89 | ± | 12.0 |
| Contr_l + CrossE_L + PMI_L | GCN | 59.78 0.90 | ± | 61.62 0.89 | ± | 64.17 1.71 | ± | 61.39 0.43 | ± | 10.25 |
| Contr_l + CrossE_L + PMI_L + PR_L + Triplet_L | GCN | 59.79 1.13 | ± | 61.94 0.80 | ± | 64.33 1.78 | ± | 60.84 0.58 | ± | 10.0 |
| Contr_l + CrossE_L + PMI_L + Triplet_L | GCN | 60.29 0.78 | ± | 62.03 0.74 | ± | 63.96 1.28 | ± | 60.90 0.73 | ± | 9.25 |
| Contr_l + PMI_L | GCN | 59.05 1.79 | ± | 61.69 0.88 | ± | 63.06 1.36 | ± | 61.28 1.09 | ± | 5.2 |
| CrossE_L + PMI_L | GCN | 60.75 1.53 | ± | 63.51 1.18 | ± | 63.08 0.92 | ± | 60.79 0.53 | ± | 14.0 |
| CrossE_L + PMI_L + Triplet_L | GCN | 60.34 1.28 | ± | 61.06 0.42 | ± | 63.32 1.37 | ± | 61.68 0.70 | ± | 8.1 |
| PMI_L | GCN | 60.83 1.92 | ± | 61.21 1.44 | ± | 63.57 1.08 | ± | 60.40 1.17 | ± | 8.0 |
| PMI_L + Triplet_L | GCN | 61.37 1.54 | ± | 62.03 0.93 | ± | 62.58 1.01 | ± | 60.90 0.55 | ± | 14.125 |
| Triplet_L | GCN | 59.68 1.03 | ± | 61.66 1.38 | ± | 64.57 1.84 | ± | 61.54 1.37 | ± | 12.125 |

*1.2.4 Clustering Quality Metrics .* Clustering Quality Metrics evaluate how well the node embeddings form distinct and meaningful groups. Silhouette measures how similar a node is to its own cluster compared to others, Calinski-Harabasz assesses the separation and compactness of clusters, and kNN Consistency checks whether nearest neighbors in embedding space belong to the same ground-truth class, indicating local embedding fidelity.

Table 37. Silhouette Performance (↑): This table presents the top 10 models ranked by their average performance in terms of silhouette. For the complete list of models, refer to Supplementary Information 2 Table 37. Top-ranked results are highlighted in red, second-ranked in blue, and third-ranked in green.

| Loss Type | Model | Cora ↓ Citeseer | | Cora ↓ Bitcoin | | Citeseer ↓ Cora | | Citeseer ↓ Bitcoin | | Average Rank |
|---|---|---|---|---|---|---|---|---|---|---|
| Contr_l + CrossE_L + PMI_L | GAT | -0.24 0.48 | ± | 1.52 0.51 | ± | -0.22 0.59 | ± | 0.92 0.51 | ± | 20.875 |
| Contr_l + PMI_L | MPNN | 0.46 0.65 | ± | 0.98 1.39 | ± | 2.66 1.23 | ± | 0.54 0.83 | ± | 16.4 |
| Contr_l + Triplet_L | GAT | -0.51 0.42 | ± | 1.43 1.12 | ± | 0.01 0.82 | ± | 0.98 0.59 | ± | 23.125 |
| CrossE_L + PMI_L + Triplet_L | GAT | -0.30 0.56 | ± | 1.08 0.64 | ± | 0.14 0.64 | ± | 0.98 0.62 | ± | 20.875 |
| CrossE_L + PMI_L + Triplet_L | MPNN | 0.18 0.77 | ± | 1.02 0.95 | ± | 2.71 1.70 | ± | 0.57 0.80 | ± | 16.8 |
| CrossE_L + PR_L + Triplet_L | MPNN | -0.25 0.94 | ± | 1.04 0.79 | ± | 2.13 2.40 | ± | 0.99 0.77 | ± | 17.125 |
| PMI_L | MPNN | 0.52 0.75 | ± | 0.76 1.18 | ± | 2.25 1.37 | ± | 0.79 1.67 | ± | 19.875 |
| PMI_L + Triplet_L | GAT | -0.42 0.55 | ± | 1.12 0.52 | ± | 0.13 0.59 | ± | 1.26 0.38 | ± | 20.875 |
| PMI_L + Triplet_L | MPNN | 0.19 0.39 | ± | 0.92 0.40 | ± | 2.15 1.60 | ± | 0.41 1.77 | ± | 22.625 |
| Triplet_L | GAT | -0.13 0.45 | ± | 1.25 0.60 | ± | -0.08 0.93 | ± | 1.67 0.50 | ± | 16.6 |

Table 38. Calinski Harabasz Performance (↑): This table presents the top 10 models ranked by their average performance in terms of calinski harabasz. For the complete list of models, refer to Supplementary Information 2 Table 38. Top-ranked results are highlighted in <span style="color:red">red</span>, second-ranked in <span style="color:blue">blue</span>, and third-ranked in <span style="color:green">green</span>.

| Loss Type | Model | Cora ↓ Citeseer | Cora ↓ Bitcoin | Citeseer ↓ Cora | Citeseer ↓ Bitcoin | Average Rank |
|---|---|---|---|---|---|---|
| Contr_l + CrossE_L + PR_L | MPNN | 8354.59 ± 2933.25 | 1743.45 ± 327.10 | 11140.48 ± 1106.41 | 1785.37 ± 459.99 | 25.0 |
| Contr_l + CrossE_L + PR_L + Triplet_L | MPNN | 6596.77 ± 315.95 | 1688.64 ± 421.56 | 9058.54 ± 1830.16 | 1963.42 ± 756.79 | 35.5 |
| Contr_l + PR_L | MPNN | 8845.30 ± 579.92 | 1747.83 ± 447.73 | 11783.43 ± 5279.23 | 1891.84 ± 502.29 | 22.0 |
| Contr_l + PR_L + Triplet_L | MPNN | 6328.80 ± 1113.48 | 1972.21 ± 255.52 | 8944.50 ± 2344.77 | 1843.75 ± 621.59 | 31.0 |
| CrossE_L + PMI_L + PR_L | MPNN | 4963.28 ± 523.85 | 1915.22 ± 298.64 | 7469.40 ± 2140.05 | 2202.81 ± 577.69 | 39.75 |
| CrossE_L + PR_L | MPNN | 7991.67 ± 2938.89 | 2157.59 ± 618.77 | 12646.01 ± 2990.87 | 1991.88 ± 437.89 | <span style="background-color:#F08080">14.2</span> |
| CrossE_L + PR_L + Triplet_L | MPNN | 6627.76 ± 772.50 | 1988.58 ± 483.42 | 11845.68 ± 2634.10 | 2388.24 ± 438.70 | 21.0 |
| PMI_L | MPNN | 4936.01 ± 952.40 | 1900.87 ± 323.88 | 8619.00 ± 1000.02 | 2011.77 ± 611.40 | 37.5 |
| PR_L | MPNN | 6900.91 ± 1684.91 | 2053.16 ± 368.24 | 11675.30 ± 2947.07 | 2055.69 ± 230.16 | <span style="background-color:#77DD77">20.0</span> |
| PR_L + Triplet_L | MPNN | 7166.81 ± 2287.42 | 1963.38 ± 606.25 | 12943.92 ± 3270.82 | 2213.17 ± 757.37 | <span style="background-color:#6B6BE8">19.0</span> |

Table 39. Knn Consistency Performance (↑): This table presents the top 10 models ranked by their average performance in terms of knn consistency. For the complete list of models, refer to Supplementary Information 2 Table 39. Top-ranked results are highlighted in <span style="color:red">red</span>, second-ranked in <span style="color:blue">blue</span>, and third-ranked in <span style="color:green">green</span>.

| Loss Type | Model | Cora ↓ Citeseer | | Cora ↓ Bitcoin | | Citeseer ↓ Cora | | Citeseer ↓ Bitcoin | | Average Rank |
|---|---|---|---|---|---|---|---|---|---|---|
| Contr_l + CrossE_L + PMI_L | GAT | 70.02 | ± 0.41 | 75.04 | ± 0.09 | 83.83 | ± 0.24 | 75.00 | ± 0.26 | 57.875 |
| Contr_l + CrossE_L + PMI_L + PR_L | GAT | 69.90 | ± 0.46 | 75.05 | ± 0.34 | 83.61 | ± 0.57 | 75.08 | ± 0.17 | 59.25 |
| CrossE_L + PMI_L + PR_L + Triplet_L | MPNN | 69.70 | ± 0.13 | 75.50 | ± 0.18 | 80.69 | ± 0.58 | 75.58 | ± 0.23 | 58.625 |
| CrossE_L + PMI_L + Triplet_L | MPNN | 69.57 | ± 0.18 | 75.44 | ± 0.19 | 82.13 | ± 0.15 | 75.50 | ± 0.33 | 53.8 |
| CrossE_L + Triplet_L | MPNN | 69.72 | ± 0.27 | 75.12 | ± 0.21 | 82.09 | ± 0.28 | 75.74 | ± 0.12 | 53.6 |
| CrossE_L + Triplet_L | SAGE | 67.94 | ± 0.48 | 75.90 | ± 0.17 | 80.42 | ± 0.62 | 76.00 | ± 0.25 | 58.75 |
| PMI_L | MPNN | 69.66 | ± 0.08 | 75.55 | ± 0.21 | 81.74 | ± 0.31 | 75.28 | ± 0.34 | 59.25 |
| Triplet_L | GAT | 70.38 | ± 0.24 | 74.87 | ± 0.14 | 83.84 | ± 0.25 | 75.14 | ± 0.21 | 57.0 |
| Triplet_L | GIN | 66.92 | ± 0.44 | 76.02 | ± 0.28 | 79.81 | ± 0.66 | 76.38 | ± 0.42 | 58.375 |
| Triplet_L | SAGE | 67.40 | ± 0.31 | 76.19 | ± 0.43 | 79.95 | ± 0.58 | 76.32 | ± 0.49 | 55.8 |

*1.2.5 Semantic Coherence and Ranking:* Semantic Coherence measures how well nodes that are close in the embedding space share meaningful attributes or labels, indicating the embedding captures real-world semantics. Ranking metrics like RankMe or SelfCluster Performance evaluate how well the embeddings preserve hierarchy or structure, e.g., how well similar nodes are ranked or clustered together without supervision.

Table 40. Coherence Performance (↑): This table presents the top 10 models ranked by their average performance in terms of coherence. For the complete list of models, refer to Supplementary Information 2 Table 40. Top-ranked results are highlighted in red, second-ranked in blue, and third-ranked in green.

| Loss Type | Model | Cora ↓ Citeseer | | Cora ↓ Bitcoin | | Citeseer ↓ Cora | | Citeseer ↓ Bitcoin | | Average Rank |
|---|---|---|---|---|---|---|---|---|---|---|
| Contr_l + CrossE_L + PR_L | GIN | 100.00 0.00 | ± | 98.95 1.45 | ± | 97.64 3.89 | ± | 100.00 0.00 | ± | 10.125 |
| Contr_l + PR_L | GIN | 100.00 0.00 | ± | 100.00 0.00 | ± | 97.03 4.50 | ± | 100.00 0.00 | ± | 7.1 |
| CrossE_L | GIN | 100.00 0.00 | ± | 100.00 0.00 | ± | 100.00 0.00 | ± | 100.00 0.00 | ± | 4.5 |
| CrossE_L + PMI_L | GIN | 76.97 26.45 | ± | 100.00 0.00 | ± | 92.50 14.50 | ± | 100.00 0.01 | ± | 14.0 |
| CrossE_L + PMI_L + PR_L | GIN | 90.48 19.59 | ± | 100.00 0.00 | ± | 100.00 0.00 | ± | 100.00 0.00 | ± | 6.4 |
| CrossE_L + PMI_L + PR_L + Triplet_L | GIN | 83.87 16.35 | ± | 100.00 0.00 | ± | 93.38 8.84 | ± | 99.95 0.11 | ± | 13.5 |
| CrossE_L + PR_L | GIN | 99.19 1.81 | ± | 99.87 0.30 | ± | 100.00 0.00 | ± | 100.00 0.00 | ± | 8.375 |
| PMI_L | GIN | 87.74 24.09 | ± | 99.79 0.48 | ± | 97.05 4.69 | ± | 100.00 0.00 | ± | 12.75 |
| PR_L | GIN | 99.65 0.76 | ± | 99.99 0.02 | ± | 99.77 0.53 | ± | 99.99 0.01 | ± | 9.375 |
| PR_L + Triplet_L | GIN | 100.00 0.00 | ± | 99.57 0.96 | ± | 97.55 5.49 | ± | 99.40 1.35 | ± | 13.375 |

Table 41. Selfcluster Performance (↑): This table presents the top 10 models ranked by their average performance in terms of selfcluster. For the complete list of models, refer to Supplementary Information 2 Table 41. Top-ranked results are highlighted in red, second-ranked in blue, and third-ranked in green.

| Loss Type | Model | Cora ↓ Citeseer | | Cora ↓ Bitcoin | | Citeseer ↓ Cora | | Citeseer ↓ Bitcoin | | Average Rank |
|---|---|---|---|---|---|---|---|---|---|---|
| Contr_l + CrossE_L + PR_L | ALL | -0.79 0.00 | ± | -0.79 0.00 | ± | -0.79 0.00 | ± | -0.79 0.00 | ± | 46.8 |
| Contr_l + CrossE_L + PR_L | MPNN | -0.79 0.00 | ± | -0.79 0.00 | ± | -0.79 0.00 | ± | -0.79 0.00 | ± | 46.75 |
| Contr_l + CrossE_L + PR_L | PAGNN | -0.79 0.00 | ± | -0.79 0.00 | ± | -0.79 0.00 | ± | -0.79 0.00 | ± | 46.75 |
| Contr_l + CrossE_L + PR_L | SAGE | -0.79 0.01 | ± | -0.79 0.00 | ± | -0.79 0.00 | ± | -0.79 0.00 | ± | 46.75 |
| Contr_l + PR_L | PAGNN | -0.79 0.00 | ± | -0.79 0.00 | ± | -0.79 0.00 | ± | -0.79 0.00 | ± | 46.75 |
| CrossE_L + PMI_L + PR_L | ALL | -0.79 0.00 | ± | -0.79 0.00 | ± | -0.79 0.00 | ± | -0.79 0.00 | ± | 46.75 |
| CrossE_L + PR_L + Triplet_L | GAT | -0.79 0.00 | ± | -0.79 0.00 | ± | -0.79 0.01 | ± | -0.79 0.00 | ± | 46.75 |
| PR_L | ALL | -0.79 0.00 | ± | -0.79 0.00 | ± | -0.79 0.00 | ± | -0.79 0.00 | ± | 46.75 |
| PR_L | GAT | -0.79 0.00 | ± | -0.79 0.00 | ± | -0.79 0.00 | ± | -0.79 0.00 | ± | 46.75 |
| PR_L | GCN | -0.79 0.00 | ± | -0.79 0.00 | ± | -0.79 0.00 | ± | -0.79 0.00 | ± | 46.75 |

Table 42. Rankme Performance (↑): This table presents the top 10 models ranked by their average performance in terms of rankme. For the complete list of models, refer torefer to Supplementary Information 2 Table 42.0. Top-ranked results are highlighted in red, second-ranked in blue, and third-ranked in green.

| Loss Type | Model | Cora ↓ Citeseer | Cora ↓ Bitcoin | Citeseer ↓ Cora | Citeseer ↓ Bitcoin | Average Rank |
|---|---|---|---|---|---|---|
| Contr_l + CrossE_L + PMI_L | GAT | 454.81 ± 1.30 | 440.97 ± 1.18 | 443.67 ± 2.59 | 442.78 ± 1.80 | 7.5 |
| Contr_l + CrossE_L + PMI_L + PR_L | SAGE | 451.80 ± 0.98 | 447.45 ± 1.65 | 439.60 ± 3.29 | 444.03 ± 2.14 | 8.5 |
| Contr_l + PMI_L | GAT | 453.86 ± 1.70 | 440.47 ± 1.76 | 445.12 ± 1.78 | 441.63 ± 0.88 | 8.5 |
| CrossE_L + PMI_L | GAT | 453.79 ± 1.32 | 441.52 ± 0.81 | 444.10 ± 2.24 | 442.22 ± 1.05 | 7.75 |
| CrossE_L + PMI_L | SAGE | 448.32 ± 2.20 | 445.05 ± 1.81 | 443.52 ± 4.62 | 448.33 ± 0.84 | 9.75 |
| CrossE_L + PMI_L + PR_L | SAGE | 452.50 ± 2.08 | 448.63 ± 2.04 | 443.14 ± 4.74 | 449.12 ± 1.08 | 6.1 |
| CrossE_L + PMI_L + Triplet_L | GAT | 454.02 ± 0.73 | 441.37 ± 2.70 | 444.31 ± 1.22 | 442.13 ± 1.07 | 7.2 |
| PMI_L | GAT | 453.97 ± 1.38 | 442.60 ± 1.32 | 444.74 ± 1.47 | 442.89 ± 1.32 | 5.8 |
| PMI_L + PR_L | SAGE | 450.06 ± 3.00 | 445.57 ± 0.85 | 443.14 ± 4.11 | 448.21 ± 0.71 | 9.125 |
| PMI_L + Triplet_L | GAT | 452.83 ± 0.80 | 440.77 ± 1.11 | 445.49 ± 1.23 | 442.39 ± 2.32 | 8.25 |

**Additional Notes**

Each table includes only filtered results for top 10 Average Rank. For full table see Supplementary Information 2. For overall analysis see main manuscript.

# Supplementary Information 2 for: *Evaluating Loss Functions for Graph Neural Networks: Towards Pretraining and Generalization*

KHUSHNOOD ABBAS*, School of Computer Science and Technology, Zhoukou Normal University, China

RUIZHE HOU, School of Automation Science and Engineering,South China University of Technology, China

ZHOU WENGANG, DONG SHI, NIU LING, School of Computer Science and Technology, Zhoukou Normal University, China

SATYAKI NAN, College of Business and Computing, Georgia Southwestern State University, USA

ALIREZA ABBASI, School of Engineering and IT, University of New South Wales (UNSW Canberra), Australia

## Overview

This document provides supplementary information accompanying the manuscript titled "Evaluating Loss Functions for Graph Neural Networks: Towards Pretraining and Generalization." An additional file, Supplementary Information 1, contains the tabular results for the top 10 average-ranked models. The table numbering is consistent across both files.

## 1 Full Experimental Results

We have considered two settings one for transductive and one for inductive for evaluating the model's capability.

### 1.1 Transductive settings

*1.1.1 Node Classification results.* The generated embedding were trained and tested for node classification tasks.

Table 1. Node Cls Accuracy Performance (↑): Top-ranked results are highlighted in **1st**, second-ranked in **2nd**, and third-ranked in **3rd**.

| Loss Type | Model | CORA | | Citeseer | | Bitcoin Fraud Transaction | | Average Rank |
|---|---|---|---|---|---|---|---|---|
| Contr_l | ALL | 71.44 ± 1.90 | | 60.63 ± 2.45 | | 71.54 ± 0.37 | | 116.7 |
| Contr_l | GAT | 81.66 ± 0.38 | | 68.23 ± 1.30 | | 74.26 ± 0.73 | | 16.7 |
| Contr_l | GCN | 77.60 ± 1.86 | | 62.73 ± 2.26 | | 72.54 ± 0.75 | | 68.3 |

Continued on next page

Authors' Contact Information: Khushnood Abbas, Khushnood.abbas@zknu.edu.cn, School of Computer Science and Technology, Zhoukou Normal University, Zhoukou, Henan, China; Ruizhe Hou, auruizhe@mail.scut.edu.cn, School of Automation Science and Engineering,South China University of Technology, Guangzhou, Henan, China; Zhou Wengang, Dong Shi, Niu Ling, School of Computer Science and Technology, Zhoukou Normal University, Zhoukou, Henan, China; Satyaki Nan, satyaki.nan@gsw.edu, College of Business and Computing, Georgia Southwestern State University, Americus, GA, USA; Alireza Abbasi, a.abbasi@unsw.edu.au, School of Engineering and IT, University of New South Wales (UNSW Canberra), Canberra, ACT, Australia.

Node Cls Accuracy Continued ($\uparrow$)

| Loss Type | Model | CORA | | Citeseer | | Bitcoin Fraud Transaction | | Average Rank |
|---|---|---|---|---|---|---|---|---|
| Contr_l | GIN | 78.49 | ± | 63.99 | ± | 73.16 | ± | 40.7 |
| | | 0.17 | | 1.75 | | 1.10 | | |
| Contr_l | MPNN | 79.22 | ± | 64.08 | ± | 74.14 | ± | 28.7 |
| | | 0.97 | | 1.12 | | 0.80 | | |
| Contr_l | PAGNN | 78.04 | ± | 63.09 | ± | 71.20 | ± | 81.3 |
| | | 1.25 | | 1.91 | | 0.00 | | |
| Contr_l | SAGE | 76.57 | ± | 61.68 | ± | 74.14 | ± | 62.0 |
| | | 2.00 | | 2.50 | | 0.59 | | |
| Contr_l + CrossE_L | ALL | 72.03 | ± | 58.17 | ± | 71.32 | ± | 122.0 |
| | | 2.10 | | 2.30 | | 0.38 | | |
| Contr_l + CrossE_L | GAT | 80.70 | ± | 67.30 | ± | 72.96 | ± | 30.7 |
| | | 1.44 | | 2.04 | | 0.78 | | |
| Contr_l + CrossE_L | GCN | 78.19 | ± | 61.95 | ± | 72.44 | ± | 70.7 |
| | | 1.28 | | 1.43 | | 0.38 | | |
| Contr_l + CrossE_L | GIN | 76.64 | ± | 63.18 | ± | 74.18 | ± | 51.0 |
| | | 1.61 | | 0.80 | | 0.28 | | |
| Contr_l + CrossE_L | MPNN | 78.82 | ± | 64.14 | ± | 73.64 | ± | 32.7 |
| | | 1.92 | | 1.27 | | 0.52 | | |
| Contr_l + CrossE_L | PAGNN | 76.72 | ± | 62.61 | ± | 71.20 | ± | 93.3 |
| | | 1.03 | | 1.10 | | 0.00 | | |
| Contr_l + CrossE_L | SAGE | 76.27 | ± | 62.52 | ± | 72.36 | ± | 79.0 |
| | | 1.48 | | 1.43 | | 1.18 | | |
| Contr_l + CrossE_L + PMI_L | ALL | 72.03 | ± | 57.81 | ± | 72.54 | ± | 106.3 |
| | | 1.84 | | 2.50 | | 0.83 | | |
| Contr_l + CrossE_L + PMI_L | GAT | 80.22 | ± | 63.72 | ± | 74.66 | ± | 24.3 |
| | | 2.12 | | 0.90 | | 1.33 | | |
| Contr_l + CrossE_L + PMI_L | GCN | 71.88 | ± | 57.36 | ± | 71.88 | ± | 120.3 |
| | | 1.86 | | 1.79 | | 0.50 | | |
| Contr_l + CrossE_L + PMI_L | GIN | 70.88 | ± | 52.13 | ± | 72.20 | ± | 139.7 |
| | | 0.55 | | 1.75 | | 0.25 | | |
| Contr_l + CrossE_L + PMI_L | MPNN | 78.15 | ± | 61.41 | ± | 74.48 | ± | 49.3 |
| | | 1.59 | | 1.43 | | 0.59 | | |

Node Cls Accuracy Continued (↑)

| Loss Type | Model | CORA | | Citeseer | | Bitcoin Fraud Transaction | | Average Rank |
|---|---|---|---|---|---|---|---|---|
| Contr_l + CrossE_L + PMI_L | PAGNN | 70.85 ± 1.23 | | 49.19 ± 1.63 | | 71.20 ± 0.00 | | 163.3 |
| Contr_l + CrossE_L + PMI_L | SAGE | 61.40 ± 4.33 | | 48.80 ± 2.96 | | 78.02 ± 1.11 | | 121.0 |
| Contr_l + CrossE_L + PMI_L + PR_L | ALL | 64.43 ± 3.16 | | 53.63 ± 3.80 | | 70.90 ± 0.69 | | 174.3 |
| Contr_l + CrossE_L + PMI_L + PR_L | GAT | 79.74 ± 2.00 | | 63.18 ± 0.63 | | 73.96 ± 0.97 | | 34.3 |
| Contr_l + CrossE_L + PMI_L + PR_L | GCN | 73.14 ± 2.10 | | 56.55 ± 1.48 | | 72.92 ± 1.02 | | 98.0 |
| Contr_l + CrossE_L + PMI_L + PR_L | GIN | 70.19 ± 2.99 | | 50.06 ± 1.59 | | 71.64 ± 0.92 | | 153.0 |
| Contr_l + CrossE_L + PMI_L + PR_L | MPNN | 77.90 ± 2.02 | | 62.97 ± 2.05 | | 72.22 ± 2.01 | | 69.0 |
| Contr_l + CrossE_L + PMI_L + PR_L | PAGNN | 72.10 ± 0.97 | | 51.92 ± 1.25 | | 71.20 ± 0.00 | | 149.0 |
| Contr_l + CrossE_L + PMI_L + PR_L | SAGE | 58.52 ± 1.85 | | 49.22 ± 3.83 | | 72.18 ± 1.02 | | 161.7 |
| Contr_l + CrossE_L + PMI_L + PR_L + Triplet_L | ALL | 72.84 ± 1.98 | | 56.97 ± 3.30 | | 72.62 ± 0.87 | | 104.0 |
| Contr_l + CrossE_L + PMI_L + PR_L + Triplet_L | GAT | 79.08 ± 0.79 | | 64.17 ± 0.90 | | 75.26 ± 0.67 | | 17.0 |
| Contr_l + CrossE_L + PMI_L + PR_L + Triplet_L | GCN | 72.44 ± 0.55 | | 57.87 ± 2.13 | | 72.90 ± 0.97 | | 95.7 |
| Contr_l + CrossE_L + PMI_L + PR_L + Triplet_L | GIN | 72.18 ± 1.38 | | 55.05 ± 2.63 | | 73.06 ± 0.73 | | 109.0 |
| Contr_l + CrossE_L + PMI_L + PR_L + Triplet_L | MPNN | 77.38 ± 0.90 | | 61.32 ± 1.10 | | 74.54 ± 1.22 | | 56.0 |
| Contr_l + CrossE_L + PMI_L + PR_L + Triplet_L | PAGNN | 71.92 ± 2.18 | | 52.94 ± 0.87 | | 71.20 ± 0.00 | | 148.7 |
| Contr_l + CrossE_L + PMI_L + PR_L + Triplet_L | SAGE | 67.12 ± 2.02 | | 56.49 ± 3.74 | | 75.62 ± 1.26 | | 96.0 |

Node Cls Accuracy Continued (↑)

| Loss Type | Model | CORA | | Citeseer | | Bitcoin Fraud Transaction | | Average Rank |
|---|---|---|---|---|---|---|---|---|
| Contr_l + CrossE_L + PMI_L + Triplet_L | ALL | 75.72 | ± | 62.34 | ± | 73.22 | ± | 66.7 |
| | | 2.32 | | 1.25 | | 0.76 | | |
| Contr_l + CrossE_L + PMI_L + Triplet_L | GAT | 78.04 | ± | 63.96 | ± | 74.72 | ± | 33.0 |
| | | 0.89 | | 0.55 | | 0.79 | | |
| Contr_l + CrossE_L + PMI_L + Triplet_L | GCN | 72.66 | ± | 56.82 | ± | 73.08 | ± | 98.3 |
| | | 2.74 | | 1.94 | | 0.45 | | |
| Contr_l + CrossE_L + PMI_L + Triplet_L | GIN | 71.59 | ± | 54.62 | ± | 73.46 | ± | 109.7 |
| | | 1.49 | | 1.18 | | 1.26 | | |
| Contr_l + CrossE_L + PMI_L + Triplet_L | MPNN | 77.49 | ± | 62.40 | ± | 75.66 | ± | 40.3 |
| | | 0.61 | | 1.34 | | 0.75 | | |
| Contr_l + CrossE_L + PMI_L + Triplet_L | PAGNN | 72.07 | ± | 52.22 | ± | 71.20 | ± | 148.7 |
| | | 1.15 | | 2.03 | | 0.00 | | |
| Contr_l + CrossE_L + PMI_L + Triplet_L | SAGE | 66.79 | ± | 54.62 | ± | 76.72 | ± | 101.7 |
| | | 2.18 | | 4.07 | | 0.45 | | |
| Contr_l + CrossE_L + PR_L | ALL | 46.72 | ± | 40.00 | ± | 71.42 | ± | 183.0 |
| | | 0.66 | | 1.75 | | 0.04 | | |
| Contr_l + CrossE_L + PR_L | GAT | 65.98 | ± | 58.68 | ± | 72.22 | ± | 122.0 |
| | | 11.80 | | 3.00 | | 0.77 | | |
| Contr_l + CrossE_L + PR_L | GCN | 68.60 | ± | 56.28 | ± | 72.16 | ± | 132.7 |
| | | 2.22 | | 1.06 | | 0.62 | | |
| Contr_l + CrossE_L + PR_L | GIN | 52.95 | ± | 51.56 | ± | 71.30 | ± | 171.7 |
| | | 5.38 | | 2.75 | | 0.10 | | |
| Contr_l + CrossE_L + PR_L | MPNN | 63.03 | ± | 56.52 | ± | 72.48 | ± | 132.0 |
| | | 4.09 | | 1.01 | | 0.98 | | |
| Contr_l + CrossE_L + PR_L | PAGNN | 57.83 | ± | 42.67 | ± | 71.20 | ± | 182.7 |
| | | 2.50 | | 1.68 | | 0.00 | | |
| Contr_l + CrossE_L + PR_L | SAGE | 65.94 | ± | 52.13 | ± | 72.80 | ± | 138.7 |
| | | 9.24 | | 5.01 | | 1.54 | | |
| Contr_l + CrossE_L + PR_L + Triplet_L | ALL | 67.82 | ± | 55.19 | ± | 72.44 | ± | 133.7 |
| | | 1.31 | | 0.79 | | 0.67 | | |
| Contr_l + CrossE_L + PR_L + Triplet_L | GAT | 79.30 | ± | 64.11 | ± | 75.38 | ± | 15.0 |
| | | 0.92 | | 2.01 | | 0.63 | | |

Node Cls Accuracy Continued (↑)

| Loss Type | Model | CORA | | Citeseer | | Bitcoin Fraud Transaction | | Average Rank |
|---|---|---|---|---|---|---|---|---|
| Contr_l + CrossE_L + PR_L + Triplet_L | GCN | 74.83 ± 1.55 | | 59.01 ± 0.51 | | 72.40 ± 1.02 | | 91.3 |
| Contr_l + CrossE_L + PR_L + Triplet_L | GIN | 74.69 ± 2.44 | | 56.37 ± 2.10 | | 72.74 ± 1.64 | | 98.3 |
| Contr_l + CrossE_L + PR_L + Triplet_L | MPNN | 76.09 ± 1.08 | | 62.67 ± 2.10 | | 73.10 ± 1.10 | | 64.7 |
| Contr_l + CrossE_L + PR_L + Triplet_L | PAGNN | 67.38 ± 2.23 | | 57.45 ± 4.11 | | 71.20 ± 0.00 | | 143.0 |
| Contr_l + CrossE_L + PR_L + Triplet_L | SAGE | 74.50 ± 2.22 | | 58.92 ± 1.75 | | 74.26 ± 1.99 | | 70.7 |
| Contr_l + CrossE_L + Triplet_L | ALL | 75.33 ± 1.55 | | 60.56 ± 2.28 | | 71.67 ± 0.52 | | 97.7 |
| Contr_l + CrossE_L + Triplet_L | GAT | **81.49 ± 1.23** | | **66.94 ± 1.30** | | 73.55 ± 0.78 | | 22.7 |
| Contr_l + CrossE_L + Triplet_L | GCN | 77.38 ± 1.37 | | 62.55 ± 2.16 | | 72.69 ± 0.77 | | 68.7 |
| Contr_l + CrossE_L + Triplet_L | GIN | 79.87 ± 1.15 | | 62.88 ± 2.05 | | 74.43 ± 0.82 | | 32.3 |
| Contr_l + CrossE_L + Triplet_L | MPNN | 79.72 ± 1.11 | | 64.66 ± 1.19 | | 73.60 ± 0.91 | | 27.3 |
| Contr_l + CrossE_L + Triplet_L | PAGNN | 77.27 ± 2.02 | | 63.83 ± 1.40 | | 71.20 ± 0.00 | | 86.0 |
| Contr_l + CrossE_L + Triplet_L | SAGE | 78.71 ± 1.80 | | 62.03 ± 1.36 | | 72.87 ± 0.80 | | 59.0 |
| Contr_l + PMI_L | ALL | 72.73 ± 1.33 | | 57.00 ± 2.52 | | 72.82 ± 1.25 | | 100.3 |
| Contr_l + PMI_L | GAT | 79.67 ± 0.91 | | 63.90 ± 0.87 | | 74.92 ± 1.28 | | 22.0 |
| Contr_l + PMI_L | GCN | 72.92 ± 1.32 | | 55.41 ± 2.07 | | 72.08 ± 0.45 | | 119.7 |
| Contr_l + PMI_L | GIN | 71.18 ± 1.96 | | 53.94 ± 3.01 | | 72.32 ± 0.73 | | 132.0 |

Node Cls Accuracy Continued (↑)

| Loss Type | Model | CORA | | Citeseer | | Bitcoin Fraud Transaction | | Average Rank |
|---|---|---|---|---|---|---|---|---|
| Contr_l + PMI_L | MPNN | 78.71 | ± | 62.16 | ± | 75.14 | ± | 36.7 |
| | | 1.27 | | 1.02 | | 0.75 | | |
| Contr_l + PMI_L | PAGNN | 71.15 | ± | 49.61 | ± | 71.20 | ± | 162.3 |
| | | 1.21 | | 1.69 | | 0.00 | | |
| Contr_l + PMI_L | SAGE | 61.25 | ± | 53.15 | ± | 78.06 | ± | 109.7 |
| | | 4.64 | | 2.54 | | 0.81 | | |
| Contr_l + PMI_L + PR_L | ALL | 67.45 | ± | 55.22 | ± | 71.16 | ± | 164.3 |
| | | 0.93 | | 2.15 | | 0.83 | | |
| Contr_l + PMI_L + PR_L | GAT | 79.30 | ± | 63.33 | ± | 72.64 | ± | 50.3 |
| | | 1.95 | | 1.13 | | 1.25 | | |
| Contr_l + PMI_L + PR_L | GCN | 73.14 | ± | 56.76 | ± | 73.10 | ± | 94.7 |
| | | 0.45 | | 1.12 | | 0.58 | | |
| Contr_l + PMI_L + PR_L | GIN | 71.07 | ± | 51.65 | ± | 71.64 | ± | 147.3 |
| | | 1.62 | | 3.98 | | 0.89 | | |
| Contr_l + PMI_L + PR_L | MPNN | 78.49 | ± | 60.99 | ± | 71.58 | ± | 83.7 |
| | | 1.56 | | 1.45 | | 1.39 | | |
| Contr_l + PMI_L + PR_L | PAGNN | 71.37 | ± | 51.20 | ± | 71.20 | ± | 159.0 |
| | | 2.89 | | 2.12 | | 0.00 | | |
| Contr_l + PMI_L + PR_L | SAGE | 58.89 | ± | 52.70 | ± | 71.62 | ± | 158.7 |
| | | 2.36 | | 3.17 | | 0.27 | | |
| Contr_l + PMI_L + PR_L + Triplet_L | ALL | 72.07 | ± | 57.51 | ± | 71.70 | ± | 117.7 |
| | | 1.34 | | 1.49 | | 0.56 | | |
| Contr_l + PMI_L + PR_L + Triplet_L | GAT | 79.41 | ± | 63.27 | ± | 75.30 | ± | 22.0 |
| | | 1.29 | | 1.44 | | 0.33 | | |
| Contr_l + PMI_L + PR_L + Triplet_L | GCN | 72.99 | ± | 59.73 | ± | 72.78 | ± | 89.3 |
| | | 1.80 | | 2.76 | | 0.79 | | |
| Contr_l + PMI_L + PR_L + Triplet_L | GIN | 74.02 | ± | 57.06 | ± | 72.74 | ± | 94.7 |
| | | 1.81 | | 1.90 | | 1.06 | | |
| Contr_l + PMI_L + PR_L + Triplet_L | MPNN | 77.27 | ± | 61.95 | ± | 73.26 | ± | 65.0 |
| | | 0.80 | | 2.24 | | 0.83 | | |
| Contr_l + PMI_L + PR_L + Triplet_L | PAGNN | 73.21 | ± | 56.34 | ± | 71.20 | ± | 129.7 |
| | | 1.44 | | 1.50 | | 0.00 | | |

Node Cls Accuracy Continued (↑)

| Loss Type | Model | CORA | | | Citeseer | | | Bitcoin Fraud Transaction | | | Average Rank |
|---|---|---|---|---|---|---|---|---|---|---|---|
| Contr_l + PMI_L + PR_L + Triplet_L | SAGE | 73.40 | ± | 3.37 | 56.85 | ± | 1.61 | 75.68 | ± | 1.72 | 69.0 |
| Contr_l + PR_L | ALL | 46.87 | ± | 1.31 | 40.42 | ± | 3.40 | 71.34 | ± | 0.13 | 184.3 |
| Contr_l + PR_L | GAT | 59.78 | ± | 10.99 | 57.75 | ± | 1.12 | 72.88 | ± | 0.86 | 119.0 |
| Contr_l + PR_L | GCN | 67.64 | ± | 1.96 | 55.28 | ± | 2.71 | 72.68 | ± | 0.85 | 128.3 |
| Contr_l + PR_L | GIN | 55.43 | ± | 2.21 | 49.97 | ± | 4.00 | 71.22 | ± | 0.08 | 174.7 |
| Contr_l + PR_L | MPNN | 57.12 | ± | 1.66 | 55.71 | ± | 2.22 | 71.16 | ± | 0.63 | 172.0 |
| Contr_l + PR_L | PAGNN | 56.83 | ± | 2.66 | 44.23 | ± | 2.17 | 71.20 | ± | 0.00 | 184.7 |
| Contr_l + PR_L | SAGE | 64.84 | ± | 8.67 | 54.20 | ± | 3.98 | 72.02 | ± | 0.92 | 146.3 |
| Contr_l + PR_L + Triplet_L | ALL | 64.65 | ± | 2.10 | 55.07 | ± | 2.38 | 72.02 | ± | 0.51 | 144.0 |
| Contr_l + PR_L + Triplet_L | GAT | 79.78 | ± | 2.70 | 64.08 | ± | 1.37 | 75.38 | ± | 0.63 | 13.7 |
| Contr_l + PR_L + Triplet_L | GCN | 72.77 | ± | 3.01 | 58.41 | ± | 2.23 | 72.36 | ± | 0.43 | 102.0 |
| Contr_l + PR_L + Triplet_L | GIN | 73.25 | ± | 2.11 | 55.35 | ± | 2.26 | 73.00 | ± | 1.81 | 101.0 |
| Contr_l + PR_L + Triplet_L | MPNN | 75.13 | ± | 1.41 | 60.81 | ± | 1.34 | 73.62 | ± | 0.99 | 70.3 |
| Contr_l + PR_L + Triplet_L | PAGNN | 68.52 | ± | 1.73 | 56.64 | ± | 6.74 | 71.20 | ± | 0.00 | 148.3 |
| Contr_l + PR_L + Triplet_L | SAGE | 74.02 | ± | 1.17 | 58.44 | ± | 1.31 | 74.38 | ± | 1.39 | 72.0 |
| Contr_l + Triplet_L | ALL | 76.97 | ± | 1.53 | 62.43 | ± | 1.31 | 71.42 | ± | 0.26 | 89.0 |

Node Cls Accuracy Continued (↑)

| Loss Type | Model | CORA | | Citeseer | | Bitcoin Fraud Transaction | | Average Rank |
|---|---|---|---|---|---|---|---|---|
| Contr_l + Triplet_L | GAT | 80.89 | ± 0.91 | 66.52 | ± 1.33 | 73.54 | ± 0.93 | 23.7 |
| Contr_l + Triplet_L | GCN | 78.12 | ± 1.49 | 63.93 | ± 1.96 | 73.12 | ± 0.55 | 44.7 |
| Contr_l + Triplet_L | GIN | 78.42 | ± 0.98 | 63.33 | ± 1.94 | 74.20 | ± 0.48 | 38.3 |
| Contr_l + Triplet_L | MPNN | 78.23 | ± 1.32 | 63.72 | ± 1.30 | 74.08 | ± 1.57 | 38.3 |
| Contr_l + Triplet_L | PAGNN | 77.79 | ± 1.90 | 64.17 | ± 0.94 | 71.20 | ± 0.00 | 81.3 |
| Contr_l + Triplet_L | SAGE | 78.20 | ± 1.58 | 63.30 | ± 1.43 | 73.66 | ± 1.31 | 42.3 |
| CrossE_L | ALL | 35.24 | ± 0.00 | 21.47 | ± 0.00 | 71.20 | ± 0.00 | 196.3 |
| CrossE_L | GAT | 35.24 | ± 0.00 | 21.47 | ± 0.00 | 71.20 | ± 0.00 | 197.3 |
| CrossE_L | GCN | 35.24 | ± 0.00 | 21.47 | ± 0.00 | 71.20 | ± 0.00 | 198.3 |
| CrossE_L | GIN | 35.24 | ± 0.00 | 21.47 | ± 0.00 | 71.20 | ± 0.00 | 199.3 |
| CrossE_L | MPNN | 35.24 | ± 0.00 | 21.47 | ± 0.00 | 71.20 | ± 0.00 | 200.3 |
| CrossE_L | PAGNN | 35.24 | ± 0.00 | 21.47 | ± 0.00 | 71.20 | ± 0.00 | 201.3 |
| CrossE_L | SAGE | 35.24 | ± 0.00 | 21.47 | ± 0.00 | 71.20 | ± 0.00 | 202.3 |
| CrossE_L + PMI_L | ALL | 73.18 | ± 1.48 | 57.72 | ± 2.36 | 73.38 | ± 0.74 | 85.3 |
| CrossE_L + PMI_L | GAT | 78.71 | ± 2.18 | 64.26 | ± 0.56 | 74.78 | ± 0.86 | 25.3 |
| CrossE_L + PMI_L | GCN | 72.03 | ± 0.83 | 57.72 | ± 2.07 | 73.08 | ± 0.33 | 97.7 |

Continued on next page

Node Cls Accuracy Continued (↑)

| Loss Type | Model | CORA | | Citeseer | | Bitcoin Fraud Transaction | | Average Rank |
|-----------|-------|------|---|----------|---|----------|---|--------------|
| CrossE_L + PMI_L | GIN | 71.40 | ± | 52.64 | ± | 72.16 | ± | 137.3 |
| | | 1.88 | | 2.35 | | 0.76 | | |
| CrossE_L + PMI_L | MPNN | 77.97 | ± | 63.39 | ± | 75.94 | ± | 27.7 |
| | | 1.18 | | 0.65 | | 0.78 | | |
| CrossE_L + PMI_L | PAGNN | 72.43 | ± | 51.23 | ± | 71.20 | ± | 155.3 |
| | | 1.72 | | 3.12 | | 0.00 | | |
| CrossE_L + PMI_L | SAGE | 59.08 | ± | 47.48 | ± | 78.10 | ± | 122.3 |
| | | 3.11 | | 1.30 | | 1.80 | | |
| CrossE_L + PMI_L + PR_L | ALL | 67.05 | ± | 53.42 | ± | 70.92 | ± | 172.7 |
| | | 1.35 | | 1.40 | | 0.26 | | |
| CrossE_L + PMI_L + PR_L | GAT | 77.82 | ± | 62.46 | ± | 73.66 | ± | 53.0 |
| | | 0.77 | | 1.21 | | 2.01 | | |
| CrossE_L + PMI_L + PR_L | GCN | 72.40 | ± | 56.22 | ± | 72.56 | ± | 113.7 |
| | | 2.09 | | 1.42 | | 0.52 | | |
| CrossE_L + PMI_L + PR_L | GIN | 70.52 | ± | 49.10 | ± | 71.38 | ± | 160.0 |
| | | 0.81 | | 3.90 | | 0.20 | | |
| CrossE_L + PMI_L + PR_L | MPNN | 77.60 | ± | 61.83 | ± | 73.40 | ± | 62.0 |
| | | 1.38 | | 1.70 | | 2.21 | | |
| CrossE_L + PMI_L + PR_L | PAGNN | 72.14 | ± | 50.54 | ± | 71.20 | ± | 158.3 |
| | | 2.19 | | 1.89 | | 0.00 | | |
| CrossE_L + PMI_L + PR_L | SAGE | 61.51 | ± | 49.70 | ± | 71.90 | ± | 161.0 |
| | | 2.65 | | 1.55 | | 0.71 | | |
| CrossE_L + PMI_L + PR_L + Triplet_L | ALL | 71.66 | ± | 57.39 | ± | 72.48 | ± | 112.3 |
| | | 0.45 | | 0.91 | | 0.90 | | |
| CrossE_L + PMI_L + PR_L + Triplet_L | GAT | 79.04 | ± | 63.54 | ± | 75.00 | ± | 25.7 |
| | | 0.53 | | 0.31 | | 0.66 | | |
| CrossE_L + PMI_L + PR_L + Triplet_L | GCN | 71.07 | ± | 56.85 | ± | 72.50 | ± | 118.3 |
| | | 1.00 | | 2.06 | | 0.82 | | |
| CrossE_L + PMI_L + PR_L + Triplet_L | GIN | 72.18 | ± | 56.10 | ± | 73.02 | ± | 106.3 |
| | | 1.40 | | 2.12 | | 0.63 | | |
| CrossE_L + PMI_L + PR_L + Triplet_L | MPNN | 77.97 | ± | 62.73 | ± | 75.00 | ± | 39.0 |
| | | 1.00 | | 1.13 | | 1.32 | | |

Node Cls Accuracy Continued (↑)

| Loss Type | Model | CORA | | | Citeseer | | | Bitcoin Fraud Transaction | | | Average Rank |
|---|---|---|---|---|---|---|---|---|---|---|---|
| CrossE_L + PMI_L + PR_L + Triplet_L | PAGNN | 72.95 | ± | 0.95 | 54.29 | ± | 2.48 | 71.20 | ± | 0.00 | 143.3 |
| CrossE_L + PMI_L + PR_L + Triplet_L | SAGE | 68.93 | ± | 2.13 | 56.28 | ± | 1.92 | 75.68 | ± | 1.38 | 93.7 |
| CrossE_L + PMI_L + Triplet_L | ALL | 76.20 | ± | 1.90 | 62.01 | ± | 0.96 | 73.72 | ± | 0.67 | 62.7 |
| CrossE_L + PMI_L + Triplet_L | GAT | 79.23 | ± | 0.28 | 63.87 | ± | 0.78 | 74.78 | ± | 0.77 | 26.3 |
| CrossE_L + PMI_L + Triplet_L | GCN | 71.62 | ± | 0.68 | 57.12 | ± | 1.48 | 72.64 | ± | 0.69 | 110.0 |
| CrossE_L + PMI_L + Triplet_L | GIN | 72.51 | ± | 1.17 | 56.25 | ± | 1.93 | 73.40 | ± | 0.58 | 99.7 |
| CrossE_L + PMI_L + Triplet_L | MPNN | 78.86 | ± | 1.76 | 62.64 | ± | 1.83 | 75.16 | ± | 1.24 | 31.3 |
| CrossE_L + PMI_L + Triplet_L | PAGNN | 72.73 | ± | 0.78 | 52.28 | ± | 2.59 | 71.20 | ± | 0.00 | 151.0 |
| CrossE_L + PMI_L + Triplet_L | SAGE | 72.29 | ± | 2.03 | 55.56 | ± | 2.21 | 76.24 | ± | 1.51 | 82.3 |
| CrossE_L + PR_L | ALL | 44.76 | ± | 0.50 | 37.78 | ± | 2.66 | 71.40 | ± | 0.00 | 185.7 |
| CrossE_L + PR_L | GAT | 48.27 | ± | 3.55 | 49.34 | ± | 1.62 | 71.20 | ± | 0.00 | 189.3 |
| CrossE_L + PR_L | GCN | 57.23 | ± | 4.70 | 53.48 | ± | 2.45 | 71.28 | ± | 0.08 | 163.7 |
| CrossE_L + PR_L | GIN | 46.79 | ± | 1.56 | 45.26 | ± | 2.56 | 71.28 | ± | 0.08 | 183.7 |
| CrossE_L + PR_L | MPNN | 55.06 | ± | 11.23 | 53.54 | ± | 0.79 | 71.06 | ± | 0.30 | 181.7 |
| CrossE_L + PR_L | PAGNN | 51.92 | ± | 6.21 | 43.30 | ± | 3.02 | 71.20 | ± | 0.00 | 193.0 |
| CrossE_L + PR_L | SAGE | 49.04 | ± | 3.10 | 45.04 | ± | 4.00 | 71.66 | ± | 0.39 | 175.0 |

Node Cls Accuracy Continued (↑)

| Loss Type | | | | Model | CORA | | Citeseer | | Bitcoin Fraud Transaction | | Average Rank |
|---|---|---|---|---|---|---|---|---|---|---|---|
| CrossE_L | + | PR_L | + Triplet_L | ALL | 64.06 ± 5.88 | | 49.25 ± 3.75 | | 72.16 ± 0.79 | | 159.0 |
| CrossE_L | + | PR_L | + Triplet_L | GAT | 75.65 ± 2.78 | | 62.97 ± 1.85 | | 74.46 ± 0.88 | | 53.3 |
| CrossE_L | + | PR_L | + Triplet_L | GCN | 71.44 ± 2.90 | | 57.66 ± 0.95 | | 72.60 ± 1.06 | | 109.7 |
| CrossE_L | + | PR_L | + Triplet_L | GIN | 69.96 ± 3.31 | | 54.29 ± 2.25 | | 71.22 ± 0.28 | | 151.0 |
| CrossE_L | + | PR_L | + Triplet_L | MPNN | 70.44 ± 2.53 | | 58.35 ± 1.67 | | 72.62 ± 0.94 | | 111.3 |
| CrossE_L | + | PR_L | + Triplet_L | PAGNN | 63.28 ± 3.43 | | 51.62 ± 2.07 | | 71.20 ± 0.00 | | 175.7 |
| CrossE_L | + | PR_L | + Triplet_L | SAGE | 74.32 ± 1.54 | | 59.01 ± 3.35 | | 75.22 ± 1.68 | | 61.7 |
| CrossE_L + Triplet_L | | | | ALL | 74.83 ± 1.39 | | 61.98 ± 1.14 | | 72.98 ± 1.24 | | 74.0 |
| CrossE_L + Triplet_L | | | | GAT | 80.89 ± 0.67 | | 65.20 ± 0.83 | | 74.72 ± 1.00 | | 16.7 |
| CrossE_L + Triplet_L | | | | GCN | 76.72 ± 2.05 | | 61.32 ± 0.93 | | 72.68 ± 0.64 | | 78.3 |
| CrossE_L + Triplet_L | | | | GIN | 77.86 ± 2.25 | | 60.21 ± 1.73 | | 74.88 ± 0.55 | | 52.3 |
| CrossE_L + Triplet_L | | | | MPNN | 79.89 ± 0.95 | | 64.35 ± 1.55 | | 75.00 ± 0.74 | | 15.0 |
| CrossE_L + Triplet_L | | | | PAGNN | 78.01 ± 0.73 | | 64.86 ± 1.13 | | 71.20 ± 0.00 | | 81.7 |
| CrossE_L + Triplet_L | | | | SAGE | 78.04 ± 1.10 | | 62.91 ± 2.26 | | 75.16 ± 0.73 | | 35.7 |
| PMI_L | | | | ALL | 73.69 ± 1.99 | | 58.62 ± 2.15 | | 74.64 ± 0.62 | | 71.0 |
| PMI_L | | | | GAT | 79.41 ± 1.10 | | 63.99 ± 1.50 | | 74.82 ± 0.94 | | 23.0 |

Node Cls Accuracy Continued (↑)

| Loss Type | Model | CORA | | Citeseer | | Bitcoin Fraud Transaction | | Average Rank |
|---|---|---|---|---|---|---|---|---|
| PMI_L | GCN | 72.88 | ± | 57.06 | ± | 72.20 | ± | 109.7 |
| | | 1.49 | | 2.16 | | 0.78 | | |
| PMI_L | GIN | 70.07 | ± | 53.24 | ± | 71.50 | ± | 148.3 |
| | | 2.04 | | 2.33 | | 0.32 | | |
| PMI_L | MPNN | 77.64 | ± | 62.34 | ± | 74.84 | ± | 47.0 |
| | | 0.89 | | 1.39 | | 0.81 | | |
| PMI_L | PAGNN | 71.11 | ± | 49.85 | ± | 71.20 | ± | 168.3 |
| | | 1.81 | | 1.94 | | 0.00 | | |
| PMI_L | SAGE | 56.61 | ± | 47.54 | ± | 78.84 | ± | 124.3 |
| | | 1.73 | | 1.94 | | 0.49 | | |
| PMI_L + PR_L | ALL | 67.27 | ± | 52.97 | ± | 71.10 | ± | 172.3 |
| | | 2.04 | | 4.75 | | 0.37 | | |
| PMI_L + PR_L | GAT | 78.82 | ± | 60.87 | ± | 71.60 | ± | 82.0 |
| | | 1.37 | | 2.73 | | 0.65 | | |
| PMI_L + PR_L | GCN | 73.14 | ± | 58.38 | ± | 72.92 | ± | 89.7 |
| | | 1.05 | | 1.98 | | 0.26 | | |
| PMI_L + PR_L | GIN | 69.82 | ± | 52.10 | ± | 71.22 | ± | 158.0 |
| | | 2.12 | | 2.84 | | 0.04 | | |
| PMI_L + PR_L | MPNN | 76.57 | ± | 60.75 | ± | 71.82 | ± | 94.0 |
| | | 1.05 | | 2.10 | | 1.91 | | |
| PMI_L + PR_L | PAGNN | 71.95 | ± | 49.85 | ± | 71.20 | ± | 164.7 |
| | | 1.11 | | 2.02 | | 0.00 | | |
| PMI_L + PR_L | SAGE | 60.29 | ± | 49.70 | ± | 71.68 | ± | 163.7 |
| | | 3.11 | | 2.07 | | 0.39 | | |
| PMI_L + PR_L + Triplet_L | ALL | 71.14 | ± | 56.91 | ± | 72.06 | ± | 123.7 |
| | | 1.53 | | 1.60 | | 1.16 | | |
| PMI_L + PR_L + Triplet_L | GAT | 78.97 | ± | 63.81 | ± | 75.26 | ± | 22.7 |
| | | 1.78 | | 0.93 | | 0.25 | | |
| PMI_L + PR_L + Triplet_L | GCN | 73.32 | ± | 57.48 | ± | 72.78 | ± | 94.3 |
| | | 0.56 | | 0.67 | | 0.85 | | |
| PMI_L + PR_L + Triplet_L | GIN | 70.22 | ± | 56.49 | ± | 72.36 | ± | 126.3 |
| | | 0.93 | | 1.69 | | 0.59 | | |

Node Cls Accuracy Continued (↑)

| Loss Type | Model | CORA | | | Citeseer | | | Bitcoin Fraud Transaction | | | Average Rank |
|---|---|---|---|---|---|---|---|---|---|---|---|
| PMI_L + PR_L + Triplet_L | MPNN | 77.97 | ± | 1.74 | 61.14 | ± | 1.52 | 73.78 | ± | 1.59 | 57.3 |
| PMI_L + PR_L + Triplet_L | PAGNN | 72.95 | ± | 1.55 | 54.44 | ± | 2.18 | 71.20 | ± | 0.00 | 146.0 |
| PMI_L + PR_L + Triplet_L | SAGE | 70.48 | ± | 4.57 | 56.34 | ± | 1.75 | 74.80 | ± | 1.49 | 99.0 |
| PMI_L + Triplet_L | ALL | 76.64 | ± | 1.45 | 61.32 | ± | 0.67 | 73.44 | ± | 1.05 | 68.0 |
| PMI_L + Triplet_L | GAT | 79.78 | ± | 1.97 | 64.05 | ± | 1.11 | 74.96 | ± | 0.67 | 19.0 |
| PMI_L + Triplet_L | GCN | 73.91 | ± | 1.48 | 56.01 | ± | 2.82 | 72.92 | ± | 0.89 | 100.0 |
| PMI_L + Triplet_L | GIN | 73.80 | ± | 2.38 | 56.85 | ± | 1.55 | 73.54 | ± | 1.02 | 87.3 |
| PMI_L + Triplet_L | MPNN | 77.82 | ± | 1.44 | 63.03 | ± | 1.36 | 74.86 | ± | 1.19 | 40.7 |
| PMI_L + Triplet_L | PAGNN | 70.85 | ± | 1.16 | 51.23 | ± | 1.93 | 71.20 | ± | 0.00 | 169.3 |
| PMI_L + Triplet_L | SAGE | 68.60 | ± | 2.20 | 58.32 | ± | 1.84 | 78.42 | ± | 0.74 | 80.7 |
| PR_L | ALL | 43.47 | ± | 0.21 | 34.32 | ± | 0.80 | 71.42 | ± | 0.04 | 186.0 |
| PR_L | GAT | 45.83 | ± | 0.65 | 49.64 | ± | 3.98 | 71.20 | ± | 0.00 | 193.7 |
| PR_L | GCN | 55.16 | ± | 3.22 | 52.28 | ± | 1.04 | 71.20 | ± | 0.00 | 182.3 |
| PR_L | GIN | 47.64 | ± | 2.99 | 45.44 | ± | 2.09 | 71.38 | ± | 0.08 | 180.3 |
| PR_L | MPNN | 57.53 | ± | 7.32 | 54.05 | ± | 0.75 | 71.94 | ± | 1.70 | 153.0 |
| PR_L | PAGNN | 50.70 | ± | 7.08 | 43.54 | ± | 1.16 | 71.20 | ± | 0.00 | 196.0 |

Continued on next page

Node Cls Accuracy Continued (↑)

| Loss Type | Model | CORA | | Citeseer | | Bitcoin Fraud Transaction | | Average Rank |
|-----------|-------|------|---|----------|---|---------------------------|---|--------------|
| PR_L | SAGE | 47.38 ± 2.21 | | 44.90 ± 1.26 | | 71.62 ± 0.64 | | 178.0 |
| PR_L + Triplet_L | ALL | 46.50 ± 1.01 | | 37.33 ± 0.53 | | 71.42 ± 0.04 | | 185.0 |
| PR_L + Triplet_L | GAT | 66.79 ± 11.86 | | 57.66 ± 1.64 | | 72.02 ± 0.70 | | 130.3 |
| PR_L + Triplet_L | GCN | 68.82 ± 1.79 | | 54.56 ± 2.74 | | 72.18 ± 0.59 | | 138.0 |
| PR_L + Triplet_L | GIN | 56.79 ± 2.04 | | 51.50 ± 1.42 | | 71.30 ± 0.19 | | 170.3 |
| PR_L + Triplet_L | MPNN | 62.69 ± 6.34 | | 54.71 ± 1.66 | | 72.50 ± 1.45 | | 139.0 |
| PR_L + Triplet_L | PAGNN | 56.57 ± 3.57 | | 43.06 ± 3.02 | | 71.20 ± 0.00 | | 195.0 |
| PR_L + Triplet_L | SAGE | 48.82 ± 3.46 | | 51.50 ± 5.77 | | 73.34 ± 0.86 | | 144.0 |
| Triplet_L | ALL | 77.09 ± 1.16 | | 62.10 ± 0.61 | | 72.72 ± 0.68 | | 72.3 |
| Triplet_L | GAT | 81.70 ± 0.86 | | 65.92 ± 0.51 | | 75.28 ± 0.72 | | 8.0 |
| Triplet_L | GCN | 77.16 ± 1.42 | | 61.68 ± 2.15 | | 72.98 ± 0.88 | | 70.3 |
| Triplet_L | GIN | 77.56 ± 0.56 | | 59.49 ± 0.75 | | 75.64 ± 0.97 | | 49.3 |
| Triplet_L | MPNN | 78.97 ± 0.75 | | 63.57 ± 1.66 | | 75.16 ± 1.02 | | 25.3 |
| Triplet_L | PAGNN | 77.86 ± 2.20 | | 63.45 ± 1.80 | | 71.20 ± 0.00 | | 94.3 |
| Triplet_L | SAGE | 78.34 ± 0.95 | | 62.07 ± 1.67 | | 75.30 ± 0.74 | | 36.3 |

Table 2. Node Cls F1 Performance (↑): Top-ranked results are highlighted in **1st**, second-ranked in **2nd**, and third-ranked in **3rd**.

| Loss Type | Model | CORA | | Citeseer | | Bitcoin Fraud Transaction | | Average Rank |
|---|---|---|---|---|---|---|---|---|
| Contr_l | ALL | 64.15 | ± | 51.90 | | 29.68 | ± | 124.0 |
| | | 2.15 | | 2.17 | | 1.71 | | |
| Contr_l | GAT | 79.31 | ± | 60.52 | ± | 39.10 | ± | 15.3 |
| | | 0.32 | | 1.31 | | 1.00 | | |
| Contr_l | GCN | 74.31 | ± | 55.68 | ± | 35.98 | ± | 57.7 |
| | | 2.66 | | 2.23 | | 1.68 | | |
| Contr_l | GIN | 75.41 | ± | 55.95 | ± | 37.44 | ± | 40.0 |
| | | 0.59 | | 1.16 | | 1.50 | | |
| Contr_l | MPNN | 76.56 | ± | 56.26 | ± | 38.83 | ± | 28.3 |
| | | 1.63 | | 1.26 | | 1.74 | | |
| Contr_l | PAGNN | 73.47 | ± | 54.36 | ± | 27.73 | ± | 95.3 |
| | | 3.20 | | 1.90 | | 0.00 | | |
| Contr_l | SAGE | 72.81 | ± | 53.83 | ± | 37.39 | ± | 63.7 |
| | | 1.97 | | 1.99 | | 1.20 | | |
| Contr_l + CrossE_L | ALL | 64.46 | ± | 49.38 | ± | 28.67 | ± | 133.0 |
| | | 2.00 | | 2.03 | | 1.65 | | |
| Contr_l + CrossE_L | GAT | 77.94 | ± | 59.38 | ± | 36.49 | ± | 26.0 |
| | | 1.48 | | 1.73 | | 1.90 | | |
| Contr_l + CrossE_L | GCN | 74.86 | ± | 54.21 | ± | 35.93 | ± | 62.7 |
| | | 1.22 | | 1.56 | | 1.20 | | |
| Contr_l + CrossE_L | GIN | 73.29 | ± | 55.04 | ± | 38.64 | ± | 49.7 |
| | | 1.38 | | 1.59 | | 0.96 | | |
| Contr_l + CrossE_L | MPNN | 75.92 | ± | 56.24 | ± | 37.12 | ± | 37.3 |
| | | 1.91 | | 0.97 | | 1.36 | | |
| Contr_l + CrossE_L | PAGNN | 72.01 | ± | 53.85 | ± | 27.73 | ± | 101.7 |
| | | 2.22 | | 0.95 | | 0.00 | | |
| Contr_l + CrossE_L | SAGE | 72.24 | ± | 54.57 | ± | 33.19 | ± | 81.0 |
| | | 2.18 | | 2.16 | | 3.24 | | |
| Contr_l + CrossE_L + PMI_L | ALL | 66.32 | ± | 49.39 | ± | 35.44 | ± | 106.7 |
| | | 4.09 | | 2.71 | | 2.20 | | |
| Contr_l + CrossE_L + PMI_L | GAT | 77.60 | ± | 56.54 | ± | 40.62 | ± | 16.0 |
| | | 2.42 | | 1.02 | | 1.30 | | |

Node Cls F1 Continued (↑)

| Loss Type | Model | CORA | | Citeseer | | Bitcoin Fraud Transaction | | Average Rank |
|---|---|---|---|---|---|---|---|---|
| Contr_l + CrossE_L + PMI_L | GCN | 67.42 2.32 | ± | 50.10 1.79 | ± | 34.57 0.67 | ± | 106.0 |
| Contr_l + CrossE_L + PMI_L | GIN | 66.28 1.03 | ± | 44.40 1.77 | ± | 33.68 1.27 | ± | 134.0 |
| Contr_l + CrossE_L + PMI_L | MPNN | 75.44 1.80 | ± | 54.11 1.88 | ± | 40.49 0.68 | ± | 38.7 |
| Contr_l + CrossE_L + PMI_L | PAGNN | 65.47 2.09 | ± | 41.56 1.62 | ± | 27.73 0.00 | ± | 163.3 |
| Contr_l + CrossE_L + PMI_L | SAGE | 54.07 6.14 | ± | 41.56 2.77 | ± | 43.61 2.10 | ± | 118.0 |
| Contr_l + CrossE_L + PMI_L + PR_L | ALL | 54.76 4.28 | ± | 45.31 3.10 | ± | 30.31 0.70 | ± | 151.3 |
| Contr_l + CrossE_L + PMI_L + PR_L | GAT | 77.45 2.37 | ± | 55.62 0.63 | ± | 37.21 2.75 | ± | 34.3 |
| Contr_l + CrossE_L + PMI_L + PR_L | GCN | 69.44 2.76 | ± | 49.03 1.42 | ± | 36.10 1.68 | ± | 93.3 |
| Contr_l + CrossE_L + PMI_L + PR_L | GIN | 65.95 3.22 | ± | 42.48 1.37 | ± | 29.81 2.73 | ± | 149.7 |
| Contr_l + CrossE_L + PMI_L + PR_L | MPNN | 75.11 2.21 | ± | 54.76 2.15 | ± | 32.55 5.10 | ± | 70.7 |
| Contr_l + CrossE_L + PMI_L + PR_L | PAGNN | 66.55 1.40 | ± | 44.25 1.41 | ± | 27.73 0.00 | ± | 152.3 |
| Contr_l + CrossE_L + PMI_L + PR_L | SAGE | 50.09 4.24 | ± | 42.00 3.23 | ± | 30.73 2.79 | ± | 164.3 |
| Contr_l + CrossE_L + PMI_L + PR_L + Triplet_L | ALL | 67.18 3.12 | ± | 48.71 3.28 | ± | 34.14 2.14 | ± | 114.7 |
| Contr_l + CrossE_L + PMI_L + PR_L + Triplet_L | GAT | 76.03 0.72 | ± | 56.78 0.88 | ± | 41.23 1.15 | ± | 17.3 |
| Contr_l + CrossE_L + PMI_L + PR_L + Triplet_L | GCN | 68.22 2.03 | ± | 50.64 2.67 | ± | 36.02 1.86 | ± | 89.7 |
| Contr_l + CrossE_L + PMI_L + PR_L + Triplet_L | GIN | 67.63 2.46 | ± | 47.12 2.77 | ± | 35.36 1.23 | ± | 114.0 |

Node Cls F1 Continued (↑)

| Loss Type | Model | CORA | | | Citeseer | | | Bitcoin Fraud Transaction | | | Average Rank |
|---|---|---|---|---|---|---|---|---|---|---|---|
| Contr_l + CrossE_L + PMI_L + PR_L + Triplet_L | MPNN | 74.79 | ± | 0.41 | 53.88 | ± | 1.96 | 39.65 | ± | 2.36 | 48.3 |
| Contr_l + CrossE_L + PMI_L + PR_L + Triplet_L | PAGNN | 66.36 | ± | 2.47 | 45.17 | ± | 0.93 | 27.73 | ± | 0.00 | 150.0 |
| Contr_l + CrossE_L + PMI_L + PR_L + Triplet_L | SAGE | 61.47 | ± | 2.60 | 48.10 | ± | 3.54 | 39.52 | ± | 1.86 | 105.0 |
| Contr_l + CrossE_L + PMI_L + Triplet_L | ALL | 71.08 | ± | 3.18 | 54.05 | ± | 0.81 | 36.55 | ± | 2.27 | 66.7 |
| Contr_l + CrossE_L + PMI_L + Triplet_L | GAT | 75.18 | ± | 1.11 | 56.85 | ± | 0.99 | 40.88 | ± | 0.89 | 24.7 |
| Contr_l + CrossE_L + PMI_L + Triplet_L | GCN | 69.53 | ± | 2.25 | 49.21 | ± | 1.05 | 36.16 | ± | 1.14 | 91.0 |
| Contr_l + CrossE_L + PMI_L + Triplet_L | GIN | 66.77 | ± | 1.68 | 46.83 | ± | 1.21 | 36.37 | ± | 3.42 | 110.7 |
| Contr_l + CrossE_L + PMI_L + Triplet_L | MPNN | 74.56 | ± | 0.88 | 53.95 | ± | 1.34 | 41.48 | ± | 0.80 | 40.0 |
| Contr_l + CrossE_L + PMI_L + Triplet_L | PAGNN | 66.49 | ± | 1.88 | 44.35 | ± | 2.19 | 27.73 | ± | 0.00 | 153.0 |
| Contr_l + CrossE_L + PMI_L + Triplet_L | SAGE | 60.95 | ± | 0.99 | 46.72 | ± | 4.44 | 42.33 | ± | 0.87 | 99.0 |
| Contr_l + CrossE_L + PR_L | ALL | 22.28 | ± | 2.00 | 29.71 | ± | 2.57 | 28.36 | ± | 0.12 | 186.0 |
| Contr_l + CrossE_L + PR_L | GAT | 55.22 | ± | 19.97 | 50.11 | ± | 3.75 | 30.53 | ± | 1.92 | 131.3 |
| Contr_l + CrossE_L + PR_L | GCN | 62.54 | ± | 1.79 | 48.08 | ± | 1.30 | 33.49 | ± | 1.40 | 132.0 |
| Contr_l + CrossE_L + PR_L | GIN | 33.84 | ± | 8.75 | 43.44 | ± | 3.27 | 28.15 | ± | 0.26 | 175.7 |
| Contr_l + CrossE_L + PR_L | MPNN | 50.99 | ± | 7.39 | 47.99 | ± | 0.85 | 33.91 | ± | 3.83 | 139.0 |
| Contr_l + CrossE_L + PR_L | PAGNN | 42.25 | ± | 2.93 | 32.23 | ± | 2.14 | 27.73 | ± | 0.00 | 184.3 |

Node Cls F1 Continued (↑)

| Loss Type | Model | CORA | | Citeseer | | Bitcoin Fraud Transaction | | Average Rank |
|---|---|---|---|---|---|---|---|---|
| Contr_l + CrossE_L + PR_L | SAGE | 56.13 14.32 | ± | 43.52 4.72 | ± | 32.32 4.12 | ± | 152.3 |
| Contr_l + CrossE_L + PR_L + Triplet_L | ALL | 58.56 2.89 | ± | 46.42 0.77 | ± | 34.24 2.55 | ± | 137.3 |
| Contr_l + CrossE_L + PR_L + Triplet_L | GAT | 76.20 1.29 | ± | 56.12 2.04 | ± | 40.96 1.32 | ± | 21.3 |
| Contr_l + CrossE_L + PR_L + Triplet_L | GCN | 70.62 2.04 | ± | 51.04 1.02 | ± | 35.51 2.39 | ± | 86.3 |
| Contr_l + CrossE_L + PR_L + Triplet_L | GIN | 70.34 3.37 | ± | 47.91 1.73 | ± | 35.25 3.79 | ± | 104.7 |
| Contr_l + CrossE_L + PR_L + Triplet_L | MPNN | 73.23 0.93 | ± | 54.20 1.74 | ± | 36.74 3.09 | ± | 62.3 |
| Contr_l + CrossE_L + PR_L + Triplet_L | PAGNN | 56.84 2.12 | ± | 49.31 3.59 | ± | 27.73 0.00 | ± | 146.7 |
| Contr_l + CrossE_L + PR_L + Triplet_L | SAGE | 70.82 2.21 | ± | 50.84 1.21 | ± | 36.09 4.34 | ± | 81.7 |
| Contr_l + CrossE_L + Triplet_L | ALL | 71.04 1.82 | ± | 51.90 2.23 | ± | 31.14 1.64 | ± | 97.3 |
| Contr_l + CrossE_L + Triplet_L | GAT | **79.17 1.07** | ± | 58.99 1.44 | ± | 37.28 1.46 | ± | 23.0 |
| Contr_l + CrossE_L + Triplet_L | GCN | 74.87 1.12 | ± | 54.75 2.00 | ± | 36.19 1.41 | ± | 56.7 |
| Contr_l + CrossE_L + Triplet_L | GIN | 76.90 1.49 | ± | 55.56 1.91 | ± | 39.46 1.15 | ± | 29.3 |
| Contr_l + CrossE_L + Triplet_L | MPNN | 77.11 1.27 | ± | 56.93 1.82 | ± | 37.54 1.80 | ± | 26.7 |
| Contr_l + CrossE_L + Triplet_L | PAGNN | 73.39 2.67 | ± | 54.76 1.35 | ± | 27.73 0.00 | ± | 95.7 |
| Contr_l + CrossE_L + Triplet_L | SAGE | 75.46 2.56 | ± | 53.78 1.20 | ± | 35.08 1.60 | ± | 67.3 |
| Contr_l + PMI_L | ALL | 65.52 3.31 | ± | 48.22 2.54 | ± | 35.09 2.36 | ± | 120.0 |

Node Cls F1 Continued (↑)

| Loss Type | Model | CORA | | Citeseer | | Bitcoin Fraud Transaction | | Average Rank |
|---|---|---|---|---|---|---|---|---|
| Contr_l + PMI_L | GAT | 77.79 | ± 1.27 | 56.79 | ± 1.06 | 41.12 | ± 1.26 | 11.3 |
| Contr_l + PMI_L | GCN | 69.06 | ± 1.86 | 48.08 | ± 1.99 | 34.56 | ± 0.69 | 112.3 |
| Contr_l + PMI_L | GIN | 65.86 | ± 2.27 | 45.95 | ± 2.53 | 33.99 | ± 1.26 | 129.3 |
| Contr_l + PMI_L | MPNN | 75.91 | ± 1.42 | 54.99 | ± 1.14 | 40.86 | ± 1.47 | 29.3 |
| Contr_l + PMI_L | PAGNN | 65.26 | ± 2.18 | 41.81 | ± 1.04 | 27.73 | ± 0.00 | 165.7 |
| Contr_l + PMI_L | SAGE | 53.05 | ± 4.60 | 46.10 | ± 2.32 | 43.53 | ± 1.15 | 106.3 |
| Contr_l + PMI_L + PR_L | ALL | 60.11 | ± 1.94 | 47.03 | ± 1.98 | 31.15 | ± 2.17 | 142.3 |
| Contr_l + PMI_L + PR_L | GAT | 77.06 | ± 1.79 | 55.36 | ± 1.44 | 33.77 | ± 3.47 | 55.0 |
| Contr_l + PMI_L + PR_L | GCN | 69.93 | ± 0.73 | 49.81 | ± 1.22 | 36.28 | ± 1.10 | 86.3 |
| Contr_l + PMI_L + PR_L | GIN | 65.86 | ± 2.76 | 43.84 | ± 3.98 | 29.93 | ± 3.21 | 146.0 |
| Contr_l + PMI_L + PR_L | MPNN | 75.51 | ± 1.71 | 53.11 | ± 1.86 | 31.79 | ± 3.61 | 77.0 |
| Contr_l + PMI_L + PR_L | PAGNN | 64.28 | ± 4.33 | 43.79 | ± 1.93 | 27.73 | ± 0.00 | 163.7 |
| Contr_l + PMI_L + PR_L | SAGE | 50.98 | ± 3.44 | 43.93 | ± 2.75 | 29.83 | ± 0.84 | 161.3 |
| Contr_l + PMI_L + PR_L + Triplet_L | ALL | 65.58 | ± 2.15 | 49.04 | ± 1.17 | 31.26 | ± 2.73 | 124.0 |
| Contr_l + PMI_L + PR_L + Triplet_L | GAT | 76.60 | ± 1.40 | 55.24 | ± 1.34 | 40.47 | ± 0.97 | 27.3 |
| Contr_l + PMI_L + PR_L + Triplet_L | GCN | 69.42 | ± 1.89 | 51.42 | ± 2.52 | 35.82 | ± 1.15 | 87.7 |

Node Cls F1 Continued (↑)

| Loss Type | Model | CORA | | Citeseer | | Bitcoin Fraud Transaction | | Average Rank |
|-----------|-------|------|---|----------|---|--------------------------|---|--------------|
| Contr_l + PMI_L + PR_L + Triplet_L | GIN | 69.16 ± 2.60 | | 49.12 ± 1.30 | | 35.86 ± 1.25 | | 97.7 |
| Contr_l + PMI_L + PR_L + Triplet_L | MPNN | 74.19 ± 0.75 | | 53.42 ± 1.78 | | 36.11 ± 2.97 | | 69.7 |
| Contr_l + PMI_L + PR_L + Triplet_L | PAGNN | 68.03 ± 2.45 | | 48.28 ± 1.65 | | 27.73 ± 0.00 | | 136.0 |
| Contr_l + PMI_L + PR_L + Triplet_L | SAGE | 68.16 ± 3.08 | | 48.64 ± 1.36 | | 39.76 ± 3.09 | | 84.3 |
| Contr_l + PR_L | ALL | 22.88 ± 3.57 | | 29.40 ± 4.31 | | 28.16 ± 0.32 | | 188.0 |
| Contr_l + PR_L | GAT | 44.85 ± 19.54 | | 48.87 ± 1.42 | | 32.50 ± 2.53 | | 139.0 |
| Contr_l + PR_L | GCN | 61.13 ± 3.53 | | 47.48 ± 2.43 | | 35.21 ± 2.44 | | 128.7 |
| Contr_l + PR_L | GIN | 39.60 ± 5.00 | | 41.85 ± 4.08 | | 28.17 ± 0.21 | | 177.3 |
| Contr_l + PR_L | MPNN | 41.86 ± 4.09 | | 47.22 ± 2.55 | | 30.07 ± 1.04 | | 153.3 |
| Contr_l + PR_L | PAGNN | 41.29 ± 4.16 | | 33.58 ± 2.19 | | 27.73 ± 0.00 | | 187.3 |
| Contr_l + PR_L | SAGE | 55.01 ± 15.28 | | 45.63 ± 3.84 | | 30.08 ± 2.48 | | 151.7 |
| Contr_l + PR_L + Triplet_L | ALL | 55.36 ± 3.98 | | 46.25 ± 2.62 | | 31.43 ± 1.84 | | 146.3 |
| Contr_l + PR_L + Triplet_L | GAT | 77.61 ± 2.82 | | 56.27 ± 1.84 | | 40.98 ± 0.55 | | 14.7 |
| Contr_l + PR_L + Triplet_L | GCN | 68.84 ± 3.72 | | 50.62 ± 1.09 | | 34.51 ± 0.61 | | 99.0 |
| Contr_l + PR_L + Triplet_L | GIN | 69.80 ± 1.64 | | 47.19 ± 2.23 | | 35.27 ± 4.12 | | 107.0 |
| Contr_l + PR_L + Triplet_L | MPNN | 70.82 ± 2.93 | | 52.66 ± 1.86 | | 36.80 ± 2.45 | | 73.0 |

Node Cls F1 Continued (↑)

| Loss Type | Model | CORA | | Citeseer | | Bitcoin Fraud Transaction | | Average Rank |
|---|---|---|---|---|---|---|---|---|
| Contr_l + PR_L + Triplet_L | PAGNN | 59.70 ± 3.25 | | 48.11 ± 6.95 | | 27.73 ± 0.00 | | 155.3 |
| Contr_l + PR_L + Triplet_L | SAGE | 70.14 ± 1.91 | | 50.34 ± 1.15 | | 37.61 ± 2.70 | | 76.3 |
| Contr_l + Triplet_L | ALL | 73.05 ± 1.23 | | 53.58 ± 1.13 | | 29.46 ± 1.27 | | 95.3 |
| Contr_l + Triplet_L | GAT | 77.85 ± 0.74 | | 59.10 ± 1.51 | | 37.60 ± 1.45 | | 21.7 |
| Contr_l + Triplet_L | GCN | 74.54 ± 1.84 | | 56.22 ± 1.92 | | 36.44 ± 0.65 | | 50.3 |
| Contr_l + Triplet_L | GIN | 75.30 ± 0.73 | | 55.67 ± 2.36 | | 39.69 ± 1.08 | | 33.7 |
| Contr_l + Triplet_L | MPNN | 75.48 ± 1.73 | | 55.60 ± 0.94 | | 38.00 ± 3.01 | | 37.3 |
| Contr_l + Triplet_L | PAGNN | 74.88 ± 2.47 | | 55.46 ± 1.05 | | 27.73 ± 0.00 | | 88.7 |
| Contr_l + Triplet_L | SAGE | 75.64 ± 1.94 | | 54.78 ± 1.17 | | 36.53 ± 2.35 | | 47.0 |
| CrossE_L | ALL | 7.44 ± 0.00 | | 5.89 ± 0.00 | | 27.73 ± 0.00 | | 198.3 |
| CrossE_L | GAT | 7.44 ± 0.00 | | 5.89 ± 0.00 | | 27.73 ± 0.00 | | 199.3 |
| CrossE_L | GCN | 7.44 ± 0.00 | | 5.89 ± 0.00 | | 27.73 ± 0.00 | | 200.3 |
| CrossE_L | GIN | 7.44 ± 0.00 | | 5.89 ± 0.00 | | 27.73 ± 0.00 | | 201.3 |
| CrossE_L | MPNN | 7.44 ± 0.00 | | 5.89 ± 0.00 | | 27.73 ± 0.00 | | 202.3 |
| CrossE_L | PAGNN | 7.44 ± 0.00 | | 5.89 ± 0.00 | | 27.73 ± 0.00 | | 203.3 |
| CrossE_L | SAGE | 7.44 ± 0.00 | | 5.89 ± 0.00 | | 27.73 ± 0.00 | | 204.3 |

Node Cls F1 Continued (↑)

| Loss Type | Model | CORA | | Citeseer | | Bitcoin Fraud Transaction | | Average Rank |
|---|---|---|---|---|---|---|---|---|
| CrossE_L + PMI_L | ALL | 69.31 | ± 2.17 | 49.18 | ± 2.28 | 38.47 | ± 0.76 | 82.3 |
| CrossE_L + PMI_L | GAT | 75.68 | ± 2.99 | 57.40 | ± 1.21 | 40.69 | ± 0.95 | 19.7 |
| CrossE_L + PMI_L | GCN | 69.33 | ± 1.39 | 50.46 | ± 2.13 | 36.57 | ± 0.45 | 82.7 |
| CrossE_L + PMI_L | GIN | 67.48 | ± 0.88 | 44.87 | ± 2.18 | 33.54 | ± 1.71 | 128.0 |
| CrossE_L + PMI_L | MPNN | 74.32 | ± 1.47 | 56.21 | ± 0.65 | 41.67 | ± 1.37 | 29.7 |
| CrossE_L + PMI_L | PAGNN | 67.77 | ± 2.57 | 43.70 | ± 3.09 | 27.73 | ± 0.00 | 155.7 |
| CrossE_L + PMI_L | SAGE | 51.42 | ± 3.93 | 40.18 | ± 1.45 | 43.91 | ± 3.10 | 120.7 |
| CrossE_L + PMI_L + PR_L | ALL | 60.21 | ± 2.90 | 45.27 | ± 1.81 | 29.62 | ± 0.57 | 152.0 |
| CrossE_L + PMI_L + PR_L | GAT | 75.25 | ± 1.17 | 54.69 | ± 2.30 | 35.72 | ± 6.13 | 58.3 |
| CrossE_L + PMI_L + PR_L | GCN | 69.38 | ± 1.84 | 48.46 | ± 1.80 | 35.89 | ± 1.26 | 100.0 |
| CrossE_L + PMI_L + PR_L | GIN | 65.63 | ± 1.02 | 40.97 | ± 3.86 | 28.67 | ± 1.20 | 157.0 |
| CrossE_L + PMI_L + PR_L | MPNN | 75.17 | ± 1.45 | 53.48 | ± 1.95 | 35.69 | ± 6.58 | 67.0 |
| CrossE_L + PMI_L + PR_L | PAGNN | 67.63 | ± 4.31 | 43.25 | ± 1.13 | 27.73 | ± 0.00 | 159.3 |
| CrossE_L + PMI_L + PR_L | SAGE | 54.16 | ± 3.08 | 42.22 | ± 1.80 | 30.29 | ± 2.52 | 161.7 |
| CrossE_L + PMI_L + PR_L + Triplet_L | ALL | 65.57 | ± 2.30 | 48.91 | ± 0.77 | 35.29 | ± 1.46 | 114.0 |
| CrossE_L + PMI_L + PR_L + Triplet_L | GAT | 76.66 | ± 1.23 | 55.75 | ± 1.01 | 40.40 | ± 0.96 | 24.3 |

Node Cls F1 Continued (↑)

| Loss Type | Model | CORA | | Citeseer | | Bitcoin Fraud Transaction | | Average Rank |
|---|---|---|---|---|---|---|---|---|
| CrossE_L + PMI_L + PR_L + Triplet_L | GCN | 67.48 ± 0.97 | | 49.50 ± 2.11 | | 35.31 ± 1.95 | | 103.3 |
| CrossE_L + PMI_L + PR_L + Triplet_L | GIN | 67.45 ± 2.20 | | 48.40 ± 2.24 | | 35.69 ± 1.84 | | 108.7 |
| CrossE_L + PMI_L + PR_L + Triplet_L | MPNN | 74.94 ± 1.81 | | 54.52 ± 1.66 | | 39.99 ± 3.13 | | 42.0 |
| CrossE_L + PMI_L + PR_L + Triplet_L | PAGNN | 68.07 ± 1.30 | | 46.39 ± 2.67 | | 27.73 ± 0.00 | | 148.0 |
| CrossE_L + PMI_L + PR_L + Triplet_L | SAGE | 64.26 ± 2.51 | | 48.28 ± 1.60 | | 39.27 ± 2.95 | | 102.7 |
| CrossE_L + PMI_L + Triplet_L | ALL | 72.47 ± 2.52 | | 53.83 ± 1.45 | | 37.70 ± 1.08 | | 61.7 |
| CrossE_L + PMI_L + Triplet_L | GAT | 76.87 ± 1.04 | | 56.91 ± 1.00 | | 40.95 ± 0.81 | | 15.7 |
| CrossE_L + PMI_L + Triplet_L | GCN | 68.21 ± 1.22 | | 50.31 ± 1.42 | | 35.92 ± 1.17 | | 93.7 |
| CrossE_L + PMI_L + Triplet_L | GIN | 67.12 ± 1.69 | | 48.23 ± 1.80 | | 37.40 ± 1.91 | | 100.0 |
| CrossE_L + PMI_L + Triplet_L | MPNN | 76.13 ± 2.58 | | 54.98 ± 2.62 | | 41.07 ± 1.54 | | 26.3 |
| CrossE_L + PMI_L + Triplet_L | PAGNN | 67.19 ± 1.00 | | 44.57 ± 2.04 | | 27.73 ± 0.00 | | 157.0 |
| CrossE_L + PMI_L + Triplet_L | SAGE | 68.20 ± 1.95 | | 47.89 ± 2.12 | | 41.45 ± 2.41 | | 80.7 |
| CrossE_L + PR_L | ALL | 18.98 ± 0.33 | | 26.53 ± 3.28 | | 28.31 ± 0.12 | | 188.7 |
| CrossE_L + PR_L | GAT | 23.47 ± 6.41 | | 37.98 ± 1.62 | | 27.73 ± 0.00 | | 194.3 |
| CrossE_L + PR_L | GCN | 41.94 ± 8.73 | | 44.71 ± 1.66 | | 27.97 ± 0.29 | | 168.3 |
| CrossE_L + PR_L | GIN | 22.87 ± 4.29 | | 35.42 ± 3.93 | | 28.06 ± 0.21 | | 186.3 |

Node Cls F1 Continued (↑)

| Loss Type | | | Model | CORA | | Citeseer | | Bitcoin Fraud Transaction | | Average Rank |
|---|---|---|---|---|---|---|---|---|---|---|
| CrossE_L + PR_L | | | MPNN | 36.90 ± 21.51 | | 44.65 ± 1.04 | | 29.14 ± 0.61 | | 166.0 |
| CrossE_L + PR_L | | | PAGNN | 32.31 ± 11.23 | | 32.04 ± 3.56 | | 27.73 ± 0.00 | | 195.0 |
| CrossE_L + PR_L | | | SAGE | 26.49 ± 4.87 | | 36.49 ± 4.58 | | 29.58 ± 1.81 | | 178.7 |
| CrossE_L + PR_L + Triplet_L | | | ALL | 51.75 ± 9.64 | | 40.27 ± 3.72 | | 31.60 ± 4.07 | | 162.7 |
| CrossE_L + PR_L + Triplet_L | | | GAT | 71.30 ± 4.15 | | 54.60 ± 2.33 | | 37.71 ± 2.19 | | 57.7 |
| CrossE_L + PR_L + Triplet_L | | | GCN | 67.72 ± 3.50 | | 49.39 ± 1.00 | | 34.89 ± 2.88 | | 105.0 |
| CrossE_L + PR_L + Triplet_L | | | GIN | 64.38 ± 4.18 | | 46.63 ± 2.35 | | 28.35 ± 0.52 | | 148.0 |
| CrossE_L + PR_L + Triplet_L | | | MPNN | 63.46 ± 2.95 | | 49.89 ± 1.50 | | 34.85 ± 2.11 | | 116.3 |
| CrossE_L + PR_L + Triplet_L | | | PAGNN | 51.56 ± 5.33 | | 43.41 ± 2.19 | | 27.73 ± 0.00 | | 180.7 |
| CrossE_L + PR_L + Triplet_L | | | SAGE | 70.59 ± 2.28 | | 50.37 ± 2.99 | | 37.94 ± 3.67 | | 72.7 |
| CrossE_L + Triplet_L | | | ALL | 71.83 ± 0.96 | | 53.49 ± 1.14 | | 35.66 ± 3.41 | | 78.0 |
| CrossE_L + Triplet_L | | | GAT | **79.23 ± 1.28** | | 57.76 ± 1.09 | | 40.27 ± 1.69 | | 13.0 |
| CrossE_L + Triplet_L | | | GCN | 74.33 ± 1.70 | | 52.97 ± 1.62 | | 36.44 ± 0.71 | | 67.7 |
| CrossE_L + Triplet_L | | | GIN | 75.25 ± 1.90 | | 52.47 ± 1.31 | | 40.57 ± 0.28 | | 46.7 |
| CrossE_L + Triplet_L | | | MPNN | 76.90 ± 1.27 | | 56.30 ± 0.89 | | 40.26 ± 1.22 | | 21.7 |
| CrossE_L + Triplet_L | | | PAGNN | 74.89 ± 1.07 | | 56.45 ± 1.77 | | 27.73 ± 0.00 | | 87.3 |

Node Cls F1 Continued (↑)

| Loss Type | Model | CORA | | Citeseer | | Bitcoin Fraud Transaction | | Average Rank |
|---|---|---|---|---|---|---|---|---|
| CrossE_L + Triplet_L | SAGE | 75.58 ± 0.96 | | 54.40 ± 2.19 | | 39.67 ± 1.12 | | 39.3 |
| PMI_L | ALL | 69.30 ± 2.15 | | 50.35 ± 1.93 | | 39.30 ± 1.15 | | 76.0 |
| PMI_L | GAT | 77.20 ± 1.98 | | 57.06 ± 2.63 | | 40.94 ± 0.89 | | 13.3 |
| PMI_L | GCN | 69.41 ± 2.18 | | 49.78 ± 2.24 | | 35.14 ± 1.23 | | 98.0 |
| PMI_L | GIN | 64.58 ± 2.82 | | 45.51 ± 2.05 | | 31.67 ± 1.23 | | 139.7 |
| PMI_L | MPNN | 74.55 ± 1.02 | | 54.35 ± 1.35 | | 40.30 ± 1.32 | | 46.0 |
| PMI_L | PAGNN | 64.86 ± 2.19 | | 42.48 ± 1.57 | | 27.73 ± 0.00 | | 172.0 |
| PMI_L | SAGE | 48.96 ± 1.96 | | 40.72 ± 1.71 | | 45.28 ± 1.04 | | 121.0 |
| PMI_L + PR_L | ALL | 60.90 ± 3.00 | | 44.61 ± 4.88 | | 29.11 ± 0.53 | | 155.3 |
| PMI_L + PR_L | GAT | 76.21 ± 1.74 | | 52.60 ± 3.25 | | 30.12 ± 1.38 | | 79.7 |
| PMI_L + PR_L | GCN | 69.69 ± 1.50 | | 51.27 ± 1.66 | | 36.42 ± 0.54 | | 81.0 |
| PMI_L + PR_L | GIN | 64.82 ± 3.29 | | 44.54 ± 3.19 | | 27.78 ± 0.11 | | 156.0 |
| PMI_L + PR_L | MPNN | 73.59 ± 1.35 | | 53.00 ± 1.39 | | 31.37 ± 5.15 | | 89.0 |
| PMI_L + PR_L | PAGNN | 67.27 ± 1.60 | | 42.32 ± 2.10 | | 27.73 ± 0.00 | | 165.3 |
| PMI_L + PR_L | SAGE | 53.52 ± 4.71 | | 42.63 ± 2.43 | | 29.68 ± 1.08 | | 164.0 |
| PMI_L + PR_L + Triplet_L | ALL | 64.92 ± 2.11 | | 48.22 ± 1.52 | | 32.97 ± 1.99 | | 128.7 |

Node Cls F1 Continued (↑)

| Loss Type | Model | CORA | | Citeseer | | Bitcoin Fraud Transaction | | Average Rank |
|---|---|---|---|---|---|---|---|---|
| PMI_L + PR_L + Triplet_L | GAT | 76.16 ± 2.54 | | 56.30 ± 1.11 | | 40.20 ± 0.39 | | 25.0 |
| PMI_L + PR_L + Triplet_L | GCN | 70.04 ± 0.54 | | 50.37 ± 0.56 | | 36.21 ± 1.19 | | 83.7 |
| PMI_L + PR_L + Triplet_L | GIN | 64.40 ± 2.72 | | 48.54 ± 1.64 | | 34.84 ± 1.89 | | 122.3 |
| PMI_L + PR_L + Triplet_L | MPNN | 74.79 ± 2.14 | | 53.02 ± 1.89 | | 37.57 ± 3.68 | | 59.0 |
| PMI_L + PR_L + Triplet_L | PAGNN | 67.67 ± 2.69 | | 46.59 ± 1.89 | | 27.73 ± 0.00 | | 151.0 |
| PMI_L + PR_L + Triplet_L | SAGE | 65.60 ± 4.50 | | 48.71 ± 1.36 | | 37.77 ± 3.90 | | 98.0 |
| PMI_L + Triplet_L | ALL | 73.14 ± 1.11 | | 52.72 ± 0.68 | | 37.79 ± 1.76 | | 63.7 |
| PMI_L + Triplet_L | GAT | 76.97 ± 2.22 | | 57.31 ± 1.37 | | 41.22 ± 0.77 | | `11.7` |
| PMI_L + Triplet_L | GCN | 70.87 ± 1.59 | | 48.96 ± 3.33 | | 36.37 ± 2.17 | | 87.3 |
| PMI_L + Triplet_L | GIN | 69.54 ± 3.74 | | 48.94 ± 1.25 | | 37.65 ± 1.57 | | 84.0 |
| PMI_L + Triplet_L | MPNN | 74.74 ± 1.59 | | 55.49 ± 1.32 | | 40.47 ± 1.29 | | 37.3 |
| PMI_L + Triplet_L | PAGNN | 65.15 ± 2.16 | | 43.51 ± 1.52 | | 27.73 ± 0.00 | | 170.0 |
| PMI_L + Triplet_L | SAGE | 63.43 ± 3.31 | | 49.89 ± 1.73 | | `44.72 ± 1.15` | | 81.7 |
| PR_L | ALL | 18.49 ± 0.41 | | 22.40 ± 0.89 | | 28.36 ± 0.12 | | 188.7 |
| PR_L | GAT | 19.72 ± 0.44 | | 39.36 ± 4.46 | | 27.73 ± 0.00 | | 198.3 |
| PR_L | GCN | 38.14 ± 6.55 | | 44.05 ± 0.82 | | 27.73 ± 0.00 | | 185.3 |

Node Cls F1 Continued (↑)

| Loss Type | Model | CORA | | Citeseer | | Bitcoin Fraud Transaction | | Average Rank |
|---|---|---|---|---|---|---|---|---|
| PR_L | GIN | 25.33 | ± 7.02 | 36.90 | ± 2.01 | 28.21 | ± 0.17 | 183.3 |
| PR_L | MPNN | 42.26 | ± 10.77 | 45.16 | ± 0.88 | 33.06 | ± 4.22 | 152.0 |
| PR_L | PAGNN | 30.66 | ± 11.28 | 31.92 | ± 1.11 | 27.73 | ± 0.00 | 198.7 |
| PR_L | SAGE | 25.42 | ± 4.21 | 35.22 | ± 1.35 | 29.28 | ± 1.49 | 180.3 |
| PR_L + Triplet_L | ALL | 22.32 | ± 2.58 | 25.34 | ± 0.80 | 28.36 | ± 0.12 | 187.3 |
| PR_L + Triplet_L | GAT | 56.23 | ± 20.18 | 48.63 | ± 1.87 | 30.17 | ± 2.44 | 139.0 |
| PR_L + Triplet_L | GCN | 62.08 | ± 3.21 | 46.89 | ± 2.33 | 33.21 | ± 2.09 | 136.3 |
| PR_L + Triplet_L | GIN | 41.94 | ± 3.23 | 43.52 | ± 1.90 | 28.24 | ± 0.21 | 171.3 |
| PR_L + Triplet_L | MPNN | 50.67 | ± 10.71 | 46.42 | ± 1.84 | 33.50 | ± 3.37 | 145.7 |
| PR_L + Triplet_L | PAGNN | 40.57 | ± 6.32 | 31.78 | ± 3.56 | 27.73 | ± 0.00 | 197.3 |
| PR_L + Triplet_L | SAGE | 27.79 | ± 5.76 | 42.67 | ± 5.35 | 33.45 | ± 1.61 | 162.3 |
| Triplet_L | ALL | 73.39 | ± 1.34 | 53.67 | ± 0.64 | 35.99 | ± 2.01 | 71.0 |
| Triplet_L | GAT | 79.11 | ± 1.25 | 58.32 | ± 1.37 | 41.44 | ± 1.13 | 6.3 |
| Triplet_L | GCN | 74.67 | ± 1.75 | 53.25 | ± 1.88 | 36.54 | ± 1.41 | 63.0 |
| Triplet_L | GIN | 74.60 | ± 1.19 | 51.50 | ± 1.24 | 41.44 | ± 0.86 | 48.0 |
| Triplet_L | MPNN | 76.20 | ± 1.11 | 55.65 | ± 1.58 | 41.00 | ± 1.31 | 22.7 |

<div align="center">Node Cls F1 Continued (↑)</div>

| Loss Type | Model | CORA | | Citeseer | | Bitcoin Fraud Transaction | | Average Rank |
|-----------|-------|------|---|----------|---|---------------------------|---|--------------|
| Triplet_L | PAGNN | 74.67 | ± | 54.86 | ± | 27.73 | ± | 101.0 |
|           |       | 2.75 |   | 2.10 |   | 0.00 |   |      |
| Triplet_L | SAGE  | 75.25 | ± | 54.46 | ± | 39.47 | ± | 43.0 |
|           |       | 1.86 |   | 2.02 |   | 0.92 |   |      |

Table 3.  Node Cls Precision Performance (↑): Top-ranked results are highlighted in **1st**, second-ranked in **2nd**, and third-ranked in **3rd**.

| Loss Type | Model | CORA | | Citeseer | | Bitcoin Fraud Transaction | | Average Rank |
|-----------|-------|------|---|----------|---|---------------------------|---|--------------|
| Contr_l | ALL | 66.82 | ± | 51.84 | ± | 43.24 | ± | 140.0 |
|         |     | 3.44 |   | 4.19 |   | 3.02 |   |       |
| Contr_l | GAT | 79.27 | ± | 64.64 | ± | 47.04 | ± | 22.7 |
|         |     | 0.31 |   | 3.99 |   | 1.41 |   |      |
| Contr_l | GCN | 75.65 | ± | 61.15 | ± | 44.13 | ± | 68.7 |
|         |     | 2.84 |   | 2.15 |   | 1.56 |   |      |
| Contr_l | GIN | 75.69 | ± | 58.51 | ± | 45.09 | ± | 65.7 |
|         |     | 0.31 |   | 2.11 |   | 1.99 |   |      |
| Contr_l | MPNN | 76.89 | ± | 63.23 | ± | 46.79 | ± | 34.0 |
|         |      | 1.87 |   | 5.83 |   | 1.02 |   |      |
| Contr_l | PAGNN | 75.24 | ± | 54.51 | ± | 23.73 | ± | 107.7 |
|         |       | 2.21 |   | 4.40 |   | 0.00 |   |       |
| Contr_l | SAGE | 73.49 | ± | 57.77 | ± | 48.70 | ± | 55.3 |
|         |      | 2.28 |   | 8.03 |   | 2.38 |   |      |
| Contr_l + CrossE_L | ALL | 66.97 | ± | 48.15 | ± | 36.28 | ± | 158.7 |
|                    |     | 4.50 |   | 1.65 |   | 14.49 |   |       |
| Contr_l + CrossE_L | GAT | 78.00 | ± | 64.33 | ± | 45.10 | ± | 40.0 |
|                    |     | 1.27 |   | 3.91 |   | 1.72 |   |      |
| Contr_l + CrossE_L | GCN | 75.85 | ± | 56.36 | ± | 43.85 | ± | 84.3 |
|                    |     | 1.33 |   | 3.28 |   | 0.66 |   |      |

Node Cls Precision Continued (↑)

| Loss Type | | | Model | CORA | | Citeseer | | Bitcoin Fraud Transaction | | Average Rank |
|---|---|---|---|---|---|---|---|---|---|---|
| Contr_l + CrossE_L | | | GIN | 73.79 ± 0.76 | | 57.43 ± 4.64 | | 47.35 ± 1.47 | | 62.3 |
| Contr_l + CrossE_L | | | MPNN | 76.13 ± 1.86 | | 60.30 ± 1.97 | | 46.83 ± 0.93 | | 44.0 |
| Contr_l + CrossE_L | | | PAGNN | 74.51 ± 1.54 | | 55.91 ± 7.27 | | 23.73 ± 0.00 | | 103.3 |
| Contr_l + CrossE_L | | | SAGE | 71.89 ± 2.22 | | 57.47 ± 5.23 | | 44.92 ± 4.78 | | 89.7 |
| Contr_l + CrossE_L + PMI_L | | | ALL | 73.66 ± 1.32 | | 51.74 ± 5.92 | | 44.22 ± 1.68 | | 111.7 |
| Contr_l + CrossE_L + PMI_L | | | GAT | **79.70 ± 1.71** | | 59.71 ± 1.90 | | 46.89 ± 2.17 | | 31.7 |
| Contr_l + CrossE_L + PMI_L | | | GCN | 70.90 ± 2.58 | | 53.88 ± 3.24 | | 42.73 ± 1.34 | | 124.0 |
| Contr_l + CrossE_L + PMI_L | | | GIN | 69.34 ± 1.44 | | 47.18 ± 8.40 | | 44.33 ± 0.70 | | 145.7 |
| Contr_l + CrossE_L + PMI_L | | | MPNN | 76.51 ± 2.74 | | 57.30 ± 3.09 | | 46.60 ± 1.08 | | 51.7 |
| Contr_l + CrossE_L + PMI_L | | | PAGNN | 72.47 ± 2.30 | | 43.30 ± 2.51 | | 23.73 ± 0.00 | | 152.7 |
| Contr_l + CrossE_L + PMI_L | | | SAGE | 58.55 ± 5.78 | | 41.59 ± 4.30 | | 53.10 ± 1.21 | | 127.0 |
| Contr_l + CrossE_L + PMI_L + PR_L | | | ALL | 66.64 ± 4.95 | | 46.28 ± 2.44 | | 39.19 ± 3.83 | | 163.7 |
| Contr_l + CrossE_L + PMI_L + PR_L | | | GAT | 79.61 ± 2.68 | | 57.94 ± 1.51 | | 47.88 ± 1.49 | | 30.7 |
| Contr_l + CrossE_L + PMI_L + PR_L | | | GCN | 71.67 ± 2.06 | | 54.28 ± 6.36 | | 45.15 ± 2.25 | | 102.0 |
| Contr_l + CrossE_L + PMI_L + PR_L | | | GIN | 69.71 ± 3.42 | | 47.65 ± 7.71 | | 37.09 ± 12.40 | | 156.0 |
| Contr_l + CrossE_L + PMI_L + PR_L | | | MPNN | 76.43 ± 2.30 | | 56.74 ± 6.09 | | 42.85 ± 3.76 | | 81.0 |

Node Cls Precision Continued (↑)

| Loss Type | Model | CORA | | Citeseer | | Bitcoin Fraud Transaction | | Average Rank |
|---|---|---|---|---|---|---|---|---|
| Contr_l + CrossE_L + PMI_L + PR_L | PAGNN | 73.21 2.92 | ± | 46.20 2.05 | ± | 23.73 0.00 | ± | 144.0 |
| Contr_l + CrossE_L + PMI_L + PR_L | SAGE | 55.92 3.76 | ± | 47.04 9.01 | ± | 45.39 13.00 | ± | 148.0 |
| Contr_l + CrossE_L + PMI_L + PR_L + Triplet_L | ALL | 72.91 3.58 | ± | 52.02 9.68 | ± | 45.44 1.88 | ± | 101.0 |
| Contr_l + CrossE_L + PMI_L + PR_L + Triplet_L | GAT | 77.08 1.06 | ± | 60.38 3.72 | ± | 47.90 1.19 | ± | 29.3 |
| Contr_l + CrossE_L + PMI_L + PR_L + Triplet_L | GCN | 71.01 0.76 | ± | 55.44 2.92 | ± | 45.13 2.20 | ± | 101.0 |
| Contr_l + CrossE_L + PMI_L + PR_L + Triplet_L | GIN | 72.53 2.50 | ± | 49.63 5.97 | ± | 46.46 2.04 | ± | 101.0 |
| Contr_l + CrossE_L + PMI_L + PR_L + Triplet_L | MPNN | 75.58 0.86 | ± | 57.17 4.05 | ± | 47.15 1.75 | ± | 56.7 |
| Contr_l + CrossE_L + PMI_L + PR_L + Triplet_L | PAGNN | 74.02 3.00 | ± | 52.65 8.76 | ± | 23.73 0.00 | ± | 120.3 |
| Contr_l + CrossE_L + PMI_L + PR_L + Triplet_L | SAGE | 63.64 1.95 | ± | 52.14 6.72 | ± | 51.20 2.64 | ± | 99.0 |
| Contr_l + CrossE_L + PMI_L + Triplet_L | ALL | 74.06 3.15 | ± | 61.01 7.54 | ± | 45.87 1.04 | ± | 59.0 |
| Contr_l + CrossE_L + PMI_L + Triplet_L | GAT | 76.43 0.87 | ± | 61.24 2.96 | ± | 46.83 1.17 | ± | 38.3 |
| Contr_l + CrossE_L + PMI_L + Triplet_L | GCN | 71.74 2.90 | ± | 50.67 3.18 | ± | 45.71 0.85 | ± | 109.3 |
| Contr_l + CrossE_L + PMI_L + Triplet_L | GIN | 70.67 1.92 | ± | 52.36 6.57 | ± | 46.35 1.88 | ± | 106.0 |
| Contr_l + CrossE_L + PMI_L + Triplet_L | MPNN | 75.64 0.68 | ± | 56.98 5.56 | ± | 48.63 1.24 | ± | 49.7 |
| Contr_l + CrossE_L + PMI_L + Triplet_L | PAGNN | 72.66 1.96 | ± | 49.04 9.30 | ± | 23.73 0.00 | ± | 138.0 |
| Contr_l + CrossE_L + PMI_L + Triplet_L | SAGE | 63.12 1.20 | ± | 54.37 7.23 | ± | 50.88 1.60 | ± | 91.7 |

Node Cls Precision Continued (↑)

| Loss Type | Model | CORA | | Citeseer | | Bitcoin Fraud Transaction | | Average Rank |
|---|---|---|---|---|---|---|---|---|
| Contr_l + CrossE_L + PR_L | ALL | 32.62 ± 1.02 | | 40.00 ± 7.01 | | 53.79 ± 4.56 | | 131.7 |
| Contr_l + CrossE_L + PR_L | GAT | 64.35 ± 17.05 | | 51.42 ± 5.14 | | 53.26 ± 4.98 | | 98.0 |
| Contr_l + CrossE_L + PR_L | GCN | 68.98 ± 1.64 | | 49.71 ± 4.50 | | 44.18 ± 2.31 | | 137.3 |
| Contr_l + CrossE_L + PR_L | GIN | 58.42 ± 13.51 | | 43.98 ± 2.32 | | 43.21 ± 12.44 | | 170.0 |
| Contr_l + CrossE_L + PR_L | MPNN | 67.72 ± 9.06 | | 49.30 ± 1.71 | | 44.74 ± 1.69 | | 135.0 |
| Contr_l + CrossE_L + PR_L | PAGNN | 61.74 ± 6.33 | | 42.59 ± 5.34 | | 23.73 ± 0.00 | | 179.3 |
| Contr_l + CrossE_L + PR_L | SAGE | 63.65 ± 7.58 | | 43.19 ± 3.58 | | 45.56 ± 12.30 | | 147.7 |
| Contr_l + CrossE_L + PR_L + Triplet_L | ALL | 64.49 ± 8.24 | | 47.82 ± 5.52 | | 44.50 ± 1.92 | | 147.0 |
| Contr_l + CrossE_L + PR_L + Triplet_L | GAT | 77.05 ± 0.71 | | 59.50 ± 1.75 | | 48.40 ± 0.95 | | 31.3 |
| Contr_l + CrossE_L + PR_L + Triplet_L | GCN | 73.06 ± 1.23 | | 55.35 ± 7.16 | | 43.65 ± 2.67 | | 103.0 |
| Contr_l + CrossE_L + PR_L + Triplet_L | GIN | 73.40 ± 2.49 | | 47.44 ± 2.33 | | 44.32 ± 3.63 | | 125.7 |
| Contr_l + CrossE_L + PR_L + Triplet_L | MPNN | 75.57 ± 1.25 | | 56.65 ± 4.57 | | 45.14 ± 1.33 | | 74.0 |
| Contr_l + CrossE_L + PR_L + Triplet_L | PAGNN | 66.26 ± 5.98 | | 49.53 ± 2.15 | | 23.73 ± 0.00 | | 157.7 |
| Contr_l + CrossE_L + PR_L + Triplet_L | SAGE | 71.66 ± 1.99 | | 54.47 ± 6.82 | | 50.41 ± 2.21 | | 75.7 |
| Contr_l + CrossE_L + Triplet_L | ALL | 71.75 ± 1.33 | | 55.43 ± 7.54 | | 43.86 ± 3.99 | | 109.3 |
| Contr_l + CrossE_L + Triplet_L | GAT | 79.39 ± 1.29 | | 63.30 ± 3.66 | | 46.37 ± 1.62 | | 27.7 |

Node Cls Precision Continued (↑)

| Loss Type | Model | CORA | | Citeseer | | Bitcoin Fraud Transaction | | Average Rank |
|---|---|---|---|---|---|---|---|---|
| Contr_l + CrossE_L + Triplet_L | GCN | 75.28 | ± | 57.47 | ± | 44.44 | ± | 79.0 |
| | | 1.10 | | 3.98 | | 1.63 | | |
| Contr_l + CrossE_L + Triplet_L | GIN | 77.37 | ± | 60.61 | ± | 47.27 | ± | 31.7 |
| | | 1.49 | | 5.25 | | 1.75 | | |
| Contr_l + CrossE_L + Triplet_L | MPNN | 77.35 | ± | 58.93 | ± | 46.25 | ± | 44.7 |
| | | 1.21 | | 4.22 | | 1.70 | | |
| Contr_l + CrossE_L + Triplet_L | PAGNN | 75.40 | ± | 53.53 | ± | 23.73 | ± | 113.7 |
| | | 1.41 | | 2.00 | | 0.00 | | |
| Contr_l + CrossE_L + Triplet_L | SAGE | 75.51 | ± | 58.16 | ± | 45.81 | ± | 61.3 |
| | | 2.41 | | 5.68 | | 2.02 | | |
| Contr_l + PMI_L | ALL | 72.15 | ± | 47.30 | ± | 45.22 | ± | 121.3 |
| | | 6.58 | | 2.44 | | 3.10 | | |
| Contr_l + PMI_L | GAT | 79.03 | ± | 61.66 | ± | 47.14 | ± | 25.3 |
| | | 1.14 | | 4.93 | | 2.04 | | |
| Contr_l + PMI_L | GCN | 71.69 | ± | 51.94 | ± | 43.40 | ± | 126.0 |
| | | 1.13 | | 3.25 | | 1.29 | | |
| Contr_l + PMI_L | GIN | 69.23 | ± | 48.99 | ± | 44.68 | ± | 134.7 |
| | | 3.17 | | 5.82 | | 2.56 | | |
| Contr_l + PMI_L | MPNN | 77.05 | ± | 59.35 | ± | 47.84 | ± | 35.0 |
| | | 1.15 | | 2.29 | | 1.01 | | |
| Contr_l + PMI_L | PAGNN | 72.24 | ± | 43.48 | ± | 23.73 | ± | 156.0 |
| | | 1.23 | | 1.10 | | 0.00 | | |
| Contr_l + PMI_L | SAGE | 61.02 | ± | 50.86 | ± | 53.38 | ± | 102.0 |
| | | 1.81 | | 6.03 | | 0.86 | | |
| Contr_l + PMI_L + PR_L | ALL | 71.06 | ± | 48.42 | ± | 39.87 | ± | 145.7 |
| | | 4.33 | | 1.53 | | 3.43 | | |
| Contr_l + PMI_L + PR_L | GAT | 77.77 | ± | 62.59 | ± | 45.67 | ± | 36.0 |
| | | 2.35 | | 5.89 | | 3.97 | | |
| Contr_l + PMI_L + PR_L | GCN | 73.03 | ± | 54.72 | ± | 45.65 | ± | 88.3 |
| | | 1.73 | | 5.18 | | 1.15 | | |
| Contr_l + PMI_L + PR_L | GIN | 69.95 | ± | 49.06 | ± | 41.97 | ± | 144.7 |
| | | 1.72 | | 10.09 | | 12.25 | | |

Node Cls Precision Continued (↑)

| Loss Type | Model | CORA | | | Citeseer | | | Bitcoin Fraud Transaction | | | Average Rank |
|---|---|---|---|---|---|---|---|---|---|---|---|
| Contr_l + PMI_L + PR_L | MPNN | 76.89 | ± | 2.15 | 58.32 | ± | 6.01 | 40.76 | ± | 3.99 | 76.0 |
| Contr_l + PMI_L + PR_L | PAGNN | 72.27 | ± | 4.51 | 48.62 | ± | 7.56 | 23.73 | ± | 0.00 | 143.0 |
| Contr_l + PMI_L + PR_L | SAGE | 56.79 | ± | 3.88 | 46.74 | ± | 9.54 | 45.15 | ± | 1.83 | 149.7 |
| Contr_l + PMI_L + PR_L + Triplet_L | ALL | 72.00 | ± | 3.33 | 48.91 | ± | 0.73 | 42.34 | ± | 3.68 | 133.3 |
| Contr_l + PMI_L + PR_L + Triplet_L | GAT | 77.40 | ± | 1.58 | 57.17 | ± | 4.23 | 48.78 | ± | 1.12 | 35.0 |
| Contr_l + PMI_L + PR_L + Triplet_L | GCN | 72.20 | ± | 2.43 | 58.55 | ± | 8.40 | 44.92 | ± | 1.92 | 84.0 |
| Contr_l + PMI_L + PR_L + Triplet_L | GIN | 72.88 | ± | 1.79 | 52.97 | ± | 5.22 | 44.84 | ± | 2.69 | 103.7 |
| Contr_l + PMI_L + PR_L + Triplet_L | MPNN | 75.78 | ± | 0.80 | 55.72 | ± | 3.52 | 46.44 | ± | 0.77 | 65.7 |
| Contr_l + PMI_L + PR_L + Triplet_L | PAGNN | 74.35 | ± | 2.42 | 52.12 | ± | 7.75 | 23.73 | ± | 0.00 | 124.0 |
| Contr_l + PMI_L + PR_L + Triplet_L | SAGE | 70.18 | ± | 3.86 | 50.53 | ± | 7.24 | 50.60 | ± | 1.88 | 95.0 |
| Contr_l + PR_L | ALL | 32.00 | ± | 9.10 | 38.00 | ± | 7.54 | 45.44 | ± | 13.97 | 163.7 |
| Contr_l + PR_L | GAT | 59.33 | ± | 12.99 | 49.67 | ± | 0.61 | 52.57 | ± | 2.66 | 106.7 |
| Contr_l + PR_L | GCN | 69.45 | ± | 3.50 | 51.82 | ± | 6.88 | 44.82 | ± | 1.53 | 126.0 |
| Contr_l + PR_L | GIN | 56.81 | ± | 8.33 | 42.60 | ± | 2.73 | 44.40 | ± | 7.67 | 164.7 |
| Contr_l + PR_L | MPNN | 63.83 | ± | 6.82 | 48.73 | ± | 1.42 | 40.82 | ± | 6.48 | 155.7 |
| Contr_l + PR_L | PAGNN | 59.02 | ± | 7.39 | 43.72 | ± | 4.13 | 23.73 | ± | 0.00 | 182.3 |

Node Cls Precision Continued (↑)

| Loss Type | Model | CORA | | Citeseer | | Bitcoin Fraud Transaction | | Average Rank |
|---|---|---|---|---|---|---|---|---|
| Contr_l + PR_L | SAGE | 59.25 ± 13.56 | | 44.75 ± 3.32 | | 40.20 ± 15.10 | | 172.0 |
| Contr_l + PR_L + Triplet_L | ALL | 61.60 ± 7.36 | | 45.87 ± 1.98 | | 45.95 ± 2.19 | | 140.7 |
| Contr_l + PR_L + Triplet_L | GAT | 78.96 ± 1.49 | | 60.95 ± 3.80 | | 48.40 ± 1.25 | | 21.3 |
| Contr_l + PR_L + Triplet_L | GCN | 72.22 ± 4.24 | | 52.56 ± 2.50 | | 44.40 ± 1.40 | | 112.0 |
| Contr_l + PR_L + Triplet_L | GIN | 73.34 ± 1.39 | | 48.11 ± 4.42 | | 44.67 ± 4.82 | | 119.7 |
| Contr_l + PR_L + Triplet_L | MPNN | 74.43 ± 2.06 | | 55.84 ± 7.63 | | 46.95 ± 1.42 | | 68.3 |
| Contr_l + PR_L + Triplet_L | PAGNN | 71.26 ± 6.19 | | 55.08 ± 13.04 | | 23.73 ± 0.00 | | 129.7 |
| Contr_l + PR_L + Triplet_L | SAGE | 70.46 ± 2.43 | | 56.95 ± 7.84 | | 48.92 ± 2.64 | | 75.0 |
| Contr_l + Triplet_L | ALL | 73.14 ± 1.76 | | 52.15 ± 1.07 | | 39.50 ± 9.42 | | 122.0 |
| Contr_l + Triplet_L | GAT | 78.60 ± 0.82 | | 61.99 ± 2.52 | | 46.04 ± 1.82 | | 32.0 |
| Contr_l + Triplet_L | GCN | 75.28 ± 1.99 | | 61.37 ± 3.19 | | 45.66 ± 1.39 | | 54.7 |
| Contr_l + Triplet_L | GIN | 75.74 ± 0.53 | | 58.36 ± 3.24 | | 46.43 ± 0.89 | | 55.7 |
| Contr_l + Triplet_L | MPNN | 75.40 ± 1.67 | | 57.98 ± 3.79 | | 46.92 ± 2.76 | | 57.0 |
| Contr_l + Triplet_L | PAGNN | 76.54 ± 0.90 | | 57.80 ± 6.91 | | 23.73 ± 0.00 | | 87.7 |
| Contr_l + Triplet_L | SAGE | 75.87 ± 1.95 | | 57.62 ± 4.80 | | 47.27 ± 2.58 | | 50.0 |
| CrossE_L | ALL | 5.03 ± 0.00 | | 3.58 ± 0.00 | | 23.73 ± 0.00 | | 198.3 |

Node Cls Precision Continued (↑)

| Loss Type | Model | CORA | | Citeseer | | Bitcoin Fraud Transaction | | Average Rank |
|---|---|---|---|---|---|---|---|---|
| CrossE_L | GAT | 5.03 ± 0.00 | | 3.58 ± 0.00 | | 23.73 ± 0.00 | | 199.3 |
| CrossE_L | GCN | 5.03 ± 0.00 | | 3.58 ± 0.00 | | 23.73 ± 0.00 | | 200.3 |
| CrossE_L | GIN | 5.03 ± 0.00 | | 3.58 ± 0.00 | | 23.73 ± 0.00 | | 201.3 |
| CrossE_L | MPNN | 5.03 ± 0.00 | | 3.58 ± 0.00 | | 23.73 ± 0.00 | | 202.3 |
| CrossE_L | PAGNN | 5.03 ± 0.00 | | 3.58 ± 0.00 | | 23.73 ± 0.00 | | 203.3 |
| CrossE_L | SAGE | 5.03 ± 0.00 | | 3.58 ± 0.00 | | 23.73 ± 0.00 | | 204.3 |
| CrossE_L + PMI_L | ALL | 73.79 ± 2.46 | | 51.26 ± 3.43 | | 45.26 ± 1.51 | | 100.7 |
| CrossE_L + PMI_L | GAT | 78.19 ± 3.34 | | 61.05 ± 2.97 | | 47.09 ± 1.32 | | 30.0 |
| CrossE_L + PMI_L | GCN | 70.91 ± 1.45 | | 53.68 ± 3.85 | | 45.50 ± 1.04 | | 106.3 |
| CrossE_L + PMI_L | GIN | 71.47 ± 1.43 | | 44.67 ± 2.92 | | 43.98 ± 2.17 | | 145.7 |
| CrossE_L + PMI_L | MPNN | 75.84 ± 0.88 | | 59.70 ± 2.52 | | 49.17 ± 0.96 | | 34.7 |
| CrossE_L + PMI_L | PAGNN | 73.77 ± 1.63 | | 51.61 ± 10.49 | | 23.73 ± 0.00 | | 133.0 |
| CrossE_L + PMI_L | SAGE | 59.50 ± 3.59 | | 39.98 ± 2.00 | | 52.49 ± 1.17 | | 128.7 |
| CrossE_L + PMI_L + PR_L | ALL | 74.22 ± 4.45 | | 48.13 ± 4.12 | | 38.49 ± 2.48 | | 132.0 |
| CrossE_L + PMI_L + PR_L | GAT | 76.37 ± 0.96 | | 59.70 ± 4.13 | | 47.43 ± 1.64 | | 39.3 |
| CrossE_L + PMI_L + PR_L | GCN | 71.98 ± 1.58 | | 51.64 ± 5.39 | | 44.21 ± 1.00 | | 120.7 |

Node Cls Precision Continued (↑)

| Loss Type | Model | CORA | | Citeseer | | Bitcoin Fraud Transaction | | Average Rank |
|---|---|---|---|---|---|---|---|---|
| CrossE_L + PMI_L + PR_L | GIN | 70.05 ± 0.91 | | 45.19 ± 8.03 | | 38.60 ± 14.56 | | 159.3 |
| CrossE_L + PMI_L + PR_L | MPNN | 77.38 ± 1.33 | | 54.51 ± 3.93 | | 44.12 ± 6.47 | | 81.0 |
| CrossE_L + PMI_L + PR_L | PAGNN | 74.56 ± 1.97 | | 48.25 ± 7.12 | | 23.73 ± 0.00 | | 137.3 |
| CrossE_L + PMI_L + PR_L | SAGE | 60.08 ± 4.65 | | 43.77 ± 3.78 | | 42.64 ± 10.80 | | 169.0 |
| CrossE_L + PMI_L + PR_L + Triplet_L | ALL | 70.75 ± 1.88 | | 52.32 ± 7.57 | | 44.22 ± 2.14 | | 124.7 |
| CrossE_L + PMI_L + PR_L + Triplet_L | GAT | 78.27 ± 0.93 | | 58.16 ± 4.73 | | 47.98 ± 1.38 | | 31.3 |
| CrossE_L + PMI_L + PR_L + Triplet_L | GCN | 69.70 ± 1.16 | | 53.87 ± 4.82 | | 44.11 ± 1.60 | | 124.3 |
| CrossE_L + PMI_L + PR_L + Triplet_L | GIN | 71.58 ± 2.94 | | 54.71 ± 8.70 | | 45.79 ± 0.87 | | 96.0 |
| CrossE_L + PMI_L + PR_L + Triplet_L | MPNN | 76.24 ± 1.03 | | 56.36 ± 4.60 | | 48.04 ± 1.07 | | 48.3 |
| CrossE_L + PMI_L + PR_L + Triplet_L | PAGNN | 74.46 ± 2.65 | | 54.07 ± 9.28 | | 23.73 ± 0.00 | | 119.3 |
| CrossE_L + PMI_L + PR_L + Triplet_L | SAGE | 66.82 ± 2.69 | | 55.08 ± 7.18 | | 51.69 ± 0.85 | | 84.3 |
| CrossE_L + PMI_L + Triplet_L | ALL | 74.05 ± 3.04 | | 55.22 ± 3.87 | | 46.57 ± 1.32 | | 75.3 |
| CrossE_L + PMI_L + Triplet_L | GAT | 78.26 ± 1.67 | | 61.07 ± 4.03 | | 46.95 ± 1.21 | | 30.7 |
| CrossE_L + PMI_L + Triplet_L | GCN | 70.45 ± 0.76 | | 56.81 ± 5.54 | | 44.39 ± 1.54 | | 108.3 |
| CrossE_L + PMI_L + Triplet_L | GIN | 71.42 ± 3.07 | | 48.84 ± 3.85 | | 45.76 ± 0.53 | | 116.3 |
| CrossE_L + PMI_L + Triplet_L | MPNN | 77.07 ± 2.22 | | 58.62 ± 4.82 | | 47.75 ± 2.01 | | 36.0 |

Node Cls Precision Continued (↑)

| Loss Type | | | | Model | CORA | | Citeseer | | Bitcoin Fraud Transaction | | Average Rank |
|---|---|---|---|---|---|---|---|---|---|---|---|
| CrossE_L + PMI_L + Triplet_L | | | | PAGNN | 73.62 ± 2.08 | | 49.65 ± 7.70 | | 23.73 ± 0.00 | | 138.0 |
| CrossE_L + PMI_L + Triplet_L | | | | SAGE | 70.55 ± 2.49 | | 52.13 ± 4.93 | | 50.19 ± 1.55 | | 90.3 |
| CrossE_L + PR_L | | | | ALL | 18.75 ± 0.28 | | 35.93 ± 3.61 | | 53.79 ± 4.55 | | 135.0 |
| CrossE_L + PR_L | | | | GAT | 29.18 ± 11.99 | | 44.96 ± 6.76 | | 23.73 ± 0.00 | | 190.7 |
| CrossE_L + PR_L | | | | GCN | 67.11 ± 13.64 | | 45.42 ± 1.47 | | 42.09 ± 17.10 | | 159.3 |
| CrossE_L + PR_L | | | | GIN | 33.91 ± 10.28 | | 42.46 ± 4.44 | | 44.87 ± 13.85 | | 165.0 |
| CrossE_L + PR_L | | | | MPNN | 48.78 ± 21.98 | | 49.16 ± 1.01 | | 39.30 ± 3.68 | | 164.3 |
| CrossE_L + PR_L | | | | PAGNN | 52.58 ± 15.37 | | 39.89 ± 9.15 | | 23.73 ± 0.00 | | 195.0 |
| CrossE_L + PR_L | | | | SAGE | 39.39 ± 10.20 | | 40.37 ± 2.52 | | 51.65 ± 5.57 | | 134.0 |
| CrossE_L + PR_L + Triplet_L | | | | ALL | 61.39 ± 10.86 | | 45.14 ± 2.82 | | 49.43 ± 4.84 | | 123.7 |
| CrossE_L + PR_L + Triplet_L | | | | GAT | 76.20 ± 1.61 | | 55.82 ± 4.48 | | 49.32 ± 1.71 | | 45.7 |
| CrossE_L + PR_L + Triplet_L | | | | GCN | 71.27 ± 4.33 | | 52.60 ± 7.23 | | 44.40 ± 2.37 | | 118.3 |
| CrossE_L + PR_L + Triplet_L | | | | GIN | 69.48 ± 1.07 | | 48.09 ± 6.59 | | 40.23 ± 7.24 | | 152.0 |
| CrossE_L + PR_L + Triplet_L | | | | MPNN | 74.41 ± 2.40 | | 49.52 ± 1.68 | | 44.85 ± 1.87 | | 106.3 |
| CrossE_L + PR_L + Triplet_L | | | | PAGNN | 65.16 ± 5.85 | | 44.81 ± 2.06 | | 23.73 ± 0.00 | | 179.3 |
| CrossE_L + PR_L + Triplet_L | | | | SAGE | 71.06 ± 2.22 | | 53.81 ± 9.75 | | 52.48 ± 2.12 | | 78.7 |

Node Cls Precision Continued (↑)

| Loss Type | Model | CORA | | Citeseer | | Bitcoin Fraud Transaction | | Average Rank |
|---|---|---|---|---|---|---|---|---|
| CrossE_L + Triplet_L | ALL | 72.35 | ± | 53.35 | ± | 45.18 | ± | 100.0 |
| | | 0.67 | | 2.82 | | 1.97 | | |
| CrossE_L + Triplet_L | GAT | 79.89 | ± | 59.64 | ± | 47.16 | ± | 29.3 |
| | | 1.21 | | 1.50 | | 1.35 | | |
| CrossE_L + Triplet_L | GCN | 75.38 | ± | 54.77 | ± | 44.36 | ± | 91.0 |
| | | 2.12 | | 5.43 | | 1.42 | | |
| CrossE_L + Triplet_L | GIN | 76.41 | ± | 53.62 | ± | 47.48 | ± | 61.3 |
| | | 1.77 | | 2.77 | | 1.20 | | |
| CrossE_L + Triplet_L | MPNN | 77.67 | ± | 60.05 | ± | 47.98 | ± | 28.0 |
| | | 1.21 | | 3.42 | | 1.00 | | |
| CrossE_L + Triplet_L | PAGNN | 76.51 | ± | 57.23 | ± | 23.73 | ± | 96.0 |
| | | 1.29 | | 4.69 | | 0.00 | | |
| CrossE_L + Triplet_L | SAGE | 76.30 | ± | 58.14 | ± | 49.14 | ± | 36.3 |
| | | 1.01 | | 5.35 | | 1.20 | | |
| PMI_L | ALL | 74.50 | ± | 52.78 | ± | 47.88 | ± | 72.7 |
| | | 2.23 | | 8.18 | | 0.80 | | |
| PMI_L | GAT | 79.25 | ± | 62.25 | ± | 47.04 | ± | 25.0 |
| | | 1.81 | | 5.03 | | 1.53 | | |
| PMI_L | GCN | 71.85 | ± | 53.22 | ± | 43.44 | ± | 119.3 |
| | | 1.41 | | 4.78 | | 2.04 | | |
| PMI_L | GIN | 67.76 | ± | 48.00 | ± | 41.99 | ± | 153.3 |
| | | 3.65 | | 8.40 | | 1.23 | | |
| PMI_L | MPNN | 75.81 | ± | 60.09 | ± | 47.54 | ± | 41.3 |
| | | 1.26 | | 4.15 | | 1.33 | | |
| PMI_L | PAGNN | 71.80 | ± | 53.58 | ± | 23.73 | ± | 138.3 |
| | | 2.09 | | 8.27 | | 0.00 | | |
| PMI_L | SAGE | 54.36 | ± | 41.11 | ± | 52.91 | ± | 129.7 |
| | | 3.87 | | 2.86 | | 0.84 | | |
| PMI_L + PR_L | ALL | 73.83 | ± | 46.41 | ± | 39.64 | ± | 136.3 |
| | | 2.40 | | 2.95 | | 3.75 | | |
| PMI_L + PR_L | GAT | 77.36 | ± | 53.77 | ± | 45.09 | ± | 74.0 |
| | | 2.06 | | 4.98 | | 8.29 | | |

Node Cls Precision Continued (↑)

| Loss Type | Model | CORA | | Citeseer | | Bitcoin Fraud Transaction | | Average Rank |
|-----------|-------|------|---|----------|---|---------------------------|---|--------------|
| PMI_L + PR_L | GCN | 71.00 | ± | 57.08 | ± | 44.99 | ± | 98.0 |
| | | 1.11 | | 3.30 | | 0.64 | | |
| PMI_L + PR_L | GIN | 71.36 | ± | 49.48 | ± | 30.40 | ± | 143.0 |
| | | 2.01 | | 8.56 | | 14.92 | | |
| PMI_L + PR_L | MPNN | 75.79 | ± | 60.10 | ± | 40.16 | ± | 78.7 |
| | | 0.91 | | 6.03 | | 5.43 | | |
| PMI_L + PR_L | PAGNN | 73.00 | ± | 44.06 | ± | 23.73 | ± | 158.3 |
| | | 2.61 | | 2.36 | | 0.00 | | |
| PMI_L + PR_L | SAGE | 59.69 | ± | 45.17 | ± | 48.31 | ± | 129.0 |
| | | 6.23 | | 3.02 | | 5.79 | | |
| PMI_L + PR_L + Triplet_L | ALL | 69.05 | ± | 48.48 | ± | 43.77 | ± | 145.7 |
| | | 2.53 | | 1.93 | | 4.39 | | |
| PMI_L + PR_L + Triplet_L | GAT | 77.72 | ± | 61.48 | ± | 48.93 | ± | 18.3 |
| | | 2.84 | | 4.65 | | 1.00 | | |
| PMI_L + PR_L + Triplet_L | GCN | 71.84 | ± | 55.22 | ± | 44.83 | ± | 101.0 |
| | | 1.69 | | 3.39 | | 2.21 | | |
| PMI_L + PR_L + Triplet_L | GIN | 68.57 | ± | 56.01 | ± | 44.15 | ± | 117.3 |
| | | 1.95 | | 6.94 | | 1.21 | | |
| PMI_L + PR_L + Triplet_L | MPNN | 76.34 | ± | 60.73 | ± | 46.22 | ± | 45.0 |
| | | 1.64 | | 7.12 | | 2.21 | | |
| PMI_L + PR_L + Triplet_L | PAGNN | 75.04 | ± | 47.24 | ± | 23.73 | ± | 144.0 |
| | | 2.26 | | 1.40 | | 0.00 | | |
| PMI_L + PR_L + Triplet_L | SAGE | 67.33 | ± | 54.91 | ± | 50.51 | ± | 85.0 |
| | | 4.10 | | 7.37 | | 0.32 | | |
| PMI_L + Triplet_L | ALL | 74.23 | ± | 52.16 | ± | 45.62 | ± | 92.0 |
| | | 1.84 | | 1.89 | | 1.87 | | |
| PMI_L + Triplet_L | GAT | 78.35 | ± | 61.81 | ± | 47.15 | ± | 25.7 |
| | | 2.88 | | 3.83 | | 1.00 | | |
| PMI_L + Triplet_L | GCN | 72.87 | ± | 53.64 | ± | 44.83 | ± | 102.7 |
| | | 1.97 | | 5.40 | | 1.40 | | |
| PMI_L + Triplet_L | GIN | 71.92 | ± | 55.17 | ± | 45.98 | ± | 89.3 |
| | | 3.23 | | 8.90 | | 2.01 | | |

Node Cls Precision Continued (↑)

| Loss Type | Model | CORA | | Citeseer | | Bitcoin Fraud Transaction | | Average Rank |
|-----------|-------|------|---|----------|---|---------------------------|---|--------------|
| PMI_L + Triplet_L | MPNN | 75.95 ± 1.68 | | 58.63 ± 1.15 | | 47.48 ± 2.09 | | 43.7 |
| PMI_L + Triplet_L | PAGNN | 71.15 ± 0.90 | | 47.63 ± 6.21 | | 23.73 ± 0.00 | | 163.0 |
| PMI_L + Triplet_L | SAGE | 65.27 ± 2.56 | | 51.22 ± 5.07 | | 52.49 ± 0.96 | | 99.7 |
| PR_L | ALL | 26.35 ± 7.16 | | 33.70 ± 6.78 | | 53.79 ± 4.56 | | 136.0 |
| PR_L | GAT | 18.32 ± 0.44 | | 45.46 ± 7.41 | | 23.73 ± 0.00 | | 192.7 |
| PR_L | GCN | 62.57 ± 10.96 | | 48.77 ± 7.91 | | 23.73 ± 0.00 | | 173.3 |
| PR_L | GIN | 34.86 ± 17.41 | | 41.65 ± 1.15 | | 54.89 ± 4.97 | | 128.7 |
| PR_L | MPNN | 66.41 ± 2.18 | | 48.51 ± 0.56 | | 41.75 ± 3.99 | | 154.0 |
| PR_L | PAGNN | 52.04 ± 15.35 | | 37.38 ± 6.26 | | 23.73 ± 0.00 | | 199.0 |
| PR_L | SAGE | 37.71 ± 8.09 | | 40.20 ± 3.91 | | 42.31 ± 12.62 | | 180.0 |
| PR_L + Triplet_L | ALL | 32.90 ± 10.11 | | 34.37 ± 5.86 | | 53.79 ± 4.56 | | 134.3 |
| PR_L + Triplet_L | GAT | 68.35 ± 13.61 | | 49.72 ± 0.40 | | 48.47 ± 14.56 | | 104.7 |
| PR_L + Triplet_L | GCN | 70.56 ± 2.70 | | 55.32 ± 9.85 | | 44.51 ± 1.57 | | 110.0 |
| PR_L + Triplet_L | GIN | 61.80 ± 6.07 | | 44.06 ± 1.00 | | 48.77 ± 9.23 | | 127.0 |
| PR_L + Triplet_L | MPNN | 72.27 ± 6.73 | | 47.86 ± 1.20 | | 44.40 ± 4.14 | | 128.0 |
| PR_L + Triplet_L | PAGNN | 55.62 ± 5.03 | | 34.29 ± 6.67 | | 23.73 ± 0.00 | | 199.3 |

Node Cls Precision Continued (↑)

| Loss Type | Model | CORA | | Citeseer | | Bitcoin Fraud Transaction | | Average Rank |
|---|---|---|---|---|---|---|---|---|
| PR_L + Triplet_L | SAGE | 41.43 ± 8.49 | | 42.71 ± 4.27 | | 52.61 ± 2.73 | | 129.3 |
| Triplet_L | ALL | 74.49 ± 1.03 | | 56.22 ± 4.61 | | 44.52 ± 0.89 | | 85.3 |
| Triplet_L | GAT | 79.99 ± 1.42 | | 61.21 ± 1.97 | | 47.68 ± 0.86 | | 20.7 |
| Triplet_L | GCN | 75.85 ± 2.03 | | 57.55 ± 5.62 | | 45.05 ± 1.86 | | 68.0 |
| Triplet_L | GIN | 76.25 ± 1.08 | | 53.85 ± 4.91 | | 48.69 ± 1.88 | | 55.3 |
| Triplet_L | MPNN | 77.15 ± 1.36 | | 59.71 ± 2.46 | | 47.73 ± 1.48 | | 32.7 |
| Triplet_L | PAGNN | 76.05 ± 1.16 | | 56.43 ± 5.63 | | 23.73 ± 0.00 | | 105.7 |
| Triplet_L | SAGE | 75.90 ± 1.38 | | 58.98 ± 5.43 | | 50.05 ± 1.82 | | 34.0 |

**Table 4.** Node Cls Recall (Sensitivity) Performance (↑): Top-ranked results are highlighted in **1st**, second-ranked in **2nd**, and third-ranked in **3rd**.

| Loss Type | Model | CORA | | Citeseer | | Bitcoin Fraud Transaction | | Average Rank |
|---|---|---|---|---|---|---|---|---|
| Contr_l | ALL | 65.17 ± 1.64 | | 54.05 ± 2.25 | | 34.11 ± 0.71 | | 112.7 |
| Contr_l | GAT | 79.77 ± 0.62 | | 61.90 ± 1.07 | | 39.20 ± 0.72 | | 15.3 |
| Contr_l | GCN | 73.68 ± 2.58 | | 56.94 ± 2.10 | | 37.11 ± 0.98 | | 60.7 |
| Contr_l | GIN | 75.41 ± 0.90 | | 57.58 ± 1.47 | | 38.02 ± 1.07 | | 38.3 |

Continued on next page

Node Cls Recall (Sensitivity) Continued (↑)

| Loss Type | Model | CORA | | Citeseer | | Bitcoin Fraud Transaction | | Average Rank |
|---|---|---|---|---|---|---|---|---|
| Contr_l | MPNN | 76.78 1.51 | ± | 57.91 0.95 | ± | 39.03 1.16 | ± | 25.0 |
| Contr_l | PAGNN | 72.94 3.08 | ± | 56.58 1.89 | ± | 33.33 0.00 | ± | 93.0 |
| Contr_l | SAGE | 72.90 1.90 | ± | 55.63 2.02 | ± | 38.19 0.71 | ± | 61.7 |
| Contr_l + CrossE_L | ALL | 65.55 1.27 | ± | 51.72 1.89 | ± | 33.69 0.70 | ± | 121.0 |
| Contr_l + CrossE_L | GAT | 78.47 1.96 | ± | 60.91 1.84 | ± | 37.48 1.19 | ± | 27.0 |
| Contr_l + CrossE_L | GCN | 74.48 1.40 | ± | 55.80 1.52 | ± | 37.05 0.69 | ± | 64.3 |
| Contr_l + CrossE_L | GIN | 73.03 1.98 | ± | 56.62 1.01 | ± | 38.91 0.56 | ± | 50.3 |
| Contr_l + CrossE_L | MPNN | 76.06 2.02 | ± | 57.75 0.91 | ± | 37.94 0.81 | ± | 35.0 |
| Contr_l + CrossE_L | PAGNN | 71.17 2.47 | ± | 56.12 0.97 | ± | 33.33 0.00 | ± | 99.7 |
| Contr_l + CrossE_L | SAGE | 73.10 2.34 | ± | 56.30 1.84 | ± | 35.79 1.71 | ± | 78.0 |
| Contr_l + CrossE_L + PMI_L | ALL | 64.44 4.04 | ± | 51.44 2.39 | ± | 36.84 1.34 | ± | 108.7 |
| Contr_l + CrossE_L + PMI_L | GAT | 76.25 2.71 | ± | 57.86 0.83 | ± | 40.27 1.03 | ± | 19.7 |
| Contr_l + CrossE_L + PMI_L | GCN | 65.89 2.42 | ± | 51.60 1.58 | ± | 36.20 0.44 | ± | 106.7 |
| Contr_l + CrossE_L + PMI_L | GIN | 64.48 1.21 | ± | 46.45 1.61 | ± | 35.90 0.64 | ± | 134.7 |
| Contr_l + CrossE_L + PMI_L | MPNN | 74.97 1.56 | ± | 55.43 1.62 | ± | 40.16 0.50 | ± | 43.0 |
| Contr_l + CrossE_L + PMI_L | PAGNN | 62.23 1.63 | ± | 43.89 1.61 | ± | 33.33 0.00 | ± | 166.0 |

Node Cls Recall (Sensitivity) Continued (↑)

| Loss Type | Model | CORA | | Citeseer | | Bitcoin Fraud Transaction | | Average Rank |
|---|---|---|---|---|---|---|---|---|
| Contr_l + CrossE_L + PMI_L | SAGE | 52.34<br>5.81 | ± | 43.38<br>2.76 | ± | 42.68<br>1.59 | ± | 118.3 |
| Contr_l + CrossE_L + PMI_L + PR_L | ALL | 51.61<br>3.97 | ± | 47.69<br>3.42 | ± | 34.10<br>0.46 | ± | 155.0 |
| Contr_l + CrossE_L + PMI_L + PR_L | GAT | 76.04<br>2.33 | ± | 57.18<br>0.55 | ± | 38.12<br>1.66 | ± | 37.0 |
| Contr_l + CrossE_L + PMI_L + PR_L | GCN | 68.23<br>3.39 | ± | 50.71<br>1.37 | ± | 37.25<br>1.10 | ± | 95.7 |
| Contr_l + CrossE_L + PMI_L + PR_L | GIN | 63.95<br>3.81 | ± | 44.36<br>1.54 | ± | 34.22<br>1.31 | ± | 149.7 |
| Contr_l + CrossE_L + PMI_L + PR_L | MPNN | 74.41<br>2.31 | ± | 56.58<br>2.02 | ± | 35.60<br>3.05 | ± | 71.0 |
| Contr_l + CrossE_L + PMI_L + PR_L | PAGNN | 63.56<br>1.29 | ± | 46.31<br>1.28 | ± | 33.33<br>0.00 | ± | 156.3 |
| Contr_l + CrossE_L + PMI_L + PR_L | SAGE | 48.97<br>3.90 | ± | 43.69<br>3.33 | ± | 34.73<br>1.38 | ± | 164.0 |
| Contr_l + CrossE_L + PMI_L + PR_L + Triplet_L | ALL | 64.95<br>2.74 | ± | 51.08<br>3.21 | ± | 36.23<br>1.19 | ± | 112.3 |
| Contr_l + CrossE_L + PMI_L + PR_L + Triplet_L | GAT | 75.39<br>0.79 | ± | 58.24<br>0.88 | ± | 40.77<br>0.86 | ± | 17.7 |
| Contr_l + CrossE_L + PMI_L + PR_L + Triplet_L | GCN | 66.84<br>2.79 | ± | 52.10<br>2.46 | ± | 37.21<br>1.15 | ± | 92.3 |
| Contr_l + CrossE_L + PMI_L + PR_L + Triplet_L | GIN | 65.24<br>2.72 | ± | 48.82<br>2.71 | ± | 36.89<br>0.75 | ± | 114.0 |
| Contr_l + CrossE_L + PMI_L + PR_L + Triplet_L | MPNN | 74.36<br>0.46 | ± | 55.23<br>1.36 | ± | 39.64<br>1.60 | ± | 49.3 |
| Contr_l + CrossE_L + PMI_L + PR_L + Triplet_L | PAGNN | 63.21<br>2.37 | ± | 47.21<br>0.78 | ± | 33.33<br>0.00 | ± | 155.3 |
| Contr_l + CrossE_L + PMI_L + PR_L + Triplet_L | SAGE | 60.80<br>3.10 | ± | 50.35<br>3.37 | ± | 39.66<br>1.27 | ± | 102.7 |
| Contr_l + CrossE_L + PMI_L + Triplet_L | ALL | 70.29<br>2.69 | ± | 55.94<br>0.98 | ± | 37.59<br>1.38 | ± | 67.7 |

Node Cls Recall (Sensitivity) Continued ($\uparrow$)

| Loss Type | Model | CORA | | Citeseer | | Bitcoin Fraud Transaction | | Average Rank |
|---|---|---|---|---|---|---|---|---|
| Contr_l + CrossE_L + PMI_L + Triplet_L | GAT | 74.52 1.55 | ± | 58.17 0.72 | ± | 40.46 0.71 | ± | 25.0 |
| Contr_l + CrossE_L + PMI_L + Triplet_L | GCN | 68.42 2.05 | ± | 50.98 1.52 | ± | 37.31 0.67 | ± | 93.3 |
| Contr_l + CrossE_L + PMI_L + Triplet_L | GIN | 64.55 1.75 | ± | 48.78 1.15 | ± | 37.59 1.99 | ± | 110.3 |
| Contr_l + CrossE_L + PMI_L + Triplet_L | MPNN | 73.79 1.38 | ± | 55.75 1.13 | ± | 40.95 0.63 | ± | 41.7 |
| Contr_l + CrossE_L + PMI_L + Triplet_L | PAGNN | 63.58 1.87 | ± | 46.71 2.22 | ± | 33.33 0.00 | ± | 154.7 |
| Contr_l + CrossE_L + PMI_L + Triplet_L | SAGE | 60.84 1.58 | ± | 48.47 4.21 | ± | 41.65 0.66 | ± | 98.7 |
| Contr_l + CrossE_L + PR_L | ALL | 26.15 1.26 | ± | 33.89 2.14 | ± | 33.62 0.05 | ± | 186.0 |
| Contr_l + CrossE_L + PR_L | GAT | 54.79 19.61 | ± | 52.77 3.18 | ± | 34.66 0.95 | ± | 128.0 |
| Contr_l + CrossE_L + PR_L | GCN | 59.87 1.94 | ± | 50.51 1.01 | ± | 35.79 0.78 | ± | 130.7 |
| Contr_l + CrossE_L + PR_L | GIN | 34.38 6.79 | ± | 45.67 2.65 | ± | 33.50 0.11 | ± | 175.0 |
| Contr_l + CrossE_L + PR_L | MPNN | 47.95 5.98 | ± | 50.21 0.86 | ± | 36.21 2.07 | ± | 138.0 |
| Contr_l + CrossE_L + PR_L | PAGNN | 40.68 2.93 | ± | 36.29 1.72 | ± | 33.33 0.00 | ± | 185.0 |
| Contr_l + CrossE_L + PR_L | SAGE | 57.01 14.45 | ± | 46.05 4.83 | ± | 35.59 2.14 | ± | 149.7 |
| Contr_l + CrossE_L + PR_L + Triplet_L | ALL | 58.68 2.42 | ± | 48.79 0.83 | ± | 36.26 1.30 | ± | 134.0 |
| Contr_l + CrossE_L + PR_L + Triplet_L | GAT | 75.85 1.76 | ± | 57.92 2.07 | ± | 40.59 0.98 | ± | 19.3 |
| Contr_l + CrossE_L + PR_L + Triplet_L | GCN | 69.20 2.54 | ± | 52.94 0.77 | ± | 36.85 1.37 | ± | 87.3 |

Node Cls Recall (Sensitivity) Continued (↑)

| Loss Type | Model | CORA | | Citeseer | | Bitcoin Fraud Transaction | | Average Rank |
|---|---|---|---|---|---|---|---|---|
| Contr_l + CrossE_L + PR_L + Triplet_L | GIN | 68.92 ± 3.78 | | 49.82 ± 1.73 | | 36.87 ± 2.24 | | 104.0 |
| Contr_l + CrossE_L + PR_L + Triplet_L | MPNN | 71.53 ± 0.72 | | 56.25 ± 2.03 | | 37.72 ± 1.94 | | 63.0 |
| Contr_l + CrossE_L + PR_L + Triplet_L | PAGNN | 54.93 ± 2.54 | | 51.34 ± 3.96 | | 33.33 ± 0.00 | | 148.0 |
| Contr_l + CrossE_L + PR_L + Triplet_L | SAGE | 71.04 ± 2.34 | | 52.96 ± 1.22 | | 37.66 ± 2.64 | | 74.7 |
| Contr_l + CrossE_L + Triplet_L | ALL | 71.28 ± 1.89 | | 54.07 ± 2.24 | | 34.69 ± 0.75 | | 95.7 |
| Contr_l + CrossE_L + Triplet_L | GAT | 79.25 ± 1.07 | | 60.61 ± 1.37 | | 38.02 ± 0.96 | | 22.7 |
| Contr_l + CrossE_L + Triplet_L | GCN | 74.87 ± 1.40 | | 56.37 ± 1.90 | | 37.24 ± 0.92 | | 55.7 |
| Contr_l + CrossE_L + Triplet_L | GIN | 76.69 ± 1.54 | | 56.85 ± 1.88 | | 39.46 ± 0.80 | | 30.0 |
| Contr_l + CrossE_L + Triplet_L | MPNN | 77.26 ± 1.40 | | 58.40 ± 1.46 | | 38.19 ± 1.18 | | 24.7 |
| Contr_l + CrossE_L + Triplet_L | PAGNN | 72.35 ± 3.50 | | 57.10 ± 1.30 | | 33.33 ± 0.00 | | 94.7 |
| Contr_l + CrossE_L + Triplet_L | SAGE | 75.96 ± 2.92 | | 55.77 ± 1.13 | | 36.72 ± 0.95 | | 61.3 |
| Contr_l + PMI_L | ALL | 64.88 ± 2.76 | | 50.62 ± 2.34 | | 36.73 ± 1.44 | | 112.3 |
| Contr_l + PMI_L | GAT | 77.02 ± 1.65 | | 58.14 ± 0.87 | | 40.65 ± 1.01 | | 12.7 |
| Contr_l + PMI_L | GCN | 67.57 ± 2.21 | | 49.58 ± 1.88 | | 36.25 ± 0.41 | | 113.3 |
| Contr_l + PMI_L | GIN | 64.06 ± 1.75 | | 47.52 ± 2.75 | | 36.06 ± 0.71 | | 132.0 |
| Contr_l + PMI_L | MPNN | 75.18 ± 1.88 | | 56.08 ± 0.95 | | 40.50 ± 1.13 | | 35.3 |

Node Cls Recall (Sensitivity) Continued (↑)

| Loss Type | Model | CORA | | | Citeseer | | | Bitcoin Fraud Transaction | | | Average Rank |
|---|---|---|---|---|---|---|---|---|---|---|---|
| Contr_l + PMI_L | PAGNN | 62.24 | ± | 2.18 | 44.11 | ± | 1.33 | 33.33 | ± | 0.00 | 167.3 |
| Contr_l + PMI_L | SAGE | 51.44 | ± | 4.61 | 47.57 | ± | 2.30 | 42.56 | ± | 0.90 | 107.7 |
| Contr_l + PMI_L + PR_L | ALL | 56.32 | ± | 0.94 | 48.87 | ± | 1.89 | 34.54 | ± | 1.13 | 143.7 |
| Contr_l + PMI_L + PR_L | GAT | 76.95 | ± | 1.49 | 57.28 | ± | 1.20 | 36.14 | ± | 1.98 | 51.0 |
| Contr_l + PMI_L + PR_L | GCN | 68.08 | ± | 1.03 | 51.12 | ± | 0.91 | 37.37 | ± | 0.71 | 93.0 |
| Contr_l + PMI_L + PR_L | GIN | 63.86 | ± | 3.05 | 46.20 | ± | 3.68 | 34.29 | ± | 1.54 | 145.7 |
| Contr_l + PMI_L + PR_L | MPNN | 74.77 | ± | 1.57 | 54.65 | ± | 1.61 | 34.98 | ± | 1.99 | 80.7 |
| Contr_l + PMI_L + PR_L | PAGNN | 61.68 | ± | 3.69 | 45.67 | ± | 1.86 | 33.33 | ± | 0.00 | 166.3 |
| Contr_l + PMI_L + PR_L | SAGE | 49.57 | ± | 3.44 | 46.31 | ± | 2.85 | 34.18 | ± | 0.38 | 160.0 |
| Contr_l + PMI_L + PR_L + Triplet_L | ALL | 64.21 | ± | 1.47 | 51.66 | ± | 0.92 | 34.79 | ± | 1.25 | 120.0 |
| Contr_l + PMI_L + PR_L + Triplet_L | GAT | 76.24 | ± | 1.45 | 57.15 | ± | 1.02 | 40.23 | ± | 0.68 | 25.0 |
| Contr_l + PMI_L + PR_L + Triplet_L | GCN | 67.71 | ± | 1.83 | 53.37 | ± | 2.63 | 37.05 | ± | 0.76 | 89.7 |
| Contr_l + PMI_L + PR_L + Triplet_L | GIN | 67.65 | ± | 2.74 | 50.99 | ± | 1.53 | 37.06 | ± | 0.88 | 99.3 |
| Contr_l + PMI_L + PR_L + Triplet_L | MPNN | 73.47 | ± | 0.72 | 55.52 | ± | 1.75 | 37.42 | ± | 1.81 | 66.7 |
| Contr_l + PMI_L + PR_L + Triplet_L | PAGNN | 65.08 | ± | 2.31 | 50.42 | ± | 1.54 | 33.33 | ± | 0.00 | 139.0 |
| Contr_l + PMI_L + PR_L + Triplet_L | SAGE | 68.00 | ± | 3.12 | 51.15 | ± | 1.20 | 39.87 | ± | 2.10 | 77.7 |

Node Cls Recall (Sensitivity) Continued (↑)

| Loss Type | Model | CORA | | | Citeseer | | | Bitcoin Fraud Transaction | | | Average Rank |
|---|---|---|---|---|---|---|---|---|---|---|---|
| Contr_l + PR_L | ALL | 26.48 | ± | 2.16 | 33.97 | ± | 3.60 | 33.52 | ± | 0.15 | 187.7 |
| Contr_l + PR_L | GAT | 45.03 | ± | 18.22 | 51.95 | ± | 1.20 | 35.63 | ± | 1.24 | 131.7 |
| Contr_l + PR_L | GCN | 58.21 | ± | 2.77 | 49.39 | ± | 2.63 | 36.77 | ± | 1.34 | 128.7 |
| Contr_l + PR_L | GIN | 38.48 | ± | 4.08 | 44.13 | ± | 3.79 | 33.48 | ± | 0.06 | 177.3 |
| Contr_l + PR_L | MPNN | 39.77 | ± | 3.03 | 49.37 | ± | 2.30 | 34.11 | ± | 0.53 | 155.0 |
| Contr_l + PR_L | PAGNN | 39.55 | ± | 3.82 | 37.96 | ± | 2.06 | 33.33 | ± | 0.00 | 187.3 |
| Contr_l + PR_L | SAGE | 55.69 | ± | 14.57 | 47.94 | ± | 3.82 | 34.44 | ± | 1.20 | 148.7 |
| Contr_l + PR_L + Triplet_L | ALL | 54.61 | ± | 4.37 | 49.04 | ± | 2.34 | 34.93 | ± | 0.85 | 143.7 |
| Contr_l + PR_L + Triplet_L | GAT | 77.04 | ± | 3.90 | 58.03 | ± | 1.41 | 40.57 | ± | 0.43 | 14.0 |
| Contr_l + PR_L + Triplet_L | GCN | 66.97 | ± | 3.64 | 52.44 | ± | 1.58 | 36.30 | ± | 0.37 | 99.0 |
| Contr_l + PR_L + Triplet_L | GIN | 67.85 | ± | 1.84 | 49.17 | ± | 2.29 | 36.95 | ± | 2.47 | 108.0 |
| Contr_l + PR_L + Triplet_L | MPNN | 69.02 | ± | 2.97 | 54.56 | ± | 1.55 | 37.81 | ± | 1.53 | 74.7 |
| Contr_l + PR_L + Triplet_L | PAGNN | 57.34 | ± | 2.22 | 50.58 | ± | 6.37 | 33.33 | ± | 0.00 | 152.3 |
| Contr_l + PR_L + Triplet_L | SAGE | 70.36 | ± | 1.68 | 52.30 | ± | 1.01 | 38.40 | ± | 1.77 | 72.3 |
| Contr_l + Triplet_L | ALL | 73.34 | ± | 1.19 | 55.89 | ± | 1.18 | 33.99 | ± | 0.50 | 90.3 |
| Contr_l + Triplet_L | GAT | 77.40 | ± | 0.96 | 60.46 | ± | 1.23 | 38.19 | ± | 1.01 | 23.0 |

Node Cls Recall (Sensitivity) Continued (↑)

| Loss Type | Model | CORA | | | Citeseer | | | Bitcoin Fraud Transaction | | | Average Rank |
|---|---|---|---|---|---|---|---|---|---|---|---|
| Contr_l + Triplet_L | GCN | 74.20 | ± | 2.05 | 57.79 | ± | 1.95 | 37.45 | ± | 0.46 | 47.7 |
| Contr_l + Triplet_L | GIN | 75.30 | ± | 0.80 | 57.20 | ± | 1.98 | 39.59 | ± | 0.78 | 34.0 |
| Contr_l + Triplet_L | MPNN | 75.93 | ± | 1.86 | 57.27 | ± | 1.01 | 38.56 | ± | 1.95 | 33.7 |
| Contr_l + Triplet_L | PAGNN | 73.86 | ± | 3.43 | 57.47 | ± | 0.99 | 33.33 | ± | 0.00 | 88.3 |
| Contr_l + Triplet_L | SAGE | 76.13 | ± | 2.01 | 56.82 | ± | 1.27 | 37.65 | ± | 1.49 | 40.7 |
| CrossE_L | ALL | 14.29 | ± | 0.00 | 16.67 | ± | 0.00 | 33.33 | ± | 0.00 | 198.3 |
| CrossE_L | GAT | 14.29 | ± | 0.00 | 16.67 | ± | 0.00 | 33.33 | ± | 0.00 | 199.3 |
| CrossE_L | GCN | 14.29 | ± | 0.00 | 16.67 | ± | 0.00 | 33.33 | ± | 0.00 | 200.3 |
| CrossE_L | GIN | 14.29 | ± | 0.00 | 16.67 | ± | 0.00 | 33.33 | ± | 0.00 | 201.3 |
| CrossE_L | MPNN | 14.29 | ± | 0.00 | 16.67 | ± | 0.00 | 33.33 | ± | 0.00 | 202.3 |
| CrossE_L | PAGNN | 14.29 | ± | 0.00 | 16.67 | ± | 0.00 | 33.33 | ± | 0.00 | 203.3 |
| CrossE_L | SAGE | 14.29 | ± | 0.00 | 16.67 | ± | 0.00 | 33.33 | ± | 0.00 | 204.3 |
| CrossE_L + PMI_L | ALL | 66.57 | ± | 1.92 | 51.27 | ± | 2.27 | 38.68 | ± | 0.54 | 85.0 |
| CrossE_L + PMI_L | GAT | 74.02 | ± | 2.77 | 58.60 | ± | 0.76 | 40.32 | ± | 0.75 | 26.0 |
| CrossE_L + PMI_L | GCN | 68.44 | ± | 1.48 | 52.01 | ± | 1.91 | 37.51 | ± | 0.26 | 83.3 |
| CrossE_L + PMI_L | GIN | 65.20 | ± | 1.06 | 46.56 | ± | 2.27 | 35.82 | ± | 0.96 | 130.3 |

Node Cls Recall (Sensitivity) Continued (↑)

| Loss Type | Model | CORA | | Citeseer | | Bitcoin Fraud Transaction | | Average Rank |
|---|---|---|---|---|---|---|---|---|
| CrossE_L + PMI_L | MPNN | 73.49 ± 2.06 | | 57.28 ± 0.43 | | 41.13 ± 1.04 | | 31.3 |
| CrossE_L + PMI_L | PAGNN | 64.58 ± 2.54 | | 45.76 ± 2.95 | | 33.33 ± 0.00 | | 160.7 |
| CrossE_L + PMI_L | SAGE | 49.39 ± 3.76 | | 42.25 ± 1.35 | | 42.97 ± 2.36 | | 121.3 |
| CrossE_L + PMI_L + PR_L | ALL | 55.37 ± 2.22 | | 47.19 ± 1.51 | | 33.85 ± 0.23 | | 156.0 |
| CrossE_L + PMI_L + PR_L | GAT | 74.65 ± 1.42 | | 56.53 ± 1.58 | | 37.55 ± 3.42 | | 51.3 |
| CrossE_L + PMI_L + PR_L | GCN | 67.87 ± 1.85 | | 50.31 ± 1.46 | | 37.05 ± 0.75 | | 104.0 |
| CrossE_L + PMI_L + PR_L | GIN | 63.36 ± 1.34 | | 43.80 ± 3.41 | | 33.71 ± 0.48 | | 158.0 |
| CrossE_L + PMI_L + PR_L | MPNN | 73.92 ± 1.46 | | 55.14 ± 1.74 | | 37.53 ± 3.71 | | 63.7 |
| CrossE_L + PMI_L + PR_L | PAGNN | 64.60 ± 4.13 | | 45.02 ± 1.56 | | 33.33 ± 0.00 | | 162.3 |
| CrossE_L + PMI_L + PR_L | SAGE | 52.53 ± 2.94 | | 44.08 ± 1.67 | | 34.48 ± 1.15 | | 161.7 |
| CrossE_L + PMI_L + PR_L + Triplet_L | ALL | 63.50 ± 1.95 | | 51.40 ± 0.74 | | 36.72 ± 0.94 | | 113.3 |
| CrossE_L + PMI_L + PR_L + Triplet_L | GAT | 75.54 ± 1.62 | | 57.48 ± 0.56 | | 40.15 ± 0.69 | | 27.3 |
| CrossE_L + PMI_L + PR_L + Triplet_L | GCN | 66.28 ± 0.85 | | 50.90 ± 2.04 | | 36.76 ± 1.18 | | 105.7 |
| CrossE_L + PMI_L + PR_L + Triplet_L | GIN | 65.23 ± 2.02 | | 50.11 ± 2.08 | | 37.07 ± 1.06 | | 109.0 |
| CrossE_L + PMI_L + PR_L + Triplet_L | MPNN | 74.17 ± 2.27 | | 56.33 ± 1.24 | | 39.98 ± 2.11 | | 42.3 |
| CrossE_L + PMI_L + PR_L + Triplet_L | PAGNN | 64.91 ± 1.40 | | 48.43 ± 2.41 | | 33.33 ± 0.00 | | 152.0 |

Node Cls Recall (Sensitivity) Continued (↑)

| Loss Type | Model | CORA | | Citeseer | | Bitcoin Fraud Transaction | | Average Rank |
|---|---|---|---|---|---|---|---|---|
| CrossE_L + PMI_L + PR_L + Triplet_L | SAGE | 63.44 ± 2.73 | | 50.19 ± 1.56 | | 39.58 ± 1.86 | | 99.7 |
| CrossE_L + PMI_L + Triplet_L | ALL | 71.77 ± 2.55 | | 55.81 ± 1.02 | | 38.27 ± 0.71 | | 61.0 |
| CrossE_L + PMI_L + Triplet_L | GAT | 76.02 ± 1.15 | | 58.17 ± 0.84 | | 40.51 ± 0.63 | | 17.7 |
| CrossE_L + PMI_L + Triplet_L | GCN | 67.09 ± 1.20 | | 51.54 ± 1.34 | | 37.07 ± 0.76 | | 97.0 |
| CrossE_L + PMI_L + Triplet_L | GIN | 65.10 ± 1.34 | | 50.18 ± 1.77 | | 38.08 ± 1.20 | | 100.0 |
| CrossE_L + PMI_L + Triplet_L | MPNN | 75.78 ± 2.77 | | 56.35 ± 2.22 | | 40.64 ± 1.19 | | 29.0 |
| CrossE_L + PMI_L + Triplet_L | PAGNN | 64.46 ± 0.87 | | 46.61 ± 2.36 | | 33.33 ± 0.00 | | 159.3 |
| CrossE_L + PMI_L + Triplet_L | SAGE | 67.34 ± 1.91 | | 49.67 ± 1.92 | | 41.02 ± 1.83 | | 78.7 |
| CrossE_L + PR_L | ALL | 23.60 ± 0.48 | | 31.27 ± 2.65 | | 33.59 ± 0.03 | | 188.3 |
| CrossE_L + PR_L | GAT | 27.53 ± 4.34 | | 43.01 ± 1.63 | | 33.33 ± 0.00 | | 193.3 |
| CrossE_L + PR_L | GCN | 40.57 ± 6.96 | | 47.67 ± 1.93 | | 33.44 ± 0.12 | | 165.7 |
| CrossE_L + PR_L | GIN | 26.73 ± 2.66 | | 39.17 ± 2.65 | | 33.47 ± 0.09 | | 186.3 |
| CrossE_L + PR_L | MPNN | 37.97 ± 17.02 | | 47.35 ± 0.77 | | 33.74 ± 0.23 | | 164.7 |
| CrossE_L + PR_L | PAGNN | 33.37 ± 7.94 | | 36.74 ± 3.11 | | 33.33 ± 0.00 | | 195.0 |
| CrossE_L + PR_L | SAGE | 29.31 ± 3.83 | | 39.46 ± 3.56 | | 34.13 ± 0.75 | | 177.0 |
| CrossE_L + PR_L + Triplet_L | ALL | 51.28 ± 9.99 | | 42.82 ± 3.38 | | 35.17 ± 2.03 | | 162.3 |

Node Cls Recall (Sensitivity) Continued (↑)

| Loss Type | | | | Model | CORA | | Citeseer | | Bitcoin Fraud Transaction | | Average Rank |
|---|---|---|---|---|---|---|---|---|---|---|---|
| CrossE_L | + | PR_L | + Triplet_L | GAT | 69.10 ± 4.84 | | 56.76 ± 2.04 | | 38.46 ± 1.39 | | 57.7 |
| CrossE_L | + | PR_L | + Triplet_L | GCN | 65.72 ± 3.08 | | 51.78 ± 0.84 | | 36.61 ± 1.63 | | 102.7 |
| CrossE_L | + | PR_L | + Triplet_L | GIN | 62.15 ± 4.68 | | 48.34 ± 2.15 | | 33.54 ± 0.26 | | 150.7 |
| CrossE_L | + | PR_L | + Triplet_L | MPNN | 60.26 ± 3.12 | | 51.95 ± 1.57 | | 36.56 ± 1.29 | | 116.7 |
| CrossE_L | + | PR_L | + Triplet_L | PAGNN | 48.56 ± 5.32 | | 45.85 ± 1.91 | | 33.33 ± 0.00 | | 180.0 |
| CrossE_L | + | PR_L | + Triplet_L | SAGE | 70.63 ± 2.58 | | 52.72 ± 2.86 | | 38.78 ± 2.14 | | 69.7 |
| CrossE_L + Triplet_L | | | | ALL | 71.74 ± 1.33 | | 55.64 ± 1.20 | | 37.12 ± 1.98 | | 73.7 |
| CrossE_L + Triplet_L | | | | GAT | **78.98 ± 1.54** | | 59.20 ± 0.89 | | 40.04 ± 1.23 | | 13.7 |
| CrossE_L + Triplet_L | | | | GCN | 73.81 ± 1.64 | | 54.89 ± 1.39 | | 37.35 ± 0.51 | | 68.3 |
| CrossE_L + Triplet_L | | | | GIN | 74.52 ± 1.94 | | 54.01 ± 1.44 | | 40.24 ± 0.23 | | 48.0 |
| CrossE_L + Triplet_L | | | | MPNN | 76.60 ± 1.39 | | 57.98 ± 1.16 | | 40.05 ± 0.90 | | 20.3 |
| CrossE_L + Triplet_L | | | | PAGNN | 73.90 ± 1.32 | | 58.26 ± 1.45 | | 33.33 ± 0.00 | | 87.7 |
| CrossE_L + Triplet_L | | | | SAGE | 75.32 ± 0.99 | | 56.33 ± 2.02 | | 39.68 ± 0.79 | | 38.7 |
| PMI_L | | | | ALL | 66.25 ± 2.16 | | 52.23 ± 1.82 | | 39.38 ± 0.81 | | 78.7 |
| PMI_L | | | | GAT | 75.80 ± 2.09 | | 58.24 ± 1.94 | | 40.50 ± 0.71 | | 19.3 |
| PMI_L | | | | GCN | 68.16 ± 2.45 | | 51.15 ± 2.01 | | 36.57 ± 0.79 | | 101.3 |

Node Cls Recall (Sensitivity) Continued (↑)

| Loss Type | Model | CORA | | Citeseer | | Bitcoin Fraud Transaction | | Average Rank |
|---|---|---|---|---|---|---|---|---|
| PMI_L | GIN | 63.19 ± 2.99 | | 47.22 ± 2.18 | | 34.83 ± 0.60 | | 141.3 |
| PMI_L | MPNN | 73.86 ± 1.77 | | 55.97 ± 1.32 | | 40.07 ± 0.97 | | 46.0 |
| PMI_L | PAGNN | 61.93 ± 2.10 | | 44.40 ± 1.99 | | 33.33 ± 0.00 | | 174.0 |
| PMI_L | SAGE | 47.60 ± 1.50 | | 42.28 ± 1.89 | | 44.01 ± 0.90 | | 122.0 |
| PMI_L + PR_L | ALL | 56.21 ± 3.19 | | 47.12 ± 4.39 | | 33.75 ± 0.27 | | 156.0 |
| PMI_L + PR_L | GAT | 75.73 ± 1.75 | | 54.88 ± 2.70 | | 34.28 ± 0.70 | | 80.3 |
| PMI_L + PR_L | GCN | 69.04 ± 2.20 | | 52.74 ± 1.70 | | 37.40 ± 0.33 | | 82.0 |
| PMI_L + PR_L | GIN | 62.13 ± 3.94 | | 46.55 ± 2.93 | | 33.35 ± 0.05 | | 158.7 |
| PMI_L + PR_L | MPNN | 72.36 ± 1.86 | | 54.58 ± 1.67 | | 35.00 ± 2.95 | | 91.3 |
| PMI_L + PR_L | PAGNN | 64.31 ± 1.42 | | 44.23 ± 1.95 | | 33.33 ± 0.00 | | 168.7 |
| PMI_L + PR_L | SAGE | 51.76 ± 4.08 | | 44.27 ± 2.01 | | 34.16 ± 0.50 | | 163.3 |
| PMI_L + PR_L + Triplet_L | ALL | 63.10 ± 1.90 | | 50.88 ± 1.75 | | 35.54 ± 1.20 | | 126.7 |
| PMI_L + PR_L + Triplet_L | GAT | 75.39 ± 2.37 | | 57.90 ± 0.85 | | 40.03 ± 0.26 | | 28.0 |
| PMI_L + PR_L + Triplet_L | GCN | 68.95 ± 0.62 | | 51.84 ± 0.27 | | 37.26 ± 0.78 | | 87.3 |
| PMI_L + PR_L + Triplet_L | GIN | 62.60 ± 2.61 | | 50.44 ± 1.75 | | 36.50 ± 1.07 | | 122.7 |
| PMI_L + PR_L + Triplet_L | MPNN | 73.72 ± 2.58 | | 54.65 ± 1.62 | | 38.33 ± 2.37 | | 61.0 |

Node Cls Recall (Sensitivity) Continued (↑)

| Loss Type | Model | CORA | | Citeseer | | Bitcoin Fraud Transaction | | Average Rank |
|---|---|---|---|---|---|---|---|---|
| PMI_L + PR_L + Triplet_L | PAGNN | 64.58 ± 2.59 | | 48.63 ± 2.02 | | 33.33 ± 0.00 | | 155.0 |
| PMI_L + PR_L + Triplet_L | SAGE | 64.98 ± 5.05 | | 50.75 ± 1.12 | | 38.64 ± 2.25 | | 92.3 |
| PMI_L + Triplet_L | ALL | 72.65 ± 0.90 | | 54.84 ± 0.47 | | 38.29 ± 1.19 | | 64.3 |
| PMI_L + Triplet_L | GAT | 76.25 ± 1.95 | | 58.42 ± 1.14 | | 40.72 ± 0.61 | | 12.0 |
| PMI_L + Triplet_L | GCN | 69.64 ± 1.64 | | 50.39 ± 2.98 | | 37.41 ± 1.35 | | 93.0 |
| PMI_L + Triplet_L | GIN | 68.21 ± 4.01 | | 50.68 ± 1.45 | | 38.22 ± 1.06 | | 87.0 |
| PMI_L + Triplet_L | MPNN | 74.06 ± 1.66 | | 56.78 ± 1.28 | | 40.18 ± 1.00 | | 37.7 |
| PMI_L + Triplet_L | PAGNN | 62.35 ± 2.40 | | 45.58 ± 1.61 | | 33.33 ± 0.00 | | 172.7 |
| PMI_L + Triplet_L | SAGE | 63.30 ± 3.56 | | 51.90 ± 1.53 | | 43.54 ± 0.93 | | 77.0 |
| PR_L | ALL | 22.51 ± 0.22 | | 27.71 ± 0.83 | | 33.62 ± 0.05 | | 188.7 |
| PR_L | GAT | 24.85 ± 0.51 | | 43.55 ± 4.02 | | 33.33 ± 0.00 | | 197.0 |
| PR_L | GCN | 37.47 ± 5.34 | | 46.59 ± 0.97 | | 33.33 ± 0.00 | | 184.3 |
| PR_L | GIN | 28.40 ± 4.40 | | 39.29 ± 1.90 | | 33.56 ± 0.08 | | 183.0 |
| PR_L | MPNN | 40.53 ± 10.13 | | 47.64 ± 0.65 | | 35.66 ± 2.51 | | 151.3 |
| PR_L | PAGNN | 31.80 ± 8.43 | | 36.79 ± 1.07 | | 33.33 ± 0.00 | | 198.0 |
| PR_L | SAGE | 28.06 ± 3.34 | | 39.33 ± 1.17 | | 34.00 ± 0.73 | | 179.3 |

Node Cls Recall (Sensitivity) Continued (↑)

| Loss Type | Model | CORA | | Citeseer | | Bitcoin Fraud Transaction | | Average Rank |
|---|---|---|---|---|---|---|---|---|
| PR_L + Triplet_L | ALL | 25.94 | ± | 30.68 | ± | 33.62 | ± | 187.7 |
| | | 1.57 | | 0.58 | | 0.05 | | |
| PR_L + Triplet_L | GAT | 55.35 | ± | 51.74 | ± | 34.48 | ± | 132.7 |
| | | 19.68 | | 1.53 | | 1.12 | | |
| PR_L + Triplet_L | GCN | 59.20 | ± | 48.80 | ± | 35.71 | ± | 137.7 |
| | | 3.10 | | 2.62 | | 1.12 | | |
| PR_L + Triplet_L | GIN | 39.91 | ± | 45.81 | ± | 33.53 | ± | 171.3 |
| | | 3.00 | | 1.52 | | 0.12 | | |
| PR_L + Triplet_L | MPNN | 48.02 | ± | 48.51 | ± | 35.95 | ± | 145.0 |
| | | 9.67 | | 1.66 | | 1.91 | | |
| PR_L + Triplet_L | PAGNN | 39.12 | ± | 36.59 | ± | 33.33 | ± | 197.0 |
| | | 4.81 | | 3.15 | | 0.00 | | |
| PR_L + Triplet_L | SAGE | 29.97 | ± | 45.66 | ± | 36.13 | ± | 160.0 |
| | | 4.48 | | 5.06 | | 0.92 | | |
| Triplet_L | ALL | 72.77 | ± | 55.52 | ± | 37.17 | ± | 72.7 |
| | | 1.83 | | 0.65 | | 1.23 | | |
| Triplet_L | GAT | 78.70 | ± | 59.75 | ± | 40.92 | ± | 6.3 |
| | | 1.63 | | 0.82 | | 0.89 | | |
| Triplet_L | GCN | 73.94 | ± | 55.22 | ± | 37.48 | ± | 64.0 |
| | | 1.76 | | 2.06 | | 0.93 | | |
| Triplet_L | GIN | 73.46 | ± | 53.32 | ± | 40.92 | ± | 50.7 |
| | | 1.32 | | 0.82 | | 0.67 | | |
| Triplet_L | MPNN | 75.87 | ± | 57.19 | ± | 40.58 | ± | 23.7 |
| | | 0.87 | | 1.54 | | 1.01 | | |
| Triplet_L | PAGNN | 73.99 | ± | 56.80 | ± | 33.33 | ± | 99.0 |
| | | 4.00 | | 1.81 | | 0.00 | | |
| Triplet_L | SAGE | 75.05 | ± | 56.05 | ± | 39.57 | ± | 43.0 |
| | | 2.37 | | 1.59 | | 0.65 | | |

*1.1.2 Evaluation on Link prediction task.* In this section we keep results based on link prediction tasks.

Table 5. Lp Accuracy Performance (↑): Top-ranked results are highlighted in **1st**, second-ranked in **2nd**, and third-ranked in **3rd**.

| Loss Type | Model | CORA | | Citeseer | | Bitcoin Fraud Transaction | | Average Rank |
|-----------|-------|------|--|----------|--|---------------------------|--|--------------|
| Contr_l | ALL | 90.43 | ± | 94.31 | ± | 85.41 | ± | 88.0 |
| | | 0.43 | | 0.79 | | 0.99 | | |
| Contr_l | GAT | 96.28 | ± | 99.07 | ± | 91.21 | ± | 10.7 |
| | | 0.16 | | 0.06 | | 0.36 | | |
| Contr_l | GCN | 95.33 | ± | 98.31 | ± | 88.83 | ± | 34.3 |
| | | 0.17 | | 0.24 | | 0.62 | | |
| Contr_l | GIN | 92.79 | ± | 97.41 | ± | 86.18 | ± | 71.0 |
| | | 0.32 | | 0.15 | | 2.39 | | |
| Contr_l | MPNN | 92.89 | ± | 97.33 | ± | 85.25 | ± | 74.3 |
| | | 0.36 | | 0.29 | | 0.53 | | |
| Contr_l | PAGNN | 91.07 | ± | 93.93 | ± | 57.30 | ± | 118.0 |
| | | 0.74 | | 0.73 | | 0.58 | | |
| Contr_l | SAGE | 93.92 | ± | 97.64 | ± | 82.91 | ± | 74.7 |
| | | 0.19 | | 0.26 | | 1.41 | | |
| Contr_l + CrossE_L | ALL | 90.38 | ± | 94.66 | ± | 83.89 | ± | 90.3 |
| | | 0.65 | | 0.39 | | 1.82 | | |
| Contr_l + CrossE_L | GAT | 96.04 | ± | 99.04 | ± | 92.47 | ± | 12.3 |
| | | 0.22 | | 0.08 | | 1.76 | | |
| Contr_l + CrossE_L | GCN | 95.39 | ± | 98.31 | ± | 89.06 | ± | 32.7 |
| | | 0.35 | | 0.19 | | 0.43 | | |
| Contr_l + CrossE_L | GIN | 92.29 | ± | 96.87 | ± | 84.64 | ± | 79.3 |
| | | 0.55 | | 0.26 | | 2.17 | | |
| Contr_l + CrossE_L | MPNN | 92.83 | ± | 97.48 | ± | 86.77 | ± | 68.3 |
| | | 0.69 | | 0.13 | | 0.78 | | |
| Contr_l + CrossE_L | PAGNN | 90.62 | ± | 93.63 | ± | 56.88 | ± | 121.3 |
| | | 0.75 | | 0.38 | | 0.18 | | |
| Contr_l + CrossE_L | SAGE | 94.12 | ± | 97.85 | ± | 83.89 | ± | 71.0 |
| | | 0.45 | | 0.27 | | 2.19 | | |
| Contr_l + CrossE_L + PMI_L | ALL | 84.81 | ± | 83.37 | ± | 77.61 | ± | 137.0 |
| | | 2.74 | | 3.72 | | 0.46 | | |
| Contr_l + CrossE_L + PMI_L | GAT | 96.69 | ± | 98.89 | ± | 85.60 | ± | 31.7 |
| | | 0.23 | | 0.11 | | 1.39 | | |

Lp Accuracy Continued (↑)

| Loss Type | Model | CORA | | Citeseer | | Bitcoin Fraud Transaction | | Average Rank |
|---|---|---|---|---|---|---|---|---|
| Contr_l + CrossE_L + PMI_L | GCN | 95.61 0.52 | ± | 98.02 0.19 | ± | 87.83 0.25 | ± | 46.7 |
| Contr_l + CrossE_L + PMI_L | GIN | 84.74 2.26 | ± | 85.43 1.28 | ± | 63.79 2.49 | ± | 142.3 |
| Contr_l + CrossE_L + PMI_L | MPNN | 95.23 0.35 | ± | 98.14 0.30 | ± | 86.91 0.67 | ± | 50.0 |
| Contr_l + CrossE_L + PMI_L | PAGNN | 59.80 3.85 | ± | 73.85 0.87 | ± | 53.67 0.31 | ± | 189.7 |
| Contr_l + CrossE_L + PMI_L | SAGE | 77.63 1.28 | ± | 83.15 2.81 | ± | 57.53 4.74 | ± | 158.3 |
| Contr_l + CrossE_L + PMI_L + PR_L | ALL | 67.65 3.82 | ± | 79.81 0.34 | ± | 67.30 0.64 | ± | 161.0 |
| Contr_l + CrossE_L + PMI_L + PR_L | GAT | 95.97 0.81 | ± | 98.60 0.13 | ± | 70.78 12.78 | ± | 59.7 |
| Contr_l + CrossE_L + PMI_L + PR_L | GCN | 95.55 0.32 | ± | 98.12 0.22 | ± | 87.67 0.80 | ± | 45.0 |
| Contr_l + CrossE_L + PMI_L + PR_L | GIN | 83.61 1.76 | ± | 82.51 1.10 | ± | 54.56 0.64 | ± | 161.3 |
| Contr_l + CrossE_L + PMI_L + PR_L | MPNN | 94.45 0.30 | ± | 94.37 3.18 | ± | 60.17 15.15 | ± | 105.3 |
| Contr_l + CrossE_L + PMI_L + PR_L | PAGNN | 65.51 4.56 | ± | 73.98 0.99 | ± | 46.41 0.31 | ± | 189.7 |
| Contr_l + CrossE_L + PMI_L + PR_L | SAGE | 76.72 2.23 | ± | 81.62 3.01 | ± | 55.48 0.98 | ± | 165.0 |
| Contr_l + CrossE_L + PMI_L + PR_L + Triplet_L | ALL | 86.98 0.56 | ± | 87.03 0.80 | ± | 77.97 0.78 | ± | 126.7 |
| Contr_l + CrossE_L + PMI_L + PR_L + Triplet_L | GAT | 96.29 0.38 | ± | 98.44 0.59 | ± | 85.39 4.54 | ± | 37.0 |
| Contr_l + CrossE_L + PMI_L + PR_L + Triplet_L | GCN | 96.09 0.12 | ± | 98.25 0.12 | ± | 88.66 0.26 | ± | 30.3 |
| Contr_l + CrossE_L + PMI_L + PR_L + Triplet_L | GIN | 84.43 1.96 | ± | 88.41 0.87 | ± | 74.45 1.70 | ± | 132.3 |

Lp Accuracy Continued (↑)

| Loss Type | Model | CORA | | Citeseer | | Bitcoin Fraud Transaction | | Average Rank |
|---|---|---|---|---|---|---|---|---|
| Contr_l + CrossE_L + PMI_L + PR_L + Triplet_L | MPNN | 94.56 0.87 | ± | 95.16 1.58 | ± | 84.97 5.39 | ± | 78.7 |
| Contr_l + CrossE_L + PMI_L + PR_L + Triplet_L | PAGNN | 68.06 2.99 | ± | 74.93 0.45 | ± | 53.79 0.33 | ± | 182.0 |
| Contr_l + CrossE_L + PMI_L + PR_L + Triplet_L | SAGE | 88.18 2.40 | ± | 95.04 0.60 | ± | 57.03 0.76 | ± | 122.3 |
| Contr_l + CrossE_L + PMI_L + Triplet_L | ALL | 91.47 0.48 | ± | 96.89 0.23 | ± | 85.61 0.67 | ± | 76.7 |
| Contr_l + CrossE_L + PMI_L + Triplet_L | GAT | 97.03 0.48 | ± | 98.92 0.08 | ± | 88.67 0.47 | ± | 19.0 |
| Contr_l + CrossE_L + PMI_L + Triplet_L | GCN | 96.13 0.31 | ± | 98.18 0.19 | ± | 88.73 0.64 | ± | 30.0 |
| Contr_l + CrossE_L + PMI_L + Triplet_L | GIN | 85.64 2.42 | ± | 88.97 0.73 | ± | 74.36 2.85 | ± | 128.7 |
| Contr_l + CrossE_L + PMI_L + Triplet_L | MPNN | 95.19 0.78 | ± | 97.91 0.50 | ± | 87.50 1.07 | ± | 53.0 |
| Contr_l + CrossE_L + PMI_L + Triplet_L | PAGNN | 69.29 1.87 | ± | 74.22 1.03 | ± | 54.63 0.97 | ± | 180.0 |
| Contr_l + CrossE_L + PMI_L + Triplet_L | SAGE | 89.59 0.74 | ± | 94.17 0.79 | ± | 78.95 1.62 | ± | 105.7 |
| Contr_l + CrossE_L + PR_L | ALL | 54.87 4.76 | ± | 61.21 9.39 | ± | 81.58 0.38 | ± | 165.7 |
| Contr_l + CrossE_L + PR_L | GAT | 89.69 4.08 | ± | 92.42 2.77 | ± | 83.56 1.36 | ± | 99.0 |
| Contr_l + CrossE_L + PR_L | GCN | 89.89 0.84 | ± | 92.13 0.89 | ± | 82.94 0.54 | ± | 100.0 |
| Contr_l + CrossE_L + PR_L | GIN | 69.39 7.23 | ± | 84.31 1.54 | ± | 63.52 16.19 | ± | 156.3 |
| Contr_l + CrossE_L + PR_L | MPNN | 78.84 1.70 | ± | 88.56 0.63 | ± | 66.10 14.53 | ± | 142.3 |
| Contr_l + CrossE_L + PR_L | PAGNN | 52.29 0.31 | ± | 63.03 9.21 | ± | 54.59 2.25 | ± | 194.3 |

Lp Accuracy Continued (↑)

| Loss Type | Model | CORA | | Citeseer | | Bitcoin Fraud Transaction | | Average Rank |
|-----------|-------|------|---|----------|---|---------------------------|---|--------------|
| Contr_l + CrossE_L + PR_L | SAGE | 86.61 | ± | 88.38 | ± | 54.21 | ± | 147.3 |
| | | 8.85 | | 13.08 | | 1.73 | | |
| Contr_l + CrossE_L + PR_L + Triplet_L | ALL | 83.08 | ± | 89.37 | ± | 80.95 | ± | 123.3 |
| | | 1.24 | | 0.95 | | 2.00 | | |
| Contr_l + CrossE_L + PR_L + Triplet_L | GAT | 95.33 | ± | 97.33 | ± | 88.77 | ± | 50.3 |
| | | 0.57 | | 1.24 | | 1.35 | | |
| Contr_l + CrossE_L + PR_L + Triplet_L | GCN | 93.72 | ± | 95.73 | ± | 86.71 | ± | 73.0 |
| | | 0.78 | | 0.76 | | 2.06 | | |
| Contr_l + CrossE_L + PR_L + Triplet_L | GIN | 87.34 | ± | 92.09 | ± | 80.26 | ± | 110.7 |
| | | 0.82 | | 0.74 | | 7.24 | | |
| Contr_l + CrossE_L + PR_L + Triplet_L | MPNN | 87.67 | ± | 93.15 | ± | 81.11 | ± | 106.3 |
| | | 1.10 | | 0.33 | | 4.59 | | |
| Contr_l + CrossE_L + PR_L + Triplet_L | PAGNN | 67.60 | ± | 83.40 | ± | 54.09 | ± | 173.7 |
| | | 4.81 | | 4.52 | | 1.79 | | |
| Contr_l + CrossE_L + PR_L + Triplet_L | SAGE | 93.53 | ± | 96.10 | ± | 74.03 | ± | 95.3 |
| | | 0.52 | | 0.67 | | 3.01 | | |
| Contr_l + CrossE_L + Triplet_L | ALL | 93.18 | ± | 96.81 | ± | 88.79 | ± | 60.7 |
| | | 0.37 | | 0.23 | | 1.11 | | |
| Contr_l + CrossE_L + Triplet_L | GAT | **97.40** | ± | **99.32** | ± | **94.61** | ± | **2.7** |
| | | **0.37** | | **0.10** | | **0.85** | | |
| Contr_l + CrossE_L + Triplet_L | GCN | 96.23 | ± | 98.65 | ± | 90.40 | ± | 18.0 |
| | | 0.18 | | 0.15 | | 0.46 | | |
| Contr_l + CrossE_L + Triplet_L | GIN | 94.77 | ± | 97.92 | ± | 87.59 | ± | 55.3 |
| | | 0.55 | | 0.41 | | 2.03 | | |
| Contr_l + CrossE_L + Triplet_L | MPNN | 94.65 | ± | 98.26 | ± | 89.32 | ± | 39.3 |
| | | 0.44 | | 0.15 | | 0.75 | | |
| Contr_l + CrossE_L + Triplet_L | PAGNN | 89.49 | ± | 95.93 | ± | 57.38 | ± | 117.3 |
| | | 5.06 | | 0.27 | | 0.28 | | |
| Contr_l + CrossE_L + Triplet_L | SAGE | 95.77 | ± | 98.39 | ± | 89.47 | ± | 26.3 |
| | | 0.44 | | 0.16 | | 0.88 | | |
| Contr_l + PMI_L | ALL | 85.15 | ± | 91.36 | ± | 79.62 | ± | 116.7 |
| | | 2.71 | | 1.49 | | 1.69 | | |

Lp Accuracy Continued (↑)

| Loss Type | Model | CORA | | Citeseer | | Bitcoin Fraud Transaction | | Average Rank |
|---|---|---|---|---|---|---|---|---|
| Contr_l + PMI_L | GAT | 97.00 | ± | 98.83 | ± | 85.74 | ± | 31.7 |
| | | 0.26 | | 0.13 | | 1.25 | | |
| Contr_l + PMI_L | GCN | 95.66 | ± | 98.03 | ± | 88.21 | ± | 43.0 |
| | | 0.35 | | 0.14 | | 0.68 | | |
| Contr_l + PMI_L | GIN | 83.92 | ± | 86.78 | ± | 65.17 | ± | 141.3 |
| | | 2.76 | | 1.07 | | 2.44 | | |
| Contr_l + PMI_L | MPNN | 95.13 | ± | 98.01 | ± | 86.98 | ± | 53.3 |
| | | 0.29 | | 0.31 | | 0.47 | | |
| Contr_l + PMI_L | PAGNN | 62.72 | ± | 73.66 | ± | 53.63 | ± | 189.0 |
| | | 5.06 | | 0.09 | | 0.51 | | |
| Contr_l + PMI_L | SAGE | 79.20 | ± | 89.81 | ± | 56.04 | ± | 150.0 |
| | | 2.41 | | 2.12 | | 1.43 | | |
| Contr_l + PMI_L + PR_L | ALL | 72.70 | ± | 79.41 | ± | 69.44 | ± | 157.0 |
| | | 2.49 | | 0.45 | | 0.79 | | |
| Contr_l + PMI_L + PR_L | GAT | 95.71 | ± | 96.51 | ± | 61.31 | ± | 89.0 |
| | | 0.35 | | 2.28 | | 12.94 | | |
| Contr_l + PMI_L + PR_L | GCN | 95.92 | ± | 98.11 | ± | 88.12 | ± | 38.3 |
| | | 0.26 | | 0.14 | | 0.40 | | |
| Contr_l + PMI_L + PR_L | GIN | 84.98 | ± | 83.49 | ± | 55.82 | ± | 153.3 |
| | | 2.94 | | 2.18 | | 1.22 | | |
| Contr_l + PMI_L + PR_L | MPNN | 95.04 | ± | 92.69 | ± | 64.25 | ± | 102.3 |
| | | 0.39 | | 2.94 | | 15.71 | | |
| Contr_l + PMI_L + PR_L | PAGNN | 63.32 | ± | 73.39 | ± | 46.60 | ± | 191.3 |
| | | 4.21 | | 0.86 | | 0.35 | | |
| Contr_l + PMI_L + PR_L | SAGE | 78.31 | ± | 87.42 | ± | 56.40 | ± | 154.3 |
| | | 1.27 | | 1.52 | | 0.78 | | |
| Contr_l + PMI_L + PR_L + Triplet_L | ALL | 88.15 | ± | 90.75 | ± | 78.30 | ± | 116.3 |
| | | 1.03 | | 0.80 | | 1.36 | | |
| Contr_l + PMI_L + PR_L + Triplet_L | GAT | 95.91 | ± | 98.29 | ± | 83.69 | ± | 46.7 |
| | | 0.39 | | 0.32 | | 4.22 | | |
| Contr_l + PMI_L + PR_L + Triplet_L | GCN | 95.77 | ± | 98.12 | ± | 88.64 | ± | 38.7 |
| | | 0.30 | | 0.17 | | 0.57 | | |

Lp Accuracy Continued (↑)

| Loss Type | Model | CORA | | Citeseer | | Bitcoin Fraud Transaction | | Average Rank |
|---|---|---|---|---|---|---|---|---|
| Contr_l + PMI_L + PR_L + Triplet_L | GIN | 88.61 ± 1.06 | | 90.90 ± 1.37 | | 77.18 ± 1.14 | | 116.7 |
| Contr_l + PMI_L + PR_L + Triplet_L | MPNN | 94.44 ± 0.28 | | 94.55 ± 0.57 | | 79.85 ± 3.99 | | 89.7 |
| Contr_l + PMI_L + PR_L + Triplet_L | PAGNN | 70.67 ± 0.83 | | 79.00 ± 2.67 | | 54.92 ± 1.01 | | 173.0 |
| Contr_l + PMI_L + PR_L + Triplet_L | SAGE | 91.92 ± 0.76 | | 96.58 ± 0.43 | | 62.17 ± 4.09 | | 104.7 |
| Contr_l + PR_L | ALL | 61.27 ± 5.93 | | 55.88 ± 9.31 | | 81.67 ± 0.59 | | 163.0 |
| Contr_l + PR_L | GAT | 89.80 ± 3.07 | | 91.55 ± 0.86 | | 84.13 ± 1.10 | | 98.7 |
| Contr_l + PR_L | GCN | 89.74 ± 0.45 | | 92.87 ± 1.15 | | 85.57 ± 2.75 | | 93.3 |
| Contr_l + PR_L | GIN | 68.35 ± 8.23 | | 81.76 ± 2.35 | | 55.58 ± 19.13 | | 169.7 |
| Contr_l + PR_L | MPNN | 79.09 ± 2.04 | | 89.34 ± 0.87 | | 54.99 ± 3.10 | | 153.7 |
| Contr_l + PR_L | PAGNN | 52.08 ± 0.25 | | 68.06 ± 2.94 | | 52.54 ± 5.13 | | 197.7 |
| Contr_l + PR_L | SAGE | 86.36 ± 9.49 | | 90.35 ± 10.77 | | 55.13 ± 1.20 | | 143.3 |
| Contr_l + PR_L + Triplet_L | ALL | 78.08 ± 7.59 | | 85.82 ± 5.21 | | 81.15 ± 0.35 | | 133.3 |
| Contr_l + PR_L + Triplet_L | GAT | 94.78 ± 1.99 | | 97.35 ± 0.93 | | 88.83 ± 2.02 | | 52.7 |
| Contr_l + PR_L + Triplet_L | GCN | 92.95 ± 0.59 | | 95.51 ± 0.85 | | 85.78 ± 1.85 | | 77.7 |
| Contr_l + PR_L + Triplet_L | GIN | 87.63 ± 1.17 | | 91.03 ± 1.26 | | 78.43 ± 4.06 | | 116.0 |
| Contr_l + PR_L + Triplet_L | MPNN | 88.55 ± 2.92 | | 92.35 ± 1.15 | | 77.22 ± 4.04 | | 113.3 |

Lp Accuracy Continued (↑)

| Loss Type | Model | CORA | | Citeseer | | Bitcoin Fraud Transaction | | Average Rank |
|---|---|---|---|---|---|---|---|---|
| Contr_l + PR_L + Triplet_L | PAGNN | 61.69 ± 4.24 | | 83.78 ± 6.21 | | 53.09 ± 0.21 | | 179.3 |
| Contr_l + PR_L + Triplet_L | SAGE | 93.09 ± 0.42 | | 96.63 ± 0.43 | | 78.45 ± 2.81 | | 89.3 |
| Contr_l + Triplet_L | ALL | 93.60 ± 0.37 | | 97.03 ± 0.33 | | 90.25 ± 0.57 | | 54.3 |
| Contr_l + Triplet_L | GAT | 97.22 ± 0.38 | | 99.28 ± 0.11 | | 94.81 ± 0.97 | | 3.7 |
| Contr_l + Triplet_L | GCN | 96.29 ± 0.16 | | 98.70 ± 0.15 | | 90.28 ± 0.49 | | 17.3 |
| Contr_l + Triplet_L | GIN | 94.79 ± 0.65 | | 97.92 ± 0.45 | | 86.35 ± 1.83 | | 59.0 |
| Contr_l + Triplet_L | MPNN | 94.85 ± 0.49 | | 98.37 ± 0.18 | | 89.71 ± 0.65 | | 34.7 |
| Contr_l + Triplet_L | PAGNN | 89.77 ± 6.47 | | 96.17 ± 0.29 | | 57.59 ± 0.72 | | 114.0 |
| Contr_l + Triplet_L | SAGE | 96.03 ± 0.16 | | 98.40 ± 0.36 | | 88.97 ± 2.58 | | 24.0 |
| CrossE_L | ALL | 81.05 ± 12.97 | | 79.46 ± 4.60 | | 79.43 ± 3.81 | | 141.7 |
| CrossE_L | GAT | 80.19 ± 17.05 | | 86.12 ± 21.57 | | 76.18 ± 22.97 | | 139.0 |
| CrossE_L | GCN | 51.35 ± 0.00 | | 47.65 ± 0.00 | | 34.64 ± 0.00 | | 209.3 |
| CrossE_L | GIN | 51.37 ± 0.02 | | 47.66 ± 0.02 | | 34.64 ± 0.00 | | 209.0 |
| CrossE_L | MPNN | 85.05 ± 1.78 | | 84.88 ± 5.64 | | 34.64 ± 0.00 | | 164.7 |
| CrossE_L | PAGNN | 59.25 ± 3.66 | | 67.98 ± 3.10 | | 34.69 ± 0.04 | | 200.0 |
| CrossE_L | SAGE | 69.56 ± 5.25 | | 61.43 ± 11.03 | | 34.65 ± 0.01 | | 193.7 |

Lp Accuracy Continued (↑)

| Loss Type | Model | CORA | | Citeseer | | Bitcoin Fraud Transaction | | Average Rank |
|---|---|---|---|---|---|---|---|---|
| CrossE_L + PMI_L | ALL | 85.06 | ± 0.83 | 83.67 | ± 1.08 | 79.09 | ± 1.82 | 132.3 |
| CrossE_L + PMI_L | GAT | 97.17 | ± 0.14 | 99.03 | ± 0.06 | 84.99 | ± 0.46 | 29.7 |
| CrossE_L + PMI_L | GCN | 95.80 | ± 0.43 | 98.18 | ± 0.27 | 87.86 | ± 0.53 | 37.7 |
| CrossE_L + PMI_L | GIN | 84.55 | ± 2.62 | 84.18 | ± 0.64 | 62.19 | ± 1.82 | 145.3 |
| CrossE_L + PMI_L | MPNN | 94.95 | ± 0.36 | 98.01 | ± 0.33 | 87.24 | ± 1.10 | 54.7 |
| CrossE_L + PMI_L | PAGNN | 63.05 | ± 4.89 | 73.12 | ± 0.40 | 50.58 | ± 4.12 | 192.0 |
| CrossE_L + PMI_L | SAGE | 73.58 | ± 1.44 | 76.31 | ± 0.45 | 43.64 | ± 3.10 | 181.0 |
| CrossE_L + PMI_L + PR_L | ALL | 66.92 | ± 0.84 | 79.70 | ± 0.83 | 61.27 | ± 5.64 | 166.0 |
| CrossE_L + PMI_L + PR_L | GAT | 96.44 | ± 0.37 | 95.29 | ± 3.02 | 70.82 | ± 16.16 | 78.0 |
| CrossE_L + PMI_L + PR_L | GCN | 95.81 | ± 0.55 | 98.08 | ± 0.26 | 87.38 | ± 1.34 | 42.3 |
| CrossE_L + PMI_L + PR_L | GIN | 81.82 | ± 1.45 | 81.07 | ± 3.68 | 57.70 | ± 3.86 | 155.3 |
| CrossE_L + PMI_L + PR_L | MPNN | 94.58 | ± 1.07 | 94.60 | ± 3.97 | 71.66 | ± 17.60 | 96.3 |
| CrossE_L + PMI_L + PR_L | PAGNN | 66.19 | ± 5.94 | 73.71 | ± 0.59 | 46.19 | ± 2.10 | 190.3 |
| CrossE_L + PMI_L + PR_L | SAGE | 73.86 | ± 0.72 | 77.25 | ± 2.30 | 55.82 | ± 0.55 | 169.7 |
| CrossE_L + PMI_L + PR_L + Triplet_L | ALL | 86.54 | ± 0.10 | 85.03 | ± 0.65 | 77.99 | ± 1.19 | 129.7 |
| CrossE_L + PMI_L + PR_L + Triplet_L | GAT | 96.22 | ± 0.31 | 98.27 | ± 0.35 | 83.06 | ± 6.29 | 45.7 |

Lp Accuracy Continued (↑)

| Loss Type | Model | CORA | | Citeseer | | Bitcoin Fraud Transaction | | Average Rank |
|---|---|---|---|---|---|---|---|---|
| CrossE_L + PMI_L + PR_L + Triplet_L | GCN | 95.79 ± 0.39 | | 98.28 ± 0.13 | | 88.19 ± 0.82 | | 34.7 |
| CrossE_L + PMI_L + PR_L + Triplet_L | GIN | 85.48 ± 1.33 | | 87.59 ± 1.26 | | 74.16 ± 1.22 | | 131.3 |
| CrossE_L + PMI_L + PR_L + Triplet_L | MPNN | 95.05 ± 0.33 | | 95.39 ± 2.33 | | 85.01 ± 6.18 | | 73.0 |
| CrossE_L + PMI_L + PR_L + Triplet_L | PAGNN | 71.76 ± 0.66 | | 75.42 ± 1.17 | | 54.03 ± 0.53 | | 178.0 |
| CrossE_L + PMI_L + PR_L + Triplet_L | SAGE | 88.09 ± 3.92 | | 93.85 ± 1.08 | | 56.66 ± 0.58 | | 127.3 |
| CrossE_L + PMI_L + Triplet_L | ALL | 93.03 ± 0.56 | | 97.74 ± 0.30 | | 87.33 ± 0.76 | | 64.0 |
| CrossE_L + PMI_L + Triplet_L | GAT | 97.12 ± 0.23 | | 98.94 ± 0.10 | | 89.82 ± 0.90 | | 12.3 |
| CrossE_L + PMI_L + Triplet_L | GCN | 95.64 ± 0.46 | | 98.21 ± 0.09 | | 88.78 ± 0.41 | | 35.7 |
| CrossE_L + PMI_L + Triplet_L | GIN | 87.55 ± 1.30 | | 90.74 ± 0.46 | | 79.27 ± 0.86 | | 115.7 |
| CrossE_L + PMI_L + Triplet_L | MPNN | 94.47 ± 0.65 | | 97.90 ± 0.25 | | 88.71 ± 0.31 | | 53.3 |
| CrossE_L + PMI_L + Triplet_L | PAGNN | 71.74 ± 1.17 | | 74.64 ± 1.18 | | 55.15 ± 1.43 | | 175.7 |
| CrossE_L + PMI_L + Triplet_L | SAGE | 91.70 ± 0.63 | | 95.93 ± 0.86 | | 82.71 ± 1.48 | | 87.7 |
| CrossE_L + PR_L | ALL | 55.97 ± 6.50 | | 49.97 ± 1.42 | | 81.97 ± 0.23 | | 165.7 |
| CrossE_L + PR_L | GAT | 88.81 ± 1.97 | | 91.73 ± 0.82 | | 78.29 ± 11.54 | | 112.7 |
| CrossE_L + PR_L | GCN | 86.92 ± 1.27 | | 90.62 ± 1.53 | | 58.06 ± 21.49 | | 133.3 |
| CrossE_L + PR_L | GIN | 60.76 ± 10.30 | | 78.99 ± 3.61 | | 70.35 ± 0.63 | | 166.7 |

Lp Accuracy Continued (↑)

| Loss Type | Model | CORA | | Citeseer | | Bitcoin Fraud Transaction | | Average Rank |
|---|---|---|---|---|---|---|---|---|
| CrossE_L + PR_L | MPNN | 82.62 6.26 | ± | 87.67 0.79 | ± | 45.54 3.62 | ± | 160.7 |
| CrossE_L + PR_L | PAGNN | 51.85 0.20 | ± | 53.02 7.66 | ± | 43.24 0.42 | ± | 204.7 |
| CrossE_L + PR_L | SAGE | 66.09 2.23 | ± | 63.37 9.02 | ± | 39.44 6.62 | ± | 196.7 |
| CrossE_L + PR_L + Triplet_L | ALL | 75.05 9.45 | ± | 73.11 6.74 | ± | 79.16 2.88 | ± | 154.0 |
| CrossE_L + PR_L + Triplet_L | GAT | 91.47 2.00 | ± | 96.80 1.87 | ± | 85.13 2.09 | ± | 80.3 |
| CrossE_L + PR_L + Triplet_L | GCN | 92.05 1.14 | ± | 94.27 0.90 | ± | 86.33 1.31 | ± | 82.0 |
| CrossE_L + PR_L + Triplet_L | GIN | 84.45 3.74 | ± | 90.50 1.14 | ± | 70.73 0.62 | ± | 131.7 |
| CrossE_L + PR_L + Triplet_L | MPNN | 82.08 2.04 | ± | 90.77 0.66 | ± | 70.64 8.15 | ± | 132.7 |
| CrossE_L + PR_L + Triplet_L | PAGNN | 56.41 5.41 | ± | 76.28 3.64 | ± | 53.36 0.39 | ± | 188.3 |
| CrossE_L + PR_L + Triplet_L | SAGE | 93.77 0.88 | ± | 96.59 0.87 | ± | 64.61 10.55 | ± | 97.7 |
| CrossE_L + Triplet_L | ALL | 95.49 0.45 | ± | 98.18 0.19 | ± | 90.15 1.03 | ± | 33.3 |
| CrossE_L + Triplet_L | GAT | 98.26 0.26 | ± | 99.51 0.08 | ± | 93.34 1.24 | ± | 2.3 |
| CrossE_L + Triplet_L | GCN | 97.14 0.38 | ± | 98.90 0.19 | ± | 90.80 0.44 | ± | 10.0 |
| CrossE_L + Triplet_L | GIN | 95.90 0.60 | ± | 97.73 0.27 | ± | 86.80 1.48 | ± | 49.7 |
| CrossE_L + Triplet_L | MPNN | 96.45 0.46 | ± | 98.95 0.29 | ± | 90.86 0.60 | ± | 10.7 |
| CrossE_L + Triplet_L | PAGNN | 91.03 6.05 | ± | 97.26 0.46 | ± | 57.01 0.16 | ± | 109.3 |

Lp Accuracy Continued (↑)

| Loss Type | Model | CORA | | Citeseer | | Bitcoin Fraud Transaction | | Average Rank |
|---|---|---|---|---|---|---|---|---|
| CrossE_L + Triplet_L | SAGE | 97.40 ± 0.29 | | 98.84 ± 0.14 | | 88.85 ± 0.80 | | 14.3 |
| PMI_L | ALL | 81.36 ± 0.59 | | 84.67 ± 1.20 | | 80.49 ± 0.39 | | 132.3 |
| PMI_L | GAT | 97.31 ± 0.14 | | 98.89 ± 0.21 | | 85.25 ± 0.59 | | 30.3 |
| PMI_L | GCN | 95.80 ± 0.28 | | 97.89 ± 0.36 | | 87.91 ± 0.84 | | 45.0 |
| PMI_L | GIN | 81.31 ± 2.11 | | 84.28 ± 1.30 | | 59.83 ± 4.07 | | 151.0 |
| PMI_L | MPNN | 95.08 ± 0.21 | | 98.15 ± 0.22 | | 87.29 ± 0.91 | | 49.3 |
| PMI_L | PAGNN | 65.23 ± 3.83 | | 73.14 ± 0.93 | | 53.51 ± 0.79 | | 189.0 |
| PMI_L | SAGE | 73.46 ± 1.77 | | 75.83 ± 0.63 | | 46.12 ± 4.96 | | 181.7 |
| PMI_L + PR_L | ALL | 68.29 ± 2.68 | | 78.60 ± 3.32 | | 64.33 ± 0.59 | | 164.0 |
| PMI_L + PR_L | GAT | 95.44 ± 0.70 | | 93.84 ± 5.30 | | 59.87 ± 11.72 | | 100.3 |
| PMI_L + PR_L | GCN | 95.66 ± 0.86 | | 98.20 ± 0.11 | | 87.84 ± 0.97 | | 40.7 |
| PMI_L + PR_L | GIN | 82.37 ± 1.95 | | 82.40 ± 3.75 | | 56.19 ± 1.63 | | 157.7 |
| PMI_L + PR_L | MPNN | 94.75 ± 0.58 | | 90.42 ± 1.03 | | 60.15 ± 14.97 | | 113.7 |
| PMI_L + PR_L | PAGNN | 66.26 ± 1.07 | | 73.06 ± 1.17 | | 42.91 ± 5.19 | | 194.0 |
| PMI_L + PR_L | SAGE | 74.37 ± 2.23 | | 77.82 ± 1.41 | | 54.85 ± 1.07 | | 171.7 |
| PMI_L + PR_L + Triplet_L | ALL | 86.66 ± 0.63 | | 87.32 ± 0.89 | | 79.18 ± 1.18 | | 123.7 |

Lp Accuracy Continued (↑)

| Loss Type | Model | CORA | | Citeseer | | Bitcoin Fraud Transaction | | Average Rank |
|---|---|---|---|---|---|---|---|---|
| PMI_L + PR_L + Triplet_L | GAT | 95.93 | ± | 98.08 | ± | 79.44 | ± | 59.7 |
| | | 0.21 | | 0.43 | | 4.82 | | |
| PMI_L + PR_L + Triplet_L | GCN | 95.74 | ± | 98.01 | ± | 88.13 | ± | 43.7 |
| | | 0.34 | | 0.14 | | 0.38 | | |
| PMI_L + PR_L + Triplet_L | GIN | 86.52 | ± | 87.56 | ± | 75.30 | ± | 129.3 |
| | | 1.03 | | 0.73 | | 1.29 | | |
| PMI_L + PR_L + Triplet_L | MPNN | 94.72 | ± | 93.56 | ± | 82.50 | ± | 87.0 |
| | | 0.44 | | 0.63 | | 5.25 | | |
| PMI_L + PR_L + Triplet_L | PAGNN | 70.59 | ± | 76.13 | ± | 54.74 | ± | 176.3 |
| | | 2.53 | | 0.61 | | 0.96 | | |
| PMI_L + PR_L + Triplet_L | SAGE | 90.70 | ± | 95.33 | ± | 56.98 | ± | 116.0 |
| | | 3.25 | | 0.70 | | 0.36 | | |
| PMI_L + Triplet_L | ALL | 92.71 | ± | 97.38 | ± | 87.10 | ± | 68.7 |
| | | 0.57 | | 0.38 | | 0.51 | | |
| PMI_L + Triplet_L | GAT | 97.13 | ± | 98.97 | ± | 88.90 | ± | 13.3 |
| | | 0.23 | | 0.10 | | 0.80 | | |
| PMI_L + Triplet_L | GCN | 95.72 | ± | 98.17 | ± | 88.47 | ± | 38.0 |
| | | 0.27 | | 0.19 | | 0.33 | | |
| PMI_L + Triplet_L | GIN | 86.04 | ± | 89.19 | ± | 76.79 | ± | 126.7 |
| | | 2.19 | | 1.17 | | 1.60 | | |
| PMI_L + Triplet_L | MPNN | 94.99 | ± | 97.95 | ± | 88.26 | ± | 50.0 |
| | | 0.51 | | 0.30 | | 0.65 | | |
| PMI_L + Triplet_L | PAGNN | 69.44 | ± | 74.84 | ± | 55.21 | ± | 176.3 |
| | | 1.07 | | 0.79 | | 1.21 | | |
| PMI_L + Triplet_L | SAGE | 90.12 | ± | 95.54 | ± | 79.86 | ± | 96.0 |
| | | 1.74 | | 0.54 | | 2.08 | | |
| PR_L | ALL | 51.39 | ± | 48.33 | ± | 82.75 | ± | 167.7 |
| | | 0.04 | | 0.71 | | 0.75 | | |
| PR_L | GAT | 90.15 | ± | 91.18 | ± | 83.17 | ± | 100.0 |
| | | 1.51 | | 0.60 | | 1.57 | | |
| PR_L | GCN | 86.59 | ± | 90.68 | ± | 86.03 | ± | 103.7 |
| | | 0.97 | | 0.80 | | 0.91 | | |

Lp Accuracy Continued (↑)

| Loss Type | Model | CORA | | | Citeseer | | | Bitcoin Fraud Transaction | | | Average Rank |
|---|---|---|---|---|---|---|---|---|---|---|---|
| PR_L | GIN | 56.09 | ± | 8.69 | 79.52 | ± | 3.05 | 70.81 | ± | 0.65 | 165.3 |
| PR_L | MPNN | 80.16 | ± | 6.54 | 87.30 | ± | 1.05 | 58.26 | ± | 15.52 | 149.0 |
| PR_L | PAGNN | 51.88 | ± | 0.21 | 48.94 | ± | 0.44 | 43.39 | ± | 0.22 | 205.0 |
| PR_L | SAGE | 67.24 | ± | 1.46 | 68.37 | ± | 1.94 | 46.23 | ± | 2.07 | 191.3 |
| PR_L + Triplet_L | ALL | 54.12 | ± | 4.96 | 50.49 | ± | 1.04 | 81.77 | ± | 0.58 | 166.3 |
| PR_L + Triplet_L | GAT | 90.08 | ± | 4.26 | 91.19 | ± | 1.26 | 83.01 | ± | 0.72 | 101.0 |
| PR_L + Triplet_L | GCN | 89.54 | ± | 0.41 | 92.14 | ± | 0.78 | 82.55 | ± | 0.49 | 103.0 |
| PR_L + Triplet_L | GIN | 60.77 | ± | 9.79 | 80.23 | ± | 2.16 | 68.87 | ± | 0.74 | 164.3 |
| PR_L + Triplet_L | MPNN | 79.89 | ± | 3.80 | 88.08 | ± | 0.94 | 67.99 | ± | 14.84 | 141.7 |
| PR_L + Triplet_L | PAGNN | 52.08 | ± | 0.22 | 67.04 | ± | 3.00 | 53.33 | ± | 1.09 | 197.7 |
| PR_L + Triplet_L | SAGE | 71.93 | ± | 2.44 | 79.87 | ± | 14.88 | 53.31 | ± | 0.66 | 175.0 |
| Triplet_L | ALL | 95.25 | ± | 0.45 | 98.16 | ± | 0.16 | 89.16 | ± | 0.84 | 37.3 |
| Triplet_L | GAT | 98.71 | ± | 0.25 | 99.63 | ± | 0.06 | 92.65 | ± | 0.58 | 2.0 |
| Triplet_L | GCN | 97.12 | ± | 0.14 | 98.95 | ± | 0.17 | 90.67 | ± | 0.47 | 10.7 |
| Triplet_L | GIN | 96.43 | ± | 0.71 | 98.08 | ± | 0.37 | 86.24 | ± | 0.93 | 42.0 |
| Triplet_L | MPNN | 95.68 | ± | 0.82 | 98.76 | ± | 0.23 | 90.75 | ± | 0.31 | 23.3 |

Lp Accuracy Continued (↑)

| Loss Type | Model | CORA | | Citeseer | | Bitcoin Fraud Transaction | | Average Rank |
|-----------|-------|------|---|----------|---|---------------------------|---|--------------|
| Triplet_L | PAGNN | 91.11 | ± | 97.44 | ± | 57.39 | ± | 105.3 |
|           |       | 6.32 |   | 0.23 |   | 0.38 |   |       |
| Triplet_L | SAGE | 97.19 | ± | 98.81 | ± | 88.84 | ± | 16.3 |
|           |      | 0.38 |   | 0.27 |   | 0.48 |   |      |

Table 6. Lp Aupr Performance (↑): Top-ranked results are highlighted in **1st**, second-ranked in **2nd**, and third-ranked in **3rd**.

| Loss Type | Model | CORA | | Citeseer | | Bitcoin Fraud Transaction | | Average Rank |
|-----------|-------|------|---|----------|---|---------------------------|---|--------------|
| Contr_l | ALL | 96.03 | ± | 98.33 | ± | 87.88 | ± | 84.7 |
|         |     | 0.45 |   | 0.26 |   | 1.08 |   |      |
| Contr_l | GAT | 99.21 | ± | 99.85 | ± | 93.41 | ± | 17.0 |
|         |     | 0.06 |   | 0.03 |   | 0.32 |   |      |
| Contr_l | GCN | 99.02 | ± | 99.76 | ± | 91.87 | ± | 28.7 |
|         |     | 0.04 |   | 0.05 |   | 0.18 |   |      |
| Contr_l | GIN | 96.92 | ± | 99.17 | ± | 88.06 | ± | 74.0 |
|         |     | 0.23 |   | 0.08 |   | 3.26 |   |      |
| Contr_l | MPNN | 97.31 | ± | 99.20 | ± | 87.44 | ± | 71.7 |
|         |      | 0.32 |   | 0.14 |   | 0.73 |   |      |
| Contr_l | PAGNN | 95.28 | ± | 97.49 | ± | 36.07 | ± | 128.7 |
|         |       | 0.42 |   | 0.48 |   | 0.27 |   |       |
| Contr_l | SAGE | 98.39 | ± | 99.62 | ± | 84.83 | ± | 60.3 |
|         |      | 0.06 |   | 0.05 |   | 1.70 |   |      |
| Contr_l + CrossE_L | ALL | 96.08 | ± | 98.41 | ± | 86.76 | ± | 85.3 |
|                    |     | 0.26 |   | 0.20 |   | 2.04 |   |      |
| Contr_l + CrossE_L | GAT | 99.23 | ± | 99.86 | ± | 94.85 | ± | 11.3 |
|                    |     | 0.06 |   | 0.03 |   | 1.78 |   |      |
| Contr_l + CrossE_L | GCN | 99.04 | ± | 99.75 | ± | 91.69 | ± | 30.3 |
|                    |     | 0.10 |   | 0.04 |   | 0.33 |   |      |

Lp Aupr Continued (↑)

| Loss Type | Model | CORA | | Citeseer | | Bitcoin Fraud Transaction | | Average Rank |
|---|---|---|---|---|---|---|---|---|
| Contr_l + CrossE_L | GIN | 96.32 | ± | 98.92 | ± | 86.52 | ± | 80.0 |
| | | 0.39 | | 0.15 | | 2.44 | | |
| Contr_l + CrossE_L | MPNN | 97.26 | ± | 99.25 | ± | 89.39 | ± | 66.0 |
| | | 0.42 | | 0.13 | | 0.50 | | |
| Contr_l + CrossE_L | PAGNN | 95.02 | ± | 97.37 | ± | 36.18 | ± | 127.3 |
| | | 0.61 | | 0.40 | | 0.25 | | |
| Contr_l + CrossE_L | SAGE | 98.53 | ± | 99.66 | ± | 85.20 | ± | 58.3 |
| | | 0.21 | | 0.08 | | 2.65 | | |
| Contr_l + CrossE_L + PMI_L | ALL | 92.66 | ± | 92.20 | ± | 78.00 | ± | 127.3 |
| | | 1.35 | | 2.46 | | 1.19 | | |
| Contr_l + CrossE_L + PMI_L | GAT | 99.41 | ± | 99.84 | ± | 87.84 | ± | 32.0 |
| | | 0.05 | | 0.02 | | 1.43 | | |
| Contr_l + CrossE_L + PMI_L | GCN | 99.15 | ± | 99.73 | ± | 90.72 | ± | 35.0 |
| | | 0.18 | | 0.05 | | 0.32 | | |
| Contr_l + CrossE_L + PMI_L | GIN | 92.37 | ± | 93.74 | ± | 55.46 | ± | 140.7 |
| | | 1.42 | | 0.81 | | 2.60 | | |
| Contr_l + CrossE_L + PMI_L | MPNN | 97.49 | ± | 99.31 | ± | 89.13 | ± | 62.0 |
| | | 0.22 | | 0.11 | | 0.80 | | |
| Contr_l + CrossE_L + PMI_L | PAGNN | 76.27 | ± | 83.52 | ± | 34.89 | ± | 192.0 |
| | | 1.51 | | 0.57 | | 0.33 | | |
| Contr_l + CrossE_L + PMI_L | SAGE | 87.76 | ± | 91.02 | ± | 51.50 | ± | 154.7 |
| | | 0.83 | | 2.27 | | 6.91 | | |
| Contr_l + CrossE_L + PMI_L + PR_L | ALL | 80.40 | ± | 89.40 | ± | 57.66 | ± | 161.3 |
| | | 2.18 | | 0.54 | | 0.32 | | |
| Contr_l + CrossE_L + PMI_L + PR_L | GAT | 99.05 | ± | 99.73 | ± | 73.04 | ± | 63.7 |
| | | 0.35 | | 0.07 | | 7.78 | | |
| Contr_l + CrossE_L + PMI_L + PR_L | GCN | 99.14 | ± | 99.72 | ± | 90.38 | ± | 37.7 |
| | | 0.14 | | 0.05 | | 0.98 | | |
| Contr_l + CrossE_L + PMI_L + PR_L | GIN | 90.89 | ± | 92.10 | ± | 50.00 | ± | 150.3 |
| | | 0.88 | | 0.54 | | 3.10 | | |
| Contr_l + CrossE_L + PMI_L + PR_L | MPNN | 97.23 | ± | 97.91 | ± | 68.47 | ± | 103.7 |
| | | 0.36 | | 1.24 | | 11.16 | | |

Continued on next page

Lp Aupr Continued ($\uparrow$)

| Loss Type | Model | CORA | | Citeseer | | Bitcoin Fraud Transaction | | Average Rank |
|---|---|---|---|---|---|---|---|---|
| Contr_l + CrossE_L + PMI_L + PR_L | PAGNN | 78.69 ± 1.78 | | 82.61 ± 0.69 | | 33.16 ± 0.25 | | 193.7 |
| Contr_l + CrossE_L + PMI_L + PR_L | SAGE | 86.88 ± 1.76 | | 89.59 ± 2.59 | | 38.54 ± 2.51 | | 162.0 |
| Contr_l + CrossE_L + PMI_L + PR_L + Triplet_L | ALL | 94.00 ± 0.19 | | 94.81 ± 0.42 | | 75.07 ± 0.76 | | 119.0 |
| Contr_l + CrossE_L + PMI_L + PR_L + Triplet_L | GAT | 99.09 ± 0.21 | | 99.74 ± 0.10 | | 86.74 ± 5.33 | | 46.3 |
| Contr_l + CrossE_L + PMI_L + PR_L + Triplet_L | GCN | 99.30 ± 0.07 | | 99.72 ± 0.05 | | 91.60 ± 0.20 | | 26.3 |
| Contr_l + CrossE_L + PMI_L + PR_L + Triplet_L | GIN | 91.84 ± 1.13 | | 95.43 ± 0.43 | | 64.49 ± 2.73 | | 132.3 |
| Contr_l + CrossE_L + PMI_L + PR_L + Triplet_L | MPNN | 96.99 ± 0.64 | | 98.22 ± 0.60 | | 85.41 ± 7.62 | | 85.7 |
| Contr_l + CrossE_L + PMI_L + PR_L + Triplet_L | PAGNN | 79.61 ± 1.56 | | 82.92 ± 0.83 | | 35.22 ± 0.11 | | 186.7 |
| Contr_l + CrossE_L + PMI_L + PR_L + Triplet_L | SAGE | 94.79 ± 1.59 | | 98.45 ± 0.13 | | 48.78 ± 1.81 | | 117.7 |
| Contr_l + CrossE_L + PMI_L + Triplet_L | ALL | 96.34 ± 0.26 | | 99.10 ± 0.15 | | 87.50 ± 0.65 | | 77.0 |
| Contr_l + CrossE_L + PMI_L + Triplet_L | GAT | 99.53 ± 0.10 | | 99.86 ± 0.04 | | 91.29 ± 0.50 | | 17.7 |
| Contr_l + CrossE_L + PMI_L + Triplet_L | GCN | 99.29 ± 0.05 | | 99.76 ± 0.03 | | 91.69 ± 0.65 | | 22.0 |
| Contr_l + CrossE_L + PMI_L + Triplet_L | GIN | 92.48 ± 1.61 | | 95.74 ± 0.20 | | 68.31 ± 4.72 | | 128.0 |
| Contr_l + CrossE_L + PMI_L + Triplet_L | MPNN | 97.62 ± 0.35 | | 99.22 ± 0.09 | | 90.32 ± 1.15 | | 59.7 |
| Contr_l + CrossE_L + PMI_L + Triplet_L | PAGNN | 79.58 ± 1.03 | | 83.29 ± 0.32 | | 35.11 ± 0.24 | | 187.3 |
| Contr_l + CrossE_L + PMI_L + Triplet_L | SAGE | 95.79 ± 0.39 | | 98.12 ± 0.35 | | 77.56 ± 2.22 | | 99.3 |

Continued on next page

Lp Aupr Continued (↑)

| Loss Type | Model | CORA | | Citeseer | | Bitcoin Fraud Transaction | | Average Rank |
|---|---|---|---|---|---|---|---|---|
| Contr_l + CrossE_L + PR_L | ALL | 69.66 ± 3.66 | | 71.36 ± 8.42 | | 72.74 ± 0.69 | | 175.3 |
| Contr_l + CrossE_L + PR_L | GAT | 94.60 ± 2.45 | | 94.63 ± 2.26 | | 82.20 ± 2.93 | | 110.0 |
| Contr_l + CrossE_L + PR_L | GCN | 95.98 ± 0.35 | | 96.16 ± 0.37 | | 83.12 ± 0.60 | | 99.0 |
| Contr_l + CrossE_L + PR_L | GIN | 82.83 ± 1.82 | | 91.78 ± 1.27 | | 44.82 ± 1.58 | | 161.3 |
| Contr_l + CrossE_L + PR_L | MPNN | 86.51 ± 1.48 | | 92.22 ± 0.68 | | 69.50 ± 11.39 | | 145.0 |
| Contr_l + CrossE_L + PR_L | PAGNN | 72.28 ± 3.31 | | 72.73 ± 4.59 | | 35.63 ± 0.59 | | 195.0 |
| Contr_l + CrossE_L + PR_L | SAGE | 93.39 ± 6.39 | | 93.25 ± 9.70 | | 36.54 ± 1.50 | | 146.0 |
| Contr_l + CrossE_L + PR_L + Triplet_L | ALL | 91.04 ± 0.88 | | 96.04 ± 0.43 | | 83.24 ± 1.42 | | 114.7 |
| Contr_l + CrossE_L + PR_L + Triplet_L | GAT | 98.03 ± 0.21 | | 98.71 ± 0.24 | | 90.93 ± 2.45 | | 59.3 |
| Contr_l + CrossE_L + PR_L + Triplet_L | GCN | 98.13 ± 0.39 | | 98.58 ± 0.21 | | 88.96 ± 2.63 | | 66.7 |
| Contr_l + CrossE_L + PR_L + Triplet_L | GIN | 93.93 ± 0.37 | | 96.60 ± 0.47 | | 78.80 ± 13.19 | | 109.3 |
| Contr_l + CrossE_L + PR_L + Triplet_L | MPNN | 93.98 ± 0.44 | | 97.17 ± 0.17 | | 80.61 ± 5.36 | | 105.0 |
| Contr_l + CrossE_L + PR_L + Triplet_L | PAGNN | 81.99 ± 2.89 | | 89.27 ± 5.04 | | 36.11 ± 0.21 | | 170.3 |
| Contr_l + CrossE_L + PR_L + Triplet_L | SAGE | 97.42 ± 0.15 | | 98.43 ± 0.19 | | 71.86 ± 4.20 | | 95.0 |
| Contr_l + CrossE_L + Triplet_L | ALL | 97.58 ± 0.23 | | 99.26 ± 0.06 | | 91.76 ± 1.16 | | 50.7 |
| Contr_l + CrossE_L + Triplet_L | GAT | 99.55 ± 0.08 | | 99.90 ± 0.02 | | 96.81 ± 0.75 | | 4.7 |

Lp Aupr Continued (↑)

| Loss Type | Model | CORA | | Citeseer | | Bitcoin Fraud Transaction | | Average Rank |
|---|---|---|---|---|---|---|---|---|
| Contr_l + CrossE_L + Triplet_L | GCN | 99.32 ± 0.06 | | 99.79 ± 0.05 | | 93.52 ± 0.34 | | 15.7 |
| Contr_l + CrossE_L + Triplet_L | GIN | 98.16 ± 0.22 | | 99.44 ± 0.19 | | 90.06 ± 2.08 | | 53.0 |
| Contr_l + CrossE_L + Triplet_L | MPNN | 98.19 ± 0.38 | | 99.58 ± 0.06 | | 92.15 ± 0.84 | | 41.7 |
| Contr_l + CrossE_L + Triplet_L | PAGNN | 93.98 ± 3.46 | | 98.63 ± 0.25 | | 36.14 ± 0.15 | | 125.7 |
| Contr_l + CrossE_L + Triplet_L | SAGE | 99.13 ± 0.12 | | 99.78 ± 0.04 | | 91.68 ± 1.00 | | 27.3 |
| Contr_l + PMI_L | ALL | 92.99 ± 1.40 | | 97.01 ± 0.65 | | 80.93 ± 1.50 | | 109.0 |
| Contr_l + PMI_L | GAT | 99.52 ± 0.06 | | 99.83 ± 0.03 | | 88.05 ± 1.41 | | 31.7 |
| Contr_l + PMI_L | GCN | 99.16 ± 0.11 | | 99.70 ± 0.05 | | 91.30 ± 0.72 | | 33.7 |
| Contr_l + PMI_L | GIN | 91.48 ± 1.77 | | 94.36 ± 0.76 | | 57.27 ± 1.77 | | 139.7 |
| Contr_l + PMI_L | MPNN | 97.56 ± 0.30 | | 99.30 ± 0.17 | | 89.33 ± 0.51 | | 61.0 |
| Contr_l + PMI_L | PAGNN | 77.48 ± 1.74 | | 82.82 ± 0.86 | | 34.91 ± 0.27 | | 193.0 |
| Contr_l + PMI_L | SAGE | 88.86 ± 1.87 | | 95.48 ± 1.33 | | 52.17 ± 1.45 | | 141.3 |
| Contr_l + PMI_L + PR_L | ALL | 82.41 ± 1.38 | | 89.25 ± 0.47 | | 58.86 ± 0.51 | | 158.0 |
| Contr_l + PMI_L + PR_L | GAT | 98.98 ± 0.16 | | 99.24 ± 0.67 | | 68.21 ± 7.69 | | 81.3 |
| Contr_l + PMI_L + PR_L | GCN | 99.21 ± 0.06 | | 99.71 ± 0.06 | | 91.21 ± 0.46 | | 32.0 |
| Contr_l + PMI_L + PR_L | GIN | 91.89 ± 2.03 | | 92.53 ± 1.00 | | 50.59 ± 2.36 | | 146.0 |

Lp Aupr Continued (↑)

| Loss Type | Model | CORA | | Citeseer | | Bitcoin Fraud Transaction | | Average Rank |
|-----------|-------|------|---|----------|---|---------------------------|---|--------------|
| Contr_l + PMI_L + PR_L | MPNN | 97.32 | ± | 97.33 | ± | 68.95 | ± | 103.0 |
| | | 0.35 | | 1.10 | | 10.57 | | |
| Contr_l + PMI_L + PR_L | PAGNN | 76.96 | ± | 82.23 | ± | 33.38 | ± | 196.7 |
| | | 1.44 | | 0.72 | | 0.24 | | |
| Contr_l + PMI_L + PR_L | SAGE | 88.05 | ± | 93.74 | ± | 38.62 | ± | 153.3 |
| | | 1.15 | | 1.16 | | 1.70 | | |
| Contr_l + PMI_L + PR_L + Triplet_L | ALL | 94.83 | ± | 96.79 | ± | 79.09 | ± | 104.0 |
| | | 0.49 | | 0.32 | | 1.32 | | |
| Contr_l + PMI_L + PR_L + Triplet_L | GAT | 99.09 | ± | 99.78 | ± | 85.19 | ± | 47.0 |
| | | 0.19 | | 0.05 | | 3.67 | | |
| Contr_l + PMI_L + PR_L + Triplet_L | GCN | 99.18 | ± | 99.68 | ± | 91.53 | ± | 33.0 |
| | | 0.13 | | 0.08 | | 0.85 | | |
| Contr_l + PMI_L + PR_L + Triplet_L | GIN | 93.79 | ± | 96.51 | ± | 71.12 | ± | 119.7 |
| | | 0.52 | | 0.63 | | 1.01 | | |
| Contr_l + PMI_L + PR_L + Triplet_L | MPNN | 97.08 | ± | 98.22 | ± | 79.12 | ± | 92.0 |
| | | 0.52 | | 0.16 | | 5.23 | | |
| Contr_l + PMI_L + PR_L + Triplet_L | PAGNN | 80.78 | ± | 84.89 | ± | 35.70 | ± | 178.3 |
| | | 0.61 | | 1.51 | | 0.33 | | |
| Contr_l + PMI_L + PR_L + Triplet_L | SAGE | 97.44 | ± | 99.26 | ± | 57.40 | ± | 94.3 |
| | | 0.52 | | 0.18 | | 2.71 | | |
| Contr_l + PR_L | ALL | 73.58 | ± | 70.07 | ± | 72.63 | ± | 173.3 |
| | | 2.57 | | 6.81 | | 1.42 | | |
| Contr_l + PR_L | GAT | 94.95 | ± | 94.24 | ± | 82.85 | ± | 108.0 |
| | | 2.05 | | 0.86 | | 3.15 | | |
| Contr_l + PR_L | GCN | 95.80 | ± | 96.70 | ± | 87.60 | ± | 91.3 |
| | | 0.33 | | 0.76 | | 3.39 | | |
| Contr_l + PR_L | GIN | 81.59 | ± | 90.15 | ± | 43.86 | ± | 165.3 |
| | | 2.59 | | 1.68 | | 2.31 | | |
| Contr_l + PR_L | MPNN | 85.87 | ± | 92.69 | ± | 62.80 | ± | 147.7 |
| | | 1.59 | | 1.23 | | 0.98 | | |
| Contr_l + PR_L | PAGNN | 70.86 | ± | 73.94 | ± | 35.55 | ± | 195.3 |
| | | 2.03 | | 2.26 | | 0.31 | | |

Lp Aupr Continued (↑)

| Loss Type | Model | CORA | | Citeseer | | Bitcoin Fraud Transaction | | Average Rank |
|---|---|---|---|---|---|---|---|---|
| Contr_l + PR_L | SAGE | 93.23 | ± 6.83 | 94.28 | ± 8.56 | 36.84 | ± 2.11 | 144.0 |
| Contr_l + PR_L + Triplet_L | ALL | 87.17 | ± 5.87 | 93.96 | ± 2.58 | 83.40 | ± 0.77 | 124.3 |
| Contr_l + PR_L + Triplet_L | GAT | 97.78 | ± 0.54 | 98.72 | ± 0.10 | 90.26 | ± 3.04 | 63.0 |
| Contr_l + PR_L + Triplet_L | GCN | 97.80 | ± 0.23 | 98.52 | ± 0.23 | 87.36 | ± 2.53 | 72.0 |
| Contr_l + PR_L + Triplet_L | GIN | 93.97 | ± 0.49 | 96.14 | ± 1.05 | 72.96 | ± 10.74 | 117.7 |
| Contr_l + PR_L + Triplet_L | MPNN | 94.29 | ± 1.57 | 96.61 | ± 0.58 | 76.04 | ± 4.63 | 110.3 |
| Contr_l + PR_L + Triplet_L | PAGNN | 80.14 | ± 1.26 | 89.33 | ± 6.20 | 36.08 | ± 0.26 | 174.3 |
| Contr_l + PR_L + Triplet_L | SAGE | 97.29 | ± 0.14 | 98.52 | ± 0.12 | 77.49 | ± 3.07 | 89.0 |
| Contr_l + Triplet_L | ALL | 97.84 | ± 0.19 | 99.31 | ± 0.13 | 93.25 | ± 0.43 | 44.7 |
| Contr_l + Triplet_L | GAT | 99.53 | ± 0.06 | 99.89 | ± 0.02 | 97.04 | ± 0.67 | 6.0 |
| Contr_l + Triplet_L | GCN | 99.34 | ± 0.03 | 99.84 | ± 0.04 | 93.61 | ± 0.36 | 14.3 |
| Contr_l + Triplet_L | GIN | 98.13 | ± 0.54 | 99.42 | ± 0.17 | 88.87 | ± 2.07 | 57.7 |
| Contr_l + Triplet_L | MPNN | 98.24 | ± 0.30 | 99.59 | ± 0.06 | 92.41 | ± 0.78 | 40.3 |
| Contr_l + Triplet_L | PAGNN | 94.54 | ± 3.97 | 98.79 | ± 0.08 | 36.11 | ± 0.43 | 123.0 |
| Contr_l + Triplet_L | SAGE | 99.25 | ± 0.06 | 99.79 | ± 0.03 | 91.41 | ± 2.96 | 23.3 |
| CrossE_L | ALL | 90.28 | ± 8.81 | 87.21 | ± 5.82 | 71.58 | ± 5.88 | 146.3 |

Lp Aupr Continued (↑)

| Loss Type | Model | CORA | | Citeseer | | Bitcoin Fraud Transaction | | Average Rank |
|-----------|-------|------|---|----------|---|---------------------------|---|--------------|
| CrossE_L | GAT | 83.80 | ± | 88.36 | ± | 78.01 | ± | 144.0 |
| | | 17.17 | | 21.43 | | 22.37 | | |
| CrossE_L | GCN | 59.70 | ± | 56.02 | ± | 42.46 | ± | 194.7 |
| | | 3.33 | | 1.67 | | 2.39 | | |
| CrossE_L | GIN | 46.87 | ± | 42.05 | ± | 30.13 | ± | 209.7 |
| | | 2.68 | | 0.69 | | 1.36 | | |
| CrossE_L | MPNN | 92.51 | ± | 93.14 | ± | 37.30 | ± | 147.7 |
| | | 1.19 | | 3.82 | | 1.87 | | |
| CrossE_L | PAGNN | 77.23 | ± | 74.22 | ± | 29.87 | ± | 200.3 |
| | | 1.18 | | 3.34 | | 0.60 | | |
| CrossE_L | SAGE | 72.23 | ± | 69.45 | ± | 35.04 | ± | 200.3 |
| | | 5.89 | | 10.00 | | 0.71 | | |
| CrossE_L + PMI_L | ALL | 92.02 | ± | 92.13 | ± | 76.96 | ± | 130.7 |
| | | 0.64 | | 0.70 | | 2.06 | | |
| CrossE_L + PMI_L | GAT | 99.58 | ± | 99.88 | ± | 86.31 | ± | 28.0 |
| | | 0.03 | | 0.03 | | 1.81 | | |
| CrossE_L + PMI_L | GCN | 99.23 | ± | 99.73 | ± | 90.89 | ± | 31.3 |
| | | 0.11 | | 0.07 | | 0.42 | | |
| CrossE_L + PMI_L | GIN | 91.79 | ± | 92.63 | ± | 53.84 | ± | 145.0 |
| | | 1.95 | | 0.29 | | 1.67 | | |
| CrossE_L + PMI_L | MPNN | 97.51 | ± | 99.36 | ± | 89.48 | ± | 59.0 |
| | | 0.14 | | 0.12 | | 1.21 | | |
| CrossE_L + PMI_L | PAGNN | 77.52 | ± | 82.86 | ± | 34.48 | ± | 193.0 |
| | | 2.41 | | 0.30 | | 0.78 | | |
| CrossE_L + PMI_L | SAGE | 84.46 | ± | 85.31 | ± | 35.67 | ± | 174.3 |
| | | 0.90 | | 0.33 | | 1.27 | | |
| CrossE_L + PMI_L + PR_L | ALL | 80.53 | ± | 89.20 | ± | 55.49 | ± | 163.3 |
| | | 1.07 | | 0.22 | | 0.46 | | |
| CrossE_L + PMI_L + PR_L | GAT | 99.22 | ± | 98.85 | ± | 73.50 | ± | 72.0 |
| | | 0.20 | | 0.84 | | 8.61 | | |
| CrossE_L + PMI_L + PR_L | GCN | 99.19 | ± | 99.71 | ± | 90.18 | ± | 37.3 |
| | | 0.18 | | 0.08 | | 1.45 | | |

Lp Aupr Continued (↑)

| Loss Type | Model | CORA | | Citeseer | | Bitcoin Fraud Transaction | | Average Rank |
|---|---|---|---|---|---|---|---|---|
| CrossE_L + PMI_L + PR_L | GIN | 89.69 ± 1.48 | | 91.41 ± 2.01 | | 47.54 ± 2.33 | | 153.3 |
| CrossE_L + PMI_L + PR_L | MPNN | 97.26 ± 0.31 | | 98.00 ± 1.55 | | 74.83 ± 13.11 | | 96.7 |
| CrossE_L + PMI_L + PR_L | PAGNN | 78.88 ± 2.04 | | 82.44 ± 0.61 | | 33.20 ± 0.40 | | 193.7 |
| CrossE_L + PMI_L + PR_L | SAGE | 84.58 ± 0.79 | | 86.13 ± 1.64 | | 35.08 ± 0.73 | | 176.7 |
| CrossE_L + PMI_L + PR_L + Triplet_L | ALL | 93.29 ± 0.15 | | 93.89 ± 0.52 | | 75.16 ± 0.51 | | 124.7 |
| CrossE_L + PMI_L + PR_L + Triplet_L | GAT | 99.15 ± 0.13 | | 99.69 ± 0.04 | | 83.89 ± 7.71 | | 51.7 |
| CrossE_L + PMI_L + PR_L + Triplet_L | GCN | 99.20 ± 0.13 | | 99.76 ± 0.05 | | 91.42 ± 0.82 | | 27.0 |
| CrossE_L + PMI_L + PR_L + Triplet_L | GIN | 92.00 ± 0.96 | | 95.14 ± 0.63 | | 65.21 ± 1.57 | | 131.3 |
| CrossE_L + PMI_L + PR_L + Triplet_L | MPNN | 97.50 ± 0.29 | | 98.37 ± 0.76 | | 85.31 ± 8.18 | | 79.7 |
| CrossE_L + PMI_L + PR_L + Triplet_L | PAGNN | 81.52 ± 0.80 | | 83.39 ± 0.85 | | 35.48 ± 0.30 | | 180.0 |
| CrossE_L + PMI_L + PR_L + Triplet_L | SAGE | 94.58 ± 2.20 | | 97.94 ± 0.51 | | 46.97 ± 1.05 | | 123.3 |
| CrossE_L + PMI_L + Triplet_L | ALL | 97.13 ± 0.30 | | 99.38 ± 0.12 | | 89.30 ± 1.04 | | 65.0 |
| CrossE_L + PMI_L + Triplet_L | GAT | 99.57 ± 0.06 | | 99.85 ± 0.04 | | 92.56 ± 0.91 | | 12.0 |
| CrossE_L + PMI_L + Triplet_L | GCN | 99.15 ± 0.11 | | 99.75 ± 0.04 | | 91.85 ± 0.32 | | 26.7 |
| CrossE_L + PMI_L + Triplet_L | GIN | 93.36 ± 0.66 | | 96.59 ± 0.37 | | 76.94 ± 2.49 | | 114.0 |
| CrossE_L + PMI_L + Triplet_L | MPNN | 97.03 ± 0.33 | | 99.29 ± 0.13 | | 91.51 ± 0.48 | | 59.3 |

Lp Aupr Continued (↑)

| Loss Type | Model | CORA | | Citeseer | | Bitcoin Fraud Transaction | | Average Rank |
|---|---|---|---|---|---|---|---|---|
| CrossE_L + PMI_L + Triplet_L | PAGNN | 81.52 | ± | 83.32 | ± | 35.34 | ± | 182.0 |
| | | 0.72 | | 0.98 | | 0.36 | | |
| CrossE_L + PMI_L + Triplet_L | SAGE | 97.24 | ± | 98.84 | ± | 82.57 | ± | 81.3 |
| | | 0.45 | | 0.55 | | 2.39 | | |
| CrossE_L + PR_L | ALL | 72.58 | ± | 66.45 | ± | 73.54 | ± | 173.3 |
| | | 2.76 | | 3.46 | | 1.37 | | |
| CrossE_L + PR_L | GAT | 94.01 | ± | 94.02 | ± | 76.85 | ± | 118.7 |
| | | 1.60 | | 1.06 | | 11.27 | | |
| CrossE_L + PR_L | GCN | 93.61 | ± | 94.66 | ± | 57.71 | ± | 131.3 |
| | | 1.14 | | 0.93 | | 9.22 | | |
| CrossE_L + PR_L | GIN | 79.22 | ± | 85.79 | ± | 45.68 | ± | 173.7 |
| | | 2.99 | | 2.13 | | 1.11 | | |
| CrossE_L + PR_L | MPNN | 88.92 | ± | 91.26 | ± | 59.47 | ± | 148.7 |
| | | 4.86 | | 1.06 | | 1.18 | | |
| CrossE_L + PR_L | PAGNN | 70.12 | ± | 69.60 | ± | 32.06 | ± | 205.0 |
| | | 2.91 | | 1.62 | | 1.17 | | |
| CrossE_L + PR_L | SAGE | 78.61 | ± | 77.14 | ± | 32.94 | ± | 196.7 |
| | | 1.31 | | 2.34 | | 0.34 | | |
| CrossE_L + PR_L + Triplet_L | ALL | 83.75 | ± | 84.29 | ± | 75.84 | ± | 151.0 |
| | | 8.39 | | 4.88 | | 7.09 | | |
| CrossE_L + PR_L + Triplet_L | GAT | 96.29 | ± | 98.38 | ± | 85.33 | ± | 87.0 |
| | | 0.92 | | 0.51 | | 3.30 | | |
| CrossE_L + PR_L + Triplet_L | GCN | 97.19 | ± | 97.60 | ± | 88.29 | ± | 81.7 |
| | | 0.60 | | 0.31 | | 2.09 | | |
| CrossE_L + PR_L + Triplet_L | GIN | 91.67 | ± | 95.00 | ± | 48.22 | ± | 141.3 |
| | | 1.98 | | 0.86 | | 2.77 | | |
| CrossE_L + PR_L + Triplet_L | MPNN | 89.59 | ± | 94.92 | ± | 72.58 | ± | 132.3 |
| | | 0.89 | | 0.35 | | 5.90 | | |
| CrossE_L + PR_L + Triplet_L | PAGNN | 77.07 | ± | 82.45 | ± | 36.10 | ± | 188.3 |
| | | 2.70 | | 3.61 | | 0.31 | | |
| CrossE_L + PR_L + Triplet_L | SAGE | 97.49 | ± | 98.49 | ± | 57.72 | ± | 100.0 |
| | | 0.32 | | 0.22 | | 14.74 | | |

Lp Aupr Continued (↑)

| Loss Type | Model | CORA | | Citeseer | | Bitcoin Fraud Transaction | | Average Rank |
|---|---|---|---|---|---|---|---|---|
| CrossE_L + Triplet_L | ALL | 98.66 ± 0.20 | | 99.61 ± 0.06 | | 93.25 ± 1.28 | | 37.7 |
| CrossE_L + Triplet_L | GAT | 99.74 ± 0.06 | | 99.92 ± 0.02 | | 96.34 ± 1.05 | | 2.3 |
| CrossE_L + Triplet_L | GCN | 99.56 ± 0.08 | | 99.86 ± 0.05 | | 94.78 ± 0.16 | | 7.3 |
| CrossE_L + Triplet_L | GIN | 98.68 ± 0.23 | | 99.52 ± 0.08 | | 89.71 ± 1.67 | | 50.3 |
| CrossE_L + Triplet_L | MPNN | 98.93 ± 0.24 | | 99.70 ± 0.12 | | 94.00 ± 0.37 | | 31.3 |
| CrossE_L + Triplet_L | PAGNN | 94.92 ± 3.91 | | 99.03 ± 0.24 | | 35.83 ± 0.27 | | 120.3 |
| CrossE_L + Triplet_L | SAGE | 99.62 ± 0.06 | | 99.86 ± 0.04 | | 91.03 ± 0.98 | | 16.3 |
| PMI_L | ALL | 89.88 ± 0.44 | | 92.48 ± 0.68 | | 78.09 ± 0.64 | | 131.0 |
| PMI_L | GAT | 99.61 ± 0.04 | | 99.86 ± 0.02 | | 86.42 ± 1.99 | | 29.3 |
| PMI_L | GCN | 99.25 ± 0.07 | | 99.71 ± 0.10 | | 90.89 ± 0.84 | | 32.7 |
| PMI_L | GIN | 89.52 ± 1.35 | | 92.73 ± 0.63 | | 53.47 ± 2.05 | | 148.0 |
| PMI_L | MPNN | 97.57 ± 0.14 | | 99.33 ± 0.17 | | 89.59 ± 0.70 | | 58.3 |
| PMI_L | PAGNN | 77.45 ± 1.38 | | 82.99 ± 0.31 | | 34.94 ± 0.46 | | 192.0 |
| PMI_L | SAGE | 83.65 ± 1.61 | | 84.47 ± 0.60 | | 36.58 ± 1.35 | | 172.0 |
| PMI_L + PR_L | ALL | 81.05 ± 1.93 | | 88.33 ± 1.69 | | 56.01 ± 0.41 | | 163.0 |
| PMI_L + PR_L | GAT | 98.86 ± 0.25 | | 97.98 ± 2.26 | | 64.28 ± 1.82 | | 94.0 |

Lp Aupr Continued (↑)

| Loss Type | Model | CORA | | | Citeseer | | | Bitcoin Fraud Transaction | | | Average Rank |
|-----------|-------|------|---|------|----------|---|------|---------------------------|---|------|--------------|
| PMI_L + PR_L | GCN | 99.11 | ± | 0.25 | 99.74 | ± | 0.05 | 90.88 | ± | 0.98 | 36.0 |
| PMI_L + PR_L | GIN | 89.68 | ± | 1.61 | 92.19 | ± | 2.20 | 46.42 | ± | 0.62 | 153.0 |
| PMI_L + PR_L | MPNN | 97.15 | ± | 0.44 | 96.28 | ± | 0.52 | 66.47 | ± | 11.40 | 110.0 |
| PMI_L + PR_L | PAGNN | 78.30 | ± | 1.12 | 81.82 | ± | 0.48 | 32.75 | ± | 0.42 | 197.0 |
| PMI_L + PR_L | SAGE | 85.12 | ± | 2.07 | 86.63 | ± | 0.64 | 35.17 | ± | 0.54 | 175.3 |
| PMI_L + PR_L + Triplet_L | ALL | 93.57 | ± | 0.53 | 95.04 | ± | 0.69 | 75.94 | ± | 1.56 | 118.7 |
| PMI_L + PR_L + Triplet_L | GAT | 99.09 | ± | 0.04 | 99.67 | ± | 0.03 | 79.95 | ± | 5.52 | 59.3 |
| PMI_L + PR_L + Triplet_L | GCN | 99.15 | ± | 0.15 | 99.70 | ± | 0.06 | 90.87 | ± | 0.44 | 38.7 |
| PMI_L + PR_L + Triplet_L | GIN | 92.49 | ± | 0.80 | 95.11 | ± | 0.50 | 64.86 | ± | 2.82 | 130.7 |
| PMI_L + PR_L + Triplet_L | MPNN | 97.33 | ± | 0.22 | 97.80 | ± | 0.27 | 81.32 | ± | 7.56 | 88.7 |
| PMI_L + PR_L + Triplet_L | PAGNN | 80.59 | ± | 1.41 | 83.35 | ± | 0.83 | 35.36 | ± | 0.18 | 182.3 |
| PMI_L + PR_L + Triplet_L | SAGE | 96.60 | ± | 1.85 | 98.67 | ± | 0.31 | 49.03 | ± | 0.86 | 109.3 |
| PMI_L + Triplet_L | ALL | 96.86 | ± | 0.36 | 99.27 | ± | 0.20 | 89.15 | ± | 0.46 | 69.7 |
| PMI_L + Triplet_L | GAT | 99.54 | ± | 0.08 | 99.86 | ± | 0.04 | 91.65 | ± | 0.97 | 16.0 |
| PMI_L + Triplet_L | GCN | 99.14 | ± | 0.07 | 99.75 | ± | 0.04 | 91.39 | ± | 0.33 | 31.7 |
| PMI_L + Triplet_L | GIN | 92.71 | ± | 1.58 | 95.99 | ± | 0.54 | 71.68 | ± | 3.75 | 124.0 |

Lp Aupr Continued (↑)

| Loss Type | Model | CORA | | Citeseer | | Bitcoin Fraud Transaction | | Average Rank |
|---|---|---|---|---|---|---|---|---|
| PMI_L + Triplet_L | MPNN | 97.68 ± 0.33 | | 99.31 ± 0.10 | | 91.10 ± 0.70 | | 53.3 |
| PMI_L + Triplet_L | PAGNN | 79.71 ± 1.20 | | 83.32 ± 0.79 | | 35.22 ± 0.41 | | 185.7 |
| PMI_L + Triplet_L | SAGE | 96.12 ± 1.07 | | 98.74 ± 0.29 | | 78.58 ± 2.32 | | 90.7 |
| PR_L | ALL | 63.93 ± 3.00 | | 63.52 ± 1.84 | | 75.18 ± 1.68 | | 175.3 |
| PR_L | GAT | 95.12 ± 1.23 | | 92.99 ± 0.78 | | 79.81 ± 2.48 | | 113.3 |
| PR_L | GCN | 93.29 ± 0.43 | | 94.88 ± 0.46 | | 88.19 ± 1.11 | | 105.3 |
| PR_L | GIN | 78.63 ± 1.81 | | 85.86 ± 2.22 | | 46.32 ± 1.15 | | 174.0 |
| PR_L | MPNN | 86.73 ± 5.84 | | 90.52 ± 0.84 | | 66.24 ± 10.54 | | 150.0 |
| PR_L | PAGNN | 68.39 ± 3.43 | | 67.74 ± 1.52 | | 33.21 ± 1.00 | | 205.0 |
| PR_L | SAGE | 78.56 ± 1.24 | | 78.66 ± 1.14 | | 33.36 ± 0.74 | | 195.3 |
| PR_L + Triplet_L | ALL | 70.44 ± 3.23 | | 63.94 ± 1.46 | | 73.89 ± 1.31 | | 174.7 |
| PR_L + Triplet_L | GAT | 94.74 ± 2.54 | | 93.81 ± 1.38 | | 81.53 ± 2.82 | | 112.3 |
| PR_L + Triplet_L | GCN | 95.61 ± 0.52 | | 96.08 ± 0.68 | | 82.27 ± 1.06 | | 101.7 |
| PR_L + Triplet_L | GIN | 80.41 ± 1.56 | | 89.06 ± 1.64 | | 43.87 ± 1.07 | | 169.7 |
| PR_L + Triplet_L | MPNN | 87.17 ± 3.04 | | 91.07 ± 0.89 | | 70.42 ± 9.63 | | 146.3 |
| PR_L + Triplet_L | PAGNN | 69.99 ± 2.70 | | 72.95 ± 1.69 | | 35.37 ± 0.28 | | 197.3 |

Lp Aupr Continued (↑)

| Loss Type | Model | CORA | | Citeseer | | Bitcoin Fraud Transaction | | Average Rank |
|-----------|-------|------|---|----------|---|---------------------------|---|--------------|
| PR_L + Triplet_L | SAGE | 82.71 ± 2.23 | | 86.90 ± 10.52 | | 36.91 ± 0.83 | | 169.7 |
| Triplet_L | ALL | 98.63 ± 0.20 | | 99.62 ± 0.04 | | 92.16 ± 0.66 | | 38.7 |
| Triplet_L | GAT | 99.80 ± 0.04 | | 99.93 ± 0.02 | | 95.61 ± 0.37 | | 2.0 |
| Triplet_L | GCN | 99.55 ± 0.04 | | 99.87 ± 0.04 | | 94.47 ± 0.62 | | 7.7 |
| Triplet_L | GIN | 98.78 ± 0.39 | | 99.58 ± 0.07 | | 88.99 ± 1.18 | | 52.3 |
| Triplet_L | MPNN | 98.44 ± 0.37 | | 99.66 ± 0.08 | | 93.93 ± 0.51 | | 36.0 |
| Triplet_L | PAGNN | 94.89 ± 4.51 | | 99.02 ± 0.17 | | 36.16 ± 0.33 | | 118.7 |
| Triplet_L | SAGE | 99.56 ± 0.08 | | 99.86 ± 0.05 | | 90.75 ± 0.49 | | 21.0 |

Table 7. **Lp Auroc Performance (↑)**: Top-ranked results are highlighted in **1st**, second-ranked in **2nd**, and third-ranked in **3rd**.

| Loss Type | Model | CORA | | Citeseer | | Bitcoin Fraud Transaction | | Average Rank |
|-----------|-------|------|---|----------|---|---------------------------|---|--------------|
| Contr_l | ALL | 96.54 ± 0.32 | | 98.63 ± 0.21 | | 91.81 ± 0.70 | | 77.0 |
| Contr_l | GAT | 99.27 ± 0.04 | | 99.88 ± 0.02 | | 95.68 ± 0.29 | | 14.7 |
| Contr_l | GCN | 98.94 ± 0.06 | | 99.80 ± 0.04 | | 94.33 ± 0.11 | | 32.3 |
| Contr_l | GIN | 97.53 ± 0.17 | | 99.42 ± 0.04 | | 91.97 ± 2.19 | | 68.7 |

Lp Auroc Continued (↑)

| Loss Type | Model | CORA | | Citeseer | | Bitcoin Fraud Transaction | | Average Rank |
|---|---|---|---|---|---|---|---|---|
| Contr_l | MPNN | 97.71 | ± | 99.40 | ± | 91.34 | ± | 70.3 |
| | | 0.18 | | 0.11 | | 0.40 | | |
| Contr_l | PAGNN | 96.45 | ± | 98.32 | ± | 58.46 | ± | 117.7 |
| | | 0.44 | | 0.33 | | 0.45 | | |
| Contr_l | SAGE | 98.43 | ± | 99.68 | ± | 90.54 | ± | 60.0 |
| | | 0.04 | | 0.05 | | 0.99 | | |
| Contr_l + CrossE_L | ALL | 96.46 | ± | 98.73 | ± | 91.14 | ± | 79.3 |
| | | 0.22 | | 0.14 | | 1.48 | | |
| Contr_l + CrossE_L | GAT | 99.24 | ± | 99.90 | ± | 96.75 | ± | 10.7 |
| | | 0.08 | | 0.02 | | 1.37 | | |
| Contr_l + CrossE_L | GCN | 98.97 | ± | 99.80 | ± | 94.11 | ± | 32.7 |
| | | 0.10 | | 0.03 | | 0.25 | | |
| Contr_l + CrossE_L | GIN | 97.08 | ± | 99.22 | ± | 90.94 | ± | 75.7 |
| | | 0.33 | | 0.12 | | 1.70 | | |
| Contr_l + CrossE_L | MPNN | 97.67 | ± | 99.44 | ± | 92.79 | ± | 64.7 |
| | | 0.40 | | 0.08 | | 0.46 | | |
| Contr_l + CrossE_L | PAGNN | 96.20 | ± | 98.11 | ± | 58.57 | ± | 119.0 |
| | | 0.56 | | 0.24 | | 0.38 | | |
| Contr_l + CrossE_L | SAGE | 98.53 | ± | 99.72 | ± | 91.05 | ± | 56.0 |
| | | 0.18 | | 0.06 | | 1.74 | | |
| Contr_l + CrossE_L + PMI_L | ALL | 92.72 | ± | 91.61 | ± | 84.23 | ± | 120.3 |
| | | 1.57 | | 3.04 | | 0.84 | | |
| Contr_l + CrossE_L + PMI_L | GAT | 99.36 | ± | 99.88 | ± | 91.57 | ± | 30.3 |
| | | 0.06 | | 0.02 | | 1.02 | | |
| Contr_l + CrossE_L + PMI_L | GCN | 99.03 | ± | 99.75 | ± | 93.32 | ± | 39.3 |
| | | 0.19 | | 0.04 | | 0.26 | | |
| Contr_l + CrossE_L + PMI_L | GIN | 92.40 | ± | 93.69 | ± | 67.79 | ± | 132.7 |
| | | 1.64 | | 0.90 | | 2.80 | | |
| Contr_l + CrossE_L + PMI_L | MPNN | 98.21 | ± | 99.52 | ± | 91.95 | ± | 60.0 |
| | | 0.16 | | 0.07 | | 0.71 | | |
| Contr_l + CrossE_L + PMI_L | PAGNN | 73.46 | ± | 84.66 | ± | 55.73 | ± | 188.0 |
| | | 2.01 | | 0.55 | | 0.64 | | |

Lp Auroc Continued (↑)

| Loss Type | Model | CORA | | Citeseer | | Bitcoin Fraud Transaction | | Average Rank |
|---|---|---|---|---|---|---|---|---|
| Contr_l + CrossE_L + PMI_L | SAGE | 86.34 0.97 | ± | 90.45 2.63 | ± | 68.67 4.77 | ± | 148.0 |
| Contr_l + CrossE_L + PMI_L + PR_L | ALL | 79.16 2.29 | ± | 88.06 0.82 | ± | 65.98 0.15 | ± | 159.7 |
| Contr_l + CrossE_L + PMI_L + PR_L | GAT | 99.01 0.35 | ± | 99.80 0.06 | ± | 78.23 4.95 | ± | 61.0 |
| Contr_l + CrossE_L + PMI_L + PR_L | GCN | 99.02 0.16 | ± | 99.74 0.03 | ± | 93.08 0.66 | ± | 42.0 |
| Contr_l + CrossE_L + PMI_L + PR_L | GIN | 91.07 1.40 | ± | 91.97 0.77 | ± | 63.87 1.03 | ± | 144.3 |
| Contr_l + CrossE_L + PMI_L + PR_L | MPNN | 97.91 0.26 | ± | 98.12 1.29 | ± | 74.53 9.24 | ± | 96.7 |
| Contr_l + CrossE_L + PMI_L + PR_L | PAGNN | 76.63 2.52 | ± | 84.56 0.65 | ± | 51.41 0.52 | ± | 187.3 |
| Contr_l + CrossE_L + PMI_L + PR_L | SAGE | 85.49 1.99 | ± | 89.27 3.39 | ± | 58.78 0.59 | ± | 158.7 |
| Contr_l + CrossE_L + PMI_L + PR_L + Triplet_L | ALL | 94.42 0.17 | ± | 95.06 0.46 | ± | 79.79 0.67 | ± | 113.0 |
| Contr_l + CrossE_L + PMI_L + PR_L + Triplet_L | GAT | 99.04 0.22 | ± | 99.79 0.10 | ± | 90.15 3.74 | ± | 44.0 |
| Contr_l + CrossE_L + PMI_L + PR_L + Triplet_L | GCN | 99.22 0.09 | ± | 99.77 0.04 | ± | 94.04 0.16 | ± | 27.3 |
| Contr_l + CrossE_L + PMI_L + PR_L + Triplet_L | GIN | 91.80 1.37 | ± | 95.71 0.45 | ± | 74.73 1.83 | ± | 125.0 |
| Contr_l + CrossE_L + PMI_L + PR_L + Triplet_L | MPNN | 97.80 0.52 | ± | 98.47 0.55 | ± | 88.51 6.18 | ± | 80.3 |
| Contr_l + CrossE_L + PMI_L + PR_L + Triplet_L | PAGNN | 78.31 2.20 | ± | 85.23 0.32 | ± | 56.45 0.19 | ± | 179.7 |
| Contr_l + CrossE_L + PMI_L + PR_L + Triplet_L | SAGE | 94.85 1.68 | ± | 98.77 0.16 | ± | 65.59 1.52 | ± | 112.0 |
| Contr_l + CrossE_L + PMI_L + Triplet_L | ALL | 96.82 0.24 | ± | 99.32 0.08 | ± | 91.03 0.44 | ± | 75.3 |

Lp Auroc Continued (↑)

| Loss Type | Model | CORA | | Citeseer | | Bitcoin Fraud Transaction | | Average Rank |
|---|---|---|---|---|---|---|---|---|
| Contr_l + CrossE_L + PMI_L + Triplet_L | GAT | 99.50 ± 0.12 | | 99.89 ± 0.02 | | 94.14 ± 0.25 | | 16.3 |
| Contr_l + CrossE_L + PMI_L + Triplet_L | GCN | 99.20 ± 0.08 | | 99.79 ± 0.04 | | 94.03 ± 0.44 | | 27.0 |
| Contr_l + CrossE_L + PMI_L + Triplet_L | GIN | 92.63 ± 2.00 | | 96.12 ± 0.28 | | 78.66 ± 3.18 | | 116.7 |
| Contr_l + CrossE_L + PMI_L + Triplet_L | MPNN | 98.28 ± 0.30 | | 99.45 ± 0.07 | | 93.45 ± 0.79 | | 56.0 |
| Contr_l + CrossE_L + PMI_L + Triplet_L | PAGNN | 78.00 ± 1.07 | | 84.86 ± 0.31 | | 56.24 ± 0.37 | | 181.7 |
| Contr_l + CrossE_L + PMI_L + Triplet_L | SAGE | 95.82 ± 0.39 | | 98.44 ± 0.32 | | 86.41 ± 1.29 | | 90.3 |
| Contr_l + CrossE_L + PR_L | ALL | 65.47 ± 4.01 | | 71.06 ± 10.21 | | 68.21 ± 0.88 | | 181.7 |
| Contr_l + CrossE_L + PR_L | GAT | 91.37 ± 3.90 | | 91.31 ± 3.67 | | 82.60 ± 4.07 | | 127.7 |
| Contr_l + CrossE_L + PR_L | GCN | 94.26 ± 0.56 | | 94.47 ± 0.52 | | 85.59 ± 0.56 | | 108.3 |
| Contr_l + CrossE_L + PR_L | GIN | 76.13 ± 2.15 | | 89.93 ± 1.94 | | 56.47 ± 1.24 | | 174.7 |
| Contr_l + CrossE_L + PR_L | MPNN | 79.91 ± 2.31 | | 88.30 ± 1.47 | | 75.30 ± 9.44 | | 150.3 |
| Contr_l + CrossE_L + PR_L | PAGNN | 64.08 ± 5.04 | | 73.45 ± 5.16 | | 57.31 ± 1.02 | | 193.7 |
| Contr_l + CrossE_L + PR_L | SAGE | 91.03 ± 7.91 | | 91.72 ± 11.11 | | 58.74 ± 1.89 | | 148.7 |
| Contr_l + CrossE_L + PR_L + Triplet_L | ALL | 90.34 ± 0.80 | | 96.14 ± 0.52 | | 88.66 ± 0.98 | | 111.3 |
| Contr_l + CrossE_L + PR_L + Triplet_L | GAT | 96.67 ± 0.66 | | 97.96 ± 0.11 | | 92.60 ± 2.28 | | 78.7 |
| Contr_l + CrossE_L + PR_L + Triplet_L | GCN | 97.41 ± 0.58 | | 98.12 ± 0.23 | | 91.42 ± 2.67 | | 78.7 |

Lp Auroc Continued (↑)

| Loss Type | Model | CORA | | Citeseer | | Bitcoin Fraud Transaction | | Average Rank |
|---|---|---|---|---|---|---|---|---|
| Contr_l + CrossE_L + PR_L + Triplet_L | GIN | 92.55 0.74 | ± | 95.76 0.55 | ± | 84.24 10.74 | | 112.0 |
| Contr_l + CrossE_L + PR_L + Triplet_L | MPNN | 92.09 0.87 | ± | 96.42 0.25 | ± | 84.61 4.28 | | 110.7 |
| Contr_l + CrossE_L + PR_L + Triplet_L | PAGNN | 79.50 4.06 | ± | 90.85 3.65 | ± | 58.66 0.31 | | 162.0 |
| Contr_l + CrossE_L + PR_L + Triplet_L | SAGE | 95.81 0.17 | ± | 97.56 0.23 | ± | 83.08 2.30 | | 100.0 |
| Contr_l + CrossE_L + Triplet_L | ALL | 97.90 0.18 | ± | 99.44 0.05 | ± | 94.72 0.90 | ± | 54.3 |
| Contr_l + CrossE_L + Triplet_L | GAT | 99.58 0.09 | ± | 99.93 0.02 | ± | 98.25 0.53 | ± | 3.0 |
| Contr_l + CrossE_L + Triplet_L | GCN | 99.28 0.06 | ± | 99.85 0.03 | ± | 95.60 0.19 | ± | 16.3 |
| Contr_l + CrossE_L + Triplet_L | GIN | 98.51 0.21 | ± | 99.61 0.13 | ± | 93.56 1.45 | ± | 49.0 |
| Contr_l + CrossE_L + Triplet_L | MPNN | 98.49 0.27 | ± | 99.70 0.04 | ± | 94.78 0.64 | ± | 43.3 |
| Contr_l + CrossE_L + Triplet_L | PAGNN | 95.03 3.71 | ± | 99.15 0.09 | ± | 58.54 0.25 | ± | 117.3 |
| Contr_l + CrossE_L + Triplet_L | SAGE | 99.13 0.15 | ± | 99.82 0.03 | ± | 95.35 0.66 | ± | 20.7 |
| Contr_l + PMI_L | ALL | 92.96 1.84 | ± | 97.26 0.58 | ± | 86.16 1.10 | ± | 103.3 |
| Contr_l + PMI_L | GAT | 99.50 0.07 | ± | 99.87 0.02 | ± | 91.78 0.97 | ± | 30.3 |
| Contr_l + PMI_L | GCN | 99.04 0.12 | ± | 99.73 0.03 | ± | 93.77 0.63 | ± | 37.0 |
| Contr_l + PMI_L | GIN | 91.31 2.58 | ± | 94.46 0.90 | ± | 70.10 1.18 | ± | 132.7 |
| Contr_l + PMI_L | MPNN | 98.24 0.19 | ± | 99.51 0.11 | ± | 92.16 0.42 | ± | 58.7 |

Lp Auroc Continued (↑)

| Loss Type | Model | CORA | | Citeseer | | Bitcoin Fraud Transaction | | Average Rank |
|---|---|---|---|---|---|---|---|---|
| Contr_l + PMI_L | PAGNN | 75.25 ± 2.06 | | 84.34 ± 0.49 | | 55.79 ± 0.43 | | 187.3 |
| Contr_l + PMI_L | SAGE | 87.77 ± 2.16 | | 96.12 ± 1.23 | | 67.52 ± 1.21 | | 136.3 |
| Contr_l + PMI_L + PR_L | ALL | 82.03 ± 1.61 | | 87.81 ± 0.74 | | 66.91 ± 0.52 | | 156.7 |
| Contr_l + PMI_L + PR_L | GAT | 98.91 ± 0.17 | | 99.29 ± 0.71 | | 74.80 ± 4.70 | | 82.3 |
| Contr_l + PMI_L + PR_L | GCN | 99.12 ± 0.08 | | 99.73 ± 0.05 | | 93.69 ± 0.38 | | 35.0 |
| Contr_l + PMI_L + PR_L | GIN | 92.18 ± 2.21 | | 92.42 ± 1.28 | | 64.74 ± 1.06 | | 138.3 |
| Contr_l + PMI_L + PR_L | MPNN | 98.06 ± 0.26 | | 97.47 ± 1.13 | | 74.67 ± 8.36 | | 99.0 |
| Contr_l + PMI_L + PR_L | PAGNN | 74.75 ± 1.75 | | 84.09 ± 0.45 | | 51.80 ± 0.50 | | 192.0 |
| Contr_l + PMI_L + PR_L | SAGE | 86.98 ± 1.43 | | 94.63 ± 1.09 | | 58.97 ± 0.51 | | 145.7 |
| Contr_l + PMI_L + PR_L + Triplet_L | ALL | 95.21 ± 0.53 | | 97.16 ± 0.32 | | 84.37 ± 1.20 | | 101.7 |
| Contr_l + PMI_L + PR_L + Triplet_L | GAT | 99.04 ± 0.24 | | 99.81 ± 0.05 | | 88.82 ± 2.14 | | 45.3 |
| Contr_l + PMI_L + PR_L + Triplet_L | GCN | 99.06 ± 0.16 | | 99.73 ± 0.07 | | 93.91 ± 0.69 | | 36.0 |
| Contr_l + PMI_L + PR_L + Triplet_L | GIN | 94.38 ± 0.66 | | 96.88 ± 0.65 | | 79.15 ± 0.65 | | 108.7 |
| Contr_l + PMI_L + PR_L + Triplet_L | MPNN | 97.82 ± 0.36 | | 98.47 ± 0.15 | | 83.62 ± 3.70 | | 86.3 |
| Contr_l + PMI_L + PR_L + Triplet_L | PAGNN | 79.50 ± 0.71 | | 87.68 ± 1.71 | | 57.74 ± 0.58 | | 170.3 |
| Contr_l + PMI_L + PR_L + Triplet_L | SAGE | 97.32 ± 0.49 | | 99.38 ± 0.14 | | 73.65 ± 3.14 | | 94.3 |

Lp Auroc Continued (↑)

| Loss Type | Model | CORA | | Citeseer | | Bitcoin Fraud Transaction | | Average Rank |
|-----------|-------|------|--|----------|--|---------------------------|--|--------------|
| Contr_l + PR_L | ALL | 69.73 ± 2.60 | | 68.10 ± 8.59 | | 68.02 ± 1.78 | | 181.3 |
| Contr_l + PR_L | GAT | 91.99 ± 3.29 | | 90.84 ± 1.29 | | 83.56 ± 5.00 | | 126.0 |
| Contr_l + PR_L | GCN | 93.87 ± 0.61 | | 95.33 ± 1.21 | | 90.11 ± 3.41 | | 102.3 |
| Contr_l + PR_L | GIN | 74.43 ± 3.73 | | 88.00 ± 2.29 | | 55.24 ± 2.28 | | 182.7 |
| Contr_l + PR_L | MPNN | 78.44 ± 1.85 | | 88.79 ± 2.31 | | 69.28 ± 0.77 | | 157.3 |
| Contr_l + PR_L | PAGNN | 61.82 ± 2.92 | | 74.85 ± 3.17 | | 57.17 ± 0.64 | | 193.7 |
| Contr_l + PR_L | SAGE | 90.84 ± 8.40 | | 92.58 ± 10.25 | | 58.97 ± 1.85 | | 145.3 |
| Contr_l + PR_L + Triplet_L | ALL | 86.11 ± 5.73 | | 93.78 ± 2.84 | | 88.73 ± 0.41 | | 120.7 |
| Contr_l + PR_L + Triplet_L | GAT | 96.25 ± 0.50 | | 97.98 ± 0.12 | | 91.56 ± 3.44 | | 83.7 |
| Contr_l + PR_L + Triplet_L | GCN | 96.98 ± 0.38 | | 98.10 ± 0.33 | | 89.68 ± 2.58 | | 84.7 |
| Contr_l + PR_L + Triplet_L | GIN | 92.66 ± 0.38 | | 95.39 ± 1.65 | | 77.51 ± 10.59 | | 119.3 |
| Contr_l + PR_L + Triplet_L | MPNN | 92.70 ± 2.54 | | 95.66 ± 0.59 | | 81.01 ± 3.24 | | 114.3 |
| Contr_l + PR_L + Triplet_L | PAGNN | 76.23 ± 2.35 | | 90.64 ± 4.96 | | 58.58 ± 0.43 | | 167.7 |
| Contr_l + PR_L + Triplet_L | SAGE | 95.65 ± 0.14 | | 97.60 ± 0.12 | | 86.27 ± 2.00 | | 96.0 |
| Contr_l + Triplet_L | ALL | 98.15 ± 0.14 | | 99.48 ± 0.10 | | 95.79 ± 0.40 | | 46.3 |
| Contr_l + Triplet_L | GAT | 99.55 ± 0.07 | | 99.92 ± 0.01 | | 98.40 ± 0.51 | | 4.0 |

Lp Auroc Continued (↑)

| Loss Type | Model | CORA | | Citeseer | | Bitcoin Fraud Transaction | | Average Rank |
|---|---|---|---|---|---|---|---|---|
| Contr_l + Triplet_L | GCN | 99.29 | ± | 99.87 | ± | 95.73 | ± | 15.0 |
| | | 0.05 | | 0.03 | | 0.18 | | |
| Contr_l + Triplet_L | GIN | 98.50 | ± | 99.58 | ± | 92.72 | ± | 53.7 |
| | | 0.36 | | 0.14 | | 1.41 | | |
| Contr_l + Triplet_L | MPNN | 98.55 | ± | 99.71 | ± | 95.08 | ± | 40.3 |
| | | 0.23 | | 0.03 | | 0.57 | | |
| Contr_l + Triplet_L | PAGNN | 95.37 | ± | 99.20 | ± | 58.49 | ± | 116.7 |
| | | 4.31 | | 0.07 | | 0.72 | | |
| Contr_l + Triplet_L | SAGE | 99.26 | ± | 99.83 | ± | 95.15 | ± | 19.0 |
| | | 0.05 | | 0.03 | | 1.84 | | |
| CrossE_L | ALL | 89.56 | ± | 83.20 | ± | 71.82 | ± | 156.0 |
| | | 10.81 | | 9.28 | | 3.32 | | |
| CrossE_L | GAT | 85.76 | ± | 89.82 | ± | 85.59 | ± | 133.3 |
| | | 17.67 | | 20.13 | | 16.71 | | |
| CrossE_L | GCN | 44.42 | ± | 43.42 | ± | 46.33 | ± | 208.7 |
| | | 3.64 | | 1.50 | | 3.82 | | |
| CrossE_L | GIN | 39.07 | ± | 39.33 | ± | 38.12 | ± | 210.0 |
| | | 3.21 | | 1.32 | | 1.92 | | |
| CrossE_L | MPNN | 91.93 | ± | 93.59 | ± | 49.59 | ± | 155.7 |
| | | 1.50 | | 3.69 | | 2.58 | | |
| CrossE_L | PAGNN | 72.91 | ± | 75.80 | ± | 41.91 | ± | 199.3 |
| | | 1.16 | | 2.40 | | 1.81 | | |
| CrossE_L | SAGE | 76.37 | ± | 72.76 | ± | 49.84 | ± | 195.7 |
| | | 5.72 | | 8.12 | | 1.00 | | |
| CrossE_L + PMI_L | ALL | 91.44 | ± | 91.21 | ± | 80.94 | ± | 129.3 |
| | | 0.80 | | 0.85 | | 1.37 | | |
| CrossE_L + PMI_L | GAT | 99.55 | ± | 99.91 | ± | 89.55 | ± | 29.7 |
| | | 0.04 | | 0.02 | | 2.31 | | |
| CrossE_L + PMI_L | GCN | 99.14 | ± | 99.75 | ± | 93.47 | ± | 33.0 |
| | | 0.15 | | 0.07 | | 0.27 | | |
| CrossE_L + PMI_L | GIN | 91.88 | ± | 92.37 | ± | 65.76 | ± | 139.3 |
| | | 2.09 | | 0.33 | | 0.89 | | |

Lp Auroc Continued (↑)

| Loss Type | Model | CORA | | Citeseer | | Bitcoin Fraud Transaction | | Average Rank |
|---|---|---|---|---|---|---|---|---|
| CrossE_L + PMI_L | MPNN | 98.21 | ± | 99.53 | ± | 92.18 | ± | 58.7 |
| | | 0.13 | | 0.09 | | 0.89 | | |
| CrossE_L + PMI_L | PAGNN | 75.46 | ± | 84.27 | ± | 54.30 | ± | 189.3 |
| | | 2.77 | | 0.33 | | 2.03 | | |
| CrossE_L + PMI_L | SAGE | 82.78 | ± | 83.73 | ± | 55.29 | ± | 181.0 |
| | | 1.16 | | 0.65 | | 2.30 | | |
| CrossE_L + PMI_L + PR_L | ALL | 79.15 | ± | 87.61 | ± | 63.40 | ± | 164.7 |
| | | 0.76 | | 0.32 | | 0.46 | | |
| CrossE_L + PMI_L + PR_L | GAT | 99.14 | ± | 98.86 | ± | 76.82 | ± | 75.0 |
| | | 0.21 | | 0.88 | | 5.86 | | |
| CrossE_L + PMI_L + PR_L | GCN | 99.07 | ± | 99.75 | ± | 92.87 | ± | 38.7 |
| | | 0.22 | | 0.07 | | 0.99 | | |
| CrossE_L + PMI_L + PR_L | GIN | 89.43 | ± | 91.26 | ± | 62.26 | ± | 149.7 |
| | | 1.70 | | 2.21 | | 1.84 | | |
| CrossE_L + PMI_L + PR_L | MPNN | 97.94 | ± | 98.13 | ± | 79.35 | ± | 89.7 |
| | | 0.36 | | 1.68 | | 10.59 | | |
| CrossE_L + PMI_L + PR_L | PAGNN | 76.82 | ± | 84.34 | ± | 51.40 | ± | 188.3 |
| | | 2.91 | | 0.27 | | 0.75 | | |
| CrossE_L + PMI_L + PR_L | SAGE | 82.77 | ± | 84.66 | ± | 56.32 | ± | 175.7 |
| | | 0.88 | | 1.70 | | 1.37 | | |
| CrossE_L + PMI_L + PR_L + Triplet_L | ALL | 93.75 | ± | 93.95 | ± | 78.12 | ± | 120.0 |
| | | 0.19 | | 0.60 | | 1.19 | | |
| CrossE_L + PMI_L + PR_L + Triplet_L | GAT | 99.09 | ± | 99.75 | ± | 88.15 | ± | 49.7 |
| | | 0.17 | | 0.05 | | 5.38 | | |
| CrossE_L + PMI_L + PR_L + Triplet_L | GCN | 99.09 | ± | 99.79 | ± | 93.87 | ± | 31.3 |
| | | 0.15 | | 0.03 | | 0.61 | | |
| CrossE_L + PMI_L + PR_L + Triplet_L | GIN | 92.24 | ± | 95.37 | ± | 75.33 | ± | 122.0 |
| | | 0.92 | | 0.72 | | 1.47 | | |
| CrossE_L + PMI_L + PR_L + Triplet_L | MPNN | 98.17 | ± | 98.59 | ± | 88.15 | ± | 77.7 |
| | | 0.17 | | 0.80 | | 6.43 | | |
| CrossE_L + PMI_L + PR_L + Triplet_L | PAGNN | 80.40 | ± | 85.41 | ± | 56.91 | ± | 173.0 |
| | | 1.01 | | 0.78 | | 0.49 | | |

Lp Auroc Continued (↑)

| Loss Type | Model | CORA | | Citeseer | | Bitcoin Fraud Transaction | | Average Rank |
|---|---|---|---|---|---|---|---|---|
| CrossE_L + PMI_L + PR_L + Triplet_L | SAGE | 94.59 ± 2.71 | | 98.30 ± 0.45 | | 63.66 ± 0.57 | | 117.0 |
| CrossE_L + PMI_L + Triplet_L | ALL | 97.59 ± 0.26 | | 99.56 ± 0.09 | | 92.28 ± 0.80 | | 62.3 |
| CrossE_L + PMI_L + Triplet_L | GAT | 99.55 ± 0.07 | | 99.89 ± 0.03 | | 95.07 ± 0.51 | | 12.7 |
| CrossE_L + PMI_L + Triplet_L | GCN | 99.04 ± 0.12 | | 99.78 ± 0.03 | | 94.36 ± 0.26 | | 31.0 |
| CrossE_L + PMI_L + Triplet_L | GIN | 93.92 ± 0.83 | | 96.98 ± 0.35 | | 84.64 ± 2.16 | | 104.0 |
| CrossE_L + PMI_L + Triplet_L | MPNN | 97.77 ± 0.22 | | 99.49 ± 0.09 | | 94.42 ± 0.29 | | 54.0 |
| CrossE_L + PMI_L + Triplet_L | PAGNN | 80.57 ± 0.69 | | 85.30 ± 0.89 | | 56.73 ± 0.56 | | 174.0 |
| CrossE_L + PMI_L + Triplet_L | SAGE | 97.16 ± 0.49 | | 99.08 ± 0.41 | | 89.47 ± 1.37 | | 79.3 |
| CrossE_L + PR_L | ALL | 68.43 ± 2.83 | | 61.68 ± 4.72 | | 69.42 ± 1.52 | | 181.0 |
| CrossE_L + PR_L | GAT | 90.24 ± 2.55 | | 90.10 ± 1.94 | | 78.78 ± 8.42 | | 136.0 |
| CrossE_L + PR_L | GCN | 90.37 ± 1.90 | | 92.06 ± 1.30 | | 61.45 ± 7.25 | | 146.3 |
| CrossE_L + PR_L | GIN | 70.58 ± 4.39 | | 80.65 ± 2.28 | | 57.42 ± 1.12 | | 189.3 |
| CrossE_L + PR_L | MPNN | 83.21 ± 8.38 | | 86.77 ± 1.98 | | 64.61 ± 1.90 | | 159.3 |
| CrossE_L + PR_L | PAGNN | 60.91 ± 4.03 | | 67.46 ± 2.06 | | 48.36 ± 3.09 | | 205.3 |
| CrossE_L + PR_L | SAGE | 73.94 ± 1.38 | | 73.47 ± 2.57 | | 51.54 ± 0.95 | | 196.7 |
| CrossE_L + PR_L + Triplet_L | ALL | 82.45 ± 8.27 | | 84.22 ± 4.55 | | 77.46 ± 10.43 | | 155.7 |

Lp Auroc Continued (↑)

| Loss Type | | | | Model | CORA | | Citeseer | | Bitcoin Fraud Transaction | | Average Rank |
|---|---|---|---|---|---|---|---|---|---|---|---|
| CrossE_L | + | PR_L | + | GAT | 94.31 ± 1.22 | | 97.45 ± 0.45 | | 86.84 ± 4.25 | | 99.3 |
| Triplet_L | | | | | | | | | | | |
| CrossE_L | + | PR_L | + | GCN | 95.99 ± 0.93 | | 96.60 ± 0.37 | | 90.62 ± 2.26 | | 91.0 |
| Triplet_L | | | | | | | | | | | |
| CrossE_L | + | PR_L | + | GIN | 88.95 ± 2.50 | | 93.07 ± 1.46 | | 59.18 ± 2.01 | | 147.3 |
| Triplet_L | | | | | | | | | | | |
| CrossE_L | + | PR_L | + | MPNN | 84.89 ± 0.86 | | 93.13 ± 0.63 | | 78.40 ± 3.89 | | 134.7 |
| Triplet_L | | | | | | | | | | | |
| CrossE_L | + | PR_L | + | PAGNN | 71.72 ± 4.31 | | 84.51 ± 3.18 | | 58.72 ± 0.48 | | 181.0 |
| Triplet_L | | | | | | | | | | | |
| CrossE_L | + | PR_L | + | SAGE | 95.86 ± 0.33 | | 97.50 ± 0.23 | | 73.22 ± 10.57 | | 109.3 |
| Triplet_L | | | | | | | | | | | |
| CrossE_L + Triplet_L | | | | ALL | 98.86 ± 0.18 | | 99.73 ± 0.04 | | 95.60 ± 0.97 | | 35.3 |
| CrossE_L + Triplet_L | | | | GAT | 99.76 ± 0.06 | | 99.95 ± 0.01 | | 97.78 ± 0.64 | | 2.0 |
| CrossE_L + Triplet_L | | | | GCN | 99.54 ± 0.09 | | 99.89 ± 0.03 | | 96.80 ± 0.11 | | 8.3 |
| CrossE_L + Triplet_L | | | | GIN | 98.95 ± 0.19 | | 99.66 ± 0.05 | | 93.55 ± 1.16 | | 46.0 |
| CrossE_L + Triplet_L | | | | MPNN | 99.14 ± 0.19 | | 99.80 ± 0.08 | | 96.25 ± 0.31 | | 20.0 |
| CrossE_L + Triplet_L | | | | PAGNN | 95.67 ± 4.49 | | 99.47 ± 0.14 | | 58.00 ± 0.40 | | 113.7 |
| CrossE_L + Triplet_L | | | | SAGE | 99.62 ± 0.06 | | 99.89 ± 0.03 | | 95.34 ± 0.45 | | 10.0 |
| PMI_L | | | | ALL | 88.56 ± 0.53 | | 91.55 ± 0.85 | | 81.49 ± 0.74 | | 132.7 |
| PMI_L | | | | GAT | 99.58 ± 0.05 | | 99.88 ± 0.02 | | 89.64 ± 2.40 | | 32.0 |
| PMI_L | | | | GCN | 99.12 ± 0.08 | | 99.73 ± 0.09 | | 93.46 ± 0.68 | | 37.7 |

Lp Auroc Continued (↑)

| Loss Type | Model | CORA | | Citeseer | | Bitcoin Fraud Transaction | | Average Rank |
|---|---|---|---|---|---|---|---|---|
| PMI_L | GIN | 89.16 1.79 | ± | 92.53 0.73 | ± | 65.26 2.55 | ± | 144.3 |
| PMI_L | MPNN | 98.22 0.08 | ± | 99.54 0.11 | ± | 92.35 0.55 | ± | 57.0 |
| PMI_L | PAGNN | 75.29 1.64 | ± | 84.23 0.26 | ± | 55.70 0.92 | ± | 189.0 |
| PMI_L | SAGE | 81.90 1.71 | ± | 82.70 0.85 | ± | 57.43 2.61 | ± | 177.3 |
| PMI_L + PR_L | ALL | 79.79 1.84 | ± | 86.90 1.39 | ± | 64.05 0.28 | ± | 163.0 |
| PMI_L + PR_L | GAT | 98.79 0.27 | ± | 97.86 2.55 | ± | 69.96 1.00 | ± | 95.3 |
| PMI_L + PR_L | GCN | 98.99 0.35 | ± | 99.77 0.04 | ± | 93.45 0.69 | ± | 39.7 |
| PMI_L + PR_L | GIN | 89.39 1.74 | ± | 92.10 2.51 | ± | 61.38 0.92 | ± | 148.3 |
| PMI_L + PR_L | MPNN | 97.90 0.35 | ± | 96.38 0.52 | ± | 72.95 8.80 | ± | 104.7 |
| PMI_L + PR_L | PAGNN | 75.99 0.97 | ± | 83.62 0.61 | ± | 50.39 1.37 | ± | 193.0 |
| PMI_L + PR_L | SAGE | 83.38 2.33 | ± | 85.26 0.70 | ± | 56.16 0.85 | ± | 173.7 |
| PMI_L + PR_L + Triplet_L | ALL | 94.09 0.44 | ± | 95.26 0.64 | ± | 79.09 1.53 | ± | 115.0 |
| PMI_L + PR_L + Triplet_L | GAT | 99.03 0.06 | ± | 99.73 0.04 | ± | 85.35 3.72 | ± | 58.3 |
| PMI_L + PR_L + Triplet_L | GCN | 99.03 0.16 | ± | 99.73 0.04 | ± | 93.41 0.36 | ± | 43.3 |
| PMI_L + PR_L + Triplet_L | GIN | 92.91 0.77 | ± | 95.37 0.45 | ± | 75.17 1.68 | ± | 121.0 |
| PMI_L + PR_L + Triplet_L | MPNN | 97.97 0.18 | ± | 98.01 0.23 | ± | 84.83 6.13 | ± | 85.3 |

Lp Auroc Continued (↑)

| Loss Type | Model | CORA | | Citeseer | | Bitcoin Fraud Transaction | | Average Rank |
|---|---|---|---|---|---|---|---|---|
| PMI_L + PR_L + Triplet_L | PAGNN | 79.14 ± 1.63 | | 85.98 ± 0.91 | | 56.85 ± 0.38 | | 175.3 |
| PMI_L + PR_L + Triplet_L | SAGE | 96.47 ± 2.07 | | 98.92 ± 0.26 | | 65.97 ± 0.91 | | 105.3 |
| PMI_L + Triplet_L | ALL | 97.39 ± 0.26 | | 99.48 ± 0.13 | | 92.12 ± 0.36 | | 67.0 |
| PMI_L + Triplet_L | GAT | 99.52 ± 0.08 | | 99.89 ± 0.02 | | 94.46 ± 0.75 | | 16.0 |
| PMI_L + Triplet_L | GCN | 99.01 ± 0.10 | | 99.77 ± 0.03 | | 93.96 ± 0.24 | | 36.0 |
| PMI_L + Triplet_L | GIN | 93.03 ± 1.78 | | 96.33 ± 0.59 | | 80.67 ± 2.71 | | 111.7 |
| PMI_L + Triplet_L | MPNN | 98.26 ± 0.25 | | 99.51 ± 0.08 | | 93.93 ± 0.46 | | 51.3 |
| PMI_L + Triplet_L | PAGNN | 78.22 ± 1.47 | | 85.19 ± 0.51 | | 56.47 ± 0.65 | | 180.0 |
| PMI_L + Triplet_L | SAGE | 96.20 ± 1.01 | | 98.96 ± 0.21 | | 87.31 ± 1.09 | | 85.7 |
| PR_L | ALL | 57.24 ± 3.99 | | 56.36 ± 2.56 | | 70.82 ± 1.91 | | 183.3 |
| PR_L | GAT | 91.91 ± 2.17 | | 88.32 ± 1.27 | | 78.99 ± 3.49 | | 133.7 |
| PR_L | GCN | 89.77 ± 1.06 | | 92.51 ± 0.61 | | 89.84 ± 1.51 | | 117.7 |
| PR_L | GIN | 69.78 ± 3.04 | | 80.63 ± 2.69 | | 58.42 ± 0.67 | | 188.7 |
| PR_L | MPNN | 80.19 ± 9.53 | | 85.43 ± 1.52 | | 72.28 ± 8.32 | | 156.7 |
| PR_L | PAGNN | 58.51 ± 4.87 | | 65.27 ± 1.70 | | 51.06 ± 2.54 | | 205.0 |
| PR_L | SAGE | 74.08 ± 1.27 | | 75.15 ± 1.72 | | 52.51 ± 1.30 | | 195.0 |

Lp Auroc Continued (↑)

| Loss Type | Model | CORA | | Citeseer | | Bitcoin Fraud Transaction | | Average Rank |
|---|---|---|---|---|---|---|---|---|
| PR_L + Triplet_L | ALL | 66.08 | ± | 60.47 | ± | 69.42 | ± | 182.0 |
| | | 3.70 | | 2.65 | | 1.45 | | |
| PR_L + Triplet_L | GAT | 91.67 | ± | 90.09 | ± | 82.27 | ± | 130.0 |
| | | 4.02 | | 2.25 | | 4.86 | | |
| PR_L + Triplet_L | GCN | 93.52 | ± | 94.41 | ± | 84.62 | ± | 112.3 |
| | | 0.94 | | 1.10 | | 1.80 | | |
| PR_L + Triplet_L | GIN | 72.49 | ± | 86.99 | ± | 55.88 | ± | 184.0 |
| | | 1.77 | | 2.01 | | 0.90 | | |
| PR_L + Triplet_L | MPNN | 80.74 | ± | 86.16 | ± | 75.31 | ± | 151.7 |
| | | 4.55 | | 1.85 | | 7.67 | | |
| PR_L + Triplet_L | PAGNN | 60.68 | ± | 72.68 | ± | 56.85 | ± | 196.7 |
| | | 3.31 | | 2.37 | | 0.58 | | |
| PR_L + Triplet_L | SAGE | 78.37 | ± | 84.30 | ± | 58.97 | ± | 174.3 |
| | | 2.70 | | 12.06 | | 0.50 | | |
| Triplet_L | ALL | 98.81 | ± | 99.73 | ± | 94.81 | ± | 39.3 |
| | | 0.19 | | 0.02 | | 0.48 | | |
| Triplet_L | GAT | 99.83 | ± | 99.95 | ± | 97.31 | ± | 2.3 |
| | | 0.03 | | 0.01 | | 0.24 | | |
| Triplet_L | GCN | 99.54 | ± | 99.90 | ± | 96.56 | ± | 8.3 |
| | | 0.04 | | 0.03 | | 0.41 | | |
| Triplet_L | GIN | 99.06 | ± | 99.71 | ± | 93.07 | ± | 43.3 |
| | | 0.27 | | 0.07 | | 0.97 | | |
| Triplet_L | MPNN | 98.78 | ± | 99.76 | ± | 96.26 | ± | 31.3 |
| | | 0.29 | | 0.06 | | 0.28 | | |
| Triplet_L | PAGNN | 95.68 | ± | 99.48 | ± | 58.61 | ± | 110.7 |
| | | 4.73 | | 0.04 | | 0.53 | | |
| Triplet_L | SAGE | 99.56 | ± | 99.89 | ± | 95.29 | ± | 12.0 |
| | | 0.09 | | 0.04 | | 0.29 | | |

Table 8. Lp F1 Performance (↑): Top-ranked results are highlighted in **1st**, second-ranked in **2nd**, and third-ranked in **3rd**.

| Loss Type | Model | CORA | | Citeseer | | Bitcoin Fraud Transaction | | Average Rank |
|---|---|---|---|---|---|---|---|---|
| Contr_l | ALL | 91.00 | ± | 94.10 | ± | 79.41 | ± | 85.3 |
| | | 0.39 | | 0.77 | | 1.08 | | |
| Contr_l | GAT | 96.41 | ± | 99.02 | ± | 87.01 | ± | 10.0 |
| | | 0.15 | | 0.06 | | 0.50 | | |
| Contr_l | GCN | 95.44 | ± | 98.23 | ± | 83.49 | ± | 35.0 |
| | | 0.16 | | 0.25 | | 0.63 | | |
| Contr_l | GIN | 93.13 | ± | 97.30 | ± | 79.90 | ± | 70.0 |
| | | 0.32 | | 0.14 | | 3.41 | | |
| Contr_l | MPNN | 93.14 | ± | 97.23 | ± | 78.58 | ± | 73.0 |
| | | 0.31 | | 0.30 | | 0.71 | | |
| Contr_l | PAGNN | 91.49 | ± | 93.74 | ± | 59.11 | ± | 113.7 |
| | | 0.65 | | 0.73 | | 0.16 | | |
| Contr_l | SAGE | 94.14 | ± | 97.54 | ± | 76.52 | ± | 72.7 |
| | | 0.16 | | 0.27 | | 1.50 | | |
| Contr_l + CrossE_L | ALL | 90.91 | ± | 94.45 | ± | 77.79 | ± | 89.0 |
| | | 0.55 | | 0.38 | | 2.33 | | |
| Contr_l + CrossE_L | GAT | 96.16 | ± | 98.99 | ± | 88.98 | ± | 12.0 |
| | | 0.22 | | 0.08 | | 2.69 | | |
| Contr_l + CrossE_L | GCN | 95.53 | ± | 98.24 | ± | 83.68 | ± | 34.0 |
| | | 0.33 | | 0.20 | | 0.61 | | |
| Contr_l + CrossE_L | GIN | 92.60 | ± | 96.73 | ± | 77.96 | ± | 78.0 |
| | | 0.55 | | 0.27 | | 2.76 | | |
| Contr_l + CrossE_L | MPNN | 93.11 | ± | 97.37 | ± | 80.87 | ± | 67.0 |
| | | 0.69 | | 0.13 | | 0.78 | | |
| Contr_l + CrossE_L | PAGNN | 91.11 | ± | 93.49 | ± | 59.31 | ± | 114.0 |
| | | 0.76 | | 0.39 | | 0.09 | | |
| Contr_l + CrossE_L | SAGE | 94.32 | ± | 97.76 | ± | 77.46 | ± | 69.7 |
| | | 0.43 | | 0.28 | | 2.89 | | |
| Contr_l + CrossE_L + PMI_L | ALL | 85.63 | ± | 82.17 | ± | 68.00 | ± | 134.7 |
| | | 2.24 | | 4.32 | | 1.34 | | |
| Contr_l + CrossE_L + PMI_L | GAT | 96.77 | ± | 98.83 | ± | 78.95 | ± | 31.7 |
| | | 0.23 | | 0.12 | | 1.67 | | |

Continued on next page

Lp F1 Continued (↑)

| Loss Type | Model | CORA | | | Citeseer | | | Bitcoin Fraud Transaction | | | Average Rank |
|---|---|---|---|---|---|---|---|---|---|---|---|
| Contr_l + CrossE_L + PMI_L | GCN | 95.69 | ± | | 97.92 | ± | | 81.81 | ± | | 46.7 |
| | | 0.51 | | | 0.19 | | | 0.46 | | | |
| Contr_l + CrossE_L + PMI_L | GIN | 85.52 | ± | | 84.66 | ± | | 58.04 | ± | | 147.3 |
| | | 2.11 | | | 1.45 | | | 1.75 | | | |
| Contr_l + CrossE_L + PMI_L | MPNN | 95.43 | ± | | 98.06 | ± | | 80.33 | ± | | 50.3 |
| | | 0.34 | | | 0.31 | | | 0.99 | | | |
| Contr_l + CrossE_L + PMI_L | PAGNN | 69.90 | ± | | 76.43 | ± | | 55.34 | ± | | 186.7 |
| | | 0.76 | | | 0.44 | | | 0.37 | | | |
| Contr_l + CrossE_L + PMI_L | SAGE | 79.36 | ± | | 82.36 | ± | | 58.89 | ± | | 154.0 |
| | | 0.95 | | | 2.98 | | | 1.48 | | | |
| Contr_l + CrossE_L + PMI_L + PR_L | ALL | 73.89 | ± | | 77.69 | ± | | 55.14 | ± | | 175.0 |
| | | 1.71 | | | 1.31 | | | 0.24 | | | |
| Contr_l + CrossE_L + PMI_L + PR_L | GAT | 96.07 | ± | | 98.53 | ± | | 64.93 | ± | | 58.0 |
| | | 0.80 | | | 0.13 | | | 8.64 | | | |
| Contr_l + CrossE_L + PMI_L + PR_L | GCN | 95.65 | ± | | 98.03 | ± | | 81.71 | ± | | 45.3 |
| | | 0.31 | | | 0.23 | | | 1.11 | | | |
| Contr_l + CrossE_L + PMI_L + PR_L | GIN | 84.41 | ± | | 81.78 | ± | | 57.43 | ± | | 155.0 |
| | | 1.75 | | | 1.36 | | | 2.02 | | | |
| Contr_l + CrossE_L + PMI_L + PR_L | MPNN | 94.65 | ± | | 94.10 | ± | | 60.53 | ± | | 101.3 |
| | | 0.30 | | | 3.34 | | | 11.01 | | | |
| Contr_l + CrossE_L + PMI_L + PR_L | PAGNN | 71.87 | ± | | 76.75 | ± | | 54.50 | ± | | 184.7 |
| | | 1.90 | | | 0.79 | | | 0.43 | | | |
| Contr_l + CrossE_L + PMI_L + PR_L | SAGE | 78.77 | ± | | 80.94 | ± | | 57.91 | ± | | 159.3 |
| | | 1.69 | | | 3.32 | | | 0.55 | | | |
| Contr_l + CrossE_L + PMI_L + PR_L + Triplet_L | ALL | 87.93 | ± | | 86.76 | ± | | 64.98 | ± | | 127.7 |
| | | 0.35 | | | 0.78 | | | 1.85 | | | |
| Contr_l + CrossE_L + PMI_L + PR_L + Triplet_L | GAT | 96.38 | ± | | 98.37 | ± | | 78.86 | ± | | 36.3 |
| | | 0.37 | | | 0.62 | | | 5.22 | | | |
| Contr_l + CrossE_L + PMI_L + PR_L + Triplet_L | GCN | 96.16 | ± | | 98.16 | ± | | 83.15 | ± | | 29.7 |
| | | 0.12 | | | 0.12 | | | 0.24 | | | |
| Contr_l + CrossE_L + PMI_L + PR_L + Triplet_L | GIN | 85.15 | ± | | 88.11 | ± | | 64.68 | ± | | 133.3 |
| | | 1.84 | | | 0.89 | | | 0.69 | | | |

Lp F1 Continued (↑)

| Loss Type | Model | CORA | | Citeseer | | Bitcoin Fraud Transaction | | Average Rank |
|---|---|---|---|---|---|---|---|---|
| Contr_l + CrossE_L + PMI_L + PR_L + Triplet_L | MPNN | 94.77 0.85 | ± | 94.91 1.68 | ± | 77.05 8.24 | ± | 79.3 |
| Contr_l + CrossE_L + PMI_L + PR_L + Triplet_L | PAGNN | 73.10 1.60 | ± | 77.36 0.24 | ± | 55.93 0.14 | ± | 175.3 |
| Contr_l + CrossE_L + PMI_L + PR_L + Triplet_L | SAGE | 88.71 2.14 | ± | 94.86 0.63 | ± | 59.80 0.89 | ± | 114.7 |
| Contr_l + CrossE_L + PMI_L + Triplet_L | ALL | 91.80 0.44 | ± | 96.74 0.24 | ± | 79.07 0.76 | ± | 76.3 |
| Contr_l + CrossE_L + PMI_L + Triplet_L | GAT | 97.11 0.46 | ± | 98.87 0.08 | ± | 83.16 0.49 | ± | 18.0 |
| Contr_l + CrossE_L + PMI_L + Triplet_L | GCN | 96.23 0.31 | ± | 98.09 0.20 | ± | 82.94 1.01 | ± | 31.0 |
| Contr_l + CrossE_L + PMI_L + Triplet_L | GIN | 86.21 2.53 | ± | 88.74 0.62 | ± | 64.79 2.61 | ± | 129.0 |
| Contr_l + CrossE_L + PMI_L + Triplet_L | MPNN | 95.39 0.75 | ± | 97.82 0.51 | ± | 81.64 1.38 | ± | 52.0 |
| Contr_l + CrossE_L + PMI_L + Triplet_L | PAGNN | 72.84 0.85 | ± | 76.91 0.50 | ± | 56.10 0.17 | ± | 176.0 |
| Contr_l + CrossE_L + PMI_L + Triplet_L | SAGE | 90.01 0.63 | ± | 93.95 0.81 | ± | 71.57 1.89 | ± | 100.0 |
| Contr_l + CrossE_L + PR_L | ALL | 68.13 0.32 | ± | 70.24 4.03 | ± | 67.40 0.65 | ± | 172.3 |
| Contr_l + CrossE_L + PR_L | GAT | 89.25 4.59 | ± | 91.45 3.23 | ± | 73.93 2.37 | ± | 102.3 |
| Contr_l + CrossE_L + PR_L | GCN | 89.74 0.91 | ± | 91.34 0.99 | ± | 73.08 0.71 | ± | 102.0 |
| Contr_l + CrossE_L + PR_L | GIN | 70.88 2.75 | ± | 82.71 1.85 | ± | 54.24 1.95 | ± | 178.0 |
| Contr_l + CrossE_L + PR_L | MPNN | 76.89 2.12 | ± | 86.83 0.75 | ± | 61.81 10.67 | ± | 144.7 |
| Contr_l + CrossE_L + PR_L | PAGNN | 68.22 0.13 | ± | 69.04 3.81 | ± | 56.03 0.99 | ± | 192.0 |

Lp F1 Continued (↑)

| Loss Type | Model | CORA | | Citeseer | | Bitcoin Fraud Transaction | | Average Rank |
|---|---|---|---|---|---|---|---|---|
| Contr_l + CrossE_L + PR_L | SAGE | 87.02 | ± | 88.22 | ± | 57.66 | ± | 139.3 |
| | | 8.20 | | 12.71 | | 1.43 | | |
| Contr_l + CrossE_L + PR_L + Triplet_L | ALL | 84.23 | ± | 88.88 | ± | 74.34 | ± | 120.7 |
| | | 0.98 | | 1.04 | | 1.68 | | |
| Contr_l + CrossE_L + PR_L + Triplet_L | GAT | 95.39 | ± | 97.15 | ± | 83.24 | ± | 51.0 |
| | | 0.55 | | 1.35 | | 2.24 | | |
| Contr_l + CrossE_L + PR_L + Triplet_L | GCN | 93.76 | ± | 95.43 | ± | 79.95 | ± | 74.0 |
| | | 0.77 | | 0.83 | | 2.95 | | |
| Contr_l + CrossE_L + PR_L + Triplet_L | GIN | 87.47 | ± | 91.45 | ± | 72.17 | ± | 109.3 |
| | | 0.68 | | 0.77 | | 9.92 | | |
| Contr_l + CrossE_L + PR_L + Triplet_L | MPNN | 87.52 | ± | 92.55 | ± | 71.97 | ± | 107.7 |
| | | 1.15 | | 0.40 | | 5.96 | | |
| Contr_l + CrossE_L + PR_L + Triplet_L | PAGNN | 74.05 | ± | 83.42 | ± | 58.69 | ± | 157.3 |
| | | 3.01 | | 3.90 | | 0.16 | | |
| Contr_l + CrossE_L + PR_L + Triplet_L | SAGE | 93.63 | ± | 95.87 | ± | 68.96 | ± | 88.7 |
| | | 0.50 | | 0.71 | | 2.24 | | |
| Contr_l + CrossE_L + Triplet_L | ALL | 93.47 | ± | 96.67 | ± | 83.93 | ± | 59.3 |
| | | 0.36 | | 0.24 | | 1.52 | | |
| Contr_l + CrossE_L + Triplet_L | GAT | 97.48 | ± | 99.29 | ± | 92.17 | ± | 2.7 |
| | | 0.37 | | 0.11 | | 1.28 | | |
| Contr_l + CrossE_L + Triplet_L | GCN | 96.33 | ± | 98.58 | ± | 85.85 | ± | 18.3 |
| | | 0.17 | | 0.16 | | 0.66 | | |
| Contr_l + CrossE_L + Triplet_L | GIN | 94.98 | ± | 97.82 | ± | 82.10 | ± | 53.7 |
| | | 0.52 | | 0.43 | | 2.81 | | |
| Contr_l + CrossE_L + Triplet_L | MPNN | 94.85 | ± | 98.18 | ± | 84.47 | ± | 39.0 |
| | | 0.43 | | 0.16 | | 1.07 | | |
| Contr_l + CrossE_L + Triplet_L | PAGNN | 90.16 | ± | 95.77 | ± | 59.39 | ± | 109.3 |
| | | 4.44 | | 0.28 | | 0.18 | | |
| Contr_l + CrossE_L + Triplet_L | SAGE | 95.89 | ± | 98.32 | ± | 84.86 | ± | 25.0 |
| | | 0.43 | | 0.17 | | 1.26 | | |
| Contr_l + PMI_L | ALL | 85.64 | ± | 90.96 | ± | 71.46 | ± | 115.0 |
| | | 2.91 | | 1.49 | | 1.68 | | |

Lp F1 Continued (↑)

| Loss Type | Model | CORA | | | Citeseer | | | Bitcoin Fraud Transaction | | | Average Rank |
|---|---|---|---|---|---|---|---|---|---|---|---|
| Contr_l + PMI_L | GAT | 97.08 | ± | 0.25 | 98.77 | ± | 0.13 | 79.30 | ± | 1.63 | 31.0 |
| Contr_l + PMI_L | GCN | 95.76 | ± | 0.34 | 97.93 | ± | 0.15 | 82.42 | ± | 0.82 | 43.3 |
| Contr_l + PMI_L | GIN | 84.66 | ± | 3.02 | 86.05 | ± | 1.41 | 59.04 | ± | 1.26 | 144.0 |
| Contr_l + PMI_L | MPNN | 95.30 | ± | 0.27 | 97.92 | ± | 0.33 | 80.57 | ± | 0.48 | 54.0 |
| Contr_l + PMI_L | PAGNN | 70.98 | ± | 1.16 | 76.44 | ± | 0.33 | 55.41 | ± | 0.32 | 183.7 |
| Contr_l + PMI_L | SAGE | 80.74 | ± | 2.07 | 89.64 | ± | 2.17 | 58.64 | ± | 0.72 | 143.0 |
| Contr_l + PMI_L + PR_L | ALL | 76.43 | ± | 1.63 | 77.30 | ± | 1.00 | 56.36 | ± | 0.64 | 168.7 |
| Contr_l + PMI_L + PR_L | GAT | 95.83 | ± | 0.34 | 96.32 | ± | 2.44 | 60.80 | ± | 7.30 | 84.7 |
| Contr_l + PMI_L + PR_L | GCN | 96.02 | ± | 0.26 | 98.01 | ± | 0.14 | 82.33 | ± | 0.49 | 38.3 |
| Contr_l + PMI_L + PR_L | GIN | 85.64 | ± | 2.87 | 82.47 | ± | 2.36 | 57.63 | ± | 2.19 | 150.0 |
| Contr_l + PMI_L + PR_L | MPNN | 95.21 | ± | 0.37 | 92.26 | ± | 3.13 | 60.94 | ± | 10.90 | 99.7 |
| Contr_l + PMI_L + PR_L | PAGNN | 70.88 | ± | 1.04 | 76.36 | ± | 0.22 | 54.57 | ± | 0.32 | 188.0 |
| Contr_l + PMI_L + PR_L | SAGE | 80.01 | ± | 1.25 | 87.33 | ± | 1.54 | 58.46 | ± | 0.64 | 148.3 |
| Contr_l + PMI_L + PR_L + Triplet_L | ALL | 88.84 | ± | 0.91 | 90.53 | ± | 0.82 | 68.25 | ± | 2.01 | 112.3 |
| Contr_l + PMI_L + PR_L + Triplet_L | GAT | 96.01 | ± | 0.39 | 98.20 | ± | 0.34 | 76.86 | ± | 4.74 | 46.7 |
| Contr_l + PMI_L + PR_L + Triplet_L | GCN | 95.87 | ± | 0.28 | 98.03 | ± | 0.18 | 82.87 | ± | 1.02 | 38.7 |

Lp F1 Continued (↑)

| Loss Type | Model | CORA | | Citeseer | | Bitcoin Fraud Transaction | | Average Rank |
|---|---|---|---|---|---|---|---|---|
| Contr_l + PMI_L + PR_L + Triplet_L | GIN | 89.08 1.02 | ± | 90.64 1.47 | ± | 66.45 1.05 | ± | 116.0 |
| Contr_l + PMI_L + PR_L + Triplet_L | MPNN | 94.64 0.29 | ± | 94.29 0.59 | ± | 69.80 6.01 | ± | 89.7 |
| Contr_l + PMI_L + PR_L + Triplet_L | PAGNN | 73.87 0.46 | ± | 79.99 2.02 | ± | 57.62 0.35 | ± | 166.3 |
| Contr_l + PMI_L + PR_L + Triplet_L | SAGE | 92.19 0.72 | ± | 96.43 0.43 | ± | 61.33 1.59 | ± | 99.7 |
| Contr_l + PR_L | ALL | 69.36 1.26 | ± | 67.75 3.70 | ± | 67.42 1.20 | ± | 171.7 |
| Contr_l + PR_L | GAT | 89.49 3.40 | ± | 90.50 1.03 | ± | 74.52 2.38 | ± | 103.7 |
| Contr_l + PR_L | GCN | 89.58 0.46 | ± | 92.23 1.28 | ± | 78.26 4.16 | ± | 94.7 |
| Contr_l + PR_L | GIN | 70.03 1.23 | ± | 80.27 2.57 | ± | 52.84 1.32 | ± | 185.7 |
| Contr_l + PR_L | MPNN | 76.59 2.86 | ± | 87.83 1.04 | ± | 55.12 0.46 | ± | 161.3 |
| Contr_l + PR_L | PAGNN | 68.13 0.10 | ± | 69.45 2.27 | ± | 56.12 1.02 | ± | 191.0 |
| Contr_l + PR_L | SAGE | 86.89 8.74 | ± | 89.61 11.78 | ± | 58.16 1.58 | ± | 136.0 |
| Contr_l + PR_L + Triplet_L | ALL | 80.80 5.45 | ± | 85.74 4.38 | ± | 74.36 0.42 | ± | 127.3 |
| Contr_l + PR_L + Triplet_L | GAT | 94.81 2.04 | ± | 97.18 1.00 | ± | 83.14 3.38 | ± | 56.3 |
| Contr_l + PR_L + Triplet_L | GCN | 92.98 0.56 | ± | 95.18 0.91 | ± | 78.17 3.05 | ± | 81.3 |
| Contr_l + PR_L + Triplet_L | GIN | 87.82 1.08 | ± | 90.47 1.32 | ± | 67.45 7.91 | ± | 118.0 |
| Contr_l + PR_L + Triplet_L | MPNN | 88.45 3.02 | ± | 91.63 1.34 | ± | 67.53 5.02 | ± | 112.7 |

Lp F1 Continued (↑)

| Loss Type | Model | CORA | | Citeseer | | Bitcoin Fraud Transaction | | Average Rank |
|---|---|---|---|---|---|---|---|---|
| Contr_l + PR_L + Triplet_L | PAGNN | 71.01 ± 1.43 | | 83.86 ± 5.13 | | 58.57 ± 0.24 | | 162.3 |
| Contr_l + PR_L + Triplet_L | SAGE | 93.21 ± 0.42 | | 96.43 ± 0.47 | | 72.31 ± 2.42 | | 84.7 |
| Contr_l + Triplet_L | ALL | 93.87 ± 0.41 | | 96.90 ± 0.35 | | 85.99 ± 0.72 | | 53.0 |
| Contr_l + Triplet_L | GAT | 97.31 ± 0.37 | | 99.25 ± 0.12 | | 92.47 ± 1.39 | | 3.7 |
| Contr_l + Triplet_L | GCN | 96.38 ± 0.16 | | 98.63 ± 0.16 | | 85.77 ± 0.75 | | 18.0 |
| Contr_l + Triplet_L | GIN | 94.99 ± 0.63 | | 97.81 ± 0.48 | | 80.46 ± 2.57 | | 58.7 |
| Contr_l + Triplet_L | MPNN | 95.04 ± 0.48 | | 98.30 ± 0.19 | | 85.11 ± 1.00 | | 34.3 |
| Contr_l + Triplet_L | PAGNN | 90.53 ± 5.54 | | 96.01 ± 0.29 | | 59.63 ± 0.37 | | 107.3 |
| Contr_l + Triplet_L | SAGE | 96.16 ± 0.16 | | 98.32 ± 0.38 | | 84.27 ± 3.65 | | 23.7 |
| CrossE_L | ALL | 84.11 ± 7.66 | | 76.83 ± 6.48 | | 67.40 ± 3.87 | | 147.3 |
| CrossE_L | GAT | 84.31 ± 10.49 | | 89.41 ± 14.00 | | 74.77 ± 13.01 | | 118.7 |
| CrossE_L | GCN | 67.86 ± 0.00 | | 64.55 ± 0.00 | | 51.46 ± 0.00 | | 208.7 |
| CrossE_L | GIN | 67.87 ± 0.01 | | 64.55 ± 0.00 | | 51.46 ± 0.00 | | 209.0 |
| CrossE_L | MPNN | 85.27 ± 1.82 | | 84.86 ± 5.13 | | 51.46 ± 0.00 | | 164.3 |
| CrossE_L | PAGNN | 69.40 ± 1.06 | | 70.57 ± 1.51 | | 51.47 ± 0.01 | | 199.0 |
| CrossE_L | SAGE | 74.97 ± 2.66 | | 69.38 ± 4.08 | | 51.46 ± 0.01 | | 192.0 |

Lp F1 Continued (↑)

| Loss Type | Model | CORA | | | Citeseer | | | Bitcoin Fraud Transaction | | | Average Rank |
|---|---|---|---|---|---|---|---|---|---|---|---|
| CrossE_L + PMI_L | ALL | 85.24 | ± | 0.96 | 81.68 | ± | 1.14 | 68.03 | ± | 2.25 | 136.0 |
| CrossE_L + PMI_L | GAT | 97.25 | ± | 0.14 | 98.98 | ± | 0.07 | 77.88 | ± | 1.05 | 29.3 |
| CrossE_L + PMI_L | GCN | 95.88 | ± | 0.42 | 98.09 | ± | 0.29 | 82.05 | ± | 0.67 | 38.7 |
| CrossE_L + PMI_L | GIN | 85.24 | ± | 2.25 | 82.99 | ± | 0.47 | 56.87 | ± | 0.57 | 152.3 |
| CrossE_L + PMI_L | MPNN | 95.17 | ± | 0.35 | 97.92 | ± | 0.35 | 80.87 | ± | 1.56 | 55.0 |
| CrossE_L + PMI_L | PAGNN | 71.16 | ± | 1.61 | 76.19 | ± | 0.31 | 54.84 | ± | 0.85 | 186.3 |
| CrossE_L + PMI_L | SAGE | 76.61 | ± | 1.01 | 75.64 | ± | 0.44 | 52.86 | ± | 1.32 | 185.3 |
| CrossE_L + PMI_L + PR_L | ALL | 73.27 | ± | 0.29 | 77.11 | ± | 0.46 | 53.12 | ± | 0.62 | 183.7 |
| CrossE_L + PMI_L + PR_L | GAT | 96.54 | ± | 0.37 | 95.03 | ± | 3.19 | 65.68 | ± | 8.82 | 76.7 |
| CrossE_L + PMI_L + PR_L | GCN | 95.91 | ± | 0.52 | 97.98 | ± | 0.27 | 81.14 | ± | 1.61 | 43.3 |
| CrossE_L + PMI_L + PR_L | GIN | 82.68 | ± | 1.66 | 81.28 | ± | 3.18 | 57.40 | ± | 0.95 | 157.7 |
| CrossE_L + PMI_L + PR_L | MPNN | 94.78 | ± | 1.04 | 94.26 | ± | 4.26 | 67.80 | ± | 12.20 | 91.7 |
| CrossE_L + PMI_L + PR_L | PAGNN | 71.91 | ± | 2.32 | 76.55 | ± | 0.32 | 54.07 | ± | 1.16 | 186.0 |
| CrossE_L + PMI_L + PR_L | SAGE | 76.49 | ± | 0.57 | 76.39 | ± | 1.55 | 56.95 | ± | 1.66 | 171.7 |
| CrossE_L + PMI_L + PR_L + Triplet_L | ALL | 87.29 | ± | 0.14 | 84.84 | ± | 0.88 | 66.57 | ± | 1.14 | 130.7 |
| CrossE_L + PMI_L + PR_L + Triplet_L | GAT | 96.30 | ± | 0.31 | 98.19 | ± | 0.37 | 76.20 | ± | 7.64 | 44.7 |

Lp F1 Continued (↑)

| Loss Type | Model | CORA | | Citeseer | | Bitcoin Fraud Transaction | | Average Rank |
|-----------|-------|------|---|----------|---|---------------------------|---|--------------|
| CrossE_L + PMI_L + PR_L + Triplet_L | GCN | 95.89 | ± | 98.19 | ± | 82.73 | ± | 34.0 |
| | | 0.37 | | 0.14 | | 0.98 | | |
| CrossE_L + PMI_L + PR_L + Triplet_L | GIN | 86.20 | ± | 87.32 | ± | 64.24 | ± | 132.0 |
| | | 1.17 | | 1.33 | | 1.18 | | |
| CrossE_L + PMI_L + PR_L + Triplet_L | MPNN | 95.24 | ± | 95.14 | ± | 77.17 | ± | 74.3 |
| | | 0.32 | | 2.48 | | 8.99 | | |
| CrossE_L + PMI_L + PR_L + Triplet_L | PAGNN | 74.63 | ± | 77.44 | ± | 56.04 | ± | 171.0 |
| | | 0.63 | | 0.60 | | 0.22 | | |
| CrossE_L + PMI_L + PR_L + Triplet_L | SAGE | 88.68 | ± | 93.63 | ± | 59.18 | ± | 120.7 |
| | | 3.62 | | 1.05 | | 0.28 | | |
| CrossE_L + PMI_L + Triplet_L | ALL | 93.30 | ± | 97.63 | ± | 81.29 | ± | 63.3 |
| | | 0.56 | | 0.31 | | 1.23 | | |
| CrossE_L + PMI_L + Triplet_L | GAT | 97.20 | ± | 98.89 | ± | 84.90 | ± | 12.7 |
| | | 0.22 | | 0.10 | | 1.18 | | |
| CrossE_L + PMI_L + Triplet_L | GCN | 95.72 | ± | 98.12 | ± | 83.15 | ± | 36.7 |
| | | 0.43 | | 0.09 | | 0.59 | | |
| CrossE_L + PMI_L + Triplet_L | GIN | 88.11 | ± | 90.51 | ± | 69.83 | ± | 112.0 |
| | | 1.21 | | 0.45 | | 1.77 | | |
| CrossE_L + PMI_L + Triplet_L | MPNN | 94.65 | ± | 97.80 | ± | 83.31 | ± | 52.3 |
| | | 0.64 | | 0.26 | | 0.40 | | |
| CrossE_L + PMI_L + Triplet_L | PAGNN | 74.83 | ± | 77.25 | ± | 56.42 | ± | 169.3 |
| | | 0.51 | | 0.80 | | 0.25 | | |
| CrossE_L + PMI_L + Triplet_L | SAGE | 91.95 | ± | 95.76 | ± | 75.75 | ± | 85.7 |
| | | 0.72 | | 0.87 | | 2.23 | | |
| CrossE_L + PR_L | ALL | 68.44 | ± | 65.41 | ± | 68.27 | ± | 170.7 |
| | | 1.13 | | 0.58 | | 0.93 | | |
| CrossE_L + PR_L | GAT | 88.30 | ± | 90.69 | ± | 69.43 | ± | 111.3 |
| | | 2.34 | | 1.00 | | 9.33 | | |
| CrossE_L + PR_L | GCN | 86.49 | ± | 89.57 | ± | 55.78 | ± | 145.0 |
| | | 1.43 | | 1.76 | | 4.44 | | |
| CrossE_L + PR_L | GIN | 69.42 | ± | 75.64 | ± | 53.81 | ± | 194.7 |
| | | 1.89 | | 4.12 | | 0.80 | | |

Lp F1 Continued (↑)

| Loss Type | | | Model | CORA | | Citeseer | | Bitcoin Fraud Transaction | | Average Rank |
|---|---|---|---|---|---|---|---|---|---|---|
| CrossE_L + PR_L | | | MPNN | 81.20 ± 7.25 | | 85.69 ± 1.04 | | 52.81 ± 0.98 | | 166.3 |
| CrossE_L + PR_L | | | PAGNN | 68.04 ± 0.10 | | 65.61 ± 0.90 | | 54.42 ± 0.17 | | 201.0 |
| CrossE_L + PR_L | | | SAGE | 69.87 ± 1.01 | | 66.98 ± 2.25 | | 51.82 ± 0.75 | | 200.3 |
| CrossE_L + PR_L + Triplet_L | | | ALL | 78.54 ± 6.43 | | 76.49 ± 4.57 | | 67.69 ± 6.76 | | 151.0 |
| CrossE_L + PR_L + Triplet_L | | | GAT | 91.29 ± 2.14 | | 96.57 ± 2.04 | | 77.00 ± 3.68 | | 83.3 |
| CrossE_L + PR_L + Triplet_L | | | GCN | 92.04 ± 1.20 | | 93.80 ± 1.01 | | 79.13 ± 2.47 | | 84.0 |
| CrossE_L + PR_L + Triplet_L | | | GIN | 84.54 ± 3.63 | | 89.63 ± 1.35 | | 54.72 ± 0.83 | | 151.7 |
| CrossE_L + PR_L + Triplet_L | | | MPNN | 81.07 ± 2.31 | | 89.67 ± 0.76 | | 63.93 ± 6.34 | | 134.7 |
| CrossE_L + PR_L + Triplet_L | | | PAGNN | 69.64 ± 1.89 | | 76.76 ± 3.26 | | 58.52 ± 0.15 | | 175.7 |
| CrossE_L + PR_L + Triplet_L | | | SAGE | 93.84 ± 0.87 | | 96.37 ± 0.95 | | 63.18 ± 6.64 | | 95.3 |
| CrossE_L + Triplet_L | | | ALL | 95.64 ± 0.46 | | 98.09 ± 0.19 | | 85.69 ± 1.63 | | 33.3 |
| CrossE_L + Triplet_L | | | GAT | 98.31 ± 0.25 | | 99.49 ± 0.08 | | 90.14 ± 1.99 | | 2.3 |
| CrossE_L + Triplet_L | | | GCN | 97.21 ± 0.36 | | 98.85 ± 0.20 | | 86.69 ± 0.45 | | 10.0 |
| CrossE_L + Triplet_L | | | GIN | 96.06 ± 0.58 | | 97.62 ± 0.29 | | 81.22 ± 1.96 | | 46.7 |
| CrossE_L + Triplet_L | | | MPNN | 96.58 ± 0.43 | | 98.91 ± 0.30 | | 86.80 ± 0.87 | | 10.7 |
| CrossE_L + Triplet_L | | | PAGNN | 91.43 ± 5.73 | | 97.14 ± 0.48 | | 59.29 ± 0.18 | | 103.3 |

Lp F1 Continued (↑)

| Loss Type | Model | CORA | | | Citeseer | | | Bitcoin Fraud Transaction | | | Average Rank |
|---|---|---|---|---|---|---|---|---|---|---|---|
| CrossE_L + Triplet_L | SAGE | 97.48 | ± | 0.28 | 98.78 | ± | 0.14 | 84.26 | ± | 0.99 | 13.3 |
| PMI_L | ALL | 81.48 | ± | 0.57 | 82.62 | ± | 1.42 | 69.10 | ± | 1.16 | 136.3 |
| PMI_L | GAT | 97.39 | ± | 0.14 | 98.84 | ± | 0.22 | 77.97 | ± | 1.46 | 30.0 |
| PMI_L | GCN | 95.89 | ± | 0.26 | 97.77 | ± | 0.39 | 81.80 | ± | 1.31 | 46.7 |
| PMI_L | GIN | 82.33 | ± | 1.95 | 83.21 | ± | 1.24 | 55.92 | ± | 1.84 | 159.7 |
| PMI_L | MPNN | 95.26 | ± | 0.21 | 98.07 | ± | 0.24 | 80.85 | ± | 1.06 | 50.7 |
| PMI_L | PAGNN | 70.88 | ± | 0.96 | 76.14 | ± | 0.32 | 55.39 | ± | 0.59 | 186.7 |
| PMI_L | SAGE | 76.06 | ± | 1.28 | 74.93 | ± | 0.69 | 54.09 | ± | 1.75 | 185.0 |
| PMI_L + PR_L | ALL | 73.72 | ± | 0.97 | 76.79 | ± | 0.97 | 53.36 | ± | 0.35 | 183.3 |
| PMI_L + PR_L | GAT | 95.56 | ± | 0.67 | 93.39 | ± | 5.83 | 55.60 | ± | 1.32 | 110.7 |
| PMI_L + PR_L | GCN | 95.73 | ± | 0.86 | 98.11 | ± | 0.12 | 81.97 | ± | 1.31 | 40.7 |
| PMI_L + PR_L | GIN | 83.02 | ± | 1.77 | 82.34 | ± | 3.47 | 58.12 | ± | 0.56 | 153.7 |
| PMI_L + PR_L | MPNN | 94.92 | ± | 0.57 | 89.76 | ± | 1.04 | 60.36 | ± | 10.68 | 108.0 |
| PMI_L + PR_L | PAGNN | 71.33 | ± | 0.46 | 75.85 | ± | 0.56 | 53.19 | ± | 1.17 | 190.3 |
| PMI_L + PR_L | SAGE | 76.95 | ± | 1.85 | 76.99 | ± | 0.98 | 56.04 | ± | 0.77 | 170.3 |
| PMI_L + PR_L + Triplet_L | ALL | 87.61 | ± | 0.43 | 86.94 | ± | 0.94 | 67.68 | ± | 2.72 | 124.3 |

Lp F1 Continued (↑)

| Loss Type | Model | CORA | | | Citeseer | | | Bitcoin Fraud Transaction | | | Average Rank |
|---|---|---|---|---|---|---|---|---|---|---|---|
| PMI_L + PR_L + Triplet_L | GAT | 96.05 | ± | 0.22 | 98.00 | ± | 0.44 | 71.97 | ± | 5.63 | 57.7 |
| PMI_L + PR_L + Triplet_L | GCN | 95.83 | ± | 0.33 | 97.91 | ± | 0.15 | 82.23 | ± | 0.59 | 44.3 |
| PMI_L + PR_L + Triplet_L | GIN | 87.16 | ± | 1.03 | 87.22 | ± | 0.56 | 64.46 | ± | 1.51 | 130.0 |
| PMI_L + PR_L + Triplet_L | MPNN | 94.91 | ± | 0.40 | 93.21 | ± | 0.65 | 72.74 | ± | 8.56 | 86.7 |
| PMI_L + PR_L + Triplet_L | PAGNN | 73.49 | ± | 1.21 | 78.04 | ± | 0.85 | 56.22 | ± | 0.19 | 170.7 |
| PMI_L + PR_L + Triplet_L | SAGE | 91.03 | ± | 3.04 | 95.15 | ± | 0.69 | 59.51 | ± | 0.28 | 108.7 |
| PMI_L + Triplet_L | ALL | 92.99 | ± | 0.50 | 97.26 | ± | 0.39 | 81.02 | ± | 0.70 | 68.0 |
| PMI_L + Triplet_L | GAT | 97.21 | ± | 0.23 | 98.92 | ± | 0.10 | 83.71 | ± | 1.21 | 14.0 |
| PMI_L + Triplet_L | GCN | 95.80 | ± | 0.26 | 98.08 | ± | 0.20 | 82.74 | ± | 0.55 | 38.7 |
| PMI_L + Triplet_L | GIN | 86.74 | ± | 2.05 | 88.99 | ± | 1.08 | 66.59 | ± | 1.62 | 125.7 |
| PMI_L + Triplet_L | MPNN | 95.17 | ± | 0.52 | 97.87 | ± | 0.31 | 82.63 | ± | 0.89 | 50.7 |
| PMI_L + Triplet_L | PAGNN | 72.84 | ± | 0.84 | 77.13 | ± | 0.37 | 56.19 | ± | 0.34 | 174.7 |
| PMI_L + Triplet_L | SAGE | 90.61 | ± | 1.53 | 95.34 | ± | 0.56 | 72.70 | ± | 1.61 | 92.7 |
| PR_L | ALL | 67.88 | ± | 0.02 | 64.78 | ± | 0.22 | 70.18 | ± | 1.57 | 172.0 |
| PR_L | GAT | 89.89 | ± | 1.71 | 89.98 | ± | 0.73 | 72.63 | ± | 2.95 | 106.0 |
| PR_L | GCN | 86.13 | ± | 0.89 | 89.56 | ± | 1.00 | 78.84 | ± | 1.15 | 109.0 |

Lp F1 Continued (↑)

| Loss Type | Model | CORA | | Citeseer | | Bitcoin Fraud Transaction | | Average Rank |
|---|---|---|---|---|---|---|---|---|
| PR_L | GIN | 68.40 | ± | 75.80 | ± | 55.03 | ± | 192.3 |
| | | 0.63 | | 3.81 | | 0.89 | | |
| PR_L | MPNN | 77.78 | ± | 85.23 | ± | 59.25 | ± | 150.3 |
| | | 8.11 | | 1.22 | | 10.26 | | |
| PR_L | PAGNN | 68.05 | ± | 65.06 | ± | 54.57 | ± | 201.0 |
| | | 0.10 | | 0.22 | | 0.08 | | |
| PR_L | SAGE | 70.15 | ± | 68.10 | ± | 52.39 | ± | 198.3 |
| | | 0.87 | | 1.63 | | 0.51 | | |
| PR_L + Triplet_L | ALL | 68.35 | ± | 65.60 | ± | 68.40 | ± | 170.7 |
| | | 0.83 | | 0.49 | | 1.40 | | |
| PR_L + Triplet_L | GAT | 89.67 | ± | 90.05 | ± | 73.14 | ± | 105.0 |
| | | 4.74 | | 1.55 | | 2.08 | | |
| PR_L + Triplet_L | GCN | 89.38 | ± | 91.34 | ± | 72.46 | ± | 105.0 |
| | | 0.49 | | 0.92 | | 0.83 | | |
| PR_L + Triplet_L | GIN | 68.82 | ± | 79.00 | ± | 52.88 | ± | 188.3 |
| | | 0.52 | | 2.01 | | 0.77 | | |
| PR_L + Triplet_L | MPNN | 77.87 | ± | 86.16 | ± | 62.80 | ± | 144.0 |
| | | 4.63 | | 1.16 | | 9.30 | | |
| PR_L + Triplet_L | PAGNN | 68.14 | ± | 68.20 | ± | 55.98 | ± | 193.0 |
| | | 0.09 | | 1.98 | | 0.97 | | |
| PR_L + Triplet_L | SAGE | 73.37 | ± | 79.58 | ± | 56.40 | ± | 169.7 |
| | | 2.17 | | 14.80 | | 0.70 | | |
| Triplet_L | ALL | 95.43 | ± | 98.08 | ± | 84.24 | ± | 38.0 |
| | | 0.44 | | 0.16 | | 1.06 | | |
| Triplet_L | GAT | 98.76 | ± | 99.61 | ± | 89.21 | ± | 2.0 |
| | | 0.24 | | 0.06 | | 0.89 | | |
| Triplet_L | GCN | 97.20 | ± | 98.90 | ± | 86.27 | ± | 10.7 |
| | | 0.15 | | 0.18 | | 0.94 | | |
| Triplet_L | GIN | 96.57 | ± | 97.98 | ± | 80.23 | ± | 41.0 |
| | | 0.66 | | 0.39 | | 1.40 | | |
| Triplet_L | MPNN | 95.83 | ± | 98.71 | ± | 86.59 | ± | 23.0 |
| | | 0.79 | | 0.23 | | 0.51 | | |

Lp F1 Continued (↑)

| Loss Type | Model | CORA | | Citeseer | | Bitcoin Fraud Transaction | | Average Rank |
|-----------|-------|------|---|----------|---|---------------------------|---|--------------|
| Triplet_L | PAGNN | 91.57 | ± | 97.32 | ± | 59.50 | ± | 99.7 |
| | | 5.86 | | 0.24 | | 0.22 | | |
| Triplet_L | SAGE | 97.27 | ± | 98.76 | ± | 84.15 | ± | 15.7 |
| | | 0.36 | | 0.28 | | 0.66 | | |

Table 9. Lp Precision Performance (↑): Top-ranked results are highlighted in **1st**, second-ranked in **2nd**, and third-ranked in **3rd**.

| Loss Type | Model | CORA | | Citeseer | | Bitcoin Fraud Transaction | | Average Rank |
|-----------|-------|------|---|----------|---|---------------------------|---|--------------|
| Contr_l | ALL | 88.04 | ± | 93.18 | ± | 77.81 | ± | 109.3 |
| | | 1.19 | | 1.73 | | 2.68 | | |
| Contr_l | GAT | 95.69 | ± | 98.72 | ± | 89.10 | ± | 19.0 |
| | | 0.45 | | 0.20 | | 0.93 | | |
| Contr_l | GCN | 95.69 | ± | 98.04 | ± | 85.66 | ± | 39.0 |
| | | 0.59 | | 0.20 | | 2.19 | | |
| Contr_l | GIN | 91.17 | ± | 96.69 | ± | 80.66 | ± | 86.3 |
| | | 0.29 | | 0.53 | | 4.09 | | |
| Contr_l | MPNN | 92.37 | ± | 96.51 | ± | 79.11 | ± | 87.0 |
| | | 0.88 | | 0.28 | | 1.15 | | |
| Contr_l | PAGNN | 89.69 | ± | 92.18 | ± | 44.23 | ± | 131.3 |
| | | 1.45 | | 1.62 | | 0.28 | | |
| Contr_l | SAGE | 93.26 | ± | 97.00 | ± | 73.16 | ± | 87.7 |
| | | 0.59 | | 0.38 | | 3.11 | | |
| Contr_l + CrossE_L | ALL | 88.38 | ± | 93.56 | ± | 74.54 | ± | 111.3 |
| | | 1.30 | | 0.86 | | 2.89 | | |
| Contr_l + CrossE_L | GAT | 95.81 | ± | 98.70 | ± | 90.09 | ± | 18.3 |
| | | 0.48 | | 0.16 | | 1.92 | | |
| Contr_l + CrossE_L | GCN | 95.20 | ± | 97.92 | ± | 86.60 | ± | 40.7 |
| | | 0.66 | | 0.30 | | 1.52 | | |

Lp Precision Continued (↑)

| Loss Type | Model | CORA | | Citeseer | | Bitcoin Fraud Transaction | | Average Rank |
|---|---|---|---|---|---|---|---|---|
| Contr_l + CrossE_L | GIN | 91.36 | ± | 96.34 | ± | 77.70 | ± | 92.0 |
| | | 0.96 | | 0.33 | | 4.00 | | |
| Contr_l + CrossE_L | MPNN | 91.84 | ± | 96.93 | ± | 81.13 | ± | 82.3 |
| | | 0.45 | | 0.26 | | 2.31 | | |
| Contr_l + CrossE_L | PAGNN | 88.69 | ± | 91.18 | ± | 44.06 | ± | 136.3 |
| | | 0.84 | | 0.40 | | 0.10 | | |
| Contr_l + CrossE_L | SAGE | 93.56 | ± | 97.37 | ± | 75.24 | ± | 82.7 |
| | | 0.39 | | 0.21 | | 3.34 | | |
| Contr_l + CrossE_L + PMI_L | ALL | 83.60 | ± | 83.65 | ± | 67.37 | ± | 138.7 |
| | | 3.89 | | 2.88 | | 1.31 | | |
| Contr_l + CrossE_L + PMI_L | GAT | 96.92 | ± | 98.58 | ± | 80.14 | ± | 35.3 |
| | | 0.35 | | 0.23 | | 3.10 | | |
| Contr_l + CrossE_L + PMI_L | GCN | 96.44 | ± | 98.13 | ± | 84.88 | ± | 34.0 |
| | | 0.70 | | 0.45 | | 1.10 | | |
| Contr_l + CrossE_L + PMI_L | GIN | 83.45 | ± | 85.04 | ± | 48.59 | ± | 148.0 |
| | | 2.61 | | 3.18 | | 2.41 | | |
| Contr_l + CrossE_L + PMI_L | MPNN | 94.03 | ± | 97.43 | ± | 83.79 | ± | 63.7 |
| | | 0.60 | | 0.54 | | 1.90 | | |
| Contr_l + CrossE_L + PMI_L | PAGNN | 57.10 | ± | 67.02 | ± | 41.54 | ± | 190.7 |
| | | 3.19 | | 1.24 | | 0.23 | | |
| Contr_l + CrossE_L + PMI_L | SAGE | 75.49 | ± | 82.13 | ± | 44.60 | ± | 158.0 |
| | | 2.11 | | 2.79 | | 3.24 | | |
| Contr_l + CrossE_L + PMI_L + PR_L | ALL | 63.50 | ± | 82.13 | ± | 52.57 | ± | 161.3 |
| | | 4.18 | | 2.85 | | 0.91 | | |
| Contr_l + CrossE_L + PMI_L + PR_L | GAT | 96.00 | ± | 98.28 | ± | 61.42 | ± | 64.7 |
| | | 0.58 | | 0.26 | | 18.61 | | |
| Contr_l + CrossE_L + PMI_L + PR_L | GCN | 95.88 | ± | 98.18 | ± | 84.09 | ± | 40.0 |
| | | 0.46 | | 0.33 | | 1.86 | | |
| Contr_l + CrossE_L + PMI_L + PR_L | GIN | 82.47 | ± | 81.17 | ± | 42.49 | ± | 164.7 |
| | | 1.82 | | 0.70 | | 0.81 | | |
| Contr_l + CrossE_L + PMI_L + PR_L | MPNN | 93.69 | ± | 93.92 | ± | 50.47 | ± | 113.7 |
| | | 0.44 | | 3.34 | | 20.02 | | |

Lp Precision Continued (↑)

| Loss Type | Model | CORA | | Citeseer | | Bitcoin Fraud Transaction | | Average Rank |
|---|---|---|---|---|---|---|---|---|
| Contr_l + CrossE_L + PMI_L + PR_L | PAGNN | 62.29 4.19 | ± | 66.84 0.92 | ± | 38.61 0.24 | ± | 191.0 |
| Contr_l + CrossE_L + PMI_L + PR_L | SAGE | 74.19 2.79 | ± | 79.89 2.84 | ± | 43.07 0.58 | ± | 167.0 |
| Contr_l + CrossE_L + PMI_L + PR_L + Triplet_L | ALL | 84.02 1.63 | ± | 84.50 1.46 | ± | 72.28 1.23 | ± | 132.3 |
| Contr_l + CrossE_L + PMI_L + PR_L + Triplet_L | GAT | 96.43 0.36 | ± | 98.11 0.48 | ± | 80.36 8.68 | ± | 47.0 |
| Contr_l + CrossE_L + PMI_L + PR_L + Triplet_L | GCN | 96.79 0.24 | ± | 98.28 0.18 | ± | 85.70 1.43 | ± | 24.3 |
| Contr_l + CrossE_L + PMI_L + PR_L + Triplet_L | GIN | 83.55 3.02 | ± | 86.25 1.22 | ± | 62.34 3.38 | ± | 137.7 |
| Contr_l + CrossE_L + PMI_L + PR_L + Triplet_L | MPNN | 93.70 0.92 | ± | 95.05 1.35 | ± | 81.76 8.71 | ± | 86.0 |
| Contr_l + CrossE_L + PMI_L + PR_L + Triplet_L | PAGNN | 64.59 3.01 | ± | 67.92 0.64 | ± | 41.76 0.18 | ± | 183.0 |
| Contr_l + CrossE_L + PMI_L + PR_L + Triplet_L | SAGE | 87.24 3.19 | ± | 93.69 0.58 | ± | 44.23 0.57 | ± | 134.7 |
| Contr_l + CrossE_L + PMI_L + Triplet_L | ALL | 90.58 0.79 | ± | 96.59 0.35 | ± | 79.75 1.74 | ± | 90.3 |
| Contr_l + CrossE_L + PMI_L + Triplet_L | GAT | 96.99 0.58 | ± | 98.70 0.06 | ± | 85.76 1.76 | ± | 15.0 |
| Contr_l + CrossE_L + PMI_L + Triplet_L | GCN | 96.32 0.66 | ± | 98.37 0.17 | ± | 87.26 1.97 | ± | 22.0 |
| Contr_l + CrossE_L + PMI_L + Triplet_L | GIN | 84.88 1.63 | ± | 86.49 1.62 | ± | 62.13 4.77 | ± | 134.0 |
| Contr_l + CrossE_L + PMI_L + Triplet_L | MPNN | 93.91 1.02 | ± | 97.20 0.67 | ± | 83.20 2.40 | ± | 69.3 |
| Contr_l + CrossE_L + PMI_L + Triplet_L | PAGNN | 66.86 2.57 | ± | 67.12 1.37 | ± | 42.21 0.49 | ± | 181.7 |
| Contr_l + CrossE_L + PMI_L + Triplet_L | SAGE | 88.80 1.51 | ± | 93.02 0.91 | ± | 67.35 2.62 | ± | 118.0 |

Lp Precision Continued (↑)

| Loss Type | Model | CORA | | Citeseer | | Bitcoin Fraud Transaction | | Average Rank |
|---|---|---|---|---|---|---|---|---|
| Contr_l + CrossE_L + PR_L | ALL | 53.87 ± 3.47 | | 56.35 ± 6.60 | | 87.13 ± 1.62 | | 139.0 |
| Contr_l + CrossE_L + PR_L | GAT | 94.81 ± 1.76 | | 97.87 ± 0.15 | | 81.96 ± 1.80 | | 59.3 |
| Contr_l + CrossE_L + PR_L | GCN | 93.59 ± 0.65 | | 96.08 ± 0.95 | | 80.70 ± 2.23 | | 84.7 |
| Contr_l + CrossE_L + PR_L | GIN | 72.59 ± 10.47 | | 87.19 ± 2.97 | | 54.11 ± 11.11 | | 145.7 |
| Contr_l + CrossE_L + PR_L | MPNN | 87.47 ± 1.88 | | 96.17 ± 1.05 | | 55.27 ± 17.96 | | 118.3 |
| Contr_l + CrossE_L + PR_L | PAGNN | 51.84 ± 0.16 | | 59.23 ± 8.55 | | 42.22 ± 1.09 | | 195.3 |
| Contr_l + CrossE_L + PR_L | SAGE | 87.64 ± 9.69 | | 88.08 ± 14.76 | | 42.42 ± 1.12 | | 145.7 |
| Contr_l + CrossE_L + PR_L + Triplet_L | ALL | 80.86 ± 1.84 | | 88.58 ± 1.14 | | 70.04 ± 4.06 | | 131.3 |
| Contr_l + CrossE_L + PR_L + Triplet_L | GAT | 96.79 ± 0.73 | | 98.41 ± 0.60 | | 86.12 ± 2.02 | | 20.0 |
| Contr_l + CrossE_L + PR_L + Triplet_L | GCN | 95.71 ± 0.83 | | 97.41 ± 0.65 | | 83.90 ± 3.95 | | 53.3 |
| Contr_l + CrossE_L + PR_L + Triplet_L | GIN | 89.03 ± 1.87 | | 94.33 ± 1.46 | | 70.80 ± 10.30 | | 110.0 |
| Contr_l + CrossE_L + PR_L + Triplet_L | MPNN | 91.06 ± 1.30 | | 96.03 ± 0.49 | | 74.82 ± 8.70 | | 97.7 |
| Contr_l + CrossE_L + PR_L + Triplet_L | PAGNN | 63.16 ± 4.14 | | 80.35 ± 6.37 | | 42.67 ± 0.82 | | 175.7 |
| Contr_l + CrossE_L + PR_L + Triplet_L | SAGE | 94.73 ± 0.63 | | 96.83 ± 0.88 | | 59.11 ± 3.74 | | 90.0 |
| Contr_l + CrossE_L + Triplet_L | ALL | 92.04 ± 0.72 | | 96.25 ± 0.51 | | 83.41 ± 2.16 | | 79.7 |
| Contr_l + CrossE_L + Triplet_L | GAT | 96.96 ± 0.32 | | 99.03 ± 0.11 | | 92.60 ± 1.17 | | 5.3 |

Continued on next page

Lp Precision Continued (↑)

| Loss Type | Model | CORA | | Citeseer | | Bitcoin Fraud Transaction | | Average Rank |
|---|---|---|---|---|---|---|---|---|
| Contr_l + CrossE_L + Triplet_L | GCN | 96.14 ± 0.45 | | 98.20 ± 0.28 | | 87.68 ± 1.07 | | 24.3 |
| Contr_l + CrossE_L + Triplet_L | GIN | 93.60 ± 0.81 | | 97.43 ± 0.51 | | 82.10 ± 3.39 | | 72.0 |
| Contr_l + CrossE_L + Triplet_L | MPNN | 93.82 ± 0.38 | | 97.76 ± 0.20 | | 85.12 ± 1.56 | | 56.7 |
| Contr_l + CrossE_L + Triplet_L | PAGNN | 87.60 ± 5.81 | | 94.92 ± 0.56 | | 44.33 ± 0.18 | | 131.0 |
| Contr_l + CrossE_L + Triplet_L | SAGE | 95.54 ± 0.60 | | 97.95 ± 0.23 | | 84.59 ± 1.73 | | 45.3 |
| Contr_l + PMI_L | ALL | 84.77 ± 1.35 | | 90.82 ± 2.20 | | 69.60 ± 3.41 | | 125.7 |
| Contr_l + PMI_L | GAT | 97.19 ± 0.43 | | 98.64 ± 0.08 | | 79.83 ± 2.65 | | 33.0 |
| Contr_l + PMI_L | GCN | 96.13 ± 0.49 | | 98.08 ± 0.38 | | 85.32 ± 2.17 | | 36.7 |
| Contr_l + PMI_L | GIN | 82.75 ± 1.97 | | 86.41 ± 0.96 | | 49.96 ± 2.40 | | 145.3 |
| Contr_l + PMI_L | MPNN | 94.40 ± 0.73 | | 97.32 ± 0.63 | | 83.47 ± 2.01 | | 63.0 |
| Contr_l + PMI_L | PAGNN | 59.85 ± 5.03 | | 66.62 ± 0.41 | | 41.54 ± 0.30 | | 191.0 |
| Contr_l + PMI_L | SAGE | 77.09 ± 2.76 | | 86.97 ± 2.58 | | 43.51 ± 0.84 | | 154.0 |
| Contr_l + PMI_L + PR_L | ALL | 68.79 ± 2.69 | | 81.53 ± 2.74 | | 55.81 ± 1.27 | | 156.7 |
| Contr_l + PMI_L + PR_L | GAT | 95.75 ± 0.28 | | 96.22 ± 1.65 | | 50.83 ± 16.69 | | 91.3 |
| Contr_l + PMI_L + PR_L | GCN | 96.24 ± 0.54 | | 98.31 ± 0.28 | | 84.98 ± 1.18 | | 32.3 |
| Contr_l + PMI_L + PR_L | GIN | 84.05 ± 2.60 | | 83.62 ± 3.90 | | 43.15 ± 0.93 | | 155.3 |

Lp Precision Continued (↑)

| Loss Type | Model | CORA | | Citeseer | | Bitcoin Fraud Transaction | | Average Rank |
|---|---|---|---|---|---|---|---|---|
| Contr_l + PMI_L + PR_L | MPNN | 94.34 ± 0.62 | | 92.92 ± 2.48 | | 55.38 ± 20.05 | | 108.0 |
| Contr_l + PMI_L + PR_L | PAGNN | 60.27 ± 3.83 | | 66.26 ± 1.35 | | 38.69 ± 0.20 | | 192.7 |
| Contr_l + PMI_L + PR_L | SAGE | 75.96 ± 1.46 | | 84.00 ± 1.83 | | 43.63 ± 0.53 | | 159.0 |
| Contr_l + PMI_L + PR_L + Triplet_L | ALL | 86.08 ± 1.40 | | 88.34 ± 1.07 | | 69.60 ± 4.54 | | 126.7 |
| Contr_l + PMI_L + PR_L + Triplet_L | GAT | 96.05 ± 0.23 | | 98.17 ± 0.20 | | 77.00 ± 9.59 | | 53.3 |
| Contr_l + PMI_L + PR_L + Triplet_L | GCN | 96.05 ± 0.44 | | 97.93 ± 0.42 | | 86.78 ± 1.61 | | 34.3 |
| Contr_l + PMI_L + PR_L + Triplet_L | GIN | 87.74 ± 1.18 | | 88.87 ± 1.03 | | 67.84 ± 2.65 | | 122.7 |
| Contr_l + PMI_L + PR_L + Triplet_L | MPNN | 93.87 ± 0.33 | | 94.16 ± 0.87 | | 72.68 ± 6.55 | | 97.3 |
| Contr_l + PMI_L + PR_L + Triplet_L | PAGNN | 68.13 ± 1.49 | | 73.53 ± 3.84 | | 42.74 ± 0.54 | | 175.0 |
| Contr_l + PMI_L + PR_L + Triplet_L | SAGE | 91.64 ± 1.15 | | 96.19 ± 1.01 | | 47.73 ± 3.09 | | 113.7 |
| Contr_l + PR_L | ALL | 59.26 ± 5.11 | | 52.78 ± 6.06 | | 87.96 ± 3.93 | | 135.0 |
| Contr_l + PR_L | GAT | 94.40 ± 1.28 | | 97.37 ± 0.40 | | 83.88 ± 1.14 | | 61.7 |
| Contr_l + PR_L | GCN | 93.60 ± 0.66 | | 96.03 ± 0.92 | | 81.86 ± 4.42 | | 83.3 |
| Contr_l + PR_L | GIN | 72.30 ± 11.18 | | 83.02 ± 4.16 | | 47.94 ± 12.23 | | 157.3 |
| Contr_l + PR_L | MPNN | 89.88 ± 0.67 | | 96.26 ± 0.95 | | 42.26 ± 1.63 | | 126.0 |
| Contr_l + PR_L | PAGNN | 51.74 ± 0.14 | | 63.93 ± 3.09 | | 41.49 ± 2.10 | | 197.3 |

Lp Precision Continued (↑)

| Loss Type | Model | CORA | | Citeseer | | Bitcoin Fraud Transaction | | Average Rank |
|---|---|---|---|---|---|---|---|---|
| Contr_l + PR_L | SAGE | 86.75 ± 10.06 | | 91.01 ± 11.13 | | 42.95 ± 0.94 | | 143.0 |
| Contr_l + PR_L + Triplet_L | ALL | 74.88 ± 8.88 | | 83.71 ± 7.71 | | 70.34 ± 0.80 | | 140.7 |
| Contr_l + PR_L + Triplet_L | GAT | 96.41 ± 1.25 | | 98.41 ± 0.66 | | 86.94 ± 2.40 | | 22.0 |
| Contr_l + PR_L + Triplet_L | GCN | 95.11 ± 1.10 | | 97.50 ± 0.92 | | 83.37 ± 2.50 | | 58.0 |
| Contr_l + PR_L + Triplet_L | GIN | 88.90 ± 1.70 | | 91.63 ± 2.17 | | 69.78 ± 3.63 | | 116.7 |
| Contr_l + PR_L + Triplet_L | MPNN | 91.55 ± 2.54 | | 95.64 ± 0.67 | | 67.16 ± 7.24 | | 106.0 |
| Contr_l + PR_L + Triplet_L | PAGNN | 58.40 ± 3.59 | | 81.18 ± 8.27 | | 42.19 ± 0.15 | | 182.0 |
| Contr_l + PR_L + Triplet_L | SAGE | 94.09 ± 0.35 | | 97.56 ± 0.25 | | 65.52 ± 4.50 | | 84.0 |
| Contr_l + Triplet_L | ALL | 92.41 ± 0.48 | | 96.34 ± 0.29 | | 85.73 ± 1.54 | | 68.3 |
| Contr_l + Triplet_L | GAT | 96.72 ± 0.53 | | 98.89 ± 0.16 | | 93.04 ± 1.57 | | 7.3 |
| Contr_l + Triplet_L | GCN | 96.45 ± 0.27 | | 98.42 ± 0.29 | | 87.08 ± 1.67 | | 19.0 |
| Contr_l + Triplet_L | GIN | 93.91 ± 0.64 | | 97.96 ± 0.44 | | 79.84 ± 2.99 | | 67.7 |
| Contr_l + Triplet_L | MPNN | 94.06 ± 0.66 | | 97.76 ± 0.19 | | 85.32 ± 0.85 | | 54.0 |
| Contr_l + Triplet_L | PAGNN | 87.70 ± 7.58 | | 95.34 ± 0.73 | | 44.49 ± 0.45 | | 127.0 |
| Contr_l + Triplet_L | SAGE | 95.51 ± 0.22 | | 98.18 ± 0.28 | | 83.39 ± 3.95 | | 46.0 |
| CrossE_L | ALL | 79.26 ± 13.60 | | 82.19 ± 3.04 | | 76.00 ± 8.86 | | 135.0 |

Lp Precision Continued (↑)

| Loss Type | Model | CORA | | Citeseer | | Bitcoin Fraud Transaction | | Average Rank |
|---|---|---|---|---|---|---|---|---|
| CrossE_L | GAT | 77.74 | ± | 84.78 | ± | 71.02 | ± | 136.7 |
| | | 16.03 | | 20.93 | | 20.30 | | |
| CrossE_L | GCN | 51.35 | ± | 47.65 | ± | 34.64 | ± | 209.0 |
| | | 0.00 | | 0.00 | | 0.00 | | |
| CrossE_L | GIN | 51.36 | ± | 47.66 | ± | 34.64 | ± | 208.7 |
| | | 0.01 | | 0.01 | | 0.00 | | |
| CrossE_L | MPNN | 86.24 | ± | 81.90 | ± | 34.64 | ± | 168.3 |
| | | 2.07 | | 6.86 | | 0.00 | | |
| CrossE_L | PAGNN | 56.83 | ± | 63.10 | ± | 34.66 | ± | 200.7 |
| | | 3.33 | | 3.74 | | 0.01 | | |
| CrossE_L | SAGE | 65.47 | ± | 58.10 | ± | 34.64 | ± | 196.3 |
| | | 5.04 | | 10.27 | | 0.01 | | |
| CrossE_L + PMI_L | ALL | 86.47 | ± | 87.90 | ± | 72.59 | ± | 123.0 |
| | | 0.92 | | 2.83 | | 4.47 | | |
| CrossE_L + PMI_L | GAT | 97.19 | ± | 98.77 | ± | 79.56 | ± | 31.7 |
| | | 0.22 | | 0.18 | | 1.50 | | |
| CrossE_L + PMI_L | GCN | 96.58 | ± | 98.30 | ± | 84.14 | ± | 31.3 |
| | | 0.93 | | 0.18 | | 1.66 | | |
| CrossE_L + PMI_L | GIN | 83.90 | ± | 85.17 | ± | 47.09 | ± | 147.0 |
| | | 3.43 | | 2.40 | | 1.52 | | |
| CrossE_L + PMI_L | MPNN | 93.55 | ± | 97.60 | ± | 84.21 | ± | 64.3 |
| | | 0.42 | | 0.42 | | 2.46 | | |
| CrossE_L + PMI_L | PAGNN | 59.95 | ± | 65.92 | ± | 40.24 | ± | 193.3 |
| | | 4.79 | | 0.49 | | 1.74 | | |
| CrossE_L + PMI_L | SAGE | 70.29 | ± | 74.17 | ± | 37.25 | ± | 182.7 |
| | | 1.79 | | 1.06 | | 1.31 | | |
| CrossE_L + PMI_L + PR_L | ALL | 62.65 | ± | 83.52 | ± | 46.47 | ± | 166.0 |
| | | 1.10 | | 3.16 | | 3.87 | | |
| CrossE_L + PMI_L + PR_L | GAT | 96.39 | ± | 95.29 | ± | 65.24 | ± | 84.0 |
| | | 0.35 | | 2.70 | | 21.62 | | |
| CrossE_L + PMI_L + PR_L | GCN | 96.12 | ± | 98.11 | ± | 84.38 | ± | 39.7 |
| | | 0.88 | | 0.31 | | 3.47 | | |

Lp Precision Continued (↑)

| Loss Type | Model | CORA | | Citeseer | | Bitcoin Fraud Transaction | | Average Rank |
|---|---|---|---|---|---|---|---|---|
| CrossE_L + PMI_L + PR_L | GIN | 80.87 | ± | 77.38 | ± | 44.42 | ± | 160.7 |
| | | 0.81 | | 5.40 | | 2.51 | | |
| CrossE_L + PMI_L + PR_L | MPNN | 93.62 | ± | 94.70 | ± | 64.90 | ± | 105.0 |
| | | 1.01 | | 3.22 | | 22.59 | | |
| CrossE_L + PMI_L + PR_L | PAGNN | 63.60 | ± | 66.59 | ± | 38.40 | ± | 190.0 |
| | | 6.04 | | 0.78 | | 0.98 | | |
| CrossE_L + PMI_L + PR_L | SAGE | 71.09 | ± | 75.91 | ± | 42.98 | ± | 171.0 |
| | | 1.15 | | 4.20 | | 0.58 | | |
| CrossE_L + PMI_L + PR_L + Triplet_L | ALL | 84.73 | ± | 81.98 | ± | 70.39 | ± | 135.7 |
| | | 0.70 | | 0.53 | | 2.76 | | |
| CrossE_L + PMI_L + PR_L + Triplet_L | GAT | 96.52 | ± | 97.86 | ± | 75.72 | ± | 54.7 |
| | | 0.38 | | 0.40 | | 11.05 | | |
| CrossE_L + PMI_L + PR_L + Triplet_L | GCN | 96.37 | ± | 98.51 | ± | 83.94 | ± | 31.3 |
| | | 0.76 | | 0.12 | | 2.41 | | |
| CrossE_L + PMI_L + PR_L + Triplet_L | GIN | 84.29 | ± | 85.08 | ± | 61.77 | ± | 137.7 |
| | | 1.96 | | 1.35 | | 2.01 | | |
| CrossE_L + PMI_L + PR_L + Triplet_L | MPNN | 94.11 | ± | 95.19 | ± | 82.17 | ± | 80.7 |
| | | 0.44 | | 1.96 | | 10.56 | | |
| CrossE_L + PMI_L + PR_L + Triplet_L | PAGNN | 69.27 | ± | 68.90 | ± | 41.90 | ± | 179.0 |
| | | 0.69 | | 1.90 | | 0.30 | | |
| CrossE_L + PMI_L + PR_L + Triplet_L | SAGE | 86.80 | ± | 92.50 | ± | 43.93 | ± | 138.7 |
| | | 4.37 | | 1.92 | | 0.32 | | |
| CrossE_L + PMI_L + Triplet_L | ALL | 92.11 | ± | 97.43 | ± | 83.21 | ± | 72.3 |
| | | 0.51 | | 0.61 | | 1.30 | | |
| CrossE_L + PMI_L + Triplet_L | GAT | **97.23** | ± | 98.68 | ± | 87.38 | ± | 8.7 |
| | | **0.43** | | 0.07 | | 2.34 | | |
| CrossE_L + PMI_L + Triplet_L | GCN | 96.47 | ± | 98.36 | ± | 86.65 | ± | 22.3 |
| | | 0.82 | | 0.25 | | 0.86 | | |
| CrossE_L + PMI_L + Triplet_L | GIN | 86.52 | ± | 88.50 | ± | 70.40 | ± | 123.0 |
| | | 1.52 | | 1.06 | | 1.30 | | |
| CrossE_L + PMI_L + Triplet_L | MPNN | 93.93 | ± | 97.20 | ± | 85.38 | ± | 60.7 |
| | | 0.47 | | 0.44 | | 1.33 | | |

Lp Precision Continued (↑)

| Loss Type | Model | CORA | | | Citeseer | | | Bitcoin Fraud Transaction | | | Average Rank |
|---|---|---|---|---|---|---|---|---|---|---|---|
| CrossE_L + PMI_L + Triplet_L | PAGNN | 69.05 | ± | 2.08 | 67.50 | ± | 1.29 | 42.55 | ± | 0.69 | 176.7 |
| CrossE_L + PMI_L + Triplet_L | SAGE | 91.49 | ± | 1.25 | 95.12 | ± | 1.19 | 73.63 | ± | 1.80 | 101.3 |
| CrossE_L + PR_L | ALL | 54.94 | ± | 5.14 | 48.79 | ± | 0.71 | 87.55 | ± | 2.22 | 138.3 |
| CrossE_L + PR_L | GAT | 94.99 | ± | 0.41 | 97.65 | ± | 0.37 | 74.55 | ± | 17.62 | 69.3 |
| CrossE_L + PR_L | GCN | 92.06 | ± | 1.09 | 95.28 | ± | 1.42 | 52.79 | ± | 17.29 | 112.3 |
| CrossE_L + PR_L | GIN | 61.95 | ± | 13.15 | 84.70 | ± | 5.50 | 58.46 | ± | 1.20 | 157.7 |
| CrossE_L + PR_L | MPNN | 89.56 | ± | 2.21 | 95.83 | ± | 1.24 | 37.80 | ± | 1.51 | 136.7 |
| CrossE_L + PR_L | PAGNN | 51.61 | ± | 0.11 | 51.54 | ± | 6.51 | 37.70 | ± | 0.17 | 204.3 |
| CrossE_L + PR_L | SAGE | 64.43 | ± | 2.83 | 60.83 | ± | 7.75 | 35.95 | ± | 1.85 | 194.7 |
| CrossE_L + PR_L + Triplet_L | ALL | 72.42 | ± | 10.89 | 66.93 | ± | 9.21 | 72.49 | ± | 2.67 | 150.7 |
| CrossE_L + PR_L + Triplet_L | GAT | 95.66 | ± | 0.80 | 98.13 | ± | 1.09 | 82.71 | ± | 2.21 | 49.0 |
| CrossE_L + PR_L + Triplet_L | GCN | 94.69 | ± | 0.79 | 96.89 | ± | 0.50 | 83.86 | ± | 0.92 | 64.7 |
| CrossE_L + PR_L + Triplet_L | GIN | 86.48 | ± | 4.23 | 93.35 | ± | 1.18 | 59.00 | ± | 1.33 | 126.7 |
| CrossE_L + PR_L + Triplet_L | MPNN | 88.60 | ± | 2.52 | 96.01 | ± | 0.77 | 58.26 | ± | 13.69 | 116.7 |
| CrossE_L + PR_L + Triplet_L | PAGNN | 54.56 | ± | 3.96 | 72.15 | ± | 4.09 | 42.29 | ± | 0.14 | 186.0 |
| CrossE_L + PR_L + Triplet_L | SAGE | 95.34 | ± | 1.10 | 97.55 | ± | 0.59 | 51.10 | ± | 10.09 | 84.7 |

Lp Precision Continued (↑)

| Loss Type | Model | CORA | | Citeseer | | Bitcoin Fraud Transaction | | Average Rank |
|-----------|-------|------|--|----------|--|---------------------------|--|--------------|
| CrossE_L + Triplet_L | ALL | 94.81 ± 0.27 | | 97.68 ± 0.42 | | 86.24 ± 1.69 | | 46.7 |
| CrossE_L + Triplet_L | GAT | 97.86 ± 0.31 | | 99.19 ± 0.13 | | 92.20 ± 0.95 | | 2.3 |
| CrossE_L + Triplet_L | GCN | 97.18 ± 0.56 | | 98.59 ± 0.45 | | 86.97 ± 1.92 | | 12.3 |
| CrossE_L + Triplet_L | GIN | 94.86 ± 0.55 | | 97.72 ± 0.19 | | 80.14 ± 2.54 | | 63.7 |
| CrossE_L + Triplet_L | MPNN | 95.50 ± 0.71 | | 98.41 ± 0.33 | | 86.83 ± 0.93 | | 30.3 |
| CrossE_L + Triplet_L | PAGNN | 89.98 ± 5.98 | | 96.67 ± 0.53 | | 44.12 ± 0.11 | | 117.3 |
| CrossE_L + Triplet_L | SAGE | 96.98 ± 0.38 | | 98.51 ± 0.23 | | 82.55 ± 2.00 | | 30.7 |
| PMI_L | ALL | 83.18 ± 1.53 | | 89.86 ± 1.86 | | 76.56 ± 1.24 | | 122.7 |
| PMI_L | GAT | 97.17 ± 0.27 | | 98.51 ± 0.43 | | 80.77 ± 1.46 | | 32.7 |
| PMI_L | GCN | 96.45 ± 0.67 | | 98.15 ± 0.34 | | 85.51 ± 2.32 | | 30.3 |
| PMI_L | GIN | 80.05 ± 2.43 | | 84.97 ± 3.57 | | 45.37 ± 3.39 | | 152.3 |
| PMI_L | MPNN | 94.27 ± 0.46 | | 97.49 ± 0.28 | | 84.71 ± 2.55 | | 57.3 |
| PMI_L | PAGNN | 62.71 ± 4.25 | | 66.06 ± 1.36 | | 41.49 ± 0.52 | | 191.0 |
| PMI_L | SAGE | 70.92 ± 2.10 | | 74.09 ± 1.09 | | 38.48 ± 2.08 | | 180.7 |
| PMI_L + PR_L | ALL | 64.42 ± 3.20 | | 81.42 ± 9.13 | | 48.79 ± 0.69 | | 164.7 |
| PMI_L + PR_L | GAT | 95.63 ± 0.95 | | 94.29 ± 4.56 | | 49.79 ± 12.35 | | 101.7 |

Lp Precision Continued (↑)

| Loss Type | Model | CORA | | Citeseer | | Bitcoin Fraud Transaction | | Average Rank |
|---|---|---|---|---|---|---|---|---|
| PMI_L + PR_L | GCN | 96.55 | ± | 98.18 | ± | 84.35 | ± | 33.0 |
| | | 0.55 | | 0.20 | | 2.53 | | |
| PMI_L + PR_L | GIN | 82.19 | ± | 79.19 | ± | 43.49 | ± | 163.3 |
| | | 2.63 | | 4.82 | | 0.77 | | |
| PMI_L + PR_L | MPNN | 94.20 | ± | 91.58 | ± | 50.61 | ± | 112.7 |
| | | 0.51 | | 1.80 | | 20.32 | | |
| PMI_L + PR_L | PAGNN | 63.32 | ± | 66.29 | ± | 37.25 | ± | 193.0 |
| | | 1.20 | | 1.83 | | 1.70 | | |
| PMI_L + PR_L | SAGE | 71.54 | ± | 76.24 | ± | 42.29 | ± | 173.3 |
| | | 2.14 | | 2.49 | | 0.64 | | |
| PMI_L + PR_L + Triplet_L | ALL | 83.77 | ± | 85.37 | ± | 73.10 | ± | 129.3 |
| | | 1.54 | | 1.16 | | 1.08 | | |
| PMI_L + PR_L + Triplet_L | GAT | 95.89 | ± | 97.63 | ± | 69.11 | ± | 70.0 |
| | | 0.20 | | 0.72 | | 9.13 | | |
| PMI_L + PR_L + Triplet_L | GCN | 96.31 | ± | 97.90 | ± | 85.44 | ± | 37.0 |
| | | 0.42 | | 0.38 | | 0.88 | | |
| PMI_L + PR_L + Triplet_L | GIN | 85.30 | ± | 85.50 | ± | 64.31 | ± | 134.0 |
| | | 1.23 | | 1.91 | | 2.19 | | |
| PMI_L + PR_L + Triplet_L | MPNN | 93.98 | ± | 93.68 | ± | 78.61 | ± | 92.7 |
| | | 0.78 | | 1.04 | | 8.03 | | |
| PMI_L + PR_L + Triplet_L | PAGNN | 68.71 | ± | 69.47 | ± | 42.29 | ± | 178.0 |
| | | 3.66 | | 0.66 | | 0.50 | | |
| PMI_L + PR_L + Triplet_L | SAGE | 90.44 | ± | 94.15 | ± | 44.15 | ± | 127.3 |
| | | 3.80 | | 1.19 | | 0.24 | | |
| PMI_L + Triplet_L | ALL | 91.96 | ± | 96.98 | ± | 82.64 | ± | 78.7 |
| | | 1.18 | | 0.62 | | 1.36 | | |
| PMI_L + Triplet_L | GAT | 97.04 | ± | 98.72 | ± | 85.19 | ± | 17.3 |
| | | 0.41 | | 0.18 | | 1.43 | | |
| PMI_L + Triplet_L | GCN | 96.68 | ± | 98.34 | ± | 85.94 | ± | 23.0 |
| | | 0.47 | | 0.46 | | 0.67 | | |
| PMI_L + Triplet_L | GIN | 84.70 | ± | 86.60 | ± | 66.57 | ± | 132.7 |
| | | 2.59 | | 2.00 | | 3.30 | | |

Lp Precision Continued (↑)

| Loss Type | Model | CORA | | Citeseer | | Bitcoin Fraud Transaction | | Average Rank |
|---|---|---|---|---|---|---|---|---|
| PMI_L + Triplet_L | MPNN | 94.25 ± 0.48 | | 96.98 ± 0.42 | | 84.79 ± 1.61 | | 62.0 |
| PMI_L + Triplet_L | PAGNN | 67.00 ± 1.10 | | 68.10 ± 1.55 | | 42.51 ± 0.67 | | 177.7 |
| PMI_L + Triplet_L | SAGE | 88.74 ± 2.66 | | 94.93 ± 0.68 | | 68.87 ± 4.00 | | 113.0 |
| PR_L | ALL | 51.37 ± 0.02 | | 47.97 ± 0.34 | | 87.50 ± 2.61 | | 142.3 |
| PR_L | GAT | 94.82 ± 0.64 | | 98.09 ± 0.33 | | 83.10 ± 2.54 | | 53.3 |
| PR_L | GCN | 91.94 ± 2.13 | | 96.03 ± 0.62 | | 83.07 ± 2.68 | | 85.0 |
| PR_L | GIN | 57.02 ± 11.63 | | 86.64 ± 3.96 | | 59.07 ± 1.52 | | 155.0 |
| PR_L | MPNN | 89.49 ± 2.31 | | 95.61 ± 1.85 | | 49.55 ± 20.47 | | 120.0 |
| PR_L | PAGNN | 51.63 ± 0.11 | | 48.27 ± 0.22 | | 37.79 ± 0.09 | | 204.7 |
| PR_L | SAGE | 66.01 ± 2.08 | | 65.71 ± 2.76 | | 37.81 ± 0.80 | | 190.3 |
| PR_L + Triplet_L | ALL | 53.32 ± 3.70 | | 49.05 ± 0.53 | | 85.53 ± 0.49 | | 145.7 |
| PR_L + Triplet_L | GAT | 94.92 ± 1.35 | | 97.34 ± 0.61 | | 80.81 ± 1.84 | | 66.3 |
| PR_L + Triplet_L | GCN | 93.38 ± 0.32 | | 96.12 ± 0.80 | | 80.06 ± 2.86 | | 87.0 |
| PR_L + Triplet_L | GIN | 62.89 ± 14.22 | | 80.20 ± 3.50 | | 55.64 ± 1.39 | | 163.3 |
| PR_L + Triplet_L | MPNN | 89.13 ± 3.15 | | 96.48 ± 0.95 | | 58.78 ± 18.83 | | 109.7 |
| PR_L + Triplet_L | PAGNN | 51.73 ± 0.11 | | 63.36 ± 3.60 | | 41.59 ± 0.50 | | 197.0 |

Lp Precision Continued (↑)

| Loss Type | Model | CORA | | Citeseer | | Bitcoin Fraud Transaction | | Average Rank |
|---|---|---|---|---|---|---|---|---|
| PR_L + Triplet_L | SAGE | 71.60 ± 2.71 | | 78.91 ± 17.00 | | 41.69 ± 0.43 | | 175.0 |
| Triplet_L | ALL | 94.20 ± 0.49 | | 97.77 ± 0.26 | | 84.93 ± 2.14 | | 53.7 |
| Triplet_L | GAT | 98.24 ± 0.41 | | 99.38 ± 0.08 | | 90.73 ± 0.99 | | 2.0 |
| Triplet_L | GCN | 96.95 ± 0.24 | | 98.80 ± 0.07 | | 87.97 ± 1.08 | | 8.3 |
| Triplet_L | GIN | 95.50 ± 1.11 | | 97.88 ± 0.34 | | 79.85 ± 1.50 | | 60.0 |
| Triplet_L | MPNN | 95.17 ± 0.98 | | 98.35 ± 0.25 | | 86.98 ± 0.64 | | 31.7 |
| Triplet_L | PAGNN | 89.72 ± 6.51 | | 97.05 ± 0.28 | | 44.36 ± 0.23 | | 113.7 |
| Triplet_L | SAGE | 96.96 ± 0.59 | | 98.65 ± 0.22 | | 82.83 ± 0.92 | | 28.7 |

Table 10. **Lp Recall Performance** (↑): Top-ranked results are highlighted in **1st**, second-ranked in **2nd**, and third-ranked in **3rd**.

| Loss Type | Model | CORA | | Citeseer | | Bitcoin Fraud Transaction | | Average Rank |
|---|---|---|---|---|---|---|---|---|
| Contr_l | ALL | 94.19 ± 1.43 | | 95.06 ± 0.79 | | 81.17 ± 1.70 | | 94.3 |
| Contr_l | GAT | 97.14 ± 0.45 | | 99.32 ± 0.21 | | 85.02 ± 0.49 | | 31.0 |
| Contr_l | GCN | 95.20 ± 0.53 | | 98.41 ± 0.35 | | 81.48 ± 0.99 | | 70.3 |
| Contr_l | GIN | 95.19 ± 0.66 | | 97.92 ± 0.40 | | 79.23 ± 3.72 | | 85.7 |

Continued on next page

Lp Recall Continued (↑)

| Loss Type | Model | CORA | | Citeseer | | Bitcoin Fraud Transaction | | Average Rank |
|---|---|---|---|---|---|---|---|---|
| Contr_l | MPNN | 93.94 ± 0.56 | | 97.96 ± 0.40 | | 78.08 ± 0.94 | | 92.7 |
| Contr_l | PAGNN | 93.38 ± 0.86 | | 95.38 ± 1.40 | | 89.07 ± 1.71 | | 74.7 |
| Contr_l | SAGE | 95.03 ± 0.43 | | 98.08 ± 0.21 | | 80.33 ± 2.34 | | 79.0 |
| Contr_l + CrossE_L | ALL | 93.61 ± 0.86 | | 95.37 ± 0.42 | | 81.38 ± 2.41 | | 94.7 |
| Contr_l + CrossE_L | GAT | 96.52 ± 0.60 | | 99.29 ± 0.13 | | 87.91 ± 3.47 | | 28.0 |
| Contr_l + CrossE_L | GCN | 95.86 ± 0.34 | | 98.56 ± 0.25 | | 81.00 ± 1.41 | | 64.0 |
| Contr_l + CrossE_L | GIN | 93.89 ± 1.39 | | 97.11 ± 0.42 | | 78.28 ± 2.03 | | 100.0 |
| Contr_l + CrossE_L | MPNN | 94.43 ± 1.15 | | 97.81 ± 0.34 | | 80.67 ± 1.20 | | 84.7 |
| Contr_l + CrossE_L | PAGNN | 93.69 ± 1.62 | | 95.93 ± 0.62 | | 90.73 ± 0.39 | | 69.3 |
| Contr_l + CrossE_L | SAGE | 95.09 ± 0.60 | | 98.15 ± 0.44 | | 79.84 ± 2.72 | | 79.3 |
| Contr_l + CrossE_L + PMI_L | ALL | 87.84 ± 0.97 | | 80.88 ± 6.51 | | 68.78 ± 3.74 | | 165.0 |
| Contr_l + CrossE_L + PMI_L | GAT | 96.62 ± 0.50 | | 99.09 ± 0.37 | | 77.84 ± 0.82 | | 61.7 |
| Contr_l + CrossE_L + PMI_L | GCN | 94.96 ± 0.66 | | 97.71 ± 0.18 | | 78.97 ± 1.45 | | 92.0 |
| Contr_l + CrossE_L + PMI_L | GIN | 87.75 ± 2.58 | | 84.50 ± 4.06 | | 72.21 ± 2.45 | | 159.3 |
| Contr_l + CrossE_L + PMI_L | MPNN | 96.86 ± 0.61 | | 98.70 ± 0.24 | | 77.20 ± 1.87 | | 66.0 |
| Contr_l + CrossE_L + PMI_L | PAGNN | 90.77 ± 6.10 | | 88.99 ± 1.70 | | 82.86 ± 0.92 | | 115.3 |

Lp Recall Continued (↑)

| Loss Type | | | Model | CORA | | Citeseer | | Bitcoin Fraud Transaction | | Average Rank |
|---|---|---|---|---|---|---|---|---|---|---|
| Contr_l + CrossE_L + PMI_L | | | SAGE | 83.73 ± 2.20 | | 82.62 ± 3.60 | | 87.50 ± 5.42 | | 133.0 |
| Contr_l + CrossE_L + PMI_L + PR_L | | | ALL | 88.82 ± 4.01 | | 73.98 ± 4.69 | | 57.99 ± 0.82 | | 179.3 |
| Contr_l + CrossE_L + PMI_L + PR_L | | | GAT | 96.16 ± 1.10 | | 98.79 ± 0.20 | | 73.54 ± 8.50 | | 77.3 |
| Contr_l + CrossE_L + PMI_L + PR_L | | | GCN | 95.42 ± 0.55 | | 97.87 ± 0.52 | | 79.47 ± 1.25 | | 82.0 |
| Contr_l + CrossE_L + PMI_L + PR_L | | | GIN | 86.50 ± 2.68 | | 82.43 ± 2.60 | | 88.66 ± 6.05 | | 121.0 |
| Contr_l + CrossE_L + PMI_L + PR_L | | | MPNN | 95.65 ± 0.66 | | 94.29 ± 3.49 | | 82.38 ± 4.22 | | 85.7 |
| Contr_l + CrossE_L + PMI_L + PR_L | | | PAGNN | 85.46 ± 4.52 | | 90.11 ± 1.04 | | 92.65 ± 1.10 | | 102.0 |
| Contr_l + CrossE_L + PMI_L + PR_L | | | SAGE | 84.01 ± 0.81 | | 82.13 ± 5.14 | | 88.39 ± 1.36 | | 131.0 |
| Contr_l + CrossE_L + PMI_L + PR_L + Triplet_L | | | ALL | 92.26 ± 1.56 | | 89.19 ± 1.49 | | 59.08 ± 2.99 | | 150.0 |
| Contr_l + CrossE_L + PMI_L + PR_L + Triplet_L | | | GAT | 96.34 ± 0.46 | | 98.63 ± 1.00 | | 77.71 ± 2.35 | | 70.0 |
| Contr_l + CrossE_L + PMI_L + PR_L + Triplet_L | | | GCN | 95.54 ± 0.32 | | 98.05 ± 0.09 | | 80.78 ± 1.16 | | 72.7 |
| Contr_l + CrossE_L + PMI_L + PR_L + Triplet_L | | | GIN | 86.93 ± 3.00 | | 90.05 ± 1.29 | | 67.46 ± 2.91 | | 151.0 |
| Contr_l + CrossE_L + PMI_L + PR_L + Triplet_L | | | MPNN | 95.87 ± 1.19 | | 94.77 ± 2.02 | | 72.87 ± 7.94 | | 107.3 |
| Contr_l + CrossE_L + PMI_L + PR_L + Triplet_L | | | PAGNN | 84.34 ± 1.36 | | 89.87 ± 0.75 | | 84.64 ± 0.28 | | 123.7 |
| Contr_l + CrossE_L + PMI_L + PR_L + Triplet_L | | | SAGE | 90.26 ± 1.21 | | 96.07 ± 0.87 | | 92.26 ± 1.88 | | 75.7 |
| Contr_l + CrossE_L + PMI_L + Triplet_L | | | ALL | 93.07 ± 0.59 | | 96.90 ± 0.29 | | 78.43 ± 0.96 | | 103.0 |

Lp Recall Continued (↑)

| Loss Type | Model | CORA | | | Citeseer | | | Bitcoin Fraud Transaction | | | Average Rank |
|---|---|---|---|---|---|---|---|---|---|---|---|
| Contr_l + CrossE_L + PMI_L + Triplet_L | GAT | 97.24 | ± | 0.56 | 99.04 | ± | 0.21 | 80.75 | ± | 0.98 | 48.0 |
| Contr_l + CrossE_L + PMI_L + Triplet_L | GCN | 96.15 | ± | 0.67 | 97.81 | ± | 0.35 | 79.07 | ± | 2.10 | 79.0 |
| Contr_l + CrossE_L + PMI_L + Triplet_L | GIN | 87.61 | ± | 3.81 | 91.14 | ± | 0.83 | 67.84 | ± | 0.36 | 145.3 |
| Contr_l + CrossE_L + PMI_L + Triplet_L | MPNN | 96.92 | ± | 0.93 | 98.44 | ± | 0.43 | 80.16 | ± | 0.71 | 57.7 |
| Contr_l + CrossE_L + PMI_L + Triplet_L | PAGNN | 80.12 | ± | 2.20 | 90.10 | ± | 1.28 | 83.68 | ± | 1.47 | 132.3 |
| Contr_l + CrossE_L + PMI_L + Triplet_L | SAGE | 91.28 | ± | 0.80 | 94.89 | ± | 0.75 | 76.43 | ± | 2.54 | 120.7 |
| Contr_l + CrossE_L + PR_L | ALL | 93.84 | ± | 8.59 | 94.52 | ± | 4.61 | 54.97 | ± | 0.90 | 135.0 |
| Contr_l + CrossE_L + PR_L | GAT | 84.53 | ± | 7.64 | 85.97 | ± | 5.85 | 67.35 | ± | 2.76 | 171.3 |
| Contr_l + CrossE_L + PR_L | GCN | 86.21 | ± | 1.50 | 87.04 | ± | 1.04 | 66.84 | ± | 1.77 | 166.3 |
| Contr_l + CrossE_L + PR_L | GIN | 72.41 | ± | 14.42 | 78.81 | ± | 3.73 | 61.15 | ± | 21.72 | 196.3 |
| Contr_l + CrossE_L + PR_L | MPNN | 68.65 | ± | 2.91 | 79.15 | ± | 1.06 | 74.69 | ± | 9.66 | 182.3 |
| Contr_l + CrossE_L + PR_L | PAGNN | 99.73 | ± | 0.07 | 85.25 | ± | 10.49 | 83.60 | ± | 5.80 | 82.7 |
| Contr_l + CrossE_L + PR_L | SAGE | 86.46 | ± | 6.72 | 88.47 | ± | 10.44 | 89.99 | ± | 2.09 | 111.0 |
| Contr_l + CrossE_L + PR_L + Triplet_L | ALL | 87.90 | ± | 0.56 | 89.21 | ± | 1.83 | 79.42 | ± | 2.41 | 132.3 |
| Contr_l + CrossE_L + PR_L + Triplet_L | GAT | 94.03 | ± | 0.44 | 95.94 | ± | 2.10 | 80.62 | ± | 3.54 | 94.3 |
| Contr_l + CrossE_L + PR_L + Triplet_L | GCN | 91.89 | ± | 0.84 | 93.52 | ± | 1.02 | 76.40 | ± | 2.69 | 122.3 |

Lp Recall Continued (↑)

| Loss Type | Model | CORA | | | Citeseer | | | Bitcoin Fraud Transaction | | | Average Rank |
|---|---|---|---|---|---|---|---|---|---|---|---|
| Contr_l + CrossE_L + PR_L + Triplet_L | GIN | 86.00 | ± | | 88.76 | ± | | 73.63 | ± | | 152.7 |
| | | 0.76 | | | 1.04 | | | 9.66 | | | |
| Contr_l + CrossE_L + PR_L + Triplet_L | MPNN | 84.26 | ± | | 89.32 | ± | | 69.49 | ± | | 161.3 |
| | | 1.68 | | | 1.01 | | | 4.03 | | | |
| Contr_l + CrossE_L + PR_L + Triplet_L | PAGNN | 89.63 | ± | | 86.92 | ± | | 94.10 | ± | | 100.0 |
| | | 1.06 | | | 1.17 | | | 3.67 | | | |
| Contr_l + CrossE_L + PR_L + Triplet_L | SAGE | 92.57 | ± | | 94.94 | ± | | 82.98 | ± | | 95.3 |
| | | 0.44 | | | 0.64 | | | 1.60 | | | |
| Contr_l + CrossE_L + Triplet_L | ALL | 94.95 | ± | | 97.09 | ± | | 84.48 | ± | | 76.7 |
| | | 0.89 | | | 0.50 | | | 1.77 | | | |
| Contr_l + CrossE_L + Triplet_L | GAT | 98.00 | ± | | 99.55 | ± | | 91.77 | ± | | 12.0 |
| | | 0.69 | | | 0.19 | | | 1.91 | | | |
| Contr_l + CrossE_L + Triplet_L | GCN | 96.53 | ± | | 98.97 | ± | | 84.11 | ± | | 42.3 |
| | | 0.45 | | | 0.13 | | | 0.97 | | | |
| Contr_l + CrossE_L + Triplet_L | GIN | 96.42 | ± | | 98.23 | ± | | 82.15 | ± | | 57.7 |
| | | 0.68 | | | 0.49 | | | 2.93 | | | |
| Contr_l + CrossE_L + Triplet_L | MPNN | 95.91 | ± | | 98.60 | ± | | 83.85 | ± | | 54.0 |
| | | 0.70 | | | 0.34 | | | 1.41 | | | |
| Contr_l + CrossE_L + Triplet_L | PAGNN | 92.95 | ± | | 96.64 | ± | | 89.94 | ± | | 71.7 |
| | | 2.97 | | | 0.45 | | | 0.34 | | | |
| Contr_l + CrossE_L + Triplet_L | SAGE | 96.25 | ± | | 98.68 | ± | | 85.16 | ± | | 44.3 |
| | | 0.81 | | | 0.29 | | | 1.92 | | | |
| Contr_l + PMI_L | ALL | 86.61 | ± | | 91.12 | ± | | 73.55 | ± | | 142.7 |
| | | 4.73 | | | 0.93 | | | 1.93 | | | |
| Contr_l + PMI_L | GAT | 96.97 | ± | | 98.91 | ± | | 78.81 | ± | | 58.7 |
| | | 0.20 | | | 0.23 | | | 1.38 | | | |
| Contr_l + PMI_L | GCN | 95.39 | ± | | 97.79 | ± | | 79.75 | ± | | 84.3 |
| | | 0.56 | | | 0.51 | | | 1.21 | | | |
| Contr_l + PMI_L | GIN | 86.77 | ± | | 85.76 | ± | | 72.35 | ± | | 159.0 |
| | | 5.11 | | | 3.27 | | | 2.88 | | | |
| Contr_l + PMI_L | MPNN | 96.23 | ± | | 98.53 | ± | | 77.92 | ± | | 71.7 |
| | | 0.46 | | | 0.39 | | | 1.52 | | | |

Lp Recall Continued (↑)

| Loss Type | Model | CORA | | Citeseer | | Bitcoin Fraud Transaction | | Average Rank |
|---|---|---|---|---|---|---|---|---|
| Contr_l + PMI_L | PAGNN | 88.46 ± 8.10 | | 89.68 ± 1.59 | | 83.15 ± 0.77 | | 115.0 |
| Contr_l + PMI_L | SAGE | 84.77 ± 1.81 | | 92.54 ± 2.96 | | 89.93 ± 1.05 | | 105.0 |
| Contr_l + PMI_L + PR_L | ALL | 86.04 ± 0.21 | | 73.70 ± 4.00 | | 56.93 ± 0.76 | | 188.7 |
| Contr_l + PMI_L + PR_L | GAT | 95.90 ± 0.41 | | 96.43 ± 3.25 | | 82.70 ± 10.71 | | 74.3 |
| Contr_l + PMI_L + PR_L | GCN | 95.82 ± 0.71 | | 97.72 ± 0.20 | | 79.85 ± 0.62 | | 79.7 |
| Contr_l + PMI_L + PR_L | GIN | 87.32 ± 3.46 | | 81.65 ± 5.05 | | 86.96 ± 7.08 | | 123.3 |
| Contr_l + PMI_L + PR_L | MPNN | 96.10 ± 0.58 | | 91.63 ± 3.87 | | 75.96 ± 14.95 | | 104.7 |
| Contr_l + PMI_L + PR_L | PAGNN | 86.69 ± 5.75 | | 90.16 ± 2.02 | | 92.59 ± 0.95 | | 97.0 |
| Contr_l + PMI_L + PR_L | SAGE | 84.59 ± 2.55 | | 90.96 ± 2.00 | | 88.55 ± 0.91 | | 109.0 |
| Contr_l + PMI_L + PR_L + Triplet_L | ALL | 91.80 ± 0.71 | | 92.85 ± 1.18 | | 67.50 ± 5.78 | | 134.3 |
| Contr_l + PMI_L + PR_L + Triplet_L | GAT | 95.98 ± 0.64 | | 98.23 ± 0.52 | | 77.20 ± 1.08 | | 80.7 |
| Contr_l + PMI_L + PR_L + Triplet_L | GCN | 95.69 ± 0.23 | | 98.13 ± 0.29 | | 79.36 ± 2.40 | | 77.7 |
| Contr_l + PMI_L + PR_L + Triplet_L | GIN | 90.48 ± 1.22 | | 92.48 ± 2.07 | | 65.22 ± 2.13 | | 142.7 |
| Contr_l + PMI_L + PR_L + Triplet_L | MPNN | 95.41 ± 0.62 | | 94.44 ± 0.73 | | 67.27 ± 6.33 | | 118.0 |
| Contr_l + PMI_L + PR_L + Triplet_L | PAGNN | 80.76 ± 2.45 | | 87.83 ± 0.77 | | 88.43 ± 1.65 | | 128.7 |
| Contr_l + PMI_L + PR_L + Triplet_L | SAGE | 92.75 ± 0.90 | | 96.67 ± 0.32 | | 86.29 ± 3.68 | | 78.3 |

Lp Recall Continued (↑)

| Loss Type | Model | CORA | | Citeseer | | Bitcoin Fraud Transaction | | Average Rank |
|---|---|---|---|---|---|---|---|---|
| Contr_l + PR_L | ALL | 85.03 ± 9.13 | | 95.93 ± 4.84 | | 54.78 ± 2.53 | | 155.0 |
| Contr_l + PR_L | GAT | 85.16 ± 5.40 | | 84.54 ± 1.63 | | 67.12 ± 3.77 | | 171.7 |
| Contr_l + PR_L | GCN | 85.90 ± 0.72 | | 88.71 ± 1.71 | | 74.96 ± 4.01 | | 152.7 |
| Contr_l + PR_L | GIN | 71.62 ± 15.61 | | 77.95 ± 4.52 | | 70.66 ± 26.82 | | 188.0 |
| Contr_l + PR_L | MPNN | 66.81 ± 4.16 | | 80.76 ± 1.35 | | 79.83 ± 6.38 | | 168.0 |
| Contr_l + PR_L | PAGNN | 99.76 ± 0.08 | | 76.05 ± 1.34 | | 87.43 ± 6.84 | | 83.0 |
| Contr_l + PR_L | SAGE | 87.08 ± 7.34 | | 88.28 ± 12.40 | | 90.08 ± 3.51 | | 108.0 |
| Contr_l + PR_L + Triplet_L | ALL | 88.24 ± 1.06 | | 88.18 ± 1.16 | | 78.88 ± 1.15 | | 138.0 |
| Contr_l + PR_L + Triplet_L | GAT | 93.29 ± 2.90 | | 95.98 ± 1.41 | | 79.72 ± 4.56 | | 101.0 |
| Contr_l + PR_L + Triplet_L | GCN | 90.96 ± 0.64 | | 92.97 ± 1.03 | | 73.64 ± 4.10 | | 128.0 |
| Contr_l + PR_L + Triplet_L | GIN | 86.77 ± 0.68 | | 89.39 ± 2.04 | | 65.73 ± 12.02 | | 157.3 |
| Contr_l + PR_L + Triplet_L | MPNN | 85.60 ± 4.23 | | 87.95 ± 2.02 | | 68.03 ± 3.02 | | 163.3 |
| Contr_l + PR_L + Triplet_L | PAGNN | 91.04 ± 3.96 | | 87.06 ± 1.93 | | 95.71 ± 0.55 | | 95.3 |
| Contr_l + PR_L + Triplet_L | SAGE | 92.35 ± 0.54 | | 95.32 ± 0.79 | | 80.91 ± 1.46 | | 101.0 |
| Contr_l + Triplet_L | ALL | 95.38 ± 1.23 | | 97.46 ± 0.55 | | 86.26 ± 0.80 | | 66.3 |
| Contr_l + Triplet_L | GAT | 97.90 ± 0.41 | | 99.61 ± 0.19 | | 91.90 ± 1.37 | | 12.0 |

Lp Recall Continued (↑)

| Loss Type | Model | CORA | | Citeseer | | Bitcoin Fraud Transaction | | Average Rank |
|---|---|---|---|---|---|---|---|---|
| Contr_l + Triplet_L | GCN | 96.32 ± 0.33 | | 98.85 ± 0.30 | | 84.53 ± 1.95 | | 44.7 |
| Contr_l + Triplet_L | GIN | 96.09 ± 0.83 | | 97.67 ± 0.71 | | 81.11 ± 2.39 | | 72.7 |
| Contr_l + Triplet_L | MPNN | 96.04 ± 0.79 | | 98.86 ± 0.40 | | 84.90 ± 1.43 | | 46.7 |
| Contr_l + Triplet_L | PAGNN | 93.68 ± 3.05 | | 96.69 ± 0.56 | | 90.41 ± 0.20 | | 67.3 |
| Contr_l + Triplet_L | SAGE | 96.82 ± 0.37 | | 98.47 ± 0.60 | | 85.19 ± 3.69 | | 42.3 |
| CrossE_L | ALL | 91.49 ± 4.14 | | 72.61 ± 10.01 | | 60.88 ± 1.37 | | 173.7 |
| CrossE_L | GAT | 94.47 ± 5.61 | | 97.81 ± 1.51 | | 84.82 ± 8.41 | | 72.0 |
| CrossE_L | GCN | 100.00 ± 0.00 | | 100.00 ± 0.00 | | 100.00 ± 0.00 | | 1.0 |
| CrossE_L | GIN | 100.00 ± 0.00 | | 100.00 ± 0.00 | | 100.00 ± 0.00 | | 2.0 |
| CrossE_L | MPNN | 84.37 ± 2.51 | | 88.25 ± 4.46 | | 100.00 ± 0.00 | | 109.7 |
| CrossE_L | PAGNN | 90.02 ± 7.94 | | 80.37 ± 2.98 | | 99.97 ± 0.04 | | 105.7 |
| CrossE_L | SAGE | 88.23 ± 3.85 | | 89.57 ± 10.14 | | 100.00 ± 0.00 | | 92.7 |
| CrossE_L + PMI_L | ALL | 84.06 ± 1.98 | | 76.36 ± 2.19 | | 64.17 ± 2.74 | | 189.7 |
| CrossE_L + PMI_L | GAT | 97.30 ± 0.26 | | 99.20 ± 0.16 | | 76.34 ± 2.84 | | 59.0 |
| CrossE_L + PMI_L | GCN | 95.21 ± 0.92 | | 97.87 ± 0.62 | | 80.10 ± 1.22 | | 81.3 |
| CrossE_L + PMI_L | GIN | 86.66 ± 1.40 | | 81.01 ± 1.95 | | 71.90 ± 2.07 | | 167.0 |

Lp Recall Continued (↑)

| Loss Type | Model | CORA | | Citeseer | | Bitcoin Fraud Transaction | | Average Rank |
|---|---|---|---|---|---|---|---|---|
| CrossE_L + PMI_L | MPNN | 96.85 ± 0.54 | | 98.24 ± 0.48 | | 77.82 ± 1.88 | | 70.0 |
| CrossE_L + PMI_L | PAGNN | 88.39 ± 5.72 | | 90.27 ± 0.97 | | 86.48 ± 4.94 | | 102.3 |
| CrossE_L + PMI_L | SAGE | 84.22 ± 1.51 | | 77.21 ± 1.47 | | 91.12 ± 2.13 | | 131.3 |
| CrossE_L + PMI_L + PR_L | ALL | 88.29 ± 2.12 | | 71.73 ± 2.02 | | 63.27 ± 8.51 | | 178.3 |
| CrossE_L + PMI_L + PR_L | GAT | 96.68 ± 0.53 | | 94.78 ± 3.72 | | 73.45 ± 11.79 | | 97.3 |
| CrossE_L + PMI_L + PR_L | GCN | 95.72 ± 0.58 | | 97.86 ± 0.61 | | 78.21 ± 1.32 | | 84.7 |
| CrossE_L + PMI_L + PR_L | GIN | 84.62 ± 3.32 | | 85.99 ± 4.70 | | 82.46 ± 9.79 | | 139.0 |
| CrossE_L + PMI_L + PR_L | MPNN | 95.98 ± 1.41 | | 93.87 ± 5.50 | | 76.88 ± 6.22 | | 101.7 |
| CrossE_L + PMI_L + PR_L | PAGNN | 83.58 ± 4.90 | | 90.03 ± 0.89 | | 91.39 ± 2.41 | | 110.7 |
| CrossE_L + PMI_L + PR_L | SAGE | 82.81 ± 1.68 | | 77.07 ± 1.91 | | 84.51 ± 5.32 | | 149.0 |
| CrossE_L + PMI_L + PR_L + Triplet_L | ALL | 90.02 ± 1.01 | | 87.93 ± 2.25 | | 63.20 ± 1.10 | | 158.7 |
| CrossE_L + PMI_L + PR_L + Triplet_L | GAT | 96.09 ± 0.59 | | 98.54 ± 0.46 | | 77.00 ± 4.40 | | 76.7 |
| CrossE_L + PMI_L + PR_L + Triplet_L | GCN | 95.41 ± 0.35 | | 97.89 ± 0.25 | | 81.60 ± 1.00 | | 73.0 |
| CrossE_L + PMI_L + PR_L + Triplet_L | GIN | 88.22 ± 1.50 | | 89.71 ± 1.82 | | 66.96 ± 1.27 | | 151.7 |
| CrossE_L + PMI_L + PR_L + Triplet_L | MPNN | 96.40 ± 0.53 | | 95.10 ± 3.07 | | 72.78 ± 7.73 | | 99.0 |
| CrossE_L + PMI_L + PR_L + Triplet_L | PAGNN | 80.91 ± 1.26 | | 88.50 ± 2.22 | | 84.57 ± 0.28 | | 135.7 |

Lp Recall Continued (↑)

| Loss Type | Model | CORA | | Citeseer | | Bitcoin Fraud Transaction | | Average Rank |
|---|---|---|---|---|---|---|---|---|
| CrossE_L + PMI_L + PR_L + Triplet_L | SAGE | 90.69 ± 3.16 | | 94.81 ± 0.77 | | 90.69 ± 1.11 | | 82.7 |
| CrossE_L + PMI_L + Triplet_L | ALL | 94.52 ± 0.93 | | 97.83 ± 0.30 | | 79.50 ± 2.10 | | 88.3 |
| CrossE_L + PMI_L + Triplet_L | GAT | 97.18 ± 0.44 | | 99.11 ± 0.17 | | 82.59 ± 1.00 | | 40.7 |
| CrossE_L + PMI_L + Triplet_L | GCN | 94.99 ± 0.13 | | 97.88 ± 0.20 | | 79.92 ± 0.67 | | 82.7 |
| CrossE_L + PMI_L + Triplet_L | GIN | 89.76 ± 1.20 | | 92.62 ± 1.10 | | 69.35 ± 3.46 | | 137.7 |
| CrossE_L + PMI_L + Triplet_L | MPNN | 95.39 ± 0.84 | | 98.41 ± 0.33 | | 81.37 ± 1.29 | | 69.7 |
| CrossE_L + PMI_L + Triplet_L | PAGNN | 81.81 ± 3.14 | | 90.33 ± 0.79 | | 83.81 ± 2.39 | | 128.3 |
| CrossE_L + PMI_L + Triplet_L | SAGE | 92.47 ± 2.26 | | 96.43 ± 0.59 | | 78.00 ± 2.81 | | 108.3 |
| CrossE_L + PR_L | ALL | 92.63 ± 10.12 | | 99.25 ± 0.50 | | 56.01 ± 2.10 | | 107.3 |
| CrossE_L + PR_L | GAT | 82.56 ± 4.12 | | 84.68 ± 1.68 | | 67.06 ± 4.31 | | 178.3 |
| CrossE_L + PR_L | GCN | 81.58 ± 2.15 | | 84.51 ± 2.19 | | 72.23 ± 25.35 | | 175.3 |
| CrossE_L + PR_L | GIN | 85.94 ± 18.52 | | 68.43 ± 4.01 | | 49.86 ± 0.91 | | 193.0 |
| CrossE_L + PR_L | MPNN | 74.66 ± 11.22 | | 77.51 ± 1.89 | | 87.85 ± 2.75 | | 145.0 |
| CrossE_L + PR_L | PAGNN | 99.80 ± 0.06 | | 93.68 ± 12.76 | | 97.79 ± 0.09 | | 40.3 |
| CrossE_L + PR_L | SAGE | 76.47 ± 2.17 | | 77.25 ± 12.97 | | 93.86 ± 8.43 | | 135.0 |
| CrossE_L + PR_L + Triplet_L | ALL | 86.69 ± 0.83 | | 90.33 ± 2.92 | | 64.29 ± 11.92 | | 155.3 |

Lp Recall Continued (↑)

| Loss Type | | | | Model | CORA | | Citeseer | | Bitcoin Fraud Transaction | | Average Rank |
|---|---|---|---|---|---|---|---|---|---|---|---|
| CrossE_L | + | PR_L | + | GAT | 87.35 | ± | 95.07 | ± | 72.13 | ± | 136.0 |
| Triplet_L | | | | | 3.67 | | 3.02 | | 5.34 | | |
| CrossE_L | + | PR_L | + | GCN | 89.55 | ± | 90.90 | ± | 74.99 | ± | 134.0 |
| Triplet_L | | | | | 1.87 | | 1.54 | | 4.31 | | |
| CrossE_L | + | PR_L | + | GIN | 82.72 | ± | 86.22 | ± | 51.05 | ± | 185.3 |
| Triplet_L | | | | | 3.39 | | 2.30 | | 1.56 | | |
| CrossE_L | + | PR_L | + | MPNN | 74.79 | ± | 84.12 | ± | 73.09 | ± | 177.7 |
| Triplet_L | | | | | 3.33 | | 0.83 | | 3.59 | | |
| CrossE_L | + | PR_L | + | PAGNN | 96.89 | ± | 82.04 | ± | 94.97 | ± | 72.0 |
| Triplet_L | | | | | 4.14 | | 2.40 | | 1.24 | | |
| CrossE_L | + | PR_L | + | SAGE | 92.39 | ± | 95.23 | ± | 85.02 | ± | 88.3 |
| Triplet_L | | | | | 0.99 | | 1.38 | | 3.69 | | |
| CrossE_L + Triplet_L | | | | ALL | 96.49 | ± | 98.51 | ± | 85.20 | ± | 43.7 |
| | | | | | 0.84 | | 0.30 | | 2.91 | | |
| CrossE_L + Triplet_L | | | | GAT | 98.77 | ± | 99.79 | ± | 88.25 | ± | 16.7 |
| | | | | | 0.28 | | 0.06 | | 3.68 | | |
| CrossE_L + Triplet_L | | | | GCN | 97.25 | ± | 99.11 | ± | 86.45 | ± | 28.7 |
| | | | | | 0.58 | | 0.19 | | 1.27 | | |
| CrossE_L + Triplet_L | | | | GIN | 97.29 | ± | 97.51 | ± | 82.35 | ± | 60.3 |
| | | | | | 0.65 | | 0.61 | | 1.70 | | |
| CrossE_L + Triplet_L | | | | MPNN | 97.70 | ± | 99.41 | ± | 86.78 | ± | 22.7 |
| | | | | | 0.40 | | 0.35 | | 0.87 | | |
| CrossE_L + Triplet_L | | | | PAGNN | 92.94 | ± | 97.62 | ± | 90.38 | ± | 67.0 |
| | | | | | 5.56 | | 0.60 | | 0.50 | | |
| CrossE_L + Triplet_L | | | | SAGE | 97.99 | ± | 99.06 | ± | 86.08 | ± | 27.7 |
| | | | | | 0.26 | | 0.29 | | 1.49 | | |
| PMI_L | | | | ALL | 79.90 | ± | 76.49 | ± | 63.03 | ± | 196.7 |
| | | | | | 1.63 | | 2.03 | | 2.45 | | |
| PMI_L | | | | GAT | 97.61 | ± | 99.18 | ± | 75.47 | ± | 58.7 |
| | | | | | 0.24 | | 0.33 | | 3.70 | | |
| PMI_L | | | | GCN | 95.35 | ± | 97.41 | ± | 78.48 | ± | 92.7 |
| | | | | | 0.24 | | 0.90 | | 2.48 | | |

Lp Recall Continued (↑)

| Loss Type | Model | CORA | | Citeseer | | Bitcoin Fraud Transaction | | Average Rank |
|---|---|---|---|---|---|---|---|---|
| PMI_L | GIN | 84.77 | ± 2.13 | 81.71 | ± 3.45 | 73.29 | ± 2.63 | 169.0 |
| PMI_L | MPNN | 96.27 | ± 0.59 | 98.66 | ± 0.37 | 77.36 | ± 0.73 | 70.7 |
| PMI_L | PAGNN | 82.27 | ± 6.48 | 89.93 | ± 1.80 | 83.32 | ± 0.71 | 131.7 |
| PMI_L | SAGE | 82.03 | ± 0.99 | 75.83 | ± 1.61 | 91.35 | ± 2.21 | 136.3 |
| PMI_L + PR_L | ALL | 86.46 | ± 3.49 | 73.99 | ± 7.88 | 58.89 | ± 0.46 | 186.0 |
| PMI_L + PR_L | GAT | 95.50 | ± 0.74 | 92.65 | ± 7.73 | 72.71 | ± 21.68 | 114.7 |
| PMI_L + PR_L | GCN | 94.94 | ± 1.35 | 98.04 | ± 0.36 | 79.78 | ± 1.82 | 83.3 |
| PMI_L + PR_L | GIN | 83.92 | ± 1.97 | 85.96 | ± 4.18 | 87.79 | ± 4.43 | 128.3 |
| PMI_L + PR_L | MPNN | 95.67 | ± 0.68 | 88.03 | ± 1.00 | 81.87 | ± 5.03 | 102.3 |
| PMI_L + PR_L | PAGNN | 81.70 | ± 1.13 | 88.78 | ± 2.96 | 93.40 | ± 4.29 | 116.0 |
| PMI_L + PR_L | SAGE | 83.24 | ± 1.47 | 77.82 | ± 1.09 | 83.09 | ± 2.27 | 151.0 |
| PMI_L + PR_L + Triplet_L | ALL | 91.86 | ± 1.28 | 88.60 | ± 1.51 | 63.10 | ± 4.21 | 152.3 |
| PMI_L + PR_L + Triplet_L | GAT | 96.20 | ± 0.51 | 98.38 | ± 0.48 | 75.41 | ± 1.94 | 79.7 |
| PMI_L + PR_L + Triplet_L | GCN | 95.35 | ± 0.31 | 97.93 | ± 0.51 | 79.25 | ± 1.01 | 84.0 |
| PMI_L + PR_L + Triplet_L | GIN | 89.12 | ± 1.87 | 89.06 | ± 1.17 | 64.62 | ± 0.94 | 153.7 |
| PMI_L + PR_L + Triplet_L | MPNN | 95.86 | ± 0.33 | 92.76 | ± 0.91 | 67.75 | ± 8.98 | 116.0 |

Lp Recall Continued (↑)

| Loss Type | Model | CORA | | Citeseer | | Bitcoin Fraud Transaction | | Average Rank |
|---|---|---|---|---|---|---|---|---|
| PMI_L + PR_L + Triplet_L | PAGNN | 79.27 ± 3.54 | | 89.08 ± 2.54 | | 83.86 ± 1.31 | | 136.3 |
| PMI_L + PR_L + Triplet_L | SAGE | 91.65 ± 2.43 | | 96.18 ± 0.30 | | 91.28 ± 0.39 | | 74.3 |
| PMI_L + Triplet_L | ALL | 94.04 ± 0.37 | | 97.55 ± 0.37 | | 79.48 ± 1.36 | | 93.7 |
| PMI_L + Triplet_L | GAT | 97.39 ± 0.43 | | 99.12 ± 0.08 | | 82.30 ± 1.70 | | 40.0 |
| PMI_L + Triplet_L | GCN | 94.93 ± 0.34 | | 97.82 ± 0.37 | | 79.77 ± 0.98 | | 87.3 |
| PMI_L + Triplet_L | GIN | 88.93 ± 2.55 | | 91.54 ± 1.10 | | 66.70 ± 1.70 | | 145.0 |
| PMI_L + Triplet_L | MPNN | 96.11 ± 1.19 | | 98.78 ± 0.33 | | 80.61 ± 1.28 | | 61.0 |
| PMI_L + Triplet_L | PAGNN | 79.80 ± 1.04 | | 89.04 ± 2.82 | | 82.88 ± 1.23 | | 140.0 |
| PMI_L + Triplet_L | SAGE | 92.59 ± 0.70 | | 95.75 ± 0.51 | | 77.22 ± 2.61 | | 112.0 |
| PR_L | ALL | **100.00 ± 0.00** | | 99.70 ± 0.44 | | 58.66 ± 2.48 | | 69.7 |
| PR_L | GAT | 85.50 ± 3.06 | | 83.12 ± 1.17 | | 64.59 ± 4.18 | | 176.0 |
| PR_L | GCN | 81.03 ± 0.95 | | 83.93 ± 1.86 | | 75.09 ± 1.79 | | 171.3 |
| PR_L | GIN | 92.26 ± 16.80 | | 67.41 ± 4.08 | | 51.57 ± 2.08 | | 176.7 |
| PR_L | MPNN | 69.30 ± 12.17 | | 76.90 ± 1.50 | | 81.87 ± 7.08 | | 164.3 |
| PR_L | PAGNN | 99.81 ± 0.05 | | 99.76 ± 0.10 | | 98.13 ± 0.22 | | **5.0** |
| PR_L | SAGE | 74.96 ± 2.63 | | 70.85 ± 3.56 | | 85.35 ± 1.98 | | 154.7 |

Lp Recall Continued (↑)

| Loss Type | Model | CORA | | Citeseer | | Bitcoin Fraud Transaction | | Average Rank |
|---|---|---|---|---|---|---|---|---|
| PR_L + Triplet_L | ALL | 96.23 ± 7.09 | | 99.06 ± 0.30 | | 57.01 ± 1.94 | | 89.7 |
| PR_L + Triplet_L | GAT | 85.22 ± 8.05 | | 83.81 ± 2.83 | | 66.98 ± 4.34 | | 173.7 |
| PR_L + Triplet_L | GCN | 85.71 ± 1.06 | | 87.03 ± 1.58 | | 66.31 ± 2.69 | | 169.0 |
| PR_L + Triplet_L | GIN | 84.28 ± 20.79 | | 77.96 ± 2.80 | | 50.43 ± 1.32 | | 191.3 |
| PR_L + Triplet_L | MPNN | 69.20 ± 5.60 | | 77.83 ± 1.51 | | 72.75 ± 9.09 | | 186.7 |
| PR_L + Triplet_L | PAGNN | 99.78 ± 0.07 | | 74.04 ± 3.08 | | 85.72 ± 4.36 | | 86.7 |
| PR_L + Triplet_L | SAGE | 75.28 ± 2.76 | | 80.66 ± 13.20 | | 87.18 ± 2.02 | | 143.0 |
| Triplet_L | ALL | 96.70 ± 0.83 | | 98.39 ± 0.17 | | 83.59 ± 1.06 | | 50.0 |
| Triplet_L | GAT | 99.27 ± 0.15 | | **99.84 ± 0.05** | | 87.76 ± 1.52 | | 17.3 |
| Triplet_L | GCN | 97.45 ± 0.45 | | 99.00 ± 0.35 | | 84.69 ± 2.58 | | 34.3 |
| Triplet_L | GIN | 97.66 ± 0.57 | | 98.09 ± 0.67 | | 80.64 ± 2.19 | | 55.7 |
| Triplet_L | MPNN | 96.49 ± 0.91 | | 99.08 ± 0.30 | | 86.22 ± 1.16 | | 35.7 |
| Triplet_L | PAGNN | 93.52 ± 5.16 | | 97.59 ± 0.45 | | 90.34 ± 0.48 | | 66.0 |
| Triplet_L | SAGE | 97.59 ± 0.16 | | 98.85 ± 0.44 | | 85.52 ± 0.82 | | 32.7 |

Table 11. Lp Specificity Performance (↑): Top-ranked results are highlighted in **1st**, second-ranked in **2nd**, and third-ranked in **3rd**.

| Loss Type | Model | CORA | | Citeseer | | Bitcoin Fraud Transaction | | Average Rank |
|---|---|---|---|---|---|---|---|---|
| Contr_l | ALL | 86.46 | ± | 93.64 | ± | 87.66 | ± | 112.3 |
| | | 1.66 | | 1.75 | | 2.00 | | |
| Contr_l | GAT | 95.38 | ± | 98.83 | ± | 94.48 | ± | 21.7 |
| | | 0.53 | | 0.19 | | 0.53 | | |
| Contr_l | GCN | 95.47 | ± | 98.21 | ± | 92.73 | ± | 40.7 |
| | | 0.67 | | 0.18 | | 1.40 | | |
| Contr_l | GIN | 90.26 | ± | 96.95 | ± | 89.87 | ± | 89.7 |
| | | 0.37 | | 0.52 | | 2.47 | | |
| Contr_l | MPNN | 91.80 | ± | 96.77 | ± | 89.06 | ± | 91.0 |
| | | 1.06 | | 0.26 | | 0.79 | | |
| Contr_l | PAGNN | 88.64 | ± | 92.60 | ± | 40.46 | ± | 133.3 |
| | | 1.85 | | 1.68 | | 1.78 | | |
| Contr_l | SAGE | 92.75 | ± | 97.24 | ± | 84.26 | ± | 90.7 |
| | | 0.72 | | 0.36 | | 2.68 | | |
| Contr_l + CrossE_L | ALL | 86.98 | ± | 94.02 | ± | 85.22 | ± | 113.7 |
| | | 1.73 | | 0.86 | | 2.10 | | |
| Contr_l + CrossE_L | GAT | 95.54 | ± | 98.80 | ± | 94.89 | ± | 20.7 |
| | | 0.55 | | 0.15 | | 0.90 | | |
| Contr_l + CrossE_L | GCN | 94.89 | ± | 98.09 | ± | 93.33 | ± | 42.0 |
| | | 0.74 | | 0.28 | | 0.95 | | |
| Contr_l + CrossE_L | GIN | 90.61 | ± | 96.64 | ± | 88.01 | ± | 97.0 |
| | | 1.24 | | 0.31 | | 2.61 | | |
| Contr_l + CrossE_L | MPNN | 91.14 | ± | 97.18 | ± | 90.00 | ± | 86.3 |
| | | 0.49 | | 0.25 | | 1.64 | | |
| Contr_l + CrossE_L | PAGNN | 87.38 | ± | 91.55 | ± | 38.93 | ± | 141.0 |
| | | 1.15 | | 0.41 | | 0.43 | | |
| Contr_l + CrossE_L | SAGE | 93.09 | ± | 97.59 | ± | 86.03 | ± | 85.0 |
| | | 0.42 | | 0.19 | | 2.18 | | |
| Contr_l + CrossE_L + PMI_L | ALL | 81.62 | ± | 85.63 | ± | 82.29 | ± | 136.7 |
| | | 5.31 | | 2.66 | | 1.94 | | |
| Contr_l + CrossE_L + PMI_L | GAT | 96.76 | ± | 98.70 | ± | 89.71 | ± | 35.3 |
| | | 0.38 | | 0.22 | | 2.02 | | |

Lp Specificity Continued (↑)

| Loss Type | Model | CORA | | Citeseer | | Bitcoin Fraud Transaction | | Average Rank |
|-----------|-------|------|---|----------|---|--------------------------|---|--------------|
| Contr_l + CrossE_L + PMI_L | GCN | 96.30 0.76 | ± | 98.31 0.41 | ± | 92.53 0.77 | ± | 33.3 |
| Contr_l + CrossE_L + PMI_L | GIN | 81.56 3.40 | ± | 86.27 3.77 | ± | 59.32 4.24 | ± | 145.3 |
| Contr_l + CrossE_L + PMI_L | MPNN | 93.51 0.72 | ± | 97.63 0.52 | ± | 92.05 1.25 | ± | 66.0 |
| Contr_l + CrossE_L + PMI_L | PAGNN | 27.10 14.17 | ± | 60.06 2.95 | ± | 38.19 0.53 | ± | 187.0 |
| Contr_l + CrossE_L + PMI_L | SAGE | 71.19 3.86 | ± | 83.63 2.71 | ± | 41.63 10.00 | ± | 158.3 |
| Contr_l + CrossE_L + PMI_L + PR_L | ALL | 45.30 11.59 | ± | 85.12 3.82 | ± | 72.23 1.38 | ± | 156.7 |
| Contr_l + CrossE_L + PMI_L + PR_L | GAT | 95.77 0.60 | ± | 98.42 0.24 | ± | 69.32 23.44 | ± | 67.7 |
| Contr_l + CrossE_L + PMI_L + PR_L | GCN | 95.67 0.52 | ± | 98.35 0.31 | ± | 92.01 1.10 | ± | 42.3 |
| Contr_l + CrossE_L + PMI_L + PR_L | GIN | 80.56 2.44 | ± | 82.59 0.87 | ± | 36.48 2.53 | ± | 166.7 |
| Contr_l + CrossE_L + PMI_L + PR_L | MPNN | 93.19 0.52 | ± | 94.44 3.09 | ± | 48.39 25.39 | ± | 117.0 |
| Contr_l + CrossE_L + PMI_L + PR_L | PAGNN | 44.46 13.36 | ± | 59.29 1.69 | ± | 21.90 0.26 | ± | 192.0 |
| Contr_l + CrossE_L + PMI_L + PR_L | SAGE | 69.03 4.31 | ± | 81.15 3.00 | ± | 38.04 1.86 | ± | 168.3 |
| Contr_l + CrossE_L + PMI_L + PR_L + Triplet_L | ALL | 81.42 2.59 | ± | 85.07 1.81 | ± | 87.98 0.97 | ± | 131.3 |
| Contr_l + CrossE_L + PMI_L + PR_L + Triplet_L | GAT | 96.23 0.38 | ± | 98.27 0.44 | ± | 89.46 6.19 | ± | 50.0 |
| Contr_l + CrossE_L + PMI_L + PR_L + Triplet_L | GCN | 96.66 0.27 | ± | 98.44 0.17 | ± | 92.84 0.94 | ± | 25.3 |
| Contr_l + CrossE_L + PMI_L + PR_L + Triplet_L | GIN | 81.80 4.15 | ± | 86.92 1.37 | ± | 78.16 4.07 | ± | 138.0 |

Lp Specificity Continued (↑)

| Loss Type | Model | CORA | | Citeseer | | Bitcoin Fraud Transaction | | Average Rank |
|---|---|---|---|---|---|---|---|---|
| Contr_l + CrossE_L + PMI_L + PR_L + Triplet_L | MPNN | 93.19 1.03 | ± | 95.51 1.20 | ± | 91.38 4.12 | ± | 85.3 |
| Contr_l + CrossE_L + PMI_L + PR_L + Triplet_L | PAGNN | 50.87 7.24 | ± | 61.34 1.41 | ± | 37.44 0.60 | ± | 182.0 |
| Contr_l + CrossE_L + PMI_L + PR_L + Triplet_L | SAGE | 85.98 3.86 | ± | 94.11 0.55 | ± | 38.35 0.54 | ± | 140.3 |
| Contr_l + CrossE_L + PMI_L + Triplet_L | ALL | 89.77 0.95 | ± | 96.88 0.34 | ± | 89.42 1.21 | ± | 93.0 |
| Contr_l + CrossE_L + PMI_L + Triplet_L | GAT | 96.81 0.63 | ± | 98.81 0.05 | ± | 92.87 1.11 | ± | 15.7 |
| Contr_l + CrossE_L + PMI_L + Triplet_L | GCN | 96.11 0.74 | ± | 98.53 0.16 | ± | 93.85 1.20 | ± | 21.7 |
| Contr_l + CrossE_L + PMI_L + Triplet_L | GIN | 83.56 1.60 | ± | 87.01 1.90 | ± | 77.82 4.32 | ± | 135.7 |
| Contr_l + CrossE_L + PMI_L + Triplet_L | MPNN | 93.36 1.18 | ± | 97.42 0.63 | ± | 91.39 1.41 | ± | 72.7 |
| Contr_l + CrossE_L + PMI_L + Triplet_L | PAGNN | 57.85 5.85 | ± | 59.75 3.04 | ± | 39.23 2.24 | ± | 179.0 |
| Contr_l + CrossE_L + PMI_L + Triplet_L | SAGE | 87.82 1.92 | ± | 93.52 0.86 | ± | 80.29 2.48 | ± | 120.7 |
| Contr_l + CrossE_L + PR_L | ALL | 13.72 18.85 | ± | 30.89 22.00 | ± | 95.68 0.66 | ± | 136.0 |
| Contr_l + CrossE_L + PR_L | GAT | 95.14 1.52 | ± | 98.30 0.09 | ± | 92.16 0.66 | ± | 47.3 |
| Contr_l + CrossE_L + PR_L | GCN | 93.77 0.65 | ± | 96.77 0.78 | ± | 91.48 1.39 | ± | 75.3 |
| Contr_l + CrossE_L + PR_L | GIN | 66.20 28.80 | ± | 89.33 3.08 | ± | 64.78 36.25 | ± | 145.0 |
| Contr_l + CrossE_L + PR_L | MPNN | 89.60 1.75 | ± | 97.12 0.80 | ± | 61.55 25.13 | ± | 111.3 |
| Contr_l + CrossE_L + PR_L | PAGNN | 2.21 0.71 | ± | 42.80 25.94 | ± | 39.21 6.21 | ± | 191.7 |

Continued on next page

Lp Specificity Continued (↑)

| Loss Type | Model | CORA | | Citeseer | | Bitcoin Fraud Transaction | | Average Rank |
|---|---|---|---|---|---|---|---|---|
| Contr_l + CrossE_L + PR_L | SAGE | 86.76 | ± | 88.29 | ± | 35.24 | ± | 150.0 |
| | | 11.15 | | 15.50 | | 1.82 | | |
| Contr_l + CrossE_L + PR_L + Triplet_L | ALL | 77.99 | ± | 89.51 | ± | 81.76 | ± | 133.7 |
| | | 2.62 | | 1.20 | | 4.00 | | |
| Contr_l + CrossE_L + PR_L + Triplet_L | GAT | 96.71 | ± | 98.59 | ± | 93.09 | ± | 19.3 |
| | | 0.76 | | 0.50 | | 1.18 | | |
| Contr_l + CrossE_L + PR_L + Triplet_L | GCN | 95.65 | ± | 97.74 | ± | 92.18 | ± | 51.3 |
| | | 0.85 | | 0.56 | | 2.11 | | |
| Contr_l + CrossE_L + PR_L + Triplet_L | GIN | 88.76 | ± | 95.12 | ± | 83.77 | ± | 111.0 |
| | | 2.15 | | 1.35 | | 6.10 | | |
| Contr_l + CrossE_L + PR_L + Triplet_L | MPNN | 91.26 | ± | 96.64 | ± | 87.28 | ± | 96.7 |
| | | 1.36 | | 0.46 | | 5.31 | | |
| Contr_l + CrossE_L + PR_L + Triplet_L | PAGNN | 44.34 | ± | 80.20 | ± | 32.88 | ± | 182.7 |
| | | 9.40 | | 7.82 | | 4.68 | | |
| Contr_l + CrossE_L + PR_L + Triplet_L | SAGE | 94.56 | ± | 97.17 | ± | 69.29 | ± | 92.3 |
| | | 0.66 | | 0.80 | | 5.16 | | |
| Contr_l + CrossE_L + Triplet_L | ALL | 91.32 | ± | 96.55 | ± | 91.07 | ± | 89.0 |
| | | 0.91 | | 0.49 | | 1.36 | | |
| Contr_l + CrossE_L + Triplet_L | GAT | 96.76 | ± | 99.11 | ± | 96.11 | ± | 6.0 |
| | | 0.35 | | 0.11 | | 0.64 | | |
| Contr_l + CrossE_L + Triplet_L | GCN | 95.91 | ± | 98.35 | ± | 93.73 | ± | 27.3 |
| | | 0.51 | | 0.26 | | 0.63 | | |
| Contr_l + CrossE_L + Triplet_L | GIN | 93.03 | ± | 97.64 | ± | 90.46 | ± | 76.0 |
| | | 0.95 | | 0.47 | | 2.07 | | |
| Contr_l + CrossE_L + Triplet_L | MPNN | 93.33 | ± | 97.95 | ± | 92.22 | ± | 60.7 |
| | | 0.43 | | 0.19 | | 0.94 | | |
| Contr_l + CrossE_L + Triplet_L | PAGNN | 85.83 | ± | 95.28 | ± | 40.12 | ± | 133.0 |
| | | 7.52 | | 0.55 | | 0.47 | | |
| Contr_l + CrossE_L + Triplet_L | SAGE | 95.25 | ± | 98.12 | ± | 91.76 | ± | 51.3 |
| | | 0.69 | | 0.21 | | 1.14 | | |
| Contr_l + PMI_L | ALL | 83.61 | ± | 91.59 | ± | 82.84 | ± | 125.0 |
| | | 1.12 | | 2.15 | | 2.93 | | |

Continued on next page

Lp Specificity Continued (↑)

| Loss Type | Model | CORA | | | Citeseer | | | Bitcoin Fraud Transaction | | | Average Rank |
|---|---|---|---|---|---|---|---|---|---|---|---|
| Contr_l + PMI_L | GAT | 97.04 | ± | 0.46 | 98.76 | ± | 0.08 | 89.41 | ± | 1.66 | 33.3 |
| Contr_l + PMI_L | GCN | 95.94 | ± | 0.54 | 98.25 | ± | 0.36 | 92.69 | ± | 1.31 | 36.7 |
| Contr_l + PMI_L | GIN | 80.91 | ± | 2.49 | 87.70 | ± | 1.36 | 61.36 | ± | 4.92 | 143.0 |
| Contr_l + PMI_L | MPNN | 93.96 | ± | 0.84 | 97.53 | ± | 0.60 | 91.78 | ± | 1.33 | 64.3 |
| Contr_l + PMI_L | PAGNN | 35.55 | ± | 18.73 | 59.08 | ± | 1.45 | 37.97 | ± | 0.93 | 188.0 |
| Contr_l + PMI_L | SAGE | 73.32 | ± | 3.79 | 87.32 | ± | 2.85 | 38.07 | ± | 2.46 | 158.0 |
| Contr_l + PMI_L + PR_L | ALL | 58.62 | ± | 5.19 | 84.60 | ± | 3.70 | 76.07 | ± | 1.36 | 153.7 |
| Contr_l + PMI_L + PR_L | GAT | 95.51 | ± | 0.30 | 96.57 | ± | 1.43 | 49.97 | ± | 25.12 | 97.0 |
| Contr_l + PMI_L + PR_L | GCN | 96.04 | ± | 0.61 | 98.47 | ± | 0.26 | 92.51 | ± | 0.71 | 32.7 |
| Contr_l + PMI_L + PR_L | GIN | 82.51 | ± | 2.87 | 85.18 | ± | 4.89 | 39.32 | ± | 4.24 | 155.0 |
| Contr_l + PMI_L + PR_L | MPNN | 93.92 | ± | 0.72 | 93.66 | ± | 2.20 | 58.04 | ± | 29.30 | 109.7 |
| Contr_l + PMI_L + PR_L | PAGNN | 38.65 | ± | 14.64 | 58.12 | ± | 3.45 | 22.22 | ± | 0.68 | 193.7 |
| Contr_l + PMI_L + PR_L | SAGE | 71.69 | ± | 2.67 | 84.20 | ± | 2.02 | 39.35 | ± | 0.94 | 160.7 |
| Contr_l + PMI_L + PR_L + Triplet_L | ALL | 84.30 | ± | 1.79 | 88.83 | ± | 1.17 | 84.03 | ± | 4.15 | 126.0 |
| Contr_l + PMI_L + PR_L + Triplet_L | GAT | 95.83 | ± | 0.25 | 98.34 | ± | 0.18 | 87.13 | ± | 6.51 | 55.7 |
| Contr_l + PMI_L + PR_L + Triplet_L | GCN | 95.85 | ± | 0.48 | 98.11 | ± | 0.39 | 93.57 | ± | 1.06 | 33.7 |

Lp Specificity Continued (↑)

| Loss Type | Model | CORA | | Citeseer | | Bitcoin Fraud Transaction | | Average Rank |
|---|---|---|---|---|---|---|---|---|
| Contr_l + PMI_L + PR_L + Triplet_L | GIN | 86.64 ± 1.39 | | 89.47 ± 0.90 | | 83.51 ± 2.45 | | 122.0 |
| Contr_l + PMI_L + PR_L + Triplet_L | MPNN | 93.42 ± 0.39 | | 94.66 ± 0.85 | | 86.52 ± 3.65 | | 96.7 |
| Contr_l + PMI_L + PR_L + Triplet_L | PAGNN | 60.02 ± 3.95 | | 70.96 ± 5.53 | | 37.16 ± 2.29 | | 178.7 |
| Contr_l + PMI_L + PR_L + Triplet_L | SAGE | 91.06 ± 1.35 | | 96.50 ± 0.96 | | 49.38 ± 8.20 | | 118.0 |
| Contr_l + PR_L | ALL | 36.20 ± 21.56 | | 19.42 ± 22.15 | | 95.93 ± 1.62 | | 132.7 |
| Contr_l + PR_L | GAT | 94.70 ± 1.13 | | 97.93 ± 0.30 | | 93.15 ± 0.73 | | 45.3 |
| Contr_l + PR_L | GCN | 93.79 ± 0.69 | | 96.67 ± 0.76 | | 91.19 ± 2.15 | | 76.7 |
| Contr_l + PR_L | GIN | 64.90 ± 33.25 | | 85.24 ± 4.68 | | 47.59 ± 43.48 | | 157.3 |
| Contr_l + PR_L | MPNN | 92.07 ± 0.57 | | 97.15 ± 0.73 | | 41.83 ± 8.10 | | 110.3 |
| Contr_l + PR_L | PAGNN | 1.76 ± 0.59 | | 60.79 ± 4.74 | | 34.06 ± 11.34 | | 193.0 |
| Contr_l + PR_L | SAGE | 85.59 ± 11.77 | | 92.24 ± 9.35 | | 36.60 ± 0.68 | | 147.7 |
| Contr_l + PR_L + Triplet_L | ALL | 67.35 ± 15.25 | | 83.68 ± 9.64 | | 82.36 ± 0.87 | | 144.0 |
| Contr_l + PR_L + Triplet_L | GAT | 96.34 ± 1.21 | | 98.59 ± 0.57 | | 93.66 ± 1.09 | | 18.7 |
| Contr_l + PR_L + Triplet_L | GCN | 95.05 ± 1.18 | | 97.83 ± 0.80 | | 92.21 ± 1.23 | | 54.7 |
| Contr_l + PR_L + Triplet_L | GIN | 88.55 ± 1.92 | | 92.52 ± 2.21 | | 85.16 ± 0.82 | | 115.3 |
| Contr_l + PR_L + Triplet_L | MPNN | 91.66 ± 2.56 | | 96.35 ± 0.53 | | 82.09 ± 4.70 | | 103.7 |

Lp Specificity Continued (↑)

| Loss Type | Model | CORA | | Citeseer | | Bitcoin Fraud Transaction | | Average Rank |
|---|---|---|---|---|---|---|---|---|
| Contr_l + PR_L + Triplet_L | PAGNN | 30.71 ± 12.89 | | 80.80 ± 11.01 | | 30.49 ± 0.17 | | 185.7 |
| Contr_l + PR_L + Triplet_L | SAGE | 93.88 ± 0.36 | | 97.83 ± 0.22 | | 77.14 ± 4.77 | | 85.0 |
| Contr_l + Triplet_L | ALL | 91.73 ± 0.66 | | 96.63 ± 0.27 | | 92.37 ± 0.98 | | 77.3 |
| Contr_l + Triplet_L | GAT | 96.50 ± 0.59 | | 98.98 ± 0.15 | | 96.35 ± 0.85 | | 7.7 |
| Contr_l + Triplet_L | GCN | 96.26 ± 0.30 | | 98.56 ± 0.27 | | 93.33 ± 1.10 | | 22.3 |
| Contr_l + Triplet_L | GIN | 93.42 ± 0.71 | | 98.15 ± 0.40 | | 89.12 ± 1.72 | | 71.0 |
| Contr_l + Triplet_L | MPNN | 93.60 ± 0.77 | | 97.93 ± 0.18 | | 92.26 ± 0.48 | | 58.0 |
| Contr_l + Triplet_L | PAGNN | 85.65 ± 10.14 | | 95.69 ± 0.72 | | 40.19 ± 1.19 | | 131.7 |
| Contr_l + Triplet_L | SAGE | 95.19 ± 0.25 | | 98.34 ± 0.25 | | 90.98 ± 2.23 | | 53.0 |
| CrossE_L | ALL | 70.02 ± 30.43 | | 85.70 ± 2.87 | | 89.26 ± 5.83 | | 131.0 |
| CrossE_L | GAT | 65.12 ± 37.26 | | 75.48 ± 42.28 | | 71.61 ± 39.54 | | 156.7 |
| CrossE_L | GCN | 0.00 ± 0.00 | | 0.00 ± 0.00 | | 0.00 ± 0.00 | | 209.3 |
| CrossE_L | GIN | 0.04 ± 0.04 | | 0.01 ± 0.03 | | 0.00 ± 0.00 | | 209.0 |
| CrossE_L | MPNN | 85.76 ± 2.41 | | 81.82 ± 8.47 | | 0.00 ± 0.00 | | 168.3 |
| CrossE_L | PAGNN | 26.77 ± 15.24 | | 56.71 ± 8.24 | | 0.08 ± 0.08 | | 200.3 |
| CrossE_L | SAGE | 49.85 ± 13.76 | | 35.81 ± 29.89 | | 0.01 ± 0.01 | | 196.3 |

Lp Specificity Continued (↑)

| Loss Type | Model | CORA | | | Citeseer | | | Bitcoin Fraud Transaction | | | Average Rank |
|---|---|---|---|---|---|---|---|---|---|---|---|
| CrossE_L + PMI_L | ALL | 86.10 | ± | 1.27 | 90.34 | ± | 2.70 | 87.00 | ± | 2.94 | 118.0 |
| CrossE_L + PMI_L | GAT | 97.03 | ± | 0.24 | 98.87 | ± | 0.17 | 89.57 | ± | 1.28 | 30.3 |
| CrossE_L + PMI_L | GCN | 96.42 | ± | 1.02 | 98.46 | ± | 0.17 | 91.97 | ± | 1.04 | 34.3 |
| CrossE_L + PMI_L | GIN | 82.33 | ± | 4.38 | 87.07 | ± | 2.67 | 57.05 | ± | 3.83 | 143.0 |
| CrossE_L + PMI_L | MPNN | 92.96 | ± | 0.49 | 97.80 | ± | 0.39 | 92.24 | ± | 1.43 | 66.0 |
| CrossE_L + PMI_L | PAGNN | 36.29 | ± | 15.91 | 57.50 | ± | 1.25 | 31.54 | ± | 8.84 | 192.7 |
| CrossE_L + PMI_L | SAGE | 62.34 | ± | 3.49 | 75.49 | ± | 1.74 | 18.48 | ± | 5.22 | 182.0 |
| CrossE_L + PMI_L + PR_L | ALL | 44.36 | ± | 3.79 | 86.96 | ± | 3.21 | 60.21 | ± | 13.12 | 157.7 |
| CrossE_L + PMI_L + PR_L | GAT | 96.18 | ± | 0.37 | 95.75 | ± | 2.41 | 69.42 | ± | 30.90 | 87.7 |
| CrossE_L + PMI_L + PR_L | GCN | 95.91 | ± | 0.98 | 98.28 | ± | 0.29 | 92.25 | ± | 2.17 | 40.0 |
| CrossE_L + PMI_L + PR_L | GIN | 78.87 | ± | 1.30 | 76.60 | ± | 7.92 | 44.58 | ± | 11.04 | 160.3 |
| CrossE_L + PMI_L + PR_L | MPNN | 93.09 | ± | 1.12 | 95.26 | ± | 2.81 | 68.89 | ± | 29.96 | 110.7 |
| CrossE_L + PMI_L + PR_L | PAGNN | 47.82 | ± | 17.33 | 58.86 | ± | 1.78 | 22.23 | ± | 3.45 | 190.7 |
| CrossE_L + PMI_L + PR_L | SAGE | 64.40 | ± | 2.48 | 77.42 | ± | 5.69 | 40.61 | ± | 2.74 | 166.0 |
| CrossE_L + PMI_L + PR_L + Triplet_L | ALL | 82.87 | ± | 1.13 | 82.40 | ± | 1.00 | 85.84 | ± | 2.06 | 134.7 |
| CrossE_L + PMI_L + PR_L + Triplet_L | GAT | 96.34 | ± | 0.42 | 98.03 | ± | 0.38 | 86.27 | ± | 7.54 | 57.3 |

Lp Specificity Continued (↑)

| Loss Type | Model | CORA | | Citeseer | | Bitcoin Fraud Transaction | | Average Rank |
|---|---|---|---|---|---|---|---|---|
| CrossE_L + PMI_L + PR_L + Triplet_L | GCN | 96.20 | ± | 98.65 | ± | 91.68 | ± | 34.3 |
| | | 0.83 | | 0.11 | | 1.47 | | |
| CrossE_L + PMI_L + PR_L + Triplet_L | GIN | 82.59 | ± | 85.67 | ± | 77.98 | ± | 139.3 |
| | | 2.63 | | 1.43 | | 1.97 | | |
| CrossE_L + PMI_L + PR_L + Triplet_L | MPNN | 93.63 | ± | 95.64 | ± | 91.49 | ± | 80.0 |
| | | 0.51 | | 1.74 | | 5.39 | | |
| CrossE_L + PMI_L + PR_L + Triplet_L | PAGNN | 62.10 | ± | 63.51 | ± | 37.84 | ± | 177.7 |
| | | 1.44 | | 3.96 | | 0.93 | | |
| CrossE_L + PMI_L + PR_L + Triplet_L | SAGE | 85.34 | ± | 92.97 | ± | 38.62 | ± | 143.3 |
| | | 5.17 | | 1.99 | | 1.32 | | |
| CrossE_L + PMI_L + Triplet_L | ALL | 91.45 | ± | 97.65 | ± | 91.48 | ± | 76.0 |
| | | 0.59 | | 0.57 | | 0.82 | | |
| CrossE_L + PMI_L + Triplet_L | GAT | <mark>97.07</mark> | <mark>±</mark> | 98.79 | ± | 93.65 | ± | 10.0 |
| | | <mark>0.47</mark> | | 0.06 | | 1.35 | | |
| CrossE_L + PMI_L + Triplet_L | GCN | 96.32 | ± | 98.51 | ± | 93.47 | ± | 22.0 |
| | | 0.90 | | 0.24 | | 0.48 | | |
| CrossE_L + PMI_L + Triplet_L | GIN | 85.22 | ± | 89.03 | ± | 84.53 | ± | 124.3 |
| | | 1.83 | | 1.24 | | 1.35 | | |
| CrossE_L + PMI_L + Triplet_L | MPNN | 93.50 | ± | 97.42 | ± | 92.60 | ± | 62.3 |
| | | 0.49 | | 0.42 | | 0.90 | | |
| CrossE_L + PMI_L + Triplet_L | PAGNN | 61.10 | ± | 60.36 | ± | 39.96 | ± | 174.0 |
| | | 5.39 | | 2.51 | | 3.41 | | |
| CrossE_L + PMI_L + Triplet_L | SAGE | 90.88 | ± | 95.49 | ± | 85.20 | ± | 105.0 |
| | | 1.64 | | 1.14 | | 0.91 | | |
| CrossE_L + PR_L | ALL | 17.27 | ± | 5.10 | ± | 95.74 | ± | 136.7 |
| | | 23.91 | | 2.99 | | 1.05 | | |
| CrossE_L + PR_L | GAT | 95.40 | ± | 98.14 | ± | 84.25 | ± | 66.0 |
| | | 0.52 | | 0.29 | | 18.31 | | |
| CrossE_L + PR_L | GCN | 92.57 | ± | 96.19 | ± | 50.55 | ± | 113.7 |
| | | 1.07 | | 1.13 | | 46.26 | | |
| CrossE_L + PR_L | GIN | 34.17 | ± | 88.61 | ± | 81.20 | ± | 150.0 |
| | | 40.47 | | 4.57 | | 0.91 | | |

Lp Specificity Continued (↑)

| Loss Type | Model | CORA | | Citeseer | | Bitcoin Fraud Transaction | | Average Rank |
|---|---|---|---|---|---|---|---|---|
| CrossE_L + PR_L | MPNN | 91.02 ± 1.16 | | 96.91 ± 0.99 | | 23.11 ± 6.95 | | 129.0 |
| CrossE_L + PR_L | PAGNN | 1.23 ± 0.39 | | 16.00 ± 26.15 | | 14.32 ± 0.68 | | 205.0 |
| CrossE_L + PR_L | SAGE | 55.13 ± 6.49 | | 50.73 ± 28.62 | | 10.60 ± 14.54 | | 194.0 |
| CrossE_L + PR_L + Triplet_L | ALL | 62.76 ± 19.71 | | 57.44 ± 15.21 | | 87.04 ± 2.85 | | 153.3 |
| CrossE_L + PR_L + Triplet_L | GAT | 95.82 ± 0.75 | | 98.37 ± 0.93 | | 92.02 ± 1.04 | | 41.0 |
| CrossE_L + PR_L + Triplet_L | GCN | 94.70 ± 0.80 | | 97.34 ± 0.40 | | 92.35 ± 0.65 | | 59.0 |
| CrossE_L + PR_L + Triplet_L | GIN | 86.29 ± 4.51 | | 94.40 ± 1.09 | | 81.16 ± 1.39 | | 120.0 |
| CrossE_L + PR_L + Triplet_L | MPNN | 89.79 ± 2.64 | | 96.82 ± 0.62 | | 69.34 ± 14.14 | | 110.0 |
| CrossE_L + PR_L + Triplet_L | PAGNN | 13.69 ± 15.48 | | 71.03 ± 5.04 | | 31.30 ± 1.22 | | 190.7 |
| CrossE_L + PR_L + Triplet_L | SAGE | 95.22 ± 1.14 | | 97.83 ± 0.51 | | 53.79 ± 17.90 | | 85.7 |
| CrossE_L + Triplet_L | ALL | 94.43 ± 0.30 | | 97.87 ± 0.40 | | 92.78 ± 1.08 | | 51.3 |
| CrossE_L + Triplet_L | GAT | 97.72 ± 0.33 | | 99.26 ± 0.12 | | 96.04 ± 0.52 | | 2.3 |
| CrossE_L + Triplet_L | GCN | 97.02 ± 0.61 | | 98.71 ± 0.42 | | 93.10 ± 1.24 | | 14.3 |
| CrossE_L + Triplet_L | GIN | 94.43 ± 0.60 | | 97.93 ± 0.18 | | 89.16 ± 1.63 | | 69.0 |
| CrossE_L + Triplet_L | MPNN | 95.13 ± 0.80 | | 98.53 ± 0.30 | | 93.02 ± 0.51 | | 34.7 |
| CrossE_L + Triplet_L | PAGNN | 89.02 ± 6.67 | | 96.94 ± 0.50 | | 39.32 ± 0.33 | | 122.3 |

Lp Specificity Continued (↑)

| Loss Type | Model | CORA | | Citeseer | | Bitcoin Fraud Transaction | | Average Rank |
|---|---|---|---|---|---|---|---|---|
| CrossE_L + Triplet_L | SAGE | 96.78 | ± 0.42 | 98.64 | ± 0.21 | 90.32 | ± 1.40 | 33.7 |
| PMI_L | ALL | 82.90 | ± 2.12 | 92.11 | ± 1.60 | 89.74 | ± 1.02 | 115.0 |
| PMI_L | GAT | 97.00 | ± 0.30 | 98.63 | ± 0.41 | 90.43 | ± 1.30 | 32.7 |
| PMI_L | GCN | 96.29 | ± 0.74 | 98.33 | ± 0.32 | 92.91 | ± 1.43 | 30.3 |
| PMI_L | GIN | 77.65 | ± 3.13 | 86.62 | ± 4.31 | 52.70 | ± 7.48 | 149.3 |
| PMI_L | MPNN | 93.82 | ± 0.55 | 97.68 | ± 0.27 | 92.56 | ± 1.54 | 56.7 |
| PMI_L | PAGNN | 47.24 | ± 14.41 | 57.85 | ± 3.35 | 37.70 | ± 0.98 | 187.0 |
| PMI_L | SAGE | 64.41 | ± 3.58 | 75.83 | ± 1.73 | 22.14 | ± 8.65 | 180.0 |
| PMI_L + PR_L | ALL | 49.10 | ± 9.03 | 82.80 | ± 13.38 | 67.22 | ± 1.05 | 160.0 |
| PMI_L + PR_L | GAT | 95.38 | ± 1.04 | 94.93 | ± 4.16 | 53.07 | ± 29.28 | 102.0 |
| PMI_L + PR_L | GCN | 96.42 | ± 0.56 | 98.35 | ± 0.19 | 92.11 | ± 1.57 | 35.0 |
| PMI_L + PR_L | GIN | 80.72 | ± 3.43 | 79.15 | ± 6.36 | 39.43 | ± 4.72 | 162.0 |
| PMI_L + PR_L | MPNN | 93.78 | ± 0.55 | 92.61 | ± 1.73 | 48.63 | ± 25.52 | 115.3 |
| PMI_L + PR_L | PAGNN | 49.97 | ± 3.22 | 58.75 | ± 4.54 | 16.15 | ± 10.14 | 191.7 |
| PMI_L + PR_L | SAGE | 65.00 | ± 3.11 | 77.81 | ± 3.37 | 39.88 | ± 2.25 | 167.0 |
| PMI_L + PR_L + Triplet_L | ALL | 81.17 | ± 2.37 | 86.16 | ± 1.31 | 87.70 | ± 0.83 | 129.7 |

Lp Specificity Continued (↑)

| Loss Type | Model | CORA | | | Citeseer | | | Bitcoin Fraud Transaction | | | Average Rank |
|---|---|---|---|---|---|---|---|---|---|---|---|
| PMI_L + PR_L + Triplet_L | GAT | 95.65 | ± | 0.23 | 97.82 | ± | 0.68 | 81.57 | ± | 6.37 | 73.7 |
| PMI_L + PR_L + Triplet_L | GCN | 96.15 | ± | 0.45 | 98.08 | ± | 0.36 | 92.83 | ± | 0.54 | 37.0 |
| PMI_L + PR_L + Triplet_L | GIN | 83.77 | ± | 1.69 | 86.20 | ± | 2.29 | 80.96 | ± | 1.56 | 135.0 |
| PMI_L + PR_L + Triplet_L | MPNN | 93.51 | ± | 0.89 | 94.30 | ± | 1.01 | 90.32 | ± | 3.50 | 90.0 |
| PMI_L + PR_L + Triplet_L | PAGNN | 61.42 | ± | 8.31 | 64.34 | ± | 1.83 | 39.31 | ± | 2.15 | 174.0 |
| PMI_L + PR_L + Triplet_L | SAGE | 89.70 | ± | 4.35 | 94.55 | ± | 1.16 | 38.79 | ± | 0.47 | 133.0 |
| PMI_L + Triplet_L | ALL | 91.30 | ± | 1.40 | 97.23 | ± | 0.59 | 91.13 | ± | 0.92 | 82.0 |
| PMI_L + Triplet_L | GAT | 96.86 | ± | 0.45 | 98.83 | ± | 0.17 | 92.41 | ± | 0.84 | 18.3 |
| PMI_L + Triplet_L | GCN | 96.56 | ± | 0.50 | 98.50 | ± | 0.43 | 93.08 | ± | 0.42 | 22.7 |
| PMI_L + Triplet_L | GIN | 82.98 | ± | 3.29 | 87.06 | ± | 2.26 | 82.13 | ± | 2.69 | 131.3 |
| PMI_L + Triplet_L | MPNN | 93.80 | ± | 0.59 | 97.20 | ± | 0.40 | 92.31 | ± | 1.00 | 63.3 |
| PMI_L + Triplet_L | PAGNN | 58.49 | ± | 2.03 | 61.92 | ± | 3.87 | 40.55 | ± | 2.48 | 172.7 |
| PMI_L + Triplet_L | SAGE | 87.52 | ± | 3.29 | 95.34 | ± | 0.64 | 81.25 | ± | 4.28 | 114.7 |
| PR_L | ALL | 0.09 | ± | 0.08 | 1.55 | ± | 1.77 | 95.52 | ± | 1.17 | 141.0 |
| PR_L | GAT | 95.06 | ± | 0.67 | 98.52 | ± | 0.25 | 93.02 | ± | 1.20 | 35.7 |
| PR_L | GCN | 92.45 | ± | 2.18 | 96.83 | ± | 0.54 | 91.83 | ± | 1.67 | 78.7 |

Lp Specificity Continued (↑)

| Loss Type | Model | CORA | | Citeseer | | Bitcoin Fraud Transaction | | Average Rank |
|---|---|---|---|---|---|---|---|---|
| PR_L | GIN | 17.90 ± 35.53 | | 90.54 ± 2.80 | | 81.01 ± 1.80 | | 149.7 |
| PR_L | MPNN | 91.62 ± 1.08 | | 96.76 ± 1.40 | | 45.74 ± 27.26 | | 114.7 |
| PR_L | PAGNN | 1.28 ± 0.43 | | 2.66 ± 0.77 | | 14.38 ± 0.42 | | 205.3 |
| PR_L | SAGE | 59.08 ± 4.99 | | 66.12 ± 5.20 | | 25.49 ± 4.14 | | 183.0 |
| PR_L + Triplet_L | ALL | 9.67 ± 17.68 | | 6.28 ± 1.97 | | 94.89 ± 0.26 | | 139.0 |
| PR_L + Triplet_L | GAT | 95.22 ± 1.12 | | 97.91 ± 0.51 | | 91.51 ± 1.47 | | 56.3 |
| PR_L + Triplet_L | GCN | 93.58 ± 0.39 | | 96.79 ± 0.70 | | 91.16 ± 1.80 | | 77.7 |
| PR_L + Triplet_L | GIN | 35.94 ± 42.03 | | 82.30 ± 4.37 | | 78.65 ± 1.45 | | 160.0 |
| PR_L + Triplet_L | MPNN | 91.18 ± 2.21 | | 97.42 ± 0.70 | | 65.47 ± 26.85 | | 104.3 |
| PR_L + Triplet_L | PAGNN | 1.72 ± 0.50 | | 60.67 ± 6.95 | | 36.16 ± 3.63 | | 192.7 |
| PR_L + Triplet_L | SAGE | 68.38 ± 4.11 | | 79.14 ± 18.03 | | 35.35 ± 1.38 | | 173.0 |
| Triplet_L | ALL | 93.72 ± 0.57 | | 97.95 ± 0.25 | | 92.11 ± 1.32 | | 59.3 |
| Triplet_L | GAT | 98.13 ± 0.44 | | 99.43 ± 0.07 | | 95.24 ± 0.57 | | 3.3 |
| Triplet_L | GCN | 96.77 ± 0.27 | | 98.90 ± 0.07 | | 93.84 ± 0.76 | | 9.7 |
| Triplet_L | GIN | 95.14 ± 1.27 | | 98.07 ± 0.31 | | 89.20 ± 1.02 | | 63.3 |
| Triplet_L | MPNN | 94.83 ± 1.09 | | 98.48 ± 0.23 | | 93.15 ± 0.43 | | 36.0 |

Lp Specificity Continued (↑)

| Loss Type | Model | CORA | | Citeseer | | Bitcoin Fraud Transaction | | Average Rank |
|-----------|-------|------|---|----------|---|---------|---|--------------|
| Triplet_L | PAGNN | 88.56 ± 7.58 | | 97.30 ± 0.27 | | 39.92 ± 0.69 | | 118.7 |
| Triplet_L | SAGE | 96.76 ± 0.65 | | 98.77 ± 0.20 | | 90.60 ± 0.60 | | 32.3 |

*1.1.3 Evaluation on Embedding–Adjacency Alignment.*

Table 12. Cosine-Adj Corr Performance (↑): Top-ranked results are highlighted in 1st, second-ranked in 2nd, and third-ranked in 3rd.

| Loss Type | Model | CORA | | Citeseer | | Bitcoin Fraud Transaction | | Average Rank |
|-----------|-------|------|---|----------|---|---------|---|--------------|
| Contr_l | ALL | 7.66 ± 0.21 | | 6.05 ± 0.21 | | 3.66 ± 0.10 | | 87.3 |
| Contr_l | GAT | 12.04 ± 0.45 | | 12.47 ± 0.27 | | 5.90 ± 0.16 | | 17.0 |
| Contr_l | GCN | 10.96 ± 0.13 | | 10.91 ± 0.32 | | 5.36 ± 0.22 | | 29.3 |
| Contr_l | GIN | 10.47 ± 0.43 | | 10.61 ± 0.45 | | 3.91 ± 0.53 | | 49.0 |
| Contr_l | MPNN | 9.15 ± 0.29 | | 8.86 ± 0.59 | | 3.86 ± 0.06 | | 68.3 |
| Contr_l | PAGNN | 8.42 ± 0.62 | | 6.55 ± 0.29 | | 1.00 ± 0.02 | | 105.3 |
| Contr_l | SAGE | 9.43 ± 0.28 | | 8.94 ± 0.74 | | 3.42 ± 0.17 | | 69.7 |
| Contr_l + CrossE_L | ALL | 7.44 ± 0.33 | | 6.14 ± 0.06 | | 3.48 ± 0.19 | | 89.3 |
| Contr_l + CrossE_L | GAT | 11.93 ± 0.71 | | 12.89 ± 0.14 | | 6.18 ± 0.51 | | 15.0 |

Continued on next page

Cosine-Adj Corr Continued (↑)

| Loss Type | Model | CORA | | Citeseer | | Bitcoin Fraud Transaction | | Average Rank |
|-----------|-------|------|--|----------|--|---------------------------|--|--------------|
| Contr_l + CrossE_L | GCN | 11.10 | ± | 10.59 | ± | 5.26 | ± | 31.0 |
| | | 0.17 | | 0.26 | | 0.21 | | |
| Contr_l + CrossE_L | GIN | 9.80 | ± | 9.33 | ± | 3.61 | ± | 63.3 |
| | | 0.39 | | 0.51 | | 0.46 | | |
| Contr_l + CrossE_L | MPNN | 9.39 | ± | 9.09 | ± | 4.14 | ± | 63.0 |
| | | 0.64 | | 0.35 | | 0.10 | | |
| Contr_l + CrossE_L | PAGNN | 8.28 | ± | 6.36 | ± | 1.01 | ± | 106.7 |
| | | 0.46 | | 0.20 | | 0.01 | | |
| Contr_l + CrossE_L | SAGE | 9.54 | ± | 9.53 | ± | 3.41 | ± | 66.0 |
| | | 0.48 | | 0.66 | | 0.31 | | |
| Contr_l + CrossE_L + PMI_L | ALL | 5.55 | ± | 3.61 | ± | 2.60 | ± | 121.7 |
| | | 0.55 | | 0.60 | | 0.08 | | |
| Contr_l + CrossE_L + PMI_L | GAT | 12.45 | ± | 12.71 | ± | 4.08 | ± | 28.0 |
| | | 0.41 | | 0.48 | | 0.30 | | |
| Contr_l + CrossE_L + PMI_L | GCN | 12.40 | ± | 11.59 | ± | 4.97 | ± | 23.0 |
| | | 0.59 | | 0.28 | | 0.09 | | |
| Contr_l + CrossE_L + PMI_L | GIN | 6.44 | ± | 4.11 | ± | 1.29 | ± | 123.7 |
| | | 0.75 | | 0.22 | | 0.22 | | |
| Contr_l + CrossE_L + PMI_L | MPNN | 10.55 | ± | 9.91 | ± | 4.41 | ± | 47.3 |
| | | 0.76 | | 0.57 | | 0.25 | | |
| Contr_l + CrossE_L + PMI_L | PAGNN | 2.75 | ± | 2.90 | ± | 0.43 | ± | 168.3 |
| | | 0.20 | | 0.07 | | 0.02 | | |
| Contr_l + CrossE_L + PMI_L | SAGE | 5.71 | ± | 5.56 | ± | 1.27 | ± | 119.0 |
| | | 0.16 | | 0.61 | | 0.33 | | |
| Contr_l + CrossE_L + PMI_L + PR_L | ALL | 3.20 | ± | 2.95 | ± | 0.75 | ± | 157.7 |
| | | 0.41 | | 0.24 | | 0.02 | | |
| Contr_l + CrossE_L + PMI_L + PR_L | GAT | 10.99 | ± | 11.08 | ± | 2.08 | ± | 60.3 |
| | | 1.31 | | 0.46 | | 0.74 | | |
| Contr_l + CrossE_L + PMI_L + PR_L | GCN | 12.20 | ± | 11.21 | ± | 4.79 | ± | 32.3 |
| | | 0.63 | | 0.34 | | 0.21 | | |
| Contr_l + CrossE_L + PMI_L + PR_L | GIN | 6.05 | ± | 3.74 | ± | 0.92 | ± | 134.0 |
| | | 0.67 | | 0.29 | | 0.05 | | |

Cosine-Adj Corr Continued (↑)

| Loss Type | Model | CORA | | Citeseer | | Bitcoin Fraud Transaction | | Average Rank |
|---|---|---|---|---|---|---|---|---|
| Contr_l + CrossE_L + PMI_L + PR_L | MPNN | 9.40 0.48 | ± | 6.15 1.55 | ± | 1.70 1.44 | ± | 97.3 |
| Contr_l + CrossE_L + PMI_L + PR_L | PAGNN | 3.16 0.24 | ± | 3.00 0.08 | ± | 0.27 0.05 | ± | 165.3 |
| Contr_l + CrossE_L + PMI_L + PR_L | SAGE | 5.52 0.48 | ± | 5.10 0.54 | ± | 0.58 0.05 | ± | 134.3 |
| Contr_l + CrossE_L + PMI_L + PR_L + Triplet_L | ALL | 6.91 0.11 | ± | 4.43 0.06 | ± | 2.19 0.14 | ± | 111.0 |
| Contr_l + CrossE_L + PMI_L + PR_L + Triplet_L | GAT | 10.90 0.94 | ± | 10.21 2.30 | ± | 4.02 0.97 | ± | 47.7 |
| Contr_l + CrossE_L + PMI_L + PR_L + Triplet_L | GCN | 12.55 0.26 | ± | 11.51 0.38 | ± | 5.21 0.04 | ± | 18.7 |
| Contr_l + CrossE_L + PMI_L + PR_L + Triplet_L | GIN | 6.26 0.36 | ± | 4.96 0.22 | ± | 1.90 0.20 | ± | 115.0 |
| Contr_l + CrossE_L + PMI_L + PR_L + Triplet_L | MPNN | 9.63 0.64 | ± | 6.22 1.00 | ± | 3.98 1.21 | ± | 73.7 |
| Contr_l + CrossE_L + PMI_L + PR_L + Triplet_L | PAGNN | 3.29 0.28 | ± | 3.10 0.05 | ± | 0.52 0.01 | ± | 158.7 |
| Contr_l + CrossE_L + PMI_L + PR_L + Triplet_L | SAGE | 6.87 0.63 | ± | 6.72 0.34 | ± | 1.07 0.10 | ± | 109.7 |
| Contr_l + CrossE_L + PMI_L + Triplet_L | ALL | 7.80 0.26 | ± | 7.20 0.23 | ± | 4.01 0.13 | ± | 76.7 |
| Contr_l + CrossE_L + PMI_L + Triplet_L | GAT | 13.23 0.91 | ± | 12.32 0.28 | ± | 4.97 0.09 | ± | 16.0 |
| Contr_l + CrossE_L + PMI_L + Triplet_L | GCN | 12.92 0.60 | ± | 11.49 0.34 | ± | 5.20 0.34 | ± | 17.7 |
| Contr_l + CrossE_L + PMI_L + Triplet_L | GIN | 7.05 0.69 | ± | 5.19 0.24 | ± | 2.17 0.24 | ± | 106.0 |
| Contr_l + CrossE_L + PMI_L + Triplet_L | MPNN | 10.18 0.34 | ± | 9.48 1.12 | ± | 4.76 0.28 | ± | 50.3 |
| Contr_l + CrossE_L + PMI_L + Triplet_L | PAGNN | 3.34 0.16 | ± | 2.97 0.02 | ± | 0.50 0.01 | ± | 160.3 |

Cosine-Adj Corr Continued (↑)

| Loss Type | Model | CORA | | Citeseer | | Bitcoin Fraud Transaction | | Average Rank |
|---|---|---|---|---|---|---|---|---|
| Contr_l + CrossE_L + PMI_L + Triplet_L | SAGE | 7.43 0.18 | ± | 6.66 0.43 | ± | 2.95 0.22 | ± | 91.3 |
| Contr_l + CrossE_L + PR_L | ALL | -0.32 0.62 | ± | 1.52 0.76 | ± | -3.26 0.22 | ± | 194.3 |
| Contr_l + CrossE_L + PR_L | GAT | 4.15 3.66 | ± | 2.34 2.82 | ± | -0.04 0.26 | ± | 168.7 |
| Contr_l + CrossE_L + PR_L | GCN | 3.37 0.29 | ± | 2.56 0.13 | ± | 2.03 0.08 | ± | 144.0 |
| Contr_l + CrossE_L + PR_L | GIN | 0.77 0.66 | ± | 2.31 0.37 | ± | -2.93 0.26 | ± | 188.7 |
| Contr_l + CrossE_L + PR_L | MPNN | 1.48 0.27 | ± | 1.23 0.23 | ± | 1.93 1.32 | ± | 161.0 |
| Contr_l + CrossE_L + PR_L | PAGNN | 0.63 0.70 | ± | 1.32 0.38 | ± | 0.81 0.05 | ± | 176.3 |
| Contr_l + CrossE_L + PR_L | SAGE | 4.88 2.85 | ± | 3.64 1.61 | ± | 0.43 0.67 | ± | 150.0 |
| Contr_l + CrossE_L + PR_L + Triplet_L | ALL | 4.63 0.19 | ± | 4.33 0.07 | ± | 2.68 0.17 | ± | 118.7 |
| Contr_l + CrossE_L + PR_L + Triplet_L | GAT | 9.43 1.90 | ± | 7.72 2.23 | ± | 3.31 0.96 | ± | 75.0 |
| Contr_l + CrossE_L + PR_L + Triplet_L | GCN | 6.29 0.81 | ± | 5.58 0.48 | ± | 3.83 1.25 | ± | 94.7 |
| Contr_l + CrossE_L + PR_L + Triplet_L | GIN | 5.72 0.25 | ± | 4.74 0.28 | ± | 2.34 1.35 | ± | 113.7 |
| Contr_l + CrossE_L + PR_L + Triplet_L | MPNN | 3.94 0.65 | ± | 4.18 0.27 | ± | 2.97 0.85 | ± | 120.7 |
| Contr_l + CrossE_L + PR_L + Triplet_L | PAGNN | 2.67 0.32 | ± | 3.40 0.67 | ± | 0.95 0.02 | ± | 155.7 |
| Contr_l + CrossE_L + PR_L + Triplet_L | SAGE | 8.30 0.38 | ± | 6.16 0.71 | ± | 2.21 0.20 | ± | 96.3 |
| Contr_l + CrossE_L + Triplet_L | ALL | 8.61 0.34 | ± | 7.22 0.26 | ± | 4.33 0.15 | ± | 70.3 |

Cosine-Adj Corr Continued (↑)

| Loss Type | Model | CORA | | Citeseer | | Bitcoin Fraud Transaction | | Average Rank |
|---|---|---|---|---|---|---|---|---|
| Contr_l + CrossE_L + Triplet_L | GAT | 13.67 ± 0.83 | | 12.92 ± 0.53 | | 7.01 ± 0.24 | | 3.0 |
| Contr_l + CrossE_L + Triplet_L | GCN | 12.17 ± 0.16 | | 11.39 ± 0.24 | | 5.79 ± 0.20 | | 22.0 |
| Contr_l + CrossE_L + Triplet_L | GIN | 11.73 ± 0.58 | | 10.58 ± 0.81 | | 4.04 ± 0.49 | | 42.7 |
| Contr_l + CrossE_L + Triplet_L | MPNN | 10.47 ± 0.43 | | 10.06 ± 0.36 | | 4.72 ± 0.20 | | 44.7 |
| Contr_l + CrossE_L + Triplet_L | PAGNN | 8.03 ± 1.86 | | 7.61 ± 0.20 | | 0.99 ± 0.02 | | 101.7 |
| Contr_l + CrossE_L + Triplet_L | SAGE | 10.80 ± 0.62 | | 9.78 ± 0.37 | | 4.32 ± 0.21 | | 48.3 |
| Contr_l + PMI_L | ALL | 5.67 ± 0.71 | | 4.92 ± 0.24 | | 2.87 ± 0.17 | | 108.3 |
| Contr_l + PMI_L | GAT | 12.88 ± 0.27 | | 12.26 ± 0.72 | | 4.13 ± 0.30 | | 26.7 |
| Contr_l + PMI_L | GCN | 12.99 ± 0.34 | | 11.32 ± 0.29 | | 5.08 ± 0.29 | | 21.3 |
| Contr_l + PMI_L | GIN | 6.06 ± 0.80 | | 4.38 ± 0.37 | | 1.46 ± 0.10 | | 122.7 |
| Contr_l + PMI_L | MPNN | 10.04 ± 0.41 | | 9.86 ± 0.62 | | 4.43 ± 0.11 | | 51.7 |
| Contr_l + PMI_L | PAGNN | 2.94 ± 0.20 | | 2.89 ± 0.03 | | 0.45 ± 0.03 | | 168.3 |
| Contr_l + PMI_L | SAGE | 5.93 ± 0.41 | | 6.20 ± 0.52 | | 1.11 ± 0.07 | | 115.7 |
| Contr_l + PMI_L + PR_L | ALL | 3.84 ± 0.30 | | 2.90 ± 0.17 | | 0.82 ± 0.03 | | 154.0 |
| Contr_l + PMI_L + PR_L | GAT | 10.27 ± 0.62 | | 7.80 ± 2.53 | | 1.69 ± 0.75 | | 81.7 |
| Contr_l + PMI_L + PR_L | GCN | 12.85 ± 0.15 | | 10.78 ± 0.38 | | 4.97 ± 0.22 | | 26.3 |

Cosine-Adj Corr Continued (↑)

| Loss Type | Model | CORA | | Citeseer | | Bitcoin Fraud Transaction | | Average Rank |
|---|---|---|---|---|---|---|---|---|
| Contr_l + PMI_L + PR_L | GIN | 6.30 | ± | 3.86 | ± | 0.98 | ± | 130.0 |
| | | 0.60 | | 0.46 | | 0.04 | | |
| Contr_l + PMI_L + PR_L | MPNN | 10.06 | ± | 5.52 | ± | 1.73 | ± | 94.7 |
| | | 0.68 | | 2.00 | | 1.38 | | |
| Contr_l + PMI_L + PR_L | PAGNN | 2.90 | ± | 2.93 | ± | 0.32 | ± | 169.7 |
| | | 0.28 | | 0.05 | | 0.04 | | |
| Contr_l + PMI_L + PR_L | SAGE | 5.72 | ± | 5.07 | ± | 0.58 | ± | 133.0 |
| | | 0.28 | | 0.55 | | 0.04 | | |
| Contr_l + PMI_L + PR_L + Triplet_L | ALL | 6.96 | ± | 5.09 | ± | 2.64 | ± | 103.0 |
| | | 0.27 | | 0.07 | | 0.18 | | |
| Contr_l + PMI_L + PR_L + Triplet_L | GAT | 10.15 | ± | 7.77 | ± | 3.29 | ± | 69.0 |
| | | 0.56 | | 0.96 | | 0.64 | | |
| Contr_l + PMI_L + PR_L + Triplet_L | GCN | 11.51 | ± | 9.84 | ± | 5.05 | ± | 37.0 |
| | | 0.81 | | 0.64 | | 0.34 | | |
| Contr_l + PMI_L + PR_L + Triplet_L | GIN | 7.36 | ± | 5.59 | ± | 2.32 | ± | 101.7 |
| | | 0.52 | | 0.31 | | 0.10 | | |
| Contr_l + PMI_L + PR_L + Triplet_L | MPNN | 9.46 | ± | 5.82 | ± | 2.73 | ± | 87.3 |
| | | 0.43 | | 0.15 | | 0.87 | | |
| Contr_l + PMI_L + PR_L + Triplet_L | PAGNN | 3.45 | ± | 3.44 | ± | 0.66 | ± | 152.0 |
| | | 0.15 | | 0.27 | | 0.04 | | |
| Contr_l + PMI_L + PR_L + Triplet_L | SAGE | 7.62 | ± | 7.15 | ± | 1.69 | ± | 100.0 |
| | | 0.49 | | 0.32 | | 0.24 | | |
| Contr_l + PR_L | ALL | -0.85 | ± | 0.93 | ± | -3.26 | ± | 200.0 |
| | | 0.73 | | 1.04 | | 0.30 | | |
| Contr_l + PR_L | GAT | 2.04 | ± | 1.19 | ± | -0.04 | ± | 185.0 |
| | | 3.81 | | 0.15 | | 0.24 | | |
| Contr_l + PR_L | GCN | 3.13 | ± | 3.17 | ± | 3.36 | ± | 129.3 |
| | | 0.31 | | 0.51 | | 1.34 | | |
| Contr_l + PR_L | GIN | 0.96 | ± | 2.25 | ± | -2.92 | ± | 188.0 |
| | | 0.15 | | 0.69 | | 0.31 | | |
| Contr_l + PR_L | MPNN | 0.99 | ± | 1.17 | ± | 1.12 | ± | 169.3 |
| | | 0.45 | | 0.29 | | 0.06 | | |

Cosine-Adj Corr Continued (↑)

| Loss Type | Model | CORA | | Citeseer | | Bitcoin Fraud Transaction | | Average Rank |
|---|---|---|---|---|---|---|---|---|
| Contr_l + PR_L | PAGNN | 0.25 | ± | 1.26 | ± | 0.79 | ± | 177.7 |
| | | 0.42 | | 0.36 | | 0.03 | | |
| Contr_l + PR_L | SAGE | 5.02 | ± | 4.36 | ± | 0.66 | ± | 139.7 |
| | | 2.77 | | 2.14 | | 0.52 | | |
| Contr_l + PR_L + Triplet_L | ALL | 4.03 | ± | 3.65 | ± | 2.66 | ± | 125.7 |
| | | 1.06 | | 0.62 | | 0.07 | | |
| Contr_l + PR_L + Triplet_L | GAT | 9.24 | ± | 7.79 | ± | 3.19 | ± | 76.3 |
| | | 3.24 | | 2.22 | | 1.05 | | |
| Contr_l + PR_L + Triplet_L | GCN | 5.37 | ± | 5.44 | ± | 3.08 | ± | 106.7 |
| | | 0.44 | | 0.45 | | 1.22 | | |
| Contr_l + PR_L + Triplet_L | GIN | 5.81 | ± | 4.20 | ± | 1.69 | ± | 124.0 |
| | | 0.32 | | 0.24 | | 1.15 | | |
| Contr_l + PR_L + Triplet_L | MPNN | 4.10 | ± | 3.47 | ± | 2.39 | ± | 129.0 |
| | | 1.14 | | 0.69 | | 0.58 | | |
| Contr_l + PR_L + Triplet_L | PAGNN | 2.38 | ± | 3.58 | ± | 0.96 | ± | 154.3 |
| | | 0.33 | | 0.88 | | 0.01 | | |
| Contr_l + PR_L + Triplet_L | SAGE | 7.83 | ± | 6.88 | ± | 2.49 | ± | 91.3 |
| | | 0.44 | | 0.53 | | 0.18 | | |
| Contr_l + Triplet_L | ALL | 8.98 | ± | 7.41 | ± | 4.62 | ± | 66.0 |
| | | 0.21 | | 0.16 | | 0.12 | | |
| Contr_l + Triplet_L | GAT | 13.02 | ± | 12.76 | ± | 6.95 | ± | 6.7 |
| | | 0.70 | | 0.61 | | 0.16 | | |
| Contr_l + Triplet_L | GCN | 12.20 | ± | 11.45 | ± | 5.78 | ± | 21.0 |
| | | 0.28 | | 0.15 | | 0.18 | | |
| Contr_l + Triplet_L | GIN | 11.49 | ± | 10.05 | ± | 3.70 | ± | 49.7 |
| | | 0.47 | | 0.72 | | 0.45 | | |
| Contr_l + Triplet_L | MPNN | 10.48 | ± | 10.23 | ± | 4.81 | ± | 41.7 |
| | | 0.47 | | 0.29 | | 0.14 | | |
| Contr_l + Triplet_L | PAGNN | 8.35 | ± | 7.74 | ± | 0.98 | ± | 100.7 |
| | | 2.18 | | 0.13 | | 0.02 | | |
| Contr_l + Triplet_L | SAGE | 10.92 | ± | 9.65 | ± | 4.20 | ± | 48.7 |
| | | 0.27 | | 0.66 | | 0.51 | | |

Cosine-Adj Corr Continued (↑)

| Loss Type | Model | CORA | | Citeseer | | Bitcoin Fraud Transaction | | Average Rank |
|---|---|---|---|---|---|---|---|---|
| CrossE_L | ALL | 2.99 0.76 | ± | 1.63 2.22 | ± | -3.54 0.19 | ± | 184.7 |
| CrossE_L | GAT | 2.56 1.48 | ± | 2.45 1.45 | ± | 1.75 1.14 | ± | 155.0 |
| CrossE_L | GCN | -6.60 0.24 | ± | -3.87 0.05 | ± | -4.11 0.01 | ± | 210.0 |
| CrossE_L | GIN | -5.28 0.36 | ± | -2.88 0.07 | ± | -3.89 0.03 | ± | 209.0 |
| CrossE_L | MPNN | 2.43 0.15 | ± | 2.90 0.15 | ± | -3.75 0.17 | ± | 182.3 |
| CrossE_L | PAGNN | 2.09 0.16 | ± | 1.52 0.16 | ± | -0.69 0.14 | ± | 184.0 |
| CrossE_L | SAGE | 1.64 0.32 | ± | 0.50 0.01 | ± | -0.03 0.03 | ± | 190.0 |
| CrossE_L + PMI_L | ALL | 5.35 0.31 | ± | 3.43 0.18 | ± | 2.69 0.28 | ± | 123.0 |
| CrossE_L + PMI_L | GAT | 13.23 0.59 | ± | 13.03 0.25 | ± | 3.85 0.24 | ± | 25.3 |
| CrossE_L + PMI_L | GCN | 12.91 0.49 | ± | 11.51 0.41 | ± | 4.89 0.24 | ± | 22.0 |
| CrossE_L + PMI_L | GIN | 6.40 0.62 | ± | 3.67 0.14 | ± | 1.08 0.06 | ± | 128.3 |
| CrossE_L + PMI_L | MPNN | 9.72 0.31 | ± | 9.57 0.70 | ± | 4.49 0.34 | ± | 53.7 |
| CrossE_L + PMI_L | PAGNN | 3.07 0.35 | ± | 2.88 0.04 | ± | 0.37 0.06 | ± | 169.3 |
| CrossE_L + PMI_L | SAGE | 4.93 0.24 | ± | 4.43 0.13 | ± | 0.29 0.11 | ± | 146.3 |
| CrossE_L + PMI_L + PR_L | ALL | 3.13 0.08 | ± | 2.81 0.05 | ± | 0.62 0.02 | ± | 163.7 |
| CrossE_L + PMI_L + PR_L | GAT | 11.07 0.94 | ± | 7.61 3.22 | ± | 1.99 0.85 | ± | 75.3 |

Cosine-Adj Corr Continued (↑)

| Loss Type | Model | CORA | | Citeseer | | Bitcoin Fraud Transaction | | Average Rank |
|---|---|---|---|---|---|---|---|---|
| CrossE_L + PMI_L + PR_L | GCN | 12.34 0.49 | ± | 11.38 0.47 | ± | 4.70 0.54 | ± | 30.7 |
| CrossE_L + PMI_L + PR_L | GIN | 5.64 0.27 | ± | 3.69 0.48 | ± | 0.84 0.15 | ± | 137.7 |
| CrossE_L + PMI_L + PR_L | MPNN | 9.55 0.74 | ± | 6.64 2.24 | ± | 2.47 1.65 | ± | 85.7 |
| CrossE_L + PMI_L + PR_L | PAGNN | 3.20 0.31 | ± | 2.98 0.15 | ± | 0.21 0.10 | ± | 166.3 |
| CrossE_L + PMI_L + PR_L | SAGE | 5.11 0.24 | ± | 4.75 0.37 | ± | 0.42 0.05 | ± | 141.7 |
| CrossE_L + PMI_L + PR_L + Triplet_L | ALL | 6.66 0.09 | ± | 4.20 0.14 | ± | 2.28 0.05 | ± | 114.3 |
| CrossE_L + PMI_L + PR_L + Triplet_L | GAT | 10.73 0.59 | ± | 9.28 1.60 | ± | 3.57 1.20 | ± | 58.3 |
| CrossE_L + PMI_L + PR_L + Triplet_L | GCN | 12.44 0.37 | ± | 11.43 0.30 | ± | 5.15 0.22 | ± | 23.3 |
| CrossE_L + PMI_L + PR_L + Triplet_L | GIN | 6.36 0.27 | ± | 4.82 0.32 | ± | 1.93 0.14 | ± | 113.7 |
| CrossE_L + PMI_L + PR_L + Triplet_L | MPNN | 10.00 0.45 | ± | 6.68 1.60 | ± | 4.01 1.28 | ± | 69.7 |
| CrossE_L + PMI_L + PR_L + Triplet_L | PAGNN | 3.49 0.11 | ± | 3.10 0.13 | ± | 0.51 0.02 | ± | 157.3 |
| CrossE_L + PMI_L + PR_L + Triplet_L | SAGE | 6.94 0.49 | ± | 6.28 0.33 | ± | 0.91 0.05 | ± | 115.7 |
| CrossE_L + PMI_L + Triplet_L | ALL | 9.06 0.30 | ± | 8.57 0.34 | ± | 4.46 0.20 | ± | 63.7 |
| CrossE_L + PMI_L + Triplet_L | GAT | 13.31 0.35 | ± | 12.57 0.35 | ± | 5.36 0.26 | ± | 10.0 |
| CrossE_L + PMI_L + Triplet_L | GCN | 12.78 0.60 | ± | 11.80 0.23 | ± | 5.33 0.25 | ± | 16.0 |
| CrossE_L + PMI_L + Triplet_L | GIN | 7.37 0.40 | ± | 5.69 0.26 | ± | 2.66 0.16 | ± | 98.3 |

Cosine-Adj Corr Continued (↑)

| Loss Type | | | | Model | CORA | | Citeseer | | Bitcoin Fraud Transaction | | Average Rank |
|---|---|---|---|---|---|---|---|---|---|---|---|
| CrossE_L | + | PMI_L | + | MPNN | 9.90 | ± | 9.51 | ± | 5.22 | ± | 45.0 |
| Triplet_L | | | | | 0.40 | | 0.56 | | 0.15 | | |
| CrossE_L | + | PMI_L | + | PAGNN | 3.63 | ± | 3.00 | ± | 0.54 | ± | 156.3 |
| Triplet_L | | | | | 0.24 | | 0.08 | | 0.01 | | |
| CrossE_L | + | PMI_L | + | SAGE | 7.99 | ± | 7.30 | ± | 3.44 | ± | 80.3 |
| Triplet_L | | | | | 0.17 | | 0.35 | | 0.24 | | |
| CrossE_L + PR_L | | | | ALL | -2.13 | ± | 0.40 | ± | -3.43 | ± | 204.7 |
| | | | | | 0.64 | | 0.28 | | 0.08 | | |
| CrossE_L + PR_L | | | | GAT | -0.60 | ± | -0.21 | ± | -0.72 | ± | 200.0 |
| | | | | | 2.25 | | 0.25 | | 0.30 | | |
| CrossE_L + PR_L | | | | GCN | 2.04 | ± | 1.43 | ± | -2.90 | ± | 186.7 |
| | | | | | 0.25 | | 0.17 | | 0.77 | | |
| CrossE_L + PR_L | | | | GIN | 0.15 | ± | 0.95 | ± | -3.62 | ± | 198.7 |
| | | | | | 0.38 | | 0.53 | | 0.11 | | |
| CrossE_L + PR_L | | | | MPNN | 1.36 | ± | 1.01 | ± | 0.67 | ± | 178.0 |
| | | | | | 2.38 | | 0.15 | | 0.19 | | |
| CrossE_L + PR_L | | | | PAGNN | -0.02 | ± | 0.41 | ± | 0.13 | ± | 193.7 |
| | | | | | 0.35 | | 0.37 | | 0.12 | | |
| CrossE_L + PR_L | | | | SAGE | -0.84 | ± | -0.06 | ± | -1.94 | ± | 200.3 |
| | | | | | 1.07 | | 0.86 | | 0.34 | | |
| CrossE_L | + | PR_L | + | ALL | 3.34 | ± | 2.50 | ± | 0.60 | ± | 163.3 |
| Triplet_L | | | | | 1.15 | | 0.44 | | 1.83 | | |
| CrossE_L | + | PR_L | + | GAT | 4.08 | ± | 6.99 | ± | 1.14 | ± | 119.0 |
| Triplet_L | | | | | 2.76 | | 3.60 | | 0.94 | | |
| CrossE_L | + | PR_L | + | GCN | 4.44 | ± | 4.28 | ± | 3.29 | ± | 116.0 |
| Triplet_L | | | | | 0.63 | | 0.61 | | 1.16 | | |
| CrossE_L | + | PR_L | + | GIN | 3.87 | ± | 3.52 | ± | -2.12 | ± | 162.3 |
| Triplet_L | | | | | 0.68 | | 0.29 | | 0.25 | | |
| CrossE_L | + | PR_L | + | MPNN | 1.97 | ± | 2.03 | ± | 1.98 | ± | 156.3 |
| Triplet_L | | | | | 0.18 | | 0.17 | | 0.61 | | |
| CrossE_L | + | PR_L | + | PAGNN | 1.68 | ± | 2.47 | ± | 0.92 | ± | 167.7 |
| Triplet_L | | | | | 0.54 | | 0.37 | | 0.03 | | |

Cosine-Adj Corr Continued (↑)

| Loss Type | Model | CORA | | Citeseer | | Bitcoin Fraud Transaction | | Average Rank |
|---|---|---|---|---|---|---|---|---|
| CrossE_L + PR_L + Triplet_L | SAGE | 8.36 ± 0.96 | | 6.91 ± 1.08 | | 1.54 ± 0.67 | | 98.0 |
| CrossE_L + Triplet_L | ALL | 10.47 ± 0.33 | | 9.17 ± 0.29 | | 4.93 ± 0.16 | | 48.0 |
| CrossE_L + Triplet_L | GAT | 14.66 ± 0.66 | | 13.65 ± 0.13 | | 6.35 ± 0.49 | | 2.3 |
| CrossE_L + Triplet_L | GCN | 13.42 ± 0.48 | | 12.20 ± 0.46 | | 6.01 ± 0.06 | | 9.0 |
| CrossE_L + Triplet_L | GIN | 11.93 ± 0.59 | | 9.50 ± 0.12 | | 3.34 ± 0.46 | | 55.3 |
| CrossE_L + Triplet_L | MPNN | 12.34 ± 0.55 | | 11.71 ± 0.74 | | 5.34 ± 0.14 | | 19.0 |
| CrossE_L + Triplet_L | PAGNN | 8.83 ± 2.52 | | 8.71 ± 0.38 | | 0.97 ± 0.01 | | 97.3 |
| CrossE_L + Triplet_L | SAGE | 12.47 ± 0.37 | | 10.49 ± 0.41 | | 3.88 ± 0.10 | | 40.7 |
| PMI_L | ALL | 4.48 ± 0.10 | | 3.52 ± 0.19 | | 2.82 ± 0.10 | | 124.0 |
| PMI_L | GAT | 13.86 ± 0.74 | | 12.92 ± 0.82 | | 3.93 ± 0.27 | | 22.3 |
| PMI_L | GCN | 12.86 ± 0.15 | | 11.45 ± 0.47 | | 4.96 ± 0.38 | | 23.0 |
| PMI_L | GIN | 5.57 ± 0.70 | | 3.78 ± 0.19 | | 1.05 ± 0.19 | | 132.7 |
| PMI_L | MPNN | 10.13 ± 0.62 | | 10.11 ± 0.39 | | 4.49 ± 0.21 | | 48.0 |
| PMI_L | PAGNN | 2.98 ± 0.25 | | 2.88 ± 0.04 | | 0.40 ± 0.04 | | 169.7 |
| PMI_L | SAGE | 4.62 ± 0.42 | | 4.23 ± 0.20 | | 0.40 ± 0.13 | | 148.0 |
| PMI_L + PR_L | ALL | 3.29 ± 0.25 | | 2.78 ± 0.14 | | 0.62 ± 0.01 | | 162.7 |

Cosine-Adj Corr Continued (↑)

| Loss Type | Model | CORA | | Citeseer | | Bitcoin Fraud Transaction | | Average Rank |
|-----------|-------|------|---|----------|---|---------------------------|---|--------------|
| PMI_L + PR_L | GAT | 9.56 | ± | 6.85 | ± | 1.03 | ± | 97.3 |
| | | 0.69 | | 3.31 | | 0.06 | | |
| PMI_L + PR_L | GCN | 11.64 | ± | 11.26 | ± | 4.83 | ± | 33.3 |
| | | 0.58 | | 0.55 | | 0.28 | | |
| PMI_L + PR_L | GIN | 5.57 | ± | 3.88 | ± | 0.80 | ± | 138.0 |
| | | 0.52 | | 0.58 | | 0.05 | | |
| PMI_L + PR_L | MPNN | 9.53 | ± | 4.44 | ± | 1.49 | ± | 106.0 |
| | | 0.44 | | 0.17 | | 1.41 | | |
| PMI_L + PR_L | PAGNN | 3.02 | ± | 2.93 | ± | 0.11 | ± | 170.3 |
| | | 0.17 | | 0.07 | | 0.03 | | |
| PMI_L + PR_L | SAGE | 5.11 | ± | 4.81 | ± | 0.39 | ± | 142.3 |
| | | 0.40 | | 0.23 | | 0.06 | | |
| PMI_L + PR_L + Triplet_L | ALL | 6.73 | ± | 4.51 | ± | 2.38 | ± | 109.3 |
| | | 0.10 | | 0.15 | | 0.18 | | |
| PMI_L + PR_L + Triplet_L | GAT | 10.28 | ± | 8.86 | ± | 2.76 | ± | 69.0 |
| | | 0.31 | | 1.51 | | 0.89 | | |
| PMI_L + PR_L + Triplet_L | GCN | 12.22 | ± | 10.71 | ± | 4.84 | ± | 32.0 |
| | | 0.30 | | 0.87 | | 0.11 | | |
| PMI_L + PR_L + Triplet_L | GIN | 6.45 | ± | 4.81 | ± | 1.92 | ± | 113.7 |
| | | 0.27 | | 0.13 | | 0.20 | | |
| PMI_L + PR_L + Triplet_L | MPNN | 9.80 | ± | 5.35 | ± | 3.21 | ± | 84.3 |
| | | 0.38 | | 0.13 | | 1.26 | | |
| PMI_L + PR_L + Triplet_L | PAGNN | 3.44 | ± | 3.20 | ± | 0.54 | ± | 156.0 |
| | | 0.13 | | 0.11 | | 0.01 | | |
| PMI_L + PR_L + Triplet_L | SAGE | 7.71 | ± | 6.96 | ± | 1.10 | ± | 103.7 |
| | | 0.38 | | 0.26 | | 0.08 | | |
| PMI_L + Triplet_L | ALL | 8.76 | ± | 8.11 | ± | 4.45 | ± | 65.7 |
| | | 0.30 | | 0.43 | | 0.17 | | |
| PMI_L + Triplet_L | GAT | 13.38 | ± | 12.67 | ± | 5.10 | ± | 12.7 |
| | | 0.38 | | 0.40 | | 0.33 | | |
| PMI_L + Triplet_L | GCN | 12.51 | ± | 11.53 | ± | 5.17 | ± | 20.0 |
| | | 0.31 | | 0.26 | | 0.24 | | |

Cosine-Adj Corr Continued (↑)

| Loss Type | Model | CORA | | Citeseer | | Bitcoin Fraud Transaction | | Average Rank |
|---|---|---|---|---|---|---|---|---|
| PMI_L + Triplet_L | GIN | 6.75 ± 0.62 | | 5.28 ± 0.32 | | 2.30 ± 0.18 | | 106.0 |
| PMI_L + Triplet_L | MPNN | 10.09 ± 0.49 | | 9.78 ± 0.89 | | 4.99 ± 0.20 | | 45.3 |
| PMI_L + Triplet_L | PAGNN | 3.30 ± 0.18 | | 2.97 ± 0.04 | | 0.52 ± 0.01 | | 160.7 |
| PMI_L + Triplet_L | SAGE | 7.63 ± 0.51 | | 7.18 ± 0.23 | | 3.12 ± 0.20 | | 86.3 |
| PR_L | ALL | -1.60 ± 0.53 | | -0.15 ± 0.42 | | -3.38 ± 0.12 | | 205.3 |
| PR_L | GAT | -1.39 ± 1.88 | | 0.09 ± 0.42 | | -0.66 ± 0.51 | | 200.0 |
| PR_L | GCN | 1.86 ± 0.41 | | 1.61 ± 0.08 | | 1.29 ± 0.68 | | 163.3 |
| PR_L | GIN | -0.04 ± 0.37 | | 0.91 ± 0.30 | | -3.51 ± 0.19 | | 200.0 |
| PR_L | MPNN | 1.67 ± 1.77 | | 0.98 ± 0.11 | | 1.53 ± 1.05 | | 166.3 |
| PR_L | PAGNN | -0.14 ± 0.57 | | 0.29 ± 0.20 | | 0.23 ± 0.09 | | 194.3 |
| PR_L | SAGE | -1.09 ± 1.08 | | -0.68 ± 0.71 | | -1.62 ± 0.20 | | 202.0 |
| PR_L + Triplet_L | ALL | -0.88 ± 0.99 | | 0.56 ± 0.17 | | -3.36 ± 0.16 | | 201.7 |
| PR_L + Triplet_L | GAT | 4.69 ± 4.20 | | 1.09 ± 0.27 | | 0.04 ± 0.73 | | 171.3 |
| PR_L + Triplet_L | GCN | 2.96 ± 0.50 | | 2.65 ± 0.16 | | 1.92 ± 0.26 | | 150.7 |
| PR_L + Triplet_L | GIN | 0.79 ± 0.18 | | 2.23 ± 0.36 | | -2.96 ± 0.28 | | 189.3 |
| PR_L + Triplet_L | MPNN | 1.31 ± 0.63 | | 0.95 ± 0.36 | | 1.82 ± 0.92 | | 165.7 |

Continued on next page

Cosine-Adj Corr Continued (↑)

| Loss Type | Model | CORA | | Citeseer | | Bitcoin Fraud Transaction | | Average Rank |
|---|---|---|---|---|---|---|---|---|
| PR_L + Triplet_L | PAGNN | 0.16 0.22 | ± | 0.85 0.21 | ± | 0.72 0.04 | ± | 182.3 |
| PR_L + Triplet_L | SAGE | 0.10 0.86 | ± | 2.55 2.66 | ± | -0.27 0.64 | ± | 185.7 |
| Triplet_L | ALL | 10.38 0.42 | ± | 9.28 0.14 | ± | 4.81 0.17 | ± | 49.7 |
| Triplet_L | GAT | 15.89 0.65 | ± | 14.08 0.16 | ± | 6.20 0.26 | ± | 2.0 |
| Triplet_L | GCN | 13.43 0.30 | ± | 12.32 0.39 | ± | 5.71 0.30 | ± | 9.3 |
| Triplet_L | GIN | 12.52 0.92 | ± | 9.93 0.65 | ± | 3.53 0.17 | ± | 45.0 |
| Triplet_L | MPNN | 11.39 0.83 | ± | 11.37 0.50 | ± | 5.19 0.15 | ± | 29.0 |
| Triplet_L | PAGNN | 8.89 2.67 | ± | 8.76 0.31 | ± | 0.97 0.01 | ± | 97.0 |
| Triplet_L | SAGE | 12.34 0.42 | ± | 10.48 0.57 | ± | 4.10 0.08 | ± | 40.0 |

Table 13. Dot-Adj Corr Performance (↑): Top-ranked results are highlighted in **1st**, second-ranked in **2nd**, and third-ranked in **3rd**.

| Loss Type | Model | CORA | | Citeseer | | Bitcoin Fraud Transaction | | Average Rank |
|---|---|---|---|---|---|---|---|---|
| Contr_l | ALL | 7.66 0.21 | ± | 6.05 0.21 | ± | 3.66 0.10 | ± | 87.3 |
| Contr_l | GAT | 12.04 0.45 | ± | 12.47 0.27 | ± | 5.90 0.16 | ± | 17.0 |
| Contr_l | GCN | 10.96 0.13 | ± | 10.91 0.32 | ± | 5.36 0.22 | ± | 29.3 |

Dot-Adj Corr Continued (↑)

| Loss Type | Model | CORA | | | Citeseer | | | Bitcoin Fraud Transaction | | | Average Rank |
|---|---|---|---|---|---|---|---|---|---|---|---|
| Contr_l | GIN | 10.47 | ± | 0.43 | 10.61 | ± | 0.45 | 3.91 | ± | 0.53 | 49.0 |
| Contr_l | MPNN | 9.15 | ± | 0.29 | 8.86 | ± | 0.59 | 3.86 | ± | 0.06 | 68.3 |
| Contr_l | PAGNN | 8.42 | ± | 0.62 | 6.55 | ± | 0.29 | 1.00 | ± | 0.02 | 105.3 |
| Contr_l | SAGE | 9.43 | ± | 0.28 | 8.94 | ± | 0.74 | 3.42 | ± | 0.17 | 69.7 |
| Contr_l + CrossE_L | ALL | 7.44 | ± | 0.33 | 6.14 | ± | 0.06 | 3.48 | ± | 0.19 | 89.3 |
| Contr_l + CrossE_L | GAT | 11.93 | ± | 0.71 | 12.89 | ± | 0.14 | 6.18 | ± | 0.51 | 15.0 |
| Contr_l + CrossE_L | GCN | 11.10 | ± | 0.17 | 10.59 | ± | 0.26 | 5.26 | ± | 0.21 | 31.0 |
| Contr_l + CrossE_L | GIN | 9.80 | ± | 0.39 | 9.33 | ± | 0.51 | 3.61 | ± | 0.46 | 63.3 |
| Contr_l + CrossE_L | MPNN | 9.39 | ± | 0.64 | 9.09 | ± | 0.35 | 4.14 | ± | 0.10 | 63.0 |
| Contr_l + CrossE_L | PAGNN | 8.28 | ± | 0.46 | 6.36 | ± | 0.20 | 1.01 | ± | 0.01 | 106.7 |
| Contr_l + CrossE_L | SAGE | 9.54 | ± | 0.48 | 9.53 | ± | 0.66 | 3.41 | ± | 0.31 | 66.0 |
| Contr_l + CrossE_L + PMI_L | ALL | 5.55 | ± | 0.55 | 3.61 | ± | 0.60 | 2.60 | ± | 0.08 | 121.7 |
| Contr_l + CrossE_L + PMI_L | GAT | 12.45 | ± | 0.41 | 12.71 | ± | 0.48 | 4.08 | ± | 0.30 | 28.0 |
| Contr_l + CrossE_L + PMI_L | GCN | 12.40 | ± | 0.59 | 11.59 | ± | 0.28 | 4.97 | ± | 0.08 | 23.0 |
| Contr_l + CrossE_L + PMI_L | GIN | 6.44 | ± | 0.75 | 4.11 | ± | 0.22 | 1.29 | ± | 0.22 | 123.7 |
| Contr_l + CrossE_L + PMI_L | MPNN | 10.55 | ± | 0.76 | 9.91 | ± | 0.57 | 4.41 | ± | 0.25 | 47.3 |

Dot-Adj Corr Continued (↑)

| Loss Type | Model | CORA | | Citeseer | | Bitcoin Fraud Transaction | | Average Rank |
|---|---|---|---|---|---|---|---|---|
| Contr_l + CrossE_L + PMI_L | PAGNN | 2.75 0.20 | ± | 2.90 0.07 | ± | 0.43 0.02 | ± | 168.7 |
| Contr_l + CrossE_L + PMI_L | SAGE | 5.71 0.16 | ± | 5.56 0.61 | ± | 1.27 0.33 | ± | 119.0 |
| Contr_l + CrossE_L + ALL PMI_L + PR_L | ALL | 3.20 0.41 | ± | 2.95 0.24 | ± | 0.75 0.02 | ± | 158.0 |
| Contr_l + CrossE_L + PMI_L + PR_L | GAT | 10.99 1.31 | ± | 11.08 0.46 | ± | 2.08 0.74 | ± | 60.3 |
| Contr_l + CrossE_L + PMI_L + PR_L | GCN | 12.20 0.63 | ± | 11.21 0.34 | ± | 4.79 0.21 | ± | 32.3 |
| Contr_l + CrossE_L + PMI_L + PR_L | GIN | 6.05 0.67 | ± | 3.74 0.29 | ± | 0.92 0.05 | ± | 134.0 |
| Contr_l + CrossE_L + PMI_L + PR_L | MPNN | 9.40 0.48 | ± | 6.15 1.55 | ± | 1.70 1.44 | ± | 97.3 |
| Contr_l + CrossE_L + PMI_L + PR_L | PAGNN | 3.16 0.24 | ± | 3.00 0.08 | ± | 0.27 0.05 | ± | 165.7 |
| Contr_l + CrossE_L + PMI_L + PR_L | SAGE | 5.52 0.48 | ± | 5.10 0.54 | ± | 0.58 0.05 | ± | 134.3 |
| Contr_l + CrossE_L + PMI_L + PR_L + Triplet_L | ALL | 6.91 0.11 | ± | 4.43 0.06 | ± | 2.19 0.14 | ± | 111.0 |
| Contr_l + CrossE_L + PMI_L + PR_L + Triplet_L | GAT | 10.90 0.94 | ± | 10.21 2.30 | ± | 4.02 0.97 | ± | 47.7 |
| Contr_l + CrossE_L + PMI_L + PR_L + Triplet_L | GCN | 12.55 0.26 | ± | 11.51 0.38 | ± | 5.21 0.04 | ± | 18.7 |
| Contr_l + CrossE_L + PMI_L + PR_L + Triplet_L | GIN | 6.26 0.36 | ± | 4.96 0.22 | ± | 1.90 0.20 | ± | 115.3 |
| Contr_l + CrossE_L + PMI_L + PR_L + Triplet_L | MPNN | 9.63 0.64 | ± | 6.22 1.00 | ± | 3.98 1.21 | ± | 73.7 |
| Contr_l + CrossE_L + PMI_L + PR_L + Triplet_L | PAGNN | 3.29 0.28 | ± | 3.10 0.05 | ± | 0.52 0.01 | ± | 159.0 |
| Contr_l + CrossE_L + PMI_L + PR_L + Triplet_L | SAGE | 6.87 0.63 | ± | 6.72 0.34 | ± | 1.07 0.10 | ± | 109.7 |

Dot-Adj Corr Continued (↑)

| Loss Type | Model | CORA | | | Citeseer | | | Bitcoin Fraud Transaction | | | Average Rank |
|---|---|---|---|---|---|---|---|---|---|---|---|
| Contr_l + CrossE_L + PMI_L + Triplet_L | ALL | 7.80 | ± | 0.26 | 7.20 | ± | 0.23 | 4.01 | ± | 0.13 | 76.7 |
| Contr_l + CrossE_L + PMI_L + Triplet_L | GAT | 13.23 | ± | 0.91 | 12.32 | ± | 0.28 | 4.97 | ± | 0.09 | 16.0 |
| Contr_l + CrossE_L + PMI_L + Triplet_L | GCN | 12.92 | ± | 0.60 | 11.49 | ± | 0.34 | 5.20 | ± | 0.34 | 17.7 |
| Contr_l + CrossE_L + PMI_L + Triplet_L | GIN | 7.05 | ± | 0.69 | 5.19 | ± | 0.24 | 2.17 | ± | 0.24 | 106.0 |
| Contr_l + CrossE_L + PMI_L + Triplet_L | MPNN | 10.18 | ± | 0.34 | 9.48 | ± | 1.12 | 4.76 | ± | 0.28 | 50.3 |
| Contr_l + CrossE_L + PMI_L + Triplet_L | PAGNN | 3.34 | ± | 0.16 | 2.97 | ± | 0.02 | 0.50 | ± | 0.01 | 160.7 |
| Contr_l + CrossE_L + PMI_L + Triplet_L | SAGE | 7.43 | ± | 0.18 | 6.66 | ± | 0.43 | 2.95 | ± | 0.22 | 91.3 |
| Contr_l + CrossE_L + PR_L | ALL | -0.32 | ± | 0.62 | 1.52 | ± | 0.76 | -3.26 | ± | 0.22 | 194.3 |
| Contr_l + CrossE_L + PR_L | GAT | 4.15 | ± | 3.66 | 2.34 | ± | 2.82 | -0.04 | ± | 0.26 | 168.7 |
| Contr_l + CrossE_L + PR_L | GCN | 3.37 | ± | 0.29 | 2.56 | ± | 0.13 | 2.03 | ± | 0.08 | 144.3 |
| Contr_l + CrossE_L + PR_L | GIN | 0.77 | ± | 0.66 | 2.31 | ± | 0.37 | -2.93 | ± | 0.26 | 188.7 |
| Contr_l + CrossE_L + PR_L | MPNN | 1.48 | ± | 0.27 | 1.23 | ± | 0.23 | 1.93 | ± | 1.32 | 161.0 |
| Contr_l + CrossE_L + PR_L | PAGNN | 0.63 | ± | 0.70 | 1.32 | ± | 0.38 | 0.81 | ± | 0.05 | 176.3 |
| Contr_l + CrossE_L + PR_L | SAGE | 4.88 | ± | 2.85 | 3.64 | ± | 1.61 | 0.43 | ± | 0.67 | 150.0 |
| Contr_l + CrossE_L + PR_L + Triplet_L | ALL | 4.63 | ± | 0.19 | 4.33 | ± | 0.07 | 2.68 | ± | 0.17 | 118.7 |
| Contr_l + CrossE_L + PR_L + Triplet_L | GAT | 9.43 | ± | 1.90 | 7.72 | ± | 2.23 | 3.31 | ± | 0.96 | 75.0 |

Dot-Adj Corr Continued (↑)

| Loss Type | Model | CORA | | Citeseer | | Bitcoin Fraud Transaction | | Average Rank |
|---|---|---|---|---|---|---|---|---|
| Contr_l + CrossE_L + PR_L + Triplet_L | GCN | 6.29 | ± | 5.58 | ± | 3.83 | ± | 94.7 |
| | | 0.81 | | 0.48 | | 1.25 | | |
| Contr_l + CrossE_L + PR_L + Triplet_L | GIN | 5.72 | ± | 4.74 | ± | 2.34 | ± | 113.7 |
| | | 0.25 | | 0.28 | | 1.35 | | |
| Contr_l + CrossE_L + PR_L + Triplet_L | MPNN | 3.94 | ± | 4.18 | ± | 2.97 | ± | 120.7 |
| | | 0.65 | | 0.27 | | 0.85 | | |
| Contr_l + CrossE_L + PR_L + Triplet_L | PAGNN | 2.67 | ± | 3.40 | ± | 0.95 | ± | 155.7 |
| | | 0.32 | | 0.67 | | 0.02 | | |
| Contr_l + CrossE_L + PR_L + Triplet_L | SAGE | 8.30 | ± | 6.16 | ± | 2.21 | ± | 96.3 |
| | | 0.38 | | 0.71 | | 0.20 | | |
| Contr_l + CrossE_L + Triplet_L | ALL | 8.61 | ± | 7.22 | ± | 4.33 | ± | 70.3 |
| | | 0.34 | | 0.26 | | 0.15 | | |
| Contr_l + CrossE_L + Triplet_L | GAT | 13.67 | ± | 12.92 | ± | 7.01 | ± | 3.0 |
| | | 0.83 | | 0.53 | | 0.24 | | |
| Contr_l + CrossE_L + Triplet_L | GCN | 12.17 | ± | 11.39 | ± | 5.79 | ± | 22.0 |
| | | 0.16 | | 0.24 | | 0.20 | | |
| Contr_l + CrossE_L + Triplet_L | GIN | 11.73 | ± | 10.58 | ± | 4.04 | ± | 42.7 |
| | | 0.58 | | 0.81 | | 0.49 | | |
| Contr_l + CrossE_L + Triplet_L | MPNN | 10.47 | ± | 10.06 | ± | 4.72 | ± | 44.7 |
| | | 0.43 | | 0.36 | | 0.20 | | |
| Contr_l + CrossE_L + Triplet_L | PAGNN | 8.03 | ± | 7.61 | ± | 0.99 | ± | 101.7 |
| | | 1.86 | | 0.20 | | 0.02 | | |
| Contr_l + CrossE_L + Triplet_L | SAGE | 10.80 | ± | 9.78 | ± | 4.32 | ± | 48.3 |
| | | 0.62 | | 0.37 | | 0.21 | | |
| Contr_l + PMI_L | ALL | 5.67 | ± | 4.92 | ± | 2.87 | ± | 108.3 |
| | | 0.71 | | 0.24 | | 0.17 | | |
| Contr_l + PMI_L | GAT | 12.88 | ± | 12.26 | ± | 4.13 | ± | 26.7 |
| | | 0.27 | | 0.72 | | 0.30 | | |
| Contr_l + PMI_L | GCN | 12.99 | ± | 11.32 | ± | 5.08 | ± | 21.3 |
| | | 0.34 | | 0.29 | | 0.29 | | |
| Contr_l + PMI_L | GIN | 6.06 | ± | 4.38 | ± | 1.46 | ± | 122.7 |
| | | 0.80 | | 0.37 | | 0.10 | | |

Dot-Adj Corr Continued (↑)

| Loss Type | Model | CORA | | Citeseer | | Bitcoin Fraud Transaction | | Average Rank |
|---|---|---|---|---|---|---|---|---|
| Contr_l + PMI_L | MPNN | 10.04 0.41 | ± | 9.86 0.62 | ± | 4.43 0.11 | ± | 51.7 |
| Contr_l + PMI_L | PAGNN | 2.94 0.20 | ± | 2.89 0.03 | ± | 0.45 0.03 | ± | 168.7 |
| Contr_l + PMI_L | SAGE | 5.93 0.41 | ± | 6.20 0.52 | ± | 1.11 0.07 | ± | 115.7 |
| Contr_l + PMI_L + PR_L | ALL | 3.84 0.30 | ± | 2.90 0.17 | ± | 0.82 0.03 | ± | 154.3 |
| Contr_l + PMI_L + PR_L | GAT | 10.27 0.62 | ± | 7.80 2.53 | ± | 1.69 0.75 | ± | 81.7 |
| Contr_l + PMI_L + PR_L | GCN | 12.85 0.15 | ± | 10.78 0.38 | ± | 4.97 0.22 | ± | 26.3 |
| Contr_l + PMI_L + PR_L | GIN | 6.30 0.60 | ± | 3.86 0.46 | ± | 0.98 0.04 | ± | 130.0 |
| Contr_l + PMI_L + PR_L | MPNN | 10.06 0.68 | ± | 5.52 2.00 | ± | 1.73 1.38 | ± | 94.7 |
| Contr_l + PMI_L + PR_L | PAGNN | 2.90 0.28 | ± | 2.93 0.05 | ± | 0.32 0.04 | ± | 170.0 |
| Contr_l + PMI_L + PR_L | SAGE | 5.72 0.28 | ± | 5.07 0.55 | ± | 0.58 0.04 | ± | 133.0 |
| Contr_l + PMI_L + PR_L + Triplet_L | ALL | 6.96 0.27 | ± | 5.09 0.07 | ± | 2.64 0.18 | ± | 103.0 |
| Contr_l + PMI_L + PR_L + Triplet_L | GAT | 10.15 0.56 | ± | 7.77 0.96 | ± | 3.29 0.64 | ± | 69.0 |
| Contr_l + PMI_L + PR_L + Triplet_L | GCN | 11.51 0.81 | ± | 9.84 0.64 | ± | 5.05 0.34 | ± | 37.0 |
| Contr_l + PMI_L + PR_L + Triplet_L | GIN | 7.36 0.52 | ± | 5.59 0.31 | ± | 2.32 0.10 | ± | 101.7 |
| Contr_l + PMI_L + PR_L + Triplet_L | MPNN | 9.46 0.43 | ± | 5.82 0.15 | ± | 2.73 0.87 | ± | 87.3 |
| Contr_l + PMI_L + PR_L + Triplet_L | PAGNN | 3.45 0.15 | ± | 3.44 0.27 | ± | 0.66 0.04 | ± | 152.0 |

Dot-Adj Corr Continued (↑)

| Loss Type | Model | CORA | | Citeseer | | Bitcoin Fraud Transaction | | Average Rank |
|-----------|-------|------|--|----------|--|---------------------------|--|--------------|
| Contr_l + PMI_L + PR_L + Triplet_L | SAGE | 7.62 0.49 | ± | 7.15 0.32 | ± | 1.69 0.24 | ± | 100.0 |
| Contr_l + PR_L | ALL | -0.85 0.73 | ± | 0.93 1.04 | ± | -3.26 0.29 | ± | 200.0 |
| Contr_l + PR_L | GAT | 2.04 3.81 | ± | 1.19 0.15 | ± | -0.04 0.24 | ± | 185.0 |
| Contr_l + PR_L | GCN | 3.13 0.31 | ± | 3.17 0.51 | ± | 3.36 1.34 | ± | 129.7 |
| Contr_l + PR_L | GIN | 0.96 0.15 | ± | 2.25 0.69 | ± | -2.92 0.31 | ± | 188.0 |
| Contr_l + PR_L | MPNN | 0.99 0.45 | ± | 1.17 0.29 | ± | 1.12 0.06 | ± | 169.3 |
| Contr_l + PR_L | PAGNN | 0.25 0.42 | ± | 1.26 0.36 | ± | 0.79 0.03 | ± | 177.7 |
| Contr_l + PR_L | SAGE | 5.02 2.77 | ± | 4.36 2.14 | ± | 0.66 0.52 | ± | 139.7 |
| Contr_l + PR_L + Triplet_L | ALL | 4.03 1.06 | ± | 3.65 0.62 | ± | 2.66 0.07 | ± | 125.7 |
| Contr_l + PR_L + Triplet_L | GAT | 9.24 3.24 | ± | 7.79 2.22 | ± | 3.19 1.05 | ± | 76.3 |
| Contr_l + PR_L + Triplet_L | GCN | 5.37 0.44 | ± | 5.44 0.45 | ± | 3.08 1.22 | ± | 106.7 |
| Contr_l + PR_L + Triplet_L | GIN | 5.81 0.32 | ± | 4.20 0.24 | ± | 1.69 1.15 | ± | 124.0 |
| Contr_l + PR_L + Triplet_L | MPNN | 4.10 1.14 | ± | 3.47 0.69 | ± | 2.39 0.58 | ± | 129.0 |
| Contr_l + PR_L + Triplet_L | PAGNN | 2.38 0.33 | ± | 3.58 0.88 | ± | 0.96 0.01 | ± | 154.0 |
| Contr_l + PR_L + Triplet_L | SAGE | 7.83 0.44 | ± | 6.88 0.53 | ± | 2.49 0.18 | ± | 91.3 |
| Contr_l + Triplet_L | ALL | 8.98 0.21 | ± | 7.41 0.16 | ± | 4.62 0.12 | ± | 66.0 |

Dot-Adj Corr Continued (↑)

| Loss Type | Model | CORA | | Citeseer | | Bitcoin Fraud Transaction | | Average Rank |
|---|---|---|---|---|---|---|---|---|
| Contr_l + Triplet_L | GAT | 13.02 ± 0.70 | | 12.76 ± 0.61 | | 6.95 ± 0.16 | | 6.7 |
| Contr_l + Triplet_L | GCN | 12.20 ± 0.28 | | 11.45 ± 0.15 | | 5.78 ± 0.18 | | 21.0 |
| Contr_l + Triplet_L | GIN | 11.49 ± 0.47 | | 10.05 ± 0.72 | | 3.70 ± 0.45 | | 49.7 |
| Contr_l + Triplet_L | MPNN | 10.48 ± 0.47 | | 10.23 ± 0.29 | | 4.81 ± 0.14 | | 41.7 |
| Contr_l + Triplet_L | PAGNN | 8.35 ± 2.18 | | 7.74 ± 0.13 | | 0.98 ± 0.02 | | 100.7 |
| Contr_l + Triplet_L | SAGE | 10.92 ± 0.27 | | 9.65 ± 0.66 | | 4.20 ± 0.51 | | 48.7 |
| CrossE_L | ALL | 2.99 ± 0.76 | | 1.63 ± 2.22 | | -3.54 ± 0.19 | | 184.7 |
| CrossE_L | GAT | 2.12 ± 1.83 | | 3.18 ± 1.79 | | 1.96 ± 1.11 | | 145.7 |
| CrossE_L | GCN | -6.60 ± 0.24 | | -3.87 ± 0.05 | | -4.11 ± 0.01 | | 210.0 |
| CrossE_L | GIN | -5.26 ± 0.37 | | -2.88 ± 0.07 | | -3.90 ± 0.03 | | 209.0 |
| CrossE_L | MPNN | 2.43 ± 0.15 | | 2.90 ± 0.15 | | -3.74 ± 0.17 | | 182.3 |
| CrossE_L | PAGNN | 2.09 ± 0.16 | | 1.52 ± 0.16 | | -0.69 ± 0.14 | | 184.0 |
| CrossE_L | SAGE | 1.31 ± 0.68 | | 0.50 ± 0.01 | | -0.03 ± 0.04 | | 190.7 |
| CrossE_L + PMI_L | ALL | 5.35 ± 0.31 | | 3.43 ± 0.18 | | 2.69 ± 0.28 | | 123.0 |
| CrossE_L + PMI_L | GAT | 13.23 ± 0.59 | | 13.03 ± 0.25 | | 3.85 ± 0.24 | | 25.3 |
| CrossE_L + PMI_L | GCN | 12.91 ± 0.49 | | 11.51 ± 0.41 | | 4.89 ± 0.24 | | 22.0 |

Dot-Adj Corr Continued (↑)

| Loss Type | Model | CORA | | Citeseer | | Bitcoin Fraud Transaction | | Average Rank |
|---|---|---|---|---|---|---|---|---|
| CrossE_L + PMI_L | GIN | 6.40 | ± | 3.67 | ± | 1.08 | ± | 128.3 |
| | | 0.62 | | 0.14 | | 0.06 | | |
| CrossE_L + PMI_L | MPNN | 9.72 | ± | 9.57 | ± | 4.49 | ± | 53.7 |
| | | 0.31 | | 0.70 | | 0.34 | | |
| CrossE_L + PMI_L | PAGNN | 3.07 | ± | 2.88 | ± | 0.37 | ± | 169.7 |
| | | 0.35 | | 0.04 | | 0.06 | | |
| CrossE_L + PMI_L | SAGE | 4.93 | ± | 4.43 | ± | 0.29 | ± | 146.3 |
| | | 0.24 | | 0.13 | | 0.11 | | |
| CrossE_L + PMI_L + PR_L | ALL | 3.13 | ± | 2.81 | ± | 0.62 | ± | 164.0 |
| | | 0.08 | | 0.05 | | 0.02 | | |
| CrossE_L + PMI_L + PR_L | GAT | 11.07 | ± | 7.61 | ± | 1.99 | ± | 75.3 |
| | | 0.94 | | 3.22 | | 0.85 | | |
| CrossE_L + PMI_L + PR_L | GCN | 12.34 | ± | 11.38 | ± | 4.70 | ± | 30.7 |
| | | 0.49 | | 0.47 | | 0.54 | | |
| CrossE_L + PMI_L + PR_L | GIN | 5.64 | ± | 3.69 | ± | 0.84 | ± | 137.7 |
| | | 0.27 | | 0.48 | | 0.15 | | |
| CrossE_L + PMI_L + PR_L | MPNN | 9.55 | ± | 6.64 | ± | 2.47 | ± | 85.7 |
| | | 0.74 | | 2.24 | | 1.65 | | |
| CrossE_L + PMI_L + PR_L | PAGNN | 3.20 | ± | 2.98 | ± | 0.21 | ± | 166.7 |
| | | 0.31 | | 0.15 | | 0.10 | | |
| CrossE_L + PMI_L + PR_L | SAGE | 5.11 | ± | 4.75 | ± | 0.42 | ± | 141.7 |
| | | 0.24 | | 0.37 | | 0.05 | | |
| CrossE_L + PMI_L + PR_L + Triplet_L | ALL | 6.66 | ± | 4.20 | ± | 2.28 | ± | 114.3 |
| | | 0.09 | | 0.14 | | 0.05 | | |
| CrossE_L + PMI_L + PR_L + Triplet_L | GAT | 10.73 | ± | 9.28 | ± | 3.57 | ± | 58.3 |
| | | 0.59 | | 1.60 | | 1.20 | | |
| CrossE_L + PMI_L + PR_L + Triplet_L | GCN | 12.44 | ± | 11.43 | ± | 5.15 | ± | 23.3 |
| | | 0.37 | | 0.30 | | 0.22 | | |
| CrossE_L + PMI_L + PR_L + Triplet_L | GIN | 6.36 | ± | 4.82 | ± | 1.93 | ± | 114.0 |
| | | 0.27 | | 0.32 | | 0.14 | | |
| CrossE_L + PMI_L + PR_L + Triplet_L | MPNN | 10.00 | ± | 6.68 | ± | 4.01 | ± | 69.7 |
| | | 0.45 | | 1.60 | | 1.28 | | |

Dot-Adj Corr Continued (↑)

| Loss Type | Model | CORA | | Citeseer | | Bitcoin Fraud Transaction | | Average Rank |
|---|---|---|---|---|---|---|---|---|
| CrossE_L + PMI_L + PR_L + Triplet_L | PAGNN | 3.49 0.11 | ± | 3.10 0.13 | ± | 0.51 0.02 | ± | 157.7 |
| CrossE_L + PMI_L + PR_L + Triplet_L | SAGE | 6.94 0.49 | ± | 6.28 0.33 | ± | 0.91 0.05 | ± | 115.7 |
| CrossE_L + PMI_L + Triplet_L | ALL | 9.06 0.30 | ± | 8.57 0.34 | ± | 4.46 0.20 | ± | 63.7 |
| CrossE_L + PMI_L + Triplet_L | GAT | 13.31 0.35 | ± | 12.57 0.35 | ± | 5.36 0.26 | ± | 10.0 |
| CrossE_L + PMI_L + Triplet_L | GCN | 12.78 0.60 | ± | 11.80 0.23 | ± | 5.33 0.25 | ± | 16.0 |
| CrossE_L + PMI_L + Triplet_L | GIN | 7.37 0.40 | ± | 5.69 0.26 | ± | 2.66 0.16 | ± | 98.3 |
| CrossE_L + PMI_L + Triplet_L | MPNN | 9.90 0.40 | ± | 9.51 0.56 | ± | 5.22 0.15 | ± | 45.0 |
| CrossE_L + PMI_L + Triplet_L | PAGNN | 3.63 0.24 | ± | 3.00 0.08 | ± | 0.54 0.01 | ± | 156.7 |
| CrossE_L + PMI_L + Triplet_L | SAGE | 7.99 0.17 | ± | 7.30 0.35 | ± | 3.44 0.24 | ± | 80.3 |
| CrossE_L + PR_L | ALL | -2.13 0.64 | ± | 0.40 0.28 | ± | -3.43 0.07 | ± | 204.7 |
| CrossE_L + PR_L | GAT | -0.60 2.25 | ± | -0.21 0.25 | ± | -0.72 0.30 | ± | 200.0 |
| CrossE_L + PR_L | GCN | 2.04 0.25 | ± | 1.43 0.17 | ± | -2.90 0.77 | ± | 186.7 |
| CrossE_L + PR_L | GIN | 0.15 0.38 | ± | 0.95 0.53 | ± | -3.62 0.11 | ± | 198.7 |
| CrossE_L + PR_L | MPNN | 1.36 2.38 | ± | 1.01 0.15 | ± | 0.67 0.19 | ± | 177.7 |
| CrossE_L + PR_L | PAGNN | -0.02 0.35 | ± | 0.41 0.37 | ± | 0.13 0.12 | ± | 193.7 |
| CrossE_L + PR_L | SAGE | -0.84 1.07 | ± | -0.06 0.86 | ± | -1.94 0.34 | ± | 200.3 |

Dot-Adj Corr Continued (↑)

| Loss Type | | | | Model | CORA | | Citeseer | | Bitcoin Fraud Transaction | | Average Rank |
|---|---|---|---|---|---|---|---|---|---|---|---|
| CrossE_L | + | PR_L | + | ALL | 3.34 | ± | 2.50 | ± | 0.60 | ± | 163.7 |
| Triplet_L | | | | | 1.15 | | 0.44 | | 1.83 | | |
| CrossE_L | + | PR_L | + | GAT | 4.08 | ± | 6.99 | ± | 1.14 | ± | 119.0 |
| Triplet_L | | | | | 2.76 | | 3.60 | | 0.94 | | |
| CrossE_L | + | PR_L | + | GCN | 4.44 | ± | 4.28 | ± | 3.29 | ± | 116.0 |
| Triplet_L | | | | | 0.63 | | 0.61 | | 1.16 | | |
| CrossE_L | + | PR_L | + | GIN | 3.87 | ± | 3.52 | ± | -2.12 | ± | 162.3 |
| Triplet_L | | | | | 0.68 | | 0.29 | | 0.25 | | |
| CrossE_L | + | PR_L | + | MPNN | 1.97 | ± | 2.03 | ± | 1.98 | ± | 156.3 |
| Triplet_L | | | | | 0.18 | | 0.16 | | 0.61 | | |
| CrossE_L | + | PR_L | + | PAGNN | 1.68 | ± | 2.47 | ± | 0.92 | ± | 168.0 |
| Triplet_L | | | | | 0.54 | | 0.37 | | 0.03 | | |
| CrossE_L | + | PR_L | + | SAGE | 8.36 | ± | 6.91 | ± | 1.54 | ± | 98.0 |
| Triplet_L | | | | | 0.96 | | 1.08 | | 0.67 | | |
| CrossE_L + Triplet_L | | | | ALL | 10.47 | ± | 9.17 | ± | 4.93 | ± | 48.0 |
| | | | | | 0.33 | | 0.29 | | 0.16 | | |
| CrossE_L + Triplet_L | | | | GAT | 14.66 | ± | 13.65 | ± | 6.35 | ± | 2.3 |
| | | | | | 0.66 | | 0.13 | | 0.49 | | |
| CrossE_L + Triplet_L | | | | GCN | 13.42 | ± | 12.20 | ± | 6.01 | ± | 9.0 |
| | | | | | 0.48 | | 0.46 | | 0.06 | | |
| CrossE_L + Triplet_L | | | | GIN | 11.93 | ± | 9.50 | ± | 3.34 | ± | 55.3 |
| | | | | | 0.59 | | 0.12 | | 0.46 | | |
| CrossE_L + Triplet_L | | | | MPNN | 12.34 | ± | 11.71 | ± | 5.34 | ± | 19.0 |
| | | | | | 0.55 | | 0.74 | | 0.14 | | |
| CrossE_L + Triplet_L | | | | PAGNN | 8.83 | ± | 8.71 | ± | 0.97 | ± | 97.3 |
| | | | | | 2.52 | | 0.38 | | 0.01 | | |
| CrossE_L + Triplet_L | | | | SAGE | 12.47 | ± | 10.49 | ± | 3.88 | ± | 40.7 |
| | | | | | 0.37 | | 0.41 | | 0.10 | | |
| PMI_L | | | | ALL | 4.48 | ± | 3.52 | ± | 2.82 | ± | 124.0 |
| | | | | | 0.10 | | 0.19 | | 0.10 | | |
| PMI_L | | | | GAT | 13.86 | ± | 12.92 | ± | 3.93 | ± | 22.3 |
| | | | | | 0.74 | | 0.82 | | 0.27 | | |

Dot-Adj Corr Continued (↑)

| Loss Type | Model | CORA | | Citeseer | | Bitcoin Fraud Transaction | | Average Rank |
|---|---|---|---|---|---|---|---|---|
| PMI_L | GCN | 12.86 | ± | 11.45 | ± | 4.96 | ± | 23.0 |
| | | 0.15 | | 0.47 | | 0.38 | | |
| PMI_L | GIN | 5.57 | ± | 3.78 | ± | 1.05 | ± | 132.7 |
| | | 0.70 | | 0.19 | | 0.19 | | |
| PMI_L | MPNN | 10.13 | ± | 10.11 | ± | 4.49 | ± | 48.0 |
| | | 0.62 | | 0.39 | | 0.21 | | |
| PMI_L | PAGNN | 2.98 | ± | 2.88 | ± | 0.40 | ± | 170.0 |
| | | 0.25 | | 0.04 | | 0.04 | | |
| PMI_L | SAGE | 4.62 | ± | 4.23 | ± | 0.40 | ± | 148.0 |
| | | 0.42 | | 0.20 | | 0.13 | | |
| PMI_L + PR_L | ALL | 3.29 | ± | 2.78 | ± | 0.62 | ± | 163.0 |
| | | 0.25 | | 0.14 | | 0.01 | | |
| PMI_L + PR_L | GAT | 9.56 | ± | 6.85 | ± | 1.03 | ± | 97.3 |
| | | 0.69 | | 3.31 | | 0.06 | | |
| PMI_L + PR_L | GCN | 11.64 | ± | 11.26 | ± | 4.83 | ± | 33.3 |
| | | 0.58 | | 0.55 | | 0.28 | | |
| PMI_L + PR_L | GIN | 5.57 | ± | 3.88 | ± | 0.80 | ± | 138.0 |
| | | 0.52 | | 0.58 | | 0.05 | | |
| PMI_L + PR_L | MPNN | 9.53 | ± | 4.44 | ± | 1.49 | ± | 106.0 |
| | | 0.44 | | 0.17 | | 1.41 | | |
| PMI_L + PR_L | PAGNN | 3.02 | ± | 2.93 | ± | 0.11 | ± | 170.7 |
| | | 0.17 | | 0.07 | | 0.03 | | |
| PMI_L + PR_L | SAGE | 5.11 | ± | 4.81 | ± | 0.39 | ± | 142.3 |
| | | 0.40 | | 0.23 | | 0.06 | | |
| PMI_L + PR_L + Triplet_L | ALL | 6.73 | ± | 4.51 | ± | 2.38 | ± | 109.3 |
| | | 0.10 | | 0.15 | | 0.18 | | |
| PMI_L + PR_L + Triplet_L | GAT | 10.28 | ± | 8.86 | ± | 2.76 | ± | 69.0 |
| | | 0.31 | | 1.51 | | 0.89 | | |
| PMI_L + PR_L + Triplet_L | GCN | 12.22 | ± | 10.71 | ± | 4.84 | ± | 32.0 |
| | | 0.30 | | 0.87 | | 0.11 | | |
| PMI_L + PR_L + Triplet_L | GIN | 6.45 | ± | 4.81 | ± | 1.92 | ± | 114.0 |
| | | 0.27 | | 0.13 | | 0.20 | | |

Dot-Adj Corr Continued (↑)

| Loss Type | Model | CORA | | Citeseer | | Bitcoin Fraud Transaction | | Average Rank |
|---|---|---|---|---|---|---|---|---|
| PMI_L + PR_L + Triplet_L | MPNN | 9.80 | ± | 5.35 | ± | 3.21 | ± | 84.3 |
| | | 0.38 | | 0.13 | | 1.26 | | |
| PMI_L + PR_L + Triplet_L | PAGNN | 3.44 | ± | 3.20 | ± | 0.54 | ± | 156.0 |
| | | 0.13 | | 0.11 | | 0.01 | | |
| PMI_L + PR_L + Triplet_L | SAGE | 7.71 | ± | 6.96 | ± | 1.10 | ± | 103.7 |
| | | 0.38 | | 0.26 | | 0.08 | | |
| PMI_L + Triplet_L | ALL | 8.76 | ± | 8.11 | ± | 4.45 | ± | 65.7 |
| | | 0.30 | | 0.43 | | 0.17 | | |
| PMI_L + Triplet_L | GAT | 13.38 | ± | 12.67 | ± | 5.10 | ± | 12.7 |
| | | 0.38 | | 0.40 | | 0.33 | | |
| PMI_L + Triplet_L | GCN | 12.51 | ± | 11.53 | ± | 5.17 | ± | 20.0 |
| | | 0.31 | | 0.26 | | 0.24 | | |
| PMI_L + Triplet_L | GIN | 6.75 | ± | 5.28 | ± | 2.30 | ± | 106.0 |
| | | 0.62 | | 0.32 | | 0.18 | | |
| PMI_L + Triplet_L | MPNN | 10.09 | ± | 9.78 | ± | 4.99 | ± | 45.3 |
| | | 0.49 | | 0.89 | | 0.20 | | |
| PMI_L + Triplet_L | PAGNN | 3.30 | ± | 2.97 | ± | 0.52 | ± | 161.0 |
| | | 0.18 | | 0.04 | | 0.01 | | |
| PMI_L + Triplet_L | SAGE | 7.63 | ± | 7.18 | ± | 3.12 | ± | 86.3 |
| | | 0.51 | | 0.23 | | 0.20 | | |
| PR_L | ALL | -1.60 | ± | -0.15 | ± | -3.38 | ± | 205.3 |
| | | 0.53 | | 0.42 | | 0.12 | | |
| PR_L | GAT | -1.39 | ± | 0.09 | ± | -0.66 | ± | 200.0 |
| | | 1.88 | | 0.42 | | 0.51 | | |
| PR_L | GCN | 1.86 | ± | 1.60 | ± | 1.29 | ± | 163.3 |
| | | 0.41 | | 0.08 | | 0.68 | | |
| PR_L | GIN | -0.04 | ± | 0.91 | ± | -3.51 | ± | 200.0 |
| | | 0.37 | | 0.30 | | 0.19 | | |
| PR_L | MPNN | 1.67 | ± | 0.98 | ± | 1.53 | ± | 166.3 |
| | | 1.77 | | 0.11 | | 1.05 | | |
| PR_L | PAGNN | -0.14 | ± | 0.29 | ± | 0.23 | ± | 194.3 |
| | | 0.57 | | 0.20 | | 0.09 | | |

Continued on next page

Dot-Adj Corr Continued (↑)

| Loss Type | Model | CORA | | Citeseer | | Bitcoin Fraud Transaction | | Average Rank |
|---|---|---|---|---|---|---|---|---|
| PR_L | SAGE | -1.09 1.08 | ± | -0.68 0.71 | ± | -1.62 0.20 | ± | 202.0 |
| PR_L + Triplet_L | ALL | -0.88 0.99 | ± | 0.56 0.17 | ± | -3.36 0.16 | ± | 201.7 |
| PR_L + Triplet_L | GAT | 4.69 4.20 | ± | 1.09 0.27 | ± | 0.04 0.73 | ± | 171.3 |
| PR_L + Triplet_L | GCN | 2.96 0.50 | ± | 2.65 0.16 | ± | 1.92 0.26 | ± | 151.3 |
| PR_L + Triplet_L | GIN | 0.79 0.18 | ± | 2.23 0.36 | ± | -2.96 0.28 | ± | 189.3 |
| PR_L + Triplet_L | MPNN | 1.31 0.63 | ± | 0.95 0.36 | ± | 1.82 0.92 | ± | 166.0 |
| PR_L + Triplet_L | PAGNN | 0.16 0.22 | ± | 0.85 0.21 | ± | 0.72 0.04 | ± | 182.3 |
| PR_L + Triplet_L | SAGE | 0.10 0.86 | ± | 2.55 2.66 | ± | -0.27 0.64 | ± | 186.0 |
| Triplet_L | ALL | 10.38 0.42 | ± | 9.28 0.14 | ± | 4.81 0.17 | ± | 49.7 |
| Triplet_L | GAT | 15.89 0.65 | ± | 14.08 0.16 | ± | 6.20 0.26 | ± | 2.0 |
| Triplet_L | GCN | 13.43 0.30 | ± | 12.32 0.39 | ± | 5.71 0.30 | ± | 9.3 |
| Triplet_L | GIN | 12.52 0.92 | ± | 9.93 0.65 | ± | 3.53 0.17 | ± | 45.0 |
| Triplet_L | MPNN | 11.39 0.83 | ± | 11.37 0.50 | ± | 5.19 0.15 | ± | 29.0 |
| Triplet_L | PAGNN | 8.89 2.67 | ± | 8.76 0.31 | ± | 0.97 0.01 | ± | 97.0 |
| Triplet_L | SAGE | 12.34 0.42 | ± | 10.48 0.57 | ± | 4.10 0.08 | ± | 40.0 |

Table 14. Euclidean-Adj Corr Performance (↑): Top-ranked results are highlighted in **1st**, second-ranked in **2nd**, and third-ranked in **3rd**.

| Loss Type | Model | CORA | | Citeseer | | Bitcoin Fraud Transaction | | Average Rank |
|---|---|---|---|---|---|---|---|---|
| Contr_l | ALL | 9.88 ± 0.25 | | 8.53 ± 0.37 | | 4.86 ± 0.17 | | 87.0 |
| Contr_l | GAT | 16.39 ± 0.50 | | 18.34 ± 0.40 | | 7.92 ± 0.23 | | 18.3 |
| Contr_l | GCN | 15.36 ± 0.16 | | 16.22 ± 0.47 | | 7.38 ± 0.29 | | 28.3 |
| Contr_l | GIN | 13.10 ± 0.44 | | 14.81 ± 0.47 | | 5.06 ± 0.78 | | 57.0 |
| Contr_l | MPNN | 12.11 ± 0.36 | | 13.00 ± 0.82 | | 5.02 ± 0.10 | | 69.3 |
| Contr_l | PAGNN | 10.59 ± 0.73 | | 8.77 ± 0.47 | | 0.72 ± 0.01 | | 113.7 |
| Contr_l | SAGE | 12.62 ± 0.29 | | 13.00 ± 1.06 | | 4.31 ± 0.26 | | 72.7 |
| Contr_l + CrossE_L | ALL | 9.74 ± 0.34 | | 8.71 ± 0.14 | | 4.59 ± 0.29 | | 89.3 |
| Contr_l + CrossE_L | GAT | 16.33 ± 0.79 | | 18.93 ± 0.15 | | 8.38 ± 0.76 | | 15.0 |
| Contr_l + CrossE_L | GCN | 15.46 ± 0.25 | | 15.80 ± 0.35 | | 7.25 ± 0.26 | | 29.3 |
| Contr_l + CrossE_L | GIN | 12.30 ± 0.48 | | 13.14 ± 0.73 | | 4.64 ± 0.64 | | 71.3 |
| Contr_l + CrossE_L | MPNN | 12.33 ± 0.71 | | 13.26 ± 0.46 | | 5.45 ± 0.14 | | 62.3 |
| Contr_l + CrossE_L | PAGNN | 10.39 ± 0.53 | | 8.45 ± 0.32 | | 0.74 ± 0.01 | | 114.3 |
| Contr_l + CrossE_L | SAGE | 12.89 ± 0.57 | | 13.94 ± 0.94 | | 4.34 ± 0.44 | | 67.3 |
| Contr_l + CrossE_L + PMI_L | ALL | 7.24 ± 0.85 | | 4.79 ± 0.98 | | 3.32 ± 0.13 | | 121.7 |

Continued on next page

Euclidean-Adj Corr Continued (↑)

| Loss Type | Model | CORA | | Citeseer | | Bitcoin Fraud Transaction | | Average Rank |
|---|---|---|---|---|---|---|---|---|
| Contr_l + CrossE_L + PMI_L | GAT | 17.36 0.51 | ± | 18.66 0.66 | ± | 5.36 0.41 | ± | 28.0 |
| Contr_l + CrossE_L + PMI_L | GCN | 17.39 0.83 | ± | 17.23 0.41 | ± | 6.84 0.13 | ± | 21.3 |
| Contr_l + CrossE_L + PMI_L | GIN | 8.26 0.93 | ± | 5.65 0.34 | ± | 1.48 0.23 | ± | 124.3 |
| Contr_l + CrossE_L + PMI_L | MPNN | 13.82 0.82 | ± | 14.47 0.78 | ± | 5.62 0.33 | ± | 50.7 |
| Contr_l + CrossE_L + PMI_L | PAGNN | 3.43 0.24 | ± | 3.39 0.11 | ± | 0.20 0.02 | ± | 176.7 |
| Contr_l + CrossE_L + PMI_L | SAGE | 6.32 0.22 | ± | 6.39 0.76 | ± | 1.31 0.42 | ± | 129.3 |
| Contr_l + CrossE_L + PMI_L + PR_L | ALL | 3.80 0.55 | ± | 3.71 0.39 | ± | 0.96 0.02 | ± | 156.7 |
| Contr_l + CrossE_L + PMI_L + PR_L | GAT | 15.38 1.79 | ± | 16.32 0.64 | ± | 2.76 1.21 | ± | 59.0 |
| Contr_l + CrossE_L + PMI_L + PR_L | GCN | 17.17 0.88 | ± | 16.76 0.49 | ± | 6.60 0.28 | ± | 29.7 |
| Contr_l + CrossE_L + PMI_L + PR_L | GIN | 7.72 0.71 | ± | 4.87 0.49 | ± | 0.89 0.13 | ± | 136.3 |
| Contr_l + CrossE_L + PMI_L + PR_L | MPNN | 12.50 0.55 | ± | 9.11 2.41 | ± | 2.12 1.87 | ± | 97.0 |
| Contr_l + CrossE_L + PMI_L + PR_L | PAGNN | 3.96 0.29 | ± | 3.52 0.10 | ± | 0.12 0.02 | ± | 171.7 |
| Contr_l + CrossE_L + PMI_L + PR_L | SAGE | 6.09 0.55 | ± | 5.92 0.61 | ± | 0.40 0.06 | ± | 144.3 |
| Contr_l + CrossE_L + PMI_L + PR_L + Triplet_L | ALL | 8.82 0.16 | ± | 5.97 0.11 | ± | 2.90 0.26 | ± | 111.7 |
| Contr_l + CrossE_L + PMI_L + PR_L + Triplet_L | GAT | 15.30 1.20 | ± | 15.09 3.25 | ± | 5.33 1.37 | ± | 45.0 |
| Contr_l + CrossE_L + PMI_L + PR_L + Triplet_L | GCN | 17.65 0.35 | ± | 17.17 0.56 | ± | 7.17 0.05 | ± | 18.0 |

Euclidean-Adj Corr Continued (↑)

| Loss Type | Model | CORA | | Citeseer | | Bitcoin Fraud Transaction | | Average Rank |
|---|---|---|---|---|---|---|---|---|
| Contr_l + CrossE_L + PMI_L + PR_L + Triplet_L | GIN | 8.07 0.37 | ± | 6.87 0.32 | ± | 2.23 0.23 | ± | 118.0 |
| Contr_l + CrossE_L + PMI_L + PR_L + Triplet_L | MPNN | 12.71 0.75 | ± | 9.29 1.58 | ± | 5.07 1.57 | ± | 73.3 |
| Contr_l + CrossE_L + PMI_L + PR_L + Triplet_L | PAGNN | 4.09 0.34 | ± | 3.67 0.08 | ± | 0.25 0.02 | ± | 165.3 |
| Contr_l + CrossE_L + SAGE PMI_L + PR_L + Triplet_L | SAGE | 8.52 0.91 | ± | 8.96 0.50 | ± | 0.99 0.12 | ± | 115.0 |
| Contr_l + CrossE_L + PMI_L + Triplet_L | ALL | 10.45 0.32 | ± | 10.72 0.34 | ± | 5.33 0.19 | ± | 75.0 |
| Contr_l + CrossE_L + PMI_L + Triplet_L | GAT | 18.39 1.16 | ± | 18.11 0.41 | ± | 6.52 0.13 | ± | 19.0 |
| Contr_l + CrossE_L + PMI_L + Triplet_L | GCN | 18.15 0.80 | ± | 17.10 0.49 | ± | 7.15 0.47 | ± | 16.7 |
| Contr_l + CrossE_L + PMI_L + Triplet_L | GIN | 8.91 0.89 | ± | 7.19 0.30 | ± | 2.53 0.33 | ± | 109.0 |
| Contr_l + CrossE_L + PMI_L + Triplet_L | MPNN | 13.51 0.44 | ± | 13.90 1.46 | ± | 6.06 0.35 | ± | 52.7 |
| Contr_l + CrossE_L + PMI_L + Triplet_L | PAGNN | 4.15 0.22 | ± | 3.48 0.02 | ± | 0.25 0.01 | ± | 167.0 |
| Contr_l + CrossE_L + PMI_L + Triplet_L | SAGE | 9.18 0.18 | ± | 8.79 0.60 | ± | 3.44 0.27 | ± | 96.7 |
| Contr_l + CrossE_L + PR_L | ALL | 0.64 0.25 | ± | 1.87 1.02 | ± | -2.85 0.18 | ± | 197.3 |
| Contr_l + CrossE_L + PR_L | GAT | 6.32 5.22 | ± | 4.33 4.03 | ± | 0.99 0.55 | ± | 144.7 |
| Contr_l + CrossE_L + PR_L | GCN | 5.35 0.39 | ± | 4.73 0.17 | ± | 3.00 0.10 | ± | 133.0 |
| Contr_l + CrossE_L + PR_L | GIN | 1.61 0.58 | ± | 3.64 0.45 | ± | -1.90 0.21 | ± | 183.0 |
| Contr_l + CrossE_L + PR_L | MPNN | 2.08 0.30 | ± | 2.12 0.29 | ± | 2.43 1.73 | ± | 162.3 |

Euclidean-Adj Corr Continued (↑)

| Loss Type | Model | CORA | | Citeseer | | Bitcoin Fraud Transaction | | Average Rank |
|---|---|---|---|---|---|---|---|---|
| Contr_l + CrossE_L + PR_L | PAGNN | 1.17 ± 0.88 | | 1.78 ± 0.55 | | 0.59 ± 0.04 | | 182.0 |
| Contr_l + CrossE_L + PR_L | SAGE | 7.28 ± 3.50 | | 5.55 ± 2.36 | | 0.53 ± 0.33 | | 140.3 |
| Contr_l + CrossE_L + PR_L + Triplet_L | ALL | 6.42 ± 0.32 | | 6.20 ± 0.15 | | 3.65 ± 0.26 | | 114.3 |
| Contr_l + CrossE_L + PR_L + Triplet_L | GAT | 14.16 ± 2.60 | | 12.32 ± 3.39 | | 4.91 ± 1.15 | | 62.3 |
| Contr_l + CrossE_L + PR_L + Triplet_L | GCN | 9.73 ± 1.23 | | 9.29 ± 0.82 | | 5.38 ± 1.61 | | 80.0 |
| Contr_l + CrossE_L + PR_L + Triplet_L | GIN | 8.24 ± 0.44 | | 7.33 ± 0.42 | | 3.17 ± 1.52 | | 107.3 |
| Contr_l + CrossE_L + PR_L + Triplet_L | MPNN | 6.18 ± 1.12 | | 6.95 ± 0.41 | | 3.81 ± 1.11 | | 111.7 |
| Contr_l + CrossE_L + PR_L + Triplet_L | PAGNN | 3.60 ± 0.42 | | 4.54 ± 1.02 | | 0.63 ± 0.04 | | 162.0 |
| Contr_l + CrossE_L + PR_L + Triplet_L | SAGE | 11.84 ± 0.62 | | 9.25 ± 1.07 | | 2.62 ± 0.28 | | 94.7 |
| Contr_l + CrossE_L + Triplet_L | ALL | 11.74 ± 0.45 | | 10.79 ± 0.42 | | 5.88 ± 0.22 | | 68.0 |
| Contr_l + CrossE_L + Triplet_L | GAT | 18.84 ± 1.06 | | 19.18 ± 0.78 | | 9.48 ± 0.38 | | 3.3 |
| Contr_l + CrossE_L + Triplet_L | GCN | 17.07 ± 0.22 | | 17.06 ± 0.35 | | 7.98 ± 0.29 | | 20.0 |
| Contr_l + CrossE_L + Triplet_L | GIN | 15.21 ± 0.76 | | 15.26 ± 1.18 | | 5.24 ± 0.67 | | 46.0 |
| Contr_l + CrossE_L + Triplet_L | MPNN | 14.09 ± 0.64 | | 14.99 ± 0.54 | | 6.23 ± 0.28 | | 45.7 |
| Contr_l + CrossE_L + Triplet_L | PAGNN | 10.17 ± 2.37 | | 10.59 ± 0.33 | | 0.74 ± 0.02 | | 108.0 |
| Contr_l + CrossE_L + Triplet_L | SAGE | 14.81 ± 0.83 | | 14.44 ± 0.58 | | 5.61 ± 0.32 | | 48.0 |

Euclidean-Adj Corr Continued (↑)

| Loss Type | Model | CORA | | | Citeseer | | | Bitcoin Fraud Transaction | | | Average Rank |
|---|---|---|---|---|---|---|---|---|---|---|---|
| Contr_l + PMI_L | ALL | 7.41 | ± | | 6.92 | ± | | 3.72 | ± | | 106.7 |
| | | 1.05 | | | 0.45 | | | 0.24 | | | |
| Contr_l + PMI_L | GAT | 17.95 | ± | | 18.00 | ± | | 5.42 | ± | | 27.7 |
| | | 0.41 | | | 1.02 | | | 0.42 | | | |
| Contr_l + PMI_L | GCN | 18.19 | ± | | 16.86 | ± | | 7.01 | ± | | 19.7 |
| | | 0.43 | | | 0.44 | | | 0.39 | | | |
| Contr_l + PMI_L | GIN | 7.78 | ± | | 6.06 | ± | | 1.66 | ± | | 124.3 |
| | | 1.00 | | | 0.53 | | | 0.11 | | | |
| Contr_l + PMI_L | MPNN | 13.32 | ± | | 14.40 | ± | | 5.66 | ± | | 52.7 |
| | | 0.53 | | | 0.78 | | | 0.11 | | | |
| Contr_l + PMI_L | PAGNN | 3.67 | ± | | 3.35 | ± | | 0.22 | ± | | 175.3 |
| | | 0.28 | | | 0.06 | | | 0.02 | | | |
| Contr_l + PMI_L | SAGE | 6.60 | ± | | 7.60 | ± | | 1.15 | ± | | 125.0 |
| | | 0.51 | | | 0.58 | | | 0.08 | | | |
| Contr_l + PMI_L + PR_L | ALL | 4.61 | ± | | 3.63 | ± | | 1.07 | ± | | 151.7 |
| | | 0.39 | | | 0.27 | | | 0.05 | | | |
| Contr_l + PMI_L + PR_L | GAT | 14.48 | ± | | 11.55 | ± | | 2.13 | ± | | 81.7 |
| | | 0.92 | | | 3.76 | | | 1.22 | | | |
| Contr_l + PMI_L + PR_L | GCN | 18.05 | ± | | 16.15 | ± | | 6.85 | ± | | 23.7 |
| | | 0.20 | | | 0.55 | | | 0.31 | | | |
| Contr_l + PMI_L + PR_L | GIN | 8.09 | ± | | 5.08 | ± | | 0.97 | ± | | 133.3 |
| | | 0.80 | | | 0.77 | | | 0.07 | | | |
| Contr_l + PMI_L + PR_L | MPNN | 13.28 | ± | | 8.07 | ± | | 2.16 | ± | | 95.7 |
| | | 0.72 | | | 3.07 | | | 1.82 | | | |
| Contr_l + PMI_L + PR_L | PAGNN | 3.60 | ± | | 3.42 | ± | | 0.15 | ± | | 177.0 |
| | | 0.35 | | | 0.08 | | | 0.01 | | | |
| Contr_l + PMI_L + PR_L | SAGE | 6.35 | ± | | 6.33 | ± | | 0.40 | ± | | 141.7 |
| | | 0.33 | | | 0.49 | | | 0.03 | | | |
| Contr_l + PMI_L + PR_L + Triplet_L | ALL | 9.04 | ± | | 7.03 | ± | | 3.44 | ± | | 102.7 |
| | | 0.41 | | | 0.11 | | | 0.27 | | | |
| Contr_l + PMI_L + PR_L + Triplet_L | GAT | 14.33 | ± | | 11.81 | ± | | 4.34 | ± | | 67.7 |
| | | 0.71 | | | 1.37 | | | 1.03 | | | |

Euclidean-Adj Corr Continued (↑)

| Loss Type | Model | CORA | | Citeseer | | Bitcoin Fraud Transaction | | Average Rank |
|---|---|---|---|---|---|---|---|---|
| Contr_l + PMI_L + PR_L + Triplet_L | GCN | 16.27 1.06 | ± | 14.84 0.94 | ± | 6.98 0.45 | ± | 33.0 |
| Contr_l + PMI_L + PR_L + Triplet_L | GIN | 9.41 0.59 | ± | 7.68 0.41 | ± | 2.75 0.14 | ± | 104.7 |
| Contr_l + PMI_L + PR_L + Triplet_L | MPNN | 12.58 0.52 | ± | 8.58 0.30 | ± | 3.46 1.13 | ± | 88.0 |
| Contr_l + PMI_L + PR_L + Triplet_L | PAGNN | 4.31 0.18 | ± | 4.16 0.36 | ± | 0.34 0.03 | ± | 160.7 |
| Contr_l + PMI_L + PR_L + Triplet_L | SAGE | 10.08 0.77 | ± | 10.05 0.53 | ± | 1.79 0.28 | ± | 101.0 |
| Contr_l + PR_L | ALL | 0.49 0.31 | ± | 1.29 1.18 | ± | -2.81 0.34 | ± | 201.3 |
| Contr_l + PR_L | GAT | 4.94 4.09 | ± | 2.84 0.65 | ± | 1.12 0.54 | ± | 156.7 |
| Contr_l + PR_L | GCN | 5.01 0.42 | ± | 5.58 0.77 | ± | 4.78 1.73 | ± | 116.7 |
| Contr_l + PR_L | GIN | 1.76 0.41 | ± | 3.47 0.99 | ± | -1.96 0.16 | ± | 185.3 |
| Contr_l + PR_L | MPNN | 1.47 0.31 | ± | 2.24 0.38 | ± | 1.38 0.09 | ± | 169.7 |
| Contr_l + PR_L | PAGNN | 0.70 0.57 | ± | 1.79 0.43 | ± | 0.56 0.04 | ± | 183.7 |
| Contr_l + PR_L | SAGE | 7.36 3.39 | ± | 6.67 3.00 | ± | 0.66 0.28 | ± | 132.0 |
| Contr_l + PR_L + Triplet_L | ALL | 5.48 1.60 | ± | 5.27 0.98 | ± | 3.62 0.09 | ± | 122.7 |
| Contr_l + PR_L + Triplet_L | GAT | 13.84 4.66 | ± | 12.50 3.23 | ± | 4.78 1.30 | ± | 64.3 |
| Contr_l + PR_L + Triplet_L | GCN | 8.31 0.76 | ± | 9.02 0.87 | ± | 4.45 1.56 | ± | 93.7 |
| Contr_l + PR_L + Triplet_L | GIN | 8.30 0.42 | ± | 6.50 0.37 | ± | 2.40 1.32 | ± | 115.0 |

Euclidean-Adj Corr Continued (↑)

| Loss Type | Model | CORA | | Citeseer | | Bitcoin Fraud Transaction | | Average Rank |
|---|---|---|---|---|---|---|---|---|
| Contr_l + PR_L + Triplet_L | MPNN | 6.37 1.64 | ± | 5.87 1.19 | ± | 3.05 0.77 | ± | 121.3 |
| Contr_l + PR_L + Triplet_L | PAGNN | 3.31 0.41 | ± | 4.81 1.29 | ± | 0.66 0.02 | ± | 159.3 |
| Contr_l + PR_L + Triplet_L | SAGE | 11.14 0.64 | ± | 10.37 0.87 | ± | 3.04 0.27 | ± | 90.0 |
| Contr_l + Triplet_L | ALL | 12.33 0.31 | ± | 11.13 0.28 | ± | 6.30 0.17 | ± | 64.0 |
| Contr_l + Triplet_L | GAT | 18.08 0.96 | ± | 18.97 0.89 | ± | 9.41 0.25 | ± | 7.0 |
| Contr_l + Triplet_L | GCN | 17.17 0.32 | ± | 17.16 0.24 | ± | 7.96 0.25 | ± | 19.0 |
| Contr_l + Triplet_L | GIN | 14.93 0.74 | ± | 14.54 1.11 | ± | 4.79 0.61 | ± | 53.3 |
| Contr_l + Triplet_L | MPNN | 14.20 0.68 | ± | 15.28 0.44 | ± | 6.34 0.21 | ± | 42.7 |
| Contr_l + Triplet_L | PAGNN | 10.65 2.75 | ± | 10.77 0.24 | ± | 0.73 0.02 | ± | 106.3 |
| Contr_l + Triplet_L | SAGE | 15.05 0.38 | ± | 14.25 1.08 | ± | 5.44 0.78 | ± | 49.0 |
| CrossE_L | ALL | 4.68 1.32 | ± | 3.31 2.25 | ± | -1.47 0.50 | ± | 176.3 |
| CrossE_L | GAT | 4.38 2.78 | ± | 4.90 2.83 | ± | 3.17 1.64 | ± | 132.3 |
| CrossE_L | GCN | -5.24 0.49 | ± | -3.01 0.14 | ± | -3.69 0.08 | ± | 209.7 |
| CrossE_L | GIN | -5.18 0.98 | ± | -2.90 0.15 | ± | -4.16 0.04 | ± | 209.3 |
| CrossE_L | MPNN | 4.11 0.35 | ± | 4.56 0.66 | ± | -3.54 0.24 | ± | 174.0 |
| CrossE_L | PAGNN | 3.00 0.26 | ± | 2.25 0.22 | ± | -0.51 0.08 | ± | 185.7 |

Euclidean-Adj Corr Continued (↑)

| Loss Type | Model | CORA | | Citeseer | | Bitcoin Fraud Transaction | | Average Rank |
|---|---|---|---|---|---|---|---|---|
| CrossE_L | SAGE | 2.46 0.85 | ± | 0.63 0.12 | ± | 0.00 0.02 | ± | 192.7 |
| CrossE_L + PMI_L | ALL | 7.19 0.39 | ± | 4.57 0.31 | ± | 3.51 0.39 | ± | 122.7 |
| CrossE_L + PMI_L | GAT | 18.42 0.76 | ± | 19.13 0.36 | ± | 5.13 0.27 | ± | 23.7 |
| CrossE_L + PMI_L | GCN | 18.12 0.69 | ± | 17.18 0.66 | ± | 6.75 0.33 | ± | 19.3 |
| CrossE_L + PMI_L | GIN | 8.14 0.80 | ± | 4.94 0.24 | ± | 1.27 0.08 | ± | 130.3 |
| CrossE_L + PMI_L | MPNN | 12.92 0.41 | ± | 14.04 0.93 | ± | 5.72 0.44 | ± | 55.7 |
| CrossE_L + PMI_L | PAGNN | 3.78 0.47 | ± | 3.35 0.07 | ± | 0.16 0.05 | ± | 176.7 |
| CrossE_L + PMI_L | SAGE | 5.41 0.25 | ± | 4.97 0.11 | ± | 0.19 0.11 | ± | 155.7 |
| CrossE_L + PMI_L + PR_L | ALL | 3.74 0.10 | ± | 3.50 0.08 | ± | 0.76 0.02 | ± | 161.3 |
| CrossE_L + PMI_L + PR_L | GAT | 15.55 1.19 | ± | 11.16 4.81 | ± | 2.76 1.38 | ± | 73.0 |
| CrossE_L + PMI_L + PR_L | GCN | 17.35 0.69 | ± | 16.99 0.70 | ± | 6.47 0.76 | ± | 28.3 |
| CrossE_L + PMI_L + PR_L | GIN | 7.22 0.32 | ± | 4.81 0.91 | ± | 0.76 0.27 | ± | 140.3 |
| CrossE_L + PMI_L + PR_L | MPNN | 12.70 0.91 | ± | 9.80 3.45 | ± | 3.12 2.13 | ± | 85.3 |
| CrossE_L + PMI_L + PR_L | PAGNN | 4.02 0.39 | ± | 3.49 0.21 | ± | 0.10 0.04 | ± | 172.3 |
| CrossE_L + PMI_L + PR_L | SAGE | 5.60 0.27 | ± | 5.34 0.39 | ± | 0.22 0.05 | ± | 152.3 |
| CrossE_L + PMI_L + PR_L + Triplet_L | ALL | 8.52 0.15 | ± | 5.60 0.21 | ± | 3.12 0.07 | ± | 113.7 |

Euclidean-Adj Corr Continued (↑)

| Loss Type | Model | CORA | | Citeseer | | Bitcoin Fraud Transaction | | Average Rank |
|-----------|-------|------|---|----------|---|---------------------------|---|--------------|
| CrossE_L + PMI_L + PR_L + Triplet_L | GAT | 15.12 0.79 | ± | 13.80 2.25 | ± | 4.68 1.72 | ± | 58.3 |
| CrossE_L + PMI_L + PR_L + Triplet_L | GCN | 17.50 0.49 | ± | 17.04 0.41 | ± | 7.08 0.32 | ± | 21.0 |
| CrossE_L + PMI_L + PR_L + Triplet_L | GIN | 8.19 0.35 | ± | 6.64 0.46 | ± | 2.26 0.15 | ± | 117.0 |
| CrossE_L + PMI_L + PR_L + Triplet_L | MPNN | 13.23 0.47 | ± | 9.92 2.45 | ± | 5.11 1.67 | ± | 69.7 |
| CrossE_L + PMI_L + PR_L + Triplet_L | PAGNN | 4.40 0.11 | ± | 3.70 0.19 | ± | 0.25 0.02 | ± | 163.3 |
| CrossE_L + PMI_L + PR_L + Triplet_L | SAGE | 8.53 0.94 | ± | 8.25 0.43 | ± | 0.80 0.07 | ± | 119.0 |
| CrossE_L + PMI_L + Triplet_L | ALL | 12.13 0.40 | ± | 12.82 0.55 | ± | 5.93 0.28 | ± | 62.3 |
| CrossE_L + PMI_L + Triplet_L | GAT | 18.47 0.49 | ± | 18.45 0.50 | ± | 7.05 0.33 | ± | 12.0 |
| CrossE_L + PMI_L + Triplet_L | GCN | 17.94 0.86 | ± | 17.58 0.33 | ± | 7.31 0.35 | ± | 15.3 |
| CrossE_L + PMI_L + Triplet_L | GIN | 9.29 0.40 | ± | 7.91 0.34 | ± | 3.20 0.21 | ± | 100.3 |
| CrossE_L + PMI_L + Triplet_L | MPNN | 13.01 0.45 | ± | 13.92 0.77 | ± | 6.65 0.18 | ± | 51.7 |
| CrossE_L + PMI_L + Triplet_L | PAGNN | 4.51 0.27 | ± | 3.53 0.13 | ± | 0.28 0.01 | ± | 162.3 |
| CrossE_L + PMI_L + Triplet_L | SAGE | 10.16 0.21 | ± | 9.75 0.54 | ± | 4.07 0.34 | ± | 87.3 |
| CrossE_L + PR_L | ALL | -0.16 0.25 | ± | 0.48 0.35 | ± | -3.16 0.10 | ± | 207.0 |
| CrossE_L + PR_L | GAT | 1.57 0.52 | ± | 1.44 0.39 | ± | -0.17 0.26 | ± | 192.7 |
| CrossE_L + PR_L | GCN | 3.00 0.38 | ± | 2.78 0.25 | ± | -1.75 1.03 | ± | 187.0 |

Continued on next page

Euclidean-Adj Corr Continued (↑)

| Loss Type | Model | CORA | | Citeseer | | Bitcoin Fraud Transaction | | Average Rank |
|---|---|---|---|---|---|---|---|---|
| CrossE_L + PR_L | GIN | 0.71 ± 0.26 | | 1.49 ± 0.51 | | -2.97 ± 0.10 | | 199.0 |
| CrossE_L + PR_L | MPNN | 2.47 ± 3.34 | | 1.56 ± 0.18 | | 0.84 ± 0.20 | | 173.7 |
| CrossE_L + PR_L | PAGNN | 0.44 ± 0.49 | | 0.74 ± 0.37 | | 0.07 ± 0.05 | | 198.3 |
| CrossE_L + PR_L | SAGE | 0.94 ± 0.41 | | 1.28 ± 0.34 | | -1.07 ± 0.29 | | 197.0 |
| CrossE_L + PR_L + Triplet_L | ALL | 4.67 ± 1.81 | | 3.36 ± 0.85 | | 1.30 ± 2.05 | | 153.7 |
| CrossE_L + PR_L + Triplet_L | GAT | 6.45 ± 4.00 | | 11.25 ± 5.33 | | 2.36 ± 1.17 | | 107.7 |
| CrossE_L + PR_L + Triplet_L | GCN | 7.05 ± 0.95 | | 7.40 ± 1.07 | | 4.73 ± 1.43 | | 103.0 |
| CrossE_L + PR_L + Triplet_L | GIN | 6.00 ± 1.05 | | 5.82 ± 0.54 | | -1.07 ± 0.23 | | 155.3 |
| CrossE_L + PR_L + Triplet_L | MPNN | 3.06 ± 0.15 | | 3.47 ± 0.24 | | 2.54 ± 0.81 | | 152.7 |
| CrossE_L + PR_L + Triplet_L | PAGNN | 2.44 ± 0.66 | | 3.25 ± 0.50 | | 0.59 ± 0.05 | | 175.7 |
| CrossE_L + PR_L + Triplet_L | SAGE | 11.95 ± 1.43 | | 10.36 ± 1.67 | | 1.75 ± 0.90 | | 96.7 |
| CrossE_L + Triplet_L | ALL | 14.58 ± 0.53 | | 13.94 ± 0.45 | | 6.66 ± 0.25 | | 44.3 |
| CrossE_L + Triplet_L | GAT | 20.27 ± 0.87 | | 20.17 ± 0.22 | | 8.43 ± 0.68 | | 2.3 |
| CrossE_L + Triplet_L | GCN | 18.86 ± 0.62 | | 18.25 ± 0.66 | | 8.28 ± 0.08 | | 7.7 |
| CrossE_L + Triplet_L | GIN | 15.75 ± 0.81 | | 13.64 ± 0.17 | | 4.39 ± 0.53 | | 57.7 |
| CrossE_L + Triplet_L | MPNN | 16.73 ± 0.81 | | 17.50 ± 1.19 | | 6.94 ± 0.21 | | 23.0 |

Continued on next page

Euclidean-Adj Corr Continued (↑)

| Loss Type | Model | CORA | | | Citeseer | | Bitcoin Fraud Transaction | | Average Rank |
|---|---|---|---|---|---|---|---|---|---|
| CrossE_L + Triplet_L | PAGNN | 11.28 3.13 | ± | 12.16 0.62 | ± | 0.73 0.01 | ± | 103.3 | | |
| CrossE_L + Triplet_L | SAGE | 17.12 0.55 | ± | 15.09 0.69 | ± | 4.82 0.16 | ± | 44.7 | | |
| PMI_L | ALL | 5.93 0.15 | ± | 4.73 0.34 | ± | 3.68 0.13 | ± | 125.7 | | |
| PMI_L | GAT | 19.22 0.98 | ± | 18.90 1.16 | ± | 5.24 0.30 | ± | 22.3 | | |
| PMI_L | GCN | 18.05 0.23 | ± | 17.04 0.69 | ± | 6.85 0.51 | ± | 21.7 | | |
| PMI_L | GIN | 7.04 0.86 | ± | 5.10 0.34 | ± | 1.24 0.20 | ± | 134.3 | | |
| PMI_L | MPNN | 13.43 0.73 | ± | 14.73 0.56 | ± | 5.72 0.28 | ± | 50.3 | | |
| PMI_L | PAGNN | 3.69 0.30 | ± | 3.34 0.06 | ± | 0.17 0.03 | ± | 177.3 | | |
| PMI_L | SAGE | 5.12 0.45 | ± | 4.77 0.21 | ± | 0.27 0.11 | ± | 155.3 | | |
| PMI_L + PR_L | ALL | 3.96 0.39 | ± | 3.41 0.27 | ± | 0.78 0.02 | ± | 162.7 | | |
| PMI_L + PR_L | GAT | 13.54 0.98 | ± | 10.01 5.06 | ± | 1.24 0.10 | ± | 92.3 | | |
| PMI_L + PR_L | GCN | 16.44 0.81 | ± | 16.81 0.78 | ± | 6.65 0.37 | ± | 31.3 | | |
| PMI_L + PR_L | GIN | 7.17 0.71 | ± | 5.15 1.06 | ± | 0.65 0.07 | ± | 141.3 | | |
| PMI_L + PR_L | MPNN | 12.70 0.60 | ± | 6.37 0.28 | ± | 1.83 1.82 | ± | 106.0 | | |
| PMI_L + PR_L | PAGNN | 3.82 0.23 | ± | 3.45 0.11 | ± | 0.06 0.02 | ± | 175.7 | | |
| PMI_L + PR_L | SAGE | 5.62 0.45 | ± | 5.41 0.26 | ± | 0.21 0.05 | ± | 152.0 | | |

Euclidean-Adj Corr Continued (↑)

| Loss Type | Model | CORA | | Citeseer | | Bitcoin Fraud Transaction | | Average Rank |
|-----------|-------|------|---|----------|---|---------------------------|---|--------------|
| PMI_L + PR_L + Triplet_L | ALL | 8.57 ± 0.17 | | 6.10 ± 0.26 | | 3.23 ± 0.35 | | 108.7 |
| PMI_L + PR_L + Triplet_L | GAT | 14.52 ± 0.41 | | 13.25 ± 2.19 | | 3.54 ± 1.29 | | 66.7 |
| PMI_L + PR_L + Triplet_L | GCN | 17.20 ± 0.45 | | 16.01 ± 1.22 | | 6.66 ± 0.15 | | 29.7 |
| PMI_L + PR_L + Triplet_L | GIN | 8.30 ± 0.32 | | 6.59 ± 0.19 | | 2.25 ± 0.25 | | 116.7 |
| PMI_L + PR_L + Triplet_L | MPNN | 13.04 ± 0.49 | | 7.89 ± 0.26 | | 4.08 ± 1.65 | | 84.0 |
| PMI_L + PR_L + Triplet_L | PAGNN | 4.31 ± 0.20 | | 3.83 ± 0.14 | | 0.27 ± 0.02 | | 162.3 |
| PMI_L + PR_L + Triplet_L | SAGE | 9.73 ± 0.92 | | 9.27 ± 0.46 | | 1.01 ± 0.09 | | 109.7 |
| PMI_L + Triplet_L | ALL | 11.70 ± 0.41 | | 12.04 ± 0.66 | | 5.91 ± 0.24 | | 66.0 |
| PMI_L + Triplet_L | GAT | 18.55 ± 0.50 | | 18.60 ± 0.57 | | 6.69 ± 0.44 | | 14.3 |
| PMI_L + Triplet_L | GCN | 17.60 ± 0.45 | | 17.23 ± 0.36 | | 7.12 ± 0.32 | | 18.3 |
| PMI_L + Triplet_L | GIN | 8.61 ± 0.87 | | 7.34 ± 0.42 | | 2.71 ± 0.23 | | 108.0 |
| PMI_L + Triplet_L | MPNN | 13.39 ± 0.57 | | 14.23 ± 1.27 | | 6.37 ± 0.27 | | 50.0 |
| PMI_L + Triplet_L | PAGNN | 4.10 ± 0.24 | | 3.48 ± 0.06 | | 0.26 ± 0.00 | | 167.3 |
| PMI_L + Triplet_L | SAGE | 9.48 ± 0.78 | | 9.55 ± 0.34 | | 3.61 ± 0.28 | | 91.7 |
| PR_L | ALL | -0.82 ± 0.35 | | -0.15 ± 0.29 | | -3.13 ± 0.14 | | 207.3 |
| PR_L | GAT | 1.40 ± 0.37 | | 1.24 ± 0.26 | | -0.16 ± 0.33 | | 195.0 |

Euclidean-Adj Corr Continued (↑)

| Loss Type | Model | CORA | | Citeseer | | Bitcoin Fraud Transaction | | Average Rank |
|---|---|---|---|---|---|---|---|---|
| PR_L | GCN | 2.78 | ± | 2.96 | ± | 2.38 | ± | 159.0 |
| | | 0.50 | | 0.09 | | 0.87 | | |
| PR_L | GIN | 0.55 | ± | 1.43 | ± | -2.68 | ± | 200.0 |
| | | 0.20 | | 0.33 | | 0.14 | | |
| PR_L | MPNN | 2.58 | ± | 1.52 | ± | 1.91 | ± | 166.3 |
| | | 2.76 | | 0.12 | | 1.42 | | |
| PR_L | PAGNN | 0.12 | ± | 0.59 | ± | 0.13 | ± | 198.7 |
| | | 0.80 | | 0.20 | | 0.05 | | |
| PR_L | SAGE | 0.95 | ± | 1.12 | ± | -0.73 | ± | 197.0 |
| | | 0.47 | | 0.13 | | 0.17 | | |
| PR_L + Triplet_L | ALL | 0.46 | ± | 0.66 | ± | -3.03 | ± | 204.3 |
| | | 0.51 | | 0.23 | | 0.23 | | |
| PR_L + Triplet_L | GAT | 7.11 | ± | 2.71 | ± | 1.10 | ± | 150.7 |
| | | 6.05 | | 0.89 | | 0.92 | | |
| PR_L + Triplet_L | GCN | 4.85 | ± | 4.79 | ± | 2.88 | ± | 134.3 |
| | | 0.58 | | 0.17 | | 0.30 | | |
| PR_L + Triplet_L | GIN | 1.51 | ± | 3.45 | ± | -1.97 | ± | 188.0 |
| | | 0.17 | | 0.48 | | 0.20 | | |
| PR_L + Triplet_L | MPNN | 2.18 | ± | 1.77 | ± | 2.33 | ± | 164.7 |
| | | 1.12 | | 0.25 | | 1.22 | | |
| PR_L + Triplet_L | PAGNN | 0.56 | ± | 1.40 | ± | 0.48 | ± | 187.7 |
| | | 0.37 | | 0.27 | | 0.05 | | |
| PR_L + Triplet_L | SAGE | 1.74 | ± | 4.25 | ± | 0.25 | ± | 173.0 |
| | | 0.47 | | 3.61 | | 0.29 | | |
| Triplet_L | ALL | 14.40 | ± | 14.11 | ± | 6.47 | ± | 46.7 |
| | | 0.59 | | 0.18 | | 0.25 | | |
| Triplet_L | GAT | 21.86 | ± | 20.80 | ± | 8.18 | ± | 2.7 |
| | | 0.84 | | 0.23 | | 0.35 | | |
| Triplet_L | GCN | 18.88 | ± | 18.44 | ± | 7.88 | ± | 8.3 |
| | | 0.39 | | 0.57 | | 0.37 | | |
| Triplet_L | GIN | 16.50 | ± | 14.39 | ± | 4.51 | ± | 52.0 |
| | | 1.27 | | 1.00 | | 0.25 | | |

Euclidean-Adj Corr Continued (↑)

| Loss Type | Model | CORA | | Citeseer | | Bitcoin Fraud Transaction | | Average Rank |
|-----------|-------|------|---|----------|---|---------------------------|---|--------------|
| Triplet_L | MPNN  | 15.24 ± 1.20 | | 16.91 ± 0.76 | | 6.75 ± 0.21 | | 32.3 |
| Triplet_L | PAGNN | 11.37 ± 3.38 | | 12.28 ± 0.51 | | 0.71 ± 0.01 | | 103.3 |
| Triplet_L | SAGE  | 16.88 ± 0.64 | | 15.01 ± 0.94 | | 5.05 ± 0.08 | | 44.3 |

Table 15. Graph Reconstruction Bce Loss Performance (↓): Top-ranked results are highlighted in **1st**, second-ranked in **2nd**, and third-ranked in **3rd**.

| Loss Type | Model | CORA | | Citeseer | | Bitcoin Fraud Transaction | | Average Rank |
|-----------|-------|------|---|----------|---|---------------------------|---|--------------|
| Contr_l | ALL | 54.22 ± 0.20 | | 53.77 ± 0.15 | | 57.29 ± 0.10 | | 13.0 |
| Contr_l | GAT | 54.02 ± 0.30 | | 53.57 ± 0.13 | | 56.86 ± 0.08 | | 5.0 |
| Contr_l | GCN | 54.14 ± 0.13 | | 53.55 ± 0.32 | | 57.72 ± 0.32 | | 8.3 |
| Contr_l | GIN | 54.23 ± 0.22 | | 53.56 ± 0.17 | | 58.44 ± 0.34 | | 19.0 |
| Contr_l | MPNN | 54.16 ± 0.32 | | 53.27 ± 0.43 | | 57.67 ± 0.11 | | 5.3 |
| Contr_l | PAGNN | 55.25 ± 0.28 | | 54.54 ± 0.21 | | 68.54 ± 0.40 | | 83.3 |
| Contr_l | SAGE | 54.54 ± 0.11 | | 53.99 ± 0.42 | | 58.07 ± 0.19 | | 26.7 |
| Contr_l + CrossE_L | ALL | 54.50 ± 0.23 | | 54.13 ± 0.30 | | 57.59 ± 0.12 | | 23.3 |
| Contr_l + CrossE_L | GAT | 54.12 ± 0.45 | | 53.90 ± 0.10 | | 57.20 ± 0.28 | | 12.3 |

Graph Reconstruction Bce Loss Continued (↓)

| Loss Type | Model | CORA | | Citeseer | | Bitcoin Fraud Transaction | | Average Rank |
|---|---|---|---|---|---|---|---|---|
| Contr_l + CrossE_L | GCN | 54.46 ± 0.25 | ± | 53.41 ± 0.13 | ± | 57.72 ± 0.09 | ± | 11.3 |
| Contr_l + CrossE_L | GIN | 54.26 ± 0.20 | ± | 53.60 ± 0.19 | ± | 59.01 ± 0.91 | ± | 26.3 |
| Contr_l + CrossE_L | MPNN | 54.62 ± 0.43 | ± | 53.37 ± 0.39 | ± | 57.75 ± 0.14 | ± | 15.3 |
| Contr_l + CrossE_L | PAGNN | 55.00 ± 0.29 | ± | 54.56 ± 0.21 | ± | 68.68 ± 0.15 | ± | 82.3 |
| Contr_l + CrossE_L | SAGE | 54.49 ± 0.24 | ± | 54.37 ± 0.55 | ± | 58.26 ± 0.29 | ± | 36.3 |
| Contr_l + CrossE_L + PMI_L | ALL | 56.97 ± 0.35 | ± | 56.71 ± 0.83 | ± | 59.73 ± 0.28 | ± | 89.7 |
| Contr_l + CrossE_L + PMI_L | GAT | 56.03 ± 0.59 | ± | 54.58 ± 0.22 | ± | 58.55 ± 0.14 | ± | 59.0 |
| Contr_l + CrossE_L + PMI_L | GCN | 55.60 ± 0.11 | ± | 54.19 ± 0.06 | ± | 58.05 ± 0.17 | ± | 40.3 |
| Contr_l + CrossE_L + PMI_L | GIN | 57.72 ± 1.53 | ± | 56.06 ± 0.55 | ± | 65.75 ± 1.39 | ± | 109.0 |
| Contr_l + CrossE_L + PMI_L | MPNN | 56.43 ± 1.15 | ± | 53.62 ± 0.18 | ± | 58.78 ± 0.37 | ± | 48.0 |
| Contr_l + CrossE_L + PMI_L | PAGNN | 64.70 ± 0.66 | ± | 58.26 ± 0.23 | ± | 70.42 ± 0.33 | ± | 146.7 |
| Contr_l + CrossE_L + PMI_L | SAGE | 66.00 ± 0.13 | ± | 64.22 ± 0.96 | ± | 65.23 ± 1.54 | ± | 142.7 |
| Contr_l + CrossE_L + PMI_L + PR_L | ALL | 61.62 ± 1.10 | ± | 58.39 ± 0.81 | ± | 70.18 ± 0.36 | ± | 138.7 |
| Contr_l + CrossE_L + PMI_L + PR_L | GAT | 56.16 ± 0.34 | ± | 54.54 ± 0.24 | ± | 62.85 ± 1.64 | ± | 77.3 |
| Contr_l + CrossE_L + PMI_L + PR_L | GCN | 55.26 ± 0.38 | ± | 54.24 ± 0.12 | ± | 58.41 ± 0.16 | ± | 44.0 |
| Contr_l + CrossE_L + PMI_L + PR_L | GIN | 57.62 ± 1.42 | ± | 56.63 ± 0.59 | ± | 66.79 ± 0.92 | ± | 114.3 |

Graph Reconstruction Bce Loss Continued (↓)

| Loss Type | Model | CORA | | Citeseer | | Bitcoin Fraud Transaction | | Average Rank |
|---|---|---|---|---|---|---|---|---|
| Contr_l + CrossE_L + PMI_L + PR_L | MPNN | 57.21 1.08 | ± | 54.26 0.56 | ± | 64.96 3.40 | ± | 85.7 |
| Contr_l + CrossE_L + PMI_L + PR_L | PAGNN | 64.34 0.51 | ± | 58.23 0.41 | ± | 71.93 0.61 | ± | 148.7 |
| Contr_l + CrossE_L + PMI_L + PR_L | SAGE | 66.27 0.63 | ± | 63.41 2.32 | ± | 69.43 0.39 | ± | 155.0 |
| Contr_l + CrossE_L + PMI_L + PR_L + Triplet_L | ALL | 56.04 0.55 | ± | 55.67 0.27 | ± | 61.30 0.36 | ± | 82.3 |
| Contr_l + CrossE_L + PMI_L + PR_L + Triplet_L | GAT | 55.82 0.20 | ± | 54.85 1.05 | ± | 60.06 1.77 | ± | 69.3 |
| Contr_l + CrossE_L + PMI_L + PR_L + Triplet_L | GCN | 55.42 0.26 | ± | 53.97 0.25 | ± | 58.57 0.20 | ± | 42.3 |
| Contr_l + CrossE_L + PMI_L + PR_L + Triplet_L | GIN | 58.13 1.09 | ± | 55.35 0.31 | ± | 64.20 0.88 | ± | 102.3 |
| Contr_l + CrossE_L + PMI_L + PR_L + Triplet_L | MPNN | 57.63 0.75 | ± | 54.15 0.03 | ± | 59.52 1.42 | ± | 69.3 |
| Contr_l + CrossE_L + PMI_L + PR_L + Triplet_L | PAGNN | 63.62 0.82 | ± | 58.00 0.09 | ± | 69.88 0.10 | ± | 139.0 |
| Contr_l + CrossE_L + PMI_L + PR_L + Triplet_L | SAGE | 58.88 1.20 | ± | 57.09 0.11 | ± | 65.73 0.55 | ± | 118.7 |
| Contr_l + CrossE_L + PMI_L + Triplet_L | ALL | 54.97 0.34 | ± | 54.37 0.22 | ± | 57.92 0.14 | ± | 38.0 |
| Contr_l + CrossE_L + PMI_L + Triplet_L | GAT | 56.16 0.36 | ± | 54.39 0.16 | ± | 58.21 0.11 | ± | 54.0 |
| Contr_l + CrossE_L + PMI_L + Triplet_L | GCN | 55.18 0.26 | ± | 54.26 0.31 | ± | 58.12 0.18 | ± | 40.0 |
| Contr_l + CrossE_L + PMI_L + Triplet_L | GIN | 56.86 0.77 | ± | 55.23 0.18 | ± | 63.30 0.55 | ± | 89.7 |
| Contr_l + CrossE_L + PMI_L + Triplet_L | MPNN | 57.42 0.70 | ± | 53.56 0.29 | ± | 58.62 0.26 | ± | 51.7 |
| Contr_l + CrossE_L + PMI_L + Triplet_L | PAGNN | 63.22 0.64 | ± | 58.18 0.40 | ± | 69.98 0.37 | ± | 138.3 |

Graph Reconstruction Bce Loss Continued (↓)

| Loss Type | Model | CORA | | Citeseer | | Bitcoin Fraud Transaction | | Average Rank |
|---|---|---|---|---|---|---|---|---|
| Contr_l + CrossE_L + PMI_L + Triplet_L | SAGE | 58.85 ± 1.09 | | 56.70 ± 0.22 | | 60.31 ± 0.36 | | 101.3 |
| Contr_l + CrossE_L + PR_L | ALL | 79.63 ± 0.60 | | 71.95 ± 4.43 | | 84.31 ± 0.62 | | 197.7 |
| Contr_l + CrossE_L + PR_L | GAT | 70.49 ± 5.62 | | 73.69 ± 5.90 | | 79.86 ± 0.64 | | 183.7 |
| Contr_l + CrossE_L + PR_L | GCN | 66.21 ± 1.00 | | 68.20 ± 1.04 | | 65.07 ± 0.87 | | 147.0 |
| Contr_l + CrossE_L + PR_L | GIN | 76.89 ± 2.72 | | 69.07 ± 3.42 | | 82.10 ± 0.42 | | 187.0 |
| Contr_l + CrossE_L + PR_L | MPNN | 73.02 ± 1.38 | | 72.94 ± 1.81 | | 64.03 ± 2.82 | | 153.7 |
| Contr_l + CrossE_L + PR_L | PAGNN | 77.40 ± 0.53 | | 72.58 ± 2.11 | | 68.64 ± 0.63 | | 169.3 |
| Contr_l + CrossE_L + PR_L | SAGE | 69.07 ± 5.82 | | 69.51 ± 3.84 | | 73.21 ± 4.79 | | 173.7 |
| Contr_l + CrossE_L + PR_L + Triplet_L | ALL | 64.53 ± 0.83 | | 59.46 ± 0.59 | | 64.93 ± 0.15 | | 131.3 |
| Contr_l + CrossE_L + PR_L + Triplet_L | GAT | 62.27 ± 0.60 | | 61.84 ± 0.89 | | 66.59 ± 2.91 | | 133.7 |
| Contr_l + CrossE_L + PR_L + Triplet_L | GCN | 60.94 ± 0.83 | | 60.57 ± 0.43 | | 61.00 ± 2.92 | | 114.7 |
| Contr_l + CrossE_L + PR_L + Triplet_L | GIN | 62.56 ± 1.01 | | 61.52 ± 0.53 | | 68.20 ± 5.33 | | 137.0 |
| Contr_l + CrossE_L + PR_L + Triplet_L | MPNN | 64.57 ± 1.12 | | 62.92 ± 0.36 | | 60.67 ± 1.00 | | 126.3 |
| Contr_l + CrossE_L + PR_L + Triplet_L | PAGNN | 70.74 ± 1.25 | | 62.72 ± 2.41 | | 68.17 ± 0.24 | | 154.0 |
| Contr_l + CrossE_L + PR_L + Triplet_L | SAGE | 62.88 ± 0.15 | | 62.68 ± 0.95 | | 65.14 ± 0.72 | | 132.0 |
| Contr_l + CrossE_L + Triplet_L | ALL | 54.30 ± 0.40 | | 54.11 ± 0.45 | | 57.28 ± 0.24 | | 17.7 |

Graph Reconstruction Bce Loss Continued (↓)

| Loss Type | Model | CORA | Citeseer | Bitcoin Fraud Transaction | Average Rank |
|---|---|---|---|---|---|
| Contr_l + CrossE_L + Triplet_L | GAT | 54.47 ± 0.48 | 53.69 ± 0.32 | 57.16 ± 0.17 | 13.3 |
| Contr_l + CrossE_L + Triplet_L | GCN | 54.44 ± 0.24 | 53.69 ± 0.34 | 57.96 ± 0.46 | 18.7 |
| Contr_l + CrossE_L + Triplet_L | GIN | 54.57 ± 0.38 | 53.92 ± 0.43 | 59.71 ± 0.98 | 41.0 |
| Contr_l + CrossE_L + Triplet_L | MPNN | 54.54 ± 0.54 | 53.61 ± 0.49 | 57.86 ± 0.17 | 19.3 |
| Contr_l + CrossE_L + Triplet_L | PAGNN | 55.99 ± 1.74 | 54.75 ± 0.24 | 70.66 ± 0.34 | 101.3 |
| Contr_l + CrossE_L + Triplet_L | SAGE | 54.80 ± 0.36 | 54.06 ± 0.29 | 58.36 ± 0.15 | 34.0 |
| Contr_l + PMI_L | ALL | 56.48 ± 0.44 | 55.62 ± 0.26 | 59.26 ± 0.25 | 77.3 |
| Contr_l + PMI_L | GAT | 56.28 ± 0.35 | 54.45 ± 0.22 | 58.56 ± 0.22 | 60.3 |
| Contr_l + PMI_L | GCN | 55.62 ± 0.30 | 54.13 ± 0.19 | 58.17 ± 0.22 | 41.0 |
| Contr_l + PMI_L | GIN | 57.79 ± 0.63 | 55.60 ± 0.22 | 64.66 ± 0.71 | 102.3 |
| Contr_l + PMI_L | MPNN | 57.36 ± 0.92 | 53.53 ± 0.16 | 58.94 ± 0.18 | 52.3 |
| Contr_l + PMI_L | PAGNN | 64.23 ± 0.95 | 58.25 ± 0.26 | 70.29 ± 0.33 | 144.7 |
| Contr_l + PMI_L | SAGE | 65.74 ± 0.61 | 60.57 ± 1.39 | 65.83 ± 0.49 | 139.3 |
| Contr_l + PMI_L + PR_L | ALL | 59.83 ± 0.61 | 58.46 ± 0.51 | 69.30 ± 0.54 | 134.7 |
| Contr_l + PMI_L + PR_L | GAT | 56.08 ± 0.11 | 55.53 ± 0.92 | 63.82 ± 1.90 | 87.3 |
| Contr_l + PMI_L + PR_L | GCN | 55.33 ± 0.34 | 54.12 ± 0.36 | 58.43 ± 0.52 | 42.0 |

Graph Reconstruction Bce Loss Continued (↓)

| Loss Type | Model | CORA | | | Citeseer | | | Bitcoin Fraud Transaction | | | Average Rank |
|---|---|---|---|---|---|---|---|---|---|---|---|
| Contr_l + PMI_L + PR_L | GIN | 57.78 | ± | | 56.44 | ± | | 66.38 | ± | | 113.0 |
| | | 1.12 | | | 0.83 | | | 0.77 | | | |
| Contr_l + PMI_L + PR_L | MPNN | 56.97 | ± | | 54.93 | ± | | 64.98 | ± | | 94.3 |
| | | 1.08 | | | 0.61 | | | 2.88 | | | |
| Contr_l + PMI_L + PR_L | PAGNN | 64.73 | ± | | 58.52 | ± | | 71.46 | ± | | 151.3 |
| | | 0.59 | | | 0.34 | | | 0.47 | | | |
| Contr_l + PMI_L + PR_L | SAGE | 65.71 | ± | | 59.82 | ± | | 69.14 | ± | | 147.0 |
| | | 0.60 | | | 1.58 | | | 0.61 | | | |
| Contr_l + PMI_L + PR_L + Triplet_L | ALL | 55.73 | ± | | 55.35 | ± | | 60.04 | ± | | 71.7 |
| | | 0.36 | | | 0.16 | | | 0.18 | | | |
| Contr_l + PMI_L + PR_L + Triplet_L | GAT | 56.07 | ± | | 56.19 | ± | | 62.26 | ± | | 86.3 |
| | | 0.74 | | | 1.28 | | | 1.26 | | | |
| Contr_l + PMI_L + PR_L + Triplet_L | GCN | 55.55 | ± | | 54.56 | ± | | 58.77 | ± | | 57.3 |
| | | 0.26 | | | 0.94 | | | 0.54 | | | |
| Contr_l + PMI_L + PR_L + Triplet_L | GIN | 56.03 | ± | | 55.63 | ± | | 63.00 | ± | | 85.0 |
| | | 0.57 | | | 0.31 | | | 0.41 | | | |
| Contr_l + PMI_L + PR_L + Triplet_L | MPNN | 56.79 | ± | | 54.77 | ± | | 60.41 | ± | | 79.0 |
| | | 1.36 | | | 0.23 | | | 0.75 | | | |
| Contr_l + PMI_L + PR_L + Triplet_L | PAGNN | 63.11 | ± | | 57.79 | ± | | 68.61 | ± | | 131.0 |
| | | 0.83 | | | 0.56 | | | 0.47 | | | |
| Contr_l + PMI_L + PR_L + Triplet_L | SAGE | 56.93 | ± | | 56.05 | ± | | 62.77 | ± | | 94.0 |
| | | 0.38 | | | 0.45 | | | 1.04 | | | |
| Contr_l + PR_L | ALL | 80.05 | ± | | 74.43 | ± | | 84.15 | ± | | 201.0 |
| | | 0.47 | | | 5.43 | | | 0.90 | | | |
| Contr_l + PR_L | GAT | 74.54 | ± | | 76.05 | ± | | 79.85 | ± | | 189.0 |
| | | 6.67 | | | 1.40 | | | 0.41 | | | |
| Contr_l + PR_L | GCN | 66.92 | ± | | 66.03 | ± | | 61.80 | ± | | 138.0 |
| | | 1.27 | | | 2.59 | | | 4.01 | | | |
| Contr_l + PR_L | GIN | 76.28 | ± | | 69.37 | ± | | 82.67 | ± | | 188.0 |
| | | 0.77 | | | 2.94 | | | 0.66 | | | |
| Contr_l + PR_L | MPNN | 74.59 | ± | | 74.00 | ± | | 66.47 | ± | | 163.3 |
| | | 2.29 | | | 1.70 | | | 0.60 | | | |

Graph Reconstruction Bce Loss Continued (↓)

| Loss Type | Model | CORA | | Citeseer | | Bitcoin Fraud Transaction | | Average Rank |
|---|---|---|---|---|---|---|---|---|
| Contr_l + PR_L | PAGNN | 77.50 | ± 0.54 | 73.15 | ± 2.25 | 68.69 | ± 0.52 | 171.3 |
| Contr_l + PR_L | SAGE | 68.45 | ± 5.84 | 67.60 | ± 5.63 | 71.78 | ± 4.75 | 169.0 |
| Contr_l + PR_L + Triplet_L | ALL | 66.76 | ± 1.73 | 61.69 | ± 1.69 | 65.30 | ± 0.45 | 140.7 |
| Contr_l + PR_L + Triplet_L | GAT | 63.48 | ± 1.66 | 62.31 | ± 0.85 | 67.24 | ± 2.64 | 139.0 |
| Contr_l + PR_L + Triplet_L | GCN | 61.73 | ± 0.71 | 60.84 | ± 0.63 | 62.57 | ± 3.18 | 119.7 |
| Contr_l + PR_L + Triplet_L | GIN | 63.16 | ± 0.75 | 62.34 | ± 0.93 | 69.28 | ± 3.11 | 143.3 |
| Contr_l + PR_L + Triplet_L | MPNN | 65.43 | ± 0.85 | 63.87 | ± 0.92 | 61.47 | ± 0.92 | 130.7 |
| Contr_l + PR_L + Triplet_L | PAGNN | 72.16 | ± 0.95 | 62.49 | ± 2.88 | 68.54 | ± 0.34 | 155.3 |
| Contr_l + PR_L + Triplet_L | SAGE | 63.47 | ± 0.20 | 62.57 | ± 0.34 | 64.97 | ± 0.41 | 132.0 |
| Contr_l + Triplet_L | ALL | 54.37 | ± 0.23 | 53.79 | ± 0.20 | 57.30 | ± 0.20 | 16.0 |
| Contr_l + Triplet_L | GAT | 54.23 | ± 0.26 | 53.54 | ± 0.35 | 57.35 | ± 0.20 | 7.3 |
| Contr_l + Triplet_L | GCN | 54.43 | ± 0.15 | 53.66 | ± 0.32 | 58.16 | ± 0.44 | 19.0 |
| Contr_l + Triplet_L | GIN | 54.48 | ± 0.05 | 53.89 | ± 0.22 | 60.57 | ± 1.06 | 42.0 |
| Contr_l + Triplet_L | MPNN | 54.32 | ± 0.19 | 53.54 | ± 0.40 | 57.85 | ± 0.31 | 11.7 |
| Contr_l + Triplet_L | PAGNN | 56.49 | ± 2.61 | 54.81 | ± 0.27 | 70.71 | ± 0.38 | 108.7 |
| Contr_l + Triplet_L | SAGE | 54.74 | ± 0.16 | 54.02 | ± 0.31 | 58.64 | ± 0.27 | 37.7 |

Graph Reconstruction Bce Loss Continued (↓)

| Loss Type | Model | CORA | | | Citeseer | | | Bitcoin Fraud Transaction | | | Average Rank |
|---|---|---|---|---|---|---|---|---|---|---|---|
| CrossE_L | ALL | 79.23 | ± | 0.14 | 79.19 | ± | 0.14 | 78.98 | ± | 0.12 | 198.0 |
| CrossE_L | GAT | 78.92 | ± | 0.15 | 78.97 | ± | 0.17 | 79.14 | ± | 0.20 | 196.3 |
| CrossE_L | GCN | 78.51 | ± | 0.25 | 78.35 | ± | 0.17 | 78.38 | ± | 0.13 | 193.0 |
| CrossE_L | GIN | 78.94 | ± | 0.19 | 78.78 | ± | 0.17 | 78.73 | ± | 0.35 | 195.3 |
| CrossE_L | MPNN | 79.01 | ± | 0.09 | 79.06 | ± | 0.07 | 79.17 | ± | 0.13 | 198.3 |
| CrossE_L | PAGNN | 79.31 | ± | 0.14 | 79.52 | ± | 0.03 | 79.60 | ± | 0.13 | 201.0 |
| CrossE_L | SAGE | 78.91 | ± | 0.06 | 79.01 | ± | 0.15 | 78.97 | ± | 0.20 | 195.3 |
| CrossE_L + PMI_L | ALL | 57.25 | ± | 0.51 | 56.52 | ± | 0.47 | 60.36 | ± | 0.48 | 93.0 |
| CrossE_L + PMI_L | GAT | 56.90 | ± | 0.41 | 54.73 | ± | 0.31 | 59.03 | ± | 0.38 | 73.3 |
| CrossE_L + PMI_L | GCN | 55.38 | ± | 0.18 | 54.33 | ± | 0.24 | 58.38 | ± | 0.45 | 46.7 |
| CrossE_L + PMI_L | GIN | 57.85 | ± | 1.32 | 56.22 | ± | 0.41 | 66.80 | ± | 0.29 | 114.7 |
| CrossE_L + PMI_L | MPNN | 58.03 | ± | 0.73 | 53.67 | ± | 0.22 | 59.00 | ± | 0.43 | 62.0 |
| CrossE_L + PMI_L | PAGNN | 63.75 | ± | 0.67 | 58.44 | ± | 0.20 | 70.54 | ± | 0.55 | 145.3 |
| CrossE_L + PMI_L | SAGE | 66.93 | ± | 0.54 | 65.51 | ± | 0.49 | 69.48 | ± | 0.51 | 158.7 |
| CrossE_L + PMI_L + PR_L | ALL | 61.49 | ± | 0.17 | 58.68 | ± | 0.31 | 71.46 | ± | 0.30 | 144.0 |
| CrossE_L + PMI_L + PR_L | GAT | 56.42 | ± | 0.56 | 55.52 | ± | 1.19 | 63.30 | ± | 2.59 | 88.3 |

Graph Reconstruction Bce Loss Continued (↓)

| Loss Type | Model | CORA | | Citeseer | | Bitcoin Fraud Transaction | | Average Rank |
|---|---|---|---|---|---|---|---|---|
| CrossE_L + PMI_L + PR_L | GCN | 55.61 ± 0.32 | | 54.24 ± 0.29 | | 58.26 ± 0.52 | | 45.3 |
| CrossE_L + PMI_L + PR_L | GIN | 57.88 ± 0.71 | | 56.82 ± 0.79 | | 67.49 ± 0.69 | | 120.0 |
| CrossE_L + PMI_L + PR_L | MPNN | 57.94 ± 1.32 | | 54.93 ± 0.59 | | 63.22 ± 4.70 | | 96.7 |
| CrossE_L + PMI_L + PR_L | PAGNN | 63.96 ± 1.12 | | 58.52 ± 0.23 | | 72.64 ± 1.05 | | 150.7 |
| CrossE_L + PMI_L + PR_L | SAGE | 66.05 ± 0.46 | | 65.02 ± 0.50 | | 70.26 ± 0.40 | | 159.0 |
| CrossE_L + PMI_L + PR_L + Triplet_L | ALL | 55.82 ± 0.28 | | 55.95 ± 0.12 | | 61.76 ± 0.13 | | 82.0 |
| CrossE_L + PMI_L + PR_L + Triplet_L | GAT | 56.22 ± 0.39 | | 55.05 ± 0.84 | | 60.98 ± 2.03 | | 78.3 |
| CrossE_L + PMI_L + PR_L + Triplet_L | GCN | 55.51 ± 0.22 | | 54.15 ± 0.11 | | 58.28 ± 0.38 | | 42.0 |
| CrossE_L + PMI_L + PR_L + Triplet_L | GIN | 57.34 ± 0.82 | | 55.66 ± 0.34 | | 64.01 ± 0.46 | | 98.7 |
| CrossE_L + PMI_L + PR_L + Triplet_L | MPNN | 57.51 ± 0.89 | | 54.27 ± 0.49 | | 59.62 ± 1.51 | | 72.7 |
| CrossE_L + PMI_L + PR_L + Triplet_L | PAGNN | 63.59 ± 0.60 | | 58.21 ± 0.28 | | 69.81 ± 0.13 | | 139.3 |
| CrossE_L + PMI_L + PR_L + Triplet_L | SAGE | 59.84 ± 2.88 | | 57.44 ± 0.75 | | 66.70 ± 0.55 | | 123.3 |
| CrossE_L + PMI_L + Triplet_L | ALL | 54.41 ± 0.32 | | 53.71 ± 0.30 | | 57.66 ± 0.33 | | 15.7 |
| CrossE_L + PMI_L + Triplet_L | GAT | 56.15 ± 0.64 | | 54.34 ± 0.23 | | 58.63 ± 0.45 | | 57.7 |
| CrossE_L + PMI_L + Triplet_L | GCN | 55.33 ± 0.22 | | 54.00 ± 0.11 | | 58.29 ± 0.18 | | 38.3 |
| CrossE_L + PMI_L + Triplet_L | GIN | 56.45 ± 0.55 | | 55.13 ± 0.28 | | 63.02 ± 0.79 | | 85.7 |

Graph Reconstruction Bce Loss Continued (↓)

| Loss Type | Model | CORA | Citeseer | Bitcoin Fraud Transaction | Average Rank |
|---|---|---|---|---|---|
| CrossE_L + PMI_L + Triplet_L | MPNN | 56.46 ± 0.41 | 53.38 ± 0.28 | 58.38 ± 0.20 | 38.7 |
| CrossE_L + PMI_L + Triplet_L | PAGNN | 62.42 ± 0.95 | 58.14 ± 0.24 | 69.76 ± 0.30 | 135.0 |
| CrossE_L + PMI_L + Triplet_L | SAGE | 57.17 ± 0.30 | 56.38 ± 0.46 | 59.81 ± 0.23 | 89.7 |
| CrossE_L + PR_L | ALL | 80.65 ± 0.35 | 78.24 ± 0.93 | 82.94 ± 0.81 | 204.3 |
| CrossE_L + PR_L | GAT | 77.85 ± 3.49 | 79.89 ± 0.42 | 79.78 ± 1.32 | 198.0 |
| CrossE_L + PR_L | GCN | 70.62 ± 1.33 | 72.58 ± 1.21 | 77.74 ± 1.27 | 178.7 |
| CrossE_L + PR_L | GIN | 78.47 ± 1.44 | 74.87 ± 4.03 | 81.13 ± 0.22 | 194.0 |
| CrossE_L + PR_L | MPNN | 78.14 ± 1.14 | 74.40 ± 1.19 | 70.75 ± 2.04 | 181.0 |
| CrossE_L + PR_L | PAGNN | 78.48 ± 0.78 | 77.00 ± 1.80 | 73.73 ± 1.44 | 189.3 |
| CrossE_L + PR_L | SAGE | 79.33 ± 0.53 | 78.25 ± 1.51 | 80.01 ± 0.55 | 200.0 |
| CrossE_L + PR_L + Triplet_L | ALL | 70.39 ± 2.28 | 66.94 ± 0.93 | 75.47 ± 7.57 | 173.0 |
| CrossE_L + PR_L + Triplet_L | GAT | 69.10 ± 1.98 | 65.59 ± 1.74 | 75.18 ± 3.14 | 171.0 |
| CrossE_L + PR_L + Triplet_L | GCN | 64.07 ± 1.16 | 63.88 ± 0.46 | 62.76 ± 3.72 | 130.0 |
| CrossE_L + PR_L + Triplet_L | GIN | 67.30 ± 2.58 | 65.51 ± 1.01 | 81.27 ± 0.57 | 175.7 |
| CrossE_L + PR_L + Triplet_L | MPNN | 71.74 ± 1.24 | 68.53 ± 1.29 | 63.52 ± 1.85 | 149.0 |
| CrossE_L + PR_L + Triplet_L | PAGNN | 74.92 ± 1.24 | 66.54 ± 1.46 | 68.26 ± 0.40 | 161.7 |

Graph Reconstruction Bce Loss Continued (↓)

| Loss Type | Model | CORA | | Citeseer | | Bitcoin Fraud Transaction | | Average Rank |
|-----------|-------|------|---|----------|---|---------------------------|---|--------------|
| CrossE_L + PR_L + Triplet_L | SAGE | 64.83 ± 0.64 | | 64.90 ± 1.08 | | 69.05 ± 1.92 | | 152.0 |
| CrossE_L + Triplet_L | ALL | 54.51 ± 0.60 | | 53.75 ± 0.36 | | 57.97 ± 0.28 | | 23.0 |
| CrossE_L + Triplet_L | GAT | 55.00 ± 0.12 | | 54.15 ± 0.17 | | 59.75 ± 0.94 | | 48.7 |
| CrossE_L + Triplet_L | GCN | 54.58 ± 0.09 | | 53.78 ± 0.14 | | 59.80 ± 0.26 | | 41.0 |
| CrossE_L + Triplet_L | GIN | 55.06 ± 0.40 | | 54.71 ± 0.29 | | 66.02 ± 2.24 | | 78.7 |
| CrossE_L + Triplet_L | MPNN | 54.50 ± 0.27 | | 53.60 ± 0.40 | | 58.88 ± 0.32 | | 29.3 |
| CrossE_L + Triplet_L | PAGNN | 56.77 ± 2.79 | | 54.64 ± 0.30 | | 72.96 ± 0.24 | | 111.0 |
| CrossE_L + Triplet_L | SAGE | 55.16 ± 0.22 | | 54.22 ± 0.18 | | 61.60 ± 0.16 | | 59.0 |
| PMI_L | ALL | 58.29 ± 0.21 | | 56.34 ± 0.61 | | 60.19 ± 0.23 | | 97.3 |
| PMI_L | GAT | 56.57 ± 0.32 | | 54.68 ± 0.34 | | 58.89 ± 0.42 | | 69.3 |
| PMI_L | GCN | 55.15 ± 0.20 | | 54.35 ± 0.37 | | 58.42 ± 0.43 | | 45.7 |
| PMI_L | GIN | 57.35 ± 0.83 | | 56.26 ± 0.45 | | 66.96 ± 1.08 | | 111.7 |
| PMI_L | MPNN | 57.13 ± 0.85 | | 53.70 ± 0.25 | | 59.10 ± 0.32 | | 57.0 |
| PMI_L | PAGNN | 64.08 ± 0.74 | | 58.28 ± 0.23 | | 70.82 ± 0.27 | | 147.3 |
| PMI_L | SAGE | 67.79 ± 0.87 | | 65.73 ± 0.60 | | 69.00 ± 0.43 | | 158.0 |
| PMI_L + PR_L | ALL | 61.09 ± 0.73 | | 58.79 ± 0.59 | | 71.75 ± 0.28 | | 144.3 |

Graph Reconstruction Bce Loss Continued (↓)

| Loss Type | Model | CORA | | Citeseer | | Bitcoin Fraud Transaction | | Average Rank |
|---|---|---|---|---|---|---|---|---|
| PMI_L + PR_L | GAT | 56.32 | ± | 56.34 | ± | 66.45 | ± | 102.3 |
| | | 0.22 | | 1.56 | | 0.38 | | |
| PMI_L + PR_L | GCN | 55.73 | ± | 54.28 | ± | 58.65 | ± | 53.7 |
| | | 0.36 | | 0.28 | | 0.38 | | |
| PMI_L + PR_L | GIN | 58.45 | ± | 56.66 | ± | 67.59 | ± | 121.3 |
| | | 0.44 | | 0.86 | | 0.30 | | |
| PMI_L + PR_L | MPNN | 57.80 | ± | 55.24 | ± | 66.59 | ± | 107.3 |
| | | 0.77 | | 0.32 | | 4.02 | | |
| PMI_L + PR_L | PAGNN | 65.06 | ± | 58.50 | ± | 73.57 | ± | 155.0 |
| | | 0.98 | | 0.25 | | 0.53 | | |
| PMI_L + PR_L | SAGE | 66.15 | ± | 64.84 | ± | 70.57 | ± | 160.0 |
| | | 0.75 | | 0.37 | | 0.52 | | |
| PMI_L + PR_L + Triplet_L | ALL | 56.00 | ± | 55.68 | ± | 61.44 | ± | 81.7 |
| | | 0.72 | | 0.29 | | 0.36 | | |
| PMI_L + PR_L + Triplet_L | GAT | 55.89 | ± | 55.24 | ± | 62.71 | ± | 79.3 |
| | | 0.26 | | 0.50 | | 1.83 | | |
| PMI_L + PR_L + Triplet_L | GCN | 55.30 | ± | 54.11 | ± | 58.65 | ± | 44.3 |
| | | 0.36 | | 0.26 | | 0.59 | | |
| PMI_L + PR_L + Triplet_L | GIN | 57.62 | ± | 55.63 | ± | 64.36 | ± | 101.7 |
| | | 0.71 | | 0.37 | | 0.88 | | |
| PMI_L + PR_L + Triplet_L | MPNN | 57.91 | ± | 54.61 | ± | 60.26 | ± | 84.7 |
| | | 0.49 | | 0.16 | | 1.73 | | |
| PMI_L + PR_L + Triplet_L | PAGNN | 63.61 | ± | 58.15 | ± | 69.71 | ± | 138.3 |
| | | 0.53 | | 0.26 | | 0.22 | | |
| PMI_L + PR_L + Triplet_L | SAGE | 59.10 | ± | 56.93 | ± | 65.53 | ± | 118.0 |
| | | 2.52 | | 0.21 | | 0.51 | | |
| PMI_L + Triplet_L | ALL | 54.77 | ± | 53.72 | ± | 57.76 | ± | 24.0 |
| | | 0.25 | | 0.19 | | 0.10 | | |
| PMI_L + Triplet_L | GAT | 56.02 | ± | 54.47 | ± | 58.38 | ± | 55.0 |
| | | 0.38 | | 0.24 | | 0.37 | | |
| PMI_L + Triplet_L | GCN | 55.32 | ± | 54.20 | ± | 58.17 | ± | 40.0 |
| | | 0.23 | | 0.30 | | 0.19 | | |

Graph Reconstruction Bce Loss Continued (↓)

| Loss Type | Model | CORA | | Citeseer | | Bitcoin Fraud Transaction | | Average Rank |
|---|---|---|---|---|---|---|---|---|
| PMI_L + Triplet_L | GIN | 57.14 ± 0.62 | | 55.31 ± 0.38 | | 63.43 ± 0.60 | | 93.3 |
| PMI_L + Triplet_L | MPNN | 56.81 ± 0.61 | | 53.57 ± 0.32 | | 58.44 ± 0.14 | | 46.3 |
| PMI_L + Triplet_L | PAGNN | 63.45 ± 0.45 | | 58.22 ± 0.28 | | 70.00 ± 0.21 | | 139.7 |
| PMI_L + Triplet_L | SAGE | 58.28 ± 0.53 | | 56.68 ± 0.21 | | 60.50 ± 0.14 | | 101.3 |
| PR_L | ALL | 80.91 ± 0.45 | | 79.58 ± 0.78 | | 83.11 ± 0.91 | | 208.7 |
| PR_L | GAT | 80.18 ± 0.54 | | 79.27 ± 0.94 | | 80.36 ± 1.53 | | 204.3 |
| PR_L | GCN | 71.39 ± 2.23 | | 71.14 ± 0.85 | | 79.18 ± 0.34 | | 181.0 |
| PR_L | GIN | 78.93 ± 1.39 | | 75.25 ± 1.78 | | 81.30 ± 1.07 | | 197.0 |
| PR_L | MPNN | 76.12 ± 1.32 | | 74.85 ± 1.01 | | 65.66 ± 2.75 | | 163.3 |
| PR_L | PAGNN | 78.59 ± 0.92 | | 77.53 ± 0.67 | | 72.92 ± 0.87 | | 189.0 |
| PR_L | SAGE | 79.52 ± 0.60 | | 79.35 ± 0.41 | | 79.61 ± 0.28 | | 201.7 |
| PR_L + Triplet_L | ALL | 79.66 ± 1.03 | | 77.30 ± 0.91 | | 83.19 ± 1.26 | | 203.3 |
| PR_L + Triplet_L | GAT | 70.49 ± 4.92 | | 76.20 ± 1.49 | | 78.99 ± 2.66 | | 184.7 |
| PR_L + Triplet_L | GCN | 67.94 ± 2.32 | | 68.01 ± 1.80 | | 66.03 ± 1.51 | | 152.7 |
| PR_L + Triplet_L | GIN | 76.00 ± 1.44 | | 69.11 ± 1.64 | | 82.21 ± 0.28 | | 186.7 |
| PR_L + Triplet_L | MPNN | 74.94 ± 2.01 | | 74.82 ± 2.21 | | 64.27 ± 3.25 | | 158.0 |

Graph Reconstruction Bce Loss Continued (↓)

| Loss Type | Model | CORA | | Citeseer | | Bitcoin Fraud Transaction | | Average Rank |
|---|---|---|---|---|---|---|---|---|
| PR_L + Triplet_L | PAGNN | 77.28 ± 0.34 | | 75.59 ± 0.67 | | 69.15 ± 0.43 | | 175.3 |
| PR_L + Triplet_L | SAGE | 78.32 ± 1.29 | | 73.65 ± 5.64 | | 77.47 ± 3.23 | | 185.7 |
| Triplet_L | ALL | 54.20 ± 0.30 | | 53.54 ± 0.33 | | 57.61 ± 0.13 | | 7.7 |
| Triplet_L | GAT | 55.02 ± 0.17 | | 54.20 ± 0.29 | | 59.38 ± 0.68 | | 48.0 |
| Triplet_L | GCN | 54.54 ± 0.20 | | 53.71 ± 0.08 | | 60.41 ± 0.53 | | 41.7 |
| Triplet_L | GIN | 55.06 ± 0.26 | | 54.36 ± 0.32 | | 65.07 ± 0.38 | | 71.0 |
| Triplet_L | MPNN | 54.56 ± 0.25 | | 53.35 ± 0.26 | | 59.43 ± 0.35 | | 29.3 |
| Triplet_L | PAGNN | 56.63 ± 3.04 | | 54.41 ± 0.12 | | 73.10 ± 0.28 | | 108.0 |
| Triplet_L | SAGE | 55.05 ± 0.17 | | 54.20 ± 0.19 | | 61.07 ± 0.20 | | 56.0 |

### 1.1.4 Evaluation on Clustering Quality Tasks.

Table 16. Silhouette Performance (↑): Top-ranked results are highlighted in **1st**, second-ranked in **2nd**, and third-ranked in **3rd**.

| Loss Type | Model | CORA | | Citeseer | | Bitcoin Fraud Transaction | | Average Rank |
|---|---|---|---|---|---|---|---|---|
| Contr_l | ALL | 13.75 ± 1.88 | | 5.39 ± 0.84 | | -2.96 ± 0.72 | | 29.3 |
| Contr_l | GAT | 12.59 ± 0.76 | | 4.15 ± 0.19 | | -1.39 ± 0.12 | | 29.7 |

Continued on next page

Silhouette Continued (↑)

| Loss Type | Model | CORA | | Citeseer | | Bitcoin Fraud Transaction | | Average Rank |
|---|---|---|---|---|---|---|---|---|
| Contr_l | GCN | 7.52 ± 0.69 | | 3.01 ± 1.16 | | -1.09 ± 0.78 | | 42.0 |
| Contr_l | GIN | 15.02 ± 1.24 | | 4.83 ± 0.72 | | -3.61 ± 1.11 | | 32.7 |
| Contr_l | MPNN | 13.47 ± 1.41 | | 5.35 ± 0.98 | | -2.63 ± 0.43 | | 29.0 |
| Contr_l | PAGNN | 13.63 ± 1.32 | | 5.57 ± 1.11 | | -19.34 ± 0.67 | | 69.0 |
| Contr_l | SAGE | 14.02 ± 1.65 | | 4.26 ± 1.34 | | -4.11 ± 1.34 | | 37.7 |
| Contr_l + CrossE_L | ALL | 12.46 ± 3.07 | | 5.24 ± 1.79 | | -3.03 ± 1.38 | | 33.3 |
| Contr_l + CrossE_L | GAT | 11.94 ± 0.95 | | 4.18 ± 0.23 | | -1.56 ± 0.20 | | 31.0 |
| Contr_l + CrossE_L | GCN | 9.42 ± 1.26 | | 2.28 ± 0.48 | | -1.34 ± 0.47 | | 42.7 |
| Contr_l + CrossE_L | GIN | 14.17 ± 0.88 | | 4.65 ± 0.45 | | -3.05 ± 2.12 | | 31.3 |
| Contr_l + CrossE_L | MPNN | 14.34 ± 2.80 | | 5.85 ± 0.74 | | -3.21 ± 1.01 | | 29.3 |
| Contr_l + CrossE_L | PAGNN | 13.85 ± 0.47 | | 5.21 ± 0.84 | | -21.54 ± 0.60 | | 73.0 |
| Contr_l + CrossE_L | SAGE | 12.93 ± 1.73 | | 4.79 ± 0.61 | | -5.26 ± 1.30 | | 40.3 |
| Contr_l + CrossE_L + PMI_L | ALL | 1.55 ± 3.43 | | -7.37 ± 4.29 | | -3.58 ± 1.21 | | 126.0 |
| Contr_l + CrossE_L + PMI_L | GAT | 3.68 ± 1.37 | | 1.19 ± 0.24 | | -0.36 ± 0.49 | | 56.3 |
| Contr_l + CrossE_L + PMI_L | GCN | 1.94 ± 0.74 | | 0.36 ± 0.25 | | -1.33 ± 0.72 | | 87.0 |
| Contr_l + CrossE_L + PMI_L | GIN | 1.75 ± 1.72 | | -6.06 ± 1.95 | | -10.27 ± 2.40 | | 137.0 |

Silhouette Continued (↑)

| Loss Type | | | Model | CORA | | Citeseer | | Bitcoin Fraud Transaction | | Average Rank |
|---|---|---|---|---|---|---|---|---|---|---|
| Contr_l + CrossE_L + PMI_L | | | MPNN | 6.34 0.99 | ± | 1.24 0.50 | ± | -0.53 0.58 | ± | 51.7 |
| Contr_l + CrossE_L + PMI_L | | | PAGNN | -8.40 1.23 | ± | -12.22 0.69 | ± | -16.07 0.04 | ± | 188.7 |
| Contr_l + CrossE_L + PMI_L | | | SAGE | -0.92 0.29 | ± | -0.90 0.32 | ± | -10.53 1.56 | ± | 129.3 |
| Contr_l + CrossE_L + PMI_L + PR_L | | | ALL | -8.37 5.63 | ± | -11.56 3.10 | ± | -18.81 1.07 | ± | 192.7 |
| Contr_l + CrossE_L + PMI_L + PR_L | | | GAT | 5.74 1.05 | ± | 1.11 0.17 | ± | -3.53 2.86 | ± | 76.3 |
| Contr_l + CrossE_L + PMI_L + PR_L | | | GCN | 2.35 0.45 | ± | 0.22 0.25 | ± | -0.40 0.33 | ± | 72.3 |
| Contr_l + CrossE_L + PMI_L + PR_L | | | GIN | 2.20 2.36 | ± | -7.10 1.43 | ± | -15.59 3.07 | ± | 146.3 |
| Contr_l + CrossE_L + PMI_L + PR_L | | | MPNN | 6.05 1.30 | ± | -2.21 2.84 | ± | -14.09 7.42 | ± | 117.0 |
| Contr_l + CrossE_L + PMI_L + PR_L | | | PAGNN | -6.42 1.41 | ± | -10.57 1.30 | ± | -16.01 0.06 | ± | 177.0 |
| Contr_l + CrossE_L + PMI_L + PR_L | | | SAGE | -0.67 0.23 | ± | -1.26 0.90 | ± | -14.22 0.56 | ± | 139.0 |
| Contr_l + CrossE_L + PMI_L + PR_L + Triplet_L | | | ALL | 8.62 1.44 | ± | -2.37 0.73 | ± | -6.35 2.02 | ± | 89.3 |
| Contr_l + CrossE_L + PMI_L + PR_L + Triplet_L | | | GAT | 6.29 1.37 | ± | 1.11 1.18 | ± | -0.36 0.60 | ± | 49.7 |
| Contr_l + CrossE_L + PMI_L + PR_L + Triplet_L | | | GCN | 2.55 0.88 | ± | 0.53 0.23 | ± | -0.83 0.12 | ± | 72.3 |
| Contr_l + CrossE_L + PMI_L + PR_L + Triplet_L | | | GIN | 2.18 1.48 | ± | -2.57 1.02 | ± | -7.46 0.84 | ± | 123.7 |
| Contr_l + CrossE_L + PMI_L + PR_L + Triplet_L | | | MPNN | 7.03 1.62 | ± | -1.11 1.54 | ± | -3.11 3.40 | ± | 80.7 |
| Contr_l + CrossE_L + PMI_L + PR_L + Triplet_L | | | PAGNN | -7.80 0.87 | ± | -9.56 0.70 | ± | -16.11 0.05 | ± | 180.7 |

Silhouette Continued (↑)

| Loss Type | Model | CORA | | Citeseer | | Bitcoin Fraud Transaction | | Average Rank |
|---|---|---|---|---|---|---|---|---|
| Contr_l + CrossE_L + PMI_L + PR_L + Triplet_L | SAGE | 2.05 3.07 | ± | 0.03 1.37 | ± | -13.20 0.46 | ± | 121.7 |
| Contr_l + CrossE_L + PMI_L + Triplet_L | ALL | 10.90 1.30 | ± | 3.49 0.74 | ± | -2.60 0.80 | ± | 38.7 |
| Contr_l + CrossE_L + PMI_L + Triplet_L | GAT | 4.03 0.57 | ± | 1.40 0.37 | ± | -0.50 0.47 | ± | 55.7 |
| Contr_l + CrossE_L + PMI_L + Triplet_L | GCN | 2.45 0.64 | ± | 0.20 0.27 | ± | -0.77 0.50 | ± | 77.7 |
| Contr_l + CrossE_L + PMI_L + Triplet_L | GIN | 3.77 1.87 | ± | -1.93 1.41 | ± | -6.90 1.73 | ± | 108.7 |
| Contr_l + CrossE_L + PMI_L + Triplet_L | MPNN | 6.19 0.38 | ± | 1.29 0.57 | ± | -0.54 0.91 | ± | 51.7 |
| Contr_l + CrossE_L + PMI_L + Triplet_L | PAGNN | -7.33 1.18 | ± | -11.56 0.34 | ± | -16.12 0.04 | ± | 184.3 |
| Contr_l + CrossE_L + PMI_L + Triplet_L | SAGE | 0.64 1.65 | ± | -0.90 0.71 | ± | -4.07 1.86 | ± | 113.0 |
| Contr_l + CrossE_L + PR_L | ALL | -8.00 1.53 | ± | -15.65 3.72 | ± | -12.06 2.81 | ± | 181.7 |
| Contr_l + CrossE_L + PR_L | GAT | 3.03 7.60 | ± | -5.99 5.33 | ± | -5.08 6.60 | ± | 117.0 |
| Contr_l + CrossE_L + PR_L | GCN | -0.12 2.19 | ± | -5.76 1.22 | ± | -0.44 1.13 | ± | 102.7 |
| Contr_l + CrossE_L + PR_L | GIN | -4.86 1.77 | ± | -7.63 0.48 | ± | -15.30 1.08 | ± | 164.0 |
| Contr_l + CrossE_L + PR_L | MPNN | -1.11 0.98 | ± | -8.23 1.22 | ± | -12.56 10.69 | ± | 155.3 |
| Contr_l + CrossE_L + PR_L | PAGNN | -9.21 2.07 | ± | -11.43 0.91 | ± | -16.91 1.70 | ± | 191.3 |
| Contr_l + CrossE_L + PR_L | SAGE | 2.64 5.27 | ± | -2.19 2.10 | ± | -10.85 4.40 | ± | 121.0 |
| Contr_l + CrossE_L + PR_L + Triplet_L | ALL | 2.25 2.33 | ± | -0.27 0.89 | ± | -2.83 1.15 | ± | 96.7 |

Silhouette Continued (↑)

| Loss Type | Model | CORA | | | Citeseer | | | Bitcoin Fraud Transaction | | | Average Rank |
|---|---|---|---|---|---|---|---|---|---|---|---|
| Contr_l + CrossE_L + PR_L + Triplet_L | GAT | 6.85 | ± | 2.00 | 1.97 | ± | 1.97 | -0.11 | ± | 0.90 | 36.7 |
| Contr_l + CrossE_L + PR_L + Triplet_L | GCN | 3.16 | ± | 0.62 | -0.93 | ± | 0.53 | 0.08 | ± | 0.84 | 70.0 |
| Contr_l + CrossE_L + PR_L + Triplet_L | GIN | 6.13 | ± | 1.90 | -0.15 | ± | 0.81 | -3.94 | ± | 3.85 | 85.7 |
| Contr_l + CrossE_L + PR_L + Triplet_L | MPNN | 6.37 | ± | 1.00 | 0.48 | ± | 1.06 | -4.57 | ± | 4.70 | 77.7 |
| Contr_l + CrossE_L + PR_L + Triplet_L | PAGNN | -5.50 | ± | 1.32 | -3.38 | ± | 4.41 | -20.00 | ± | 0.92 | 169.0 |
| Contr_l + CrossE_L + PR_L + Triplet_L | SAGE | 8.31 | ± | 0.93 | 1.39 | ± | 1.48 | -7.14 | ± | 1.42 | 70.3 |
| Contr_l + CrossE_L + Triplet_L | ALL | 12.83 | ± | 1.09 | 4.52 | ± | 0.67 | -2.50 | ± | 0.83 | 30.7 |
| Contr_l + CrossE_L + Triplet_L | GAT | 9.55 | ± | 0.46 | 3.30 | ± | 0.17 | -1.31 | ± | 0.27 | 35.0 |
| Contr_l + CrossE_L + Triplet_L | GCN | 8.16 | ± | 0.53 | 2.73 | ± | 0.65 | -1.23 | ± | 0.62 | 40.3 |
| Contr_l + CrossE_L + Triplet_L | GIN | 12.30 | ± | 0.83 | 3.50 | ± | 0.23 | -1.97 | ± | 0.77 | 33.7 |
| Contr_l + CrossE_L + Triplet_L | MPNN | 12.70 | ± | 1.07 | 4.04 | ± | 0.28 | -2.13 | ± | 0.60 | 33.0 |
| Contr_l + CrossE_L + Triplet_L | PAGNN | 9.53 | ± | 7.15 | 4.26 | ± | 0.69 | -20.60 | ± | 0.72 | 81.3 |
| Contr_l + CrossE_L + Triplet_L | SAGE | 11.43 | ± | 1.08 | 3.69 | ± | 0.33 | -3.25 | ± | 1.07 | 41.3 |
| Contr_l + PMI_L | ALL | 1.56 | ± | 5.50 | -1.38 | ± | 0.71 | -3.93 | ± | 1.48 | 112.0 |
| Contr_l + PMI_L | GAT | 3.47 | ± | 0.82 | 1.35 | ± | 0.23 | -0.36 | ± | 0.62 | 56.3 |
| Contr_l + PMI_L | GCN | 1.61 | ± | 0.22 | 0.12 | ± | 0.25 | -0.50 | ± | 0.37 | 81.7 |

Silhouette Continued (↑)

| Loss Type | Model | CORA | | Citeseer | | Bitcoin Fraud Transaction | | Average Rank |
|---|---|---|---|---|---|---|---|---|
| Contr_l + PMI_L | GIN | -0.16 ± | | -4.47 ± | | -9.75 ± | | 137.7 |
| | | 2.18 | | 2.07 | | 1.88 | | |
| Contr_l + PMI_L | MPNN | 6.63 ± | | 1.67 ± | | -0.27 ± | | 39.3 |
| | | 0.88 | | 0.20 | | 0.60 | | |
| Contr_l + PMI_L | PAGNN | -8.54 ± | | -12.22 ± | | -16.10 ± | | 189.7 |
| | | 0.92 | | 0.38 | | 0.03 | | |
| Contr_l + PMI_L | SAGE | -0.96 ± | | -1.35 ± | | -11.74 ± | | 135.3 |
| | | 0.74 | | 0.65 | | 0.74 | | |
| Contr_l + PMI_L + PR_L | ALL | -3.65 ± | | -12.02 ± | | -18.17 ± | | 182.3 |
| | | 3.06 | | 2.14 | | 0.63 | | |
| Contr_l + PMI_L + PR_L | GAT | 5.84 ± | | 0.09 ± | | -5.71 ± | | 91.0 |
| | | 1.27 | | 1.85 | | 3.55 | | |
| Contr_l + PMI_L + PR_L | GCN | 2.16 ± | | 0.32 ± | | -1.04 ± | | 81.7 |
| | | 0.96 | | 0.16 | | 0.58 | | |
| Contr_l + PMI_L + PR_L | GIN | 2.66 ± | | -7.10 ± | | -17.83 ± | | 151.0 |
| | | 2.38 | | 2.86 | | 3.25 | | |
| Contr_l + PMI_L + PR_L | MPNN | 7.78 ± | | -3.31 ± | | -13.90 ± | | 110.0 |
| | | 0.73 | | 2.72 | | 7.13 | | |
| Contr_l + PMI_L + PR_L | PAGNN | -8.64 ± | | -11.15 ± | | -16.04 ± | | 185.3 |
| | | 1.71 | | 0.73 | | 0.03 | | |
| Contr_l + PMI_L + PR_L | SAGE | -0.67 ± | | -2.12 ± | | -14.36 ± | | 142.3 |
| | | 0.65 | | 1.23 | | 0.54 | | |
| Contr_l + PMI_L + PR_L + Triplet_L | ALL | 7.89 ± | | 0.03 ± | | -6.35 ± | | 81.3 |
| | | 1.33 | | 0.80 | | 1.07 | | |
| Contr_l + PMI_L + PR_L + Triplet_L | GAT | 7.40 ± | | 0.81 ± | | 0.65 ± | | 40.3 |
| | | 1.04 | | 0.92 | | 0.32 | | |
| Contr_l + PMI_L + PR_L + Triplet_L | GCN | 2.94 ± | | 0.80 ± | | -0.65 ± | | 68.7 |
| | | 1.72 | | 0.39 | | 0.33 | | |
| Contr_l + PMI_L + PR_L + Triplet_L | GIN | 6.89 ± | | -0.16 ± | | -7.14 ± | | 91.7 |
| | | 2.14 | | 1.52 | | 1.23 | | |
| Contr_l + PMI_L + PR_L + Triplet_L | MPNN | 7.64 ± | | 0.10 ± | | -6.70 ± | | 84.0 |
| | | 1.35 | | 0.76 | | 3.93 | | |

Silhouette Continued (↑)

| Loss Type | Model | CORA | | Citeseer | | Bitcoin Fraud Transaction | | Average Rank |
|---|---|---|---|---|---|---|---|---|
| Contr_l + PMI_L + PR_L + Triplet_L | PAGNN | -6.92 ± 1.12 | | -6.49 ± 2.18 | | -16.20 ± 0.18 | | 173.7 |
| Contr_l + PMI_L + PR_L + Triplet_L | SAGE | 7.38 ± 1.33 | | 1.27 ± 1.10 | | -9.10 ± 1.35 | | 80.0 |
| Contr_l + PR_L | ALL | -7.08 ± 1.13 | | -15.21 ± 2.63 | | -11.12 ± 1.11 | | 175.7 |
| Contr_l + PR_L | GAT | -0.46 ± 6.39 | | -8.59 ± 1.35 | | -2.11 ± 8.39 | | 125.7 |
| Contr_l + PR_L | GCN | -0.62 ± 1.53 | | -3.89 ± 2.06 | | -0.08 ± 1.37 | | 98.0 |
| Contr_l + PR_L | GIN | -5.63 ± 2.32 | | -8.20 ± 2.37 | | -17.05 ± 3.02 | | 175.7 |
| Contr_l + PR_L | MPNN | -2.96 ± 1.61 | | -8.03 ± 1.89 | | -14.89 ± 0.91 | | 160.3 |
| Contr_l + PR_L | PAGNN | -10.74 ± 2.27 | | -11.11 ± 0.54 | | -16.68 ± 1.57 | | 192.3 |
| Contr_l + PR_L | SAGE | 2.48 ± 5.20 | | -1.21 ± 3.96 | | -11.38 ± 6.08 | | 119.0 |
| Contr_l + PR_L + Triplet_L | ALL | -0.18 ± 5.11 | | -3.38 ± 4.02 | | -2.52 ± 1.47 | | 116.0 |
| Contr_l + PR_L + Triplet_L | GAT | 6.59 ± 3.55 | | 1.54 ± 2.35 | | -0.11 ± 1.25 | | 40.0 |
| Contr_l + PR_L + Triplet_L | GCN | 2.02 ± 1.39 | | -1.62 ± 0.75 | | 0.05 ± 0.73 | | 82.3 |
| Contr_l + PR_L + Triplet_L | GIN | 4.48 ± 2.60 | | -0.87 ± 1.05 | | -6.27 ± 3.64 | | 98.7 |
| Contr_l + PR_L + Triplet_L | MPNN | 5.96 ± 1.40 | | -2.14 ± 1.98 | | -5.51 ± 3.60 | | 100.3 |
| Contr_l + PR_L + Triplet_L | PAGNN | -5.21 ± 0.27 | | -2.90 ± 4.73 | | -20.67 ± 0.37 | | 168.3 |
| Contr_l + PR_L + Triplet_L | SAGE | 8.14 ± 0.99 | | 1.60 ± 1.03 | | -5.39 ± 1.30 | | 62.7 |

Continued on next page

Silhouette Continued (↑)

| Loss Type | Model | CORA | | Citeseer | | Bitcoin Fraud Transaction | | Average Rank |
|---|---|---|---|---|---|---|---|---|
| Contr_l + Triplet_L | ALL | 12.97 | ± | 4.12 | ± | -2.12 | ± | 31.3 |
| | | 0.86 | | 0.43 | | 0.41 | | |
| Contr_l + Triplet_L | GAT | 9.16 | ± | 3.19 | ± | -0.97 | ± | 33.3 |
| | | 0.64 | | 0.18 | | 0.25 | | |
| Contr_l + Triplet_L | GCN | 7.65 | ± | 2.71 | ± | -1.09 | ± | 42.3 |
| | | 1.16 | | 0.51 | | 0.58 | | |
| Contr_l + Triplet_L | GIN | 11.57 | ± | 3.16 | ± | -2.55 | ± | 38.7 |
| | | 0.57 | | 0.09 | | 1.68 | | |
| Contr_l + Triplet_L | MPNN | 11.79 | ± | 3.71 | ± | -1.69 | ± | 33.0 |
| | | 0.48 | | 0.39 | | 0.34 | | |
| Contr_l + Triplet_L | PAGNN | 10.11 | ± | 4.38 | ± | -20.24 | ± | 79.0 |
| | | 7.81 | | 0.52 | | 0.64 | | |
| Contr_l + Triplet_L | SAGE | 10.34 | ± | 3.15 | ± | -3.94 | ± | 48.0 |
| | | 0.59 | | 0.36 | | 0.94 | | |
| CrossE_L | ALL | -9.56 | ± | -3.27 | ± | -1.31 | ± | 130.3 |
| | | 1.93 | | 1.99 | | 2.79 | | |
| CrossE_L | GAT | -4.28 | ± | -3.88 | ± | 1.35 | ± | 101.0 |
| | | 1.08 | | 0.53 | | 0.81 | | |
| CrossE_L | GCN | -5.30 | ± | -5.02 | ± | -13.30 | ± | 156.3 |
| | | 1.21 | | 1.06 | | 2.75 | | |
| CrossE_L | GIN | -9.15 | ± | -17.13 | ± | -17.72 | ± | 199.0 |
| | | 1.24 | | 1.35 | | 0.20 | | |
| CrossE_L | MPNN | -8.30 | ± | -6.64 | ± | -21.48 | ± | 184.7 |
| | | 1.01 | | 1.59 | | 2.44 | | |
| CrossE_L | PAGNN | -4.49 | ± | -4.99 | ± | -16.13 | ± | 163.0 |
| | | 1.69 | | 1.24 | | 0.04 | | |
| CrossE_L | SAGE | -5.86 | ± | -21.49 | ± | -12.27 | ± | 178.0 |
| | | 1.85 | | 6.28 | | 1.94 | | |
| CrossE_L + PMI_L | ALL | -0.54 | ± | -10.03 | ± | -2.82 | ± | 132.0 |
| | | 1.35 | | 0.99 | | 0.84 | | |
| CrossE_L + PMI_L | GAT | 3.61 | ± | 1.12 | ± | -0.60 | ± | 61.7 |
| | | 0.79 | | 0.25 | | 1.26 | | |

Silhouette Continued (↑)

| Loss Type | Model | CORA | | Citeseer | | Bitcoin Fraud Transaction | | Average Rank |
|---|---|---|---|---|---|---|---|---|
| CrossE_L + PMI_L | GCN | 1.45 | ± | 0.28 | ± | -0.64 | ± | 83.0 |
| | | 0.57 | | 0.27 | | 0.41 | | |
| CrossE_L + PMI_L | GIN | 0.81 | ± | -8.93 | ± | -11.62 | ± | 149.3 |
| | | 1.69 | | 1.37 | | 1.46 | | |
| CrossE_L + PMI_L | MPNN | 5.62 | ± | 0.98 | ± | -0.74 | ± | 61.3 |
| | | 1.75 | | 0.66 | | 0.29 | | |
| CrossE_L + PMI_L | PAGNN | -7.45 | ± | -12.10 | ± | -16.03 | ± | 185.3 |
| | | 2.40 | | 0.78 | | 0.06 | | |
| CrossE_L + PMI_L | SAGE | -1.07 | ± | -0.74 | ± | -7.23 | ± | 125.7 |
| | | 0.22 | | 0.20 | | 0.61 | | |
| CrossE_L + PMI_L + PR_L | ALL | -10.82 | ± | -13.40 | ± | -22.15 | ± | 206.0 |
| | | 1.32 | | 0.64 | | 0.57 | | |
| CrossE_L + PMI_L + PR_L | GAT | 5.02 | ± | -1.47 | ± | -5.31 | ± | 100.3 |
| | | 0.43 | | 2.75 | | 6.89 | | |
| CrossE_L + PMI_L + PR_L | GCN | 2.34 | ± | 0.24 | ± | -1.11 | ± | 82.7 |
| | | 1.22 | | 0.26 | | 0.65 | | |
| CrossE_L + PMI_L + PR_L | GIN | 1.51 | ± | -8.11 | ± | -17.74 | ± | 162.0 |
| | | 0.50 | | 3.03 | | 3.82 | | |
| CrossE_L + PMI_L + PR_L | MPNN | 6.66 | ± | -2.50 | ± | -8.55 | ± | 105.0 |
| | | 1.90 | | 3.96 | | 11.23 | | |
| CrossE_L + PMI_L + PR_L | PAGNN | -6.41 | ± | -10.76 | ± | -15.98 | ± | 176.7 |
| | | 3.11 | | 2.65 | | 0.12 | | |
| CrossE_L + PMI_L + PR_L | SAGE | -0.71 | ± | -0.71 | ± | -14.15 | ± | 135.3 |
| | | 0.41 | | 0.36 | | 0.36 | | |
| CrossE_L + PMI_L + PR_L + Triplet_L | ALL | 7.03 | ± | -3.95 | ± | -6.36 | ± | 101.7 |
| | | 0.74 | | 1.10 | | 0.39 | | |
| CrossE_L + PMI_L + PR_L + Triplet_L | GAT | 5.56 | ± | 1.25 | ± | -0.39 | ± | 53.7 |
| | | 1.60 | | 0.58 | | 0.84 | | |
| CrossE_L + PMI_L + PR_L + Triplet_L | GCN | 2.01 | ± | 0.47 | ± | -1.05 | ± | 82.7 |
| | | 1.17 | | 0.13 | | 0.71 | | |
| CrossE_L + PMI_L + PR_L + Triplet_L | GIN | 3.31 | ± | -2.48 | ± | -8.24 | ± | 116.3 |
| | | 2.37 | | 1.09 | | 1.70 | | |

Silhouette Continued (↑)

| Loss Type | Model | CORA | | Citeseer | | Bitcoin Fraud Transaction | | Average Rank |
|---|---|---|---|---|---|---|---|---|
| CrossE_L + PMI_L + PR_L + Triplet_L | MPNN | 5.72 1.17 | ± | -0.60 2.33 | ± | -2.63 4.87 | ± | 84.0 |
| CrossE_L + PMI_L + PR_L + Triplet_L | PAGNN | -6.28 0.73 | ± | -9.27 1.37 | ± | -16.12 0.04 | ± | 176.3 |
| CrossE_L + PMI_L + PR_L + Triplet_L | SAGE | 2.24 2.72 | ± | -0.24 1.54 | ± | -11.86 0.83 | ± | 118.7 |
| CrossE_L + PMI_L + Triplet_L | ALL | 10.66 0.94 | ± | 3.16 0.39 | ± | -1.92 0.28 | ± | 37.7 |
| CrossE_L + PMI_L + Triplet_L | GAT | 4.06 0.89 | ± | 1.44 0.19 | ± | -0.46 0.46 | ± | 53.0 |
| CrossE_L + PMI_L + Triplet_L | GCN | 1.47 0.16 | ± | 0.38 0.20 | ± | -0.68 0.44 | ± | 81.0 |
| CrossE_L + PMI_L + Triplet_L | GIN | 3.15 2.84 | ± | -0.79 1.10 | ± | -5.43 2.79 | ± | 101.7 |
| CrossE_L + PMI_L + Triplet_L | MPNN | 7.53 1.10 | ± | 1.32 0.55 | ± | -0.96 0.42 | ± | 49.3 |
| CrossE_L + PMI_L + Triplet_L | PAGNN | -6.76 1.35 | ± | -11.04 0.78 | ± | -16.13 0.02 | ± | 182.3 |
| CrossE_L + PMI_L + Triplet_L | SAGE | 2.98 1.27 | ± | 0.16 0.55 | ± | -3.17 0.37 | ± | 90.7 |
| CrossE_L + PR_L | ALL | -5.83 1.21 | ± | -19.27 1.25 | ± | -10.66 4.36 | ± | 172.7 |
| CrossE_L + PR_L | GAT | -4.59 3.08 | ± | -8.74 1.14 | ± | -14.85 21.04 | ± | 165.3 |
| CrossE_L + PR_L | GCN | -0.23 1.44 | ± | -10.09 1.90 | ± | -13.40 2.50 | ± | 156.7 |
| CrossE_L + PR_L | GIN | -10.25 2.67 | ± | -12.25 1.94 | ± | -14.92 2.61 | ± | 188.3 |
| CrossE_L + PR_L | MPNN | -4.17 4.49 | ± | -13.48 0.79 | ± | -25.19 5.64 | ± | 191.3 |
| CrossE_L + PR_L | PAGNN | -11.67 2.06 | ± | -12.00 0.92 | ± | -15.37 0.23 | ± | 188.7 |

Silhouette Continued (↑)

| Loss Type | Model | CORA | | | Citeseer | | | Bitcoin Fraud Transaction | | | Average Rank |
|---|---|---|---|---|---|---|---|---|---|---|---|
| CrossE_L + PR_L | SAGE | -4.94 | ± | 1.32 | -4.45 | ± | 0.81 | -6.31 | ± | 1.38 | 140.3 |
| CrossE_L + PR_L + Triplet_L | ALL | -2.78 | ± | 6.45 | -9.76 | ± | 5.20 | -11.88 | ± | 11.42 | 158.7 |
| CrossE_L + PR_L + Triplet_L | GAT | 3.13 | ± | 2.44 | 0.41 | ± | 3.37 | -0.44 | ± | 1.37 | 65.7 |
| CrossE_L + PR_L + Triplet_L | GCN | 2.13 | ± | 0.99 | -2.92 | ± | 1.31 | 0.53 | ± | 1.15 | 85.3 |
| CrossE_L + PR_L + Triplet_L | GIN | 2.32 | ± | 1.61 | -1.69 | ± | 1.33 | -15.53 | ± | 2.11 | 132.3 |
| CrossE_L + PR_L + Triplet_L | MPNN | 3.14 | ± | 0.74 | -5.32 | ± | 0.44 | -7.34 | ± | 2.50 | 123.0 |
| CrossE_L + PR_L + Triplet_L | PAGNN | -7.70 | ± | 0.66 | -8.18 | ± | 2.07 | -18.70 | ± | 1.06 | 183.0 |
| CrossE_L + PR_L + Triplet_L | SAGE | 6.44 | ± | 1.75 | 1.75 | ± | 1.25 | -6.52 | ± | 1.46 | 73.7 |
| CrossE_L + Triplet_L | ALL | 9.42 | ± | 0.88 | 2.58 | ± | 0.08 | -1.33 | ± | 0.41 | 41.0 |
| CrossE_L + Triplet_L | GAT | 6.10 | ± | 0.43 | 2.03 | ± | 0.07 | -0.14 | ± | 0.24 | 41.0 |
| CrossE_L + Triplet_L | GCN | 5.34 | ± | 0.45 | 1.41 | ± | 0.23 | -0.69 | ± | 0.22 | 56.3 |
| CrossE_L + Triplet_L | GIN | 7.81 | ± | 0.68 | 1.44 | ± | 0.17 | -0.97 | ± | 0.65 | 46.0 |
| CrossE_L + Triplet_L | MPNN | 8.58 | ± | 0.25 | 2.29 | ± | 0.30 | -0.94 | ± | 0.69 | 38.0 |
| CrossE_L + Triplet_L | PAGNN | 8.18 | ± | 6.71 | 3.24 | ± | 0.48 | -21.36 | ± | 0.69 | 87.7 |
| CrossE_L + Triplet_L | SAGE | 7.02 | ± | 0.41 | 2.51 | ± | 0.34 | -4.90 | ± | 2.12 | 62.7 |
| PMI_L | ALL | -4.27 | ± | 1.63 | -9.51 | ± | 1.40 | -2.98 | ± | 0.47 | 138.0 |

Silhouette Continued (↑)

| Loss Type | Model | CORA | | Citeseer | | Bitcoin Fraud Transaction | | Average Rank |
|---|---|---|---|---|---|---|---|---|
| PMI_L | GAT | 3.54 | ± | 1.10 | ± | -0.48 | ± | 61.7 |
| | | 0.89 | | 0.27 | | 1.01 | | |
| PMI_L | GCN | 2.02 | ± | 0.01 | ± | -1.28 | ± | 91.0 |
| | | 1.02 | | 0.42 | | 1.06 | | |
| PMI_L | GIN | -1.77 | ± | -8.44 | ± | -11.28 | ± | 153.3 |
| | | 3.54 | | 1.59 | | 1.78 | | |
| PMI_L | MPNN | 6.98 | ± | 1.43 | ± | -0.27 | ± | 40.3 |
| | | 0.98 | | 0.28 | | 0.21 | | |
| PMI_L | PAGNN | -7.34 | ± | -12.20 | ± | -16.02 | ± | 184.7 |
| | | 1.63 | | 0.32 | | 0.06 | | |
| PMI_L | SAGE | -1.43 | ± | -1.19 | ± | -7.11 | ± | 128.0 |
| | | 0.42 | | 0.22 | | 1.37 | | |
| PMI_L + PR_L | ALL | -9.83 | ± | -13.10 | ± | -21.43 | ± | 203.0 |
| | | 1.37 | | 1.98 | | 0.84 | | |
| PMI_L + PR_L | GAT | 6.18 | ± | -3.73 | ± | -10.74 | ± | 113.7 |
| | | 0.69 | | 5.12 | | 2.37 | | |
| PMI_L + PR_L | GCN | 2.44 | ± | 0.36 | ± | -0.83 | ± | 76.3 |
| | | 1.18 | | 0.34 | | 0.75 | | |
| PMI_L + PR_L | GIN | 2.13 | ± | -7.12 | ± | -20.21 | ± | 159.7 |
| | | 3.59 | | 3.61 | | 1.13 | | |
| PMI_L + PR_L | MPNN | 6.93 | ± | -4.99 | ± | -17.69 | ± | 130.0 |
| | | 1.15 | | 1.16 | | 9.38 | | |
| PMI_L + PR_L | PAGNN | -6.72 | ± | -10.22 | ± | -16.15 | ± | 181.3 |
| | | 1.11 | | 1.05 | | 0.08 | | |
| PMI_L + PR_L | SAGE | -0.48 | ± | -0.80 | ± | -14.00 | ± | 133.7 |
| | | 0.63 | | 0.44 | | 0.43 | | |
| PMI_L + PR_L + Triplet_L | ALL | 7.86 | ± | -2.72 | ± | -6.42 | ± | 94.3 |
| | | 0.99 | | 1.46 | | 1.92 | | |
| PMI_L + PR_L + Triplet_L | GAT | 6.19 | ± | 0.81 | ± | 0.40 | ± | 47.7 |
| | | 0.89 | | 0.75 | | 0.93 | | |
| PMI_L + PR_L + Triplet_L | GCN | 3.37 | ± | 0.11 | ± | -0.47 | ± | 69.3 |
| | | 0.93 | | 0.30 | | 0.35 | | |

Silhouette Continued (↑)

| Loss Type | Model | CORA | | Citeseer | | Bitcoin Fraud Transaction | | Average Rank |
|---|---|---|---|---|---|---|---|---|
| PMI_L + PR_L + Triplet_L | GIN | 3.90 ± 1.16 | | -2.27 ± 0.84 | | -7.78 ± 1.02 | | 112.7 |
| PMI_L + PR_L + Triplet_L | MPNN | 6.36 ± 0.82 | | -1.78 ± 0.53 | | -5.17 ± 5.36 | | 94.3 |
| PMI_L + PR_L + Triplet_L | PAGNN | -6.33 ± 1.41 | | -8.27 ± 1.09 | | -16.11 ± 0.06 | | 174.0 |
| PMI_L + PR_L + Triplet_L | SAGE | 3.60 ± 2.79 | | 0.29 ± 0.67 | | -11.54 ± 0.69 | | 104.7 |
| PMI_L + Triplet_L | ALL | 11.08 ± 0.32 | | 3.15 ± 0.83 | | -1.98 ± 0.45 | | 38.3 |
| PMI_L + Triplet_L | GAT | 4.34 ± 0.47 | | 1.38 ± 0.20 | | -0.71 ± 0.69 | | 58.7 |
| PMI_L + Triplet_L | GCN | 1.72 ± 0.58 | | 0.30 ± 0.33 | | -0.92 ± 0.59 | | 83.7 |
| PMI_L + Triplet_L | GIN | 2.55 ± 5.11 | | -1.25 ± 1.70 | | -6.49 ± 1.30 | | 110.7 |
| PMI_L + Triplet_L | MPNN | 5.76 ± 1.02 | | 1.81 ± 0.45 | | -1.19 ± 0.59 | | 57.0 |
| PMI_L + Triplet_L | PAGNN | -7.41 ± 1.22 | | -11.74 ± 0.46 | | -16.13 ± 0.03 | | 187.3 |
| PMI_L + Triplet_L | SAGE | 2.46 ± 2.57 | | 0.31 ± 0.71 | | -3.30 ± 0.42 | | 91.7 |
| PR_L | ALL | -10.67 ± 2.84 | | -22.17 ± 2.59 | | -6.85 ± 4.51 | | 177.7 |
| PR_L | GAT | -4.92 ± 2.75 | | -10.85 ± 0.97 | | -16.63 ± 23.20 | | 179.0 |
| PR_L | GCN | -1.81 ± 1.27 | | -11.64 ± 1.37 | | 1.57 ± 2.20 | | 115.0 |
| PR_L | GIN | -10.04 ± 1.46 | | -13.06 ± 1.37 | | -11.99 ± 1.97 | | 183.0 |
| PR_L | MPNN | -3.18 ± 3.64 | | -12.67 ± 1.78 | | -14.07 ± 6.38 | | 171.0 |

Silhouette Continued (↑)

| Loss Type | Model | CORA | | Citeseer | | Bitcoin Fraud Transaction | | Average Rank |
|---|---|---|---|---|---|---|---|---|
| PR_L | PAGNN | -14.22 ± 4.15 | | -12.95 ± 1.24 | | -15.74 ± 0.40 | | 192.7 |
| PR_L | SAGE | -4.37 ± 1.24 | | -5.69 ± 1.61 | | -4.04 ± 1.64 | | 135.3 |
| PR_L + Triplet_L | ALL | -7.08 ± 0.62 | | -19.29 ± 2.18 | | -8.44 ± 4.01 | | 174.3 |
| PR_L + Triplet_L | GAT | 2.18 ± 6.72 | | -8.79 ± 1.93 | | -4.58 ± 5.32 | | 128.7 |
| PR_L + Triplet_L | GCN | 0.85 ± 1.93 | | -5.81 ± 2.08 | | 0.66 ± 1.79 | | 95.7 |
| PR_L + Triplet_L | GIN | -5.77 ± 0.59 | | -6.97 ± 0.76 | | -14.44 ± 2.49 | | 163.3 |
| PR_L + Triplet_L | MPNN | -1.05 ± 2.70 | | -10.13 ± 1.72 | | -12.92 ± 9.84 | | 159.3 |
| PR_L + Triplet_L | PAGNN | -11.26 ± 2.21 | | -11.67 ± 0.74 | | -16.32 ± 1.59 | | 194.3 |
| PR_L + Triplet_L | SAGE | -4.70 ± 1.97 | | -3.08 ± 3.90 | | -7.10 ± 2.33 | | 139.7 |
| Triplet_L | ALL | 9.85 ± 0.60 | | 2.62 ± 0.20 | | -1.45 ± 0.40 | | 40.0 |
| Triplet_L | GAT | 5.68 ± 0.27 | | 1.99 ± 0.12 | | -0.01 ± 0.24 | | 42.7 |
| Triplet_L | GCN | 6.00 ± 0.67 | | 1.58 ± 0.24 | | -0.31 ± 0.35 | | 45.0 |
| Triplet_L | GIN | 7.98 ± 0.77 | | 1.54 ± 0.24 | | -0.77 ± 0.78 | | 42.0 |
| Triplet_L | MPNN | 9.57 ± 0.27 | | 2.44 ± 0.18 | | -1.04 ± 0.45 | | 36.7 |
| Triplet_L | PAGNN | 7.66 ± 7.48 | | 2.95 ± 0.36 | | -20.11 ± 0.66 | | 91.0 |
| Triplet_L | SAGE | 7.09 ± 0.42 | | 2.43 ± 0.26 | | -3.83 ± 1.06 | | 59.3 |

**Table 17.** Calinski Harabasz Performance (↑): Top-ranked results are highlighted in **1st**, second-ranked in **2nd**, and third-ranked in **3rd**.

| Loss Type | Model | CORA | Citeseer | Bitcoin Fraud Transaction | Average Rank |
|---|---|---|---|---|---|
| Contr_l | ALL | 33999.80 ± 1554.88 | 20866.42 ± 2637.95 | 1599.91 ± 497.12 | 20.0 |
| Contr_l | GAT | 21223.49 ± 751.97 | 9846.52 ± 331.72 | 1395.32 ± 217.35 | 64.3 |
| Contr_l | GCN | 17722.33 ± 500.81 | 10327.50 ± 1607.04 | 1233.48 ± 318.83 | 71.0 |
| Contr_l | GIN | 24042.89 ± 292.92 | 11027.25 ± 323.93 | 2237.69 ± 636.62 | 34.3 |
| Contr_l | MPNN | 28655.93 ± 1127.06 | 15089.31 ± 1261.35 | 2573.91 ± 972.52 | 12.0 |
| Contr_l | PAGNN | 28226.88 ± 2385.56 | 18500.06 ± 2009.69 | 107.32 ± 3.72 | 73.0 |
| Contr_l | SAGE | 28642.23 ± 1348.12 | 13833.98 ± 712.98 | 1513.99 ± 585.44 | 32.0 |
| Contr_l + CrossE_L | ALL | 34582.58 ± 2397.13 | 19808.53 ± 3260.27 | 1265.13 ± 475.16 | 28.3 |
| Contr_l + CrossE_L | GAT | 20771.32 ± 614.24 | 9513.84 ± 336.53 | 1143.45 ± 395.01 | 76.3 |
| Contr_l + CrossE_L | GCN | 20139.96 ± 1081.18 | 9592.66 ± 1057.50 | 1449.42 ± 312.10 | 66.0 |
| Contr_l + CrossE_L | GIN | 24083.73 ± 754.84 | 12194.74 ± 1561.17 | 2492.66 ± 1152.14 | 26.7 |
| Contr_l + CrossE_L | MPNN | 29319.66 ± 1433.96 | 15618.32 ± 630.39 | 1682.53 ± 535.11 | 23.3 |
| Contr_l + CrossE_L | PAGNN | 27524.86 ± 991.73 | 18420.21 ± 2106.53 | 112.47 ± 0.69 | 71.7 |
| Contr_l + CrossE_L | SAGE | 26715.22 ± 689.40 | 13313.83 ± 430.31 | 936.96 ± 474.49 | 56.0 |
| Contr_l + CrossE_L + PMI_L | ALL | 26316.19 ± 2353.56 | 9033.07 ± 3972.82 | 2499.66 ± 791.90 | 43.3 |

Calinski Harabasz Continued (↑)

| Loss Type | | | | Model | CORA | Citeseer | Bitcoin Fraud Transaction | Average Rank |
|---|---|---|---|---|---|---|---|---|
| Contr_l + CrossE_L + PMI_L | | | | GAT | 10029.85 ± 1146.83 | 5846.02 ± 426.25 | 2060.56 ± 455.08 | 118.7 |
| Contr_l + CrossE_L + PMI_L | | | | GCN | 7000.05 ± 1515.47 | 4065.49 ± 546.74 | 1158.26 ± 159.14 | 156.3 |
| Contr_l + CrossE_L + PMI_L | | | | GIN | 13494.13 ± 4531.11 | 7187.12 ± 1159.91 | 1332.37 ± 429.82 | 108.3 |
| Contr_l + CrossE_L + PMI_L | | | | MPNN | 11911.75 ± 1246.43 | 6266.69 ± 314.76 | 2204.80 ± 466.00 | 112.3 |
| Contr_l + CrossE_L + PMI_L | | | | PAGNN | 8125.00 ± 1203.89 | 6810.13 ± 150.69 | 116.26 ± 3.32 | 167.3 |
| Contr_l + CrossE_L + PMI_L | | | | SAGE | 2414.02 ± 126.05 | 1850.31 ± 582.15 | 1567.97 ± 1285.14 | 155.0 |
| Contr_l + CrossE_L + PMI_L + PR_L | | | | ALL | 6762.25 ± 2469.19 | 6490.33 ± 607.43 | 268.42 ± 52.71 | 175.0 |
| Contr_l + CrossE_L + PMI_L + PR_L | | | | GAT | 13998.05 ± 3283.37 | 6942.92 ± 697.02 | 1093.95 ± 815.37 | 118.3 |
| Contr_l + CrossE_L + PMI_L + PR_L | | | | GCN | 8305.93 ± 1328.37 | 4193.78 ± 420.20 | 1381.10 ± 341.36 | 140.7 |
| Contr_l + CrossE_L + PMI_L + PR_L | | | | GIN | 14993.14 ± 3491.83 | 6024.22 ± 651.78 | 257.98 ± 156.74 | 150.3 |
| Contr_l + CrossE_L + PMI_L + PR_L | | | | MPNN | 13109.78 ± 2090.56 | 8323.31 ± 461.99 | 555.34 ± 397.44 | 128.7 |
| Contr_l + CrossE_L + PMI_L + PR_L | | | | PAGNN | 7878.23 ± 776.37 | 6800.85 ± 227.47 | 96.23 ± 2.97 | 174.0 |
| Contr_l + CrossE_L + PMI_L + PR_L | | | | SAGE | 2464.12 ± 414.21 | 3774.16 ± 4427.16 | 222.81 ± 65.01 | 192.3 |
| Contr_l + CrossE_L + PMI_L + PR_L + Triplet_L | | | | ALL | 22409.59 ± 2619.43 | 10371.77 ± 486.86 | 986.38 ± 210.52 | 72.3 |
| Contr_l + CrossE_L + PMI_L + PR_L + Triplet_L | | | | GAT | 13993.71 ± 2396.63 | 7706.67 ± 1159.54 | 1981.81 ± 555.70 | 88.7 |
| Contr_l + CrossE_L + PMI_L + PR_L + Triplet_L | | | | GCN | 7624.64 ± 969.45 | 4830.70 ± 942.85 | 1076.59 ± 139.87 | 155.0 |

Calinski Harabasz Continued (↑)

| Loss Type | Model | CORA | Citeseer | Bitcoin Fraud Transaction | Average Rank |
|---|---|---|---|---|---|
| Contr_l + CrossE_L + PMI_L + PR_L + Triplet_L | GIN | 12912.53 ± 2386.39 | 7968.41 ± 1073.81 | 1395.68 ± 470.53 | 104.0 |
| Contr_l + CrossE_L + PMI_L + PR_L + Triplet_L | MPNN | 13851.71 ± 2849.64 | 8492.27 ± 416.27 | 1404.56 ± 443.80 | 95.0 |
| Contr_l + CrossE_L + PMI_L + PR_L + Triplet_L | PAGNN | 7890.22 ± 961.28 | 7033.58 ± 299.15 | 108.79 ± 1.35 | 167.7 |
| Contr_l + CrossE_L + PMI_L + PR_L + Triplet_L | SAGE | 15451.66 ± 6421.44 | 10687.31 ± 1513.63 | 530.91 ± 160.20 | 104.0 |
| Contr_l + CrossE_L + PMI_L + Triplet_L | ALL | 28131.10 ± 1789.24 | 15135.16 ± 2657.31 | 2327.37 ± 594.73 | 15.7 |
| Contr_l + CrossE_L + PMI_L + Triplet_L | GAT | 9913.30 ± 1679.78 | 6777.48 ± 667.76 | 2034.27 ± 277.40 | 112.0 |
| Contr_l + CrossE_L + PMI_L + Triplet_L | GCN | 7538.24 ± 306.22 | 4064.36 ± 786.08 | 984.80 ± 209.14 | 161.7 |
| Contr_l + CrossE_L + PMI_L + Triplet_L | GIN | 14726.69 ± 1640.38 | 9489.34 ± 1767.53 | 2555.30 ± 1316.93 | 65.3 |
| Contr_l + CrossE_L + PMI_L + Triplet_L | MPNN | 12104.18 ± 1430.90 | 7151.55 ± 941.65 | 1916.20 ± 816.24 | 103.0 |
| Contr_l + CrossE_L + PMI_L + Triplet_L | PAGNN | 8222.99 ± 658.88 | 6695.18 ± 201.17 | 118.01 ± 1.14 | 168.3 |
| Contr_l + CrossE_L + PMI_L + Triplet_L | SAGE | 12620.14 ± 3703.43 | 9653.10 ± 1961.69 | 2264.74 ± 984.10 | 77.3 |
| Contr_l + CrossE_L + PR_L | ALL | 22941.68 ± 4835.73 | 10740.28 ± 4173.50 | 555.97 ± 135.90 | 82.0 |
| Contr_l + CrossE_L + PR_L | GAT | 30960.49 ± 15502.52 | 10711.01 ± 1656.44 | 1248.34 ± 498.19 | 48.3 |
| Contr_l + CrossE_L + PR_L | GCN | 23707.95 ± 2916.97 | 7932.54 ± 902.67 | 815.32 ± 148.13 | 94.0 |
| Contr_l + CrossE_L + PR_L | GIN | 22401.19 ± 7397.89 | 11371.90 ± 3115.31 | 541.23 ± 81.19 | 81.0 |

Continued on next page

Calinski Harabasz Continued (↑)

| Loss Type | Model | CORA | Citeseer | Bitcoin Fraud Transaction | Average Rank |
|---|---|---|---|---|---|
| Contr_l + CrossE_L + PR_L | MPNN | 42722.11 ± 5399.12 | 16117.28 ± 2041.15 | 746.88 ± 435.99 | 49.0 |
| Contr_l + CrossE_L + PR_L | PAGNN | 11095.67 ± 893.96 | 11675.43 ± 2017.60 | 62.52 ± 8.22 | 132.0 |
| Contr_l + CrossE_L + PR_L | SAGE | 22541.85 ± 9401.96 | 12418.07 ± 5969.28 | 464.50 ± 343.86 | 80.3 |
| Contr_l + CrossE_L + PR_L + Triplet_L | ALL | 26190.76 ± 3829.57 | 15076.12 ± 1234.50 | 2743.79 ± 1095.69 | 14.3 |
| Contr_l + CrossE_L + PR_L + Triplet_L | GAT | 16351.83 ± 1862.25 | 9740.18 ± 1347.69 | 1756.82 ± 262.09 | 66.3 |
| Contr_l + CrossE_L + PR_L + Triplet_L | GCN | 15527.65 ± 2255.34 | 7003.93 ± 656.11 | 943.30 ± 401.00 | 119.0 |
| Contr_l + CrossE_L + PR_L + Triplet_L | GIN | 21679.95 ± 2044.52 | 9756.98 ± 936.15 | 1337.24 ± 625.61 | 65.0 |
| Contr_l + CrossE_L + PR_L + Triplet_L | MPNN | 33314.56 ± 8366.43 | 12876.56 ± 828.46 | 1115.49 ± 590.23 | 44.3 |
| Contr_l + CrossE_L + PR_L + Triplet_L | PAGNN | 15990.73 ± 4793.31 | 12551.68 ± 2474.72 | 81.18 ± 2.10 | 110.0 |
| Contr_l + CrossE_L + PR_L + Triplet_L | SAGE | 22363.12 ± 2057.66 | 13700.16 ± 1946.72 | 1095.36 ± 984.91 | 56.7 |
| Contr_l + CrossE_L + Triplet_L | ALL | 27000.08 ± 1817.96 | 15874.57 ± 1507.06 | 1428.80 ± 275.46 | 29.7 |
| Contr_l + CrossE_L + Triplet_L | GAT | 15784.30 ± 968.74 | 8559.39 ± 560.96 | 922.00 ± 188.95 | 106.0 |
| Contr_l + CrossE_L + Triplet_L | GCN | 16362.07 ± 732.58 | 9232.54 ± 770.30 | 1412.63 ± 325.14 | 77.7 |
| Contr_l + CrossE_L + Triplet_L | GIN | 18829.34 ± 1516.56 | 9689.63 ± 1236.98 | 2208.00 ± 585.67 | 55.3 |
| Contr_l + CrossE_L + Triplet_L | MPNN | 23291.11 ± 1750.47 | 11820.45 ± 725.43 | 1599.46 ± 305.53 | 44.0 |
| Contr_l + CrossE_L + Triplet_L | PAGNN | 22166.19 ± 5837.86 | 14188.11 ± 834.19 | 110.75 ± 2.13 | 86.7 |

Calinski Harabasz Continued (↑)

| Loss Type | Model | CORA | Citeseer | Bitcoin Fraud Transaction | Average Rank |
|-----------|-------|------|----------|---------------------------|--------------|
| Contr_l + CrossE_L + Triplet_L | SAGE | 21644.64 ± 868.72 | 11548.33 ± 528.99 | 1179.01 ± 267.88 | 60.3 |
| Contr_l + PMI_L | ALL | 27155.80 ± 5178.65 | 15382.98 ± 2055.08 | 2467.89 ± 805.42 | 14.7 |
| Contr_l + PMI_L | GAT | 9924.65 ± 1072.43 | 6767.69 ± 1075.50 | 1919.26 ± 211.38 | 113.7 |
| Contr_l + PMI_L | GCN | 6285.48 ± 659.22 | 3998.00 ± 873.54 | 957.10 ± 252.74 | 167.7 |
| Contr_l + PMI_L | GIN | 13322.43 ± 2799.91 | 7906.82 ± 922.59 | 1673.24 ± 693.04 | 97.0 |
| Contr_l + PMI_L | MPNN | 12827.58 ± 1034.78 | 7215.29 ± 956.98 | 2366.74 ± 1069.19 | 92.0 |
| Contr_l + PMI_L | PAGNN | 8523.26 ± 886.62 | 6829.41 ± 251.29 | 115.89 ± 5.07 | 165.3 |
| Contr_l + PMI_L | SAGE | 2531.93 ± 538.35 | 6184.87 ± 3143.61 | 684.58 ± 213.85 | 172.0 |
| Contr_l + PMI_L + PR_L | ALL | 10233.07 ± 2052.92 | 6584.02 ± 662.49 | 516.44 ± 165.63 | 155.7 |
| Contr_l + PMI_L + PR_L | GAT | 14978.31 ± 2628.56 | 8595.28 ± 1129.44 | 590.55 ± 540.71 | 118.0 |
| Contr_l + PMI_L + PR_L | GCN | 7104.81 ± 538.30 | 5420.32 ± 818.47 | 976.70 ± 69.79 | 161.7 |
| Contr_l + PMI_L + PR_L | GIN | 15266.93 ± 3838.35 | 6725.78 ± 1634.51 | 317.78 ± 244.87 | 141.3 |
| Contr_l + PMI_L + PR_L | MPNN | 14387.65 ± 2442.14 | 8040.74 ± 1065.78 | 468.58 ± 240.89 | 128.0 |
| Contr_l + PMI_L + PR_L | PAGNN | 7782.24 ± 688.45 | 6857.89 ± 102.84 | 97.83 ± 2.21 | 173.3 |
| Contr_l + PMI_L + PR_L | SAGE | 3000.60 ± 769.21 | 8041.21 ± 2966.65 | 212.45 ± 26.70 | 163.7 |
| Contr_l + PMI_L + PR_L + Triplet_L | ALL | 25410.90 ± 1735.91 | 12574.90 ± 915.14 | 868.41 ± 239.82 | 61.7 |

Calinski Harabasz Continued (↑)

| Loss Type | Model | CORA | Citeseer | Bitcoin Fraud Transaction | Average Rank |
|---|---|---|---|---|---|
| Contr_l + PMI_L + PR_L + Triplet_L | GAT | 18714.94 ± 1712.70 | 8191.12 ± 956.82 | 1462.01 ± 368.51 | 79.0 |
| Contr_l + PMI_L + PR_L + Triplet_L | GCN | 9177.09 ± 2120.70 | 7116.87 ± 691.86 | 1110.40 ± 173.14 | 130.0 |
| Contr_l + PMI_L + PR_L + Triplet_L | GIN | 20357.38 ± 3244.23 | 9077.03 ± 942.22 | 1592.04 ± 310.16 | 69.3 |
| Contr_l + PMI_L + PR_L + Triplet_L | MPNN | 16224.21 ± 2259.90 | 10422.18 ± 928.82 | 1044.65 ± 360.74 | 84.7 |
| Contr_l + PMI_L + PR_L + Triplet_L | PAGNN | 7390.06 ± 263.06 | 8381.83 ± 954.85 | 104.44 ± 2.32 | 162.7 |
| Contr_l + PMI_L + PR_L + Triplet_L | SAGE | 24684.55 ± 2836.03 | 12751.06 ± 770.93 | 1220.28 ± 608.91 | 47.7 |
| Contr_l + PR_L | ALL | 18890.93 ± 6349.24 | 9661.72 ± 3234.84 | 606.17 ± 87.22 | 96.7 |
| Contr_l + PR_L | GAT | 25767.51 ± 18130.14 | 9376.47 ± 2123.15 | 1627.37 ± 310.53 | 52.7 |
| Contr_l + PR_L | GCN | 23674.58 ± 2483.87 | 8340.41 ± 2682.44 | 1005.60 ± 302.56 | 84.7 |
| Contr_l + PR_L | GIN | 24863.59 ± 8133.08 | 10768.22 ± 3604.50 | 547.55 ± 204.87 | 78.0 |
| Contr_l + PR_L | MPNN | 34764.46 ± 10988.41 | 14065.32 ± 2164.39 | 391.93 ± 82.18 | 63.7 |
| Contr_l + PR_L | PAGNN | 12003.40 ± 836.76 | 11188.88 ± 858.43 | 63.33 ± 12.28 | 132.7 |
| Contr_l + PR_L | SAGE | 23484.00 ± 9436.75 | 12580.61 ± 4945.11 | 354.68 ± 232.93 | 78.3 |
| Contr_l + PR_L + Triplet_L | ALL | 24655.94 ± 6414.50 | 13289.51 ± 1725.12 | 3197.13 ± 830.95 | 19.7 |
| Contr_l + PR_L + Triplet_L | GAT | 17602.36 ± 4525.57 | 9366.94 ± 722.27 | 1807.67 ± 324.42 | 65.0 |

Calinski Harabasz Continued (↑)

| Loss Type | Model | CORA | Citeseer | Bitcoin Fraud Transaction | Average Rank |
|---|---|---|---|---|---|
| Contr_l + PR_L + Triplet_L | GCN | 17068.15 ± 2000.11 | 6888.89 ± 1013.40 | 882.15 ± 174.33 | 114.7 |
| Contr_l + PR_L + Triplet_L | GIN | 19088.89 ± 4027.37 | 10009.48 ± 760.02 | 2049.46 ± 1082.16 | 54.0 |
| Contr_l + PR_L + Triplet_L | MPNN | 30881.43 ± 9588.49 | 12929.37 ± 2193.18 | 1070.46 ± 308.67 | 47.3 |
| Contr_l + PR_L + Triplet_L | PAGNN | 13372.05 ± 2619.52 | 12259.64 ± 1582.12 | 79.41 ± 4.61 | 121.3 |
| Contr_l + PR_L + Triplet_L | SAGE | 23376.59 ± 2379.72 | 11740.92 ± 1147.75 | 1390.26 ± 396.58 | 48.3 |
| Contr_l + Triplet_L | ALL | 25726.39 ± 2284.07 | 14417.78 ± 635.03 | 1090.04 ± 288.13 | 48.0 |
| Contr_l + Triplet_L | GAT | 16096.97 ± 1293.13 | 8669.56 ± 546.71 | 962.99 ± 115.11 | 101.7 |
| Contr_l + Triplet_L | GCN | 15678.69 ± 1345.86 | 9204.60 ± 945.86 | 1123.42 ± 227.97 | 94.0 |
| Contr_l + Triplet_L | GIN | 18313.37 ± 748.22 | 9661.93 ± 509.24 | <mark>2774.79 ± 603.68</mark> | 49.7 |
| Contr_l + Triplet_L | MPNN | 22110.98 ± 1368.34 | 11447.90 ± 1073.44 | 1313.94 ± 317.42 | 56.0 |
| Contr_l + Triplet_L | PAGNN | 21625.21 ± 6522.00 | 13989.06 ± 693.60 | 114.42 ± 4.96 | 88.3 |
| Contr_l + Triplet_L | SAGE | 20389.92 ± 360.64 | 11416.24 ± 524.63 | 1096.20 ± 292.41 | 66.3 |
| CrossE_L | ALL | 3435.38 ± 2817.76 | 5206.33 ± 1583.14 | 749.08 ± 177.84 | 173.3 |
| CrossE_L | GAT | 4560.58 ± 1525.73 | 2661.05 ± 418.34 | 2381.54 ± 528.46 | 136.7 |
| CrossE_L | GCN | 982.00 ± 356.36 | 818.82 ± 108.94 | 752.13 ± 419.40 | 184.7 |
| CrossE_L | GIN | 2348.53 ± 2437.40 | 2487.52 ± 178.75 | 636.18 ± 40.73 | 183.7 |

Calinski Harabasz Continued (↑)

| Loss Type | Model | CORA | Citeseer | Bitcoin Fraud Transaction | Average Rank |
|---|---|---|---|---|---|
| CrossE_L | MPNN | 3544.51 ± 177.72 | 6159.57 ± 1953.17 | 513.98 ± 221.39 | 176.0 |
| CrossE_L | PAGNN | 5563.76 ± 731.44 | 4532.62 ± 494.46 | 235.18 ± 5.29 | 186.0 |
| CrossE_L | SAGE | 2317.83 ± 281.62 | 509.18 ± 52.22 | 728.81 ± 783.27 | 185.3 |
| CrossE_L + PMI_L | ALL | 7520.46 ± 1318.87 | 5865.95 ± 298.13 | 1866.05 ± 580.67 | 130.3 |
| CrossE_L + PMI_L | GAT | 8609.62 ± 797.77 | 5528.14 ± 569.92 | 2284.60 ± 845.66 | 120.0 |
| CrossE_L + PMI_L | GCN | 6209.96 ± 378.18 | 4028.71 ± 367.02 | 1220.94 ± 177.47 | 156.3 |
| CrossE_L + PMI_L | GIN | 12021.39 ± 1795.75 | 6629.31 ± 337.38 | 837.75 ± 339.31 | 141.7 |
| CrossE_L + PMI_L | MPNN | 12400.71 ± 2030.03 | 6515.73 ± 983.11 | 1830.65 ± 444.67 | 112.0 |
| CrossE_L + PMI_L | PAGNN | 7825.79 ± 580.04 | 6780.57 ± 205.73 | 114.50 ± 8.30 | 171.3 |
| CrossE_L + PMI_L | SAGE | 2029.22 ± 343.81 | 1473.30 ± 170.88 | 805.02 ± 291.31 | 182.3 |
| CrossE_L + PMI_L + PR_L | ALL | 5641.33 ± 466.62 | 5776.62 ± 65.31 | 202.54 ± 26.02 | 184.0 |
| CrossE_L + PMI_L + PR_L | GAT | 12187.79 ± 918.68 | 8176.66 ± 1021.40 | 1218.44 ± 827.84 | 110.7 |
| CrossE_L + PMI_L + PR_L | GCN | 7838.60 ± 1420.46 | 4090.52 ± 631.72 | 1009.04 ± 291.75 | 158.0 |
| CrossE_L + PMI_L + PR_L | GIN | 16142.02 ± 2691.19 | 6277.69 ± 569.27 | 399.29 ± 527.81 | 141.0 |
| CrossE_L + PMI_L + PR_L | MPNN | 13954.37 ± 3157.07 | 7420.93 ± 946.44 | 1053.27 ± 815.24 | 116.3 |
| CrossE_L + PMI_L + PR_L | PAGNN | 8018.23 ± 2002.06 | 6846.71 ± 258.66 | 91.84 ± 3.23 | 172.0 |

Calinski Harabasz Continued (↑)

| Loss Type | Model | CORA | Citeseer | Bitcoin Fraud Transaction | Average Rank |
|-----------|-------|------|----------|---------------------------|--------------|
| CrossE_L + PMI_L + PR_L | SAGE | 2837.17 ± 252.58 | 1659.71 ± 215.15 | 208.00 ± 63.81 | 194.7 |
| CrossE_L + PMI_L + PR_L + Triplet_L | ALL | 19574.20 ± 1461.74 | 10181.99 ± 1041.71 | 1226.23 ± 406.59 | 69.3 |
| CrossE_L + PMI_L + PR_L + Triplet_L | GAT | 13761.21 ± 2265.53 | 8002.05 ± 293.87 | 1599.65 ± 657.58 | 95.7 |
| CrossE_L + PMI_L + PR_L + Triplet_L | GCN | 7491.42 ± 1253.82 | 4722.35 ± 785.61 | 1069.61 ± 287.60 | 157.3 |
| CrossE_L + PMI_L + PR_L + Triplet_L | GIN | 15914.45 ± 3152.94 | 8331.14 ± 666.34 | 1251.49 ± 700.01 | 92.7 |
| CrossE_L + PMI_L + PR_L + Triplet_L | MPNN | 12149.14 ± 986.66 | 8517.58 ± 1258.38 | 1397.79 ± 536.97 | 102.7 |
| CrossE_L + PMI_L + PR_L + Triplet_L | PAGNN | 7841.09 ± 751.36 | 7238.85 ± 317.16 | 108.10 ± 1.05 | 165.7 |
| CrossE_L + PMI_L + PR_L + Triplet_L | SAGE | 13780.88 ± 5549.44 | 10823.79 ± 2744.73 | 467.32 ± 134.57 | 109.7 |
| CrossE_L + PMI_L + Triplet_L | ALL | 22660.30 ± 1741.97 | 11380.85 ± 837.06 | 2256.77 ± 305.62 | 36.0 |
| CrossE_L + PMI_L + Triplet_L | GAT | 9765.70 ± 993.25 | 6642.51 ± 438.06 | 1961.82 ± 280.55 | 115.0 |
| CrossE_L + PMI_L + Triplet_L | GCN | 7244.26 ± 1009.12 | 4140.44 ± 153.85 | 1135.51 ± 267.73 | 156.3 |
| CrossE_L + PMI_L + Triplet_L | GIN | 12658.27 ± 2593.68 | 8668.76 ± 1259.75 | 2131.36 ± 1094.70 | 86.0 |
| CrossE_L + PMI_L + Triplet_L | MPNN | 14833.11 ± 535.14 | 7551.55 ± 1016.13 | 2122.16 ± 557.89 | 85.7 |
| CrossE_L + PMI_L + Triplet_L | PAGNN | 7672.53 ± 758.43 | 7187.92 ± 141.67 | 116.73 ± 1.80 | 164.7 |
| CrossE_L + PMI_L + Triplet_L | SAGE | 16548.09 ± 3477.77 | 9713.24 ± 1205.05 | 1758.09 ± 299.67 | 65.3 |
| CrossE_L + PR_L | ALL | 18215.78 ± 4897.35 | 13630.95 ± 1779.47 | 527.96 ± 44.18 | 84.0 |

Continued on next page

Calinski Harabasz Continued (↑)

| Loss Type | Model | CORA | Citeseer | Bitcoin Fraud Transaction | Average Rank |
|---|---|---|---|---|---|
| CrossE_L + PR_L | GAT | 23360.95 ± 18110.54 | 6153.28 ± 1268.98 | 534.87 ± 606.42 | 122.0 |
| CrossE_L + PR_L | GCN | 47621.37 ± 8824.25 | 11019.34 ± 1099.46 | 308.30 ± 66.91 | 73.3 |
| CrossE_L + PR_L | GIN | 17014.76 ± 6739.24 | 14068.37 ± 3290.72 | 621.44 ± 199.93 | 81.0 |
| CrossE_L + PR_L | MPNN | 17507.33 ± 8433.11 | 15859.40 ± 1795.85 | 317.21 ± 33.87 | 83.7 |
| CrossE_L + PR_L | PAGNN | 10403.89 ± 1197.62 | 9096.98 ± 2123.87 | 111.52 ± 22.17 | 144.7 |
| CrossE_L + PR_L | SAGE | 6966.26 ± 1286.79 | 2468.39 ± 684.93 | 773.21 ± 211.00 | 174.3 |
| CrossE_L + PR_L + Triplet_L | ALL | 19737.26 ± 9021.11 | 10256.28 ± 4854.88 | 2598.76 ± 1540.62 | 43.3 |
| CrossE_L + PR_L + Triplet_L | GAT | 29972.48 ± 7523.17 | 9293.00 ± 579.77 | 1605.75 ± 327.14 | 49.3 |
| CrossE_L + PR_L + Triplet_L | GCN | 19233.05 ± 3541.85 | 6617.58 ± 588.44 | 1047.03 ± 309.44 | 110.7 |
| CrossE_L + PR_L + Triplet_L | GIN | 19991.01 ± 2620.73 | 9212.00 ± 550.50 | 700.89 ± 278.82 | 96.3 |
| CrossE_L + PR_L + Triplet_L | MPNN | 40659.66 ± 3276.94 | 17530.56 ± 1188.88 | 809.53 ± 184.17 | 46.3 |
| CrossE_L + PR_L + Triplet_L | PAGNN | 13335.22 ± 2533.70 | 9996.39 ± 1140.64 | 80.14 ± 3.86 | 132.7 |
| CrossE_L + PR_L + Triplet_L | SAGE | 19499.33 ± 1612.47 | 11341.37 ± 1567.28 | 1064.64 ± 531.60 | 71.7 |
| CrossE_L + Triplet_L | ALL | 16966.27 ± 1902.44 | 9107.35 ± 625.86 | 1721.54 ± 453.18 | 71.7 |
| CrossE_L + Triplet_L | GAT | 11097.33 ± 899.03 | 6415.05 ± 233.72 | 1182.63 ± 294.99 | 132.3 |

Calinski Harabasz Continued (↑)

| Loss Type | Model | CORA | Citeseer | Bitcoin Fraud Transaction | Average Rank |
|---|---|---|---|---|---|
| CrossE_L + Triplet_L | GCN | 11544.87 ± 595.87 | 6387.93 ± 366.40 | 1324.10 ± 189.11 | 128.3 |
| CrossE_L + Triplet_L | GIN | 13187.24 ± 951.81 | 7020.21 ± 292.62 | 2338.58 ± 350.87 | 93.3 |
| CrossE_L + Triplet_L | MPNN | 15191.42 ± 914.52 | 7671.90 ± 939.77 | 1719.05 ± 415.45 | 89.7 |
| CrossE_L + Triplet_L | PAGNN | 17066.67 ± 4443.26 | 10801.62 ± 1449.04 | 108.98 ± 1.80 | 108.0 |
| CrossE_L + Triplet_L | SAGE | 13881.00 ± 801.82 | 9066.30 ± 554.19 | 1375.73 ± 270.05 | 93.3 |
| PMI_L | ALL | 6151.28 ± 670.81 | 5593.08 ± 277.38 | 1974.75 ± 750.52 | 135.0 |
| PMI_L | GAT | 8226.01 ± 1220.07 | 5433.50 ± 772.63 | 2521.63 ± 658.75 | 118.7 |
| PMI_L | GCN | 7155.13 ± 1058.65 | 3447.88 ± 854.48 | 1060.08 ± 96.74 | 163.3 |
| PMI_L | GIN | 15914.01 ± 3193.62 | 6784.74 ± 375.27 | 895.23 ± 503.85 | 121.7 |
| PMI_L | MPNN | 13343.50 ± 1266.43 | 6446.06 ± 662.09 | 1630.98 ± 815.84 | 112.0 |
| PMI_L | PAGNN | 7947.61 ± 1222.01 | 6561.48 ± 258.18 | 117.30 ± 3.70 | 171.3 |
| PMI_L | SAGE | 2046.74 ± 67.48 | 1404.40 ± 85.47 | 884.58 ± 180.73 | 179.3 |
| PMI_L + PR_L | ALL | 5574.35 ± 393.07 | 6068.59 ± 618.09 | 205.82 ± 26.59 | 182.3 |
| PMI_L + PR_L | GAT | 16384.05 ± 2016.02 | 7460.26 ± 965.09 | 297.49 ± 107.24 | 127.0 |
| PMI_L + PR_L | GCN | 8614.72 ± 1452.69 | 4475.82 ± 956.53 | 979.08 ± 275.47 | 154.0 |
| PMI_L + PR_L | GIN | 15884.10 ± 2383.46 | 6823.50 ± 1738.22 | 140.75 ± 15.19 | 141.0 |

Calinski Harabasz Continued (↑)

| Loss Type | Model | CORA | Citeseer | Bitcoin Fraud Transaction | Average Rank |
|---|---|---|---|---|---|
| PMI_L + PR_L | MPNN | 13660.21 ± 1716.10 | 8095.57 ± 378.90 | 452.87 ± 496.33 | 132.0 |
| PMI_L + PR_L | PAGNN | 7518.36 ± 684.36 | 6989.21 ± 301.49 | 88.55 ± 1.42 | 175.0 |
| PMI_L + PR_L | SAGE | 3038.60 ± 772.84 | 1626.77 ± 385.95 | 196.45 ± 37.16 | 195.3 |
| PMI_L + PR_L + Triplet_L | ALL | 21965.71 ± 1652.38 | 10147.74 ± 527.97 | 828.87 ± 245.87 | 81.7 |
| PMI_L + PR_L + Triplet_L | GAT | 15685.17 ± 1242.40 | 7813.41 ± 459.72 | 1323.01 ± 242.55 | 97.0 |
| PMI_L + PR_L + Triplet_L | GCN | 9144.45 ± 1149.18 | 5443.55 ± 1642.15 | 1145.71 ± 259.35 | 143.7 |
| PMI_L + PR_L + Triplet_L | GIN | 15564.53 ± 2931.41 | 9208.33 ± 1035.60 | 1711.20 ± 777.13 | 77.0 |
| PMI_L + PR_L + Triplet_L | MPNN | 13276.63 ± 2104.80 | 9204.65 ± 489.25 | 1205.46 ± 611.03 | 99.7 |
| PMI_L + PR_L + Triplet_L | PAGNN | 8350.96 ± 703.27 | 7150.82 ± 113.79 | 105.17 ± 1.85 | 165.3 |
| PMI_L + PR_L + Triplet_L | SAGE | 16202.55 ± 7170.33 | 10059.73 ± 827.88 | 674.81 ± 312.09 | 99.3 |
| PMI_L + Triplet_L | ALL | 23923.04 ± 1830.79 | 12575.87 ± 397.09 | 2306.15 ± 163.35 | 27.7 |
| PMI_L + Triplet_L | GAT | 10230.60 ± 841.99 | 6379.65 ± 625.52 | 2031.97 ± 415.71 | 116.3 |
| PMI_L + Triplet_L | GCN | 7340.87 ± 821.02 | 3910.53 ± 671.61 | 1145.53 ± 305.18 | 157.3 |
| PMI_L + Triplet_L | GIN | 16204.91 ± 2952.99 | 9170.81 ± 2017.81 | 2299.47 ± 615.73 | 65.3 |
| PMI_L + Triplet_L | MPNN | 12782.15 ± 725.20 | 7670.54 ± 1267.88 | 1700.99 ± 462.88 | 100.7 |
| PMI_L + Triplet_L | PAGNN | 7716.90 ± 1051.33 | 7047.53 ± 278.82 | 118.28 ± 1.23 | 165.0 |

Calinski Harabasz Continued (↑)

| Loss Type | Model | CORA | Citeseer | Bitcoin Fraud Transaction | Average Rank |
|---|---|---|---|---|---|
| PMI_L + Triplet_L | SAGE | 14236.29 ± 3044.90 | 10175.41 ± 851.19 | 2697.27 ± 742.10 | 60.0 |
| PR_L | ALL | 14858.10 ± 3396.76 | 10860.66 ± 1466.30 | 540.19 ± 80.53 | 102.7 |
| PR_L | GAT | 12313.02 ± 2537.39 | 6958.42 ± 2319.29 | 461.80 ± 350.72 | 147.0 |
| PR_L | GCN | 43394.23 ± 16594.70 | 10782.26 ± 1362.22 | 1177.70 ± 533.41 | 47.7 |
| PR_L | GIN | 14651.04 ± 6850.02 | 14268.96 ± 2421.08 | 727.60 ± 92.50 | 88.3 |
| PR_L | MPNN | 28955.72 ± 7594.74 | 15390.33 ± 1067.28 | 507.95 ± 424.72 | 61.3 |
| PR_L | PAGNN | 10414.08 ± 1649.03 | 8663.63 ± 1345.96 | 123.38 ± 3.50 | 143.7 |
| PR_L | SAGE | 6253.20 ± 1051.70 | 2234.23 ± 380.32 | 835.09 ± 173.06 | 173.7 |
| PR_L + Triplet_L | ALL | 20608.82 ± 8374.88 | 13674.47 ± 1243.05 | 600.98 ± 95.91 | 76.0 |
| PR_L + Triplet_L | GAT | 29434.41 ± 14970.24 | 9306.14 ± 2182.79 | 1324.54 ± 564.08 | 56.7 |
| PR_L + Triplet_L | GCN | 22385.70 ± 4850.28 | 7545.84 ± 1552.09 | 846.37 ± 304.92 | 99.3 |
| PR_L + Triplet_L | GIN | 23196.90 ± 3258.37 | 10570.74 ± 2630.45 | 756.27 ± 182.14 | 77.7 |
| PR_L + Triplet_L | MPNN | 32548.01 ± 8204.72 | 14451.15 ± 3240.73 | 747.43 ± 414.98 | 53.7 |
| PR_L + Triplet_L | PAGNN | 12919.65 ± 1058.83 | 9918.58 ± 514.56 | 75.62 ± 21.07 | 135.3 |
| PR_L + Triplet_L | SAGE | 10819.74 ± 4382.79 | 7780.35 ± 5528.34 | 798.24 ± 255.70 | 133.3 |

Calinski Harabasz Continued (↑)

| Loss Type | Model | CORA | Citeseer | Bitcoin Fraud Transaction | Average Rank |
|-----------|-------|------|----------|---------------------------|--------------|
| Triplet_L | ALL | 17380.98 ± 1541.17 | 8998.69 ± 167.45 | 1744.64 ± 330.38 | 71.3 |
| Triplet_L | GAT | 9612.20 ± 638.13 | 5806.41 ± 356.72 | 1427.11 ± 179.13 | 131.7 |
| Triplet_L | GCN | 12155.05 ± 453.14 | 6527.63 ± 501.18 | 1150.79 ± 180.69 | 130.0 |
| Triplet_L | GIN | 12827.87 ± 1490.94 | 7217.78 ± 1149.76 | 2999.23 ± 586.81 | 87.3 |
| Triplet_L | MPNN | 17316.45 ± 1349.06 | 8161.19 ± 810.14 | 1833.58 ± 231.91 | 75.0 |
| Triplet_L | PAGNN | 16817.63 ± 4908.89 | 10200.80 ± 856.41 | 108.94 ± 1.43 | 113.0 |
| Triplet_L | SAGE | 14102.30 ± 1046.44 | 8849.24 ± 646.34 | 1419.47 ± 191.44 | 90.3 |

Table 18. Knn Consistency Performance (↑): Top-ranked results are highlighted in **1st**, second-ranked in **2nd**, and third-ranked in **3rd**.

| Loss Type | Model | CORA | Citeseer | Bitcoin Fraud Transaction | Average Rank |
|-----------|-------|------|----------|---------------------------|--------------|
| Contr_l | ALL | 78.76 ± 0.71 | 66.82 ± 0.34 | 72.95 ± 0.37 | 139.7 |
| Contr_l | GAT | 85.84 ± 0.27 | 71.85 ± 0.26 | 74.23 ± 0.09 | 44.7 |
| Contr_l | GCN | 84.52 ± 0.11 | 70.49 ± 0.77 | 73.55 ± 0.22 | 66.0 |
| Contr_l | GIN | 81.64 ± 0.62 | 69.80 ± 0.54 | 75.34 ± 0.73 | 75.0 |
| Contr_l | MPNN | 82.96 ± 0.27 | 70.24 ± 0.47 | 74.77 ± 0.38 | 66.0 |

Knn Consistency Continued (↑)

| Loss Type | Model | CORA | | Citeseer | | Bitcoin Fraud Transaction | | Average Rank |
|---|---|---|---|---|---|---|---|---|
| Contr_l | PAGNN | 80.20 | ± 0.65 | 68.33 | ± 0.91 | 63.76 | ± 0.23 | 140.3 |
| Contr_l | SAGE | 84.42 | ± 0.57 | 70.42 | ± 0.79 | 74.68 | ± 0.30 | 57.7 |
| Contr_l + CrossE_L | ALL | 78.33 | ± 0.49 | 66.19 | ± 0.33 | 72.59 | ± 0.75 | 142.0 |
| Contr_l + CrossE_L | GAT | 85.62 | ± 0.16 | 71.88 | ± 0.31 | 74.10 | ± 0.31 | 46.3 |
| Contr_l + CrossE_L | GCN | 84.99 | ± 0.16 | 70.19 | ± 0.63 | 73.45 | ± 0.15 | 67.3 |
| Contr_l + CrossE_L | GIN | 80.59 | ± 0.42 | 68.93 | ± 0.36 | 75.42 | ± 0.63 | 87.3 |
| Contr_l + CrossE_L | MPNN | 83.21 | ± 0.87 | 70.45 | ± 0.27 | 74.18 | ± 0.19 | 68.7 |
| Contr_l + CrossE_L | PAGNN | 79.65 | ± 0.56 | 67.50 | ± 0.55 | 63.76 | ± 0.12 | 146.0 |
| Contr_l + CrossE_L | SAGE | 83.98 | ± 0.41 | 70.71 | ± 0.51 | 74.32 | ± 0.38 | 57.7 |
| Contr_l + CrossE_L + PMI_L | ALL | 77.45 | ± 1.50 | 66.05 | ± 0.81 | 74.62 | ± 0.12 | 125.0 |
| Contr_l + CrossE_L + PMI_L | GAT | 84.35 | ± 0.35 | 70.62 | ± 0.32 | 75.11 | ± 0.21 | 47.0 |
| Contr_l + CrossE_L + PMI_L | GCN | 82.28 | ± 0.47 | 68.22 | ± 0.33 | 73.66 | ± 0.21 | 110.0 |
| Contr_l + CrossE_L + PMI_L | GIN | 76.61 | ± 0.58 | 63.50 | ± 0.75 | 75.41 | ± 0.26 | 124.3 |
| Contr_l + CrossE_L + PMI_L | MPNN | 82.05 | ± 0.17 | 69.69 | ± 0.21 | 75.49 | ± 0.38 | 66.3 |
| Contr_l + CrossE_L + PMI_L | PAGNN | 73.65 | ± 0.52 | 59.71 | ± 0.47 | 64.02 | ± 0.13 | 184.3 |
| Contr_l + CrossE_L + PMI_L | SAGE | 61.02 | ± 1.52 | 55.88 | ± 2.12 | 78.03 | ± 0.89 | 139.0 |

Knn Consistency Continued (↑)

| Loss Type | Model | CORA | | Citeseer | | Bitcoin Fraud Transaction | | Average Rank |
|---|---|---|---|---|---|---|---|---|
| Contr_l + CrossE_L + PMI_L + PR_L | ALL | 74.76 1.59 | ± | 64.34 0.40 | ± | 75.66 0.19 | ± | 118.3 |
| Contr_l + CrossE_L + PMI_L + PR_L | GAT | 84.44 0.32 | ± | 70.52 0.27 | ± | 75.36 0.55 | ± | 43.3 |
| Contr_l + CrossE_L + PMI_L + PR_L | GCN | 82.88 0.17 | ± | 68.32 0.34 | ± | 73.51 0.23 | ± | 103.7 |
| Contr_l + CrossE_L + PMI_L + PR_L | GIN | 75.96 1.03 | ± | 62.05 0.92 | ± | 75.10 0.38 | ± | 137.7 |
| Contr_l + CrossE_L + PMI_L + PR_L | MPNN | 81.87 0.24 | ± | 68.82 0.62 | ± | 75.87 0.19 | ± | 66.7 |
| Contr_l + CrossE_L + PMI_L + PR_L | PAGNN | 74.19 0.75 | ± | 59.58 0.57 | ± | 63.86 0.11 | ± | 185.7 |
| Contr_l + CrossE_L + PMI_L + PR_L | SAGE | 61.23 1.85 | ± | 56.70 2.96 | ± | 78.34 0.68 | ± | 135.7 |
| Contr_l + CrossE_L + PMI_L + PR_L + Triplet_L | ALL | 77.30 0.70 | ± | 65.16 0.72 | ± | 74.31 0.20 | ± | 132.0 |
| Contr_l + CrossE_L + PMI_L + PR_L + Triplet_L | GAT | 84.72 0.26 | ± | 70.58 0.39 | ± | 74.97 0.23 | ± | 45.3 |
| Contr_l + CrossE_L + PMI_L + PR_L + Triplet_L | GCN | 82.70 0.47 | ± | 68.48 0.35 | ± | 73.42 0.29 | ± | 103.7 |
| Contr_l + CrossE_L + PMI_L + PR_L + Triplet_L | GIN | 76.28 0.53 | ± | 64.42 0.77 | ± | 75.59 0.18 | ± | 113.7 |
| Contr_l + CrossE_L + PMI_L + PR_L + Triplet_L | MPNN | 82.17 0.36 | ± | 68.89 0.32 | ± | 75.46 0.30 | ± | 77.0 |
| Contr_l + CrossE_L + PMI_L + PR_L + Triplet_L | PAGNN | 73.94 0.92 | ± | 60.42 0.37 | ± | 63.94 0.07 | ± | 180.7 |
| Contr_l + CrossE_L + PMI_L + PR_L + Triplet_L | SAGE | 75.36 2.75 | ± | 67.37 1.08 | ± | 78.04 0.83 | ± | 94.3 |
| Contr_l + CrossE_L + PMI_L + Triplet_L | ALL | 80.33 0.43 | ± | 69.35 0.53 | ± | 74.56 0.34 | ± | 96.3 |
| Contr_l + CrossE_L + PMI_L + Triplet_L | GAT | 84.57 0.17 | ± | 70.64 0.38 | ± | 75.10 0.23 | ± | 44.0 |

Knn Consistency Continued (↑)

| Loss Type | Model | CORA | | Citeseer | | Bitcoin Fraud Transaction | | Average Rank |
|---|---|---|---|---|---|---|---|---|
| Contr_l + CrossE_L + PMI_L + Triplet_L | GCN | 82.62 | ± | 68.46 | ± | 73.59 | ± | 102.0 |
| | | 0.47 | | 0.50 | | 0.27 | | |
| Contr_l + CrossE_L + PMI_L + Triplet_L | GIN | 76.60 | ± | 64.81 | ± | 75.57 | ± | 113.0 |
| | | 0.50 | | 0.24 | | 0.55 | | |
| Contr_l + CrossE_L + PMI_L + Triplet_L | MPNN | 82.31 | ± | 69.33 | ± | 75.50 | ± | 67.0 |
| | | 0.34 | | 0.28 | | 0.37 | | |
| Contr_l + CrossE_L + PMI_L + Triplet_L | PAGNN | 74.34 | ± | 60.38 | ± | 63.91 | ± | 179.7 |
| | | 1.26 | | 0.60 | | 0.19 | | |
| Contr_l + CrossE_L + PMI_L + Triplet_L | SAGE | 75.84 | ± | 66.15 | ± | 76.25 | ± | 99.0 |
| | | 2.57 | | 0.92 | | 0.94 | | |
| Contr_l + CrossE_L + PR_L | ALL | 74.59 | ± | 58.68 | ± | 75.54 | ± | 140.7 |
| | | 0.82 | | 2.63 | | 0.41 | | |
| Contr_l + CrossE_L + PR_L | GAT | 83.74 | ± | 70.20 | ± | 74.07 | ± | 71.0 |
| | | 1.09 | | 0.50 | | 0.23 | | |
| Contr_l + CrossE_L + PR_L | GCN | 82.41 | ± | 68.64 | ± | 73.18 | ± | 110.0 |
| | | 0.43 | | 0.15 | | 0.15 | | |
| Contr_l + CrossE_L + PR_L | GIN | 74.87 | ± | 63.27 | ± | 75.43 | ± | 132.0 |
| | | 1.51 | | 0.84 | | 0.65 | | |
| Contr_l + CrossE_L + PR_L | MPNN | 80.28 | ± | 69.02 | ± | 75.52 | ± | 81.7 |
| | | 0.19 | | 0.46 | | 0.45 | | |
| Contr_l + CrossE_L + PR_L | PAGNN | 73.80 | ± | 58.49 | ± | 63.82 | ± | 193.0 |
| | | 0.78 | | 0.68 | | 0.13 | | |
| Contr_l + CrossE_L + PR_L | SAGE | 76.71 | ± | 64.35 | ± | 78.16 | ± | 98.0 |
| | | 7.50 | | 5.31 | | 0.80 | | |
| Contr_l + CrossE_L + PR_L + Triplet_L | ALL | 74.37 | ± | 64.64 | ± | 73.69 | ± | 152.7 |
| | | 0.47 | | 0.44 | | 0.59 | | |
| Contr_l + CrossE_L + PR_L + Triplet_L | GAT | 85.17 | ± | 71.00 | ± | 74.48 | ± | 45.7 |
| | | 0.11 | | 0.55 | | 0.14 | | |
| Contr_l + CrossE_L + PR_L + Triplet_L | GCN | 83.24 | ± | 69.01 | ± | 73.41 | ± | 92.7 |
| | | 0.35 | | 0.39 | | 0.31 | | |
| Contr_l + CrossE_L + PR_L + Triplet_L | GIN | 78.78 | ± | 66.64 | ± | 75.60 | ± | 96.7 |
| | | 0.74 | | 0.73 | | 0.37 | | |

Knn Consistency Continued (↑)

| Loss Type | Model | CORA | | Citeseer | | Bitcoin Fraud Transaction | | Average Rank |
|---|---|---|---|---|---|---|---|---|
| Contr_l + CrossE_L + PR_L + Triplet_L | MPNN | 81.41 | ± | 69.26 | ± | 75.51 | ± | 75.0 |
| | | 0.16 | | 0.44 | | 0.23 | | |
| Contr_l + CrossE_L + PR_L + Triplet_L | PAGNN | 74.21 | ± | 63.23 | ± | 63.71 | ± | 182.7 |
| | | 0.75 | | 2.07 | | 0.09 | | |
| Contr_l + CrossE_L + PR_L + Triplet_L | SAGE | 83.46 | ± | 69.10 | ± | 75.54 | ± | 56.7 |
| | | 0.34 | | 0.81 | | 0.59 | | |
| Contr_l + CrossE_L + Triplet_L | ALL | 80.87 | ± | 68.58 | ± | 73.34 | ± | 121.7 |
| | | 0.78 | | 0.60 | | 0.36 | | |
| Contr_l + CrossE_L + Triplet_L | GAT | 85.54 | ± | 71.53 | ± | 74.20 | ± | 46.3 |
| | | 0.23 | | 0.37 | | 0.22 | | |
| Contr_l + CrossE_L + Triplet_L | GCN | 84.74 | ± | 70.43 | ± | 73.67 | ± | 62.3 |
| | | 0.27 | | 0.40 | | 0.19 | | |
| Contr_l + CrossE_L + Triplet_L | GIN | 82.51 | ± | 69.63 | ± | 75.60 | ± | 53.7 |
| | | 0.50 | | 0.27 | | 0.42 | | |
| Contr_l + CrossE_L + Triplet_L | MPNN | 83.57 | ± | 70.49 | ± | 74.81 | ± | 57.3 |
| | | 0.30 | | 0.29 | | 0.39 | | |
| Contr_l + CrossE_L + Triplet_L | PAGNN | 79.56 | ± | 69.00 | ± | 63.76 | ± | 132.3 |
| | | 1.74 | | 0.47 | | 0.15 | | |
| Contr_l + CrossE_L + Triplet_L | SAGE | 84.89 | ± | 70.34 | ± | 74.77 | ± | 51.7 |
| | | 0.48 | | 0.47 | | 0.29 | | |
| Contr_l + PMI_L | ALL | 77.17 | ± | 66.06 | ± | 74.62 | ± | 126.3 |
| | | 0.71 | | 0.64 | | 0.25 | | |
| Contr_l + PMI_L | GAT | 84.18 | ± | 70.55 | ± | 75.10 | ± | 50.0 |
| | | 0.22 | | 0.29 | | 0.19 | | |
| Contr_l + PMI_L | GCN | 82.15 | ± | 68.21 | ± | 73.41 | ± | 117.7 |
| | | 0.24 | | 0.34 | | 0.15 | | |
| Contr_l + PMI_L | GIN | 76.03 | ± | 63.90 | ± | 75.45 | ± | 124.0 |
| | | 0.56 | | 0.48 | | 0.43 | | |
| Contr_l + PMI_L | MPNN | 82.18 | ± | 69.57 | ± | 75.58 | ± | 61.7 |
| | | 0.34 | | 0.48 | | 0.38 | | |
| Contr_l + PMI_L | PAGNN | 73.52 | ± | 59.27 | ± | 63.82 | ± | 193.3 |
| | | 0.86 | | 0.50 | | 0.04 | | |

Knn Consistency Continued (↑)

| Loss Type | Model | CORA | | Citeseer | | Bitcoin Fraud Transaction | | Average Rank |
|---|---|---|---|---|---|---|---|---|
| Contr_l + PMI_L | SAGE | 61.31 ± 2.57 | | 62.19 ± 1.68 | | 77.65 ± 0.44 | | 127.7 |
| Contr_l + PMI_L + PR_L | ALL | 74.71 ± 1.20 | | 64.45 ± 0.56 | | 75.49 ± 0.41 | | 126.0 |
| Contr_l + PMI_L + PR_L | GAT | 84.62 ± 0.35 | | 70.25 ± 0.53 | | 75.47 ± 0.42 | | 42.3 |
| Contr_l + PMI_L + PR_L | GCN | 82.57 ± 0.36 | | 68.35 ± 0.43 | | 73.47 ± 0.23 | | 105.0 |
| Contr_l + PMI_L + PR_L | GIN | 75.75 ± 0.60 | | 63.42 ± 1.00 | | 75.59 ± 0.29 | | 121.0 |
| Contr_l + PMI_L + PR_L | MPNN | 82.37 ± 0.44 | | 68.31 ± 0.58 | | 75.60 ± 0.44 | | 75.0 |
| Contr_l + PMI_L + PR_L | PAGNN | 73.00 ± 1.44 | | 59.44 ± 0.73 | | 63.94 ± 0.15 | | 189.7 |
| Contr_l + PMI_L + PR_L | SAGE | 62.06 ± 2.09 | | 61.85 ± 2.37 | | 77.55 ± 0.56 | | 128.7 |
| Contr_l + PMI_L + PR_L + Triplet_L | ALL | 77.10 ± 0.68 | | 65.55 ± 0.79 | | 73.89 ± 0.20 | | 136.3 |
| Contr_l + PMI_L + PR_L + Triplet_L | GAT | 84.50 ± 0.23 | | 70.39 ± 0.46 | | 74.89 ± 0.27 | | 54.7 |
| Contr_l + PMI_L + PR_L + Triplet_L | GCN | 82.94 ± 0.47 | | 69.05 ± 0.41 | | 73.35 ± 0.28 | | 96.3 |
| Contr_l + PMI_L + PR_L + Triplet_L | GIN | 77.45 ± 0.34 | | 65.97 ± 1.23 | | 75.75 ± 0.28 | | 97.7 |
| Contr_l + PMI_L + PR_L + Triplet_L | MPNN | 82.38 ± 0.30 | | 68.70 ± 0.12 | | 75.30 ± 0.32 | | 81.3 |
| Contr_l + PMI_L + PR_L + Triplet_L | PAGNN | 74.65 ± 0.59 | | 61.65 ± 0.69 | | 63.92 ± 0.10 | | 174.7 |
| Contr_l + PMI_L + PR_L + Triplet_L | SAGE | 81.23 ± 0.56 | | 68.71 ± 0.60 | | 77.26 ± 1.39 | | 67.3 |
| Contr_l + PR_L | ALL | 74.78 ± 1.34 | | 60.19 ± 3.24 | | 75.45 ± 0.23 | | 139.7 |

Knn Consistency Continued (↑)

| Loss Type | Model | CORA | | Citeseer | | Bitcoin Fraud Transaction | | Average Rank |
|---|---|---|---|---|---|---|---|---|
| Contr_l + PR_L | GAT | 83.26 | ± 0.97 | 69.94 | ± 0.15 | 74.25 | ± 0.39 | 71.0 |
| Contr_l + PR_L | GCN | 82.33 | ± 0.23 | 68.64 | ± 0.48 | 73.47 | ± 0.35 | 107.0 |
| Contr_l + PR_L | GIN | 74.31 | ± 1.19 | 62.61 | ± 0.88 | 75.64 | ± 0.37 | 128.7 |
| Contr_l + PR_L | MPNN | 80.75 | ± 0.14 | 69.00 | ± 0.26 | 75.68 | ± 0.31 | 73.3 |
| Contr_l + PR_L | PAGNN | 73.68 | ± 0.41 | 58.70 | ± 0.57 | 63.74 | ± 0.15 | 196.3 |
| Contr_l + PR_L | SAGE | 77.29 | ± 6.67 | 65.75 | ± 4.12 | 77.92 | ± 0.82 | 90.7 |
| Contr_l + PR_L + Triplet_L | ALL | 73.64 | ± 1.48 | 63.48 | ± 1.56 | 73.32 | ± 0.41 | 171.7 |
| Contr_l + PR_L + Triplet_L | GAT | 84.79 | ± 0.84 | 71.04 | ± 0.44 | 74.45 | ± 0.10 | 47.7 |
| Contr_l + PR_L + Triplet_L | GCN | 82.96 | ± 0.32 | 68.88 | ± 0.46 | 73.36 | ± 0.17 | 98.7 |
| Contr_l + PR_L + Triplet_L | GIN | 78.52 | ± 0.92 | 65.61 | ± 0.74 | 75.66 | ± 0.45 | 99.3 |
| Contr_l + PR_L + Triplet_L | MPNN | 81.53 | ± 0.63 | 69.01 | ± 0.24 | 75.35 | ± 0.19 | 84.3 |
| Contr_l + PR_L + Triplet_L | PAGNN | 74.69 | ± 0.33 | 63.50 | ± 2.95 | 63.71 | ± 0.19 | 177.3 |
| Contr_l + PR_L + Triplet_L | SAGE | 82.95 | ± 0.47 | 69.24 | ± 0.93 | 75.69 | ± 0.72 | 51.0 |
| Contr_l + Triplet_L | ALL | 81.22 | ± 0.35 | 68.65 | ± 0.34 | 73.24 | ± 0.18 | 120.0 |
| Contr_l + Triplet_L | GAT | 85.39 | ± 0.24 | 71.53 | ± 0.19 | 74.24 | ± 0.17 | 46.0 |
| Contr_l + Triplet_L | GCN | 84.75 | ± 0.24 | 70.53 | ± 0.66 | 73.56 | ± 0.13 | 61.0 |

Knn Consistency Continued (↑)

| Loss Type | Model | CORA | | | Citeseer | | | Bitcoin Fraud Transaction | | | Average Rank |
|---|---|---|---|---|---|---|---|---|---|---|---|
| Contr_l + Triplet_L | GIN | 81.69 | ± | 0.62 | 69.23 | ± | 0.57 | 75.88 | ± | 0.35 | 61.3 |
| Contr_l + Triplet_L | MPNN | 83.29 | ± | 0.18 | 70.32 | ± | 0.30 | 74.81 | ± | 0.19 | 61.7 |
| Contr_l + Triplet_L | PAGNN | 80.08 | ± | 2.02 | 68.96 | ± | 0.35 | 63.84 | ± | 0.24 | 129.0 |
| Contr_l + Triplet_L | SAGE | 84.90 | ± | 0.22 | 70.54 | ± | 0.69 | 74.68 | ± | 0.59 | 48.7 |
| CrossE_L | ALL | 75.06 | ± | 4.54 | 64.40 | ± | 0.58 | 73.87 | ± | 0.37 | 149.0 |
| CrossE_L | GAT | 80.71 | ± | 2.25 | 68.21 | ± | 1.05 | 73.68 | ± | 0.29 | 120.3 |
| CrossE_L | GCN | 60.44 | ± | 2.85 | 49.90 | ± | 2.45 | 72.98 | ± | 0.43 | 196.7 |
| CrossE_L | GIN | 52.01 | ± | 1.34 | 45.27 | ± | 1.46 | 66.92 | ± | 4.55 | 199.7 |
| CrossE_L | MPNN | 74.43 | ± | 1.65 | 61.11 | ± | 2.33 | 74.79 | ± | 0.55 | 152.0 |
| CrossE_L | PAGNN | 68.70 | ± | 0.71 | 58.59 | ± | 0.29 | 63.91 | ± | 0.23 | 193.7 |
| CrossE_L | SAGE | 48.94 | ± | 1.76 | 47.03 | ± | 2.59 | 77.61 | ± | 2.36 | 144.7 |
| CrossE_L + PMI_L | ALL | 78.16 | ± | 0.38 | 65.51 | ± | 0.58 | 75.32 | ± | 0.15 | 115.7 |
| CrossE_L + PMI_L | GAT | 84.40 | ± | 0.42 | 70.55 | ± | 0.27 | 74.91 | ± | 0.29 | 51.0 |
| CrossE_L + PMI_L | GCN | 82.34 | ± | 0.52 | 68.26 | ± | 0.55 | 73.58 | ± | 0.19 | 108.7 |
| CrossE_L + PMI_L | GIN | 76.02 | ± | 0.82 | 63.66 | ± | 0.20 | 75.76 | ± | 0.16 | 112.0 |
| CrossE_L + PMI_L | MPNN | 82.51 | ± | 0.52 | 69.29 | ± | 0.23 | 75.53 | ± | 0.18 | 62.0 |

Knn Consistency Continued (↑)

| Loss Type | Model | CORA | | Citeseer | | Bitcoin Fraud Transaction | | Average Rank |
|---|---|---|---|---|---|---|---|---|
| CrossE_L + PMI_L | PAGNN | 73.97 | ± 0.67 | 59.51 | ± 0.53 | 63.89 | ± 0.13 | 186.3 |
| CrossE_L + PMI_L | SAGE | 58.15 | ± 2.16 | 53.81 | ± 1.30 | 77.38 | ± 0.66 | 143.3 |
| CrossE_L + PMI_L + PR_L | ALL | 75.82 | ± 0.61 | 64.14 | ± 0.48 | 75.61 | ± 0.26 | 116.0 |
| CrossE_L + PMI_L + PR_L | GAT | 84.53 | ± 0.19 | 69.82 | ± 0.24 | 75.22 | ± 0.81 | 53.0 |
| CrossE_L + PMI_L + PR_L | GCN | 82.66 | ± 0.32 | 68.14 | ± 0.18 | 73.51 | ± 0.19 | 108.0 |
| CrossE_L + PMI_L + PR_L | GIN | 75.27 | ± 0.31 | 61.86 | ± 0.71 | 75.35 | ± 0.42 | 137.0 |
| CrossE_L + PMI_L + PR_L | MPNN | 82.09 | ± 0.26 | 68.83 | ± 0.75 | 75.65 | ± 0.49 | 69.3 |
| CrossE_L + PMI_L + PR_L | PAGNN | 73.93 | ± 1.46 | 59.55 | ± 0.80 | 63.90 | ± 0.13 | 186.0 |
| CrossE_L + PMI_L + PR_L | SAGE | 61.54 | ± 1.74 | 55.60 | ± 1.07 | 78.30 | ± 0.74 | 136.0 |
| CrossE_L + PMI_L + PR_L + Triplet_L | ALL | 77.46 | ± 0.64 | 65.13 | ± 0.48 | 74.70 | ± 0.26 | 127.0 |
| CrossE_L + PMI_L + PR_L + Triplet_L | GAT | 84.74 | ± 0.33 | 70.43 | ± 0.30 | 75.01 | ± 0.19 | 48.7 |
| CrossE_L + PMI_L + PR_L + Triplet_L | GCN | 82.47 | ± 0.51 | 68.35 | ± 0.51 | 73.23 | ± 0.23 | 112.7 |
| CrossE_L + PMI_L + PR_L + Triplet_L | GIN | 76.56 | ± 0.66 | 64.35 | ± 0.62 | 75.60 | ± 0.30 | 113.7 |
| CrossE_L + PMI_L + PR_L + Triplet_L | MPNN | 82.02 | ± 0.35 | 69.19 | ± 0.95 | 75.35 | ± 0.48 | 79.3 |
| CrossE_L + PMI_L + PR_L + Triplet_L | PAGNN | 74.49 | ± 0.52 | 60.21 | ± 0.38 | 63.90 | ± 0.08 | 179.7 |
| CrossE_L + PMI_L + PR_L + Triplet_L | SAGE | 75.11 | ± 5.50 | 66.37 | ± 0.92 | 78.23 | ± 0.70 | 95.7 |

Knn Consistency Continued (↑)

| Loss Type | | | | Model | CORA | | Citeseer | | Bitcoin Fraud Transaction | | Average Rank |
|---|---|---|---|---|---|---|---|---|---|---|---|
| CrossE_L | + | PMI_L | + Triplet_L | ALL | 81.18 0.46 | ± | 69.48 0.22 | ± | 74.66 0.43 | ± | 91.3 |
| CrossE_L | + | PMI_L | + Triplet_L | GAT | 84.65 0.29 | ± | 70.70 0.28 | ± | 75.10 0.25 | ± | 43.7 |
| CrossE_L | + | PMI_L | + Triplet_L | GCN | 81.97 0.41 | ± | 68.35 0.25 | ± | 73.38 0.17 | ± | 117.7 |
| CrossE_L | + | PMI_L | + Triplet_L | GIN | 76.89 0.35 | ± | 65.23 0.57 | ± | 76.03 0.39 | ± | 100.0 |
| CrossE_L | + | PMI_L | + Triplet_L | MPNN | 82.36 0.17 | ± | 69.56 0.26 | ± | 75.58 0.24 | ± | 59.3 |
| CrossE_L | + | PMI_L | + Triplet_L | PAGNN | 74.83 0.70 | ± | 60.68 0.69 | ± | 63.85 0.19 | ± | 176.7 |
| CrossE_L | + | PMI_L | + Triplet_L | SAGE | 79.53 0.99 | ± | 67.50 1.04 | ± | 76.21 0.35 | ± | 87.0 |
| CrossE_L + PR_L | | | | ALL | 75.51 0.56 | ± | 64.01 0.66 | ± | 75.65 0.16 | ± | 117.0 |
| CrossE_L + PR_L | | | | GAT | 83.17 0.46 | ± | 69.53 0.30 | ± | 74.16 0.45 | ± | 78.7 |
| CrossE_L + PR_L | | | | GCN | 81.86 0.53 | ± | 68.20 0.39 | ± | 72.87 0.38 | ± | 126.7 |
| CrossE_L + PR_L | | | | GIN | 73.62 1.26 | ± | 62.76 0.41 | ± | 75.84 0.22 | ± | 127.3 |
| CrossE_L + PR_L | | | | MPNN | 80.84 0.71 | ± | 68.69 0.52 | ± | 75.67 0.25 | ± | 77.7 |
| CrossE_L + PR_L | | | | PAGNN | 73.09 0.70 | ± | 59.63 0.37 | ± | 63.78 0.15 | ± | 193.0 |
| CrossE_L + PR_L | | | | SAGE | 61.95 1.43 | ± | 55.11 1.08 | ± | <mark>78.50 0.74</mark> | ± | 135.7 |
| CrossE_L | + | PR_L | + Triplet_L | ALL | 73.35 1.79 | ± | 59.83 2.11 | ± | 74.54 0.72 | ± | 166.0 |
| CrossE_L | + | PR_L | + Triplet_L | GAT | 84.14 0.43 | ± | 70.80 0.84 | ± | 74.38 0.19 | ± | 56.7 |

Knn Consistency Continued (↑)

| Loss Type | | | | Model | CORA | | Citeseer | | Bitcoin Fraud Transaction | | Average Rank |
|---|---|---|---|---|---|---|---|---|---|---|---|
| CrossE_L | + | PR_L | + | GCN | 82.65 ± 0.25 | | 68.62 ± 0.23 | | 73.42 ± 0.39 | | 104.3 |
| Triplet_L | | | | | | | | | | | |
| CrossE_L | + | PR_L | + | GIN | 77.11 ± 1.35 | | 65.96 ± 0.95 | | 75.78 ± 0.43 | | 98.3 |
| Triplet_L | | | | | | | | | | | |
| CrossE_L | + | PR_L | + | MPNN | 81.15 ± 0.22 | | 68.76 ± 0.22 | | 75.37 ± 0.36 | | 88.3 |
| Triplet_L | | | | | | | | | | | |
| CrossE_L | + | PR_L | + | PAGNN | 73.88 ± 0.75 | | 60.68 ± 1.15 | | 63.82 ± 0.22 | | 186.0 |
| Triplet_L | | | | | | | | | | | |
| CrossE_L | + | PR_L | + | SAGE | 83.12 ± 1.20 | | 69.32 ± 0.88 | | 77.19 ± 0.82 | | 42.3 |
| Triplet_L | | | | | | | | | | | |
| CrossE_L + Triplet_L | | | | ALL | 81.98 ± 0.23 | | 69.23 ± 0.41 | | 73.99 ± 0.34 | | 97.0 |
| CrossE_L + Triplet_L | | | | GAT | 85.07 ± 0.16 | | 71.07 ± 0.33 | | 74.90 ± 0.09 | | 39.7 |
| CrossE_L + Triplet_L | | | | GCN | 84.24 ± 0.19 | | 69.93 ± 0.27 | | 73.45 ± 0.23 | | 77.0 |
| CrossE_L + Triplet_L | | | | GIN | 81.62 ± 0.39 | | 68.73 ± 0.17 | | 76.25 ± 0.38 | | 67.3 |
| CrossE_L + Triplet_L | | | | MPNN | 83.47 ± 0.35 | | 70.26 ± 0.46 | | 75.22 ± 0.18 | | 55.7 |
| CrossE_L + Triplet_L | | | | PAGNN | 80.25 ± 1.61 | | 69.22 ± 0.21 | | 63.77 ± 0.15 | | 126.7 |
| CrossE_L + Triplet_L | | | | SAGE | 84.79 ± 0.21 | | 70.55 ± 0.57 | | 75.25 ± 0.19 | | 40.7 |
| PMI_L | | | | ALL | 77.77 ± 0.23 | | 65.64 ± 0.59 | | 75.38 ± 0.53 | | 112.0 |
| PMI_L | | | | GAT | 84.42 ± 0.16 | | 70.47 ± 0.38 | | 74.93 ± 0.32 | | 52.7 |
| PMI_L | | | | GCN | 82.53 ± 0.65 | | 67.87 ± 0.28 | | 73.38 ± 0.31 | | 114.3 |
| PMI_L | | | | GIN | 76.30 ± 0.69 | | 63.31 ± 0.65 | | 75.42 ± 0.19 | | 126.3 |

Knn Consistency Continued ($\uparrow$)

| Loss Type | Model | CORA | | Citeseer | | Bitcoin Fraud Transaction | | Average Rank |
|---|---|---|---|---|---|---|---|---|
| PMI_L | MPNN | 82.34 | ± | 69.53 | ± | 75.52 | ± | 63.3 |
| | | 0.38 | | 0.10 | | 0.31 | | |
| PMI_L | PAGNN | 73.71 | ± | 59.46 | ± | 63.89 | ± | 188.7 |
| | | 0.94 | | 0.28 | | 0.12 | | |
| PMI_L | SAGE | 55.91 | ± | 53.17 | ± | 77.83 | ± | 142.3 |
| | | 1.41 | | 1.06 | | 0.72 | | |
| PMI_L + PR_L | ALL | 76.26 | ± | 64.04 | ± | 75.52 | ± | 120.0 |
| | | 0.39 | | 1.26 | | 0.13 | | |
| PMI_L + PR_L | GAT | 84.21 | ± | 69.64 | ± | 75.94 | ± | 36.0 |
| | | 0.23 | | 0.96 | | 0.31 | | |
| PMI_L + PR_L | GCN | 82.34 | ± | 68.35 | ± | 73.27 | ± | 115.3 |
| | | 0.38 | | 0.26 | | 0.26 | | |
| PMI_L + PR_L | GIN | 75.28 | ± | 62.65 | ± | 74.95 | ± | 141.3 |
| | | 0.83 | | 1.09 | | 0.48 | | |
| PMI_L + PR_L | MPNN | 82.16 | ± | 68.12 | ± | 75.77 | ± | 76.3 |
| | | 0.22 | | 0.10 | | 0.52 | | |
| PMI_L + PR_L | PAGNN | 74.01 | ± | 58.86 | ± | 63.82 | ± | 190.7 |
| | | 0.53 | | 0.82 | | 0.14 | | |
| PMI_L + PR_L | SAGE | 61.26 | ± | 55.23 | ± | 77.80 | ± | 139.7 |
| | | 2.06 | | 1.28 | | 0.68 | | |
| PMI_L + PR_L + Triplet_L | ALL | 77.06 | ± | 65.18 | ± | 74.63 | ± | 130.7 |
| | | 0.80 | | 0.32 | | 0.35 | | |
| PMI_L + PR_L + Triplet_L | GAT | 84.73 | ± | 70.45 | ± | 75.15 | ± | 45.7 |
| | | 0.10 | | 0.14 | | 0.11 | | |
| PMI_L + PR_L + Triplet_L | GCN | 83.09 | ± | 68.69 | ± | 73.43 | ± | 98.7 |
| | | 0.50 | | 0.33 | | 0.17 | | |
| PMI_L + PR_L + Triplet_L | GIN | 75.94 | ± | 64.91 | ± | 76.07 | ± | 104.7 |
| | | 0.13 | | 0.37 | | 0.38 | | |
| PMI_L + PR_L + Triplet_L | MPNN | 82.26 | ± | 68.63 | ± | 75.68 | ± | 70.3 |
| | | 0.41 | | 0.25 | | 0.21 | | |
| PMI_L + PR_L + Triplet_L | PAGNN | 74.30 | ± | 60.44 | ± | 63.85 | ± | 182.3 |
| | | 0.41 | | 0.69 | | 0.24 | | |

Knn Consistency Continued (↑)

| Loss Type | Model | CORA | | Citeseer | | Bitcoin Fraud Transaction | | Average Rank |
|---|---|---|---|---|---|---|---|---|
| PMI_L + PR_L + Triplet_L | SAGE | 77.68 ± 5.56 | | 67.21 ± 1.08 | | 78.21 ± 0.54 | | 84.3 |
| PMI_L + Triplet_L | ALL | 80.93 ± 0.38 | | 69.58 ± 0.23 | | 74.72 ± 0.25 | | 88.7 |
| PMI_L + Triplet_L | GAT | 84.69 ± 0.28 | | 70.59 ± 0.32 | | 75.08 ± 0.26 | | 44.7 |
| PMI_L + Triplet_L | GCN | 82.67 ± 0.24 | | 68.47 ± 0.43 | | 73.34 ± 0.26 | | 107.7 |
| PMI_L + Triplet_L | GIN | 77.10 ± 1.00 | | 65.70 ± 0.55 | | 75.48 ± 0.73 | | 112.0 |
| PMI_L + Triplet_L | MPNN | 82.27 ± 0.13 | | 69.54 ± 0.29 | | 75.55 ± 0.26 | | 62.7 |
| PMI_L + Triplet_L | PAGNN | 74.41 ± 0.47 | | 60.36 ± 0.62 | | 63.94 ± 0.17 | | 178.7 |
| PMI_L + Triplet_L | SAGE | 76.92 ± 2.13 | | 68.12 ± 0.57 | | 77.08 ± 0.49 | | 90.3 |
| PR_L | ALL | 75.73 ± 0.52 | | 64.46 ± 0.64 | | 75.52 ± 0.17 | | 120.3 |
| PR_L | GAT | 82.69 ± 0.52 | | 69.71 ± 0.14 | | 74.23 ± 0.30 | | 78.0 |
| PR_L | GCN | 81.68 ± 0.27 | | 67.93 ± 0.19 | | 73.00 ± 0.23 | | 128.0 |
| PR_L | GIN | 74.26 ± 1.04 | | 62.73 ± 0.61 | | 75.74 ± 0.05 | | 125.7 |
| PR_L | MPNN | 80.32 ± 0.95 | | 68.73 ± 0.50 | | 75.81 ± 0.31 | | 75.7 |
| PR_L | PAGNN | 73.03 ± 0.93 | | 59.33 ± 0.46 | | 63.78 ± 0.17 | | 195.7 |
| PR_L | SAGE | 61.22 ± 1.61 | | 55.94 ± 0.87 | | 78.75 ± 0.34 | | 135.7 |
| PR_L + Triplet_L | ALL | 74.97 ± 0.64 | | 61.87 ± 0.70 | | 75.31 ± 0.24 | | 139.3 |

Knn Consistency Continued (↑)

| Loss Type | Model | CORA | | Citeseer | | Bitcoin Fraud Transaction | | Average Rank |
|---|---|---|---|---|---|---|---|---|
| PR_L + Triplet_L | GAT | 83.77 ± 1.10 | | 69.97 ± 0.16 | | 74.03 ± 0.20 | | 71.7 |
| PR_L + Triplet_L | GCN | 82.38 ± 0.39 | | 68.21 ± 0.34 | | 73.16 ± 0.31 | | 117.7 |
| PR_L + Triplet_L | GIN | 74.45 ± 0.54 | | 62.61 ± 0.22 | | 75.83 ± 0.30 | | 122.3 |
| PR_L + Triplet_L | MPNN | 80.73 ± 0.42 | | 69.05 ± 0.58 | | 75.59 ± 0.12 | | 77.0 |
| PR_L + Triplet_L | PAGNN | 73.13 ± 1.10 | | 58.61 ± 0.54 | | 63.75 ± 0.27 | | 198.3 |
| PR_L + Triplet_L | SAGE | 64.03 ± 2.70 | | 61.49 ± 6.12 | | 78.76 ± 0.45 | | 124.0 |
| Triplet_L | ALL | 82.04 ± 0.50 | | 69.41 ± 0.25 | | 73.94 ± 0.31 | | 94.0 |
| Triplet_L | GAT | 84.93 ± 0.14 | | 71.11 ± 0.35 | | 75.13 ± 0.15 | | 36.0 |
| Triplet_L | GCN | 84.40 ± 0.24 | | 69.89 ± 0.51 | | 73.56 ± 0.13 | | 74.3 |
| Triplet_L | GIN | 81.80 ± 0.60 | | 68.54 ± 0.34 | | 76.31 ± 0.28 | | 69.3 |
| Triplet_L | MPNN | 83.24 ± 0.42 | | 70.21 ± 0.32 | | 75.34 ± 0.28 | | 56.3 |
| Triplet_L | PAGNN | 80.21 ± 2.20 | | 69.17 ± 0.13 | | 63.75 ± 0.20 | | 129.3 |
| Triplet_L | SAGE | 84.69 ± 0.21 | | 70.64 ± 0.39 | | 75.38 ± 0.17 | | 37.3 |

1.1.5   *Semantic Coherence and Ranking.*

Table 19. Coherence Performance (↑): Top-ranked results are highlighted in **1st**, second-ranked in **2nd**, and third-ranked in **3rd**.

| Loss Type | Model | CORA | | Citeseer | | Bitcoin Fraud Transaction | | Average Rank |
|---|---|---|---|---|---|---|---|---|
| Contr_l | ALL | 33.42 ± 5.44 | | 28.80 ± 7.81 | | 24.18 ± 7.50 | | 103.0 |
| Contr_l | GAT | 17.07 ± 0.58 | | 10.07 ± 0.44 | | 9.20 ± 0.63 | | 189.7 |
| Contr_l | GCN | 54.43 ± 8.15 | | 27.68 ± 6.82 | | 48.60 ± 43.46 | | 72.3 |
| Contr_l | GIN | 81.14 ± 27.72 | | 68.44 ± 29.50 | | 93.91 ± 13.61 | | 23.3 |
| Contr_l | MPNN | 26.55 ± 2.55 | | 11.50 ± 0.64 | | 13.86 ± 1.28 | | 159.3 |
| Contr_l | PAGNN | 27.38 ± 4.06 | | 18.36 ± 1.17 | | 32.44 ± 5.59 | | 123.3 |
| Contr_l | SAGE | 28.30 ± 3.43 | | 17.19 ± 1.84 | | 21.14 ± 4.42 | | 133.0 |
| Contr_l + CrossE_L | ALL | 46.16 ± 7.41 | | 23.29 ± 4.15 | | 28.61 ± 6.03 | | 100.7 |
| Contr_l + CrossE_L | GAT | 16.93 ± 1.15 | | 10.10 ± 0.44 | | 9.61 ± 1.69 | | 188.3 |
| Contr_l + CrossE_L | GCN | 28.20 ± 4.17 | | 37.95 ± 8.31 | | 11.96 ± 3.04 | | 122.0 |
| Contr_l + CrossE_L | GIN | 82.91 ± 23.27 | | 72.17 ± 15.08 | | 99.65 ± 0.58 | | 17.0 |
| Contr_l + CrossE_L | MPNN | 26.50 ± 3.35 | | 12.39 ± 1.62 | | 15.09 ± 3.35 | | 156.3 |
| Contr_l + CrossE_L | PAGNN | 26.30 ± 2.78 | | 20.22 ± 2.48 | | 34.76 ± 7.63 | | 119.7 |
| Contr_l + CrossE_L | SAGE | 24.16 ± 1.86 | | 13.61 ± 0.73 | | 22.61 ± 6.11 | | 149.7 |
| Contr_l + CrossE_L + PMI_L | ALL | 35.62 ± 6.02 | | 25.58 ± 4.66 | | 25.55 ± 3.59 | | 105.7 |
| Contr_l + CrossE_L + PMI_L | GAT | 12.48 ± 0.12 | | 8.31 ± 0.66 | | 9.88 ± 0.94 | | 200.0 |

Coherence Continued (↑)

| Loss Type | | | Model | CORA | | Citeseer | | Bitcoin Fraud Transaction | | Average Rank |
|---|---|---|---|---|---|---|---|---|---|---|
| Contr_l + CrossE_L + PMI_L | | | GCN | 67.42 ± 17.88 | | 64.17 ± 22.60 | | 66.07 ± 23.26 | | 37.7 |
| Contr_l + CrossE_L + PMI_L | | | GIN | 73.48 ± 26.30 | | 78.19 ± 19.01 | | 99.63 ± 0.82 | | 18.3 |
| Contr_l + CrossE_L + PMI_L | | | MPNN | 25.97 ± 1.65 | | 13.12 ± 1.34 | | 10.89 ± 0.37 | | 164.3 |
| Contr_l + CrossE_L + PMI_L | | | PAGNN | 29.79 ± 2.72 | | 28.31 ± 5.14 | | 46.01 ± 15.83 | | 88.7 |
| Contr_l + CrossE_L + PMI_L | | | SAGE | 15.91 ± 1.86 | | 12.75 ± 1.75 | | 10.76 ± 0.64 | | 181.0 |
| Contr_l + CrossE_L + PMI_L + PR_L | | | ALL | 46.58 ± 4.54 | | 29.22 ± 5.09 | | 27.95 ± 4.24 | | 90.3 |
| Contr_l + CrossE_L + PMI_L + PR_L | | | GAT | 12.51 ± 1.09 | | 8.95 ± 0.75 | | 16.05 ± 4.01 | | 183.7 |
| Contr_l + CrossE_L + PMI_L + PR_L | | | GCN | 71.85 ± 11.17 | | 70.29 ± 29.18 | | 51.49 ± 19.90 | | 36.3 |
| Contr_l + CrossE_L + PMI_L + PR_L | | | GIN | 80.88 ± 25.04 | | 86.76 ± 18.68 | | 97.76 ± 5.01 | | 14.3 |
| Contr_l + CrossE_L + PMI_L + PR_L | | | MPNN | 25.11 ± 1.71 | | 13.72 ± 2.51 | | 23.62 ± 6.14 | | 147.0 |
| Contr_l + CrossE_L + PMI_L + PR_L | | | PAGNN | 28.58 ± 7.38 | | 27.21 ± 6.14 | | 30.43 ± 5.60 | | 110.0 |
| Contr_l + CrossE_L + PMI_L + PR_L | | | SAGE | 16.43 ± 1.47 | | 11.49 ± 0.58 | | 18.91 ± 3.45 | | 169.3 |
| Contr_l + CrossE_L + PMI_L + PR_L + Triplet_L | | | ALL | 28.09 ± 5.57 | | 25.18 ± 7.70 | | 20.81 ± 3.76 | | 125.0 |
| Contr_l + CrossE_L + PMI_L + PR_L + Triplet_L | | | GAT | 12.61 ± 0.89 | | 9.20 ± 1.52 | | 10.41 ± 1.65 | | 193.7 |
| Contr_l + CrossE_L + PMI_L + PR_L + Triplet_L | | | GCN | 65.91 ± 14.44 | | 60.18 ± 23.85 | | 57.64 ± 19.64 | | 42.7 |
| Contr_l + CrossE_L + PMI_L + PR_L + Triplet_L | | | GIN | 69.96 ± 28.32 | | 83.02 ± 17.65 | | 100.00 ± 0.00 | | 15.3 |

Coherence Continued (↑)

| Loss Type | Model | CORA | | Citeseer | | Bitcoin Fraud Transaction | | Average Rank |
|---|---|---|---|---|---|---|---|---|
| Contr_l + CrossE_L + PMI_L + PR_L + Triplet_L | MPNN | 25.72 | ± 2.22 | 13.77 | ± 2.26 | 11.90 | ± 1.45 | 159.3 |
| Contr_l + CrossE_L + PMI_L + PR_L + Triplet_L | PAGNN | 30.48 | ± 3.72 | 22.84 | ± 2.51 | 35.82 | ± 11.31 | 103.0 |
| Contr_l + CrossE_L + PMI_L + PR_L + Triplet_L | SAGE | 19.00 | ± 3.56 | 13.72 | ± 0.87 | 13.48 | ± 1.29 | 165.0 |
| Contr_l + CrossE_L + PMI_L + Triplet_L | ALL | 27.55 | ± 1.90 | 21.06 | ± 4.40 | 20.31 | ± 0.75 | 131.7 |
| Contr_l + CrossE_L + PMI_L + Triplet_L | GAT | 11.83 | ± 0.97 | 8.40 | ± 0.59 | 9.13 | ± 0.99 | 204.3 |
| Contr_l + CrossE_L + PMI_L + Triplet_L | GCN | 56.76 | ± 11.37 | 42.05 | ± 7.40 | 59.17 | ± 11.33 | 56.0 |
| Contr_l + CrossE_L + PMI_L + Triplet_L | GIN | 51.21 | ± 15.58 | 79.16 | ± 19.98 | 89.29 | ± 13.13 | 38.0 |
| Contr_l + CrossE_L + PMI_L + Triplet_L | MPNN | 26.14 | ± 2.43 | 12.71 | ± 2.03 | 8.60 | ± 1.27 | 174.3 |
| Contr_l + CrossE_L + PMI_L + Triplet_L | PAGNN | 28.83 | ± 8.42 | 27.03 | ± 0.71 | 32.53 | ± 2.68 | 106.7 |
| Contr_l + CrossE_L + PMI_L + Triplet_L | SAGE | 19.52 | ± 0.76 | 14.02 | ± 2.30 | 11.22 | ± 1.60 | 167.7 |
| Contr_l + CrossE_L + PR_L | ALL | 36.51 | ± 6.10 | 36.72 | ± 6.54 | 56.45 | ± 7.10 | 70.3 |
| Contr_l + CrossE_L + PR_L | GAT | 29.51 | ± 11.95 | 23.88 | ± 7.65 | 33.54 | ± 6.19 | 106.7 |
| Contr_l + CrossE_L + PR_L | GCN | 60.08 | ± 18.45 | 58.29 | ± 25.25 | 23.67 | ± 4.74 | 72.0 |
| Contr_l + CrossE_L + PR_L | GIN | 99.40 | ± 1.35 | 100.00 | ± 0.00 | 98.65 | ± 3.02 | 8.0 |
| Contr_l + CrossE_L + PR_L | MPNN | 60.43 | ± 13.66 | 44.43 | ± 8.58 | 16.43 | ± 4.26 | 85.0 |
| Contr_l + CrossE_L + PR_L | PAGNN | 29.46 | ± 3.96 | 49.17 | ± 11.20 | 42.66 | ± 5.33 | 77.7 |

Continued on next page

Coherence Continued (↑)

| Loss Type | Model | CORA | | Citeseer | | Bitcoin Fraud Transaction | | Average Rank |
|---|---|---|---|---|---|---|---|---|
| Contr_l + CrossE_L + PR_L | SAGE | 34.78 9.74 | ± | 21.79 5.47 | ± | 29.47 6.58 | ± | 107.3 |
| Contr_l + CrossE_L + PR_L + Triplet_L | ALL | 30.45 11.91 | ± | 27.78 7.09 | ± | 42.57 16.66 | ± | 91.0 |
| Contr_l + CrossE_L + PR_L + Triplet_L | GAT | 16.89 1.88 | ± | 12.02 1.60 | ± | 13.09 3.62 | ± | 174.3 |
| Contr_l + CrossE_L + PR_L + Triplet_L | GCN | 53.26 17.78 | ± | 35.94 16.56 | ± | 46.94 23.35 | ± | 66.3 |
| Contr_l + CrossE_L + PR_L + Triplet_L | GIN | 100.00 0.00 | ± | 69.58 21.85 | ± | 99.96 0.10 | ± | 13.3 |
| Contr_l + CrossE_L + PR_L + Triplet_L | MPNN | 46.22 7.34 | ± | 27.52 4.30 | ± | 9.90 1.60 | ± | 120.0 |
| Contr_l + CrossE_L + PR_L + Triplet_L | PAGNN | 35.50 10.43 | ± | 43.83 6.97 | ± | 27.15 4.11 | ± | 87.7 |
| Contr_l + CrossE_L + PR_L + Triplet_L | SAGE | 22.54 2.24 | ± | 17.42 4.49 | ± | 42.98 11.20 | ± | 123.3 |
| Contr_l + CrossE_L + Triplet_L | ALL | 38.56 4.97 | ± | 19.10 2.26 | ± | 21.40 6.68 | ± | 116.0 |
| Contr_l + CrossE_L + Triplet_L | GAT | 16.06 1.00 | ± | 9.67 0.46 | ± | 7.58 0.88 | ± | 197.3 |
| Contr_l + CrossE_L + Triplet_L | GCN | 31.61 9.66 | ± | 25.21 12.80 | ± | 33.45 20.29 | ± | 102.3 |
| Contr_l + CrossE_L + Triplet_L | GIN | 64.89 25.17 | ± | 71.65 24.35 | ± | 90.93 19.67 | ± | 32.3 |
| Contr_l + CrossE_L + Triplet_L | MPNN | 21.26 2.26 | ± | 10.76 0.58 | ± | 9.27 1.38 | ± | 184.0 |
| Contr_l + CrossE_L + Triplet_L | PAGNN | 23.34 2.91 | ± | 16.65 3.20 | ± | 32.16 8.97 | ± | 134.7 |
| Contr_l + CrossE_L + Triplet_L | SAGE | 20.81 2.59 | ± | 14.16 1.27 | ± | 10.40 0.84 | ± | 169.3 |
| Contr_l + PMI_L | ALL | 36.74 8.18 | ± | 20.86 5.86 | ± | 34.55 6.15 | ± | 99.3 |

Coherence Continued (↑)

| Loss Type | Model | CORA | | Citeseer | | Bitcoin Fraud Transaction | | Average Rank |
|---|---|---|---|---|---|---|---|---|
| Contr_l + PMI_L | GAT | 11.94 ± 0.88 | | 8.53 ± 0.54 | | 9.72 ± 0.72 | | 200.0 |
| Contr_l + PMI_L | GCN | 66.18 ± 8.80 | | 76.73 ± 23.31 | | 70.08 ± 12.55 | | 32.3 |
| Contr_l + PMI_L | GIN | 63.74 ± 21.23 | | 64.81 ± 18.12 | | 87.99 ± 16.62 | | 36.0 |
| Contr_l + PMI_L | MPNN | 29.20 ± 4.03 | | 13.84 ± 2.19 | | 11.48 ± 2.09 | | 148.0 |
| Contr_l + PMI_L | PAGNN | 29.47 ± 7.43 | | 32.46 ± 4.12 | | 40.27 ± 8.49 | | 90.0 |
| Contr_l + PMI_L | SAGE | 16.56 ± 1.46 | | 12.87 ± 3.29 | | 11.02 ± 1.10 | | 177.0 |
| Contr_l + PMI_L + PR_L | ALL | 51.38 ± 5.17 | | 29.53 ± 2.93 | | 29.06 ± 5.13 | | 86.7 |
| Contr_l + PMI_L + PR_L | GAT | 13.80 ± 0.87 | | 10.80 ± 2.57 | | 19.93 ± 5.74 | | 173.0 |
| Contr_l + PMI_L + PR_L | GCN | 56.33 ± 6.13 | | 52.92 ± 14.36 | | 49.44 ± 19.61 | | 53.0 |
| Contr_l + PMI_L + PR_L | GIN | 71.50 ± 21.81 | | 76.28 ± 23.98 | | 87.08 ± 14.81 | | 27.0 |
| Contr_l + PMI_L + PR_L | MPNN | 26.79 ± 3.08 | | 15.49 ± 1.84 | | 20.87 ± 5.69 | | 139.7 |
| Contr_l + PMI_L + PR_L | PAGNN | 26.49 ± 3.84 | | 25.57 ± 1.07 | | 25.97 ± 3.93 | | 124.3 |
| Contr_l + PMI_L + PR_L | SAGE | 16.97 ± 1.38 | | 14.86 ± 2.90 | | 18.85 ± 2.49 | | 156.3 |
| Contr_l + PMI_L + PR_L + Triplet_L | ALL | 27.25 ± 2.67 | | 25.04 ± 7.31 | | 22.79 ± 4.94 | | 126.0 |
| Contr_l + PMI_L + PR_L + Triplet_L | GAT | 15.96 ± 3.18 | | 11.20 ± 1.56 | | 12.10 ± 1.40 | | 180.7 |
| Contr_l + PMI_L + PR_L + Triplet_L | GCN | 63.39 ± 23.38 | | 47.00 ± 28.95 | | 31.85 ± 13.94 | | 66.7 |

Coherence Continued (↑)

| Loss Type | Model | CORA | | Citeseer | | Bitcoin Fraud Transaction | | Average Rank |
|---|---|---|---|---|---|---|---|---|
| Contr_l + PMI_L + PR_L + Triplet_L | GIN | 51.50 ± 13.15 | | 53.70 ± 6.32 | | 90.18 ± 21.97 | | 45.3 |
| Contr_l + PMI_L + PR_L + Triplet_L | MPNN | 25.71 ± 3.13 | | 14.80 ± 1.97 | | 11.38 ± 1.42 | | 157.3 |
| Contr_l + PMI_L + PR_L + Triplet_L | PAGNN | 30.03 ± 13.03 | | 21.60 ± 3.45 | | 37.81 ± 10.54 | | 104.0 |
| Contr_l + PMI_L + PR_L + Triplet_L | SAGE | 22.77 ± 2.31 | | 14.50 ± 1.21 | | 18.67 ± 3.46 | | 152.7 |
| Contr_l + PR_L | ALL | 53.03 ± 9.42 | | 43.13 ± 11.42 | | 53.31 ± 9.64 | | 59.0 |
| Contr_l + PR_L | GAT | 37.58 ± 19.40 | | 27.58 ± 4.36 | | 33.61 ± 5.16 | | 91.0 |
| Contr_l + PR_L | GCN | 71.17 ± 7.72 | | 35.80 ± 13.17 | | 33.66 ± 9.81 | | 64.7 |
| Contr_l + PR_L | GIN | 100.00 ± 0.00 | | 84.12 ± 12.29 | | 100.00 ± 0.00 | | 5.7 |
| Contr_l + PR_L | MPNN | 65.32 ± 10.21 | | 41.12 ± 5.01 | | 15.84 ± 2.41 | | 85.7 |
| Contr_l + PR_L | PAGNN | 32.21 ± 4.44 | | 48.95 ± 9.48 | | 48.27 ± 11.93 | | 71.3 |
| Contr_l + PR_L | SAGE | 34.79 ± 8.27 | | 25.65 ± 9.92 | | 33.83 ± 9.73 | | 95.7 |
| Contr_l + PR_L + Triplet_L | ALL | 35.05 ± 4.87 | | 29.48 ± 4.67 | | 30.36 ± 3.99 | | 93.7 |
| Contr_l + PR_L + Triplet_L | GAT | 16.36 ± 2.86 | | 12.34 ± 2.50 | | 14.20 ± 1.91 | | 172.0 |
| Contr_l + PR_L + Triplet_L | GCN | 45.20 ± 12.89 | | 52.67 ± 28.78 | | 41.96 ± 16.46 | | 63.3 |
| Contr_l + PR_L + Triplet_L | GIN | 99.75 ± 0.57 | | 88.51 ± 22.65 | | 95.42 ± 10.25 | | 12.0 |
| Contr_l + PR_L + Triplet_L | MPNN | 41.23 ± 11.92 | | 25.75 ± 7.02 | | 10.13 ± 0.81 | | 123.7 |

Coherence Continued (↑)

| Loss Type | Model | CORA | | Citeseer | | Bitcoin Fraud Transaction | | Average Rank |
|---|---|---|---|---|---|---|---|---|
| Contr_l + PR_L + Triplet_L | PAGNN | 27.95 ± 5.45 | | 43.18 ± 12.67 | | 27.94 ± 1.60 | | 101.3 |
| Contr_l + PR_L + Triplet_L | SAGE | 25.19 ± 10.87 | | 19.70 ± 3.81 | | 50.06 ± 9.52 | | 114.3 |
| Contr_l + Triplet_L | ALL | 32.74 ± 2.63 | | 21.55 ± 1.26 | | 15.40 ± 2.02 | | 125.0 |
| Contr_l + Triplet_L | GAT | 14.32 ± 0.94 | | 9.49 ± 0.40 | | 7.36 ± 0.42 | | 200.3 |
| Contr_l + Triplet_L | GCN | 32.47 ± 7.67 | | 20.19 ± 6.87 | | 23.86 ± 25.32 | | 118.0 |
| Contr_l + Triplet_L | GIN | 40.90 ± 5.84 | | 86.15 ± 20.36 | | 84.91 ± 32.54 | | 41.0 |
| Contr_l + Triplet_L | MPNN | 21.62 ± 2.24 | | 10.69 ± 0.61 | | 9.34 ± 1.08 | | 183.7 |
| Contr_l + Triplet_L | PAGNN | 23.91 ± 4.49 | | 16.42 ± 1.34 | | 33.27 ± 12.92 | | 132.3 |
| Contr_l + Triplet_L | SAGE | 21.10 ± 1.92 | | 13.48 ± 1.19 | | 10.63 ± 0.68 | | 171.7 |
| CrossE_L | ALL | 39.94 ± 4.84 | | 33.13 ± 9.61 | | 58.70 ± 20.42 | | 70.0 |
| CrossE_L | GAT | 25.79 ± 1.93 | | 25.13 ± 2.44 | | 15.23 ± 2.19 | | 138.3 |
| CrossE_L | GCN | 97.61 ± 2.49 | | 99.94 ± 0.14 | | 99.46 ± 0.44 | | 8.3 |
| CrossE_L | GIN | 100.00 ± 0.01 | | 100.00 ± 0.00 | | 100.00 ± 0.00 | | 2.7 |
| CrossE_L | MPNN | 54.53 ± 12.03 | | 32.51 ± 4.35 | | 91.03 ± 1.72 | | 55.0 |
| CrossE_L | PAGNN | 57.26 ± 12.15 | | 40.03 ± 10.25 | | 98.74 ± 0.65 | | 46.7 |
| CrossE_L | SAGE | 33.24 ± 8.25 | | 18.70 ± 3.09 | | 29.17 ± 32.46 | | 113.3 |

Coherence Continued (↑)

| Loss Type | Model | CORA | | Citeseer | | Bitcoin Fraud Transaction | | Average Rank |
|---|---|---|---|---|---|---|---|---|
| CrossE_L + PMI_L | ALL | 25.45 | ± | 19.95 | ± | 13.69 | ± | 147.0 |
| | | 3.28 | | 1.88 | | 1.36 | | |
| CrossE_L + PMI_L | GAT | 11.64 | ± | 8.32 | ± | 10.24 | ± | 200.0 |
| | | 0.56 | | 0.83 | | 0.72 | | |
| CrossE_L + PMI_L | GCN | 76.22 | ± | 60.96 | ± | 72.37 | ± | 30.3 |
| | | 15.68 | | 21.62 | | 21.41 | | |
| CrossE_L + PMI_L | GIN | 87.33 | ± | 83.87 | ± | 92.47 | ± | 17.0 |
| | | 17.01 | | 21.31 | | 16.85 | | |
| CrossE_L + PMI_L | MPNN | 28.65 | ± | 14.41 | ± | 12.39 | ± | 147.0 |
| | | 2.04 | | 1.60 | | 1.71 | | |
| CrossE_L + PMI_L | PAGNN | 31.07 | ± | 29.43 | ± | 36.38 | ± | 91.0 |
| | | 7.65 | | 4.88 | | 13.60 | | |
| CrossE_L + PMI_L | SAGE | 22.93 | ± | 11.58 | ± | 8.66 | ± | 181.7 |
| | | 5.38 | | 2.43 | | 0.87 | | |
| CrossE_L + PMI_L + PR_L | ALL | 41.85 | ± | 31.37 | ± | 29.59 | ± | 88.7 |
| | | 6.50 | | 3.81 | | 4.49 | | |
| CrossE_L + PMI_L + PR_L | GAT | 12.38 | ± | 11.28 | ± | 18.33 | ± | 177.0 |
| | | 0.66 | | 2.66 | | 8.59 | | |
| CrossE_L + PMI_L + PR_L | GCN | 64.91 | ± | 54.66 | ± | 68.71 | ± | 41.7 |
| | | 11.42 | | 20.56 | | 23.30 | | |
| CrossE_L + PMI_L + PR_L | GIN | 76.92 | ± | 88.29 | ± | 94.18 | ± | 16.7 |
| | | 28.23 | | 17.36 | | 7.92 | | |
| CrossE_L + PMI_L + PR_L | MPNN | 32.15 | ± | 17.42 | ± | 17.04 | ± | 129.0 |
| | | 3.97 | | 6.10 | | 7.83 | | |
| CrossE_L + PMI_L + PR_L | PAGNN | 25.03 | ± | 25.83 | ± | 29.00 | ± | 124.0 |
| | | 2.40 | | 3.34 | | 4.97 | | |
| CrossE_L + PMI_L + PR_L | SAGE | 15.58 | ± | 9.96 | ± | 22.38 | ± | 171.7 |
| | | 2.60 | | 0.72 | | 6.25 | | |
| CrossE_L + PMI_L + PR_L + Triplet_L | ALL | 26.55 | ± | 22.89 | ± | 16.58 | ± | 136.0 |
| | | 4.45 | | 1.25 | | 2.46 | | |
| CrossE_L + PMI_L + PR_L + Triplet_L | GAT | 12.40 | ± | 9.62 | ± | 10.88 | ± | 192.0 |
| | | 0.62 | | 0.82 | | 1.29 | | |

Coherence Continued (↑)

| Loss Type | Model | CORA | | Citeseer | | Bitcoin Fraud Transaction | | Average Rank |
|---|---|---|---|---|---|---|---|---|
| CrossE_L + PMI_L + PR_L + Triplet_L | GCN | 69.64 10.29 | ± | 57.99 24.22 | ± | 58.39 13.33 | ± | 40.7 |
| CrossE_L + PMI_L + PR_L + Triplet_L | GIN | 66.69 27.81 | ± | 75.26 10.79 | ± | 96.67 6.51 | ± | 25.7 |
| CrossE_L + PMI_L + PR_L + Triplet_L | MPNN | 25.25 1.81 | ± | 15.88 5.86 | ± | 11.22 2.69 | ± | 158.0 |
| CrossE_L + PMI_L + PR_L + Triplet_L | PAGNN | 28.36 4.72 | ± | 23.85 4.74 | ± | 39.73 8.33 | ± | 105.7 |
| CrossE_L + PMI_L + PR_L + Triplet_L | SAGE | 19.28 2.50 | ± | 14.55 2.24 | ± | 13.53 0.95 | ± | 161.0 |
| CrossE_L + PMI_L + Triplet_L | ALL | 26.81 2.73 | ± | 23.47 3.32 | ± | 18.29 2.77 | ± | 133.3 |
| CrossE_L + PMI_L + Triplet_L | GAT | 11.86 0.78 | ± | 8.48 0.49 | ± | 8.84 0.41 | ± | 204.0 |
| CrossE_L + PMI_L + Triplet_L | GCN | 57.87 7.55 | ± | 68.96 24.76 | ± | 67.60 8.98 | ± | 42.3 |
| CrossE_L + PMI_L + Triplet_L | GIN | 65.50 28.42 | ± | 83.98 14.17 | ± | 100.00 0.00 | ± | 18.7 |
| CrossE_L + PMI_L + Triplet_L | MPNN | 26.88 2.38 | ± | 11.87 1.31 | ± | 9.51 1.65 | ± | 170.0 |
| CrossE_L + PMI_L + Triplet_L | PAGNN | 28.69 7.73 | ± | 25.56 2.72 | ± | 34.06 2.85 | ± | 105.7 |
| CrossE_L + PMI_L + Triplet_L | SAGE | 20.22 1.98 | ± | 13.49 1.42 | ± | 10.03 0.66 | ± | 174.7 |
| CrossE_L + PR_L | ALL | 59.34 7.97 | ± | 27.56 6.59 | ± | 69.59 7.87 | ± | 63.7 |
| CrossE_L + PR_L | GAT | 62.30 11.49 | ± | 41.93 6.85 | ± | 68.20 17.14 | ± | 51.3 |
| CrossE_L + PR_L | GCN | 74.28 15.94 | ± | 60.31 19.40 | ± | 99.52 0.26 | ± | 24.7 |
| CrossE_L + PR_L | GIN | 98.24 3.92 | ± | 95.73 7.25 | ± | 99.70 0.67 | ± | 7.0 |

Continued on next page

Coherence Continued (↑)

| Loss Type | Model | CORA | | Citeseer | | Bitcoin Fraud Transaction | | Average Rank |
|---|---|---|---|---|---|---|---|---|
| CrossE_L + PR_L | MPNN | 53.31 9.26 | ± | 50.99 7.54 | ± | 24.60 4.12 | ± | 76.7 |
| CrossE_L + PR_L | PAGNN | 31.31 2.25 | ± | 42.36 6.08 | ± | 40.60 7.85 | ± | 80.0 |
| CrossE_L + PR_L | SAGE | 51.45 7.95 | ± | 34.71 11.62 | ± | 48.83 14.51 | ± | 67.3 |
| CrossE_L + PR_L + Triplet_L | ALL | 44.25 13.82 | ± | 34.79 4.55 | ± | 44.58 8.75 | ± | 72.3 |
| CrossE_L + PR_L + Triplet_L | GAT | 22.40 2.76 | ± | 11.86 2.17 | ± | 23.00 3.43 | ± | 157.3 |
| CrossE_L + PR_L + Triplet_L | GCN | 64.80 20.33 | ± | 41.68 17.78 | ± | 32.66 9.10 | ± | 69.0 |
| CrossE_L + PR_L + Triplet_L | GIN | 91.56 11.62 | ± | 95.32 6.71 | ± | 99.99 0.01 | ± | 7.7 |
| CrossE_L + PR_L + Triplet_L | MPNN | 54.16 6.71 | ± | 28.18 6.20 | ± | 12.36 1.72 | ± | 106.0 |
| CrossE_L + PR_L + Triplet_L | PAGNN | 28.90 3.16 | ± | 40.98 4.17 | ± | 28.92 8.24 | ± | 99.0 |
| CrossE_L + PR_L + Triplet_L | SAGE | 20.33 3.87 | ± | 19.25 2.52 | ± | 34.85 13.80 | ± | 129.3 |
| CrossE_L + Triplet_L | ALL | 33.03 3.20 | ± | 27.84 3.06 | ± | 11.27 0.78 | ± | 121.7 |
| CrossE_L + Triplet_L | GAT | 13.17 1.25 | ± | 8.99 0.74 | ± | 8.44 1.10 | ± | 201.0 |
| CrossE_L + Triplet_L | GCN | 34.53 9.26 | ± | 36.48 18.58 | ± | 26.16 14.14 | ± | 95.0 |
| CrossE_L + Triplet_L | GIN | 96.95 6.66 | ± | 73.97 32.49 | ± | 53.79 30.33 | ± | 29.3 |
| CrossE_L + Triplet_L | MPNN | 17.02 2.23 | ± | 9.35 1.03 | ± | 7.22 0.60 | ± | 196.3 |
| CrossE_L + Triplet_L | PAGNN | 23.82 5.44 | ± | 17.17 3.69 | ± | 38.80 13.59 | ± | 126.3 |

Coherence Continued (↑)

| Loss Type | Model | CORA | | Citeseer | | Bitcoin Fraud Transaction | | Average Rank |
|---|---|---|---|---|---|---|---|---|
| CrossE_L + Triplet_L | SAGE | 15.10 ± 1.38 | | 11.95 ± 0.63 | | 8.69 ± 0.53 | | 190.7 |
| PMI_L | ALL | 28.68 ± 3.48 | | 18.15 ± 2.48 | | 13.43 ± 2.20 | | 139.7 |
| PMI_L | GAT | 11.37 ± 0.80 | | 8.24 ± 0.57 | | 9.99 ± 0.63 | | 202.3 |
| PMI_L | GCN | 75.19 ± 7.13 | | 67.94 ± 21.27 | | 81.33 ± 10.95 | | 29.3 |
| PMI_L | GIN | 66.22 ± 20.44 | | 89.20 ± 16.38 | | 96.65 ± 7.49 | | 20.7 |
| PMI_L | MPNN | 27.65 ± 2.32 | | 14.65 ± 1.27 | | 11.82 ± 1.45 | | 150.3 |
| PMI_L | PAGNN | 29.52 ± 2.33 | | 34.47 ± 5.77 | | 37.71 ± 10.24 | | 89.7 |
| PMI_L | SAGE | 20.90 ± 2.14 | | 12.17 ± 1.50 | | 8.13 ± 0.96 | | 183.7 |
| PMI_L + PR_L | ALL | 37.25 ± 12.07 | | 34.36 ± 8.61 | | 32.76 ± 3.39 | | 86.7 |
| PMI_L + PR_L | GAT | 13.62 ± 1.42 | | 13.21 ± 5.20 | | 28.16 ± 4.49 | | 158.7 |
| PMI_L + PR_L | GCN | 62.11 ± 21.08 | | 64.52 ± 23.15 | | 49.05 ± 20.76 | | 47.0 |
| PMI_L + PR_L | GIN | 79.19 ± 27.93 | | 90.80 ± 10.93 | | 95.55 ± 6.12 | | 14.7 |
| PMI_L + PR_L | MPNN | 27.99 ± 2.79 | | 17.35 ± 2.31 | | 26.60 ± 9.61 | | 128.7 |
| PMI_L + PR_L | PAGNN | 28.97 ± 5.22 | | 26.48 ± 3.56 | | 31.59 ± 3.58 | | 108.0 |
| PMI_L + PR_L | SAGE | 16.48 ± 1.87 | | 10.88 ± 1.66 | | 20.82 ± 5.33 | | 168.3 |
| PMI_L + PR_L + Triplet_L | ALL | 27.72 ± 4.12 | | 22.25 ± 2.27 | | 19.51 ± 2.00 | | 130.0 |

Coherence Continued (↑)

| Loss Type | Model | CORA | | Citeseer | | Bitcoin Fraud Transaction | | Average Rank |
|---|---|---|---|---|---|---|---|---|
| PMI_L + PR_L + Triplet_L | GAT | 13.21 ± 0.66 | | 9.50 ± 0.86 | | 12.59 ± 2.30 | | 185.7 |
| PMI_L + PR_L + Triplet_L | GCN | 65.15 ± 10.12 | | 50.52 ± 14.65 | | 60.88 ± 33.98 | | 44.7 |
| PMI_L + PR_L + Triplet_L | GIN | 79.61 ± 21.77 | | 82.17 ± 30.09 | | 99.33 ± 1.50 | | 15.7 |
| PMI_L + PR_L + Triplet_L | MPNN | 30.86 ± 7.52 | | 13.88 ± 1.20 | | 10.84 ± 2.52 | | 147.0 |
| PMI_L + PR_L + Triplet_L | PAGNN | 30.41 ± 5.80 | | 21.83 ± 1.88 | | 34.54 ± 8.48 | | 105.7 |
| PMI_L + PR_L + Triplet_L | SAGE | 18.67 ± 1.78 | | 13.19 ± 0.83 | | 13.98 ± 2.65 | | 166.3 |
| PMI_L + Triplet_L | ALL | 25.83 ± 2.74 | | 20.58 ± 1.99 | | 18.73 ± 1.37 | | 139.3 |
| PMI_L + Triplet_L | GAT | 12.12 ± 1.09 | | 8.49 ± 0.51 | | 9.37 ± 0.53 | | 201.3 |
| PMI_L + Triplet_L | GCN | 72.93 ± 14.67 | | 72.40 ± 22.19 | | 71.12 ± 18.88 | | 29.0 |
| PMI_L + Triplet_L | GIN | 68.19 ± 23.25 | | 78.03 ± 30.28 | | 87.96 ± 16.85 | | 28.0 |
| PMI_L + Triplet_L | MPNN | 27.18 ± 4.88 | | 14.53 ± 2.56 | | 9.60 ± 1.82 | | 160.3 |
| PMI_L + Triplet_L | PAGNN | 26.35 ± 2.24 | | 27.84 ± 1.73 | | 32.85 ± 2.16 | | 111.7 |
| PMI_L + Triplet_L | SAGE | 17.99 ± 1.39 | | 13.50 ± 0.96 | | 10.17 ± 0.49 | | 175.7 |
| PR_L | ALL | 58.48 ± 10.53 | | 44.96 ± 14.42 | | 70.70 ± 9.22 | | 49.0 |
| PR_L | GAT | 68.72 ± 17.70 | | 42.68 ± 10.04 | | 59.81 ± 20.52 | | 47.0 |
| PR_L | GCN | 71.21 ± 11.00 | | 60.59 ± 22.36 | | 72.04 ± 21.38 | | 33.7 |

Coherence Continued (↑)

| Loss Type | Model | CORA | | Citeseer | | Bitcoin Fraud Transaction | | Average Rank |
|-----------|-------|------|---|----------|---|---------------------------|---|--------------|
| PR_L | GIN | 100.00 ± 0.00 | | 95.16 ± 7.14 | | 100.00 ± 0.00 | | 5.0 |
| PR_L | MPNN | 50.45 ± 8.95 | | 54.38 ± 7.75 | | 17.34 ± 4.21 | | 85.3 |
| PR_L | PAGNN | 43.28 ± 9.36 | | 50.41 ± 6.18 | | 34.02 ± 6.30 | | 71.0 |
| PR_L | SAGE | 46.32 ± 13.10 | | 47.93 ± 6.69 | | 45.32 ± 7.78 | | 62.3 |
| PR_L + Triplet_L | ALL | 41.48 ± 7.39 | | 31.15 ± 6.71 | | 60.61 ± 5.05 | | 69.3 |
| PR_L + Triplet_L | GAT | 28.68 ± 14.88 | | 27.39 ± 2.90 | | 31.81 ± 8.64 | | 108.3 |
| PR_L + Triplet_L | GCN | 61.60 ± 20.26 | | 45.34 ± 10.91 | | 26.40 ± 13.77 | | 74.0 |
| PR_L + Triplet_L | GIN | 99.95 ± 0.10 | | 81.82 ± 16.55 | | 100.00 ± 0.00 | | 9.7 |
| PR_L + Triplet_L | MPNN | 44.54 ± 5.91 | | 42.79 ± 5.38 | | 13.47 ± 2.69 | | 98.3 |
| PR_L + Triplet_L | PAGNN | 34.17 ± 8.35 | | 46.61 ± 6.92 | | 42.29 ± 10.82 | | 72.7 |
| PR_L + Triplet_L | SAGE | 39.65 ± 6.20 | | 28.00 ± 8.85 | | 41.07 ± 4.42 | | 82.3 |
| Triplet_L | ALL | 27.54 ± 0.72 | | 28.98 ± 3.13 | | 13.25 ± 2.57 | | 128.3 |
| Triplet_L | GAT | 12.83 ± 1.46 | | 8.90 ± 0.86 | | 7.91 ± 0.63 | | 202.7 |
| Triplet_L | GCN | 33.97 ± 7.90 | | 38.17 ± 10.24 | | 39.92 ± 33.10 | | 80.0 |
| Triplet_L | GIN | 75.83 ± 34.08 | | 79.49 ± 17.32 | | 49.23 ± 19.91 | | 32.0 |
| Triplet_L | MPNN | 18.66 ± 2.82 | | 10.15 ± 0.94 | | 7.74 ± 0.35 | | 191.7 |

Coherence Continued (↑)

| Loss Type | Model | CORA | | Citeseer | | Bitcoin Fraud Transaction | | Average Rank |
|-----------|-------|------|---|----------|---|---------------------------|---|--------------|
| Triplet_L | PAGNN | 22.98 | ± | 13.70 | ± | 36.47 | ± | 134.3 |
|           |       | 2.49 |   | 2.89 |   | 19.68 |   |       |
| Triplet_L | SAGE | 16.15 | ± | 13.12 | ± | 8.64 | ± | 187.0 |
|           |      | 1.65 |   | 0.94 |   | 1.42 |   |       |

Table 20. Selfcluster Performance (↑): Top-ranked results are highlighted in **1st**, second-ranked in **2nd**, and third-ranked in **3rd**.

| Loss Type | Model | CORA | | Citeseer | | Bitcoin Fraud Transaction | | Average Rank |
|-----------|-------|------|---|----------|---|---------------------------|---|--------------|
| Contr_l | ALL | -0.81 | ± | -0.80 | ± | -0.80 | ± | 71.7 |
|         |     | 0.00 |   | 0.00 |   | 0.00 |   |      |
| Contr_l | GAT | -0.81 | ± | -0.81 | ± | -0.80 | ± | 93.3 |
|         |     | 0.00 |   | 0.00 |   | 0.00 |   |      |
| Contr_l | GCN | -0.81 | ± | -0.81 | ± | -0.80 | ± | 94.3 |
|         |     | 0.00 |   | 0.00 |   | 0.00 |   |      |
| Contr_l | GIN | -0.81 | ± | -0.81 | ± | -0.80 | ± | 95.3 |
|         |     | 0.00 |   | 0.00 |   | 0.00 |   |      |
| Contr_l | MPNN | -0.81 | ± | -0.81 | ± | -0.80 | ± | 96.3 |
|         |      | 0.00 |   | 0.00 |   | 0.00 |   |      |
| Contr_l | PAGNN | -0.81 | ± | -0.81 | ± | -0.79 | ± | 67.0 |
|         |       | 0.00 |   | 0.00 |   | 0.00 |   |      |
| Contr_l | SAGE | -0.81 | ± | -0.81 | ± | -0.80 | ± | 98.0 |
|         |      | 0.00 |   | 0.00 |   | 0.00 |   |      |
| Contr_l + CrossE_L | ALL | -0.81 | ± | -0.80 | ± | -0.80 | ± | 76.3 |
|                    |     | 0.00 |   | 0.00 |   | 0.00 |   |      |
| Contr_l + CrossE_L | GAT | -0.81 | ± | -0.81 | ± | -0.80 | ± | 99.7 |
|                    |     | 0.00 |   | 0.00 |   | 0.00 |   |      |
| Contr_l + CrossE_L | GCN | -0.81 | ± | -0.81 | ± | -0.80 | ± | 100.7 |
|                    |     | 0.00 |   | 0.00 |   | 0.00 |   |       |

Selfcluster Continued (↑)

| Loss Type | Model | CORA | | Citeseer | | Bitcoin Fraud Transaction | | Average Rank |
|---|---|---|---|---|---|---|---|---|
| Contr_l + CrossE_L | GIN | -0.81 | ± | -0.81 | ± | -0.80 | ± | 101.7 |
| | | 0.00 | | 0.00 | | 0.00 | | |
| Contr_l + CrossE_L | MPNN | -0.81 | ± | -0.81 | ± | -0.80 | ± | 102.7 |
| | | 0.00 | | 0.00 | | 0.00 | | |
| Contr_l + CrossE_L | PAGNN | -0.81 | ± | -0.81 | ± | -0.79 | ± | 71.7 |
| | | 0.00 | | 0.01 | | 0.00 | | |
| Contr_l + CrossE_L | SAGE | -0.81 | ± | -0.81 | ± | -0.80 | ± | 104.3 |
| | | 0.00 | | 0.00 | | 0.00 | | |
| Contr_l + CrossE_L + PMI_L | ALL | -0.80 | ± | -0.80 | ± | -0.80 | ± | 63.0 |
| | | 0.00 | | 0.01 | | 0.00 | | |
| Contr_l + CrossE_L + PMI_L | GAT | -0.82 | ± | -0.81 | ± | -0.80 | ± | 135.7 |
| | | 0.00 | | 0.00 | | 0.00 | | |
| Contr_l + CrossE_L + PMI_L | GCN | -0.82 | ± | -0.81 | ± | -0.80 | ± | 136.7 |
| | | 0.00 | | 0.00 | | 0.00 | | |
| Contr_l + CrossE_L + PMI_L | GIN | -0.81 | ± | -0.80 | ± | -0.80 | ± | 82.3 |
| | | 0.00 | | 0.00 | | 0.00 | | |
| Contr_l + CrossE_L + PMI_L | MPNN | -0.81 | ± | -0.81 | ± | -0.80 | ± | 107.7 |
| | | 0.00 | | 0.00 | | 0.00 | | |
| Contr_l + CrossE_L + PMI_L | PAGNN | -0.80 | ± | -0.79 | ± | -0.79 | ± | 14.3 |
| | | 0.00 | | 0.00 | | 0.00 | | |
| Contr_l + CrossE_L + PMI_L | SAGE | -0.81 | ± | -0.81 | ± | -0.80 | ± | 108.7 |
| | | 0.00 | | 0.00 | | 0.00 | | |
| Contr_l + CrossE_L + PMI_L + PR_L | ALL | -0.79 | ± | -0.79 | ± | -0.79 | ± | 2.3 |
| | | 0.01 | | 0.01 | | 0.00 | | |
| Contr_l + CrossE_L + PMI_L + PR_L | GAT | -0.81 | ± | -0.81 | ± | -0.80 | ± | 109.7 |
| | | 0.00 | | 0.00 | | 0.01 | | |
| Contr_l + CrossE_L + PMI_L + PR_L | GCN | -0.82 | ± | -0.81 | ± | -0.80 | ± | 140.0 |
| | | 0.01 | | 0.00 | | 0.00 | | |
| Contr_l + CrossE_L + PMI_L + PR_L | GIN | -0.81 | ± | -0.80 | ± | -0.79 | ± | 51.7 |
| | | 0.00 | | 0.00 | | 0.00 | | |
| Contr_l + CrossE_L + PMI_L + PR_L | MPNN | -0.81 | ± | -0.80 | ± | -0.79 | ± | 52.7 |
| | | 0.00 | | 0.01 | | 0.00 | | |

Selfcluster Continued (↑)

| Loss Type | Model | CORA | | Citeseer | | Bitcoin Fraud Transaction | | Average Rank |
|---|---|---|---|---|---|---|---|---|
| Contr_l + CrossE_L + PMI_L + PR_L | PAGNN | -0.80 0.00 | ± | -0.79 0.01 | ± | -0.79 0.00 | | 16.7 |
| Contr_l + CrossE_L + PMI_L + PR_L | SAGE | -0.81 0.00 | ± | -0.81 0.00 | ± | -0.79 0.00 | ± | 79.0 |
| Contr_l + CrossE_L + PMI_L + PR_L + Triplet_L | ALL | -0.81 0.00 | ± | -0.80 0.00 | ± | -0.80 0.00 | ± | 87.3 |
| Contr_l + CrossE_L + PMI_L + PR_L + Triplet_L | GAT | -0.81 0.00 | ± | -0.81 0.00 | ± | -0.80 0.00 | ± | 113.3 |
| Contr_l + CrossE_L + PMI_L + PR_L + Triplet_L | GCN | -0.82 0.00 | ± | -0.81 0.00 | ± | -0.80 0.00 | ± | 142.3 |
| Contr_l + CrossE_L + PMI_L + PR_L + Triplet_L | GIN | -0.81 0.00 | ± | -0.80 0.00 | ± | -0.80 0.00 | ± | 89.3 |
| Contr_l + CrossE_L + PMI_L + PR_L + Triplet_L | MPNN | -0.81 0.00 | ± | -0.80 0.01 | ± | -0.80 0.00 | ± | 90.3 |
| Contr_l + CrossE_L + PMI_L + PR_L + Triplet_L | PAGNN | -0.80 0.00 | ± | -0.80 0.00 | ± | -0.79 0.00 | ± | 36.3 |
| Contr_l + CrossE_L + PMI_L + PR_L + Triplet_L | SAGE | -0.81 0.00 | ± | -0.81 0.00 | ± | -0.79 0.00 | ± | 82.3 |
| Contr_l + CrossE_L + PMI_L + Triplet_L | ALL | -0.81 0.00 | ± | -0.81 0.00 | ± | -0.80 0.00 | ± | 117.0 |
| Contr_l + CrossE_L + PMI_L + Triplet_L | GAT | -0.82 0.00 | ± | -0.81 0.00 | ± | -0.80 0.00 | ± | 145.0 |
| Contr_l + CrossE_L + PMI_L + Triplet_L | GCN | -0.82 0.00 | ± | -0.81 0.00 | ± | -0.80 0.00 | ± | 146.0 |
| Contr_l + CrossE_L + PMI_L + Triplet_L | GIN | -0.81 0.00 | ± | -0.80 0.00 | ± | -0.80 0.00 | ± | 93.3 |
| Contr_l + CrossE_L + PMI_L + Triplet_L | MPNN | -0.81 0.00 | ± | -0.81 0.00 | ± | -0.80 0.00 | ± | 120.0 |
| Contr_l + CrossE_L + PMI_L + Triplet_L | PAGNN | -0.80 0.00 | ± | -0.79 0.00 | ± | -0.79 0.00 | ± | 19.0 |
| Contr_l + CrossE_L + PMI_L + Triplet_L | SAGE | -0.81 0.00 | ± | -0.81 0.00 | ± | -0.80 0.00 | ± | 121.0 |

Selfcluster Continued (↑)

| Loss Type | Model | CORA | | Citeseer | | Bitcoin Fraud Transaction | | Average Rank |
|---|---|---|---|---|---|---|---|---|
| Contr_l + CrossE_L + PR_L | ALL | -0.79 ± 0.00 | | -0.80 0.01 | ± | -0.79 0.00 | ± | 25.0 |
| Contr_l + CrossE_L + PR_L | GAT | -0.80 0.01 | ± | -0.79 0.00 | ± | -0.79 0.00 | ± | 20.3 |
| Contr_l + CrossE_L + PR_L | GCN | -0.80 0.00 | ± | -0.80 0.00 | ± | -0.80 0.00 | ± | 74.7 |
| Contr_l + CrossE_L + PR_L | GIN | -0.79 ± 0.00 | | -0.80 0.01 | ± | -0.79 0.00 | ± | 26.7 |
| Contr_l + CrossE_L + PR_L | MPNN | -0.79 0.00 | ± | -0.79 0.00 | ± | -0.80 0.01 | ± | 43.0 |
| Contr_l + CrossE_L + PR_L | PAGNN | -0.79 0.00 | ± | -0.79 0.00 | ± | -0.79 0.00 | ± | 9.0 |
| Contr_l + CrossE_L + PR_L | SAGE | -0.80 0.01 | ± | -0.80 0.00 | ± | -0.79 0.00 | ± | 41.7 |
| Contr_l + CrossE_L + PR_L + Triplet_L | ALL | -0.80 0.00 | ± | -0.80 0.00 | ± | -0.80 0.00 | ± | 77.0 |
| Contr_l + CrossE_L + PR_L + Triplet_L | GAT | -0.81 0.00 | ± | -0.81 0.00 | ± | -0.80 0.00 | ± | 123.0 |
| Contr_l + CrossE_L + PR_L + Triplet_L | GCN | -0.81 0.00 | ± | -0.80 0.01 | ± | -0.80 0.00 | ± | 99.0 |
| Contr_l + CrossE_L + PR_L + Triplet_L | GIN | -0.81 0.00 | ± | -0.80 0.00 | ± | -0.80 0.00 | ± | 100.0 |
| Contr_l + CrossE_L + PR_L + Triplet_L | MPNN | -0.80 0.00 | ± | -0.80 0.00 | ± | -0.80 0.00 | ± | 79.7 |
| Contr_l + CrossE_L + PR_L + Triplet_L | PAGNN | -0.80 0.00 | ± | -0.80 0.00 | ± | -0.79 0.00 | ± | 44.7 |
| Contr_l + CrossE_L + PR_L + Triplet_L | SAGE | -0.81 0.00 | ± | -0.80 0.00 | ± | -0.80 0.00 | ± | 102.0 |
| Contr_l + CrossE_L + Triplet_L | ALL | -0.81 0.00 | ± | -0.81 0.00 | ± | -0.80 0.00 | ± | 126.3 |
| Contr_l + CrossE_L + Triplet_L | GAT | -0.82 0.00 | ± | -0.81 0.00 | ± | -0.80 0.00 | ± | 152.3 |

Selfcluster Continued (↑)

| Loss Type | Model | CORA | | Citeseer | | Bitcoin Fraud Transaction | | Average Rank |
|---|---|---|---|---|---|---|---|---|
| Contr_l + CrossE_L + Triplet_L | GCN | -0.81 ± 0.00 | | -0.81 ± 0.00 | | -0.80 ± 0.00 | | 128.0 |
| Contr_l + CrossE_L + Triplet_L | GIN | -0.81 ± 0.00 | | -0.81 ± 0.00 | | -0.80 ± 0.00 | | 129.0 |
| Contr_l + CrossE_L + Triplet_L | MPNN | -0.81 ± 0.00 | | -0.81 ± 0.00 | | -0.80 ± 0.00 | | 130.0 |
| Contr_l + CrossE_L + Triplet_L | PAGNN | -0.81 ± 0.00 | | -0.81 ± 0.00 | | -0.79 ± 0.00 | | 93.3 |
| Contr_l + CrossE_L + Triplet_L | SAGE | -0.81 ± 0.00 | | -0.81 ± 0.00 | | -0.80 ± 0.00 | | 131.7 |
| Contr_l + PMI_L | ALL | -0.80 ± 0.00 | | -0.80 ± 0.00 | | -0.80 ± 0.00 | | 84.0 |
| Contr_l + PMI_L | GAT | -0.82 ± 0.00 | | -0.81 ± 0.00 | | -0.80 ± 0.00 | | 156.7 |
| Contr_l + PMI_L | GCN | -0.82 ± 0.00 | | -0.81 ± 0.00 | | -0.80 ± 0.00 | | 157.7 |
| Contr_l + PMI_L | GIN | -0.81 ± 0.00 | | -0.80 ± 0.00 | | -0.80 ± 0.00 | | 108.3 |
| Contr_l + PMI_L | MPNN | -0.81 ± 0.00 | | -0.81 ± 0.00 | | -0.80 ± 0.00 | | 135.0 |
| Contr_l + PMI_L | PAGNN | -0.80 ± 0.00 | | -0.79 ± 0.00 | | -0.79 ± 0.00 | | 25.7 |
| Contr_l + PMI_L | SAGE | -0.81 ± 0.00 | | -0.81 ± 0.00 | | -0.79 ± 0.00 | | 97.0 |
| Contr_l + PMI_L + PR_L | ALL | -0.80 ± 0.00 | | -0.79 ± 0.00 | | -0.79 ± 0.00 | | 27.0 |
| Contr_l + PMI_L + PR_L | GAT | -0.81 ± 0.00 | | -0.81 ± 0.00 | | -0.79 ± 0.00 | | 98.3 |
| Contr_l + PMI_L + PR_L | GCN | -0.82 ± 0.00 | | -0.81 ± 0.00 | | -0.80 ± 0.00 | | 160.3 |
| Contr_l + PMI_L + PR_L | GIN | -0.81 ± 0.00 | | -0.80 ± 0.00 | | -0.79 ± 0.00 | | 72.7 |

Selfcluster Continued (↑)

| Loss Type | Model | CORA | | Citeseer | | Bitcoin Fraud Transaction | | Average Rank |
|---|---|---|---|---|---|---|---|---|
| Contr_l + PMI_L + PR_L | MPNN | -0.81 | ± | -0.80 | ± | -0.79 | ± | 73.7 |
| | | 0.00 | | 0.00 | | 0.00 | | |
| Contr_l + PMI_L + PR_L | PAGNN | -0.80 | ± | -0.79 | ± | -0.79 | ± | 29.0 |
| | | 0.00 | | 0.00 | | 0.00 | | |
| Contr_l + PMI_L + PR_L | SAGE | -0.81 | ± | -0.81 | ± | -0.79 | ± | 101.3 |
| | | 0.00 | | 0.01 | | 0.00 | | |
| Contr_l + PMI_L + PR_L + Triplet_L | ALL | -0.81 | ± | -0.80 | ± | -0.80 | ± | 112.7 |
| | | 0.00 | | 0.00 | | 0.00 | | |
| Contr_l + PMI_L + PR_L + Triplet_L | GAT | -0.81 | ± | -0.81 | ± | -0.80 | ± | 140.0 |
| | | 0.00 | | 0.00 | | 0.00 | | |
| Contr_l + PMI_L + PR_L + Triplet_L | GCN | -0.81 | ± | -0.81 | ± | -0.80 | ± | 141.0 |
| | | 0.00 | | 0.00 | | 0.00 | | |
| Contr_l + PMI_L + PR_L + Triplet_L | GIN | -0.81 | ± | -0.80 | ± | -0.80 | ± | 115.0 |
| | | 0.00 | | 0.00 | | 0.00 | | |
| Contr_l + PMI_L + PR_L + Triplet_L | MPNN | -0.81 | ± | -0.80 | ± | -0.80 | ± | 116.0 |
| | | 0.00 | | 0.00 | | 0.00 | | |
| Contr_l + PMI_L + PR_L + Triplet_L | PAGNN | -0.80 | ± | -0.80 | ± | -0.79 | ± | 52.7 |
| | | 0.00 | | 0.00 | | 0.00 | | |
| Contr_l + PMI_L + PR_L + Triplet_L | SAGE | -0.81 | ± | -0.81 | ± | -0.80 | ± | 143.3 |
| | | 0.00 | | 0.00 | | 0.00 | | |
| Contr_l + PR_L | ALL | -0.79 | ± | -0.79 | ± | -0.79 | ± | 15.0 |
| | | 0.00 | | 0.01 | | 0.00 | | |
| Contr_l + PR_L | GAT | -0.80 | ± | -0.79 | ± | -0.79 | ± | 31.7 |
| | | 0.01 | | 0.00 | | 0.00 | | |
| Contr_l + PR_L | GCN | -0.80 | ± | -0.80 | ± | -0.80 | ± | 92.7 |
| | | 0.00 | | 0.00 | | 0.00 | | |
| Contr_l + PR_L | GIN | -0.79 | ± | -0.79 | ± | -0.79 | ± | 16.7 |
| | | 0.00 | | 0.01 | | 0.00 | | |
| Contr_l + PR_L | MPNN | -0.79 | ± | -0.79 | ± | -0.79 | ± | 17.7 |
| | | 0.00 | | 0.00 | | 0.00 | | |
| Contr_l + PR_L | PAGNN | -0.79 | ± | -0.79 | ± | -0.79 | ± | 18.7 |
| | | 0.00 | | 0.00 | | 0.00 | | |

Selfcluster Continued (↑)

| Loss Type | Model | CORA | | Citeseer | | Bitcoin Fraud Transaction | | Average Rank |
|---|---|---|---|---|---|---|---|---|
| Contr_l + PR_L | SAGE | -0.80 | ± | -0.80 | ± | -0.79 | ± | 56.3 |
| | | 0.00 | | 0.00 | | 0.00 | | |
| Contr_l + PR_L + Triplet_L | ALL | -0.80 | ± | -0.80 | ± | -0.80 | ± | 94.3 |
| | | 0.00 | | 0.00 | | 0.00 | | |
| Contr_l + PR_L + Triplet_L | GAT | -0.81 | ± | -0.81 | ± | -0.80 | ± | 145.0 |
| | | 0.00 | | 0.01 | | 0.00 | | |
| Contr_l + PR_L + Triplet_L | GCN | -0.81 | ± | -0.80 | ± | -0.80 | ± | 120.3 |
| | | 0.01 | | 0.00 | | 0.00 | | |
| Contr_l + PR_L + Triplet_L | GIN | -0.81 | ± | -0.80 | ± | -0.80 | ± | 121.3 |
| | | 0.00 | | 0.00 | | 0.00 | | |
| Contr_l + PR_L + Triplet_L | MPNN | -0.80 | ± | -0.80 | ± | -0.80 | ± | 97.0 |
| | | 0.00 | | 0.00 | | 0.00 | | |
| Contr_l + PR_L + Triplet_L | PAGNN | -0.80 | ± | -0.80 | ± | -0.79 | ± | 59.3 |
| | | 0.00 | | 0.00 | | 0.00 | | |
| Contr_l + PR_L + Triplet_L | SAGE | -0.81 | ± | -0.80 | ± | -0.80 | ± | 123.3 |
| | | 0.00 | | 0.00 | | 0.00 | | |
| Contr_l + Triplet_L | ALL | -0.81 | ± | -0.81 | ± | -0.80 | ± | 148.3 |
| | | 0.00 | | 0.00 | | 0.00 | | |
| Contr_l + Triplet_L | GAT | -0.82 | ± | -0.81 | ± | -0.80 | ± | 168.0 |
| | | 0.01 | | 0.00 | | 0.00 | | |
| Contr_l + Triplet_L | GCN | -0.81 | ± | -0.81 | ± | -0.80 | ± | 150.0 |
| | | 0.00 | | 0.00 | | 0.00 | | |
| Contr_l + Triplet_L | GIN | -0.81 | ± | -0.81 | ± | -0.80 | ± | 151.0 |
| | | 0.00 | | 0.00 | | 0.00 | | |
| Contr_l + Triplet_L | MPNN | -0.81 | ± | -0.81 | ± | -0.80 | ± | 152.0 |
| | | 0.00 | | 0.00 | | 0.00 | | |
| Contr_l + Triplet_L | PAGNN | -0.81 | ± | -0.81 | ± | -0.79 | ± | 112.7 |
| | | 0.00 | | 0.00 | | 0.00 | | |
| Contr_l + Triplet_L | SAGE | -0.81 | ± | -0.81 | ± | -0.80 | ± | 153.7 |
| | | 0.00 | | 0.00 | | 0.00 | | |
| CrossE_L | ALL | -0.79 | ± | -0.79 | ± | -0.79 | ± | 20.7 |
| | | 0.00 | | 0.00 | | 0.00 | | |

Selfcluster Continued (↑)

| Loss Type | Model | CORA | | Citeseer | | Bitcoin Fraud Transaction | | Average Rank |
|---|---|---|---|---|---|---|---|---|
| CrossE_L | GAT | -0.79 0.00 | ± | -0.79 0.00 | ± | -0.79 0.00 | ± | 21.7 |
| CrossE_L | GCN | -0.79 0.00 | ± | -0.79 0.00 | ± | -0.79 0.00 | ± | 22.7 |
| CrossE_L | GIN | -0.79 0.00 | ± | -0.79 0.00 | ± | -0.79 0.00 | ± | 23.7 |
| CrossE_L | MPNN | -0.79 0.00 | ± | -0.79 0.00 | ± | -0.79 0.00 | ± | 24.7 |
| CrossE_L | PAGNN | -0.79 0.00 | ± | -0.79 0.00 | ± | -0.79 0.00 | ± | 25.7 |
| CrossE_L | SAGE | -0.79 0.00 | ± | -0.79 0.00 | ± | -0.79 0.00 | ± | 26.7 |
| CrossE_L + PMI_L | ALL | -0.80 0.01 | ± | -0.80 0.00 | ± | -0.80 0.00 | ± | 101.3 |
| CrossE_L + PMI_L | GAT | -0.82 0.00 | ± | -0.81 0.00 | ± | -0.80 0.00 | ± | 172.3 |
| CrossE_L + PMI_L | GCN | -0.82 0.00 | ± | -0.81 0.00 | ± | -0.80 0.00 | ± | 173.3 |
| CrossE_L + PMI_L | GIN | -0.81 0.00 | ± | -0.80 0.00 | ± | -0.80 0.00 | ± | 129.7 |
| CrossE_L + PMI_L | MPNN | -0.81 0.00 | ± | -0.81 0.00 | ± | -0.80 0.00 | ± | 157.0 |
| CrossE_L + PMI_L | PAGNN | -0.80 0.00 | ± | -0.79 0.00 | ± | -0.79 0.00 | ± | 42.3 |
| CrossE_L + PMI_L | SAGE | -0.81 0.00 | ± | -0.81 0.00 | ± | -0.80 0.00 | ± | 158.0 |
| CrossE_L + PMI_L + PR_L | ALL | -0.80 0.00 | ± | -0.79 0.00 | ± | -0.79 0.00 | ± | 43.3 |
| CrossE_L + PMI_L + PR_L | GAT | -0.81 0.00 | ± | -0.80 0.01 | ± | -0.80 0.01 | ± | 132.0 |
| CrossE_L + PMI_L + PR_L | GCN | -0.82 0.00 | ± | -0.81 0.00 | ± | -0.80 0.00 | ± | 176.3 |

Selfcluster Continued (↑)

| Loss Type | Model | CORA | | Citeseer | | Bitcoin Fraud Transaction | | Average Rank |
|---|---|---|---|---|---|---|---|---|
| CrossE_L + PMI_L + PR_L | GIN | -0.81 | ± | -0.80 | ± | -0.79 | ± | 93.3 |
| | | 0.00 | | 0.00 | | 0.00 | | |
| CrossE_L + PMI_L + PR_L | MPNN | -0.81 | ± | -0.81 | ± | -0.80 | ± | 160.7 |
| | | 0.00 | | 0.01 | | 0.01 | | |
| CrossE_L + PMI_L + PR_L | PAGNN | -0.80 | ± | -0.79 | ± | -0.79 | ± | 44.7 |
| | | 0.00 | | 0.00 | | 0.00 | | |
| CrossE_L + PMI_L + PR_L | SAGE | -0.81 | ± | -0.81 | ± | -0.79 | ± | 122.0 |
| | | 0.00 | | 0.00 | | 0.00 | | |
| CrossE_L + PMI_L + PR_L + Triplet_L | ALL | -0.81 | ± | -0.80 | ± | -0.80 | ± | 135.0 |
| | | 0.00 | | 0.00 | | 0.00 | | |
| CrossE_L + PMI_L + PR_L + Triplet_L | GAT | -0.81 | ± | -0.81 | ± | -0.80 | ± | 163.0 |
| | | 0.00 | | 0.00 | | 0.00 | | |
| CrossE_L + PMI_L + PR_L + Triplet_L | GCN | -0.82 | ± | -0.81 | ± | -0.80 | ± | 179.3 |
| | | 0.00 | | 0.00 | | 0.00 | | |
| CrossE_L + PMI_L + PR_L + Triplet_L | GIN | -0.81 | ± | -0.80 | ± | -0.80 | ± | 137.0 |
| | | 0.00 | | 0.00 | | 0.00 | | |
| CrossE_L + PMI_L + PR_L + Triplet_L | MPNN | -0.81 | ± | -0.81 | ± | -0.80 | ± | 165.3 |
| | | 0.00 | | 0.01 | | 0.00 | | |
| CrossE_L + PMI_L + PR_L + Triplet_L | PAGNN | -0.80 | ± | -0.80 | ± | -0.79 | ± | 68.3 |
| | | 0.00 | | 0.00 | | 0.00 | | |
| CrossE_L + PMI_L + PR_L + Triplet_L | SAGE | -0.81 | ± | -0.81 | ± | -0.79 | ± | 125.7 |
| | | 0.00 | | 0.00 | | 0.00 | | |
| CrossE_L + PMI_L + Triplet_L | ALL | -0.81 | ± | -0.81 | ± | -0.80 | ± | 167.0 |
| | | 0.00 | | 0.00 | | 0.00 | | |
| CrossE_L + PMI_L + Triplet_L | GAT | -0.82 | ± | -0.81 | ± | -0.80 | ± | 182.3 |
| | | 0.00 | | 0.00 | | 0.00 | | |
| CrossE_L + PMI_L + Triplet_L | GCN | -0.82 | ± | -0.81 | ± | -0.80 | ± | 183.3 |
| | | 0.00 | | 0.00 | | 0.00 | | |
| CrossE_L + PMI_L + Triplet_L | GIN | -0.81 | ± | -0.80 | ± | -0.80 | ± | 140.7 |
| | | 0.00 | | 0.01 | | 0.00 | | |
| CrossE_L + PMI_L + Triplet_L | MPNN | -0.81 | ± | -0.81 | ± | -0.80 | ± | 170.0 |
| | | 0.00 | | 0.00 | | 0.00 | | |

<div align="center">Selfcluster Continued (↑)</div>

| Loss Type | | | | Model | CORA | | Citeseer | | Bitcoin Fraud Transaction | | Average Rank |
|---|---|---|---|---|---|---|---|---|---|---|---|
| CrossE_L + PMI_L + Triplet_L | | | | PAGNN | -0.80 ± 0.00 | ± | -0.79 0.00 | ± | -0.79 0.00 | ± | 47.0 |
| CrossE_L + PMI_L + Triplet_L | | | | SAGE | -0.81 ± 0.00 | ± | -0.81 0.00 | ± | -0.80 0.00 | ± | 171.0 |
| CrossE_L + PR_L | | | | ALL | -0.79 ± 0.00 | ± | -0.79 0.00 | ± | -0.79 0.00 | ± | 31.7 |
| CrossE_L + PR_L | | | | GAT | -0.79 ± 0.00 | ± | -0.79 0.00 | ± | -0.79 0.00 | ± | 32.7 |
| CrossE_L + PR_L | | | | GCN | -0.79 ± 0.00 | ± | -0.79 0.00 | ± | -0.79 0.00 | ± | 33.7 |
| CrossE_L + PR_L | | | | GIN | -0.79 ± 0.00 | ± | -0.79 0.00 | ± | -0.79 0.00 | ± | 34.7 |
| CrossE_L + PR_L | | | | MPNN | -0.79 ± 0.00 | ± | -0.79 0.00 | ± | -0.79 0.00 | ± | 35.7 |
| CrossE_L + PR_L | | | | PAGNN | -0.79 ± 0.00 | ± | -0.79 0.00 | ± | -0.79 0.00 | ± | 36.7 |
| CrossE_L + PR_L | | | | SAGE | -0.79 ± 0.00 | ± | -0.79 0.00 | ± | -0.79 0.00 | ± | 37.7 |
| CrossE_L + PR_L + Triplet_L | | | | ALL | -0.80 ± 0.00 | ± | -0.80 0.00 | ± | -0.79 0.01 | ± | 73.0 |
| CrossE_L + PR_L + Triplet_L | | | | GAT | -0.80 ± 0.00 | ± | -0.80 0.00 | ± | -0.79 0.00 | ± | 74.0 |
| CrossE_L + PR_L + Triplet_L | | | | GCN | -0.80 ± 0.00 | ± | -0.80 0.00 | ± | -0.80 0.00 | ± | 114.0 |
| CrossE_L + PR_L + Triplet_L | | | | GIN | -0.80 ± 0.00 | ± | -0.80 0.00 | ± | -0.79 0.00 | ± | 75.7 |
| CrossE_L + PR_L + Triplet_L | | | | MPNN | -0.79 ± 0.00 | ± | -0.79 0.00 | ± | -0.80 0.00 | ± | 78.7 |
| CrossE_L + PR_L + Triplet_L | | | | PAGNN | -0.79 ± 0.00 | ± | -0.80 0.00 | ± | -0.79 0.00 | ± | 61.7 |
| CrossE_L + PR_L + Triplet_L | | | | SAGE | -0.81 ± 0.00 | ± | -0.80 0.00 | ± | -0.79 0.01 | ± | 106.3 |

Selfcluster Continued (↑)

| Loss Type | Model | CORA | | Citeseer | | Bitcoin Fraud Transaction | | Average Rank |
|---|---|---|---|---|---|---|---|---|
| CrossE_L + Triplet_L | ALL | -0.81 | ± | -0.81 | ± | -0.80 | ± | 173.0 |
| | | 0.00 | | 0.00 | | 0.00 | | |
| CrossE_L + Triplet_L | GAT | -0.82 | ± | -0.81 | ± | -0.80 | ± | 187.3 |
| | | 0.00 | | 0.00 | | 0.00 | | |
| CrossE_L + Triplet_L | GCN | -0.82 | ± | -0.81 | ± | -0.80 | ± | 188.3 |
| | | 0.00 | | 0.00 | | 0.00 | | |
| CrossE_L + Triplet_L | GIN | -0.81 | ± | -0.81 | ± | -0.80 | ± | 175.3 |
| | | 0.00 | | 0.00 | | 0.00 | | |
| CrossE_L + Triplet_L | MPNN | -0.81 | ± | -0.81 | ± | -0.80 | ± | 176.3 |
| | | 0.01 | | 0.00 | | 0.00 | | |
| CrossE_L + Triplet_L | PAGNN | -0.81 | ± | -0.81 | ± | -0.79 | ± | 137.0 |
| | | 0.00 | | 0.00 | | 0.00 | | |
| CrossE_L + Triplet_L | SAGE | -0.82 | ± | -0.81 | ± | -0.80 | ± | 191.0 |
| | | 0.00 | | 0.00 | | 0.00 | | |
| PMI_L | ALL | -0.80 | ± | -0.80 | ± | -0.80 | ± | 118.7 |
| | | 0.00 | | 0.00 | | 0.00 | | |
| PMI_L | GAT | -0.82 | ± | -0.81 | ± | -0.80 | ± | 192.3 |
| | | 0.00 | | 0.00 | | 0.00 | | |
| PMI_L | GCN | -0.82 | ± | -0.81 | ± | -0.80 | ± | 193.3 |
| | | 0.00 | | 0.00 | | 0.00 | | |
| PMI_L | GIN | -0.81 | ± | -0.80 | ± | -0.80 | ± | 150.7 |
| | | 0.00 | | 0.00 | | 0.00 | | |
| PMI_L | MPNN | -0.81 | ± | -0.81 | ± | -0.80 | ± | 181.0 |
| | | 0.00 | | 0.00 | | 0.00 | | |
| PMI_L | PAGNN | -0.80 | ± | -0.79 | ± | -0.79 | ± | 56.7 |
| | | 0.00 | | 0.00 | | 0.00 | | |
| PMI_L | SAGE | -0.81 | ± | -0.81 | ± | -0.80 | ± | 182.0 |
| | | 0.00 | | 0.00 | | 0.00 | | |
| PMI_L + PR_L | ALL | -0.80 | ± | -0.79 | ± | -0.79 | ± | 57.7 |
| | | 0.00 | | 0.00 | | 0.00 | | |
| PMI_L + PR_L | GAT | -0.81 | ± | -0.81 | ± | -0.79 | ± | 141.3 |
| | | 0.00 | | 0.01 | | 0.00 | | |

Selfcluster Continued (↑)

| Loss Type | Model | CORA | | Citeseer | | Bitcoin Fraud Transaction | | Average Rank |
|---|---|---|---|---|---|---|---|---|
| PMI_L + PR_L | GCN | -0.81 | ± | -0.81 | ± | -0.80 | ± | 183.7 |
| | | 0.00 | | 0.00 | | 0.00 | | |
| PMI_L + PR_L | GIN | -0.81 | ± | -0.80 | ± | -0.79 | ± | 112.3 |
| | | 0.00 | | 0.00 | | 0.00 | | |
| PMI_L + PR_L | MPNN | -0.81 | ± | -0.80 | ± | -0.79 | ± | 113.3 |
| | | 0.00 | | 0.00 | | 0.00 | | |
| PMI_L + PR_L | PAGNN | -0.80 | ± | -0.80 | ± | -0.79 | ± | 82.3 |
| | | 0.00 | | 0.01 | | 0.00 | | |
| PMI_L + PR_L | SAGE | -0.81 | ± | -0.81 | ± | -0.79 | ± | 144.7 |
| | | 0.00 | | 0.00 | | 0.00 | | |
| PMI_L + PR_L + Triplet_L | ALL | -0.81 | ± | -0.80 | ± | -0.80 | ± | 156.0 |
| | | 0.00 | | 0.00 | | 0.00 | | |
| PMI_L + PR_L + Triplet_L | GAT | -0.81 | ± | -0.81 | ± | -0.80 | ± | 186.7 |
| | | 0.00 | | 0.00 | | 0.00 | | |
| PMI_L + PR_L + Triplet_L | GCN | -0.82 | ± | -0.81 | ± | -0.80 | ± | 198.3 |
| | | 0.00 | | 0.00 | | 0.00 | | |
| PMI_L + PR_L + Triplet_L | GIN | -0.81 | ± | -0.80 | ± | -0.80 | ± | 158.0 |
| | | 0.00 | | 0.00 | | 0.00 | | |
| PMI_L + PR_L + Triplet_L | MPNN | -0.81 | ± | -0.80 | ± | -0.80 | ± | 159.0 |
| | | 0.00 | | 0.00 | | 0.00 | | |
| PMI_L + PR_L + Triplet_L | PAGNN | -0.80 | ± | -0.80 | ± | -0.79 | ± | 84.7 |
| | | 0.00 | | 0.00 | | 0.00 | | |
| PMI_L + PR_L + Triplet_L | SAGE | -0.81 | ± | -0.81 | ± | -0.79 | ± | 148.0 |
| | | 0.00 | | 0.00 | | 0.00 | | |
| PMI_L + Triplet_L | ALL | -0.81 | ± | -0.81 | ± | -0.80 | ± | 190.3 |
| | | 0.00 | | 0.00 | | 0.00 | | |
| PMI_L + Triplet_L | GAT | -0.82 | ± | -0.81 | ± | -0.80 | ± | 201.0 |
| | | 0.00 | | 0.00 | | 0.00 | | |
| PMI_L + Triplet_L | GCN | -0.82 | ± | -0.81 | ± | -0.80 | ± | 202.0 |
| | | 0.00 | | 0.00 | | 0.00 | | |
| PMI_L + Triplet_L | GIN | -0.81 | ± | -0.80 | ± | -0.80 | ± | 162.0 |
| | | 0.00 | | 0.00 | | 0.00 | | |

Selfcluster Continued (↑)

| Loss Type | Model | CORA | | Citeseer | | Bitcoin Fraud Transaction | | Average Rank |
|---|---|---|---|---|---|---|---|---|
| PMI_L + Triplet_L | MPNN | -0.81 | ± | -0.81 | ± | -0.80 | ± | 193.3 |
| | | 0.00 | | 0.00 | | 0.00 | | |
| PMI_L + Triplet_L | PAGNN | -0.80 | ± | -0.79 | ± | -0.79 | ± | 61.7 |
| | | 0.00 | | 0.00 | | 0.00 | | |
| PMI_L + Triplet_L | SAGE | -0.81 | ± | -0.81 | ± | -0.80 | ± | 194.3 |
| | | 0.00 | | 0.00 | | 0.00 | | |
| PR_L | ALL | -0.79 | ± | -0.79 | ± | -0.79 | ± | 46.0 |
| | | 0.00 | | 0.00 | | 0.00 | | |
| PR_L | GAT | -0.79 | ± | -0.79 | ± | -0.79 | ± | 47.0 |
| | | 0.00 | | 0.00 | | 0.00 | | |
| PR_L | GCN | -0.79 | ± | -0.79 | ± | -0.79 | ± | 48.0 |
| | | 0.00 | | 0.00 | | 0.00 | | |
| PR_L | GIN | -0.79 | ± | -0.79 | ± | -0.79 | ± | 49.0 |
| | | 0.00 | | 0.00 | | 0.00 | | |
| PR_L | MPNN | -0.79 | ± | -0.79 | ± | -0.79 | ± | 50.0 |
| | | 0.00 | | 0.00 | | 0.00 | | |
| PR_L | PAGNN | -0.79 | ± | -0.79 | ± | -0.79 | ± | 51.0 |
| | | 0.00 | | 0.00 | | 0.00 | | |
| PR_L | SAGE | -0.79 | ± | -0.79 | ± | -0.79 | ± | 52.0 |
| | | 0.00 | | 0.00 | | 0.00 | | |
| PR_L + Triplet_L | ALL | -0.79 | ± | -0.79 | ± | -0.79 | ± | 53.0 |
| | | 0.00 | | 0.00 | | 0.00 | | |
| PR_L + Triplet_L | GAT | -0.80 | ± | -0.79 | ± | -0.79 | ± | 68.0 |
| | | 0.01 | | 0.00 | | 0.00 | | |
| PR_L + Triplet_L | GCN | -0.80 | ± | -0.80 | ± | -0.80 | ± | 130.3 |
| | | 0.00 | | 0.00 | | 0.00 | | |
| PR_L + Triplet_L | GIN | -0.79 | ± | -0.80 | ± | -0.79 | ± | 76.3 |
| | | 0.00 | | 0.00 | | 0.00 | | |
| PR_L + Triplet_L | MPNN | -0.79 | ± | -0.79 | ± | -0.80 | ± | 95.3 |
| | | 0.00 | | 0.00 | | 0.01 | | |
| PR_L + Triplet_L | PAGNN | -0.79 | ± | -0.79 | ± | -0.79 | ± | 56.0 |
| | | 0.00 | | 0.00 | | 0.00 | | |

Selfcluster Continued (↑)

| Loss Type | Model | CORA | | Citeseer | | Bitcoin Fraud Transaction | | Average Rank |
|-----------|-------|------|---|----------|---|---------------------------|---|--------------|
| PR_L + Triplet_L | SAGE | -0.79 ± 0.00 | | -0.79 ± 0.01 | | -0.79 ± 0.00 | | 57.0 |
| Triplet_L | ALL | -0.81 ± 0.00 | | -0.81 ± 0.00 | | -0.80 ± 0.00 | | 196.0 |
| Triplet_L | GAT | -0.82 ± 0.00 | | -0.81 ± 0.00 | | -0.80 ± 0.00 | | 206.0 |
| Triplet_L | GCN | -0.82 ± 0.00 | | -0.81 ± 0.00 | | -0.80 ± 0.00 | | 207.0 |
| Triplet_L | GIN | -0.82 ± 0.00 | | -0.81 ± 0.00 | | -0.80 ± 0.00 | | 208.0 |
| Triplet_L | MPNN | -0.81 ± 0.00 | | -0.81 ± 0.00 | | -0.80 ± 0.00 | | 199.0 |
| Triplet_L | PAGNN | -0.81 ± 0.00 | | -0.81 ± 0.00 | | -0.79 ± 0.00 | | 158.7 |
| Triplet_L | SAGE | -0.82 ± 0.01 | | -0.81 ± 0.00 | | -0.80 ± 0.00 | | 210.0 |

Table 21. Rankme Performance (↑): Top-ranked results are highlighted in **1st**, second-ranked in **2nd**, and third-ranked in **3rd**.

| Loss Type | Model | CORA | | Citeseer | | Bitcoin Fraud Transaction | | Average Rank |
|-----------|-------|------|---|----------|---|---------------------------|---|--------------|
| Contr_l | ALL | 264.22 ± 5.76 | | 288.18 ± 8.56 | | 270.55 ± 8.41 | | 177.0 |
| Contr_l | GAT | 428.42 ± 1.70 | | 446.72 ± 0.76 | | 424.66 ± 1.93 | | 24.7 |
| Contr_l | GCN | 414.09 ± 1.91 | | 429.18 ± 2.60 | | 418.62 ± 2.28 | | 52.3 |
| Contr_l | GIN | 384.08 ± 2.37 | | 402.84 ± 3.41 | | 339.31 ± 8.71 | | 113.3 |

Continued on next page

Rankme Continued (↑)

| Loss Type | Model | CORA | Citeseer | Bitcoin Fraud Transaction | Average Rank |
|---|---|---|---|---|---|
| Contr_l | MPNN | 391.57 ± 5.68 | 414.87 ± 8.44 | 380.63 ± 6.87 | 92.3 |
| Contr_l | PAGNN | 348.78 ± 2.20 | 350.00 ± 2.26 | 176.96 ± 11.85 | 164.7 |
| Contr_l | SAGE | 375.62 ± 2.77 | 394.10 ± 2.66 | 369.56 ± 5.52 | 111.3 |
| Contr_l + CrossE_L | ALL | 259.36 ± 9.74 | 260.10 ± 14.94 | 271.83 ± 26.32 | 181.7 |
| Contr_l + CrossE_L | GAT | 423.87 ± 3.17 | 440.53 ± 0.42 | 406.94 ± 11.27 | 44.3 |
| Contr_l + CrossE_L | GCN | 408.23 ± 3.54 | 426.79 ± 1.39 | 413.29 ± 2.68 | 66.7 |
| Contr_l + CrossE_L | GIN | 374.03 ± 2.03 | 387.71 ± 6.37 | 328.89 ± 6.29 | 129.7 |
| Contr_l + CrossE_L | MPNN | 381.82 ± 11.37 | 406.28 ± 6.06 | 350.48 ± 11.44 | 107.7 |
| Contr_l + CrossE_L | PAGNN | 344.66 ± 2.90 | 337.74 ± 3.59 | 146.93 ± 6.12 | 174.7 |
| Contr_l + CrossE_L | SAGE | 362.95 ± 2.76 | 376.53 ± 7.40 | 343.29 ± 13.02 | 131.7 |
| Contr_l + CrossE_L + PMI_L | ALL | 320.45 ± 8.03 | 374.89 ± 5.99 | 353.39 ± 2.85 | 141.0 |
| Contr_l + CrossE_L + PMI_L | GAT | 449.49 ± 1.34 | 457.64 ± 0.96 | 435.66 ± 1.17 | 7.0 |
| Contr_l + CrossE_L + PMI_L | GCN | 420.95 ± 1.17 | 430.74 ± 1.86 | 415.96 ± 3.32 | 48.3 |
| Contr_l + CrossE_L + PMI_L | GIN | 395.29 ± 5.70 | 392.04 ± 3.52 | 360.59 ± 6.62 | 107.3 |
| Contr_l + CrossE_L + PMI_L | MPNN | 410.41 ± 3.29 | 429.58 ± 1.95 | 420.38 ± 2.29 | 48.3 |
| Contr_l + CrossE_L + PMI_L | PAGNN | 355.04 ± 4.57 | 351.93 ± 3.00 | 178.02 ± 3.36 | 160.7 |

Rankme Continued (↑)

| Loss Type | Model | CORA | Citeseer | Bitcoin Fraud Transaction | Average Rank |
|---|---|---|---|---|---|
| Contr_l + CrossE_L + PMI_L | SAGE | 446.21 ± 2.78 | 449.46 ± 2.65 | 415.68 ± 11.53 | 28.7 |
| Contr_l + CrossE_L + PMI_L + PR_L | ALL | 316.31 ± 14.01 | 369.88 ± 3.57 | 349.78 ± 2.26 | 145.7 |
| Contr_l + CrossE_L + PMI_L + PR_L | GAT | 446.09 ± 3.18 | 455.93 ± 0.88 | 424.65 ± 5.56 | 15.3 |
| Contr_l + CrossE_L + PMI_L + PR_L | GCN | 422.00 ± 1.94 | 428.61 ± 1.60 | 420.49 ± 2.67 | 42.7 |
| Contr_l + CrossE_L + PMI_L + PR_L | GIN | 389.75 ± 4.35 | 373.01 ± 14.56 | 319.14 ± 24.42 | 130.7 |
| Contr_l + CrossE_L + PMI_L + PR_L | MPNN | 408.77 ± 2.61 | 419.94 ± 6.03 | 387.70 ± 17.76 | 77.0 |
| Contr_l + CrossE_L + PMI_L + PR_L | PAGNN | 355.90 ± 1.96 | 354.46 ± 3.34 | 165.13 ± 5.92 | 165.0 |
| Contr_l + CrossE_L + PMI_L + PR_L | SAGE | 446.90 ± 2.63 | 449.09 ± 3.58 | 356.19 ± 12.72 | 49.0 |
| Contr_l + CrossE_L + PMI_L + PR_L + Triplet_L | ALL | 315.81 ± 8.56 | 342.53 ± 9.18 | 337.06 ± 12.28 | 159.7 |
| Contr_l + CrossE_L + PMI_L + PR_L + Triplet_L | GAT | 447.12 ± 2.17 | 453.59 ± 5.24 | 435.61 ± 4.92 | 10.7 |
| Contr_l + CrossE_L + PMI_L + PR_L + Triplet_L | GCN | 421.50 ± 2.12 | 430.89 ± 1.61 | 420.47 ± 4.10 | 37.3 |
| Contr_l + CrossE_L + PMI_L + PR_L + Triplet_L | GIN | 397.20 ± 6.15 | 395.28 ± 1.67 | 371.70 ± 7.73 | 99.3 |
| Contr_l + CrossE_L + PMI_L + PR_L + Triplet_L | MPNN | 406.01 ± 0.52 | 421.28 ± 0.87 | 418.94 ± 6.44 | 62.3 |
| Contr_l + CrossE_L + PMI_L + PR_L + Triplet_L | PAGNN | 354.79 ± 6.38 | 364.02 ± 5.32 | 173.10 ± 6.23 | 158.7 |
| Contr_l + CrossE_L + PMI_L + PR_L + Triplet_L | SAGE | 429.93 ± 10.15 | 430.06 ± 1.95 | 393.03 ± 10.80 | 54.3 |
| Contr_l + CrossE_L + PMI_L + Triplet_L | ALL | 342.83 ± 1.08 | 375.54 ± 4.02 | 358.37 ± 2.67 | 135.0 |

Rankme Continued (↑)

| Loss Type | Model | CORA | Citeseer | Bitcoin Fraud Transaction | Average Rank |
|---|---|---|---|---|---|
| Contr_l + CrossE_L + PMI_L + Triplet_L | GAT | 449.78 ± 1.96 | 457.25 ± 0.46 | 439.97 ± 1.07 | 5.3 |
| Contr_l + CrossE_L + PMI_L + Triplet_L | GCN | 423.20 ± 3.65 | 430.98 ± 1.04 | 420.47 ± 2.77 | 35.3 |
| Contr_l + CrossE_L + PMI_L + Triplet_L | GIN | 399.14 ± 3.80 | 397.21 ± 4.37 | 372.24 ± 6.44 | 96.0 |
| Contr_l + CrossE_L + PMI_L + Triplet_L | MPNN | 408.81 ± 1.26 | 428.49 ± 4.32 | 425.33 ± 1.90 | 49.7 |
| Contr_l + CrossE_L + PMI_L + Triplet_L | PAGNN | 360.33 ± 8.07 | 359.76 ± 3.34 | 175.30 ± 6.16 | 152.7 |
| Contr_l + CrossE_L + PMI_L + Triplet_L | SAGE | 430.79 ± 5.04 | 431.87 ± 3.47 | 406.72 ± 1.21 | 46.0 |
| Contr_l + CrossE_L + PR_L | ALL | 181.43 ± 5.90 | 227.60 ± 14.57 | 158.21 ± 15.41 | 197.7 |
| Contr_l + CrossE_L + PR_L | GAT | 379.28 ± 22.88 | 394.82 ± 16.66 | 338.48 ± 5.73 | 121.0 |
| Contr_l + CrossE_L + PR_L | GCN | 398.25 ± 4.73 | 406.36 ± 1.49 | 409.08 ± 3.17 | 81.0 |
| Contr_l + CrossE_L + PR_L | GIN | 263.95 ± 28.81 | 329.09 ± 16.80 | 269.91 ± 14.60 | 173.7 |
| Contr_l + CrossE_L + PR_L | MPNN | 305.46 ± 13.59 | 337.21 ± 14.07 | 402.74 ± 9.37 | 140.7 |
| Contr_l + CrossE_L + PR_L | PAGNN | 264.54 ± 7.63 | 268.16 ± 5.89 | 177.43 ± 4.47 | 184.0 |
| Contr_l + CrossE_L + PR_L | SAGE | 358.34 ± 19.70 | 392.45 ± 16.70 | 353.45 ± 25.46 | 123.7 |
| Contr_l + CrossE_L + PR_L + Triplet_L | ALL | 262.08 ± 6.63 | 296.53 ± 10.68 | 265.93 ± 9.31 | 179.0 |
| Contr_l + CrossE_L + PR_L + Triplet_L | GAT | 417.11 ± 7.51 | 434.95 ± 3.24 | 414.09 ± 4.82 | 47.7 |
| Contr_l + CrossE_L + PR_L + Triplet_L | GCN | 412.67 ± 2.80 | 422.83 ± 1.96 | 418.14 ± 5.70 | 59.7 |

Rankme Continued (↑)

| Loss Type | Model | CORA | | Citeseer | | Bitcoin Fraud Transaction | | Average Rank |
|---|---|---|---|---|---|---|---|---|
| Contr_l + CrossE_L + PR_L + Triplet_L | GIN | 358.03 | ± 5.46 | 374.56 | ± 2.65 | 363.74 | ± 11.18 | 129.0 |
| Contr_l + CrossE_L + PR_L + Triplet_L | MPNN | 371.01 | ± 5.92 | 387.94 | ± 2.63 | 413.69 | ± 3.25 | 101.7 |
| Contr_l + CrossE_L + PR_L + Triplet_L | PAGNN | 295.51 | ± 5.69 | 324.35 | ± 11.05 | 167.68 | ± 8.33 | 180.0 |
| Contr_l + CrossE_L + PR_L + Triplet_L | SAGE | 370.24 | ± 3.62 | 395.38 | ± 6.61 | 369.07 | ± 6.64 | 112.0 |
| Contr_l + CrossE_L + Triplet_L | ALL | 280.48 | ± 13.24 | 287.97 | ± 14.05 | 289.53 | ± 11.71 | 174.0 |
| Contr_l + CrossE_L + Triplet_L | GAT | 430.45 | ± 1.90 | 446.61 | ± 1.36 | 428.94 | ± 1.74 | 23.0 |
| Contr_l + CrossE_L + Triplet_L | GCN | 415.52 | ± 1.54 | 427.69 | ± 1.36 | 418.63 | ± 2.17 | 54.7 |
| Contr_l + CrossE_L + Triplet_L | GIN | 389.77 | ± 2.53 | 399.97 | ± 2.94 | 348.17 | ± 5.24 | 108.7 |
| Contr_l + CrossE_L + Triplet_L | MPNN | 396.77 | ± 4.60 | 414.57 | ± 5.33 | 396.20 | ± 10.55 | 85.7 |
| Contr_l + CrossE_L + Triplet_L | PAGNN | 363.40 | ± 3.82 | 358.24 | ± 2.28 | 145.03 | ± 10.08 | 159.3 |
| Contr_l + CrossE_L + Triplet_L | SAGE | 383.85 | ± 3.12 | 397.93 | ± 2.63 | 376.06 | ± 2.76 | 101.7 |
| Contr_l + PMI_L | ALL | 322.70 | ± 5.57 | 358.59 | ± 7.05 | 341.71 | ± 14.29 | 151.7 |
| Contr_l + PMI_L | GAT | `449.80` | ± `0.97` | 457.47 | ± 0.71 | 437.18 | ± 1.43 | 5.0 |
| Contr_l + PMI_L | GCN | 423.57 | ± 3.04 | 428.95 | ± 1.79 | 417.26 | ± 2.50 | 49.0 |
| Contr_l + PMI_L | GIN | 398.18 | ± 5.38 | 395.03 | ± 5.54 | 361.70 | ± 7.12 | 102.0 |
| Contr_l + PMI_L | MPNN | 407.49 | ± 3.52 | 428.70 | ± 1.76 | 419.90 | ± 2.49 | 54.7 |

Rankme Continued (↑)

| Loss Type | Model | CORA | Citeseer | Bitcoin Fraud Transaction | Average Rank |
|---|---|---|---|---|---|
| Contr_l + PMI_L | PAGNN | 354.54 ± 9.33 | 348.37 ± 5.26 | 174.23 ± 8.23 | 165.3 |
| Contr_l + PMI_L | SAGE | 445.05 ± 1.10 | 441.98 ± 5.81 | 412.37 ± 3.94 | 35.3 |
| Contr_l + PMI_L + PR_L | ALL | 316.85 ± 14.51 | 370.23 ± 2.25 | 350.22 ± 2.01 | 144.3 |
| Contr_l + PMI_L + PR_L | GAT | 445.26 ± 2.14 | 446.62 ± 7.52 | 418.78 ± 8.33 | 27.0 |
| Contr_l + PMI_L + PR_L | GCN | 421.45 ± 2.32 | 427.03 ± 1.09 | 418.46 ± 2.81 | 53.3 |
| Contr_l + PMI_L + PR_L | GIN | 390.10 ± 5.92 | 375.78 ± 12.01 | 324.83 ± 19.67 | 127.3 |
| Contr_l + PMI_L + PR_L | MPNN | 407.71 ± 3.18 | 415.10 ± 5.97 | 387.97 ± 18.82 | 80.7 |
| Contr_l + PMI_L + PR_L | PAGNN | 347.21 ± 9.11 | 353.69 ± 8.89 | 167.35 ± 11.23 | 167.3 |
| Contr_l + PMI_L + PR_L | SAGE | 446.38 ± 0.38 | 436.37 ± 10.06 | 340.32 ± 11.65 | 60.0 |
| Contr_l + PMI_L + PR_L + Triplet_L | ALL | 307.92 ± 9.17 | 321.74 ± 9.81 | 315.22 ± 10.93 | 167.7 |
| Contr_l + PMI_L + PR_L + Triplet_L | GAT | 436.04 ± 6.28 | 440.94 ± 5.75 | 429.52 ± 4.15 | 22.3 |
| Contr_l + PMI_L + PR_L + Triplet_L | GCN | 419.04 ± 2.61 | 428.42 ± 2.19 | 420.05 ± 0.94 | 48.0 |
| Contr_l + PMI_L + PR_L + Triplet_L | GIN | 393.28 ± 6.27 | 388.09 ± 7.49 | 378.64 ± 1.98 | 104.7 |
| Contr_l + PMI_L + PR_L + Triplet_L | MPNN | 408.51 ± 3.82 | 416.43 ± 2.91 | 409.15 ± 8.09 | 72.7 |
| Contr_l + PMI_L + PR_L + Triplet_L | PAGNN | 359.07 ± 8.20 | 366.53 ± 5.89 | 177.17 ± 4.84 | 151.7 |
| Contr_l + PMI_L + PR_L + Triplet_L | SAGE | 405.95 ± 7.38 | 416.89 ± 6.17 | 385.72 ± 9.38 | 81.7 |

Rankme Continued (↑)

| Loss Type | Model | CORA | Citeseer | Bitcoin Fraud Transaction | Average Rank |
|---|---|---|---|---|---|
| Contr_l + PR_L | ALL | 188.24 ± 11.33 | 228.79 ± 13.38 | 166.15 ± 28.21 | 195.7 |
| Contr_l + PR_L | GAT | 362.51 ± 30.88 | 393.67 ± 9.04 | 339.43 ± 7.17 | 126.7 |
| Contr_l + PR_L | GCN | 395.43 ± 4.02 | 411.16 ± 6.22 | 410.27 ± 7.77 | 81.7 |
| Contr_l + PR_L | GIN | 267.59 ± 18.08 | 328.92 ± 21.16 | 265.20 ± 5.37 | 174.3 |
| Contr_l + PR_L | MPNN | 283.34 ± 9.40 | 339.64 ± 11.67 | 398.64 ± 3.23 | 141.7 |
| Contr_l + PR_L | PAGNN | 262.29 ± 11.15 | 265.84 ± 6.74 | 172.88 ± 9.54 | 187.7 |
| Contr_l + PR_L | SAGE | 355.22 ± 15.82 | 378.82 ± 11.64 | 343.89 ± 17.24 | 136.3 |
| Contr_l + PR_L + Triplet_L | ALL | 260.00 ± 11.29 | 306.46 ± 7.29 | 262.18 ± 7.91 | 179.3 |
| Contr_l + PR_L + Triplet_L | GAT | 418.38 ± 4.38 | 435.41 ± 3.13 | 414.35 ± 4.79 | 46.3 |
| Contr_l + PR_L + Triplet_L | GCN | 412.19 ± 2.76 | 422.09 ± 2.29 | 411.60 ± 5.33 | 65.7 |
| Contr_l + PR_L + Triplet_L | GIN | 360.17 ± 11.20 | 372.27 ± 4.91 | 370.14 ± 3.88 | 126.0 |
| Contr_l + PR_L + Triplet_L | MPNN | 376.21 ± 9.32 | 390.84 ± 4.02 | 411.37 ± 4.32 | 99.0 |
| Contr_l + PR_L + Triplet_L | PAGNN | 309.78 ± 10.93 | 330.31 ± 22.31 | 166.56 ± 7.95 | 178.0 |
| Contr_l + PR_L + Triplet_L | SAGE | 371.45 ± 2.27 | 393.57 ± 4.10 | 368.88 ± 4.38 | 113.7 |
| Contr_l + Triplet_L | ALL | 281.90 ± 10.02 | 299.16 ± 10.42 | 278.18 ± 7.51 | 173.0 |
| Contr_l + Triplet_L | GAT | 432.36 ± 2.82 | 446.49 ± 2.48 | 427.31 ± 4.19 | 23.0 |

Rankme Continued (↑)

| Loss Type | Model | CORA | Citeseer | Bitcoin Fraud Transaction | Average Rank |
|-----------|-------|------|----------|---------------------------|--------------|
| Contr_l + Triplet_L | GCN | 415.05 ± 1.17 | 427.72 ± 2.00 | 418.93 ± 2.33 | 53.3 |
| Contr_l + Triplet_L | GIN | 390.73 ± 2.93 | 398.87 ± 1.52 | 343.71 ± 4.57 | 111.3 |
| Contr_l + Triplet_L | MPNN | 399.84 ± 2.16 | 416.58 ± 3.13 | 400.44 ± 10.12 | 80.3 |
| Contr_l + Triplet_L | PAGNN | 363.48 ± 4.51 | 355.98 ± 4.03 | 137.01 ± 12.10 | 161.3 |
| Contr_l + Triplet_L | SAGE | 382.55 ± 2.52 | 397.98 ± 3.66 | 375.22 ± 8.55 | 102.0 |
| CrossE_L | ALL | 34.09 ± 9.91 | 33.59 ± 15.37 | 10.82 ± 4.12 | 204.7 |
| CrossE_L | GAT | 3.95 ± 2.29 | 5.50 ± 2.95 | 7.01 ± 4.99 | 208.0 |
| CrossE_L | GCN | 17.34 ± 5.01 | 16.07 ± 4.42 | 14.56 ± 3.86 | 205.7 |
| CrossE_L | GIN | 2.65 ± 2.12 | 2.48 ± 1.16 | 1.60 ± 0.39 | 209.7 |
| CrossE_L | MPNN | 16.43 ± 1.90 | 18.06 ± 3.64 | 6.90 ± 3.68 | 207.0 |
| CrossE_L | PAGNN | 21.11 ± 2.35 | 28.62 ± 2.92 | 13.35 ± 2.55 | 205.0 |
| CrossE_L | SAGE | 2.65 ± 0.99 | 7.56 ± 4.07 | 2.60 ± 1.08 | 209.0 |
| CrossE_L + PMI_L | ALL | 371.04 ± 6.88 | 391.04 ± 2.08 | 396.87 ± 1.69 | 105.3 |
| CrossE_L + PMI_L | GAT | 449.94 ± 0.74 | 458.30 ± 0.32 | 436.27 ± 1.44 | 4.7 |
| CrossE_L + PMI_L | GCN | 423.03 ± 2.96 | 429.15 ± 2.59 | 417.88 ± 1.37 | 48.0 |
| CrossE_L + PMI_L | GIN | 397.84 ± 4.57 | 388.04 ± 5.14 | 361.68 ± 4.82 | 107.7 |

Continued on next page

Rankme Continued (↑)

| Loss Type | Model | CORA | Citeseer | Bitcoin Fraud Transaction | Average Rank |
|-----------|-------|------|----------|---------------------------|--------------|
| CrossE_L + PMI_L | MPNN | 404.84 ± 1.43 | 427.21 ± 1.80 | 419.02 ± 2.85 | 61.0 |
| CrossE_L + PMI_L | PAGNN | 359.31 ± 4.53 | 352.44 ± 2.33 | 177.49 ± 15.12 | 156.3 |
| CrossE_L + PMI_L | SAGE | 443.11 ± 3.50 | 454.44 ± 3.00 | 437.10 ± 2.26 | 14.0 |
| CrossE_L + PMI_L + PR_L | ALL | 334.99 ± 5.93 | 372.44 ± 2.80 | 345.20 ± 5.05 | 144.0 |
| CrossE_L + PMI_L + PR_L | GAT | 448.12 ± 1.62 | 448.08 ± 8.03 | 418.86 ± 15.21 | 23.3 |
| CrossE_L + PMI_L + PR_L | GCN | 422.42 ± 3.57 | 430.45 ± 2.17 | 418.90 ± 3.22 | 42.3 |
| CrossE_L + PMI_L + PR_L | GIN | 388.40 ± 6.93 | 369.99 ± 18.76 | 296.30 ± 34.71 | 133.7 |
| CrossE_L + PMI_L + PR_L | MPNN | 405.76 ± 3.41 | 416.51 ± 6.74 | 396.22 ± 31.17 | 80.0 |
| CrossE_L + PMI_L + PR_L | PAGNN | 358.63 ± 8.73 | 355.88 ± 3.85 | 161.92 ± 17.42 | 161.7 |
| CrossE_L + PMI_L + PR_L | SAGE | 448.69 ± 2.21 | 452.69 ± 4.43 | 331.19 ± 25.21 | 55.0 |
| CrossE_L + PMI_L + PR_L + Triplet_L | ALL | 335.72 ± 12.35 | 350.53 ± 6.71 | 354.06 ± 7.22 | 148.0 |
| CrossE_L + PMI_L + PR_L + Triplet_L | GAT | 446.64 ± 1.62 | 452.39 ± 3.50 | 434.67 ± 3.88 | 13.0 |
| CrossE_L + PMI_L + PR_L + Triplet_L | GCN | 422.03 ± 2.98 | 430.63 ± 2.19 | 420.47 ± 1.84 | 38.3 |
| CrossE_L + PMI_L + PR_L + Triplet_L | GIN | 396.15 ± 4.19 | 395.04 ± 5.54 | 376.23 ± 5.62 | 97.7 |
| CrossE_L + PMI_L + PR_L + Triplet_L | MPNN | 409.36 ± 1.86 | 422.41 ± 4.42 | 419.12 ± 9.23 | 57.7 |
| CrossE_L + PMI_L + PR_L + Triplet_L | PAGNN | 356.30 ± 7.73 | 362.65 ± 3.76 | 178.11 ± 6.75 | 154.0 |

Rankme Continued (↑)

| Loss Type | Model | CORA | Citeseer | Bitcoin Fraud Transaction | Average Rank |
|-----------|-------|------|----------|---------------------------|--------------|
| CrossE_L + PMI_L + PR_L + Triplet_L | SAGE | 432.98 ± 5.33 | 433.05 ± 4.32 | 386.81 ± 13.34 | 49.3 |
| CrossE_L + PMI_L + Triplet_L | ALL | 357.01 ± 4.28 | 383.71 ± 4.12 | 364.54 ± 5.82 | 126.3 |
| CrossE_L + PMI_L + Triplet_L | GAT | 449.59 ± 0.25 | 457.62 ± 0.51 | 441.40 ± 0.70 | 4.0 |
| CrossE_L + PMI_L + Triplet_L | GCN | 423.71 ± 4.73 | 431.52 ± 0.85 | 419.62 ± 2.67 | 35.7 |
| CrossE_L + PMI_L + Triplet_L | GIN | 402.12 ± 2.76 | 402.09 ± 3.10 | 374.04 ± 6.32 | 92.0 |
| CrossE_L + PMI_L + Triplet_L | MPNN | 409.79 ± 1.82 | 429.79 ± 1.67 | 427.89 ± 1.53 | 43.7 |
| CrossE_L + PMI_L + Triplet_L | PAGNN | 361.94 ± 9.19 | 355.12 ± 5.60 | 177.79 ± 6.72 | 153.7 |
| CrossE_L + PMI_L + Triplet_L | SAGE | 427.20 ± 3.08 | 431.00 ± 3.19 | 411.87 ± 2.72 | 45.7 |
| CrossE_L + PR_L | ALL | 157.72 ± 12.21 | 208.97 ± 24.59 | 144.92 ± 3.10 | 200.3 |
| CrossE_L + PR_L | GAT | 299.07 ± 38.72 | 322.33 ± 11.03 | 140.59 ± 60.12 | 185.0 |
| CrossE_L + PR_L | GCN | 359.50 ± 11.74 | 378.70 ± 6.91 | 116.32 ± 44.28 | 155.3 |
| CrossE_L + PR_L | GIN | 237.84 ± 23.41 | 274.04 ± 23.98 | 195.22 ± 8.52 | 185.0 |
| CrossE_L + PR_L | MPNN | 278.87 ± 41.67 | 308.39 ± 9.03 | 384.04 ± 4.26 | 152.0 |
| CrossE_L + PR_L | PAGNN | 241.11 ± 31.20 | 265.65 ± 19.92 | 168.86 ± 14.48 | 191.3 |
| CrossE_L + PR_L | SAGE | 370.89 ± 21.35 | 386.47 ± 27.32 | 347.74 ± 14.95 | 125.7 |
| CrossE_L + PR_L + Triplet_L | ALL | 254.07 ± 15.62 | 278.04 ± 4.88 | 243.84 ± 26.83 | 182.3 |

Rankme Continued (↑)

| Loss Type | | | Model | CORA | Citeseer | Bitcoin Fraud Transaction | Average Rank |
|---|---|---|---|---|---|---|---|
| CrossE_L | + PR_L | + Triplet_L | GAT | 410.57 ± 3.97 | 432.84 ± 1.98 | 391.33 ± 10.39 | 60.7 |
| CrossE_L | + PR_L | + Triplet_L | GCN | 403.39 ± 3.36 | 417.79 ± 3.33 | 414.06 ± 7.48 | 72.7 |
| CrossE_L | + PR_L | + Triplet_L | GIN | 345.29 ± 9.71 | 366.41 ± 3.07 | 308.86 ± 10.13 | 150.7 |
| CrossE_L | + PR_L | + Triplet_L | MPNN | 336.25 ± 6.21 | 375.09 ± 5.71 | 407.99 ± 4.54 | 121.0 |
| CrossE_L | + PR_L | + Triplet_L | PAGNN | 275.79 ± 16.54 | 305.84 ± 11.66 | 167.55 ± 3.36 | 185.0 |
| CrossE_L | + PR_L | + Triplet_L | SAGE | 379.44 ± 3.07 | 396.03 ± 4.56 | 376.77 ± 6.15 | 103.7 |
| CrossE_L + Triplet_L | | | ALL | 323.53 ± 12.87 | 338.27 ± 9.14 | 331.71 ± 19.96 | 159.3 |
| CrossE_L + Triplet_L | | | GAT | 442.33 ± 1.10 | 452.14 ± 1.13 | 437.08 ± 2.92 | 15.3 |
| CrossE_L + Triplet_L | | | GCN | 420.39 ± 2.51 | 430.88 ± 1.12 | 419.61 ± 2.46 | 42.3 |
| CrossE_L + Triplet_L | | | GIN | 394.70 ± 2.70 | 399.88 ± 4.01 | 344.08 ± 5.79 | 108.0 |
| CrossE_L + Triplet_L | | | MPNN | 416.57 ± 1.34 | 429.70 ± 1.54 | 422.65 ± 1.39 | 43.3 |
| CrossE_L + Triplet_L | | | PAGNN | 374.74 ± 6.43 | 375.91 ± 3.60 | 133.61 ± 11.47 | 150.3 |
| CrossE_L + Triplet_L | | | SAGE | 403.85 ± 1.33 | 416.59 ± 2.53 | 392.92 ± 1.95 | 81.7 |
| PMI_L | | | ALL | 368.50 ± 2.25 | 396.75 ± 4.09 | 398.50 ± 2.73 | 102.0 |
| PMI_L | | | GAT | 451.40 ± 1.45 | 458.38 ± 0.87 | 436.68 ± 1.67 | 3.7 |
| PMI_L | | | GCN | 421.31 ± 3.24 | 429.05 ± 2.15 | 417.79 ± 2.43 | 51.0 |

Rankme Continued (↑)

| Loss Type | Model | CORA | | Citeseer | | Bitcoin Fraud Transaction | | Average Rank |
|---|---|---|---|---|---|---|---|---|
| PMI_L | GIN | 394.05 | ± 3.55 | 389.86 | ± 5.12 | 356.16 | ± 4.45 | 110.7 |
| PMI_L | MPNN | 406.93 | ± 1.83 | 427.68 | ± 0.64 | 418.56 | ± 2.07 | 62.0 |
| PMI_L | PAGNN | 357.42 | ± 5.60 | 352.22 | ± 3.10 | 184.47 | ± 6.41 | 157.3 |
| PMI_L | SAGE | 441.90 | ± 2.94 | 451.47 | ± 2.22 | 440.49 | ± 2.21 | 14.3 |
| PMI_L + PR_L | ALL | 342.45 | ± 10.31 | 359.06 | ± 21.87 | 348.55 | ± 4.69 | 145.0 |
| PMI_L + PR_L | GAT | 443.62 | ± 3.22 | 442.44 | ± 12.10 | 402.32 | ± 3.31 | 40.7 |
| PMI_L + PR_L | GCN | 420.75 | ± 3.04 | 430.30 | ± 3.41 | 418.57 | ± 3.96 | 47.0 |
| PMI_L + PR_L | GIN | 390.93 | ± 4.20 | 375.36 | ± 24.70 | 268.59 | ± 12.86 | 131.0 |
| PMI_L + PR_L | MPNN | 406.01 | ± 1.74 | 411.12 | ± 4.09 | 373.14 | ± 25.37 | 88.3 |
| PMI_L + PR_L | PAGNN | 355.82 | ± 6.41 | 355.20 | ± 4.93 | 151.17 | ± 9.10 | 166.0 |
| PMI_L + PR_L | SAGE | 446.20 | ± 3.36 | 453.11 | ± 3.37 | 336.59 | ± 13.09 | 56.3 |
| PMI_L + PR_L + Triplet_L | ALL | 322.76 | ± 10.86 | 341.46 | ± 5.62 | 348.00 | ± 13.60 | 153.7 |
| PMI_L + PR_L + Triplet_L | GAT | 444.67 | ± 0.92 | 450.81 | ± 3.85 | 430.38 | ± 3.70 | 17.0 |
| PMI_L + PR_L + Triplet_L | GCN | 421.06 | ± 2.23 | 429.67 | ± 2.88 | 419.37 | ± 2.73 | 44.3 |
| PMI_L + PR_L + Triplet_L | GIN | 393.34 | ± 3.30 | 394.36 | ± 3.19 | 373.73 | ± 1.80 | 102.7 |
| PMI_L + PR_L + Triplet_L | MPNN | 407.89 | ± 2.20 | 418.35 | ± 1.96 | 415.49 | ± 8.26 | 67.3 |

Rankme Continued ($\uparrow$)

| Loss Type | Model | CORA | Citeseer | Bitcoin Fraud Transaction | Average Rank |
|---|---|---|---|---|---|
| PMI_L + PR_L + Triplet_L | PAGNN | 357.33 ± 4.22 | 364.07 ± 3.94 | 174.41 ± 4.15 | 155.3 |
| PMI_L + PR_L + Triplet_L | SAGE | 427.24 ± 8.69 | 433.19 ± 4.30 | 381.36 ± 10.53 | 52.3 |
| PMI_L + Triplet_L | ALL | 354.67 ± 4.58 | 389.04 ± 2.48 | 369.87 ± 3.91 | 125.3 |
| PMI_L + Triplet_L | GAT | 449.19 ± 0.62 | 457.69 ± 0.43 | 440.38 ± 0.66 | 4.7 |
| PMI_L + Triplet_L | GCN | 420.35 ± 1.59 | 428.99 ± 2.12 | 418.55 ± 2.88 | 51.7 |
| PMI_L + Triplet_L | GIN | 395.58 ± 3.72 | 400.91 ± 3.92 | 374.43 ± 3.93 | 95.0 |
| PMI_L + Triplet_L | MPNN | 409.01 ± 1.72 | 429.61 ± 1.64 | 426.41 ± 2.34 | 46.0 |
| PMI_L + Triplet_L | PAGNN | 357.75 ± 5.58 | 356.12 ± 1.94 | 172.49 ± 3.99 | 158.7 |
| PMI_L + Triplet_L | SAGE | 431.04 ± 5.25 | 432.58 ± 1.54 | 414.11 ± 2.43 | 40.0 |
| PR_L | ALL | 149.72 ± 4.92 | 172.84 ± 8.33 | 144.63 ± 3.77 | 201.7 |
| PR_L | GAT | 277.38 ± 11.25 | 332.67 ± 24.59 | 194.71 ± 16.43 | 174.7 |
| PR_L | GCN | 354.38 ± 8.83 | 389.26 ± 2.54 | 223.38 ± 30.30 | 145.0 |
| PR_L | GIN | 229.47 ± 28.97 | 274.73 ± 14.08 | 223.20 ± 7.54 | 184.7 |
| PR_L | MPNN | 289.14 ± 23.09 | 309.39 ± 8.17 | 394.90 ± 8.56 | 147.3 |
| PR_L | PAGNN | 244.77 ± 30.39 | 264.05 ± 7.55 | 168.17 ± 5.70 | 191.7 |
| PR_L | SAGE | 367.48 ± 23.79 | 370.24 ± 18.33 | 346.32 ± 2.77 | 132.7 |

Rankme Continued (↑)

| Loss Type | Model | CORA | Citeseer | Bitcoin Fraud Transaction | Average Rank |
|-----------|-------|------|----------|---------------------------|--------------|
| PR_L + Triplet_L | ALL | 181.79 ± 11.00 | 213.73 ± 11.62 | 144.81 ± 21.91 | 199.7 |
| PR_L + Triplet_L | GAT | 388.20 ± 25.89 | 390.05 ± 10.26 | 323.78 ± 21.12 | 124.0 |
| PR_L + Triplet_L | GCN | 393.36 ± 5.60 | 405.05 ± 3.31 | 406.81 ± 3.49 | 86.3 |
| PR_L + Triplet_L | GIN | 280.25 ± 17.57 | 337.81 ± 10.77 | 267.20 ± 10.58 | 170.3 |
| PR_L + Triplet_L | MPNN | 307.55 ± 20.41 | 326.91 ± 8.36 | 404.07 ± 8.16 | 141.7 |
| PR_L + Triplet_L | PAGNN | 261.09 ± 10.26 | 256.97 ± 8.11 | 169.20 ± 7.04 | 190.3 |
| PR_L + Triplet_L | SAGE | 376.72 ± 10.51 | 396.86 ± 8.63 | 357.08 ± 8.16 | 110.7 |
| Triplet_L | ALL | 331.63 ± 6.48 | 358.83 ± 12.13 | 347.20 ± 5.98 | 148.0 |
| Triplet_L | GAT | 444.68 ± 0.94 | 453.75 ± 0.48 | 442.15 ± 0.94 | 9.7 |
| Triplet_L | GCN | 422.85 ± 1.06 | 431.43 ± 1.24 | 419.78 ± 2.46 | 37.0 |
| Triplet_L | GIN | 404.57 ± 3.20 | 409.32 ± 2.80 | 350.25 ± 3.58 | 97.0 |
| Triplet_L | MPNN | 417.04 ± 0.84 | 432.52 ± 0.78 | 423.63 ± 0.96 | 38.0 |
| Triplet_L | PAGNN | 379.62 ± 7.62 | 383.32 ± 2.48 | 155.35 ± 5.41 | 144.0 |
| Triplet_L | SAGE | 410.29 ± 1.01 | 423.35 ± 1.96 | 405.08 ± 1.81 | 68.7 |

## 1.2 Inductive Results

### 1.2.1 Assessment of Learned Representations through Node Classification.

Table 22. Node Cls Accuracy Performance (↑): This table presents models (Loss function and GNN) ranked by their average performance in terms of node cls accuracy. Top-ranked results are highlighted in red, second-ranked in blue, and third-ranked in green.

| Loss Type | Model | Cora ↓ Citeseer | | Cora ↓ Bitcoin | | Citeseer ↓ Cora | | Citeseer ↓ Bitcoin | | Average Rank |
|---|---|---|---|---|---|---|---|---|---|---|
| Contr_l | ALL | 44.47 | ± 2.71 | 71.18 | ± 0.04 | 62.07 | ± 4.38 | 71.16 | ± 0.09 | 178.625 |
| Contr_l | GAT | 62.16 | ± 0.72 | 75.32 | ± 0.84 | 76.83 | ± 1.68 | 74.96 | ± 0.67 | 29.125 |
| Contr_l | GCN | 57.72 | ± 0.50 | 73.04 | ± 0.70 | 72.07 | ± 1.34 | 73.58 | ± 0.70 | 67.125 |
| Contr_l | GIN | 52.88 | ± 3.47 | 71.48 | ± 0.47 | 71.85 | ± 3.11 | 71.62 | ± 0.60 | 100.375 |
| Contr_l | MPNN | 59.46 | ± 2.01 | 74.88 | ± 0.66 | 73.84 | ± 1.44 | 74.34 | ± 0.68 | 50.75 |
| Contr_l | PAGNN | 53.84 | ± 2.04 | 71.20 | ± 0.00 | 68.71 | ± 4.27 | 71.20 | ± 0.00 | 129.125 |
| Contr_l | SAGE | 47.57 | ± 2.64 | 75.06 | ± 1.68 | 63.03 | ± 4.69 | 72.68 | ± 1.11 | 113.125 |
| Contr_l + CrossE_L | ALL | 48.23 | ± 2.60 | 71.12 | ± 0.38 | 56.90 | ± 6.26 | 71.18 | ± 0.04 | 184.0 |
| Contr_l + CrossE_L | GAT | 61.53 | ± 1.36 | 75.52 | ± 0.49 | 75.13 | ± 2.42 | 74.86 | ± 0.78 | 33.75 |
| Contr_l + CrossE_L | GCN | 56.64 | ± 1.63 | 73.42 | ± 0.61 | 72.43 | ± 2.05 | 72.94 | ± 0.92 | 71.875 |
| Contr_l + CrossE_L | GIN | 52.88 | ± 2.34 | 71.10 | ± 0.33 | 69.30 | ± 1.37 | 71.44 | ± 0.47 | 129.5 |
| Contr_l + CrossE_L | MPNN | 60.27 | ± 1.48 | 74.64 | ± 0.59 | 74.28 | ± 1.68 | 73.76 | ± 0.27 | 51.0 |
| Contr_l + CrossE_L | PAGNN | 53.69 | ± 2.45 | 71.20 | ± 0.00 | 69.11 | ± 2.37 | 71.22 | ± 0.04 | 122.125 |
| Contr_l + CrossE_L | SAGE | 48.41 | ± 3.58 | 72.54 | ± 0.80 | 61.11 | ± 1.31 | 71.34 | ± 0.23 | 135.125 |
| Contr_l + CrossE_L + PMI_L | ALL | 52.46 | ± 2.88 | 71.18 | ± 0.04 | 70.19 | ± 4.42 | 71.52 | ± 0.44 | 125.0 |

Table 22. Results for Node Cls Accuracy (↑) (continued)

| Loss Type | Model | Cora ↓ Citeseer | | Cora ↓ Bitcoin | | Citeseer ↓ Cora | | Citeseer ↓ Bitcoin | | Average Rank |
|---|---|---|---|---|---|---|---|---|---|---|
| Contr_l + CrossE_L + PMI_L | GAT | 62.61 | ± 2.23 | 75.86 | ± 0.58 | 76.13 | ± 1.64 | 75.28 | ± 0.70 | 22.125 |
| Contr_l + CrossE_L + PMI_L | GCN | 56.64 | ± 2.36 | 72.50 | ± 0.60 | 68.52 | ± 2.02 | 73.44 | ± 1.07 | 82.625 |
| Contr_l + CrossE_L + PMI_L | GIN | 50.90 | ± 1.20 | 71.16 | ± 0.23 | 64.98 | ± 3.30 | 71.18 | ± 0.30 | 161.0 |
| Contr_l + CrossE_L + PMI_L | MPNN | 62.76 | ± 0.73 | 74.98 | ± 0.73 | 76.24 | ± 1.95 | 75.50 | ± 0.87 | 27.875 |
| Contr_l + CrossE_L + PMI_L | PAGNN | 50.93 | ± 2.42 | 71.20 | ± 0.00 | 60.59 | ± 2.66 | 71.20 | ± 0.00 | 155.625 |
| Contr_l + CrossE_L + PMI_L | SAGE | 47.24 | ± 2.12 | 77.98 | ± 1.87 | 57.09 | ± 3.43 | 76.76 | ± 1.83 | 92.0 |
| Contr_l + CrossE_L + PMI_L + PR_L | ALL | 43.57 | ± 1.29 | 71.40 | ± 0.00 | 65.79 | ± 1.97 | 71.14 | ± 0.05 | 154.625 |
| Contr_l + CrossE_L + PMI_L + PR_L | GAT | 62.67 | ± 1.36 | 75.26 | ± 0.92 | 76.72 | ± 2.17 | 75.12 | ± 0.58 | 27.75 |
| Contr_l + CrossE_L + PMI_L + PR_L | GCN | 57.93 | ± 1.33 | 72.96 | ± 0.94 | 70.07 | ± 1.34 | 73.28 | ± 0.67 | 74.25 |
| Contr_l + CrossE_L + PMI_L + PR_L | GIN | 51.20 | ± 2.86 | 71.22 | ± 0.15 | 61.55 | ± 4.20 | 71.18 | ± 0.04 | 152.125 |
| Contr_l + CrossE_L + PMI_L + PR_L | MPNN | 61.62 | ± 1.12 | 75.74 | ± 0.85 | 76.09 | ± 1.84 | 72.88 | ± 1.09 | 42.375 |
| Contr_l + CrossE_L + PMI_L + PR_L | PAGNN | 52.64 | ± 2.32 | 71.20 | ± 0.00 | 62.55 | ± 2.81 | 71.20 | ± 0.00 | 144.5 |
| Contr_l + CrossE_L + PMI_L + PR_L | SAGE | 48.53 | ± 2.91 | 78.66 | ± 1.43 | 57.05 | ± 0.79 | 77.04 | ± 1.36 | 87.875 |
| Contr_l + CrossE_L + PMI_L + PR_L + Triplet_L | ALL | 51.05 | ± 2.02 | 71.28 | ± 0.16 | 65.87 | ± 3.23 | 71.32 | ± 0.33 | 125.625 |
| Contr_l + CrossE_L + PMI_L + PR_L + Triplet_L | GAT | 63.00 | ± 1.56 | 75.62 | ± 0.36 | 76.97 | ± 1.37 | 75.46 | ± 0.44 | 19.375 |

Table 22. Results for Node Cls Accuracy (↑) (continued)

| Loss Type | Model | Cora ↓ Citeseer | ± | Cora ↓ Bitcoin | ± | Citeseer ↓ Cora | ± | Citeseer ↓ Bitcoin | ± | Average Rank |
|---|---|---|---|---|---|---|---|---|---|---|
| Contr_l + CrossE_L + PMI_L + PR_L + Triplet_L | GCN | 56.79 | ± 3.09 | 73.44 | ± 0.50 | 69.71 | ± 2.12 | 72.72 | ± 0.40 | 79.0 |
| Contr_l + CrossE_L + PMI_L + PR_L + Triplet_L | GIN | 50.84 | ± 1.71 | 71.24 | ± 0.32 | 67.57 | ± 2.18 | 71.12 | ± 0.29 | 144.125 |
| Contr_l + CrossE_L + PMI_L + PR_L + Triplet_L | MPNN | 60.39 | ± 1.13 | 76.22 | ± 0.99 | 77.31 | ± 1.71 | 73.48 | ± 1.41 | 32.75 |
| Contr_l + CrossE_L + PMI_L + PR_L + Triplet_L | PAGNN | 52.58 | ± 2.50 | 71.20 | ± 0.00 | 64.76 | ± 1.88 | 71.20 | ± 0.00 | 141.25 |
| Contr_l + CrossE_L + PMI_L + PR_L + Triplet_L | SAGE | 49.76 | ± 1.62 | 77.20 | ± 0.23 | 61.33 | ± 1.78 | 78.60 | ± 1.74 | 77.25 |
| Contr_l + CrossE_L + PMI_L + Triplet_L | ALL | 54.90 | ± 2.25 | 72.12 | ± 1.00 | 69.78 | ± 1.22 | 73.38 | ± 0.98 | 84.75 |
| Contr_l + CrossE_L + PMI_L + Triplet_L | GAT | 64.02 | ± 1.41 | 75.58 | ± 0.37 | 77.68 | ± 1.56 | 75.52 | ± 0.54 | 14.25 |
| Contr_l + CrossE_L + PMI_L + Triplet_L | GCN | 57.30 | ± 2.10 | 73.48 | ± 0.19 | 68.38 | ± 1.81 | 73.12 | ± 1.05 | 78.625 |
| Contr_l + CrossE_L + PMI_L + Triplet_L | GIN | 53.21 | ± 1.76 | 71.22 | ± 0.33 | 69.04 | ± 3.54 | 71.12 | ± 0.38 | 132.0 |
| Contr_l + CrossE_L + PMI_L + Triplet_L | MPNN | 61.38 | ± 1.50 | 75.52 | ± 0.82 | 75.94 | ± 1.89 | 74.64 | ± 0.50 | 34.5 |
| Contr_l + CrossE_L + PMI_L + Triplet_L | PAGNN | 52.64 | ± 2.45 | 71.20 | ± 0.00 | 63.14 | ± 1.45 | 71.20 | ± 0.00 | 143.0 |
| Contr_l + CrossE_L + PMI_L + Triplet_L | SAGE | 47.54 | ± 4.18 | 77.08 | ± 0.67 | 60.04 | ± 1.78 | 77.62 | ± 0.68 | 86.0 |
| Contr_l + CrossE_L + PR_L | ALL | 32.91 | ± 2.78 | 71.20 | ± 0.00 | 49.63 | ± 4.52 | 71.20 | ± 0.00 | 182.125 |
| Contr_l + CrossE_L + PR_L | GAT | 57.78 | ± 2.31 | 75.20 | ± 0.60 | 58.86 | ± 7.56 | 74.24 | ± 0.76 | 81.625 |

Table 22.  Results for Node Cls Accuracy (↑) (continued)

| Loss Type | Model | Cora ↓ Citeseer | | Cora ↓ Bitcoin | | Citeseer ↓ Cora | | Citeseer ↓ Bitcoin | | Average Rank |
|---|---|---|---|---|---|---|---|---|---|---|
| Contr_l + CrossE_L + PR_L | GCN | 54.38 | ± 2.36 | 72.80 | ± 0.74 | 68.23 | ± 2.41 | 72.78 | ± 0.67 | 94.625 |
| Contr_l + CrossE_L + PR_L | GIN | 44.56 | ± 0.76 | 71.28 | ± 0.18 | 62.51 | ± 2.98 | 71.20 | ± 0.00 | 154.5 |
| Contr_l + CrossE_L + PR_L | MPNN | 54.08 | ± 2.37 | 71.48 | ± 0.36 | 67.90 | ± 4.62 | 71.62 | ± 0.45 | 106.0 |
| Contr_l + CrossE_L + PR_L | PAGNN | 46.64 | ± 3.00 | 71.20 | ± 0.00 | 57.71 | ± 3.94 | 71.20 | ± 0.00 | 171.0 |
| Contr_l + CrossE_L + PR_L | SAGE | 44.26 | ± 4.21 | 73.96 | ± 2.12 | 54.21 | ± 3.58 | 75.50 | ± 1.37 | 116.875 |
| Contr_l + CrossE_L + PR_L + Triplet_L | ALL | 42.07 | ± 1.68 | 71.20 | ± 0.00 | 55.65 | ± 6.36 | 71.22 | ± 0.04 | 171.875 |
| Contr_l + CrossE_L + PR_L + Triplet_L | GAT | 60.18 | ± 2.85 | 75.40 | ± 0.35 | 74.13 | ± 1.55 | 75.06 | ± 0.64 | 38.875 |
| Contr_l + CrossE_L + PR_L + Triplet_L | GCN | 54.89 | ± 2.62 | 73.14 | ± 0.97 | 67.30 | ± 1.99 | 73.18 | ± 0.62 | 89.75 |
| Contr_l + CrossE_L + PR_L + Triplet_L | GIN | 51.74 | ± 2.51 | 71.20 | ± 0.16 | 68.38 | ± 1.20 | 71.22 | ± 0.16 | 132.625 |
| Contr_l + CrossE_L + PR_L + Triplet_L | MPNN | 58.47 | ± 1.25 | 73.62 | ± 0.61 | 72.88 | ± 2.08 | 72.20 | ± 0.82 | 68.625 |
| Contr_l + CrossE_L + PR_L + Triplet_L | PAGNN | 49.46 | ± 2.09 | 71.20 | ± 0.00 | 62.99 | ± 1.97 | 71.20 | ± 0.00 | 153.625 |
| Contr_l + CrossE_L + PR_L + Triplet_L | SAGE | 44.90 | ± 2.31 | 72.50 | ± 0.57 | 57.05 | ± 2.75 | 74.04 | ± 0.71 | 132.25 |
| Contr_l + CrossE_L + Triplet_L | ALL | 48.92 | ± 4.05 | 70.99 | ± 0.31 | 62.36 | ± 2.88 | 71.10 | ± 0.23 | 175.75 |
| Contr_l + CrossE_L + Triplet_L | GAT | 61.48 | ± 1.69 | 75.63 | ± 0.66 | 75.81 | ± 1.54 | 75.06 | ± 0.38 | 29.875 |
| Contr_l + CrossE_L + Triplet_L | GCN | 58.00 | ± 1.97 | 72.77 | ± 0.60 | 71.94 | ± 1.78 | 73.75 | ± 0.62 | 68.25 |
| Contr_l + CrossE_L + Triplet_L | GIN | 53.95 | ± 2.57 | 71.56 | ± 0.49 | 71.59 | ± 1.99 | 71.13 | ± 0.54 | 115.75 |

Table 22. Results for Node Cls Accuracy (↑) (continued)

| Loss Type | Model | Cora ↓ Citeseer | | Cora ↓ Bitcoin | | Citeseer ↓ Cora | | Citeseer ↓ Bitcoin | | Average Rank |
|---|---|---|---|---|---|---|---|---|---|---|
| Contr_l + CrossE_L + Triplet_L | MPNN | 59.84 1.84 | ± | 74.27 0.96 | ± | 74.45 1.66 | ± | 74.76 0.63 | ± | 47.625 |
| Contr_l + CrossE_L + Triplet_L | PAGNN | 54.31 1.71 | ± | 71.20 0.00 | ± | 70.83 1.57 | ± | 71.21 0.03 | ± | 115.25 |
| Contr_l + CrossE_L + Triplet_L | SAGE | 50.86 2.71 | ± | 73.53 1.15 | ± | 64.02 1.62 | ± | 74.51 1.41 | ± | 97.375 |
| Contr_l + PMI_L | ALL | 52.04 2.20 | ± | 71.22 0.04 | ± | 67.97 5.04 | ± | 71.76 0.46 | ± | 119.125 |
| Contr_l + PMI_L | GAT | 63.15 0.81 | ± | 75.48 0.48 | ± | 76.05 2.32 | ± | 74.78 0.56 | ± | 29.0 |
| Contr_l + PMI_L | GCN | 57.33 2.69 | ± | 73.22 0.70 | ± | 70.81 1.88 | ± | 73.36 0.68 | ± | 70.625 |
| Contr_l + PMI_L | GIN | 51.89 3.01 | ± | 71.42 0.24 | ± | 65.17 3.19 | ± | 70.88 0.40 | ± | 142.625 |
| Contr_l + PMI_L | MPNN | 61.08 0.82 | ± | 75.50 1.13 | ± | 76.42 1.34 | ± | 74.62 0.84 | ± | 34.25 |
| Contr_l + PMI_L | PAGNN | 52.46 1.56 | ± | 71.20 0.00 | ± | 56.97 3.06 | ± | 71.20 0.00 | ± | 156.25 |
| Contr_l + PMI_L | SAGE | 46.91 1.28 | ± | **78.80 1.75** | ± | 58.45 4.18 | ± | 76.98 2.15 | ± | 89.125 |
| Contr_l + PMI_L + PR_L | ALL | 47.00 1.71 | ± | 71.40 0.00 | ± | 63.10 1.88 | ± | 71.16 0.17 | ± | 155.125 |
| Contr_l + PMI_L + PR_L | GAT | 63.87 1.13 | ± | 75.70 0.61 | ± | 75.02 1.72 | ± | 74.38 0.19 | ± | 30.625 |
| Contr_l + PMI_L + PR_L | GCN | 56.07 1.59 | ± | 73.54 0.90 | ± | 70.59 2.69 | ± | 73.26 1.03 | ± | 72.375 |
| Contr_l + PMI_L + PR_L | GIN | 50.75 0.79 | ± | 71.24 0.05 | ± | 61.11 5.93 | ± | 71.22 0.08 | ± | 141.75 |
| Contr_l + PMI_L + PR_L | MPNN | 60.81 1.53 | ± | 75.44 0.92 | ± | 75.50 1.55 | ± | 72.40 1.10 | ± | 53.375 |
| Contr_l + PMI_L + PR_L | PAGNN | 50.75 1.99 | ± | 71.20 0.00 | ± | 61.26 4.17 | ± | 71.20 0.00 | ± | 155.75 |

Table 22. Results for Node Cls Accuracy (↑) (continued)

| Loss Type | Model | Cora ↓ Citeseer | | Cora ↓ Bitcoin | | Citeseer ↓ Cora | | Citeseer ↓ Bitcoin | | Average Rank |
|---|---|---|---|---|---|---|---|---|---|---|
| Contr_l + PMI_L + PR_L | SAGE | 47.69 1.66 | ± | 78.00 2.21 | ± | 56.72 4.36 | ± | 74.56 1.97 | ± | 101.25 |
| Contr_l + PMI_L + PR_L + Triplet_L | ALL | 50.78 1.77 | ± | 71.18 0.08 | ± | 61.70 2.56 | ± | 71.24 0.13 | ± | 152.75 |
| Contr_l + PMI_L + PR_L + Triplet_L | GAT | 61.47 0.62 | ± | 75.44 0.75 | ± | 74.46 1.85 | ± | 74.62 0.98 | ± | 38.5 |
| Contr_l + PMI_L + PR_L + Triplet_L | GCN | 55.71 0.44 | ± | 72.86 0.70 | ± | 70.30 1.84 | ± | 73.46 0.60 | ± | 77.5 |
| Contr_l + PMI_L + PR_L + Triplet_L | GIN | 52.31 1.91 | ± | 71.16 0.26 | ± | 67.49 2.08 | ± | 71.06 0.27 | ± | 156.0 |
| Contr_l + PMI_L + PR_L + Triplet_L | MPNN | 62.64 2.35 | ± | 74.96 0.82 | ± | 76.53 3.53 | ± | 71.88 0.92 | ± | 51.25 |
| Contr_l + PMI_L + PR_L + Triplet_L | PAGNN | 52.43 1.67 | ± | 71.20 0.00 | ± | 68.97 2.05 | ± | 71.20 0.00 | ± | 133.875 |
| Contr_l + PMI_L + PR_L + Triplet_L | SAGE | 49.58 3.63 | ± | 77.62 1.67 | ± | 60.19 1.90 | ± | 73.42 0.69 | ± | 96.125 |
| Contr_l + PR_L | ALL | 34.74 1.65 | ± | 71.20 0.00 | ± | 47.45 2.14 | ± | 71.20 0.00 | ± | 182.125 |
| Contr_l + PR_L | GAT | 55.73 3.18 | ± | 74.72 0.73 | ± | 59.45 5.80 | ± | 74.76 0.24 | ± | 84.125 |
| Contr_l + PR_L | GCN | 52.88 2.20 | ± | 73.20 0.33 | ± | 65.31 4.15 | ± | 72.94 0.61 | ± | 101.0 |
| Contr_l + PR_L | GIN | 45.76 4.10 | ± | 71.20 0.00 | ± | 61.37 6.04 | ± | 71.20 0.17 | ± | 164.125 |
| Contr_l + PR_L | MPNN | 50.33 2.59 | ± | 71.20 0.00 | ± | 65.24 4.75 | ± | 72.16 1.00 | ± | 134.75 |
| Contr_l + PR_L | PAGNN | 44.81 3.66 | ± | 71.20 0.00 | ± | 56.94 2.88 | ± | 71.20 0.00 | ± | 174.125 |
| Contr_l + PR_L | SAGE | 45.50 2.23 | ± | 73.00 0.38 | ± | 54.80 3.74 | ± | 76.02 2.19 | ± | 119.75 |
| Contr_l + PR_L + Triplet_L | ALL | 42.58 1.99 | ± | 71.16 0.09 | ± | 56.02 3.86 | ± | 71.20 0.07 | ± | 185.25 |

Table 22. Results for Node Cls Accuracy (↑) (continued)

| Loss Type | Model | Cora ↓ Citeseer | | Cora ↓ Bitcoin | | Citeseer ↓ Cora | | Citeseer ↓ Bitcoin | | Average Rank |
|---|---|---|---|---|---|---|---|---|---|---|
| Contr_l + PR_L + Triplet_L | GAT | 58.47 ± 1.58 | | 75.30 ± 0.91 | | 72.99 ± 3.41 | | 75.22 ± 0.95 | | 41.5 |
| Contr_l + PR_L + Triplet_L | GCN | 54.95 ± 2.51 | | 73.76 ± 0.50 | | 68.78 ± 1.90 | | 72.62 ± 1.00 | | 85.625 |
| Contr_l + PR_L + Triplet_L | GIN | 51.77 ± 2.64 | | 71.48 ± 0.65 | | 67.45 ± 4.53 | | 71.20 ± 0.41 | | 128.125 |
| Contr_l + PR_L + Triplet_L | MPNN | 58.68 ± 1.48 | | 73.58 ± 0.72 | | 72.91 ± 2.24 | | 72.26 ± 1.34 | | 68.0 |
| Contr_l + PR_L + Triplet_L | PAGNN | 51.62 ± 2.15 | | 71.20 ± 0.00 | | 65.13 ± 2.98 | | 71.20 ± 0.00 | | 145.375 |
| Contr_l + PR_L + Triplet_L | SAGE | 47.45 ± 3.93 | | 74.42 ± 1.31 | | 58.26 ± 1.40 | | 74.82 ± 1.89 | | 111.75 |
| Contr_l + Triplet_L | ALL | 48.26 ± 1.74 | | 71.06 ± 0.05 | | 62.73 ± 4.30 | | 71.10 ± 0.29 | | 176.25 |
| Contr_l + Triplet_L | GAT | 63.75 ± 1.46 | | 75.14 ± 0.50 | | 77.38 ± 1.06 | | 75.66 ± 0.61 | | 20.375 |
| Contr_l + Triplet_L | GCN | 55.47 ± 1.12 | | 73.10 ± 0.72 | | 72.33 ± 0.79 | | 72.94 ± 0.77 | | 78.375 |
| Contr_l + Triplet_L | GIN | 53.87 ± 1.24 | | 71.38 ± 0.49 | | 69.85 ± 2.30 | | 71.22 ± 0.54 | | 107.625 |
| Contr_l + Triplet_L | MPNN | 60.15 ± 1.30 | | 73.84 ± 0.74 | | 75.79 ± 1.39 | | 74.14 ± 0.40 | | 50.5 |
| Contr_l + Triplet_L | PAGNN | 55.62 ± 0.89 | | 71.20 ± 0.00 | | 69.52 ± 2.28 | | 71.20 ± 0.00 | | 121.875 |
| Contr_l + Triplet_L | SAGE | 49.31 ± 1.74 | | 72.14 ± 0.23 | | 64.02 ± 2.10 | | 72.60 ± 0.35 | | 122.125 |
| CrossE_L | ALL | 22.73 ± 1.73 | | 71.20 ± 0.00 | | 35.24 ± 0.00 | | 71.20 ± 0.00 | | 186.125 |
| CrossE_L | GAT | 21.47 ± 0.00 | | 71.20 ± 0.00 | | 35.24 ± 0.00 | | 71.20 ± 0.00 | | 186.875 |
| CrossE_L | GCN | 21.47 ± 0.00 | | 71.20 ± 0.00 | | 35.24 ± 0.00 | | 71.20 ± 0.00 | | 186.875 |

Table 22.  Results for Node Cls Accuracy (↑) (continued)

| Loss Type | Model | Cora ↓ Citeseer | | Cora ↓ Bitcoin | | Citeseer ↓ Cora | | Citeseer ↓ Bitcoin | | Average Rank |
|---|---|---|---|---|---|---|---|---|---|---|
| CrossE_L | GIN | 21.47 | ± 0.00 | 71.20 | ± 0.00 | 35.24 | ± 0.00 | 71.20 | ± 0.00 | 186.875 |
| CrossE_L | MPNN | 21.47 | ± 0.00 | 71.20 | ± 0.00 | 35.24 | ± 0.00 | 71.20 | ± 0.00 | 186.875 |
| CrossE_L | PAGNN | 25.44 | ± 3.65 | 71.20 | ± 0.00 | 35.24 | ± 0.00 | 71.20 | ± 0.00 | 185.875 |
| CrossE_L | SAGE | 21.47 | ± 0.00 | 71.20 | ± 0.00 | 35.24 | ± 0.00 | 71.20 | ± 0.00 | 186.875 |
| CrossE_L + PMI_L | ALL | 58.29 | ± 1.60 | 71.40 | ± 0.17 | 69.23 | ± 2.28 | 71.46 | ± 0.18 | 94.375 |
| CrossE_L + PMI_L | GAT | 62.19 | ± 3.60 | 74.64 | ± 0.53 | 75.83 | ± 1.40 | 75.02 | ± 0.94 | 36.625 |
| CrossE_L + PMI_L | GCN | 56.85 | ± 1.18 | 73.46 | ± 0.99 | 72.29 | ± 2.69 | 72.64 | ± 0.69 | 73.125 |
| CrossE_L + PMI_L | GIN | 52.97 | ± 2.77 | 71.08 | ± 0.23 | 65.68 | ± 2.55 | 71.24 | ± 0.17 | 138.625 |
| CrossE_L + PMI_L | MPNN | 61.20 | ± 0.87 | 75.84 | ± 1.01 | 77.53 | ± 2.04 | 75.02 | ± 0.86 | 24.0 |
| CrossE_L + PMI_L | PAGNN | 53.15 | ± 2.51 | 71.20 | ± 0.00 | 58.89 | ± 2.99 | 71.20 | ± 0.00 | 150.125 |
| CrossE_L + PMI_L | SAGE | 45.71 | ± 2.70 | 77.36 | ± 2.39 | 59.04 | ± 1.62 | 78.16 | ± 0.78 | 89.375 |
| CrossE_L + PMI_L + PR_L | ALL | 46.97 | ± 1.24 | 71.42 | ± 0.04 | 65.72 | ± 3.27 | 71.26 | ± 0.09 | 133.75 |
| CrossE_L + PMI_L + PR_L | GAT | 64.14 | ± 1.98 | 75.26 | ± 0.38 | 73.58 | ± 5.52 | 74.26 | ± 0.29 | 37.625 |
| CrossE_L + PMI_L + PR_L | GCN | 56.70 | ± 3.20 | 72.94 | ± 0.69 | 71.51 | ± 1.28 | 73.60 | ± 0.73 | 71.25 |
| CrossE_L + PMI_L + PR_L | GIN | 51.14 | ± 2.13 | 71.16 | ± 0.13 | 59.04 | ± 8.71 | 71.10 | ± 0.10 | 174.75 |
| CrossE_L + PMI_L + PR_L | MPNN | 62.04 | ± 1.40 | 76.18 | ± 1.01 | 76.53 | ± 1.22 | 76.04 | ± 0.80 | 17.75 |

Table 22. Results for Node Cls Accuracy (↑) (continued)

| Loss Type | Model | Cora ↓ Citeseer | | Cora ↓ Bitcoin | | Citeseer ↓ Cora | | Citeseer ↓ Bitcoin | | Average Rank |
|---|---|---|---|---|---|---|---|---|---|---|
| CrossE_L + PMI_L + PR_L | PAGNN | 53.45 2.47 | ± | 71.20 0.00 | ± | 62.21 2.56 | ± | 71.20 0.00 | ± | 142.875 |
| CrossE_L + PMI_L + PR_L | SAGE | 47.51 1.94 | ± | 78.34 0.56 | ± | 58.08 1.15 | ± | 77.50 1.52 | ± | 88.5 |
| CrossE_L + PMI_L + PR_L + Triplet_L | ALL | 52.34 1.54 | ± | 71.20 0.10 | ± | 63.28 4.93 | ± | 71.24 0.11 | ± | 135.75 |
| CrossE_L + PMI_L + PR_L + Triplet_L | GAT | 63.93 2.01 | ± | 75.34 0.65 | ± | 76.16 1.85 | ± | 74.72 0.56 | ± | 28.0 |
| CrossE_L + PMI_L + PR_L + Triplet_L | GCN | 55.65 1.65 | ± | 73.76 0.53 | ± | 69.89 2.55 | ± | 73.26 0.44 | ± | 74.25 |
| CrossE_L + PMI_L + PR_L + Triplet_L | GIN | 52.25 1.12 | ± | 71.24 0.15 | ± | 66.38 3.54 | ± | 71.26 0.17 | ± | 123.0 |
| CrossE_L + PMI_L + PR_L + Triplet_L | MPNN | 62.37 1.54 | ± | 75.56 0.87 | ± | 77.16 1.12 | ± | 74.98 0.83 | ± | 24.75 |
| CrossE_L + PMI_L + PR_L + Triplet_L | PAGNN | 53.72 1.04 | ± | 71.20 0.00 | ± | 63.21 2.88 | ± | 71.20 0.00 | ± | 139.375 |
| CrossE_L + PMI_L + PR_L + Triplet_L | SAGE | 47.12 2.24 | ± | 77.42 0.75 | ± | 58.60 2.10 | ± | 78.46 2.28 | ± | 88.25 |
| CrossE_L + PMI_L + Triplet_L | ALL | 54.68 1.12 | ± | 72.74 0.91 | ± | 69.34 2.14 | ± | 73.34 0.83 | ± | 85.625 |
| CrossE_L + PMI_L + Triplet_L | GAT | 63.42 2.18 | ± | 75.30 0.53 | ± | 76.83 0.43 | ± | 75.60 0.63 | ± | 21.375 |
| CrossE_L + PMI_L + Triplet_L | GCN | 55.44 1.27 | ± | 72.46 0.89 | ± | 70.11 1.62 | ± | 73.02 1.18 | ± | 86.75 |
| CrossE_L + PMI_L + Triplet_L | GIN | 51.74 1.87 | ± | 71.44 0.34 | ± | 67.90 1.08 | ± | 71.06 0.41 | ± | 138.375 |
| CrossE_L + PMI_L + Triplet_L | MPNN | 61.65 1.52 | ± | 76.06 1.19 | ± | 77.20 1.10 | ± | 74.40 0.58 | ± | 27.25 |
| CrossE_L + PMI_L + Triplet_L | PAGNN | 52.13 1.98 | ± | 71.20 0.00 | ± | 61.62 3.75 | ± | 71.20 0.00 | ± | 149.625 |
| CrossE_L + PMI_L + Triplet_L | SAGE | 48.62 1.24 | ± | 76.52 1.61 | ± | 60.88 2.36 | ± | 77.08 1.19 | ± | 83.0 |

Table 22. Results for Node Cls Accuracy (↑) (continued)

| Loss Type | Model | Cora ↓ Citeseer | | Cora ↓ Bitcoin | | Citeseer ↓ Cora | | Citeseer ↓ Bitcoin | | Average Rank |
|---|---|---|---|---|---|---|---|---|---|---|
| CrossE_L + PR_L | ALL | 32.34 ± 2.42 | | 71.20 ± 0.00 | | 46.86 ± 2.57 | | 71.20 ± 0.00 | | 183.125 |
| CrossE_L + PR_L | GAT | 47.48 ± 8.10 | | 71.62 ± 0.22 | | 37.53 ± 1.56 | | 71.68 ± 0.32 | | 150.375 |
| CrossE_L + PR_L | GCN | 52.19 ± 3.88 | | 72.74 ± 1.26 | | 55.87 ± 2.99 | | 72.68 ± 0.64 | | 127.5 |
| CrossE_L + PR_L | GIN | 41.95 ± 4.75 | | 71.20 ± 0.00 | | 52.88 ± 6.16 | | 71.20 ± 0.00 | | 179.875 |
| CrossE_L + PR_L | MPNN | 52.28 ± 3.68 | | 71.20 ± 0.00 | | 60.70 ± 6.57 | | 71.22 ± 0.04 | | 144.125 |
| CrossE_L + PR_L | PAGNN | 43.04 ± 3.90 | | 71.20 ± 0.00 | | 55.79 ± 5.41 | | 71.20 ± 0.00 | | 177.375 |
| CrossE_L + PR_L | SAGE | 42.16 ± 4.63 | | 73.72 ± 1.65 | | 45.42 ± 3.65 | | 74.44 ± 2.32 | | 129.75 |
| CrossE_L + PR_L + Triplet_L | ALL | 44.02 ± 4.63 | | 71.26 ± 0.13 | | 51.00 ± 3.59 | | 71.22 ± 0.04 | | 163.25 |
| CrossE_L + PR_L + Triplet_L | GAT | 60.21 ± 1.55 | | 75.64 ± 0.63 | | 72.21 ± 2.69 | | 75.22 ± 0.56 | | 38.125 |
| CrossE_L + PR_L + Triplet_L | GCN | 56.58 ± 3.31 | | 74.06 ± 0.40 | | 69.52 ± 2.61 | | 73.30 ± 1.28 | | 71.75 |
| CrossE_L + PR_L + Triplet_L | GIN | 51.83 ± 4.00 | | 70.94 ± 0.27 | | 66.53 ± 2.41 | | 71.16 ± 0.17 | | 158.75 |
| CrossE_L + PR_L + Triplet_L | MPNN | 56.43 ± 1.65 | | 72.86 ± 0.90 | | 72.44 ± 1.00 | | 73.52 ± 1.16 | | 70.125 |
| CrossE_L + PR_L + Triplet_L | PAGNN | 47.39 ± 2.14 | | 71.20 ± 0.00 | | 63.06 ± 3.90 | | 71.20 ± 0.00 | | 158.625 |
| CrossE_L + PR_L + Triplet_L | SAGE | 48.59 ± 1.99 | | 73.50 ± 1.14 | | 58.45 ± 1.37 | | 72.52 ± 0.85 | | 127.0 |
| CrossE_L + Triplet_L | ALL | 50.72 ± 3.30 | | 71.12 ± 0.29 | | 68.52 ± 2.42 | | 70.88 ± 0.35 | | 161.625 |
| CrossE_L + Triplet_L | GAT | 63.36 ± 1.98 | | 75.84 ± 0.31 | | 77.38 ± 1.15 | | 76.04 ± 0.27 | | 13.125 |

Table 22. Results for Node Cls Accuracy (↑) (continued)

| Loss Type | Model | Cora ↓ Citeseer | | Cora ↓ Bitcoin | | Citeseer ↓ Cora | | Citeseer ↓ Bitcoin | | Average Rank |
|---|---|---|---|---|---|---|---|---|---|---|
| CrossE_L + Triplet_L | GCN | 55.86 | ± 0.89 | 73.22 | ± 0.60 | 71.55 | ± 1.93 | 73.04 | ± 0.75 | 76.875 |
| CrossE_L + Triplet_L | GIN | 54.17 | ± 2.14 | 71.02 | ± 0.86 | 72.36 | ± 0.66 | 71.40 | ± 0.57 | 118.0 |
| CrossE_L + Triplet_L | MPNN | 60.42 | ± 1.97 | 74.64 | ± 0.77 | 76.72 | ± 1.64 | 73.08 | ± 0.53 | 50.125 |
| CrossE_L + Triplet_L | PAGNN | 55.07 | ± 1.95 | 71.20 | ± 0.00 | 72.22 | ± 1.56 | 71.18 | ± 0.04 | 122.375 |
| CrossE_L + Triplet_L | SAGE | 53.90 | ± 3.13 | 73.32 | ± 1.75 | 65.50 | ± 1.58 | 73.30 | ± 0.39 | 91.0 |
| PMI_L | ALL | 56.70 | ± 1.75 | 71.34 | ± 0.11 | 69.00 | ± 4.26 | 72.76 | ± 0.97 | 93.75 |
| PMI_L | GAT | 63.16 | ± 2.77 | 75.70 | ± 0.59 | 76.01 | ± 1.47 | 75.46 | ± 0.30 | 21.75 |
| PMI_L | GCN | 56.18 | ± 1.44 | 73.24 | ± 0.87 | 68.41 | ± 2.98 | 73.20 | ± 0.43 | 82.25 |
| PMI_L | GIN | 52.34 | ± 2.78 | 70.96 | ± 0.27 | 64.87 | ± 4.81 | 71.12 | ± 0.24 | 161.125 |
| PMI_L | MPNN | 61.32 | ± 1.30 | 75.30 | ± 0.82 | 75.72 | ± 1.21 | 75.14 | ± 1.10 | 34.75 |
| PMI_L | PAGNN | 53.57 | ± 0.39 | 71.20 | ± 0.00 | 58.34 | ± 2.37 | 71.20 | ± 0.00 | 150.375 |
| PMI_L | SAGE | 44.44 | ± 3.20 | **78.68** | **± 2.22** | 54.24 | ± 3.82 | 77.64 | ± 1.73 | 96.25 |
| PMI_L + PR_L | ALL | 48.59 | ± 3.44 | 71.40 | ± 0.00 | 62.29 | ± 5.59 | 71.16 | ± 0.05 | 153.25 |
| PMI_L + PR_L | GAT | 62.82 | ± 2.71 | 75.38 | ± 0.68 | 74.79 | ± 3.29 | 74.46 | ± 1.15 | 35.875 |
| PMI_L + PR_L | GCN | 56.25 | ± 1.49 | 73.02 | ± 0.86 | 70.96 | ± 1.37 | 73.04 | ± 0.77 | 78.0 |
| PMI_L + PR_L | GIN | 52.04 | ± 0.74 | 71.34 | ± 0.26 | 61.81 | ± 5.19 | 71.02 | ± 0.29 | 150.25 |

Table 22.  Results for Node Cls Accuracy (↑) (continued)

| Loss Type | Model | Cora ↓ Citeseer | Cora ↓ Bitcoin | Citeseer ↓ Cora | Citeseer ↓ Bitcoin | Average Rank |
|---|---|---|---|---|---|---|
| PMI_L + PR_L | MPNN | 60.57 ± 2.33 | 75.52 ± 0.89 | 73.87 ± 1.63 | 71.84 ± 0.80 | 56.25 |
| PMI_L + PR_L | PAGNN | 53.21 ± 0.99 | 71.20 ± 0.00 | 64.76 ± 3.47 | 71.20 ± 0.00 | 138.75 |
| PMI_L + PR_L | SAGE | 48.83 ± 1.68 | 77.62 ± 2.33 | 58.38 ± 2.50 | 78.96 ± 1.81 | 83.875 |
| PMI_L + PR_L + Triplet_L | ALL | 49.40 ± 3.66 | 71.26 ± 0.05 | 64.54 ± 1.53 | 71.08 ± 0.16 | 152.625 |
| PMI_L + PR_L + Triplet_L | GAT | 62.73 ± 2.81 | 75.00 ± 0.81 | 76.13 ± 1.47 | 75.44 ± 0.90 | 29.625 |
| PMI_L + PR_L + Triplet_L | GCN | 55.47 ± 2.08 | 72.36 ± 0.73 | 69.67 ± 1.88 | 73.30 ± 0.42 | 84.875 |
| PMI_L + PR_L + Triplet_L | GIN | 53.75 ± 1.54 | 71.20 ± 0.21 | 66.35 ± 3.00 | 71.24 ± 0.15 | 125.5 |
| PMI_L + PR_L + Triplet_L | MPNN | 61.71 ± 2.06 | 75.08 ± 0.50 | 75.27 ± 1.65 | 72.44 ± 0.55 | 53.875 |
| PMI_L + PR_L + Triplet_L | PAGNN | 51.89 ± 0.97 | 71.20 ± 0.00 | 66.46 ± 3.01 | 71.20 ± 0.00 | 140.75 |
| PMI_L + PR_L + Triplet_L | SAGE | 49.19 ± 2.70 | 78.28 ± 1.43 | 59.41 ± 3.97 | 76.52 ± 1.58 | 83.25 |
| PMI_L + Triplet_L | ALL | 55.94 ± 2.66 | 71.04 ± 0.30 | 71.77 ± 1.85 | 73.24 ± 0.38 | 104.0 |
| PMI_L + Triplet_L | GAT | 63.60 ± 1.22 | 75.12 ± 0.53 | 76.09 ± 2.38 | 75.00 ± 0.82 | 29.875 |
| PMI_L + Triplet_L | GCN | 54.66 ± 1.62 | 72.94 ± 0.88 | 70.30 ± 1.71 | 73.04 ± 0.92 | 84.5 |
| PMI_L + Triplet_L | GIN | 53.06 ± 2.49 | 71.36 ± 0.18 | 69.52 ± 2.97 | 71.16 ± 0.09 | 126.375 |
| PMI_L + Triplet_L | MPNN | 61.29 ± 2.12 | 75.80 ± 0.72 | 77.71 ± 1.59 | 75.48 ± 0.87 | 20.5 |
| PMI_L + Triplet_L | PAGNN | 51.80 ± 2.06 | 71.20 ± 0.00 | 61.77 ± 3.98 | 71.20 ± 0.00 | 150.625 |

Table 22. Results for Node Cls Accuracy (↑) (continued)

| Loss Type | Model | Cora ↓ Citeseer | | Cora ↓ Bitcoin | | Citeseer ↓ Cora | | Citeseer ↓ Bitcoin | | Average Rank |
|---|---|---|---|---|---|---|---|---|---|---|
| PMI_L + Triplet_L | SAGE | 49.13 ± 1.83 | | 78.90 ± 1.51 | | 59.48 ± 3.61 | | 78.12 ± 1.51 | | 79.625 |
| PR_L | ALL | 30.42 ± 0.90 | | 71.20 ± 0.00 | | 43.17 ± 1.18 | | 71.20 ± 0.00 | | 183.875 |
| PR_L | GAT | 49.28 ± 2.01 | | 71.62 ± 0.46 | | 37.23 ± 1.96 | | 72.02 ± 0.79 | | 144.875 |
| PR_L | GCN | 50.54 ± 2.73 | | 71.92 ± 0.23 | | 59.78 ± 2.53 | | 71.98 ± 0.49 | | 131.5 |
| PR_L | GIN | 38.74 ± 4.08 | | 71.20 ± 0.00 | | 49.74 ± 5.91 | | 71.22 ± 0.04 | | 174.375 |
| PR_L | MPNN | 51.44 ± 3.37 | | 71.20 ± 0.00 | | 59.48 ± 3.78 | | 71.20 ± 0.00 | | 156.0 |
| PR_L | PAGNN | 41.35 ± 4.62 | | 71.20 ± 0.00 | | 58.12 ± 2.66 | | 71.20 ± 0.00 | | 175.375 |
| PR_L | SAGE | 42.79 ± 3.79 | | 73.12 ± 1.08 | | 40.77 ± 4.09 | | 75.92 ± 0.77 | | 125.25 |
| PR_L + Triplet_L | ALL | 34.41 ± 2.23 | | 71.20 ± 0.00 | | 48.71 ± 2.14 | | 71.20 ± 0.00 | | 182.125 |
| PR_L + Triplet_L | GAT | 58.08 ± 3.24 | | 75.02 ± 1.05 | | 57.71 ± 4.04 | | 74.46 ± 0.40 | | 82.75 |
| PR_L + Triplet_L | GCN | 54.26 ± 2.50 | | 73.28 ± 0.97 | | 65.76 ± 3.53 | | 72.78 ± 0.33 | | 94.625 |
| PR_L + Triplet_L | GIN | 48.35 ± 3.49 | | 71.18 ± 0.04 | | 63.58 ± 1.22 | | 71.30 ± 0.41 | | 152.625 |
| PR_L + Triplet_L | MPNN | 54.26 ± 1.94 | | 71.36 ± 0.36 | | 60.37 ± 3.77 | | 71.70 ± 0.73 | | 121.25 |
| PR_L + Triplet_L | PAGNN | 45.40 ± 2.38 | | 71.20 ± 0.00 | | 56.13 ± 2.85 | | 71.20 ± 0.00 | | 174.125 |
| PR_L + Triplet_L | SAGE | 45.26 ± 1.31 | | 73.46 ± 1.10 | | 49.81 ± 6.02 | | 74.24 ± 1.75 | | 128.5 |
| Triplet_L | ALL | 53.75 ± 2.82 | | 71.24 ± 0.33 | | 69.74 ± 3.11 | | 73.22 ± 1.41 | | 97.75 |

Table 22. Results for Node Cls Accuracy (↑) (continued)

| Loss Type | Model | Cora ↓ Citeseer | | Cora ↓ Bitcoin | | Citeseer ↓ Cora | | Citeseer ↓ Bitcoin | | Average Rank |
|---|---|---|---|---|---|---|---|---|---|---|
| Triplet_L | GAT | 63.99 ± 1.05 | | 75.50 ± 0.78 | | 78.49 ± 1.53 | | 75.36 ± 0.48 | | 16.875 |
| Triplet_L | GCN | 57.72 ± 2.67 | | 73.28 ± 0.66 | | 72.66 ± 2.44 | | 73.00 ± 0.63 | | 69.25 |
| Triplet_L | GIN | 55.92 ± 1.39 | | 71.30 ± 0.47 | | 71.03 ± 1.15 | | 71.54 ± 0.34 | | 96.0 |
| Triplet_L | MPNN | 61.71 ± 1.32 | | 74.96 ± 1.25 | | 78.01 ± 2.05 | | 74.66 ± 0.84 | | 32.5 |
| Triplet_L | PAGNN | 57.36 ± 2.10 | | 71.20 ± 0.00 | | 72.80 ± 2.25 | | 71.20 ± 0.00 | | 108.625 |
| Triplet_L | SAGE | 53.21 ± 1.09 | | 76.14 ± 1.11 | | 63.66 ± 2.75 | | 76.06 ± 1.03 | | 64.75 |

Table 23. Node Cls F1 Performance (↑): This table presents models (Loss function and GNN) ranked by their average performance in terms of node cls f1. Top-ranked results are highlighted in red, second-ranked in blue, and third-ranked in green.

| Loss Type | Model | Cora ↓ Citeseer | | Cora ↓ Bitcoin | | Citeseer ↓ Cora | | Citeseer ↓ Bitcoin | | Average Rank |
|---|---|---|---|---|---|---|---|---|---|---|
| Contr_l | ALL | 37.09 ± 3.12 | | 27.72 ± 0.01 | | 52.22 ± 6.88 | | 27.90 ± 0.29 | | 170.375 |
| Contr_l | GAT | 54.37 ± 1.17 | | 41.00 ± 0.96 | | 74.04 ± 1.84 | | 40.29 ± 0.89 | | 26.125 |
| Contr_l | GCN | 50.07 ± 1.33 | | 36.31 ± 0.98 | | 69.23 ± 1.81 | | 37.24 ± 1.45 | | 58.875 |
| Contr_l | GIN | 45.26 ± 3.18 | | 29.86 ± 0.97 | | 65.63 ± 4.07 | | 32.16 ± 0.88 | | 101.375 |
| Contr_l | MPNN | 51.12 ± 1.69 | | 39.39 ± 0.85 | | 69.63 ± 2.92 | | 38.11 ± 1.79 | | 48.875 |

Table 23. Results for Node Cls F1 (↑) (continued)

| Loss Type | Model | Cora ↓ Citeseer | | Cora ↓ Bitcoin | | Citeseer ↓ Cora | | Citeseer ↓ Bitcoin | | Average Rank |
|---|---|---|---|---|---|---|---|---|---|---|
| Contr_l | PAGNN | 46.01 1.53 | ± | 27.73 0.00 | ± | 62.46 9.08 | ± | 27.73 0.00 | ± | 139.375 |
| Contr_l | SAGE | 40.08 2.37 | ± | 38.07 3.08 | ± | 56.86 6.07 | ± | 32.82 2.89 | ± | 111.75 |
| Contr_l + CrossE_L | ALL | 40.68 2.55 | ± | 28.06 0.40 | ± | 43.55 10.61 | ± | 27.77 0.09 | ± | 162.25 |
| Contr_l + CrossE_L | GAT | 53.89 1.12 | ± | 41.06 0.86 | ± | 71.30 2.69 | ± | 40.44 0.84 | ± | 32.125 |
| Contr_l + CrossE_L | GCN | 49.18 1.62 | ± | 37.22 0.52 | ± | 69.16 2.90 | ± | 36.16 1.33 | ± | 63.375 |
| Contr_l + CrossE_L | GIN | 45.34 2.46 | ± | 28.45 0.22 | ± | 64.00 2.29 | ± | 30.68 1.37 | ± | 109.125 |
| Contr_l + CrossE_L | MPNN | 52.14 1.16 | ± | 39.10 0.59 | ± | 69.83 3.38 | ± | 37.12 0.93 | ± | 50.875 |
| Contr_l + CrossE_L | PAGNN | 45.83 2.33 | ± | 27.73 0.00 | ± | 61.56 4.15 | ± | 27.78 0.11 | ± | 134.75 |
| Contr_l + CrossE_L | SAGE | 41.41 3.08 | ± | 32.37 2.07 | ± | 53.57 3.72 | ± | 28.42 0.58 | ± | 133.375 |
| Contr_l + CrossE_L + PMI_L | ALL | 44.92 3.17 | ± | 27.77 0.10 | ± | 62.46 8.53 | ± | 29.29 1.36 | ± | 121.875 |
| Contr_l + CrossE_L + PMI_L | GAT | 55.07 1.99 | ± | 42.26 0.73 | ± | 73.20 1.66 | ± | 41.31 0.74 | ± | 16.5 |
| Contr_l + CrossE_L + PMI_L | GCN | 48.81 2.17 | ± | 35.45 0.43 | ± | 64.29 2.36 | ± | 36.85 2.31 | ± | 75.0 |
| Contr_l + CrossE_L + PMI_L | GIN | 43.58 1.29 | ± | 28.38 0.42 | ± | 56.44 6.93 | ± | 28.97 0.76 | ± | 129.0 |
| Contr_l + CrossE_L + PMI_L | MPNN | 54.76 1.43 | ± | 40.19 1.10 | ± | 72.35 1.80 | ± | 40.69 1.12 | ± | 32.625 |
| Contr_l + CrossE_L + PMI_L | PAGNN | 43.51 2.12 | ± | 27.73 0.00 | ± | 45.68 4.14 | ± | 27.73 0.00 | ± | 169.875 |
| Contr_l + CrossE_L + PMI_L | SAGE | 40.06 2.22 | ± | 44.83 2.47 | ± | 49.31 4.34 | ± | 42.17 3.30 | ± | 84.75 |

Table 23. Results for Node Cls F1 (↑) (continued)

| Loss Type | Model | Cora ↓ Citeseer | | Cora ↓ Bitcoin | | Citeseer ↓ Cora | | Citeseer ↓ Bitcoin | | Average Rank |
|---|---|---|---|---|---|---|---|---|---|---|
| Contr_l + CrossE_L + PMI_L + PR_L | ALL | 34.66 | ± | 28.22 | ± | 55.96 | ± | 27.89 | ± | 151.875 |
| | | 1.85 | | 0.00 | | 5.30 | | 0.39 | | |
| Contr_l + CrossE_L + PMI_L + PR_L | GAT | 54.80 | ± | 40.99 | ± | 74.04 | ± | 40.81 | ± | 22.5 |
| | | 1.65 | | 1.38 | | 3.39 | | 0.84 | | |
| Contr_l + CrossE_L + PMI_L + PR_L | GCN | 49.99 | ± | 35.91 | ± | 66.26 | ± | 37.08 | ± | 66.5 |
| | | 1.40 | | 1.50 | | 2.61 | | 0.98 | | |
| Contr_l + CrossE_L + PMI_L + PR_L | GIN | 43.63 | ± | 28.22 | ± | 49.93 | ± | 27.90 | ± | 144.875 |
| | | 2.76 | | 0.33 | | 10.00 | | 0.29 | | |
| Contr_l + CrossE_L + PMI_L + PR_L | MPNN | 53.58 | ± | 41.34 | ± | 71.92 | ± | 33.78 | ± | 47.0 |
| | | 0.55 | | 0.85 | | 1.97 | | 2.53 | | |
| Contr_l + CrossE_L + PMI_L + PR_L | PAGNN | 45.10 | ± | 27.73 | ± | 48.82 | ± | 27.73 | ± | 160.5 |
| | | 1.99 | | 0.00 | | 2.75 | | | | |
| Contr_l + CrossE_L + PMI_L + PR_L | SAGE | 40.66 | ± | 45.57 ± | | 48.27 | ± | 42.78 | ± | 83.625 |
| | | 2.76 | | 2.09 | | 1.72 | | 2.36 | | |
| Contr_l + CrossE_L + PMI_L + PR_L + Triplet_L | ALL | 43.21 | ± | 27.97 | ± | 56.50 | ± | 28.62 | ± | 136.875 |
| | | 2.07 | | 0.35 | | 7.10 | | 0.94 | | |
| Contr_l + CrossE_L + PMI_L + PR_L + Triplet_L | GAT | 54.78 | ± | 41.61 | ± | 74.45 | ± | 41.74 | ± | 14.0 |
| | | 1.65 | | 0.57 | | 1.70 | | 0.76 | | |
| Contr_l + CrossE_L + PMI_L + PR_L + Triplet_L | GCN | 49.64 | ± | 37.11 | ± | 65.65 | ± | 35.60 | ± | 71.5 |
| | | 3.03 | | 1.19 | | 2.39 | | 0.85 | | |
| Contr_l + CrossE_L + PMI_L + PR_L + Triplet_L | GIN | 43.27 | ± | 28.40 | ± | 62.86 | ± | 28.28 | ± | 126.875 |
| | | 1.61 | | 0.51 | | 2.25 | | 0.38 | | |
| Contr_l + CrossE_L + PMI_L + PR_L + Triplet_L | MPNN | 53.36 | ± | 41.55 | ± | 73.08 | ± | 35.79 | ± | 40.5 |
| | | 1.02 | | 1.55 | | 1.74 | | 3.19 | | |
| Contr_l + CrossE_L + PMI_L + PR_L + Triplet_L | PAGNN | 44.92 | ± | 27.73 | ± | 53.68 | ± | 27.73 | ± | 153.125 |
| | | 1.94 | | 0.00 | | 2.90 | | 0.00 | | |

Table 23. Results for Node Cls F1 (↑) (continued)

| Loss Type | Model | Cora ↓ Citeseer | | Cora ↓ Bitcoin | | Citeseer ↓ Cora | | Citeseer ↓ Bitcoin | | Average Rank |
|---|---|---|---|---|---|---|---|---|---|---|
| Contr_l + CrossE_L + PMI_L + PR_L + Triplet_L | SAGE | 42.42 | ± 1.64 | 42.75 | 0.30 | 53.43 | ± 2.99 | 45.24 | ± 2.19 | 72.75 |
| Contr_l + CrossE_L + PMI_L + Triplet_L | ALL | 46.64 | 2.22 | 30.72 | 2.59 | 64.18 | ± 1.54 | 35.01 | ± 2.15 | 91.875 |
| Contr_l + CrossE_L + PMI_L + Triplet_L | GAT | 56.40 | ± 1.83 | 41.39 | 0.60 | 75.16 | ± 1.56 | 41.49 | ± 0.65 | 12.0 |
| Contr_l + CrossE_L + PMI_L + Triplet_L | GCN | 49.19 | 1.86 | 37.01 | 0.73 | 64.18 | ± 2.03 | 35.96 | ± 1.99 | 73.375 |
| Contr_l + CrossE_L + PMI_L + Triplet_L | GIN | 45.53 | 1.51 | 28.96 | 1.08 | 63.61 | ± 4.78 | 28.91 | ± 1.02 | 110.125 |
| Contr_l + CrossE_L + PMI_L + Triplet_L | MPNN | 53.33 | 1.86 | 40.69 | 1.54 | 72.51 | ± 1.34 | 39.13 | ± 0.81 | 37.5 |
| Contr_l + CrossE_L + PMI_L + Triplet_L | PAGNN | 44.69 | 1.97 | 27.73 | 0.00 | 50.42 | ± 1.92 | 27.73 | ± 0.00 | 160.0 |
| Contr_l + CrossE_L + PMI_L + Triplet_L | SAGE | 40.70 | 4.05 | 42.85 | 1.28 | 52.44 | ± 2.99 | 43.76 | ± 1.19 | 77.75 |
| Contr_l + CrossE_L + PR_L | ALL | 19.07 | 3.63 | 27.73 | 0.00 | 27.88 | ± 8.37 | 27.73 | ± 0.00 | 192.5 |
| Contr_l + CrossE_L + PR_L | GAT | 48.42 | 3.07 | 40.40 | 0.84 | 43.55 | ± 13.31 | 36.19 | ± 2.02 | 93.0 |
| Contr_l + CrossE_L + PR_L | GCN | 46.10 | 2.61 | 35.16 | 1.69 | 61.34 | ± 3.50 | 33.91 | ± 1.84 | 96.0 |
| Contr_l + CrossE_L + PR_L | GIN | 34.91 | 1.87 | 27.93 | 0.44 | 50.66 | ± 4.44 | 27.86 | ± 0.12 | 160.0 |
| Contr_l + CrossE_L + PR_L | MPNN | 45.49 | 2.75 | 28.45 | 0.84 | 60.07 | ± 7.46 | 28.82 | ± 1.18 | 118.125 |
| Contr_l + CrossE_L + PR_L | PAGNN | 36.95 | 3.23 | 27.73 | 0.00 | 42.59 | ± 6.54 | 27.73 | ± 0.00 | 183.75 |
| Contr_l + CrossE_L + PR_L | SAGE | 36.09 | 3.47 | 34.98 | 4.39 | 39.53 | ± 6.12 | 38.54 | ± 2.72 | 129.75 |
| Contr_l + CrossE_L + PR_L + Triplet_L | ALL | 33.67 | 3.12 | 27.73 | 0.00 | 41.17 | ± 11.21 | 27.78 | ± 0.11 | 181.25 |

Table 23. Results for Node Cls F1 (↑) (continued)

| Loss Type | Model | Cora ↓ Citeseer | | Cora ↓ Bitcoin | | Citeseer ↓ Cora | | Citeseer ↓ Bitcoin | | Average Rank |
|---|---|---|---|---|---|---|---|---|---|---|
| Contr_l + CrossE_L + PR_L + Triplet_L | GAT | 52.33 2.44 | ± | 40.29 0.48 | ± | 70.60 2.71 | ± | 40.28 1.22 | ± | 41.625 |
| Contr_l + CrossE_L + PR_L + Triplet_L | GCN | 46.91 2.85 | ± | 35.81 1.88 | ± | 62.59 3.56 | ± | 36.28 1.31 | ± | 83.375 |
| Contr_l + CrossE_L + PR_L + Triplet_L | GIN | 43.98 2.28 | ± | 27.95 0.25 | ± | 61.69 2.75 | ± | 29.50 0.91 | ± | 123.75 |
| Contr_l + CrossE_L + PR_L + Triplet_L | MPNN | 50.33 1.22 | ± | 34.95 1.79 | ± | 67.72 3.52 | ± | 32.64 1.97 | ± | 74.875 |
| Contr_l + CrossE_L + PR_L + Triplet_L | PAGNN | 41.08 2.37 | ± | 27.73 0.00 | ± | 50.48 3.68 | ± | 27.73 0.00 | ± | 168.75 |
| Contr_l + CrossE_L + PR_L + Triplet_L | SAGE | 37.27 2.55 | ± | 32.00 1.52 | ± | 48.29 4.57 | ± | 35.62 1.81 | ± | 134.25 |
| Contr_l + CrossE_L + Triplet_L | ALL | 40.58 4.85 | ± | 27.90 0.26 | ± | 54.26 4.64 | ± | 28.16 0.47 | ± | 147.875 |
| Contr_l + CrossE_L + Triplet_L | GAT | 53.72 1.87 | ± | 41.49 0.93 | ± | 72.83 1.97 | ± | 40.70 0.54 | ± | 25.5 |
| Contr_l + CrossE_L + Triplet_L | GCN | 50.33 1.85 | ± | 35.87 1.09 | ± | 67.86 1.79 | ± | 37.62 0.65 | ± | 59.375 |
| Contr_l + CrossE_L + Triplet_L | GIN | 46.25 2.39 | ± | 30.70 1.05 | ± | 67.32 2.87 | ± | 31.51 0.56 | ± | 92.75 |
| Contr_l + CrossE_L + Triplet_L | MPNN | 51.78 1.79 | ± | 37.82 1.90 | ± | 70.60 2.15 | ± | 38.46 1.55 | ± | 48.375 |
| Contr_l + CrossE_L + Triplet_L | PAGNN | 46.51 1.51 | ± | 27.73 0.00 | ± | 66.09 2.44 | ± | 27.75 0.08 | ± | 126.625 |
| Contr_l + CrossE_L + Triplet_L | SAGE | 43.52 2.20 | ± | 34.75 2.40 | ± | 58.29 3.11 | ± | 37.49 2.64 | ± | 100.5 |
| Contr_l + PMI_L | ALL | 43.70 1.91 | ± | 27.78 0.11 | ± | 61.27 9.18 | ± | 29.63 1.64 | ± | 127.25 |
| Contr_l + PMI_L | GAT | 55.41 1.51 | ± | 41.11 0.73 | ± | 72.90 3.77 | ± | 40.65 1.02 | ± | 24.5 |
| Contr_l + PMI_L | GCN | 49.73 2.41 | ± | 36.78 0.66 | ± | 66.05 2.80 | ± | 36.66 0.47 | ± | 67.25 |

Table 23. Results for Node Cls F1 (↑) (continued)

| Loss Type | Model | Cora ↓ Citeseer | ± | Cora ↓ Bitcoin | ± | Citeseer ↓ Cora | ± | Citeseer ↓ Bitcoin | ± | Average Rank |
|---|---|---|---|---|---|---|---|---|---|---|
| Contr_l + PMI_L | GIN | 44.74 2.87 | ± | 29.59 1.23 | ± | 57.34 6.15 | ± | 28.83 1.09 | ± | 121.0 |
| Contr_l + PMI_L | MPNN | 52.98 0.90 | ± | 40.81 1.82 | ± | 73.34 1.54 | ± | 39.67 1.12 | ± | 33.125 |
| Contr_l + PMI_L | PAGNN | 44.80 1.55 | ± | 27.73 0.00 | ± | 41.37 4.31 | ± | 27.73 0.00 | ± | 167.5 |
| Contr_l + PMI_L | SAGE | 39.85 1.46 | ± | 45.40 2.38 | ± | 50.27 4.82 | ± | 43.49 3.07 | ± | 82.5 |
| Contr_l + PMI_L + PR_L | ALL | 39.15 0.91 | ± | 28.31 0.12 | ± | 51.05 5.66 | ± | 27.90 0.21 | ± | 151.25 |
| Contr_l + PMI_L + PR_L | GAT | 55.73 1.18 | ± | 41.94 0.82 | ± | 70.36 1.69 | ± | 37.92 0.59 | ± | 29.5 |
| Contr_l + PMI_L + PR_L | GCN | 48.84 1.91 | ± | 36.93 1.86 | ± | 66.56 2.67 | ± | 36.90 1.55 | ± | 66.25 |
| Contr_l + PMI_L + PR_L | GIN | 43.55 0.75 | ± | 27.91 0.30 | ± | 49.68 11.60 | ± | 27.87 0.22 | ± | 148.875 |
| Contr_l + PMI_L + PR_L | MPNN | 53.19 1.98 | ± | 40.43 1.93 | ± | 71.18 1.94 | ± | 32.36 2.87 | ± | 56.25 |
| Contr_l + PMI_L + PR_L | PAGNN | 43.21 1.39 | ± | 27.73 0.00 | ± | 47.47 7.57 | ± | 27.73 0.00 | ± | 169.875 |
| Contr_l + PMI_L + PR_L | SAGE | 40.45 1.65 | ± | 44.50 3.24 | ± | 48.21 4.53 | ± | 37.91 4.41 | ± | 97.0 |
| Contr_l + PMI_L + PR_L + Triplet_L | ALL | 43.15 1.30 | ± | 27.86 0.21 | ± | 49.10 4.28 | ± | 28.04 0.26 | ± | 151.75 |
| Contr_l + PMI_L + PR_L + Triplet_L | GAT | 53.67 1.17 | ± | 41.26 1.23 | ± | 71.11 2.55 | ± | 38.60 2.18 | ± | 35.875 |
| Contr_l + PMI_L + PR_L + Triplet_L | GCN | 48.31 0.54 | ± | 36.15 1.47 | ± | 66.88 2.96 | ± | 36.44 1.38 | ± | 71.25 |
| Contr_l + PMI_L + PR_L + Triplet_L | GIN | 44.34 2.05 | ± | 28.61 1.13 | ± | 61.13 1.81 | ± | 28.89 0.59 | ± | 121.75 |
| Contr_l + PMI_L + PR_L + Triplet_L | MPNN | 54.19 2.24 | ± | 40.02 1.09 | ± | 73.51 3.44 | ± | 30.67 1.78 | ± | 52.5 |

Table 23. Results for Node Cls F1 (↑) (continued)

| Loss Type | Model | Cora ↓ Citeseer | | Cora ↓ Bitcoin | | Citeseer ↓ Cora | | Citeseer ↓ Bitcoin | | Average Rank |
|---|---|---|---|---|---|---|---|---|---|---|
| Contr_l + PMI_L + PR_L + Triplet_L | PAGNN | 44.75 1.39 | ± | 27.73 0.00 | ± | 59.32 5.66 | ± | 27.73 0.00 | ± | 149.0 |
| Contr_l + PMI_L + PR_L + Triplet_L | SAGE | 42.00 3.86 | ± | 43.28 2.33 | ± | 51.32 4.38 | ± | 34.65 1.84 | ± | 98.5 |
| Contr_l + PR_L | ALL | 22.14 1.90 | ± | 27.73 0.00 | ± | 23.25 5.74 | ± | 27.73 0.00 | ± | 193.0 |
| Contr_l + PR_L | GAT | 46.37 4.12 | ± | 39.54 1.05 | ± | 45.25 8.26 | ± | 37.57 0.83 | ± | 94.375 |
| Contr_l + PR_L | GCN | 44.87 2.22 | ± | 35.81 1.39 | ± | 57.93 8.11 | ± | 34.63 1.08 | ± | 101.625 |
| Contr_l + PR_L | GIN | 36.62 4.81 | ± | 27.73 0.00 | ± | 50.51 8.94 | ± | 28.42 0.67 | ± | 164.0 |
| Contr_l + PR_L | MPNN | 41.21 2.92 | ± | 27.73 0.00 | ± | 53.96 8.34 | ± | 30.42 2.32 | ± | 146.625 |
| Contr_l + PR_L | PAGNN | 34.75 3.99 | ± | 27.73 0.00 | ± | 40.72 3.94 | ± | 27.73 0.00 | ± | 187.0 |
| Contr_l + PR_L | SAGE | 37.92 2.12 | ± | 32.70 1.04 | ± | 41.41 8.23 | ± | 39.39 3.56 | ± | 126.25 |
| Contr_l + PR_L + Triplet_L | ALL | 33.69 2.36 | ± | 27.72 0.02 | ± | 40.63 7.75 | ± | 27.77 0.11 | ± | 188.75 |
| Contr_l + PR_L + Triplet_L | GAT | 50.05 1.36 | ± | 40.34 1.33 | ± | 67.09 5.73 | ± | 40.91 1.48 | ± | 47.0 |
| Contr_l + PR_L + Triplet_L | GCN | 47.59 1.79 | ± | 37.57 1.45 | ± | 63.93 1.70 | ± | 34.58 2.51 | ± | 81.375 |
| Contr_l + PR_L + Triplet_L | GIN | 43.91 2.67 | ± | 30.35 1.67 | ± | 61.73 5.89 | ± | 29.30 0.69 | ± | 115.25 |
| Contr_l + PR_L + Triplet_L | MPNN | 50.00 1.37 | ± | 34.94 1.28 | ± | 67.59 3.15 | ± | 32.63 3.61 | ± | 77.0 |
| Contr_l + PR_L + Triplet_L | PAGNN | 43.51 2.35 | ± | 27.73 0.00 | ± | 54.46 6.47 | ± | 27.73 0.00 | ± | 158.625 |
| Contr_l + PR_L + Triplet_L | SAGE | 39.72 3.79 | ± | 36.84 3.34 | ± | 50.09 1.68 | ± | 37.60 3.97 | ± | 112.875 |

Table 23. Results for Node Cls F1 (↑) (continued)

| Loss Type | Model | Cora ↓ Citeseer | | Cora ↓ Bitcoin | | Citeseer ↓ Cora | | Citeseer ↓ Bitcoin | | Average Rank |
|---|---|---|---|---|---|---|---|---|---|---|
| Contr_l + Triplet_L | ALL | 40.25 | ± 1.79 | 27.87 | ± 0.28 | 55.32 | ± 7.05 | 27.97 | ± 0.22 | 148.75 |
| Contr_l + Triplet_L | GAT | 56.10 | ± 1.42 | 40.56 | ± 0.84 | 74.32 | ± 1.42 | 41.49 | ± 1.08 | 19.25 |
| Contr_l + Triplet_L | GCN | 47.82 | ± 1.60 | 36.55 | ± 0.85 | 69.63 | ± 0.94 | 37.05 | ± 1.80 | 65.125 |
| Contr_l + Triplet_L | GIN | 45.77 | ± 1.11 | 29.87 | ± 1.65 | 64.59 | ± 3.85 | 31.80 | ± 0.73 | 99.875 |
| Contr_l + Triplet_L | MPNN | 51.53 | ± 1.52 | 37.22 | ± 1.56 | 71.71 | ± 2.12 | 37.98 | ± 0.97 | 48.375 |
| Contr_l + Triplet_L | PAGNN | 47.48 | ± 0.70 | 27.73 | ± 0.00 | 63.53 | ± 3.45 | 27.73 | ± 0.00 | 134.75 |
| Contr_l + Triplet_L | SAGE | 42.48 | ± 1.67 | 31.96 | ± 0.55 | 57.40 | ± 4.06 | 33.35 | ± 0.96 | 116.75 |
| CrossE_L | ALL | 7.62±2.36 | | 27.73 | ± 0.00 | 7.44±0.00 | | 27.73 | ± 0.00 | 196.5 |
| CrossE_L | GAT | 5.89±0.00 | | 27.73 | ± 0.00 | 7.44±0.00 | | 27.73 | ± 0.00 | 197.25 |
| CrossE_L | GCN | 5.89±0.00 | | 27.73 | ± 0.00 | 7.44±0.00 | | 27.73 | ± 0.00 | 197.25 |
| CrossE_L | GIN | 5.89±0.00 | | 27.73 | ± 0.00 | 7.44±0.00 | | 27.73 | ± 0.00 | 197.25 |
| CrossE_L | MPNN | 5.89±0.00 | | 27.73 | ± 0.00 | 7.44±0.00 | | 27.73 | ± 0.00 | 197.25 |
| CrossE_L | PAGNN | 10.81 | ± 4.52 | 27.73 | ± 0.00 | 7.44±0.00 | | 27.73 | ± 0.00 | 196.25 |
| CrossE_L | SAGE | 5.89±0.00 | | 27.73 | ± 0.00 | 7.44±0.00 | | 27.73 | ± 0.00 | 197.25 |
| CrossE_L + PMI_L | ALL | 50.20 | ± 1.62 | 28.22 | ± 0.42 | 62.00 | ± 4.83 | 28.75 | ± 0.73 | 105.875 |
| CrossE_L + PMI_L | GAT | 54.22 | ± 3.61 | 40.21 | ± 0.83 | 72.71 | ± 2.35 | 40.94 | ± 0.94 | 31.0 |

Table 23. Results for Node Cls F1 (↑) (continued)

| Loss Type | Model | Cora ↓ Citeseer | | Cora ↓ Bitcoin | | Citeseer ↓ Cora | | Citeseer ↓ Bitcoin | | Average Rank |
|---|---|---|---|---|---|---|---|---|---|---|
| CrossE_L + PMI_L | GCN | 49.23 1.31 | ± | 36.81 1.86 | ± | 67.79 3.85 | ± | 35.97 0.86 | ± | 66.75 |
| CrossE_L + PMI_L | GIN | 45.41 3.05 | ± | 28.84 1.35 | ± | 56.19 1.92 | ± | 28.68 0.91 | ± | 121.75 |
| CrossE_L + PMI_L | MPNN | 53.39 1.39 | ± | 41.12 1.63 | ± | 74.73 1.63 | ± | 39.94 0.96 | ± | 27.0 |
| CrossE_L + PMI_L | PAGNN | 45.33 2.51 | ± | 27.73 0.00 | ± | 43.58 3.52 | ± | 27.73 0.00 | ± | 163.75 |
| CrossE_L + PMI_L | SAGE | 38.62 2.52 | ± | 43.36 3.32 | ± | 51.72 1.62 | ± | 45.00 1.59 | ± | 81.25 |
| CrossE_L + PMI_L + PR_L | ALL | 38.50 1.59 | ± | 28.27 0.11 | ± | 55.41 5.63 | ± | 27.92 0.20 | ± | 146.625 |
| CrossE_L + PMI_L + PR_L | GAT | 56.97 1.78 | ± | 41.14 0.58 | ± | 69.30 8.24 | ± | 37.90 0.80 | ± | 33.0 |
| CrossE_L + PMI_L + PR_L | GCN | 48.98 2.64 | ± | 35.98 1.85 | ± | 67.28 2.14 | ± | 37.46 1.90 | ± | 64.75 |
| CrossE_L + PMI_L + PR_L | GIN | 43.65 2.20 | ± | 28.02 0.51 | ± | 46.74 14.38 | ± | 28.77 0.74 | ± | 145.75 |
| CrossE_L + PMI_L + PR_L | MPNN | 54.65 2.44 | ± | 41.72 1.52 | ± | 71.92 1.34 | ± | 41.49 1.21 | ± | 21.125 |
| CrossE_L + PMI_L + PR_L | PAGNN | 45.75 2.32 | ± | 27.73 0.00 | ± | 47.97 4.14 | ± | 27.73 0.00 | ± | 158.75 |
| CrossE_L + PMI_L + PR_L | SAGE | 40.69 1.55 | ± | 44.70 1.02 | ± | 50.97 2.49 | ± | 43.45 2.27 | ± | 79.25 |
| CrossE_L + PMI_L + PR_L + Triplet_L | ALL | 44.77 1.50 | ± | 27.91 0.27 | ± | 52.29 7.11 | ± | 28.09 0.26 | ± | 138.0 |
| CrossE_L + PMI_L + PR_L + Triplet_L | GAT | 56.44 2.71 | ± | 41.36 1.14 | ± | 72.74 2.13 | ± | 40.50 1.14 | ± | 20.875 |
| CrossE_L + PMI_L + PR_L + Triplet_L | GCN | 48.20 1.43 | ± | 37.47 0.85 | ± | 65.21 3.53 | ± | 36.99 0.25 | ± | 69.75 |
| CrossE_L + PMI_L + PR_L + Triplet_L | GIN | 44.54 1.12 | ± | 28.40 0.45 | ± | 60.03 6.23 | ± | 28.81 0.81 | ± | 123.875 |

Table 23. Results for Node Cls F1 (↑) (continued)

| Loss Type | Model | Cora ↓ Citeseer | | Cora ↓ Bitcoin | | Citeseer ↓ Cora | | Citeseer ↓ Bitcoin | | Average Rank |
|---|---|---|---|---|---|---|---|---|---|---|
| CrossE_L + PMI_L + PR_L + Triplet_L | MPNN | 54.93 ± 1.67 | | 41.26 ± 1.32 | | 72.72 ± 1.31 | | 39.50 ± 1.66 | | 27.25 |
| CrossE_L + PMI_L + PR_L + Triplet_L | PAGNN | 45.82 ± 0.94 | | 27.73 ± 0.00 | | 49.40 ± 4.29 | | 27.73 ± 0.00 | | 155.25 |
| CrossE_L + PMI_L + PR_L + Triplet_L | SAGE | 39.95 ± 2.22 | | 43.11 ± 1.31 | | 49.16 ± 2.63 | | 44.57 ± 2.64 | | 84.875 |
| CrossE_L + PMI_L + Triplet_L | ALL | 46.59 ± 1.66 | | 33.28 ± 1.96 | | 64.00 ± 4.13 | | 35.98 ± 1.66 | | 88.625 |
| CrossE_L + PMI_L + Triplet_L | GAT | 55.80 ± 3.11 | | 41.29 ± 0.77 | | 74.00 ± 0.84 | | 41.48 ± 0.95 | | 17.25 |
| CrossE_L + PMI_L + Triplet_L | GCN | 48.15 ± 0.83 | | 35.39 ± 1.57 | | 65.94 ± 1.11 | | 36.70 ± 1.75 | | 76.125 |
| CrossE_L + PMI_L + Triplet_L | GIN | 44.78 ± 2.00 | | 28.99 ± 0.89 | | 61.98 ± 1.98 | | 29.06 ± 0.69 | | 114.625 |
| CrossE_L + PMI_L + Triplet_L | MPNN | 53.82 ± 1.64 | | 41.34 ± 1.66 | | 74.20 ± 0.84 | | 38.99 ± 1.01 | | 26.875 |
| CrossE_L + PMI_L + Triplet_L | PAGNN | 44.73 ± 2.08 | | 27.73 ± 0.00 | | 48.04 ± 7.29 | | 27.73 ± 0.00 | | 163.75 |
| CrossE_L + PMI_L + Triplet_L | SAGE | 41.33 ± 1.46 | | 42.23 ± 2.78 | | 53.66 ± 2.82 | | 42.79 ± 1.94 | | 77.0 |
| CrossE_L + PR_L | ALL | 18.56 ± 3.39 | | 27.73 ± 0.00 | | 23.26 ± 5.14 | | 27.73 ± 0.00 | | 193.5 |
| CrossE_L + PR_L | GAT | 36.39 ± 9.08 | | 28.96 ± 0.67 | | 11.76 ± 2.51 | | 29.06 ± 0.98 | | 159.5 |
| CrossE_L + PR_L | GCN | 43.38 ± 4.53 | | 32.59 ± 3.60 | | 39.09 ± 4.59 | | 32.59 ± 1.74 | | 134.75 |
| CrossE_L + PR_L | GIN | 32.31 ± 4.97 | | 27.73 ± 0.00 | | 34.30 ± 13.22 | | 27.73 ± 0.00 | | 190.25 |
| CrossE_L + PR_L | MPNN | 42.02 ± 4.81 | | 27.73 ± 0.00 | | 46.22 ± 11.17 | | 27.82 ± 0.21 | | 165.375 |
| CrossE_L + PR_L | PAGNN | 32.99 ± 4.16 | | 27.73 ± 0.00 | | 40.29 ± 7.59 | | 27.73 ± 0.00 | | 189.0 |

Table 23. Results for Node Cls F1 (↑) (continued)

| Loss Type | Model | Cora ↓ Citeseer | | Cora ↓ Bitcoin | | Citeseer ↓ Cora | | Citeseer ↓ Bitcoin | | Average Rank |
|---|---|---|---|---|---|---|---|---|---|---|
| CrossE_L + PR_L | SAGE | 32.80 4.51 | ± | 34.09 4.18 | ± | 24.95 5.90 | ± | 35.24 5.10 | ± | 146.0 |
| CrossE_L + PR_L + Triplet_L | ALL | 35.11 5.87 | ± | 27.88 0.33 | ± | 31.21 8.21 | ± | 27.78 0.11 | ± | 174.375 |
| CrossE_L + PR_L + Triplet_L | GAT | 51.47 1.49 | ± | 40.96 0.93 | ± | 67.24 4.29 | ± | 40.46 0.97 | ± | 44.0 |
| CrossE_L + PR_L + Triplet_L | GCN | 48.50 3.07 | ± | 37.90 1.19 | ± | 63.95 4.12 | ± | 36.22 2.37 | ± | 72.5 |
| CrossE_L + PR_L + Triplet_L | GIN | 43.79 4.12 | ± | 28.62 1.13 | ± | 59.85 2.99 | ± | 28.16 0.17 | ± | 127.375 |
| CrossE_L + PR_L + Triplet_L | MPNN | 48.16 1.82 | ± | 32.75 2.47 | ± | 67.16 1.07 | ± | 34.94 2.24 | ± | 82.25 |
| CrossE_L + PR_L + Triplet_L | PAGNN | 37.76 3.61 | ± | 27.73 0.00 | ± | 51.95 7.39 | ± | 27.73 0.00 | ± | 172.0 |
| CrossE_L + PR_L + Triplet_L | SAGE | 41.00 1.58 | ± | 35.08 2.73 | ± | 48.86 1.50 | ± | 32.07 2.60 | ± | 130.25 |
| CrossE_L + Triplet_L | ALL | 43.30 3.64 | ± | 28.56 0.91 | ± | 61.68 4.29 | ± | 29.15 1.06 | ± | 123.5 |
| CrossE_L + Triplet_L | GAT | 55.21 2.04 | ± | 41.79 0.46 | ± | 75.24 1.06 | ± | 41.76 0.32 | ± | 11.0 |
| CrossE_L + Triplet_L | GCN | 48.25 1.17 | ± | 36.63 0.72 | ± | 68.14 2.75 | ± | 36.27 1.90 | ± | 68.25 |
| CrossE_L + Triplet_L | GIN | 46.84 2.84 | ± | 29.60 1.42 | ± | 67.64 1.66 | ± | 31.27 0.46 | ± | 92.0 |
| CrossE_L + Triplet_L | MPNN | 52.94 2.46 | ± | 38.78 1.22 | ± | 73.31 0.49 | ± | 36.92 0.80 | ± | 45.5 |
| CrossE_L + Triplet_L | PAGNN | 47.22 1.79 | ± | 27.73 0.00 | ± | 66.33 2.64 | ± | 27.72 0.01 | ± | 134.875 |
| CrossE_L + Triplet_L | SAGE | 46.85 2.91 | ± | 35.24 3.70 | ± | 59.80 1.04 | ± | 34.42 1.38 | ± | 94.75 |
| PMI_L | ALL | 48.57 1.64 | ± | 28.25 0.20 | ± | 62.58 4.43 | ± | 32.16 2.37 | ± | 100.625 |

Continued on next page

Table 23. Results for Node Cls F1 (↑) (continued)

| Loss Type | Model | Cora ↓ Citeseer | | Cora ↓ Bitcoin | | Citeseer ↓ Cora | | Citeseer ↓ Bitcoin | | Average Rank |
|---|---|---|---|---|---|---|---|---|---|---|
| PMI_L | GAT | 55.82 ± 2.24 | | 41.65 ± 0.74 | | 73.12 ± 2.28 | | 41.59 ± 0.50 | | 15.25 |
| PMI_L | GCN | 48.15 ± 1.25 | | 36.85 ± 1.22 | | 64.12 ± 2.63 | | 36.01 ± 1.28 | | 77.375 |
| PMI_L | GIN | 44.70 ± 2.43 | | 28.27 ± 0.51 | | 57.75 ± 6.16 | | 28.78 ± 0.59 | | 126.875 |
| PMI_L | MPNN | 53.69 ± 1.10 | | 40.81 ± 1.03 | | 72.23 ± 1.42 | | 40.78 ± 1.67 | | 31.625 |
| PMI_L | PAGNN | 45.73 ± 0.36 | | 27.73 ± 0.00 | | 41.61 ± 3.83 | | 27.73 ± 0.00 | | 163.25 |
| PMI_L | SAGE | 37.75 ± 2.97 | | 45.55 ± 2.64 | | 45.25 ± 5.66 | | 43.80 ± 3.31 | | 89.75 |
| PMI_L + PR_L | ALL | 40.66 ± 4.23 | | 28.22 ± 0.00 | | 50.29 ± 10.38 | | 27.76 ± 0.11 | | 155.0 |
| PMI_L + PR_L | GAT | 54.62 ± 3.07 | | 41.36 ± 0.77 | | 70.64 ± 4.59 | | 39.42 ± 1.49 | | 30.625 |
| PMI_L + PR_L | GCN | 48.41 ± 1.86 | | 36.49 ± 0.91 | | 67.29 ± 1.87 | | 36.55 ± 1.17 | | 68.75 |
| PMI_L + PR_L | GIN | 44.17 ± 0.86 | | 28.63 ± 0.87 | | 50.35 ± 9.73 | | 28.84 ± 0.59 | | 133.25 |
| PMI_L + PR_L | MPNN | 52.51 ± 1.94 | | 40.51 ± 1.26 | | 69.13 ± 2.47 | | 31.12 ± 1.62 | | 62.25 |
| PMI_L + PR_L | PAGNN | 45.65 ± 1.15 | | 27.73 ± 0.00 | | 52.79 ± 6.20 | | 27.73 ± 0.00 | | 151.5 |
| PMI_L + PR_L | SAGE | 41.76 ± 2.03 | | 43.58 ± 3.86 | | 49.16 ± 2.64 | | 45.95 ± 3.01 | | 78.875 |
| PMI_L + PR_L + Triplet_L | ALL | 41.80 ± 3.40 | | 28.01 ± 0.30 | | 54.62 ± 2.35 | | 27.83 ± 0.20 | | 145.25 |
| PMI_L + PR_L + Triplet_L | GAT | 54.76 ± 2.78 | | 40.65 ± 1.31 | | 72.72 ± 1.39 | | 41.36 ± 0.95 | | 27.0 |
| PMI_L + PR_L + Triplet_L | GCN | 48.26 ± 2.11 | | 35.60 ± 1.08 | | 65.29 ± 2.16 | | 37.24 ± 1.43 | | 73.375 |

Table 23. Results for Node Cls F1 (↑) (continued)

| Loss Type | Model | Cora ↓ Citeseer | | Cora ↓ Bitcoin | | Citeseer ↓ Cora | | Citeseer ↓ Bitcoin | | Average Rank |
|---|---|---|---|---|---|---|---|---|---|---|
| PMI_L + PR_L + Triplet_L | GIN | 46.15 ± 1.11 | | 29.04 ± 0.84 | | 59.06 ± 4.17 | | 28.27 ± 0.23 | | 116.0 |
| PMI_L + PR_L + Triplet_L | MPNN | 54.05 ± 2.03 | | 40.27 ± 0.64 | | 71.30 ± 2.06 | | 32.21 ± 1.15 | | 54.125 |
| PMI_L + PR_L + Triplet_L | PAGNN | 44.38 ± 0.98 | | 27.73 ± 0.00 | | 55.91 ± 6.10 | | 27.73 ± 0.00 | | 154.0 |
| PMI_L + PR_L + Triplet_L | SAGE | 42.15 ± 2.31 | | 44.84 ± 1.93 | | 52.29 ± 5.29 | | 41.67 ± 2.63 | | 74.875 |
| PMI_L + Triplet_L | ALL | 47.81 ± 2.46 | | 28.72 ± 0.61 | | 66.53 ± 2.90 | | 36.36 ± 1.26 | | 85.375 |
| PMI_L + Triplet_L | GAT | 56.07 ± 1.98 | | 40.72 ± 0.63 | | 73.14 ± 3.85 | | 40.93 ± 1.48 | | 22.75 |
| PMI_L + Triplet_L | GCN | 47.59 ± 1.03 | | 35.70 ± 1.45 | | 65.68 ± 1.66 | | 35.80 ± 2.08 | | 81.125 |
| PMI_L + Triplet_L | GIN | 45.31 ± 2.19 | | 29.27 ± 0.48 | | 63.10 ± 4.95 | | 29.40 ± 0.79 | | 109.25 |
| PMI_L + Triplet_L | MPNN | 53.35 ± 2.41 | | 40.99 ± 1.38 | | 74.72 ± 1.58 | | 40.73 ± 1.51 | | 26.625 |
| PMI_L + Triplet_L | PAGNN | 44.06 ± 1.60 | | 27.73 ± 0.00 | | 47.26 ± 5.70 | | 27.73 ± 0.00 | | 166.25 |
| PMI_L + Triplet_L | SAGE | 42.04 ± 1.68 | | **45.55 ± 1.80** | | 52.11 ± 3.03 | | 44.38 ± 2.01 | | 72.625 |
| PR_L | ALL | 15.38 ± 1.56 | | 27.73 ± 0.00 | | 17.84 ± 2.92 | | 27.73 ± 0.00 | | 194.25 |
| PR_L | GAT | 37.67 ± 2.43 | | 28.87 ± 1.09 | | 11.02 ± 3.41 | | 29.86 ± 2.09 | | 156.5 |
| PR_L | GCN | 41.29 ± 3.33 | | 31.23 ± 0.71 | | 45.30 ± 4.36 | | 31.23 ± 1.47 | | 138.0 |
| PR_L | GIN | 27.74 ± 4.37 | | 27.73 ± 0.00 | | 28.08 ± 9.68 | | 27.78 ± 0.11 | | 185.0 |
| PR_L | MPNN | 41.63 ± 4.55 | | 27.73 ± 0.00 | | 45.51 ± 7.04 | | 27.73 ± 0.00 | | 174.0 |

Table 23. Results for Node Cls F1 (↑) (continued)

| Loss Type | Model | Cora ↓ Citeseer | | Cora ↓ Bitcoin | | Citeseer ↓ Cora | | Citeseer ↓ Bitcoin | | Average Rank |
|---|---|---|---|---|---|---|---|---|---|---|
| PR_L | PAGNN | 29.95 | ± | 27.73 | ± | 42.99 | ± | 27.73 | ± | 187.25 |
| | | 5.59 | | 0.00 | | 5.01 | | 0.00 | | |
| PR_L | SAGE | 33.95 | ± | 33.41 | ± | 16.84 | ± | 38.96 | ± | 135.25 |
| | | 4.58 | | 2.68 | | 6.40 | | 1.51 | | |
| PR_L + Triplet_L | ALL | 21.77 | ± | 27.73 | ± | 27.24 | ± | 27.73 | ± | 192.5 |
| | | 3.01 | | 0.00 | | 3.28 | | 0.00 | | |
| PR_L + Triplet_L | GAT | 48.93 | ± | 39.10 | ± | 42.25 | ± | 36.36 | ± | 94.0 |
| | | 3.66 | | 1.74 | | 6.70 | | 0.72 | | |
| PR_L + Triplet_L | GCN | 46.39 | ± | 35.76 | ± | 58.93 | ± | 34.64 | ± | 94.75 |
| | | 2.09 | | 2.21 | | 6.14 | | 1.14 | | |
| PR_L + Triplet_L | GIN | 39.72 | ± | 27.77 | ± | 54.81 | ± | 28.49 | ± | 149.0 |
| | | 4.29 | | 0.09 | | 3.01 | | 1.02 | | |
| PR_L + Triplet_L | MPNN | 45.34 | ± | 28.12 | ± | 46.49 | ± | 29.18 | ± | 136.125 |
| | | 2.18 | | 0.86 | | 7.17 | | 2.14 | | |
| PR_L + Triplet_L | PAGNN | 34.51 | ± | 27.73 | ± | 40.76 | ± | 27.73 | ± | 187.25 |
| | | 2.65 | | 0.00 | | 4.06 | | 0.00 | | |
| PR_L + Triplet_L | SAGE | 37.10 | ± | 33.75 | ± | 33.83 | ± | 36.19 | ± | 138.375 |
| | | 1.35 | | 2.46 | | 12.38 | | 3.93 | | |
| Triplet_L | ALL | 45.77 | ± | 28.69 | ± | 65.56 | ± | 36.07 | ± | 94.625 |
| | | 2.38 | | 0.46 | | 4.43 | | 2.22 | | |
| Triplet_L | GAT | 55.98 | ± | 41.28 | ± | **75.77** | ± | 41.42 | ± | 14.5 |
| | | 1.52 | | 1.31 | | **2.12** | | 0.63 | | |
| Triplet_L | GCN | 50.62 | ± | 36.64 | ± | 68.42 | ± | 37.02 | ± | 58.75 |
| | | 3.25 | | 1.08 | | 3.22 | | 1.08 | | |
| Triplet_L | GIN | 48.03 | ± | 31.08 | ± | 67.20 | ± | 31.86 | ± | 89.0 |
| | | 1.47 | | 1.21 | | 1.51 | | 1.23 | | |
| Triplet_L | MPNN | 53.77 | ± | 39.58 | ± | 74.42 | ± | 39.31 | ± | 33.5 |
| | | 1.68 | | 2.60 | | 2.53 | | 1.70 | | |
| Triplet_L | PAGNN | 49.20 | ± | 27.73 | ± | 67.95 | ± | 27.73 | ± | 119.75 |
| | | 2.07 | | 0.00 | | 3.17 | | 0.00 | | |
| Triplet_L | SAGE | 45.69 | ± | 40.92 | ± | 58.35 | ± | 40.75 | ± | 69.5 |
| | | 1.67 | | 1.74 | | 3.51 | | 1.81 | | |

Table 24. Node Cls Precision Performance (↑): This table presents models (Loss function and GNN) ranked by their average performance in terms of node cls precision. Top-ranked results are highlighted in <span style="color:red">red</span>, second-ranked in <span style="color:blue">blue</span>, and third-ranked in <span style="color:green">green</span>.

| Loss Type | Model | Cora ↓ Citeseer | | Cora ↓ Bitcoin | | Citeseer ↓ Cora | | Citeseer ↓ Bitcoin | | Average Rank |
|---|---|---|---|---|---|---|---|---|---|---|
| Contr_l | ALL | 40.39 ± 10.18 | | 23.73 ± 0.00 | | 61.57 ± 4.64 | | 28.74 ± 7.47 | | 169.625 |
| Contr_l | GAT | 58.04 ± 6.60 | | 48.19 ± 1.38 | | 77.62 ± 1.52 | | 47.87 ± 1.18 | | 41.375 |
| Contr_l | GCN | 53.18 ± 3.99 | | 45.57 ± 1.88 | | 72.71 ± 1.99 | | 46.57 ± 1.61 | | 78.625 |
| Contr_l | GIN | 48.52 ± 8.82 | | 42.82 ± 3.45 | | 72.71 ± 4.08 | | 42.48 ± 2.40 | | 100.25 |
| Contr_l | MPNN | 53.17 ± 3.86 | | 48.66 ± 1.61 | | 73.11 ± 1.89 | | 48.21 ± 0.70 | | 54.25 |
| Contr_l | PAGNN | 53.00 ± 7.96 | | 23.73 ± 0.00 | | 67.80 ± 6.57 | | 23.73 ± 0.00 | | 138.375 |
| Contr_l | SAGE | 39.62 ± 2.13 | | 51.08 ± 2.08 | | 61.06 ± 6.12 | | 47.58 ± 3.50 | | 108.75 |
| Contr_l + CrossE_L | ALL | 39.98 ± 2.18 | | 34.74 ± 12.67 | | 52.49 ± 11.93 | | 25.95 ± 4.97 | | 170.75 |
| Contr_l + CrossE_L | GAT | 58.43 ± 2.04 | | 48.78 ± 1.43 | | 74.15 ± 3.06 | | 47.49 ± 1.33 | | 46.125 |
| Contr_l + CrossE_L | GCN | 55.07 ± 5.57 | | 46.07 ± 1.52 | | 72.30 ± 2.99 | | 45.18 ± 2.20 | | 79.875 |
| Contr_l + CrossE_L | GIN | 47.76 ± 8.54 | | 39.90 ± 6.75 | | 70.06 ± 3.00 | | 41.00 ± 5.02 | | 114.5 |
| Contr_l + CrossE_L | MPNN | 58.28 ± 7.75 | | 48.20 ± 1.54 | | 73.74 ± 0.89 | | 47.34 ± 0.85 | | 53.75 |
| Contr_l + CrossE_L | PAGNN | 46.13 ± 1.73 | | 23.73 ± 0.00 | | 67.19 ± 6.97 | | 30.40 ± 14.92 | | 151.125 |
| Contr_l + CrossE_L | SAGE | 43.91 ± 3.87 | | 48.02 ± 3.20 | | 61.15 ± 4.57 | | 44.20 ± 13.25 | | 127.125 |
| Contr_l + CrossE_L + PMI_L | ALL | 48.04 ± 6.79 | | 27.06 ± 7.46 | | 73.58 ± 1.54 | | 42.54 ± 6.69 | | 108.25 |

Continued on next page

Table 24. Results for Node Cls Precision (↑) (continued)

| Loss Type | Model | Cora ↓ Citeseer | | Cora ↓ Bitcoin | | Citeseer ↓ Cora | | Citeseer ↓ Bitcoin | Average Rank |
|---|---|---|---|---|---|---|---|---|---|
| Contr_l + CrossE_L + PMI_L | GAT | 58.31 2.31 | ± | 48.42 0.87 | ± | 76.75 1.33 | ± | 47.83 1.21 | 39.75 |
| Contr_l + CrossE_L + PMI_L | GCN | 50.05 2.18 | ± | 44.47 1.73 | ± | 67.49 1.63 | ± | 46.09 1.85 | 104.25 |
| Contr_l + CrossE_L + PMI_L | GIN | 46.90 7.50 | ± | 39.70 4.86 | ± | 66.08 4.99 | ± | 39.39 4.43 | 130.25 |
| Contr_l + CrossE_L + PMI_L | MPNN | 59.10 4.44 | ± | 47.98 1.01 | ± | 75.84 1.56 | ± | 49.02 1.47 | 39.0 |
| Contr_l + CrossE_L + PMI_L | PAGNN | 46.35 2.02 | ± | 23.73 0.00 | ± | 67.35 5.32 | ± | 23.73 0.00 | 156.625 |
| Contr_l + CrossE_L + PMI_L | SAGE | 40.12 3.13 | ± | 50.87 2.11 | ± | 56.02 3.00 | ± | 50.73 1.53 | 102.75 |
| Contr_l + CrossE_L + PMI_L + PR_L | ALL | 42.93 3.00 | ± | 57.11 0.00 | ± | 74.39 3.11 | ± | 26.70 6.65 | 93.375 |
| Contr_l + CrossE_L + PMI_L + PR_L | GAT | 59.16 3.14 | ± | 48.00 1.33 | ± | 78.36 2.10 | ± | 47.82 0.83 | 39.25 |
| Contr_l + CrossE_L + PMI_L + PR_L | GCN | 54.02 3.50 | ± | 45.51 2.51 | ± | 70.37 3.58 | ± | 45.83 1.83 | 87.375 |
| Contr_l + CrossE_L + PMI_L + PR_L | GIN | 47.69 9.04 | ± | 38.77 9.96 | ± | 63.22 5.58 | ± | 30.41 9.15 | 135.875 |
| Contr_l + CrossE_L + PMI_L + PR_L | MPNN | 56.15 3.93 | ± | 49.09 1.78 | ± | 76.57 1.36 | ± | 47.40 2.66 | 42.75 |
| Contr_l + CrossE_L + PMI_L + PR_L | PAGNN | 47.19 1.88 | ± | 23.73 0.00 | ± | 73.33 3.21 | ± | 23.73 0.00 | 135.875 |
| Contr_l + CrossE_L + PMI_L + PR_L | SAGE | 40.85 2.15 | ± | 51.78 1.34 | ± | 55.67 2.80 | ± | 50.96 1.24 | 98.0 |
| Contr_l + CrossE_L + PMI_L + PR_L + Triplet_L | ALL | 45.71 1.94 | ± | 37.09 18.29 | ± | 67.82 11.24 | ± | 41.78 6.75 | 130.875 |
| Contr_l + CrossE_L + PMI_L + PR_L + Triplet_L | GAT | 58.18 4.16 | ± | 48.41 0.53 | ± | 77.17 1.50 | ± | 47.94 0.79 | 38.125 |

Table 24. Results for Node Cls Precision (↑) (continued)

| Loss Type | Model | Cora ↓ Citeseer | | Cora ↓ Bitcoin | | Citeseer ↓ Cora | | Citeseer ↓ Bitcoin | | Average Rank |
|---|---|---|---|---|---|---|---|---|---|---|
| Contr_l + CrossE_L + PMI_L + PR_L + Triplet_L | GCN | 52.50 4.36 | ± | 46.17 0.90 | ± | 69.30 3.18 | ± | 45.00 1.17 | ± | 94.5 |
| Contr_l + CrossE_L + PMI_L + PR_L + Triplet_L | GIN | 47.03 8.50 | ± | 42.45 9.58 | ± | 68.70 1.38 | ± | 38.86 5.98 | ± | 124.0 |
| Contr_l + CrossE_L + PMI_L + PR_L + Triplet_L | MPNN | 56.53 2.93 | ± | 50.10 1.06 | ± | 75.82 1.56 | ± | 47.07 2.58 | ± | 43.875 |
| Contr_l + CrossE_L + PMI_L + PR_L + Triplet_L | PAGNN | 47.16 2.19 | ± | 23.73 0.00 | ± | 71.80 4.58 | ± | 23.73 0.00 | ± | 139.875 |
| Contr_l + CrossE_L + PMI_L + PR_L + Triplet_L | SAGE | 42.15 2.36 | ± | 51.68 0.61 | ± | 59.39 4.39 | ± | 52.05 1.91 | ± | 89.75 |
| Contr_l + CrossE_L + PMI_L + Triplet_L | ALL | 46.16 1.98 | ± | 47.86 5.73 | ± | 70.67 2.55 | ± | 48.37 2.35 | ± | 87.0 |
| Contr_l + CrossE_L + PMI_L + Triplet_L | GAT | 58.78 3.73 | ± | 48.54 0.81 | ± | 78.42 1.47 | ± | 48.35 1.23 | ± | 30.5 |
| Contr_l + CrossE_L + PMI_L + Triplet_L | GCN | 51.18 5.00 | ± | 46.36 0.45 | ± | 67.96 1.92 | ± | 45.96 2.33 | ± | 95.625 |
| Contr_l + CrossE_L + PMI_L + Triplet_L | GIN | 46.19 2.87 | ± | 39.53 10.34 | ± | 70.19 3.26 | ± | 37.73 5.49 | ± | 125.25 |
| Contr_l + CrossE_L + PMI_L + Triplet_L | MPNN | 56.41 2.83 | ± | 49.02 0.95 | ± | 74.87 2.21 | ± | 48.08 0.89 | ± | 44.25 |
| Contr_l + CrossE_L + PMI_L + Triplet_L | PAGNN | 47.89 1.75 | ± | 23.73 0.00 | ± | 76.60 2.68 | ± | 23.73 0.00 | ± | 123.375 |
| Contr_l + CrossE_L + PMI_L + Triplet_L | SAGE | 42.20 5.70 | ± | 51.30 1.74 | ± | 58.30 2.16 | ± | 51.45 0.81 | ± | 91.625 |
| Contr_l + CrossE_L + PR_L | ALL | 23.04 3.45 | ± | 23.73 0.00 | ± | 44.40 18.19 | ± | 23.73 0.00 | ± | 193.125 |
| Contr_l + CrossE_L + PR_L | GAT | 48.30 2.13 | ± | 48.45 1.03 | ± | 65.02 8.83 | ± | 51.24 0.53 | ± | 77.625 |

Continued on next page

                                                                

Table 24.  Results for Node Cls Precision (↑) (continued)

| Loss Type | Model | Cora ↓ Citeseer | | Cora ↓ Bitcoin | | Citeseer ↓ Cora | | Citeseer ↓ Bitcoin | | Average Rank |
|---|---|---|---|---|---|---|---|---|---|---|
| Contr_l + CrossE_L + PR_L | GCN | 51.79 9.15 | ± | 45.40 1.86 | ± | 70.05 5.42 | ± | 46.96 1.76 | ± | 88.5 |
| Contr_l + CrossE_L + PR_L | GIN | 36.69 3.10 | ± | 30.42 14.95 | ± | 61.50 8.92 | ± | 33.74 9.14 | ± | 163.75 |
| Contr_l + CrossE_L + PR_L | MPNN | 45.98 2.53 | ± | 41.58 16.93 | ± | 73.53 5.52 | ± | 49.39 14.54 | ± | 86.5 |
| Contr_l + CrossE_L + PR_L | PAGNN | 43.95 3.51 | ± | 23.73 0.00 | ± | 56.53 5.71 | ± | 23.73 0.00 | ± | 176.125 |
| Contr_l + CrossE_L + PR_L | SAGE | 36.57 3.20 | ± | 51.13 4.92 | ± | 52.07 10.15 | ± | 52.62 1.93 | ± | 103.0 |
| Contr_l + CrossE_L + PR_L + Triplet_L | ALL | 37.88 5.62 | ± | 23.73 0.00 | ± | 53.76 12.72 | ± | 30.40 14.92 | ± | 180.125 |
| Contr_l + CrossE_L + PR_L + Triplet_L | GAT | 57.53 7.80 | ± | 49.23 0.86 | ± | 76.29 1.67 | ± | 48.12 0.78 | ± | 37.0 |
| Contr_l + CrossE_L + PR_L + Triplet_L | GCN | 49.82 5.58 | ± | 46.09 2.14 | ± | 70.29 1.74 | ± | 45.96 1.23 | ± | 92.875 |
| Contr_l + CrossE_L + PR_L + Triplet_L | GIN | 47.39 8.17 | ± | 37.08 14.51 | ± | 67.44 4.37 | ± | 40.79 2.16 | ± | 127.75 |
| Contr_l + CrossE_L + PR_L + Triplet_L | MPNN | 59.06 9.02 | ± | 50.42 1.51 | ± | 72.30 2.83 | ± | 46.24 5.08 | ± | 48.875 |
| Contr_l + CrossE_L + PR_L + Triplet_L | PAGNN | 44.39 1.77 | ± | 23.73 0.00 | ± | 63.22 3.58 | ± | 23.73 0.00 | ± | 166.0 |
| Contr_l + CrossE_L + PR_L + Triplet_L | SAGE | 36.48 2.66 | ± | 49.29 2.42 | ± | 58.01 4.19 | ± | 51.45 1.17 | ± | 104.375 |
| Contr_l + CrossE_L + Triplet_L | ALL | 40.78 5.53 | ± | 30.41 10.33 | ± | 59.11 3.68 | ± | 34.22 8.81 | ± | 162.625 |
| Contr_l + CrossE_L + Triplet_L | GAT | 58.68 5.62 | ± | 48.52 0.93 | ± | 75.24 2.05 | ± | 47.82 0.84 | ± | 43.375 |
| Contr_l + CrossE_L + Triplet_L | GCN | 53.69 4.76 | ± | 45.03 1.79 | ± | 70.95 1.71 | ± | 46.79 1.50 | ± | 83.0 |
| Contr_l + CrossE_L + Triplet_L | GIN | 48.44 6.44 | ± | 42.82 2.93 | ± | 71.21 1.44 | ± | 40.76 2.04 | ± | 106.875 |

Table 24. Results for Node Cls Precision (↑) (continued)

| Loss Type | Model | Cora ↓ Citeseer | | Cora ↓ Bitcoin | | Citeseer ↓ Cora | | Citeseer ↓ Bitcoin | | Average Rank |
|---|---|---|---|---|---|---|---|---|---|---|
| Contr_l + CrossE_L + Triplet_L | MPNN | 54.39 5.66 | ± | 48.26 1.79 | ± | 73.61 2.56 | ± | 49.39 0.86 | ± | 49.25 |
| Contr_l + CrossE_L + Triplet_L | PAGNN | 47.42 1.73 | ± | 23.73 0.00 | ± | 71.14 1.89 | ± | 27.07 10.55 | ± | 134.375 |
| Contr_l + CrossE_L + Triplet_L | SAGE | 46.76 6.78 | ± | 49.58 2.34 | ± | 63.42 2.97 | ± | 49.50 3.04 | ± | 82.5 |
| Contr_l + PMI_L | ALL | 44.85 2.68 | ± | 30.40 14.92 | ± | 72.66 2.83 | ± | 53.41 5.61 | ± | 90.625 |
| Contr_l + PMI_L | GAT | 57.61 3.66 | ± | 48.51 0.71 | ± | 76.11 3.09 | ± | 47.10 0.68 | ± | 47.0 |
| Contr_l + PMI_L | GCN | 53.70 8.14 | ± | 45.77 1.74 | ± | 69.83 2.47 | ± | 46.45 2.09 | ± | 86.25 |
| Contr_l + PMI_L | GIN | 49.51 8.14 | ± | 42.88 1.45 | ± | 64.97 3.99 | ± | 35.15 7.41 | ± | 124.25 |
| Contr_l + PMI_L | MPNN | 56.60 2.98 | ± | 48.84 1.56 | ± | 75.66 1.63 | ± | 47.46 1.32 | ± | 46.5 |
| Contr_l + PMI_L | PAGNN | 47.96 1.54 | ± | 23.73 0.00 | ± | 62.98 12.52 | ± | 23.73 0.00 | ± | 155.5 |
| Contr_l + PMI_L | SAGE | 39.03 1.45 | ± | 52.36 1.88 | ± | 57.93 2.65 | ± | 49.67 2.62 | ± | 99.625 |
| Contr_l + PMI_L + PR_L | ALL | 45.12 1.85 | ± | 53.79 4.55 | ± | 72.10 5.97 | ± | 35.40 13.97 | ± | 89.875 |
| Contr_l + PMI_L + PR_L | GAT | 61.04 1.52 | ± | 48.31 0.94 | ± | 76.83 2.15 | ± | 48.68 0.58 | ± | 31.375 |
| Contr_l + PMI_L + PR_L | GCN | 53.96 5.85 | ± | 46.48 1.41 | ± | 70.11 3.83 | ± | 45.66 2.06 | ± | 85.75 |
| Contr_l + PMI_L + PR_L | GIN | 43.44 0.99 | ± | 34.41 15.37 | ± | 63.37 5.82 | ± | 34.85 15.73 | ± | 149.25 |
| Contr_l + PMI_L + PR_L | MPNN | 56.93 4.46 | ± | 49.05 0.78 | ± | 74.66 0.25 | ± | 46.15 6.44 | ± | 53.125 |
| Contr_l + PMI_L + PR_L | PAGNN | 46.53 0.57 | ± | 23.73 0.00 | ± | 63.07 11.57 | ± | 23.73 0.00 | ± | 161.375 |

Table 24. Results for Node Cls Precision (↑) (continued)

| Loss Type | Model | Cora ↓ Citeseer | | Cora ↓ Bitcoin | | Citeseer ↓ Cora | | Citeseer ↓ Bitcoin | | Average Rank |
|---|---|---|---|---|---|---|---|---|---|---|
| Contr_l + PMI_L + PR_L | SAGE | 42.46 | ± 2.13 | 51.23 | ± 1.98 | 56.73 | ± 1.26 | 48.62 | ± 2.77 | 101.25 |
| Contr_l + PMI_L + PR_L + Triplet_L | ALL | 43.92 | ± 2.32 | 30.40 | ± 9.96 | 55.77 | ± 3.37 | 46.61 | ± 15.22 | 144.875 |
| Contr_l + PMI_L + PR_L + Triplet_L | GAT | 59.59 | ± 5.85 | 48.22 | ± 0.93 | 74.90 | ± 1.47 | 48.48 | ± 0.96 | 40.75 |
| Contr_l + PMI_L + PR_L + Triplet_L | GCN | 56.91 | ± 6.41 | 44.96 | ± 1.24 | 70.86 | ± 2.39 | 46.84 | ± 1.12 | 78.0 |
| Contr_l + PMI_L + PR_L + Triplet_L | GIN | 44.26 | ± 1.15 | 35.23 | ± 7.28 | 69.22 | ± 2.80 | 39.82 | ± 3.71 | 133.875 |
| Contr_l + PMI_L + PR_L + Triplet_L | MPNN | 54.99 | ± 4.12 | 48.10 | ± 1.32 | 76.50 | ± 3.65 | 45.67 | ± 7.79 | 60.5 |
| Contr_l + PMI_L + PR_L + Triplet_L | PAGNN | 46.68 | ± 0.37 | 23.73 | ± 0.00 | 71.87 | ± 2.45 | 23.73 | ± 0.00 | 141.25 |
| Contr_l + PMI_L + PR_L + Triplet_L | SAGE | 44.29 | ± 7.08 | 52.28 | ± 2.67 | 59.23 | ± 3.04 | 49.79 | ± 1.82 | 89.0 |
| Contr_l + PR_L | ALL | 26.22 | ± 6.90 | 23.73 | ± 0.00 | 30.67 | ± 16.17 | 23.73 | ± 0.00 | 193.625 |
| Contr_l + PR_L | GAT | 49.19 | ± 4.95 | 47.99 | ± 1.74 | 65.70 | ± 9.59 | 51.12 | ± 0.92 | 79.0 |
| Contr_l + PR_L | GCN | 49.52 | ± 8.28 | 46.66 | ± 1.06 | 69.56 | ± 6.62 | 47.07 | ± 2.85 | 88.875 |
| Contr_l + PR_L | GIN | 38.05 | ± 4.97 | 23.73 | ± 0.00 | 60.16 | ± 7.21 | 35.25 | ± 11.50 | 171.375 |
| Contr_l + PR_L | MPNN | 45.30 | ± 2.01 | 23.73 | ± 0.00 | 71.24 | ± 6.73 | 50.78 | ± 7.89 | 105.0 |
| Contr_l + PR_L | PAGNN | 37.82 | ± 8.14 | 23.73 | ± 0.00 | 54.84 | ± 7.19 | 23.73 | ± 0.00 | 187.125 |
| Contr_l + PR_L | SAGE | 41.52 | ± 6.05 | 52.56 | ± 2.18 | 51.52 | ± 7.73 | 52.96 | ± 3.48 | 92.75 |
| Contr_l + PR_L + Triplet_L | ALL | 39.70 | ± 4.88 | 23.73 | ± 0.01 | 58.07 | ± 12.43 | 30.40 | ± 14.92 | 174.625 |

Table 24. Results for Node Cls Precision (↑) (continued)

| Loss Type | Model | Cora ↓ Citeseer | | Cora ↓ Bitcoin | | Citeseer ↓ Cora | | Citeseer ↓ Bitcoin | | Average Rank |
|---|---|---|---|---|---|---|---|---|---|---|
| Contr_l + PR_L + Triplet_L | GAT | 55.99 ± 8.98 | | 48.77 ± 1.40 | | 72.86 ± 3.48 | | 47.94 ± 1.23 | | 51.0 |
| Contr_l + PR_L + Triplet_L | GCN | 50.84 ± 2.95 | | 46.82 ± 0.79 | | 68.87 ± 2.45 | | 44.88 ± 2.91 | | 96.625 |
| Contr_l + PR_L + Triplet_L | GIN | 44.56 ± 3.65 | | 41.01 ± 6.70 | | 68.54 ± 4.98 | | 40.85 ± 4.16 | | 129.25 |
| Contr_l + PR_L + Triplet_L | MPNN | 49.59 ± 1.23 | | 49.88 ± 1.95 | | 74.39 ± 2.17 | | 44.50 ± 3.00 | | 69.75 |
| Contr_l + PR_L + Triplet_L | PAGNN | 47.56 ± 1.78 | | 23.73 ± 0.00 | | 66.09 ± 5.75 | | 23.73 ± 0.00 | | 153.125 |
| Contr_l + PR_L + Triplet_L | SAGE | 39.72 ± 3.75 | | 50.60 ± 0.82 | | 60.19 ± 4.66 | | 50.11 ± 3.60 | | 99.0 |
| Contr_l + Triplet_L | ALL | 40.24 ± 2.40 | | 28.45 ± 6.50 | | 62.98 ± 3.46 | | 35.68 ± 10.21 | | 157.625 |
| Contr_l + Triplet_L | GAT | 58.43 ± 4.15 | | 48.11 ± 0.74 | | 76.76 ± 1.30 | | 48.61 ± 0.72 | | 36.875 |
| Contr_l + Triplet_L | GCN | 49.75 ± 3.84 | | 45.59 ± 1.76 | | 72.59 ± 1.35 | | 44.64 ± 1.10 | | 92.5 |
| Contr_l + Triplet_L | GIN | 44.87 ± 1.28 | | 38.62 ± 9.56 | | 69.34 ± 2.04 | | 40.97 ± 2.48 | | 127.75 |
| Contr_l + Triplet_L | MPNN | 54.10 ± 7.90 | | 47.40 ± 1.30 | | 73.46 ± 1.91 | | 47.65 ± 0.47 | | 63.75 |
| Contr_l + Triplet_L | PAGNN | 47.75 ± 1.46 | | 23.73 ± 0.00 | | 69.47 ± 2.65 | | 23.73 ± 0.00 | | 145.25 |
| Contr_l + Triplet_L | SAGE | 45.76 ± 5.69 | | 46.25 ± 2.08 | | 61.75 ± 2.84 | | 47.00 ± 1.78 | | 117.5 |
| CrossE_L | ALL | 9.26±7.99 | | 23.73 ± 0.00 | | 5.03±0.00 | | 23.73 ± 0.00 | | 196.875 |
| CrossE_L | GAT | 3.58±0.00 | | 23.73 ± 0.00 | | 5.03±0.00 | | 23.73 ± 0.00 | | 197.625 |
| CrossE_L | GCN | 3.58±0.00 | | 23.73 ± 0.00 | | 5.03±0.00 | | 23.73 ± 0.00 | | 197.625 |

Table 24. Results for Node Cls Precision (↑) (continued)

| Loss Type | Model | Cora ↓ Citeseer | Cora ↓ Bitcoin | Citeseer ↓ Cora | Citeseer ↓ Bitcoin | Average Rank |
|---|---|---|---|---|---|---|
| CrossE_L | GIN | 3.58±0.00 | 23.73 ± 0.00 | 5.03±0.00 | 23.73 ± 0.00 | 197.625 |
| CrossE_L | MPNN | 3.58±0.00 | 23.73 ± 0.00 | 5.03±0.00 | 23.73 ± 0.00 | 197.625 |
| CrossE_L | PAGNN | 12.26 ± 7.93 | 23.73 ± 0.00 | 5.03±0.00 | 23.73 ± 0.00 | 196.625 |
| CrossE_L | SAGE | 3.58±0.00 | 23.73 ± 0.00 | 5.03±0.00 | 23.73 ± 0.00 | 197.625 |
| CrossE_L + PMI_L | ALL | 53.64 ± 7.94 | 57.11 ± 0.04 | 71.65 ± 7.78 | 48.99 ± 4.67 | 42.0 |
| CrossE_L + PMI_L | GAT | 57.67 ± 4.27 | 47.09 ± 0.75 | 75.53 ± 2.46 | 47.51 ± 1.63 | 55.0 |
| CrossE_L + PMI_L | GCN | 52.58 ± 5.25 | 46.36 ± 1.70 | 71.88 ± 4.23 | 44.54 ± 1.72 | 85.75 |
| CrossE_L + PMI_L | GIN | 48.33 ± 8.80 | 36.62 ± 7.41 | 65.31 ± 2.34 | 40.15 ± 4.16 | 127.75 |
| CrossE_L + PMI_L | MPNN | 55.08 ± 3.37 | 49.54 ± 1.75 | 76.21 ± 1.74 | 48.58 ± 2.24 | 37.25 |
| CrossE_L + PMI_L | PAGNN | 49.20 ± 1.09 | 23.73 ± 0.00 | 64.95 ± 6.37 | 23.73 ± 0.00 | 150.125 |
| CrossE_L + PMI_L | SAGE | 40.42 ± 5.16 | 50.88 ± 2.59 | 59.65 ± 3.63 | 51.26 ± 0.35 | 94.75 |
| CrossE_L + PMI_L + PR_L | ALL | 47.61 ± 0.56 | 57.12 ± 0.01 | 70.94 ± 6.41 | 40.41 ± 16.69 | 81.0 |
| CrossE_L + PMI_L + PR_L | GAT | 63.09 ± 3.03 | 47.92 ± 0.69 | 75.93 ± 3.26 | 48.31 ± 1.02 | 40.25 |
| CrossE_L + PMI_L + PR_L | GCN | 53.76 ± 4.18 | 45.23 ± 1.15 | 70.80 ± 1.57 | 46.30 ± 1.11 | 84.5 |
| CrossE_L + PMI_L + PR_L | GIN | 44.19 ± 1.21 | 31.89 ± 11.27 | 57.04 ± 12.18 | 38.42 ± 2.34 | 155.75 |
| CrossE_L + PMI_L + PR_L | MPNN | 58.08 ± 5.72 | 49.77 ± 1.31 | 75.28 ± 1.32 | 49.65 ± 1.05 | 31.25 |

Table 24. Results for Node Cls Precision (↑) (continued)

| Loss Type | Model | Cora ↓ Citeseer | ± | Cora ↓ Bitcoin | ± | Citeseer ↓ Cora | ± | Citeseer ↓ Bitcoin | ± | Average Rank |
|---|---|---|---|---|---|---|---|---|---|---|
| CrossE_L + PMI_L + PR_L | PAGNN | 48.92 0.97 | ± | 23.73 0.00 | ± | 65.83 12.24 | ± | 23.73 0.00 | ± | 149.125 |
| CrossE_L + PMI_L + PR_L | SAGE | 49.59 5.59 | ± | 52.36 1.47 | ± | 58.31 2.67 | ± | 51.39 1.57 | ± | 69.25 |
| CrossE_L + PMI_L + PR_L + Triplet_L | ALL | 46.68 0.86 | ± | 32.63 12.20 | ± | 65.52 8.95 | ± | 41.43 12.58 | ± | 133.875 |
| CrossE_L + PMI_L + PR_L + Triplet_L | GAT | 61.04 3.78 | ± | 47.90 0.77 | ± | 76.47 2.44 | ± | 47.06 0.58 | ± | 46.625 |
| CrossE_L + PMI_L + PR_L + Triplet_L | GCN | 53.94 5.48 | ± | 46.97 1.33 | ± | 69.57 2.86 | ± | 45.86 1.57 | ± | 84.75 |
| CrossE_L + PMI_L + PR_L + Triplet_L | GIN | 45.12 2.07 | ± | 38.37 9.13 | ± | 68.25 4.10 | ± | 43.58 8.41 | ± | 128.125 |
| CrossE_L + PMI_L + PR_L + Triplet_L | MPNN | 58.83 3.14 | ± | 48.52 1.12 | ± | 75.83 1.11 | ± | 48.70 0.96 | ± | 35.125 |
| CrossE_L + PMI_L + PR_L + Triplet_L | PAGNN | 47.80 0.73 | ± | 23.73 0.00 | ± | 70.38 8.30 | ± | 23.73 0.00 | ± | 140.875 |
| CrossE_L + PMI_L + PR_L + Triplet_L | SAGE | 41.42 3.80 | ± | 51.80 0.63 | ± | 56.16 2.39 | ± | 52.77 3.38 | ± | 92.25 |
| CrossE_L + PMI_L + Triplet_L | ALL | 50.22 6.73 | ± | 47.69 2.86 | ± | 68.33 5.31 | ± | 46.73 1.74 | ± | 90.0 |
| CrossE_L + PMI_L + Triplet_L | GAT | 56.70 4.47 | ± | 47.88 0.77 | ± | 77.12 0.96 | ± | 48.45 0.82 | ± | 43.375 |
| CrossE_L + PMI_L + Triplet_L | GCN | 49.89 2.63 | ± | 44.46 2.21 | ± | 69.53 2.07 | ± | 45.05 2.38 | ± | 102.5 |
| CrossE_L + PMI_L + Triplet_L | GIN | 48.56 6.78 | ± | 45.91 8.12 | ± | 68.65 1.56 | ± | 39.14 4.99 | ± | 112.5 |
| CrossE_L + PMI_L + Triplet_L | MPNN | 59.49 4.51 | ± | 49.81 1.65 | ± | 76.39 1.40 | ± | 47.53 1.31 | ± | 33.25 |
| CrossE_L + PMI_L + Triplet_L | PAGNN | 46.80 2.29 | ± | 23.73 0.00 | ± | 72.04 9.40 | ± | 23.73 0.00 | ± | 140.125 |
| CrossE_L + PMI_L + Triplet_L | SAGE | 40.48 1.25 | ± | 50.01 1.51 | ± | 60.76 3.48 | ± | 51.11 1.29 | ± | 96.125 |

Table 24. Results for Node Cls Precision (↑) (continued)

| Loss Type | Model | Cora ↓ Citeseer | | Cora ↓ Bitcoin | | Citeseer ↓ Cora | | Citeseer ↓ Bitcoin | | Average Rank |
|---|---|---|---|---|---|---|---|---|---|---|
| CrossE_L + PR_L | ALL | 23.43 | ± 1.70 | 23.73 | ± 0.00 | 35.27 | ± 20.78 | 23.73 | ± 0.00 | 193.625 |
| CrossE_L + PR_L | GAT | 43.28 | ± 8.61 | 53.42 | ± 5.28 | 26.29 | ± 13.47 | 54.95 | ± 3.00 | 92.0 |
| CrossE_L + PR_L | GCN | 45.51 | ± 7.01 | 48.47 | ± 2.79 | 54.02 | ± 7.35 | 49.40 | ± 2.25 | 104.0 |
| CrossE_L + PR_L | GIN | 35.58 | ± 8.07 | 23.73 | ± 0.00 | 44.72 | ± 16.61 | 23.73 | ± 0.00 | 191.125 |
| CrossE_L + PR_L | MPNN | 46.38 | ± 2.50 | 23.73 | ± 0.00 | 60.50 | ± 6.12 | 28.18 | ± 9.95 | 159.375 |
| CrossE_L + PR_L | PAGNN | 37.07 | ± 10.15 | 23.73 | ± 0.00 | 52.83 | ± 7.05 | 23.73 | ± 0.00 | 188.125 |
| CrossE_L + PR_L | SAGE | 36.98 | ± 4.24 | 54.02 | ± 2.77 | 42.84 | ± 8.79 | 55.69 | ± 0.99 | 98.25 |
| CrossE_L + PR_L + Triplet_L | ALL | 42.44 | ± 4.27 | 30.41 | ± 14.94 | 45.62 | ± 5.33 | 30.40 | ± 14.92 | 167.875 |
| CrossE_L + PR_L + Triplet_L | GAT | 52.11 | ± 4.51 | 49.15 | ± 1.16 | 75.50 | ± 2.04 | 48.45 | ± 0.93 | 47.125 |
| CrossE_L + PR_L + Triplet_L | GCN | 50.55 | ± 8.05 | 47.51 | ± 0.66 | 70.02 | ± 3.20 | 46.11 | ± 2.89 | 88.0 |
| CrossE_L + PR_L + Triplet_L | GIN | 43.27 | ± 3.70 | 33.18 | ± 6.94 | 68.98 | ± 2.23 | 39.66 | ± 5.52 | 137.75 |
| CrossE_L + PR_L + Triplet_L | MPNN | 47.90 | ± 1.39 | 50.14 | ± 2.51 | 73.96 | ± 1.21 | 49.07 | ± 3.15 | 54.0 |
| CrossE_L + PR_L + Triplet_L | PAGNN | 43.13 | ± 4.08 | 23.73 | ± 0.00 | 62.27 | ± 7.75 | 23.73 | ± 0.00 | 170.875 |
| CrossE_L + PR_L + Triplet_L | SAGE | 41.05 | ± 1.25 | 48.53 | ± 1.89 | 59.49 | ± 1.40 | 48.80 | ± 5.59 | 108.0 |
| CrossE_L + Triplet_L | ALL | 46.26 | ± 7.41 | 35.94 | ± 8.50 | 65.41 | ± 5.19 | 36.76 | ± 3.50 | 139.0 |
| CrossE_L + Triplet_L | GAT | 58.39 | ± 6.90 | 48.79 | ± 0.45 | 77.58 | ± 0.90 | 49.39 | ± 0.54 | 27.25 |

Table 24. Results for Node Cls Precision (↑) (continued)

| Loss Type | Model | Cora ↓ Citeseer | | Cora ↓ Bitcoin | | Citeseer ↓ Cora | | Citeseer ↓ Bitcoin | | Average Rank |
|---|---|---|---|---|---|---|---|---|---|---|
| CrossE_L + Triplet_L | GCN | 50.44 | ± | 46.01 | ± | 70.68 | ± | 45.58 | ± | 92.5 |
| | | 4.85 | | 1.84 | | 2.72 | | 1.77 | | |
| CrossE_L + Triplet_L | GIN | 51.17 | ± | 38.36 | ± | 70.44 | ± | 42.47 | ± | 103.75 |
| | | 10.19 | | 6.42 | | 1.87 | | 3.86 | | |
| CrossE_L + Triplet_L | MPNN | 54.83 | ± | 48.49 | ± | 76.02 | ± | 45.30 | ± | 59.75 |
| | | 5.48 | | 1.48 | | 1.88 | | 1.49 | | |
| CrossE_L + Triplet_L | PAGNN | 53.13 | ± | 23.73 | ± | 70.34 | ± | 23.73 | ± | 129.875 |
| | | 9.22 | | 0.00 | | 4.06 | | 0.00 | | |
| CrossE_L + Triplet_L | SAGE | 52.36 | ± | 46.74 | ± | 64.64 | ± | 49.76 | ± | 80.5 |
| | | 6.24 | | 3.87 | | 1.39 | | 2.31 | | |
| PMI_L | ALL | 53.24 | ± | 48.89 | ± | 71.75 | ± | 51.11 | ± | 47.375 |
| | | 7.80 | | 7.78 | | 3.03 | | 1.44 | | |
| PMI_L | GAT | 62.23 | ± | 48.60 | ± | 76.52 | ± | 48.03 | ± | 32.25 |
| | | 2.12 | | 1.07 | | 2.19 | | 0.44 | | |
| PMI_L | GCN | 50.83 | ± | 45.69 | ± | 67.54 | ± | 46.39 | ± | 98.25 |
| | | 7.12 | | 1.84 | | 3.96 | | 1.11 | | |
| PMI_L | GIN | 47.75 | ± | 33.97 | ± | 68.15 | ± | 39.52 | ± | 127.125 |
| | | 7.65 | | 7.21 | | 5.28 | | 4.17 | | |
| PMI_L | MPNN | 58.17 | ± | 48.34 | ± | 74.18 | ± | 47.82 | ± | 49.75 |
| | | 2.31 | | 1.40 | | 1.14 | | 1.62 | | |
| PMI_L | PAGNN | 48.09 | ± | 23.73 | ± | 62.03 | ± | 23.73 | ± | 156.375 |
| | | 1.14 | | 0.00 | | 11.37 | | 0.00 | | |
| PMI_L | SAGE | 38.59 | ± | 51.79 | ± | 54.91 | ± | 51.33 | ± | 99.75 |
| | | 3.99 | | 2.63 | | 4.15 | | 1.26 | | |
| PMI_L + PR_L | ALL | 46.58 | ± | 57.11 | ± | 73.32 | ± | 27.06 | ± | 87.0 |
| | | 2.38 | | 0.00 | | 3.65 | | 7.46 | | |
| PMI_L + PR_L | GAT | 58.45 | ± | 48.02 | ± | 75.16 | ± | 47.23 | ± | 51.375 |
| | | 6.86 | | 1.09 | | 3.51 | | 2.07 | | |
| PMI_L + PR_L | GCN | 50.09 | ± | 45.44 | ± | 70.33 | ± | 45.25 | ± | 97.25 |
| | | 4.26 | | 2.19 | | 2.59 | | 1.70 | | |
| PMI_L + PR_L | GIN | 44.37 | ± | 39.68 | ± | 62.73 | ± | 38.65 | ± | 142.75 |
| | | 0.64 | | 9.79 | | 10.88 | | 4.68 | | |

Table 24. Results for Node Cls Precision (↑) (continued)

| Loss Type | Model | Cora ↓ Citeseer | | Cora ↓ Bitcoin | | Citeseer ↓ Cora | | Citeseer ↓ Bitcoin | | Average Rank |
|---|---|---|---|---|---|---|---|---|---|---|
| PMI_L + PR_L | MPNN | 56.06 | ± 4.50 | 49.28 | ± 1.51 | 74.66 | ± 2.40 | 44.24 | ± 3.80 | 60.125 |
| PMI_L + PR_L | PAGNN | 47.23 | ± 0.82 | 23.73 | ± 0.00 | 79.33 | ± 1.56 | 23.73 | ± 0.00 | 123.125 |
| PMI_L + PR_L | SAGE | 44.73 | ± 5.04 | 51.26 | ± 1.64 | 57.82 | ± 4.50 | 51.94 | ± 0.96 | 87.25 |
| PMI_L + PR_L + Triplet_L | ALL | 43.19 | ± 3.18 | 38.87 | ± 14.71 | 66.05 | ± 6.38 | 32.62 | ± 14.51 | 144.5 |
| PMI_L + PR_L + Triplet_L | GAT | 57.60 | ± 3.40 | 47.60 | ± 1.13 | 76.64 | ± 2.42 | 48.18 | ± 1.50 | 45.75 |
| PMI_L + PR_L + Triplet_L | GCN | 50.46 | ± 3.04 | 43.79 | ± 1.79 | 68.10 | ± 1.82 | 45.63 | ± 0.53 | 104.25 |
| PMI_L + PR_L + Triplet_L | GIN | 51.26 | ± 7.80 | 38.07 | ± 8.35 | 68.53 | ± 3.63 | 44.88 | ± 7.62 | 108.125 |
| PMI_L + PR_L + Triplet_L | MPNN | 58.26 | ± 4.34 | 48.25 | ± 0.99 | 75.31 | ± 2.80 | 47.84 | ± 2.67 | 47.0 |
| PMI_L + PR_L + Triplet_L | PAGNN | 47.35 | ± 0.92 | 23.73 | ± 0.00 | 70.43 | ± 8.03 | 23.73 | ± 0.00 | 142.875 |
| PMI_L + PR_L + Triplet_L | SAGE | 46.34 | ± 5.83 | 51.72 | ± 1.38 | 59.43 | ± 5.44 | 50.88 | ± 2.02 | 83.25 |
| PMI_L + Triplet_L | ALL | 50.87 | ± 7.00 | 38.01 | ± 4.76 | 72.86 | ± 1.62 | 46.16 | ± 0.75 | 90.375 |
| PMI_L + Triplet_L | GAT | 60.02 | ± 5.61 | 47.94 | ± 1.03 | 75.94 | ± 3.26 | 47.38 | ± 0.90 | 47.0 |
| PMI_L + Triplet_L | GCN | 51.57 | ± 2.99 | 45.95 | ± 2.73 | 69.54 | ± 2.54 | 45.72 | ± 2.07 | 93.5 |
| PMI_L + Triplet_L | GIN | 48.36 | ± 7.83 | 43.43 | ± 3.56 | 68.02 | ± 4.58 | 41.06 | ± 2.80 | 115.5 |
| PMI_L + Triplet_L | MPNN | 55.83 | ± 3.58 | 49.56 | ± 0.82 | 76.23 | ± 1.34 | 48.90 | ± 1.15 | 34.75 |
| PMI_L + Triplet_L | PAGNN | 47.09 | ± 1.19 | 23.73 | ± 0.00 | 67.82 | ± 8.66 | 23.73 | ± 0.00 | 152.75 |

Table 24. Results for Node Cls Precision (↑) (continued)

| Loss Type | Model | Cora ↓ Citeseer | | Cora ↓ Bitcoin | | Citeseer ↓ Cora | | Citeseer ↓ Bitcoin | | Average Rank |
|---|---|---|---|---|---|---|---|---|---|---|
| PMI_L + Triplet_L | SAGE | 43.58 ± 4.99 | | 52.54 ± 1.71 | | 57.31 ± 2.68 | | 51.97 ± 1.62 | | 87.0 |
| PR_L | ALL | 17.67 ± 7.84 | | 23.73 ± 0.00 | | 20.67 ± 12.97 | | 23.73 ± 0.00 | | 195.375 |
| PR_L | GAT | 42.41 ± 8.21 | | 47.18 ± 13.77 | | 26.63 ± 12.90 | | 49.08 ± 14.28 | | 123.0 |
| PR_L | GCN | 42.49 ± 2.32 | | 46.36 ± 3.48 | | 61.55 ± 7.86 | | 46.88 ± 4.20 | | 124.5 |
| PR_L | GIN | 30.44 ± 2.59 | | 23.73 ± 0.00 | | 42.53 ± 14.29 | | 30.40 ± 14.92 | | 185.125 |
| PR_L | MPNN | 44.97 ± 3.94 | | 23.73 ± 0.00 | | 62.18 ± 8.42 | | 23.73 ± 0.00 | | 166.625 |
| PR_L | PAGNN | 29.63 ± 6.69 | | 23.73 ± 0.00 | | 56.49 ± 6.46 | | 23.73 ± 0.00 | | 187.875 |
| PR_L | SAGE | 35.74 ± 3.35 | | 51.21 ± 4.69 | | 28.02 ± 6.51 | | **54.19 ± 1.56** | | 105.0 |
| PR_L + Triplet_L | ALL | 29.24 ± 8.14 | | 23.73 ± 0.00 | | 44.47 ± 13.96 | | 23.73 ± 0.00 | | 192.125 |
| PR_L + Triplet_L | GAT | 49.40 ± 1.61 | | 49.34 ± 1.58 | | 63.45 ± 9.02 | | 52.30 ± 1.33 | | 68.0 |
| PR_L + Triplet_L | GCN | 49.36 ± 4.26 | | 46.72 ± 1.77 | | 69.06 ± 2.94 | | 46.22 ± 1.43 | | 95.0 |
| PR_L + Triplet_L | GIN | 41.48 ± 3.09 | | 25.95 ± 4.97 | | 66.66 ± 7.76 | | 37.92 ± 9.66 | | 150.25 |
| PR_L + Triplet_L | MPNN | 47.29 ± 0.85 | | 30.44 ± 15.00 | | 67.34 ± 6.91 | | 45.83 ± 14.12 | | 123.625 |
| PR_L + Triplet_L | PAGNN | 41.07 ± 7.98 | | 23.73 ± 0.00 | | 61.44 ± 1.86 | | 23.73 ± 0.00 | | 175.375 |
| PR_L + Triplet_L | SAGE | 39.66 ± 1.30 | | 52.47 ± 1.90 | | 49.90 ± 10.03 | | 50.79 ± 2.82 | | 101.75 |
| Triplet_L | ALL | 51.07 ± 5.71 | | 42.89 ± 5.49 | | 69.22 ± 3.76 | | 46.13 ± 3.22 | | 97.875 |

Table 24. Results for Node Cls Precision (↑) (continued)

| Loss Type | Model | Cora ↓ Citeseer | | Cora ↓ Bitcoin | | Citeseer ↓ Cora | | Citeseer ↓ Bitcoin | | Average Rank |
|---|---|---|---|---|---|---|---|---|---|---|
| Triplet_L | GAT | 62.04 | ± 5.31 | 48.34 | ± 1.86 | 78.04 | ± 0.79 | 47.91 | ± 0.68 | 33.25 |
| Triplet_L | GCN | 54.76 | ± 5.53 | 46.02 | ± 1.45 | 71.68 | ± 3.09 | 44.93 | ± 1.34 | 83.75 |
| Triplet_L | GIN | 49.00 | ± 4.86 | 41.08 | ± 2.34 | 71.20 | ± 1.43 | 42.19 | ± 1.46 | 105.0 |
| Triplet_L | MPNN | 60.16 | ± 5.33 | 48.34 | ± 1.45 | 76.96 | ± 2.64 | 47.87 | ± 1.02 | 35.625 |
| Triplet_L | PAGNN | 55.45 | ± 9.78 | 23.73 | ± 0.00 | 71.73 | ± 2.43 | 23.73 | ± 0.00 | 121.125 |
| Triplet_L | SAGE | 48.30 | ± 4.68 | 50.75 | ± 1.43 | 62.57 | ± 2.86 | 50.78 | ± 1.43 | 74.25 |

Table 25. Node Cls Recall (Sensitivity) Performance (↑): This table presents models (Loss function and GNN) ranked by their average performance in terms of node cls recall (sensitivity). Top-ranked results are highlighted in red, second-ranked in blue, and third-ranked in green.

| Loss Type | Model | Cora ↓ Citeseer | | Cora ↓ Bitcoin | | Citeseer ↓ Cora | | Citeseer ↓ Bitcoin | | Average Rank |
|---|---|---|---|---|---|---|---|---|---|---|
| Contr_l | ALL | 39.18 | ± 2.63 | 33.32 | ± 0.02 | 51.12 | ± 6.53 | 33.37 | ± 0.10 | 170.25 |
| Contr_l | GAT | 55.91 | ± 0.76 | 40.59 | ± 0.74 | 71.87 | ± 2.12 | 40.06 | ± 0.66 | 27.875 |
| Contr_l | GCN | 51.76 | ± 0.68 | 37.37 | ± 0.64 | 67.12 | ± 2.25 | 38.00 | ± 0.94 | 59.5 |
| Contr_l | GIN | 47.11 | ± 3.14 | 34.14 | ± 0.51 | 63.39 | ± 3.96 | 35.06 | ± 0.54 | 101.75 |
| Contr_l | MPNN | 53.00 | ± 1.92 | 39.46 | ± 0.61 | 68.13 | ± 3.07 | 38.64 | ± 1.13 | 48.5 |

Table 25.  Results for Node Cls Recall (Sensitivity) (↑) (continued)

| Loss Type | Model | Cora ↓ Citeseer | | Cora ↓ Bitcoin | | Citeseer ↓ Cora | | Citeseer ↓ Bitcoin | | Average Rank |
|---|---|---|---|---|---|---|---|---|---|---|
| Contr_l | PAGNN | 47.76 ± 1.78 | | 33.33 ± 0.00 | | 60.86 ± 8.10 | | 33.33 ± 0.00 | | 138.25 |
| Contr_l | SAGE | 42.01 ± 2.33 | | 38.79 ± 2.06 | | 55.07 ± 5.91 | | 35.70 ± 1.52 | | 112.25 |
| Contr_l + CrossE_L | ALL | 42.71 ± 2.60 | | 33.41 ± 0.24 | | 44.47 ± 8.87 | | 33.34 ± 0.01 | | 161.75 |
| Contr_l + CrossE_L | GAT | 55.36 ± 1.26 | | 40.65 ± 0.63 | | 69.49 ± 2.33 | | 40.15 ± 0.65 | | 32.25 |
| Contr_l + CrossE_L | GCN | 50.86 ± 1.58 | | 37.94 ± 0.40 | | 67.29 ± 2.74 | | 37.27 ± 0.89 | | 65.125 |
| Contr_l + CrossE_L | GIN | 47.02 ± 2.30 | | 33.52 ± 0.17 | | 62.68 ± 1.93 | | 34.43 ± 0.66 | | 111.75 |
| Contr_l + CrossE_L | MPNN | 53.93 ± 1.20 | | 39.24 ± 0.42 | | 68.35 ± 3.61 | | 37.96 ± 0.54 | | 50.0 |
| Contr_l + CrossE_L | PAGNN | 47.67 ± 2.45 | | 33.33 ± 0.00 | | 59.82 ± 3.33 | | 33.35 ± 0.05 | | 134.625 |
| Contr_l + CrossE_L | SAGE | 42.97 ± 2.99 | | 35.46 ± 1.08 | | 52.21 ± 2.49 | | 33.60 ± 0.25 | | 133.5 |
| Contr_l + CrossE_L + PMI_L | ALL | 46.28 ± 3.04 | | 33.34 ± 0.04 | | 60.40 ± 7.49 | | 33.97 ± 0.62 | | 122.875 |
| Contr_l + CrossE_L + PMI_L | GAT | 56.58 ± 2.06 | | 41.56 ± 0.60 | | 71.50 ± 1.85 | | 40.80 ± 0.58 | | 17.0 |
| Contr_l + CrossE_L + PMI_L | GCN | 50.60 ± 2.16 | | 36.80 ± 0.27 | | 62.84 ± 2.08 | | 37.78 ± 1.44 | | 75.25 |
| Contr_l + CrossE_L + PMI_L | GIN | 45.05 ± 1.37 | | 33.52 ± 0.18 | | 54.49 ± 5.95 | | 33.73 ± 0.36 | | 132.5 |
| Contr_l + CrossE_L + PMI_L | MPNN | 56.48 ± 0.87 | | 40.00 ± 0.82 | | 70.93 ± 2.06 | | 40.38 ± 0.84 | | 31.5 |
| Contr_l + CrossE_L + PMI_L | PAGNN | 45.09 ± 2.24 | | 33.33 ± 0.00 | | 44.58 ± 3.67 | | 33.33 ± 0.00 | | 169.75 |
| Contr_l + CrossE_L + PMI_L | SAGE | 41.95 ± 2.08 | | 43.72 ± 2.07 | | 47.51 ± 4.07 | | 41.63 ± 2.51 | | 86.25 |

Table 25. Results for Node Cls Recall (Sensitivity) (↑) (continued)

| Loss Type | Model | Cora ↓ Citeseer | | Cora ↓ Bitcoin | | Citeseer ↓ Cora | | Citeseer ↓ Bitcoin | | Average Rank |
|---|---|---|---|---|---|---|---|---|---|---|
| Contr_l + CrossE_L + PMI_L + PR_L | ALL | 37.25 ± 1.76 | | 33.57 ± 0.00 | ± | 53.10 ± 4.52 | | 33.36 ± 0.11 | ± | 151.875 |
| Contr_l + CrossE_L + PMI_L + PR_L | GAT | 56.40 ± 1.53 | | 40.59 ± 1.06 | ± | 71.59 ± 4.12 | | 40.44 ± 0.63 | ± | 25.0 |
| Contr_l + CrossE_L + PMI_L + PR_L | GCN | 51.90 ± 1.26 | | 37.15 ± 0.98 | ± | 64.41 ± 2.65 | | 37.84 ± 0.64 | ± | 66.875 |
| Contr_l + CrossE_L + PMI_L + PR_L | GIN | 45.52 ± 2.99 | | 33.49 ± 0.13 | ± | 48.63 ± 8.01 | | 33.38 ± 0.10 | ± | 144.125 |
| Contr_l + CrossE_L + PMI_L + PR_L | MPNN | 55.32 ± 0.93 | | 40.85 ± 0.66 | ± | 69.89 ± 1.66 | | 36.16 ± 1.41 | ± | 47.125 |
| Contr_l + CrossE_L + PMI_L + PR_L | PAGNN | 46.68 ± 2.28 | | 33.33 ± 0.00 | ± | 47.62 ± 2.89 | | 33.33 ± 0.00 | ± | 159.5 |
| Contr_l + CrossE_L + PMI_L + PR_L | SAGE | 42.85 ± 2.50 | | **44.34 ± 1.79** | ± | 46.78 ± 0.93 | | 42.04 ± 1.83 | ± | 83.0 |
| Contr_l + CrossE_L + PMI_L + PR_L + Triplet_L | ALL | 45.18 ± 1.86 | | 33.44 ± 0.17 | ± | 54.26 ± 5.80 | | 33.67 ± 0.43 | ± | 134.125 |
| Contr_l + CrossE_L + PMI_L + PR_L + Triplet_L | GAT | 56.56 ± 1.60 | | 41.05 ± 0.44 | ± | 72.64 ± 1.95 | | 41.15 ± 0.59 | ± | 14.625 |
| Contr_l + CrossE_L + PMI_L + PR_L + Triplet_L | GCN | 51.16 ± 3.00 | | 37.89 ± 0.75 | ± | 63.91 ± 2.13 | | 36.93 ± 0.48 | ± | 72.75 |
| Contr_l + CrossE_L + PMI_L + PR_L + Triplet_L | GIN | 44.96 ± 1.64 | | 33.56 ± 0.26 | ± | 60.41 ± 2.51 | | 33.48 ± 0.21 | ± | 128.875 |
| Contr_l + CrossE_L + PMI_L + PR_L + Triplet_L | MPNN | 54.53 ± 0.89 | | 41.05 ± 1.18 | ± | 71.59 ± 1.82 | | 37.27 ± 1.91 | ± | 39.625 |
| Contr_l + CrossE_L + PMI_L + PR_L + Triplet_L | PAGNN | 46.48 ± 2.21 | | 33.33 ± 0.00 | ± | 51.27 ± 2.84 | | 33.33 ± 0.00 | ± | 153.625 |

Table 25. Results for Node Cls Recall (Sensitivity) (↑) (continued)

| Loss Type | Model | Cora ↓ Citeseer | | Cora ↓ Bitcoin | | Citeseer ↓ Cora | | Citeseer ↓ Bitcoin | | Average Rank |
|---|---|---|---|---|---|---|---|---|---|---|
| Contr_l + CrossE_L + PMI_L + PR_L + Triplet_L | SAGE | 44.56 1.74 | ± | 41.95 0.23 | | 51.68 2.90 | ± | 44.01 1.81 | ± | 72.5 |
| Contr_l + CrossE_L + PMI_L + Triplet_L | ALL | 48.88 2.09 | ± | 34.71 1.33 | | 61.56 1.96 | ± | 36.84 1.23 | ± | 92.75 |
| Contr_l + CrossE_L + PMI_L + Triplet_L | GAT | 57.89 1.41 | ± | 40.89 0.46 | | 73.12 1.60 | ± | 40.96 0.49 | ± | 12.875 |
| Contr_l + CrossE_L + PMI_L + Triplet_L | GCN | 51.13 1.82 | ± | 37.84 0.42 | | 62.31 1.99 | ± | 37.23 1.27 | ± | 74.5 |
| Contr_l + CrossE_L + PMI_L + Triplet_L | GIN | 47.22 1.57 | ± | 33.74 0.43 | | 61.11 4.49 | ± | 33.68 0.48 | ± | 112.125 |
| Contr_l + CrossE_L + PMI_L + Triplet_L | MPNN | 54.93 1.65 | ± | 40.40 1.09 | | 71.06 1.05 | ± | 39.27 0.57 | ± | 36.5 |
| Contr_l + CrossE_L + PMI_L + Triplet_L | PAGNN | 46.38 2.21 | ± | 33.33 0.00 | | 48.12 1.87 | ± | 33.33 0.00 | ± | 160.25 |
| Contr_l + CrossE_L + PMI_L + Triplet_L | SAGE | 42.60 3.92 | ± | 42.06 0.94 | | 50.99 2.82 | ± | 42.77 0.97 | ± | 78.5 |
| Contr_l + CrossE_L + PR_L | ALL | 25.85 2.69 | ± | 33.33 0.00 | | 30.51 6.08 | ± | 33.33 0.00 | ± | 191.5 |
| Contr_l + CrossE_L + PR_L | GAT | 51.25 2.36 | ± | 40.16 0.62 | | 42.19 11.87 | ± | 37.61 1.20 | ± | 89.0 |
| Contr_l + CrossE_L + PR_L | GCN | 48.23 2.57 | ± | 36.74 0.97 | | 59.15 2.71 | ± | 36.17 0.97 | ± | 96.75 |
| Contr_l + CrossE_L + PR_L | GIN | 38.52 1.10 | ± | 33.42 0.21 | | 49.52 4.67 | ± | 33.37 0.04 | ± | 158.375 |
| Contr_l + CrossE_L + PR_L | MPNN | 47.68 2.41 | ± | 33.67 0.41 | | 56.96 6.89 | ± | 33.85 0.55 | ± | 114.75 |
| Contr_l + CrossE_L + PR_L | PAGNN | 40.51 2.99 | ± | 33.33 0.00 | | 41.13 5.53 | ± | 33.33 0.00 | ± | 181.5 |
| Contr_l + CrossE_L + PR_L | SAGE | 38.67 3.52 | ± | 37.05 2.57 | | 38.77 4.99 | ± | 39.12 1.80 | ± | 128.125 |
| Contr_l + CrossE_L + PR_L + Triplet_L | ALL | 36.26 2.14 | ± | 33.33 0.00 | | 41.38 8.30 | ± | 33.35 0.05 | ± | 179.75 |

Table 25. Results for Node Cls Recall (Sensitivity) (↑) (continued)

| Loss Type | Model | Cora ↓ Citeseer | | Cora ↓ Bitcoin | | Citeseer ↓ Cora | | Citeseer ↓ Bitcoin | | Average Rank |
|---|---|---|---|---|---|---|---|---|---|---|
| Contr_l + CrossE_L + PR_L + Triplet_L | GAT | 54.07 2.31 | ± | 40.10 0.34 | ± | 67.68 3.19 | ± | 40.08 0.88 | ± | 42.25 |
| Contr_l + CrossE_L + PR_L + Triplet_L | GCN | 49.00 2.73 | ± | 37.15 1.17 | ± | 59.82 3.54 | ± | 37.39 0.79 | ± | 85.5 |
| Contr_l + CrossE_L + PR_L + Triplet_L | GIN | 45.79 2.21 | ± | 33.40 0.14 | ± | 59.92 2.54 | ± | 33.92 0.32 | ± | 124.125 |
| Contr_l + CrossE_L + PR_L + Triplet_L | MPNN | 52.15 1.30 | ± | 36.86 0.95 | ± | 65.75 3.14 | ± | 35.47 1.00 | ± | 74.25 |
| Contr_l + CrossE_L + PR_L + Triplet_L | PAGNN | 43.52 1.88 | ± | 33.33 0.00 | ± | 48.91 3.29 | ± | 33.33 0.00 | ± | 167.5 |
| Contr_l + CrossE_L + PR_L + Triplet_L | SAGE | 39.42 2.12 | ± | 35.29 0.77 | ± | 46.28 3.79 | ± | 37.28 1.05 | ± | 134.75 |
| Contr_l + CrossE_L + Triplet_L | ALL | 43.11 4.13 | ± | 33.31 0.12 | ± | 53.34 4.28 | ± | 33.43 0.20 | ± | 159.125 |
| Contr_l + CrossE_L + Triplet_L | GAT | 55.34 1.69 | ± | 40.97 0.73 | ± | 71.39 2.13 | ± | 40.35 0.40 | ± | 25.5 |
| Contr_l + CrossE_L + Triplet_L | GCN | 52.10 1.82 | ± | 37.09 0.66 | ± | 66.17 1.74 | ± | 38.22 0.47 | ± | 61.0 |
| Contr_l + CrossE_L + Triplet_L | GIN | 48.03 2.38 | ± | 34.47 0.54 | ± | 65.54 3.32 | ± | 34.64 0.28 | ± | 93.75 |
| Contr_l + CrossE_L + Triplet_L | MPNN | 53.46 1.58 | ± | 38.46 1.23 | ± | 68.95 2.03 | ± | 38.91 1.01 | ± | 48.375 |
| Contr_l + CrossE_L + Triplet_L | PAGNN | 48.26 1.63 | ± | 33.33 0.00 | ± | 64.50 2.71 | ± | 33.34 0.04 | ± | 125.375 |
| Contr_l + CrossE_L + Triplet_L | SAGE | 45.35 2.36 | ± | 36.76 1.39 | ± | 56.68 2.90 | ± | 38.35 1.65 | ± | 99.5 |
| Contr_l + PMI_L | ALL | 45.42 1.93 | ± | 33.35 0.05 | ± | 59.20 8.50 | ± | 34.18 0.71 | ± | 127.25 |
| Contr_l + PMI_L | GAT | 57.00 1.20 | ± | 40.67 0.56 | ± | 71.12 3.77 | ± | 40.31 0.76 | ± | 24.75 |
| Contr_l + PMI_L | GCN | 51.35 2.49 | ± | 37.65 0.50 | ± | 64.24 2.66 | ± | 37.62 0.37 | ± | 68.0 |

Continued on next page

Table 25. Results for Node Cls Recall (Sensitivity) (↑) (continued)

| Loss Type | Model | Cora ↓ Citeseer | | Cora ↓ Bitcoin | | Citeseer ↓ Cora | | Citeseer ↓ Bitcoin | | Average Rank |
|---|---|---|---|---|---|---|---|---|---|---|
| Contr_l + PMI_L | GIN | 46.09 | ± 2.87 | 34.03 | ± 0.51 | 55.80 | ± 5.63 | 33.57 | ± 0.49 | 123.75 |
| Contr_l + PMI_L | MPNN | 54.50 | ± 0.74 | 40.49 | ± 1.34 | 72.12 | ± 1.50 | 39.61 | ± 0.83 | 32.375 |
| Contr_l + PMI_L | PAGNN | 46.48 | ± 1.29 | 33.33 | ± 0.00 | 40.80 | ± 3.35 | 33.33 | ± 0.00 | 166.875 |
| Contr_l + PMI_L | SAGE | 41.74 | ± 1.43 | 44.14 | ± 1.94 | 48.18 | ± 4.78 | 42.63 | ± 2.42 | 84.0 |
| Contr_l + PMI_L + PR_L | ALL | 40.98 | ± 0.90 | 33.59 | ± 0.03 | 48.97 | ± 3.78 | 33.37 | ± 0.13 | 151.5 |
| Contr_l + PMI_L + PR_L | GAT | 57.53 | ± 0.94 | 41.31 | ± 0.65 | 67.60 | ± 1.67 | 38.51 | ± 0.36 | 29.875 |
| Contr_l + PMI_L + PR_L | GCN | 50.52 | ± 1.53 | 37.82 | ± 1.19 | 64.46 | ± 2.33 | 37.74 | ± 1.05 | 67.75 |
| Contr_l + PMI_L + PR_L | GIN | 45.15 | ± 0.65 | 33.41 | ± 0.12 | 48.38 | ± 10.24 | 33.38 | ± 0.10 | 147.875 |
| Contr_l + PMI_L + PR_L | MPNN | 54.79 | ± 1.73 | 40.24 | ± 1.35 | 69.27 | ± 2.67 | 35.43 | ± 1.49 | 56.0 |
| Contr_l + PMI_L + PR_L | PAGNN | 44.70 | ± 1.70 | 33.33 | ± 0.00 | 45.88 | ± 5.47 | 33.33 | ± 0.00 | 170.0 |
| Contr_l + PMI_L + PR_L | SAGE | 42.47 | ± 1.68 | 43.49 | ± 2.73 | 46.85 | ± 5.16 | 38.72 | ± 2.86 | 96.0 |
| Contr_l + PMI_L + PR_L + Triplet_L | ALL | 44.89 | ± 1.24 | 33.37 | ± 0.09 | 49.16 | ± 4.39 | 33.45 | ± 0.10 | 147.125 |
| Contr_l + PMI_L + PR_L + Triplet_L | GAT | 55.35 | ± 0.81 | 40.79 | ± 0.94 | 69.07 | ± 2.93 | 38.98 | ± 1.43 | 35.5 |
| Contr_l + PMI_L + PR_L + Triplet_L | GCN | 49.96 | ± 0.43 | 37.26 | ± 0.93 | 64.76 | ± 2.92 | 37.54 | ± 0.85 | 72.5 |
| Contr_l + PMI_L + PR_L + Triplet_L | GIN | 46.32 | ± 1.80 | 33.61 | ± 0.46 | 58.66 | ± 2.19 | 33.65 | ± 0.14 | 121.5 |
| Contr_l + PMI_L + PR_L + Triplet_L | MPNN | 56.05 | ± 2.16 | 39.88 | ± 0.81 | 71.73 | ± 3.18 | 34.58 | ± 0.95 | 52.25 |

Table 25. Results for Node Cls Recall (Sensitivity) (↑) (continued)

| Loss Type | Model | Cora ↓ Citeseer | | Cora ↓ Bitcoin | | Citeseer ↓ Cora | | Citeseer ↓ Bitcoin | | Average Rank |
|---|---|---|---|---|---|---|---|---|---|---|
| Contr_l + PMI_L + PR_L + Triplet_L | PAGNN | 46.61 1.52 | ± | 33.33 0.00 | ± | 57.05 4.47 | ± | 33.33 0.00 | ± | 146.75 |
| Contr_l + PMI_L + PR_L + Triplet_L | SAGE | 44.16 3.53 | ± | 42.41 1.81 | ± | 49.87 3.92 | ± | 36.67 0.96 | ± | 97.75 |
| Contr_l + PR_L | ALL | 27.77 1.53 | ± | 33.33 0.00 | ± | 27.21 3.29 | ± | 33.33 0.00 | ± | 191.75 |
| Contr_l + PR_L | GAT | 49.43 3.25 | ± | 39.54 0.76 | ± | 43.59 7.77 | ± | 38.40 0.48 | ± | 91.5 |
| Contr_l + PR_L | GCN | 46.82 2.20 | ± | 37.17 0.77 | ± | 55.18 7.11 | ± | 36.52 0.61 | ± | 101.0 |
| Contr_l + PR_L | GIN | 39.64 3.99 | ± | 33.33 0.00 | ± | 48.67 8.36 | ± | 33.55 0.23 | ± | 163.75 |
| Contr_l + PR_L | MPNN | 43.96 2.66 | ± | 33.33 0.00 | ± | 51.60 7.55 | ± | 34.60 1.18 | ± | 145.0 |
| Contr_l + PR_L | PAGNN | 38.59 3.28 | ± | 33.33 0.00 | ± | 39.61 3.25 | ± | 33.33 0.00 | ± | 186.0 |
| Contr_l + PR_L | SAGE | 39.91 2.16 | ± | 35.71 0.52 | ± | 40.76 7.07 | ± | 39.69 2.38 | ± | 127.25 |
| Contr_l + PR_L + Triplet_L | ALL | 36.47 2.02 | ± | 33.31 0.04 | ± | 40.97 7.17 | ± | 33.35 0.06 | ± | 186.875 |
| Contr_l + PR_L + Triplet_L | GAT | 52.04 1.44 | ± | 40.13 0.99 | ± | 65.15 5.52 | ± | 40.53 1.13 | ± | 45.125 |
| Contr_l + PR_L + Triplet_L | GCN | 49.17 2.05 | ± | 38.22 0.90 | ± | 61.73 1.42 | ± | 36.45 1.44 | ± | 81.75 |
| Contr_l + PR_L + Triplet_L | GIN | 45.70 2.40 | ± | 34.32 0.82 | ± | 59.25 5.55 | ± | 33.85 0.33 | ± | 116.875 |
| Contr_l + PR_L + Triplet_L | MPNN | 52.00 1.22 | ± | 36.83 0.78 | ± | 65.10 3.58 | ± | 35.53 1.96 | ± | 77.25 |
| Contr_l + PR_L + Triplet_L | PAGNN | 45.40 2.28 | ± | 33.33 0.00 | ± | 52.27 5.91 | ± | 33.33 0.00 | ± | 157.5 |
| Contr_l + PR_L + Triplet_L | SAGE | 42.04 3.69 | ± | 38.04 2.01 | ± | 48.94 2.16 | ± | 38.52 2.35 | ± | 108.625 |

Table 25. Results for Node Cls Recall (Sensitivity) (↑) (continued)

| Loss Type | Model | Cora ↓ Citeseer | | Cora ↓ Bitcoin | | Citeseer ↓ Cora | | Citeseer ↓ Bitcoin | | Average Rank |
|---|---|---|---|---|---|---|---|---|---|---|
| Contr_l + Triplet_L | ALL | 42.32 | ± 1.57 | 33.33 | ± 0.08 | 54.04 | ± 6.70 | 33.37 | ± 0.17 | 156.25 |
| Contr_l + Triplet_L | GAT | 57.60 | ± 1.28 | 40.26 | ± 0.62 | 72.83 | ± 1.82 | 40.97 | ± 0.83 | 19.125 |
| Contr_l + Triplet_L | GCN | 49.58 | ± 1.34 | 37.51 | ± 0.57 | 68.12 | ± 1.00 | 37.78 | ± 1.20 | 65.25 |
| Contr_l + Triplet_L | GIN | 47.72 | ± 1.03 | 34.13 | ± 0.71 | 63.09 | ± 4.01 | 34.79 | ± 0.42 | 100.25 |
| Contr_l + Triplet_L | MPNN | 53.58 | ± 1.32 | 38.04 | ± 1.01 | 70.81 | ± 2.19 | 38.51 | ± 0.61 | 48.0 |
| Contr_l + Triplet_L | PAGNN | 49.35 | ± 0.83 | 33.33 | ± 0.00 | 62.22 | ± 2.99 | 33.33 | ± 0.00 | 132.75 |
| Contr_l + Triplet_L | SAGE | 44.06 | ± 1.58 | 35.15 | ± 0.24 | 55.92 | ± 4.01 | 35.86 | ± 0.46 | 117.5 |
| CrossE_L | ALL | 17.62 | ± 1.30 | 33.33 | ± 0.00 | 14.29 | ± 0.00 | 33.33 | ± 0.00 | 195.75 |
| CrossE_L | GAT | 16.67 | ± 0.00 | 33.33 | ± 0.00 | 14.29 | ± 0.00 | 33.33 | ± 0.00 | 196.5 |
| CrossE_L | GCN | 16.67 | ± 0.00 | 33.33 | ± 0.00 | 14.29 | ± 0.00 | 33.33 | ± 0.00 | 196.5 |
| CrossE_L | GIN | 16.67 | ± 0.00 | 33.33 | ± 0.00 | 14.29 | ± 0.00 | 33.33 | ± 0.00 | 196.5 |
| CrossE_L | MPNN | 16.67 | ± 0.00 | 33.33 | ± 0.00 | 14.29 | ± 0.00 | 33.33 | ± 0.00 | 196.5 |
| CrossE_L | PAGNN | 19.65 | ± 2.74 | 33.33 | ± 0.00 | 14.29 | ± 0.00 | 33.33 | ± 0.00 | 195.5 |
| CrossE_L | SAGE | 16.67 | ± 0.00 | 33.33 | ± 0.00 | 14.29 | ± 0.00 | 33.33 | ± 0.00 | 196.5 |
| CrossE_L + PMI_L | ALL | 52.03 | ± 1.64 | 33.57 | ± 0.20 | 59.10 | ± 3.59 | 33.76 | ± 0.31 | 103.25 |
| CrossE_L + PMI_L | GAT | 55.92 | ± 3.46 | 39.98 | ± 0.62 | 71.09 | ± 2.24 | 40.52 | ± 0.74 | 31.125 |

Continued on next page

Table 25. Results for Node Cls Recall (Sensitivity) (↑) (continued)

| Loss Type | Model | Cora ↓ Citeseer | | Cora ↓ Bitcoin | | Citeseer ↓ Cora | | Citeseer ↓ Bitcoin | | Average Rank |
|---|---|---|---|---|---|---|---|---|---|---|
| CrossE_L + PMI_L | GCN | 50.95 | ± | 37.75 | ± | 65.85 | ± | 37.10 | ± | 69.0 |
| | | 1.23 | | 1.25 | | 3.60 | | 0.57 | | |
| CrossE_L + PMI_L | GIN | 47.01 | ± | 33.67 | ± | 54.61 | ± | 33.66 | ± | 121.875 |
| | | 2.92 | | 0.41 | | 2.47 | | 0.34 | | |
| CrossE_L + PMI_L | MPNN | 54.80 | ± | 40.73 | ± | 73.80 | ± | 39.83 | ± | 27.0 |
| | | 1.02 | | 1.22 | | 1.76 | | 0.71 | | |
| CrossE_L + PMI_L | PAGNN | 47.20 | ± | 33.33 | ± | 42.74 | ± | 33.33 | ± | 162.75 |
| | | 2.58 | | 0.00 | | 3.14 | | 0.00 | | |
| CrossE_L + PMI_L | SAGE | 40.30 | ± | 42.52 | ± | 49.63 | ± | 43.86 | ± | 82.5 |
| | | 2.54 | | 2.69 | | 1.54 | | 1.43 | | |
| CrossE_L + PMI_L + PR_L | ALL | 40.25 | ± | 33.59 | ± | 52.42 | ± | 33.42 | ± | 146.25 |
| | | 1.34 | | 0.05 | | 5.09 | | 0.10 | | |
| CrossE_L + PMI_L + PR_L | GAT | 58.05 | ± | 40.68 | ± | 66.95 | ± | 38.48 | ± | 34.5 |
| | | 1.90 | | 0.45 | | 8.72 | | 0.47 | | |
| CrossE_L + PMI_L + PR_L | GCN | 50.87 | ± | 37.20 | ± | 65.67 | ± | 38.15 | ± | 64.25 |
| | | 2.87 | | 1.09 | | 2.72 | | 1.16 | | |
| CrossE_L + PMI_L + PR_L | GIN | 45.50 | ± | 33.41 | ± | 45.74 | ± | 33.63 | ± | 147.25 |
| | | 2.16 | | 0.21 | | 12.69 | | 0.24 | | |
| CrossE_L + PMI_L + PR_L | MPNN | 56.08 | ± | 41.17 | ± | 70.23 | ± | 40.99 | ± | 21.25 |
| | | 1.65 | | 1.15 | | 2.03 | | 0.92 | | |
| CrossE_L + PMI_L + PR_L | PAGNN | 47.51 | ± | 33.33 | ± | 46.93 | ± | 33.33 | ± | 157.75 |
| | | 2.60 | | 0.00 | | 3.13 | | 0.00 | | |
| CrossE_L + PMI_L + PR_L | SAGE | 42.34 | ± | 43.54 | ± | 49.05 | ± | 42.55 | ± | 81.0 |
| | | 1.83 | | 0.88 | | 2.50 | | 1.76 | | |
| CrossE_L + PMI_L + PR_L + Triplet_L | ALL | 46.47 | ± | 33.39 | ± | 50.54 | ± | 33.46 | ± | 137.75 |
| | | 1.51 | | 0.12 | | 6.72 | | 0.11 | | |
| CrossE_L + PMI_L + PR_L + Triplet_L | GAT | 57.92 | ± | 40.86 | ± | 70.50 | ± | 40.19 | ± | 23.0 |
| | | 2.28 | | 0.86 | | 1.84 | | 0.85 | | |
| CrossE_L + PMI_L + PR_L + Triplet_L | GCN | 49.82 | ± | 38.15 | ± | 63.55 | ± | 37.78 | ± | 69.75 |
| | | 1.50 | | 0.54 | | 3.68 | | 0.11 | | |
| CrossE_L + PMI_L + PR_L + Triplet_L | GIN | 46.33 | ± | 33.56 | ± | 57.54 | ± | 33.71 | ± | 122.5 |
| | | 0.89 | | 0.17 | | 5.95 | | 0.29 | | |

Table 25. Results for Node Cls Recall (Sensitivity) (↑) (continued)

| Loss Type | Model | Cora ↓ Citeseer | | Cora ↓ Bitcoin | | Citeseer ↓ Cora | | Citeseer ↓ Bitcoin | | Average Rank |
|---|---|---|---|---|---|---|---|---|---|---|
| CrossE_L + PMI_L + PR_L + Triplet_L | MPNN | 56.11 1.59 | ± | 40.80 1.00 | ± | 71.04 1.74 | ± | 39.57 1.16 | ± | 28.875 |
| CrossE_L + PMI_L + PR_L + Triplet_L | PAGNN | 47.81 1.03 | ± | 33.33 0.00 | ± | 48.25 3.19 | ± | 33.33 0.00 | ± | 154.0 |
| CrossE_L + PMI_L + PR_L + Triplet_L | SAGE | 41.72 2.30 | ± | 42.25 1.03 | ± | 48.29 2.08 | ± | 43.41 2.11 | ± | 84.75 |
| CrossE_L + PMI_L + Triplet_L | ALL | 48.56 1.28 | ± | 35.88 1.08 | ± | 62.29 3.77 | ± | 37.27 1.03 | ± | 88.625 |
| CrossE_L + PMI_L + Triplet_L | GAT | 57.37 2.52 | ± | 40.80 0.59 | ± | 72.12 1.13 | ± | 40.95 0.73 | ± | 17.25 |
| CrossE_L + PMI_L + Triplet_L | GCN | 49.72 0.96 | ± | 36.78 0.88 | ± | 64.27 0.90 | ± | 37.58 1.19 | ± | 78.375 |
| CrossE_L + PMI_L + Triplet_L | GIN | 46.09 1.96 | ± | 33.83 0.41 | ± | 59.76 1.78 | ± | 33.71 0.32 | ± | 117.875 |
| CrossE_L + PMI_L + Triplet_L | MPNN | 55.38 1.46 | ± | 40.89 1.25 | ± | 72.88 0.82 | ± | 39.16 0.68 | ± | 25.375 |
| CrossE_L + PMI_L + Triplet_L | PAGNN | 46.26 2.10 | ± | 33.33 0.00 | ± | 46.42 5.47 | ± | 33.33 0.00 | ± | 163.5 |
| CrossE_L + PMI_L + Triplet_L | SAGE | 43.15 1.50 | ± | 41.63 2.15 | ± | 51.72 2.11 | ± | 42.02 1.49 | ± | 77.25 |
| CrossE_L + PR_L | ALL | 25.33 2.30 | ± | 33.33 0.00 | ± | 27.14 3.60 | ± | 33.33 0.00 | ± | 192.75 |
| CrossE_L + PR_L | GAT | 40.48 8.13 | ± | 33.89 0.30 | ± | 16.68 1.55 | ± | 33.95 0.44 | ± | 155.5 |
| CrossE_L + PR_L | GCN | 45.77 3.95 | ± | 35.66 1.83 | ± | 38.34 3.33 | ± | 35.58 0.85 | ± | 131.75 |
| CrossE_L + PR_L | GIN | 35.89 4.69 | ± | 33.33 0.00 | ± | 35.07 9.87 | ± | 33.33 0.00 | ± | 189.5 |
| CrossE_L + PR_L | MPNN | 45.61 3.82 | ± | 33.33 0.00 | ± | 44.71 9.32 | ± | 33.37 0.08 | ± | 159.5 |
| CrossE_L + PR_L | PAGNN | 36.97 3.72 | ± | 33.33 0.00 | ± | 39.35 6.46 | ± | 33.33 0.00 | ± | 187.5 |

Table 25. Results for Node Cls Recall (Sensitivity) (↑) (continued)

| Loss Type | Model | Cora ↓ Citeseer | | Cora ↓ Bitcoin | | Citeseer ↓ Cora | | Citeseer ↓ Bitcoin | | Average Rank |
|---|---|---|---|---|---|---|---|---|---|---|
| CrossE_L + PR_L | SAGE | 36.02 | ± 4.45 | 36.62 | ± 2.41 | 26.60 | ± 4.66 | 37.34 | ± 2.94 | 144.0 |
| CrossE_L + PR_L + Triplet_L | ALL | 37.87 | ± 4.67 | 33.40 | ± 0.16 | 32.57 | ± 5.69 | 33.35 | ± 0.05 | 174.375 |
| CrossE_L + PR_L + Triplet_L | GAT | 53.68 | ± 1.47 | 40.58 | ± 0.70 | 64.57 | ± 4.76 | 40.21 | ± 0.71 | 44.75 |
| CrossE_L + PR_L + Triplet_L | GCN | 50.56 | ± 3.12 | 38.46 | ± 0.76 | 61.53 | ± 4.43 | 37.41 | ± 1.51 | 72.875 |
| CrossE_L + PR_L + Triplet_L | GIN | 45.88 | ± 3.96 | 33.53 | ± 0.37 | 57.50 | ± 2.61 | 33.45 | ± 0.11 | 129.625 |
| CrossE_L + PR_L + Triplet_L | MPNN | 50.12 | ± 1.77 | 35.72 | ± 1.21 | 64.83 | ± 1.40 | 36.83 | ± 1.33 | 82.0 |
| CrossE_L + PR_L + Triplet_L | PAGNN | 41.14 | ± 2.55 | 33.33 | ± 0.00 | 49.91 | ± 6.26 | 33.33 | ± 0.00 | 170.5 |
| CrossE_L + PR_L + Triplet_L | SAGE | 43.04 | ± 1.67 | 36.92 | ± 1.55 | 47.63 | ± 1.36 | 35.35 | ± 1.28 | 128.5 |
| CrossE_L + Triplet_L | ALL | 45.28 | ± 3.19 | 33.57 | ± 0.28 | 60.67 | ± 3.63 | 33.68 | ± 0.41 | 123.5 |
| CrossE_L + Triplet_L | GAT | 56.95 | ± 1.89 | 41.19 | ± 0.36 | 73.72 | ± 1.61 | 41.17 | ± 0.24 | **11.75** |
| CrossE_L + Triplet_L | GCN | 50.15 | ± 0.94 | 37.58 | ± 0.45 | 66.51 | ± 2.88 | 37.38 | ± 1.18 | 69.0 |
| CrossE_L + Triplet_L | GIN | 48.44 | ± 2.19 | 33.89 | ± 0.79 | 66.21 | ± 1.61 | 34.63 | ± 0.24 | 92.75 |
| CrossE_L + Triplet_L | MPNN | 54.40 | ± 2.07 | 39.06 | ± 0.84 | 71.76 | ± 0.44 | 37.71 | ± 0.51 | 45.625 |
| CrossE_L + Triplet_L | PAGNN | 49.01 | ± 1.81 | 33.33 | ± 0.00 | 64.93 | ± 2.15 | 33.32 | ± 0.02 | 133.375 |
| CrossE_L + Triplet_L | SAGE | 48.27 | ± 2.77 | 36.99 | ± 2.25 | 57.99 | ± 0.88 | 36.53 | ± 0.69 | 94.0 |
| PMI_L | ALL | 50.49 | ± 1.75 | 33.55 | ± 0.10 | 59.42 | ± 4.89 | 35.45 | ± 1.23 | 101.75 |

Table 25. Results for Node Cls Recall (Sensitivity) (↑) (continued)

| Loss Type | Model | Cora ↓ Citeseer | | Cora ↓ Bitcoin | | Citeseer ↓ Cora | | Citeseer ↓ Bitcoin | | Average Rank |
|---|---|---|---|---|---|---|---|---|---|---|
| PMI_L | GAT | 57.28 ± 2.38 | | 41.08 ± 0.58 | | 71.09 ± 2.30 | | 41.03 ± 0.39 | | 16.875 |
| PMI_L | GCN | 50.20 ± 1.33 | | 37.70 ± 0.84 | | 62.42 ± 2.06 | | 37.27 ± 0.74 | | 76.625 |
| PMI_L | GIN | 46.41 ± 2.51 | | 33.42 ± 0.14 | | 55.02 ± 5.80 | | 33.64 ± 0.21 | | 129.125 |
| PMI_L | MPNN | 55.09 ± 1.18 | | 40.45 ± 0.79 | | 71.45 ± 2.08 | | 40.43 ± 1.23 | | 30.375 |
| PMI_L | PAGNN | 47.71 ± 0.26 | | 33.33 ± 0.00 | | 41.08 ± 2.91 | | 33.33 ± 0.00 | | 162.125 |
| PMI_L | SAGE | 39.46 ± 2.98 | | 44.29 ± 2.23 | | 44.20 ± 5.32 | | 42.95 ± 2.53 | | 90.75 |
| PMI_L + PR_L | ALL | 42.59 ± 3.71 | | 33.57 ± 0.00 | | 47.82 ± 8.41 | | 33.33 ± 0.04 | | 159.875 |
| PMI_L + PR_L | GAT | 56.43 ± 2.82 | | 40.85 ± 0.60 | | 68.31 ± 4.65 | | 39.43 ± 1.09 | | 31.125 |
| PMI_L + PR_L | GCN | 50.19 ± 1.65 | | 37.46 ± 0.63 | | 65.58 ± 1.91 | | 37.49 ± 0.78 | | 69.75 |
| PMI_L + PR_L | GIN | 45.99 ± 0.77 | | 33.68 ± 0.38 | | 48.62 ± 9.05 | | 33.62 ± 0.25 | | 134.75 |
| PMI_L + PR_L | MPNN | 54.20 ± 2.09 | | 40.26 ± 0.91 | | 66.92 ± 2.46 | | 34.73 ± 0.88 | | 62.125 |
| PMI_L + PR_L | PAGNN | 47.41 ± 1.11 | | 33.33 ± 0.00 | | 50.03 ± 5.33 | | 33.33 ± 0.00 | | 152.25 |
| PMI_L + PR_L | SAGE | 43.39 ± 1.64 | | 42.80 ± 3.17 | | 47.32 ± 2.47 | | 44.76 ± 2.56 | | 80.75 |
| PMI_L + PR_L + Triplet_L | ALL | 43.60 ± 3.39 | | 33.44 ± 0.12 | | 52.99 ± 2.38 | | 33.32 ± 0.10 | | 157.5 |
| PMI_L + PR_L + Triplet_L | GAT | 56.51 ± 2.64 | | 40.33 ± 0.97 | | 70.59 ± 1.18 | | 40.85 ± 0.74 | | 27.5 |
| PMI_L + PR_L + Triplet_L | GCN | 49.72 ± 2.12 | | 36.84 ± 0.71 | | 63.87 ± 2.70 | | 37.95 ± 0.86 | | 76.125 |

Continued on next page

Table 25. Results for Node Cls Recall (Sensitivity) (↑) (continued)

| Loss Type | Model | Cora ↓ Citeseer | | Cora ↓ Bitcoin | | Citeseer ↓ Cora | | Citeseer ↓ Bitcoin | | Average Rank |
|---|---|---|---|---|---|---|---|---|---|---|
| PMI_L + PR_L + Triplet_L | GIN | 47.96 1.16 | ± | 33.76 0.28 | ± | 56.56 4.55 | ± | 33.52 0.07 | ± | 117.25 |
| PMI_L + PR_L + Triplet_L | MPNN | 55.66 1.81 | ± | 40.06 0.47 | ± | 69.12 1.65 | ± | 35.34 0.62 | ± | 55.0 |
| PMI_L + PR_L + Triplet_L | PAGNN | 46.09 1.03 | ± | 33.33 0.00 | ± | 53.25 5.12 | ± | 33.33 0.00 | ± | 153.5 |
| PMI_L + PR_L + Triplet_L | SAGE | 43.91 2.45 | ± | 43.68 1.62 | ± | 50.53 4.69 | ± | 41.20 1.92 | ± | 76.25 |
| PMI_L + Triplet_L | ALL | 49.88 2.32 | ± | 33.59 0.29 | ± | 64.26 3.17 | ± | 37.45 0.74 | ± | 87.0 |
| PMI_L + Triplet_L | GAT | 57.47 1.60 | ± | 40.37 0.48 | ± | 71.46 3.86 | ± | 40.53 1.13 | ± | 23.625 |
| PMI_L + Triplet_L | GCN | 49.11 1.41 | ± | 37.05 0.85 | ± | 63.96 1.71 | ± | 37.14 1.25 | ± | 82.625 |
| PMI_L + Triplet_L | GIN | 47.16 2.11 | ± | 33.89 0.18 | ± | 61.42 4.30 | ± | 33.87 0.25 | ± | 109.25 |
| PMI_L + Triplet_L | MPNN | 54.97 2.03 | ± | 40.63 1.00 | ± | 73.81 1.77 | ± | 40.43 1.11 | ± | 24.875 |
| PMI_L + Triplet_L | PAGNN | 45.67 2.08 | ± | 33.33 0.00 | ± | 46.10 4.93 | ± | 33.33 0.00 | ± | 166.25 |
| PMI_L + Triplet_L | SAGE | 43.79 1.67 | ± | 44.24 1.51 | ± | 50.68 2.94 | ± | 43.28 1.69 | ± | 73.25 |
| PR_L | ALL | 23.45 0.87 | ± | 33.33 0.00 | ± | 22.66 1.58 | ± | 33.33 0.00 | ± | 193.5 |
| PR_L | GAT | 42.51 2.04 | ± | 33.86 0.53 | ± | 16.29 1.97 | ± | 34.35 1.01 | ± | 152.0 |
| PR_L | GCN | 44.15 2.68 | ± | 34.79 0.25 | ± | 44.47 4.24 | ± | 34.82 0.69 | ± | 135.125 |
| PR_L | GIN | 32.39 4.08 | ± | 33.33 0.00 | ± | 30.36 7.25 | ± | 33.35 0.05 | ± | 184.75 |
| PR_L | MPNN | 44.69 3.56 | ± | 33.33 0.00 | ± | 43.54 5.93 | ± | 33.33 0.00 | ± | 172.5 |

Table 25. Results for Node Cls Recall (Sensitivity) (↑) (continued)

| Loss Type | Model | Cora ↓ Citeseer | | Cora ↓ Bitcoin | | Citeseer ↓ Cora | | Citeseer ↓ Bitcoin | | Average Rank |
|---|---|---|---|---|---|---|---|---|---|---|
| PR_L | PAGNN | 34.87 | ± 4.91 | 33.33 | ± 0.00 | 41.63 | ± 3.86 | 33.33 | ± 0.00 | 186.5 |
| PR_L | SAGE | 37.29 | ± 3.58 | 36.08 | ± 1.44 | 20.48 | ± 4.79 | 39.39 | ± 0.98 | 134.5 |
| PR_L + Triplet_L | ALL | 27.43 | ± 2.20 | 33.33 | ± 0.00 | 29.13 | ± 1.98 | 33.33 | ± 0.00 | 191.75 |
| PR_L + Triplet_L | GAT | 51.45 | ± 3.19 | 39.32 | ± 1.21 | 40.93 | ± 5.68 | 37.71 | ± 0.44 | 91.625 |
| PR_L + Triplet_L | GCN | 48.30 | ± 1.95 | 37.18 | ± 1.37 | 56.59 | ± 5.39 | 36.49 | ± 0.58 | 94.0 |
| PR_L + Triplet_L | GIN | 42.14 | ± 3.38 | 33.34 | ± 0.01 | 52.06 | ± 2.03 | 33.61 | ± 0.48 | 148.375 |
| PR_L + Triplet_L | MPNN | 47.71 | ± 1.93 | 33.52 | ± 0.42 | 44.35 | ± 6.40 | 34.01 | ± 1.01 | 133.375 |
| PR_L + Triplet_L | PAGNN | 38.86 | ± 2.44 | 33.33 | ± 0.00 | 39.20 | ± 3.61 | 33.33 | ± 0.00 | 186.0 |
| PR_L + Triplet_L | SAGE | 39.56 | ± 1.15 | 36.31 | ± 1.37 | 33.84 | ± 10.24 | 37.69 | ± 2.36 | 136.5 |
| Triplet_L | ALL | 47.88 | ± 2.64 | 33.66 | ± 0.22 | 64.04 | ± 4.09 | 37.30 | ± 1.45 | 93.25 |
| Triplet_L | GAT | 57.56 | ± 1.30 | 40.81 | ± 1.01 | 74.23 | ± 2.21 | 40.90 | ± 0.49 | 14.0 |
| Triplet_L | GCN | 52.03 | ± 2.99 | 37.59 | ± 0.71 | 66.96 | ± 3.29 | 37.75 | ± 0.72 | 59.875 |
| Triplet_L | GIN | 49.81 | ± 1.36 | 34.53 | ± 0.57 | 65.38 | ± 1.81 | 34.92 | ± 0.59 | 88.5 |
| Triplet_L | MPNN | 55.33 | ± 1.45 | 39.65 | ± 1.74 | 72.81 | ± 2.43 | 39.41 | ± 1.15 | 34.25 |
| Triplet_L | PAGNN | 50.98 | ± 1.89 | 33.33 | ± 0.00 | 67.10 | ± 3.29 | 33.33 | ± 0.00 | 118.5 |
| Triplet_L | SAGE | 47.24 | ± 1.24 | 40.61 | ± 1.27 | 56.53 | ± 3.51 | 40.50 | ± 1.33 | 69.25 |

### 1.2.2 Assessment of Learned Representations through Link Prediction (LP).

Table 26. Lp Accuracy Performance (↑): This table presents models (Loss function and GNN) ranked by their average performance in terms of lp accuracy. Top-ranked results are highlighted in red, second-ranked in blue, and third-ranked in green.

| Loss Type | Model | Cora ↓ Citeseer | | Cora ↓ Bitcoin | | Citeseer ↓ Cora | | Citeseer ↓ Bitcoin | | Average Rank |
|---|---|---|---|---|---|---|---|---|---|---|
| Contr_l | ALL | 90.97 1.54 | ± | 64.19 3.32 | ± | 87.11 1.11 | ± | 46.68 16.56 | ± | 131.25 |
| Contr_l | GAT | 98.82 0.17 | ± | 88.60 0.80 | ± | 97.75 0.15 | ± | 89.80 0.26 | ± | 18.375 |
| Contr_l | GCN | 98.14 0.13 | ± | 88.67 0.58 | ± | 96.49 0.37 | ± | 88.92 0.53 | ± | 39.375 |
| Contr_l | GIN | 93.81 0.71 | ± | 70.57 0.64 | ± | 90.66 1.17 | ± | 75.00 1.15 | ± | 106.0 |
| Contr_l | MPNN | 97.81 0.32 | ± | 87.33 0.65 | ± | 92.45 1.47 | ± | 87.74 0.80 | ± | 74.625 |
| Contr_l | PAGNN | 84.14 1.84 | ± | 35.11 0.24 | ± | 83.70 1.61 | ± | 35.58 0.35 | ± | 158.25 |
| Contr_l | SAGE | 95.67 0.33 | ± | 69.41 4.43 | ± | 93.84 1.38 | ± | 75.82 5.69 | ± | 99.5 |
| Contr_l + CrossE_L | ALL | 91.32 1.08 | ± | 69.09 1.71 | ± | 85.71 1.65 | ± | 53.90 12.96 | ± | 128.5 |
| Contr_l + CrossE_L | GAT | 98.72 0.16 | ± | 88.77 0.46 | ± | 97.27 0.34 | ± | 88.61 0.45 | ± | 29.0 |
| Contr_l + CrossE_L | GCN | 98.35 0.10 | ± | 88.41 0.88 | ± | 96.19 0.28 | ± | 89.23 0.21 | ± | 36.625 |
| Contr_l + CrossE_L | GIN | 89.91 2.26 | ± | 60.96 14.76 | ± | 90.57 1.03 | ± | 71.35 0.88 | ± | 119.25 |
| Contr_l + CrossE_L | MPNN | 97.71 0.37 | ± | 87.60 0.86 | ± | 92.85 0.74 | ± | 87.67 0.73 | ± | 74.875 |
| Contr_l + CrossE_L | PAGNN | 83.14 1.47 | ± | 35.35 0.21 | ± | 82.64 1.33 | ± | 35.39 0.25 | ± | 160.5 |
| Contr_l + CrossE_L | SAGE | 95.41 0.53 | ± | 64.57 1.56 | ± | 94.42 1.05 | ± | 65.20 3.36 | ± | 109.0 |

Continued on next page

Table 26. Results for Lp Accuracy (↑) (continued)

| Loss Type | Model | Cora ↓ Citeseer | | Cora ↓ Bitcoin | | Citeseer ↓ Cora | | Citeseer ↓ Bitcoin | | Average Rank |
|---|---|---|---|---|---|---|---|---|---|---|
| Contr_l + CrossE_L + PMI_L | ALL | 80.08 4.92 | ± | 78.79 0.72 | ± | 86.01 0.75 | ± | 81.84 0.40 | ± | 119.375 |
| Contr_l + CrossE_L + PMI_L | GAT | 99.01 0.19 | ± | 89.18 0.75 | ± | 97.58 0.23 | ± | 89.44 0.36 | ± | 10.625 |
| Contr_l + CrossE_L + PMI_L | GCN | 98.24 0.15 | ± | 89.00 0.31 | ± | 95.95 0.38 | ± | 88.69 0.68 | ± | 39.25 |
| Contr_l + CrossE_L + PMI_L | GIN | 84.01 2.42 | ± | 34.65 0.01 | ± | 81.61 1.93 | ± | 48.07 18.17 | ± | 159.375 |
| Contr_l + CrossE_L + PMI_L | MPNN | 98.43 0.19 | ± | 88.86 0.63 | ± | 93.59 0.59 | ± | 88.62 0.41 | ± | 45.75 |
| Contr_l + CrossE_L + PMI_L | PAGNN | 64.08 1.62 | ± | 34.68 0.09 | ± | 65.37 1.83 | ± | 34.66 0.01 | ± | 191.875 |
| Contr_l + CrossE_L + PMI_L | SAGE | 78.64 0.82 | ± | 50.13 3.60 | ± | 78.16 2.26 | ± | 56.29 0.44 | ± | 153.5 |
| Contr_l + CrossE_L + PMI_L + PR_L | ALL | 59.11 3.37 | ± | 82.30 0.10 | ± | 84.72 1.32 | ± | 82.34 0.15 | ± | 132.25 |
| Contr_l + CrossE_L + PMI_L + PR_L | GAT | 98.90 0.25 | ± | 89.24 1.28 | ± | 97.37 0.23 | ± | 88.78 1.29 | ± | 19.125 |
| Contr_l + CrossE_L + PMI_L + PR_L | GCN | 98.19 0.11 | ± | 89.13 0.38 | ± | 95.95 0.49 | ± | 88.54 0.70 | ± | 41.5 |
| Contr_l + CrossE_L + PMI_L + PR_L | GIN | 77.43 3.04 | ± | 34.64 0.01 | ± | 73.92 3.74 | ± | 35.00 0.20 | ± | 178.0 |
| Contr_l + CrossE_L + PMI_L + PR_L | MPNN | 98.47 0.11 | ± | 87.97 0.63 | ± | 88.76 3.36 | ± | 85.96 1.85 | ± | 72.0 |
| Contr_l + CrossE_L + PMI_L + PR_L | PAGNN | 66.93 2.41 | ± | 34.87 0.14 | ± | 64.62 3.22 | ± | 46.04 0.29 | ± | 178.875 |
| Contr_l + CrossE_L + PMI_L + PR_L | SAGE | 77.56 1.42 | ± | 44.53 1.88 | ± | 75.63 2.24 | ± | 44.54 4.17 | ± | 164.75 |
| Contr_l + CrossE_L + PMI_L + PR_L + Triplet_L | ALL | 73.91 2.82 | ± | 76.44 0.71 | ± | 85.47 1.71 | ± | 78.80 0.90 | ± | 128.75 |

Table 26. Results for Lp Accuracy (↑) (continued)

| Loss Type | Model | Cora ↓ Citeseer | ± | Cora ↓ Bitcoin | ± | Citeseer ↓ Cora | ± | Citeseer ↓ Bitcoin | ± | Average Rank |
|---|---|---|---|---|---|---|---|---|---|---|
| Contr_l + CrossE_L + PMI_L + PR_L + Triplet_L | GAT | 98.99 0.18 | ± | 88.99 1.01 | ± | 96.87 0.78 | ± | 88.94 0.71 | ± | 21.0 |
| Contr_l + CrossE_L + PMI_L + PR_L + Triplet_L | GCN | 98.24 0.19 | ± | 88.72 0.73 | ± | 95.99 0.21 | ± | 89.10 0.42 | ± | 37.625 |
| Contr_l + CrossE_L + PMI_L + PR_L + Triplet_L | GIN | 80.32 3.42 | ± | 47.62 17.75 | ± | 82.48 1.55 | ± | 34.72 0.05 | ± | 161.25 |
| Contr_l + CrossE_L + PMI_L + PR_L + Triplet_L | MPNN | 98.27 0.21 | ± | 88.59 0.19 | ± | 90.26 2.32 | ± | 86.03 0.99 | ± | 68.625 |
| Contr_l + CrossE_L + PMI_L + PR_L + Triplet_L | PAGNN | 66.64 2.16 | ± | 34.92 0.17 | ± | 64.20 1.44 | ± | 34.97 0.28 | ± | 184.375 |
| Contr_l + CrossE_L + PMI_L + PR_L + Triplet_L | SAGE | 88.43 3.84 | ± | 70.31 1.79 | ± | 89.49 0.27 | ± | 68.94 2.71 | ± | 117.75 |
| Contr_l + CrossE_L + PMI_L + Triplet_L | ALL | 94.12 1.22 | ± | 81.17 0.68 | ± | 90.93 0.76 | ± | 81.74 0.25 | ± | 98.0 |
| Contr_l + CrossE_L + PMI_L + Triplet_L | GAT | 99.04 0.07 | ± | 89.50 0.35 | ± | 97.33 0.13 | ± | 89.27 0.53 | ± | 9.25 |
| Contr_l + CrossE_L + PMI_L + Triplet_L | GCN | 98.34 0.23 | ± | 89.54 0.50 | ± | 95.51 0.74 | ± | 89.06 0.67 | ± | 29.75 |
| Contr_l + CrossE_L + PMI_L + Triplet_L | GIN | 85.04 1.78 | ± | 61.57 15.08 | ± | 83.61 2.91 | ± | 67.77 1.23 | ± | 138.75 |
| Contr_l + CrossE_L + PMI_L + Triplet_L | MPNN | 98.43 0.16 | ± | 89.08 0.30 | ± | 93.26 0.93 | ± | 88.19 0.59 | ± | 48.25 |
| Contr_l + CrossE_L + PMI_L + Triplet_L | PAGNN | 66.35 0.49 | ± | 34.64 0.00 | ± | 65.62 1.63 | ± | 34.66 0.02 | ± | 194.125 |
| Contr_l + CrossE_L + PMI_L + Triplet_L | SAGE | 90.31 1.60 | ± | 63.19 3.93 | ± | 87.76 1.02 | ± | 69.24 0.60 | ± | 123.75 |

Table 26. Results for Lp Accuracy (↑) (continued)

| Loss Type | Model | Cora ↓ Citeseer | | Cora ↓ Bitcoin | | Citeseer ↓ Cora | | Citeseer ↓ Bitcoin | | Average Rank |
|---|---|---|---|---|---|---|---|---|---|---|
| Contr_l + CrossE_L + PR_L | ALL | 59.27 2.17 | ± | 78.23 1.64 | ± | 65.62 5.74 | ± | 81.14 0.22 | ± | 148.125 |
| Contr_l + CrossE_L + PR_L | GAT | 97.71 1.51 | ± | 88.51 0.46 | ± | 96.81 0.34 | ± | 87.91 0.75 | ± | 54.25 |
| Contr_l + CrossE_L + PR_L | GCN | 97.45 0.27 | ± | 87.13 0.43 | ± | 94.72 0.47 | ± | 88.07 1.09 | ± | 69.375 |
| Contr_l + CrossE_L + PR_L | GIN | 88.53 2.19 | ± | 34.64 0.00 | ± | 74.49 2.46 | ± | 34.65 0.01 | ± | 174.375 |
| Contr_l + CrossE_L + PR_L | MPNN | 91.36 5.42 | ± | 88.67 0.55 | ± | 84.60 3.75 | ± | 88.73 0.40 | ± | 82.5 |
| Contr_l + CrossE_L + PR_L | PAGNN | 68.25 4.34 | ± | 46.00 0.49 | ± | 53.42 3.47 | ± | 46.41 0.50 | ± | 177.25 |
| Contr_l + CrossE_L + PR_L | SAGE | 91.07 7.58 | ± | 66.58 3.19 | ± | 85.50 8.92 | ± | 60.98 4.20 | ± | 130.75 |
| Contr_l + CrossE_L + PR_L + Triplet_L | ALL | 79.32 2.17 | ± | 34.67 0.06 | ± | 85.55 1.23 | ± | 59.58 14.00 | ± | 154.125 |
| Contr_l + CrossE_L + PR_L + Triplet_L | GAT | 98.76 0.18 | ± | 88.26 1.21 | ± | 97.22 0.72 | ± | 89.20 0.46 | ± | 28.625 |
| Contr_l + CrossE_L + PR_L + Triplet_L | GCN | 97.98 0.29 | ± | 88.74 0.94 | ± | 95.43 0.38 | ± | 88.66 1.20 | ± | 49.25 |
| Contr_l + CrossE_L + PR_L + Triplet_L | GIN | 90.46 1.43 | ± | 59.17 13.71 | ± | 89.09 0.84 | ± | 69.77 1.44 | ± | 122.625 |
| Contr_l + CrossE_L + PR_L + Triplet_L | MPNN | 97.08 0.81 | ± | 88.68 0.40 | ± | 89.28 0.69 | ± | 86.41 1.31 | ± | 75.5 |
| Contr_l + CrossE_L + PR_L + Triplet_L | PAGNN | 73.87 2.50 | ± | 34.92 0.10 | ± | 67.06 3.17 | ± | 44.59 4.54 | ± | 174.625 |
| Contr_l + CrossE_L + PR_L + Triplet_L | SAGE | 96.42 0.37 | ± | 71.60 2.25 | ± | 93.11 0.91 | ± | 62.47 4.39 | ± | 103.75 |
| Contr_l + CrossE_L + Triplet_L | ALL | 93.32 1.45 | ± | 70.99 0.99 | ± | 89.64 1.02 | ± | 67.70 1.40 | ± | 113.5 |
| Contr_l + CrossE_L + Triplet_L | GAT | 98.95 0.19 | ± | 88.78 0.77 | ± | 97.61 0.28 | ± | 89.22 0.42 | ± | 17.25 |

Continued on next page

Table 26. Results for Lp Accuracy (↑) (continued)

| Loss Type | Model | Cora ↓ Citeseer | | Cora ↓ Bitcoin | | Citeseer ↓ Cora | | Citeseer ↓ Bitcoin | | Average Rank |
|---|---|---|---|---|---|---|---|---|---|---|
| Contr_l + CrossE_L + Triplet_L | GCN | 98.54 | ± 0.11 | 88.76 | ± 0.74 | 96.38 | ± 0.41 | 88.46 | ± 0.47 | 35.375 |
| Contr_l + CrossE_L + Triplet_L | GIN | 95.21 | ± 0.68 | 72.21 | ± 0.72 | 91.98 | ± 1.07 | 76.06 | ± 0.54 | 99.5 |
| Contr_l + CrossE_L + Triplet_L | MPNN | 97.97 | ± 0.24 | 87.26 | ± 0.69 | 93.82 | ± 0.57 | 88.29 | ± 0.53 | 68.75 |
| Contr_l + CrossE_L + Triplet_L | PAGNN | 81.05 | ± 6.78 | 34.87 | ± 0.08 | 87.02 | ± 1.39 | 35.43 | ± 0.32 | 155.125 |
| Contr_l + CrossE_L + Triplet_L | SAGE | 96.78 | ± 0.48 | 71.35 | ± 3.67 | 95.05 | ± 0.42 | 77.12 | ± 2.56 | 89.0 |
| Contr_l + PMI_L | ALL | 82.71 | ± 8.04 | 47.96 | ± 18.26 | 87.14 | ± 1.28 | 79.01 | ± 0.70 | 128.5 |
| Contr_l + PMI_L | GAT | 99.02 | ± 0.19 | 89.71 | ± 0.57 | 97.45 | ± 0.20 | 89.33 | ± 0.84 | 7.5 |
| Contr_l + PMI_L | GCN | 98.25 | ± 0.20 | 88.65 | ± 0.64 | 95.94 | ± 0.50 | 89.13 | ± 0.76 | 39.5 |
| Contr_l + PMI_L | GIN | 83.31 | ± 4.28 | 69.68 | ± 0.86 | 80.94 | ± 1.43 | 61.28 | ± 14.90 | 140.0 |
| Contr_l + PMI_L | MPNN | 98.40 | ± 0.18 | 87.66 | ± 0.69 | 93.86 | ± 0.39 | 88.33 | ± 0.27 | 59.125 |
| Contr_l + PMI_L | PAGNN | 65.34 | ± 1.92 | 34.64 | ± 0.00 | 63.97 | ± 0.54 | 34.66 | ± 0.01 | 197.0 |
| Contr_l + PMI_L | SAGE | 78.99 | ± 2.03 | 52.89 | ± 3.93 | 82.99 | ± 3.24 | 59.11 | ± 1.07 | 148.25 |
| Contr_l + PMI_L + PR_L | ALL | 57.74 | ± 1.60 | 79.41 | ± 0.42 | 84.07 | ± 1.11 | 82.46 | ± 0.48 | 135.25 |
| Contr_l + PMI_L + PR_L | GAT | 98.76 | ± 0.30 | 89.30 | ± 0.47 | 94.14 | ± 3.75 | 83.13 | ± 1.61 | 45.625 |
| Contr_l + PMI_L + PR_L | GCN | 98.38 | ± 0.23 | 89.08 | ± 0.91 | 95.98 | ± 0.42 | 88.70 | ± 0.61 | 33.875 |
| Contr_l + PMI_L + PR_L | GIN | 81.57 | ± 5.37 | 34.64 | ± 0.00 | 77.60 | ± 3.70 | 34.94 | ± 0.16 | 174.0 |

Table 26. Results for Lp Accuracy (↑) (continued)

| Loss Type | Model | Cora ↓ Citeseer | | Cora ↓ Bitcoin | | Citeseer ↓ Cora | | Citeseer ↓ Bitcoin | | Average Rank |
|---|---|---|---|---|---|---|---|---|---|---|
| Contr_l + PMI_L + PR_L | MPNN | 98.54 0.08 | ± | 88.96 0.40 | ± | 86.69 3.92 | ± | 85.78 0.57 | ± | 63.625 |
| Contr_l + PMI_L + PR_L | PAGNN | 64.66 1.66 | ± | 34.64 0.00 | ± | 63.65 0.52 | ± | 34.66 0.01 | ± | 197.5 |
| Contr_l + PMI_L + PR_L | SAGE | 78.47 0.93 | ± | 51.42 4.19 | ± | 80.85 2.03 | ± | 53.63 2.47 | ± | 153.25 |
| Contr_l + PMI_L + PR_L + Triplet_L | ALL | 79.46 3.01 | ± | 34.64 0.00 | ± | 86.75 1.14 | ± | 71.58 2.06 | ± | 148.5 |
| Contr_l + PMI_L + PR_L + Triplet_L | GAT | 98.54 0.49 | ± | 88.57 0.67 | ± | 95.80 1.05 | ± | 85.41 2.13 | ± | 52.125 |
| Contr_l + PMI_L + PR_L + Triplet_L | GCN | 98.20 0.22 | ± | 89.27 0.55 | ± | 96.02 0.53 | ± | 88.75 0.64 | ± | 34.125 |
| Contr_l + PMI_L + PR_L + Triplet_L | GIN | 85.06 0.83 | ± | 70.18 0.67 | ± | 85.94 1.99 | ± | 47.65 17.73 | ± | 133.25 |
| Contr_l + PMI_L + PR_L + Triplet_L | MPNN | 98.54 0.16 | ± | 88.27 0.66 | ± | 88.48 1.06 | ± | 85.09 1.35 | ± | 70.125 |
| Contr_l + PMI_L + PR_L + Triplet_L | PAGNN | 67.74 1.53 | ± | 34.65 0.01 | ± | 64.67 2.53 | ± | 34.75 0.22 | ± | 188.625 |
| Contr_l + PMI_L + PR_L + Triplet_L | SAGE | 93.28 1.10 | ± | 65.13 0.82 | ± | 90.20 1.84 | ± | 68.82 4.48 | ± | 116.25 |
| Contr_l + PR_L | ALL | 61.24 4.60 | ± | 81.36 0.34 | ± | 70.78 4.97 | ± | 82.07 0.63 | ± | 142.75 |
| Contr_l + PR_L | GAT | 98.09 0.67 | ± | 87.90 0.13 | ± | 96.76 0.24 | ± | 87.74 0.81 | ± | 56.25 |
| Contr_l + PR_L | GCN | 97.48 0.37 | ± | 87.99 0.46 | ± | 94.71 0.60 | ± | 88.56 0.75 | ± | 63.375 |
| Contr_l + PR_L | GIN | 86.85 4.19 | ± | 45.57 6.16 | ± | 70.02 12.11 | ± | 46.67 16.39 | ± | 157.125 |
| Contr_l + PR_L | MPNN | 91.33 2.14 | ± | 88.49 0.63 | ± | 86.57 1.82 | ± | 88.91 0.31 | ± | 81.0 |
| Contr_l + PR_L | PAGNN | 63.92 9.22 | ± | 42.08 5.71 | ± | 52.40 0.73 | ± | 46.71 0.42 | ± | 182.0 |

Table 26. Results for Lp Accuracy (↑) (continued)

| Loss Type | Model | Cora ↓ Citeseer | | Cora ↓ Bitcoin | | Citeseer ↓ Cora | | Citeseer ↓ Bitcoin | | Average Rank |
|---|---|---|---|---|---|---|---|---|---|---|
| Contr_l + PR_L | SAGE | 91.10 | ± 7.24 | 68.17 | ± 3.23 | 88.08 | ± 7.36 | 46.78 | ± 1.06 | 127.0 |
| Contr_l + PR_L + Triplet_L | ALL | 74.34 | ± 11.40 | 47.48 | ± 17.57 | 84.89 | ± 2.55 | 34.65 | ± 0.02 | 166.5 |
| Contr_l + PR_L + Triplet_L | GAT | 98.75 | ± 0.20 | 89.05 | ± 0.66 | 97.17 | ± 0.63 | 87.96 | ± 1.06 | 31.0 |
| Contr_l + PR_L + Triplet_L | GCN | 97.86 | ± 0.29 | 88.56 | ± 1.00 | 95.56 | ± 0.36 | 89.17 | ± 0.68 | 48.5 |
| Contr_l + PR_L + Triplet_L | GIN | 92.43 | ± 2.01 | 68.81 | ± 2.37 | 86.11 | ± 3.66 | 68.74 | ± 1.02 | 122.5 |
| Contr_l + PR_L + Triplet_L | MPNN | 96.04 | ± 2.37 | 88.07 | ± 0.39 | 87.19 | ± 3.53 | 87.29 | ± 0.49 | 86.0 |
| Contr_l + PR_L + Triplet_L | PAGNN | 71.36 | ± 2.93 | 34.71 | ± 0.06 | 70.02 | ± 7.26 | 35.92 | ± 0.39 | 178.625 |
| Contr_l + PR_L + Triplet_L | SAGE | 96.17 | ± 0.28 | 68.04 | ± 3.72 | 93.90 | ± 0.76 | 68.25 | ± 2.17 | 105.25 |
| Contr_l + Triplet_L | ALL | 94.45 | ± 1.00 | 66.21 | ± 2.25 | 90.04 | ± 0.92 | 67.26 | ± 1.33 | 117.25 |
| Contr_l + Triplet_L | GAT | 98.97 | ± 0.10 | 89.04 | ± 0.84 | 97.65 | ± 0.31 | 88.73 | ± 0.92 | 19.875 |
| Contr_l + Triplet_L | GCN | 98.41 | ± 0.10 | 88.73 | ± 0.51 | 96.45 | ± 0.50 | 89.18 | ± 0.48 | 29.375 |
| Contr_l + Triplet_L | GIN | 94.97 | ± 0.61 | 70.60 | ± 1.00 | 91.82 | ± 1.80 | 73.61 | ± 0.98 | 104.0 |
| Contr_l + Triplet_L | MPNN | 98.09 | ± 0.27 | 86.93 | ± 1.54 | 94.53 | ± 0.43 | 87.90 | ± 0.47 | 68.375 |
| Contr_l + Triplet_L | PAGNN | 82.57 | ± 6.33 | 35.73 | ± 0.37 | 86.60 | ± 1.09 | 37.97 | ± 5.73 | 152.25 |
| Contr_l + Triplet_L | SAGE | 97.02 | ± 0.30 | 71.11 | ± 3.46 | 94.82 | ± 0.71 | 73.04 | ± 2.34 | 91.875 |
| CrossE_L | ALL | 85.81 | ± 19.03 | 81.97 | ± 0.69 | 55.74 | ± 6.48 | 81.98 | ± 0.19 | 132.5 |

Continued on next page

Table 26. Results for Lp Accuracy (↑) (continued)

| Loss Type | Model | Cora ↓ Citeseer | | Cora ↓ Bitcoin | | Citeseer ↓ Cora | | Citeseer ↓ Bitcoin | | Average Rank |
|---|---|---|---|---|---|---|---|---|---|---|
| CrossE_L | GAT | 85.08 | ± 21.17 | 42.93 | ± 1.67 | 81.24 | ± 17.34 | 85.59 | ± 0.29 | 135.0 |
| CrossE_L | GCN | 52.23 | ± 10.25 | 75.98 | ± 0.67 | 51.35 | ± 0.00 | 34.64 | ± 0.00 | 182.75 |
| CrossE_L | GIN | 47.67 | ± 0.02 | 34.64 | ± 0.00 | 51.35 | ± 0.00 | 34.64 | ± 0.00 | 208.0 |
| CrossE_L | MPNN | 88.09 | ± 2.91 | 34.65 | ± 0.02 | 85.69 | ± 1.72 | 34.64 | ± 0.00 | 164.25 |
| CrossE_L | PAGNN | 69.51 | ± 2.20 | 34.64 | ± 0.00 | 57.97 | ± 8.60 | 34.65 | ± 0.01 | 197.375 |
| CrossE_L | SAGE | 74.42 | ± 6.68 | 48.27 | ± 0.58 | 60.00 | ± 6.88 | 35.42 | ± 0.47 | 174.5 |
| CrossE_L + PMI_L | ALL | 77.41 | ± 2.27 | 82.57 | ± 0.18 | 85.87 | ± 0.96 | 82.64 | ± 0.39 | 117.75 |
| CrossE_L + PMI_L | GAT | 99.03 | ± 0.14 | 89.19 | ± 0.10 | 97.56 | ± 0.24 | 89.18 | ± 0.18 | 12.375 |
| CrossE_L + PMI_L | GCN | 98.08 | ± 0.23 | 88.61 | ± 0.69 | 95.91 | ± 0.40 | 88.58 | ± 0.54 | 50.5 |
| CrossE_L + PMI_L | GIN | 81.79 | ± 2.82 | 40.74 | ± 13.62 | 77.84 | ± 2.08 | 47.21 | ± 17.03 | 158.5 |
| CrossE_L + PMI_L | MPNN | 98.25 | ± 0.15 | 88.60 | ± 0.44 | 93.43 | ± 1.28 | 88.23 | ± 0.39 | 58.5 |
| CrossE_L + PMI_L | PAGNN | 64.56 | ± 3.49 | 46.19 | ± 0.46 | 64.58 | ± 0.65 | 34.66 | ± 0.01 | 186.125 |
| CrossE_L + PMI_L | SAGE | 75.92 | ± 1.48 | 47.50 | ± 3.27 | 73.53 | ± 0.91 | 41.04 | ± 6.03 | 165.25 |
| CrossE_L + PMI_L + PR_L | ALL | 55.85 | ± 0.39 | 82.56 | ± 0.28 | 83.86 | ± 0.96 | 82.75 | ± 0.42 | 132.75 |
| CrossE_L + PMI_L + PR_L | GAT | 98.99 | ± 0.17 | 89.56 | ± 0.21 | 93.74 | ± 3.16 | 84.09 | ± 1.45 | 43.625 |
| CrossE_L + PMI_L + PR_L | GCN | 98.25 | ± 0.31 | 88.65 | ± 0.67 | 95.93 | ± 0.23 | 89.00 | ± 0.47 | 40.875 |

Table 26. Results for Lp Accuracy (↑) (continued)

| Loss Type | Model | Cora ↓ Citeseer | ± | Cora ↓ Bitcoin | ± | Citeseer ↓ Cora | ± | Citeseer ↓ Bitcoin | ± | Average Rank |
|---|---|---|---|---|---|---|---|---|---|---|
| CrossE_L + PMI_L + PR_L | GIN | 79.57 2.65 | ± | 70.20 0.76 | ± | 78.20 2.19 | ± | 41.10 14.36 | ± | 150.75 |
| CrossE_L + PMI_L + PR_L | MPNN | 98.35 0.25 | ± | 88.11 0.33 | ± | 88.49 5.73 | ± | 87.86 0.61 | ± | 69.875 |
| CrossE_L + PMI_L + PR_L | PAGNN | 66.42 2.50 | ± | 34.80 0.15 | ± | 63.25 2.72 | ± | 34.66 0.01 | ± | 191.5 |
| CrossE_L + PMI_L + PR_L | SAGE | 76.28 1.43 | ± | 46.19 1.61 | ± | 73.87 1.10 | ± | 47.75 1.56 | ± | 161.625 |
| CrossE_L + PMI_L + PR_L + Triplet_L | ALL | 74.16 2.55 | ± | 81.38 0.56 | ± | 86.04 0.53 | ± | 76.29 0.33 | ± | 125.5 |
| CrossE_L + PMI_L + PR_L + Triplet_L | GAT | 98.98 0.06 | ± | 89.49 0.28 | ± | 96.60 1.11 | ± | 88.80 0.70 | ± | 18.25 |
| CrossE_L + PMI_L + PR_L + Triplet_L | GCN | 98.19 0.31 | ± | 88.88 0.64 | ± | 95.74 0.52 | ± | 88.51 0.42 | ± | 46.125 |
| CrossE_L + PMI_L + PR_L + Triplet_L | GIN | 82.20 3.01 | ± | 51.53 15.44 | ± | 82.05 2.96 | ± | 41.25 14.62 | ± | 153.25 |
| CrossE_L + PMI_L + PR_L + Triplet_L | MPNN | 98.33 0.08 | ± | 88.53 0.93 | ± | 90.19 2.95 | ± | 88.22 0.53 | ± | 63.375 |
| CrossE_L + PMI_L + PR_L + Triplet_L | PAGNN | 68.94 1.17 | ± | 46.20 0.08 | ± | 64.23 0.82 | ± | 34.75 0.21 | ± | 180.625 |
| CrossE_L + PMI_L + PR_L + Triplet_L | SAGE | 88.73 5.10 | ± | 67.10 3.29 | ± | 87.05 1.05 | ± | 67.79 2.17 | ± | 125.75 |
| CrossE_L + PMI_L + Triplet_L | ALL | 96.03 0.35 | ± | 82.13 0.62 | ± | 91.78 1.36 | ± | 82.43 0.37 | ± | 91.5 |
| CrossE_L + PMI_L + Triplet_L | GAT | 99.01 0.10 | ± | 89.24 1.10 | ± | 97.48 0.23 | ± | 89.42 0.34 | ± | 10.5 |
| CrossE_L + PMI_L + Triplet_L | GCN | 98.23 0.17 | ± | 89.20 0.33 | ± | 96.04 0.21 | ± | 88.62 0.82 | ± | 37.25 |
| CrossE_L + PMI_L + Triplet_L | GIN | 84.41 1.54 | ± | 68.87 0.70 | ± | 85.14 1.12 | ± | 69.33 0.57 | ± | 130.75 |
| CrossE_L + PMI_L + Triplet_L | MPNN | 98.25 0.29 | ± | 88.12 0.59 | ± | 94.09 0.61 | ± | 88.80 0.70 | ± | 54.25 |

Table 26. Results for Lp Accuracy (↑) (continued)

| Loss Type | Model | Cora ↓ Citeseer | | Cora ↓ Bitcoin | | Citeseer ↓ Cora | | Citeseer ↓ Bitcoin | | Average Rank |
|-----------|-------|---------|---|---------|---|---------|---|---------|---|--------------|
| CrossE_L + PMI_L + Triplet_L | PAGNN | 67.67 | ± 1.68 | 34.86 | ± 0.11 | 66.22 | ± 2.02 | 34.84 | ± 0.24 | 183.75 |
| CrossE_L + PMI_L + Triplet_L | SAGE | 92.80 | ± 0.98 | 68.93 | ± 1.73 | 89.91 | ± 0.79 | 73.11 | ± 1.36 | 112.25 |
| CrossE_L + PR_L | ALL | 58.55 | ± 5.16 | 81.89 | ± 0.39 | 67.65 | ± 8.38 | 82.37 | ± 0.38 | 144.0 |
| CrossE_L + PR_L | GAT | 98.11 | ± 0.37 | 87.38 | ± 0.65 | 96.18 | ± 0.33 | 87.32 | ± 0.50 | 60.75 |
| CrossE_L + PR_L | GCN | 94.90 | ± 1.75 | 87.04 | ± 0.71 | 94.60 | ± 1.01 | 86.74 | ± 1.25 | 79.5 |
| CrossE_L + PR_L | GIN | 86.21 | ± 1.82 | 45.59 | ± 6.13 | 63.04 | ± 13.29 | 37.30 | ± 5.92 | 166.25 |
| CrossE_L + PR_L | MPNN | 93.79 | ± 4.62 | 88.69 | ± 0.32 | 80.84 | ± 3.62 | 89.31 | ± 0.45 | 78.5 |
| CrossE_L + PR_L | PAGNN | 63.33 | ± 8.88 | 34.64 | ± 0.00 | 51.56 | ± 0.10 | 34.64 | ± 0.01 | 204.625 |
| CrossE_L + PR_L | SAGE | 74.34 | ± 1.17 | 45.62 | ± 1.48 | 71.26 | ± 1.93 | 45.88 | ± 1.15 | 167.625 |
| CrossE_L + PR_L + Triplet_L | ALL | 71.08 | ± 12.53 | 34.64 | ± 0.00 | 78.35 | ± 2.88 | 34.65 | ± 0.01 | 186.375 |
| CrossE_L + PR_L + Triplet_L | GAT | 98.40 | ± 0.43 | 87.83 | ± 0.77 | 96.93 | ± 0.73 | 89.48 | ± 0.45 | 33.75 |
| CrossE_L + PR_L + Triplet_L | GCN | 97.74 | ± 0.23 | 88.66 | ± 0.47 | 95.16 | ± 0.77 | 89.86 | ± 0.71 | 43.25 |
| CrossE_L + PR_L + Triplet_L | GIN | 91.79 | ± 2.89 | 51.91 | ± 15.84 | 85.24 | ± 1.84 | 68.37 | ± 0.85 | 130.25 |
| CrossE_L + PR_L + Triplet_L | MPNN | 95.33 | ± 2.01 | 88.84 | ± 0.55 | 83.62 | ± 3.39 | 88.94 | ± 0.62 | 75.125 |
| CrossE_L + PR_L + Triplet_L | PAGNN | 73.41 | ± 2.44 | 46.34 | ± 0.40 | 59.76 | ± 5.45 | 40.17 | ± 5.43 | 176.5 |
| CrossE_L + PR_L + Triplet_L | SAGE | 96.85 | ± 0.77 | 69.00 | ± 3.90 | 94.59 | ± 1.19 | 68.08 | ± 3.09 | 100.5 |

Table 26. Results for Lp Accuracy (↑) (continued)

| Loss Type | Model | Cora ↓ Citeseer | | Cora ↓ Bitcoin | | Citeseer ↓ Cora | | Citeseer ↓ Bitcoin | | Average Rank |
|---|---|---|---|---|---|---|---|---|---|---|
| CrossE_L + Triplet_L | ALL | 96.67 0.10 | ± | 74.40 0.97 | ± | 91.95 1.34 | ± | 73.45 0.72 | ± | 97.875 |
| CrossE_L + Triplet_L | GAT | 99.16 0.10 | ± | 89.78 0.40 | ± | 97.67 0.28 | ± | 89.79 0.74 | ± | 2.25 |
| CrossE_L + Triplet_L | GCN | 98.52 0.19 | ± | 88.85 0.32 | ± | 96.60 0.28 | ± | 88.68 0.77 | ± | 30.875 |
| CrossE_L + Triplet_L | GIN | 95.59 0.45 | ± | 71.12 0.82 | ± | 91.40 0.89 | ± | 72.89 0.66 | ± | 103.25 |
| CrossE_L + Triplet_L | MPNN | 98.36 0.20 | ± | 88.50 0.49 | ± | 94.55 0.56 | ± | 87.53 0.47 | ± | 58.0 |
| CrossE_L + Triplet_L | PAGNN | 84.15 6.02 | ± | 34.65 0.01 | ± | 86.45 1.45 | ± | 34.69 0.02 | ± | 161.375 |
| CrossE_L + Triplet_L | SAGE | 97.93 0.27 | ± | 75.56 2.11 | ± | 95.75 0.39 | ± | 74.85 1.36 | ± | 83.25 |
| PMI_L | ALL | 71.91 1.95 | ± | 83.07 0.38 | ± | 86.39 1.08 | ± | 82.79 0.30 | ± | 119.25 |
| PMI_L | GAT | 99.14 0.13 | ± | 89.44 0.95 | ± | 97.53 0.31 | ± | 89.74 0.39 | ± | 5.75 |
| PMI_L | GCN | 98.06 0.18 | ± | 88.51 0.20 | ± | 95.71 0.25 | ± | 89.10 0.62 | ± | 48.25 |
| PMI_L | GIN | 79.83 2.30 | ± | 34.67 0.04 | ± | 78.56 1.53 | ± | 34.75 0.13 | ± | 172.375 |
| PMI_L | MPNN | 98.51 0.21 | ± | 88.04 0.78 | ± | 94.11 0.47 | ± | 88.07 0.70 | ± | 56.875 |
| PMI_L | PAGNN | 66.14 2.35 | ± | 34.64 0.00 | ± | 66.35 2.16 | ± | 34.84 0.25 | ± | 190.5 |
| PMI_L | SAGE | 75.51 2.01 | ± | 48.65 3.39 | ± | 71.70 1.47 | ± | 44.06 2.45 | ± | 164.5 |
| PMI_L + PR_L | ALL | 57.12 2.46 | ± | 82.42 0.52 | ± | 83.44 0.71 | ± | 82.60 0.49 | ± | 134.25 |
| PMI_L + PR_L | GAT | 98.63 0.40 | ± | 88.62 1.00 | ± | 90.89 7.49 | ± | 87.57 1.50 | ± | 58.0 |

Continued on next page

Table 26. Results for Lp Accuracy (↑) (continued)

| Loss Type | Model | Cora ↓ Citeseer | | Cora ↓ Bitcoin | | Citeseer ↓ Cora | | Citeseer ↓ Bitcoin | | Average Rank |
|---|---|---|---|---|---|---|---|---|---|---|
| PMI_L + PR_L | GCN | 98.13 | ± 0.30 | 89.28 | ± 0.77 | 95.72 | ± 0.86 | 88.22 | ± 0.86 | 43.375 |
| PMI_L + PR_L | GIN | 78.58 | ± 3.43 | 34.65 | ± 0.02 | 76.92 | ± 7.28 | 55.82 | ± 19.28 | 165.625 |
| PMI_L + PR_L | MPNN | 98.30 | ± 0.34 | 88.19 | ± 0.71 | 84.54 | ± 2.18 | 85.91 | ± 1.14 | 83.25 |
| PMI_L + PR_L | PAGNN | 66.47 | ± 2.05 | 34.68 | ± 0.08 | 63.09 | ± 0.99 | 34.76 | ± 0.22 | 189.75 |
| PMI_L + PR_L | SAGE | 76.57 | ± 1.04 | 44.40 | ± 4.99 | 73.08 | ± 0.95 | 45.14 | ± 3.08 | 166.5 |
| PMI_L + PR_L + Triplet_L | ALL | 75.81 | ± 1.48 | 65.22 | ± 17.31 | 86.66 | ± 1.29 | 63.91 | ± 16.36 | 139.75 |
| PMI_L + PR_L + Triplet_L | GAT | 98.92 | ± 0.14 | 89.60 ± 0.60 | | 96.80 | ± 0.79 | 88.53 | ± 0.75 | 22.0 |
| PMI_L + PR_L + Triplet_L | GCN | 98.32 | ± 0.12 | 89.23 | ± 1.03 | 95.90 | ± 0.30 | 88.35 | ± 0.72 | 38.875 |
| PMI_L + PR_L + Triplet_L | GIN | 83.05 | ± 3.35 | 60.03 | ± 14.25 | 81.85 | ± 2.22 | 34.84 | ± 0.07 | 155.75 |
| PMI_L + PR_L + Triplet_L | MPNN | 98.45 | ± 0.07 | 88.55 | ± 0.46 | 87.58 | ± 1.23 | 86.90 | ± 1.01 | 68.5 |
| PMI_L + PR_L + Triplet_L | PAGNN | 67.46 | ± 2.61 | 34.88 | ± 0.08 | 64.47 | ± 1.83 | 34.86 | ± 0.29 | 184.5 |
| PMI_L + PR_L + Triplet_L | SAGE | 91.35 | ± 3.17 | 70.83 | ± 2.34 | 89.41 | ± 1.42 | 68.35 | ± 1.80 | 115.125 |
| PMI_L + Triplet_L | ALL | 95.63 | ± 0.79 | 74.86 | ± 1.40 | 91.95 | ± 0.84 | 83.16 | ± 0.42 | 92.875 |
| PMI_L + Triplet_L | GAT | 99.05 | ± 0.09 | 89.20 | ± 0.79 | 97.51 | ± 0.16 | 89.48 | ± 0.35 | 9.25 |
| PMI_L + Triplet_L | GCN | 98.27 | ± 0.05 | 88.98 | ± 0.78 | 96.09 | ± 0.41 | 88.56 | ± 0.38 | 38.75 |
| PMI_L + Triplet_L | GIN | 88.11 | ± 1.30 | 59.39 | ± 13.85 | 84.36 | ± 1.53 | 69.61 | ± 0.71 | 133.5 |

Table 26. Results for Lp Accuracy (↑) (continued)

| Loss Type | Model | Cora ↓ Citeseer | | Cora ↓ Bitcoin | | Citeseer ↓ Cora | | Citeseer ↓ Bitcoin | | Average Rank |
|---|---|---|---|---|---|---|---|---|---|---|
| PMI_L + Triplet_L | MPNN | 98.32 | ± 0.26 | 88.20 | ± 0.66 | 93.76 | ± 0.66 | 88.67 | ± 0.67 | 55.875 |
| PMI_L + Triplet_L | PAGNN | 65.91 | ± 2.52 | 34.65 | ± 0.02 | 67.31 | ± 2.63 | 34.66 | ± 0.02 | 191.125 |
| PMI_L + Triplet_L | SAGE | 90.46 | ± 1.90 | 59.70 | ± 4.88 | 90.49 | ± 0.60 | 71.88 | ± 1.14 | 118.875 |
| PR_L | ALL | 51.41 | ± 2.15 | 82.08 | ± 0.24 | 58.08 | ± 5.14 | 82.84 | ± 0.29 | 148.75 |
| PR_L | GAT | 97.87 | ± 0.49 | 86.33 | ± 0.62 | 96.29 | ± 0.51 | 87.24 | ± 0.93 | 64.75 |
| PR_L | GCN | 96.55 | ± 0.74 | 86.70 | ± 0.48 | 94.46 | ± 0.68 | 86.09 | ± 1.74 | 78.25 |
| PR_L | GIN | 85.85 | ± 2.28 | 37.43 | ± 6.23 | 70.56 | ± 10.69 | 42.92 | ± 7.57 | 161.375 |
| PR_L | MPNN | 92.56 | ± 6.05 | 88.72 | ± 0.53 | 82.40 | ± 2.70 | 88.96 | ± 0.31 | 80.875 |
| PR_L | PAGNN | 59.59 | ± 10.86 | 34.66 | ± 0.04 | 51.80 | ± 0.14 | 34.76 | ± 0.00 | 196.625 |
| PR_L | SAGE | 74.37 | ± 1.46 | 46.20 | ± 1.45 | 73.03 | ± 2.19 | 46.53 | ± 1.87 | 164.875 |
| PR_L + Triplet_L | ALL | 58.00 | ± 4.29 | 68.30 | ± 18.87 | 61.17 | ± 3.00 | 81.84 | ± 0.25 | 158.625 |
| PR_L + Triplet_L | GAT | 98.02 | ± 1.28 | 84.47 | ± 0.78 | 96.08 | ± 1.05 | 87.40 | ± 0.44 | 64.25 |
| PR_L + Triplet_L | GCN | 97.03 | ± 0.59 | 87.80 | ± 0.94 | 94.82 | ± 0.82 | 88.20 | ± 0.09 | 67.375 |
| PR_L + Triplet_L | GIN | 85.85 | ± 3.91 | 34.64 | ± 0.00 | 68.17 | ± 12.23 | 34.67 | ± 0.02 | 176.625 |
| PR_L + Triplet_L | MPNN | 93.11 | ± 5.53 | 89.15 | ± 0.67 | 84.84 | ± 4.10 | 89.00 | ± 0.43 | 71.625 |
| PR_L + Triplet_L | PAGNN | 63.90 | ± 8.75 | 34.79 | ± 0.05 | 51.91 | ± 0.22 | 34.93 | ± 0.20 | 192.75 |

Table 26.  Results for Lp Accuracy (↑) (continued)

| Loss Type | Model | Cora ↓ Citeseer | | Cora ↓ Bitcoin | | Citeseer ↓ Cora | | Citeseer ↓ Bitcoin | | Average Rank |
|---|---|---|---|---|---|---|---|---|---|---|
| PR_L + Triplet_L | SAGE | 78.42 | ± 1.74 | 47.46 | ± 0.63 | 81.89 | ± 9.51 | 45.17 | ± 1.23 | 157.75 |
| Triplet_L | ALL | 96.09 | ± 0.63 | 70.83 | ± 1.67 | 92.00 | ± 1.06 | 83.36 | ± 0.76 | 93.375 |
| Triplet_L | GAT | 99.24 | ± 0.12 | 89.25 | ± 0.46 | 97.83 | ± 0.26 | 89.52 | ± 0.79 | 5.0 |
| Triplet_L | GCN | 98.65 | ± 0.16 | 89.27 | ± 0.74 | 96.46 | ± 0.49 | 89.28 | ± 0.59 | 17.625 |
| Triplet_L | GIN | 96.61 | ± 0.52 | 71.02 | ± 0.68 | 91.45 | ± 1.57 | 74.15 | ± 0.71 | 100.0 |
| Triplet_L | MPNN | 98.49 | ± 0.24 | 88.71 | ± 0.64 | 94.68 | ± 0.47 | 88.79 | ± 1.31 | 40.75 |
| Triplet_L | PAGNN | 84.72 | ± 5.90 | 35.19 | ± 0.42 | 88.12 | ± 0.48 | 49.02 | ± 0.17 | 141.75 |
| Triplet_L | SAGE | 97.69 | ± 0.38 | 77.76 | ± 1.48 | 95.40 | ± 0.48 | 78.48 | ± 1.21 | 84.5 |

Table 27.  Lp Aupr Performance (↑): This table presents models (Loss function and GNN) ranked by their average performance in terms of lp aupr. Top-ranked results are highlighted in red, second-ranked in blue, and third-ranked in green.

| Loss Type | Model | Cora ↓ Citeseer | | Cora ↓ Bitcoin | | Citeseer ↓ Cora | | Citeseer ↓ Bitcoin | | Average Rank |
|---|---|---|---|---|---|---|---|---|---|---|
| Contr_l | ALL | 96.65 | ± 0.95 | 49.85 | ± 2.48 | 94.03 | ± 0.64 | 51.16 | ± 3.01 | 121.75 |
| Contr_l | GAT | 99.87 | ± 0.03 | 91.65 | ± 0.96 | 99.70 | ± 0.03 | 92.66 | ± 0.49 | 19.375 |
| Contr_l | GCN | 99.77 | ± 0.03 | 92.11 | ± 0.53 | 99.45 | ± 0.13 | 92.12 | ± 0.40 | 30.625 |

Table 27. Results for Lp Aupr (↑) (continued)

| Loss Type | Model | Cora ↓ Citeseer | | Cora ↓ Bitcoin | | Citeseer ↓ Cora | | Citeseer ↓ Bitcoin | | Average Rank |
|---|---|---|---|---|---|---|---|---|---|---|
| Contr_l | GIN | 98.13 | ± | 46.66 | ± | 96.01 | ± | 52.89 | ± | 115.5 |
| | | 0.28 | | 0.31 | | 0.48 | | 1.28 | | |
| Contr_l | MPNN | 99.37 | ± | 90.16 | ± | 96.65 | ± | 90.68 | ± | 76.875 |
| | | 0.13 | | 1.05 | | 0.66 | | 0.78 | | |
| Contr_l | PAGNN | 89.30 | ± | 31.67 | ± | 91.36 | ± | 32.26 | ± | 165.625 |
| | | 2.64 | | 0.37 | | 0.86 | | 0.09 | | |
| Contr_l | SAGE | 99.18 | ± | 59.42 | ± | 98.65 | ± | 60.09 | ± | 92.875 |
| | | 0.09 | | 1.22 | | 0.52 | | 0.78 | | |
| Contr_l + CrossE_L | ALL | 97.03 | ± | 62.92 | ± | 93.17 | ± | 44.19 | ± | 120.0 |
| | | 0.59 | | 1.06 | | 0.99 | | 1.54 | | |
| Contr_l + CrossE_L | GAT | 99.88 | ± | 91.59 | ± | 99.60 | ± | 91.70 | ± | 31.125 |
| | | 0.02 | | 0.72 | | 0.07 | | 0.59 | | |
| Contr_l + CrossE_L | GCN | 99.79 | ± | 91.58 | ± | 99.38 | ± | 92.19 | ± | 34.875 |
| | | 0.04 | | 0.91 | | 0.10 | | 0.25 | | |
| Contr_l + CrossE_L | GIN | 96.17 | ± | 41.63 | ± | 95.80 | ± | 45.72 | ± | 128.625 |
| | | 1.46 | | 0.50 | | 0.54 | | 0.19 | | |
| Contr_l + CrossE_L | MPNN | 99.35 | ± | 90.45 | ± | 96.90 | ± | 90.42 | ± | 76.375 |
| | | 0.08 | | 1.16 | | 0.32 | | 0.85 | | |
| Contr_l + CrossE_L | PAGNN | 87.41 | ± | 31.55 | ± | 90.53 | ± | 31.27 | ± | 172.25 |
| | | 1.77 | | 0.23 | | 0.82 | | 0.18 | | |
| Contr_l + CrossE_L | SAGE | 99.04 | ± | 49.97 | ± | 98.85 | ± | 48.25 | ± | 104.75 |
| | | 0.13 | | 1.48 | | 0.36 | | 1.15 | | |
| Contr_l + CrossE_L + PMI_L | ALL | 88.11 | ± | 68.10 | ± | 92.70 | ± | 72.76 | ± | 120.25 |
| | | 4.62 | | 1.28 | | 0.87 | | 1.01 | | |
| Contr_l + CrossE_L + PMI_L | GAT | **99.91** | **±** | 92.42 | ± | 99.66 | ± | 92.32 | ± | 11.25 |
| | | **0.04** | | 0.61 | | 0.05 | | 0.35 | | |
| Contr_l + CrossE_L + PMI_L | GCN | 99.79 | ± | 92.44 | ± | 99.30 | ± | 91.91 | ± | 30.375 |
| | | 0.02 | | 0.31 | | 0.06 | | 0.57 | | |
| Contr_l + CrossE_L + PMI_L | GIN | 92.36 | ± | 41.96 | ± | 90.28 | ± | 42.27 | ± | 151.0 |
| | | 2.03 | | 0.73 | | 1.14 | | 0.72 | | |
| Contr_l + CrossE_L + PMI_L | MPNN | 99.48 | ± | 91.78 | ± | 97.03 | ± | 91.44 | ± | 60.625 |
| | | 0.13 | | 0.85 | | 0.33 | | 0.57 | | |

Table 27.  Results for Lp Aupr (↑) (continued)

| Loss Type | Model | Cora ↓ Citeseer | | Cora ↓ Bitcoin | | Citeseer ↓ Cora | | Citeseer ↓ Bitcoin | | Average Rank |
|---|---|---|---|---|---|---|---|---|---|---|
| Contr_l + CrossE_L + PMI_L | PAGNN | 70.64 | ± 1.87 | 30.84 | ± 0.26 | 79.40 | ± 1.71 | 29.70 | ± 0.18 | 198.5 |
| Contr_l + CrossE_L + PMI_L | SAGE | 87.38 | ± 0.79 | 38.20 | ± 0.52 | 87.60 | ± 2.24 | 41.37 | ± 0.45 | 162.25 |
| Contr_l + CrossE_L + PMI_L + PR_L | ALL | 70.07 | ± 6.52 | 72.41 | ± 0.52 | 91.44 | ± 0.80 | 73.20 | ± 0.42 | 134.5 |
| Contr_l + CrossE_L + PMI_L + PR_L | GAT | 99.89 | ± 0.06 | 92.19 | ± 1.36 | 99.66 | ± 0.04 | 91.95 | ± 1.44 | 21.0 |
| Contr_l + CrossE_L + PMI_L + PR_L | GCN | 99.78 | ± 0.04 | 92.29 | ± 0.54 | 99.30 | ± 0.12 | 91.83 | ± 0.97 | 34.5 |
| Contr_l + CrossE_L + PMI_L + PR_L | GIN | 87.86 | ± 1.98 | 40.27 | ± 0.77 | 85.67 | ± 1.72 | 39.34 | ± 0.76 | 163.75 |
| Contr_l + CrossE_L + PMI_L + PR_L | MPNN | 99.49 | ± 0.09 | 90.73 | ± 0.89 | 94.78 | ± 1.48 | 88.08 | ± 2.37 | 81.875 |
| Contr_l + CrossE_L + PMI_L + PR_L | PAGNN | 72.00 | ± 2.36 | 31.19 | ± 0.15 | 77.65 | ± 3.22 | 31.41 | ± 0.12 | 194.625 |
| Contr_l + CrossE_L + PMI_L + PR_L | SAGE | 86.25 | ± 1.70 | 36.28 | ± 0.54 | 85.68 | ± 2.21 | 37.50 | ± 1.28 | 169.0 |
| Contr_l + CrossE_L + PMI_L + PR_L + Triplet_L | ALL | 83.24 | ± 3.27 | 65.34 | ± 1.11 | 92.79 | ± 0.78 | 68.57 | ± 1.88 | 127.625 |
| Contr_l + CrossE_L + PMI_L + PR_L + Triplet_L | GAT | 99.90 | ± 0.03 | 91.86 | ± 1.04 | 99.51 | ± 0.19 | 92.00 | ± 0.78 | 25.5 |
| Contr_l + CrossE_L + PMI_L + PR_L + Triplet_L | GCN | 99.79 | ± 0.04 | 92.23 | ± 0.78 | 99.30 | ± 0.06 | 92.31 | ± 0.55 | 29.125 |
| Contr_l + CrossE_L + PMI_L + PR_L + Triplet_L | GIN | 89.76 | ± 2.30 | 43.16 | ± 0.98 | 91.30 | ± 0.64 | 41.26 | ± 0.55 | 151.5 |
| Contr_l + CrossE_L + PMI_L + PR_L + Triplet_L | MPNN | 99.43 | ± 0.09 | 91.59 | ± 0.15 | 95.37 | ± 1.20 | 88.12 | ± 1.11 | 77.375 |

Table 27. Results for Lp Aupr (↑) (continued)

| Loss Type | Model | Cora ↓ Citeseer | | Cora ↓ Bitcoin | | Citeseer ↓ Cora | | Citeseer ↓ Bitcoin | | Average Rank |
|---|---|---|---|---|---|---|---|---|---|---|
| Contr_l + CrossE_L + PMI_L + PR_L + Triplet_L | PAGNN | 71.61 1.23 | ± | 30.58 0.24 | ± | 78.54 0.92 | ± | 30.61 0.23 | ± | 197.0 |
| Contr_l + CrossE_L + PMI_L + PR_L + Triplet_L | SAGE | 95.23 2.46 | ± | 53.26 1.68 | ± | 96.42 0.24 | ± | 51.00 1.42 | ± | 115.875 |
| Contr_l + CrossE_L + PMI_L + Triplet_L | ALL | 98.34 0.45 | ± | 73.74 0.70 | ± | 96.05 0.44 | ± | 76.02 0.72 | ± | 95.0 |
| Contr_l + CrossE_L + PMI_L + Triplet_L | GAT | 99.91 0.02 | ± | 92.69 0.23 | ± | 99.63 0.03 | ± | 92.45 0.40 | ± | 9.5 |
| Contr_l + CrossE_L + PMI_L + Triplet_L | GCN | 99.80 0.05 | ± | 92.97 0.65 | ± | 99.17 0.18 | ± | 92.34 0.67 | ± | 24.125 |
| Contr_l + CrossE_L + PMI_L + Triplet_L | GIN | 93.07 1.40 | ± | 43.27 0.38 | ± | 91.54 1.84 | ± | 44.21 0.84 | ± | 142.75 |
| Contr_l + CrossE_L + PMI_L + Triplet_L | MPNN | 99.49 0.09 | ± | 92.14 0.57 | ± | 96.89 0.29 | ± | 91.26 0.69 | ± | 58.875 |
| Contr_l + CrossE_L + PMI_L + Triplet_L | PAGNN | 71.91 1.37 | ± | 29.95 0.13 | ± | 79.34 1.21 | ± | 30.23 0.15 | ± | 197.75 |
| Contr_l + CrossE_L + PMI_L + Triplet_L | SAGE | 96.46 1.01 | ± | 49.62 1.55 | ± | 95.35 0.73 | ± | 54.76 1.36 | ± | 117.75 |
| Contr_l + CrossE_L + PR_L | ALL | 69.77 2.83 | ± | 66.31 1.80 | ± | 79.31 7.26 | ± | 69.60 0.47 | ± | 149.25 |
| Contr_l + CrossE_L + PR_L | GAT | 99.59 0.46 | ± | 91.46 0.54 | ± | 99.50 0.05 | ± | 90.38 0.92 | ± | 50.5 |
| Contr_l + CrossE_L + PR_L | GCN | 99.60 0.06 | ± | 89.93 0.79 | ± | 98.90 0.17 | ± | 91.27 1.31 | ± | 62.375 |
| Contr_l + CrossE_L + PR_L | GIN | 95.19 1.48 | ± | 48.83 1.12 | ± | 86.19 0.82 | ± | 42.54 1.51 | ± | 145.25 |
| Contr_l + CrossE_L + PR_L | MPNN | 95.14 4.19 | ± | 91.37 0.77 | ± | 91.63 3.06 | ± | 91.57 0.50 | ± | 93.5 |
| Contr_l + CrossE_L + PR_L | PAGNN | 77.69 4.14 | ± | 32.68 0.32 | ± | 68.58 4.80 | ± | 32.10 0.27 | ± | 188.25 |

Table 27. Results for Lp Aupr (↑) (continued)

| Loss Type | Model | Cora ↓ Citeseer | | Cora ↓ Bitcoin | | Citeseer ↓ Cora | | Citeseer ↓ Bitcoin | | Average Rank |
|---|---|---|---|---|---|---|---|---|---|---|
| Contr_l + CrossE_L + PR_L | SAGE | 96.11 5.78 | ± | 52.15 1.07 | ± | 93.63 6.56 | ± | 49.29 1.21 | ± | 123.625 |
| Contr_l + CrossE_L + PR_L + Triplet_L | ALL | 88.70 1.75 | ± | 51.89 1.41 | ± | 92.99 0.59 | ± | 56.08 1.71 | ± | 129.5 |
| Contr_l + CrossE_L + PR_L + Triplet_L | GAT | 99.87 0.03 | ± | 91.40 1.16 | ± | 99.58 0.19 | ± | 92.52 0.36 | ± | 26.0 |
| Contr_l + CrossE_L + PR_L + Triplet_L | GCN | 99.75 0.05 | ± | 91.84 0.94 | ± | 99.16 0.12 | ± | 91.45 1.57 | ± | 47.75 |
| Contr_l + CrossE_L + PR_L + Triplet_L | GIN | 96.67 0.77 | ± | 41.28 0.58 | ± | 94.91 0.44 | ± | 44.65 0.57 | ± | 131.0 |
| Contr_l + CrossE_L + PR_L + Triplet_L | MPNN | 99.22 0.27 | ± | 91.74 0.52 | ± | 95.34 0.41 | ± | 89.01 1.81 | ± | 77.875 |
| Contr_l + CrossE_L + PR_L + Triplet_L | PAGNN | 79.38 2.82 | ± | 31.57 0.13 | ± | 81.73 2.45 | ± | 32.34 0.12 | ± | 182.0 |
| Contr_l + CrossE_L + PR_L + Triplet_L | SAGE | 99.39 0.08 | ± | 56.07 1.09 | ± | 98.34 0.42 | ± | 55.85 1.20 | ± | 95.0 |
| Contr_l + CrossE_L + Triplet_L | ALL | 97.94 0.70 | ± | 59.88 2.03 | ± | 95.63 0.56 | ± | 47.27 1.10 | ± | 112.125 |
| Contr_l + CrossE_L + Triplet_L | GAT | 99.91 0.02 | ± | 91.88 0.80 | ± | 99.67 0.04 | ± | 92.31 0.45 | ± | 17.125 |
| Contr_l + CrossE_L + Triplet_L | GCN | 99.83 0.02 | ± | 92.11 0.75 | ± | 99.41 0.10 | ± | 91.77 0.40 | ± | 31.0 |
| Contr_l + CrossE_L + Triplet_L | GIN | 98.67 0.26 | ± | 47.45 0.70 | ± | 96.47 0.62 | ± | 50.52 0.50 | ± | 113.5 |
| Contr_l + CrossE_L + Triplet_L | MPNN | 99.42 0.09 | ± | 89.92 0.85 | ± | 97.27 0.35 | ± | 91.37 0.81 | ± | 71.25 |
| Contr_l + CrossE_L + Triplet_L | PAGNN | 85.41 7.83 | ± | 31.90 0.21 | ± | 93.27 0.87 | ± | 32.26 0.22 | ± | 163.875 |
| Contr_l + CrossE_L + Triplet_L | SAGE | 99.47 0.10 | ± | 55.25 1.18 | ± | 99.07 0.12 | ± | 58.78 1.07 | ± | 89.5 |
| Contr_l + PMI_L | ALL | 90.72 5.95 | ± | 53.23 4.14 | ± | 94.01 0.71 | ± | 70.89 1.09 | ± | 119.875 |

Table 27. Results for Lp Aupr (↑) (continued)

| Loss Type | Model | Cora ↓ Citeseer | | Cora ↓ Bitcoin | | Citeseer ↓ Cora | | Citeseer ↓ Bitcoin | | Average Rank |
|---|---|---|---|---|---|---|---|---|---|---|
| Contr_l + PMI_L | GAT | 99.90 ± 0.03 | | 92.83 ± 0.80 | | 99.65 ± 0.03 | | 92.27 ± 0.99 | | 12.0 |
| Contr_l + PMI_L | GCN | 99.78 ± 0.05 | | 91.88 ± 0.82 | | 99.30 ± 0.18 | | 92.28 ± 1.04 | | 34.625 |
| Contr_l + PMI_L | GIN | 91.92 ± 3.29 | | 44.53 ± 0.36 | | 89.85 ± 0.91 | | 42.64 ± 0.54 | | 147.5 |
| Contr_l + PMI_L | MPNN | 99.50 ± 0.06 | | 89.98 ± 0.61 | | 97.21 ± 0.13 | | 91.69 ± 0.31 | | 65.0 |
| Contr_l + PMI_L | PAGNN | 71.75 ± 1.77 | | 30.14 ± 0.07 | | 78.30 ± 0.89 | | 29.66 ± 0.11 | | 200.75 |
| Contr_l + PMI_L | SAGE | 87.33 ± 1.98 | | 39.46 ± 1.60 | | 91.80 ± 2.68 | | 43.45 ± 0.63 | | 153.5 |
| Contr_l + PMI_L + PR_L | ALL | 68.57 ± 2.43 | | 68.06 ± 0.89 | | 91.20 ± 0.56 | | 74.11 ± 0.54 | | 137.25 |
| Contr_l + PMI_L + PR_L | GAT | 99.88 ± 0.05 | | 92.53 ± 0.48 | | 98.52 ± 1.51 | | 85.25 ± 1.91 | | 45.5 |
| Contr_l + PMI_L + PR_L | GCN | 99.81 ± 0.05 | | 92.35 ± 0.81 | | 99.29 ± 0.12 | | 91.90 ± 0.76 | | 30.875 |
| Contr_l + PMI_L + PR_L | GIN | 91.05 ± 3.77 | | 45.40 ± 0.68 | | 87.82 ± 2.38 | | 39.90 ± 1.86 | | 153.5 |
| Contr_l + PMI_L + PR_L | MPNN | 99.50 ± 0.07 | | 91.93 ± 0.54 | | 93.74 ± 1.94 | | 87.92 ± 0.70 | | 74.625 |
| Contr_l + PMI_L + PR_L | PAGNN | 70.74 ± 1.25 | | 31.39 ± 0.27 | | 77.76 ± 1.47 | | 31.19 ± 0.19 | | 196.0 |
| Contr_l + PMI_L + PR_L | SAGE | 87.08 ± 0.86 | | 39.52 ± 1.56 | | 89.58 ± 1.51 | | 41.83 ± 0.93 | | 159.75 |
| Contr_l + PMI_L + PR_L + Triplet_L | ALL | 87.51 ± 3.16 | | 50.60 ± 2.36 | | 93.74 ± 0.46 | | 62.32 ± 1.43 | | 126.875 |
| Contr_l + PMI_L + PR_L + Triplet_L | GAT | 99.85 ± 0.07 | | 91.49 ± 0.98 | | 99.23 ± 0.28 | | 88.04 ± 2.59 | | 51.375 |
| Contr_l + PMI_L + PR_L + Triplet_L | GCN | 99.79 ± 0.03 | | 92.75 ± 0.72 | | 99.31 ± 0.12 | | 92.08 ± 0.71 | | 25.5 |

Continued on next page

Table 27. Results for Lp Aupr (↑) (continued)

| Loss Type | Model | Cora ↓ Citeseer | | Cora ↓ Bitcoin | | Citeseer ↓ Cora | | Citeseer ↓ Bitcoin | | Average Rank |
|---|---|---|---|---|---|---|---|---|---|---|
| Contr_l + PMI_L + PR_L + Triplet_L | GIN | 92.93 0.59 | ± | 44.49 0.21 | ± | 93.06 0.79 | ± | 40.97 0.75 | ± | 142.0 |
| Contr_l + PMI_L + PR_L + Triplet_L | MPNN | 99.50 0.14 | ± | 91.10 0.48 | ± | 94.94 0.36 | ± | 87.10 1.27 | ± | 79.75 |
| Contr_l + PMI_L + PR_L + Triplet_L | PAGNN | 73.18 1.66 | ± | 31.52 0.27 | ± | 78.41 2.52 | ± | 31.02 0.21 | ± | 191.25 |
| Contr_l + PMI_L + PR_L + Triplet_L | SAGE | 98.10 0.53 | ± | 58.05 1.73 | ± | 97.02 0.95 | ± | 52.05 1.13 | ± | 104.875 |
| Contr_l + PR_L | ALL | 72.35 3.42 | ± | 70.01 0.59 | ± | 82.96 3.27 | ± | 71.16 1.06 | ± | 141.75 |
| Contr_l + PR_L | GAT | 99.72 0.15 | ± | 90.87 0.28 | ± | 99.47 0.10 | ± | 90.20 0.79 | ± | 52.375 |
| Contr_l + PR_L | GCN | 99.62 0.11 | ± | 91.07 0.52 | ± | 98.92 0.24 | ± | 91.70 0.86 | ± | 54.25 |
| Contr_l + PR_L | GIN | 93.81 3.92 | ± | 50.78 0.51 | ± | 84.77 4.99 | ± | 40.37 0.73 | ± | 148.0 |
| Contr_l + PR_L | MPNN | 95.43 1.59 | ± | 91.42 0.99 | ± | 93.47 1.55 | ± | 92.09 0.49 | ± | 80.875 |
| Contr_l + PR_L | PAGNN | 75.28 4.31 | ± | 32.73 0.32 | ± | 70.75 0.85 | ± | 32.42 0.22 | ± | 187.0 |
| Contr_l + PR_L | SAGE | 96.17 5.56 | ± | 54.03 1.10 | ± | 95.12 5.68 | ± | 40.72 1.83 | ± | 125.375 |
| Contr_l + PR_L + Triplet_L | ALL | 84.04 9.73 | ± | 52.64 1.70 | ± | 92.54 1.77 | ± | 54.39 2.20 | ± | 136.5 |
| Contr_l + PR_L + Triplet_L | GAT | 99.87 0.04 | ± | 92.41 0.67 | ± | 99.59 0.16 | ± | 90.83 1.06 | ± | 28.625 |
| Contr_l + PR_L + Triplet_L | GCN | 99.72 0.07 | ± | 91.74 1.02 | ± | 99.20 0.08 | ± | 92.40 0.77 | ± | 39.0 |
| Contr_l + PR_L + Triplet_L | GIN | 97.54 0.94 | ± | 46.12 0.86 | ± | 93.47 2.09 | ± | 42.82 0.65 | ± | 129.625 |
| Contr_l + PR_L + Triplet_L | MPNN | 98.77 1.05 | ± | 91.04 0.34 | ± | 94.04 2.61 | ± | 90.28 0.37 | ± | 85.25 |

Table 27. Results for Lp Aupr (↑) (continued)

| Loss Type | Model | Cora ↓ Citeseer | | Cora ↓ Bitcoin | | Citeseer ↓ Cora | | Citeseer ↓ Bitcoin | | Average Rank |
|---|---|---|---|---|---|---|---|---|---|---|
| Contr_l + PR_L + Triplet_L | PAGNN | 77.89 3.00 | ± | 32.00 0.13 | ± | 82.75 5.75 | ± | 32.61 0.11 | ± | 179.75 |
| Contr_l + PR_L + Triplet_L | SAGE | 99.30 0.11 | ± | 58.55 0.88 | ± | 98.65 0.28 | ± | 56.55 1.13 | ± | 93.625 |
| Contr_l + Triplet_L | ALL | 98.38 0.31 | ± | 46.67 1.79 | ± | 95.63 0.67 | ± | 45.68 1.52 | ± | 119.875 |
| Contr_l + Triplet_L | GAT | 99.91 0.02 | ± | 92.27 0.80 | ± | 99.68 0.02 | ± | 91.50 0.85 | ± | 20.75 |
| Contr_l + Triplet_L | GCN | 99.82 0.03 | ± | 92.11 0.51 | ± | 99.43 0.10 | ± | 92.29 0.53 | ± | 26.125 |
| Contr_l + Triplet_L | GIN | 98.68 0.18 | ± | 44.03 0.79 | ± | 96.45 0.98 | ± | 47.15 0.50 | ± | 118.5 |
| Contr_l + Triplet_L | MPNN | 99.51 0.08 | ± | 89.62 2.05 | ± | 97.56 0.24 | ± | 90.94 0.46 | ± | 68.75 |
| Contr_l + Triplet_L | PAGNN | 87.02 6.86 | ± | 32.07 0.34 | ± | 92.97 0.86 | ± | 32.11 0.51 | ± | 164.125 |
| Contr_l + Triplet_L | SAGE | 99.55 0.07 | ± | 52.28 2.75 | ± | 98.99 0.23 | ± | 54.10 1.70 | ± | 90.125 |
| CrossE_L | ALL | 93.66 10.40 | ± | 81.68 1.26 | ± | 72.06 7.61 | ± | 76.78 0.46 | ± | 127.5 |
| CrossE_L | GAT | 87.84 20.37 | ± | 41.99 0.43 | ± | 84.56 18.76 | ± | 86.02 0.46 | ± | 141.5 |
| CrossE_L | GCN | 69.28 4.98 | ± | 66.27 0.56 | ± | 49.95 1.04 | ± | 43.69 1.03 | ± | 166.0 |
| CrossE_L | GIN | 44.56 2.66 | ± | 29.86 0.12 | ± | 43.39 0.70 | ± | 29.35 0.33 | ± | 210.0 |
| CrossE_L | MPNN | 95.25 1.50 | ± | 34.71 0.45 | ± | 93.11 1.39 | ± | 34.95 0.28 | ± | 150.625 |
| CrossE_L | PAGNN | 76.08 1.71 | ± | 30.57 0.15 | ± | 78.13 0.67 | ± | 30.46 0.07 | ± | 194.25 |
| CrossE_L | SAGE | 76.61 8.56 | ± | 35.67 0.21 | ± | 67.21 9.29 | ± | 32.94 0.21 | ± | 186.0 |

Table 27. Results for Lp Aupr (↑) (continued)

| Loss Type | Model | Cora ↓ Citeseer | | Cora ↓ Bitcoin | | Citeseer ↓ Cora | | Citeseer ↓ Bitcoin | | Average Rank |
|---|---|---|---|---|---|---|---|---|---|---|
| CrossE_L + PMI_L | ALL | 87.15 | ± 2.37 | 74.19 | ± 0.67 | 92.40 | ± 1.00 | 73.94 | ± | 120.0 |
| CrossE_L + PMI_L | GAT | 99.91 | ± 0.02 | 92.06 | ± 0.46 | 99.67 | ± 0.03 | 92.07 | ± 0.15 | 18.375 |
| CrossE_L + PMI_L | GCN | 99.76 | ± 0.03 | 91.95 | ± 0.68 | 99.30 | ± 0.09 | 92.02 | ± 0.51 | 37.5 |
| CrossE_L + PMI_L | GIN | 91.00 | ± 2.03 | 42.56 | ± 0.76 | 88.37 | ± 1.02 | 43.47 | ± 1.25 | 150.875 |
| CrossE_L + PMI_L | MPNN | 99.44 | ± 0.09 | 91.48 | ± 0.17 | 96.99 | ± 0.44 | 91.07 | ± 0.40 | 67.25 |
| CrossE_L + PMI_L | PAGNN | 70.76 | ± 3.66 | 31.98 | ± 0.21 | 78.72 | ± 0.71 | 29.84 | ± 0.15 | 195.0 |
| CrossE_L + PMI_L | SAGE | 84.99 | ± 1.06 | 36.54 | ± 0.78 | 84.27 | ± 0.72 | 36.41 | ± 0.56 | 171.25 |
| CrossE_L + PMI_L + PR_L | ALL | 65.87 | ± 0.78 | 73.91 | ± 1.09 | 90.96 | ± 0.56 | 74.68 | ± 0.18 | 135.25 |
| CrossE_L + PMI_L + PR_L | GAT | 99.90 | ± 0.03 | 92.53 | ± 0.21 | 98.36 | ± 1.15 | 85.74 | ± 1.30 | 44.0 |
| CrossE_L + PMI_L + PR_L | GCN | 99.80 | ± 0.02 | 92.18 | ± 0.62 | 99.31 | ± 0.07 | 92.35 | ± 0.49 | 27.0 |
| CrossE_L + PMI_L + PR_L | GIN | 89.28 | ± 1.84 | 44.54 | ± 0.88 | 88.29 | ± 1.56 | 42.33 | ± 0.67 | 151.75 |
| CrossE_L + PMI_L + PR_L | MPNN | 99.44 | ± 0.13 | 90.69 | ± 0.77 | 94.55 | ± 2.94 | 90.43 | ± 0.74 | 81.25 |
| CrossE_L + PMI_L + PR_L | PAGNN | 71.97 | ± 3.20 | 30.35 | ± 0.34 | 77.05 | ± 4.26 | 30.33 | ± 0.09 | 199.75 |
| CrossE_L + PMI_L + PR_L | SAGE | 85.45 | ± 0.95 | 37.06 | ± 0.41 | 84.15 | ± 0.68 | 36.84 | ± 1.06 | 170.0 |
| CrossE_L + PMI_L + PR_L + Triplet_L | ALL | 83.27 | ± 2.13 | 71.09 | ± 1.53 | 92.74 | ± 0.42 | 65.52 | ± 1.14 | 126.0 |
| CrossE_L + PMI_L + PR_L + Triplet_L | GAT | 99.90 | ± 0.01 | 92.55 | ± 0.30 | 99.44 | ± 0.31 | 91.67 | ± 1.09 | 22.875 |

Table 27. Results for Lp Aupr (↑) (continued)

| Loss Type | Model | Cora ↓ Citeseer | ± | Cora ↓ Bitcoin | ± | Citeseer ↓ Cora | ± | Citeseer ↓ Bitcoin | ± | Average Rank |
|---|---|---|---|---|---|---|---|---|---|---|
| CrossE_L + PMI_L + PR_L + Triplet_L | GCN | 99.77 0.06 | ± | 92.32 0.51 | ± | 99.23 0.10 | ± | 91.53 0.57 | ± | 39.125 |
| CrossE_L + PMI_L + PR_L + Triplet_L | GIN | 91.37 2.02 | ± | 42.73 0.35 | ± | 90.64 1.41 | ± | 41.72 0.32 | ± | 150.75 |
| CrossE_L + PMI_L + PR_L + Triplet_L | MPNN | 99.49 0.06 | ± | 91.14 0.91 | ± | 95.48 1.40 | ± | 90.97 0.86 | ± | 72.875 |
| CrossE_L + PMI_L + PR_L + Triplet_L | PAGNN | 74.61 1.68 | ± | 31.78 0.07 | ± | 78.04 0.95 | ± | 30.86 0.20 | ± | 190.75 |
| CrossE_L + PMI_L + PR_L + Triplet_L | SAGE | 95.27 3.72 | ± | 52.96 1.18 | ± | 94.95 0.67 | ± | 50.13 1.32 | ± | 120.5 |
| CrossE_L + PMI_L + Triplet_L | ALL | 98.91 0.14 | ± | 75.09 1.01 | ± | 96.29 0.55 | ± | 76.98 0.80 | ± | 91.0 |
| CrossE_L + PMI_L + Triplet_L | GAT | 99.90 0.01 | ± | 92.04 1.13 | ± | 99.65 0.04 | ± | 92.43 0.55 | ± | 17.5 |
| CrossE_L + PMI_L + Triplet_L | GCN | 99.80 0.02 | ± | 92.70 0.27 | ± | 99.33 0.08 | ± | 91.87 1.05 | ± | 26.75 |
| CrossE_L + PMI_L + Triplet_L | GIN | 92.89 1.05 | ± | 44.46 0.81 | ± | 92.53 0.68 | ± | 43.47 0.48 | ± | 141.125 |
| CrossE_L + PMI_L + Triplet_L | MPNN | 99.48 0.08 | ± | 90.77 0.91 | ± | 97.18 0.20 | ± | 91.69 1.25 | ± | 64.875 |
| CrossE_L + PMI_L + Triplet_L | PAGNN | 72.57 1.80 | ± | 31.44 0.21 | ± | 79.90 1.20 | ± | 30.82 0.12 | ± | 190.875 |
| CrossE_L + PMI_L + Triplet_L | SAGE | 97.83 0.39 | ± | 53.78 1.84 | ± | 96.65 0.39 | ± | 56.92 1.35 | ± | 105.375 |
| CrossE_L + PR_L | ALL | 70.14 3.08 | ± | 70.57 1.06 | ± | 79.26 6.63 | ± | 72.03 1.12 | ± | 147.0 |
| CrossE_L + PR_L | GAT | 99.74 0.11 | ± | 90.19 0.75 | ± | 99.31 0.07 | ± | 89.77 0.96 | ± | 58.125 |
| CrossE_L + PR_L | GCN | 98.61 0.76 | ± | 89.36 1.13 | ± | 98.82 0.38 | ± | 88.96 1.85 | ± | 80.0 |
| CrossE_L + PR_L | GIN | 93.70 1.99 | ± | 51.27 0.23 | ± | 79.62 7.13 | ± | 52.45 0.37 | ± | 141.5 |

Table 27. Results for Lp Aupr (↑) (continued)

| Loss Type | Model | Cora ↓ Citeseer | | Cora ↓ Bitcoin | | Citeseer ↓ Cora | | Citeseer ↓ Bitcoin | | Average Rank |
|---|---|---|---|---|---|---|---|---|---|---|
| CrossE_L + PR_L | MPNN | 96.41 | ± 3.98 | 91.46 | ± 0.61 | 88.12 | ± 3.00 | 92.30 | ± 0.58 | 87.75 |
| CrossE_L + PR_L | PAGNN | 75.92 | ± 3.82 | 31.17 | ± 0.12 | 67.42 | ± 3.10 | 30.37 | ± 0.07 | 197.0 |
| CrossE_L + PR_L | SAGE | 83.40 | ± 0.53 | 41.91 | ± 1.18 | 82.69 | ± 1.26 | 39.86 | ± 1.99 | 168.75 |
| CrossE_L + PR_L + Triplet_L | ALL | 79.18 | ± 11.59 | 45.43 | ± 2.70 | 88.21 | ± 1.65 | 46.27 | ± 1.81 | 154.5 |
| CrossE_L + PR_L + Triplet_L | GAT | 99.80 | ± 0.09 | 90.45 | ± 0.88 | 99.53 | ± 0.18 | 92.74 ± 0.39 | | 31.875 |
| CrossE_L + PR_L + Triplet_L | GCN | 99.67 | ± 0.04 | 91.89 | ± 0.66 | 99.09 | ± 0.24 | 93.30 ± 0.54 | | 36.125 |
| CrossE_L + PR_L + Triplet_L | GIN | 97.25 | ± 1.58 | 41.53 | ± 0.62 | 93.11 | ± 1.57 | 43.19 | ± 0.57 | 135.875 |
| CrossE_L + PR_L + Triplet_L | MPNN | 98.25 | ± 0.97 | 91.71 | ± 0.72 | 91.59 | ± 2.41 | 91.90 | ± 0.53 | 80.875 |
| CrossE_L + PR_L + Triplet_L | PAGNN | 80.53 | ± 3.06 | 32.13 | ± 0.25 | 76.99 | ± 4.65 | 32.21 | ± 0.15 | 185.0 |
| CrossE_L + PR_L + Triplet_L | SAGE | 99.48 | ± 0.21 | 58.04 | ± 2.24 | 98.88 | ± 0.44 | 52.10 | ± 1.50 | 92.75 |
| CrossE_L + Triplet_L | ALL | 99.11 | ± 0.12 | 61.66 | ± 0.61 | 96.50 | ± 0.58 | 52.83 | ± 1.04 | 101.25 |
| CrossE_L + Triplet_L | GAT | 99.92 ± 0.02 | | 92.86 ± 0.62 | ± | 99.70 ± 0.02 | | 92.72 | ± 0.83 | 2.625 |
| CrossE_L + Triplet_L | GCN | 99.83 | ± 0.03 | 92.20 | ± 0.28 | 99.49 | ± 0.06 | 92.04 | ± 1.05 | 25.125 |
| CrossE_L + Triplet_L | GIN | 98.79 | ± 0.15 | 44.21 | ± 0.54 | 96.18 | ± 0.44 | 45.74 | ± 0.76 | 119.375 |
| CrossE_L + Triplet_L | MPNN | 99.49 | ± 0.15 | 91.07 | ± 0.51 | 97.73 | ± 0.26 | 89.64 | ± 0.77 | 68.75 |
| CrossE_L + Triplet_L | PAGNN | 89.18 | ± 6.61 | 31.19 | ± 0.37 | 92.82 | ± 0.76 | 32.14 | ± 0.47 | 164.875 |

Table 27. Results for Lp Aupr (↑) (continued)

| Loss Type | Model | Cora ↓ Citeseer | | Cora ↓ Bitcoin | | Citeseer ↓ Cora | | Citeseer ↓ Bitcoin | | Average Rank |
|---|---|---|---|---|---|---|---|---|---|---|
| CrossE_L + Triplet_L | SAGE | 99.70 | ± | 55.35 | ± | 99.31 | ± | 54.92 | ± | 80.125 |
| | | 0.06 | | 1.63 | | 0.13 | | 0.69 | | |
| PMI_L | ALL | 83.23 | ± | 75.59 | ± | 92.69 | ± | 75.02 | ± | 121.75 |
| | | 1.51 | | 1.19 | | 0.51 | | 0.81 | | |
| PMI_L | GAT | 99.90 | ± | 92.41 | ± | 99.69 | ± | 92.70 | ± | 9.25 |
| | | 0.02 | | 1.02 | | 0.07 | | 0.48 | | |
| PMI_L | GCN | 99.77 | ± | 91.86 | ± | 99.23 | ± | 92.44 | ± | 35.25 |
| | | 0.03 | | 0.49 | | 0.09 | | 0.97 | | |
| PMI_L | GIN | 89.88 | ± | 41.15 | ± | 88.77 | ± | 40.68 | ± | 157.25 |
| | | 1.83 | | 0.31 | | 1.05 | | 0.62 | | |
| PMI_L | MPNN | 99.51 | ± | 90.80 | ± | 96.91 | ± | 90.61 | ± | 68.75 |
| | | 0.08 | | 0.88 | | 0.32 | | 0.98 | | |
| PMI_L | PAGNN | 72.69 | ± | 31.36 | ± | 79.88 | ± | 30.15 | ± | 192.75 |
| | | 1.81 | | 0.31 | | 1.01 | | 0.24 | | |
| PMI_L | SAGE | 84.49 | ± | 37.67 | ± | 82.29 | ± | 35.46 | ± | 173.375 |
| | | 1.40 | | 1.13 | | 1.08 | | 0.55 | | |
| PMI_L + PR_L | ALL | 67.79 | ± | 72.19 | ± | 90.65 | ± | 74.25 | ± | 136.5 |
| | | 3.64 | | 1.76 | | 0.42 | | 1.01 | | |
| PMI_L + PR_L | GAT | 99.85 | ± | 91.46 | ± | 96.82 | ± | 90.18 | ± | 58.375 |
| | | 0.08 | | 1.26 | | 3.54 | | 1.78 | | |
| PMI_L + PR_L | GCN | 99.77 | ± | 92.73 | ± | 99.26 | ± | 91.34 | ± | 36.625 |
| | | 0.05 | | 0.80 | | 0.24 | | 0.97 | | |
| PMI_L + PR_L | GIN | 88.42 | ± | 41.01 | ± | 87.94 | ± | 42.47 | ± | 157.75 |
| | | 1.98 | | 0.60 | | 2.93 | | 0.84 | | |
| PMI_L + PR_L | MPNN | 99.45 | ± | 90.69 | ± | 92.25 | ± | 87.90 | ± | 92.125 |
| | | 0.15 | | 0.81 | | 1.17 | | 1.84 | | |
| PMI_L + PR_L | PAGNN | 72.96 | ± | 30.43 | ± | 76.05 | ± | 29.95 | ± | 199.5 |
| | | 2.16 | | 0.28 | | 0.83 | | 0.09 | | |
| PMI_L + PR_L | SAGE | 85.28 | ± | 36.82 | ± | 83.78 | ± | 35.46 | ± | 171.625 |
| | | 0.99 | | 0.79 | | 1.14 | | 0.77 | | |
| PMI_L + PR_L + Triplet_L | ALL | 84.68 | ± | 58.89 | ± | 93.37 | ± | 60.02 | ± | 127.25 |
| | | 1.98 | | 4.66 | | 0.64 | | 0.96 | | |

Table 27. Results for Lp Aupr (↑) (continued)

| Loss Type | Model | Cora ↓ Citeseer | | Cora ↓ Bitcoin | | Citeseer ↓ Cora | | Citeseer ↓ Bitcoin | | Average Rank |
|---|---|---|---|---|---|---|---|---|---|---|
| PMI_L + PR_L + Triplet_L | GAT | 99.90 | ± 0.03 | 92.71 | ± 0.71 | 99.48 | ± 0.20 | 91.58 | ± 1.03 | 21.5 |
| PMI_L + PR_L + Triplet_L | GCN | 99.80 | ± 0.02 | 92.81 | ± 0.92 | 99.32 | ± 0.11 | 91.30 | ± 0.76 | 30.5 |
| PMI_L + PR_L + Triplet_L | GIN | 91.58 | ± 2.42 | 43.61 | ± 0.48 | 90.85 | ± 0.86 | 38.75 | ± 0.72 | 152.5 |
| PMI_L + PR_L + Triplet_L | MPNN | 99.46 | ± 0.05 | 91.78 | ± 0.71 | 94.10 | ± 0.72 | 89.17 | ± 1.07 | 76.125 |
| PMI_L + PR_L + Triplet_L | PAGNN | 72.68 | ± 1.69 | 31.44 | ± 0.14 | 77.99 | ± 1.69 | 31.05 | ± 0.20 | 193.125 |
| PMI_L + PR_L + Triplet_L | SAGE | 96.97 | ± 2.21 | 52.99 | ± 0.94 | 96.42 | ± 0.72 | 50.01 | ± 0.95 | 113.375 |
| PMI_L + Triplet_L | ALL | 98.81 | ± 0.18 | 62.56 | ± 1.92 | 96.38 | ± 0.50 | 77.56 | ± 0.48 | 94.75 |
| PMI_L + Triplet_L | GAT | 99.90 | ± 0.02 | 92.23 | ± 0.86 | 99.66 | ± 0.03 | **92.77 ± 0.65** | | 11.75 |
| PMI_L + Triplet_L | GCN | 99.78 | ± 0.02 | 92.40 | ± 0.74 | 99.35 | ± 0.11 | 91.51 | ± 0.52 | 34.0 |
| PMI_L + Triplet_L | GIN | 95.30 | ± 0.80 | 44.16 | ± 0.48 | 92.17 | ± 1.21 | 45.27 | ± 0.73 | 138.0 |
| PMI_L + Triplet_L | MPNN | 99.52 | ± 0.06 | 90.92 | ± 0.66 | 97.21 | ± 0.19 | 91.24 | ± 0.88 | 64.375 |
| PMI_L + Triplet_L | PAGNN | 71.42 | ± 2.64 | 30.93 | ± 0.18 | 80.88 | ± 1.09 | 29.98 | ± 0.18 | 195.75 |
| PMI_L + Triplet_L | SAGE | 96.57 | ± 1.14 | 50.47 | ± 1.33 | 96.99 | ± 0.25 | 55.42 | ± 0.83 | 109.875 |
| PR_L | ALL | 63.39 | ± 2.01 | 72.06 | ± 0.96 | 71.37 | ± 5.45 | 73.93 | ± 1.04 | 151.5 |
| PR_L | GAT | 99.69 | ± 0.11 | 88.87 | ± 0.67 | 99.37 | ± 0.16 | 89.56 | ± 1.07 | 60.125 |
| PR_L | GCN | 99.25 | ± 0.32 | 88.84 | ± 0.31 | 98.76 | ± 0.28 | 88.01 | ± 2.10 | 78.5 |

Continued on next page

Table 27. Results for Lp Aupr (↑) (continued)

| Loss Type | Model | Cora ↓ Citeseer | | Cora ↓ Bitcoin | | Citeseer ↓ Cora | | Citeseer ↓ Bitcoin | | Average Rank |
|---|---|---|---|---|---|---|---|---|---|---|
| PR_L | GIN | 92.49 2.39 | ± | 52.15 0.21 | ± | 82.63 5.36 | ± | 52.05 0.29 | ± | 141.5 |
| PR_L | MPNN | 95.51 5.09 | ± | 91.57 0.36 | ± | 89.33 2.33 | ± | 91.84 0.54 | ± | 91.25 |
| PR_L | PAGNN | 73.91 4.99 | ± | 30.83 0.23 | ± | 65.08 1.53 | ± | 31.65 0.17 | ± | 196.75 |
| PR_L | SAGE | 83.24 1.34 | ± | 36.36 0.43 | ± | 83.60 1.26 | ± | 38.77 1.09 | ± | 172.875 |
| PR_L + Triplet_L | ALL | 69.31 4.34 | ± | 62.40 4.37 | ± | 74.41 4.16 | ± | 70.89 0.25 | ± | 153.875 |
| PR_L + Triplet_L | GAT | 99.69 0.32 | ± | 85.07 1.63 | ± | 99.28 0.34 | ± | 89.97 0.49 | ± | 63.375 |
| PR_L + Triplet_L | GCN | 99.51 0.16 | ± | 90.68 1.04 | ± | 98.90 0.27 | ± | 91.05 0.56 | ± | 63.375 |
| PR_L + Triplet_L | GIN | 92.98 3.47 | ± | 48.26 0.72 | ± | 84.75 4.58 | ± | 40.60 1.05 | ± | 151.0 |
| PR_L + Triplet_L | MPNN | 96.09 4.51 | ± | 92.18 0.77 | ± | 91.94 3.33 | ± | 92.17 0.50 | ± | 76.875 |
| PR_L + Triplet_L | PAGNN | 75.33 3.63 | ± | 32.06 0.23 | ± | 69.09 2.92 | ± | 31.96 0.11 | ± | 190.25 |
| PR_L + Triplet_L | SAGE | 86.68 1.68 | ± | 40.37 0.71 | ± | 90.30 7.13 | ± | 41.34 1.66 | ± | 159.75 |
| Triplet_L | ALL | 98.93 0.21 | ± | 50.20 1.30 | ± | 96.56 0.39 | ± | 79.27 1.05 | ± | 99.5 |
| Triplet_L | GAT | 99.93 0.02 | ± | 92.30 0.78 | ± | 99.71 0.04 | ± | 92.51 1.08 | ± | 7.75 |
| Triplet_L | GCN | 99.86 0.02 | ± | 92.71 0.77 | ± | 99.43 0.12 | ± | 92.48 0.53 | ± | 16.0 |
| Triplet_L | GIN | 99.05 0.16 | ± | 44.93 0.12 | ± | 96.18 0.89 | ± | 48.23 0.65 | ± | 115.875 |
| Triplet_L | MPNN | 99.55 0.10 | ± | 91.60 0.62 | ± | 97.54 0.18 | ± | 91.77 1.71 | ± | 54.25 |

Table 27. Results for Lp Aupr (↑) (continued)

| Loss Type | Model | Cora ↓ Citeseer | | Cora ↓ Bitcoin | | Citeseer ↓ Cora | | Citeseer ↓ Bitcoin | | Average Rank |
|---|---|---|---|---|---|---|---|---|---|---|
| Triplet_L | PAGNN | 90.13 ± 6.51 | ± | 32.07 0.40 | ± | 93.67 0.47 | ± | 33.50 0.29 | ± | 154.875 |
| Triplet_L | SAGE | 99.67 0.08 | ± | 59.94 1.51 | ± | 99.17 0.15 | ± | 61.87 1.15 | ± | 80.5 |

Table 28. Lp Auroc Performance (↑): This table presents models (Loss function and GNN) ranked by their average performance in terms of lp auroc. Top-ranked results are highlighted in red, second-ranked in blue, and third-ranked in green.

| Loss Type | Model | Cora ↓ Citeseer | | Cora ↓ Bitcoin | | Citeseer ↓ Cora | | Citeseer ↓ Bitcoin | | Average Rank |
|---|---|---|---|---|---|---|---|---|---|---|
| Contr_l | ALL | 96.91 0.88 | ± | 61.08 1.41 | ± | 94.01 0.79 | ± | 59.42 0.86 | ± | 123.75 |
| Contr_l | GAT | 99.88 0.03 | ± | 94.24 0.85 | ± | 99.71 0.03 | ± | 95.10 0.47 | ± | 18.875 |
| Contr_l | GCN | 99.79 0.02 | ± | 94.44 0.44 | ± | 99.41 0.14 | ± | 94.52 0.44 | ± | 31.625 |
| Contr_l | GIN | 98.38 0.29 | ± | 59.89 0.46 | ± | 96.33 0.61 | ± | 67.08 0.99 | ± | 112.875 |
| Contr_l | MPNN | 99.54 0.09 | ± | 93.21 0.78 | ± | 97.16 0.74 | ± | 93.69 0.60 | ± | 72.0 |
| Contr_l | PAGNN | 91.60 1.77 | ± | 47.49 0.84 | ± | 91.50 0.87 | ± | 49.03 0.13 | ± | 161.625 |
| Contr_l | SAGE | 99.18 0.08 | ± | 77.49 2.24 | ± | 98.49 0.57 | ± | 81.54 3.26 | ± | 84.5 |
| Contr_l + CrossE_L | ALL | 97.14 0.75 | ± | 73.48 1.40 | ± | 93.13 1.05 | ± | 61.91 1.91 | ± | 117.125 |
| Contr_l + CrossE_L | GAT | 99.90 0.01 | ± | 94.20 0.58 | ± | 99.61 0.06 | ± | 94.31 0.57 | ± | 29.25 |

Continued on next page

Table 28. Results for Lp Auroc (↑) (continued)

| Loss Type | Model | Cora ↓ Citeseer | | Cora ↓ Bitcoin | | Citeseer ↓ Cora | | Citeseer ↓ Bitcoin | | Average Rank |
|---|---|---|---|---|---|---|---|---|---|---|
| Contr_l + CrossE_L | GCN | 99.81 | ± 0.04 | 94.01 | ± 0.71 | 99.32 | ± 0.11 | 94.60 | ± 0.16 | 36.0 |
| Contr_l + CrossE_L | GIN | 96.53 | ± 1.35 | 55.08 | ± 0.61 | 96.24 | ± 0.52 | 60.20 | ± 0.36 | 128.5 |
| Contr_l + CrossE_L | MPNN | 99.52 | ± 0.05 | 93.18 | ± 0.83 | 97.40 | ± 0.30 | 93.33 | ± 0.81 | 74.0 |
| Contr_l + CrossE_L | PAGNN | 90.43 | ± 1.17 | 47.45 | ± 0.44 | 90.53 | ± 1.00 | 47.29 | ± 0.37 | 167.75 |
| Contr_l + CrossE_L | SAGE | 99.06 | ± 0.17 | 74.07 | ± 1.56 | 98.73 | ± 0.42 | 73.90 | ± 1.83 | 91.0 |
| Contr_l + CrossE_L + PMI_L | ALL | 89.03 | ± 4.25 | 65.84 | ± 0.85 | 91.31 | ± 1.58 | 67.96 | ± 1.40 | 131.5 |
| Contr_l + CrossE_L + PMI_L | GAT | 99.92 | ± 0.03 | 94.88 | ± 0.35 | 99.66 | ± 0.06 | 94.84 | ± 0.33 | 11.0 |
| Contr_l + CrossE_L + PMI_L | GCN | 99.81 | ± 0.03 | 94.70 | ± 0.31 | 99.23 | ± 0.09 | 94.31 | ± 0.38 | 33.75 |
| Contr_l + CrossE_L + PMI_L | GIN | 92.51 | ± 2.04 | 54.62 | ± 0.84 | 89.25 | ± 1.35 | 55.28 | ± 0.49 | 153.25 |
| Contr_l + CrossE_L + PMI_L | MPNN | 99.65 | ± 0.07 | 94.32 | ± 0.76 | 97.62 | ± 0.31 | 93.81 | ± 0.50 | 59.875 |
| Contr_l + CrossE_L + PMI_L | PAGNN | 73.48 | ± 1.40 | 45.06 | ± 0.57 | 77.79 | ± 1.80 | 41.54 | ± 0.38 | 198.375 |
| Contr_l + CrossE_L + PMI_L | SAGE | 86.30 | ± 0.79 | 60.37 | ± 1.04 | 86.65 | ± 2.32 | 64.72 | ± 0.42 | 144.75 |
| Contr_l + CrossE_L + PMI_L + PR_L | ALL | 69.67 | ± 7.05 | 67.83 | ± 0.50 | 89.18 | ± 0.99 | 68.12 | ± 0.25 | 145.625 |
| Contr_l + CrossE_L + PMI_L + PR_L | GAT | 99.91 | ± 0.04 | 94.58 | ± 1.15 | 99.65 | ± 0.04 | 94.53 | ± 1.08 | 20.125 |
| Contr_l + CrossE_L + PMI_L + PR_L | GCN | 99.79 | ± 0.05 | 94.56 | ± 0.46 | 99.23 | ± 0.16 | 94.26 | ± 0.80 | 38.75 |
| Contr_l + CrossE_L + PMI_L + PR_L | GIN | 87.75 | ± 2.08 | 53.15 | ± 0.87 | 84.10 | ± 2.30 | 48.09 | ± 0.40 | 171.25 |

Table 28. Results for Lp Auroc (↑) (continued)

| Loss Type | Model | Cora ↓ Citeseer | | Cora ↓ Bitcoin | | Citeseer ↓ Cora | | Citeseer ↓ Bitcoin | | Average Rank |
|---|---|---|---|---|---|---|---|---|---|---|
| Contr_l + CrossE_L + PMI_L + PR_L | MPNN | 99.67 | ± 0.03 | 93.29 | ± 0.86 | 94.80 | ± 2.04 | 90.93 | ± 2.01 | 79.875 |
| Contr_l + CrossE_L + PMI_L + PR_L | PAGNN | 75.68 | ± 2.33 | 46.77 | ± 0.25 | 76.19 | ± 3.43 | 47.83 | ± 0.21 | 192.0 |
| Contr_l + CrossE_L + PMI_L + PR_L | SAGE | 84.86 | ± 1.99 | 56.81 | ± 0.93 | 84.43 | ± 2.66 | 58.35 | ± 2.14 | 157.375 |
| Contr_l + CrossE_L + PMI_L + PR_L + Triplet_L | ALL | 83.89 | ± 2.50 | 62.38 | ± 0.79 | 91.61 | ± 0.71 | 65.45 | ± 1.85 | 138.125 |
| Contr_l + CrossE_L + PMI_L + PR_L + Triplet_L | GAT | 99.92 | ± 0.02 | 94.42 | ± 1.04 | 99.50 | ± 0.21 | 94.39 | ± 0.73 | 23.5 |
| Contr_l + CrossE_L + PMI_L + PR_L + Triplet_L | GCN | 99.81 | ± 0.03 | 94.64 | ± 0.54 | 99.25 | ± 0.06 | 94.51 | ± 0.47 | 31.125 |
| Contr_l + CrossE_L + PMI_L + PR_L + Triplet_L | GIN | 89.85 | ± 2.47 | 55.03 | ± 0.83 | 90.65 | ± 0.90 | 54.19 | ± 0.56 | 156.625 |
| Contr_l + CrossE_L + PMI_L + PR_L + Triplet_L | MPNN | 99.62 | ± 0.06 | 94.01 | ± 0.19 | 95.59 | ± 1.43 | 91.36 | ± 0.83 | 77.5 |
| Contr_l + CrossE_L + PMI_L + PR_L + Triplet_L | PAGNN | 75.81 | ± 1.56 | 44.38 | ± 0.58 | 77.05 | ± 1.31 | 45.08 | ± 0.61 | 194.25 |
| Contr_l + CrossE_L + PMI_L + PR_L + Triplet_L | SAGE | 95.44 | ± 2.54 | 77.22 | ± 1.27 | 96.21 | ± 0.22 | 75.86 | ± 1.84 | 103.0 |
| Contr_l + CrossE_L + PMI_L + Triplet_L | ALL | 98.52 | ± 0.42 | 71.27 | ± 0.79 | 96.35 | ± 0.49 | 74.35 | ± 1.27 | 101.5 |
| Contr_l + CrossE_L + PMI_L + Triplet_L | GAT | 99.92 | ± 0.01 | 95.15 | ± 0.20 | 99.63 | ± 0.03 | 94.85 | ± 0.32 | 8.875 |
| Contr_l + CrossE_L + PMI_L + Triplet_L | GCN | 99.82 | ± 0.05 | 95.16 | ± 0.42 | 99.09 | ± 0.23 | 94.77 | ± 0.52 | 24.375 |

Table 28. Results for Lp Auroc (↑) (continued)

| Loss Type | Model | Cora ↓ Citeseer | ± | Cora ↓ Bitcoin | ± | Citeseer ↓ Cora | ± | Citeseer ↓ Bitcoin | ± | Average Rank |
|---|---|---|---|---|---|---|---|---|---|---|
| Contr_l + CrossE_L + PMI_L + Triplet_L | GIN | 93.27 1.47 | ± | 56.18 0.34 | ± | 91.06 2.09 | ± | 57.02 0.47 | ± | 144.0 |
| Contr_l + CrossE_L + PMI_L + Triplet_L | MPNN | 99.67 0.05 | ± | 94.41 0.64 | ± | 97.46 0.41 | ± | 93.81 0.68 | ± | 57.25 |
| Contr_l + CrossE_L + PMI_L + Triplet_L | PAGNN | 75.49 0.85 | ± | 42.18 0.27 | ± | 77.97 1.06 | ± | 43.62 0.59 | ± | 195.75 |
| Contr_l + CrossE_L + PMI_L + Triplet_L | SAGE | 96.66 0.97 | ± | 72.09 2.41 | ± | 95.04 0.78 | ± | 76.72 0.85 | ± | 105.0 |
| Contr_l + CrossE_L + PR_L | ALL | 70.23 3.19 | ± | 61.76 1.02 | ± | 77.14 6.84 | ± | 64.66 0.41 | ± | 161.0 |
| Contr_l + CrossE_L + PR_L | GAT | 99.57 0.54 | ± | 94.19 0.40 | ± | 99.49 0.06 | ± | 92.96 0.69 | ± | 53.25 |
| Contr_l + CrossE_L + PR_L | GCN | 99.61 0.07 | ± | 92.34 0.97 | ± | 98.76 0.19 | ± | 93.58 1.02 | ± | 68.5 |
| Contr_l + CrossE_L + PR_L | GIN | 94.64 2.08 | ± | 54.37 0.47 | ± | 83.17 1.24 | ± | 53.02 0.36 | ± | 158.5 |
| Contr_l + CrossE_L + PR_L | MPNN | 93.24 6.25 | ± | 93.94 0.61 | ± | 88.74 5.17 | ± | 94.11 0.40 | ± | 97.25 |
| Contr_l + CrossE_L + PR_L | PAGNN | 77.64 5.29 | ± | 49.89 0.55 | ± | 60.62 7.32 | ± | 49.08 0.51 | ± | 187.25 |
| Contr_l + CrossE_L + PR_L | SAGE | 95.87 6.41 | ± | 75.29 2.19 | ± | 92.82 7.44 | ± | 71.40 2.69 | ± | 112.5 |
| Contr_l + CrossE_L + PR_L + Triplet_L | ALL | 89.52 1.64 | ± | 57.58 0.67 | ± | 92.67 0.89 | ± | 62.24 0.86 | ± | 139.25 |
| Contr_l + CrossE_L + PR_L + Triplet_L | GAT | 99.89 0.03 | ± | 94.34 0.84 | ± | 99.57 0.23 | ± | **95.21 0.32** | ± | 20.0 |
| Contr_l + CrossE_L + PR_L + Triplet_L | GCN | 99.76 0.05 | ± | 94.06 0.71 | ± | 99.06 0.12 | ± | 93.92 1.55 | ± | 50.25 |
| Contr_l + CrossE_L + PR_L + Triplet_L | GIN | 96.89 0.74 | ± | 55.09 0.77 | ± | 95.17 0.44 | ± | 59.09 0.65 | ± | 130.75 |
| Contr_l + CrossE_L + PR_L + Triplet_L | MPNN | 99.31 0.32 | ± | 94.47 0.61 | ± | 95.24 0.58 | ± | 92.34 1.52 | ± | 73.25 |

Table 28. Results for Lp Auroc (↑) (continued)

| Loss Type | Model | Cora ↓ Citeseer | | Cora ↓ Bitcoin | | Citeseer ↓ Cora | | Citeseer ↓ Bitcoin | | Average Rank |
|---|---|---|---|---|---|---|---|---|---|---|
| Contr_l + CrossE_L + PR_L + Triplet_L | PAGNN | 83.08 | ± 2.23 | 47.81 | ± 0.27 | 79.07 | ± 2.79 | 49.45 | ± 0.22 | 179.25 |
| Contr_l + CrossE_L + PR_L + Triplet_L | SAGE | 99.42 | ± 0.09 | 78.51 | ± 1.44 | 98.15 | ± 0.47 | 73.97 | ± 2.03 | 87.0 |
| Contr_l + CrossE_L + Triplet_L | ALL | 98.05 | ± 0.77 | 72.63 | ± 1.62 | 95.84 | ± 0.59 | 64.62 | ± 0.99 | 109.25 |
| Contr_l + CrossE_L + Triplet_L | GAT | 99.92 | ± 0.02 | 94.43 | ± 0.67 | 99.68 | ± 0.05 | 94.80 | ± 0.37 | 15.75 |
| Contr_l + CrossE_L + Triplet_L | GCN | 99.85 | ± 0.02 | 94.39 | ± 0.69 | 99.37 | ± 0.11 | 94.33 | ± 0.35 | 31.625 |
| Contr_l + CrossE_L + Triplet_L | GIN | 98.90 | ± 0.21 | 61.26 | ± 0.67 | 96.99 | ± 0.59 | 65.93 | ± 0.31 | 108.25 |
| Contr_l + CrossE_L + Triplet_L | MPNN | 99.59 | ± 0.06 | 92.84 | ± 0.72 | 97.82 | ± 0.31 | 93.99 | ± 0.65 | 68.25 |
| Contr_l + CrossE_L + Triplet_L | PAGNN | 88.44 | ± 6.44 | 48.09 | ± 0.43 | 93.85 | ± 0.88 | 49.26 | ± 0.41 | 158.5 |
| Contr_l + CrossE_L + Triplet_L | SAGE | 99.49 | ± 0.11 | 78.28 | ± 2.11 | 98.95 | ± 0.14 | 81.55 | ± 1.85 | 79.25 |
| Contr_l + PMI_L | ALL | 91.20 | ± 5.46 | 57.65 | ± 1.51 | 93.69 | ± 0.98 | 69.02 | ± 1.15 | 129.75 |
| Contr_l + PMI_L | GAT | 99.92 | ± 0.02 | 95.19 | ± 0.70 | 99.65 | ± 0.02 | 94.73 | ± 0.81 | 9.25 |
| Contr_l + PMI_L | GCN | 99.79 | ± 0.05 | 94.37 | ± 0.61 | 99.26 | ± 0.20 | 94.67 | ± 0.90 | 34.125 |
| Contr_l + PMI_L | GIN | 91.73 | ± 3.73 | 56.79 | ± 0.47 | 88.94 | ± 1.00 | 55.59 | ± 0.31 | 151.125 |
| Contr_l + PMI_L | MPNN | 99.66 | ± 0.04 | 92.55 | ± 0.54 | 97.77 | ± 0.20 | 94.35 | ± 0.34 | 62.875 |
| Contr_l + PMI_L | PAGNN | 75.06 | ± 1.93 | 42.77 | ± 0.25 | 76.76 | ± 0.80 | 41.54 | ± 0.22 | 199.125 |
| Contr_l + PMI_L | SAGE | 86.44 | ± 2.38 | 61.60 | ± 2.59 | 91.40 | ± 2.85 | 67.80 | ± 0.88 | 135.625 |

Table 28. Results for Lp Auroc (↑) (continued)

| Loss Type | Model | Cora ↓ Citeseer | | Cora ↓ Bitcoin | | Citeseer ↓ Cora | | Citeseer ↓ Bitcoin | | Average Rank |
|---|---|---|---|---|---|---|---|---|---|---|
| Contr_l + PMI_L + PR_L | ALL | 68.73 ± 3.03 | | 63.76 0.97 | ± | 89.03 0.71 | ± | 69.19 0.81 | ± | 148.625 |
| Contr_l + PMI_L + PR_L | GAT | 99.89 ± 0.04 | | 95.01 0.40 | ± | 98.29 1.87 | ± | 89.36 1.29 | ± | 44.75 |
| Contr_l + PMI_L + PR_L | GCN | 99.83 ± 0.04 | | 94.59 0.69 | ± | 99.23 0.11 | ± | 94.41 0.63 | ± | 31.625 |
| Contr_l + PMI_L + PR_L | GIN | 90.74 ± 4.72 | | 55.26 0.51 | ± | 86.43 2.43 | ± | 50.69 1.02 | ± | 160.5 |
| Contr_l + PMI_L + PR_L | MPNN | 99.68 ± 0.04 | | 94.38 0.49 | ± | 93.43 2.41 | ± | 91.15 0.59 | ± | 72.75 |
| Contr_l + PMI_L + PR_L | PAGNN | 73.97 ± 1.29 | | 46.85 0.59 | ± | 76.28 1.52 | ± | 46.94 0.36 | ± | 194.75 |
| Contr_l + PMI_L + PR_L | SAGE | 86.16 ± 1.21 | | 61.61 2.32 | ± | 89.08 1.61 | ± | 64.15 1.45 | ± | 142.0 |
| Contr_l + PMI_L + PR_L + Triplet_L | ALL | 88.36 ± 2.74 | | 56.82 1.08 | ± | 93.13 0.65 | ± | 63.88 1.04 | ± | 139.625 |
| Contr_l + PMI_L + PR_L + Triplet_L | GAT | 99.86 ± 0.07 | | 94.08 0.88 | ± | 99.17 0.32 | ± | 91.78 1.80 | ± | 49.625 |
| Contr_l + PMI_L + PR_L + Triplet_L | GCN | 99.80 ± 0.03 | | 95.04 0.44 | ± | 99.26 0.16 | ± | 94.48 0.53 | ± | 27.125 |
| Contr_l + PMI_L + PR_L + Triplet_L | GIN | 93.38 ± 0.42 | | 57.02 0.13 | ± | 92.62 0.92 | ± | 54.43 0.45 | ± | 142.25 |
| Contr_l + PMI_L + PR_L + Triplet_L | MPNN | 99.67 ± 0.08 | | 93.70 0.24 | ± | 94.84 0.41 | ± | 90.51 1.16 | ± | 77.875 |
| Contr_l + PMI_L + PR_L + Triplet_L | PAGNN | 77.14 ± 1.48 | | 47.49 0.47 | ± | 76.77 3.11 | ± | 46.49 0.46 | ± | 189.625 |
| Contr_l + PMI_L + PR_L + Triplet_L | SAGE | 98.16 ± 0.47 | | 75.17 0.96 | ± | 96.67 1.04 | ± | 75.28 2.53 | ± | 99.5 |
| Contr_l + PR_L | ALL | 72.85 ± 4.09 | | 64.67 0.88 | ± | 81.37 2.76 | ± | 66.41 0.89 | ± | 154.75 |
| Contr_l + PR_L | GAT | 99.73 ± 0.17 | | 93.68 0.40 | ± | 99.43 0.12 | ± | 92.79 0.86 | ± | 51.625 |

Table 28. Results for Lp Auroc (↑) (continued)

| Loss Type | Model | Cora ↓ Citeseer | | Cora ↓ Bitcoin | | Citeseer ↓ Cora | | Citeseer ↓ Bitcoin | | Average Rank |
|---|---|---|---|---|---|---|---|---|---|---|
| Contr_l + PR_L | GCN | 99.62 | ± 0.11 | 93.33 | ± 0.60 | 98.81 | ± 0.29 | 93.87 | ± 0.77 | 62.625 |
| Contr_l + PR_L | GIN | 92.56 | ± 6.10 | 54.90 | ± 0.14 | 80.25 | ± 8.04 | 54.29 | ± 0.66 | 161.75 |
| Contr_l + PR_L | MPNN | 93.72 | ± 2.48 | 94.06 | ± 0.98 | 91.81 | ± 2.85 | 94.61 | ± 0.46 | 82.0 |
| Contr_l + PR_L | PAGNN | 74.45 | ± 5.40 | 50.20 | ± 0.53 | 62.70 | ± 1.52 | 50.04 | ± 0.44 | 188.75 |
| Contr_l + PR_L | SAGE | 95.92 | ± 6.23 | 76.35 | ± 2.17 | 94.45 | ± 6.53 | 59.41 | ± 1.53 | 116.25 |
| Contr_l + PR_L + Triplet_L | ALL | 84.92 | ± 9.41 | 58.76 | ± 0.62 | 91.71 | ± 2.41 | 58.47 | ± 0.86 | 144.75 |
| Contr_l + PR_L + Triplet_L | GAT | 99.89 | ± 0.03 | 94.99 | ± 0.65 | 99.58 | ± 0.19 | 93.32 | ± 0.88 | 27.625 |
| Contr_l + PR_L + Triplet_L | GCN | 99.73 | ± 0.07 | 94.05 | ± 0.88 | 99.10 | ± 0.13 | 94.69 | ± 0.77 | 42.875 |
| Contr_l + PR_L + Triplet_L | GIN | 97.78 | ± 0.94 | 59.38 | ± 0.57 | 93.41 | ± 2.41 | 56.24 | ± 0.49 | 129.125 |
| Contr_l + PR_L + Triplet_L | MPNN | 98.71 | ± 1.42 | 93.65 | ± 0.36 | 93.07 | ± 4.22 | 93.33 | ± 0.31 | 86.375 |
| Contr_l + PR_L + Triplet_L | PAGNN | 81.10 | ± 2.62 | 48.31 | ± 0.25 | 80.55 | ± 6.93 | 49.44 | ± 0.15 | 179.5 |
| Contr_l + PR_L + Triplet_L | SAGE | 99.34 | ± 0.09 | 77.69 | ± 2.27 | 98.53 | ± 0.32 | 76.09 | ± 1.25 | 85.0 |
| Contr_l + Triplet_L | ALL | 98.58 | ± 0.35 | 63.18 | ± 2.03 | 95.94 | ± 0.61 | 61.95 | ± 1.44 | 114.25 |
| Contr_l + Triplet_L | GAT | 99.92 | ± 0.01 | 94.71 | ± 0.75 | 99.69 | ± 0.03 | 94.24 | ± 0.67 | 19.0 |
| Contr_l + Triplet_L | GCN | 99.84 | ± 0.03 | 94.50 | ± 0.33 | 99.40 | ± 0.11 | 94.66 | ± 0.34 | 26.25 |
| Contr_l + Triplet_L | GIN | 98.88 | ± 0.19 | 57.96 | ± 0.69 | 96.96 | ± 0.96 | 61.87 | ± 0.42 | 115.25 |

Table 28. Results for Lp Auroc (↑) (continued)

| Loss Type | Model | Cora ↓ Citeseer | | Cora ↓ Bitcoin | | Citeseer ↓ Cora | | Citeseer ↓ Bitcoin | | Average Rank |
|---|---|---|---|---|---|---|---|---|---|---|
| Contr_l + Triplet_L | MPNN | 99.66 | ± 0.06 | 92.60 | ± 1.69 | 98.14 | ± 0.20 | 93.50 | ± 0.34 | 68.5 |
| Contr_l + Triplet_L | PAGNN | 89.86 | ± 5.56 | 48.57 | ± 0.92 | 93.43 | ± 0.81 | 48.87 | ± 0.97 | 157.375 |
| Contr_l + Triplet_L | SAGE | 99.58 | ± 0.08 | 76.56 | ± 3.08 | 98.85 | ± 0.27 | 78.93 | ± 1.63 | 80.5 |
| CrossE_L | ALL | 92.55 | ± 13.30 | 85.63 | ± 1.52 | 64.03 | ± 11.71 | 75.29 | ± 0.33 | 131.5 |
| CrossE_L | GAT | 89.71 | ± 18.97 | 61.12 | ± 0.66 | 86.05 | ± 19.88 | 91.78 | ± 0.30 | 130.125 |
| CrossE_L | GCN | 59.33 | ± 6.41 | 67.57 | ± 0.37 | 34.82 | ± 0.59 | 49.16 | ± 1.49 | 179.0 |
| CrossE_L | GIN | 43.16 | ± 3.68 | 39.39 | ± 0.23 | 34.02 | ± 1.05 | 38.01 | ± 0.71 | 210.0 |
| CrossE_L | MPNN | 95.50 | ± 1.59 | 48.38 | ± 0.75 | 93.01 | ± 1.53 | 47.20 | ± 0.82 | 154.25 |
| CrossE_L | PAGNN | 78.31 | ± 1.72 | 45.20 | ± 0.29 | 71.69 | ± 0.90 | 42.93 | ± 0.10 | 196.25 |
| CrossE_L | SAGE | 82.56 | ± 6.72 | 55.07 | ± 0.31 | 68.30 | ± 8.46 | 47.71 | ± 0.49 | 183.75 |
| CrossE_L + PMI_L | ALL | 87.40 | ± 2.22 | 69.72 | ± 0.77 | 90.85 | ± 0.84 | 68.77 | ± 0.93 | 131.5 |
| CrossE_L + PMI_L | GAT | 99.93 | ± 0.02 | 94.66 | ± 0.32 | 99.68 | ± 0.03 | 94.40 | ± 0.21 | 15.75 |
| CrossE_L + PMI_L | GCN | 99.78 | ± 0.02 | 94.35 | ± 0.54 | 99.23 | ± 0.10 | 94.56 | ± 0.36 | 37.875 |
| CrossE_L + PMI_L | GIN | 91.07 | ± 2.45 | 55.16 | ± 0.60 | 87.06 | ± 1.14 | 55.67 | ± 0.96 | 155.25 |
| CrossE_L + PMI_L | MPNN | 99.64 | ± 0.05 | 93.80 | ± 0.11 | 97.53 | ± 0.51 | 93.61 | ± 0.49 | 66.375 |
| CrossE_L + PMI_L | PAGNN | 74.08 | ± 3.57 | 49.25 | ± 0.43 | 77.20 | ± 0.65 | 41.96 | ± 0.36 | 192.25 |

Table 28. Results for Lp Auroc (↑) (continued)

| Loss Type | Model | Cora ↓ Citeseer | | Cora ↓ Bitcoin | | Citeseer ↓ Cora | | Citeseer ↓ Bitcoin | | Average Rank |
|---|---|---|---|---|---|---|---|---|---|---|
| CrossE_L + PMI_L | SAGE | 83.23 | ± | 57.67 | ± | 82.45 | ± | 56.59 | ± | 160.25 |
| | | 1.24 | | 1.40 | | 0.69 | | 0.91 | | |
| CrossE_L + PMI_L + PR_L | ALL | 64.97 | ± | 68.84 | ± | 88.74 | ± | 69.97 | ± | 146.375 |
| | | 0.82 | | 0.97 | | 0.66 | | 0.32 | | |
| CrossE_L + PMI_L + PR_L | GAT | 99.91 | ± | 94.93 | ± | 98.07 | ± | 89.06 | ± | 46.125 |
| | | 0.02 | | 0.05 | | 1.39 | | 1.19 | | |
| CrossE_L + PMI_L + PR_L | GCN | 99.81 | ± | 94.65 | ± | 99.27 | ± | 94.70 | ± | 27.5 |
| | | 0.02 | | 0.53 | | 0.07 | | 0.36 | | |
| CrossE_L + PMI_L + PR_L | GIN | 88.95 | ± | 56.49 | ± | 86.70 | ± | 55.20 | ± | 158.0 |
| | | 2.02 | | 0.51 | | 2.21 | | 0.62 | | |
| CrossE_L + PMI_L + PR_L | MPNN | 99.63 | ± | 93.31 | ± | 94.60 | ± | 92.90 | ± | 79.5 |
| | | 0.07 | | 0.62 | | 3.56 | | 0.63 | | |
| CrossE_L + PMI_L + PR_L | PAGNN | 75.75 | ± | 43.87 | ± | 75.08 | ± | 44.10 | ± | 197.5 |
| | | 2.99 | | 1.05 | | 5.26 | | 0.37 | | |
| CrossE_L + PMI_L + PR_L | SAGE | 83.91 | ± | 58.36 | ± | 82.29 | ± | 58.17 | ± | 156.75 |
| | | 1.06 | | 0.93 | | 0.88 | | 1.71 | | |
| CrossE_L + PMI_L + PR_L + Triplet_L | ALL | 84.06 | ± | 66.25 | ± | 91.40 | ± | 62.79 | ± | 137.875 |
| | | 1.84 | | 1.57 | | 0.60 | | 1.59 | | |
| CrossE_L + PMI_L + PR_L + Triplet_L | GAT | 99.91 | ± | 94.99 | ± | 99.41 | ± | 94.36 | ± | 20.5 |
| | | 0.01 | | 0.16 | | 0.36 | | 0.83 | | |
| CrossE_L + PMI_L + PR_L + Triplet_L | GCN | 99.80 | ± | 94.68 | ± | 99.17 | ± | 93.93 | ± | 38.875 |
| | | 0.06 | | 0.29 | | 0.13 | | 0.50 | | |
| CrossE_L + PMI_L + PR_L + Triplet_L | GIN | 91.34 | ± | 55.67 | ± | 89.92 | ± | 54.56 | ± | 152.875 |
| | | 2.20 | | 0.31 | | 1.96 | | 0.33 | | |
| CrossE_L + PMI_L + PR_L + Triplet_L | MPNN | 99.65 | ± | 93.59 | ± | 95.60 | ± | 93.64 | ± | 72.25 |
| | | 0.03 | | 1.12 | | 1.80 | | 0.74 | | |
| CrossE_L + PMI_L + PR_L + Triplet_L | PAGNN | 78.46 | ± | 48.63 | ± | 76.62 | ± | 46.00 | ± | 187.25 |
| | | 1.29 | | 0.15 | | 0.86 | | 0.48 | | |
| CrossE_L + PMI_L + PR_L + Triplet_L | SAGE | 95.37 | ± | 75.75 | ± | 94.60 | ± | 74.92 | ± | 108.375 |
| | | 3.97 | | 2.28 | | 0.71 | | 1.65 | | |
| CrossE_L + PMI_L + Triplet_L | ALL | 99.10 | ± | 72.08 | ± | 96.77 | ± | 76.22 | ± | 95.375 |
| | | 0.11 | | 0.89 | | 0.63 | | 0.75 | | |

Table 28. Results for Lp Auroc (↑) (continued)

| Loss Type | Model | Cora ↓ Citeseer | ± | Cora ↓ Bitcoin | ± | Citeseer ↓ Cora | ± | Citeseer ↓ Bitcoin | ± | Average Rank |
|---|---|---|---|---|---|---|---|---|---|---|
| CrossE_L + PMI_L + Triplet_L | GAT | 99.92 | ± | 94.63 | ± | 99.65 | ± | 95.01 | ± | 13.25 |
| | | 0.01 | | 0.76 | | 0.04 | | 0.28 | | |
| CrossE_L + PMI_L + Triplet_L | GCN | 99.82 | ± | 95.03 | ± | 99.27 | ± | 94.26 | ± | 28.375 |
| | | 0.02 | | 0.31 | | 0.10 | | 0.82 | | |
| CrossE_L + PMI_L + Triplet_L | GIN | 93.17 | ± | 56.81 | ± | 92.21 | ± | 56.77 | ± | 141.125 |
| | | 1.06 | | 0.31 | | 0.77 | | 0.52 | | |
| CrossE_L + PMI_L + Triplet_L | MPNN | 99.65 | ± | 93.27 | ± | 97.79 | ± | 94.11 | ± | 64.375 |
| | | 0.05 | | 0.88 | | 0.21 | | 1.12 | | |
| CrossE_L + PMI_L + Triplet_L | PAGNN | 76.70 | ± | 47.55 | ± | 78.68 | ± | 45.95 | ± | 188.0 |
| | | 1.78 | | 0.34 | | 1.18 | | 0.19 | | |
| CrossE_L + PMI_L + Triplet_L | SAGE | 97.95 | ± | 75.87 | ± | 96.41 | ± | 80.06 | ± | 96.75 |
| | | 0.43 | | 1.86 | | 0.40 | | 0.43 | | |
| CrossE_L + PR_L | ALL | 69.95 | ± | 65.47 | ± | 78.02 | ± | 67.08 | ± | 157.125 |
| | | 4.65 | | 1.03 | | 7.54 | | 1.35 | | |
| CrossE_L + PR_L | GAT | 99.75 | ± | 92.78 | ± | 99.24 | ± | 92.32 | ± | 59.75 |
| | | 0.11 | | 0.64 | | 0.13 | | 1.04 | | |
| CrossE_L + PR_L | GCN | 98.46 | ± | 91.53 | ± | 98.63 | ± | 91.31 | ± | 81.25 |
| | | 0.86 | | 1.33 | | 0.54 | | 1.74 | | |
| CrossE_L + PR_L | GIN | 92.61 | ± | 55.03 | ± | 73.10 | ± | 55.38 | ± | 164.875 |
| | | 3.30 | | 0.19 | | 10.45 | | 0.43 | | |
| CrossE_L + PR_L | MPNN | 94.79 | ± | 94.10 | ± | 82.52 | ± | 94.71 | ± | 88.75 |
| | | 6.43 | | 0.48 | | 4.98 | | 0.47 | | |
| CrossE_L + PR_L | PAGNN | 74.77 | ± | 45.85 | ± | 56.89 | ± | 42.80 | ± | 201.0 |
| | | 4.57 | | 0.25 | | 3.91 | | 0.10 | | |
| CrossE_L + PR_L | SAGE | 81.41 | ± | 57.70 | ± | 80.23 | ± | 56.70 | ± | 163.5 |
| | | 0.57 | | 1.15 | | 1.50 | | 1.13 | | |
| CrossE_L + PR_L + Triplet_L | ALL | 80.78 | ± | 51.14 | ± | 86.11 | ± | 57.00 | ± | 167.5 |
| | | 10.85 | | 1.41 | | 2.53 | | 1.03 | | |
| CrossE_L + PR_L + Triplet_L | GAT | 99.82 | ± | 93.01 | ± | 99.53 | ± | 95.29 | ± | 31.625 |
| | | 0.09 | | 0.72 | | 0.19 | | 0.24 | | |
| CrossE_L + PR_L + Triplet_L | GCN | 99.68 | ± | 94.29 | ± | 99.00 | ± | 95.59 | ± | 38.375 |
| | | 0.07 | | 0.53 | | 0.28 | | 0.62 | | |

Table 28. Results for Lp Auroc (↑) (continued)

| Loss Type | Model | Cora ↓ Citeseer | | Cora ↓ Bitcoin | | Citeseer ↓ Cora | | Citeseer ↓ Bitcoin | | Average Rank |
|---|---|---|---|---|---|---|---|---|---|---|
| CrossE_L + PR_L + Triplet_L | GIN | 97.31 | ± 1.77 | 55.99 | ± 0.63 | 92.59 | ± 2.50 | 57.37 | ± 0.56 | 136.0 |
| CrossE_L + PR_L + Triplet_L | MPNN | 97.89 | ± 1.43 | 93.97 | ± 0.76 | 89.03 | ± 3.91 | 94.32 | ± 0.49 | 87.75 |
| CrossE_L + PR_L + Triplet_L | PAGNN | 82.80 | ± 2.88 | 49.09 | ± 0.48 | 72.99 | ± 6.28 | 49.23 | ± 0.27 | 183.25 |
| CrossE_L + PR_L + Triplet_L | SAGE | 99.52 | ± 0.20 | 78.06 | ± 2.81 | 98.80 | ± 0.48 | 76.04 | ± 2.21 | 82.125 |
| CrossE_L + Triplet_L | ALL | 99.32 | ± 0.05 | 70.63 | ± 0.53 | 97.02 | ± 0.63 | 67.68 | ± 0.53 | 99.5 |
| CrossE_L + Triplet_L | GAT | 99.93 ± 0.02 | | 95.30 ± 0.47 | | 99.71 ± 0.02 | | 95.13 | ± 0.61 | 2.5 |
| CrossE_L + Triplet_L | GCN | 99.85 | ± 0.03 | 94.52 | ± 0.44 | 99.45 | ± 0.07 | 94.29 | ± 0.79 | 29.625 |
| CrossE_L + Triplet_L | GIN | 99.01 | ± 0.12 | 57.75 | ± 0.59 | 96.69 | ± 0.44 | 59.62 | ± 0.67 | 117.125 |
| CrossE_L + Triplet_L | MPNN | 99.67 | ± 0.08 | 93.43 | ± 0.52 | 98.25 | ± 0.24 | 92.06 | ± 1.01 | 67.375 |
| CrossE_L + Triplet_L | PAGNN | 91.35 | ± 5.37 | 45.34 | ± 1.02 | 93.41 | ± 1.00 | 47.89 | ± 0.98 | 160.875 |
| CrossE_L + Triplet_L | SAGE | 99.75 | ± 0.05 | 80.80 | ± 1.61 | 99.24 | ± 0.14 | 80.53 | ± 0.84 | 66.75 |
| PMI_L | ALL | 82.78 | ± 1.59 | 71.08 | ± 1.35 | 91.19 | ± 0.71 | 70.33 | ± 0.68 | 132.75 |
| PMI_L | GAT | 99.92 | ± 0.02 | 94.89 | ± 0.72 | 99.68 | ± 0.07 | 95.10 | ± 0.39 | 8.375 |
| PMI_L | GCN | 99.79 | ± 0.03 | 94.30 | ± 0.41 | 99.16 | ± 0.11 | 94.81 | ± 0.81 | 36.5 |
| PMI_L | GIN | 89.39 | ± 1.75 | 54.35 | ± 0.40 | 87.39 | ± 0.97 | 51.79 | ± 0.83 | 163.25 |
| PMI_L | MPNN | 99.67 | ± 0.05 | 93.36 | ± 0.72 | 97.67 | ± 0.30 | 93.11 | ± 0.91 | 67.625 |

Table 28. Results for Lp Auroc (↑) (continued)

| Loss Type | Model | Cora ↓ Citeseer | | Cora ↓ Bitcoin | | Citeseer ↓ Cora | | Citeseer ↓ Bitcoin | | Average Rank |
|---|---|---|---|---|---|---|---|---|---|---|
| PMI_L | PAGNN | 75.99 | ± 2.07 | 46.92 | ± 0.63 | 78.25 | ± 1.12 | 42.84 | ± 0.68 | 191.75 |
| PMI_L | SAGE | 82.93 | ± 1.72 | 59.36 | ± 2.05 | 80.31 | ± 1.13 | 55.66 | ± 1.13 | 160.875 |
| PMI_L + PR_L | ALL | 67.49 | ± 3.85 | 66.97 | ± 1.76 | 88.29 | ± 0.41 | 69.36 | ± 1.10 | 147.5 |
| PMI_L + PR_L | GAT | 99.86 | ± 0.08 | 93.97 | ± 0.92 | 96.29 | ± 4.21 | 93.30 | ± 1.22 | 60.75 |
| PMI_L + PR_L | GCN | 99.79 | ± 0.05 | 94.88 | ± 0.47 | 99.19 | ± 0.27 | 93.83 | ± 0.82 | 38.625 |
| PMI_L + PR_L | GIN | 88.40 | ± 2.20 | 54.39 | ± 0.70 | 86.16 | ± 3.83 | 54.97 | ± 0.53 | 163.25 |
| PMI_L + PR_L | MPNN | 99.63 | ± 0.10 | 93.17 | ± 0.68 | 91.91 | ± 1.34 | 90.83 | ± 1.57 | 89.625 |
| PMI_L + PR_L | PAGNN | 76.59 | ± 1.98 | 44.01 | ± 0.84 | 74.43 | ± 0.86 | 42.35 | ± 0.16 | 198.0 |
| PMI_L + PR_L | SAGE | 83.82 | ± 1.16 | 57.83 | ± 1.74 | 81.99 | ± 1.16 | 55.66 | ± 1.36 | 160.625 |
| PMI_L + PR_L + Triplet_L | ALL | 85.42 | ± 1.40 | 58.99 | ± 2.42 | 92.30 | ± 0.76 | 59.93 | ± 0.83 | 141.5 |
| PMI_L + PR_L + Triplet_L | GAT | 99.91 | ± 0.02 | 95.12 | ± 0.67 | 99.46 | ± 0.24 | 94.23 | ± 0.81 | 21.375 |
| PMI_L + PR_L + Triplet_L | GCN | 99.82 | ± 0.02 | 95.14 | ± 0.70 | 99.28 | ± 0.13 | 93.90 | ± 0.64 | 29.625 |
| PMI_L + PR_L + Triplet_L | GIN | 91.95 | ± 2.32 | 56.10 | ± 0.36 | 90.00 | ± 1.16 | 51.70 | ± 0.47 | 152.75 |
| PMI_L + PR_L + Triplet_L | MPNN | 99.64 | ± 0.03 | 94.06 | ± 0.57 | 93.95 | ± 0.73 | 92.22 | ± 1.11 | 76.875 |
| PMI_L + PR_L + Triplet_L | PAGNN | 76.56 | ± 2.08 | 47.31 | ± 0.29 | 76.17 | ± 1.83 | 46.67 | ± 0.43 | 192.0 |
| PMI_L + PR_L + Triplet_L | SAGE | 97.01 | ± 2.32 | 77.70 | ± 1.67 | 96.16 | ± 0.83 | 75.24 | ± 0.97 | 101.0 |

Table 28. Results for Lp Auroc (↑) (continued)

| Loss Type | Model | Cora ↓ Citeseer | | Cora ↓ Bitcoin | | Citeseer ↓ Cora | | Citeseer ↓ Bitcoin | | Average Rank |
|---|---|---|---|---|---|---|---|---|---|---|
| PMI_L + Triplet_L | ALL | 99.01 | ± 0.18 | 65.80 | ± 0.79 | 96.89 | ± 0.49 | 75.51 | ± 0.74 | 100.125 |
| PMI_L + Triplet_L | GAT | 99.92 | ± 0.02 | 94.79 | ± 0.70 | 99.66 | ± 0.02 | 94.98 | ± 0.64 | 10.625 |
| PMI_L + Triplet_L | GCN | 99.80 | ± 0.02 | 94.74 | ± 0.54 | 99.30 | ± 0.13 | 94.13 | ± 0.41 | 33.25 |
| PMI_L + Triplet_L | GIN | 95.47 | ± 0.77 | 56.52 | ± 0.67 | 91.61 | ± 1.38 | 57.81 | ± 0.48 | 139.125 |
| PMI_L + Triplet_L | MPNN | 99.67 | ± 0.04 | 93.46 | ± 0.35 | 97.76 | ± 0.22 | 93.54 | ± 1.01 | 65.125 |
| PMI_L + Triplet_L | PAGNN | 74.96 | ± 2.46 | 45.46 | ± 0.50 | 79.46 | ± 1.32 | 42.95 | ± 0.50 | 193.0 |
| PMI_L + Triplet_L | SAGE | 96.77 | ± 1.07 | 70.18 | ± 3.07 | 96.75 | ± 0.29 | 78.70 | ± 0.87 | 101.5 |
| PR_L | ALL | 59.59 | ± 3.25 | 66.82 | ± 1.23 | 68.39 | ± 7.32 | 69.27 | ± 1.43 | 160.25 |
| PR_L | GAT | 99.70 | ± 0.12 | 91.37 | ± 0.68 | 99.33 | ± 0.18 | 92.10 | ± 0.91 | 60.25 |
| PR_L | GCN | 99.16 | ± 0.44 | 91.06 | ± 0.21 | 98.55 | ± 0.42 | 90.17 | ± 1.84 | 80.5 |
| PR_L | GIN | 90.29 | ± 4.09 | 55.34 | ± 0.27 | 76.06 | ± 8.05 | 55.59 | ± 0.25 | 165.875 |
| PR_L | MPNN | 93.42 | ± 8.16 | 94.13 | ± 0.29 | 84.59 | ± 3.55 | 94.34 | ± 0.67 | 93.0 |
| PR_L | PAGNN | 72.00 | ± 6.55 | 44.36 | ± 0.33 | 54.42 | ± 1.54 | 47.79 | ± 0.32 | 200.75 |
| PR_L | SAGE | 81.36 | ± 1.46 | 56.97 | ± 0.76 | 81.49 | ± 1.39 | 58.37 | ± 0.90 | 161.25 |
| PR_L + Triplet_L | ALL | 69.03 | ± 5.42 | 60.25 | ± 2.00 | 72.69 | ± 4.85 | 65.82 | ± 0.60 | 166.0 |
| PR_L + Triplet_L | GAT | 99.69 | ± 0.35 | 87.43 | ± 1.78 | 99.18 | ± 0.46 | 92.54 | ± 0.50 | 64.25 |

Table 28. Results for Lp Auroc (↑) (continued)

| Loss Type | Model | Cora ↓ Citeseer | | Cora ↓ Bitcoin | | Citeseer ↓ Cora | | Citeseer ↓ Bitcoin | | Average Rank |
|---|---|---|---|---|---|---|---|---|---|---|
| PR_L + Triplet_L | GCN | 99.50 0.20 | ± | 92.93 1.03 | ± | 98.75 0.36 | ± | 93.32 0.70 | ± | 70.625 |
| PR_L + Triplet_L | GIN | 91.34 5.21 | ± | 54.26 0.28 | ± | 81.10 7.02 | ± | 53.88 0.39 | ± | 164.875 |
| PR_L + Triplet_L | MPNN | 94.49 7.16 | ± | 94.66 0.61 | ± | 89.18 5.53 | ± | 94.59 0.29 | ± | 79.25 |
| PR_L + Triplet_L | PAGNN | 74.72 4.39 | ± | 48.49 0.43 | ± | 60.20 3.95 | ± | 48.50 0.24 | ± | 192.75 |
| PR_L + Triplet_L | SAGE | 85.40 2.05 | ± | 58.68 0.71 | ± | 89.22 7.97 | ± | 58.29 0.41 | ± | 149.25 |
| Triplet_L | ALL | 99.10 0.20 | ± | 64.81 0.75 | ± | 97.12 0.42 | ± | 78.37 0.99 | ± | 97.375 |
| Triplet_L | GAT | 99.94 0.02 | ± | 94.84 0.60 | ± | 99.72 0.03 | ± | 94.93 0.78 | ± | 7.0 |
| Triplet_L | GCN | 99.88 0.02 | ± | 94.99 0.68 | ± | 99.39 0.14 | ± | 94.85 0.52 | ± | 17.25 |
| Triplet_L | GIN | 99.26 0.13 | ± | 58.91 0.16 | ± | 96.76 0.87 | ± | 62.33 0.62 | ± | 111.75 |
| Triplet_L | MPNN | 99.70 0.06 | ± | 94.04 0.61 | ± | 98.14 0.15 | ± | 94.31 1.45 | ± | 53.625 |
| Triplet_L | PAGNN | 92.01 5.20 | ± | 48.37 0.98 | ± | 94.42 0.32 | ± | 51.39 0.47 | ± | 151.25 |
| Triplet_L | SAGE | 99.70 0.11 | ± | 82.35 1.40 | ± | 99.08 0.19 | ± | 83.94 0.73 | ± | 70.0 |

Table 29. Lp F1 Performance (↑): This table presents models (Loss function and GNN) ranked by their average performance in terms of lp f1. Top-ranked results are highlighted in <span style="color:red">red</span>, second-ranked in <span style="color:blue">blue</span>, and third-ranked in <span style="color:green">green</span>.

| Loss Type | Model | Cora ↓ Citeseer | | Cora ↓ Bitcoin | | Citeseer ↓ Cora | | Citeseer ↓ Bitcoin | | Average Rank |
|---|---|---|---|---|---|---|---|---|---|---|
| Contr_l | ALL | 90.58 ± 1.55 | | 53.32 ± 1.55 | ± | 87.68 ± 1.05 | | 51.87 ± 0.56 | ± | 133.875 |
| Contr_l | GAT | 98.76 ± 0.18 | | 83.36 ± 1.26 | ± | 97.82 ± 0.15 | ± | 84.85 ± 0.47 | ± | 16.75 |
| Contr_l | GCN | 98.05 ± 0.14 | | 83.45 ± 0.73 | ± | 96.58 ± 0.36 | ± | 83.77 ± 0.80 | ± | 37.0 |
| Contr_l | GIN | 93.55 ± 0.74 | | 55.78 ± 0.77 | ± | 91.09 ± 1.08 | ± | 63.10 ± 0.83 | ± | 111.75 |
| Contr_l | MPNN | 97.71 ± 0.34 | | 81.53 ± 1.12 | ± | 92.66 ± 1.50 | ± | 82.18 ± 1.12 | ± | 73.25 |
| Contr_l | PAGNN | 84.46 ± 1.48 | | 51.56 ± 0.04 | ± | 84.65 ± 1.25 | ± | 51.66 ± 0.06 | ± | 154.5 |
| Contr_l | SAGE | 95.43 ± 0.35 | | 64.68 ± 2.93 | ± | 94.02 ± 1.33 | ± | 70.31 ± 4.29 | ± | 92.25 |
| Contr_l + CrossE_L | ALL | 90.99 ± 1.17 | | 60.03 ± 1.08 | ± | 86.46 ± 1.33 | ± | 53.04 ± 1.04 | ± | 126.5 |
| Contr_l + CrossE_L | GAT | 98.66 ± 0.17 | | 83.44 ± 0.76 | ± | 97.34 ± 0.35 | ± | 83.27 ± 0.75 | ± | 28.875 |
| Contr_l + CrossE_L | GCN | 98.25 ± 0.11 | | 82.80 ± 1.24 | ± | 96.29 ± 0.29 | ± | 83.96 ± 0.34 | ± | 38.625 |
| Contr_l + CrossE_L | GIN | 89.71 ± 2.14 | | 52.00 ± 1.06 | ± | 90.88 ± 1.07 | ± | 57.09 ± 0.41 | ± | 127.25 |
| Contr_l + CrossE_L | MPNN | 97.60 ± 0.39 | | 81.73 ± 1.28 | ± | 93.10 ± 0.70 | ± | 81.78 ± 1.05 | ± | 74.75 |
| Contr_l + CrossE_L | PAGNN | 83.50 ± 1.21 | | 51.60 ± 0.04 | ± | 83.51 ± 1.41 | ± | 51.61 ± 0.02 | ± | 157.875 |
| Contr_l + CrossE_L | SAGE | 95.15 ± 0.56 | | 62.14 ± 1.27 | ± | 94.57 ± 1.01 | ± | 64.19 ± 2.03 | ± | 99.375 |
| Contr_l + CrossE_L + PMI_L | ALL | 80.85 ± 4.16 | | 62.76 ± 0.74 | ± | 85.81 ± 1.17 | ± | 68.24 ± 0.59 | ± | 125.5 |

Table 29. Results for Lp F1 (↑) (continued)

| Loss Type | Model | Cora ↓ Citeseer | ± | Cora ↓ Bitcoin | ± | Citeseer ↓ Cora | ± | Citeseer ↓ Bitcoin | ± | Average Rank |
|---|---|---|---|---|---|---|---|---|---|---|
| Contr_l + CrossE_L + PMI_L | GAT | 98.96 | ± | 84.22 | ± | 97.65 | ± | 84.55 | ± | 9.625 |
| | | 0.20 | | 0.94 | | 0.23 | | 0.37 | | |
| Contr_l + CrossE_L + PMI_L | GCN | 98.16 | ± | 83.75 | ± | 96.06 | ± | 83.27 | ± | 39.875 |
| | | 0.16 | | 0.42 | | 0.37 | | 0.89 | | |
| Contr_l + CrossE_L + PMI_L | GIN | 84.07 | ± | 51.46 | ± | 82.26 | ± | 51.54 | ± | 166.875 |
| | | 2.16 | | 0.00 | | 1.59 | | 0.06 | | |
| Contr_l + CrossE_L + PMI_L | MPNN | 98.36 | ± | 83.41 | ± | 93.80 | ± | 83.07 | ± | 49.75 |
| | | 0.19 | | 1.01 | | 0.59 | | 0.69 | | |
| Contr_l + CrossE_L + PMI_L | PAGNN | 70.31 | ± | 51.46 | ± | 73.06 | ± | 51.47 | ± | 194.125 |
| | | 0.45 | | 0.01 | | 0.36 | | 0.01 | | |
| Contr_l + CrossE_L + PMI_L | SAGE | 77.94 | ± | 54.63 | ± | 79.90 | ± | 56.71 | ± | 148.375 |
| | | 1.00 | | 1.21 | | 1.86 | | 0.25 | | |
| Contr_l + CrossE_L + PMI_L + PR_L | ALL | 69.21 | ± | 68.07 | ± | 84.14 | ± | 68.89 | ± | 134.75 |
| | | 1.57 | | 0.28 | | 1.52 | | 0.29 | | |
| Contr_l + CrossE_L + PMI_L + PR_L | GAT | 98.84 | ± | 84.16 | ± | 97.44 | ± | 83.68 | ± | 19.0 |
| | | 0.26 | | 1.77 | | 0.22 | | 1.79 | | |
| Contr_l + CrossE_L + PMI_L + PR_L | GCN | 98.11 | ± | 83.83 | ± | 96.05 | ± | 83.17 | ± | 41.375 |
| | | 0.12 | | 0.53 | | 0.49 | | 1.12 | | |
| Contr_l + CrossE_L + PMI_L + PR_L | GIN | 78.52 | ± | 51.46 | ± | 77.11 | ± | 51.53 | ± | 174.75 |
| | | 2.19 | | 0.00 | | 2.31 | | 0.05 | | |
| Contr_l + CrossE_L + PMI_L + PR_L | MPNN | 98.41 | ± | 82.05 | ± | 89.09 | ± | 79.20 | ± | 72.5 |
| | | 0.11 | | 0.96 | | 3.35 | | 2.75 | | |
| Contr_l + CrossE_L + PMI_L + PR_L | PAGNN | 71.96 | ± | 51.48 | ± | 73.22 | ± | 51.72 | ± | 179.25 |
| | | 1.26 | | 0.02 | | 1.04 | | 0.07 | | |
| Contr_l + CrossE_L + PMI_L + PR_L | SAGE | 76.71 | ± | 52.85 | ± | 78.07 | ± | 52.95 | ± | 158.5 |
| | | 1.71 | | 0.57 | | 2.03 | | 1.40 | | |
| Contr_l + CrossE_L + PMI_L + PR_L + Triplet_L | ALL | 75.33 | ± | 59.24 | ± | 85.69 | ± | 63.04 | ± | 138.0 |
| | | 1.34 | | 1.09 | | 1.47 | | 1.64 | | |
| Contr_l + CrossE_L + PMI_L + PR_L + Triplet_L | GAT | 98.94 | ± | 83.73 | ± | 96.95 | ± | 83.88 | ± | 20.375 |
| | | 0.19 | | 1.40 | | 0.77 | | 1.04 | | |

Table 29. Results for Lp F1 (↑) (continued)

| Loss Type | Model | Cora ↓ Citeseer | | Cora ↓ Bitcoin | | Citeseer ↓ Cora | | Citeseer ↓ Bitcoin | | Average Rank |
|---|---|---|---|---|---|---|---|---|---|---|
| Contr_l + CrossE_L + PMI_L + PR_L + Triplet_L | GCN | 98.15 0.20 | ± | 83.46 0.98 | ± | 96.09 0.20 | ± | 83.89 0.68 | ± | 36.875 |
| Contr_l + CrossE_L + PMI_L + PR_L + Triplet_L | GIN | 80.90 2.93 | ± | 51.91 0.64 | ± | 83.41 1.34 | ± | 51.48 0.01 | ± | 161.5 |
| Contr_l + CrossE_L + PMI_L + PR_L + Triplet_L | MPNN | 98.19 0.23 | ± | 83.20 0.19 | ± | 90.56 2.28 | ± | 79.55 1.25 | ± | 68.0 |
| Contr_l + CrossE_L + PMI_L + PR_L + Triplet_L | PAGNN | 72.41 1.24 | ± | 51.51 0.03 | ± | 73.15 0.42 | ± | 51.49 0.02 | ± | 182.375 |
| Contr_l + CrossE_L + PMI_L + PR_L + Triplet_L | SAGE | 88.14 3.73 | ± | 65.72 1.06 | ± | 89.89 0.33 | ± | 64.91 1.84 | ± | 110.75 |
| Contr_l + CrossE_L + PMI_L + Triplet_L | ALL | 93.84 1.23 | ± | 67.94 1.21 | ± | 91.15 0.74 | ± | 69.73 0.69 | ± | 96.5 |
| Contr_l + CrossE_L + PMI_L + Triplet_L | GAT | 99.00 0.07 | ± | 84.77 0.40 | ± | 97.39 0.13 | ± | 84.34 0.60 | ± | 8.0 |
| Contr_l + CrossE_L + PMI_L + Triplet_L | GCN | 98.26 0.24 | ± | 84.51 0.85 | ± | 95.59 0.75 | ± | 83.88 1.02 | ± | 30.0 |
| Contr_l + CrossE_L + PMI_L + Triplet_L | GIN | 84.87 1.83 | ± | 53.27 1.19 | ± | 84.15 2.73 | ± | 52.72 0.76 | ± | 144.25 |
| Contr_l + CrossE_L + PMI_L + Triplet_L | MPNN | 98.36 0.16 | ± | 83.76 0.50 | ± | 93.50 0.92 | ± | 82.65 0.88 | ± | 48.875 |
| Contr_l + CrossE_L + PMI_L + Triplet_L | PAGNN | 71.81 0.43 | ± | 51.46 0.00 | ± | 73.34 0.20 | ± | 51.47 0.01 | ± | 190.875 |
| Contr_l + CrossE_L + PMI_L + Triplet_L | SAGE | 90.03 1.56 | ± | 61.29 2.38 | ± | 88.39 0.99 | ± | 64.77 0.85 | ± | 115.625 |
| Contr_l + CrossE_L + PR_L | ALL | 69.13 0.86 | ± | 62.22 2.06 | ± | 72.93 1.93 | ± | 66.58 0.43 | ± | 154.875 |
| Contr_l + CrossE_L + PR_L | GAT | 97.57 1.63 | ± | 83.00 0.81 | ± | 96.89 0.33 | ± | 81.98 0.88 | ± | 54.875 |

Continued on next page

Table 29. Results for Lp F1 (↑) (continued)

| Loss Type | Model | Cora ↓ Citeseer | | Cora ↓ Bitcoin | | Citeseer ↓ Cora | | Citeseer ↓ Bitcoin | | Average Rank |
|---|---|---|---|---|---|---|---|---|---|---|
| Contr_l + CrossE_L + PR_L | GCN | 97.31 | ± 0.30 | 81.11 | ± 0.80 | 94.86 | ± 0.46 | 82.54 | ± 1.48 | 69.5 |
| Contr_l + CrossE_L + PR_L | GIN | 87.66 | ± 2.40 | 51.46 | ± 0.00 | 75.65 | ± 1.20 | 51.46 | ± 0.01 | 176.0 |
| Contr_l + CrossE_L + PR_L | MPNN | 90.20 | ± 6.53 | 83.07 | ± 0.80 | 84.01 | ± 4.40 | 83.36 | ± 0.51 | 89.25 |
| Contr_l + CrossE_L + PR_L | PAGNN | 70.91 | ± 3.85 | 51.80 | ± 0.23 | 68.70 | ± 1.42 | 52.11 | ± 0.19 | 180.25 |
| Contr_l + CrossE_L + PR_L | SAGE | 90.65 | ± 7.93 | 64.05 | ± 2.24 | 86.63 | ± 7.37 | 60.16 | ± 2.20 | 118.75 |
| Contr_l + CrossE_L + PR_L + Triplet_L | ALL | 80.41 | ± 1.58 | 51.47 | ± 0.02 | 85.92 | ± 1.12 | 52.68 | ± 0.98 | 154.75 |
| Contr_l + CrossE_L + PR_L + Triplet_L | GAT | 98.70 | ± 0.18 | 82.92 | ± 1.50 | 97.29 | ± 0.71 | 84.26 | ± 0.63 | 26.875 |
| Contr_l + CrossE_L + PR_L + Triplet_L | GCN | 97.87 | ± 0.30 | 83.30 | ± 1.32 | 95.55 | ± 0.36 | 83.19 | ± 2.10 | 52.625 |
| Contr_l + CrossE_L + PR_L + Triplet_L | GIN | 90.21 | ± 1.31 | 52.08 | ± 0.54 | 89.35 | ± 0.83 | 56.52 | ± 1.04 | 129.25 |
| Contr_l + CrossE_L + PR_L + Triplet_L | MPNN | 96.91 | ± 0.88 | 83.50 | ± 0.72 | 89.64 | ± 0.65 | 80.33 | ± 2.06 | 73.0 |
| Contr_l + CrossE_L + PR_L + Triplet_L | PAGNN | 76.25 | ± 1.58 | 51.51 | ± 0.02 | 72.90 | ± 1.03 | 51.96 | ± 0.09 | 174.625 |
| Contr_l + CrossE_L + PR_L + Triplet_L | SAGE | 96.24 | ± 0.39 | 66.96 | ± 1.69 | 93.32 | ± 0.92 | 60.59 | ± 1.90 | 97.5 |
| Contr_l + CrossE_L + Triplet_L | ALL | 93.01 | ± 1.50 | 60.73 | ± 0.84 | 90.05 | ± 0.94 | 57.15 | ± 0.97 | 114.25 |
| Contr_l + CrossE_L + Triplet_L | GAT | 98.90 | ± 0.19 | 83.69 | ± 1.10 | 97.67 | ± 0.27 | 84.20 | ± 0.51 | 16.5 |
| Contr_l + CrossE_L + Triplet_L | GCN | 98.47 | ± 0.12 | 83.41 | ± 0.98 | 96.48 | ± 0.40 | 83.16 | ± 0.57 | 35.625 |
| Contr_l + CrossE_L + Triplet_L | GIN | 95.00 | ± 0.69 | 57.77 | ± 1.11 | 92.26 | ± 1.04 | 63.92 | ± 0.49 | 106.5 |

Table 29. Results for Lp F1 (↑) (continued)

| Loss Type | Model | Cora ↓ Citeseer | | Cora ↓ Bitcoin | | Citeseer ↓ Cora | | Citeseer ↓ Bitcoin | | Average Rank |
|---|---|---|---|---|---|---|---|---|---|---|
| Contr_l + CrossE_L + Triplet_L | MPNN | 97.88 | ± 0.25 | 81.15 | ± 1.17 | 94.04 | ± 0.52 | 82.73 | ± 1.00 | 68.5 |
| Contr_l + CrossE_L + Triplet_L | PAGNN | 82.35 | ± 4.79 | 51.53 | ± 0.03 | 87.63 | ± 1.28 | 51.62 | ± 0.06 | 151.5 |
| Contr_l + CrossE_L + Triplet_L | SAGE | 96.62 | ± 0.50 | 66.56 | ± 2.80 | 95.19 | ± 0.40 | 70.99 | ± 2.08 | 83.0 |
| Contr_l + PMI_L | ALL | 83.23 | ± 6.82 | 51.66 | ± 0.27 | 87.39 | ± 1.28 | 64.26 | ± 1.00 | 136.0 |
| Contr_l + PMI_L | GAT | 98.97 | ± 0.20 | 84.91 | ± 0.98 | 97.53 | ± 0.19 | 84.30 | ± 1.09 | 7.75 |
| Contr_l + PMI_L | GCN | 98.16 | ± 0.20 | 83.40 | ± 0.92 | 96.04 | ± 0.51 | 84.00 | ± 1.30 | 38.125 |
| Contr_l + PMI_L | GIN | 83.06 | ± 4.21 | 53.52 | ± 0.19 | 81.97 | ± 1.32 | 52.50 | ± 0.78 | 150.75 |
| Contr_l + PMI_L | MPNN | 98.33 | ± 0.18 | 81.44 | ± 0.64 | 94.08 | ± 0.39 | 82.98 | ± 0.22 | 59.0 |
| Contr_l + PMI_L | PAGNN | 71.45 | ± 0.91 | 51.46 | ± 0.00 | 72.84 | ± 0.15 | 51.47 | ± 0.00 | 194.875 |
| Contr_l + PMI_L | SAGE | 78.15 | ± 2.24 | 55.00 | ± 1.38 | 84.32 | ± 2.85 | 58.97 | ± 0.68 | 141.125 |
| Contr_l + PMI_L + PR_L | ALL | 68.73 | ± 0.72 | 63.51 | ± 0.92 | 83.62 | ± 1.30 | 69.38 | ± 0.42 | 140.75 |
| Contr_l + PMI_L + PR_L | GAT | 98.70 | ± 0.32 | 84.39 | ± 0.73 | 94.28 | ± 3.65 | 75.91 | ± 2.14 | 45.875 |
| Contr_l + PMI_L + PR_L | GCN | 98.30 | ± 0.24 | 83.81 | ± 1.21 | 96.08 | ± 0.42 | 83.41 | ± 0.94 | 34.125 |
| Contr_l + PMI_L + PR_L | GIN | 81.76 | ± 5.17 | 51.46 | ± 0.00 | 79.52 | ± 2.36 | 51.52 | ± 0.05 | 171.375 |
| Contr_l + PMI_L + PR_L | MPNN | 98.48 | ± 0.08 | 83.58 | ± 0.61 | 87.20 | ± 3.65 | 79.34 | ± 0.94 | 62.75 |
| Contr_l + PMI_L + PR_L | PAGNN | 70.81 | ± 0.65 | 51.46 | ± 0.00 | 72.94 | ± 0.24 | 51.47 | ± 0.01 | 194.625 |

Table 29. Results for Lp F1 (↑) (continued)

| Loss Type | Model | Cora ↓ Citeseer | | Cora ↓ Bitcoin | | Citeseer ↓ Cora | | Citeseer ↓ Bitcoin | | Average Rank |
|---|---|---|---|---|---|---|---|---|---|---|
| Contr_l + PMI_L + PR_L | SAGE | 77.86 1.12 | ± | 55.00 1.52 | ± | 82.23 1.57 | ± | 56.90 1.06 | ± | 146.625 |
| Contr_l + PMI_L + PR_L + Triplet_L | ALL | 79.68 2.60 | ± | 51.46 0.00 | ± | 87.15 1.04 | ± | 55.01 1.28 | ± | 152.0 |
| Contr_l + PMI_L + PR_L + Triplet_L | GAT | 98.47 0.52 | ± | 83.31 1.24 | ± | 95.91 1.02 | ± | 79.00 2.91 | ± | 50.125 |
| Contr_l + PMI_L + PR_L + Triplet_L | GCN | 98.10 0.23 | ± | 84.29 0.59 | ± | 96.12 0.52 | ± | 83.64 0.92 | ± | 34.0 |
| Contr_l + PMI_L + PR_L + Triplet_L | GIN | 85.07 0.74 | ± | 54.27 0.82 | ± | 86.27 1.90 | ± | 51.64 0.27 | ± | 142.25 |
| Contr_l + PMI_L + PR_L + Triplet_L | MPNN | 98.47 0.16 | ± | 82.70 0.80 | ± | 88.88 0.86 | ± | 78.33 1.58 | ± | 70.5 |
| Contr_l + PMI_L + PR_L + Triplet_L | PAGNN | 72.89 0.81 | ± | 51.46 0.00 | ± | 73.26 0.75 | ± | 51.47 0.01 | ± | 189.375 |
| Contr_l + PMI_L + PR_L + Triplet_L | SAGE | 92.95 1.10 | ± | 62.40 0.62 | ± | 90.51 1.67 | ± | 64.48 2.54 | ± | 108.5 |
| Contr_l + PR_L | ALL | 70.00 1.65 | ± | 66.78 0.69 | ± | 75.00 1.28 | ± | 68.15 0.81 | ± | 144.375 |
| Contr_l + PR_L | GAT | 97.99 0.72 | ± | 82.18 0.19 | ± | 96.84 0.24 | ± | 81.84 1.09 | ± | 56.375 |
| Contr_l + PR_L | GCN | 97.34 0.39 | ± | 82.31 0.63 | ± | 94.84 0.59 | ± | 83.12 1.13 | ± | 63.0 |
| Contr_l + PR_L | GIN | 85.73 5.08 | ± | 51.86 0.39 | ± | 74.33 6.45 | ± | 51.59 0.18 | ± | 161.75 |
| Contr_l + PR_L | MPNN | 90.36 2.50 | ± | 83.10 1.14 | ± | 86.43 2.01 | ± | 83.77 0.35 | ± | 80.5 |
| Contr_l + PR_L | PAGNN | 69.00 3.25 | ± | 51.82 0.10 | ± | 68.23 0.22 | ± | 52.56 0.18 | ± | 181.0 |
| Contr_l + PR_L | SAGE | 90.52 7.84 | ± | 65.08 2.26 | ± | 88.61 6.85 | ± | 54.88 0.80 | ± | 117.25 |
| Contr_l + PR_L + Triplet_L | ALL | 77.94 7.61 | ± | 51.53 0.13 | ± | 85.17 2.63 | ± | 51.46 0.01 | ± | 169.875 |

Table 29. Results for Lp F1 (↑) (continued)

| Loss Type | Model | Cora ↓ Citeseer | | Cora ↓ Bitcoin | | Citeseer ↓ Cora | | Citeseer ↓ Bitcoin | | Average Rank |
|---|---|---|---|---|---|---|---|---|---|---|
| Contr_l + PR_L + Triplet_L | GAT | 98.69 | ± 0.21 | 84.08 | ± 1.00 | 97.25 | ± 0.61 | 82.54 | ± 1.42 | 29.125 |
| Contr_l + PR_L + Triplet_L | GCN | 97.75 | ± 0.31 | 83.05 | ± 1.28 | 95.66 | ± 0.36 | 84.28 | ± 1.10 | 47.25 |
| Contr_l + PR_L + Triplet_L | GIN | 92.14 | ± 2.00 | 55.75 | ± 0.95 | 86.82 | ± 3.24 | 53.35 | ± 0.40 | 126.75 |
| Contr_l + PR_L + Triplet_L | MPNN | 95.75 | ± 2.65 | 82.43 | ± 0.59 | 87.18 | ± 4.07 | 81.68 | ± 0.50 | 85.75 |
| Contr_l + PR_L + Triplet_L | PAGNN | 74.38 | ± 1.66 | 51.47 | ± 0.00 | 74.90 | ± 3.77 | 51.72 | ± 0.08 | 175.25 |
| Contr_l + PR_L + Triplet_L | SAGE | 95.96 | ± 0.27 | 64.58 | ± 2.63 | 94.08 | ± 0.70 | 63.65 | ± 1.61 | 97.125 |
| Contr_l + Triplet_L | ALL | 94.20 | ± 1.02 | 55.51 | ± 1.31 | 90.37 | ± 0.84 | 54.85 | ± 1.16 | 117.5 |
| Contr_l + Triplet_L | GAT | 98.92 | ± 0.10 | 84.06 | ± 1.13 | 97.71 | ± 0.30 | 83.54 | ± 1.07 | 17.75 |
| Contr_l + Triplet_L | GCN | 98.34 | ± 0.10 | 83.43 | ± 0.68 | 96.53 | ± 0.48 | 83.88 | ± 0.67 | 31.0 |
| Contr_l + Triplet_L | GIN | 94.74 | ± 0.64 | 54.92 | ± 0.65 | 92.16 | ± 1.69 | 59.78 | ± 0.73 | 112.25 |
| Contr_l + Triplet_L | MPNN | 98.00 | ± 0.29 | 80.87 | ± 2.08 | 94.72 | ± 0.43 | 82.20 | ± 0.40 | 67.5 |
| Contr_l + Triplet_L | PAGNN | 83.58 | ± 4.74 | 51.72 | ± 0.11 | 87.15 | ± 1.03 | 51.61 | ± 0.05 | 149.75 |
| Contr_l + Triplet_L | SAGE | 96.86 | ± 0.32 | 65.86 | ± 3.06 | 94.94 | ± 0.71 | 67.59 | ± 1.55 | 87.5 |
| CrossE_L | ALL | 88.47 | ± 12.58 | 72.73 | ± 0.85 | 69.23 | ± 2.20 | 70.85 | ± 0.34 | 126.75 |
| CrossE_L | GAT | 88.20 | ± 13.71 | 53.29 | ± 0.48 | 85.46 | ± 10.52 | 79.03 | ± 0.61 | 123.0 |
| CrossE_L | GCN | 64.69 | ± 0.31 | 62.84 | ± 0.66 | 67.86 | ± 0.00 | 51.46 | ± 0.00 | 185.25 |

Continued on next page

Table 29.  Results for Lp F1 (↑) (continued)

| Loss Type | Model | Cora ↓ Citeseer | | Cora ↓ Bitcoin | | Citeseer ↓ Cora | | Citeseer ↓ Bitcoin | | Average Rank |
|---|---|---|---|---|---|---|---|---|---|---|
| CrossE_L | GIN | 64.56 | ± 0.01 | 51.46 | ± 0.00 | 67.86 | ± 0.00 | 51.46 | ± | 206.375 |
| CrossE_L | MPNN | 87.64 | ± 3.19 | 51.46 | ± 0.00 | 86.03 | ± 1.75 | 51.46 | ± 0.00 | 165.0 |
| CrossE_L | PAGNN | 71.33 | ± 1.01 | 51.46 | ± 0.00 | 68.10 | ± 0.18 | 51.46 | ± 0.01 | 200.5 |
| CrossE_L | SAGE | 76.57 | ± 3.95 | 56.84 | ± 0.25 | 70.35 | ± 2.66 | 51.62 | ± 0.08 | 166.875 |
| CrossE_L + PMI_L | ALL | 78.40 | ± 1.70 | 69.59 | ± 0.72 | 85.64 | ± 1.06 | 69.49 | ± 0.95 | 120.75 |
| CrossE_L + PMI_L | GAT | 98.99 | ± 0.14 | 83.99 | ± 0.22 | 97.63 | ± 0.23 | 83.94 | ± 0.49 | 13.75 |
| CrossE_L + PMI_L | GCN | 97.99 | ± 0.24 | 83.28 | ± 0.83 | 96.01 | ± 0.40 | 83.41 | ± 0.71 | 47.875 |
| CrossE_L + PMI_L | GIN | 81.88 | ± 2.88 | 51.54 | ± 0.16 | 79.42 | ± 1.12 | 51.75 | ± 0.42 | 161.875 |
| CrossE_L + PMI_L | MPNN | 98.17 | ± 0.16 | 82.95 | ± 0.59 | 93.67 | ± 1.24 | 82.63 | ± 0.36 | 60.875 |
| CrossE_L + PMI_L | PAGNN | 71.02 | ± 1.56 | 53.06 | ± 0.27 | 72.92 | ± 0.17 | 51.47 | ± 0.00 | 183.75 |
| CrossE_L + PMI_L | SAGE | 75.24 | ± 1.17 | 53.19 | ± 0.94 | 76.31 | ± 0.30 | 51.85 | ± 0.45 | 163.5 |
| CrossE_L + PMI_L + PR_L | ALL | 67.98 | ± 0.16 | 69.26 | ± 0.76 | 83.36 | ± 1.10 | 70.21 | ± 0.59 | 136.25 |
| CrossE_L + PMI_L + PR_L | GAT | 98.94 | ± 0.18 | 84.60 | ± 0.32 | 93.90 | ± 3.06 | 76.68 | ± 1.66 | 43.875 |
| CrossE_L + PMI_L + PR_L | GCN | 98.16 | ± 0.33 | 83.47 | ± 0.89 | 96.04 | ± 0.23 | 83.77 | ± 0.67 | 38.625 |
| CrossE_L + PMI_L + PR_L | GIN | 79.88 | ± 2.41 | 53.92 | ± 0.61 | 79.41 | ± 2.20 | 51.53 | ± 0.14 | 159.375 |
| CrossE_L + PMI_L + PR_L | MPNN | 98.28 | ± 0.26 | 82.13 | ± 0.48 | 89.02 | ± 5.34 | 81.93 | ± 0.91 | 70.875 |

Table 29. Results for Lp F1 (↑) (continued)

| Loss Type | Model | Cora ↓ Citeseer | | Cora ↓ Bitcoin | | Citeseer ↓ Cora | | Citeseer ↓ Bitcoin | | Average Rank |
|---|---|---|---|---|---|---|---|---|---|---|
| CrossE_L + PMI_L + PR_L | PAGNN | 71.91 1.21 | ± | 51.48 0.02 | ± | 72.57 0.88 | ± | 51.47 0.01 | ± | 190.5 |
| CrossE_L + PMI_L + PR_L | SAGE | 75.92 1.13 | ± | 53.65 0.79 | ± | 76.11 0.69 | ± | 53.63 0.67 | ± | 157.125 |
| CrossE_L + PMI_L + PR_L + Triplet_L | ALL | 75.67 1.39 | ± | 66.84 1.47 | ± | 86.01 0.59 | ± | 58.48 0.83 | ± | 132.25 |
| CrossE_L + PMI_L + PR_L + Triplet_L | GAT | 98.93 0.07 | ± | 84.45 0.45 | ± | 96.68 1.10 | ± | 83.61 1.20 | ± | 19.0 |
| CrossE_L + PMI_L + PR_L + Triplet_L | GCN | 98.10 0.33 | ± | 83.78 0.68 | ± | 95.85 0.50 | ± | 83.03 0.68 | ± | 45.125 |
| CrossE_L + PMI_L + PR_L + Triplet_L | GIN | 82.23 2.81 | ± | 51.65 0.24 | ± | 82.89 2.70 | ± | 51.50 0.06 | ± | 162.375 |
| CrossE_L + PMI_L + PR_L + Triplet_L | MPNN | 98.26 0.08 | ± | 82.83 1.39 | ± | 90.58 2.83 | ± | 82.76 0.86 | ± | 62.875 |
| CrossE_L + PMI_L + PR_L + Triplet_L | PAGNN | 73.36 0.48 | ± | 52.06 0.05 | ± | 73.03 0.17 | ± | 51.47 0.01 | ± | 181.125 |
| CrossE_L + PMI_L + PR_L + Triplet_L | SAGE | 88.49 5.12 | ± | 64.30 2.18 | ± | 87.68 0.89 | ± | 63.84 1.50 | ± | 116.375 |
| CrossE_L + PMI_L + Triplet_L | ALL | 95.83 0.38 | ± | 69.82 0.91 | ± | 92.07 1.30 | ± | 71.11 0.58 | ± | 89.75 |
| CrossE_L + PMI_L + Triplet_L | GAT | 98.96 0.10 | ± | 84.17 1.58 | ± | 97.55 0.23 | ± | 84.56 0.59 | ± | 10.875 |
| CrossE_L + PMI_L + Triplet_L | GCN | 98.15 0.18 | ± | 84.04 0.59 | ± | 96.14 0.20 | ± | 83.17 1.19 | ± | 38.75 |
| CrossE_L + PMI_L + Triplet_L | GIN | 84.46 1.43 | ± | 53.14 0.45 | ± | 85.57 1.12 | ± | 53.33 0.74 | ± | 143.0 |
| CrossE_L + PMI_L + Triplet_L | MPNN | 98.17 0.31 | ± | 82.30 0.90 | ± | 94.30 0.58 | ± | 83.48 1.30 | ± | 54.875 |
| CrossE_L + PMI_L + Triplet_L | PAGNN | 72.87 0.92 | ± | 51.49 0.03 | ± | 73.64 0.33 | ± | 51.47 0.01 | ± | 184.5 |
| CrossE_L + PMI_L + Triplet_L | SAGE | 92.48 1.00 | ± | 64.42 1.70 | ± | 90.27 0.68 | ± | 68.15 0.68 | ± | 105.375 |

Table 29. Results for Lp F1 (↑) (continued)

| Loss Type | Model | Cora ↓ Citeseer | | Cora ↓ Bitcoin | | Citeseer ↓ Cora | | Citeseer ↓ Bitcoin | | Average Rank |
|---|---|---|---|---|---|---|---|---|---|---|
| CrossE_L + PR_L | ALL | 68.67 | ± 2.01 | 67.30 | ± 0.56 | 74.36 | ± 3.50 | 68.31 | ± 0.62 | 145.75 |
| CrossE_L + PR_L | GAT | 98.01 | ± 0.39 | 81.27 | ± 0.53 | 96.27 | ± 0.32 | 81.04 | ± 0.97 | 61.25 |
| CrossE_L + PR_L | GCN | 94.55 | ± 1.89 | 80.45 | ± 1.24 | 94.73 | ± 1.04 | 80.24 | ± 1.78 | 80.5 |
| CrossE_L + PR_L | GIN | 85.08 | ± 2.23 | 51.78 | ± 0.25 | 72.02 | ± 4.92 | 51.50 | ± 0.09 | 169.875 |
| CrossE_L + PR_L | MPNN | 93.02 | ± 5.58 | 83.31 | ± 0.62 | 79.51 | ± 4.22 | 84.15 | ± 0.60 | 81.625 |
| CrossE_L + PR_L | PAGNN | 68.81 | ± 3.17 | 51.46 | ± 0.00 | 67.94 | ± 0.04 | 51.46 | ± 0.00 | 203.75 |
| CrossE_L + PR_L | SAGE | 73.57 | ± 0.48 | 54.46 | ± 0.47 | 74.55 | ± 1.29 | 54.34 | ± 0.74 | 159.0 |
| CrossE_L + PR_L + Triplet_L | ALL | 76.15 | ± 8.21 | 51.46 | ± 0.00 | 78.32 | ± 3.06 | 51.46 | ± 0.01 | 184.25 |
| CrossE_L + PR_L + Triplet_L | GAT | 98.33 | ± 0.45 | 82.06 | ± 1.20 | 97.01 | ± 0.71 | 84.58 | ± 0.58 | 33.625 |
| CrossE_L + PR_L + Triplet_L | GCN | 97.62 | ± 0.24 | 83.32 | ± 0.63 | 95.28 | ± 0.76 | 85.16 | ± 1.05 | 42.75 |
| CrossE_L + PR_L + Triplet_L | GIN | 91.34 | ± 3.12 | 52.11 | ± 0.75 | 85.74 | ± 2.10 | 54.40 | ± 0.82 | 135.125 |
| CrossE_L + PR_L + Triplet_L | MPNN | 94.97 | ± 2.23 | 83.55 | ± 0.80 | 83.37 | ± 3.93 | 83.82 | ± 0.94 | 75.75 |
| CrossE_L + PR_L + Triplet_L | PAGNN | 75.19 | ± 1.92 | 52.11 | ± 0.21 | 71.11 | ± 2.50 | 51.71 | ± 0.12 | 174.875 |
| CrossE_L + PR_L + Triplet_L | SAGE | 96.69 | ± 0.81 | 65.69 | ± 2.86 | 94.74 | ± 1.16 | 64.54 | ± 2.57 | 90.875 |
| CrossE_L + Triplet_L | ALL | 96.52 | ± 0.10 | 62.94 | ± 0.60 | 92.24 | ± 1.31 | 61.70 | ± 0.37 | 101.5 |
| CrossE_L + Triplet_L | GAT | 99.12 | ± 0.11 | 84.97 | ± 0.52 | 97.74 | ± 0.28 | 85.03 | ± 0.86 | 2.0 |

Table 29.  Results for Lp F1 (↑) (continued)

| Loss Type | Model | Cora ↓ Citeseer | | Cora ↓ Bitcoin | | Citeseer ↓ Cora | | Citeseer ↓ Bitcoin | | Average Rank |
|---|---|---|---|---|---|---|---|---|---|---|
| CrossE_L + Triplet_L | GCN | 98.45 | ± 0.20 | 83.66 | ± 0.41 | 96.69 | ± 0.29 | 83.40 | ± 1.37 | 30.0 |
| CrossE_L + Triplet_L | GIN | 95.38 | ± 0.45 | 55.05 | ± 1.00 | 91.67 | ± 0.91 | 57.79 | ± 0.91 | 112.0 |
| CrossE_L + Triplet_L | MPNN | 98.28 | ± 0.20 | 82.78 | ± 0.71 | 94.74 | ± 0.56 | 81.30 | ± 0.73 | 59.0 |
| CrossE_L + Triplet_L | PAGNN | 84.71 | ± 4.66 | 51.46 | ± 0.01 | 87.00 | ± 1.43 | 51.48 | ± 0.01 | 159.625 |
| CrossE_L + Triplet_L | SAGE | 97.82 | ± 0.28 | 70.01 | ± 1.93 | 95.87 | ± 0.39 | 69.48 | ± 1.16 | 76.5 |
| PMI_L | ALL | 73.90 | ± 1.29 | 70.96 | ± 1.10 | 86.21 | ± 1.12 | 70.43 | ± 0.68 | 121.5 |
| PMI_L | GAT | 99.10 | ± 0.14 | 84.51 | ± 1.35 | 97.61 | ± 0.29 | 84.92 | ± 0.55 | 5.125 |
| PMI_L | GCN | 97.96 | ± 0.19 | 83.12 | ± 0.43 | 95.83 | ± 0.24 | 84.24 | ± 1.05 | 45.0 |
| PMI_L | GIN | 80.02 | ± 2.03 | 51.47 | ± 0.01 | 79.83 | ± 0.92 | 51.48 | ± 0.02 | 171.0 |
| PMI_L | MPNN | 98.44 | ± 0.21 | 82.26 | ± 0.97 | 94.32 | ± 0.46 | 82.14 | ± 1.11 | 57.25 |
| PMI_L | PAGNN | 71.72 | ± 1.16 | 51.46 | ± 0.00 | 73.14 | ± 0.25 | 51.47 | ± 0.01 | 192.125 |
| PMI_L | SAGE | 75.07 | ± 1.51 | 53.99 | ± 1.36 | 74.72 | ± 0.97 | 52.65 | ± 0.82 | 161.5 |
| PMI_L + PR_L | ALL | 68.55 | ± 0.93 | 68.18 | ± 0.93 | 82.85 | ± 0.76 | 69.69 | ± 1.24 | 137.5 |
| PMI_L + PR_L | GAT | 98.56 | ± 0.42 | 83.12 | ± 1.54 | 91.40 | ± 6.92 | 81.92 | ± 2.13 | 58.125 |
| PMI_L + PR_L | GCN | 98.03 | ± 0.32 | 84.44 | ± 1.13 | 95.83 | ± 0.75 | 82.70 | ± 0.98 | 43.125 |
| PMI_L + PR_L | GIN | 79.22 | ± 2.69 | 51.46 | ± 0.00 | 79.13 | ± 3.91 | 52.07 | ± 0.63 | 169.125 |

Table 29. Results for Lp F1 (↑) (continued)

| Loss Type | Model | Cora ↓ Citeseer | | Cora ↓ Bitcoin | | Citeseer ↓ Cora | | Citeseer ↓ Bitcoin | | Average Rank |
|---|---|---|---|---|---|---|---|---|---|---|
| PMI_L + PR_L | MPNN | 98.22 | ± 0.35 | 82.19 | ± 0.94 | 85.20 | ± 1.95 | 79.12 | ± 2.15 | 83.5 |
| PMI_L + PR_L | PAGNN | 71.92 | ± 1.03 | 51.46 | ± 0.01 | 72.75 | ± 0.26 | 51.48 | ± 0.01 | 191.625 |
| PMI_L + PR_L | SAGE | 75.77 | ± 1.03 | 53.14 | ± 1.29 | 76.08 | ± 0.71 | 52.39 | ± 0.37 | 162.125 |
| PMI_L + PR_L + Triplet_L | ALL | 76.68 | ± 0.73 | 54.82 | ± 2.99 | 86.73 | ± 1.40 | 52.85 | ± 1.18 | 143.5 |
| PMI_L + PR_L + Triplet_L | GAT | 98.86 | ± 0.15 | 84.79 | ± 0.81 | 96.88 | ± 0.77 | 83.35 | ± 0.96 | 19.5 |
| PMI_L + PR_L + Triplet_L | GCN | 98.24 | ± 0.12 | 84.29 | ± 1.48 | 96.01 | ± 0.30 | 82.81 | ± 1.01 | 38.5 |
| PMI_L + PR_L + Triplet_L | GIN | 83.41 | ± 2.89 | 52.34 | ± 1.07 | 82.74 | ± 1.77 | 51.51 | ± 0.02 | 157.875 |
| PMI_L + PR_L + Triplet_L | MPNN | 98.39 | ± 0.07 | 83.11 | ± 0.74 | 87.98 | ± 1.14 | 80.82 | ± 1.44 | 68.0 |
| PMI_L + PR_L + Triplet_L | PAGNN | 72.37 | ± 1.55 | 51.48 | ± 0.02 | 73.03 | ± 0.37 | 51.47 | ± 0.01 | 187.625 |
| PMI_L + PR_L + Triplet_L | SAGE | 90.95 | ± 3.44 | 66.43 | ± 1.97 | 89.84 | ± 1.34 | 64.29 | ± 0.86 | 108.25 |
| PMI_L + Triplet_L | ALL | 95.45 | ± 0.82 | 60.32 | ± 1.06 | 92.22 | ± 0.79 | 71.97 | ± 0.37 | 97.25 |
| PMI_L + Triplet_L | GAT | 99.01 | ± 0.09 | 84.31 | ± 1.12 | 97.58 | ± 0.16 | 84.67 | ± 0.63 | 7.375 |
| PMI_L + Triplet_L | GCN | 98.18 | ± 0.06 | 83.89 | ± 0.95 | 96.20 | ± 0.42 | 83.18 | ± 0.58 | 37.0 |
| PMI_L + Triplet_L | GIN | 87.79 | ± 1.13 | 52.29 | ± 0.57 | 85.02 | ± 1.41 | 54.01 | ± 0.76 | 141.25 |
| PMI_L + Triplet_L | MPNN | 98.25 | ± 0.27 | 82.39 | ± 0.76 | 93.98 | ± 0.62 | 82.96 | ± 0.97 | 58.875 |
| PMI_L + Triplet_L | PAGNN | 71.71 | ± 1.27 | 51.46 | ± 0.01 | 73.58 | ± 0.52 | 51.47 | ± 0.01 | 191.125 |

Table 29. Results for Lp F1 (↑) (continued)

| Loss Type | Model | Cora ↓ Citeseer | | Cora ↓ Bitcoin | | Citeseer ↓ Cora | | Citeseer ↓ Bitcoin | | Average Rank |
|---|---|---|---|---|---|---|---|---|---|---|
| PMI_L + Triplet_L | SAGE | 90.03 | ± | 59.60 | ± | 90.84 | ± | 67.02 | ± | 111.875 |
| | | 1.95 | | 2.73 | | 0.54 | | 1.25 | | |
| PR_L | ALL | 65.98 | ± | 67.97 | ± | 70.40 | ± | 69.64 | ± | 151.25 |
| | | 0.81 | | 0.74 | | 2.06 | | 0.47 | | |
| PR_L | GAT | 97.76 | ± | 79.79 | ± | 96.39 | ± | 80.92 | ± | 64.75 |
| | | 0.52 | | 1.03 | | 0.49 | | 1.20 | | |
| PR_L | GCN | 96.36 | ± | 80.13 | ± | 94.57 | ± | 79.06 | ± | 79.375 |
| | | 0.80 | | 0.52 | | 0.72 | | 2.68 | | |
| PR_L | GIN | 84.50 | ± | 51.54 | ± | 72.93 | ± | 51.72 | ± | 166.75 |
| | | 2.54 | | 0.18 | | 3.41 | | 0.28 | | |
| PR_L | MPNN | 91.53 | ± | 83.30 | ± | 81.38 | ± | 83.72 | ± | 85.625 |
| | | 7.23 | | 0.60 | | 3.28 | | 0.58 | | |
| PR_L | PAGNN | 67.80 | ± | 51.46 | ± | 68.03 | ± | 51.48 | ± | 200.125 |
| | | 3.61 | | 0.00 | | 0.05 | | 0.00 | | |
| PR_L | SAGE | 73.44 | ± | 53.65 | ± | 75.74 | ± | 54.25 | ± | 159.375 |
| | | 1.29 | | 0.54 | | 1.33 | | 0.96 | | |
| PR_L + Triplet_L | ALL | 68.64 | ± | 58.39 | ± | 72.06 | ± | 67.94 | ± | 159.5 |
| | | 1.91 | | 4.30 | | 1.31 | | 0.21 | | |
| PR_L + Triplet_L | GAT | 97.91 | ± | 75.73 | ± | 96.17 | ± | 81.40 | ± | 64.25 |
| | | 1.37 | | 1.74 | | 1.05 | | 0.56 | | |
| PR_L + Triplet_L | GCN | 96.88 | ± | 81.92 | ± | 94.93 | ± | 82.59 | ± | 67.75 |
| | | 0.62 | | 1.31 | | 0.82 | | 0.31 | | |
| PR_L + Triplet_L | GIN | 84.48 | ± | 51.46 | ± | 74.34 | ± | 51.47 | ± | 177.125 |
| | | 4.60 | | 0.00 | | 4.75 | | 0.01 | | |
| PR_L + Triplet_L | MPNN | 92.15 | ± | 83.91 | ± | 84.36 | ± | 83.63 | ± | 75.25 |
| | | 6.75 | | 0.82 | | 4.63 | | 0.70 | | |
| PR_L + Triplet_L | PAGNN | 68.92 | ± | 51.50 | ± | 68.06 | ± | 51.51 | ± | 191.875 |
| | | 3.08 | | 0.01 | | 0.07 | | 0.03 | | |
| PR_L + Triplet_L | SAGE | 77.17 | ± | 54.73 | ± | 83.31 | ± | 53.78 | ± | 149.0 |
| | | 1.86 | | 0.59 | | 8.44 | | 0.37 | | |
| Triplet_L | ALL | 95.89 | ± | 58.45 | ± | 92.32 | ± | 72.88 | ± | 96.5 |
| | | 0.66 | | 0.99 | | 1.03 | | 0.83 | | |

Continued on next page

Table 29. Results for Lp F1 (↑) (continued)

| Loss Type | Model | Cora ↓ Citeseer | | Cora ↓ Bitcoin | | Citeseer ↓ Cora | | Citeseer ↓ Bitcoin | | Average Rank |
|-----------|-------|-----------------|---|----------------|---|-----------------|---|---------------------|---|--------------|
| Triplet_L | GAT | 99.20 ± 0.12 | | 84.31 ± 0.77 | | 97.89 ± 0.25 | | 84.64 ± 1.18 | | 4.875 |
| Triplet_L | GCN | 98.58 ± 0.17 | | 84.18 ± 0.99 | | 96.56 ± 0.47 | | 84.25 ± 0.85 | | 19.25 |
| Triplet_L | GIN | 96.44 ± 0.55 | | 56.55 ± 0.26 | | 91.83 ± 1.50 | | 60.47 ± 0.88 | ± | 107.25 |
| Triplet_L | MPNN | 98.43 ± 0.24 | | 83.23 ± 0.84 | | 94.86 ± 0.46 | | 83.22 ± 1.98 | ± | 45.125 |
| Triplet_L | PAGNN | 85.24 ± 4.63 | | 51.56 ± 0.06 | | 88.62 ± 0.35 | | 51.88 ± 0.09 | ± | 142.625 |
| Triplet_L | SAGE | 97.57 ± 0.40 | | 71.56 ± 1.31 | | 95.52 ± 0.47 | | 73.03 ± 0.99 | ± | 76.125 |

Table 30. Lp Precision Performance (↑): This table presents models (Loss function and GNN) ranked by their average performance in terms of lp precision. Top-ranked results are highlighted in red, second-ranked in blue, and third-ranked in green.

| Loss Type | Model | Cora ↓ Citeseer | | Cora ↓ Bitcoin | | Citeseer ↓ Cora | | Citeseer ↓ Bitcoin | | Average Rank |
|-----------|-------|-----------------|---|----------------|---|-----------------|---|---------------------|---|--------------|
| Contr_l | ALL | 90.08 ± 2.21 | ± | 48.97 ± 3.77 | ± | 86.12 ± 1.36 | ± | 40.68 ± 8.66 | ± | 135.75 |
| Contr_l | GAT | 98.69 ± 0.21 | ± | 84.32 ± 1.19 | | 97.61 ± 0.33 | | 87.43 ± 1.16 | ± | 27.375 |
| Contr_l | GCN | 98.09 ± 0.20 | | 84.55 ± 1.79 | ± | 96.70 ± 0.53 | | 85.10 ± 1.37 | ± | 47.125 |
| Contr_l | GIN | 92.92 ± 1.13 | | 58.21 ± 1.43 | ± | 89.33 ± 1.94 | | 64.81 ± 3.10 | | 109.0 |
| Contr_l | MPNN | 97.52 ± 0.24 | | 82.29 ± 0.64 | ± | 92.35 ± 0.79 | | 82.77 ± 1.70 | | 84.75 |

Table 30.  Results for Lp Precision (↑) (continued)

| Loss Type | Model | Cora ↓ Citeseer | Cora ↓ Bitcoin | Citeseer ↓ Cora | Citeseer ↓ Bitcoin | Average Rank |
|---|---|---|---|---|---|---|
| Contr_l | PAGNN | 79.41 ± 2.87 | 34.77 ± 0.06 | 82.09 ± 2.59 | 34.91 ± 0.09 | 160.25 |
| Contr_l | SAGE | 96.01 ± 0.46 | 54.34 ± 4.69 | 93.79 ± 1.60 | 62.12 ± 6.72 | 102.25 |
| Contr_l + CrossE_L | ALL | 90.10 ± 1.04 | 54.47 ± 2.22 | 84.40 ± 2.77 | 42.90 ± 6.21 | 133.75 |
| Contr_l + CrossE_L | GAT | 98.61 ± 0.43 | 85.27 ± 0.69 | 97.47 ± 0.21 | 84.78 ± 0.74 | 33.0 |
| Contr_l + CrossE_L | GCN | 98.62 ± 0.21 | 85.22 ± 1.90 | 96.35 ± 0.27 | 86.77 ± 0.84 | 28.125 |
| Contr_l + CrossE_L | GIN | 87.55 ± 3.46 | 49.63 ± 8.59 | 90.19 ± 0.34 | 59.45 ± 2.03 | 118.75 |
| Contr_l + CrossE_L | MPNN | 97.37 ± 0.50 | 83.48 ± 1.50 | 92.26 ± 0.85 | 83.81 ± 1.47 | 82.75 |
| Contr_l + CrossE_L | PAGNN | 78.33 ± 2.44 | 34.84 ± 0.05 | 81.51 ± 1.20 | 34.85 ± 0.05 | 161.75 |
| Contr_l + CrossE_L | SAGE | 95.95 ± 0.71 | 49.36 ± 1.27 | 94.58 ± 1.28 | 50.00 ± 2.64 | 109.875 |
| Contr_l + CrossE_L + PMI_L | ALL | 75.09 ± 5.81 | 80.20 ± 2.73 | 89.48 ± 1.79 | 86.73 ± 3.16 | 93.25 |
| Contr_l + CrossE_L + PMI_L | GAT | 98.82 ± 0.09 | 85.17 ± 1.94 | 97.53 ± 0.33 | 85.78 ± 1.47 | 25.375 |
| Contr_l + CrossE_L + PMI_L | GCN | 98.08 ± 0.27 | 85.76 ± 0.88 | 96.14 ± 0.41 | 85.45 ± 1.64 | 40.125 |
| Contr_l + CrossE_L + PMI_L | GIN | 80.18 ± 3.11 | 34.65 ± 0.01 | 81.68 ± 3.02 | 42.47 ± 10.67 | 156.375 |
| Contr_l + CrossE_L + PMI_L | MPNN | 97.78 ± 0.38 | 86.17 ± 1.62 | 93.11 ± 0.38 | 85.71 ± 0.87 | 48.375 |
| Contr_l + CrossE_L + PMI_L | PAGNN | 58.08 ± 1.58 | 34.65 ± 0.02 | 60.94 ± 1.88 | 34.65 ± 0.00 | 194.125 |
| Contr_l + CrossE_L + PMI_L | SAGE | 76.75 ± 1.30 | 39.95 ± 1.60 | 75.86 ± 2.74 | 43.17 ± 0.24 | 153.75 |

Table 30.  Results for Lp Precision (↑) (continued)

| Loss Type | Model | Cora ↓ Citeseer | | Cora ↓ Bitcoin | | Citeseer ↓ Cora | | Citeseer ↓ Bitcoin | | Average Rank |
|-----------|-------|------|---|------|---|------|---|------|---|------|
| Contr_l + CrossE_L + PMI_L + PR_L | ALL | 54.08 2.25 | ± | 90.77 1.20 | ± | 90.04 1.15 | ± | 88.44 0.92 | ± | 75.5 |
| Contr_l + CrossE_L + PMI_L + PR_L | GAT | 98.90 0.12 | ± | 85.91 2.48 | ± | 97.27 0.32 | ± | 84.36 2.33 | ± | 30.125 |
| Contr_l + CrossE_L + PMI_L + PR_L | GCN | 98.19 0.34 | ± | 86.53 0.98 | ± | 96.27 0.42 | ± | 84.66 0.76 | ± | 36.375 |
| Contr_l + CrossE_L + PMI_L + PR_L | GIN | 72.15 4.01 | ± | 34.64 0.00 | ± | 70.61 4.53 | ± | 34.74 0.06 | ± | 180.125 |
| Contr_l + CrossE_L + PMI_L + PR_L | MPNN | 97.83 0.25 | ± | 84.91 1.02 | ± | 88.70 2.71 | ± | 81.32 2.89 | ± | 83.5 |
| Contr_l + CrossE_L + PMI_L + PR_L | PAGNN | 60.48 2.25 | ± | 34.70 0.04 | ± | 60.15 3.27 | ± | 37.48 0.11 | ± | 180.5 |
| Contr_l + CrossE_L + PMI_L + PR_L | SAGE | 75.86 1.40 | ± | 37.47 0.74 | ± | 72.58 2.00 | ± | 37.57 1.67 | ± | 163.875 |
| Contr_l + CrossE_L + PMI_L + PR_L + Triplet_L | ALL | 69.04 4.72 | ± | 74.00 2.13 | ± | 86.89 2.86 | ± | 79.64 2.64 | ± | 125.25 |
| Contr_l + CrossE_L + PMI_L + PR_L + Triplet_L | GAT | 98.79 0.20 | ± | 85.83 2.20 | ± | 96.98 0.59 | ± | 84.74 1.63 | ± | 31.0 |
| Contr_l + CrossE_L + PMI_L + PR_L + Triplet_L | GCN | 98.12 0.35 | ± | 84.91 1.52 | ± | 96.24 0.48 | ± | 85.98 0.97 | ± | 41.375 |
| Contr_l + CrossE_L + PMI_L + PR_L + Triplet_L | GIN | 75.46 3.85 | ± | 41.78 9.77 | ± | 81.27 2.21 | ± | 34.67 0.02 | ± | 162.0 |
| Contr_l + CrossE_L + PMI_L + PR_L + Triplet_L | MPNN | 97.79 0.19 | ± | 84.91 0.89 | ± | 90.07 1.95 | ± | 80.81 2.60 | ± | 79.75 |
| Contr_l + CrossE_L + PMI_L + PR_L + Triplet_L | PAGNN | 59.82 1.85 | ± | 34.72 0.05 | ± | 59.52 1.29 | ± | 34.72 0.06 | ± | 185.875 |

Table 30. Results for Lp Precision (↑) (continued)

| Loss Type | Model | Cora ↓ Citeseer | | Cora ↓ Bitcoin | | Citeseer ↓ Cora | | Citeseer ↓ Bitcoin | | Average Rank |
|---|---|---|---|---|---|---|---|---|---|---|
| Contr_l + CrossE_L + PMI_L + PR_L + Triplet_L | SAGE | 86.52 4.88 | ± | 54.85 1.86 | ± | 88.82 0.62 | ± | 53.47 2.76 | ± | 124.25 |
| Contr_l + CrossE_L + PMI_L + Triplet_L | ALL | 93.79 1.89 | ± | 82.83 1.71 | ± | 91.37 1.54 | ± | 81.88 0.85 | ± | 93.5 |
| Contr_l + CrossE_L + PMI_L + Triplet_L | GAT | 98.85 0.10 | ± | 85.27 1.19 | ± | 97.56 0.30 | ± | 85.36 1.81 | ± | 25.5 |
| Contr_l + CrossE_L + PMI_L + Triplet_L | GCN | 98.22 0.37 | ± | 86.72 0.85 | ± | 96.18 0.29 | ± | 85.62 1.03 | ± | 28.75 |
| Contr_l + CrossE_L + PMI_L + Triplet_L | GIN | 81.93 1.74 | ± | 50.47 8.97 | ± | 83.63 3.12 | ± | 53.67 1.91 | ± | 137.0 |
| Contr_l + CrossE_L + PMI_L + Triplet_L | MPNN | 97.96 0.17 | ± | 86.42 1.33 | ± | 92.63 0.78 | ± | 84.11 1.24 | ± | 55.25 |
| Contr_l + CrossE_L + PMI_L + Triplet_L | PAGNN | 59.77 0.51 | ± | 34.64 0.00 | ± | 61.03 1.79 | ± | 34.65 0.01 | ± | 195.125 |
| Contr_l + CrossE_L + PMI_L + Triplet_L | SAGE | 88.45 2.37 | ± | 48.40 3.28 | ± | 86.14 0.97 | ± | 53.69 0.60 | ± | 131.125 |
| Contr_l + CrossE_L + PR_L | ALL | 54.16 1.41 | ± | 78.28 4.97 | ± | 62.05 5.62 | ± | 86.23 0.51 | ± | 127.25 |
| Contr_l + CrossE_L + PR_L | GAT | 98.08 0.93 | ± | 85.13 1.40 | ± | 96.83 0.34 | ± | 84.86 2.53 | ± | 44.125 |
| Contr_l + CrossE_L + PR_L | GCN | 97.68 0.40 | ± | 82.48 1.05 | ± | 94.87 0.51 | ± | 83.82 2.18 | ± | 76.5 |
| Contr_l + CrossE_L + PR_L | GIN | 89.88 2.34 | ± | 34.64 0.00 | ± | 74.64 4.31 | ± | 34.65 0.01 | ± | 171.625 |
| Contr_l + CrossE_L + PR_L | MPNN | 95.59 2.21 | ± | 86.13 1.23 | ± | 89.27 1.89 | ± | 85.39 1.46 | ± | 66.5 |
| Contr_l + CrossE_L + PR_L | PAGNN | 62.99 3.81 | ± | 37.50 0.24 | ± | 52.53 2.04 | ± | 37.74 0.23 | ± | 179.25 |
| Contr_l + CrossE_L + PR_L | SAGE | 90.48 7.90 | ± | 51.17 2.73 | ± | 84.42 10.36 | ± | 46.72 2.93 | ± | 134.25 |
| Contr_l + CrossE_L + PR_L + Triplet_L | ALL | 73.48 3.31 | ± | 34.65 0.02 | ± | 86.07 1.65 | ± | 47.50 7.45 | ± | 157.625 |

Table 30. Results for Lp Precision (↑) (continued)

| Loss Type | Model | Cora ↓ Citeseer | ± | Cora ↓ Bitcoin | ± | Citeseer ↓ Cora | ± | Citeseer ↓ Bitcoin | ± | Average Rank |
|---|---|---|---|---|---|---|---|---|---|---|
| Contr_l + CrossE_L + PR_L + Triplet_L | GAT | 98.69 0.25 | ± | 83.77 2.83 | ± | 97.27 0.61 | ± | 85.13 1.08 | ± | 41.5 |
| Contr_l + CrossE_L + PR_L + Triplet_L | GCN | 98.19 0.18 | ± | 85.73 1.79 | ± | 95.47 0.77 | ± | 85.39 1.05 | ± | 41.0 |
| Contr_l + CrossE_L + PR_L + Triplet_L | GIN | 88.45 2.59 | ± | 46.88 6.86 | ± | 89.61 0.86 | ± | 56.44 2.29 | ± | 122.125 |
| Contr_l + CrossE_L + PR_L + Triplet_L | MPNN | 97.41 0.50 | ± | 84.34 1.05 | ± | 89.02 1.35 | ± | 80.52 1.56 | ± | 89.875 |
| Contr_l + CrossE_L + PR_L + Triplet_L | PAGNN | 67.48 3.32 | ± | 34.72 0.02 | ± | 63.65 4.38 | ± | 37.24 1.17 | ± | 175.625 |
| Contr_l + CrossE_L + PR_L + Triplet_L | SAGE | 96.35 0.61 | ± | 56.20 2.45 | ± | 92.94 0.92 | ± | 47.88 3.20 | ± | 107.0 |
| Contr_l + CrossE_L + Triplet_L | ALL | 92.76 1.79 | ± | 57.24 1.51 | ± | 88.92 1.51 | ± | 52.96 1.82 | ± | 117.5 |
| Contr_l + CrossE_L + Triplet_L | GAT | 98.88 0.22 | ± | 84.33 1.62 | ± | 97.59 0.41 | ± | 85.59 1.47 | ± | 31.25 |
| Contr_l + CrossE_L + Triplet_L | GCN | 98.67 0.25 | ± | 85.38 1.96 | ± | 96.34 0.47 | ± | 84.14 1.80 | ± | 41.125 |
| Contr_l + CrossE_L + Triplet_L | GIN | 94.62 1.24 | ± | 61.03 1.42 | ± | 91.40 1.10 | ± | 66.93 1.38 | ± | 100.5 |
| Contr_l + CrossE_L + Triplet_L | MPNN | 97.54 0.44 | ± | 83.31 1.54 | ± | 93.11 1.02 | ± | 84.58 1.24 | ± | 76.25 |
| Contr_l + CrossE_L + Triplet_L | PAGNN | 75.86 8.37 | ± | 34.71 0.02 | ± | 85.83 1.63 | ± | 34.86 0.08 | ± | 161.5 |
| Contr_l + CrossE_L + Triplet_L | SAGE | 96.83 0.69 | ± | 56.19 4.07 | ± | 94.90 0.73 | ± | 63.62 3.85 | ± | 94.5 |
| Contr_l + PMI_L | ALL | 79.49 10.79 | ± | 42.49 10.89 | ± | 88.07 1.81 | ± | 78.44 2.29 | ± | 128.0 |
| Contr_l + PMI_L | GAT | 98.96 0.17 | ± | 86.28 1.15 | ± | 97.06 0.24 | ± | 86.05 1.99 | ± | 14.75 |
| Contr_l + PMI_L | GCN | 98.36 0.43 | ± | 84.54 1.58 | ± | 96.23 0.22 | ± | 85.68 0.91 | ± | 42.5 |

Continued on next page

Table 30. Results for Lp Precision (↑) (continued)

| Loss Type | Model | Cora ↓ Citeseer | Cora ↓ Bitcoin | Citeseer ↓ Cora | Citeseer ↓ Bitcoin | Average Rank |
|---|---|---|---|---|---|---|
| Contr_l + PMI_L | GIN | 80.66 ± 4.91 | 57.16 ± 1.84 | 79.71 ± 1.49 | 50.12 ± 8.87 | 137.75 |
| Contr_l + PMI_L | MPNN | 97.91 ± 0.35 | 85.17 ± 2.66 | 93.10 ± 0.63 | 83.88 ± 1.70 | 66.875 |
| Contr_l + PMI_L | PAGNN | 58.88 ± 1.69 | 34.64 ± 0.00 | 59.43 ± 0.57 | 34.65 ± 0.00 | 198.125 |
| Contr_l + PMI_L | SAGE | 77.44 ± 2.15 | 41.21 ± 2.05 | 80.20 ± 3.52 | 45.21 ± 0.74 | 150.25 |
| Contr_l + PMI_L + PR_L | ALL | 53.10 ± 0.96 | 82.28 ± 1.18 | 88.55 ± 1.26 | 87.89 ± 3.21 | 106.125 |
| Contr_l + PMI_L + PR_L | GAT | 98.75 ± 0.33 | 85.35 ± 1.55 | 94.59 ± 3.72 | 75.21 ± 2.81 | 56.75 |
| Contr_l + PMI_L + PR_L | GCN | 98.23 ± 0.53 | 86.26 ± 2.37 | 96.17 ± 0.65 | 84.92 ± 1.27 | 35.75 |
| Contr_l + PMI_L + PR_L | GIN | 77.54 ± 5.61 | 34.64 ± 0.00 | 75.62 ± 5.67 | 34.72 ± 0.05 | 173.875 |
| Contr_l + PMI_L + PR_L | MPNN | 97.94 ± 0.24 | 86.23 ± 1.27 | 86.39 ± 4.59 | 79.89 ± 0.83 | 76.75 |
| Contr_l + PMI_L + PR_L | PAGNN | 58.44 ± 1.50 | 34.64 ± 0.00 | 59.04 ± 0.47 | 34.65 ± 0.00 | 198.625 |
| Contr_l + PMI_L + PR_L | SAGE | 76.37 ± 1.19 | 40.61 ± 2.10 | 78.70 ± 3.05 | 42.00 ± 1.35 | 152.75 |
| Contr_l + PMI_L + PR_L + Triplet_L | ALL | 75.65 ± 4.18 | 34.64 ± 0.00 | 86.84 ± 1.54 | 61.31 ± 4.71 | 150.375 |
| Contr_l + PMI_L + PR_L + Triplet_L | GAT | 98.56 ± 0.41 | 84.23 ± 0.79 | 96.03 ± 1.10 | 78.93 ± 3.75 | 63.75 |
| Contr_l + PMI_L + PR_L + Triplet_L | GCN | 98.30 ± 0.46 | 85.60 ± 2.05 | 96.20 ± 0.51 | 84.36 ± 1.17 | 43.375 |
| Contr_l + PMI_L + PR_L + Triplet_L | GIN | 81.25 ± 1.49 | 57.96 ± 1.52 | 86.62 ± 2.78 | 41.82 ± 9.82 | 132.5 |
| Contr_l + PMI_L + PR_L + Triplet_L | MPNN | 98.05 ± 0.36 | 84.67 ± 1.97 | 88.27 ± 2.25 | 79.02 ± 3.02 | 85.75 |

Table 30. Results for Lp Precision (↑) (continued)

| Loss Type | Model | Cora ↓ Citeseer | | Cora ↓ Bitcoin | | Citeseer ↓ Cora | | Citeseer ↓ Bitcoin | | Average Rank |
|---|---|---|---|---|---|---|---|---|---|---|
| Contr_l + PMI_L + PR_L + Triplet_L | PAGNN | 60.83 1.42 | ± | 34.64 0.01 | ± | 60.07 2.34 | ± | 34.67 0.05 | ± | 191.375 |
| Contr_l + PMI_L + PR_L + Triplet_L | SAGE | 93.06 1.71 | ± | 49.82 0.69 | ± | 90.25 2.74 | ± | 53.72 4.58 | ± | 115.75 |
| Contr_l + PR_L | ALL | 55.75 3.39 | ± | 87.31 0.88 | ± | 68.50 8.23 | ± | 88.60 2.46 | ± | 96.75 |
| Contr_l + PR_L | GAT | 98.12 0.54 | ± | 83.95 1.19 | ± | 96.90 0.24 | ± | 84.06 2.20 | ± | 56.125 |
| Contr_l + PR_L | GCN | 97.59 0.40 | ± | 84.09 1.41 | ± | 94.98 0.77 | ± | 85.05 1.40 | ± | 66.25 |
| Contr_l + PR_L | GIN | 87.99 2.74 | ± | 37.58 1.69 | ± | 70.24 12.89 | ± | 40.42 7.89 | ± | 154.25 |
| Contr_l + PR_L | MPNN | 95.81 1.55 | ± | 84.55 0.90 | ± | 89.67 1.31 | ± | 84.99 1.02 | ± | 79.875 |
| Contr_l + PR_L | PAGNN | 60.01 7.12 | ± | 36.56 1.44 | ± | 51.91 0.40 | ± | 38.00 0.20 | ± | 181.25 |
| Contr_l + PR_L | SAGE | 91.20 7.11 | ± | 52.68 3.27 | ± | 87.56 7.64 | ± | 38.85 0.59 | ± | 130.5 |
| Contr_l + PR_L + Triplet_L | ALL | 68.84 12.40 | ± | 41.60 9.51 | ± | 85.80 2.66 | ± | 34.65 0.01 | ± | 165.375 |
| Contr_l + PR_L + Triplet_L | GAT | 98.42 0.26 | ± | 84.72 1.21 | ± | 97.04 0.38 | ± | 82.99 2.08 | ± | 50.0 |
| Contr_l + PR_L + Triplet_L | GCN | 98.03 0.33 | ± | 85.45 2.45 | ± | 96.13 0.54 | ± | 84.78 1.00 | ± | 47.875 |
| Contr_l + PR_L + Triplet_L | GIN | 91.37 2.83 | ± | 55.17 3.93 | ± | 84.97 4.41 | ± | 55.32 1.82 | ± | 126.75 |
| Contr_l + PR_L + Triplet_L | MPNN | 96.76 0.48 | ± | 84.19 2.03 | ± | 88.95 1.90 | ± | 81.64 1.75 | ± | 92.75 |
| Contr_l + PR_L + Triplet_L | PAGNN | 65.07 3.50 | ± | 34.66 0.01 | ± | 67.30 8.46 | ± | 35.00 0.11 | ± | 179.25 |
| Contr_l + PR_L + Triplet_L | SAGE | 96.34 0.95 | ± | 52.62 3.48 | ± | 93.85 1.25 | ± | 52.84 2.27 | ± | 106.625 |

Table 30. Results for Lp Precision (↑) (continued)

| Loss Type | Model | Cora ↓ Citeseer | Cora ↓ Bitcoin | Citeseer ↓ Cora | Citeseer ↓ Bitcoin | Average Rank |
|---|---|---|---|---|---|---|
| Contr_l + Triplet_L | ALL | 93.92 ± 1.40 | 51.18 ± 2.62 | 89.80 ± 1.35 | 52.56 ± 1.82 | 118.375 |
| Contr_l + Triplet_L | GAT | 98.72 ± 0.23 | 84.80 ± 1.77 | 97.74 ± 0.21 | 84.71 ± 2.80 | 36.5 |
| Contr_l + Triplet_L | GCN | 98.48 ± 0.21 | 85.05 ± 1.47 | 96.70 ± 0.69 | 86.74 ± 1.49 | 30.625 |
| Contr_l + Triplet_L | GIN | 94.30 ± 0.68 | 58.66 ± 2.01 | 90.81 ± 2.15 | 63.43 ± 2.37 | 104.75 |
| Contr_l + Triplet_L | MPNN | 97.84 ± 0.17 | 82.14 ± 3.19 | 94.01 ± 0.39 | 83.91 ± 2.07 | 76.625 |
| Contr_l + Triplet_L | PAGNN | 77.22 ± 7.79 | 34.96 ± 0.12 | 85.84 ± 1.22 | 35.51 ± 1.49 | 157.5 |
| Contr_l + Triplet_L | SAGE | 97.29 ± 0.34 | 55.94 ± 3.73 | 95.19 ± 0.56 | 58.11 ± 3.08 | 94.5 |
| CrossE_L | ALL | 84.77 ± 19.60 | 76.43 ± 1.86 | 54.35 ± 4.77 | 80.58 ± 0.71 | 133.75 |
| CrossE_L | GAT | 85.04 ± 21.08 | 37.20 ± 0.62 | 78.21 ± 16.40 | 79.76 ± 2.17 | 140.25 |
| CrossE_L | GCN | 53.03 ± 12.03 | 67.77 ± 1.82 | 51.35 ± 0.00 | 34.64 ± 0.00 | 182.0 |
| CrossE_L | GIN | 47.66 ± 0.01 | 34.64 ± 0.00 | 51.35 ± 0.00 | 34.64 ± 0.00 | 207.625 |
| CrossE_L | MPNN | 86.55 ± 2.66 | 34.64 ± 0.00 | 86.23 ± 1.58 | 34.64 ± 0.00 | 167.0 |
| CrossE_L | PAGNN | 64.85 ± 3.14 | 34.64 ± 0.00 | 58.26 ± 9.40 | 34.65 ± 0.01 | 195.125 |
| CrossE_L | SAGE | 69.20 ± 7.23 | 39.98 ± 0.26 | 57.53 ± 5.76 | 34.86 ± 0.12 | 175.125 |
| CrossE_L + PMI_L | ALL | 72.16 ± 2.98 | 88.00 ± 1.69 | 89.52 ± 0.81 | 88.81 ± 1.52 | 68.875 |
| CrossE_L + PMI_L | GAT | 98.84 ± 0.24 | 86.29 ± 0.94 | 97.53 ± 0.20 | 86.34 ± 1.07 | 14.0 |

Table 30. Results for Lp Precision (↑) (continued)

| Loss Type | Model | Cora ↓ Citeseer | | Cora ↓ Bitcoin | | Citeseer ↓ Cora | | Citeseer ↓ Bitcoin | | Average Rank |
|---|---|---|---|---|---|---|---|---|---|---|
| CrossE_L + PMI_L | GCN | 98.05 | ± | 84.87 | ± | 96.19 | ± | 84.02 | ± | 57.75 |
| | | 0.31 | | 2.33 | | 0.55 | | 1.61 | | |
| CrossE_L + PMI_L | GIN | 77.83 | ± | 37.65 | ± | 76.30 | ± | 41.09 | ± | 156.125 |
| | | 2.62 | | 6.72 | | 4.42 | | 8.77 | | |
| CrossE_L + PMI_L | MPNN | 97.63 | ± | 86.11 | ± | 92.66 | ± | 84.58 | ± | 59.5 |
| | | 0.22 | | 1.92 | | 1.09 | | 2.00 | | |
| CrossE_L + PMI_L | PAGNN | 58.40 | ± | 38.02 | ± | 60.04 | ± | 34.65 | ± | 187.5 |
| | | 2.94 | | 0.23 | | 0.63 | | 0.00 | | |
| CrossE_L + PMI_L | SAGE | 73.86 | ± | 38.54 | ± | 70.66 | ± | 36.31 | ± | 164.75 |
| | | 2.42 | | 1.30 | | 1.61 | | 1.60 | | |
| CrossE_L + PMI_L + PR_L | ALL | 51.95 | ± | 88.96 | ± | 88.55 | ± | 87.42 | ± | 84.625 |
| | | 0.24 | | 1.09 | | 0.92 | | 2.26 | | |
| CrossE_L + PMI_L + PR_L | GAT | 98.93 | ± | 86.51 | ± | 93.98 | ± | 78.08 | ± | 47.625 |
| | | 0.14 | | 0.85 | | 3.18 | | 3.34 | | |
| CrossE_L + PMI_L + PR_L | GCN | 98.20 | ± | 84.34 | ± | 95.99 | ± | 85.75 | ± | 46.875 |
| | | 0.37 | | 1.56 | | 0.26 | | 1.60 | | |
| CrossE_L + PMI_L + PR_L | GIN | 75.40 | ± | 58.11 | ± | 77.14 | ± | 38.14 | ± | 148.5 |
| | | 3.31 | | 1.52 | | 2.83 | | 7.80 | | |
| CrossE_L + PMI_L + PR_L | MPNN | 97.89 | ± | 85.74 | ± | 87.63 | ± | 84.60 | ± | 70.875 |
| | | 0.42 | | 1.54 | | 6.23 | | 1.19 | | |
| CrossE_L + PMI_L + PR_L | PAGNN | 59.89 | ± | 34.68 | ± | 58.98 | ± | 34.65 | ± | 192.75 |
| | | 2.11 | | 0.04 | | 2.16 | | 0.00 | | |
| CrossE_L + PMI_L + PR_L | SAGE | 73.60 | ± | 38.25 | ± | 71.80 | ± | 38.73 | ± | 161.625 |
| | | 2.10 | | 0.73 | | 1.57 | | 0.71 | | |
| CrossE_L + PMI_L + PR_L + Triplet_L | ALL | 68.98 | ± | 87.21 | ± | 88.65 | ± | 74.34 | ± | 100.25 |
| | | 3.90 | | 0.80 | | 1.45 | | 1.14 | | |
| CrossE_L + PMI_L + PR_L + Triplet_L | GAT | 98.84 | ± | 86.67 | ± | 96.97 | ± | 84.75 | ± | 24.625 |
| | | 0.16 | | 1.11 | | 0.89 | | 0.41 | | |
| CrossE_L + PMI_L + PR_L + Triplet_L | GCN | 98.28 | ± | 84.78 | ± | 95.85 | ± | 85.06 | ± | 48.75 |
| | | 0.20 | | 2.22 | | 0.66 | | 1.72 | | |
| CrossE_L + PMI_L + PR_L + Triplet_L | GIN | 78.63 | ± | 42.07 | ± | 81.36 | ± | 38.36 | ± | 150.25 |
| | | 3.86 | | 6.82 | | 3.66 | | 8.26 | | |

Table 30. Results for Lp Precision (↑) (continued)

| Loss Type | Model | Cora ↓ Citeseer | | Cora ↓ Bitcoin | | Citeseer ↓ Cora | | Citeseer ↓ Bitcoin | | Average Rank |
|---|---|---|---|---|---|---|---|---|---|---|
| CrossE_L + PMI_L + PR_L + Triplet_L | MPNN | 97.95 | ± | 86.05 | ± | 89.36 | ± | 83.97 | ± | 65.875 |
| | | 0.25 | | 1.82 | | 2.82 | | 1.76 | | |
| CrossE_L + PMI_L + PR_L + Triplet_L | PAGNN | 62.08 | ± | 37.65 | ± | 59.61 | ± | 34.67 | ± | 182.25 |
| | | 1.30 | | 0.03 | | 0.90 | | 0.04 | | |
| CrossE_L + PMI_L + PR_L + Triplet_L | SAGE | 86.42 | ± | 51.67 | ± | 85.80 | ± | 52.33 | ± | 134.625 |
| | | 5.51 | | 2.90 | | 1.75 | | 2.16 | | |
| CrossE_L + PMI_L + Triplet_L | ALL | 95.93 | ± | 84.18 | ± | 91.21 | ± | 82.59 | ± | 89.25 |
| | | 0.24 | | 2.04 | | 1.56 | | 1.23 | | |
| CrossE_L + PMI_L + Triplet_L | GAT | 98.94 | ± | 85.85 | ± | 97.36 | ± | 85.49 | ± | 20.375 |
| | | 0.28 | | 1.87 | | 0.31 | | 0.94 | | |
| CrossE_L + PMI_L + Triplet_L | GCN | 98.17 | ± | 86.08 | ± | 96.21 | ± | 85.28 | ± | 35.75 |
| | | 0.25 | | 1.03 | | 0.41 | | 1.76 | | |
| CrossE_L + PMI_L + Triplet_L | GIN | 80.53 | ± | 55.58 | ± | 85.32 | ± | 56.43 | ± | 131.25 |
| | | 2.09 | | 1.47 | | 1.10 | | 1.08 | | |
| CrossE_L + PMI_L + Triplet_L | MPNN | 97.63 | ± | 85.08 | ± | 93.54 | ± | 85.30 | ± | 60.125 |
| | | 0.41 | | 1.39 | | 0.91 | | 0.54 | | |
| CrossE_L + PMI_L + Triplet_L | PAGNN | 60.77 | ± | 34.70 | ± | 61.61 | ± | 34.69 | ± | 184.25 |
| | | 1.63 | | 0.03 | | 2.32 | | 0.05 | | |
| CrossE_L + PMI_L + Triplet_L | SAGE | 92.06 | ± | 53.42 | ± | 89.52 | ± | 57.87 | ± | 115.625 |
| | | 1.23 | | 1.72 | | 1.55 | | 1.70 | | |
| CrossE_L + PR_L | ALL | 53.91 | ± | 89.89 | ± | 64.48 | ± | 90.60 | ± | 96.5 |
| | | 3.15 | | 1.81 | | 8.78 | | 1.84 | | |
| CrossE_L + PR_L | GAT | 98.29 | ± | 83.78 | ± | 96.57 | ± | 84.07 | ± | 54.75 |
| | | 0.22 | | 2.79 | | 0.47 | | 1.61 | | |
| CrossE_L + PR_L | GCN | 96.15 | ± | 84.28 | ± | 94.85 | ± | 83.00 | ± | 79.5 |
| | | 1.52 | | 1.84 | | 0.53 | | 2.53 | | |
| CrossE_L + PR_L | GIN | 87.68 | ± | 37.56 | ± | 63.46 | ± | 35.33 | ± | 162.0 |
| | | 1.65 | | 1.64 | | 15.07 | | 1.53 | | |
| CrossE_L + PR_L | MPNN | 96.85 | ± | 85.23 | ± | 87.89 | ± | 86.55 | ± | 67.375 |
| | | 0.93 | | 0.87 | | 2.88 | | 1.73 | | |
| CrossE_L + PR_L | PAGNN | 59.33 | ± | 34.64 | ± | 51.46 | ± | 34.64 | ± | 203.25 |
| | | 7.12 | | 0.00 | | 0.06 | | 0.00 | | |

Table 30. Results for Lp Precision (↑) (continued)

| Loss Type | Model | Cora ↓ Citeseer | | Cora ↓ Bitcoin | | Citeseer ↓ Cora | | Citeseer ↓ Bitcoin | | Average Rank |
|---|---|---|---|---|---|---|---|---|---|---|
| CrossE_L + PR_L | SAGE | 72.41 ± 2.67 | | 38.37 ± 0.58 | | 68.44 ± 2.15 | | 38.40 ± 0.56 | | 165.0 |
| CrossE_L + PR_L + Triplet_L | ALL | 66.07 ± 14.60 | | 34.64 ± 0.00 | | 80.61 ± 2.98 | | 34.65 ± 0.01 | | 183.375 |
| CrossE_L + PR_L + Triplet_L | GAT | 98.15 ± 0.57 | | 83.86 ± 1.17 | | 96.87 ± 0.91 | | 86.00 ± 1.43 | | 43.25 |
| CrossE_L + PR_L + Triplet_L | GCN | 97.79 ± 0.19 | | 84.97 ± 1.24 | | 95.24 ± 0.48 | | 86.36 ± 1.31 | | 50.25 |
| CrossE_L + PR_L + Triplet_L | GIN | 91.71 ± 2.85 | | 42.57 ± 7.43 | | 85.06 ± 2.86 | | 54.39 ± 1.38 | | 131.75 |
| CrossE_L + PR_L + Triplet_L | MPNN | 96.73 ± 0.61 | | 85.38 ± 1.02 | | 86.66 ± 2.22 | | 84.98 ± 0.87 | | 77.125 |
| CrossE_L + PR_L + Triplet_L | PAGNN | 67.79 ± 2.67 | | 37.71 ± 0.20 | | 56.61 ± 3.85 | | 36.03 ± 1.38 | | 177.5 |
| CrossE_L + PR_L + Triplet_L | SAGE | 96.84 ± 0.72 | | 53.52 ± 3.83 | | 94.58 ± 1.23 | | 52.56 ± 2.96 | | 103.5 |
| CrossE_L + Triplet_L | ALL | 96.22 ± 0.19 | | 63.22 ± 1.94 | | 91.36 ± 1.05 | | 61.75 ± 1.71 | | 100.0 |
| CrossE_L + Triplet_L | GAT | 98.82 ± 0.14 | | 86.62 ± 1.26 | | 97.32 ± 0.34 | | 86.55 ± 2.32 | | **13.25** |
| CrossE_L + Triplet_L | GCN | 98.51 ± 0.31 | | 85.07 ± 1.45 | | 96.61 ± 0.24 | | 84.69 ± 1.26 | | 42.75 |
| CrossE_L + Triplet_L | GIN | 95.34 ± 1.06 | | 59.79 ± 1.70 | | 91.16 ± 0.88 | | 62.80 ± 1.64 | | 103.0 |
| CrossE_L + Triplet_L | MPNN | 98.05 ± 0.25 | | 86.04 ± 1.32 | | 94.04 ± 0.67 | | 84.72 ± 2.27 | | 50.375 |
| CrossE_L + Triplet_L | PAGNN | 79.80 ± 7.69 | | 34.65 ± 0.01 | | 85.78 ± 1.75 | | 34.66 ± 0.01 | | 165.5 |
| CrossE_L + Triplet_L | SAGE | 98.05 ± 0.32 | | 61.02 ± 2.70 | | 95.71 ± 0.49 | | 60.00 ± 1.75 | | 82.0 |
| PMI_L | ALL | 66.40 ± 2.33 | | 87.48 ± 1.26 | | 89.89 ± 1.44 | | 87.03 ± 1.22 | | 72.0 |

Table 30. Results for Lp Precision (↑) (continued)

| Loss Type | Model | Cora ↓ Citeseer | | Cora ↓ Bitcoin | | Citeseer ↓ Cora | | Citeseer ↓ Bitcoin | | Average Rank |
|---|---|---|---|---|---|---|---|---|---|---|
| PMI_L | GAT | 98.93 | ± | 85.96 | ± | 97.30 | ± | 86.53 | ± | 16.0 |
| | | 0.15 | | 1.67 | | 0.66 | | 1.26 | | |
| PMI_L | GCN | 98.09 | ± | 84.63 | ± | 95.63 | ± | 84.37 | ± | 58.625 |
| | | 0.33 | | 0.68 | | 0.33 | | 1.07 | | |
| PMI_L | GIN | 75.94 | ± | 34.65 | ± | 77.38 | ± | 34.67 | ± | 173.125 |
| | | 3.24 | | 0.02 | | 2.98 | | 0.03 | | |
| PMI_L | MPNN | 98.02 | ± | 84.68 | ± | 93.39 | ± | 85.34 | ± | 59.25 |
| | | 0.25 | | 2.15 | | 0.50 | | 1.08 | | |
| PMI_L | PAGNN | 59.66 | ± | 34.64 | ± | 62.22 | ± | 34.69 | ± | 190.875 |
| | | 2.10 | | 0.00 | | 2.97 | | 0.05 | | |
| PMI_L | SAGE | 73.05 | ± | 39.20 | ± | 69.06 | ± | 37.27 | ± | 165.5 |
| | | 2.93 | | 1.53 | | 1.76 | | 0.95 | | |
| PMI_L + PR_L | ALL | 52.75 | ± | 91.44 | ± | 88.52 | ± | 87.94 | ± | 82.75 |
| | | 1.54 | | 1.45 | | 1.26 | | 2.20 | | |
| PMI_L + PR_L | GAT | 98.63 | ± | 85.49 | ± | 90.24 | ± | 82.63 | ± | 59.75 |
| | | 0.28 | | 1.52 | | 8.87 | | 2.80 | | |
| PMI_L + PR_L | GCN | 98.13 | ± | 85.01 | ± | 95.97 | ± | 84.31 | ± | 54.625 |
| | | 0.46 | | 1.94 | | 0.65 | | 2.56 | | |
| PMI_L + PR_L | GIN | 74.15 | ± | 34.65 | ± | 76.48 | ± | 48.62 | ± | 163.375 |
| | | 5.08 | | 0.01 | | 10.17 | | 12.76 | | |
| PMI_L + PR_L | MPNN | 97.71 | ± | 86.13 | ± | 83.90 | ± | 81.18 | ± | 83.125 |
| | | 0.43 | | 2.15 | | 2.75 | | 0.72 | | |
| PMI_L + PR_L | PAGNN | 59.92 | ± | 34.65 | ± | 58.61 | ± | 34.67 | ± | 191.625 |
| | | 1.87 | | 0.02 | | 0.86 | | 0.05 | | |
| PMI_L + PR_L | SAGE | 74.73 | ± | 37.65 | ± | 69.99 | ± | 37.51 | ± | 165.75 |
| | | 1.35 | | 1.96 | | 1.08 | | 0.96 | | |
| PMI_L + PR_L + Triplet_L | ALL | 71.08 | ± | 58.96 | ± | 88.59 | ± | 55.68 | ± | 130.25 |
| | | 2.83 | | 15.23 | | 0.88 | | 11.80 | | |
| PMI_L + PR_L + Triplet_L | GAT | 98.90 | ± | 85.96 | ± | 97.03 | ± | 83.91 | ± | 33.125 |
| | | 0.30 | | 1.31 | | 0.81 | | 2.19 | | |
| PMI_L + PR_L + Triplet_L | GCN | 98.21 | ± | 85.19 | ± | 96.04 | ± | 84.72 | ± | 47.125 |
| | | 0.50 | | 2.01 | | 0.62 | | 1.52 | | |

Continued on next page

Table 30. Results for Lp Precision (↑) (continued)

| Loss Type | Model | Cora ↓ Citeseer | | Cora ↓ Bitcoin | | Citeseer ↓ Cora | | Citeseer ↓ Bitcoin | | Average Rank |
|---|---|---|---|---|---|---|---|---|---|---|
| PMI_L + PR_L + Triplet_L | GIN | 78.56 | ± 4.76 | 48.20 | ± 8.02 | 81.11 | ± 3.87 | 34.70 | ± 0.02 | 155.5 |
| PMI_L + PR_L + Triplet_L | MPNN | 97.89 | ± 0.12 | 85.08 | ± 1.72 | 87.48 | ± 1.54 | 82.04 | ± 2.11 | 83.375 |
| PMI_L + PR_L + Triplet_L | PAGNN | 60.87 | ± 2.28 | 34.70 | ± 0.02 | 59.94 | ± 1.84 | 34.69 | ± 0.05 | 185.25 |
| PMI_L + PR_L + Triplet_L | SAGE | 90.46 | ± 2.87 | 55.34 | ± 2.54 | 88.56 | ± 1.51 | 52.86 | ± 1.85 | 122.25 |
| PMI_L + Triplet_L | ALL | 94.86 | ± 0.95 | 66.74 | ± 3.36 | 91.57 | ± 1.10 | 85.07 | ± 1.75 | 83.5 |
| PMI_L + Triplet_L | GAT | 98.85 | ± 0.24 | 84.90 | ± 1.34 | 97.39 | ± 0.28 | 85.49 | ± 0.31 | 30.0 |
| PMI_L + Triplet_L | GCN | 98.13 | ± 0.19 | 85.08 | ± 2.18 | 96.14 | ± 0.43 | 84.76 | ± 1.08 | 48.5 |
| PMI_L + Triplet_L | GIN | 86.12 | ± 2.73 | 47.20 | ± 7.08 | 83.73 | ± 2.04 | 56.80 | ± 1.35 | 135.5 |
| PMI_L + Triplet_L | MPNN | 97.62 | ± 0.50 | 85.32 | ± 2.01 | 93.25 | ± 0.95 | 86.62 | ± 1.43 | 51.25 |
| PMI_L + Triplet_L | PAGNN | 59.44 | ± 2.28 | 34.65 | ± 0.01 | 63.18 | ± 3.13 | 34.65 | ± 0.01 | 191.625 |
| PMI_L + Triplet_L | SAGE | 89.79 | ± 2.47 | 45.90 | ± 3.70 | 89.81 | ± 0.89 | 56.47 | ± 1.25 | 121.0 |
| PR_L | ALL | 49.54 | ± 1.13 | 89.27 | ± 1.78 | 55.46 | ± 3.26 | 89.97 | ± 0.97 | 104.25 |
| PR_L | GAT | 97.79 | ± 0.30 | 81.75 | ± 0.93 | 96.29 | ± 0.69 | 84.05 | ± 2.34 | 68.25 |
| PR_L | GCN | 97.02 | ± 0.55 | 83.11 | ± 1.50 | 95.07 | ± 0.63 | 82.62 | ± 3.17 | 79.75 |
| PR_L | GIN | 88.39 | ± 3.02 | 35.39 | ± 1.67 | 74.27 | ± 12.75 | 36.86 | ± 2.04 | 157.75 |
| PR_L | MPNN | 96.04 | ± 1.98 | 85.55 | ± 1.65 | 88.74 | ± 1.20 | 85.61 | ± 0.98 | 69.25 |

Continued on next page

Table 30. Results for Lp Precision (↑) (continued)

| Loss Type | Model | Cora ↓ Citeseer | | Cora ↓ Bitcoin | | Citeseer ↓ Cora | | Citeseer ↓ Bitcoin | | Average Rank |
|---|---|---|---|---|---|---|---|---|---|---|
| PR_L | PAGNN | 56.51 | ± 8.21 | 34.64 | ± 0.07 | 51.59 | ± 0.01 | 34.68 | ± 0.01 | 199.125 |
| PR_L | SAGE | 72.63 | ± 2.43 | 38.25 | ± 0.63 | 70.54 | ± 2.85 | 38.56 | ± 0.85 | 163.375 |
| PR_L + Triplet_L | ALL | 53.45 | ± 2.89 | 66.27 | ± 18.26 | 57.24 | ± 2.05 | 87.54 | ± 1.66 | 129.5 |
| PR_L + Triplet_L | GAT | 98.05 | ± 0.84 | 82.55 | ± 1.59 | 96.42 | ± 0.79 | 83.33 | ± 1.11 | 64.25 |
| PR_L + Triplet_L | GCN | 97.11 | ± 0.71 | 84.24 | ± 2.24 | 95.39 | ± 0.51 | 84.46 | ± 0.94 | 70.25 |
| PR_L + Triplet_L | GIN | 88.09 | ± 3.00 | 34.64 | ± 0.00 | 67.79 | ± 13.38 | 34.66 | ± 0.01 | 174.25 |
| PR_L + Triplet_L | MPNN | 96.88 | ± 1.51 | 86.32 | ± 2.28 | 89.12 | ± 2.55 | 86.29 | ± 1.03 | 56.5 |
| PR_L + Triplet_L | PAGNN | 59.94 | ± 6.96 | 34.69 | ± 0.01 | 51.64 | ± 0.12 | 34.72 | ± 0.04 | 189.75 |
| PR_L + Triplet_L | SAGE | 77.83 | ± 2.02 | 39.01 | ± 0.29 | 80.23 | ± 10.43 | 37.99 | ± 0.48 | 154.875 |
| Triplet_L | ALL | 96.02 | ± 0.82 | 57.87 | ± 2.80 | 91.05 | ± 0.90 | 83.79 | ± 2.61 | 95.75 |
| Triplet_L | GAT | 99.02 ± 0.24 | | 85.24 | ± 0.89 | 97.64 ± 0.61 | | 85.94 | ± 1.41 | 19.5 |
| Triplet_L | GCN | 98.57 | ± 0.10 | 86.05 | ± 1.68 | 96.35 | ± 0.71 | 85.83 | ± 1.55 | 26.75 |
| Triplet_L | GIN | 96.34 | ± 0.76 | 58.89 | ± 1.41 | 90.09 | ± 1.56 | 64.32 | ± 1.35 | 101.625 |
| Triplet_L | MPNN | 97.91 | ± 0.54 | 85.76 | ± 1.69 | 94.03 | ± 0.60 | 86.49 | ± 2.38 | 44.375 |
| Triplet_L | PAGNN | 80.42 | ± 7.86 | 34.79 | ± 0.10 | 87.26 | ± 1.43 | 38.55 | ± 0.08 | 148.75 |
| Triplet_L | SAGE | 97.91 | ± 0.46 | 64.35 | ± 2.36 | 95.40 | ± 0.31 | 64.64 | ± 2.08 | 82.25 |

Table 31. Lp Recall Performance (↑): This table presents models (Loss function and GNN) ranked by their average performance in terms of lp recall. Top-ranked results are highlighted in <span style="color:red">red</span>, second-ranked in <span style="color:blue">blue</span>, and third-ranked in <span style="color:green">green</span>.

| Loss Type | Model | Cora ↓ Citeseer | | Cora ↓ Bitcoin | | Citeseer ↓ Cora | | Citeseer ↓ Bitcoin | | Average Rank |
|---|---|---|---|---|---|---|---|---|---|---|
| Contr_l | ALL | 91.11 ± 1.56 | | 58.91 ± 2.95 | | 89.31 ± 1.59 | | 82.49 ± 24.13 | | 135.875 |
| Contr_l | GAT | 98.82 ± 0.40 | | 82.44 ± 2.01 | | 98.02 ± 0.46 | | 82.45 ± 1.51 | | 56.0 |
| Contr_l | GCN | 98.01 ± 0.27 | | 82.41 ± 1.30 | | 96.47 ± 0.41 | | 82.50 ± 1.63 | | 70.625 |
| Contr_l | GIN | 94.21 ± 1.13 | | 53.61 ± 1.86 | | 92.95 ± 1.53 | | 61.74 ± 3.52 | | 149.25 |
| Contr_l | MPNN | 97.89 ± 0.47 | | 80.81 ± 1.93 | | 93.00 ± 2.36 | | 81.64 ± 1.55 | | 106.25 |
| Contr_l | PAGNN | 90.28 ± 1.52 | | 99.68 ± 0.26 | | 87.42 ± 0.72 | | 99.38 ± 0.33 | | 90.75 |
| Contr_l | SAGE | 94.85 ± 0.60 | | 80.38 ± 3.52 | | 94.24 ± 1.34 | | 81.60 ± 2.54 | | 114.375 |
| Contr_l + CrossE_L | ALL | 91.91 ± 1.72 | | 66.94 ± 1.35 | | 88.67 ± 0.87 | | 74.57 ± 17.99 | | 152.0 |
| Contr_l + CrossE_L | GAT | 98.71 ± 0.35 | | 81.68 ± 1.32 | | 97.20 ± 0.80 | | 81.84 ± 1.43 | | 70.75 |
| Contr_l + CrossE_L | GCN | 97.89 ± 0.34 | | 80.54 ± 1.32 | | 96.24 ± 0.69 | | 81.32 ± 0.96 | | 92.375 |
| Contr_l + CrossE_L | GIN | 92.03 ± 1.43 | | 60.81 ± 21.95 | | 91.58 ± 1.90 | | 55.03 ± 2.23 | | 153.5 |
| Contr_l + CrossE_L | MPNN | 97.83 ± 0.35 | | 80.08 ± 1.60 | | 93.95 ± 0.64 | | 79.86 ± 1.31 | | 115.0 |
| Contr_l + CrossE_L | PAGNN | 89.45 ± 1.57 | | 99.46 ± 0.23 | | 85.65 ± 2.57 | | 99.44 ± 0.33 | | 97.75 |
| Contr_l + CrossE_L | SAGE | 94.37 ± 0.60 | | 83.94 ± 2.87 | | 94.56 ± 0.82 | | 89.78 ± 1.35 | | 76.25 |
| Contr_l + CrossE_L + PMI_L | ALL | 87.73 ± 2.64 | | 51.57 ± 0.25 | | 82.57 ± 3.85 | | 56.32 ± 1.80 | | 189.375 |

Table 31. Results for Lp Recall (↑) (continued)

| Loss Type | Model | Cora ↓ Citeseer | | Cora ↓ Bitcoin | | Citeseer ↓ Cora | | Citeseer ↓ Bitcoin | | Average Rank |
|---|---|---|---|---|---|---|---|---|---|---|
| Contr_l + CrossE_L + PMI_L | GAT | 99.10 | ± 0.31 | 83.31 | ± 0.93 | 97.77 | ± 0.49 | 83.39 | ± 1.04 | 43.25 |
| Contr_l + CrossE_L + PMI_L | GCN | 98.24 | ± 0.32 | 81.85 | ± 0.71 | 95.98 | ± 0.60 | 81.22 | ± 0.64 | 84.0 |
| Contr_l + CrossE_L + PMI_L | GIN | 88.41 | ± 1.52 | 100.00 | ± 0.00 | 82.90 | ± 1.21 | 79.66 | ± 27.71 | 127.25 |
| Contr_l + CrossE_L + PMI_L | MPNN | 98.95 | ± 0.25 | 80.86 | ± 2.06 | 94.50 | ± 0.90 | 80.61 | ± 1.39 | 93.875 |
| Contr_l + CrossE_L + PMI_L | PAGNN | 89.19 | ± 2.54 | 99.95 | ± 0.12 | 91.37 | ± 3.24 | 100.00 | ± 0.00 | 76.75 |
| Contr_l + CrossE_L + PMI_L | SAGE | 79.23 | ± 2.36 | 86.53 | ± 2.65 | 84.42 | ± 1.35 | 82.65 | ± 1.17 | 129.0 |
| Contr_l + CrossE_L + PMI_L + PR_L | ALL | 96.23 | ± 1.68 | 54.47 | ± 0.73 | 79.00 | ± 2.36 | 56.42 | ± 0.60 | 170.5 |
| Contr_l + CrossE_L + PMI_L + PR_L | GAT | 98.78 | ± 0.50 | 82.51 | ± 1.56 | 97.61 | ± 0.30 | 83.04 | ± 1.69 | 53.5 |
| Contr_l + CrossE_L + PMI_L + PR_L | GCN | 98.02 | ± 0.57 | 81.30 | ± 0.79 | 95.82 | ± 0.83 | 81.74 | ± 1.56 | 87.75 |
| Contr_l + CrossE_L + PMI_L + PR_L | GIN | 86.33 | ± 2.31 | 100.00 | ± 0.01 | 85.15 | ± 1.38 | 99.75 | ± 0.19 | 95.625 |
| Contr_l + CrossE_L + PMI_L + PR_L | MPNN | 98.99 | ± 0.24 | 79.39 | ± 1.10 | 89.49 | ± 4.03 | 77.22 | ± 3.04 | 118.375 |
| Contr_l + CrossE_L + PMI_L + PR_L | PAGNN | 88.91 | ± 1.74 | 99.75 | ± 0.15 | 93.98 | ± 3.93 | 83.43 | ± 0.36 | 87.875 |
| Contr_l + CrossE_L + PMI_L + PR_L | SAGE | 77.61 | ± 2.84 | 89.71 | ± 1.28 | 84.47 | ± 2.68 | 89.95 | ± 3.36 | 119.25 |
| Contr_l + CrossE_L + PMI_L + PR_L + Triplet_L | ALL | 83.28 | ± 3.28 | 49.42 | ± 1.27 | 84.55 | ± 1.13 | 52.22 | ± 2.09 | 195.5 |
| Contr_l + CrossE_L + PMI_L + PR_L + Triplet_L | GAT | 99.09 | ± 0.21 | 81.76 | ± 1.52 | 96.92 | ± 0.96 | 83.07 | ± 1.72 | 55.75 |

Continued on next page

Table 31. Results for Lp Recall (↑) (continued)

| Loss Type | Model | Cora ↓ Citeseer | | Cora ↓ Bitcoin | | Citeseer ↓ Cora | | Citeseer ↓ Bitcoin | | Average Rank |
|---|---|---|---|---|---|---|---|---|---|---|
| Contr_l + CrossE_L + PMI_L + PR_L + Triplet_L | GCN | 98.18 0.26 | ± | 82.06 0.63 | ± | 95.94 0.39 | ± | 81.91 1.36 | ± | 79.875 |
| Contr_l + CrossE_L + PMI_L + PR_L + Triplet_L | GIN | 87.24 1.73 | ± | 81.09 25.91 | ± | 85.70 1.74 | ± | 99.95 0.04 | ± | 121.0 |
| Contr_l + CrossE_L + PMI_L + PR_L + Triplet_L | MPNN | 98.60 0.58 | ± | 81.57 0.69 | ± | 91.08 2.90 | ± | 78.40 1.81 | ± | 109.125 |
| Contr_l + CrossE_L + PMI_L + PR_L + Triplet_L | PAGNN | 91.79 1.28 | ± | 99.76 0.15 | ± | 94.95 1.86 | ± | 99.60 0.37 | ± | 64.875 |
| Contr_l + CrossE_L + PMI_L + PR_L + Triplet_L | SAGE | 89.87 2.62 | ± | 82.07 1.68 | ± | 90.99 1.14 | ± | 82.74 1.56 | ± | 114.625 |
| Contr_l + CrossE_L + PMI_L + Triplet_L | ALL | 93.89 0.68 | ± | 57.61 1.42 | ± | 90.96 1.68 | ± | 60.74 1.33 | ± | 151.375 |
| Contr_l + CrossE_L + PMI_L + Triplet_L | GAT | 99.15 0.18 | ± | 84.29 0.77 | ± | 97.24 0.49 | ± | 83.39 1.24 | ± | 40.125 |
| Contr_l + CrossE_L + PMI_L + Triplet_L | GCN | 98.30 0.36 | ± | 82.43 1.63 | ± | 95.03 1.44 | ± | 82.23 1.38 | ± | 78.25 |
| Contr_l + CrossE_L + PMI_L + Triplet_L | GIN | 88.04 2.21 | ± | 62.45 21.00 | ± | 84.69 2.37 | ± | 51.84 0.55 | ± | 179.75 |
| Contr_l + CrossE_L + PMI_L + Triplet_L | MPNN | 98.76 0.17 | ± | 81.29 1.61 | ± | 94.38 1.26 | ± | 81.27 1.47 | ± | 91.0 |
| Contr_l + CrossE_L + PMI_L + Triplet_L | PAGNN | 89.96 1.77 | ± | **100.00 0.00** | ± | 92.06 3.40 | ± | **100.00 0.00** | ± | 69.25 |
| Contr_l + CrossE_L + PMI_L + Triplet_L | SAGE | 91.67 0.78 | ± | 83.81 2.08 | ± | 90.78 1.44 | ± | 81.64 2.32 | ± | 108.25 |
| Contr_l + CrossE_L + PR_L | ALL | 95.62 1.93 | ± | 51.68 0.87 | ± | 89.53 6.37 | ± | 54.23 0.46 | ± | 159.25 |
| Contr_l + CrossE_L + PR_L | GAT | 97.08 2.48 | ± | 81.02 2.06 | ± | 96.96 0.41 | ± | 79.35 1.65 | ± | 97.5 |

Table 31. Results for Lp Recall (↑) (continued)

| Loss Type | Model | Cora ↓ Citeseer | | Cora ↓ Bitcoin | | Citeseer ↓ Cora | | Citeseer ↓ Bitcoin | | Average Rank |
|---|---|---|---|---|---|---|---|---|---|---|
| Contr_l + CrossE_L + PR_L | GCN | 96.96 0.67 | ± | 79.83 1.94 | ± | 94.86 0.61 | ± | 81.31 1.31 | ± | 108.5 |
| Contr_l + CrossE_L + PR_L | GIN | 85.59 3.00 | ± | 100.00 0.00 | ± | 77.01 3.61 | ± | **100.00 0.00** | ± | 101.25 |
| Contr_l + CrossE_L + PR_L | MPNN | 85.68 9.98 | ± | 80.22 1.06 | ± | 79.54 7.04 | ± | 81.45 1.25 | ± | 160.75 |
| Contr_l + CrossE_L + PR_L | PAGNN | 81.16 4.33 | ± | 83.76 0.00 | ± | 99.38 1.12 | ± | 84.16 0.28 | ± | 82.5 |
| Contr_l + CrossE_L + PR_L | SAGE | 90.82 8.01 | ± | 85.73 1.71 | ± | 89.22 3.73 | ± | 84.74 2.44 | ± | 96.25 |
| Contr_l + CrossE_L + PR_L + Triplet_L | ALL | 88.91 1.50 | ± | 99.99 0.03 | ± | 85.77 0.83 | ± | 64.48 19.92 | ± | 129.5 |
| Contr_l + CrossE_L + PR_L + Triplet_L | GAT | 98.71 0.27 | ± | 82.14 1.15 | ± | 97.32 0.94 | ± | 83.41 0.80 | ± | 56.625 |
| Contr_l + CrossE_L + PR_L + Triplet_L | GCN | 97.56 0.47 | ± | 81.01 1.20 | ± | 95.64 0.51 | ± | 81.15 3.54 | ± | 100.5 |
| Contr_l + CrossE_L + PR_L + Triplet_L | GIN | 92.07 0.41 | ± | 63.80 20.26 | ± | 89.10 1.03 | ± | 56.70 1.83 | ± | 156.75 |
| Contr_l + CrossE_L + PR_L + Triplet_L | MPNN | 96.43 1.28 | ± | 82.70 1.80 | ± | 90.28 1.35 | ± | 80.17 2.96 | ± | 112.625 |
| Contr_l + CrossE_L + PR_L + Triplet_L | PAGNN | 87.84 2.47 | ± | 99.76 0.14 | ± | 86.26 7.78 | ± | 86.48 6.82 | ± | 102.875 |
| Contr_l + CrossE_L + PR_L + Triplet_L | SAGE | 96.14 0.34 | ± | 82.93 1.86 | ± | 93.72 1.62 | ± | 82.93 3.08 | ± | 88.0 |
| Contr_l + CrossE_L + Triplet_L | ALL | 93.29 1.65 | ± | 64.73 1.62 | ± | 91.23 1.14 | ± | 62.15 1.81 | ± | 145.5 |
| Contr_l + CrossE_L + Triplet_L | GAT | 98.92 0.34 | ± | 83.09 1.60 | ± | 97.77 0.47 | ± | 82.88 1.14 | ± | 48.875 |
| Contr_l + CrossE_L + Triplet_L | GCN | 98.26 0.26 | ± | 81.56 1.23 | ± | 96.61 0.50 | ± | 82.26 1.59 | ± | 74.375 |
| Contr_l + CrossE_L + Triplet_L | GIN | 95.40 0.71 | ± | 54.88 1.86 | ± | 93.15 1.39 | ± | 61.19 1.24 | ± | 144.75 |

Table 31. Results for Lp Recall (↑) (continued)

| Loss Type | Model | Cora ↓ Citeseer | | Cora ↓ Bitcoin | | Citeseer ↓ Cora | | Citeseer ↓ Bitcoin | | Average Rank |
|---|---|---|---|---|---|---|---|---|---|---|
| Contr_l + CrossE_L + Triplet_L | MPNN | 98.22 0.26 | ± | 79.15 2.44 | ± | 95.00 0.69 | ± | 81.02 2.41 | ± | 105.0 |
| Contr_l + CrossE_L + Triplet_L | PAGNN | 90.79 1.93 | ± | 99.92 0.06 | ± | 89.51 1.42 | ± | 99.41 0.30 | ± | 83.375 |
| Contr_l + CrossE_L + Triplet_L | SAGE | 96.41 0.68 | ± | 81.88 1.57 | ± | 95.49 0.54 | ± | 80.54 1.74 | ± | 100.5 |
| Contr_l + PMI_L | ALL | 87.97 3.64 | ± | 80.01 27.41 | ± | 86.75 2.22 | ± | 54.44 1.20 | ± | 167.625 |
| Contr_l + PMI_L | GAT | 98.98 0.32 | ± | 83.63 2.14 | ± | 98.00 0.27 | ± | 82.63 0.43 | ± | 45.625 |
| Contr_l + PMI_L | GCN | 97.96 0.41 | ± | 82.31 1.67 | ± | 95.85 1.03 | ± | 82.41 2.30 | ± | 78.25 |
| Contr_l + PMI_L | GIN | 85.66 3.87 | ± | 50.39 1.51 | ± | 84.38 1.46 | ± | 61.34 21.63 | ± | 186.75 |
| Contr_l + PMI_L | MPNN | 98.76 0.14 | ± | 78.10 1.42 | ± | 95.09 0.91 | ± | 82.15 1.60 | ± | 90.875 |
| Contr_l + PMI_L | PAGNN | 90.92 1.32 | ± | 100.00 0.00 | ± | 94.10 1.52 | ± | 100.00 0.00 | ± | 61.375 |
| Contr_l + PMI_L | SAGE | 78.93 3.17 | ± | 82.89 2.59 | ± | 88.91 2.20 | ± | 84.79 0.94 | ± | 121.5 |
| Contr_l + PMI_L + PR_L | ALL | 97.44 0.90 | ± | 51.73 1.18 | ± | 79.25 2.28 | ± | 57.37 1.43 | ± | 167.5 |
| Contr_l + PMI_L + PR_L | GAT | 98.65 0.38 | ± | 83.49 1.85 | ± | 93.97 3.72 | ± | 76.64 1.98 | ± | 93.0 |
| Contr_l + PMI_L + PR_L | GCN | 98.37 0.41 | ± | 81.54 1.37 | ± | 95.99 0.85 | ± | 81.98 1.61 | ± | 78.875 |
| Contr_l + PMI_L + PR_L | GIN | 86.56 5.39 | ± | 100.00 0.00 | ± | 84.22 2.93 | ± | 99.77 0.07 | ± | 97.0 |
| Contr_l + PMI_L + PR_L | MPNN | 99.02 0.19 | ± | 81.11 1.41 | ± | 88.07 3.16 | ± | 78.80 1.54 | ± | 111.5 |
| Contr_l + PMI_L + PR_L | PAGNN | 89.89 1.53 | ± | 100.00 0.00 | ± | 95.42 1.25 | ± | 100.00 0.00 | ± | 58.25 |

Table 31. Results for Lp Recall (↑) (continued)

| Loss Type | Model | Cora ↓ Citeseer | | Cora ↓ Bitcoin | | Citeseer ↓ Cora | | Citeseer ↓ Bitcoin | | Average Rank |
|---|---|---|---|---|---|---|---|---|---|---|
| Contr_l + PMI_L + PR_L | SAGE | 79.45 | ± 2.32 | 85.48 | ± 3.02 | 86.24 | ± 2.86 | 88.26 | ± 1.14 | 116.25 |
| Contr_l + PMI_L + ALL PR_L + Triplet_L | ALL | 84.36 | ± 3.39 | 100.00 | ± 0.00 | 87.47 | ± 0.70 | 50.05 | ± 1.23 | 139.0 |
| Contr_l + PMI_L + PR_L + Triplet_L | GAT | 98.38 | ± 0.68 | 82.45 | ± 2.64 | 95.79 | ± 1.16 | 79.15 | ± 3.33 | 88.125 |
| Contr_l + PMI_L + PR_L + Triplet_L | GCN | 97.90 | ± 0.38 | 83.07 | ± 0.99 | 96.05 | ± 0.63 | 82.93 | ± 1.11 | 67.625 |
| Contr_l + PMI_L + PR_L + Triplet_L | GIN | 89.30 | ± 1.49 | 51.11 | ± 2.28 | 85.98 | ± 2.38 | 80.50 | ± 26.70 | 166.25 |
| Contr_l + PMI_L + PR_L + Triplet_L | MPNN | 98.90 | ± 0.07 | 80.86 | ± 1.29 | 89.54 | ± 1.07 | 77.71 | ± 0.76 | 113.0 |
| Contr_l + PMI_L + PR_L + Triplet_L | PAGNN | 91.00 | ± 1.53 | 100.00 | ± 0.00 | 94.14 | ± 3.48 | 99.86 | ± 0.30 | 64.25 |
| Contr_l + PMI_L + PR_L + Triplet_L | SAGE | 92.85 | ± 0.53 | 83.51 | ± 1.64 | 90.79 | ± 0.77 | 81.20 | ± 2.94 | 111.125 |
| Contr_l + PR_L | ALL | 94.53 | ± 4.06 | 54.07 | ± 0.80 | 84.86 | ± 9.79 | 55.39 | ± 0.26 | 168.75 |
| Contr_l + PR_L | GAT | 97.87 | ± 1.04 | 80.51 | ± 1.31 | 96.78 | ± 0.44 | 79.79 | ± 1.81 | 96.5 |
| Contr_l + PR_L | GCN | 97.10 | ± 0.46 | 80.63 | ± 1.33 | 94.70 | ± 0.81 | 81.30 | ± 1.54 | 105.5 |
| Contr_l + PR_L | GIN | 83.72 | ± 7.17 | 84.55 | ± 8.64 | 82.01 | ± 11.92 | 81.90 | ± 24.75 | 137.25 |
| Contr_l + PR_L | MPNN | 85.54 | ± 3.51 | 81.74 | ± 2.33 | 83.47 | ± 3.21 | 82.58 | ± 0.59 | 141.5 |
| Contr_l + PR_L | PAGNN | 83.10 | ± 9.78 | 89.89 | ± 8.53 | 99.54 | ± 0.53 | 85.20 | ± 0.21 | 74.875 |
| Contr_l + PR_L | SAGE | 89.87 | ± 8.55 | 85.35 | ± 2.08 | 89.72 | ± 6.12 | 93.41 | ± 1.67 | 93.875 |
| Contr_l + PR_L + Triplet_L | ALL | 91.52 | ± 4.09 | 80.51 | ± 26.69 | 84.63 | ± 3.82 | 100.00 | ± 0.00 | 112.875 |

Table 31. Results for Lp Recall (↑) (continued)

| Loss Type | Model | Cora ↓ Citeseer | | Cora ↓ Bitcoin | | Citeseer ↓ Cora | | Citeseer ↓ Bitcoin | | Average Rank |
|---|---|---|---|---|---|---|---|---|---|---|
| Contr_l + PR_L + Triplet_L | GAT | 98.96 | ± 0.39 | 83.46 | ± 1.70 | 97.46 | ± 0.91 | 82.12 | ± 1.03 | 53.0 |
| Contr_l + PR_L + Triplet_L | GCN | 97.47 | ± 0.38 | 80.81 | ± 1.11 | 95.19 | ± 0.62 | 83.79 | ± 1.93 | 86.625 |
| Contr_l + PR_L + Triplet_L | GIN | 92.95 | ± 1.43 | 56.62 | ± 2.21 | 88.79 | ± 2.17 | 51.58 | ± 1.31 | 164.0 |
| Contr_l + PR_L + Triplet_L | MPNN | 94.85 | ± 4.68 | 80.83 | ± 2.31 | 85.72 | ± 7.12 | 81.77 | ± 1.45 | 129.875 |
| Contr_l + PR_L + Triplet_L | PAGNN | 87.02 | ± 1.77 | 99.94 | ± 0.08 | 86.08 | ± 8.22 | 99.08 | ± 0.31 | 98.5 |
| Contr_l + PR_L + Triplet_L | SAGE | 95.60 | ± 0.62 | 83.80 | ± 1.48 | 94.31 | ± 0.31 | 80.12 | ± 2.11 | 99.875 |
| Contr_l + Triplet_L | ALL | 94.49 | ± 0.79 | 60.75 | ± 1.30 | 90.96 | ± 0.71 | 57.39 | ± 1.22 | 149.875 |
| Contr_l + Triplet_L | GAT | 99.12 | ± 0.13 | 83.34 | ± 1.09 | 97.69 | ± 0.44 | 82.46 | ± 1.19 | 49.25 |
| Contr_l + Triplet_L | GCN | 98.19 | ± 0.11 | 81.89 | ± 1.19 | 96.37 | ± 0.57 | 81.23 | ± 1.28 | 81.0 |
| Contr_l + Triplet_L | GIN | 95.19 | ± 0.91 | 51.67 | ± 0.48 | 93.57 | ± 1.39 | 56.58 | ± 1.29 | 150.25 |
| Contr_l + Triplet_L | MPNN | 98.16 | ± 0.55 | 79.71 | ± 2.45 | 95.44 | ± 0.88 | 80.60 | ± 1.33 | 104.5 |
| Contr_l + Triplet_L | PAGNN | 91.54 | ± 0.97 | 99.37 | ± 0.14 | 88.51 | ± 1.22 | 95.46 | ± 8.74 | 87.0 |
| Contr_l + Triplet_L | SAGE | 96.43 | ± 0.51 | 80.18 | ± 2.39 | 94.69 | ± 1.09 | 80.93 | ± 1.49 | 112.75 |
| CrossE_L | ALL | 95.44 | ± 3.51 | 69.41 | ± 1.35 | 96.33 | ± 5.27 | 63.22 | ± 0.70 | 120.75 |
| CrossE_L | GAT | 95.07 | ± 5.26 | 93.93 | ± 1.01 | 96.59 | ± 2.21 | 78.44 | ± 3.08 | 84.0 |
| CrossE_L | GCN | 91.60 | ± 18.77 | 58.63 | ± 1.53 | 100.00 | ± 0.00 | 100.00 | ± 0.00 | 80.375 |

Table 31. Results for Lp Recall (↑) (continued)

| Loss Type | Model | Cora ↓ Citeseer | Cora ↓ Bitcoin | Citeseer ↓ Cora | Citeseer ↓ Bitcoin | Average Rank |
|---|---|---|---|---|---|---|
| CrossE_L | GIN | 100.00 ± 0.00 | 100.00 ± 0.00 | 100.00 ± 0.00 | 100.00 ± 0.00 | 5.375 |
| CrossE_L | MPNN | 88.82 ± 4.66 | 100.00 ± 0.01 | 85.84 ± 2.32 | 100.00 ± 0.00 | 86.0 |
| CrossE_L | PAGNN | 79.54 ± 3.58 | 100.00 ± 0.00 | 87.28 ± 17.30 | 100.00 ± 0.00 | 92.75 |
| CrossE_L | SAGE | 86.51 ± 4.49 | 98.31 ± 0.31 | 91.61 ± 5.73 | 99.45 ± 0.49 | 89.75 |
| CrossE_L + PMI_L | ALL | 85.90 ± 0.68 | 57.60 ± 1.67 | 82.09 ± 1.58 | 57.10 ± 1.54 | 186.5 |
| CrossE_L + PMI_L | GAT | 99.14 ± 0.16 | 81.83 ± 1.15 | 97.73 ± 0.40 | 81.71 ± 1.81 | 61.5 |
| CrossE_L + PMI_L | GCN | 97.93 ± 0.23 | 81.80 ± 1.51 | 95.84 ± 0.73 | 82.85 ± 1.54 | 78.0 |
| CrossE_L + PMI_L | GIN | 86.38 ± 3.44 | 90.84 ± 20.49 | 83.14 ± 3.43 | 81.40 ± 25.30 | 130.75 |
| CrossE_L + PMI_L | MPNN | 98.73 ± 0.13 | 80.07 ± 1.74 | 94.69 ± 1.46 | 80.82 ± 1.64 | 99.625 |
| CrossE_L + PMI_L | PAGNN | 90.89 ± 3.21 | 87.79 ± 0.57 | 92.86 ± 1.07 | 100.00 ± 0.00 | 75.5 |
| CrossE_L + PMI_L | SAGE | 76.74 ± 1.67 | 86.00 ± 3.14 | 83.00 ± 1.58 | 91.54 ± 8.09 | 122.25 |
| CrossE_L + PMI_L + PR_L | ALL | 98.35 ± 0.69 | 56.72 ± 1.27 | 78.78 ± 1.76 | 58.70 ± 1.13 | 156.0 |
| CrossE_L + PMI_L + PR_L | GAT | 98.96 ± 0.27 | 82.79 ± 0.98 | 93.84 ± 3.20 | 75.39 ± 1.14 | 92.375 |
| CrossE_L + PMI_L + PR_L | GCN | 98.12 ± 0.52 | 82.62 ± 1.11 | 96.10 ± 0.24 | 81.92 ± 1.61 | 73.875 |
| CrossE_L + PMI_L + PR_L | GIN | 85.01 ± 2.43 | 50.31 ± 0.37 | 81.99 ± 4.41 | 90.26 ± 21.70 | 157.75 |
| CrossE_L + PMI_L + PR_L | MPNN | 98.66 ± 0.38 | 78.85 ± 1.53 | 90.54 ± 4.95 | 79.44 ± 1.25 | 119.25 |

Table 31. Results for Lp Recall (↑) (continued)

| Loss Type | Model | Cora ↓ Citeseer | | Cora ↓ Bitcoin | | Citeseer ↓ Cora | | Citeseer ↓ Bitcoin | | Average Rank |
|---|---|---|---|---|---|---|---|---|---|---|
| CrossE_L + PMI_L + PR_L | PAGNN | 90.07 0.77 | ± | 99.84 0.15 | ± | 94.52 3.14 | ± | 100.00 0.00 | ± | 66.0 |
| CrossE_L + PMI_L + PR_L | SAGE | 78.42 0.86 | ± | 89.89 1.70 | ± | 81.00 0.99 | ± | 87.19 1.58 | ± | 124.625 |
| CrossE_L + PMI_L + PR_L + Triplet_L | ALL | 84.17 4.30 | ± | 54.21 2.00 | ± | 83.56 1.71 | ± | 48.21 1.27 | ± | 194.0 |
| CrossE_L + PMI_L + PR_L + Triplet_L | GAT | 99.03 0.15 | ± | 82.35 1.30 | ± | 96.39 1.55 | ± | 82.53 2.12 | ± | 58.75 |
| CrossE_L + PMI_L + PR_L + Triplet_L | GCN | 97.93 0.53 | ± | 82.85 1.29 | ± | 95.86 0.59 | ± | 81.16 2.12 | ± | 84.25 |
| CrossE_L + PMI_L + PR_L + Triplet_L | GIN | 86.24 2.20 | ± | 74.57 23.22 | ± | 84.53 2.66 | ± | 90.00 22.26 | ± | 141.75 |
| CrossE_L + PMI_L + PR_L + Triplet_L | MPNN | 98.57 0.14 | ± | 79.86 1.80 | ± | 91.83 2.97 | ± | 81.64 2.26 | ± | 105.25 |
| CrossE_L + PMI_L + PR_L + Triplet_L | PAGNN | 89.70 1.50 | ± | 84.32 0.17 | ± | 94.31 2.33 | ± | 99.87 0.30 | ± | 80.625 |
| CrossE_L + PMI_L + PR_L + Triplet_L | SAGE | 90.67 4.76 | ± | 85.27 1.30 | ± | 89.66 1.02 | ± | 81.94 1.20 | ± | 108.25 |
| CrossE_L + PMI_L + Triplet_L | ALL | 95.74 0.78 | ± | 59.66 1.06 | ± | 92.96 1.51 | ± | 62.45 0.92 | ± | 139.75 |
| CrossE_L + PMI_L + Triplet_L | GAT | 98.98 0.30 | ± | 82.55 1.39 | ± | 97.74 0.44 | ± | 83.67 1.57 | ± | 46.5 |
| CrossE_L + PMI_L + Triplet_L | GCN | 98.12 0.16 | ± | 82.12 1.62 | ± | 96.07 0.33 | ± | 81.19 1.65 | ± | 83.375 |
| CrossE_L + PMI_L + Triplet_L | GIN | 88.81 1.38 | ± | 50.96 1.62 | ± | 85.85 1.56 | ± | 50.57 1.30 | ± | 185.0 |
| CrossE_L + PMI_L + Triplet_L | MPNN | 98.72 0.30 | ± | 79.73 1.48 | ± | 95.07 0.66 | ± | 81.78 2.69 | ± | 92.5 |
| CrossE_L + PMI_L + Triplet_L | PAGNN | 91.08 2.15 | ± | 99.78 0.05 | ± | 91.80 4.12 | ± | 99.75 0.34 | ± | 75.625 |
| CrossE_L + PMI_L + Triplet_L | SAGE | 92.90 0.93 | ± | 81.15 2.02 | ± | 91.04 0.57 | ± | 83.01 2.38 | ± | 110.25 |

Table 31. Results for Lp Recall (↑) (continued)

| Loss Type | Model | Cora ↓ Citeseer | | Cora ↓ Bitcoin | | Citeseer ↓ Cora | | Citeseer ↓ Bitcoin | | Average Rank |
|---|---|---|---|---|---|---|---|---|---|---|
| CrossE_L + PR_L | ALL | 94.93 | ± 3.10 | 53.80 | ± 0.59 | 89.74 | ± 7.23 | 54.84 | ± 0.86 | 158.5 |
| CrossE_L + PR_L | GAT | 97.74 | ± 0.61 | 78.99 | ± 1.64 | 95.98 | ± 0.42 | 78.30 | ± 2.49 | 110.75 |
| CrossE_L + PR_L | GCN | 93.02 | ± 2.35 | 77.02 | ± 2.57 | 94.63 | ± 1.81 | 77.71 | ± 2.19 | 131.625 |
| CrossE_L + PR_L | GIN | 82.72 | ± 3.85 | 84.26 | ± 8.80 | 89.01 | ± 12.88 | 96.07 | ± 8.76 | 109.125 |
| CrossE_L + PR_L | MPNN | 89.84 | ± 9.48 | 81.50 | ± 1.65 | 72.63 | ± 5.18 | 81.93 | ± 1.55 | 146.5 |
| CrossE_L + PR_L | PAGNN | 83.76 | ± 9.32 | 100.00 | ± 0.00 | 99.94 | ± 0.06 | 100.00 | ± 0.00 | 52.5 |
| CrossE_L + PR_L | SAGE | 74.92 | ± 2.46 | 93.81 | ± 1.16 | 81.89 | ± 0.89 | 92.97 | ± 1.99 | 121.25 |
| CrossE_L + PR_L + Triplet_L | ALL | 92.47 | ± 4.26 | 100.00 | ± 0.00 | 76.27 | ± 4.59 | 100.00 | ± 0.00 | 86.25 |
| CrossE_L + PR_L + Triplet_L | GAT | 98.51 | ± 0.39 | 80.36 | ± 1.73 | 97.16 | ± 0.65 | 83.23 | ± 1.09 | 71.0 |
| CrossE_L + PR_L + Triplet_L | GCN | 97.46 | ± 0.45 | 81.75 | ± 0.86 | 95.34 | ± 1.07 | 84.01 | ± 1.46 | 80.0 |
| CrossE_L + PR_L + Triplet_L | GIN | 90.99 | ± 3.78 | 75.06 | ± 22.84 | 86.66 | ± 5.05 | 54.45 | ± 1.36 | 164.0 |
| CrossE_L + PR_L + Triplet_L | MPNN | 93.34 | ± 4.07 | 81.79 | ± 0.96 | 80.48 | ± 6.35 | 82.69 | ± 1.16 | 126.75 |
| CrossE_L + PR_L + Triplet_L | PAGNN | 84.47 | ± 1.71 | 84.26 | ± 0.28 | 95.97 | ± 1.84 | 92.46 | ± 8.11 | 84.875 |
| CrossE_L + PR_L + Triplet_L | SAGE | 96.53 | ± 0.93 | 85.31 | ± 2.51 | 94.90 | ± 1.29 | 83.67 | ± 1.37 | 71.125 |
| CrossE_L + Triplet_L | ALL | 96.82 | ± 0.31 | 62.71 | ± 1.28 | 93.14 | ± 1.68 | 61.75 | ± 2.24 | 135.5 |
| CrossE_L + Triplet_L | GAT | 99.42 | ± 0.15 | 83.40 | ± 0.98 | 98.17 | ± 0.49 | 83.60 | ± 0.79 | 38.25 |

Table 31. Results for Lp Recall (↑) (continued)

| Loss Type | Model | Cora ↓ Citeseer | | Cora ↓ Bitcoin | | Citeseer ↓ Cora | | Citeseer ↓ Bitcoin | | Average Rank |
|---|---|---|---|---|---|---|---|---|---|---|
| CrossE_L + Triplet_L | GCN | 98.38 0.18 | ± | 82.33 1.42 | ± | 96.77 0.59 | ± | 82.22 3.02 | ± | 67.875 |
| CrossE_L + Triplet_L | GIN | 95.43 0.45 | ± | 51.05 1.59 | ± | 92.20 1.53 | ± | 53.59 1.93 | ± | 155.25 |
| CrossE_L + Triplet_L | MPNN | 98.52 0.19 | ± | 79.78 1.11 | ± | 95.45 1.17 | ± | 78.25 2.58 | ± | 105.0 |
| CrossE_L + Triplet_L | PAGNN | 90.67 1.67 | ± | 100.00 0.00 | ± | 88.28 2.22 | ± | 100.00 0.00 | ± | 76.625 |
| CrossE_L + Triplet_L | SAGE | 97.60 0.43 | ± | 82.26 2.91 | ± | 96.02 0.69 | ± | 82.56 1.39 | ± | 77.375 |
| PMI_L | ALL | 83.37 0.58 | ± | 59.73 1.99 | ± | 82.84 1.58 | ± | 59.17 1.17 | ± | 186.0 |
| PMI_L | GAT | 99.26 0.18 | ± | 83.11 1.20 | ± | 97.92 0.24 | ± | 83.38 1.24 | ± | 42.25 |
| PMI_L | GCN | 97.84 0.41 | ± | 81.67 1.26 | ± | 96.02 0.46 | ± | 84.14 2.19 | ± | 72.875 |
| PMI_L | GIN | 84.71 3.25 | ± | 99.99 0.01 | ± | 82.57 2.05 | ± | 99.91 0.12 | ± | 103.625 |
| PMI_L | MPNN | 98.87 0.28 | ± | 80.01 1.02 | ± | 95.27 0.84 | ± | 79.19 1.61 | ± | 99.375 |
| PMI_L | PAGNN | 90.01 1.84 | ± | 100.00 0.00 | ± | 89.15 4.90 | ± | 99.74 0.35 | ± | 80.75 |
| PMI_L | SAGE | 77.27 1.22 | ± | 86.85 1.99 | ± | 81.43 1.09 | ± | 89.74 2.17 | ± | 125.25 |
| PMI_L + PR_L | ALL | 97.98 1.48 | ± | 54.35 0.78 | ± | 77.87 1.27 | ± | 57.77 2.20 | ± | 161.875 |
| PMI_L + PR_L | GAT | 98.48 0.68 | ± | 80.90 1.98 | ± | 92.68 4.90 | ± | 81.27 2.46 | ± | 102.125 |
| PMI_L + PR_L | GCN | 97.94 0.22 | ± | 83.89 1.10 | ± | 95.69 0.96 | ± | 81.20 0.78 | ± | 79.375 |
| PMI_L + PR_L | GIN | 85.35 3.64 | ± | 100.00 0.01 | ± | 83.48 6.75 | ± | 68.75 28.48 | ± | 136.5 |

Table 31. Results for Lp Recall (↑) (continued)

| Loss Type | Model | Cora ↓ Citeseer | | Cora ↓ Bitcoin | | Citeseer ↓ Cora | | Citeseer ↓ Bitcoin | | Average Rank |
|---|---|---|---|---|---|---|---|---|---|---|
| PMI_L + PR_L | MPNN | 98.74 | ± 0.32 | 78.62 | ± 1.32 | 86.57 | ± 1.70 | 77.22 | ± 3.91 | 128.875 |
| PMI_L + PR_L | PAGNN | 90.00 | ± 1.41 | 99.96 | ± 0.08 | 95.93 | ± 1.63 | 99.88 | ± 0.28 | 61.0 |
| PMI_L + PR_L | SAGE | 76.85 | ± 1.42 | 90.75 | ± 3.71 | 83.35 | ± 0.88 | 87.09 | ± 4.46 | 122.0 |
| PMI_L + PR_L + Triplet_L | ALL | 83.40 | ± 2.19 | 59.16 | ± 22.88 | 84.97 | ± 2.15 | 57.77 | ± 23.63 | 182.5 |
| PMI_L + PR_L + Triplet_L | GAT | 98.83 | ± 0.27 | 83.65 | ± 0.70 | 96.73 | ± 0.79 | 82.85 | ± 1.78 | 51.625 |
| PMI_L + PR_L + Triplet_L | GCN | 98.26 | ± 0.44 | 83.46 | ± 2.04 | 95.99 | ± 0.80 | 81.01 | ± 1.22 | 76.625 |
| PMI_L + PR_L + Triplet_L | GIN | 89.02 | ± 1.08 | 62.97 | ± 20.75 | 84.60 | ± 2.59 | 99.89 | ± 0.09 | 132.0 |
| PMI_L + PR_L + Triplet_L | MPNN | 98.89 | ± 0.12 | 81.28 | ± 2.04 | 88.49 | ± 0.89 | 79.69 | ± 2.14 | 110.625 |
| PMI_L + PR_L + Triplet_L | PAGNN | 89.31 | ± 1.62 | 99.73 | ± 0.05 | 93.66 | ± 3.52 | 99.70 | ± 0.43 | 79.0 |
| PMI_L + PR_L + Triplet_L | SAGE | 91.48 | ± 4.33 | 83.19 | ± 1.80 | 91.17 | ± 1.19 | 82.17 | ± 2.03 | 106.75 |
| PMI_L + Triplet_L | ALL | 96.04 | ± 0.84 | 55.12 | ± 1.31 | 92.88 | ± 0.71 | 62.38 | ± 0.40 | 142.25 |
| PMI_L + Triplet_L | GAT | 99.17 | ± 0.20 | 83.73 | ± 1.08 | 97.78 | ± 0.44 | 83.87 | ± 1.29 | 37.25 |
| PMI_L + Triplet_L | GCN | 98.24 | ± 0.19 | 82.76 | ± 0.85 | 96.26 | ± 0.91 | 81.69 | ± 1.26 | 73.375 |
| PMI_L + Triplet_L | GIN | 89.59 | ± 1.19 | 63.91 | ± 20.22 | 86.38 | ± 1.93 | 51.53 | ± 1.36 | 172.75 |
| PMI_L + Triplet_L | MPNN | 98.89 | ± 0.28 | 79.69 | ± 0.60 | 94.73 | ± 0.64 | 79.61 | ± 1.01 | 101.875 |
| PMI_L + Triplet_L | PAGNN | 90.58 | ± 3.39 | 100.00 | ± 0.00 | 88.53 | ± 4.79 | 100.00 | ± 0.00 | 76.25 |

Table 31. Results for Lp Recall (↑) (continued)

| Loss Type | Model | Cora ↓ Citeseer | | Cora ↓ Bitcoin | | Citeseer ↓ Cora | | Citeseer ↓ Bitcoin | | Average Rank |
|---|---|---|---|---|---|---|---|---|---|---|
| PMI_L + Triplet_L | SAGE | 90.29 | ± | 85.38 | ± | 91.90 | ± | 82.49 | ± | 100.125 |
| | | 1.93 | | 2.39 | | 0.43 | | 2.70 | | |
| PR_L | ALL | 98.80 | ± | 54.91 | ± | 96.71 | ± | 56.80 | ± | 110.5 |
| | | 1.01 | | 1.47 | | 2.77 | | 0.42 | | |
| PR_L | GAT | 97.74 | ± | 77.94 | ± | 96.50 | ± | 78.06 | ± | 108.625 |
| | | 0.80 | | 1.67 | | 0.34 | | 1.28 | | |
| PR_L | GCN | 95.72 | ± | 77.37 | ± | 94.10 | ± | 75.87 | ± | 130.625 |
| | | 1.26 | | 0.45 | | 1.77 | | 3.50 | | |
| PR_L | GIN | 81.02 | ± | 95.99 | ± | 75.36 | ± | 88.15 | ± | 124.0 |
| | | 3.50 | | 8.96 | | 14.08 | | 10.82 | | |
| PR_L | MPNN | 87.86 | ± | 81.18 | ± | 75.23 | ± | 81.94 | ± | 150.875 |
| | | 11.48 | | 0.46 | | 4.82 | | 1.56 | | |
| PR_L | PAGNN | 87.67 | ± | 99.98 | ± | 99.87 | ± | 99.91 | ± | 53.625 |
| | | 11.29 | | 0.04 | | 0.06 | | 0.01 | | |
| PR_L | SAGE | 74.37 | ± | 89.84 | ± | 81.88 | ± | 91.48 | ± | 124.25 |
| | | 2.47 | | 1.25 | | 1.58 | | 2.26 | | |
| PR_L + Triplet_L | ALL | 96.14 | ± | 60.49 | ± | 97.36 | ± | 55.53 | ± | 122.375 |
| | | 1.84 | | 22.11 | | 1.22 | | 0.72 | | |
| PR_L + Triplet_L | GAT | 97.78 | ± | 70.05 | ± | 95.94 | ± | 79.57 | ± | 111.125 |
| | | 1.90 | | 3.52 | | 1.57 | | 0.45 | | |
| PR_L + Triplet_L | GCN | 96.65 | ± | 79.76 | ± | 94.47 | ± | 80.83 | ± | 115.5 |
| | | 0.57 | | 1.81 | | 1.38 | | 1.41 | | |
| PR_L + Triplet_L | GIN | 81.21 | ± | 100.00 | ± | 86.42 | ± | 100.00 | ± | 93.25 |
| | | 5.96 | | 0.00 | | 11.46 | | 0.00 | | |
| PR_L + Triplet_L | MPNN | 88.27 | ± | 81.68 | ± | 80.16 | ± | 81.16 | ± | 152.5 |
| | | 10.72 | | 1.45 | | 6.42 | | 1.45 | | |
| PR_L + Triplet_L | PAGNN | 82.92 | ± | 99.92 | ± | 99.78 | ± | 99.76 | ± | 62.875 |
| | | 9.67 | | 0.02 | | 0.14 | | 0.23 | | |
| PR_L + Triplet_L | SAGE | 76.54 | ± | 91.69 | ± | 86.79 | ± | 92.05 | ± | 111.5 |
| | | 2.25 | | 2.36 | | 6.19 | | 1.10 | | |
| Triplet_L | ALL | 95.76 | ± | 59.16 | ± | 93.63 | ± | 64.52 | ± | 137.875 |
| | | 0.84 | | 1.58 | | 1.19 | | 0.77 | | |

Table 31. Results for Lp Recall (↑) (continued)

| Loss Type | Model | Cora ↓ Citeseer | | Cora ↓ Bitcoin | | Citeseer ↓ Cora | | Citeseer ↓ Bitcoin | | Average Rank |
|---|---|---|---|---|---|---|---|---|---|---|
| Triplet_L | GAT | 99.38 ± 0.10 | | 83.43 ± 1.59 | | 98.14 ± 0.15 | | 83.40 ± 1.68 | | 39.25 |
| Triplet_L | GCN | 98.60 0.26 | | 82.41 0.98 | | 96.78 0.56 | | 82.76 1.53 | | 61.375 |
| Triplet_L | GIN | 96.55 0.73 | | 54.43 0.80 | | 93.64 1.46 | | 57.07 0.62 | | 142.75 |
| Triplet_L | MPNN | 98.94 0.17 | | 80.87 1.03 | | 95.71 0.83 | | 80.21 2.32 | | 88.5 |
| Triplet_L | PAGNN | 91.09 1.12 | | 99.58 0.42 | | 90.05 1.10 | | 79.33 0.33 | | 112.25 |
| Triplet_L | SAGE | 97.24 0.46 | | 80.67 1.66 | | 95.65 0.73 | | 84.04 2.89 | | 85.0 |

Table 32. Lp Specificity Performance (↑): This table presents models (Loss function and GNN) ranked by their average performance in terms of lp specificity. Top-ranked results are highlighted in <span style="color:red">red</span>, second-ranked in <span style="color:blue">blue</span>, and third-ranked in <span style="color:green">green</span>.

| Loss Type | Model | Cora ↓ Citeseer | | Cora ↓ Bitcoin | | Citeseer ↓ Cora | | Citeseer ↓ Bitcoin | | Average Rank |
|---|---|---|---|---|---|---|---|---|---|---|
| Contr_l | ALL | 90.84 2.29 | | 66.99 6.39 | | 84.79 1.77 | | 27.70 38.12 | | 132.5 |
| Contr_l | GAT | 98.81 0.20 | | 91.87 0.71 | | 97.47 ± 0.36 | | 93.70 0.75 | | 30.25 |
| Contr_l | GCN | 98.26 0.19 | | 91.99 1.16 | | 96.52 0.58 | | 92.33 0.93 | | 52.0 |
| Contr_l | GIN | 93.45 1.14 | | 79.56 1.67 | | 88.24 2.37 | | 82.03 3.44 | | 112.25 |
| Contr_l | MPNN | 97.73 0.22 | | 90.78 0.40 | | 91.88 0.76 | | 90.97 1.08 | | 87.25 |

Table 32. Results for Lp Specificity (↑) (continued)

| Loss Type | Model | Cora ↓ Citeseer | | Cora ↓ Bitcoin | | Citeseer ↓ Cora | | Citeseer ↓ Bitcoin | | Average Rank |
|---|---|---|---|---|---|---|---|---|---|---|
| Contr_l | PAGNN | 78.55 | ± 4.06 | 0.89±0.50 | | 79.77 | ± 3.62 | 1.76±0.70 | | 160.25 |
| Contr_l | SAGE | 96.41 | ± 0.43 | 63.60 | ± 7.93 | 93.41 | ± 1.75 | 72.76 | ± 9.37 | 105.25 |
| Contr_l + CrossE_L | ALL | 90.80 | ± 1.01 | 70.22 | ± 3.13 | 82.59 | ± 3.74 | 42.94 | ± 29.34 | 132.25 |
| Contr_l + CrossE_L | GAT | 98.73 | ± 0.40 | 92.52 | ± 0.43 | 97.34 | ± 0.24 | 92.21 | ± 0.50 | 36.125 |
| Contr_l + CrossE_L | GCN | 98.75 | ± 0.20 | 92.58 | ± 1.07 | 96.15 | ± 0.31 | 93.42 | ± 0.53 | 28.25 |
| Contr_l + CrossE_L | GIN | 87.98 | ± 3.80 | 61.04 | ± 34.19 | 89.49 | ± 0.28 | 80.00 | ± 2.49 | 118.25 |
| Contr_l + CrossE_L | MPNN | 97.59 | ± 0.46 | 91.59 | ± 0.86 | 91.68 | ± 0.96 | 91.81 | ± 0.89 | 83.5 |
| Contr_l + CrossE_L | PAGNN | 77.38 | ± 3.34 | 1.37±0.43 | | 79.48 | ± 1.70 | 1.44±0.55 | | 163.0 |
| Contr_l + CrossE_L | SAGE | 96.37 | ± 0.65 | 54.30 | ± 2.90 | 94.27 | ± 1.39 | 52.18 | ± 5.51 | 111.5 |
| Contr_l + CrossE_L + PMI_L | ALL | 73.11 | ± 7.89 | 93.22 | ± 1.18 | 89.65 | ± 2.59 | 95.37 | ± 1.38 | 71.5 |
| Contr_l + CrossE_L + PMI_L | GAT | 98.93 | ± 0.08 | 92.28 | ± 1.21 | 97.38 | ± 0.37 | 92.65 | ± 0.98 | 32.0 |
| Contr_l + CrossE_L + PMI_L | GCN | 98.25 | ± 0.25 | 92.79 | ± 0.55 | 95.93 | ± 0.45 | 92.65 | ± 0.97 | 42.125 |
| Contr_l + CrossE_L + PMI_L | GIN | 80.00 | ± 3.80 | 0.01±0.01 | | 80.25 | ± 4.01 | 31.32 | ± 42.49 | 157.75 |
| Contr_l + CrossE_L + PMI_L | MPNN | 97.95 | ± 0.36 | 93.10 | ± 1.01 | 92.62 | ± 0.39 | 92.87 | ± 0.58 | 49.625 |
| Contr_l + CrossE_L + PMI_L | PAGNN | 41.22 | ± 5.33 | 0.09±0.20 | | 37.93 | ± 7.17 | 0.02±0.02 | | 192.625 |
| Contr_l + CrossE_L + PMI_L | SAGE | 78.11 | ± 2.00 | 30.82 | ± 6.70 | 71.55 | ± 3.95 | 42.32 | ± 1.20 | 151.0 |

Table 32. Results for Lp Specificity (↑) (continued)

| Loss Type | Model | Cora ↓ Citeseer | | Cora ↓ Bitcoin | | Citeseer ↓ Cora | | Citeseer ↓ Bitcoin | | Average Rank |
|---|---|---|---|---|---|---|---|---|---|---|
| Contr_l + CrossE_L + PMI_L + PR_L | ALL | 25.33 | ± 7.48 | 97.06 ± 0.47 | | 90.77 | ± 1.16 | 96.09 | ± 0.39 | 73.0 |
| Contr_l + CrossE_L + PMI_L + PR_L | GAT | 99.00 | ± 0.11 | 92.80 | ± 1.45 | 97.11 | ± 0.35 | 91.82 | ± 1.39 | 34.375 |
| Contr_l + CrossE_L + PMI_L + PR_L | GCN | 98.36 | ± 0.32 | 93.29 | ± 0.59 | 96.08 | ± 0.45 | 92.15 | ± 0.35 | 36.5 |
| Contr_l + CrossE_L + PMI_L + PR_L | GIN | 69.33 | ± 6.65 | 0.01±0.02 | | 62.07 | ± 8.78 | 0.68±0.39 | | 179.25 |
| Contr_l + CrossE_L + PMI_L + PR_L | MPNN | 98.00 | ± 0.24 | 92.52 | ± 0.53 | 88.00 | ± 2.71 | 90.59 | ± 1.58 | 82.625 |
| Contr_l + CrossE_L + PMI_L + PR_L | PAGNN | 46.91 | ± 5.75 | 0.47±0.29 | | 33.63 | ± 10.75 | 26.22 | ± 0.61 | 178.25 |
| Contr_l + CrossE_L + PMI_L + PR_L | SAGE | 77.50 | ± 1.67 | 20.58 | ± 3.48 | 66.29 | ± 2.86 | 20.47 | ± 7.76 | 161.25 |
| Contr_l + CrossE_L + PMI_L + PR_L + Triplet_L | ALL | 65.39 | ± 8.34 | 90.77 | ± 1.09 | 86.44 | ± 3.48 | 92.89 | ± 1.23 | 105.375 |
| Contr_l + CrossE_L + PMI_L + PR_L + Triplet_L | GAT | 98.89 | ± 0.18 | 92.82 | ± 1.25 | 96.81 | ± 0.61 | 92.05 | ± 1.05 | 35.375 |
| Contr_l + CrossE_L + PMI_L + PR_L + Triplet_L | GCN | 98.29 | ± 0.33 | 92.26 | ± 0.88 | 96.04 | ± 0.53 | 92.91 | ± 0.63 | 43.5 |
| Contr_l + CrossE_L + PMI_L + PR_L + Triplet_L | GIN | 74.02 | ± 5.09 | 29.88 | ± 40.89 | 79.08 | ± 3.06 | 0.15±0.10 | | 164.75 |
| Contr_l + CrossE_L + PMI_L + PR_L + Triplet_L | MPNN | 97.97 | ± 0.19 | 92.31 | ± 0.58 | 89.40 | ± 2.08 | 90.08 | ± 1.71 | 82.75 |
| Contr_l + CrossE_L + PMI_L + PR_L + Triplet_L | PAGNN | 43.74 | ± 4.63 | 0.55±0.33 | | 31.73 | ± 4.89 | 0.70±0.62 | | 186.5 |

Table 32. Results for Lp Specificity (↑) (continued)

| Loss Type | Model | Cora ↓ Citeseer | | Cora ↓ Bitcoin | | Citeseer ↓ Cora | | Citeseer ↓ Bitcoin | | Average Rank |
|---|---|---|---|---|---|---|---|---|---|---|
| Contr_l + CrossE_L + PMI_L + PR_L + Triplet_L | SAGE | 87.12 5.14 | ± | 64.08 3.41 | ± | 87.90 0.87 | ± | 61.63 4.57 | ± | 127.75 |
| Contr_l + CrossE_L + PMI_L + Triplet_L | ALL | 94.32 1.80 | ± | 93.66 0.75 | ± | 90.89 1.90 | ± | 92.87 0.53 | ± | 57.375 |
| Contr_l + CrossE_L + PMI_L + Triplet_L | GAT | 98.95 0.09 | ± | 92.27 0.78 | ± | 97.43 0.34 | ± | 92.39 1.22 | ± | 34.125 |
| Contr_l + CrossE_L + PMI_L + Triplet_L | GCN | 98.38 0.34 | ± | 93.30 0.55 | ± | 96.02 0.31 | ± | 92.67 0.58 | ± | 30.125 |
| Contr_l + CrossE_L + PMI_L + Triplet_L | GIN | 82.31 1.77 | ± | 61.11 34.20 | ± | 82.46 3.55 | ± | 76.21 2.11 | ± | 132.0 |
| Contr_l + CrossE_L + PMI_L + Triplet_L | MPNN | 98.13 0.15 | ± | 93.21 0.87 | ± | 92.08 0.84 | ± | 91.85 0.79 | ± | 57.25 |
| Contr_l + CrossE_L + PMI_L + Triplet_L | PAGNN | 44.85 2.07 | ± | 0.00±0.00 | ± | 37.71 6.93 | ± | 0.03±0.03 | ± | 194.5 |
| Contr_l + CrossE_L + PMI_L + Triplet_L | SAGE | 89.07 2.44 | ± | 52.26 6.55 | ± | 84.58 1.18 | ± | 62.66 1.40 | ± | 134.125 |
| Contr_l + CrossE_L + PR_L | ALL | 26.17 5.52 | ± | 92.30 2.21 | ± | 40.38 18.35 | ± | 95.41 0.18 | ± | 114.625 |
| Contr_l + CrossE_L + PR_L | GAT | 98.28 0.82 | ± | 92.48 0.95 | ± | 96.64 0.37 | ± | 92.44 1.61 | ± | 43.0 |
| Contr_l + CrossE_L + PR_L | GCN | 97.90 0.38 | ± | 91.00 0.80 | ± | 94.58 0.55 | ± | 91.66 1.30 | ± | 78.0 |
| Contr_l + CrossE_L + PR_L | GIN | 91.21 2.10 | ± | 0.00±0.00 | ± | 71.84 8.36 | ± | 0.01±0.02 | ± | 171.625 |
| Contr_l + CrossE_L + PR_L | MPNN | 96.53 1.40 | ± | 93.15 0.72 | ± | 89.93 1.88 | ± | 92.59 0.98 | ± | 62.5 |
| Contr_l + CrossE_L + PR_L | PAGNN | 56.49 5.33 | ± | 25.99 0.75 | ± | 4.90±8.33 | ± | 26.40 0.87 | ± | 174.625 |
| Contr_l + CrossE_L + PR_L | SAGE | 91.30 7.21 | ± | 56.43 4.94 | ± | 81.58 14.47 | ± | 48.39 7.28 | ± | 136.75 |
| Contr_l + CrossE_L + PR_L + Triplet_L | ALL | 70.58 4.99 | ± | 0.05±0.10 | ± | 85.32 1.93 | ± | 56.98 31.97 | ± | 156.5 |

Table 32. Results for Lp Specificity (↑) (continued)

| Loss Type | Model | Cora ↓ Citeseer | | Cora ↓ Bitcoin | | Citeseer ↓ Cora | | Citeseer ↓ Bitcoin | | Average Rank |
|---|---|---|---|---|---|---|---|---|---|---|
| Contr_l + CrossE_L + PR_L + Triplet_L | GAT | 98.80 0.23 | ± | 91.51 1.80 | ± | 97.11 0.64 | ± | 92.27 0.68 | ± | 46.625 |
| Contr_l + CrossE_L + PR_L + Triplet_L | GCN | 98.36 0.16 | ± | 92.84 1.01 | ± | 95.20 0.86 | ± | 92.64 0.60 | ± | 41.0 |
| Contr_l + CrossE_L + PR_L + Triplet_L | GIN | 88.99 2.83 | ± | 56.72 31.72 | ± | 89.09 0.95 | ± | 76.69 2.80 | ± | 122.5 |
| Contr_l + CrossE_L + PR_L + Triplet_L | MPNN | 97.66 0.43 | ± | 91.85 0.75 | ± | 88.22 1.72 | ± | 89.72 0.84 | ± | 95.375 |
| Contr_l + CrossE_L + PR_L + Triplet_L | PAGNN | 61.16 6.31 | ± | 0.56±0.22 | | 46.80 14.10 | ± | 22.39 10.57 | ± | 173.5 |
| Contr_l + CrossE_L + PR_L + Triplet_L | SAGE | 96.69 0.57 | ± | 65.59 3.78 | ± | 92.48 1.05 | ± | 51.62 8.27 | ± | 109.0 |
| Contr_l + CrossE_L + Triplet_L | ALL | 93.35 1.72 | ± | 74.31 1.98 | ± | 87.97 1.86 | ± | 70.64 2.73 | ± | 118.0 |
| Contr_l + CrossE_L + Triplet_L | GAT | 98.98 0.21 | ± | 91.80 1.01 | ± | 97.45 0.44 | ± | 92.58 0.97 | ± | 37.0 |
| Contr_l + CrossE_L + Triplet_L | GCN | 98.80 0.23 | ± | 92.57 1.20 | ± | 96.13 0.51 | ± | 91.74 1.23 | ± | 45.25 |
| Contr_l + CrossE_L + Triplet_L | GIN | 95.05 1.22 | ± | 81.39 1.36 | ± | 90.74 1.26 | ± | 83.94 1.29 | ± | 100.5 |
| Contr_l + CrossE_L + Triplet_L | MPNN | 97.74 0.42 | ± | 91.57 1.08 | ± | 92.57 1.23 | ± | 92.15 0.90 | ± | 77.25 |
| Contr_l + CrossE_L + Triplet_L | PAGNN | 72.19 13.99 | ± | 0.38±0.14 | | 84.38 2.03 | ± | 1.52±0.63 | | 165.75 |
| Contr_l + CrossE_L + Triplet_L | SAGE | 97.12 0.65 | ± | 65.77 5.87 | ± | 94.58 0.82 | ± | 75.32 4.45 | ± | 99.0 |
| Contr_l + PMI_L | ALL | 77.92 14.29 | ± | 30.97 42.47 | ± | 87.55 2.27 | ± | 92.04 1.12 | ± | 121.5 |
| Contr_l + PMI_L | GAT | **99.05 0.16** | ± | 92.94 0.78 | ± | 96.87 0.26 | ± | 92.88 1.14 | ± | 19.75 |
| Contr_l + PMI_L | GCN | 98.51 0.40 | ± | 92.00 1.03 | ± | 96.04 0.25 | ± | 92.69 0.58 | ± | 45.5 |

Continued on next page

Table 32. Results for Lp Specificity (↑) (continued)

| Loss Type | Model | Cora ↓ Citeseer | | Cora ↓ Bitcoin | | Citeseer ↓ Cora | | Citeseer ↓ Bitcoin | | Average Rank |
|---|---|---|---|---|---|---|---|---|---|---|
| Contr_l + PMI_L | GIN | 81.18 | ± 5.18 | 79.90 | ± 2.10 | 77.31 | ± 1.92 | 61.25 | ± 34.26 | 134.0 |
| Contr_l + PMI_L | MPNN | 98.08 | ± 0.33 | 92.73 | ± 1.72 | 92.56 | ± 0.77 | 91.60 | ± 1.20 | 66.125 |
| Contr_l + PMI_L | PAGNN | 42.06 | ± 4.70 | 0.00±0.00 | | 32.16 | ± 2.63 | 0.02±0.01 | | 198.5 |
| Contr_l + PMI_L | SAGE | 79.04 | ± 2.36 | 37.00 | ± 7.26 | 76.74 | ± 4.57 | 45.50 | ± 1.74 | 146.75 |
| Contr_l + PMI_L + PR_L | ALL | 21.59 | ± 3.54 | 94.09 | ± 0.51 | 89.16 | ± 1.40 | 95.75 | ± 1.37 | 81.25 |
| Contr_l + PMI_L + PR_L | GAT | 98.86 | ± 0.30 | 92.38 | ± 1.03 | 94.32 | ± 3.95 | 86.58 | ± 1.82 | 60.25 |
| Contr_l + PMI_L + PR_L | GCN | 98.38 | ± 0.49 | 93.08 | ± 1.41 | 95.96 | ± 0.74 | 92.27 | ± 0.81 | 39.5 |
| Contr_l + PMI_L + PR_L | GIN | 77.02 | ± 6.37 | 0.00±0.00 | | 70.61 | ± 9.83 | 0.57±0.28 | | 174.875 |
| Contr_l + PMI_L + PR_L | MPNN | 98.10 | ± 0.23 | 93.12 | ± 0.81 | 85.23 | ± 5.37 | 89.48 | ± 0.55 | 79.25 |
| Contr_l + PMI_L + PR_L | PAGNN | 41.68 | ± 4.52 | 0.00±0.00 | | 30.11 | ± 2.15 | 0.02±0.01 | | 199.5 |
| Contr_l + PMI_L + PR_L | SAGE | 77.59 | ± 1.81 | 33.36 | ± 7.78 | 75.15 | ± 5.31 | 35.27 | ± 4.27 | 150.5 |
| Contr_l + PMI_L + PR_L + Triplet_L | ALL | 74.99 | ± 6.00 | 0.00±0.00 | | 85.99 | ± 1.79 | 82.99 | ± 3.72 | 149.625 |
| Contr_l + PMI_L + PR_L + Triplet_L | GAT | 98.69 | ± 0.37 | 91.81 | ± 0.63 | 95.82 | ± 1.17 | 88.73 | ± 2.41 | 66.125 |
| Contr_l + PMI_L + PR_L + Triplet_L | GCN | 98.46 | ± 0.43 | 92.56 | ± 1.27 | 96.00 | ± 0.54 | 91.84 | ± 0.69 | 49.125 |
| Contr_l + PMI_L + PR_L + Triplet_L | GIN | 81.20 | ± 2.08 | 80.29 | ± 2.06 | 85.89 | ± 3.38 | 30.23 | ± 41.28 | 130.75 |
| Contr_l + PMI_L + PR_L + Triplet_L | MPNN | 98.21 | ± 0.34 | 92.21 | ± 1.27 | 87.36 | ± 2.84 | 89.00 | ± 2.01 | 87.75 |

Table 32. Results for Lp Specificity (↑) (continued)

| Loss Type | Model | Cora ↓ Citeseer | Cora ↓ Bitcoin | Citeseer ↓ Cora | Citeseer ↓ Bitcoin | Average Rank |
|---|---|---|---|---|---|---|
| Contr_l + PMI_L + PR_L + Triplet_L | PAGNN | 46.56 ± 3.80 | 0.01±0.01 | 33.56 ± 8.78 | 0.24±0.49 | 190.25 |
| Contr_l + PMI_L + PR_L + Triplet_L | SAGE | 93.67 ± 1.66 | 55.39 ± 1.71 | 89.58 ± 3.18 | 62.25 ± 8.26 | 120.0 |
| Contr_l + PR_L | ALL | 30.92 ± 12.32 | 95.83 ± 0.33 | 55.91 ± 20.38 | 96.20 ± 0.91 | 96.25 |
| Contr_l + PR_L | GAT | 98.30 ± 0.49 | 91.82 ± 0.84 | 96.73 ± 0.26 | 91.94 ± 1.40 | 56.75 |
| Contr_l + PR_L | GCN | 97.82 ± 0.36 | 91.89 ± 0.93 | 94.72 ± 0.84 | 92.42 ± 0.81 | 66.25 |
| Contr_l + PR_L | GIN | 89.70 ± 1.77 | 24.91 ± 13.99 | 57.38 ± 33.92 | 27.99 ± 38.20 | 153.0 |
| Contr_l + PR_L | MPNN | 96.61 ± 1.19 | 92.07 ± 0.67 | 89.84 ± 1.37 | 92.26 ± 0.66 | 80.875 |
| Contr_l + PR_L | PAGNN | 46.46 ± 25.75 | 16.73 ± 13.25 | 2.63±2.05 | 26.31 ± 0.69 | 179.75 |
| Contr_l + PR_L | SAGE | 92.21 ± 6.06 | 59.07 ± 5.46 | 86.36 ± 8.81 | 22.07 ± 1.44 | 134.0 |
| Contr_l + PR_L + Triplet_L | ALL | 58.70 ± 23.81 | 29.98 ± 41.03 | 85.16 ± 3.20 | 0.02±0.02 | 165.875 |
| Contr_l + PR_L + Triplet_L | GAT | 98.56 ± 0.25 | 92.01 ± 0.78 | 96.87 ± 0.40 | 91.06 ± 1.23 | 54.75 |
| Contr_l + PR_L + Triplet_L | GCN | 98.22 ± 0.30 | 92.67 ± 1.49 | 95.95 ± 0.60 | 92.02 ± 0.64 | 53.125 |
| Contr_l + PR_L + Triplet_L | GIN | 91.96 ± 2.86 | 75.27 ± 4.79 | 83.28 ± 5.51 | 77.84 ± 2.20 | 122.75 |
| Contr_l + PR_L + Triplet_L | MPNN | 97.12 ± 0.30 | 91.91 ± 1.42 | 88.74 ± 2.30 | 90.22 ± 1.32 | 94.875 |
| Contr_l + PR_L + Triplet_L | PAGNN | 57.10 ± 7.05 | 0.13±0.14 | 53.07 ± 21.46 | 2.43±0.76 | 179.0 |
| Contr_l + PR_L + Triplet_L | SAGE | 96.68 ± 0.90 | 59.68 ± 5.84 | 93.47 ± 1.41 | 61.96 ± 3.69 | 107.25 |

Table 32. Results for Lp Specificity (↑) (continued)

| Loss Type | Model | Cora ↓ Citeseer | | Cora ↓ Bitcoin | | Citeseer ↓ Cora | | Citeseer ↓ Bitcoin | | Average Rank |
|---|---|---|---|---|---|---|---|---|---|---|
| Contr_l + Triplet_L | ALL | 94.42 | ± 1.34 | 69.10 | ± 3.93 | 89.07 | ± 1.59 | 72.49 | ± 2.10 | 114.75 |
| Contr_l + Triplet_L | GAT | 98.83 | ± 0.21 | 92.07 | ± 1.06 | 97.61 | ± 0.22 | 92.05 | ± 1.78 | 42.0 |
| Contr_l + Triplet_L | GCN | 98.62 | ± 0.19 | 92.35 | ± 0.93 | 96.52 | ± 0.74 | 93.40 | ± 0.91 | 31.5 |
| Contr_l + Triplet_L | GIN | 94.76 | ± 0.65 | 80.64 | ± 1.77 | 89.98 | ± 2.44 | 82.63 | ± 1.99 | 103.75 |
| Contr_l + Triplet_L | MPNN | 98.03 | ± 0.16 | 90.76 | ± 2.01 | 93.58 | ± 0.47 | 91.77 | ± 1.35 | 78.5 |
| Contr_l + Triplet_L | PAGNN | 74.41 | ± 12.25 | 1.99±0.64 | | 84.58 | ± 1.48 | 7.49 | ± 13.40 | 159.625 |
| Contr_l + Triplet_L | SAGE | 97.55 | ± 0.32 | 66.30 | ± 4.83 | 94.95 | ± 0.59 | 68.87 | ± 4.35 | 97.25 |
| CrossE_L | ALL | 77.05 | ± 38.04 | 88.62 | ± 1.27 | 12.90 | ± 18.82 | 91.92 | ± 0.43 | 133.0 |
| CrossE_L | GAT | 75.99 | ± 42.50 | 15.90 | ± 3.08 | 65.04 | ± 37.63 | 89.37 | ± 1.85 | 148.75 |
| CrossE_L | GCN | 16.39 | ± 36.66 | 85.17 | ± 1.57 | 0.00±0.00 | | 0.00±0.00 | | 182.75 |
| CrossE_L | GIN | 0.03±0.04 | | 0.00±0.00 | | 0.00±0.01 | | 0.00±0.00 | | 208.125 |
| CrossE_L | MPNN | 87.42 | ± 2.63 | 0.01±0.03 | | 85.53 | ± 1.69 | 0.00±0.00 | | 165.125 |
| CrossE_L | PAGNN | 60.38 | ± 6.98 | 0.00±0.00 | | 27.02 | ± 35.93 | 0.01±0.01 | | 195.875 |
| CrossE_L | SAGE | 63.43 | ± 15.53 | 21.74 | ± 0.97 | 26.63 | ± 19.68 | 1.48±0.97 | | 178.75 |
| CrossE_L + PMI_L | ALL | 69.68 | ± 4.47 | 95.81 | ± 0.82 | 89.85 | ± 0.83 | 96.17 | ± 0.65 | 67.5 |
| CrossE_L + PMI_L | GAT | 98.94 | ± 0.22 | 93.10 | ± 0.64 | 97.39 | ± 0.21 | 93.13 | ± 0.78 | 16.625 |
| CrossE_L + PMI_L | GCN | 98.23 | ± 0.29 | 92.22 | ± 1.47 | 95.99 | ± 0.61 | 91.62 | ± 1.13 | 61.5 |

Table 32. Results for Lp Specificity (↑) (continued)

| Loss Type | Model | Cora ↓ Citeseer | | Cora ↓ Bitcoin | | Citeseer ↓ Cora | | Citeseer ↓ Bitcoin | | Average Rank |
|---|---|---|---|---|---|---|---|---|---|---|
| CrossE_L + PMI_L | GIN | 77.60 | ± 2.63 | 14.18 | ± 31.71 | 72.25 | ± 7.37 | 29.09 | ± 39.46 | 157.5 |
| CrossE_L + PMI_L | MPNN | 97.82 | ± 0.21 | 93.12 | ± 1.21 | 92.09 | ± 1.16 | 92.15 | ± 1.35 | 59.25 |
| CrossE_L + PMI_L | PAGNN | 40.60 | ± 8.99 | 24.13 | ± 0.74 | 34.72 | ± 2.40 | 0.03±0.01 | | 186.375 |
| CrossE_L + PMI_L | SAGE | 75.16 | ± 3.38 | 27.10 | ± 6.43 | 63.53 | ± 3.52 | 14.26 | ± 13.49 | 162.75 |
| CrossE_L + PMI_L + PR_L | ALL | 17.16 | ± 1.23 | 96.26 | ± 0.47 | 89.24 | ± 0.98 | 95.49 | ± 0.97 | 80.5 |
| CrossE_L + PMI_L + PR_L | GAT | 99.03 | ± 0.13 | 93.15 | ± 0.57 | 93.64 | ± 3.44 | 88.70 | ± 2.24 | 48.625 |
| CrossE_L + PMI_L + PR_L | GCN | 98.36 | ± 0.34 | 91.85 | ± 1.00 | 95.76 | ± 0.28 | 92.76 | ± 1.06 | 51.125 |
| CrossE_L + PMI_L + PR_L | GIN | 74.62 | ± 4.30 | 80.74 | ± 1.24 | 74.19 | ± 4.56 | 15.04 | ± 33.48 | 148.25 |
| CrossE_L + PMI_L + PR_L | MPNN | 98.07 | ± 0.39 | 93.02 | ± 0.98 | 86.32 | ± 7.49 | 92.33 | ± 0.71 | 67.875 |
| CrossE_L + PMI_L + PR_L | PAGNN | 44.87 | ± 6.02 | 0.32±0.31 | | 30.25 | ± 8.72 | 0.03±0.02 | | 191.375 |
| CrossE_L + PMI_L + PR_L | SAGE | 74.33 | ± 2.82 | 23.03 | ± 2.80 | 66.35 | ± 2.90 | 26.84 | ± 2.87 | 161.25 |
| CrossE_L + PMI_L + PR_L + Triplet_L | ALL | 65.05 | ± 7.95 | 95.78 | ± 0.39 | 88.66 | ± 1.80 | 91.17 | ± 0.66 | 95.5 |
| CrossE_L + PMI_L + PR_L + Triplet_L | GAT | 98.94 | ± 0.15 | 93.28 | ± 0.72 | 96.82 | ± 0.92 | 92.13 | ± 0.24 | 27.625 |
| CrossE_L + PMI_L + PR_L + Triplet_L | GCN | 98.43 | ± 0.18 | 92.08 | ± 1.47 | 95.62 | ± 0.72 | 92.41 | ± 1.21 | 52.0 |
| CrossE_L + PMI_L + PR_L + Triplet_L | GIN | 78.53 | ± 4.42 | 39.33 | ± 35.93 | 79.43 | ± 4.62 | 15.41 | ± 34.17 | 152.0 |
| CrossE_L + PMI_L + PR_L + Triplet_L | MPNN | 98.12 | ± 0.23 | 93.12 | ± 1.04 | 88.46 | ± 3.08 | 91.70 | ± 1.20 | 69.25 |

Table 32. Results for Lp Specificity (↑) (continued)

| Loss Type | Model | Cora ↓ Citeseer | | Cora ↓ Bitcoin | | Citeseer ↓ Cora | | Citeseer ↓ Bitcoin | | Average Rank |
|---|---|---|---|---|---|---|---|---|---|---|
| CrossE_L + PMI_L + PR_L + Triplet_L | PAGNN | 50.05 | ± 3.50 | 25.99 | ± 0.19 | 32.47 | ± 4.09 | 0.23±0.48 | | 180.5 |
| CrossE_L + PMI_L + PR_L + Triplet_L | SAGE | 86.96 | ± 5.49 | 57.46 | ± 5.29 | 84.30 | ± 2.32 | 60.30 | ± 3.61 | 136.5 |
| CrossE_L + PMI_L + Triplet_L | ALL | 96.30 | ± 0.24 | 94.04 | ± 0.96 | 90.53 | ± 1.79 | 93.01 | ± 0.65 | 55.5 |
| CrossE_L + PMI_L + Triplet_L | GAT | 99.04 | ± 0.26 | 92.78 | ± 1.00 | 97.21 | ± 0.34 | 92.46 | ± 0.68 | 25.125 |
| CrossE_L + PMI_L + Triplet_L | GCN | 98.34 | ± 0.23 | 92.95 | ± 0.71 | 96.00 | ± 0.46 | 92.55 | ± 1.03 | 37.875 |
| CrossE_L + PMI_L + Triplet_L | GIN | 80.41 | ± 2.51 | 78.36 | ± 1.85 | 84.40 | ± 1.29 | 79.28 | ± 1.19 | 125.75 |
| CrossE_L + PMI_L + Triplet_L | MPNN | 97.82 | ± 0.39 | 92.57 | ± 0.86 | 93.06 | ± 1.05 | 92.52 | ± 0.48 | 59.25 |
| CrossE_L + PMI_L + Triplet_L | PAGNN | 46.35 | ± 4.48 | 0.45±0.20 | | 39.23 | ± 8.48 | 0.43±0.56 | | 184.625 |
| CrossE_L + PMI_L + Triplet_L | SAGE | 92.70 | ± 1.18 | 62.45 | ± 2.16 | 88.72 | ± 1.87 | 67.87 | ± 3.17 | 120.25 |
| CrossE_L + PR_L | ALL | 25.43 | ± 12.33 | 96.78 | ± 0.65 | 44.34 | ± 24.61 | 96.97 | ± 0.68 | 96.75 |
| CrossE_L + PR_L | GAT | 98.45 | ± 0.19 | 91.83 | ± 1.81 | 96.40 | ± 0.51 | 92.10 | ± 1.16 | 53.75 |
| CrossE_L + PR_L | GCN | 96.62 | ± 1.31 | 92.35 | ± 1.23 | 94.57 | ± 0.56 | 91.53 | ± 1.57 | 75.375 |
| CrossE_L + PR_L | GIN | 89.39 | ± 1.70 | 25.09 | ± 14.03 | 35.61 | ± 40.90 | 6.14 | ± 13.70 | 161.0 |
| CrossE_L + PR_L | MPNN | 97.38 | ± 0.65 | 92.50 | ± 0.62 | 89.52 | ± 2.13 | 93.23 | ± 1.11 | 63.375 |
| CrossE_L + PR_L | PAGNN | 44.72 | ± 24.79 | 0.00±0.00 | | 0.49±0.25 | | 0.00±0.01 | | 203.0 |
| CrossE_L + PR_L | SAGE | 73.81 | ± 4.30 | 20.07 | ± 2.77 | 60.04 | ± 4.12 | 20.92 | ± 2.08 | 167.0 |

Continued on next page

Table 32. Results for Lp Specificity (↑) (continued)

| Loss Type | Model | Cora ↓ Citeseer | | Cora ↓ Bitcoin | | Citeseer ↓ Cora | | Citeseer ↓ Bitcoin | | Average Rank |
|---|---|---|---|---|---|---|---|---|---|---|
| CrossE_L + PR_L + Triplet_L | ALL | 51.61 ± 27.14 | | 0.00±0.00 | | 80.56 ± 3.76 | | 0.01±0.01 | | 184.125 |
| CrossE_L + PR_L + Triplet_L | GAT | 98.31 ± 0.52 | | 91.79 ± 0.68 | | 96.68 ± 0.98 | | 92.80 ± 0.92 | | 46.5 |
| CrossE_L + PR_L + Triplet_L | GCN | 97.99 ± 0.17 | | 92.33 ± 0.77 | | 94.97 ± 0.48 | | 92.96 ± 0.78 | | 51.625 |
| CrossE_L + PR_L + Triplet_L | GIN | 92.52 ± 2.53 | | 39.64 ± 36.34 | | 83.74 ± 4.18 | | 75.76 ± 1.59 | | 130.0 |
| CrossE_L + PR_L + Triplet_L | MPNN | 97.13 ± 0.54 | | 92.57 ± 0.58 | | 86.93 ± 2.34 | | 92.25 ± 0.46 | | 78.625 |
| CrossE_L + PR_L + Triplet_L | PAGNN | 63.33 ± 4.24 | | 26.23 ± 0.62 | | 21.54 ± 13.00 | | 12.45 ± 12.61 | | 175.75 |
| CrossE_L + PR_L + Triplet_L | SAGE | 97.13 ± 0.65 | | 60.36 ± 6.30 | | 94.25 ± 1.33 | | 59.81 ± 4.14 | | 105.625 |
| CrossE_L + Triplet_L | ALL | 96.53 ± 0.19 | | 80.60 ± 2.05 | | 90.70 ± 1.10 | | 79.64 ± 2.24 | | 101.375 |
| CrossE_L + Triplet_L | GAT | 98.92 ± 0.13 | | 93.16 ± 0.78 | | 97.14 ± 0.37 | | 93.08 ± 1.41 | | 17.0 |
| CrossE_L + Triplet_L | GCN | 98.64 ± 0.28 | | 92.32 ± 0.99 | | 96.42 ± 0.27 | | 92.10 ± 0.93 | | 45.5 |
| CrossE_L + Triplet_L | GIN | 95.74 ± 1.04 | | 81.76 ± 1.56 | | 90.56 ± 1.03 | | 83.13 ± 1.63 | | 100.75 |
| CrossE_L + Triplet_L | MPNN | 98.22 ± 0.23 | | 93.13 ± 0.80 | | 93.61 ± 0.81 | | 92.46 ± 1.58 | | 46.5 |
| CrossE_L + Triplet_L | PAGNN | 78.22 ± 11.43 | | 0.01±0.02 | | 84.52 ± 2.21 | | 0.08±0.04 | | 166.75 |
| CrossE_L + Triplet_L | SAGE | 98.23 ± 0.29 | | 72.01 ± 3.55 | | 95.46 ± 0.55 | | 70.77 ± 2.28 | | 86.5 |
| PMI_L | ALL | 61.47 ± 4.05 | | 95.45 ± 0.64 | | 90.15 ± 1.57 | | 95.31 ± 0.55 | | 70.75 |
| PMI_L | GAT | 99.02 ± 0.14 | | 92.80 ± 0.91 | | 97.12 ± 0.72 | | 93.11 ± 0.83 | | 19.375 |

Table 32. Results for Lp Specificity (↑) (continued)

| Loss Type | Model | Cora ↓ Citeseer | Cora ↓ Bitcoin | Citeseer ↓ Cora | Citeseer ↓ Bitcoin | Average Rank |
|---|---|---|---|---|---|---|
| PMI_L | GCN | 98.27 ± 0.30 | 92.13 ± 0.51 | 95.37 ± 0.38 | 91.72 ± 0.78 | 63.25 |
| PMI_L | GIN | 75.39 ± 4.57 | 0.04±0.07 | 74.32 ± 4.84 | 0.22±0.26 | 173.25 |
| PMI_L | MPNN | 98.18 ± 0.24 | 92.30 ± 1.29 | 92.88 ± 0.58 | 92.78 ± 0.60 | 55.625 |
| PMI_L | PAGNN | 44.40 ± 5.80 | 0.00±0.00 | 42.28 ± 9.60 | 0.43±0.57 | 191.0 |
| PMI_L | SAGE | 73.91 ± 4.08 | 28.39 ± 5.81 | 61.42 ± 3.43 | 19.84 ± 4.62 | 164.0 |
| PMI_L + PR_L | ALL | 19.91 ± 6.02 | 97.30 ± 0.48 | 89.32 ± 1.35 | 95.76 ± 1.00 | 78.5 |
| PMI_L + PR_L | GAT | 98.76 ± 0.25 | 92.72 ± 0.82 | 89.00 ± 10.35 | 90.90 ± 1.73 | 66.5 |
| PMI_L + PR_L | GCN | 98.30 ± 0.42 | 92.14 ± 1.20 | 95.76 ± 0.68 | 91.94 ± 1.57 | 58.875 |
| PMI_L + PR_L | GIN | 72.42 ± 7.76 | 0.02±0.03 | 70.00 ± 21.23 | 48.97 ± 44.60 | 166.125 |
| PMI_L + PR_L | MPNN | 97.90 ± 0.41 | 93.26 ± 1.26 | 82.40 ± 3.51 | 90.51 ± 0.58 | 82.375 |
| PMI_L + PR_L | PAGNN | 45.04 ± 4.93 | 0.08±0.17 | 28.41 ± 3.67 | 0.25±0.48 | 190.25 |
| PMI_L + PR_L | SAGE | 76.32 ± 1.73 | 19.84 ± 9.57 | 62.24 ± 1.99 | 22.90 ± 7.02 | 162.75 |
| PMI_L + PR_L + Triplet_L | ALL | 68.90 ± 4.69 | 68.43 ± 38.52 | 88.45 ± 0.87 | 67.17 ± 37.55 | 134.0 |
| PMI_L + PR_L + Triplet_L | GAT | 99.00 ± 0.27 | 92.75 ± 0.77 | 96.87 ± 0.86 | 91.54 ± 1.47 | 38.375 |
| PMI_L + PR_L + Triplet_L | GCN | 98.37 ± 0.47 | 92.29 ± 1.26 | 95.81 ± 0.71 | 92.24 ± 0.91 | 51.25 |
| PMI_L + PR_L + Triplet_L | GIN | 77.61 ± 5.95 | 58.48 ± 32.77 | 78.94 ± 5.68 | 0.36±0.15 | 155.5 |

Table 32. Results for Lp Specificity (↑) (continued)

| Loss Type | Model | Cora ↓ Citeseer | Cora ↓ Bitcoin | Citeseer ↓ Cora | Citeseer ↓ Bitcoin | Average Rank |
|---|---|---|---|---|---|---|
| PMI_L + PR_L + Triplet_L | MPNN | 98.06 ± 0.11 | 92.41 ± 1.13 | 86.62 ± 1.76 | 90.72 ± 1.38 | 84.5 |
| PMI_L + PR_L + Triplet_L | PAGNN | 47.56 ± 5.51 | 0.50±0.14 | 33.66 ± 7.39 | 0.49±0.67 | 183.75 |
| PMI_L + PR_L + Triplet_L | SAGE | 91.23 ± 2.52 | 64.28 ± 3.50 | 87.57 ± 1.70 | 61.02 ± 3.73 | 125.0 |
| PMI_L + Triplet_L | ALL | 95.26 ± 0.89 | 85.32 ± 2.60 | 90.96 ± 1.27 | 94.18 ± 0.84 | 74.75 |
| PMI_L + Triplet_L | GAT | 98.95 ± 0.22 | 92.10 ± 0.76 | 97.23 ± 0.32 | 92.45 ± 0.25 | 35.875 |
| PMI_L + Triplet_L | GCN | 98.29 ± 0.18 | 92.28 ± 1.34 | 95.92 ± 0.49 | 92.20 ± 0.72 | 53.75 |
| PMI_L + Triplet_L | GIN | 86.76 ± 3.13 | 57.00 ± 31.91 | 82.23 ± 2.71 | 79.19 ± 1.43 | 133.25 |
| PMI_L + Triplet_L | MPNN | 97.80 ± 0.47 | 92.71 ± 1.19 | 92.75 ± 1.09 | 93.47 ± 0.78 | 51.375 |
| PMI_L + Triplet_L | PAGNN | 43.45 ± 6.99 | 0.02±0.03 | 44.92 ± 10.45 | 0.03±0.02 | 191.0 |
| PMI_L + Triplet_L | SAGE | 90.62 ± 2.46 | 46.08 ± 8.32 | 88.99 ± 1.06 | 66.26 ± 2.04 | 126.0 |
| PR_L | ALL | 8.26±4.91 | 96.48 ± 0.73 | 17.30 ± 13.10 | 96.64 ± 0.36 | 104.25 |
| PR_L | GAT | 97.99 ± 0.26 | 90.77 ± 0.59 | 96.07 ± 0.74 | 92.11 ± 1.44 | 66.5 |
| PR_L | GCN | 97.32 ± 0.49 | 91.65 ± 0.92 | 94.84 ± 0.76 | 91.50 ± 1.81 | 81.25 |
| PR_L | GIN | 90.26 ± 2.77 | 6.39 ± 14.28 | 65.49 ± 36.27 | 18.95 ± 17.31 | 157.0 |
| PR_L | MPNN | 96.84 ± 1.27 | 92.71 ± 1.01 | 89.96 ± 0.77 | 92.69 ± 0.69 | 65.25 |
| PR_L | PAGNN | 34.03 ± 30.39 | 0.03±0.07 | 1.07±0.35 | 0.23±0.01 | 196.875 |

Table 32. Results for Lp Specificity (↑) (continued)

| Loss Type | Model | Cora ↓ Citeseer | | Cora ↓ Bitcoin | | Citeseer ↓ Cora | | Citeseer ↓ Bitcoin | | Average Rank |
|---|---|---|---|---|---|---|---|---|---|---|
| PR_L | SAGE | 74.37 | ± 3.59 | 23.07 | ± 2.67 | 63.68 | ± 5.54 | 22.70 | ± 3.30 | 163.0 |
| PR_L + Triplet_L | ALL | 23.29 | ± 9.71 | 72.43 | ± 40.55 | 22.98 | ± 7.38 | 95.79 | ± 0.71 | 132.5 |
| PR_L + Triplet_L | GAT | 98.23 | ± 0.74 | 92.12 | ± 1.12 | 96.24 | ± 0.82 | 91.56 | ± 0.68 | 60.25 |
| PR_L + Triplet_L | GCN | 97.38 | ± 0.65 | 92.06 | ± 1.42 | 95.18 | ± 0.53 | 92.10 | ± 0.71 | 70.625 |
| PR_L + Triplet_L | GIN | 90.09 | ± 2.11 | 0.00±0.00 | | 48.91 | ± 37.11 | 0.04±0.03 | | 174.625 |
| PR_L + Triplet_L | MPNN | 97.52 | ± 0.86 | 93.10 | ± 1.39 | 89.77 | ± 1.91 | 93.15 | ± 0.66 | 55.75 |
| PR_L + Triplet_L | PAGNN | 46.59 | ± 24.77 | 0.27±0.08 | | 1.38±0.60 | | 0.56±0.42 | | 189.5 |
| PR_L + Triplet_L | SAGE | 80.14 | ± 2.04 | 24.01 | ± 1.89 | 76.72 | ± 13.17 | 20.32 | ± 2.37 | 155.0 |
| Triplet_L | ALL | 96.38 | ± 0.77 | 77.02 | ± 3.24 | 90.28 | ± 0.94 | 93.34 | ± 1.37 | 79.875 |
| Triplet_L | GAT | 99.11 ± 0.22 | | 92.34 | ± 0.60 | 97.49 ± 0.67 | | 92.76 | ± 0.86 | 24.375 |
| Triplet_L | GCN | 98.69 | ± 0.09 | 92.90 | ± 0.99 | 96.13 | ± 0.79 | 92.74 | ± 0.98 | 30.75 |
| Triplet_L | GIN | 96.66 | ± 0.72 | 79.82 | ± 1.46 | 89.13 | ± 1.73 | 83.21 | ± 0.85 | 104.5 |
| Triplet_L | MPNN | 98.08 | ± 0.52 | 92.87 | ± 1.03 | 93.59 | ± 0.70 | 93.34 | ± 1.27 | 44.5 |
| Triplet_L | PAGNN | 78.92 | ± 11.27 | 1.05±0.87 | | 86.08 | ± 1.95 | 32.95 | ± 0.39 | 147.5 |
| Triplet_L | SAGE | 98.11 | ± 0.41 | 76.22 | ± 2.67 | 95.13 | ± 0.32 | 75.53 | ± 2.90 | 87.5 |

### 1.2.3    Embedding–Adjacency Alignment.

Table 33. Cosine-Adj Corr Performance (↑): This table presents models (Loss function and GNN) ranked by their average performance in terms of cosine-adj corr. Top-ranked results are highlighted in red, second-ranked in blue, and third-ranked in green.

| Loss Type | Model | Cora ↓ Citeseer | Cora ↓ Bitcoin | Citeseer ↓ Cora | Citeseer ↓ Bitcoin | Average Rank |
|---|---|---|---|---|---|---|
| Contr_l | ALL | 3.68±0.32 | −0.26 ± 0.07 | 4.19±0.85 | −0.69 ± 0.24 | 148.25 |
| Contr_l | GAT | 9.11±1.10 | 4.09±0.47 | 9.64±0.57 | 4.64±0.13 | 24.25 |
| Contr_l | GCN | 8.43±0.67 | 4.46±0.27 | 9.32±0.78 | 4.60±0.28 | 27.75 |
| Contr_l | GIN | 5.87±0.35 | −0.63 ± 0.10 | 5.51±1.03 | 0.87±0.10 | 111.25 |
| Contr_l | MPNN | 6.83±0.53 | 3.51±0.31 | 6.36±1.15 | 3.60±0.22 | 67.125 |
| Contr_l | PAGNN | 3.57±0.18 | −0.03 ± 0.03 | 3.31±0.68 | 0.04±0.03 | 140.75 |
| Contr_l | SAGE | 5.74±0.81 | 1.97±0.17 | 7.65±0.83 | 2.29±0.24 | 74.75 |
| Contr_l + CrossE_L | ALL | 3.97±0.50 | 1.33±0.11 | 3.82±0.78 | 0.43±0.27 | 123.0 |
| Contr_l + CrossE_L | GAT | 9.10±0.66 | 4.12±0.16 | 9.81±0.66 | 4.26±0.25 | 29.25 |
| Contr_l + CrossE_L | GCN | 8.52±0.42 | 4.44±0.35 | 9.19±0.78 | 4.57±0.26 | 29.375 |
| Contr_l + CrossE_L | GIN | 4.76±0.29 | −1.65 ± 0.20 | 5.16±0.53 | −0.33 ± 0.06 | 135.5 |
| Contr_l + CrossE_L | MPNN | 6.88±0.79 | 3.56±0.27 | 6.46±0.57 | 3.75±0.27 | 65.125 |
| Contr_l + CrossE_L | PAGNN | 3.48±0.17 | −0.00 ± 0.01 | 3.03±0.23 | −0.03 ± 0.00 | 143.625 |
| Contr_l + CrossE_L | SAGE | 6.26±0.45 | 1.68±0.08 | 6.93±0.34 | 1.59±0.09 | 85.0 |
| Contr_l + CrossE_L + PMI_L | ALL | 3.05±0.31 | −1.81 ± 0.28 | 3.19±0.61 | −1.94 ± 0.19 | 171.5 |
| Contr_l + CrossE_L + PMI_L | GAT | 10.86 ± 0.42 | 4.68±0.24 | 10.32 ± 0.77 | 4.76±0.30 | 8.5 |
| Contr_l + CrossE_L + PMI_L | GCN | 8.98±0.46 | 4.75±0.12 | 8.68±0.74 | 4.58±0.15 | 23.25 |
| Contr_l + CrossE_L + PMI_L | GIN | 3.92±0.48 | −1.37 ± 0.19 | 3.30±0.27 | −1.13 ± 0.09 | 157.5 |
| Contr_l + CrossE_L + PMI_L | MPNN | 8.78±0.55 | 4.32±0.30 | 6.15±0.46 | 4.25±0.20 | 45.625 |

Continued on next page

Table 33. Results for Cosine-Adj Corr (↑) (continued)

| Loss Type | Model | Cora ↓ Citeseer | Cora ↓ Bitcoin | Citeseer ↓ Cora | Citeseer ↓ Bitcoin | Average Rank |
|---|---|---|---|---|---|---|
| Contr_l + CrossE_L + PMI_L | PAGNN | 1.92±0.08 | −0.05 ± 0.01 | 2.12±0.04 | −0.07 ± 0.01 | 172.875 |
| Contr_l + CrossE_L + PMI_L | SAGE | 4.74±0.27 | 0.49±0.07 | 5.24±0.44 | 0.75±0.09 | 107.125 |
| Contr_l + CrossE_L + PMI_L + PR_L | ALL | 1.54±0.50 | −3.48 ± 0.02 | 2.44±0.14 | −2.33 ± 0.12 | 192.0 |
| Contr_l + CrossE_L + PMI_L + PR_L | GAT | 9.91±1.22 | 4.54±0.47 | 10.41 ± 0.70 | 4.52±0.64 | 17.125 |
| Contr_l + CrossE_L + PMI_L + PR_L | GCN | 8.88±0.64 | 4.74±0.21 | 8.61±1.03 | 4.47±0.42 | 26.125 |
| Contr_l + CrossE_L + PMI_L + PR_L | GIN | 3.13±0.20 | −1.68 ± 0.28 | 2.83±0.63 | −0.36 ± 0.04 | 163.25 |
| Contr_l + CrossE_L + PMI_L + PR_L | MPNN | 8.48±0.52 | 3.90±0.33 | 4.68±1.04 | 1.98±0.18 | 75.0 |
| Contr_l + CrossE_L + PMI_L + PR_L | PAGNN | 2.16±0.18 | −0.03 ± 0.00 | 2.24±0.07 | −0.01 ± 0.00 | 160.75 |
| Contr_l + CrossE_L + PMI_L + PR_L | SAGE | 4.50±0.29 | 0.22±0.05 | 4.85±0.60 | 0.38±0.18 | 115.5 |
| Contr_l + CrossE_L + PMI_L + PR_L + Triplet_L | ALL | 2.72±0.16 | −2.56 ± 0.10 | 2.89±0.22 | −1.74 ± 0.26 | 175.625 |
| Contr_l + CrossE_L + PMI_L + PR_L + Triplet_L | GAT | 10.15 ± 0.69 | 4.40±0.49 | 9.18±0.65 | 4.49±0.34 | 23.25 |
| Contr_l + CrossE_L + PMI_L + PR_L + Triplet_L | GCN | 9.04±0.75 | 4.68±0.21 | 8.85±0.57 | 4.77±0.24 | 21.625 |
| Contr_l + CrossE_L + PMI_L + PR_L + Triplet_L | GIN | 3.54±0.21 | −1.92 ± 0.17 | 4.01±0.28 | −1.22 ± 0.08 | 162.0 |
| Contr_l + CrossE_L + PMI_L + PR_L + Triplet_L | MPNN | 8.35±0.26 | 3.92±0.18 | 5.27±1.27 | 2.37±0.15 | 68.5 |

Table 33. Results for Cosine-Adj Corr (↑) (continued)

| Loss Type | Model | Cora ↓ Citeseer | Cora ↓ Bitcoin | Citeseer ↓ Cora | Citeseer ↓ Bitcoin | Average Rank |
|---|---|---|---|---|---|---|
| Contr_l + CrossE_L + PMI_L + PR_L + Triplet_L | PAGNN | 2.14±0.15 | −0.05 ± 0.01 | 2.34±0.10 | −0.06 ± 0.00 | 165.125 |
| Contr_l + CrossE_L + PMI_L + PR_L + Triplet_L | SAGE | 5.20±1.00 | 1.82±0.09 | 6.58±0.30 | 1.65±0.11 | 89.375 |
| Contr_l + CrossE_L + PMI_L + Triplet_L | ALL | 4.97±0.49 | −0.43 ± 0.39 | 4.58±0.75 | 0.96±0.40 | 122.75 |
| Contr_l + CrossE_L + PMI_L + Triplet_L | GAT | 10.40 ± 0.46 | 4.75±0.25 | 10.17 ± 0.62 | 4.61±0.19 | 10.375 |
| Contr_l + CrossE_L + PMI_L + Triplet_L | GCN | 9.08±0.52 | 4.77±0.23 | 8.61±0.79 | 4.89±0.44 | 18.0 |
| Contr_l + CrossE_L + PMI_L + Triplet_L | GIN | 4.03±0.55 | −1.52 ± 0.05 | 3.96±0.77 | −1.08 ± 0.06 | 153.5 |
| Contr_l + CrossE_L + PMI_L + Triplet_L | MPNN | 8.31±0.49 | 4.22±0.20 | 6.61±0.77 | 4.03±0.22 | 50.75 |
| Contr_l + CrossE_L + PMI_L + Triplet_L | PAGNN | 2.11±0.06 | −0.05 ± 0.00 | 2.25±0.06 | −0.07 ± 0.01 | 168.375 |
| Contr_l + CrossE_L + PMI_L + Triplet_L | SAGE | 5.62±0.66 | 1.50±0.13 | 6.27±0.49 | 1.81±0.07 | 90.5 |
| Contr_l + CrossE_L + PR_L | ALL | 1.49±0.17 | −3.68 ± 0.06 | 2.08±0.21 | −3.40 ± 0.07 | 200.75 |
| Contr_l + CrossE_L + PR_L | GAT | 5.35±1.80 | 3.70±0.13 | 6.25±0.80 | 2.78±0.40 | 75.75 |
| Contr_l + CrossE_L + PR_L | GCN | 5.13±0.33 | 2.85±0.25 | 5.90±0.29 | 3.10±0.30 | 81.0 |
| Contr_l + CrossE_L + PR_L | GIN | 2.23±0.34 | −0.36 ± 0.04 | 2.75±0.33 | −0.56 ± 0.08 | 164.25 |
| Contr_l + CrossE_L + PR_L | MPNN | 3.39±1.76 | 2.83±0.41 | 4.12±1.55 | 2.90±0.31 | 105.5 |
| Contr_l + CrossE_L + PR_L | PAGNN | 1.40±0.51 | 0.03±0.00 | 0.44±0.84 | 0.03±0.00 | 166.0 |

Table 33. Results for Cosine-Adj Corr (↑) (continued)

| Loss Type | Model | Cora ↓ Citeseer | Cora ↓ Bitcoin | Citeseer ↓ Cora | Citeseer ↓ Bitcoin | Average Rank |
|---|---|---|---|---|---|---|
| Contr_l + CrossE_L + PR_L | SAGE | 4.39±0.93 | 1.70±0.11 | 4.55±1.27 | 1.40±0.15 | 110.625 |
| Contr_l + CrossE_L + PR_L + Triplet_L | ALL | 2.73±0.25 | −1.85 ± 0.34 | 2.91±0.59 | −0.68 ± 0.19 | 167.0 |
| Contr_l + CrossE_L + PR_L + Triplet_L | GAT | 7.46±0.74 | 3.69±0.26 | 8.03±1.49 | 4.20±0.14 | 52.75 |
| Contr_l + CrossE_L + PR_L + Triplet_L | GCN | 6.97±0.42 | 3.86±0.32 | 7.34±0.76 | 3.69±0.61 | 58.75 |
| Contr_l + CrossE_L + PR_L + Triplet_L | GIN | 4.46±0.28 | −1.41 ± 0.04 | 4.42±0.23 | −0.66 ± 0.12 | 144.25 |
| Contr_l + CrossE_L + PR_L + Triplet_L | MPNN | 5.76±0.31 | 3.33±0.15 | 5.31±0.62 | 2.52±0.27 | 79.5 |
| Contr_l + CrossE_L + PR_L + Triplet_L | PAGNN | 2.28±0.17 | 0.00±0.00 | 2.16±0.13 | 0.04±0.01 | 156.25 |
| Contr_l + CrossE_L + PR_L + Triplet_L | SAGE | 5.86±0.17 | 1.94±0.10 | 6.55±0.80 | 1.70±0.15 | 84.375 |
| Contr_l + CrossE_L + Triplet_L | ALL | 4.47±0.69 | 1.31±0.11 | 4.76±0.96 | 0.54±0.14 | 111.75 |
| Contr_l + CrossE_L + Triplet_L | GAT | 9.60±0.62 | 4.14±0.35 | 9.99±0.34 | 4.44±0.22 | 24.125 |
| Contr_l + CrossE_L + Triplet_L | GCN | 8.99±0.97 | 4.30±0.44 | 8.95±0.78 | 4.58±0.20 | 29.0 |
| Contr_l + CrossE_L + Triplet_L | GIN | 6.18±0.52 | −0.41 ± 0.15 | 5.82±0.99 | 0.68±0.06 | 109.75 |
| Contr_l + CrossE_L + Triplet_L | MPNN | 7.08±0.57 | 3.27±0.25 | 6.50±0.53 | 3.85±0.17 | 65.125 |
| Contr_l + CrossE_L + Triplet_L | PAGNN | 3.52±0.70 | −0.02 ± 0.01 | 4.40±0.67 | 0.05±0.04 | 133.75 |
| Contr_l + CrossE_L + Triplet_L | SAGE | 6.60±0.46 | 2.03±0.12 | 8.17±0.89 | 2.25±0.14 | 70.0 |
| Contr_l + PMI_L | ALL | 3.24±0.51 | −2.11 ± 0.08 | 3.67±0.89 | −0.47 ± 0.17 | 161.25 |

Table 33. Results for Cosine-Adj Corr (↑) (continued)

| Loss Type | Model | Cora ↓ Citeseer | Cora ↓ Bitcoin | Citeseer ↓ Cora | Citeseer ↓ Bitcoin | Average Rank |
|---|---|---|---|---|---|---|
| Contr_l + PMI_L | GAT | 10.64 ± 1.14 | 4.72±0.33 | 10.67 ± 0.52 | 4.68±0.31 | 8.0 |
| Contr_l + PMI_L | GCN | 9.42±1.08 | 4.63±0.35 | 9.20±0.82 | 4.77±0.45 | 17.125 |
| Contr_l + PMI_L | GIN | 3.58±0.65 | −1.90 ± 0.10 | 3.39±0.42 | −1.06 ± 0.05 | 161.25 |
| Contr_l + PMI_L | MPNN | 8.28±0.49 | 3.59±0.19 | 6.87±0.82 | 4.13±0.13 | 55.75 |
| Contr_l + PMI_L | PAGNN | 2.06±0.15 | −0.05 ± 0.00 | 2.10±0.03 | −0.06 ± 0.00 | 171.25 |
| Contr_l + PMI_L | SAGE | 4.61±0.33 | 0.60±0.15 | 5.85±0.71 | 0.95±0.04 | 103.875 |
| Contr_l + PMI_L + PR_L | ALL | 1.51±0.21 | −3.45 ± 0.16 | 2.46±0.11 | −2.37 ± 0.13 | 191.875 |
| Contr_l + PMI_L + PR_L | GAT | 9.83±1.07 | 4.59±0.23 | 7.47±2.20 | 2.46±0.15 | 40.875 |
| Contr_l + PMI_L + PR_L | GCN | 8.46±0.34 | 4.31±0.40 | 8.49±1.19 | 4.30±0.38 | 39.0 |
| Contr_l + PMI_L + PR_L | GIN | 3.46±0.72 | −1.10 ± 0.13 | 2.62±0.66 | −0.38 ± 0.04 | 161.25 |
| Contr_l + PMI_L + PR_L | MPNN | 8.59±0.40 | 4.07±0.14 | 4.71±1.21 | 1.41±0.12 | 76.0 |
| Contr_l + PMI_L + PR_L | PAGNN | 1.98±0.11 | −0.04 ± 0.01 | 2.20±0.11 | −0.03 ± 0.00 | 166.625 |
| Contr_l + PMI_L + PR_L | SAGE | 4.63±0.23 | 0.61±0.16 | 4.99±0.61 | 0.95±0.10 | 107.75 |
| Contr_l + PMI_L + PR_L + Triplet_L | ALL | 3.11±0.24 | −2.14 ± 0.16 | 2.76±0.29 | −0.98 ± 0.23 | 168.875 |
| Contr_l + PMI_L + PR_L + Triplet_L | GAT | 8.05±1.05 | 4.03±0.50 | 8.08±1.24 | 3.30±0.45 | 53.125 |
| Contr_l + PMI_L + PR_L + Triplet_L | GCN | 9.08±0.70 | 4.72±0.20 | 8.64±0.77 | 4.75±0.09 | 21.875 |
| Contr_l + PMI_L + PR_L + Triplet_L | GIN | 4.08±0.18 | −1.77 ± 0.08 | 3.55±0.27 | −0.86 ± 0.08 | 155.25 |
| Contr_l + PMI_L + PR_L + Triplet_L | MPNN | 8.78±0.37 | 3.91±0.19 | 4.92±0.29 | 1.84±0.27 | 71.625 |

Table 33. Results for Cosine-Adj Corr (↑) (continued)

| Loss Type | Model | Cora ↓ Citeseer | Cora ↓ Bitcoin | Citeseer ↓ Cora | Citeseer ↓ Bitcoin | Average Rank |
|---|---|---|---|---|---|---|
| Contr_l + PMI_L + PR_L + Triplet_L | PAGNN | 2.24±0.14 | −0.01 ± 0.01 | 2.46±0.11 | −0.03 ± 0.01 | 155.875 |
| Contr_l + PMI_L + PR_L + Triplet_L | SAGE | 5.07±0.63 | 1.82±0.07 | 6.08±0.68 | 1.68±0.14 | 92.375 |
| Contr_l + PR_L | ALL | 1.59±0.19 | −3.50 ± 0.06 | 2.51±0.42 | −3.24 ± 0.05 | 192.125 |
| Contr_l + PR_L | GAT | 5.03±1.24 | 3.60±0.10 | 5.15±0.54 | 2.46±0.35 | 85.125 |
| Contr_l + PR_L | GCN | 5.14±0.56 | 3.24±0.36 | 5.96±0.93 | 3.40±0.31 | 78.25 |
| Contr_l + PR_L | GIN | 2.13±0.86 | −0.30 ± 0.02 | 2.05±1.03 | −1.15 ± 0.08 | 179.25 |
| Contr_l + PR_L | MPNN | 3.17±0.66 | 1.59±1.17 | 4.67±1.08 | 3.32±0.31 | 107.625 |
| Contr_l + PR_L | PAGNN | 1.07±0.51 | 0.05±0.00 | 0.54±0.37 | 0.04±0.00 | 165.5 |
| Contr_l + PR_L | SAGE | 4.44±1.16 | 1.79±0.10 | 5.26±1.25 | 0.72±0.11 | 104.75 |
| Contr_l + PR_L + Triplet_L | ALL | 2.40±0.56 | −2.43 ± 0.10 | 2.52±0.38 | −1.41 ± 0.15 | 179.0 |
| Contr_l + PR_L + Triplet_L | GAT | 7.31±0.90 | 3.94±0.24 | 8.22±1.31 | 3.18±0.27 | 55.5 |
| Contr_l + PR_L + Triplet_L | GCN | 6.60±0.32 | 3.77±0.44 | 7.65±0.64 | 3.89±0.41 | 58.0 |
| Contr_l + PR_L + Triplet_L | GIN | 4.74±0.34 | −0.90 ± 0.11 | 4.31±0.77 | −1.39 ± 0.13 | 146.75 |
| Contr_l + PR_L + Triplet_L | MPNN | 5.77±1.05 | 3.08±0.10 | 5.01±0.66 | 2.96±0.10 | 81.5 |
| Contr_l + PR_L + Triplet_L | PAGNN | 2.11±0.23 | 0.00±0.00 | 2.30±0.34 | 0.04±0.01 | 157.125 |
| Contr_l + PR_L + Triplet_L | SAGE | 5.16±0.28 | 1.92±0.13 | 6.88±0.56 | 1.76±0.10 | 85.75 |
| Contr_l + Triplet_L | ALL | 4.62±0.20 | 0.39±0.29 | 5.06±1.22 | 0.15±0.21 | 112.75 |
| Contr_l + Triplet_L | GAT | 9.45±0.27 | 4.13±0.42 | 10.08 ± 0.42 | 4.31±0.23 | 26.0 |
| Contr_l + Triplet_L | GCN | 8.60±0.57 | 4.38±0.26 | 9.17±0.58 | 4.92±0.18 | 24.375 |

Table 33. Results for Cosine-Adj Corr (↑) (continued)

| Loss Type | Model | Cora ↓ Citeseer | Cora ↓ Bitcoin | Citeseer ↓ Cora | Citeseer ↓ Bitcoin | Average Rank |
|---|---|---|---|---|---|---|
| Contr_l + Triplet_L | GIN | 6.15±0.18 | −0.88 ± 0.12 | 5.98±1.25 | −0.01 ± 0.05 | 115.375 |
| Contr_l + Triplet_L | MPNN | 7.34±0.63 | 3.36±0.43 | 7.65±0.62 | 3.90±0.23 | 57.5 |
| Contr_l + Triplet_L | PAGNN | 3.64±0.67 | 0.02±0.04 | 3.71±0.25 | −0.02 ± 0.06 | 136.25 |
| Contr_l + Triplet_L | SAGE | 6.60±0.54 | 1.78±0.16 | 7.92±0.82 | 1.97±0.09 | 75.5 |
| CrossE_L | ALL | 3.36±1.41 | 2.39±0.11 | −0.44 ± 3.28 | −1.39 ± 0.38 | 156.0 |
| CrossE_L | GAT | 2.39±1.13 | 0.42±0.08 | 2.35±1.29 | 1.92±0.45 | 136.875 |
| CrossE_L | GCN | −2.96 ± 0.64 | −3.83 ± 0.05 | −6.88 ± 0.03 | −4.08 ± 0.01 | 209.75 |
| CrossE_L | GIN | −2.65 ± 0.23 | −4.00 ± 0.00 | −5.53 ± 0.24 | −3.84 ± 0.00 | 209.25 |
| CrossE_L | MPNN | 2.58±0.52 | −2.14 ± 0.28 | 3.28±0.31 | −3.03 ± 0.30 | 176.25 |
| CrossE_L | PAGNN | 0.49±0.00 | 0.00±0.00 | 0.92±0.36 | −0.29 ± 0.01 | 176.5 |
| CrossE_L | SAGE | 0.49±0.00 | 0.04±0.01 | 1.46±0.89 | −0.02 ± 0.00 | 168.25 |
| CrossE_L + PMI_L | ALL | 3.07±0.25 | −2.84 ± 0.06 | 2.70±0.18 | −2.03 ± 0.07 | 178.5 |
| CrossE_L + PMI_L | GAT | 10.67 ± 0.70 | 4.72±0.10 | 10.42 ± 0.79 | 4.80±0.17 | 6.25 |
| CrossE_L + PMI_L | GCN | 8.52±0.64 | 4.28±0.33 | 9.20±0.57 | 4.83±0.23 | 26.5 |
| CrossE_L + PMI_L | GIN | 3.69±0.38 | −1.63 ± 0.09 | 3.34±0.46 | −1.05 ± 0.17 | 158.0 |
| CrossE_L + PMI_L | MPNN | 7.96±0.44 | 3.93±0.23 | 6.46±0.92 | 4.10±0.25 | 57.625 |
| CrossE_L + PMI_L | PAGNN | 2.03±0.24 | 0.05±0.00 | 2.17±0.09 | −0.05 ± 0.01 | 162.125 |
| CrossE_L + PMI_L | SAGE | 4.24±0.23 | 0.25±0.09 | 4.80±0.14 | 0.25±0.06 | 118.25 |
| CrossE_L + PMI_L + PR_L | ALL | 1.23±0.05 | −3.50 ± 0.03 | 2.43±0.15 | −2.29 ± 0.05 | 194.625 |

Continued on next page

Table 33. Results for Cosine-Adj Corr (↑) (continued)

| Loss Type | Model | Cora ↓ Citeseer | Cora ↓ Bitcoin | Citeseer ↓ Cora | Citeseer ↓ Bitcoin | Average Rank |
|---|---|---|---|---|---|---|
| CrossE_L + PMI_L + PR_L | GAT | 10.49 ± 0.53 | 4.68±0.19 | 7.07±2.45 | 2.11±0.35 | 40.5 |
| CrossE_L + PMI_L + PR_L | GCN | 8.74±0.45 | 4.63±0.36 | 8.89±0.80 | 4.89±0.24 | 22.375 |
| CrossE_L + PMI_L + PR_L | GIN | 3.23±0.40 | −2.05 ± 0.16 | 2.74±0.70 | −1.19 ± 0.12 | 170.5 |
| CrossE_L + PMI_L + PR_L | MPNN | 8.33±0.58 | 3.89±0.15 | 5.17±1.48 | 3.71±0.23 | 64.5 |
| CrossE_L + PMI_L + PR_L | PAGNN | 2.16±0.22 | −0.05 ± 0.00 | 2.10±0.15 | −0.05 ± 0.00 | 168.0 |
| CrossE_L + PMI_L + PR_L | SAGE | 4.45±0.23 | 0.33±0.08 | 4.75±0.16 | 0.27±0.12 | 116.25 |
| CrossE_L + PMI_L + PR_L + Triplet_L | ALL | 2.75±0.17 | −2.73 ± 0.10 | 2.71±0.53 | −1.74 ± 0.16 | 177.5 |
| CrossE_L + PMI_L + PR_L + Triplet_L | GAT | 10.50 ± 1.51 | 4.56±0.15 | 9.00±1.32 | 4.36±0.28 | 22.875 |
| CrossE_L + PMI_L + PR_L + Triplet_L | GCN | 9.09±0.42 | 4.45±0.27 | 8.52±0.81 | 4.25±0.27 | 32.875 |
| CrossE_L + PMI_L + PR_L + Triplet_L | GIN | 3.62±0.49 | −1.38 ± 0.10 | 3.75±0.85 | −1.55 ± 0.12 | 160.25 |
| CrossE_L + PMI_L + PR_L + Triplet_L | MPNN | 7.76±0.27 | 3.87±0.32 | 5.61±1.07 | 3.63±0.18 | 65.75 |
| CrossE_L + PMI_L + PR_L + Triplet_L | PAGNN | 2.35±0.08 | 0.01±0.00 | 2.28±0.05 | −0.05 ± 0.00 | 157.5 |
| CrossE_L + PMI_L + PR_L + Triplet_L | SAGE | 5.57±0.74 | 1.73±0.11 | 5.97±0.60 | 1.60±0.09 | 93.5 |
| CrossE_L + PMI_L + Triplet_L | ALL | 5.39±0.36 | 0.01±0.15 | 4.55±0.51 | 1.64±0.15 | 111.25 |
| CrossE_L + PMI_L + Triplet_L | GAT | 10.41 ± 0.63 | 4.56±0.39 | 10.29 ± 0.53 | 4.80±0.14 | 11.875 |
| CrossE_L + PMI_L + Triplet_L | GCN | 8.99±0.68 | 4.97±0.23 | 9.31±0.55 | 4.54±0.27 | 18.125 |

Continued on next page

Table 33. Results for Cosine-Adj Corr (↑) (continued)

| Loss Type | Model | Cora ↓ Citeseer | Cora ↓ Bitcoin | Citeseer ↓ Cora | Citeseer ↓ Bitcoin | Average Rank |
|---|---|---|---|---|---|---|
| CrossE_L + PMI_L + Triplet_L | GIN | 4.09±0.28 | −1.72 ± 0.09 | 4.02±0.37 | −1.09 ± 0.17 | 154.0 |
| CrossE_L + PMI_L + Triplet_L | MPNN | 8.07±0.95 | 4.10±0.40 | 6.80±0.30 | 4.05±0.30 | 52.5 |
| CrossE_L + PMI_L + Triplet_L | PAGNN | 2.22±0.14 | −0.01 ± 0.01 | 2.32±0.07 | −0.04 ± 0.01 | 158.875 |
| CrossE_L + PMI_L + Triplet_L | SAGE | 6.19±0.34 | 1.79±0.13 | 6.69±0.59 | 2.06±0.05 | 80.25 |
| CrossE_L + PR_L | ALL | 1.37±0.41 | −3.42 ± 0.07 | 2.72±0.68 | −3.46 ± 0.04 | 191.375 |
| CrossE_L + PR_L | GAT | 4.16±0.90 | 2.79±0.27 | 4.98±0.60 | 2.42±0.20 | 96.0 |
| CrossE_L + PR_L | GCN | 2.87±0.39 | 1.80±0.35 | 4.39±0.92 | 1.73±0.38 | 117.625 |
| CrossE_L + PR_L | GIN | 1.52±0.30 | −0.24 ± 0.02 | 1.42±1.59 | −0.25 ± 0.02 | 177.25 |
| CrossE_L + PR_L | MPNN | 3.76±2.21 | 1.88±0.77 | 2.15±1.30 | 2.68±0.73 | 121.875 |
| CrossE_L + PR_L | PAGNN | 0.86±0.36 | 0.02±0.00 | −0.55 ± 0.21 | −0.05 ± 0.00 | 172.875 |
| CrossE_L + PR_L | SAGE | 2.13±0.22 | 0.61±0.10 | 3.24±0.15 | 0.50±0.07 | 140.125 |
| CrossE_L + PR_L + Triplet_L | ALL | 2.22±0.67 | −3.46 ± 0.14 | 2.16±0.17 | −1.20 ± 0.13 | 187.25 |
| CrossE_L + PR_L + Triplet_L | GAT | 5.70±1.30 | 2.81±0.29 | 7.30±1.48 | 4.16±0.11 | 65.25 |
| CrossE_L + PR_L + Triplet_L | GCN | 6.02±0.70 | 3.81±0.29 | 7.22±0.85 | 4.27±0.43 | 58.0 |
| CrossE_L + PR_L + Triplet_L | GIN | 4.26±0.47 | −1.21 ± 0.19 | 4.37±0.57 | −1.07 ± 0.08 | 147.75 |
| CrossE_L + PR_L + Triplet_L | MPNN | 4.73±0.85 | 3.27±0.33 | 4.42±0.92 | 2.94±0.41 | 91.5 |
| CrossE_L + PR_L + Triplet_L | PAGNN | 2.04±0.33 | 0.03±0.01 | 1.92±0.42 | 0.04±0.00 | 160.625 |
| CrossE_L + PR_L + Triplet_L | SAGE | 6.06±0.76 | 1.96±0.21 | 7.14±0.63 | 1.73±0.12 | 79.5 |

Table 33. Results for Cosine-Adj Corr (↑) (continued)

| Loss Type | Model | Cora ↓ Citeseer | Cora ↓ Bitcoin | Citeseer ↓ Cora | Citeseer ↓ Bitcoin | Average Rank |
|---|---|---|---|---|---|---|
| CrossE_L + Triplet_L | ALL | 5.69±0.38 | 1.09±0.09 | 5.26±1.21 | 0.85±0.10 | 100.375 |
| CrossE_L + Triplet_L | GAT | 10.17 ± 0.50 | 4.52±0.19 | 10.11 ± 0.57 | 4.59±0.27 | 16.75 |
| CrossE_L + Triplet_L | GCN | 9.16±0.76 | 4.62±0.31 | 9.37±0.68 | 4.37±0.40 | 22.5 |
| CrossE_L + Triplet_L | GIN | 6.41±0.40 | −1.15 ± 0.10 | 5.54±0.32 | −0.55 ± 0.10 | 124.25 |
| CrossE_L + Triplet_L | MPNN | 8.24±0.79 | 4.03±0.24 | 7.52±0.58 | 3.94±0.29 | 50.375 |
| CrossE_L + Triplet_L | PAGNN | 3.91±0.75 | −0.05 ± 0.00 | 4.08±0.56 | −0.31 ± 0.05 | 143.875 |
| CrossE_L + Triplet_L | SAGE | 8.29±0.10 | 1.94±0.08 | 8.99±0.77 | 1.92±0.05 | 65.0 |
| PMI_L | ALL | 2.57±0.13 | −2.94 ± 0.11 | 2.71±0.24 | −1.96 ± 0.16 | 180.375 |
| PMI_L | GAT | 10.59 ± 0.68 | `4.83±0.31` | 10.12 ± 0.38 | 4.89±0.27 | `4.75` |
| PMI_L | GCN | 8.79±0.96 | 4.61±0.40 | 8.96±0.32 | `5.08±0.57` | 20.75 |
| PMI_L | GIN | 3.36±0.23 | −1.72 ± 0.14 | 3.31±0.38 | −1.19 ± 0.09 | 164.625 |
| PMI_L | MPNN | 8.69±0.92 | 4.14±0.27 | 6.66±0.66 | 3.98±0.41 | 48.125 |
| PMI_L | PAGNN | 2.13±0.17 | −0.03 ± 0.00 | 2.15±0.04 | −0.06 ± 0.01 | 167.5 |
| PMI_L | SAGE | 4.06±0.27 | 0.38±0.11 | 4.35±0.34 | 0.19±0.10 | 123.25 |
| PMI_L + PR_L | ALL | 1.43±0.29 | −3.54 ± 0.02 | 2.41±0.22 | −2.25 ± 0.11 | 193.75 |
| PMI_L + PR_L | GAT | 9.14±1.11 | 3.96±0.36 | 7.43±2.52 | 3.60±0.42 | 45.5 |
| PMI_L + PR_L | GCN | 8.75±0.95 | 4.78±0.18 | 8.79±1.47 | 4.52±0.36 | 24.625 |
| PMI_L + PR_L | GIN | 3.27±0.38 | −1.41 ± 0.14 | 2.85±0.66 | −1.46 ± 0.12 | 167.125 |
| PMI_L + PR_L | MPNN | 8.21±0.79 | 3.96±0.27 | 3.89±0.56 | 1.20±0.15 | 87.625 |
| PMI_L + PR_L | PAGNN | 2.16±0.17 | −0.04 ± 0.01 | 2.21±0.05 | −0.07 ± 0.00 | 166.375 |
| PMI_L + PR_L | SAGE | 4.31±0.16 | 0.32±0.11 | 4.73±0.25 | 0.11±0.09 | 118.75 |

Table 33. Results for Cosine-Adj Corr (↑) (continued)

| Loss Type | Model | Cora ↓ Citeseer | Cora ↓ Bitcoin | Citeseer ↓ Cora | Citeseer ↓ Bitcoin | Average Rank |
|---|---|---|---|---|---|---|
| PMI_L + PR_L + Triplet_L | ALL | 2.82±0.12 | −2.82 ± 0.22 | 2.71±0.27 | −1.47 ± 0.19 | 176.875 |
| PMI_L + PR_L + Triplet_L | GAT | 9.33±0.54 | 4.73±0.28 | 9.05±1.43 | 4.12±0.37 | 24.5 |
| PMI_L + PR_L + Triplet_L | GCN | 8.60±0.58 | 4.73±0.43 | 8.87±0.92 | 4.38±0.35 | 28.125 |
| PMI_L + PR_L + Triplet_L | GIN | 3.82±0.33 | −1.66 ± 0.15 | 3.15±0.57 | −1.19 ± 0.11 | 161.75 |
| PMI_L + PR_L + Triplet_L | MPNN | 8.22±0.91 | 4.17±0.15 | 4.46±0.67 | 2.10±0.22 | 74.25 |
| PMI_L + PR_L + Triplet_L | PAGNN | 2.19±0.15 | −0.02 ± 0.01 | 2.30±0.08 | −0.03 ± 0.00 | 159.625 |
| PMI_L + PR_L + Triplet_L | SAGE | 5.64±0.44 | 1.78±0.06 | 6.76±0.73 | 1.60±0.04 | 88.25 |
| PMI_L + Triplet_L | ALL | 5.57±0.29 | −0.40 ± 0.17 | 5.11±0.82 | 1.64±0.15 | 113.125 |
| PMI_L + Triplet_L | GAT | 10.48 ± 0.62 | 4.66±0.23 | 10.84 ± 0.37 | 4.78±0.42 | 9.0 |
| PMI_L + Triplet_L | GCN | 8.42±0.61 | 4.60±0.33 | 9.55±0.70 | 4.62±0.18 | 25.25 |
| PMI_L + Triplet_L | GIN | 4.21±0.21 | −1.44 ± 0.16 | 3.90±0.83 | −0.97 ± 0.14 | 151.0 |
| PMI_L + Triplet_L | MPNN | 8.30±0.77 | 4.02±0.15 | 6.69±0.71 | 4.38±0.34 | 49.25 |
| PMI_L + Triplet_L | PAGNN | 2.07±0.16 | −0.03 ± 0.00 | 2.19±0.07 | −0.05 ± 0.00 | 166.375 |
| PMI_L + Triplet_L | SAGE | 5.79±0.76 | 1.40±0.15 | 6.97±0.29 | 1.96±0.05 | 83.75 |
| PR_L | ALL | 0.72±0.26 | −3.58 ± 0.03 | 1.72±0.90 | −3.54 ± 0.05 | 204.375 |
| PR_L | GAT | 4.43±0.56 | 2.65±0.24 | 5.25±0.31 | 2.67±0.23 | 91.0 |
| PR_L | GCN | 2.58±0.74 | 1.59±0.37 | 4.70±0.37 | 1.89±0.27 | 117.5 |
| PR_L | GIN | 1.37±0.30 | −0.21 ± 0.03 | 1.03±0.94 | −0.28 ± 0.02 | 179.125 |
| PR_L | MPNN | 3.77±2.79 | 2.91±0.55 | 2.67±0.89 | 2.59±0.25 | 112.0 |

Continued on next page

Table 33. Results for Cosine-Adj Corr (↑) (continued)

| Loss Type | Model | Cora ↓ Citeseer | Cora ↓ Bitcoin | Citeseer ↓ Cora | Citeseer ↓ Bitcoin | Average Rank |
|-----------|-------|-----------------|----------------|-----------------|--------------------|--------------|
| PR_L | PAGNN | 0.59±0.79 | −0.01 ± 0.00 | −0.50 ± 0.13 | 0.01±0.00 | 172.25 |
| PR_L | SAGE | 2.36±0.19 | 0.50±0.07 | 3.46±0.30 | 0.62±0.07 | 135.0 |
| PR_L + Triplet_L | ALL | 1.38±0.31 | −3.39 ± 0.16 | 2.25±0.37 | −3.54 ± 0.09 | 196.5 |
| PR_L + Triplet_L | GAT | 5.51±2.05 | 1.96±0.22 | 4.88±0.66 | 2.39±0.20 | 91.125 |
| PR_L + Triplet_L | GCN | 4.53±0.41 | 2.90±0.38 | 5.49±0.26 | 2.99±0.26 | 85.5 |
| PR_L + Triplet_L | GIN | 1.99±0.46 | −0.40 ± 0.03 | 2.71±0.59 | −1.09 ± 0.03 | 172.875 |
| PR_L + Triplet_L | MPNN | 3.96±2.44 | 3.09±0.25 | 3.70±1.74 | 3.41±0.09 | 100.5 |
| PR_L + Triplet_L | PAGNN | 1.07±0.28 | 0.01±0.00 | 0.06±0.42 | 0.03±0.00 | 168.75 |
| PR_L + Triplet_L | SAGE | 2.59±0.41 | 0.68±0.05 | 4.61±1.37 | 0.59±0.03 | 126.0 |
| Triplet_L | ALL | 5.53±0.60 | 0.28±0.11 | 6.05±0.47 | 1.97±0.22 | 94.625 |
| Triplet_L | GAT | 10.54 ± 0.70 | 4.32±0.30 | 9.97±0.65 | 4.46±0.44 | 19.875 |
| Triplet_L | GCN | 9.53±0.41 | 4.87±0.49 | 9.03±1.02 | 4.80±0.45 | 13.0 |
| Triplet_L | GIN | 7.13±0.31 | −0.79 ± 0.05 | 5.77±0.76 | −0.03 ± 0.14 | 115.0 |
| Triplet_L | MPNN | 8.66±0.87 | 4.04±0.24 | 7.27±0.99 | 4.12±0.47 | 45.875 |
| Triplet_L | PAGNN | 4.01±0.72 | −0.02 ± 0.03 | 4.24±0.42 | 0.07±0.03 | 131.0 |
| Triplet_L | SAGE | 7.95±1.39 | 2.34±0.10 | 9.41±0.64 | 2.50±0.07 | 58.75 |

Table 34. Dot-Adj Corr Performance (↑): This table presents models (Loss function and GNN) ranked by their average performance in terms of dot-adj corr. Top-ranked results are highlighted in red, second-ranked in blue, and third-ranked in green.

| Loss Type | Model | Cora ↓ Citeseer | Cora ↓ Bitcoin | Citeseer ↓ Cora | Citeseer ↓ Bitcoin | Average Rank |
|-----------|-------|-----------------|----------------|-----------------|--------------------|--------------|

Table 34. Results for Dot-Adj Corr (↑) (continued)

| Loss Type | Model | Cora ↓ Citeseer | Cora ↓ Bitcoin | Citeseer ↓ Cora | Citeseer ↓ Bitcoin | Average Rank |
|---|---|---|---|---|---|---|
| Contr_l | ALL | 3.68±0.32 | −0.26 ± 0.07 | 4.19±0.85 | −0.69 ± 0.24 | 148.25 |
| Contr_l | GAT | 9.11±1.10 | 4.09±0.47 | 9.64±0.57 | 4.64±0.13 | 24.25 |
| Contr_l | GCN | 8.43±0.67 | 4.46±0.27 | 9.32±0.78 | 4.60±0.28 | 27.75 |
| Contr_l | GIN | 5.87±0.35 | −0.63 ± 0.10 | 5.51±1.03 | 0.87±0.10 | 111.25 |
| Contr_l | MPNN | 6.83±0.53 | 3.51±0.31 | 6.36±1.15 | 3.60±0.22 | 67.125 |
| Contr_l | PAGNN | 3.57±0.18 | −0.03 ± 0.03 | 3.31±0.68 | 0.04±0.03 | 140.75 |
| Contr_l | SAGE | 5.74±0.81 | 1.97±0.17 | 7.65±0.83 | 2.29±0.24 | 75.0 |
| Contr_l + CrossE_L | ALL | 3.97±0.50 | 1.33±0.11 | 3.82±0.78 | 0.43±0.27 | 123.0 |
| Contr_l + CrossE_L | GAT | 9.10±0.66 | 4.12±0.16 | 9.81±0.66 | 4.26±0.25 | 29.25 |
| Contr_l + CrossE_L | GCN | 8.52±0.42 | 4.44±0.35 | 9.19±0.78 | 4.57±0.26 | 29.25 |
| Contr_l + CrossE_L | GIN | 4.76±0.29 | −1.65 ± 0.20 | 5.16±0.53 | −0.33 ± 0.06 | 135.5 |
| Contr_l + CrossE_L | MPNN | 6.88±0.79 | 3.56±0.27 | 6.46±0.57 | 3.75±0.27 | 65.125 |
| Contr_l + CrossE_L | PAGNN | 3.48±0.17 | −0.00 ± 0.01 | 3.03±0.23 | −0.03 ± 0.00 | 143.625 |
| Contr_l + CrossE_L | SAGE | 6.26±0.45 | 1.68±0.08 | 6.93±0.34 | 1.59±0.09 | 85.0 |
| Contr_l + CrossE_L + PMI_L | ALL | 3.05±0.31 | −1.81 ± 0.28 | 3.19±0.61 | −1.94 ± 0.19 | 171.5 |
| Contr_l + CrossE_L + PMI_L | GAT | 10.86 ± 0.42 | 4.68±0.24 | 10.32 ± 0.77 | 4.76±0.30 | 8.5 |
| Contr_l + CrossE_L + PMI_L | GCN | 8.98±0.46 | 4.75±0.12 | 8.68±0.74 | 4.57±0.15 | 23.5 |
| Contr_l + CrossE_L + PMI_L | GIN | 3.92±0.48 | −1.37 ± 0.19 | 3.30±0.27 | −1.13 ± 0.09 | 157.5 |
| Contr_l + CrossE_L + PMI_L | MPNN | 8.78±0.55 | 4.32±0.30 | 6.15±0.46 | 4.25±0.20 | 45.625 |
| Contr_l + CrossE_L + PMI_L | PAGNN | 1.92±0.08 | −0.05 ± 0.01 | 2.12±0.04 | −0.07 ± 0.01 | 172.875 |

Table 34. Results for Dot-Adj Corr (↑) (continued)

| Loss Type | Model | Cora ↓ Citeseer | Cora ↓ Bitcoin | Citeseer ↓ Cora | Citeseer ↓ Bitcoin | Average Rank |
|---|---|---|---|---|---|---|
| Contr_l + CrossE_L + PMI_L | SAGE | 4.74±0.27 | 0.49±0.07 | 5.24±0.44 | 0.75±0.09 | 107.125 |
| Contr_l + CrossE_L + PMI_L + PR_L | ALL | 1.54±0.50 | −3.48 ± 0.02 | 2.44±0.14 | −2.33 ± 0.12 | 192.25 |
| Contr_l + CrossE_L + PMI_L + PR_L | GAT | 9.91±1.22 | 4.54±0.47 | 10.41 ± 0.70 | 4.52±0.64 | 17.125 |
| Contr_l + CrossE_L + PMI_L + PR_L | GCN | 8.88±0.64 | 4.74±0.21 | 8.61±1.03 | 4.47±0.42 | 26.125 |
| Contr_l + CrossE_L + PMI_L + PR_L | GIN | 3.13±0.20 | −1.68 ± 0.28 | 2.83±0.63 | −0.36 ± 0.04 | 163.25 |
| Contr_l + CrossE_L + PMI_L + PR_L | MPNN | 8.48±0.52 | 3.90±0.33 | 4.68±1.04 | 1.98±0.18 | 75.25 |
| Contr_l + CrossE_L + PMI_L + PR_L | PAGNN | 2.16±0.18 | −0.03 ± 0.00 | 2.24±0.07 | −0.01 ± 0.00 | 160.75 |
| Contr_l + CrossE_L + PMI_L + PR_L | SAGE | 4.50±0.29 | 0.22±0.05 | 4.85±0.60 | 0.38±0.18 | 115.25 |
| Contr_l + CrossE_L + PMI_L + PR_L + Triplet_L | ALL | 2.72±0.16 | −2.56 ± 0.10 | 2.89±0.22 | −1.74 ± 0.26 | 175.875 |
| Contr_l + CrossE_L + PMI_L + PR_L + Triplet_L | GAT | 10.15 ± 0.69 | 4.40±0.49 | 9.18±0.65 | 4.49±0.34 | 23.25 |
| Contr_l + CrossE_L + PMI_L + PR_L + Triplet_L | GCN | 9.04±0.75 | 4.68±0.21 | 8.85±0.57 | 4.77±0.24 | 21.625 |
| Contr_l + CrossE_L + PMI_L + PR_L + Triplet_L | GIN | 3.54±0.21 | −1.92 ± 0.17 | 4.01±0.28 | −1.22 ± 0.08 | 162.0 |
| Contr_l + CrossE_L + PMI_L + PR_L + Triplet_L | MPNN | 8.35±0.26 | 3.92±0.18 | 5.27±1.27 | 2.37±0.15 | 68.625 |

Table 34. Results for Dot-Adj Corr (↑) (continued)

| Loss Type | Model | Cora ↓ Citeseer | Cora ↓ Bitcoin | Citeseer ↓ Cora | Citeseer ↓ Bitcoin | Average Rank |
|---|---|---|---|---|---|---|
| Contr_l + CrossE_L + PMI_L + PR_L + Triplet_L | PAGNN | 2.14±0.15 | −0.05 ± 0.01 | 2.34±0.10 | −0.06 ± 0.00 | 165.125 |
| Contr_l + CrossE_L + PMI_L + PR_L + Triplet_L | SAGE | 5.20±1.00 | 1.82±0.09 | 6.58±0.30 | 1.65±0.11 | 89.375 |
| Contr_l + CrossE_L + PMI_L + Triplet_L | ALL | 4.97±0.49 | −0.43 ± 0.39 | 4.58±0.75 | 0.96±0.40 | 122.75 |
| Contr_l + CrossE_L + PMI_L + Triplet_L | GAT | 10.40 ± 0.46 | 4.75±0.25 | 10.17 ± 0.62 | 4.61±0.19 | 10.375 |
| Contr_l + CrossE_L + PMI_L + Triplet_L | GCN | 9.08±0.52 | 4.77±0.23 | 8.61±0.79 | 4.89±0.44 | 18.0 |
| Contr_l + CrossE_L + PMI_L + Triplet_L | GIN | 4.03±0.55 | −1.52 ± 0.05 | 3.96±0.77 | −1.08 ± 0.06 | 153.5 |
| Contr_l + CrossE_L + PMI_L + Triplet_L | MPNN | 8.31±0.49 | 4.22±0.20 | 6.61±0.77 | 4.03±0.22 | 50.75 |
| Contr_l + CrossE_L + PMI_L + Triplet_L | PAGNN | 2.11±0.06 | −0.05 ± 0.00 | 2.25±0.06 | −0.07 ± 0.01 | 168.375 |
| Contr_l + CrossE_L + PMI_L + Triplet_L | SAGE | 5.62±0.66 | 1.50±0.13 | 6.27±0.49 | 1.81±0.07 | 90.5 |
| Contr_l + CrossE_L + PR_L | ALL | 1.49±0.17 | −3.68 ± 0.06 | 2.08±0.21 | −3.40 ± 0.07 | 200.75 |
| Contr_l + CrossE_L + PR_L | GAT | 5.35±1.80 | 3.70±0.13 | 6.25±0.80 | 2.78±0.40 | 75.75 |
| Contr_l + CrossE_L + PR_L | GCN | 5.13±0.33 | 2.85±0.25 | 5.90±0.29 | 3.10±0.30 | 81.0 |
| Contr_l + CrossE_L + PR_L | GIN | 2.23±0.34 | −0.36 ± 0.04 | 2.75±0.33 | −0.56 ± 0.08 | 164.25 |
| Contr_l + CrossE_L + PR_L | MPNN | 3.39±1.76 | 2.83±0.41 | 4.12±1.55 | 2.90±0.31 | 105.5 |
| Contr_l + CrossE_L + PR_L | PAGNN | 1.40±0.51 | 0.03±0.00 | 0.44±0.84 | 0.03±0.00 | 166.0 |

Table 34. Results for Dot-Adj Corr (↑) (continued)

| Loss Type | Model | Cora ↓ Citeseer | Cora ↓ Bitcoin | Citeseer ↓ Cora | Citeseer ↓ Bitcoin | Average Rank |
|---|---|---|---|---|---|---|
| Contr_l + CrossE_L + PR_L | SAGE | 4.39±0.93 | 1.70±0.11 | 4.55±1.27 | 1.40±0.15 | 110.625 |
| Contr_l + CrossE_L + PR_L + Triplet_L | ALL | 2.73±0.25 | −1.85 ± 0.34 | 2.91±0.59 | −0.68 ± 0.19 | 167.25 |
| Contr_l + CrossE_L + PR_L + Triplet_L | GAT | 7.46±0.74 | 3.69±0.26 | 8.03±1.49 | 4.20±0.14 | 52.75 |
| Contr_l + CrossE_L + PR_L + Triplet_L | GCN | 6.97±0.42 | 3.86±0.32 | 7.34±0.76 | 3.69±0.61 | 58.75 |
| Contr_l + CrossE_L + PR_L + Triplet_L | GIN | 4.46±0.28 | −1.41 ± 0.04 | 4.42±0.23 | −0.66 ± 0.12 | 144.25 |
| Contr_l + CrossE_L + PR_L + Triplet_L | MPNN | 5.76±0.31 | 3.33±0.15 | 5.31±0.62 | 2.52±0.27 | 79.5 |
| Contr_l + CrossE_L + PR_L + Triplet_L | PAGNN | 2.28±0.17 | 0.00±0.00 | 2.16±0.13 | 0.04±0.01 | 156.25 |
| Contr_l + CrossE_L + PR_L + Triplet_L | SAGE | 5.86±0.17 | 1.94±0.10 | 6.55±0.80 | 1.70±0.15 | 84.375 |
| Contr_l + CrossE_L + Triplet_L | ALL | 4.47±0.69 | 1.31±0.11 | 4.76±0.96 | 0.54±0.14 | 111.75 |
| Contr_l + CrossE_L + Triplet_L | GAT | 9.60±0.62 | 4.14±0.35 | 9.99±0.34 | 4.44±0.22 | 24.125 |
| Contr_l + CrossE_L + Triplet_L | GCN | 8.99±0.97 | 4.30±0.44 | 8.95±0.78 | 4.58±0.20 | 28.875 |
| Contr_l + CrossE_L + Triplet_L | GIN | 6.18±0.52 | −0.41 ± 0.15 | 5.82±0.99 | 0.68±0.06 | 109.75 |
| Contr_l + CrossE_L + Triplet_L | MPNN | 7.08±0.57 | 3.27±0.25 | 6.50±0.53 | 3.85±0.17 | 65.125 |
| Contr_l + CrossE_L + Triplet_L | PAGNN | 3.52±0.70 | −0.02 ± 0.01 | 4.40±0.67 | 0.05±0.04 | 133.75 |
| Contr_l + CrossE_L + Triplet_L | SAGE | 6.60±0.46 | 2.03±0.12 | 8.17±0.89 | 2.25±0.14 | 70.25 |
| Contr_l + PMI_L | ALL | 3.24±0.51 | −2.11 ± 0.08 | 3.67±0.89 | −0.47 ± 0.17 | 161.25 |

Table 34. Results for Dot-Adj Corr (↑) (continued)

| Loss Type | Model | Cora ↓ Citeseer | Cora ↓ Bitcoin | Citeseer ↓ Cora | Citeseer ↓ Bitcoin | Average Rank |
|---|---|---|---|---|---|---|
| Contr_l + PMI_L | GAT | 10.64 ± 1.14 | 4.72±0.33 | 10.67 ± 0.52 | 4.68±0.31 | 8.0 |
| Contr_l + PMI_L | GCN | 9.42±1.08 | 4.63±0.35 | 9.20±0.82 | 4.77±0.45 | 17.125 |
| Contr_l + PMI_L | GIN | 3.58±0.65 | −1.90 ± 0.10 | 3.39±0.42 | −1.06 ± 0.05 | 161.25 |
| Contr_l + PMI_L | MPNN | 8.28±0.49 | 3.59±0.19 | 6.87±0.82 | 4.13±0.13 | 55.75 |
| Contr_l + PMI_L | PAGNN | 2.06±0.15 | −0.05 ± 0.00 | 2.10±0.03 | −0.06 ± 0.00 | 171.25 |
| Contr_l + PMI_L | SAGE | 4.61±0.33 | 0.60±0.15 | 5.85±0.71 | 0.95±0.04 | 103.875 |
| Contr_l + PMI_L + PR_L | ALL | 1.51±0.21 | −3.45 ± 0.15 | 2.46±0.11 | −2.37 ± 0.13 | 192.125 |
| Contr_l + PMI_L + PR_L | GAT | 9.83±1.07 | 4.59±0.23 | 7.47±2.20 | 2.46±0.15 | 40.875 |
| Contr_l + PMI_L + PR_L | GCN | 8.46±0.34 | 4.31±0.40 | 8.49±1.19 | 4.30±0.38 | 39.0 |
| Contr_l + PMI_L + PR_L | GIN | 3.46±0.72 | −1.10 ± 0.13 | 2.62±0.66 | −0.38 ± 0.04 | 161.25 |
| Contr_l + PMI_L + PR_L | MPNN | 8.59±0.40 | 4.07±0.14 | 4.71±1.21 | 1.41±0.12 | 76.0 |
| Contr_l + PMI_L + PR_L | PAGNN | 1.98±0.11 | −0.04 ± 0.01 | 2.20±0.11 | −0.03 ± 0.00 | 166.625 |
| Contr_l + PMI_L + PR_L | SAGE | 4.63±0.23 | 0.61±0.16 | 4.99±0.61 | 0.95±0.10 | 107.75 |
| Contr_l + PMI_L + PR_L + Triplet_L | ALL | 3.11±0.24 | −2.14 ± 0.16 | 2.76±0.29 | −0.98 ± 0.23 | 168.875 |
| Contr_l + PMI_L + PR_L + Triplet_L | GAT | 8.05±1.05 | 4.03±0.50 | 8.08±1.24 | 3.30±0.45 | 53.125 |
| Contr_l + PMI_L + PR_L + Triplet_L | GCN | 9.08±0.70 | 4.72±0.20 | 8.64±0.77 | 4.75±0.09 | 21.875 |
| Contr_l + PMI_L + PR_L + Triplet_L | GIN | 4.08±0.18 | −1.77 ± 0.08 | 3.55±0.27 | −0.86 ± 0.08 | 155.25 |
| Contr_l + PMI_L + PR_L + Triplet_L | MPNN | 8.78±0.37 | 3.91±0.19 | 4.92±0.29 | 1.84±0.27 | 71.625 |

Table 34. Results for Dot-Adj Corr (↑) (continued)

| Loss Type | Model | Cora ↓ Citeseer | Cora ↓ Bitcoin | Citeseer ↓ Cora | Citeseer ↓ Bitcoin | Average Rank |
|---|---|---|---|---|---|---|
| Contr_l + PMI_L + PR_L + Triplet_L | PAGNN | 2.24±0.14 | −0.01 ± 0.01 | 2.46±0.11 | −0.03 ± 0.01 | 156.125 |
| Contr_l + PMI_L + PR_L + Triplet_L | SAGE | 5.07±0.63 | 1.82±0.07 | 6.08±0.68 | 1.68±0.14 | 92.375 |
| Contr_l + PR_L | ALL | 1.59±0.19 | −3.50 ± 0.06 | 2.51±0.42 | −3.24 ± 0.05 | 192.125 |
| Contr_l + PR_L | GAT | 5.03±1.24 | 3.60±0.10 | 5.15±0.54 | 2.46±0.35 | 85.125 |
| Contr_l + PR_L | GCN | 5.14±0.56 | 3.24±0.36 | 5.96±0.93 | 3.40±0.31 | 78.25 |
| Contr_l + PR_L | GIN | 2.13±0.86 | −0.30 ± 0.02 | 2.05±1.03 | −1.15 ± 0.08 | 179.25 |
| Contr_l + PR_L | MPNN | 3.17±0.66 | 1.59±1.17 | 4.67±1.08 | 3.32±0.31 | 107.625 |
| Contr_l + PR_L | PAGNN | 1.07±0.51 | 0.05±0.00 | 0.54±0.37 | 0.04±0.00 | 165.5 |
| Contr_l + PR_L | SAGE | 4.44±1.16 | 1.79±0.10 | 5.26±1.25 | 0.72±0.11 | 104.75 |
| Contr_l + PR_L + Triplet_L | ALL | 2.40±0.56 | −2.43 ± 0.10 | 2.52±0.38 | −1.41 ± 0.15 | 179.25 |
| Contr_l + PR_L + Triplet_L | GAT | 7.31±0.90 | 3.94±0.24 | 8.22±1.31 | 3.18±0.27 | 55.5 |
| Contr_l + PR_L + Triplet_L | GCN | 6.60±0.32 | 3.77±0.44 | 7.65±0.64 | 3.89±0.41 | 58.0 |
| Contr_l + PR_L + Triplet_L | GIN | 4.74±0.34 | −0.90 ± 0.11 | 4.31±0.77 | −1.39 ± 0.13 | 146.75 |
| Contr_l + PR_L + Triplet_L | MPNN | 5.77±1.05 | 3.08±0.10 | 5.01±0.66 | 2.96±0.10 | 81.5 |
| Contr_l + PR_L + Triplet_L | PAGNN | 2.11±0.23 | 0.00±0.00 | 2.30±0.34 | 0.04±0.01 | 157.125 |
| Contr_l + PR_L + Triplet_L | SAGE | 5.16±0.28 | 1.92±0.13 | 6.88±0.56 | 1.76±0.10 | 85.75 |
| Contr_l + Triplet_L | ALL | 4.62±0.20 | 0.39±0.29 | 5.06±1.22 | 0.15±0.21 | 112.5 |
| Contr_l + Triplet_L | GAT | 9.45±0.27 | 4.13±0.42 | 10.08 ± 0.42 | 4.31±0.23 | 26.0 |
| Contr_l + Triplet_L | GCN | 8.60±0.57 | 4.38±0.26 | 9.17±0.58 | 4.92±0.18 | 24.375 |

Table 34. Results for Dot-Adj Corr (↑) (continued)

| Loss Type | Model | Cora ↓ Citeseer | Cora ↓ Bitcoin | Citeseer ↓ Cora | Citeseer ↓ Bitcoin | Average Rank |
|---|---|---|---|---|---|---|
| Contr_l + Triplet_L | GIN | 6.15±0.18 | −0.88 ± 0.12 | 5.98±1.25 | −0.01 ± 0.05 | 115.375 |
| Contr_l + Triplet_L | MPNN | 7.34±0.63 | 3.36±0.43 | 7.65±0.62 | 3.90±0.23 | 57.5 |
| Contr_l + Triplet_L | PAGNN | 3.64±0.67 | 0.02±0.04 | 3.71±0.25 | −0.02 ± 0.06 | 136.25 |
| Contr_l + Triplet_L | SAGE | 6.60±0.54 | 1.78±0.16 | 7.92±0.82 | 1.97±0.09 | 75.75 |
| CrossE_L | ALL | 3.36±1.41 | 2.39±0.11 | −0.44 ± 3.28 | −1.39 ± 0.38 | 156.0 |
| CrossE_L | GAT | 2.97±1.70 | 0.14±0.02 | 2.47±1.66 | 2.37±0.18 | 132.375 |
| CrossE_L | GCN | −2.96 ± 0.64 | −3.83 ± 0.05 | −6.88 ± 0.03 | −4.08 ± 0.01 | 209.75 |
| CrossE_L | GIN | −2.49 ± 0.36 | −3.99 ± 0.00 | −5.53 ± 0.24 | −3.84 ± 0.00 | 209.25 |
| CrossE_L | MPNN | 2.58±0.52 | −2.14 ± 0.28 | 3.28±0.31 | −3.03 ± 0.30 | 176.5 |
| CrossE_L | PAGNN | 0.49±0.00 | 0.00±0.00 | 0.92±0.36 | −0.29 ± 0.01 | 176.5 |
| CrossE_L | SAGE | 0.49±0.00 | 0.04±0.01 | 1.09±1.04 | −0.02 ± 0.00 | 168.5 |
| CrossE_L + PMI_L | ALL | 3.07±0.25 | −2.84 ± 0.06 | 2.70±0.18 | −2.03 ± 0.07 | 178.5 |
| CrossE_L + PMI_L | GAT | 10.67 ± 0.70 | 4.72±0.10 | 10.42 ± 0.79 | 4.80±0.17 | 6.25 |
| CrossE_L + PMI_L | GCN | 8.52±0.64 | 4.28±0.33 | 9.20±0.57 | 4.83±0.23 | 26.5 |
| CrossE_L + PMI_L | GIN | 3.69±0.38 | −1.63 ± 0.09 | 3.34±0.46 | −1.05 ± 0.17 | 158.0 |
| CrossE_L + PMI_L | MPNN | 7.96±0.44 | 3.93±0.23 | 6.46±0.92 | 4.10±0.25 | 57.625 |
| CrossE_L + PMI_L | PAGNN | 2.03±0.24 | 0.05±0.00 | 2.17±0.09 | −0.05 ± 0.01 | 162.125 |
| CrossE_L + PMI_L | SAGE | 4.24±0.23 | 0.25±0.09 | 4.80±0.14 | 0.25±0.06 | 118.0 |
| CrossE_L + PMI_L + PR_L | ALL | 1.23±0.05 | −3.50 ± 0.03 | 2.43±0.15 | −2.29 ± 0.05 | 194.875 |

Continued on next page

Table 34. Results for Dot-Adj Corr (↑) (continued)

| Loss Type | Model | Cora ↓ Citeseer | Cora ↓ Bitcoin | Citeseer ↓ Cora | Citeseer ↓ Bitcoin | Average Rank |
|---|---|---|---|---|---|---|
| CrossE_L + PMI_L + PR_L | GAT | 10.49 ± 0.53 | 4.68±0.19 | 7.07±2.45 | 2.11±0.35 | 40.75 |
| CrossE_L + PMI_L + PR_L | GCN | 8.74±0.45 | 4.63±0.36 | 8.89±0.80 | 4.89±0.24 | 22.375 |
| CrossE_L + PMI_L + PR_L | GIN | 3.23±0.40 | −2.05 ± 0.16 | 2.74±0.70 | −1.19 ± 0.12 | 170.5 |
| CrossE_L + PMI_L + PR_L | MPNN | 8.33±0.58 | 3.89±0.15 | 5.17±1.48 | 3.71±0.23 | 64.5 |
| CrossE_L + PMI_L + PR_L | PAGNN | 2.16±0.22 | −0.05 ± 0.00 | 2.10±0.15 | −0.05 ± 0.00 | 168.0 |
| CrossE_L + PMI_L + PR_L | SAGE | 4.45±0.23 | 0.33±0.08 | 4.75±0.16 | 0.27±0.12 | 116.0 |
| CrossE_L + PMI_L + PR_L + Triplet_L | ALL | 2.75±0.17 | −2.73 ± 0.10 | 2.71±0.53 | −1.74 ± 0.16 | 177.75 |
| CrossE_L + PMI_L + PR_L + Triplet_L | GAT | 10.50 ± 1.51 | 4.56±0.15 | 9.00±1.32 | 4.36±0.28 | 22.875 |
| CrossE_L + PMI_L + PR_L + Triplet_L | GCN | 9.09±0.42 | 4.45±0.27 | 8.52±0.81 | 4.25±0.27 | 32.875 |
| CrossE_L + PMI_L + PR_L + Triplet_L | GIN | 3.62±0.49 | −1.38 ± 0.10 | 3.75±0.85 | −1.55 ± 0.12 | 160.25 |
| CrossE_L + PMI_L + PR_L + Triplet_L | MPNN | 7.76±0.27 | 3.87±0.32 | 5.61±1.07 | 3.63±0.18 | 65.75 |
| CrossE_L + PMI_L + PR_L + Triplet_L | PAGNN | 2.35±0.08 | 0.01±0.00 | 2.28±0.05 | −0.05 ± 0.00 | 157.5 |
| CrossE_L + PMI_L + PR_L + Triplet_L | SAGE | 5.57±0.74 | 1.73±0.11 | 5.97±0.60 | 1.60±0.09 | 93.5 |
| CrossE_L + PMI_L + Triplet_L | ALL | 5.39±0.36 | 0.01±0.15 | 4.55±0.51 | 1.64±0.15 | 111.25 |
| CrossE_L + PMI_L + Triplet_L | GAT | 10.41 ± 0.63 | 4.56±0.39 | 10.29 ± 0.53 | 4.80±0.14 | 11.875 |
| CrossE_L + PMI_L + Triplet_L | GCN | 8.99±0.68 | 4.97±0.23 | 9.31±0.55 | 4.54±0.27 | 18.125 |

Continued on next page

Table 34. Results for Dot-Adj Corr (↑) (continued)

| Loss Type | Model | Cora ↓ Citeseer | Cora ↓ Bitcoin | Citeseer ↓ Cora | Citeseer ↓ Bitcoin | Average Rank |
|-----------|-------|-----------------|----------------|-----------------|--------------------|--------------|
| CrossE_L + PMI_L + Triplet_L | GIN | 4.09±0.28 | −1.72 ± 0.09 | 4.02±0.37 | −1.09 ± 0.17 | 154.0 |
| CrossE_L + PMI_L + Triplet_L | MPNN | 8.07±0.95 | 4.10±0.40 | 6.80±0.30 | 4.05±0.30 | 52.5 |
| CrossE_L + PMI_L + Triplet_L | PAGNN | 2.22±0.14 | −0.01 ± 0.01 | 2.32±0.07 | −0.04 ± 0.01 | 158.875 |
| CrossE_L + PMI_L + Triplet_L | SAGE | 6.19±0.34 | 1.79±0.13 | 6.69±0.59 | 2.06±0.05 | 80.5 |
| CrossE_L + PR_L | ALL | 1.37±0.41 | −3.42 ± 0.07 | 2.72±0.68 | −3.46 ± 0.04 | 191.375 |
| CrossE_L + PR_L | GAT | 4.16±0.90 | 2.79±0.27 | 4.98±0.60 | 2.42±0.20 | 96.0 |
| CrossE_L + PR_L | GCN | 2.87±0.39 | 1.80±0.35 | 4.39±0.92 | 1.73±0.38 | 117.875 |
| CrossE_L + PR_L | GIN | 1.52±0.30 | −0.24 ± 0.02 | 1.42±1.59 | −0.25 ± 0.02 | 177.0 |
| CrossE_L + PR_L | MPNN | 3.76±2.21 | 1.88±0.77 | 2.15±1.30 | 2.68±0.73 | 121.875 |
| CrossE_L + PR_L | PAGNN | 0.86±0.36 | 0.02±0.00 | −0.55 ± 0.21 | −0.05 ± 0.00 | 172.875 |
| CrossE_L + PR_L | SAGE | 2.13±0.22 | 0.61±0.10 | 3.24±0.15 | 0.50±0.07 | 140.125 |
| CrossE_L + PR_L + Triplet_L | ALL | 2.22±0.67 | −3.46 ± 0.14 | 2.16±0.17 | −1.20 ± 0.13 | 187.25 |
| CrossE_L + PR_L + Triplet_L | GAT | 5.70±1.30 | 2.81±0.29 | 7.30±1.48 | 4.16±0.11 | 65.25 |
| CrossE_L + PR_L + Triplet_L | GCN | 6.02±0.70 | 3.81±0.29 | 7.22±0.85 | 4.27±0.43 | 58.0 |
| CrossE_L + PR_L + Triplet_L | GIN | 4.26±0.47 | −1.21 ± 0.19 | 4.37±0.57 | −1.07 ± 0.08 | 147.75 |
| CrossE_L + PR_L + Triplet_L | MPNN | 4.73±0.85 | 3.27±0.33 | 4.42±0.92 | 2.94±0.41 | 91.5 |
| CrossE_L + PR_L + Triplet_L | PAGNN | 2.04±0.33 | 0.03±0.01 | 1.92±0.42 | 0.04±0.00 | 160.625 |
| CrossE_L + PR_L + Triplet_L | SAGE | 6.06±0.76 | 1.96±0.21 | 7.14±0.63 | 1.73±0.12 | 79.5 |

Table 34. Results for Dot-Adj Corr (↑) (continued)

| Loss Type | Model | Cora ↓ Citeseer | Cora ↓ Bitcoin | Citeseer ↓ Cora | Citeseer ↓ Bitcoin | Average Rank |
|---|---|---|---|---|---|---|
| CrossE_L + Triplet_L | ALL | 5.69±0.38 | 1.09±0.09 | 5.26±1.21 | 0.85±0.10 | 100.375 |
| CrossE_L + Triplet_L | GAT | 10.17 ± 0.51 | 4.52±0.19 | 10.11 ± 0.57 | 4.59±0.27 | 16.75 |
| CrossE_L + Triplet_L | GCN | 9.16±0.76 | 4.62±0.31 | 9.37±0.68 | 4.37±0.40 | 22.5 |
| CrossE_L + Triplet_L | GIN | 6.41±0.40 | −1.15 ± 0.10 | 5.54±0.32 | −0.55 ± 0.10 | 124.25 |
| CrossE_L + Triplet_L | MPNN | 8.24±0.79 | 4.03±0.24 | 7.52±0.58 | 3.94±0.29 | 50.375 |
| CrossE_L + Triplet_L | PAGNN | 3.91±0.75 | −0.05 ± 0.00 | 4.08±0.56 | −0.31 ± 0.05 | 143.875 |
| CrossE_L + Triplet_L | SAGE | 8.29±0.10 | 1.94±0.08 | 8.99±0.77 | 1.92±0.05 | 65.125 |
| PMI_L | ALL | 2.57±0.13 | −2.94 ± 0.11 | 2.71±0.24 | −1.96 ± 0.16 | 180.625 |
| PMI_L | GAT | 10.59 ± 0.68 | `4.83±0.31` | 10.12 ± 0.38 | 4.89±0.27 | `4.75` |
| PMI_L | GCN | 8.79±0.96 | 4.61±0.40 | 8.96±0.32 | `5.08±0.57` | 20.75 |
| PMI_L | GIN | 3.36±0.23 | −1.72 ± 0.14 | 3.31±0.38 | −1.19 ± 0.09 | 164.625 |
| PMI_L | MPNN | 8.69±0.92 | 4.14±0.27 | 6.66±0.66 | 3.98±0.41 | 48.125 |
| PMI_L | PAGNN | 2.13±0.17 | −0.03 ± 0.00 | 2.15±0.04 | −0.06 ± 0.01 | 167.5 |
| PMI_L | SAGE | 4.06±0.27 | 0.38±0.11 | 4.35±0.34 | 0.19±0.10 | 123.0 |
| PMI_L + PR_L | ALL | 1.43±0.29 | −3.53 ± 0.01 | 2.41±0.22 | −2.25 ± 0.11 | 194.0 |
| PMI_L + PR_L | GAT | 9.14±1.11 | 3.96±0.36 | 7.43±2.52 | 3.60±0.42 | 45.5 |
| PMI_L + PR_L | GCN | 8.75±0.95 | 4.78±0.18 | 8.79±1.47 | 4.52±0.36 | 24.625 |
| PMI_L + PR_L | GIN | 3.27±0.38 | −1.41 ± 0.14 | 2.85±0.66 | −1.46 ± 0.12 | 167.125 |
| PMI_L + PR_L | MPNN | 8.21±0.79 | 3.96±0.27 | 3.89±0.56 | 1.20±0.15 | 87.625 |
| PMI_L + PR_L | PAGNN | 2.16±0.17 | −0.04 ± 0.01 | 2.21±0.05 | −0.07 ± 0.00 | 166.375 |
| PMI_L + PR_L | SAGE | 4.31±0.16 | 0.32±0.11 | 4.73±0.25 | 0.11±0.09 | 118.5 |

Table 34. Results for Dot-Adj Corr (↑) (continued)

| Loss Type | Model | Cora ↓ Citeseer | Cora ↓ Bitcoin | Citeseer ↓ Cora | Citeseer ↓ Bitcoin | Average Rank |
|---|---|---|---|---|---|---|
| PMI_L + PR_L + Triplet_L | ALL | 2.82±0.12 | −2.82 ± 0.22 | 2.71±0.27 | −1.47 ± 0.19 | 177.125 |
| PMI_L + PR_L + Triplet_L | GAT | 9.33±0.54 | 4.73±0.28 | 9.05±1.43 | 4.12±0.37 | 24.5 |
| PMI_L + PR_L + Triplet_L | GCN | 8.60±0.58 | 4.73±0.43 | 8.87±0.92 | 4.38±0.35 | 28.125 |
| PMI_L + PR_L + Triplet_L | GIN | 3.82±0.33 | −1.66 ± 0.15 | 3.15±0.57 | −1.19 ± 0.11 | 161.75 |
| PMI_L + PR_L + Triplet_L | MPNN | 8.22±0.91 | 4.17±0.15 | 4.46±0.67 | 2.10±0.22 | 74.5 |
| PMI_L + PR_L + Triplet_L | PAGNN | 2.19±0.15 | −0.02 ± 0.01 | 2.30±0.08 | −0.03 ± 0.00 | 159.625 |
| PMI_L + PR_L + Triplet_L | SAGE | 5.64±0.44 | 1.78±0.06 | 6.76±0.73 | 1.60±0.04 | 88.25 |
| PMI_L + Triplet_L | ALL | 5.57±0.29 | −0.40 ± 0.17 | 5.11±0.82 | 1.64±0.15 | 113.125 |
| PMI_L + Triplet_L | GAT | 10.48 ± 0.62 | 4.66±0.23 | 10.84 ± 0.37 | 4.78±0.42 | 9.0 |
| PMI_L + Triplet_L | GCN | 8.42±0.61 | 4.60±0.33 | 9.55±0.70 | 4.62±0.18 | 25.25 |
| PMI_L + Triplet_L | GIN | 4.21±0.21 | −1.44 ± 0.16 | 3.90±0.83 | −0.97 ± 0.14 | 151.0 |
| PMI_L + Triplet_L | MPNN | 8.30±0.77 | 4.02±0.15 | 6.69±0.71 | 4.38±0.34 | 49.25 |
| PMI_L + Triplet_L | PAGNN | 2.07±0.16 | −0.03 ± 0.00 | 2.19±0.07 | −0.05 ± 0.00 | 166.375 |
| PMI_L + Triplet_L | SAGE | 5.79±0.76 | 1.40±0.15 | 6.97±0.29 | 1.96±0.05 | 84.0 |
| PR_L | ALL | 0.72±0.26 | −3.58 ± 0.03 | 1.72±0.90 | −3.54 ± 0.05 | 204.375 |
| PR_L | GAT | 4.43±0.56 | 2.65±0.24 | 5.25±0.31 | 2.67±0.23 | 91.0 |
| PR_L | GCN | 2.58±0.74 | 1.59±0.37 | 4.70±0.37 | 1.89±0.27 | 117.75 |
| PR_L | GIN | 1.37±0.30 | −0.21 ± 0.03 | 1.03±0.94 | −0.28 ± 0.02 | 179.125 |
| PR_L | MPNN | 3.77±2.79 | 2.91±0.55 | 2.67±0.89 | 2.59±0.25 | 112.0 |

Continued on next page

Table 34. Results for Dot-Adj Corr (↑) (continued)

| Loss Type | Model | Cora ↓ Citeseer | Cora ↓ Bitcoin | Citeseer ↓ Cora | Citeseer ↓ Bitcoin | Average Rank |
|---|---|---|---|---|---|---|
| PR_L | PAGNN | 0.59±0.79 | −0.01 ± 0.00 | −0.50 ± 0.13 | 0.01±0.00 | 172.25 |
| PR_L | SAGE | 2.36±0.19 | 0.50±0.07 | 3.46±0.30 | 0.62±0.07 | 135.0 |
| PR_L + Triplet_L | ALL | 1.38±0.31 | −3.39 ± 0.16 | 2.25±0.37 | −3.54 ± 0.09 | 196.5 |
| PR_L + Triplet_L | GAT | 5.51±2.05 | 1.96±0.22 | 4.88±0.66 | 2.39±0.20 | 91.125 |
| PR_L + Triplet_L | GCN | 4.53±0.41 | 2.90±0.38 | 5.49±0.26 | 2.99±0.26 | 85.5 |
| PR_L + Triplet_L | GIN | 1.99±0.46 | −0.40 ± 0.03 | 2.71±0.59 | −1.09 ± 0.03 | 172.875 |
| PR_L + Triplet_L | MPNN | 3.96±2.44 | 3.09±0.25 | 3.70±1.74 | 3.41±0.09 | 100.5 |
| PR_L + Triplet_L | PAGNN | 1.07±0.28 | 0.01±0.00 | 0.06±0.42 | 0.03±0.00 | 168.75 |
| PR_L + Triplet_L | SAGE | 2.59±0.41 | 0.68±0.05 | 4.61±1.37 | 0.59±0.03 | 126.25 |
| Triplet_L | ALL | 5.53±0.60 | 0.28±0.11 | 6.05±0.47 | 1.97±0.22 | 94.625 |
| Triplet_L | GAT | 10.54 ± 0.70 | 4.32±0.30 | 9.97±0.65 | 4.46±0.44 | 19.875 |
| Triplet_L | GCN | 9.53±0.41 | 4.87±0.49 | 9.03±1.02 | 4.80±0.45 | 13.0 |
| Triplet_L | GIN | 7.13±0.31 | −0.79 ± 0.05 | 5.77±0.76 | −0.03 ± 0.14 | 115.0 |
| Triplet_L | MPNN | 8.66±0.87 | 4.04±0.24 | 7.27±0.99 | 4.12±0.47 | 45.875 |
| Triplet_L | PAGNN | 4.01±0.72 | −0.02 ± 0.03 | 4.24±0.42 | 0.07±0.03 | 131.0 |
| Triplet_L | SAGE | 7.95±1.39 | 2.34±0.10 | 9.41±0.64 | 2.50±0.07 | 58.75 |

Table 35. Euclidean-Adj Corr Performance (↑): This table presents models (Loss function and GNN) ranked by their average performance in terms of euclidean-adj corr. Top-ranked results are highlighted in <span style="color:red">red</span>, second-ranked in <span style="color:blue">blue</span>, and third-ranked in <span style="color:green">green</span>.

| Loss Type | Model | Cora ↓ Citeseer | Cora ↓ Bitcoin | Citeseer ↓ Cora | Citeseer ↓ Bitcoin | Average Rank |
|---|---|---|---|---|---|---|
| Contr_l | ALL | 5.66±0.34 | 0.25±0.08 | 6.19±0.99 | −0.14 ± 0.16 | 132.25 |
| Contr_l | GAT | 13.85 ± 1.55 | 5.45±0.56 | 13.84 ± 0.70 | 6.11±0.17 | 25.75 |
| Contr_l | GCN | 12.89 ± 0.98 | 6.21±0.35 | 13.49 ± 1.04 | 6.38±0.36 | 23.75 |
| Contr_l | GIN | 8.45±0.42 | −0.00 ± 0.08 | 7.83±1.18 | 1.21±0.11 | 105.875 |
| Contr_l | MPNN | 10.67 ± 0.78 | 4.65±0.39 | 9.00±1.51 | 4.78±0.28 | 67.25 |
| Contr_l | PAGNN | 4.65±0.34 | −0.10 ± 0.02 | 4.88±0.74 | −0.07 ± 0.01 | 149.25 |
| Contr_l | SAGE | 8.44±1.07 | 2.06±0.14 | 10.73 ± 1.21 | 2.27±0.16 | 81.625 |
| Contr_l + CrossE_L | ALL | 6.10±0.58 | 1.71±0.12 | 5.70±0.91 | 0.68±0.22 | 115.75 |
| Contr_l + CrossE_L | GAT | 13.85 ± 0.96 | 5.53±0.21 | 14.11 ± 0.88 | 5.69±0.32 | 28.0 |
| Contr_l + CrossE_L | GCN | 13.04 ± 0.59 | 6.16±0.47 | 13.28 ± 1.03 | 6.36±0.34 | 25.125 |
| Contr_l + CrossE_L | GIN | 6.74±0.49 | −0.87 ± 0.17 | 7.47±0.64 | 0.18±0.05 | 129.5 |
| Contr_l + CrossE_L | MPNN | 10.69 ± 1.09 | 4.71±0.33 | 9.20±0.66 | 4.94±0.33 | 65.0 |
| Contr_l + CrossE_L | PAGNN | 4.45±0.25 | −0.05 ± 0.01 | 4.56±0.26 | −0.14 ± 0.01 | 152.0 |
| Contr_l + CrossE_L | SAGE | 9.15±0.65 | 1.53±0.06 | 10.01 ± 0.54 | 1.23±0.03 | 87.875 |
| Contr_l + CrossE_L + PMI_L | ALL | 4.19±0.65 | −0.69 ± 0.27 | 4.66±0.98 | −1.14 ± 0.07 | 168.75 |

<div align="right">Continued on next page</div>

Table 35. Results for Euclidean-Adj Corr (↑) (continued)

| Loss Type | Model | Cora ↓ Citeseer | Cora ↓ Bitcoin | Citeseer ↓ Cora | Citeseer ↓ Bitcoin | Average Rank |
|---|---|---|---|---|---|---|
| Contr_l + CrossE_L + PMI_L | GAT | 16.20 ± 0.56 | 6.21±0.31 | 14.66 ± 1.02 | 6.29±0.35 | 11.625 |
| Contr_l + CrossE_L + PMI_L | GCN | 13.63 ± 0.65 | 6.55±0.15 | 12.56 ± 0.95 | 6.31±0.23 | 22.375 |
| Contr_l + CrossE_L + PMI_L | GIN | 5.39±0.58 | −0.66 ± 0.14 | 4.59±0.41 | −0.68 ± 0.05 | 159.125 |
| Contr_l + CrossE_L + PMI_L | MPNN | 13.20 ± 0.78 | 5.58±0.38 | 8.88±0.54 | 5.48±0.25 | 47.375 |
| Contr_l + CrossE_L + PMI_L | PAGNN | 2.20±0.11 | −0.10 ± 0.01 | 2.58±0.10 | −0.11 ± 0.01 | 175.875 |
| Contr_l + CrossE_L + PMI_L | SAGE | 5.38±0.27 | 0.36±0.07 | 5.91±0.54 | 0.61±0.10 | 123.875 |
| Contr_l + CrossE_L + PMI_L + PR_L | ALL | 1.66±0.63 | −3.35 ± 0.02 | 3.32±0.17 | −1.66 ± 0.07 | 195.375 |
| Contr_l + CrossE_L + PMI_L + PR_L | GAT | 14.92 ± 1.66 | 6.08±0.62 | 14.78 ± 0.89 | 5.98±0.81 | 19.25 |
| Contr_l + CrossE_L + PMI_L + PR_L | GCN | 13.48 ± 0.90 | 6.55±0.28 | 12.52 ± 1.31 | 6.22±0.57 | 24.125 |
| Contr_l + CrossE_L + PMI_L + PR_L | GIN | 4.21±0.28 | −0.91 ± 0.20 | 3.74±0.81 | −0.53 ± 0.04 | 171.625 |
| Contr_l + CrossE_L + PMI_L + PR_L | MPNN | 12.86 ± 0.70 | 5.05±0.41 | 6.80±1.48 | 3.35±0.40 | 74.5 |
| Contr_l + CrossE_L + PMI_L + PR_L | PAGNN | 2.48±0.24 | −0.09 ± 0.00 | 2.72±0.15 | −0.09 ± 0.01 | 169.25 |
| Contr_l + CrossE_L + PMI_L + PR_L | SAGE | 5.09±0.35 | 0.12±0.05 | 5.46±0.59 | 0.28±0.17 | 131.125 |
| Contr_l + CrossE_L + PMI_L + PR_L + Triplet_L | ALL | 3.43±0.30 | −1.68 ± 0.15 | 4.34±0.35 | −0.97 ± 0.14 | 178.25 |
| Contr_l + CrossE_L + PMI_L + PR_L + Triplet_L | GAT | 15.23 ± 0.99 | 5.83±0.62 | 13.22 ± 0.84 | 5.99±0.45 | 25.125 |

Table 35. Results for Euclidean-Adj Corr (↑) (continued)

| Loss Type | Model | Cora ↓ Citeseer | Cora ↓ Bitcoin | Citeseer ↓ Cora | Citeseer ↓ Bitcoin | Average Rank |
|---|---|---|---|---|---|---|
| Contr_l + CrossE_L + PMI_L + PR_L + Triplet_L | GCN | 13.69 ± 1.08 | 6.48±0.31 | 12.82 ± 0.73 | 6.62±0.32 | 18.75 |
| Contr_l + CrossE_L + PMI_L + PR_L + Triplet_L | GIN | 4.78±0.37 | −1.08 ± 0.09 | 5.55±0.38 | −0.76 ± 0.07 | 163.375 |
| Contr_l + CrossE_L + PMI_L + PR_L + Triplet_L | MPNN | 12.68 ± 0.35 | 5.12±0.19 | 7.59±1.61 | 3.63±0.20 | 69.875 |
| Contr_l + CrossE_L + PMI_L + PR_L + Triplet_L | PAGNN | 2.46±0.17 | −0.09 ± 0.00 | 2.89±0.13 | −0.13 ± 0.00 | 171.25 |
| Contr_l + CrossE_L + PMI_L + PR_L + Triplet_L | SAGE | 6.46±1.13 | 1.75±0.11 | 8.29±0.33 | 1.55±0.12 | 100.375 |
| Contr_l + CrossE_L + PMI_L + Triplet_L | ALL | 7.55±0.73 | 0.75±0.33 | 6.95±0.93 | 2.03±0.34 | 103.5 |
| Contr_l + CrossE_L + PMI_L + Triplet_L | GAT | 15.56 ± 0.63 | 6.30±0.31 | 14.44 ± 0.80 | 6.10±0.25 | 14.75 |
| Contr_l + CrossE_L + PMI_L + Triplet_L | GCN | 13.76 ± 0.76 | 6.63±0.34 | 12.44 ± 1.13 | 6.73±0.61 | 17.75 |
| Contr_l + CrossE_L + PMI_L + Triplet_L | GIN | 5.57±0.69 | −0.75 ± 0.04 | 5.57±1.03 | −0.49 ± 0.06 | 152.5 |
| Contr_l + CrossE_L + PMI_L + Triplet_L | MPNN | 12.65 ± 0.64 | 5.50±0.24 | 9.37±0.93 | 5.25±0.28 | 51.5 |
| Contr_l + CrossE_L + PMI_L + Triplet_L | PAGNN | 2.44±0.08 | −0.10 ± 0.00 | 2.75±0.09 | −0.14 ± 0.01 | 174.625 |
| Contr_l + CrossE_L + PMI_L + Triplet_L | SAGE | 6.99±0.70 | 1.46±0.12 | 7.80±0.55 | 1.79±0.09 | 101.75 |
| Contr_l + CrossE_L + PR_L | ALL | 1.83±0.26 | −2.77 ± 0.08 | 2.73±0.33 | −2.36 ± 0.12 | 198.0 |
| Contr_l + CrossE_L + PR_L | GAT | 8.82±2.29 | 5.04±0.17 | 9.75±0.91 | 3.97±0.38 | 69.0 |

Continued on next page

Table 35. Results for Euclidean-Adj Corr (↑) (continued)

| Loss Type | Model | Cora ↓ Citeseer | Cora ↓ Bitcoin | Citeseer ↓ Cora | Citeseer ↓ Bitcoin | Average Rank |
|---|---|---|---|---|---|---|
| Contr_l + CrossE_L + PR_L | GCN | 8.50±0.51 | 4.25±0.34 | 9.35±0.41 | 4.58±0.41 | 74.25 |
| Contr_l + CrossE_L + PR_L | GIN | 4.12±0.26 | −0.07 ± 0.03 | 4.02±0.46 | −0.15 ± 0.06 | 158.75 |
| Contr_l + CrossE_L + PR_L | MPNN | 5.97±2.74 | 4.31±0.23 | 6.59±1.74 | 4.40±0.14 | 94.75 |
| Contr_l + CrossE_L + PR_L | PAGNN | 2.11±0.64 | −0.01 ± 0.00 | 0.89±0.93 | −0.00 ± 0.01 | 166.5 |
| Contr_l + CrossE_L + PR_L | SAGE | 6.23±1.50 | 1.72±0.11 | 6.26±1.78 | 1.43±0.14 | 110.75 |
| Contr_l + CrossE_L + PR_L + Triplet_L | ALL | 3.88±0.35 | −0.97 ± 0.10 | 4.60±0.80 | −0.02 ± 0.16 | 158.25 |
| Contr_l + CrossE_L + PR_L + Triplet_L | GAT | 11.62 ± 1.03 | 4.98±0.35 | 11.88 ± 1.83 | 5.61±0.17 | 51.75 |
| Contr_l + CrossE_L + PR_L + Triplet_L | GCN | 10.89 ± 0.61 | 5.50±0.44 | 10.93 ± 0.97 | 5.31±0.80 | 50.875 |
| Contr_l + CrossE_L + PR_L + Triplet_L | GIN | 6.49±0.41 | −0.71 ± 0.05 | 6.57±0.32 | −0.09 ± 0.10 | 137.875 |
| Contr_l + CrossE_L + PR_L + Triplet_L | MPNN | 9.32±0.42 | 4.55±0.17 | 7.83±0.76 | 3.75±0.37 | 77.5 |
| Contr_l + CrossE_L + PR_L + Triplet_L | PAGNN | 2.93±0.22 | −0.05 ± 0.00 | 3.02±0.13 | −0.02 ± 0.01 | 159.375 |
| Contr_l + CrossE_L + PR_L + Triplet_L | SAGE | 8.59±0.27 | 1.97±0.08 | 9.04±1.10 | 1.80±0.14 | 87.25 |
| Contr_l + CrossE_L + Triplet_L | ALL | 6.95±0.81 | 1.65±0.12 | 7.07±1.11 | 0.86±0.12 | 107.5 |
| Contr_l + CrossE_L + Triplet_L | GAT | 14.58 ± 0.89 | 5.53±0.45 | 14.31 ± 0.45 | 5.89±0.28 | 24.875 |
| Contr_l + CrossE_L + Triplet_L | GCN | 13.73 ± 1.38 | 6.01±0.59 | 13.00 ± 1.07 | 6.35±0.29 | 25.0 |
| Contr_l + CrossE_L + Triplet_L | GIN | 9.03±0.71 | 0.20±0.12 | 8.29±1.18 | 1.02±0.05 | 100.25 |

Table 35. Results for Euclidean-Adj Corr (↑) (continued)

| Loss Type | Model | Cora ↓ Citeseer | Cora ↓ Bitcoin | Citeseer ↓ Cora | Citeseer ↓ Bitcoin | Average Rank |
|---|---|---|---|---|---|---|
| Contr_l + CrossE_L + Triplet_L | MPNN | 11.04 ± 0.82 | 4.38±0.30 | 9.37±0.66 | 5.09±0.23 | 64.375 |
| Contr_l + CrossE_L + Triplet_L | PAGNN | 4.43±1.07 | −0.08 ± 0.01 | 6.32±0.76 | −0.12 ± 0.03 | 145.75 |
| Contr_l + CrossE_L + Triplet_L | SAGE | 9.73±0.70 | 1.96±0.08 | 11.48 ± 1.14 | 2.21±0.09 | 76.0 |
| Contr_l + PMI_L | ALL | 4.65±1.09 | −1.18 ± 0.04 | 5.55±1.12 | 0.53±0.22 | 148.0 |
| Contr_l + PMI_L | GAT | <span style="background-color:#7ee787">15.91 ± 1.60</span> | 6.23±0.43 | <span style="background-color:#8f8fef">15.13 ± 0.65</span> | 6.19±0.41 | 11.875 |
| Contr_l + PMI_L | GCN | 14.22 ± 1.56 | 6.39±0.49 | 13.23 ± 1.12 | 6.59±0.62 | 15.5 |
| Contr_l + PMI_L | GIN | 5.02±0.94 | −0.98 ± 0.08 | 4.71±0.49 | −0.60 ± 0.04 | 162.375 |
| Contr_l + PMI_L | MPNN | 12.56 ± 0.68 | 4.66±0.23 | 9.76±0.93 | 5.35±0.14 | 56.375 |
| Contr_l + PMI_L | PAGNN | 2.37±0.21 | −0.09 ± 0.00 | 2.52±0.06 | −0.11 ± 0.00 | 174.375 |
| Contr_l + PMI_L | SAGE | 5.24±0.40 | 0.49±0.13 | 6.88±0.91 | 0.80±0.05 | 118.0 |
| Contr_l + PMI_L + PR_L | ALL | 1.63±0.25 | −3.06 ± 0.12 | 3.36±0.17 | −1.71 ± 0.08 | 195.0 |
| Contr_l + PMI_L + PR_L | GAT | 14.81 ± 1.50 | 6.11±0.30 | 10.88 ± 2.99 | 3.44±0.11 | 42.625 |
| Contr_l + PMI_L + PR_L | GCN | 12.95 ± 0.48 | 6.06±0.56 | 12.45 ± 1.57 | 5.99±0.52 | 35.375 |
| Contr_l + PMI_L + PR_L | GIN | 4.83±0.94 | −0.42 ± 0.10 | 3.54±0.87 | −0.46 ± 0.06 | 164.0 |
| Contr_l + PMI_L + PR_L | MPNN | 13.07 ± 0.54 | 5.30±0.18 | 6.72±1.64 | 2.96±0.14 | 73.0 |
| Contr_l + PMI_L + PR_L | PAGNN | 2.28±0.15 | −0.09 ± 0.01 | 2.66±0.17 | −0.10 ± 0.01 | 172.25 |
| Contr_l + PMI_L + PR_L | SAGE | 5.23±0.24 | 0.50±0.16 | 6.00±0.59 | 0.88±0.10 | 121.0 |

Table 35. Results for Euclidean-Adj Corr (↑) (continued)

| Loss Type | Model | Cora ↓ Citeseer | Cora ↓ Bitcoin | Citeseer ↓ Cora | Citeseer ↓ Bitcoin | Average Rank |
|---|---|---|---|---|---|---|
| Contr_l + PMI_L + PR_L + Triplet_L | ALL | 4.12±0.45 | −1.31 ± 0.11 | 4.41±0.43 | −0.31 ± 0.21 | 168.25 |
| Contr_l + PMI_L + PR_L + Triplet_L | GAT | 12.34 ± 1.49 | 5.41±0.62 | 11.66 ± 1.71 | 4.42±0.61 | 53.125 |
| Contr_l + PMI_L + PR_L + Triplet_L | GCN | 13.78 ± 0.95 | 6.56±0.29 | 12.60 ± 1.04 | 6.54±0.15 | 18.0 |
| Contr_l + PMI_L + PR_L + Triplet_L | GIN | 5.55±0.13 | −0.91 ± 0.06 | 5.28±0.31 | −0.51 ± 0.09 | 156.125 |
| Contr_l + PMI_L + PR_L + Triplet_L | MPNN | 13.32 ± 0.44 | 5.09±0.24 | 7.10±0.37 | 3.13±0.32 | 70.375 |
| Contr_l + PMI_L + PR_L + Triplet_L | PAGNN | 2.60±0.18 | −0.07 ± 0.00 | 3.06±0.16 | −0.11 ± 0.01 | 165.75 |
| Contr_l + PMI_L + PR_L + Triplet_L | SAGE | 6.96±0.79 | 1.91±0.09 | 8.17±0.82 | 1.66±0.11 | 97.5 |
| Contr_l + PR_L | ALL | 2.03±0.32 | −2.37 ± 0.08 | 3.38±0.61 | −1.98 ± 0.04 | 192.75 |
| Contr_l + PR_L | GAT | 8.50±1.52 | 4.92±0.13 | 8.69±0.45 | 3.65±0.38 | 76.5 |
| Contr_l + PR_L | GCN | 8.50±0.68 | 4.75±0.43 | 9.27±1.07 | 4.96±0.39 | 69.25 |
| Contr_l + PR_L | GIN | 3.87±1.18 | −0.06 ± 0.01 | 3.32±1.11 | −0.66 ± 0.05 | 166.125 |
| Contr_l + PR_L | MPNN | 5.79±0.86 | 3.63±0.88 | 7.15±1.18 | 4.64±0.19 | 92.5 |
| Contr_l + PR_L | PAGNN | 1.71±0.61 | 0.00±0.00 | 1.08±0.37 | 0.01±0.00 | 166.25 |
| Contr_l + PR_L | SAGE | 6.33±1.73 | 1.84±0.08 | 7.28±1.86 | 0.70±0.13 | 108.0 |
| Contr_l + PR_L + Triplet_L | ALL | 3.45±1.19 | −1.40 ± 0.12 | 4.06±0.71 | −0.70 ± 0.11 | 176.5 |
| Contr_l + PR_L + Triplet_L | GAT | 11.41 ± 1.23 | 5.33±0.31 | 12.10 ± 1.65 | 4.44±0.32 | 54.875 |
| Contr_l + PR_L + Triplet_L | GCN | 10.36 ± 0.42 | 5.39±0.61 | 11.42 ± 0.81 | 5.56±0.53 | 50.625 |
| Contr_l + PR_L + Triplet_L | GIN | 6.97±0.60 | −0.19 ± 0.08 | 6.28±1.02 | −0.68 ± 0.10 | 143.875 |

Table 35. Results for Euclidean-Adj Corr (↑) (continued)

| Loss Type | Model | Cora ↓ Citeseer | Cora ↓ Bitcoin | Citeseer ↓ Cora | Citeseer ↓ Bitcoin | Average Rank |
|---|---|---|---|---|---|---|
| Contr_l + PR_L + Triplet_L | MPNN | 9.21±1.54 | 4.29±0.04 | 7.45±0.85 | 4.14±0.11 | 81.75 |
| Contr_l + PR_L + Triplet_L | PAGNN | 2.72±0.30 | −0.03 ± 0.01 | 3.25±0.65 | −0.04 ± 0.00 | 159.125 |
| Contr_l + PR_L + Triplet_L | SAGE | 7.64±0.34 | 2.02±0.09 | 9.52±0.74 | 1.86±0.09 | 87.625 |
| Contr_l + Triplet_L | ALL | 7.26±0.27 | 0.73±0.23 | 7.48±1.43 | 0.55±0.17 | 108.625 |
| Contr_l + Triplet_L | GAT | 14.32 ± 0.37 | 5.55±0.54 | 14.40 ± 0.55 | 5.69±0.31 | 25.625 |
| Contr_l + Triplet_L | GCN | 13.15 ± 0.82 | 6.11±0.35 | 13.28 ± 0.81 | 6.82±0.24 | 21.5 |
| Contr_l + Triplet_L | GIN | 8.94±0.21 | −0.24 ± 0.10 | 8.51±1.60 | 0.44±0.04 | 113.5 |
| Contr_l + Triplet_L | MPNN | 11.42 ± 0.88 | 4.46±0.56 | 10.77 ± 0.73 | 5.11±0.27 | 59.5 |
| Contr_l + Triplet_L | PAGNN | 4.63±1.01 | −0.06 ± 0.02 | 5.59±0.34 | −0.19 ± 0.05 | 147.875 |
| Contr_l + Triplet_L | SAGE | 9.80±0.68 | 1.59±0.08 | 11.17 ± 1.06 | 1.81±0.05 | 80.25 |
| CrossE_L | ALL | 5.40±2.28 | 3.21±0.14 | 0.37±3.35 | 0.71±0.27 | 134.0 |
| CrossE_L | GAT | 4.79±2.79 | 0.36±0.08 | 4.38±2.63 | 3.49±0.08 | 124.625 |
| CrossE_L | GCN | −1.03 ± 1.09 | −2.29 ± 0.19 | −6.04 ± 0.06 | −3.49 ± 0.05 | 206.625 |
| CrossE_L | GIN | −2.20 ± 0.57 | −3.84 ± 0.01 | −6.20 ± 0.12 | −4.13 ± 0.02 | 210.0 |
| CrossE_L | MPNN | 4.52±0.46 | −1.23 ± 0.13 | 5.41±0.29 | −1.93 ± 0.16 | 170.25 |
| CrossE_L | PAGNN | 0.56±0.02 | 0.00±0.00 | 2.06±0.32 | −0.20 ± 0.01 | 177.125 |
| CrossE_L | SAGE | 0.56±0.02 | 0.02±0.00 | 1.90±1.22 | −0.01 ± 0.00 | 168.875 |
| CrossE_L + PMI_L | ALL | 4.01±0.40 | −1.67 ± 0.06 | 3.96±0.30 | −1.18 ± 0.03 | 179.25 |

Table 35. Results for Euclidean-Adj Corr (↑) (continued)

| Loss Type | Model | Cora ↓ Citeseer | Cora ↓ Bitcoin | Citeseer ↓ Cora | Citeseer ↓ Bitcoin | Average Rank |
|---|---|---|---|---|---|---|
| CrossE_L + PMI_L | GAT | 15.94 ± 0.98 | 6.29±0.13 | 14.84 ± 1.03 | 6.38±0.22 | 8.375 |
| CrossE_L + PMI_L | GCN | 12.95 ± 0.93 | 5.96±0.46 | 13.25 ± 0.83 | 6.65±0.32 | 26.0 |
| CrossE_L + PMI_L | GIN | 5.01±0.54 | −0.83 ± 0.07 | 4.41±0.56 | −0.61 ± 0.12 | 163.125 |
| CrossE_L + PMI_L | MPNN | 12.07 ± 0.61 | 5.12±0.28 | 9.25±1.18 | 5.33±0.29 | 57.875 |
| CrossE_L + PMI_L | PAGNN | 2.32±0.33 | −0.04 ± 0.00 | 2.61±0.12 | −0.10 ± 0.01 | 168.5 |
| CrossE_L + PMI_L | SAGE | 4.81±0.24 | 0.12±0.08 | 5.28±0.16 | 0.16±0.05 | 134.25 |
| CrossE_L + PMI_L + PR_L | ALL | 1.28±0.07 | −3.07 ± 0.03 | 3.29±0.25 | −1.72 ± 0.05 | 197.0 |
| CrossE_L + PMI_L + PR_L | GAT | 15.70 ± 0.77 | 6.23±0.25 | 10.42 ± 3.14 | 3.12±0.38 | 41.875 |
| CrossE_L + PMI_L + PR_L | GCN | 13.26 ± 0.65 | 6.43±0.47 | 12.85 ± 1.03 | 6.74±0.33 | 19.25 |
| CrossE_L + PMI_L + PR_L | GIN | 4.45±0.56 | −1.16 ± 0.08 | 3.69±1.10 | −0.64 ± 0.12 | 173.125 |
| CrossE_L + PMI_L + PR_L | MPNN | 12.63 ± 0.87 | 5.02±0.19 | 7.39±2.15 | 4.84±0.28 | 68.25 |
| CrossE_L + PMI_L + PR_L | PAGNN | 2.49±0.31 | −0.11 ± 0.01 | 2.52±0.23 | −0.10 ± 0.00 | 173.875 |
| CrossE_L + PMI_L + PR_L | SAGE | 5.02±0.27 | 0.22±0.08 | 5.25±0.15 | 0.14±0.11 | 132.625 |
| CrossE_L + PMI_L + PR_L + Triplet_L | ALL | 3.47±0.26 | −1.80 ± 0.05 | 4.05±0.78 | −1.05 ± 0.06 | 179.5 |
| CrossE_L + PMI_L + PR_L + Triplet_L | GAT | 15.76 ± 2.06 | 6.07±0.16 | 12.95 ± 1.76 | 5.77±0.39 | 24.5 |
| CrossE_L + PMI_L + PR_L + Triplet_L | GCN | 13.75 ± 0.65 | 6.19±0.39 | 12.37 ± 1.06 | 5.92±0.33 | 30.25 |
| CrossE_L + PMI_L + PR_L + Triplet_L | GIN | 4.98±0.62 | −0.64 ± 0.08 | 5.24±1.11 | −0.94 ± 0.06 | 161.625 |

Table 35. Results for Euclidean-Adj Corr (↑) (continued)

| Loss Type | Model | Cora ↓ Citeseer | Cora ↓ Bitcoin | Citeseer ↓ Cora | Citeseer ↓ Bitcoin | Average Rank |
|---|---|---|---|---|---|---|
| CrossE_L + PMI_L + PR_L + Triplet_L | MPNN | 11.88 ± 0.42 | 5.03±0.41 | 8.03±1.44 | 4.80±0.22 | 67.25 |
| CrossE_L + PMI_L + PR_L + Triplet_L | PAGNN | 2.76±0.12 | −0.06 ± 0.01 | 2.80±0.07 | −0.13 ± 0.00 | 167.125 |
| CrossE_L + PMI_L + PR_L + Triplet_L | SAGE | 6.83±1.03 | 1.69±0.08 | 7.48±0.64 | 1.49±0.10 | 103.375 |
| CrossE_L + PMI_L + Triplet_L | ALL | 8.31±0.45 | 1.23±0.15 | 7.06±0.64 | 2.60±0.15 | 100.25 |
| CrossE_L + PMI_L + Triplet_L | GAT | 15.57 ± 0.93 | 6.06±0.52 | 14.60 ± 0.72 | 6.33±0.20 | 15.375 |
| CrossE_L + PMI_L + Triplet_L | GCN | 13.61 ± 0.98 | 6.88±0.30 | 13.42 ± 0.69 | 6.26±0.41 | 17.25 |
| CrossE_L + PMI_L + Triplet_L | GIN | 5.59±0.39 | −0.88 ± 0.05 | 5.68±0.48 | −0.48 ± 0.12 | 152.0 |
| CrossE_L + PMI_L + Triplet_L | MPNN | 12.25 ± 1.30 | 5.28±0.51 | 9.68±0.36 | 5.29±0.40 | 54.75 |
| CrossE_L + PMI_L + Triplet_L | PAGNN | 2.56±0.17 | −0.07 ± 0.01 | 2.85±0.12 | −0.08 ± 0.00 | 164.875 |
| CrossE_L + PMI_L + Triplet_L | SAGE | 7.86±0.41 | 1.77±0.12 | 8.45±0.68 | 2.02±0.06 | 92.25 |
| CrossE_L + PR_L | ALL | 1.76±0.50 | −2.32 ± 0.07 | 3.63±1.01 | −2.44 ± 0.06 | 193.0 |
| CrossE_L + PR_L | GAT | 7.39±0.90 | 4.04±0.24 | 8.40±0.45 | 3.68±0.23 | 84.375 |
| CrossE_L + PR_L | GCN | 5.46±0.46 | 3.22±0.47 | 7.76±0.94 | 3.13±0.55 | 97.875 |
| CrossE_L + PR_L | GIN | 3.28±0.44 | −0.04 ± 0.02 | 2.40±2.02 | −0.04 ± 0.02 | 163.0 |
| CrossE_L + PR_L | MPNN | 6.62±3.10 | 3.95±0.32 | 4.27±1.63 | 4.44±0.35 | 103.75 |
| CrossE_L + PR_L | PAGNN | 1.55±0.48 | −0.01 ± 0.00 | −0.09 ± 0.40 | −0.05 ± 0.00 | 172.0 |
| CrossE_L + PR_L | SAGE | 3.00±0.17 | 0.60±0.10 | 3.97±0.19 | 0.47±0.08 | 144.375 |
| CrossE_L + PR_L + Triplet_L | ALL | 3.02±1.33 | −2.61 ± 0.09 | 3.14±0.48 | −0.48 ± 0.09 | 182.25 |

Table 35. Results for Euclidean-Adj Corr (↑) (continued)

| Loss Type | Model | Cora ↓ Citeseer | Cora ↓ Bitcoin | Citeseer ↓ Cora | Citeseer ↓ Bitcoin | Average Rank |
|---|---|---|---|---|---|---|
| CrossE_L + PR_L + Triplet_L | GAT | 9.24±1.74 | 4.04±0.31 | 11.01 ± 1.70 | 5.56±0.15 | 61.25 |
| CrossE_L + PR_L + Triplet_L | GCN | 9.63±0.97 | 5.43±0.38 | 10.84 ± 0.96 | 6.06±0.55 | 48.75 |
| CrossE_L + PR_L + Triplet_L | GIN | 6.43±0.73 | −0.52 ± 0.14 | 6.36±0.72 | −0.42 ± 0.07 | 143.5 |
| CrossE_L + PR_L + Triplet_L | MPNN | 8.08±1.16 | 4.53±0.29 | 6.78±1.27 | 4.44±0.21 | 85.75 |
| CrossE_L + PR_L + Triplet_L | PAGNN | 2.80±0.40 | −0.02 ± 0.00 | 2.57±0.60 | −0.02 ± 0.01 | 162.0 |
| CrossE_L + PR_L + Triplet_L | SAGE | 8.75±1.15 | 2.03±0.20 | 9.80±0.96 | 1.70±0.09 | 83.75 |
| CrossE_L + Triplet_L | ALL | 8.93±0.45 | 1.55±0.08 | 7.82±1.44 | 1.23±0.07 | 97.625 |
| CrossE_L + Triplet_L | GAT | 15.30 ± 0.73 | 5.97±0.23 | 14.49 ± 0.76 | 6.05±0.37 | 20.0 |
| CrossE_L + Triplet_L | GCN | 13.93 ± 1.08 | 6.41±0.39 | 13.58 ± 0.87 | 6.09±0.55 | 18.25 |
| CrossE_L + Triplet_L | GIN | 9.30±0.44 | −0.40 ± 0.07 | 7.91±0.44 | 0.03±0.07 | 116.5 |
| CrossE_L + Triplet_L | MPNN | 12.58 ± 1.13 | 5.26±0.30 | 10.63 ± 0.72 | 5.05±0.34 | 53.125 |
| CrossE_L + Triplet_L | PAGNN | 5.06±1.12 | −0.09 ± 0.01 | 5.95±0.70 | −0.43 ± 0.05 | 147.625 |
| CrossE_L + Triplet_L | SAGE | 11.91 ± 0.20 | 1.64±0.07 | 12.23 ± 0.95 | 1.60±0.04 | 76.75 |
| PMI_L | ALL | 3.22±0.23 | −1.87 ± 0.09 | 4.02±0.39 | −0.93 ± 0.08 | 180.375 |
| PMI_L | GAT | 15.81 ± 0.96 | 6.40±0.41 | 14.41 ± 0.53 | 6.45±0.37 | 9.125 |
| PMI_L | GCN | 13.35 ± 1.36 | 6.38±0.49 | 12.89 ± 0.40 | 7.01±0.77 | 19.0 |
| PMI_L | GIN | 4.63±0.34 | −1.00 ± 0.12 | 4.46±0.47 | −0.79 ± 0.07 | 168.875 |

Table 35. Results for Euclidean-Adj Corr (↑) (continued)

| Loss Type | Model | Cora ↓ Citeseer | Cora ↓ Bitcoin | Citeseer ↓ Cora | Citeseer ↓ Bitcoin | Average Rank |
|---|---|---|---|---|---|---|
| PMI_L | MPNN | 13.14 ± 1.29 | 5.33±0.36 | 9.52±0.76 | 5.16±0.51 | 50.5 |
| PMI_L | PAGNN | 2.47±0.23 | −0.08 ± 0.00 | 2.60±0.05 | −0.11 ± 0.01 | 171.625 |
| PMI_L | SAGE | 4.60±0.30 | 0.24±0.09 | 4.80±0.37 | 0.09±0.09 | 137.0 |
| PMI_L + PR_L | ALL | 1.54±0.38 | −3.12 ± 0.07 | 3.21±0.43 | −1.68 ± 0.05 | 197.0 |
| PMI_L + PR_L | GAT | 13.85 ± 1.54 | 5.36±0.43 | 10.58 ± 3.73 | 4.89±0.56 | 45.25 |
| PMI_L + PR_L | GCN | 13.28 ± 1.39 | 6.65±0.27 | 12.74 ± 1.95 | 6.24±0.53 | 22.5 |
| PMI_L + PR_L | GIN | 4.39±0.52 | −0.73 ± 0.10 | 3.92±0.98 | −0.90 ± 0.10 | 172.5 |
| PMI_L + PR_L | MPNN | 12.43 ± 1.12 | 5.13±0.34 | 5.63±0.84 | 2.77±0.29 | 82.25 |
| PMI_L + PR_L | PAGNN | 2.56±0.23 | −0.09 ± 0.01 | 2.64±0.08 | −0.12 ± 0.00 | 171.875 |
| PMI_L + PR_L | SAGE | 4.87±0.19 | 0.20±0.09 | 5.21±0.28 | −0.00 ± 0.09 | 135.625 |
| PMI_L + PR_L + Triplet_L | ALL | 3.60±0.19 | −2.09 ± 0.25 | 4.17±0.25 | −0.86 ± 0.15 | 177.5 |
| PMI_L + PR_L + Triplet_L | GAT | 14.12 ± 0.73 | 6.27±0.38 | 13.04 ± 1.83 | 5.48±0.47 | 26.375 |
| PMI_L + PR_L + Triplet_L | GCN | 13.11 ± 0.83 | 6.59±0.58 | 12.84 ± 1.17 | 6.07±0.46 | 25.5 |
| PMI_L + PR_L + Triplet_L | GIN | 5.17±0.58 | −0.88 ± 0.10 | 4.54±0.74 | −0.87 ± 0.06 | 164.125 |
| PMI_L + PR_L + Triplet_L | MPNN | 12.44 ± 1.22 | 5.41±0.19 | 6.49±0.93 | 3.56±0.34 | 73.375 |
| PMI_L + PR_L + Triplet_L | PAGNN | 2.54±0.20 | −0.07 ± 0.00 | 2.85±0.11 | −0.11 ± 0.00 | 167.5 |
| PMI_L + PR_L + Triplet_L | SAGE | 7.19±0.60 | 1.67±0.03 | 8.46±0.77 | 1.47±0.05 | 97.875 |

Table 35. Results for Euclidean-Adj Corr (↑) (continued)

| Loss Type | Model | Cora ↓ Citeseer | Cora ↓ Bitcoin | Citeseer ↓ Cora | Citeseer ↓ Bitcoin | Average Rank |
|---|---|---|---|---|---|---|
| PMI_L + Triplet_L | ALL | 8.44±0.39 | 0.42±0.14 | 7.64±0.97 | 2.68±0.14 | 98.875 |
| PMI_L + Triplet_L | GAT | 15.67 ± 0.91 | 6.17±0.31 | 15.35 ± 0.47 | 6.32±0.52 | 12.5 |
| PMI_L + Triplet_L | GCN | 12.84 ± 0.85 | 6.40±0.41 | 13.75 ± 0.98 | 6.37±0.26 | 21.875 |
| PMI_L + Triplet_L | GIN | 6.00±0.27 | −0.64 ± 0.13 | 5.50±1.02 | −0.36 ± 0.10 | 149.125 |
| PMI_L + Triplet_L | MPNN | 12.58 ± 1.09 | 5.19±0.19 | 9.50±0.77 | 5.63±0.44 | 52.375 |
| PMI_L + Triplet_L | PAGNN | 2.37±0.22 | −0.07 ± 0.00 | 2.71±0.11 | −0.10 ± 0.00 | 169.5 |
| PMI_L + Triplet_L | SAGE | 7.19±0.93 | 1.40±0.10 | 8.74±0.28 | 1.90±0.06 | 95.875 |
| PR_L | ALL | 0.82±0.35 | −2.62 ± 0.09 | 2.29±1.05 | −2.50 ± 0.14 | 204.25 |
| PR_L | GAT | 7.73±0.56 | 3.92±0.23 | 8.60±0.35 | 3.86±0.22 | 82.5 |
| PR_L | GCN | 5.55±0.82 | 2.96±0.32 | 7.94±0.42 | 3.16±0.36 | 96.375 |
| PR_L | GIN | 3.02±0.42 | −0.01 ± 0.02 | 2.44±1.33 | −0.02 ± 0.02 | 161.125 |
| PR_L | MPNN | 6.60±3.90 | 4.35±0.23 | 5.02±1.12 | 4.34±0.13 | 99.875 |
| PR_L | PAGNN | 1.15±0.96 | −0.02 ± 0.00 | −0.23 ± 0.21 | −0.02 ± 0.00 | 172.5 |
| PR_L | SAGE | 3.13±0.17 | 0.41±0.07 | 4.19±0.30 | 0.57±0.07 | 142.0 |
| PR_L + Triplet_L | ALL | 1.71±0.46 | −2.29 ± 0.16 | 2.85±0.55 | −2.57 ± 0.10 | 197.125 |
| PR_L + Triplet_L | GAT | 8.99±2.70 | 2.98±0.31 | 8.38±0.74 | 3.61±0.21 | 82.0 |
| PR_L + Triplet_L | GCN | 7.71±0.47 | 4.35±0.48 | 8.81±0.29 | 4.51±0.32 | 77.625 |
| PR_L + Triplet_L | GIN | 3.71±0.70 | −0.10 ± 0.02 | 3.97±0.82 | −0.52 ± 0.05 | 166.75 |
| PR_L + Triplet_L | MPNN | 6.81±3.47 | 4.55±0.18 | 6.04±2.11 | 4.62±0.09 | 90.875 |
| PR_L + Triplet_L | PAGNN | 1.71±0.43 | −0.02 ± 0.00 | 0.57±0.54 | −0.00 ± 0.00 | 168.75 |
| PR_L + Triplet_L | SAGE | 3.49±0.43 | 0.66±0.06 | 5.97±2.10 | 0.58±0.05 | 130.75 |

Table 35. Results for Euclidean-Adj Corr (↑) (continued)

| Loss Type | Model | Cora ↓ Citeseer | Cora ↓ Bitcoin | Citeseer ↓ Cora | Citeseer ↓ Bitcoin | Average Rank |
|---|---|---|---|---|---|---|
| Triplet_L | ALL | 8.57±0.87 | 0.76±0.09 | 8.71±0.65 | 2.97±0.25 | 90.0 |
| Triplet_L | GAT | 15.84 ± 0.99 | 5.74±0.36 | 14.24 ± 0.84 | 5.87±0.57 | 21.75 |
| Triplet_L | GCN | 14.48 ± 0.57 | 6.73±0.67 | 13.12 ± 1.36 | 6.65±0.60 | 11.875 |
| Triplet_L | GIN | 10.35 ± 0.43 | −0.15 ± 0.05 | 8.11±0.97 | 0.48±0.11 | 111.75 |
| Triplet_L | MPNN | 13.19 ± 1.18 | 5.27±0.27 | 10.26 ± 1.12 | 5.35±0.61 | 47.875 |
| Triplet_L | PAGNN | 5.21±1.12 | −0.11 ± 0.03 | 6.21±0.47 | −0.16 ± 0.02 | 146.125 |
| Triplet_L | SAGE | 11.37 ± 1.82 | 2.31±0.09 | 12.59 ± 0.69 | 2.48±0.08 | 69.75 |

Table 36. Graph Reconstruction Bce Loss Performance (↓): This table presents models (Loss function and GNN) ranked by their average performance in terms of graph reconstruction bce loss. Top-ranked results are highlighted in red, second-ranked in blue, and third-ranked in green.

| Loss Type | Model | Cora ↓ Citeseer | Cora ↓ Bitcoin | Citeseer ↓ Cora | Citeseer ↓ Bitcoin | Average Rank |
|---|---|---|---|---|---|---|
| Contr_l | ALL | 68.65 ± 2.51 | 77.00 ± 0.18 | 66.46 ± 3.21 | 78.03 ± 0.57 | 132.25 |
| Contr_l | GAT | 61.45 ± 2.47 | 63.65 ± 1.85 | 64.93 ± 0.99 | 62.27 ± 0.20 | 35.25 |
| Contr_l | GCN | 60.95 ± 1.47 | 62.03 ± 1.16 | 63.97 ± 1.41 | 61.63 ± 0.62 | 16.375 |
| Contr_l | GIN | 62.15 ± 1.32 | 77.18 ± 0.21 | 66.68 ± 2.75 | 71.28 ± 0.18 | 93.0 |

Table 36. Results for Graph Reconstruction Bce Loss (↓) (continued)

| Loss Type | Model | Cora ↓ Citeseer | | Cora ↓ Bitcoin | | Citeseer ↓ Cora | | Citeseer ↓ Bitcoin | | Average Rank |
|---|---|---|---|---|---|---|---|---|---|---|
| Contr_l | MPNN | 64.79 1.69 | ± | 67.76 1.44 | ± | 66.81 1.08 | ± | 69.47 1.36 | ± | 82.0 |
| Contr_l | PAGNN | 65.09 1.24 | ± | 73.98 0.18 | ± | 68.35 2.90 | ± | 73.79 0.22 | ± | 108.25 |
| Contr_l | SAGE | 63.43 2.79 | ± | 66.39 1.21 | ± | 64.84 1.53 | ± | 65.45 1.00 | ± | 51.0 |
| Contr_l + CrossE_L | ALL | 65.86 3.25 | ± | 74.18 1.00 | ± | 67.64 3.00 | ± | 74.93 0.53 | ± | 110.5 |
| Contr_l + CrossE_L | GAT | 61.03 1.24 | ± | 63.14 0.72 | ± | 64.68 1.48 | ± | 62.70 0.36 | ± | 30.125 |
| Contr_l + CrossE_L | GCN | 60.50 0.53 | ± | 61.52 0.73 | ± | 63.62 1.69 | ± | 61.90 0.89 | ± | 12.0 |
| Contr_l + CrossE_L | GIN | 64.09 1.77 | ± | 78.88 0.30 | ± | 67.85 1.61 | ± | 75.74 0.23 | ± | 123.75 |
| Contr_l + CrossE_L | MPNN | 63.07 2.42 | ± | 68.81 1.28 | ± | 66.87 2.28 | ± | 67.77 1.09 | ± | 74.75 |
| Contr_l + CrossE_L | PAGNN | 64.45 1.33 | ± | 74.82 0.06 | ± | 69.07 0.87 | ± | 74.97 0.13 | ± | 112.875 |
| Contr_l + CrossE_L | SAGE | 60.42 0.93 | ± | 66.20 0.46 | ± | 66.71 0.74 | ± | 66.98 0.52 | ± | 52.0 |
| Contr_l + CrossE_L + PMI_L | ALL | 68.77 2.21 | ± | 76.88 0.16 | ± | 65.81 1.06 | ± | 81.49 0.24 | ± | 141.0 |
| Contr_l + CrossE_L + PMI_L | GAT | 60.15 0.76 | ± | 62.84 0.70 | ± | 64.93 0.94 | ± | 61.80 0.76 | ± | 23.125 |
| Contr_l + CrossE_L + PMI_L | GCN | 59.78 0.90 | ± | 61.62 0.89 | ± | 64.17 1.71 | ± | 61.39 0.43 | ± | 10.25 |
| Contr_l + CrossE_L + PMI_L | GIN | 64.93 2.71 | ± | 79.38 0.28 | ± | 65.66 1.16 | ± | 79.61 0.18 | ± | 130.25 |
| Contr_l + CrossE_L + PMI_L | MPNN | 60.77 1.68 | ± | 65.41 0.84 | ± | 68.04 1.65 | ± | 65.35 0.67 | ± | 58.0 |
| Contr_l + CrossE_L + PMI_L | PAGNN | 66.87 0.60 | ± | 75.71 0.12 | ± | 69.06 0.24 | ± | 75.72 0.19 | ± | 136.375 |

Table 36. Results for Graph Reconstruction Bce Loss (↓) (continued)

| Loss Type | Model | Cora ↓ Citeseer | | Cora ↓ Bitcoin | | Citeseer ↓ Cora | | Citeseer ↓ Bitcoin | | Average Rank |
|---|---|---|---|---|---|---|---|---|---|---|
| Contr_l + CrossE_L + PMI_L | SAGE | 65.51 | ± 0.51 | 69.65 | ± 0.27 | 67.71 | ± 0.79 | 69.12 | ± 0.24 | 96.0 |
| Contr_l + CrossE_L + PMI_L + PR_L | ALL | 67.64 | ± 3.66 | 82.23 | ± 0.26 | 67.04 | ± 0.81 | 84.18 | ± 0.17 | 164.0 |
| Contr_l + CrossE_L + PMI_L + PR_L | GAT | 61.40 | ± 1.94 | 63.23 | ± 0.87 | 64.39 | ± 1.25 | 62.25 | ± 1.29 | 28.5 |
| Contr_l + CrossE_L + PMI_L + PR_L | GCN | 60.04 | ± 1.56 | 61.37 | ± 0.42 | 64.70 | ± 2.71 | 61.92 | ± 1.28 | 16.75 |
| Contr_l + CrossE_L + PMI_L + PR_L | GIN | 65.21 | ± 1.69 | 79.57 | ± 0.31 | 66.43 | ± 2.49 | 77.43 | ± 0.04 | 126.75 |
| Contr_l + CrossE_L + PMI_L + PR_L | MPNN | 63.09 | ± 1.03 | 66.86 | ± 1.31 | 66.93 | ± 1.04 | 75.49 | ± 0.79 | 86.5 |
| Contr_l + CrossE_L + PMI_L + PR_L | PAGNN | 66.12 | ± 0.99 | 75.49 | ± 0.15 | 68.19 | ± 0.33 | 75.25 | ± 0.22 | 124.25 |
| Contr_l + CrossE_L + PMI_L + PR_L | SAGE | 66.02 | ± 0.30 | 69.98 | ± 0.26 | 67.39 | ± 0.85 | 70.13 | ± 0.56 | 100.5 |
| Contr_l + CrossE_L + PMI_L + PR_L + Triplet_L | ALL | 64.23 | ± 0.91 | 79.86 | ± 0.36 | 66.76 | ± 0.79 | 79.42 | ± 0.48 | 135.25 |
| Contr_l + CrossE_L + PMI_L + PR_L + Triplet_L | GAT | 61.06 | ± 0.84 | 63.74 | ± 1.15 | 65.96 | ± 1.52 | 62.99 | ± 1.17 | 43.25 |
| Contr_l + CrossE_L + PMI_L + PR_L + Triplet_L | GCN | 59.79 | ± 1.13 | 61.94 | ± 0.80 | 64.33 | ± 1.78 | 60.84 | ± 0.58 | 10.0 |
| Contr_l + CrossE_L + PMI_L + PR_L + Triplet_L | GIN | 64.78 | ± 1.15 | 79.70 | ± 0.11 | 64.06 | ± 1.59 | 78.78 | ± 0.22 | 113.625 |
| Contr_l + CrossE_L + PMI_L + PR_L + Triplet_L | MPNN | 61.81 | ± 2.51 | 68.17 | ± 0.59 | 64.79 | ± 2.39 | 73.74 | ± 0.46 | 64.125 |

Table 36. Results for Graph Reconstruction Bce Loss (↓) (continued)

| Loss Type | Model | Cora ↓ Citeseer | | Cora ↓ Bitcoin | | Citeseer ↓ Cora | | Citeseer ↓ Bitcoin | | Average Rank |
|---|---|---|---|---|---|---|---|---|---|---|
| Contr_l + CrossE_L + PMI_L + PR_L + Triplet_L | PAGNN | 66.49 0.82 | ± | 76.04 0.16 | ± | 67.95 0.50 | ± | 75.97 0.08 | ± | 134.0 |
| Contr_l + CrossE_L + PMI_L + PR_L + Triplet_L | SAGE | 64.01 2.16 | ± | 66.47 0.40 | ± | 64.99 0.79 | ± | 67.42 0.56 | ± | 58.75 |
| Contr_l + CrossE_L + PMI_L + Triplet_L | ALL | 66.94 3.05 | ± | 76.83 0.51 | ± | 67.96 1.11 | ± | 72.61 0.99 | ± | 126.25 |
| Contr_l + CrossE_L + PMI_L + Triplet_L | GAT | 61.23 1.04 | ± | 62.38 0.98 | ± | 64.79 1.39 | ± | 62.49 0.54 | ± | 29.375 |
| Contr_l + CrossE_L + PMI_L + Triplet_L | GCN | 60.29 0.78 | ± | 62.03 0.74 | ± | 63.96 1.28 | ± | 60.90 0.73 | ± | 9.25 |
| Contr_l + CrossE_L + PMI_L + Triplet_L | GIN | 65.64 2.74 | ± | 79.97 0.13 | ± | 65.24 2.35 | ± | 78.89 0.21 | ± | 129.5 |
| Contr_l + CrossE_L + PMI_L + Triplet_L | MPNN | 62.04 1.35 | ± | 66.84 0.67 | ± | 66.77 1.69 | ± | 67.91 1.38 | ± | 65.5 |
| Contr_l + CrossE_L + PMI_L + Triplet_L | PAGNN | 66.80 0.95 | ± | 75.76 0.19 | ± | 68.39 0.40 | ± | 75.54 0.14 | ± | 133.875 |
| Contr_l + CrossE_L + PMI_L + Triplet_L | SAGE | 63.68 1.65 | ± | 68.02 0.70 | ± | 65.05 0.97 | ± | 66.61 0.24 | ± | 61.75 |
| Contr_l + CrossE_L + PR_L | ALL | 79.97 0.25 | ± | 80.21 0.01 | ± | 75.30 2.46 | ± | 80.49 0.01 | ± | 196.25 |
| Contr_l + CrossE_L + PR_L | GAT | 73.20 5.24 | ± | 68.93 0.24 | ± | 77.97 4.08 | ± | 78.88 0.21 | ± | 150.0 |
| Contr_l + CrossE_L + PR_L | GCN | 71.32 1.92 | ± | 70.71 1.51 | ± | 74.23 1.31 | ± | 73.81 0.69 | ± | 132.25 |
| Contr_l + CrossE_L + PR_L | GIN | 78.32 1.49 | ± | 79.04 0.04 | ± | 74.77 2.08 | ± | 79.01 0.16 | ± | 175.0 |
| Contr_l + CrossE_L + PR_L | MPNN | 77.90 2.19 | ± | 79.68 0.04 | ± | 78.09 1.16 | ± | 79.56 0.09 | ± | 184.375 |
| Contr_l + CrossE_L + PR_L | PAGNN | 76.10 0.92 | ± | 75.38 0.13 | ± | 76.02 1.41 | ± | 75.00 0.21 | ± | 148.25 |

Table 36. Results for Graph Reconstruction Bce Loss (↓) (continued)

| Loss Type | Model | Cora ↓ Citeseer | | Cora ↓ Bitcoin | | Citeseer ↓ Cora | | Citeseer ↓ Bitcoin | | Average Rank |
|---|---|---|---|---|---|---|---|---|---|---|
| Contr_l + CrossE_L + PR_L | SAGE | 71.42 | ± 4.16 | 72.04 | ± 0.22 | 74.43 | ± 2.28 | 73.37 | ± 0.27 | 132.5 |
| Contr_l + CrossE_L + PR_L + Triplet_L | ALL | 73.61 | ± 2.04 | 78.87 | ± 0.05 | 71.27 | ± 1.26 | 77.63 | ± 0.32 | 159.25 |
| Contr_l + CrossE_L + PR_L + Triplet_L | GAT | 67.51 | ± 1.24 | 67.86 | ± 0.83 | 70.67 | ± 2.77 | 67.26 | ± 0.44 | 106.375 |
| Contr_l + CrossE_L + PR_L + Triplet_L | GCN | 66.66 | ± 0.86 | 65.59 | ± 1.54 | 71.49 | ± 2.05 | 68.50 | ± 2.30 | 103.125 |
| Contr_l + CrossE_L + PR_L + Triplet_L | GIN | 68.84 | ± 1.05 | 79.42 | ± 0.07 | 70.29 | ± 0.75 | 77.29 | ± 0.33 | 155.75 |
| Contr_l + CrossE_L + PR_L + Triplet_L | MPNN | 72.82 | ± 1.30 | 75.50 | ± 0.46 | 70.38 | ± 0.79 | 76.05 | ± 0.29 | 145.5 |
| Contr_l + CrossE_L + PR_L + Triplet_L | PAGNN | 70.51 | ± 1.48 | 75.58 | ± 0.09 | 71.77 | ± 0.57 | 75.10 | ± 0.17 | 141.25 |
| Contr_l + CrossE_L + PR_L + Triplet_L | SAGE | 67.84 | ± 0.69 | 69.56 | ± 0.55 | 70.57 | ± 1.82 | 69.51 | ± 0.47 | 115.0 |
| Contr_l + CrossE_L + Triplet_L | ALL | 66.17 | ± 4.52 | 74.81 | ± 0.58 | 67.21 | ± 3.63 | 73.55 | ± 0.45 | 108.5 |
| Contr_l + CrossE_L + Triplet_L | GAT | 61.07 | ± 1.24 | 63.65 | ± 1.23 | 64.37 | ± 0.74 | 63.03 | ± 0.83 | 31.875 |
| Contr_l + CrossE_L + Triplet_L | GCN | 60.81 | ± 1.96 | 63.43 | ± 1.72 | 64.62 | ± 1.45 | 61.68 | ± 0.63 | 24.625 |
| Contr_l + CrossE_L + Triplet_L | GIN | 62.81 | ± 1.47 | 75.06 | ± 0.30 | 66.56 | ± 1.82 | 71.21 | ± 0.29 | 86.875 |
| Contr_l + CrossE_L + Triplet_L | MPNN | 65.01 | ± 1.29 | 69.92 | ± 0.62 | 67.21 | ± 1.42 | 67.97 | ± 0.68 | 89.75 |
| Contr_l + CrossE_L + Triplet_L | PAGNN | 64.86 | ± 1.23 | 74.19 | ± 0.14 | 66.29 | ± 2.16 | 74.21 | ± 0.15 | 94.5 |
| Contr_l + CrossE_L + Triplet_L | SAGE | 62.50 | ± 1.14 | 66.48 | ± 0.70 | 64.55 | ± 1.89 | 65.93 | ± 0.95 | 45.5 |
| Contr_l + PMI_L | ALL | 68.23 | ± 1.93 | 77.89 | ± 0.14 | 67.66 | ± 3.26 | 75.81 | ± 0.11 | 138.0 |

Continued on next page

Table 36. Results for Graph Reconstruction Bce Loss (↓) (continued)

| Loss Type | Model | Cora ↓ Citeseer | | Cora ↓ Bitcoin | | Citeseer ↓ Cora | | Citeseer ↓ Bitcoin | | Average Rank |
|---|---|---|---|---|---|---|---|---|---|---|
| Contr_l + PMI_L | GAT | 60.67 | ± 1.32 | 63.34 | ± 0.92 | 64.21 | ± 0.80 | 62.37 | ± 0.43 | 23.25 |
| Contr_l + PMI_L | GCN | 59.05 | ± 1.79 | 61.69 | ± 0.88 | 63.06 | ± 1.36 | 61.28 | ± 1.09 | 5.25 |
| Contr_l + PMI_L | GIN | 64.96 | ± 0.83 | 80.74 | ± 0.18 | 66.02 | ± 2.17 | 79.02 | ± 0.18 | 134.875 |
| Contr_l + PMI_L | MPNN | 62.70 | ± 2.38 | 67.09 | ± 1.09 | 66.32 | ± 1.66 | 66.75 | ± 1.13 | 62.75 |
| Contr_l + PMI_L | PAGNN | 66.30 | ± 0.75 | 75.74 | ± 0.14 | 69.07 | ± 0.33 | 75.75 | ± 0.18 | 135.125 |
| Contr_l + PMI_L | SAGE | 65.89 | ± 0.34 | 69.71 | ± 0.42 | 66.83 | ± 1.47 | 68.26 | ± 0.26 | 90.25 |
| Contr_l + PMI_L + PR_L | ALL | 68.06 | ± 2.35 | 81.13 | ± 0.53 | 67.21 | ± 0.52 | 84.74 | ± 0.21 | 165.0 |
| Contr_l + PMI_L + PR_L | GAT | 60.86 | ± 1.64 | 63.17 | ± 0.76 | 65.52 | ± 3.15 | 67.26 | ± 2.19 | 42.5 |
| Contr_l + PMI_L + PR_L | GCN | 61.94 | ± 1.16 | 63.60 | ± 1.08 | 65.76 | ± 2.20 | 63.37 | ± 1.25 | 45.125 |
| Contr_l + PMI_L + PR_L | GIN | 66.00 | ± 3.20 | 78.88 | ± 0.08 | 67.40 | ± 2.36 | 77.58 | ± 0.09 | 134.375 |
| Contr_l + PMI_L + PR_L | MPNN | 62.57 | ± 0.62 | 67.62 | ± 0.70 | 65.45 | ± 1.33 | 76.67 | ± 0.20 | 81.125 |
| Contr_l + PMI_L + PR_L | PAGNN | 66.37 | ± 0.40 | 75.53 | ± 0.13 | 68.55 | ± 0.67 | 75.31 | ± 0.25 | 128.5 |
| Contr_l + PMI_L + PR_L | SAGE | 65.53 | ± 0.51 | 69.35 | ± 0.31 | 66.04 | ± 3.08 | 69.56 | ± 0.65 | 85.25 |
| Contr_l + PMI_L + PR_L + Triplet_L | ALL | 65.32 | ± 2.56 | 78.75 | ± 0.27 | 69.17 | ± 0.90 | 77.17 | ± 0.41 | 138.0 |
| Contr_l + PMI_L + PR_L + Triplet_L | GAT | 63.54 | ± 1.74 | 64.11 | ± 1.90 | 64.44 | ± 1.11 | 65.49 | ± 1.25 | 44.75 |
| Contr_l + PMI_L + PR_L + Triplet_L | GCN | 59.68 | ± 1.13 | 62.05 | ± 0.93 | 65.50 | ± 1.28 | 61.22 | ± 0.60 | 20.875 |

Table 36. Results for Graph Reconstruction Bce Loss (↓) (continued)

| Loss Type | Model | Cora ↓ Citeseer | | Cora ↓ Bitcoin | | Citeseer ↓ Cora | | Citeseer ↓ Bitcoin | | Average Rank |
|---|---|---|---|---|---|---|---|---|---|---|
| Contr_l + PMI_L + PR_L + Triplet_L | GIN | 64.75 1.63 | ± | 80.03 0.07 | ± | 66.31 0.65 | ± | 77.78 0.21 | ± | 128.75 |
| Contr_l + PMI_L + PR_L + Triplet_L | MPNN | 62.34 1.78 | ± | 68.06 1.01 | ± | 64.88 1.51 | ± | 75.54 0.29 | ± | 74.125 |
| Contr_l + PMI_L + PR_L + Triplet_L | PAGNN | 65.91 0.84 | ± | 75.61 0.12 | ± | 67.40 0.65 | ± | 75.09 0.21 | ± | 117.125 |
| Contr_l + PMI_L + PR_L + Triplet_L | SAGE | 64.05 1.97 | ± | 67.93 0.26 | ± | 66.98 1.79 | ± | 67.86 0.92 | ± | 77.375 |
| Contr_l + PR_L | ALL | 79.83 0.23 | ± | 80.11 0.00 | ± | 76.62 3.26 | ± | 80.41 0.00 | ± | 196.25 |
| Contr_l + PR_L | GAT | 75.49 5.21 | ± | 68.42 0.32 | ± | 79.38 0.32 | ± | 78.32 0.48 | ± | 154.5 |
| Contr_l + PR_L | GCN | 70.82 2.93 | ± | 69.17 0.65 | ± | 74.15 2.74 | ± | 70.79 1.89 | ± | 124.75 |
| Contr_l + PR_L | GIN | 78.21 0.98 | ± | 79.33 0.02 | ± | 75.29 2.23 | ± | 79.60 0.09 | ± | 181.25 |
| Contr_l + PR_L | MPNN | 78.71 0.35 | ± | 79.10 0.03 | ± | 78.33 1.17 | ± | 79.58 0.04 | ± | 184.25 |
| Contr_l + PR_L | PAGNN | 76.43 1.19 | ± | 75.04 0.12 | ± | 75.93 1.79 | ± | 74.89 0.20 | ± | 144.75 |
| Contr_l + PR_L | SAGE | 70.71 4.38 | ± | 70.91 0.15 | ± | 72.42 3.70 | ± | 77.76 0.30 | ± | 143.625 |
| Contr_l + PR_L + Triplet_L | ALL | 74.39 1.85 | ± | 78.66 0.06 | ± | 73.16 1.63 | ± | 79.87 0.23 | ± | 172.5 |
| Contr_l + PR_L + Triplet_L | GAT | 68.89 2.29 | ± | 68.22 0.71 | ± | 70.75 2.23 | ± | 68.50 1.64 | ± | 113.125 |
| Contr_l + PR_L + Triplet_L | GCN | 65.78 1.76 | ± | 67.16 1.79 | ± | 69.99 1.47 | ± | 68.87 2.05 | ± | 98.25 |
| Contr_l + PR_L + Triplet_L | GIN | 69.09 0.80 | ± | 79.22 0.15 | ± | 71.30 1.69 | ± | 80.01 0.27 | ± | 169.625 |
| Contr_l + PR_L + Triplet_L | MPNN | 71.79 0.86 | ± | 77.72 0.45 | ± | 72.01 1.49 | ± | 75.35 0.31 | ± | 150.5 |

Table 36.  Results for Graph Reconstruction Bce Loss (↓) (continued)

| Loss Type | Model | Cora ↓ Citeseer | | Cora ↓ Bitcoin | | Citeseer ↓ Cora | | Citeseer ↓ Bitcoin | | Average Rank |
|---|---|---|---|---|---|---|---|---|---|---|
| Contr_l + PR_L + Triplet_L | PAGNN | 71.68 1.13 | ± | 75.73 0.08 | ± | 71.15 0.85 | ± | 74.66 0.06 | ± | 140.0 |
| Contr_l + PR_L + Triplet_L | SAGE | 69.46 0.41 | ± | 70.09 0.38 | ± | 70.03 1.82 | ± | 70.11 0.65 | ± | 120.0 |
| Contr_l + Triplet_L | ALL | 67.68 2.10 | ± | 75.11 0.37 | ± | 66.18 4.75 | ± | 75.08 0.28 | ± | 112.125 |
| Contr_l + Triplet_L | GAT | 61.20 0.65 | ± | 64.56 1.29 | ± | 64.09 1.25 | ± | 62.45 0.74 | ± | 29.5 |
| Contr_l + Triplet_L | GCN | 61.26 1.14 | ± | 62.88 1.28 | ± | 63.59 0.76 | ± | 60.99 0.60 | ± | 18.875 |
| Contr_l + Triplet_L | GIN | 62.54 1.47 | ± | 76.75 0.29 | ± | 66.54 2.76 | ± | 74.36 0.40 | ± | 96.75 |
| Contr_l + Triplet_L | MPNN | 64.55 1.89 | ± | 69.11 0.80 | ± | 65.21 2.36 | ± | 67.18 1.06 | ± | 69.75 |
| Contr_l + Triplet_L | PAGNN | 65.39 1.25 | ± | 74.07 0.13 | ± | 68.36 1.02 | ± | 74.09 0.27 | ± | 111.0 |
| Contr_l + Triplet_L | SAGE | 62.42 1.45 | ± | 65.55 0.60 | ± | 65.25 1.97 | ± | 66.14 0.88 | ± | 49.25 |
| CrossE_L | ALL | 79.22 0.14 | ± | 79.28 0.01 | ± | 79.19 0.14 | ± | 79.27 0.00 | ± | 187.5 |
| CrossE_L | GAT | 78.92 0.15 | ± | 78.69 0.00 | ± | 78.97 0.17 | ± | 78.91 0.00 | ± | 179.25 |
| CrossE_L | GCN | 78.51 0.24 | ± | 78.72 0.00 | ± | 78.35 0.17 | ± | 78.28 0.00 | ± | 174.0 |
| CrossE_L | GIN | 78.94 0.19 | ± | 79.11 0.00 | ± | 78.78 0.17 | ± | 78.63 0.00 | ± | 180.75 |
| CrossE_L | MPNN | 79.01 0.09 | ± | 78.98 0.00 | ± | 79.06 0.07 | ± | 79.12 0.00 | ± | 184.625 |
| CrossE_L | PAGNN | 78.47 0.22 | ± | 75.31 0.29 | ± | 79.52 0.04 | ± | 79.57 0.00 | ± | 175.0 |
| CrossE_L | SAGE | 78.90 0.06 | ± | 78.88 0.00 | ± | 79.01 0.15 | ± | 79.04 0.00 | ± | 181.625 |

Table 36. Results for Graph Reconstruction Bce Loss (↓) (continued)

| Loss Type | Model | Cora ↓ Citeseer | | Cora ↓ Bitcoin | | Citeseer ↓ Cora | | Citeseer ↓ Bitcoin | | Average Rank |
|---|---|---|---|---|---|---|---|---|---|---|
| CrossE_L + PMI_L | ALL | 64.16 | ± 1.59 | 81.75 | ± 0.40 | 67.70 | ± 1.03 | 83.95 | ± 0.17 | 152.25 |
| CrossE_L + PMI_L | GAT | 60.94 | ± 1.21 | 62.60 | ± 0.47 | 65.01 | ± 1.08 | 61.72 | ± 0.98 | 26.25 |
| CrossE_L + PMI_L | GCN | 60.75 | ± 1.53 | 63.51 | ± 1.18 | 63.08 | ± 0.92 | 60.79 | ± 0.53 | 14.0 |
| CrossE_L + PMI_L | GIN | 64.29 | ± 1.13 | 79.64 | ± 0.12 | 64.03 | ± 1.85 | 79.37 | ± 0.25 | 113.75 |
| CrossE_L + PMI_L | MPNN | 63.03 | ± 1.76 | 67.55 | ± 1.38 | 66.99 | ± 0.86 | 66.38 | ± 1.46 | 69.0 |
| CrossE_L + PMI_L | PAGNN | 66.22 | ± 0.97 | 75.07 | ± 0.19 | 68.68 | ± 0.55 | 75.24 | ± 0.12 | 124.75 |
| CrossE_L + PMI_L | SAGE | 66.38 | ± 0.58 | 70.03 | ± 0.30 | 67.73 | ± 0.34 | 70.39 | ± 0.12 | 107.375 |
| CrossE_L + PMI_L + PR_L | ALL | 69.92 | ± 0.44 | 81.96 | ± 0.46 | 67.41 | ± 0.77 | 85.24 | ± 0.18 | 170.875 |
| CrossE_L + PMI_L + PR_L | GAT | 60.55 | ± 0.52 | 63.37 | ± 1.00 | 66.21 | ± 4.10 | 69.60 | ± 1.95 | 49.75 |
| CrossE_L + PMI_L + PR_L | GCN | 60.69 | ± 1.00 | 62.42 | ± 1.74 | 64.61 | ± 2.33 | 60.99 | ± 0.78 | 17.625 |
| CrossE_L + PMI_L + PR_L | GIN | 66.40 | ± 1.93 | 80.36 | ± 0.14 | 67.46 | ± 1.69 | 79.51 | ± 0.24 | 156.875 |
| CrossE_L + PMI_L + PR_L | MPNN | 61.95 | ± 1.14 | 66.18 | ± 0.77 | 66.91 | ± 1.33 | 67.79 | ± 1.38 | 63.5 |
| CrossE_L + PMI_L + PR_L | PAGNN | 66.42 | ± 1.09 | 75.94 | ± 0.11 | 69.01 | ± 0.35 | 75.66 | ± 0.18 | 136.0 |
| CrossE_L + PMI_L + PR_L | SAGE | 65.80 | ± 0.43 | 69.61 | ± 0.07 | 67.80 | ± 0.71 | 69.97 | ± 0.24 | 99.75 |
| CrossE_L + PMI_L + PR_L + Triplet_L | ALL | 65.25 | ± 2.33 | 81.28 | ± 0.46 | 67.50 | ± 1.83 | 80.85 | ± 0.51 | 155.5 |
| CrossE_L + PMI_L + PR_L + Triplet_L | GAT | 60.79 | ± 3.02 | 62.85 | ± 0.73 | 65.45 | ± 1.24 | 62.53 | ± 0.40 | 32.625 |

Table 36. Results for Graph Reconstruction Bce Loss (↓) (continued)

| Loss Type | Model | Cora ↓ Citeseer | | Cora ↓ Bitcoin | | Citeseer ↓ Cora | | Citeseer ↓ Bitcoin | | Average Rank |
|---|---|---|---|---|---|---|---|---|---|---|
| CrossE_L + PMI_L + PR_L + Triplet_L | GCN | 59.76 1.02 | ± | 63.60 1.48 | ± | 64.66 1.26 | ± | 62.35 0.38 | ± | 23.625 |
| CrossE_L + PMI_L + PR_L + Triplet_L | GIN | 66.01 2.33 | ± | 79.22 0.13 | ± | 65.09 2.88 | ± | 80.19 0.22 | ± | 134.0 |
| CrossE_L + PMI_L + PR_L + Triplet_L | MPNN | 64.22 0.70 | ± | 68.34 0.93 | ± | 65.72 1.13 | ± | 71.71 0.81 | ± | 79.25 |
| CrossE_L + PMI_L + PR_L + Triplet_L | PAGNN | 65.70 0.51 | ± | 75.22 0.16 | ± | 68.24 0.31 | ± | 75.05 0.20 | ± | 117.5 |
| CrossE_L + PMI_L + PR_L + Triplet_L | SAGE | 63.79 1.96 | ± | 67.84 0.88 | ± | 65.87 0.99 | ± | 66.95 0.71 | ± | 67.25 |
| CrossE_L + PMI_L + Triplet_L | ALL | 66.38 2.33 | ± | 76.84 0.14 | ± | 70.41 0.94 | ± | 70.82 0.64 | ± | 128.625 |
| CrossE_L + PMI_L + Triplet_L | GAT | 60.41 0.71 | ± | 62.69 0.87 | ± | 64.59 0.77 | ± | 61.86 0.25 | ± | 20.5 |
| CrossE_L + PMI_L + Triplet_L | GCN | 60.34 1.28 | ± | 61.06 0.42 | ± | 63.32 1.37 | ± | 61.68 0.70 | ± | 8.125 |
| CrossE_L + PMI_L + Triplet_L | GIN | 64.46 2.48 | ± | 79.54 0.12 | ± | 65.80 1.28 | ± | 78.77 0.29 | ± | 123.5 |
| CrossE_L + PMI_L + Triplet_L | MPNN | 63.30 2.07 | ± | 64.95 1.67 | ± | 67.36 0.52 | ± | 67.87 1.22 | ± | 70.5 |
| CrossE_L + PMI_L + Triplet_L | PAGNN | 66.19 0.93 | ± | 75.71 0.09 | ± | 67.86 0.37 | ± | 75.90 0.19 | ± | 129.125 |
| CrossE_L + PMI_L + Triplet_L | SAGE | 62.24 1.01 | ± | 66.58 0.40 | ± | 65.03 1.43 | ± | 66.57 0.59 | ± | 51.0 |
| CrossE_L + PR_L | ALL | 79.96 0.12 | ± | 80.24 0.00 | ± | 79.93 0.16 | ± | 80.33 0.00 | ± | 203.5 |
| CrossE_L + PR_L | GAT | 79.35 0.41 | ± | 79.70 0.12 | ± | 79.82 0.09 | ± | 79.44 0.06 | ± | 193.375 |
| CrossE_L + PR_L | GCN | 74.15 2.91 | ± | 76.35 1.16 | ± | 78.24 0.63 | ± | 76.56 1.36 | ± | 161.0 |
| CrossE_L + PR_L | GIN | 79.47 0.25 | ± | 79.12 0.05 | ± | 77.34 4.87 | ± | 79.51 0.04 | ± | 184.625 |

Continued on next page

Table 36. Results for Graph Reconstruction Bce Loss (↓) (continued)

| Loss Type | Model | Cora ↓ Citeseer | | Cora ↓ Bitcoin | | Citeseer ↓ Cora | | Citeseer ↓ Bitcoin | | Average Rank |
|-----------|-------|-----------------|--|----------------|--|-----------------|--|--------------------|--|--------------|
| CrossE_L + PR_L | MPNN | 78.73 | ± | 79.68 | ± | 78.94 | ± | 80.12 | ± | 191.875 |
| | | 1.39 | | 0.01 | | 0.57 | | 0.02 | | |
| CrossE_L + PR_L | PAGNN | 77.62 | ± | 74.66 | ± | 79.05 | ± | 76.08 | ± | 157.0 |
| | | 0.82 | | 0.10 | | 0.55 | | 0.11 | | |
| CrossE_L + PR_L | SAGE | 78.63 | ± | 77.98 | ± | 78.69 | ± | 77.76 | ± | 172.875 |
| | | 0.25 | | 0.29 | | 0.30 | | 0.07 | | |
| CrossE_L + PR_L + Triplet_L | ALL | 74.25 | ± | 80.35 | ± | 74.63 | ± | 78.11 | ± | 176.0 |
| | | 0.77 | | 0.21 | | 0.34 | | 0.14 | | |
| CrossE_L + PR_L + Triplet_L | GAT | 73.08 | ± | 71.84 | ± | 73.87 | ± | 69.33 | ± | 127.25 |
| | | 2.72 | | 0.63 | | 3.01 | | 0.71 | | |
| CrossE_L + PR_L + Triplet_L | GCN | 68.97 | ± | 67.53 | ± | 72.18 | ± | 69.14 | ± | 113.75 |
| | | 2.24 | | 1.35 | | 2.21 | | 1.83 | | |
| CrossE_L + PR_L + Triplet_L | GIN | 71.78 | ± | 78.91 | ± | 72.36 | ± | 78.96 | ± | 165.125 |
| | | 1.38 | | 0.31 | | 0.70 | | 0.22 | | |
| CrossE_L + PR_L + Triplet_L | MPNN | 77.10 | ± | 78.93 | ± | 75.10 | ± | 78.90 | ± | 171.75 |
| | | 1.10 | | 0.18 | | 2.52 | | 0.09 | | |
| CrossE_L + PR_L + Triplet_L | PAGNN | 74.07 | ± | 75.30 | ± | 71.94 | ± | 75.44 | ± | 144.25 |
| | | 1.58 | | 0.09 | | 0.84 | | 0.13 | | |
| CrossE_L + PR_L + Triplet_L | SAGE | 68.90 | ± | 70.27 | ± | 71.40 | ± | 69.44 | ± | 120.25 |
| | | 2.12 | | 0.27 | | 1.04 | | 0.24 | | |
| CrossE_L + Triplet_L | ALL | 66.01 | ± | 74.85 | ± | 68.70 | ± | 72.57 | ± | 115.875 |
| | | 2.15 | | 0.51 | | 3.71 | | 0.54 | | |
| CrossE_L + Triplet_L | GAT | 60.99 | ± | 63.87 | ± | 65.42 | ± | 62.73 | ± | 38.375 |
| | | 0.85 | | 0.43 | | 0.88 | | 0.53 | | |
| CrossE_L + Triplet_L | GCN | 60.00 | ± | 61.57 | ± | 64.35 | ± | 62.52 | ± | 16.0 |
| | | 1.10 | | 0.74 | | 1.72 | | 0.68 | | |
| CrossE_L + Triplet_L | GIN | 63.28 | ± | 77.67 | ± | 67.37 | ± | 76.53 | ± | 116.25 |
| | | 1.36 | | 0.21 | | 0.98 | | 0.36 | | |
| CrossE_L + Triplet_L | MPNN | 63.26 | ± | 66.97 | ± | 66.70 | ± | 66.97 | ± | 67.25 |
| | | 1.65 | | 1.05 | | 1.51 | | 0.93 | | |
| CrossE_L + Triplet_L | PAGNN | 65.11 | ± | 74.67 | ± | 69.25 | ± | 74.09 | ± | 114.375 |
| | | 0.55 | | 0.11 | | 1.34 | | 0.28 | | |

Table 36. Results for Graph Reconstruction Bce Loss (↓) (continued)

| Loss Type | Model | Cora ↓ Citeseer | | Cora ↓ Bitcoin | | Citeseer ↓ Cora | | Citeseer ↓ Bitcoin | | Average Rank |
|---|---|---|---|---|---|---|---|---|---|---|
| CrossE_L + Triplet_L | SAGE | 60.84 | ± 0.59 | 66.62 | ± 0.56 | 64.12 | ± 1.68 | 66.27 | ± 0.38 | 36.375 |
| PMI_L | ALL | 64.17 | ± 0.97 | 82.40 | ± 0.24 | 67.72 | ± 1.00 | 82.46 | ± 0.48 | 153.875 |
| PMI_L | GAT | 61.04 | ± 1.28 | 62.02 | ± 0.33 | 64.68 | ± 0.68 | 61.56 | ± 0.75 | 21.875 |
| PMI_L | GCN | 60.83 | ± 1.92 | 61.21 | ± 1.44 | 63.57 | ± 1.08 | 60.40 | ± 1.17 | 8.0 |
| PMI_L | GIN | 64.17 | ± 1.73 | 79.74 | ± 0.19 | 64.71 | ± 1.75 | 79.14 | ± 0.10 | 119.625 |
| PMI_L | MPNN | 62.08 | ± 2.25 | 64.60 | ± 1.27 | 67.11 | ± 1.68 | 66.16 | ± 1.63 | 58.75 |
| PMI_L | PAGNN | 65.98 | ± 0.57 | 75.35 | ± 0.15 | 68.85 | ± 0.39 | 75.59 | ± 0.10 | 127.25 |
| PMI_L | SAGE | 66.76 | ± 0.65 | 69.75 | ± 0.36 | 68.56 | ± 1.07 | 70.44 | ± 0.24 | 112.0 |
| PMI_L + PR_L | ALL | 69.01 | ± 1.49 | 81.88 | ± 0.18 | 67.13 | ± 0.83 | 85.34 | ± 0.24 | 167.625 |
| PMI_L + PR_L | GAT | 61.65 | ± 1.20 | 65.06 | ± 1.26 | 63.96 | ± 1.62 | 66.29 | ± 1.30 | 35.875 |
| PMI_L + PR_L | GCN | 60.44 | ± 1.80 | 61.77 | ± 0.85 | 65.02 | ± 2.70 | 61.02 | ± 0.87 | 19.0 |
| PMI_L + PR_L | GIN | 64.96 | ± 0.72 | 79.93 | ± 0.17 | 66.74 | ± 2.13 | 80.59 | ± 0.30 | 144.375 |
| PMI_L + PR_L | MPNN | 62.32 | ± 1.98 | 65.92 | ± 1.08 | 65.56 | ± 0.97 | 76.79 | ± 0.45 | 76.25 |
| PMI_L + PR_L | PAGNN | 66.84 | ± 0.57 | 75.82 | ± 0.09 | 68.36 | ± 0.28 | 75.78 | ± 0.18 | 135.625 |
| PMI_L + PR_L | SAGE | 65.92 | ± 0.41 | 69.86 | ± 0.31 | 67.59 | ± 0.55 | 70.19 | ± 0.11 | 100.5 |
| PMI_L + PR_L + Triplet_L | ALL | 64.52 | ± 1.33 | 81.11 | ± 0.41 | 67.62 | ± 1.47 | 79.38 | ± 0.41 | 145.5 |

Table 36. Results for Graph Reconstruction Bce Loss (↓) (continued)

| Loss Type | Model | Cora ↓ Citeseer | | Cora ↓ Bitcoin | | Citeseer ↓ Cora | | Citeseer ↓ Bitcoin | | Average Rank |
|---|---|---|---|---|---|---|---|---|---|---|
| PMI_L + PR_L + Triplet_L | GAT | 61.90 | ± 1.30 | 62.65 | ± 0.39 | 64.76 | ± 1.61 | 63.77 | ± 0.94 | 35.0 |
| PMI_L + PR_L + Triplet_L | GCN | 61.24 | ± 1.71 | 62.63 | ± 1.67 | 64.12 | ± 2.50 | 61.74 | ± 1.11 | 22.75 |
| PMI_L + PR_L + Triplet_L | GIN | 64.82 | ± 1.65 | 79.53 | ± 0.16 | 67.13 | ± 1.99 | 79.48 | ± 0.24 | 137.875 |
| PMI_L + PR_L + Triplet_L | MPNN | 62.91 | ± 1.87 | 67.12 | ± 0.73 | 65.83 | ± 0.71 | 76.07 | ± 0.40 | 83.0 |
| PMI_L + PR_L + Triplet_L | PAGNN | 66.21 | ± 0.52 | 75.33 | ± 0.20 | 68.32 | ± 0.52 | 75.03 | ± 0.23 | 122.75 |
| PMI_L + PR_L + Triplet_L | SAGE | 63.64 | ± 1.15 | 66.92 | ± 0.39 | 64.11 | ± 1.05 | 66.83 | ± 0.33 | 50.75 |
| PMI_L + Triplet_L | ALL | 65.38 | ± 1.75 | 76.24 | ± 0.33 | 67.41 | ± 2.46 | 70.90 | ± 0.52 | 110.625 |
| PMI_L + Triplet_L | GAT | 60.95 | ± 0.74 | 62.63 | ± 0.47 | 64.05 | ± 0.63 | 62.30 | ± 1.36 | 21.25 |
| PMI_L + Triplet_L | GCN | 61.37 | ± 1.54 | 62.03 | ± 0.93 | 62.58 | ± 1.01 | 60.90 | ± 0.55 | 14.125 |
| PMI_L + Triplet_L | GIN | 65.97 | ± 3.78 | 79.78 | ± 0.22 | 65.48 | ± 2.35 | 78.45 | ± 0.35 | 130.5 |
| PMI_L + Triplet_L | MPNN | 62.46 | ± 2.02 | 65.99 | ± 0.70 | 66.35 | ± 2.10 | 64.25 | ± 1.03 | 55.0 |
| PMI_L + Triplet_L | PAGNN | 66.29 | ± 0.76 | 75.36 | ± 0.19 | 68.78 | ± 0.38 | 75.37 | ± 0.24 | 128.5 |
| PMI_L + Triplet_L | SAGE | 63.01 | ± 2.02 | 68.00 | ± 0.73 | 65.19 | ± 1.55 | 66.34 | ± 0.24 | 58.5 |
| PR_L | ALL | 80.00 | ± 0.16 | 80.09 | ± 0.00 | 79.88 | ± 0.10 | 80.34 | ± 0.01 | 203.25 |
| PR_L | GAT | 79.65 | ± 0.39 | 78.98 | ± 0.07 | 80.02 | ± 0.21 | 79.62 | ± 0.07 | 192.625 |
| PR_L | GCN | 77.37 | ± 0.99 | 75.92 | ± 1.15 | 77.76 | ± 1.26 | 74.41 | ± 3.59 | 152.25 |

Table 36. Results for Graph Reconstruction Bce Loss (↓) (continued)

| Loss Type | Model | Cora ↓ Citeseer | | Cora ↓ Bitcoin | | Citeseer ↓ Cora | | Citeseer ↓ Bitcoin | | Average Rank |
|---|---|---|---|---|---|---|---|---|---|---|
| PR_L | GIN | 79.56 ± 0.35 | | 79.62 ± 0.03 | | 79.53 ± 0.74 | | 79.46 ± 0.04 | | 192.75 |
| PR_L | MPNN | 78.76 ± 1.17 | | 79.99 ± 0.05 | | 79.23 ± 0.46 | | 79.84 ± 0.02 | | 196.125 |
| PR_L | PAGNN | 77.84 ± 1.33 | | 75.25 ± 0.09 | | 78.60 ± 0.68 | | 75.47 ± 0.17 | | 154.25 |
| PR_L | SAGE | 78.49 ± 0.64 | | 77.69 ± 0.15 | | 79.23 ± 0.24 | | 77.56 ± 0.35 | | 173.5 |
| PR_L + Triplet_L | ALL | 79.76 ± 0.36 | | 79.72 ± 0.01 | | 79.01 ± 0.26 | | 80.28 ± 0.01 | | 196.625 |
| PR_L + Triplet_L | GAT | 74.05 ± 4.87 | | 73.49 ± 1.27 | | 79.20 ± 0.61 | | 78.46 ± 0.21 | | 158.5 |
| PR_L + Triplet_L | GCN | 72.80 ± 0.80 | | 72.62 ± 0.88 | | 74.54 ± 1.95 | | 72.28 ± 1.15 | | 133.0 |
| PR_L + Triplet_L | GIN | 77.99 ± 0.75 | | 78.91 ± 0.04 | | 73.01 ± 2.19 | | 79.65 ± 0.06 | | 176.375 |
| PR_L + Triplet_L | MPNN | 78.42 ± 0.67 | | 79.77 ± 0.02 | | 78.90 ± 0.88 | | 79.74 ± 0.04 | | 190.5 |
| PR_L + Triplet_L | PAGNN | 76.66 ± 0.66 | | 75.69 ± 0.12 | | 77.54 ± 0.47 | | 75.08 ± 0.12 | | 152.125 |
| PR_L + Triplet_L | SAGE | 77.91 ± 1.05 | | 77.46 ± 0.10 | | 76.32 ± 3.57 | | 77.56 ± 0.32 | | 165.625 |
| Triplet_L | ALL | 66.03 ± 2.23 | | 74.91 ± 0.21 | | 65.60 ± 1.50 | | 71.67 ± 0.37 | | 96.75 |
| Triplet_L | GAT | 61.50 ± 1.06 | | 63.87 ± 0.83 | | 65.52 ± 1.30 | | 62.96 ± 0.78 | | 43.0 |
| Triplet_L | GCN | **59.68** ± **1.03** | | 61.66 ± 1.38 | | 64.57 ± 1.84 | | 61.54 ± 1.37 | | 12.125 |
| Triplet_L | GIN | 62.48 ± 0.70 | | 77.52 ± 0.10 | | 66.56 ± 1.62 | | 74.57 ± 0.43 | | 98.875 |
| Triplet_L | MPNN | 63.37 ± 2.30 | | 67.66 ± 1.49 | | 66.47 ± 2.26 | | 67.86 ± 0.43 | | 70.625 |

Table 36.  Results for Graph Reconstruction Bce Loss (↓) (continued)

| Loss Type | Model | Cora ↓ Citeseer | Cora ↓ Bitcoin | Citeseer ↓ Cora | Citeseer ↓ Bitcoin | Average Rank |
|---|---|---|---|---|---|---|
| Triplet_L | PAGNN | 63.69 ± 1.50 | 74.40 ± 0.12 | 68.00 ± 1.30 | 72.77 ± 0.24 | 101.0 |
| Triplet_L | SAGE | 61.87 ± 3.01 | 66.12 ± 0.34 | 63.09 ± 2.12 | 65.73 ± 0.48 | 35.25 |

### 1.2.4 Clustering Quality Metrics.

Table 37.  Silhouette Performance (↑): This table presents models (Loss function and GNN) ranked by their average performance in terms of silhouette. Top-ranked results are highlighted in red, second-ranked in blue, and third-ranked in green.

| Loss Type | Model | Cora ↓ Citeseer | Cora ↓ Bitcoin | Citeseer ↓ Cora | Citeseer ↓ Bitcoin | Average Rank |
|---|---|---|---|---|---|---|
| Contr_l | ALL | −5.27 ± 0.55 | −11.87 ± 4.30 | −5.70 ± 3.25 | −12.51 ± 5.35 | 149.0 |
| Contr_l | GAT | −0.73 ± 0.47 | 1.05±1.17 | 0.28±0.60 | 0.59±0.43 | 31.125 |
| Contr_l | GCN | −0.83 ± 0.38 | 0.90±0.23 | −0.07 ± 1.10 | 0.21±1.25 | 43.125 |
| Contr_l | GIN | −1.61 ± 1.21 | −12.30 ± 2.07 | −0.10 ± 1.65 | −9.02 ± 1.65 | 100.125 |
| Contr_l | MPNN | −0.10 ± 1.19 | 0.42±1.16 | −0.67 ± 2.17 | 0.32±1.25 | 46.625 |
| Contr_l | PAGNN | −5.26 ± 0.92 | −15.77 ± 0.11 | −5.59 ± 2.97 | −15.58 ± 0.14 | 167.25 |
| Contr_l | SAGE | −2.43 ± 0.78 | −5.88 ± 1.04 | −2.34 ± 1.60 | −7.92 ± 1.12 | 111.625 |
| Contr_l + CrossE_L | ALL | −4.33 ± 0.78 | −6.59 ± 1.06 | −7.00 ± 4.58 | −12.14 ± 4.87 | 141.75 |
| Contr_l + CrossE_L | GAT | −0.32 ± 0.35 | 1.07±0.44 | −0.46 ± 0.85 | 0.78±0.89 | 28.625 |

Continued on next page

Table 37. Results for Silhouette (↑) (continued)

| Loss Type | Model | Cora ↓ Citeseer | Cora ↓ Bitcoin | Citeseer ↓ Cora | Citeseer ↓ Bitcoin | Average Rank |
|---|---|---|---|---|---|---|
| Contr_l + CrossE_L | GCN | −0.51 ± 0.14 | 0.44±0.68 | −1.47 ± 1.25 | 0.32±0.57 | 57.0 |
| Contr_l + CrossE_L | GIN | −3.41 ± 0.61 | −17.28 ± 1.39 | −1.53 ± 3.06 | −15.50 ± 2.34 | 145.125 |
| Contr_l + CrossE_L | MPNN | −0.09 ± 0.79 | 0.39±1.25 | 0.36±0.99 | −0.94 ± 0.35 | 47.25 |
| Contr_l + CrossE_L | PAGNN | −6.15 ± 1.06 | −15.76 ± 0.12 | −6.14 ± 1.69 | −15.99 ± 0.13 | 176.75 |
| Contr_l + CrossE_L | SAGE | −2.29 ± 0.65 | −11.08 ± 0.97 | −2.76 ± 1.20 | −15.52 ± 0.92 | 132.75 |
| Contr_l + CrossE_L + PMI_L | ALL | −6.30 ± 1.82 | −9.61 ± 3.65 | −8.81 ± 2.53 | −9.84 ± 1.96 | 147.5 |
| Contr_l + CrossE_L + PMI_L | GAT | −0.24 ± 0.48 | <mark>1.52±0.51</mark> | −0.22 ± 0.59 | 0.92±0.51 | 20.875 |
| Contr_l + CrossE_L + PMI_L | GCN | −0.54 ± 0.23 | 0.38±0.29 | −0.73 ± 0.63 | −0.27 ± 1.28 | 60.5 |
| Contr_l + CrossE_L + PMI_L | GIN | −4.54 ± 0.74 | −11.27 ± 1.61 | −9.18 ± 3.12 | −14.59 ± 1.72 | 152.75 |
| Contr_l + CrossE_L + PMI_L | MPNN | 0.11±0.51 | 0.08±0.80 | 0.48±1.87 | 0.44±0.97 | 38.5 |
| Contr_l + CrossE_L + PMI_L | PAGNN | −9.74 ± 0.45 | −15.63 ± 0.03 | −21.33 ± 1.56 | −15.62 ± 0.05 | 189.75 |
| Contr_l + CrossE_L + PMI_L | SAGE | −1.01 ± 0.35 | −5.78 ± 1.03 | −1.97 ± 0.36 | −5.04 ± 0.83 | 93.5 |
| Contr_l + CrossE_L + PMI_L + PR_L | ALL | −18.67 ± 2.23 | −12.06 ± 3.67 | −14.87 ± 1.98 | −8.41 ± 1.88 | 169.0 |
| Contr_l + CrossE_L + PMI_L + PR_L | GAT | −0.56 ± 0.52 | 0.50±0.90 | −0.20 ± 1.26 | 1.51±0.83 | 37.25 |
| Contr_l + CrossE_L + PMI_L + PR_L | GCN | −0.51 ± 0.29 | 0.47±0.32 | −0.54 ± 0.95 | 0.46±0.92 | 46.5 |
| Contr_l + CrossE_L + PMI_L + PR_L | GIN | −5.96 ± 0.77 | −11.36 ± 1.59 | −11.38 ± 4.90 | −12.23 ± 1.64 | 157.0 |

Table 37. Results for Silhouette (↑) (continued)

| Loss Type | Model | Cora ↓ Citeseer | Cora ↓ Bitcoin | Citeseer ↓ Cora | Citeseer ↓ Bitcoin | Average Rank |
|---|---|---|---|---|---|---|
| Contr_l + CrossE_L + PMI_L + PR_L | MPNN | 0.81±0.56 | 1.60±0.73 | −2.03 ± 4.60 | −3.32 ± 1.25 | 47.75 |
| Contr_l + CrossE_L + PMI_L + PR_L | PAGNN | −8.55 ± 0.82 | −15.66 ± 0.02 | −17.04 ± 1.37 | −15.70 ± 0.02 | 187.0 |
| Contr_l + CrossE_L + PMI_L + PR_L | SAGE | −1.16 ± 0.29 | −4.98 ± 0.85 | −1.91 ± 0.85 | −4.23 ± 0.68 | 89.0 |
| Contr_l + CrossE_L + PMI_L + PR_L + Triplet_L | ALL | −10.26 ± 0.90 | −15.24 ± 3.35 | −9.99 ± 3.98 | −9.90 ± 1.76 | 169.5 |
| Contr_l + CrossE_L + PMI_L + PR_L + Triplet_L | GAT | −0.48 ± 0.56 | 1.24±0.51 | −1.01 ± 1.29 | 0.75±0.80 | 35.375 |
| Contr_l + CrossE_L + PMI_L + PR_L + Triplet_L | GCN | −0.69 ± 0.40 | 0.24±0.45 | −0.78 ± 0.28 | −0.18 ± 0.97 | 62.5 |
| Contr_l + CrossE_L + PMI_L + PR_L + Triplet_L | GIN | −4.88 ± 0.68 | −10.87 ± 1.70 | −5.80 ± 2.22 | −15.04 ± 1.52 | 149.25 |
| Contr_l + CrossE_L + PMI_L + PR_L + Triplet_L | MPNN | 0.31±0.63 | 0.75±1.10 | −1.65 ± 3.45 | −1.96 ± 1.16 | 55.75 |
| Contr_l + CrossE_L + PMI_L + PR_L + Triplet_L | PAGNN | −9.56 ± 0.57 | −15.63 ± 0.03 | −15.74 ± 1.63 | −15.67 ± 0.02 | 188.125 |
| Contr_l + CrossE_L + PMI_L + PR_L + Triplet_L | SAGE | −1.70 ± 0.78 | −5.28 ± 0.72 | −2.34 ± 0.59 | −5.05 ± 1.48 | 101.875 |
| Contr_l + CrossE_L + PMI_L + Triplet_L | ALL | −2.79 ± 1.34 | −5.03 ± 3.50 | −2.76 ± 1.97 | −1.18 ± 2.96 | 104.75 |
| Contr_l + CrossE_L + PMI_L + Triplet_L | GAT | −0.24 ± 0.55 | 0.59±0.61 | −0.25 ± 1.42 | 1.10±0.38 | 32.125 |
| Contr_l + CrossE_L + PMI_L + Triplet_L | GCN | −0.60 ± 0.22 | 0.46±0.88 | −1.06 ± 1.60 | 0.58±0.48 | 52.5 |

Table 37. Results for Silhouette (↑) (continued)

| Loss Type | Model | Cora ↓ Citeseer | Cora ↓ Bitcoin | Citeseer ↓ Cora | Citeseer ↓ Bitcoin | Average Rank |
|---|---|---|---|---|---|---|
| Contr_l + CrossE_L + PMI_L + Triplet_L | GIN | −3.64 ± 1.13 | −12.81 ± 1.46 | −5.88 ± 3.52 | −12.49 ± 3.30 | 147.875 |
| Contr_l + CrossE_L + PMI_L + Triplet_L | MPNN | 0.32±0.48 | 0.45±0.74 | 1.75±1.06 | 0.52±1.09 | 30.0 |
| Contr_l + CrossE_L + PMI_L + Triplet_L | PAGNN | −8.73 ± 0.80 | −15.70 ± 0.03 | −19.14 ± 1.28 | −15.62 ± 0.03 | 187.375 |
| Contr_l + CrossE_L + PMI_L + Triplet_L | SAGE | −1.77 ± 0.73 | −4.88 ± 0.26 | −2.52 ± 0.53 | −4.32 ± 1.29 | 101.375 |
| Contr_l + CrossE_L + PR_L | ALL | −13.02 ± 1.25 | −12.82 ± 3.09 | −12.58 ± 4.51 | −8.51 ± 0.81 | 169.375 |
| Contr_l + CrossE_L + PR_L | GAT | −2.29 ± 1.60 | 0.79±0.60 | −2.46 ± 1.66 | 1.89±0.86 | 63.25 |
| Contr_l + CrossE_L + PR_L | GCN | −2.27 ± 1.08 | −1.96 ± 3.14 | −2.50 ± 1.72 | 0.03±1.53 | 91.25 |
| Contr_l + CrossE_L + PR_L | GIN | −4.96 ± 0.30 | −12.05 ± 0.94 | −4.93 ± 1.58 | −12.49 ± 0.87 | 147.0 |
| Contr_l + CrossE_L + PR_L | MPNN | −2.80 ± 2.44 | 0.43±1.73 | 2.26±2.18 | 1.05±0.94 | 50.375 |
| Contr_l + CrossE_L + PR_L | PAGNN | −8.03 ± 0.74 | −15.66 ± 0.01 | −12.07 ± 1.01 | −15.60 ± 0.01 | 178.5 |
| Contr_l + CrossE_L + PR_L | SAGE | −3.35 ± 0.62 | −6.46 ± 1.34 | −4.71 ± 2.15 | −6.26 ± 0.72 | 125.125 |
| Contr_l + CrossE_L + PR_L + Triplet_L | ALL | −7.99 ± 1.23 | −17.73 ± 3.88 | −9.46 ± 3.66 | −8.05 ± 5.74 | 168.125 |
| Contr_l + CrossE_L + PR_L + Triplet_L | GAT | −1.46 ± 0.76 | 0.65±0.97 | −1.11 ± 1.53 | 0.67±0.39 | 57.25 |
| Contr_l + CrossE_L + PR_L + Triplet_L | GCN | −1.49 ± 0.66 | 0.77±1.25 | −2.17 ± 1.65 | −0.77 ± 1.73 | 73.375 |
| Contr_l + CrossE_L + PR_L + Triplet_L | GIN | −3.42 ± 0.64 | −13.21 ± 2.74 | −0.94 ± 1.93 | −12.44 ± 2.00 | 125.5 |
| Contr_l + CrossE_L + PR_L + Triplet_L | MPNN | −0.00 ± 0.38 | 0.12±0.64 | 0.78±2.19 | −0.44 ± 0.99 | 45.625 |

Continued on next page

Table 37. Results for Silhouette (↑) (continued)

| Loss Type | Model | Cora ↓ Citeseer | Cora ↓ Bitcoin | Citeseer ↓ Cora | Citeseer ↓ Bitcoin | Average Rank |
|---|---|---|---|---|---|---|
| Contr_l + CrossE_L + PR_L + Triplet_L | PAGNN | −9.17 ± 0.57 | −15.83 ± 0.01 | −8.66 ± 1.99 | −15.70 ± 0.05 | 186.125 |
| Contr_l + CrossE_L + PR_L + Triplet_L | SAGE | −2.52 ± 0.67 | −7.35 ± 0.87 | −3.08 ± 1.12 | −6.95 ± 0.44 | 120.75 |
| Contr_l + CrossE_L + Triplet_L | ALL | −3.31 ± 1.02 | −10.20 ± 3.46 | −3.39 ± 2.56 | −15.77 ± 2.90 | 145.125 |
| Contr_l + CrossE_L + Triplet_L | GAT | −0.28 ± 0.40 | 0.95±0.72 | −0.30 ± 0.54 | 0.70±1.26 | 30.125 |
| Contr_l + CrossE_L + Triplet_L | GCN | −0.52 ± 0.46 | 0.36±1.06 | −1.15 ± 1.60 | 0.16±0.78 | 59.875 |
| Contr_l + CrossE_L + Triplet_L | GIN | −2.09 ± 0.76 | −12.45 ± 2.40 | −0.86 ± 2.97 | −11.09 ± 0.71 | 113.25 |
| Contr_l + CrossE_L + Triplet_L | MPNN | −0.25 ± 0.66 | 0.96±2.61 | 0.77±1.56 | −0.11 ± 1.19 | 33.625 |
| Contr_l + CrossE_L + Triplet_L | PAGNN | −5.09 ± 2.35 | −15.80 ± 0.05 | −2.20 ± 1.13 | −15.98 ± 0.15 | 161.5 |
| Contr_l + CrossE_L + Triplet_L | SAGE | −1.61 ± 0.49 | −8.21 ± 0.67 | −0.98 ± 0.96 | −7.44 ± 0.83 | 99.5 |
| Contr_l + PMI_L | ALL | −6.78 ± 2.68 | −16.93 ± 2.82 | −5.51 ± 3.69 | −4.56 ± 1.95 | 153.5 |
| Contr_l + PMI_L | GAT | −0.01 ± 0.40 | 0.91±1.06 | −0.29 ± 0.54 | 0.25±0.74 | 34.375 |
| Contr_l + PMI_L | GCN | −0.45 ± 0.42 | 0.47±1.25 | −0.07 ± 0.71 | 0.28±0.77 | 43.75 |
| Contr_l + PMI_L | GIN | −5.38 ± 2.00 | −13.20 ± 0.80 | −9.64 ± 2.17 | −13.77 ± 1.36 | 162.75 |
| Contr_l + PMI_L | MPNN | 0.46±0.65 | 0.98±1.39 | 2.66±1.23 | 0.54±0.83 | <mark>16.375</mark> |
| Contr_l + PMI_L | PAGNN | −9.58 ± 0.60 | −15.77 ± 0.02 | −22.68 ± 0.88 | −15.67 ± 0.01 | 196.125 |
| Contr_l + PMI_L | SAGE | −1.22 ± 0.40 | −4.52 ± 1.04 | −2.27 ± 1.67 | −5.00 ± 0.58 | 94.625 |
| Contr_l + PMI_L + PR_L | ALL | −18.06 ± 1.73 | −18.69 ± 2.97 | −15.21 ± 0.91 | −8.48 ± 4.56 | 184.75 |

Table 37. Results for Silhouette (↑) (continued)

| Loss Type | Model | Cora ↓ Citeseer | Cora ↓ Bitcoin | Citeseer ↓ Cora | Citeseer ↓ Bitcoin | Average Rank |
|---|---|---|---|---|---|---|
| Contr_l + PMI_L + PR_L | GAT | −0.42 ± 0.62 | 1.07±0.69 | −2.16 ± 3.06 | −2.00 ± 3.82 | 58.75 |
| Contr_l + PMI_L + PR_L | GCN | −0.74 ± 0.18 | 0.91±0.75 | −0.75 ± 1.69 | −0.10 ± 1.37 | 51.75 |
| Contr_l + PMI_L + PR_L | GIN | −5.27 ± 0.83 | −9.77 ± 1.43 | −12.63 ± 5.15 | −12.47 ± 1.76 | 154.875 |
| Contr_l + PMI_L + PR_L | MPNN | 0.59±0.61 | 0.78±1.05 | −2.76 ± 3.12 | −1.15 ± 2.12 | 60.125 |
| Contr_l + PMI_L + PR_L | PAGNN | −8.98 ± 0.79 | −15.67 ± 0.03 | −19.31 ± 2.58 | −15.74 ± 0.05 | 191.25 |
| Contr_l + PMI_L + PR_L | SAGE | −1.13 ± 0.20 | −3.71 ± 1.17 | −2.99 ± 1.20 | −5.88 ± 1.11 | 101.5 |
| Contr_l + PMI_L + PR_L + Triplet_L | ALL | −7.98 ± 1.18 | −17.15 ± 5.07 | −9.05 ± 2.45 | −10.04 ± 5.44 | 168.75 |
| Contr_l + PMI_L + PR_L + Triplet_L | GAT | −0.59 ± 0.75 | 0.71±0.79 | −1.48 ± 0.80 | −0.06 ± 2.25 | 58.5 |
| Contr_l + PMI_L + PR_L + Triplet_L | GCN | −0.61 ± 0.25 | −0.41 ± 1.11 | −1.17 ± 1.82 | 0.71±0.50 | 58.5 |
| Contr_l + PMI_L + PR_L + Triplet_L | GIN | −3.90 ± 0.82 | −12.35 ± 2.27 | −4.79 ± 1.26 | −18.97 ± 2.42 | 158.125 |
| Contr_l + PMI_L + PR_L + Triplet_L | MPNN | 0.75±0.44 | 1.37±1.13 | −1.99 ± 2.18 | −2.15 ± 2.08 | 46.875 |
| Contr_l + PMI_L + PR_L + Triplet_L | PAGNN | −8.74 ± 0.76 | −15.74 ± 0.04 | −12.30 ± 3.71 | −15.56 ± 0.04 | 182.125 |
| Contr_l + PMI_L + PR_L + Triplet_L | SAGE | −2.44 ± 0.58 | −3.71 ± 0.40 | −3.26 ± 2.39 | −8.11 ± 1.11 | 114.75 |
| Contr_l + PR_L | ALL | −11.92 ± 0.76 | −8.52 ± 1.59 | −9.78 ± 5.43 | −5.28 ± 2.41 | 152.0 |
| Contr_l + PR_L | GAT | −2.71 ± 0.91 | 0.08±0.69 | −0.57 ± 1.82 | 2.29±1.10 | 60.125 |
| Contr_l + PR_L | GCN | −1.80 ± 0.34 | 0.73±1.62 | −2.02 ± 2.68 | −0.32 ± 1.29 | 76.5 |

Table 37. Results for Silhouette (↑) (continued)

| Loss Type | Model | Cora ↓ Citeseer | Cora ↓ Bitcoin | Citeseer ↓ Cora | Citeseer ↓ Bitcoin | Average Rank |
|---|---|---|---|---|---|---|
| Contr_l + PR_L | GIN | −6.18 ± 1.97 | −11.70 ± 0.65 | −5.30 ± 2.85 | −16.60 ± 2.05 | 163.25 |
| Contr_l + PR_L | MPNN | −1.43 ± 0.91 | −0.02 ± 1.20 | 2.51±1.09 | 0.68±0.94 | 49.0 |
| Contr_l + PR_L | PAGNN | −8.37 ± 0.77 | −15.60 ± 0.02 | −11.15 ± 0.45 | −15.68 ± 0.01 | 179.25 |
| Contr_l + PR_L | SAGE | −3.08 ± 0.52 | −6.37 ± 0.71 | −3.62 ± 0.80 | −5.72 ± 1.20 | 121.25 |
| Contr_l + PR_L + Triplet_L | ALL | −9.15 ± 3.58 | −14.76 ± 1.73 | −10.45 ± 2.08 | −13.99 ± 3.04 | 173.5 |
| Contr_l + PR_L + Triplet_L | GAT | −1.50 ± 0.26 | 0.95±0.86 | −0.42 ± 1.99 | 0.90±1.39 | 41.875 |
| Contr_l + PR_L + Triplet_L | GCN | −1.54 ± 0.81 | 0.87±1.23 | −0.32 ± 1.24 | 0.24±1.40 | 52.0 |
| Contr_l + PR_L + Triplet_L | GIN | −2.59 ± 1.04 | −10.82 ± 2.62 | −2.48 ± 3.58 | −15.77 ± 0.90 | 137.5 |
| Contr_l + PR_L + Triplet_L | MPNN | −0.47 ± 1.44 | 0.53±0.74 | 1.03±1.64 | −1.84 ± 2.24 | 48.5 |
| Contr_l + PR_L + Triplet_L | PAGNN | −8.18 ± 0.82 | −15.68 ± 0.01 | −7.98 ± 2.64 | −15.65 ± 0.07 | 175.75 |
| Contr_l + PR_L + Triplet_L | SAGE | −2.47 ± 0.58 | −5.95 ± 0.87 | −2.14 ± 0.52 | −6.34 ± 0.71 | 109.5 |
| Contr_l + Triplet_L | ALL | −3.23 ± 0.88 | −14.92 ± 2.33 | −2.44 ± 4.71 | −14.70 ± 2.50 | 140.25 |
| Contr_l + Triplet_L | GAT | −0.51 ± 0.42 | 1.43±1.12 | 0.01±0.82 | 0.98±0.59 | 23.125 |
| Contr_l + Triplet_L | GCN | −0.73 ± 0.24 | 0.85±0.54 | −0.37 ± 1.03 | −0.05 ± 0.58 | 47.25 |
| Contr_l + Triplet_L | GIN | −1.75 ± 0.34 | −15.87 ± 3.09 | −0.14 ± 3.26 | −13.21 ± 1.37 | 120.75 |
| Contr_l + Triplet_L | MPNN | −0.14 ± 0.72 | 0.97±1.34 | 3.61±1.18 | −1.89 ± 2.85 | 32.875 |

Table 37. Results for Silhouette (↑) (continued)

| Loss Type | Model | Cora ↓ Citeseer | Cora ↓ Bitcoin | Citeseer ↓ Cora | Citeseer ↓ Bitcoin | Average Rank |
|-----------|-------|-----------------|-----------------|------------------|---------------------|--------------|
| Contr_l + Triplet_L | PAGNN | −4.76 ± 2.63 | −15.76 ± 0.18 | −3.24 ± 2.58 | −16.10 ± 0.41 | 166.75 |
| Contr_l + Triplet_L | SAGE | −1.73 ± 0.34 | −10.84 ± 1.59 | −1.45 ± 1.21 | −10.10 ± 1.21 | 108.875 |
| CrossE_L | ALL | −4.47 ± 3.36 | 0.57±0.69 | −6.88 ± 5.44 | −2.15 ± 2.95 | 110.875 |
| CrossE_L | GAT | −4.24 ± 0.61 | −0.54 ± 2.93 | −3.64 ± 1.92 | 1.50±0.95 | 92.25 |
| CrossE_L | GCN | −2.79 ± 0.32 | −2.50 ± 3.56 | −7.10 ± 1.01 | −8.66 ± 4.42 | 124.125 |
| CrossE_L | GIN | −15.16 ± 2.13 | −12.55 ± 0.09 | −9.53 ± 0.75 | −14.44 ± 0.02 | 172.5 |
| CrossE_L | MPNN | −6.26 ± 1.09 | −15.45 ± 1.51 | −3.66 ± 1.01 | −16.69 ± 2.20 | 169.625 |
| CrossE_L | PAGNN | −26.36 ± 2.62 | −16.00 ± 0.00 | −3.59 ± 0.78 | −15.43 ± 0.05 | 178.75 |
| CrossE_L | SAGE | −30.72 ± 3.94 | −15.93 ± 0.02 | −3.94 ± 0.72 | −15.54 ± 0.15 | 180.75 |
| CrossE_L + PMI_L | ALL | −7.95 ± 0.58 | −5.49 ± 1.05 | −10.50 ± 1.15 | −6.98 ± 1.50 | 144.75 |
| CrossE_L + PMI_L | GAT | −0.22 ± 0.40 | 0.79±0.61 | 0.40±0.50 | 0.79±0.63 | 26.125 |
| CrossE_L + PMI_L | GCN | −0.90 ± 0.62 | 0.70±0.36 | −0.59 ± 1.16 | 0.15±0.65 | 54.375 |
| CrossE_L + PMI_L | GIN | −5.30 ± 0.55 | −12.71 ± 1.90 | −11.01 ± 2.08 | −13.17 ± 2.10 | 161.0 |
| CrossE_L + PMI_L | MPNN | 0.11±0.38 | 0.36±1.05 | 0.86±1.57 | −1.13 ± 0.44 | 45.25 |
| CrossE_L + PMI_L | PAGNN | −8.85 ± 1.18 | −15.85 ± 0.04 | −21.77 ± 1.31 | −15.67 ± 0.04 | 194.5 |
| CrossE_L + PMI_L | SAGE | −1.25 ± 0.32 | −6.06 ± 0.81 | −1.23 ± 0.33 | −4.45 ± 1.26 | 89.75 |

Table 37. Results for Silhouette (↑) (continued)

| Loss Type | Model | Cora ↓ Citeseer | Cora ↓ Bitcoin | Citeseer ↓ Cora | Citeseer ↓ Bitcoin | Average Rank |
|---|---|---|---|---|---|---|
| CrossE_L + PMI_L + PR_L | ALL | −19.23 ± 0.99 | −6.90 ± 1.54 | −14.49 ± 2.15 | −10.99 ± 2.58 | 165.125 |
| CrossE_L + PMI_L + PR_L | GAT | −0.31 ± 0.42 | 0.97±0.46 | −2.67 ± 3.08 | −0.49 ± 5.04 | 60.625 |
| CrossE_L + PMI_L + PR_L | GCN | −0.77 ± 0.49 | 0.06±0.53 | −0.87 ± 0.94 | −0.07 ± 1.27 | 65.25 |
| CrossE_L + PMI_L + PR_L | GIN | −5.07 ± 1.24 | −12.97 ± 1.68 | −12.88 ± 5.08 | −13.34 ± 1.84 | 164.25 |
| CrossE_L + PMI_L + PR_L | MPNN | <span style="background-color:#90EE90">0.61±0.50</span> | <span style="background-color:#F08080">1.64±0.30</span> | −2.58 ± 6.31 | 1.77±1.03 | 30.75 |
| CrossE_L + PMI_L + PR_L | PAGNN | −8.95 ± 1.38 | −15.62 ± 0.02 | −18.56 ± 4.52 | −15.64 ± 0.03 | 186.125 |
| CrossE_L + PMI_L + PR_L | SAGE | −0.96 ± 0.08 | −5.05 ± 0.17 | −1.62 ± 0.58 | −5.42 ± 0.40 | 90.75 |
| CrossE_L + PMI_L + PR_L + Triplet_L | ALL | −9.16 ± 1.42 | −12.57 ± 3.14 | −9.97 ± 3.58 | −15.39 ± 4.83 | 171.75 |
| CrossE_L + PMI_L + PR_L + Triplet_L | GAT | 0.16±0.27 | 1.14±0.78 | −0.54 ± 0.92 | −0.19 ± 1.26 | 34.625 |
| CrossE_L + PMI_L + PR_L + Triplet_L | GCN | −0.93 ± 0.38 | 0.12±0.86 | −0.98 ± 0.38 | 0.79±0.40 | 56.5 |
| CrossE_L + PMI_L + PR_L + Triplet_L | GIN | −4.98 ± 1.09 | −11.03 ± 0.88 | −5.83 ± 2.43 | −14.16 ± 2.14 | 149.0 |
| CrossE_L + PMI_L + PR_L + Triplet_L | MPNN | 0.24±0.63 | 0.94±1.10 | −0.64 ± 2.91 | 0.32±1.93 | 33.375 |
| CrossE_L + PMI_L + PR_L + Triplet_L | PAGNN | −8.39 ± 0.74 | −15.71 ± 0.04 | −17.20 ± 2.58 | −15.53 ± 0.02 | 183.25 |
| CrossE_L + PMI_L + PR_L + Triplet_L | SAGE | −1.52 ± 0.47 | −5.12 ± 0.98 | −2.87 ± 1.09 | −6.28 ± 2.00 | 107.5 |
| CrossE_L + PMI_L + Triplet_L | ALL | −2.11 ± 0.54 | −3.71 ± 2.32 | −1.59 ± 2.15 | −4.99 ± 3.37 | 94.75 |
| CrossE_L + PMI_L + Triplet_L | GAT | −0.30 ± 0.56 | 1.08±0.64 | 0.14±0.64 | 0.98±0.62 | 20.875 |

Table 37. Results for Silhouette (↑) (continued)

| Loss Type | Model | Cora ↓ Citeseer | Cora ↓ Bitcoin | Citeseer ↓ Cora | Citeseer ↓ Bitcoin | Average Rank |
|---|---|---|---|---|---|---|
| CrossE_L + PMI_L + Triplet_L | GCN | −0.88 ± 0.29 | 0.09±0.32 | −0.17 ± 0.46 | 1.07±0.21 | 46.625 |
| CrossE_L + PMI_L + Triplet_L | GIN | −4.23 ± 1.02 | −13.59 ± 4.06 | −6.16 ± 2.30 | −15.95 ± 3.13 | 163.25 |
| CrossE_L + PMI_L + Triplet_L | MPNN | 0.18±0.77 | 1.02±0.95 | 2.71±1.70 | 0.57±0.80 | 16.75 |
| CrossE_L + PMI_L + Triplet_L | PAGNN | −9.00 ± 0.60 | −15.77 ± 0.02 | −18.30 ± 1.45 | −15.78 ± 0.03 | 194.875 |
| CrossE_L + PMI_L + Triplet_L | SAGE | −1.41 ± 0.34 | −4.29 ± 0.80 | −2.06 ± 0.98 | −4.57 ± 1.50 | 92.25 |
| CrossE_L + PR_L | ALL | −12.26 ± 1.44 | −8.37 ± 2.50 | −3.52 ± 2.70 | −4.90 ± 3.23 | 140.0 |
| CrossE_L + PR_L | GAT | −3.57 ± 0.52 | 0.83±0.69 | −2.02 ± 0.18 | 1.94±0.99 | 64.375 |
| CrossE_L + PR_L | GCN | −4.32 ± 2.20 | 0.11±4.78 | −2.48 ± 0.99 | 0.43±2.94 | 91.5 |
| CrossE_L + PR_L | GIN | −5.68 ± 0.99 | −12.39 ± 0.65 | −8.14 ± 4.92 | −11.61 ± 0.75 | 153.875 |
| CrossE_L + PR_L | MPNN | −1.83 ± 2.43 | 1.00±0.77 | 0.51±3.01 | 0.28±1.04 | 45.875 |
| CrossE_L + PR_L | PAGNN | −8.52 ± 1.23 | −15.45 ± 0.01 | −12.52 ± 2.28 | −15.55 ± 0.02 | 176.625 |
| CrossE_L + PR_L | SAGE | −3.07 ± 0.73 | −6.46 ± 1.06 | −2.33 ± 1.18 | −6.02 ± 0.75 | 113.875 |
| CrossE_L + PR_L + Triplet_L | ALL | −10.52 ± 4.75 | −20.59 ± 6.39 | −14.49 ± 2.99 | −16.89 ± 4.60 | 201.875 |
| CrossE_L + PR_L + Triplet_L | GAT | −2.81 ± 0.88 | 0.68±1.81 | −1.89 ± 2.08 | 1.21±0.69 | 66.0 |
| CrossE_L + PR_L + Triplet_L | GCN | −2.15 ± 0.44 | 0.98±1.28 | −1.72 ± 2.03 | 0.26±0.83 | 64.125 |
| CrossE_L + PR_L + Triplet_L | GIN | −3.39 ± 0.87 | −14.24 ± 1.75 | −1.75 ± 1.87 | −13.48 ± 2.32 | 133.75 |

Continued on next page

Table 37. Results for Silhouette (↑) (continued)

| Loss Type | Model | Cora ↓ Citeseer | Cora ↓ Bitcoin | Citeseer ↓ Cora | Citeseer ↓ Bitcoin | Average Rank |
|---|---|---|---|---|---|---|
| CrossE_L + PR_L + Triplet_L | MPNN | −0.25 ± 0.94 | 1.04±0.79 | 2.13±2.40 | 0.99±0.77 | 17.125 |
| CrossE_L + PR_L + Triplet_L | PAGNN | −8.17 ± 0.71 | −15.62 ± 0.02 | −9.25 ± 0.84 | −15.61 ± 0.03 | 173.125 |
| CrossE_L + PR_L + Triplet_L | SAGE | −1.73 ± 0.80 | −6.11 ± 0.53 | −2.23 ± 0.66 | −8.58 ± 0.57 | 109.375 |
| CrossE_L + Triplet_L | ALL | −1.62 ± 0.63 | −8.88 ± 1.73 | −2.71 ± 2.78 | −13.45 ± 1.36 | 122.5 |
| CrossE_L + Triplet_L | GAT | −0.33 ± 0.27 | 0.93±0.72 | 0.44±1.25 | 0.81±0.69 | 25.5 |
| CrossE_L + Triplet_L | GCN | −0.81 ± 0.52 | 0.70±0.53 | −0.36 ± 0.80 | 0.65±1.21 | 44.625 |
| CrossE_L + Triplet_L | GIN | −1.55 ± 0.39 | −15.81 ± 2.31 | 0.84±2.05 | −14.76 ± 2.05 | 116.25 |
| CrossE_L + Triplet_L | MPNN | 0.39±0.52 | 0.76±0.78 | 3.46±1.74 | −0.74 ± 1.07 | 31.875 |
| CrossE_L + Triplet_L | PAGNN | −4.26 ± 2.41 | −15.70 ± 0.12 | −2.53 ± 1.83 | −16.48 ± 0.31 | 160.0 |
| CrossE_L + Triplet_L | SAGE | −0.89 ± 0.21 | −10.45 ± 0.94 | −1.15 ± 0.98 | −11.51 ± 0.59 | 100.875 |
| PMI_L | ALL | −9.72 ± 0.96 | −4.76 ± 3.17 | −9.37 ± 2.84 | −2.60 ± 1.83 | 138.75 |
| PMI_L | GAT | −0.48 ± 0.51 | 0.77±0.40 | −0.12 ± 1.37 | 0.88±0.77 | 33.25 |
| PMI_L | GCN | −0.85 ± 0.70 | 0.13±1.33 | −1.66 ± 0.88 | 0.23±0.82 | 67.375 |
| PMI_L | GIN | −4.84 ± 0.63 | −12.78 ± 2.43 | −9.90 ± 3.39 | −10.20 ± 1.77 | 153.75 |
| PMI_L | MPNN | 0.52±0.75 | 0.76±1.18 | 2.25±1.37 | 0.79±1.67 | 19.875 |
| PMI_L | PAGNN | −8.77 ± 1.31 | −15.74 ± 0.01 | −21.58 ± 0.57 | −15.62 ± 0.02 | 189.75 |
| PMI_L | SAGE | −1.22 ± 0.34 | −5.44 ± 0.87 | −1.38 ± 0.44 | −4.68 ± 0.94 | 89.25 |

Table 37. Results for Silhouette (↑) (continued)

| Loss Type | Model | Cora ↓ Citeseer | Cora ↓ Bitcoin | Citeseer ↓ Cora | Citeseer ↓ Bitcoin | Average Rank |
|---|---|---|---|---|---|---|
| PMI_L + PR_L | ALL | −17.03 ± 2.36 | −12.16 ± 2.98 | −14.47 ± 2.27 | −11.37 ± 1.21 | 171.25 |
| PMI_L + PR_L | GAT | −0.35 ± 0.62 | 0.67±1.14 | −2.97 ± 3.46 | 0.16±1.56 | 66.75 |
| PMI_L + PR_L | GCN | −0.95 ± 0.72 | −0.10 ± 0.30 | −0.63 ± 1.26 | −0.13 ± 1.11 | 66.625 |
| PMI_L + PR_L | GIN | −5.64 ± 1.45 | −12.02 ± 2.75 | −9.46 ± 4.10 | −19.41 ± 1.43 | 170.25 |
| PMI_L + PR_L | MPNN | 0.57±0.44 | 0.43±1.45 | −5.70 ± 3.20 | −2.23 ± 2.30 | 77.625 |
| PMI_L + PR_L | PAGNN | −7.94 ± 1.03 | −15.65 ± 0.03 | −17.80 ± 1.58 | −15.58 ± 0.03 | 180.375 |
| PMI_L + PR_L | SAGE | −1.17 ± 0.28 | −4.10 ± 0.87 | −1.21 ± 0.36 | −5.09 ± 0.85 | 86.875 |
| PMI_L + PR_L + Triplet_L | ALL | −9.79 ± 0.83 | −21.01 ± 3.67 | −9.55 ± 2.03 | −20.69 ± 5.23 | 197.25 |
| PMI_L + PR_L + Triplet_L | GAT | −0.45 ± 0.52 | 0.77±0.34 | −0.64 ± 0.95 | −0.36 ± 1.23 | 50.125 |
| PMI_L + PR_L + Triplet_L | GCN | −0.80 ± 0.44 | −0.01 ± 0.96 | −0.71 ± 0.73 | 0.06±0.66 | 63.25 |
| PMI_L + PR_L + Triplet_L | GIN | −4.27 ± 0.59 | −13.78 ± 2.67 | −6.97 ± 2.71 | −16.55 ± 1.38 | 166.5 |
| PMI_L + PR_L + Triplet_L | MPNN | 0.12±0.57 | 0.73±0.65 | −2.76 ± 3.02 | −0.78 ± 1.15 | 64.25 |
| PMI_L + PR_L + Triplet_L | PAGNN | −8.86 ± 1.00 | −15.70 ± 0.02 | −13.44 ± 2.55 | −15.62 ± 0.03 | 185.125 |
| PMI_L + PR_L + Triplet_L | SAGE | −1.71 ± 0.36 | −5.08 ± 1.01 | −3.07 ± 2.05 | −6.31 ± 0.88 | 110.75 |
| PMI_L + Triplet_L | ALL | −1.51 ± 0.77 | −12.02 ± 1.38 | −2.45 ± 2.29 | −1.20 ± 1.08 | 104.25 |
| PMI_L + Triplet_L | GAT | −0.42 ± 0.55 | 1.12±0.52 | 0.13±0.59 | 1.26±0.38 | 20.875 |

Table 37. Results for Silhouette (↑) (continued)

| Loss Type | Model | Cora ↓ Citeseer | Cora ↓ Bitcoin | Citeseer ↓ Cora | Citeseer ↓ Bitcoin | Average Rank |
|---|---|---|---|---|---|---|
| PMI_L + Triplet_L | GCN | −1.20 ± 0.38 | −0.10 ± 0.75 | −0.65 ± 0.91 | −0.05 ± 0.87 | 67.5 |
| PMI_L + Triplet_L | GIN | −4.07 ± 0.77 | −11.20 ± 3.06 | −5.69 ± 3.02 | −13.33 ± 2.04 | 143.25 |
| PMI_L + Triplet_L | MPNN | 0.19±0.39 | 0.92±0.40 | 2.15±1.60 | 0.41±1.77 | 22.625 |
| PMI_L + Triplet_L | PAGNN | −9.06 ± 0.65 | −15.77 ± 0.01 | −19.64 ± 2.04 | −15.69 ± 0.03 | 194.375 |
| PMI_L + Triplet_L | SAGE | −1.77 ± 0.49 | −4.02 ± 0.89 | −2.83 ± 0.66 | −5.36 ± 0.78 | 105.875 |
| PR_L | ALL | −16.42 ± 1.82 | −2.95 ± 2.32 | −6.49 ± 1.87 | −2.22 ± 1.06 | 135.5 |
| PR_L | GAT | −3.82 ± 0.51 | 1.25±1.00 | −2.45 ± 1.05 | 1.71±1.03 | 62.75 |
| PR_L | GCN | −4.08 ± 0.71 | 0.52±2.31 | −2.93 ± 2.06 | −0.90 ± 5.27 | 98.5 |
| PR_L | GIN | −6.01 ± 0.98 | −11.47 ± 0.62 | −4.86 ± 3.89 | −11.61 ± 0.61 | 146.375 |
| PR_L | MPNN | −1.89 ± 3.13 | 0.17±1.52 | 1.85±1.39 | 0.43±0.89 | 55.875 |
| PR_L | PAGNN | −9.50 ± 2.04 | −15.65 ± 0.01 | −14.96 ± 2.21 | −15.75 ± 0.01 | 189.75 |
| PR_L | SAGE | −2.55 ± 0.66 | −6.21 ± 0.70 | −2.88 ± 0.53 | −5.09 ± 1.38 | 114.625 |
| PR_L + Triplet_L | ALL | −12.85 ± 1.19 | −12.92 ± 2.51 | −7.65 ± 1.32 | −5.83 ± 2.53 | 159.0 |
| PR_L + Triplet_L | GAT | −3.12 ± 1.47 | −1.31 ± 2.57 | −1.95 ± 2.03 | 1.70±1.25 | 76.0 |
| PR_L + Triplet_L | GCN | −3.18 ± 0.88 | 0.09±0.82 | −2.38 ± 0.49 | 0.33±2.64 | 86.875 |
| PR_L + Triplet_L | GIN | −5.53 ± 0.79 | −12.05 ± 0.75 | −4.79 ± 1.89 | −11.82 ± 1.37 | 147.0 |
| PR_L + Triplet_L | MPNN | −1.86 ± 2.44 | 0.77±1.19 | 1.97±1.14 | −0.10 ± 1.77 | 53.5 |

Table 37. Results for Silhouette (↑) (continued)

| Loss Type | Model | Cora ↓ Citeseer | Cora ↓ Bitcoin | Citeseer ↓ Cora | Citeseer ↓ Bitcoin | Average Rank |
|---|---|---|---|---|---|---|
| PR_L + Triplet_L | PAGNN | −8.71 ± 1.38 | −15.65 ± 0.02 | −11.74 ± 1.78 | −15.60 ± 0.00 | 179.375 |
| PR_L + Triplet_L | SAGE | −3.41 ± 0.34 | −5.22 ± 1.10 | −3.44 ± 0.96 | −4.68 ± 0.86 | 116.75 |
| Triplet_L | ALL | −1.57 ± 1.59 | −12.82 ± 1.35 | −0.70 ± 2.53 | −1.87 ± 1.19 | 96.375 |
| Triplet_L | GAT | −0.13 ± 0.45 | 1.25±0.60 | −0.08 ± 0.93 | 1.67±0.50 | 16.625 |
| Triplet_L | GCN | −0.39 ± 0.38 | 0.67±0.67 | −0.67 ± 1.61 | 0.38±0.37 | 45.75 |
| Triplet_L | GIN | −0.85 ± 0.40 | −11.94 ± 2.78 | −1.51 ± 1.72 | −12.09 ± 0.90 | 106.375 |
| Triplet_L | MPNN | 0.57±0.50 | −0.88 ± 1.90 | 2.56±1.90 | 0.41±1.01 | 34.75 |
| Triplet_L | PAGNN | −4.29 ± 2.25 | −15.76 ± 0.20 | −1.90 ± 0.47 | −16.21 ± 0.23 | 155.0 |
| Triplet_L | SAGE | −1.06 ± 0.67 | −6.41 ± 0.67 | −1.58 ± 1.26 | −5.65 ± 0.85 | 94.5 |

Table 38. Calinski Harabasz Performance (↑): This table presents models (Loss function and GNN) ranked by their average performance in terms of calinski harabasz. Top-ranked results are highlighted in red, second-ranked in blue, and third-ranked in green.

| Loss Type | Model | Cora ↓ Citeseer | Cora ↓ Bitcoin | Citeseer ↓ Cora | Citeseer ↓ Bitcoin | Average Rank |
|---|---|---|---|---|---|---|
| Contr_l | ALL | 5056.31 ± 1888.38 | 339.50 ± 80.15 | 8020.35 ± 2312.85 | 725.95 ± 170.14 | 111.0 |
| Contr_l | GAT | 3016.36 ± 446.09 | 1874.95 ± 584.69 | 5474.38 ± 397.88 | 1598.64 ± 447.81 | 71.5 |

Continued on next page

Table 38. Results for Calinski Harabasz (↑) (continued)

| Loss Type | Model | Cora ↓ Citeseer | Cora ↓ Bitcoin | Citeseer ↓ Cora | Citeseer ↓ Bitcoin | Average Rank |
|-----------|-------|------|------|------|------|------|
| Contr_l | GCN | 2914.12 ± 443.46 | 1273.19 ± 232.90 | 5061.46 ± 860.74 | 1178.24 ± 129.08 | 99.5 |
| Contr_l | GIN | 5891.40 ± 1098.42 | 915.36 ± 152.42 | 7815.41 ± 2029.71 | 967.11 ± 209.16 | 87.75 |
| Contr_l | MPNN | 5365.13 ± 1824.76 | 1984.02 ± 589.76 | 7175.40 ± 865.31 | 1624.04 ± 524.10 | 44.75 |
| Contr_l | PAGNN | 7995.32 ± 937.18 | 192.91 ± 5.79 | 5884.43 ± 1345.98 | 181.18 ± 12.79 | 128.0 |
| Contr_l | SAGE | 2324.63 ± 749.89 | 609.18 ± 86.46 | 4584.84 ± 1441.20 | 538.34 ± 74.53 | 156.25 |
| Contr_l + CrossE_L | ALL | 5097.31 ± 1014.66 | 814.82 ± 442.36 | 6621.22 ± 2418.01 | 326.92 ± 86.51 | 113.25 |
| Contr_l + CrossE_L | GAT | 3280.32 ± 265.02 | 1761.00 ± 458.60 | 4931.19 ± 326.55 | 1738.41 ± 254.77 | 70.0 |
| Contr_l + CrossE_L | GCN | 3184.45 ± 540.71 | 1199.52 ± 270.00 | 4641.77 ± 768.90 | 1252.26 ± 601.55 | 101.25 |
| Contr_l + CrossE_L | GIN | 6063.65 ± 224.96 | 668.73 ± 185.57 | 7296.45 ± 1257.05 | 826.88 ± 300.78 | 98.5 |
| Contr_l + CrossE_L | MPNN | 4837.77 ± 413.82 | 1460.55 ± 344.20 | 8705.60 ± 1150.59 | 1402.38 ± 98.83 | 60.5 |
| Contr_l + CrossE_L | PAGNN | 7676.60 ± 709.06 | 195.19 ± 3.63 | 5274.55 ± 571.39 | 200.20 ± 7.49 | 123.5 |
| Contr_l + CrossE_L | SAGE | 2815.68 ± 299.82 | 407.58 ± 128.55 | 5102.13 ± 404.99 | 223.40 ± 54.64 | 149.25 |
| Contr_l + CrossE_L + PMI_L | ALL | 7196.32 ± 851.89 | 744.57 ± 115.26 | 4013.75 ± 2208.62 | 776.46 ± 224.91 | 116.0 |
| Contr_l + CrossE_L + PMI_L | GAT | 2827.18 ± 375.36 | 1922.52 ± 534.51 | 4764.69 ± 319.18 | 1696.09 ± 154.50 | 77.5 |
| Contr_l + CrossE_L + PMI_L | GCN | 3314.13 ± 426.39 | 1005.62 ± 216.95 | 4806.97 ± 798.29 | 1210.16 ± 232.85 | 103.0 |
| Contr_l + CrossE_L + PMI_L | GIN | 6422.68 ± 433.92 | 622.06 ± 141.43 | 3615.54 ± 610.01 | 773.29 ± 203.37 | 128.25 |

Table 38. Results for Calinski Harabasz (↑) (continued)

| Loss Type | Model | Cora ↓ Citeseer | Cora ↓ Bitcoin | Citeseer ↓ Cora | Citeseer ↓ Bitcoin | Average Rank |
|---|---|---|---|---|---|---|
| Contr_l + CrossE_L + PMI_L | MPNN | 4395.20 ± 309.96 | 1831.50 ± 294.19 | 7808.13 ± 426.74 | 1809.50 ± 328.16 | 49.25 |
| Contr_l + CrossE_L + PMI_L | PAGNN | 11672.89± 313.14 | 190.11 ± 0.43 | 2099.16 ± 125.65 | 188.21 ± 1.90 | 148.5 |
| Contr_l + CrossE_L + PMI_L | SAGE | 1487.67 ± 153.68 | 1185.45 ± 284.27 | 1995.47 ± 376.36 | 1013.68 ± 385.97 | 147.25 |
| Contr_l + CrossE_L + PMI_L + PR_L | ALL | 8881.69 ± 1407.05 | 467.26 ± 14.89 | 2005.04 ± 87.56 | 593.54 ± 129.42 | 136.5 |
| Contr_l + CrossE_L + PMI_L + PR_L | GAT | 2894.70 ± 359.95 | 1562.44 ± 353.35 | 4411.29 ± 596.48 | 1713.92 ± 599.10 | 93.0 |
| Contr_l + CrossE_L + PMI_L + PR_L | GCN | 3338.44 ± 457.99 | 1223.81 ± 320.30 | 5093.49 ± 1064.11 | 1216.98 ± 416.67 | 91.5 |
| Contr_l + CrossE_L + PMI_L + PR_L | GIN | 8200.56 ± 1202.64 | 650.03 ± 186.62 | 3540.36 ± 1521.43 | 375.22 ± 72.62 | 131.25 |
| Contr_l + CrossE_L + PMI_L + PR_L | MPNN | 5196.54 ± 667.28 | 1953.28 ± 412.99 | 6530.99 ± 1369.54 | 1362.98 ± 432.32 | 56.25 |
| Contr_l + CrossE_L + PMI_L + PR_L | PAGNN | 10928.40± 1049.16 | 191.58 ± 0.56 | 2443.44 ± 193.21 | 185.48 ± 0.65 | 152.0 |
| Contr_l + CrossE_L + PMI_L + PR_L | SAGE | 1208.17 ± 147.51 | 1393.77 ± 518.20 | 1812.85 ± 107.63 | 1069.69 ± 422.17 | 143.5 |
| Contr_l + CrossE_L + PMI_L + PR_L + Triplet_L | ALL | 7168.41 ± 785.30 | 472.07 ± 231.94 | 3488.40 ± 936.48 | 792.70 ± 444.49 | 129.0 |
| Contr_l + CrossE_L + PMI_L + PR_L + Triplet_L | GAT | 3147.78 ± 292.20 | 1709.31 ± 176.47 | 4481.11 ± 541.16 | 1683.06 ± 411.19 | 84.25 |
| Contr_l + CrossE_L + PMI_L + PR_L + Triplet_L | GCN | 2805.51 ± 151.28 | 1010.31 ± 204.50 | 5173.83 ± 395.68 | 1195.02 ± 311.04 | 108.5 |
| Contr_l + CrossE_L + PMI_L + PR_L + Triplet_L | GIN | 6767.95 ± 804.99 | 741.58 ± 272.25 | 4651.86 ± 1123.67 | 520.10 ± 105.78 | 119.25 |

Table 38. Results for Calinski Harabasz (↑) (continued)

| Loss Type | Model | Cora ↓ Citeseer | Cora ↓ Bitcoin | Citeseer ↓ Cora | Citeseer ↓ Bitcoin | Average Rank |
|---|---|---|---|---|---|---|
| Contr_l + CrossE_L + PMI_L + PR_L + Triplet_L | MPNN | 4729.70 ± 468.11 | 2257.26 ± 638.65 | 7250.05 ± 1537.18 | 1410.18 ± 500.59 | 51.75 |
| Contr_l + CrossE_L + PMI_L + PR_L + Triplet_L | PAGNN | 11624.18± 863.50 | 188.76 ± 0.48 | 2611.96 ± 150.59 | 186.45 ± 0.81 | 147.5 |
| Contr_l + CrossE_L + PMI_L + PR_L + Triplet_L | SAGE | 2177.41 ± 751.61 | 1023.98 ± 326.39 | 3002.33 ± 352.47 | 1343.42 ± 406.71 | 133.0 |
| Contr_l + CrossE_L + PMI_L + Triplet_L | ALL | 6359.21 ± 1427.41 | 1057.61 ± 431.84 | 6962.95 ± 1048.82 | 1220.67 ± 281.73 | 75.5 |
| Contr_l + CrossE_L + PMI_L + Triplet_L | GAT | 2944.66 ± 377.78 | 1697.24 ± 209.01 | 4762.55 ± 692.08 | 1784.44 ± 264.55 | 78.25 |
| Contr_l + CrossE_L + PMI_L + Triplet_L | GCN | 2812.29 ± 303.72 | 1106.72 ± 317.57 | 4800.00 ± 920.14 | 1095.40 ± 654.38 | 114.75 |
| Contr_l + CrossE_L + PMI_L + Triplet_L | GIN | 7305.25 ± 655.64 | 705.12 ± 192.30 | 4612.81 ± 828.89 | 673.23 ± 139.49 | 115.25 |
| Contr_l + CrossE_L + PMI_L + Triplet_L | MPNN | 4662.19 ± 325.16 | 1593.87 ± 402.65 | 8414.63 ± 862.67 | 1428.35 ± 246.43 | 61.5 |
| Contr_l + CrossE_L + PMI_L + Triplet_L | PAGNN | 11109.19± 269.30 | 194.40 ± 0.72 | 2190.40 ± 44.57 | 185.87 ± 0.97 | 149.5 |
| Contr_l + CrossE_L + PMI_L + Triplet_L | SAGE | 2311.71 ± 522.84 | 1114.31 ± 164.58 | 3416.13 ± 703.11 | 1215.00 ± 453.51 | 131.5 |
| Contr_l + CrossE_L + PR_L | ALL | 11645.43± 898.92 | 833.75 ± 167.11 | 5479.21 ± 4412.34 | 802.29 ± 152.28 | 83.0 |
| Contr_l + CrossE_L + PR_L | GAT | 2481.23 ± 230.84 | 2142.85 ± 313.99 | 4451.09 ± 377.23 | 2042.80 ± 34.04 | 80.75 |
| Contr_l + CrossE_L + PR_L | GCN | 2577.94 ± 128.78 | 1309.23 ± 513.19 | 4417.33 ± 681.01 | 1069.49 ± 352.35 | 121.75 |
| Contr_l + CrossE_L + PR_L | GIN | 6035.83 ± 767.48 | 417.42 ± 165.89 | 5919.30 ± 1406.06 | 626.17 ± 266.43 | 117.25 |
| Contr_l + CrossE_L + PR_L | MPNN | 8354.59 ± 2933.25 | 1743.45 ± 327.10 | 11140.48± 1106.41 | 1785.37 ± 459.99 | 25.0 |

Table 38. Results for Calinski Harabasz (↑) (continued)

| Loss Type | Model | Cora ↓ Citeseer | Cora ↓ Bitcoin | Citeseer ↓ Cora | Citeseer ↓ Bitcoin | Average Rank |
|---|---|---|---|---|---|---|
| Contr_l + CrossE_L + PR_L | PAGNN | 8490.39 ± 879.89 | 191.57 ± 0.61 | 8631.23 ± 1515.19 | 187.98 ± 0.27 | 113.0 |
| Contr_l + CrossE_L + PR_L | SAGE | 2611.77 ± 1334.50 | 862.15 ± 352.80 | 3042.85 ± 1216.23 | 1165.40 ± 366.84 | 139.75 |
| Contr_l + CrossE_L + ALL PR_L + Triplet_L | ALL | 8594.78 ± 2184.96 | 435.58 ± 262.48 | 4594.83 ± 1752.23 | 705.36 ± 50.61 | 116.75 |
| Contr_l + CrossE_L + GAT PR_L + Triplet_L | GAT | 2883.87 ± 347.41 | 1769.25 ± 427.79 | 4706.48 ± 675.44 | 1759.84 ± 243.48 | 80.25 |
| Contr_l + CrossE_L + GCN PR_L + Triplet_L | GCN | 2758.70 ± 361.26 | 1163.95 ± 323.71 | 3977.86 ± 253.67 | 1236.75 ± 232.95 | 119.25 |
| Contr_l + CrossE_L + GIN PR_L + Triplet_L | GIN | 5945.88 ± 427.49 | 566.17 ± 128.58 | 6320.81 ± 995.09 | 597.93 ± 125.65 | 115.25 |
| Contr_l + CrossE_L + MPNN PR_L + Triplet_L | MPNN | 6596.77 ± 315.95 | 1688.64 ± 421.56 | 9058.54 ± 1830.16 | 1963.42 ± 756.79 | 35.5 |
| Contr_l + CrossE_L + PAGNN PR_L + Triplet_L | PAGNN | 11175.42± 681.39 | 200.11 ± 0.36 | 5257.00 ± 1732.65 | 195.60 ± 1.60 | 114.5 |
| Contr_l + CrossE_L + SAGE PR_L + Triplet_L | SAGE | 2686.28 ± 941.92 | 600.01 ± 150.20 | 3578.90 ± 358.74 | 696.68 ± 147.11 | 158.5 |
| Contr_l + CrossE_L + ALL Triplet_L | ALL | 4640.79 ± 1117.76 | 742.95 ± 262.03 | 6965.11 ± 1513.92 | 604.38 ± 300.53 | 110.25 |
| Contr_l + CrossE_L + GAT Triplet_L | GAT | 3191.22 ± 484.37 | 1882.75 ± 463.59 | 5158.46 ± 51.44 | 1731.70 ± 306.08 | 65.0 |
| Contr_l + CrossE_L + GCN Triplet_L | GCN | 3134.04 ± 359.24 | 1353.88 ± 281.31 | 5190.36 ± 995.68 | 1316.50 ± 241.18 | 86.5 |
| Contr_l + CrossE_L + GIN Triplet_L | GIN | 5052.25 ± 665.63 | 865.56 ± 238.62 | 6955.06 ± 1580.64 | 930.75 ± 199.70 | 95.5 |
| Contr_l + CrossE_L + MPNN Triplet_L | MPNN | 4687.06 ± 648.90 | 2240.34 ± 773.26 | 8335.07 ± 856.28 | 1517.63 ± 518.89 | 47.75 |
| Contr_l + CrossE_L + PAGNN Triplet_L | PAGNN | 8546.61 ± 2056.41 | 197.87 ± 2.70 | 6706.20 ± 1007.98 | 184.57 ± 8.72 | 117.25 |
| Contr_l + CrossE_L + SAGE Triplet_L | SAGE | 2760.22 ± 625.21 | 609.54 ± 134.96 | 4589.72 ± 980.80 | 729.40 ± 191.58 | 145.5 |

Table 38. Results for Calinski Harabasz (↑) (continued)

| Loss Type | Model | Cora ↓ Citeseer | Cora ↓ Bitcoin | Citeseer ↓ Cora | Citeseer ↓ Bitcoin | Average Rank |
|---|---|---|---|---|---|---|
| Contr_l + PMI_L | ALL | 7145.44 ± 1239.37 | 604.79 ± 399.37 | 5522.38 ± 1335.32 | 1160.70 ± 417.23 | 94.5 |
| Contr_l + PMI_L | GAT | 3152.17 ± 202.29 | 1544.55 ± 295.26 | 4755.90 ± 238.43 | 1411.29 ± 218.37 | 88.0 |
| Contr_l + PMI_L | GCN | 3023.37 ± 351.95 | 1109.53 ± 117.97 | 5030.06 ± 657.81 | 1051.88 ± 102.63 | 108.25 |
| Contr_l + PMI_L | GIN | 7544.68 ± 1636.44 | 922.69 ± 367.48 | 3759.86 ± 637.40 | 719.36 ± 60.48 | 115.25 |
| Contr_l + PMI_L | MPNN | 4852.39 ± 419.41 | 1887.22 ± 393.45 | 8403.98 ± 1009.09 | 1650.58 ± 529.90 | 46.0 |
| Contr_l + PMI_L | PAGNN | 11526.30± 668.83 | 197.60 ± 0.71 | 1935.81 ± 95.89 | 192.37 ± 0.34 | 145.5 |
| Contr_l + PMI_L | SAGE | 1250.98 ± 159.67 | 1413.29 ± 396.55 | 2524.98 ± 657.98 | 1161.14 ± 262.39 | 134.75 |
| Contr_l + PMI_L + PR_L | ALL | 9307.33 ± 1869.13 | 397.38 ± 49.31 | 2005.37 ± 239.89 | 583.55 ± 72.10 | 137.75 |
| Contr_l + PMI_L + PR_L | GAT | 2888.92 ± 191.57 | 1766.69 ± 270.77 | 4704.98 ± 318.42 | 1616.36 ± 851.85 | 85.5 |
| Contr_l + PMI_L + PR_L | GCN | 3197.16 ± 379.50 | 1232.04 ± 366.67 | 5425.50 ± 172.28 | 1291.21 ± 254.16 | 86.25 |
| Contr_l + PMI_L + PR_L | GIN | 7263.45 ± 771.17 | 680.00 ± 199.69 | 2686.91 ± 1365.13 | 400.59 ± 132.41 | 137.25 |
| Contr_l + PMI_L + PR_L | MPNN | 4803.51 ± 565.44 | 1982.88 ± 467.48 | 6343.15 ± 1037.18 | 1598.13 ± 491.98 | 56.75 |
| Contr_l + PMI_L + PR_L | PAGNN | 11750.84± 242.32 | 188.47 ± 0.81 | 2240.98 ± 365.37 | 191.64 ± 1.11 | 147.0 |
| Contr_l + PMI_L + PR_L | SAGE | 1373.02 ± 130.94 | 1307.42 ± 670.68 | 3295.74 ± 912.37 | 874.82 ± 580.41 | 141.25 |
| Contr_l + PMI_L + PR_L + Triplet_L | ALL | 7204.72 ± 963.44 | 401.93 ± 51.87 | 4035.79 ± 1089.45 | 740.36 ± 442.15 | 128.25 |
| Contr_l + PMI_L + PR_L + Triplet_L | GAT | 3072.10 ± 554.63 | 1891.88 ± 814.54 | 5104.27 ± 585.66 | 1484.29 ± 645.51 | 72.75 |

Table 38. Results for Calinski Harabasz (↑) (continued)

| Loss Type | Model | Cora ↓ Citeseer | Cora ↓ Bitcoin | Citeseer ↓ Cora | Citeseer ↓ Bitcoin | Average Rank |
|---|---|---|---|---|---|---|
| Contr_l + PMI_L + PR_L + Triplet_L | GCN | 2714.38 ± 396.44 | 1146.07 ± 182.04 | 4826.30 ± 991.67 | 918.99 ± 225.81 | 119.25 |
| Contr_l + PMI_L + PR_L + Triplet_L | GIN | 6797.11 ± 1133.54 | 773.53 ± 304.33 | 4343.40 ± 452.12 | 740.92 ± 281.51 | 117.5 |
| Contr_l + PMI_L + PR_L + Triplet_L | MPNN | 4971.04 ± 595.06 | 1960.79 ± 247.25 | 7194.18 ± 1888.48 | 1120.30 ± 135.64 | 62.75 |
| Contr_l + PMI_L + PR_L + Triplet_L | PAGNN | 11507.21± 899.19 | 194.90 ± 1.01 | 3542.53 ± 833.54 | 187.20 ± 1.12 | 139.0 |
| Contr_l + PMI_L + PR_L + Triplet_L | SAGE | 2300.25 ± 222.42 | 1363.26 ± 384.45 | 4044.44 ± 1134.32 | 539.26 ± 156.50 | 140.0 |
| Contr_l + PR_L | ALL | 11381.02± 521.55 | 681.36 ± 280.26 | 6388.25 ± 4048.81 | 1035.35 ± 81.63 | 81.5 |
| Contr_l + PR_L | GAT | 2820.46 ± 845.77 | 1675.07 ± 420.21 | 4437.54 ± 675.74 | 2116.96 ± 334.29 | 85.0 |
| Contr_l + PR_L | GCN | 2589.68 ± 250.31 | 1118.07 ± 344.23 | 4487.98 ± 544.80 | 1051.27 ± 314.97 | 126.75 |
| Contr_l + PR_L | GIN | 6698.68 ± 1721.50 | 385.02 ± 76.53 | 7111.69 ± 2719.72 | 595.05 ± 177.07 | 110.5 |
| Contr_l + PR_L | MPNN | 8845.30 ± 579.92 | 1747.83 ± 447.73 | 11783.43± 5279.23 | 1891.84 ± 502.29 | 22.0 |
| Contr_l + PR_L | PAGNN | 8157.40 ± 664.39 | 180.87 ± 0.54 | 9762.50 ± 942.96 | 193.79 ± 0.43 | 111.5 |
| Contr_l + PR_L | SAGE | 2351.27 ± 746.33 | 722.31 ± 168.97 | 3558.47 ± 793.55 | 1366.23 ± 287.93 | 134.75 |
| Contr_l + PR_L + Triplet_L | ALL | 7623.82 ± 3647.77 | 892.44 ± 498.73 | 3536.15 ± 1248.06 | 343.54 ± 140.03 | 126.5 |
| Contr_l + PR_L + Triplet_L | GAT | 2877.90 ± 424.65 | 1762.88 ± 371.92 | 5166.01 ± 827.66 | 1715.79 ± 175.09 | 75.25 |
| Contr_l + PR_L + Triplet_L | GCN | 2879.16 ± 265.44 | 1645.81 ± 477.97 | 4805.45 ± 233.67 | 1029.68 ± 199.16 | 104.0 |
| Contr_l + PR_L + Triplet_L | GIN | 6630.99 ± 613.95 | 966.68 ± 212.38 | 6704.86 ± 1942.58 | 819.44 ± 217.22 | 89.75 |

Continued on next page

Table 38. Results for Calinski Harabasz (↑) (continued)

| Loss Type | Model | Cora ↓ Citeseer | Cora ↓ Bitcoin | Citeseer ↓ Cora | Citeseer ↓ Bitcoin | Average Rank |
|---|---|---|---|---|---|---|
| Contr_l + PR_L + Triplet_L | MPNN | 6328.80 ± 1113.48 | 1972.21 ± 255.52 | 8944.50 ± 2344.77 | 1843.75 ± 621.59 | 31.0 |
| Contr_l + PR_L + Triplet_L | PAGNN | 10974.09± 1054.16 | 193.97 ± 0.47 | 5332.71 ± 1078.13 | 186.19 ± 3.34 | 122.0 |
| Contr_l + PR_L + Triplet_L | SAGE | 2385.62 ± 376.54 | 915.50 ± 318.24 | 4060.63 ± 298.26 | 721.47 ± 210.00 | 145.5 |
| Contr_l + Triplet_L | ALL | 4281.56 ± 701.73 | 714.43 ± 516.54 | 9003.41 ± 3576.78 | 463.24 ± 209.01 | 110.5 |
| Contr_l + Triplet_L | GAT | 3164.50 ± 334.27 | 1927.66 ± 735.33 | 5369.60 ± 568.22 | 1747.64 ± 372.30 | 61.0 |
| Contr_l + Triplet_L | GCN | 2948.80 ± 566.01 | 1222.16 ± 374.37 | 5409.14 ± 486.13 | 1429.35 ± 378.47 | 87.75 |
| Contr_l + Triplet_L | GIN | 5048.74 ± 639.97 | 879.98 ± 339.83 | 7533.98 ± 2235.24 | 949.53 ± 107.15 | 92.0 |
| Contr_l + Triplet_L | MPNN | 5460.34 ± 752.87 | 1621.82 ± 562.96 | 10211.84± 1236.38 | 1476.89 ± 298.06 | 49.75 |
| Contr_l + Triplet_L | PAGNN | 8288.75 ± 1788.53 | 191.83 ± 5.25 | 6679.71 ± 1103.15 | 181.69 ± 9.18 | 124.0 |
| Contr_l + Triplet_L | SAGE | 2758.02 ± 641.64 | 414.59 ± 70.82 | 4504.52 ± 374.13 | 485.93 ± 83.87 | 158.75 |
| CrossE_L | ALL | 5196.39 ± 2637.98 | 1214.52 ± 501.07 | 11016.04± 3250.44 | 1114.72 ± 232.62 | 69.0 |
| CrossE_L | GAT | 2371.73 ± 530.95 | 2005.52 ± 1197.31 | 4596.91 ± 771.28 | 1999.15 ± 785.16 | 80.25 |
| CrossE_L | GCN | 1233.12 ± 435.95 | 1147.71 ± 697.50 | 420.60 ± 77.73 | 864.33 ± 335.20 | 156.0 |
| CrossE_L | GIN | 3614.46 ± 960.85 | 588.36 ± 4.02 | 311.71 ± 52.88 | 600.88 ± 0.60 | 160.0 |
| CrossE_L | MPNN | 4620.71 ± 1006.63 | 375.35 ± 77.51 | 6545.09 ± 1185.75 | 444.37 ± 174.22 | 129.75 |
| CrossE_L | PAGNN | 530.48 ± 7.33 | 206.60 ± 0.01 | 13518.29± 1681.51 | 193.37 ± 1.81 | 144.5 |

Table 38. Results for Calinski Harabasz (↑) (continued)

| Loss Type | Model | Cora ↓ Citeseer | Cora ↓ Bitcoin | Citeseer ↓ Cora | Citeseer ↓ Bitcoin | Average Rank |
|---|---|---|---|---|---|---|
| CrossE_L | SAGE | 546.72 ± 12.36 | 205.60 ± 0.69 | 1990.76 ± 548.43 | 204.47 ± 5.94 | 192.0 |
| CrossE_L + PMI_L | ALL | 8067.02 ± 544.85 | 714.48 ± 131.70 | 2764.70 ± 415.29 | 744.09 ± 91.19 | 122.0 |
| CrossE_L + PMI_L | GAT | 2794.91 ± 282.27 | 1418.68 ± 129.33 | 4830.89 ± 251.17 | 1717.78 ± 224.50 | 87.5 |
| CrossE_L + PMI_L | GCN | 2915.93 ± 251.70 | 1228.39 ± 504.99 | 4724.04 ± 831.95 | 1167.65 ± 312.04 | 106.0 |
| CrossE_L + PMI_L | GIN | 6952.94 ± 619.52 | 1053.07 ± 486.06 | 3855.10 ± 727.59 | 787.30 ± 213.98 | 111.5 |
| CrossE_L + PMI_L | MPNN | 4539.95 ± 243.77 | 1878.88 ± 822.91 | 7896.60 ± 924.25 | 1378.50 ± 325.87 | 58.25 |
| CrossE_L + PMI_L | PAGNN | 11360.92± 398.84 | 192.53 ± 0.81 | 2040.20 ± 190.31 | 192.38 ± 1.39 | 147.75 |
| CrossE_L + PMI_L | SAGE | 1277.77 ± 229.78 | 858.02 ± 373.78 | 1742.66 ± 218.44 | 1234.73 ± 353.64 | 151.25 |
| CrossE_L + PMI_L + PR_L | ALL | 10329.63± 718.60 | 588.81 ± 86.03 | 1989.05 ± 405.17 | 460.08 ± 72.96 | 137.75 |
| CrossE_L + PMI_L + PR_L | GAT | 2761.53 ± 464.56 | 1671.04 ± 294.16 | 4492.41 ± 682.37 | 1428.73 ± 456.13 | 98.5 |
| CrossE_L + PMI_L + PR_L | GCN | 3057.84 ± 541.66 | 1129.77 ± 332.36 | 5178.88 ± 464.89 | 1270.79 ± 458.80 | 94.75 |
| CrossE_L + PMI_L + PR_L | GIN | 7801.47 ± 1125.93 | 619.10 ± 97.05 | 2804.45 ± 1668.26 | 684.00 ± 246.42 | 129.75 |
| CrossE_L + PMI_L + PR_L | MPNN | 4963.28 ± 523.85 | 1915.22 ± 298.64 | 7469.40 ± 2140.05 | <mark>2202.81 ± 577.69</mark> | 39.75 |
| CrossE_L + PMI_L + PR_L | PAGNN | 10974.43± 770.19 | 188.11 ± 0.82 | 2427.12 ± 713.80 | 191.34 ± 0.59 | 150.75 |
| CrossE_L + PMI_L + PR_L | SAGE | 1261.76 ± 206.42 | 929.23 ± 353.97 | 1728.22 ± 297.84 | 990.62 ± 238.56 | 158.25 |
| CrossE_L + PMI_L + PR_L + Triplet_L | ALL | 7830.48 ± 806.69 | 734.57 ± 155.50 | 3361.19 ± 1455.32 | 568.43 ± 285.45 | 126.25 |

Table 38. Results for Calinski Harabasz (↑) (continued)

| Loss Type | Model | Cora ↓ Citeseer | Cora ↓ Bitcoin | Citeseer ↓ Cora | Citeseer ↓ Bitcoin | Average Rank |
|---|---|---|---|---|---|---|
| CrossE_L + PMI_L + PR_L + Triplet_L | GAT | 3345.22 ± 349.09 | 1761.81 ± 395.61 | 4860.04 ± 1020.93 | 1441.80 ± 200.59 | 76.75 |
| CrossE_L + PMI_L + PR_L + Triplet_L | GCN | 2629.34 ± 239.77 | 1088.17 ± 174.44 | 4729.37 ± 693.83 | 1125.78 ± 135.96 | 120.0 |
| CrossE_L + PMI_L + PR_L + Triplet_L | GIN | 7427.03 ± 841.15 | 674.36 ± 225.50 | 4426.37 ± 1297.67 | 569.18 ± 220.02 | 122.25 |
| CrossE_L + PMI_L + PR_L + Triplet_L | MPNN | 4819.66 ± 737.44 | 2007.30 ± 487.83 | 7135.95 ± 1150.03 | 1795.40 ± 476.55 | 43.5 |
| CrossE_L + PMI_L + PR_L + Triplet_L | PAGNN | 10597.02± 417.71 | 191.77 ± 0.63 | 2492.01 ± 207.61 | 183.07 ± 1.37 | 152.5 |
| CrossE_L + PMI_L + PR_L + Triplet_L | SAGE | 1720.75 ± 431.22 | 1106.69 ± 128.22 | 3269.58 ± 449.92 | 1602.21 ± 313.89 | 127.0 |
| CrossE_L + PMI_L + Triplet_L | ALL | 5106.03 ± 357.40 | 1229.61 ± 373.98 | 6878.03 ± 1286.17 | 1355.95 ± 101.39 | 70.25 |
| CrossE_L + PMI_L + Triplet_L | GAT | 3063.66 ± 363.41 | 1801.90 ± 197.07 | 4662.22 ± 414.04 | 1687.68 ± 305.77 | 78.75 |
| CrossE_L + PMI_L + Triplet_L | GCN | 2771.32 ± 96.37 | 999.81 ± 338.55 | 4945.23 ± 631.15 | 1264.16 ± 252.26 | 109.25 |
| CrossE_L + PMI_L + Triplet_L | GIN | 6259.89 ± 735.10 | 605.07 ± 175.98 | 4510.60 ± 772.56 | 748.93 ± 246.15 | 124.0 |
| CrossE_L + PMI_L + Triplet_L | MPNN | 4783.94 ± 597.32 | 1748.83 ± 454.33 | 8399.55 ± 894.63 | 1621.15 ± 476.57 | 53.25 |
| CrossE_L + PMI_L + Triplet_L | PAGNN | 11735.33± 506.93 | 196.21 ± 0.60 | 2343.07 ± 212.37 | 195.05 ± 1.02 | 140.0 |
| CrossE_L + PMI_L + Triplet_L | SAGE | 1895.14 ± 211.81 | 1017.28 ± 337.88 | 3311.44 ± 629.89 | 1053.42 ± 341.84 | 143.75 |
| CrossE_L + PR_L | ALL | 11814.33± 799.57 | 696.97 ± 82.04 | 9378.13 ± 1948.35 | 877.40 ± 118.69 | 70.0 |
| CrossE_L + PR_L | GAT | 2103.60 ± 107.08 | 2017.77 ± 547.31 | 4351.44 ± 304.77 | 1716.54 ± 465.41 | 91.5 |
| CrossE_L + PR_L | GCN | 2184.38 ± 437.06 | 1249.58 ± 254.56 | 3980.02 ± 683.83 | 1300.95 ± 259.82 | 120.25 |

Continued on next page

Table 38. Results for Calinski Harabasz (↑) (continued)

| Loss Type | Model | Cora ↓ Citeseer | Cora ↓ Bitcoin | Citeseer ↓ Cora | Citeseer ↓ Bitcoin | Average Rank |
|---|---|---|---|---|---|---|
| CrossE_L + PR_L | GIN | 5727.70 ± 1320.00 | 360.80 ± 56.52 | 8963.74 ± 4242.49 | 393.80 ± 45.59 | 113.5 |
| CrossE_L + PR_L | MPNN | 7991.67 ± 2938.89 | 2157.59 ± 618.77 | 12646.01± 2990.87 | 1991.88 ± 437.89 | 14.25 |
| CrossE_L + PR_L | PAGNN | 6490.38 ± 1708.57 | 169.93 ± 0.36 | 8201.63 ± 578.51 | 188.56 ± 0.25 | 126.5 |
| CrossE_L + PR_L | SAGE | 1084.71 ± 155.07 | 1088.58 ± 409.94 | 1921.65 ± 294.64 | 1132.90 ± 448.15 | 150.75 |
| CrossE_L + PR_L + Triplet_L | ALL | 9002.74 ± 2642.73 | 557.74 ± 302.34 | 2558.86 ± 637.38 | 391.42 ± 147.74 | 136.25 |
| CrossE_L + PR_L + Triplet_L | GAT | 2496.03 ± 147.56 | 1789.80 ± 382.58 | 4681.84 ± 458.09 | 1651.91 ± 240.01 | 89.75 |
| CrossE_L + PR_L + Triplet_L | GCN | 2700.67 ± 305.40 | 1563.83 ± 470.81 | 4736.82 ± 428.87 | 1600.00 ± 372.40 | 94.75 |
| CrossE_L + PR_L + Triplet_L | GIN | 6642.35 ± 1446.45 | 819.91 ± 513.69 | 5691.41 ± 778.40 | 518.00 ± 62.39 | 106.5 |
| CrossE_L + PR_L + Triplet_L | MPNN | 6627.76 ± 772.50 | 1988.58 ± 483.42 | 11845.68± 2634.10 | 2388.24 ± 438.70 | 21.0 |
| CrossE_L + PR_L + Triplet_L | PAGNN | 9744.09 ± 770.31 | 188.25 ± 0.69 | 5978.58 ± 1634.77 | 183.10 ± 0.81 | 125.5 |
| CrossE_L + PR_L + Triplet_L | SAGE | 2687.98 ± 426.93 | 741.05 ± 179.20 | 4006.55 ± 410.18 | 593.71 ± 215.08 | 151.75 |
| CrossE_L + Triplet_L | ALL | 4726.45 ± 955.30 | 673.72 ± 223.56 | 7165.27 ± 1529.53 | 896.41 ± 366.30 | 104.5 |
| CrossE_L + Triplet_L | GAT | 3064.45 ± 147.38 | 1668.36 ± 448.83 | 4904.50 ± 242.09 | 1621.85 ± 286.67 | 79.25 |
| CrossE_L + Triplet_L | GCN | 2875.50 ± 386.17 | 1317.13 ± 271.17 | 5002.28 ± 501.31 | 1187.27 ± 459.66 | 101.0 |
| CrossE_L + Triplet_L | GIN | 4921.27 ± 507.62 | 703.30 ± 221.54 | 7473.91 ± 599.43 | 881.36 ± 161.38 | 99.5 |
| CrossE_L + Triplet_L | MPNN | 4610.08 ± 364.48 | 1809.79 ± 170.91 | 10320.53± 1675.80 | 1650.20 ± 414.61 | 47.25 |

Table 38. Results for Calinski Harabasz (↑) (continued)

| Loss Type | Model | Cora ↓ Citeseer | Cora ↓ Bitcoin | Citeseer ↓ Cora | Citeseer ↓ Bitcoin | Average Rank |
|---|---|---|---|---|---|---|
| CrossE_L + Triplet_L | PAGNN | 7466.48 ± 1880.38 | 193.92 ± 2.68 | 6404.25 ± 986.22 | 174.28 ± 21.59 | 128.5 |
| CrossE_L + Triplet_L | SAGE | 3039.48 ± 756.81 | 573.55 ± 179.16 | 4366.80 ± 434.18 | 499.40 ± 82.77 | 150.5 |
| PMI_L | ALL | 8317.55 ± 498.09 | 913.33 ± 236.50 | 2985.35 ± 684.28 | 976.78 ± 186.71 | 110.5 |
| PMI_L | GAT | 2816.03 ± 375.83 | 1640.73 ± 226.20 | 4633.55 ± 433.94 | 1635.69 ± 354.54 | 91.75 |
| PMI_L | GCN | 2895.89 ± 573.52 | 1203.39 ± 240.16 | 4544.95 ± 551.46 | 1165.23 ± 430.58 | 112.0 |
| PMI_L | GIN | 7765.00 ± 1398.30 | 740.40 ± 92.23 | 3427.24 ± 600.01 | 725.93 ± 193.93 | 121.5 |
| PMI_L | MPNN | 4936.01 ± 952.40 | 1900.87 ± 323.88 | 8619.00 ± 1000.02 | 2011.77 ± 611.40 | 37.5 |
| PMI_L | PAGNN | 11283.66± 523.01 | 195.99 ± 0.64 | 1998.67 ± 105.64 | 186.93 ± 1.05 | 148.25 |
| PMI_L | SAGE | 1151.11 ± 249.36 | 1345.77 ± 263.66 | 1545.92 ± 389.30 | 1047.76 ± 296.52 | 147.75 |
| PMI_L + PR_L | ALL | 9565.78 ± 587.90 | 489.71 ± 77.95 | 1895.67 ± 701.35 | 498.69 ± 53.14 | 139.25 |
| PMI_L + PR_L | GAT | 2917.06 ± 434.02 | 1830.15 ± 160.60 | 4783.02 ± 1299.40 | 1527.10 ± 330.23 | 82.0 |
| PMI_L + PR_L | GCN | 3017.46 ± 334.57 | 1031.88 ± 255.81 | 4714.88 ± 103.40 | 1054.72 ± 230.76 | 115.0 |
| PMI_L + PR_L | GIN | 7766.81 ± 713.14 | 683.87 ± 204.06 | 3429.65 ± 1336.61 | 669.76 ± 238.76 | 125.0 |
| PMI_L + PR_L | MPNN | 4778.86 ± 576.50 | 1649.26 ± 384.33 | 6116.65 ± 1384.03 | 1340.41 ± 226.25 | 71.5 |
| PMI_L + PR_L | PAGNN | 11034.97± 787.27 | 192.72 ± 0.61 | 2257.09 ± 179.22 | 184.18 ± 0.88 | 151.5 |
| PMI_L + PR_L | SAGE | 1483.94 ± 238.75 | 1157.95 ± 489.65 | 1717.74 ± 210.79 | 1294.66 ± 465.96 | 139.75 |

Table 38. Results for Calinski Harabasz (↑) (continued)

| Loss Type | | | Model | Cora ↓ Citeseer | Cora ↓ Bitcoin | Citeseer ↓ Cora | Citeseer ↓ Bitcoin | Average Rank |
|---|---|---|---|---|---|---|---|---|
| PMI_L | + PR_L | + Triplet_L | ALL | 7452.46 ± 613.30 | 439.67 ± 896.26 | 3844.23 ± 346.61 | 482.28 ± | 135.25 |
| PMI_L | + PR_L | + Triplet_L | GAT | 2942.74 ± 316.66 | 1482.17 ± 290.21 | 5055.70 ± 634.64 | 1611.35 ± 195.52 | 84.0 |
| PMI_L | + PR_L | + Triplet_L | GCN | 3023.70 ± 347.61 | 1139.76 ± 177.74 | 4858.97 ± 311.31 | 1367.97 ± 240.13 | 96.75 |
| PMI_L | + PR_L | + Triplet_L | GIN | 7987.00 ± 755.70 | 1305.99 ± 466.99 | 4189.79 ± 971.01 | 494.94 ± 171.57 | 106.5 |
| PMI_L | + PR_L | + Triplet_L | MPNN | 4846.55 ± 372.15 | 1650.99 ± 170.13 | 6518.13 ± 1060.04 | 1261.24 ± 386.33 | 70.5 |
| PMI_L | + PR_L | + Triplet_L | PAGNN | 11169.49± 726.67 | 193.17 ± 0.49 | 2735.11 ± 341.54 | 187.56 ± 0.65 | 145.75 |
| PMI_L | + PR_L | + Triplet_L | SAGE | 2240.55 ± 504.31 | 1548.74 ± 455.49 | 3595.59 ± 737.75 | 1081.54 ± 484.09 | 125.5 |
| PMI_L + Triplet_L | | | ALL | 6288.62 ± 615.64 | 670.79 ± 184.89 | 6704.06 ± 1806.87 | 1204.08 ± 297.65 | 90.0 |
| PMI_L + Triplet_L | | | GAT | 3115.82 ± 192.70 | 1593.63 ± 353.06 | 4710.83 ± 182.67 | 1747.14 ± 257.22 | 80.75 |
| PMI_L + Triplet_L | | | GCN | 2718.53 ± 523.38 | 1092.67 ± 211.74 | 4864.71 ± 501.18 | 1213.03 ± 432.80 | 110.75 |
| PMI_L + Triplet_L | | | GIN | 6485.87 ± 272.37 | 662.47 ± 62.14 | 4396.62 ± 1110.41 | 986.44 ± 380.57 | 118.25 |
| PMI_L + Triplet_L | | | MPNN | 4259.14 ± 504.25 | 1787.84 ± 483.27 | 8375.42 ± 879.69 | 1600.51 ± 328.18 | 56.25 |
| PMI_L + Triplet_L | | | PAGNN | 11851.48± 823.14 | 195.96 ± 0.35 | 2205.97 ± 260.84 | 194.69 ± 0.99 | 141.0 |
| PMI_L + Triplet_L | | | SAGE | 1993.71 ± 168.10 | 1391.82 ± 293.36 | 3412.32 ± 328.10 | 1072.87 ± 307.85 | 132.0 |
| PR_L | | | ALL | 13704.86± 747.31 | 878.91 ± 154.27 | 10747.82± 1893.47 | 999.11 ± 176.79 | 60.5 |
| PR_L | | | GAT | 2312.31 ± 198.56 | 2165.17 ± 361.98 | 3815.55 ± 225.06 | 1895.77 ± 348.11 | 88.75 |

Continued on next page

Table 38. Results for Calinski Harabasz (↑) (continued)

| Loss Type | Model | Cora ↓ Citeseer | Cora ↓ Bitcoin | Citeseer ↓ Cora | Citeseer ↓ Bitcoin | Average Rank |
|---|---|---|---|---|---|---|
| PR_L | GCN | 2054.09 ± 168.27 | 1123.38 ± 65.67 | 4296.42 ± 1256.52 | 1286.52 ± 442.58 | 124.5 |
| PR_L | GIN | 6121.50 ± 1483.02 | 338.75 ± 27.82 | 8834.76 ± 3092.71 | 505.40 ± 129.43 | 110.75 |
| PR_L | MPNN | 6900.91 ± 1684.91 | 2053.16 ± 368.24 | 11675.30± 2947.07 | 2055.69 ± 230.16 | 20.0 |
| PR_L | PAGNN | 6556.79 ± 1703.20 | 191.02 ± 0.18 | 9997.66 ± 475.17 | 194.93 ± 0.34 | 117.25 |
| PR_L | SAGE | 1124.95 ± 168.32 | 924.80 ± 354.98 | 1775.40 ± 194.16 | 1144.54 ± 129.01 | 154.5 |
| PR_L + Triplet_L | ALL | 11491.00± 1085.43 | 528.05 ± 161.44 | 8669.00 ± 1802.91 | 702.54 ± 265.75 | 85.5 |
| PR_L + Triplet_L | GAT | 2658.81 ± 601.07 | 1848.79 ± 416.95 | 4300.68 ± 598.05 | 1963.33 ± 368.37 | 88.5 |
| PR_L + Triplet_L | GCN | 2146.97 ± 351.81 | 1380.31 ± 248.30 | 4652.43 ± 697.09 | 1198.35 ± 323.31 | 114.75 |
| PR_L + Triplet_L | GIN | 7092.61 ± 1313.03 | 476.48 ± 154.42 | 8094.38 ± 3233.32 | 750.05 ± 156.75 | 98.0 |
| PR_L + Triplet_L | MPNN | 7166.81 ± 2287.42 | 1963.38 ± 606.25 | 12943.92± 3270.82 | 2213.17 ± 757.37 | 19.0 |
| PR_L + Triplet_L | PAGNN | 8524.24 ± 1088.15 | 194.34 ± 0.46 | 9639.93 ± 669.55 | 185.65 ± 0.41 | 109.5 |
| PR_L + Triplet_L | SAGE | 1290.06 ± 400.63 | 857.95 ± 147.76 | 2467.88 ± 801.34 | 1204.72 ± 408.09 | 147.5 |
| Triplet_L | ALL | 5345.74 ± 2089.48 | 651.81 ± 179.79 | 8465.14 ± 1209.69 | 1644.25 ± 362.81 | 74.5 |
| Triplet_L | GAT | 3230.36 ± 519.85 | 1860.66 ± 506.72 | 5160.97 ± 223.38 | 1769.23 ± 375.20 | 63.5 |
| Triplet_L | GCN | 3204.63 ± 504.01 | 1127.80 ± 158.88 | 5040.11 ± 677.71 | 1156.36 ± 208.14 | 100.0 |
| Triplet_L | GIN | 5021.52 ± 456.54 | 982.95 ± 389.88 | 6426.15 ± 345.63 | 857.73 ± 252.60 | 97.75 |

Table 38. Results for Calinski Harabasz (↑) (continued)

| Loss Type | Model | Cora ↓ Citeseer | Cora ↓ Bitcoin | Citeseer ↓ Cora | Citeseer ↓ Bitcoin | Average Rank |
|-----------|-------|-----------------|----------------|-----------------|--------------------|--------------|
| Triplet_L | MPNN | 4683.69 ± 468.10 | 1446.69 ± 453.58 | 9106.47 ± 998.11 | 1973.36 ± 472.31 | 49.25 |
| Triplet_L | PAGNN | 7076.83 ± 2248.95 | 187.97 ± 5.83 | 6662.66 ± 681.67 | 165.83 ± 5.90 | 133.75 |
| Triplet_L | SAGE | 2370.44 ± 248.14 | 989.12 ± 215.98 | 3591.85 ± 616.24 | 764.80 ± 63.89 | 145.5 |

Table 39. Knn Consistency Performance (↑): This table presents models (Loss function and GNN) ranked by their average performance in terms of knn consistency. Top-ranked results are highlighted in red, second-ranked in blue, and third-ranked in green.

| Loss Type | Model | Cora ↓ Citeseer | Cora ↓ Bitcoin | Citeseer ↓ Cora | Citeseer ↓ Bitcoin | Average Rank |
|-----------|-------|-----------------|----------------|-----------------|--------------------|--------------|
| Contr_l | ALL | 59.78 ± 1.65 | 74.38 ± 0.50 | 72.90 ± 1.29 | 74.71 ± 0.29 | 156.625 |
| Contr_l | GAT | 69.31 ± 0.11 | 74.75 ± 0.24 | 83.80 ± 0.33 | 75.14 ± 0.26 | 69.875 |
| Contr_l | GCN | 68.23 ± 0.46 | 73.58 ± 0.21 | 82.53 ± 0.53 | 73.55 ± 0.14 | 101.75 |
| Contr_l | GIN | 64.92 ± 0.43 | 76.02 ± 0.44 | 79.10 ± 0.51 | 76.44 ± 0.08 | 63.375 |
| Contr_l | MPNN | 68.80 ± 0.24 | 75.07 ± 0.47 | 81.09 ± 0.14 | 75.17 ± 0.27 | 83.5 |
| Contr_l | PAGNN | 61.86 ± 0.71 | 63.78 ± 0.05 | 76.74 ± 0.45 | 63.90 ± 0.19 | 161.625 |
| Contr_l | SAGE | 64.91 ± 0.45 | 75.61 ± 0.52 | 77.91 ± 2.28 | 75.33 ± 0.23 | 88.625 |
| Contr_l + CrossE_L | ALL | 61.35 ± 1.55 | 74.56 ± 0.30 | 70.25 ± 1.97 | 72.82 ± 0.64 | 166.0 |

Table 39. Results for Knn Consistency (↑) (continued)

| Loss Type | Model | Cora ↓ Citeseer | | Cora ↓ Bitcoin | | Citeseer ↓ Cora | | Citeseer ↓ Bitcoin | | Average Rank |
|---|---|---|---|---|---|---|---|---|---|---|
| Contr_l + CrossE_L | GAT | 69.23 | ± 0.19 | 74.76 | ± 0.08 | 83.38 | ± 0.19 | 74.69 | ± 0.20 | 81.5 |
| Contr_l + CrossE_L | GCN | 68.02 | ± 0.42 | 73.33 | ± 0.17 | 82.42 | ± 0.45 | 73.43 | ± 0.19 | 112.375 |
| Contr_l + CrossE_L | GIN | 63.75 | ± 0.42 | 75.84 | ± 0.29 | 78.78 | ± 0.45 | 76.07 | ± 0.18 | 72.875 |
| Contr_l + CrossE_L | MPNN | 68.82 | ± 0.41 | 74.90 | ± 0.31 | 81.13 | ± 0.51 | 75.30 | ± 0.35 | 82.125 |
| Contr_l + CrossE_L | PAGNN | 61.09 | ± 0.72 | 63.86 | ± 0.20 | 76.37 | ± 0.79 | 63.85 | ± 0.27 | 163.5 |
| Contr_l + CrossE_L | SAGE | 64.65 | ± 0.56 | 75.43 | ± 0.34 | 78.80 | ± 1.63 | 74.98 | ± 0.36 | 102.75 |
| Contr_l + CrossE_L + PMI_L | ALL | 63.44 | ± 1.23 | 75.49 | ± 0.23 | 77.53 | ± 0.69 | 75.47 | ± 0.20 | 94.125 |
| Contr_l + CrossE_L + PMI_L | GAT | 70.02 | ± 0.41 | 75.04 | ± 0.09 | 83.83 | ± 0.24 | 75.00 | ± 0.26 | 57.875 |
| Contr_l + CrossE_L + PMI_L | GCN | 68.01 | ± 0.64 | 73.39 | ± 0.16 | 81.78 | ± 0.59 | 73.64 | ± 0.21 | 114.875 |
| Contr_l + CrossE_L + PMI_L | GIN | 63.45 | ± 0.60 | 76.07 | ± 0.14 | 75.24 | ± 0.56 | 75.76 | ± 0.26 | 82.875 |
| Contr_l + CrossE_L + PMI_L | MPNN | 69.50 | ± 0.27 | 75.45 | ± 0.22 | 81.70 | ± 0.21 | 75.49 | ± 0.25 | 60.125 |
| Contr_l + CrossE_L + PMI_L | PAGNN | 59.39 | ± 0.67 | 63.76 | ± 0.14 | 74.95 | ± 0.63 | 63.74 | ± 0.21 | 180.625 |
| Contr_l + CrossE_L + PMI_L | SAGE | 54.96 | ± 1.30 | **78.15** | ± **0.28** | 58.44 | ± 3.16 | 77.60 | ± 1.09 | 103.5 |
| Contr_l + CrossE_L + PMI_L + PR_L | ALL | 60.63 | ± 3.10 | 75.57 | ± 0.26 | 77.70 | ± 0.39 | 75.45 | ± 0.33 | 102.375 |
| Contr_l + CrossE_L + PMI_L + PR_L | GAT | 69.90 | ± 0.46 | 75.05 | ± 0.34 | 83.61 | ± 0.57 | 75.08 | ± 0.17 | 59.25 |
| Contr_l + CrossE_L + PMI_L + PR_L | GCN | 68.23 | ± 0.39 | 73.51 | ± 0.15 | 81.93 | ± 0.26 | 73.47 | ± 0.13 | 108.625 |

Table 39. Results for Knn Consistency (↑) (continued)

| Loss Type | Model | Cora ↓ Citeseer | | Cora ↓ Bitcoin | | Citeseer ↓ Cora | | Citeseer ↓ Bitcoin | | Average Rank |
|---|---|---|---|---|---|---|---|---|---|---|
| Contr_l + CrossE_L + PMI_L + PR_L | GIN | 62.61 | ± | 75.74 | ± | 73.61 | ± | 75.61 | ± | 104.875 |
| | | 0.63 | | 0.40 | | 1.07 | | 0.48 | | |
| Contr_l + CrossE_L + PMI_L + PR_L | MPNN | 69.57 | ± | 75.54 | ± | 80.73 | ± | 75.25 | ± | 67.375 |
| | | 0.35 | | 0.17 | | 0.51 | | 0.13 | | |
| Contr_l + CrossE_L + PMI_L + PR_L | PAGNN | 59.83 | ± | 63.71 | ± | 74.53 | ± | 63.70 | ± | 184.5 |
| | | 0.73 | | 0.08 | | 0.69 | | 0.11 | | |
| Contr_l + CrossE_L + PMI_L + PR_L | SAGE | 54.11 | ± | 78.14 | ± | 57.55 | ± | 77.76 | ± | 104.125 |
| | | 0.48 | | 0.67 | | 2.68 | | 0.25 | | |
| Contr_l + CrossE_L + PMI_L + PR_L + Triplet_L | ALL | 60.36 | ± | 75.11 | ± | 75.83 | ± | 75.26 | ± | 121.875 |
| | | 1.22 | | 0.29 | | 0.90 | | 0.16 | | |
| Contr_l + CrossE_L + PMI_L + PR_L + Triplet_L | GAT | 69.92 | ± | 75.03 | ± | 83.62 | ± | 74.66 | ± | 65.625 |
| | | 0.26 | | 0.08 | | 0.11 | | 0.17 | | |
| Contr_l + CrossE_L + PMI_L + PR_L + Triplet_L | GCN | 68.20 | ± | 73.51 | ± | 82.18 | ± | 73.41 | ± | 107.875 |
| | | 0.08 | | 0.28 | | 0.46 | | 0.32 | | |
| Contr_l + CrossE_L + PMI_L + PR_L + Triplet_L | GIN | 62.96 | ± | 75.86 | ± | 75.27 | ± | 75.84 | ± | 88.75 |
| | | 1.03 | | 0.53 | | 0.48 | | 0.15 | | |
| Contr_l + CrossE_L + PMI_L + PR_L + Triplet_L | MPNN | 69.69 | ± | 75.46 | ± | 80.95 | ± | 75.30 | ± | 63.625 |
| | | 0.21 | | 0.15 | | 0.66 | | 0.22 | | |
| Contr_l + CrossE_L + PMI_L + PR_L + Triplet_L | PAGNN | 59.56 | ± | 63.67 | ± | 74.68 | ± | 63.71 | ± | 186.5 |
| | | 0.34 | | 0.28 | | 0.43 | | 0.08 | | |
| Contr_l + CrossE_L + PMI_L + PR_L + Triplet_L | SAGE | 59.75 | ± | 76.87 | ± | 71.34 | ± | 77.75 | ± | 98.5 |
| | | 1.82 | | 0.32 | | 1.33 | | 0.41 | | |
| Contr_l + CrossE_L + PMI_L + Triplet_L | ALL | 66.07 | ± | 75.45 | ± | 79.16 | ± | 75.00 | ± | 97.875 |
| | | 0.61 | | 0.10 | | 0.54 | | 0.22 | | |
| Contr_l + CrossE_L + PMI_L + Triplet_L | GAT | 69.76 | ± | 74.81 | ± | 83.61 | ± | 74.71 | ± | 71.0 |
| | | 0.39 | | 0.32 | | 0.42 | | 0.18 | | |

Table 39.  Results for Knn Consistency (↑) (continued)

| Loss Type | Model | Cora ↓ Citeseer | | Cora ↓ Bitcoin | | Citeseer ↓ Cora | | Citeseer ↓ Bitcoin | | Average Rank |
|---|---|---|---|---|---|---|---|---|---|---|
| Contr_l + CrossE_L + PMI_L + Triplet_L | GCN | 68.02 0.37 | ± | 73.46 0.31 | ± | 81.49 0.52 | ± | 73.29 0.25 | ± | 119.0 |
| Contr_l + CrossE_L + PMI_L + Triplet_L | GIN | 63.53 0.47 | ± | 75.90 0.47 | ± | 75.68 1.00 | ± | 75.86 0.34 | ± | 81.375 |
| Contr_l + CrossE_L + PMI_L + Triplet_L | MPNN | 69.46 0.23 | ± | 75.08 0.32 | ± | 81.76 0.29 | ± | 75.26 0.38 | ± | 71.625 |
| Contr_l + CrossE_L + PMI_L + Triplet_L | PAGNN | 59.41 0.89 | ± | 63.75 0.17 | ± | 75.01 0.44 | ± | 63.57 0.27 | ± | 183.5 |
| Contr_l + CrossE_L + PMI_L + Triplet_L | SAGE | 61.07 0.80 | ± | 77.15 0.62 | ± | 69.74 1.10 | ± | 77.31 0.71 | ± | 96.25 |
| Contr_l + CrossE_L + PR_L | ALL | 61.71 1.15 | ± | 75.70 0.18 | ± | 71.07 1.81 | ± | 75.62 0.15 | ± | 114.375 |
| Contr_l + CrossE_L + PR_L | GAT | 69.17 0.93 | ± | 74.45 0.33 | ± | 83.64 0.47 | ± | 74.27 0.47 | ± | 83.875 |
| Contr_l + CrossE_L + PR_L | GCN | 68.10 0.81 | ± | 72.87 0.37 | ± | 82.21 0.69 | ± | 73.20 0.16 | ± | 116.5 |
| Contr_l + CrossE_L + PR_L | GIN | 62.88 0.93 | ± | 76.16 0.52 | ± | 74.28 0.59 | ± | 75.93 0.36 | ± | 89.0 |
| Contr_l + CrossE_L + PR_L | MPNN | 68.58 0.44 | ± | 74.87 0.29 | ± | 81.17 0.44 | ± | 75.17 0.34 | ± | 87.25 |
| Contr_l + CrossE_L + PR_L | PAGNN | 59.94 0.77 | ± | 63.62 0.13 | ± | 71.87 0.15 | ± | 63.80 0.07 | ± | 189.875 |
| Contr_l + CrossE_L + PR_L | SAGE | 61.22 2.79 | ± | 75.59 0.17 | ± | 68.14 5.19 | ± | 76.79 0.49 | ± | 108.875 |
| Contr_l + CrossE_L + PR_L + Triplet_L | ALL | 58.54 1.23 | ± | 75.00 0.44 | ± | 72.56 0.77 | ± | 74.54 0.44 | ± | 156.75 |
| Contr_l + CrossE_L + PR_L + Triplet_L | GAT | 69.09 0.27 | ± | 74.33 0.13 | ± | 83.31 0.30 | ± | 74.88 0.20 | ± | 84.375 |
| Contr_l + CrossE_L + PR_L + Triplet_L | GCN | 68.21 0.33 | ± | 73.34 0.25 | ± | 81.71 0.37 | ± | 73.28 0.18 | ± | 118.75 |
| Contr_l + CrossE_L + PR_L + Triplet_L | GIN | 63.68 0.52 | ± | 75.71 0.14 | ± | 77.31 0.45 | ± | 75.78 0.28 | ± | 82.625 |

Table 39. Results for Knn Consistency (↑) (continued)

| Loss Type | Model | Cora ↓ Citeseer | | Cora ↓ Bitcoin | | Citeseer ↓ Cora | | Citeseer ↓ Bitcoin | | Average Rank |
|---|---|---|---|---|---|---|---|---|---|---|
| Contr_l + CrossE_L + PR_L + Triplet_L | MPNN | 68.84 0.48 | ± | 74.75 0.14 | ± | 80.44 0.29 | ± | 75.38 0.31 | ± | 85.25 |
| Contr_l + CrossE_L + PR_L + Triplet_L | PAGNN | 60.15 0.70 | ± | 63.68 0.11 | ± | 74.63 1.01 | ± | 63.58 0.13 | ± | 184.5 |
| Contr_l + CrossE_L + PR_L + Triplet_L | SAGE | 64.68 0.34 | ± | 75.32 0.34 | ± | 74.58 0.87 | ± | 75.81 0.46 | ± | 99.625 |
| Contr_l + CrossE_L + Triplet_L | ALL | 63.53 0.98 | ± | 74.31 0.27 | ± | 74.90 1.12 | ± | 73.96 0.48 | ± | 137.625 |
| Contr_l + CrossE_L + Triplet_L | GAT | 69.58 0.36 | ± | 74.82 0.34 | ± | 83.61 0.31 | ± | 74.86 0.24 | ± | 72.25 |
| Contr_l + CrossE_L + Triplet_L | GCN | 68.49 0.33 | ± | 73.48 0.32 | ± | 82.53 0.46 | ± | 73.61 0.22 | ± | 100.625 |
| Contr_l + CrossE_L + Triplet_L | GIN | 65.87 0.70 | ± | 76.06 0.41 | ± | 79.97 0.72 | ± | 76.15 0.31 | ± | 60.25 |
| Contr_l + CrossE_L + Triplet_L | MPNN | 69.00 0.19 | ± | 75.21 0.15 | ± | 81.32 0.34 | ± | 75.07 0.31 | ± | 81.5 |
| Contr_l + CrossE_L + Triplet_L | PAGNN | 62.22 0.65 | ± | 63.76 0.20 | ± | 78.16 0.37 | ± | 63.87 0.15 | ± | 158.875 |
| Contr_l + CrossE_L + Triplet_L | SAGE | 66.42 0.83 | ± | 75.57 0.37 | ± | 79.77 0.53 | ± | 75.94 0.30 | ± | 73.375 |
| Contr_l + PMI_L | ALL | 63.13 1.71 | ± | 75.74 0.15 | ± | 77.77 0.63 | ± | 75.61 0.42 | ± | 88.0 |
| Contr_l + PMI_L | GAT | `70.03 0.13` | ± | 74.75 0.31 | ± | 83.66 0.23 | ± | 74.86 0.23 | ± | 67.125 |
| Contr_l + PMI_L | GCN | 68.20 0.33 | ± | 73.45 0.16 | ± | 81.86 0.65 | ± | 73.66 0.08 | ± | 109.125 |
| Contr_l + PMI_L | GIN | 63.46 1.23 | ± | 75.93 0.26 | ± | 75.08 0.73 | ± | 75.97 0.27 | ± | 81.125 |
| Contr_l + PMI_L | MPNN | 69.46 0.29 | ± | 75.42 0.09 | ± | 81.87 0.16 | ± | 75.28 0.26 | ± | 64.5 |
| Contr_l + PMI_L | PAGNN | 59.26 0.47 | ± | 63.69 0.09 | ± | 74.01 0.52 | ± | 63.89 0.22 | ± | 186.25 |

Table 39. Results for Knn Consistency (↑) (continued)

| Loss Type | Model | Cora ↓ Citeseer | | Cora ↓ Bitcoin | | Citeseer ↓ Cora | | Citeseer ↓ Bitcoin | | Average Rank |
|---|---|---|---|---|---|---|---|---|---|---|
| Contr_l + PMI_L | SAGE | 53.79 | ± | 78.35 | ± | 63.11 | ± | 77.69 | ± | 104.25 |
| | | 1.23 | | 0.41 | | 4.00 | | 0.28 | | |
| Contr_l + PMI_L + PR_L | ALL | 61.14 | ± | 75.56 | ± | 77.13 | ± | 75.35 | ± | 104.625 |
| | | 1.52 | | 0.08 | | 0.32 | | 0.26 | | |
| Contr_l + PMI_L + PR_L | GAT | 69.60 | ± | 75.03 | ± | 82.70 | ± | 74.66 | ± | 74.125 |
| | | 0.33 | | 0.10 | | 1.04 | | 0.48 | | |
| Contr_l + PMI_L + PR_L | GCN | 68.39 | ± | 73.35 | ± | 82.28 | ± | 73.59 | ± | 106.375 |
| | | 0.39 | | 0.18 | | 0.31 | | 0.26 | | |
| Contr_l + PMI_L + PR_L | GIN | 63.47 | ± | 75.96 | ± | 74.53 | ± | 75.80 | ± | 88.75 |
| | | 0.67 | | 0.28 | | 0.88 | | 0.22 | | |
| Contr_l + PMI_L + PR_L | MPNN | 69.41 | ± | 75.35 | ± | 79.90 | ± | 75.22 | ± | 77.375 |
| | | 0.24 | | 0.23 | | 0.83 | | 0.32 | | |
| Contr_l + PMI_L + PR_L | PAGNN | 58.76 | ± | 63.84 | ± | 74.22 | ± | 63.68 | ± | 187.25 |
| | | 0.90 | | 0.22 | | 0.39 | | 0.20 | | |
| Contr_l + PMI_L + PR_L | SAGE | 53.87 | ± | 78.13 | ± | 63.36 | ± | 76.83 | ± | 106.0 |
| | | 0.71 | | 0.36 | | 1.56 | | 0.92 | | |
| Contr_l + PMI_L + PR_L + Triplet_L | ALL | 60.32 | ± | 74.57 | ± | 74.64 | ± | 74.58 | ± | 147.875 |
| | | 1.11 | | 0.35 | | 0.84 | | 0.57 | | |
| Contr_l + PMI_L + PR_L + Triplet_L | GAT | 69.37 | ± | 74.78 | ± | 82.60 | ± | 75.18 | ± | 73.0 |
| | | 0.95 | | 0.35 | | 0.39 | | 0.23 | | |
| Contr_l + PMI_L + PR_L + Triplet_L | GCN | 68.01 | ± | 73.23 | ± | 82.10 | ± | 73.41 | ± | 116.75 |
| | | 0.34 | | 0.26 | | 0.43 | | 0.16 | | |
| Contr_l + PMI_L + PR_L + Triplet_L | GIN | 63.71 | ± | 75.89 | ± | 76.45 | ± | 75.80 | ± | 79.875 |
| | | 0.73 | | 0.37 | | 0.39 | | 0.29 | | |
| Contr_l + PMI_L + PR_L + Triplet_L | MPNN | 69.65 | ± | 75.32 | ± | 80.12 | ± | 75.31 | ± | 70.375 |
| | | 0.24 | | 0.32 | | 0.20 | | 0.27 | | |
| Contr_l + PMI_L + PR_L + Triplet_L | PAGNN | 60.11 | ± | 63.67 | ± | 74.94 | ± | 63.77 | ± | 178.875 |
| | | 0.89 | | 0.14 | | 1.09 | | 0.15 | | |
| Contr_l + PMI_L + PR_L + Triplet_L | SAGE | 63.44 | ± | 76.84 | ± | 74.17 | ± | 75.85 | ± | 87.0 |
| | | 0.61 | | 0.48 | | 1.04 | | 0.29 | | |
| Contr_l + PR_L | ALL | 62.02 | ± | 75.59 | ± | 72.17 | ± | 75.76 | ± | 112.25 |
| | | 1.46 | | 0.16 | | 1.86 | | 0.22 | | |

Table 39. Results for Knn Consistency (↑) (continued)

| Loss Type | Model | Cora ↓ Citeseer | | Cora ↓ Bitcoin | | Citeseer ↓ Cora | | Citeseer ↓ Bitcoin | | Average Rank |
|---|---|---|---|---|---|---|---|---|---|---|
| Contr_l + PR_L | GAT | 69.23 | ± 0.30 | 74.44 | ± 0.19 | 83.64 | ± 0.37 | 74.31 | ± 0.23 | 83.5 |
| Contr_l + PR_L | GCN | 68.18 | ± 0.50 | 73.24 | ± 0.36 | 82.02 | ± 0.58 | 73.07 | ± 0.18 | 118.25 |
| Contr_l + PR_L | GIN | 62.74 | ± 0.77 | 76.01 | ± 0.41 | 74.48 | ± 1.37 | 75.99 | ± 0.41 | 90.0 |
| Contr_l + PR_L | MPNN | 68.85 | ± 0.11 | 74.96 | ± 0.19 | 80.97 | ± 0.35 | 75.03 | ± 0.21 | 89.25 |
| Contr_l + PR_L | PAGNN | 59.57 | ± 0.56 | 63.70 | ± 0.12 | 71.56 | ± 0.86 | 63.74 | ± 0.15 | 193.25 |
| Contr_l + PR_L | SAGE | 60.52 | ± 2.94 | 75.69 | ± 0.28 | 70.66 | ± 3.96 | 78.15 | ± 0.77 | 104.375 |
| Contr_l + PR_L + Triplet_L | ALL | 59.24 | ± 1.25 | 75.26 | ± 0.61 | 74.62 | ± 1.26 | 75.11 | ± 0.25 | 136.5 |
| Contr_l + PR_L + Triplet_L | GAT | 69.10 | ± 0.36 | 74.21 | ± 0.30 | 83.52 | ± 0.60 | 74.55 | ± 0.12 | 86.0 |
| Contr_l + PR_L + Triplet_L | GCN | 68.21 | ± 0.24 | 73.40 | ± 0.34 | 82.36 | ± 0.20 | 73.24 | ± 0.13 | 110.75 |
| Contr_l + PR_L + Triplet_L | GIN | 64.27 | ± 0.72 | 76.13 | ± 0.30 | 77.15 | ± 0.81 | 76.15 | ± 0.42 | 68.75 |
| Contr_l + PR_L + Triplet_L | MPNN | 69.02 | ± 0.32 | 75.03 | ± 0.31 | 80.43 | ± 0.34 | 75.35 | ± 0.33 | 79.875 |
| Contr_l + PR_L + Triplet_L | PAGNN | 60.35 | ± 0.43 | 63.80 | ± 0.09 | 74.94 | ± 2.26 | 63.78 | ± 0.17 | 173.0 |
| Contr_l + PR_L + Triplet_L | SAGE | 64.52 | ± 0.36 | 75.71 | ± 0.55 | 76.18 | ± 0.36 | 75.96 | ± 0.49 | 81.0 |
| Contr_l + Triplet_L | ALL | 63.32 | ± 0.90 | 73.73 | ± 0.18 | 75.67 | ± 1.49 | 74.46 | ± 0.56 | 137.0 |
| Contr_l + Triplet_L | GAT | 69.54 | ± 0.45 | 74.59 | ± 0.42 | 83.88 | ± 0.15 | 74.89 | ± 0.13 | 72.625 |
| Contr_l + Triplet_L | GCN | 68.61 | ± 0.25 | 73.27 | ± 0.16 | 82.74 | ± 0.29 | 73.68 | ± 0.10 | 102.5 |

Table 39. Results for Knn Consistency (↑) (continued)

| Loss Type | Model | Cora ↓ Citeseer | | Cora ↓ Bitcoin | | Citeseer ↓ Cora | | Citeseer ↓ Bitcoin | | Average Rank |
|---|---|---|---|---|---|---|---|---|---|---|
| Contr_l + Triplet_L | GIN | 65.70 | ± | 75.87 | ± | 79.73 | ± | 76.15 | ± | 66.375 |
| | | 0.52 | | 0.63 | | 0.74 | | 0.18 | | |
| Contr_l + Triplet_L | MPNN | 69.40 | ± | 74.95 | ± | 81.66 | ± | 75.04 | ± | 82.75 |
| | | 0.32 | | 0.24 | | 0.21 | | 0.27 | | |
| Contr_l + Triplet_L | PAGNN | 62.28 | ± | 63.94 | ± | 77.98 | ± | 63.87 | ± | 156.125 |
| | | 0.82 | | 0.16 | | 0.85 | | 0.20 | | |
| Contr_l + Triplet_L | SAGE | 66.47 | ± | 75.55 | ± | 79.42 | ± | 75.66 | ± | 79.875 |
| | | 0.57 | | 0.35 | | 1.42 | | 0.41 | | |
| CrossE_L | ALL | 64.70 | ± | 73.79 | ± | 73.65 | ± | 73.84 | ± | 143.0 |
| | | 2.74 | | 0.09 | | 1.46 | | 0.23 | | |
| CrossE_L | GAT | 67.62 | ± | 73.39 | ± | 81.71 | ± | 73.80 | ± | 117.125 |
| | | 1.35 | | 0.26 | | 1.56 | | 0.29 | | |
| CrossE_L | GCN | 54.56 | ± | 73.81 | ± | 51.92 | ± | 73.11 | ± | 182.25 |
| | | 2.37 | | 0.31 | | 3.15 | | 0.43 | | |
| CrossE_L | GIN | 47.02 | ± | 74.16 | ± | 47.56 | ± | 72.07 | ± | 186.0 |
| | | 1.48 | | 0.44 | | 1.96 | | 0.38 | | |
| CrossE_L | MPNN | 61.70 | ± | 75.45 | ± | 74.79 | ± | 74.92 | ± | 124.875 |
| | | 1.64 | | 0.33 | | 0.73 | | 0.33 | | |
| CrossE_L | PAGNN | 53.36 | ± | 63.75 | ± | 73.12 | ± | 63.75 | ± | 193.625 |
| | | 0.53 | | 0.14 | | 0.86 | | 0.24 | | |
| CrossE_L | SAGE | 47.15 | ± | 75.68 | ± | 49.81 | ± | 75.06 | ± | 146.625 |
| | | 1.88 | | 0.66 | | 3.71 | | 0.41 | | |
| CrossE_L + PMI_L | ALL | 64.80 | ± | 75.39 | ± | 77.87 | ± | 75.36 | ± | 93.875 |
| | | 0.16 | | 0.14 | | 0.54 | | 0.29 | | |
| CrossE_L + PMI_L | GAT | 69.91 | ± | 74.89 | ± | 83.97 | ± | 74.81 | ± | 64.0 |
| | | 0.27 | | 0.27 | | 0.15 | | 0.18 | | |
| CrossE_L + PMI_L | GCN | 68.06 | ± | 73.43 | ± | 81.92 | ± | 73.53 | ± | 112.125 |
| | | 0.18 | | 0.16 | | 0.33 | | 0.32 | | |
| CrossE_L + PMI_L | GIN | 63.16 | ± | 75.98 | ± | 75.11 | ± | 75.72 | ± | 88.25 |
| | | 0.57 | | 0.69 | | 0.46 | | 0.26 | | |
| CrossE_L + PMI_L | MPNN | 69.52 | ± | 75.31 | ± | 81.89 | ± | 75.18 | ± | 68.125 |
| | | 0.34 | | 0.27 | | 0.39 | | 0.11 | | |

Table 39. Results for Knn Consistency (↑) (continued)

| Loss Type | Model | Cora ↓ Citeseer | | Cora ↓ Bitcoin | | Citeseer ↓ Cora | | Citeseer ↓ Bitcoin | | Average Rank |
|---|---|---|---|---|---|---|---|---|---|---|
| CrossE_L + PMI_L | PAGNN | 59.58 1.08 | ± | 63.88 0.05 | ± | 74.79 0.09 | ± | 63.90 0.17 | ± | 173.625 |
| CrossE_L + PMI_L | SAGE | 53.25 1.02 | ± | 77.80 0.57 | ± | 57.17 1.32 | ± | 78.39 0.48 | ± | 105.25 |
| CrossE_L + PMI_L + PR_L | ALL | 63.13 0.66 | ± | 75.35 0.21 | ± | 76.92 0.86 | ± | 75.28 0.14 | ± | 106.125 |
| CrossE_L + PMI_L + PR_L | GAT | 69.72 0.38 | ± | 74.71 0.23 | ± | 82.58 1.28 | ± | 74.48 0.26 | ± | 79.375 |
| CrossE_L + PMI_L + PR_L | GCN | 68.27 0.52 | ± | 73.41 0.12 | ± | 81.98 0.30 | ± | 73.38 0.21 | ± | 111.125 |
| CrossE_L + PMI_L + PR_L | GIN | 62.63 0.80 | ± | 75.68 0.38 | ± | 73.52 0.70 | ± | 75.73 0.51 | ± | 106.5 |
| CrossE_L + PMI_L + PR_L | MPNN | 69.89 0.32 | ± | 75.43 0.13 | ± | 80.81 1.23 | ± | 75.38 0.41 | ± | 60.875 |
| CrossE_L + PMI_L + PR_L | PAGNN | 59.38 1.01 | ± | 63.76 0.16 | ± | 74.21 1.32 | ± | 63.65 0.14 | ± | 188.5 |
| CrossE_L + PMI_L + PR_L | SAGE | 54.25 0.90 | ± | 77.82 0.43 | ± | 57.37 1.39 | ± | 77.88 0.66 | ± | 104.25 |
| CrossE_L + PMI_L + PR_L + Triplet_L | ALL | 62.51 1.24 | ± | 75.28 0.25 | ± | 76.23 0.81 | ± | 74.99 0.20 | ± | 119.75 |
| CrossE_L + PMI_L + PR_L + Triplet_L | GAT | 69.89 0.26 | ± | 75.03 0.39 | ± | 83.45 0.30 | ± | 74.95 0.18 | ± | 64.5 |
| CrossE_L + PMI_L + PR_L + Triplet_L | GCN | 68.08 0.23 | ± | 73.41 0.54 | ± | 81.81 0.40 | ± | 73.30 0.22 | ± | 116.0 |
| CrossE_L + PMI_L + PR_L + Triplet_L | GIN | 63.26 1.00 | ± | 75.88 0.28 | ± | 75.36 0.60 | ± | 75.92 0.28 | ± | 84.375 |
| CrossE_L + PMI_L + PR_L + Triplet_L | MPNN | 69.70 0.13 | ± | 75.50 0.18 | ± | 80.69 0.58 | ± | 75.58 0.23 | ± | 58.625 |
| CrossE_L + PMI_L + PR_L + Triplet_L | PAGNN | 60.14 0.38 | ± | 63.67 0.16 | ± | 74.51 0.55 | ± | 63.72 0.12 | ± | 185.125 |
| CrossE_L + PMI_L + PR_L + Triplet_L | SAGE | 59.19 2.82 | ± | 77.45 0.39 | ± | 68.85 1.32 | ± | 77.48 0.62 | ± | 103.25 |

Table 39. Results for Knn Consistency (↑) (continued)

| Loss Type | Model | Cora ↓ Citeseer | | Cora ↓ Bitcoin | | Citeseer ↓ Cora | | Citeseer ↓ Bitcoin | | Average Rank |
|---|---|---|---|---|---|---|---|---|---|---|
| CrossE_L + PMI_L + Triplet_L | ALL | 66.88 | ± 0.15 | 75.61 | ± 0.23 | 80.43 | ± 0.27 | 75.42 | ± 0.17 | 77.25 |
| CrossE_L + PMI_L + Triplet_L | GAT | 69.85 | ± 0.33 | 75.02 | ± 0.14 | 83.77 | ± 0.18 | 75.11 | ± 0.13 | 59.625 |
| CrossE_L + PMI_L + Triplet_L | GCN | 68.08 | ± 0.39 | 73.40 | ± 0.19 | 81.76 | ± 0.41 | 73.51 | ± 0.28 | 115.375 |
| CrossE_L + PMI_L + Triplet_L | GIN | 63.27 | ± 0.76 | 75.84 | ± 0.54 | 76.14 | ± 0.74 | 75.65 | ± 0.37 | 88.5 |
| CrossE_L + PMI_L + Triplet_L | MPNN | 69.57 | ± 0.18 | 75.44 | ± 0.19 | 82.13 | ± 0.15 | 75.50 | ± 0.33 | 53.75 |
| CrossE_L + PMI_L + Triplet_L | PAGNN | 60.57 | ± 0.51 | 63.71 | ± 0.13 | 74.78 | ± 0.57 | 63.57 | ± 0.22 | 181.0 |
| CrossE_L + PMI_L + Triplet_L | SAGE | 62.54 | ± 0.69 | 77.21 | ± 0.68 | 71.78 | ± 0.90 | 76.92 | ± 0.72 | 90.75 |
| CrossE_L + PR_L | ALL | 62.72 | ± 0.92 | 75.51 | ± 0.28 | 75.05 | ± 1.24 | 75.34 | ± 0.15 | 107.0 |
| CrossE_L + PR_L | GAT | 69.50 | ± 0.38 | 73.80 | ± 0.20 | 83.57 | ± 0.16 | 74.12 | ± 0.21 | 85.375 |
| CrossE_L + PR_L | GCN | 67.48 | ± 0.59 | 72.96 | ± 0.22 | 81.52 | ± 0.87 | 73.14 | ± 0.14 | 127.0 |
| CrossE_L + PR_L | GIN | 62.21 | ± 0.62 | 75.70 | ± 0.45 | 73.57 | ± 1.64 | 75.88 | ± 0.30 | 104.0 |
| CrossE_L + PR_L | MPNN | 68.95 | ± 0.76 | 75.07 | ± 0.18 | 80.90 | ± 0.80 | 75.12 | ± 0.43 | 83.875 |
| CrossE_L + PR_L | PAGNN | 59.43 | ± 0.72 | 63.72 | ± 0.21 | 73.36 | ± 0.73 | 63.64 | ± 0.20 | 191.75 |
| CrossE_L + PR_L | SAGE | 53.84 | ± 0.73 | 77.68 | ± 0.49 | 60.02 | ± 1.80 | 78.23 | ± 0.94 | 104.5 |
| CrossE_L + PR_L + Triplet_L | ALL | 59.17 | ± 2.17 | 75.67 | ± 0.12 | 72.92 | ± 1.00 | 74.92 | ± 0.12 | 137.625 |
| CrossE_L + PR_L + Triplet_L | GAT | 69.87 | ± 0.25 | 74.41 | ± 0.15 | 83.49 | ± 0.11 | 74.44 | ± 0.11 | 77.75 |

Continued on next page

Table 39. Results for Knn Consistency (↑) (continued)

| Loss Type | Model | Cora ↓ Citeseer | | Cora ↓ Bitcoin | | Citeseer ↓ Cora | | Citeseer ↓ Bitcoin | | Average Rank |
|---|---|---|---|---|---|---|---|---|---|---|
| CrossE_L + PR_L + Triplet_L | GCN | 68.23 | ± 0.31 | 73.39 | ± 0.24 | 82.36 | ± 0.21 | 73.26 | ± 0.28 | 110.625 |
| CrossE_L + PR_L + Triplet_L | GIN | 63.93 | ± 1.03 | 76.08 | ± 0.35 | 76.98 | ± 0.87 | 75.97 | ± 0.37 | 71.375 |
| CrossE_L + PR_L + Triplet_L | MPNN | 69.03 | ± 0.49 | 75.14 | ± 0.31 | 80.71 | ± 0.69 | 75.25 | ± 0.29 | 79.625 |
| CrossE_L + PR_L + Triplet_L | PAGNN | 59.54 | ± 0.66 | 63.69 | ± 0.16 | 72.99 | ± 1.00 | 63.76 | ± 0.17 | 190.25 |
| CrossE_L + PR_L + Triplet_L | SAGE | 65.80 | ± 0.73 | 75.74 | ± 0.38 | 76.77 | ± 1.82 | 75.56 | ± 0.42 | 82.75 |
| CrossE_L + Triplet_L | ALL | 66.73 | ± 0.36 | 75.39 | ± 0.42 | 79.50 | ± 0.55 | 75.40 | ± 0.35 | 87.125 |
| CrossE_L + Triplet_L | GAT | 70.07 | ± 0.22 | 74.66 | ± 0.15 | 83.87 | ± 0.14 | 75.08 | ± 0.15 | 61.625 |
| CrossE_L + Triplet_L | GCN | 68.56 | ± 0.31 | 73.47 | ± 0.17 | 82.92 | ± 0.14 | 73.51 | ± 0.35 | 101.125 |
| CrossE_L + Triplet_L | GIN | 66.17 | ± 0.53 | 75.95 | ± 0.30 | 79.76 | ± 0.51 | 76.28 | ± 0.27 | 62.5 |
| CrossE_L + Triplet_L | MPNN | 69.72 | ± 0.27 | 75.12 | ± 0.21 | 82.09 | ± 0.28 | 75.74 | ± 0.12 | 53.625 |
| CrossE_L + Triplet_L | PAGNN | 62.51 | ± 0.92 | 63.82 | ± 0.19 | 78.95 | ± 0.25 | 63.89 | ± 0.08 | 155.125 |
| CrossE_L + Triplet_L | SAGE | 67.94 | ± 0.48 | 75.90 | ± 0.17 | 80.42 | ± 0.62 | 76.00 | ± 0.25 | 58.75 |
| PMI_L | ALL | 64.67 | ± 0.39 | 75.27 | ± 0.29 | 78.01 | ± 0.56 | 75.32 | ± 0.23 | 97.25 |
| PMI_L | GAT | 69.78 | ± 0.49 | 74.82 | ± 0.31 | 83.93 | ± 0.20 | 74.91 | ± 0.34 | 64.625 |
| PMI_L | GCN | 67.87 | ± 0.23 | 73.64 | ± 0.33 | 81.51 | ± 0.43 | 73.59 | ± 0.23 | 114.75 |
| PMI_L | GIN | 63.24 | ± 0.45 | 75.90 | ± 0.38 | 74.79 | ± 0.31 | 75.90 | ± 0.42 | 87.5 |

Table 39. Results for Knn Consistency (↑) (continued)

| Loss Type | Model | Cora ↓ Citeseer | | Cora ↓ Bitcoin | | Citeseer ↓ Cora | | Citeseer ↓ Bitcoin | | Average Rank |
|---|---|---|---|---|---|---|---|---|---|---|
| PMI_L | MPNN | 69.66 | ± 0.08 | 75.55 | ± 0.31 | 81.74 | ± 0.34 | 75.28 | ± | 59.25 |
| PMI_L | PAGNN | 59.67 | ± 1.01 | 63.80 | ± 0.15 | 74.64 | ± 0.80 | 63.80 | ± 0.18 | 178.125 |
| PMI_L | SAGE | 52.36 | ± 1.64 | 78.50 | ± 0.31 | 55.13 | ± 2.76 | 77.72 | ± 0.41 | 106.5 |
| PMI_L + PR_L | ALL | 63.38 | ± 0.83 | 75.69 | ± 0.51 | 76.31 | ± 0.29 | 75.27 | ± 0.38 | 98.375 |
| PMI_L + PR_L | GAT | 69.61 | ± 0.32 | 74.63 | ± 0.23 | 81.98 | ± 1.45 | 74.82 | ± 0.15 | 83.5 |
| PMI_L + PR_L | GCN | 67.99 | ± 0.56 | 73.46 | ± 0.11 | 82.19 | ± 0.58 | 73.53 | ± 0.30 | 109.625 |
| PMI_L + PR_L | GIN | 62.51 | ± 0.79 | 75.80 | ± 0.36 | 73.96 | ± 0.57 | 75.85 | ± 0.25 | 99.875 |
| PMI_L + PR_L | MPNN | 69.57 | ± 0.30 | 75.43 | ± 0.36 | 79.78 | ± 0.36 | 75.22 | ± 0.27 | 74.5 |
| PMI_L + PR_L | PAGNN | 60.09 | ± 0.57 | 63.86 | ± 0.20 | 73.83 | ± 0.35 | 63.85 | ± 0.06 | 178.25 |
| PMI_L + PR_L | SAGE | 54.11 | ± 0.21 | 77.64 | ± 0.80 | 56.99 | ± 1.46 | 78.25 | ± 0.79 | 105.375 |
| PMI_L + PR_L + Triplet_L | ALL | 60.76 | ± 1.29 | 75.47 | ± 0.35 | 76.35 | ± 0.99 | 75.16 | ± 0.31 | 114.75 |
| PMI_L + PR_L + Triplet_L | GAT | 69.92 | ± 0.10 | 74.99 | ± 0.15 | 83.40 | ± 0.39 | 75.09 | ± 0.15 | 62.125 |
| PMI_L + PR_L + Triplet_L | GCN | 67.98 | ± 0.39 | 73.43 | ± 0.21 | 81.98 | ± 0.49 | 73.53 | ± 0.06 | 113.125 |
| PMI_L + PR_L + Triplet_L | GIN | 63.88 | ± 0.56 | 76.04 | ± 0.46 | 75.99 | ± 0.59 | 75.58 | ± 0.23 | 82.625 |
| PMI_L + PR_L + Triplet_L | MPNN | 69.76 | ± 0.37 | 75.26 | ± 0.12 | 80.27 | ± 0.28 | 75.23 | ± 0.36 | 72.5 |
| PMI_L + PR_L + Triplet_L | PAGNN | 59.80 | ± 0.46 | 63.70 | ± 0.19 | 74.61 | ± 0.67 | 63.74 | ± 0.20 | 183.5 |

Continued on next page

Table 39. Results for Knn Consistency (↑) (continued)

| Loss Type | Model | Cora ↓ Citeseer | | Cora ↓ Bitcoin | | Citeseer ↓ Cora | | Citeseer ↓ Bitcoin | | Average Rank |
|---|---|---|---|---|---|---|---|---|---|---|
| PMI_L + PR_L + Triplet_L | SAGE | 62.05 ± 2.96 | | 77.78 ± 0.29 | | 72.16 ± 1.22 | | 77.37 ± 0.61 | | 89.75 |
| PMI_L + Triplet_L | ALL | 66.95 ± 0.37 | | 75.78 ± 0.40 | | 80.25 ± 0.62 | | 75.65 ± 0.12 | | 70.125 |
| PMI_L + Triplet_L | GAT | 69.96 ± 0.17 | | 74.78 ± 0.16 | ± | **83.91 ± 0.38** | ± | 74.88 ± 0.09 | | 64.5 |
| PMI_L + Triplet_L | GCN | 68.03 ± 0.33 | | 73.40 ± 0.26 | | 82.16 ± 0.43 | | 73.55 ± 0.17 | | 110.125 |
| PMI_L + Triplet_L | GIN | 63.22 ± 0.82 | | 75.71 ± 0.38 | | 76.20 ± 0.59 | | 75.78 ± 0.31 | | 89.125 |
| PMI_L + Triplet_L | MPNN | 69.38 ± 0.32 | | 75.34 ± 0.41 | | 81.79 ± 0.38 | | 75.73 ± 0.21 | | 59.375 |
| PMI_L + Triplet_L | PAGNN | 59.63 ± 1.00 | | 63.74 ± 0.25 | | 75.69 ± 0.52 | | 63.74 ± 0.24 | | 177.375 |
| PMI_L + Triplet_L | SAGE | 60.69 ± 1.20 | | 77.91 ± 0.19 | | 71.76 ± 1.84 | | 77.06 ± 0.67 | | 93.0 |
| PR_L | ALL | 63.26 ± 0.59 | | 75.72 ± 0.19 | | 74.65 ± 0.96 | | 75.44 ± 0.19 | | 99.5 |
| PR_L | GAT | 69.41 ± 0.11 | | 73.86 ± 0.26 | | 83.59 ± 0.41 | | 74.05 ± 0.40 | | 86.125 |
| PR_L | GCN | 67.81 ± 0.43 | | 72.90 ± 0.30 | | 82.05 ± 0.58 | | 73.10 ± 0.31 | | 122.5 |
| PR_L | GIN | 62.58 ± 0.70 | | 75.87 ± 0.47 | | 74.80 ± 0.88 | | 76.17 ± 0.53 | | 88.125 |
| PR_L | MPNN | 68.85 ± 0.42 | | 74.79 ± 0.26 | | 80.99 ± 0.45 | | 74.94 ± 0.17 | | 92.75 |
| PR_L | PAGNN | 59.66 ± 0.72 | | 65.88 ± 0.27 | | 72.87 ± 0.81 | | 63.72 ± 0.20 | | 185.875 |
| PR_L | SAGE | 54.26 ± 0.86 | | 77.76 ± 0.55 | | 59.50 ± 1.20 | | 78.05 ± 0.41 | | 103.75 |
| PR_L + Triplet_L | ALL | 62.67 ± 0.85 | | 75.72 ± 0.19 | | 72.89 ± 1.47 | | 75.55 ± 0.22 | | 108.375 |

Table 39.  Results for Knn Consistency (↑) (continued)

| Loss Type | Model | Cora ↓ Citeseer | | Cora ↓ Bitcoin | | Citeseer ↓ Cora | | Citeseer ↓ Bitcoin | | Average Rank |
|---|---|---|---|---|---|---|---|---|---|---|
| PR_L + Triplet_L | GAT | 69.60 | ± 0.33 | 73.87 | ± 0.36 | 83.38 | ± 0.30 | 74.11 | ± 0.10 | 84.25 |
| PR_L + Triplet_L | GCN | 67.89 | ± 0.11 | 72.93 | ± 0.16 | 82.32 | ± 0.56 | 73.32 | ± 0.14 | 117.0 |
| PR_L + Triplet_L | GIN | 62.75 | ± 0.22 | 76.33 | ± 0.29 | 73.57 | ± 0.70 | 76.13 | ± 0.44 | 89.125 |
| PR_L + Triplet_L | MPNN | 68.91 | ± 0.22 | 75.05 | ± 0.25 | 81.10 | ± 0.48 | 74.89 | ± 0.41 | 88.25 |
| PR_L + Triplet_L | PAGNN | 59.81 | ± 0.83 | 63.77 | ± 0.14 | 72.20 | ± 0.76 | 63.76 | ± 0.21 | 186.125 |
| PR_L + Triplet_L | SAGE | 55.72 | ± 1.23 | 77.41 | ± 0.66 | 65.59 | ± 7.10 | 77.78 | ± 0.28 | 103.5 |
| Triplet_L | ALL | 66.68 | ± 0.47 | 75.13 | ± 0.32 | 79.78 | ± 0.33 | 75.28 | ± 0.24 | 93.5 |
| Triplet_L | GAT | 70.38 | ± 0.24 | 74.87 | ± 0.14 | 83.84 | ± 0.25 | 75.14 | ± 0.21 | 57.0 |
| Triplet_L | GCN | 68.85 | ± 0.39 | 73.64 | ± 0.10 | 82.89 | ± 0.34 | 73.45 | ± 0.23 | 98.875 |
| Triplet_L | GIN | 66.92 | ± 0.44 | 76.02 | ± 0.28 | 79.81 | ± 0.66 | 76.38 | ± 0.42 | 58.375 |
| Triplet_L | MPNN | 69.24 | ± 0.20 | 75.22 | ± 0.24 | 82.11 | ± 0.51 | 75.38 | ± 0.35 | 63.75 |
| Triplet_L | PAGNN | 63.14 | ± 1.07 | 63.81 | ± 0.24 | 79.33 | ± 0.54 | 63.82 | ± 0.10 | 152.5 |
| Triplet_L | SAGE | 67.40 | ± 0.31 | 76.19 | ± 0.43 | 79.95 | ± 0.58 | 76.32 | ± 0.49 | 55.75 |

*1.2.5  Semantic Coherence and Ranking:*

Table 40. Coherence Performance (↑): This table presents models (Loss function and GNN) ranked by their average performance in terms of coherence. Top-ranked results are highlighted in red, second-ranked in blue, and third-ranked in green.

| Loss Type | Model | Cora ↓ Citeseer | | Cora ↓ Bitcoin | | Citeseer ↓ Cora | | Citeseer ↓ Bitcoin | | Average Rank |
|---|---|---|---|---|---|---|---|---|---|---|
| Contr_l | ALL | 23.11 6.21 | ± | 11.07 1.52 | ± | 47.65 15.33 | ± | 16.64 3.68 | ± | 119.25 |
| Contr_l | GAT | 10.74 1.79 | ± | 12.20 2.52 | ± | 15.86 0.67 | ± | 10.67 1.22 | ± | 180.75 |
| Contr_l | GCN | 42.18 8.24 | ± | 44.31 8.67 | ± | 44.49 20.06 | ± | 26.37 8.25 | ± | 84.75 |
| Contr_l | GIN | 89.74 20.92 | ± | 96.01 8.92 | ± | 90.57 13.09 | ± | 99.96 0.09 | ± | 21.0 |
| Contr_l | MPNN | 20.77 5.04 | ± | 16.47 3.17 | ± | 30.76 5.63 | ± | 12.79 1.69 | ± | 134.5 |
| Contr_l | PAGNN | 17.18 1.64 | ± | 75.83 7.06 | ± | 33.41 3.60 | ± | 65.53 8.04 | ± | 99.25 |
| Contr_l | SAGE | 17.64 2.55 | ± | 11.83 1.70 | ± | 20.27 4.95 | ± | 12.23 2.66 | ± | 164.5 |
| Contr_l + CrossE_L | ALL | 31.52 13.31 | ± | 16.19 3.19 | ± | 51.30 11.78 | ± | 16.53 3.87 | ± | 100.0 |
| Contr_l + CrossE_L | GAT | 11.07 1.31 | ± | 10.57 0.70 | ± | 15.49 0.90 | ± | 10.99 1.08 | ± | 182.75 |
| Contr_l + CrossE_L | GCN | 27.35 7.64 | ± | 30.87 13.32 | ± | 51.09 11.37 | ± | 27.92 10.79 | ± | 89.75 |
| Contr_l + CrossE_L | GIN | 75.85 25.44 | ± | 84.34 16.21 | ± | 92.25 16.47 | ± | 96.32 8.03 | ± | 34.0 |
| Contr_l + CrossE_L | MPNN | 19.15 6.25 | ± | 14.29 1.07 | ± | 30.85 4.27 | ± | 15.51 2.41 | ± | 138.75 |
| Contr_l + CrossE_L | PAGNN | 19.54 1.58 | ± | 81.87 13.26 | ± | 43.44 3.21 | ± | 80.60 6.42 | ± | 82.5 |
| Contr_l + CrossE_L | SAGE | 16.49 2.34 | ± | 13.42 1.62 | ± | 23.11 4.08 | ± | 16.44 3.50 | ± | 152.625 |
| Contr_l + CrossE_L + PMI_L | ALL | 20.23 3.83 | ± | 19.58 4.30 | ± | 34.95 7.48 | ± | 28.80 7.00 | ± | 116.5 |

Continued on next page

Table 40. Results for Coherence (↑) (continued)

| Loss Type | Model | Cora ↓ Citeseer | Cora ↓ Bitcoin | Citeseer ↓ Cora | Citeseer ↓ Bitcoin | Average Rank |
|---|---|---|---|---|---|---|
| Contr_l + CrossE_L + PMI_L | GAT | 9.24±0.43 | 9.32±0.72 | 13.90 ± 0.67 | 9.24±0.95 | 201.25 |
| Contr_l + CrossE_L + PMI_L | GCN | 82.61 ± 20.41 | 50.24 ± 9.11 | 74.16 ± 12.23 | 71.23 ± 25.23 | 50.0 |
| Contr_l + CrossE_L + PMI_L | GIN | 82.21 ± 24.23 | 99.67 ± 0.49 | 81.35 ± 10.18 | 96.78 ± 7.20 | 26.25 |
| Contr_l + CrossE_L + PMI_L | MPNN | 18.80 ± 4.21 | 13.70 ± 2.16 | 31.81 ± 3.19 | 13.52 ± 3.32 | 140.75 |
| Contr_l + CrossE_L + PMI_L | PAGNN | 22.94 ± 2.06 | 94.84 ± 1.80 | 58.69 ± 6.57 | 86.37 ± 2.94 | 61.5 |
| Contr_l + CrossE_L + PMI_L | SAGE | 12.62 ± 1.82 | 8.93±0.71 | 22.84 ± 7.67 | 8.51±0.93 | 186.75 |
| Contr_l + CrossE_L + PMI_L + PR_L | ALL | 39.01 ± 4.82 | 73.85 ± 4.00 | 50.93 ± 6.20 | 37.22 ± 8.27 | 71.5 |
| Contr_l + CrossE_L + PMI_L + PR_L | GAT | 9.34±1.24 | 9.93±0.90 | 14.02 ± 1.13 | 9.27±0.60 | 197.25 |
| Contr_l + CrossE_L + PMI_L + PR_L | GCN | 65.44 ± 11.86 | 51.54 ± 15.11 | 71.82 ± 12.53 | 73.47 ± 15.08 | 56.5 |
| Contr_l + CrossE_L + PMI_L + PR_L | GIN | 76.20 ± 14.70 | 100.00 ± 0.00 | 95.10 ± 9.61 | 99.00 ± 2.23 | 16.75 |
| Contr_l + CrossE_L + PMI_L + PR_L | MPNN | 19.82 ± 3.96 | 17.67 ± 3.44 | 30.23 ± 1.09 | 31.49 ± 10.69 | 124.625 |
| Contr_l + CrossE_L + PMI_L + PR_L | PAGNN | 21.65 ± 2.24 | 90.56 ± 2.57 | 49.35 ± 4.91 | 90.78 ± 2.34 | 68.5 |
| Contr_l + CrossE_L + PMI_L + PR_L | SAGE | 10.43 ± 0.56 | 7.99±0.62 | 18.60 ± 2.24 | 9.43±1.17 | 193.75 |
| Contr_l + CrossE_L + PMI_L + PR_L + Triplet_L | ALL | 22.34 ± 1.74 | 36.86 ± 10.03 | 49.57 ± 9.22 | 28.00 ± 8.31 | 95.5 |
| Contr_l + CrossE_L + PMI_L + PR_L + Triplet_L | GAT | 9.35±0.30 | 10.08 ± 1.07 | 16.46 ± 2.34 | 9.25±0.71 | 193.25 |

Table 40. Results for Coherence (↑) (continued)

| Loss Type | Model | Cora ↓ Citeseer | | Cora ↓ Bitcoin | | Citeseer ↓ Cora | | Citeseer ↓ Bitcoin | | Average Rank |
|---|---|---|---|---|---|---|---|---|---|---|
| Contr_l + CrossE_L + PMI_L + PR_L + Triplet_L | GCN | 57.10 | ± 23.75 | 65.55 | 12.00 | 74.78 | ± 11.21 | 51.05 | ± 21.06 | 58.25 |
| Contr_l + CrossE_L + PMI_L + PR_L + Triplet_L | GIN | 81.22 | ± 27.57 | 100.00 ± 0.00 | | 88.68 | ± 24.25 | 98.76 | ± 2.77 | 19.25 |
| Contr_l + CrossE_L + PMI_L + PR_L + Triplet_L | MPNN | 17.83 | ± 2.22 | 17.84 | 2.49 | 31.39 | ± 9.06 | 22.98 | ± 8.81 | 132.0 |
| Contr_l + CrossE_L + PMI_L + PR_L + Triplet_L | PAGNN | 21.40 | ± 4.10 | 92.23 | 2.35 | 42.46 | ± 5.02 | 85.81 | ± 2.37 | 74.75 |
| Contr_l + CrossE_L + PMI_L + PR_L + Triplet_L | SAGE | 13.10 | ± 2.55 | 8.61±0.64 | | 20.99 | ± 3.06 | 9.97±0.71 | | 183.375 |
| Contr_l + CrossE_L + PMI_L + Triplet_L | ALL | 20.63 | ± 5.32 | 12.75 | 1.01 | 41.46 | ± 6.99 | 12.41 | ± 1.29 | 131.5 |
| Contr_l + CrossE_L + PMI_L + Triplet_L | GAT | 10.06 | ± 0.94 | 8.95±0.89 | | 14.13 | ± 0.10 | 10.06 | ± 0.85 | 195.25 |
| Contr_l + CrossE_L + PMI_L + Triplet_L | GCN | 57.75 | ± 27.47 | 53.98 | 16.55 | 57.37 | ± 7.90 | 44.56 | ± 8.68 | 66.875 |
| Contr_l + CrossE_L + PMI_L + Triplet_L | GIN | 80.50 | ± 26.18 | 100.00 ± 0.00 | | 89.78 | ± 8.22 | 97.17 | ± 6.32 | 22.25 |
| Contr_l + CrossE_L + PMI_L + Triplet_L | MPNN | 22.02 | ± 2.33 | 16.93 | 2.64 | 29.39 | ± 2.98 | 13.75 | ± 2.86 | 132.75 |
| Contr_l + CrossE_L + PMI_L + Triplet_L | PAGNN | 24.36 | ± 2.48 | 95.75 | 1.23 | 48.20 | ± 2.79 | 85.70 | ± 4.77 | 63.5 |
| Contr_l + CrossE_L + PMI_L + Triplet_L | SAGE | 14.52 | ± 2.67 | 10.29 | 1.38 | 20.58 | ± 2.79 | 8.99±0.56 | | 180.75 |
| Contr_l + CrossE_L + PR_L | ALL | 18.85 | ± 3.58 | 34.74 | 4.18 | 25.27 | ± 6.31 | 29.56 | ± 3.16 | 126.0 |
| Contr_l + CrossE_L + PR_L | GAT | 18.40 | ± 5.75 | 13.07 | 1.22 | 26.60 | ± 5.56 | 22.72 | ± 6.26 | 145.75 |

Continued on next page

Table 40. Results for Coherence (↑) (continued)

| Loss Type | Model | Cora ↓ Citeseer | | Cora ↓ Bitcoin | | Citeseer ↓ Cora | | Citeseer ↓ Bitcoin | | Average Rank |
|---|---|---|---|---|---|---|---|---|---|---|
| Contr_l + CrossE_L + PR_L | GCN | 57.59 18.68 | ± | 44.89 13.79 | ± | 73.51 16.12 | ± | 35.41 7.77 | ± | 69.5 |
| Contr_l + CrossE_L + PR_L | GIN | **100.00 0.00** | ± | 98.95 1.45 | ± | 97.64 3.89 | ± | **100.00 0.00** | ± | 10.125 |
| Contr_l + CrossE_L + PR_L | MPNN | 24.53 7.37 | ± | 48.38 5.15 | ± | 31.10 4.77 | ± | 44.76 4.67 | ± | 96.25 |
| Contr_l + CrossE_L + PR_L | PAGNN | 25.84 6.45 | ± | 93.54 2.56 | ± | 37.84 5.81 | ± | 99.26 0.41 | ± | 62.0 |
| Contr_l + CrossE_L + PR_L | SAGE | 24.33 7.53 | ± | 10.42 1.02 | ± | 30.14 6.15 | ± | 11.13 0.49 | ± | 144.875 |
| Contr_l + CrossE_L + PR_L + Triplet_L | ALL | 20.59 4.09 | ± | 18.32 2.92 | ± | 37.87 4.73 | ± | 16.76 2.40 | ± | 119.5 |
| Contr_l + CrossE_L + PR_L + Triplet_L | GAT | 13.25 1.06 | ± | 14.03 2.32 | ± | 17.79 5.26 | ± | 10.84 1.05 | ± | 170.25 |
| Contr_l + CrossE_L + PR_L + Triplet_L | GCN | 54.68 29.34 | ± | 36.91 9.96 | ± | 62.15 4.53 | ± | 85.26 18.30 | ± | 62.375 |
| Contr_l + CrossE_L + PR_L + Triplet_L | GIN | 98.64 3.04 | ± | 100.00 0.00 | ± | 98.89 2.49 | ± | 95.63 4.31 | ± | 15.0 |
| Contr_l + CrossE_L + PR_L + Triplet_L | MPNN | 18.46 5.13 | ± | 25.23 7.47 | ± | 24.91 5.58 | ± | 37.79 12.52 | ± | 128.5 |
| Contr_l + CrossE_L + PR_L + Triplet_L | PAGNN | 22.98 7.49 | ± | 94.46 2.08 | ± | 32.61 7.75 | ± | 97.83 2.09 | ± | 72.5 |
| Contr_l + CrossE_L + PR_L + Triplet_L | SAGE | 16.96 2.22 | ± | 9.92±1.10 | | 24.25 5.68 | ± | 11.52 1.91 | ± | 168.5 |
| Contr_l + CrossE_L + Triplet_L | ALL | 25.15 7.17 | ± | 12.79 2.10 | ± | 37.94 3.93 | ± | 13.99 2.30 | ± | 121.5 |
| Contr_l + CrossE_L + Triplet_L | GAT | 10.23 1.24 | ± | 11.68 1.83 | ± | 14.73 0.80 | ± | 10.17 1.42 | ± | 184.75 |
| Contr_l + CrossE_L + Triplet_L | GCN | 46.15 22.79 | ± | 29.87 13.31 | ± | 45.65 18.67 | ± | 15.83 5.66 | ± | 94.0 |
| Contr_l + CrossE_L + Triplet_L | GIN | 81.93 21.04 | ± | 88.81 12.12 | ± | 92.01 14.34 | ± | 96.27 6.96 | ± | 32.25 |

Table 40. Results for Coherence (↑) (continued)

| Loss Type | Model | Cora ↓ Citeseer | Cora ↓ Bitcoin | Citeseer ↓ Cora | Citeseer ↓ Bitcoin | Average Rank |
|---|---|---|---|---|---|---|
| Contr_l + CrossE_L + Triplet_L | MPNN | 19.27 ± 2.54 | 16.86 ± 3.01 | 29.11 ± 5.70 | 13.70 ± 3.53 | 140.75 |
| Contr_l + CrossE_L + Triplet_L | PAGNN | 19.47 ± 3.42 | 74.20 ± 5.33 | 31.57 ± 5.40 | 65.49 ± 10.61 | 96.25 |
| Contr_l + CrossE_L + Triplet_L | SAGE | 14.19 ± 1.46 | 11.31 ± 0.97 | 17.61 ± 2.93 | 10.71 ± 1.27 | 176.5 |
| Contr_l + PMI_L | ALL | 18.15 ± 1.27 | 18.29 ± 6.35 | 37.40 ± 8.87 | 16.13 ± 2.72 | 128.75 |
| Contr_l + PMI_L | GAT | 9.46±0.63 | 9.85±0.89 | 13.71 ± 0.78 | 9.96±0.51 | 196.0 |
| Contr_l + PMI_L | GCN | 85.28 ± 18.12 | 62.35 ± 22.94 | 77.74 ± 13.63 | 42.93 ± 16.10 | 52.0 |
| Contr_l + PMI_L | GIN | 70.73 ± 28.90 | 97.01 ± 3.52 | 91.41 ± 13.63 | 100.00 ± 0.00 | 22.25 |
| Contr_l + PMI_L | MPNN | 20.39 ± 1.01 | 17.52 ± 1.95 | 29.93 ± 2.16 | 12.75 ± 1.80 | 137.375 |
| Contr_l + PMI_L | PAGNN | 23.01 ± 6.51 | 96.83 ± 1.21 | 55.33 ± 5.90 | 86.56 ± 4.07 | 58.5 |
| Contr_l + PMI_L | SAGE | 12.29 ± 0.86 | 8.86±0.50 | 20.12 ± 3.14 | 7.87±0.88 | 190.75 |
| Contr_l + PMI_L + PR_L | ALL | 31.35 ± 2.88 | 65.34 ± 5.36 | 44.44 ± 6.90 | 40.36 ± 7.63 | 78.5 |
| Contr_l + PMI_L + PR_L | GAT | 9.72±0.74 | 9.48±1.21 | 17.07 ± 3.17 | 13.54 ± 1.79 | 182.0 |
| Contr_l + PMI_L + PR_L | GCN | 56.91 ± 25.35 | 59.40 ± 23.80 | 63.66 ± 14.96 | 70.24 ± 5.86 | 58.0 |
| Contr_l + PMI_L + PR_L | GIN | 74.85 ± 23.75 | 100.00 ± 0.00 | 97.61 ± 5.35 | 97.97 ± 4.52 | 18.25 |
| Contr_l + PMI_L + PR_L | MPNN | 19.69 ± 3.20 | 16.94 ± 2.77 | 30.07 ± 2.05 | 49.59 ± 6.72 | 120.5 |
| Contr_l + PMI_L + PR_L | PAGNN | 27.68 ± 8.60 | 96.02 ± 0.91 | 51.29 ± 3.26 | 86.14 ± 1.29 | 56.0 |

Table 40. Results for Coherence (↑) (continued)

| Loss Type | Model | Cora ↓ Citeseer | | Cora ↓ Bitcoin | | Citeseer ↓ Cora | | Citeseer ↓ Bitcoin | | Average Rank |
|---|---|---|---|---|---|---|---|---|---|---|
| Contr_l + PMI_L + PR_L | SAGE | 11.35 ± 1.62 | | 8.05±0.43 4.71 | | 22.18 ± 1.69 | | 12.35 ± | | 180.25 |
| Contr_l + PMI_L + PR_L + Triplet_L | ALL | 19.07 ± 1.95 | | 19.36 ± 4.30 | | 48.53 ± 4.98 | | 25.39 ± 6.65 | | 113.0 |
| Contr_l + PMI_L + PR_L + Triplet_L | GAT | 12.01 ± 2.34 | | 11.35 ± 1.10 | | 16.34 ± 1.49 | | 12.90 ± 1.00 | | 174.5 |
| Contr_l + PMI_L + PR_L + Triplet_L | GCN | 69.17 ± 24.97 | | 47.12 ± 19.53 | | 63.80 ± 14.88 | | 54.17 ± 17.12 | | 60.375 |
| Contr_l + PMI_L + PR_L + Triplet_L | GIN | 63.54 ± 35.21 | | 100.00 ± 0.00 | | 92.39 ± 7.55 | | 95.29 ± 10.52 | | 27.25 |
| Contr_l + PMI_L + PR_L + Triplet_L | MPNN | 18.70 ± 3.52 | | 14.19 ± 1.66 | | 29.37 ± 5.66 | | 29.67 ± 3.78 | | 134.75 |
| Contr_l + PMI_L + PR_L + Triplet_L | PAGNN | 21.31 ± 4.62 | | 96.37 ± 0.62 | | 41.90 ± 6.26 | | 85.91 ± 3.37 | | 71.5 |
| Contr_l + PMI_L + PR_L + Triplet_L | SAGE | 15.91 ± 1.85 | | 9.95±0.98 | | 21.07 ± 2.19 | | 12.65 ± 1.58 | | 168.5 |
| Contr_l + PR_L | ALL | 16.83 ± 2.00 | | 28.48 ± 6.38 | | 23.82 ± 7.34 | | 22.36 ± 2.97 | | 138.5 |
| Contr_l + PR_L | GAT | 27.10 ± 9.71 | | 11.80 ± 0.87 | | 30.83 ± 3.04 | | 26.23 ± 5.35 | | 122.5 |
| Contr_l + PR_L | GCN | 75.53 ± 14.95 | | 46.02 ± 21.44 | | 75.09 ± 10.26 | | 51.27 ± 23.25 | | 57.0 |
| Contr_l + PR_L | GIN | 100.00 ± 0.00 | | 100.00 ± 0.00 | | 97.03 ± 4.50 | | 100.00 ± 0.00 | | 7.125 |
| Contr_l + PR_L | MPNN | 22.43 ± 3.67 | | 60.02 ± 13.60 | | 28.39 ± 4.16 | | 35.84 ± 6.44 | | 105.25 |
| Contr_l + PR_L | PAGNN | 28.99 ± 10.19 | | 96.65 ± 0.68 | | 42.30 ± 5.15 | | 98.65 ± 0.21 | | 53.5 |
| Contr_l + PR_L | SAGE | 23.34 ± 3.48 | | 11.65 ± 1.68 | | 27.37 ± 7.57 | | 22.07 ± 3.47 | | 135.75 |
| Contr_l + PR_L + Triplet_L | ALL | 22.40 ± 8.85 | | 25.06 ± 4.55 | | 34.35 ± 6.04 | | 11.98 ± 1.99 | | 124.5 |

Table 40. Results for Coherence (↑) (continued)

| Loss Type | Model | Cora ↓ Citeseer | | Cora ↓ Bitcoin | | Citeseer ↓ Cora | | Citeseer ↓ Bitcoin | | Average Rank |
|---|---|---|---|---|---|---|---|---|---|---|
| Contr_l + PR_L + Triplet_L | GAT | 15.18 | ± 1.91 | 13.36 | ± 1.78 | 17.90 | ± 2.64 | 12.53 | ± 0.49 | 165.5 |
| Contr_l + PR_L + Triplet_L | GCN | 50.01 | ± 28.53 | 36.91 | ± 16.81 | 75.71 | ± 16.49 | 58.78 | ± 9.94 | 63.875 |
| Contr_l + PR_L + Triplet_L | GIN | 89.15 | ± 18.14 | 99.99 | ± 0.03 | 94.83 | ± 8.31 | 98.31 | ± 3.79 | 17.25 |
| Contr_l + PR_L + Triplet_L | MPNN | 21.19 | ± 1.06 | 31.24 | ± 3.83 | 24.17 | ± 2.16 | 24.37 | ± 6.78 | 125.0 |
| Contr_l + PR_L + Triplet_L | PAGNN | 18.47 | ± 3.67 | 95.00 | ± 1.81 | 32.13 | ± 6.14 | 93.77 | ± 3.34 | 86.75 |
| Contr_l + PR_L + Triplet_L | SAGE | 20.97 | ± 2.09 | 12.40 | ± 1.73 | 21.04 | ± 1.82 | 10.08 | ± 0.95 | 156.75 |
| Contr_l + Triplet_L | ALL | 20.34 | ± 3.69 | 12.28 | ± 2.07 | 38.10 | ± 4.04 | 15.88 | ± 3.27 | 129.75 |
| Contr_l + Triplet_L | GAT | 9.86±0.67 | | 11.86 | ± 1.64 | 14.43 | ± 0.31 | 9.97±0.86 | | 186.875 |
| Contr_l + Triplet_L | GCN | 35.22 | ± 10.47 | 39.41 | ± 32.60 | 35.96 | ± 8.80 | 22.91 | ± 23.93 | 96.5 |
| Contr_l + Triplet_L | GIN | 49.56 | ± 8.83 | 80.32 | ± 18.17 | 83.44 | ± 23.43 | 96.03 | ± 7.31 | 44.0 |
| Contr_l + Triplet_L | MPNN | 17.74 | ± 5.31 | 17.33 | ± 3.69 | 29.22 | ± 3.96 | 12.50 | ± 3.55 | 147.0 |
| Contr_l + Triplet_L | PAGNN | 18.39 | ± 3.80 | 79.51 | ± 6.33 | 31.68 | ± 3.83 | 68.78 | ± 8.76 | 98.25 |
| Contr_l + Triplet_L | SAGE | 14.26 | ± 1.32 | 12.05 | ± 1.79 | 17.05 | ± 2.71 | 11.58 | ± 0.60 | 172.5 |
| CrossE_L | ALL | 42.74 | ± 6.61 | 14.81 | ± 1.37 | 47.97 | ± 27.13 | 21.20 | ± 2.35 | 99.5 |
| CrossE_L | GAT | 24.69 | ± 2.27 | 15.56 | ± 1.57 | 25.34 | ± 1.23 | 15.75 | ± 1.77 | 130.125 |
| CrossE_L | GCN | 99.45 | ± 0.84 | 83.47 | ± 16.41 | 99.13 | ± 0.34 | 97.93 | ± 2.88 | 23.25 |

Table 40. Results for Coherence (↑) (continued)

| Loss Type | Model | Cora ↓ Citeseer | Cora ↓ Bitcoin | Citeseer ↓ Cora | Citeseer ↓ Bitcoin | Average Rank |
|-----------|-------|-----------------|----------------|-----------------|--------------------|--------------|
| CrossE_L | GIN | 100.00 ± 0.00 | 100.00 ± 0.00 | 100.00 ± 0.00 | 100.00 ± 0.00 | 4.5 |
| CrossE_L | MPNN | 33.11 ± 6.39 | 34.61 ± 5.39 | 36.90 ± 12.11 | 32.68 ± 4.67 | 93.25 |
| CrossE_L | PAGNN | 42.24 ± 6.64 | 97.80 ± 1.57 | 55.40 ± 14.69 | 97.35 ± 1.70 | 43.0 |
| CrossE_L | SAGE | 24.80 ± 11.17 | 13.62 ± 3.35 | 33.81 ± 6.99 | 14.19 ± 3.67 | 123.25 |
| CrossE_L + PMI_L | ALL | 14.93 ± 1.38 | 28.72 ± 3.29 | 38.43 ± 6.10 | 25.33 ± 2.67 | 122.125 |
| CrossE_L + PMI_L | GAT | 9.29±0.75 | 10.25 ± 1.06 | 13.72 ± 0.79 | 9.04±1.07 | 198.25 |
| CrossE_L + PMI_L | GCN | 71.75 ± 28.49 | 45.26 ± 15.01 | 81.87 ± 5.41 | 50.58 ± 9.92 | 56.5 |
| CrossE_L + PMI_L | GIN | 76.97 ± 26.45 | 100.00 ± 0.00 | 92.50 ± 14.50 | 100.00 ± 0.01 | 14.0 |
| CrossE_L + PMI_L | MPNN | 24.39 ± 8.04 | 20.08 ± 5.45 | 34.54 ± 7.11 | 12.13 ± 2.19 | 121.75 |
| CrossE_L + PMI_L | PAGNN | 21.41 ± 2.69 | 95.81 ± 1.18 | 52.81 ± 5.23 | 91.20 ± 2.49 | 63.0 |
| CrossE_L + PMI_L | SAGE | 11.76 ± 1.29 | 8.34±1.72 | 19.59 ± 4.21 | 8.48±1.16 | 193.75 |
| CrossE_L + PMI_L + PR_L | ALL | 33.26 ± 8.63 | 53.98 ± 6.19 | 45.69 ± 7.42 | 42.45 ± 5.27 | 77.625 |
| CrossE_L + PMI_L + PR_L | GAT | 9.26±0.55 | 9.31±0.85 | 18.64 ± 6.33 | 18.38 ± 5.47 | 177.5 |
| CrossE_L + PMI_L + PR_L | GCN | 52.09 ± 9.20 | 52.96 ± 15.63 | 80.65 ± 6.21 | 42.05 ± 18.86 | 63.25 |
| CrossE_L + PMI_L + PR_L | GIN | 90.48 ± 19.59 | 100.00 ± 0.00 | 100.00 ± 0.00 | 100.00 ± 0.00 | 6.375 |
| CrossE_L + PMI_L + PR_L | MPNN | 24.69 ± 6.00 | 15.06 ± 1.54 | 31.30 ± 2.82 | 18.35 ± 3.12 | 121.125 |

Table 40. Results for Coherence (↑) (continued)

| Loss Type | Model | Cora ↓ Citeseer | Cora ↓ Bitcoin | Citeseer ↓ Cora | Citeseer ↓ Bitcoin | Average Rank |
|---|---|---|---|---|---|---|
| CrossE_L + PMI_L + PR_L | PAGNN | 20.39 ± 4.99 | 95.13 ± 0.79 | 49.29 ± 10.92 | 87.55 ± 2.55 | 69.625 |
| CrossE_L + PMI_L + PR_L | SAGE | 10.39 ± 0.49 | 7.80±0.64 | 17.84 ± 4.29 | 7.66±0.35 | 199.5 |
| CrossE_L + PMI_L + PR_L + Triplet_L | ALL | 18.82 ± 2.49 | 27.81 ± 3.17 | 44.08 ± 5.94 | 26.17 ± 2.99 | 112.25 |
| CrossE_L + PMI_L + PR_L + Triplet_L | GAT | 9.96±2.01 | 8.99±0.67 | 15.32 ± 0.98 | 10.83 ± 1.18 | 191.75 |
| CrossE_L + PMI_L + PR_L + Triplet_L | GCN | 69.60 ± 27.84 | 59.53 ± 12.04 | 77.00 ± 18.61 | 50.30 ± 19.78 | 55.25 |
| CrossE_L + PMI_L + PR_L + Triplet_L | GIN | 83.87 ± 16.35 | 100.00 ± 0.00 | 93.38 ± 8.84 | 99.95 ± 0.11 | 13.5 |
| CrossE_L + PMI_L + PR_L + Triplet_L | MPNN | 22.17 ± 2.38 | 19.01 ± 4.29 | 28.76 ± 3.17 | 18.50 ± 3.75 | 126.75 |
| CrossE_L + PMI_L + PR_L + Triplet_L | PAGNN | 22.89 ± 6.41 | 91.05 ± 2.37 | 43.61 ± 5.73 | 79.92 ± 2.25 | 72.75 |
| CrossE_L + PMI_L + PR_L + Triplet_L | SAGE | 11.95 ± 1.54 | 9.58±1.28 | 21.05 ± 3.48 | 9.44±0.77 | 183.0 |
| CrossE_L + PMI_L + Triplet_L | ALL | 20.15 ± 0.76 | 11.43 ± 1.13 | 39.00 ± 2.68 | 10.73 ± 0.38 | 141.5 |
| CrossE_L + PMI_L + Triplet_L | GAT | 9.45±0.28 | 9.23±1.01 | 13.96 ± 0.87 | 9.21±0.64 | 200.5 |
| CrossE_L + PMI_L + Triplet_L | GCN | 69.17 ± 19.95 | 35.08 ± 12.79 | 72.21 ± 12.16 | 58.10 ± 17.63 | 62.875 |
| CrossE_L + PMI_L + Triplet_L | GIN | 63.71 ± 34.07 | 100.00 ± 0.00 | 93.04 ± 15.56 | 98.86 ± 1.34 | 21.75 |
| CrossE_L + PMI_L + Triplet_L | MPNN | 20.42 ± 4.11 | 14.45 ± 1.88 | 29.69 ± 2.81 | 15.81 ± 2.83 | 136.875 |
| CrossE_L + PMI_L + Triplet_L | PAGNN | 20.62 ± 1.00 | 94.62 ± 1.45 | 51.02 ± 5.07 | 87.69 ± 2.18 | 68.0 |
| CrossE_L + PMI_L + Triplet_L | SAGE | 12.76 ± 2.13 | 9.16±1.29 | 21.51 ± 3.46 | 8.61±1.13 | 185.75 |

Table 40. Results for Coherence (↑) (continued)

| Loss Type | Model | Cora ↓ Citeseer | Cora ↓ Bitcoin | Citeseer ↓ Cora | Citeseer ↓ Bitcoin | Average Rank |
|---|---|---|---|---|---|---|
| CrossE_L + PR_L | ALL | 17.86 ± 0.84 | 26.91 ± 3.46 | 18.36 ± 1.90 | 37.68 ± 6.53 | 135.5 |
| CrossE_L + PR_L | GAT | 38.89 ± 13.90 | 20.96 ± 6.39 | 33.55 ± 6.23 | 28.75 ± 6.96 | 100.75 |
| CrossE_L + PR_L | GCN | 67.16 ± 21.76 | 73.02 ± 17.36 | 71.97 ± 7.52 | 59.68 ± 18.88 | 54.25 |
| CrossE_L + PR_L | GIN | 99.19 ± 1.81 | 99.87 ± 0.30 | <span style="background-color:#90EE90">100.00 ± 0.00</span> | 100.00 ± 0.00 | 8.375 |
| CrossE_L + PR_L | MPNN | 24.25 ± 6.92 | 57.68 ± 8.60 | 27.53 ± 4.44 | 46.06 ± 3.36 | 101.25 |
| CrossE_L + PR_L | PAGNN | 31.61 ± 7.82 | 97.48 ± 0.46 | 38.37 ± 6.88 | 97.91 ± 0.89 | 54.75 |
| CrossE_L + PR_L | SAGE | 37.24 ± 15.37 | 13.11 ± 1.93 | 38.01 ± 9.72 | 14.13 ± 1.75 | 114.25 |
| CrossE_L + PR_L + Triplet_L | ALL | 26.72 ± 9.84 | 26.62 ± 4.39 | 27.81 ± 4.84 | 11.39 ± 1.43 | 127.75 |
| CrossE_L + PR_L + Triplet_L | GAT | 18.58 ± 4.27 | 15.46 ± 2.17 | 20.78 ± 5.61 | 11.23 ± 0.90 | 156.375 |
| CrossE_L + PR_L + Triplet_L | GCN | 63.95 ± 22.73 | 58.28 ± 22.85 | 72.22 ± 14.73 | 31.80 ± 11.89 | 66.0 |
| CrossE_L + PR_L + Triplet_L | GIN | 93.39 ± 12.66 | 99.99 ± 0.02 | 90.61 ± 13.34 | 99.47 ± 1.18 | 16.5 |
| CrossE_L + PR_L + Triplet_L | MPNN | 19.29 ± 2.23 | 32.58 ± 4.77 | 27.69 ± 6.42 | 35.75 ± 3.63 | 122.25 |
| CrossE_L + PR_L + Triplet_L | PAGNN | 25.34 ± 5.00 | 86.54 ± 2.06 | 35.85 ± 4.58 | 97.60 ± 1.36 | 68.75 |
| CrossE_L + PR_L + Triplet_L | SAGE | 15.63 ± 1.52 | 10.21 ± 1.84 | 23.11 ± 4.81 | 9.80±1.44 | 174.625 |
| CrossE_L + Triplet_L | ALL | 31.89 ± 5.36 | 9.88±0.55 | 40.04 ± 8.24 | 9.17±0.13 | 136.25 |
| CrossE_L + Triplet_L | GAT | 9.81±0.28 | 10.37 ± 0.80 | 14.54 ± 0.78 | 9.87±0.70 | 191.25 |

Table 40. Results for Coherence (↑) (continued)

| Loss Type | Model | Cora ↓ Citeseer | | Cora ↓ Bitcoin | | Citeseer ↓ Cora | | Citeseer ↓ Bitcoin | | Average Rank |
|---|---|---|---|---|---|---|---|---|---|---|
| CrossE_L + Triplet_L | GCN | 39.49 | ± 32.18 | 22.60 | ± 7.07 | 60.02 | ± 18.31 | 66.20 | ± 22.20 | 72.75 |
| CrossE_L + Triplet_L | GIN | 81.18 | ± 25.75 | 89.98 | ± 15.77 | 98.32 | ± 3.69 | 97.73 | ± 5.07 | 27.5 |
| CrossE_L + Triplet_L | MPNN | 15.88 | ± 3.33 | 12.39 | ± 1.96 | 28.94 | ± 2.92 | 12.07 | ± 2.76 | 158.75 |
| CrossE_L + Triplet_L | PAGNN | 18.50 | ± 2.49 | 75.58 | ± 10.20 | 31.04 | ± 3.66 | 54.44 | ± 13.90 | 101.75 |
| CrossE_L + Triplet_L | SAGE | 10.92 | ± 0.98 | 10.02 | ± 0.43 | 16.40 | ± 2.74 | 10.38 | ± 1.18 | 186.0 |
| PMI_L | ALL | 16.24 | ± 1.93 | 33.07 | ± 8.00 | 36.29 | ± 3.42 | 23.76 | ± 5.07 | 122.75 |
| PMI_L | GAT | 9.19±0.45 | | 8.36±0.92 | | 13.62 | ± 0.52 | 8.98±0.76 | | 206.5 |
| PMI_L | GCN | 72.80 | ± 15.32 | 94.32 | ± 8.45 | 74.67 | ± 7.01 | 51.77 | ± 20.72 | 48.0 |
| PMI_L | GIN | 87.74 | ± 24.09 | 99.79 | ± 0.48 | 97.05 | ± 4.69 | 100.00 | ± 0.00 | 12.75 |
| PMI_L | MPNN | 20.74 | ± 2.11 | 14.64 | ± 2.74 | 31.84 | ± 1.97 | 14.00 | ± 1.81 | 132.0 |
| PMI_L | PAGNN | 26.24 | ± 4.90 | 95.37 | ± 1.49 | 54.70 | ± 5.10 | 84.39 | ± 2.74 | 58.5 |
| PMI_L | SAGE | 13.67 | ± 1.53 | 8.26±1.01 | | 21.31 | ± 5.08 | 9.34±0.54 | | 185.25 |
| PMI_L + PR_L | ALL | 29.52 | ± 7.92 | 55.90 | ± 6.96 | 47.59 | ± 13.27 | 45.56 | ± 8.36 | 77.5 |
| PMI_L + PR_L | GAT | 9.99±0.60 | | 10.99 | ± 1.49 | 16.98 | ± 4.24 | 13.36 | ± 1.70 | 177.75 |
| PMI_L + PR_L | GCN | 51.48 | ± 21.33 | 43.13 | ± 18.04 | 62.65 | ± 5.26 | 59.12 | ± 27.09 | 65.5 |
| PMI_L + PR_L | GIN | 76.83 | ± 26.47 | 100.00 | ± 0.00 | 83.79 | ± 21.34 | 99.98 | ± 0.04 | 18.25 |

Table 40. Results for Coherence (↑) (continued)

| Loss Type | Model | Cora ↓ Citeseer | Cora ↓ Bitcoin | Citeseer ↓ Cora | Citeseer ↓ Bitcoin | Average Rank |
|---|---|---|---|---|---|---|
| PMI_L + PR_L | MPNN | 22.87 ± 4.83 | 15.09 ± 1.72 | 30.26 ± 6.80 | 49.02 ± 9.05 | 114.75 |
| PMI_L + PR_L | PAGNN | 19.82 ± 2.86 | 93.24 ± 2.59 | 49.48 ± 7.82 | 87.11 ± 2.68 | 72.875 |
| PMI_L + PR_L | SAGE | 11.92 ± 1.68 | 8.43±0.45 | 18.81 ± 2.89 | 7.74±0.56 | 193.75 |
| PMI_L + PR_L + Triplet_L | ALL | 22.09 ± 2.93 | 35.49 ± 12.93 | 45.05 ± 5.35 | 26.35 ± 7.53 | 100.25 |
| PMI_L + PR_L + Triplet_L | GAT | 11.30 ± 2.09 | 9.28±0.75 | 14.63 ± 1.10 | 10.19 ± 0.71 | 190.5 |
| PMI_L + PR_L + Triplet_L | GCN | 69.63 ± 10.89 | 43.50 ± 13.76 | 72.43 ± 15.09 | 49.49 ± 13.16 | 62.25 |
| PMI_L + PR_L + Triplet_L | GIN | 58.42 ± 22.48 | 100.00 ± 0.00 | 92.67 ± 15.39 | 100.00 ± 0.00 | 19.0 |
| PMI_L + PR_L + Triplet_L | MPNN | 23.29 ± 6.69 | 14.45 ± 1.17 | 28.10 ± 2.85 | 40.82 ± 6.54 | 120.375 |
| PMI_L + PR_L + Triplet_L | PAGNN | 19.73 ± 1.91 | 96.52 ± 1.47 | 39.33 ± 4.38 | 81.69 ± 2.79 | 77.5 |
| PMI_L + PR_L + Triplet_L | SAGE | 13.18 ± 1.65 | 8.55±0.87 | 18.45 ± 2.64 | 9.79±0.85 | 187.25 |
| PMI_L + Triplet_L | ALL | 18.37 ± 4.52 | 11.87 ± 3.04 | 34.77 ± 2.47 | 13.00 ± 2.21 | 144.5 |
| PMI_L + Triplet_L | GAT | 9.28±0.92 | 9.38±1.09 | 13.34 ± 0.82 | 9.58±0.84 | 199.5 |
| PMI_L + Triplet_L | GCN | 75.67 ± 26.15 | 68.33 ± 18.48 | 76.89 ± 8.76 | 60.40 ± 15.84 | 49.25 |
| PMI_L + Triplet_L | GIN | 66.24 ± 11.30 | 98.00 ± 4.47 | 100.00 ± 0.00 | 99.99 ± 0.01 | 19.25 |
| PMI_L + Triplet_L | MPNN | 19.41 ± 1.98 | 13.51 ± 2.25 | 30.34 ± 3.17 | 12.36 ± 2.16 | 145.25 |
| PMI_L + Triplet_L | PAGNN | 24.21 ± 5.18 | 96.80 ± 0.89 | 48.99 ± 6.78 | 87.17 ± 3.75 | 60.25 |

Table 40. Results for Coherence (↑) (continued)

| Loss Type | Model | Cora ↓ Citeseer | | Cora ↓ Bitcoin | | Citeseer ↓ Cora | | Citeseer ↓ Bitcoin | | Average Rank |
|---|---|---|---|---|---|---|---|---|---|---|
| PMI_L + Triplet_L | SAGE | 12.08 ± 1.42 | | 11.00 ± 0.75 | | 17.53 ± 1.61 | | 8.57±1.11 | | 186.0 |
| PR_L | ALL | 25.41 ± 9.55 | | 42.72 ± 7.75 | | 20.08 ± 3.12 | | 47.64 ± 4.32 | | 108.5 |
| PR_L | GAT | 28.19 ± 6.96 | | 25.31 ± 6.68 | | 34.88 ± 6.88 | | 25.33 ± 7.60 | | 103.875 |
| PR_L | GCN | 89.31 ± 9.56 | | 73.77 ± 11.25 | | 80.19 ± 10.60 | | 56.10 ± 14.54 | | 45.0 |
| PR_L | GIN | 99.65 ± 0.76 | | 99.99 ± 0.02 | | 99.77 ± 0.53 | | 99.99 ± 0.01 | | 9.375 |
| PR_L | MPNN | 31.14 ± 9.50 | | 45.98 ± 2.58 | | 33.22 ± 5.33 | | 51.32 ± 3.98 | | 87.0 |
| PR_L | PAGNN | 41.41 ± 12.30 | | 99.61 ± 0.07 | | 40.11 ± 10.42 | | 99.51 ± 0.19 | | 46.25 |
| PR_L | SAGE | 28.43 ± 4.64 | | 21.23 ± 4.87 | | 36.86 ± 6.01 | | 15.76 ± 2.54 | | 108.75 |
| PR_L + Triplet_L | ALL | 18.58 ± 1.41 | | 22.16 ± 4.42 | | 19.27 ± 1.55 | | 39.80 ± 4.64 | | 133.125 |
| PR_L + Triplet_L | GAT | 22.35 ± 8.06 | | 15.48 ± 1.66 | | 32.70 ± 3.93 | | 26.78 ± 4.45 | | 118.25 |
| PR_L + Triplet_L | GCN | 61.27 ± 23.42 | | 42.69 ± 21.29 | | 76.02 ± 13.20 | | 66.47 ± 13.85 | | 59.0 |
| PR_L + Triplet_L | GIN | 100.00 ± 0.00 | | 99.57 ± 0.96 | | 97.55 ± 5.49 | | 99.40 ± 1.35 | | 13.375 |
| PR_L + Triplet_L | MPNN | 22.86 ± 6.59 | | 44.55 ± 3.42 | | 27.57 ± 2.83 | | 35.59 ± 6.86 | | 110.75 |
| PR_L + Triplet_L | PAGNN | 35.88 ± 4.80 | | 94.41 ± 1.33 | | 38.88 ± 4.49 | | 98.62 ± 0.29 | | 56.0 |
| PR_L + Triplet_L | SAGE | 28.09 ± 6.18 | | 20.68 ± 5.66 | | 30.62 ± 9.51 | | 15.85 ± 1.50 | | 116.5 |
| Triplet_L | ALL | 21.86 ± 3.88 | | 10.52 ± 2.49 | | 35.01 ± 2.23 | | 11.84 ± 2.15 | | 140.5 |

Table 40.  Results for Coherence (↑) (continued)

| Loss Type | Model | Cora ↓ Citeseer | | Cora ↓ Bitcoin | | Citeseer ↓ Cora | | Citeseer ↓ Bitcoin | Average Rank |
|---|---|---|---|---|---|---|---|---|---|
| Triplet_L | GAT | 10.45 | ± 1.18 | 10.44 | ± 1.30 | 14.42 | ± | 9.56±0.55 | 190.25 |
| Triplet_L | GCN | 33.93 | ± 12.77 | 21.21 | ± 4.00 | 58.70 | ± 18.75 | 33.98 ± 16.08 | 85.0 |
| Triplet_L | GIN | 83.40 | ± 19.45 | 100.00 | ± 0.00 | 82.05 | ± 24.84 | 74.28 ± 18.52 | 28.0 |
| Triplet_L | MPNN | 17.06 | ± 2.44 | 15.30 | ± 0.94 | 30.14 | ± 3.10 | 13.33 ± 1.55 | 146.625 |
| Triplet_L | PAGNN | 16.74 | ± 2.70 | 70.56 | ± 6.27 | 30.65 | ± 3.60 | 45.26 ± 10.54 | 111.25 |
| Triplet_L | SAGE | 11.86 | ± 1.70 | 9.36±1.28 | | 14.64 | ± 1.97 | 8.49±1.18 | 195.25 |

Table 41.  Selfcluster Performance (↑): This table presents models (Loss function and GNN) ranked by their average performance in terms of selfcluster. Top-ranked results are highlighted in red, second-ranked in blue, and third-ranked in green.

| Loss Type | Model | Cora ↓ Citeseer | | Cora ↓ Bitcoin | | Citeseer ↓ Cora | | Citeseer ↓ Bitcoin | Average Rank |
|---|---|---|---|---|---|---|---|---|---|
| Contr_l | ALL | −0.80 | ± 0.00 | −0.79 ± 0.00 | | −0.80 | ± 0.00 | −0.79 ± 0.00 | 85.375 |
| Contr_l | GAT | −0.81 | ± 0.01 | −0.80 | ± 0.00 | −0.81 | ± 0.00 | −0.80 ± 0.00 | 175.5 |
| Contr_l | GCN | −0.80 | ± 0.01 | −0.80 | ± 0.00 | −0.81 | ± 0.00 | −0.80 ± 0.00 | 155.75 |
| Contr_l | GIN | −0.80 | ± 0.00 | −0.79 ± 0.00 | | −0.80 | ± 0.01 | −0.80 ± 0.00 | 111.625 |
| Contr_l | MPNN | −0.80 | ± 0.00 | −0.80 | ± 0.01 | −0.80 | ± 0.00 | −0.79 ± 0.00 | 111.625 |

Table 41.  Results for Selfcluster (↑) (continued)

| Loss Type | Model | Cora ↓ Citeseer | | Cora ↓ Bitcoin | | Citeseer ↓ Cora | | Citeseer ↓ Bitcoin | | Average Rank |
|---|---|---|---|---|---|---|---|---|---|---|
| Contr_l | PAGNN | −0.80 0.00 | ± | −0.79 0.00 | ± | −0.80 0.00 | ± | −0.79 0.00 | ± | 85.375 |
| Contr_l | SAGE | −0.80 0.00 | ± | −0.80 0.00 | ± | −0.81 0.01 | ± | −0.80 0.00 | ± | 155.75 |
| Contr_l + CrossE_L | ALL | −0.80 0.00 | ± | −0.79 0.00 | ± | −0.80 0.00 | ± | −0.79 0.00 | ± | 85.375 |
| Contr_l + CrossE_L | GAT | −0.81 0.00 | ± | −0.80 0.00 | ± | −0.81 0.01 | ± | −0.80 0.00 | ± | 175.5 |
| Contr_l + CrossE_L | GCN | −0.81 0.00 | ± | −0.80 0.00 | ± | −0.81 0.00 | ± | −0.80 0.00 | ± | 175.5 |
| Contr_l + CrossE_L | GIN | −0.80 0.00 | ± | −0.79 0.00 | ± | −0.80 0.00 | ± | −0.79 0.00 | ± | 85.375 |
| Contr_l + CrossE_L | MPNN | −0.80 0.00 | ± | −0.80 0.01 | ± | −0.80 0.00 | ± | −0.80 0.00 | ± | 137.875 |
| Contr_l + CrossE_L | PAGNN | −0.80 0.00 | ± | −0.79 0.00 | ± | −0.80 0.00 | ± | −0.79 0.00 | ± | 85.375 |
| Contr_l + CrossE_L | SAGE | −0.81 0.01 | ± | −0.80 0.00 | ± | −0.80 0.00 | ± | −0.80 0.00 | ± | 157.625 |
| Contr_l + CrossE_L + PMI_L | ALL | −0.80 0.01 | ± | −0.79 0.00 | ± | −0.80 0.00 | ± | −0.79 0.00 | ± | 85.375 |
| Contr_l + CrossE_L + PMI_L | GAT | −0.81 0.00 | ± | −0.80 0.00 | ± | −0.81 0.01 | ± | −0.80 0.00 | ± | 175.5 |
| Contr_l + CrossE_L + PMI_L | GCN | −0.81 0.00 | ± | −0.80 0.00 | ± | −0.81 0.01 | ± | −0.80 0.00 | ± | 175.5 |
| Contr_l + CrossE_L + PMI_L | GIN | −0.80 0.00 | ± | −0.79 0.00 | ± | −0.80 0.00 | ± | −0.79 0.00 | ± | 85.375 |
| Contr_l + CrossE_L + PMI_L | MPNN | −0.81 0.00 | ± | −0.80 0.00 | ± | −0.80 0.00 | ± | −0.80 0.00 | ± | 157.625 |
| Contr_l + CrossE_L + PMI_L | PAGNN | −0.80 0.00 | ± | −0.79 0.00 | ± | −0.79 0.00 | ± | −0.79 0.00 | ± | 65.875 |
| Contr_l + CrossE_L + PMI_L | SAGE | −0.81 0.00 | ± | −0.80 0.00 | ± | −0.81 0.00 | ± | −0.80 0.00 | ± | 175.5 |

Table 41. Results for Selfcluster (↑) (continued)

| Loss Type | Model | Cora ↓ Citeseer | | Cora ↓ Bitcoin | | Citeseer ↓ Cora | | Citeseer ↓ Bitcoin | | Average Rank |
|---|---|---|---|---|---|---|---|---|---|---|
| Contr_l + CrossE_L + PMI_L + PR_L | ALL | −0.79 | ± 0.00 | −0.79 0.00 | ± | −0.79 0.00 | ± | −0.79 0.00 | ± | 46.75 |
| Contr_l + CrossE_L + PMI_L + PR_L | GAT | −0.81 0.00 | ± | −0.80 0.00 | ± | −0.81 0.00 | ± | −0.80 0.00 | ± | 175.5 |
| Contr_l + CrossE_L + PMI_L + PR_L | GCN | −0.81 0.00 | ± | −0.80 0.00 | ± | −0.81 0.01 | ± | −0.80 0.00 | ± | 175.5 |
| Contr_l + CrossE_L + PMI_L + PR_L | GIN | −0.80 0.00 | ± | −0.79 0.00 | ± | −0.80 0.01 | ± | −0.79 0.00 | ± | 85.375 |
| Contr_l + CrossE_L + PMI_L + PR_L | MPNN | −0.80 0.00 | ± | −0.80 0.00 | ± | −0.80 0.00 | ± | −0.79 0.00 | ± | 111.625 |
| Contr_l + CrossE_L + PMI_L + PR_L | PAGNN | −0.80 0.00 | ± | −0.79 0.00 | ± | −0.79 0.00 | ± | −0.79 0.00 | ± | 65.875 |
| Contr_l + CrossE_L + PMI_L + PR_L | SAGE | −0.81 0.00 | ± | −0.80 0.00 | ± | −0.81 0.00 | ± | −0.80 0.00 | ± | 175.5 |
| Contr_l + CrossE_L + PMI_L + PR_L + Triplet_L | ALL | −0.80 0.00 | ± | −0.79 0.00 | ± | −0.80 0.00 | ± | −0.79 0.00 | ± | 85.375 |
| Contr_l + CrossE_L + PMI_L + PR_L + Triplet_L | GAT | −0.81 0.00 | ± | −0.80 0.00 | ± | −0.80 0.01 | ± | −0.80 0.00 | ± | 157.625 |
| Contr_l + CrossE_L + PMI_L + PR_L + Triplet_L | GCN | −0.81 0.00 | ± | −0.80 0.00 | ± | −0.81 0.00 | ± | −0.80 0.00 | ± | 175.5 |
| Contr_l + CrossE_L + PMI_L + PR_L + Triplet_L | GIN | −0.80 0.00 | ± | −0.79 0.00 | ± | −0.80 0.00 | ± | −0.79 0.00 | ± | 85.375 |
| Contr_l + CrossE_L + PMI_L + PR_L + Triplet_L | MPNN | −0.81 0.01 | ± | −0.80 0.00 | ± | −0.80 0.00 | ± | −0.79 0.00 | ± | 131.375 |
| Contr_l + CrossE_L + PMI_L + PR_L + Triplet_L | PAGNN | −0.80 0.00 | ± | −0.79 0.00 | ± | −0.79 0.00 | ± | −0.79 0.00 | ± | 65.875 |

Table 41. Results for Selfcluster (↑) (continued)

| Loss Type | Model | Cora ↓ Citeseer | | Cora ↓ Bitcoin | | Citeseer ↓ Cora | | Citeseer ↓ Bitcoin | | Average Rank |
|---|---|---|---|---|---|---|---|---|---|---|
| Contr_l + CrossE_L + PMI_L + PR_L + Triplet_L | SAGE | −0.80 0.01 | ± | −0.80 0.00 | ± | −0.81 0.00 | ± | −0.80 0.00 | ± | 155.75 |
| Contr_l + CrossE_L + PMI_L + Triplet_L | ALL | −0.80 0.00 | ± | −0.79 0.00 | ± | −0.80 0.00 | ± | −0.79 0.00 | ± | 85.375 |
| Contr_l + CrossE_L + PMI_L + Triplet_L | GAT | −0.81 0.01 | ± | −0.80 0.00 | ± | −0.81 0.00 | ± | −0.80 0.00 | ± | 175.5 |
| Contr_l + CrossE_L + PMI_L + Triplet_L | GCN | −0.81 0.00 | ± | −0.80 0.00 | ± | −0.81 0.01 | ± | −0.80 0.00 | ± | 175.5 |
| Contr_l + CrossE_L + PMI_L + Triplet_L | GIN | −0.80 0.00 | ± | −0.79 0.00 | ± | −0.80 0.00 | ± | −0.79 0.00 | ± | 85.375 |
| Contr_l + CrossE_L + PMI_L + Triplet_L | MPNN | −0.80 0.00 | ± | −0.80 0.00 | ± | −0.80 0.00 | ± | −0.80 0.00 | ± | 137.875 |
| Contr_l + CrossE_L + PMI_L + Triplet_L | PAGNN | −0.80 0.00 | ± | −0.79 0.00 | ± | −0.79 0.00 | ± | −0.79 0.00 | ± | 65.875 |
| Contr_l + CrossE_L + PMI_L + Triplet_L | SAGE | −0.81 0.01 | ± | −0.80 0.00 | ± | −0.81 0.00 | ± | −0.80 0.00 | ± | 175.5 |
| Contr_l + CrossE_L + PR_L | ALL | −0.79 0.00 | ± | −0.79 0.00 | ± | −0.79 0.00 | ± | −0.79 0.00 | ± | 46.75 |
| Contr_l + CrossE_L + PR_L | GAT | −0.80 0.01 | ± | −0.79 0.00 | ± | −0.79 0.00 | ± | −0.79 0.00 | ± | 65.875 |
| Contr_l + CrossE_L + PR_L | GCN | −0.80 0.01 | ± | −0.79 0.00 | ± | −0.79 0.01 | ± | −0.79 0.00 | ± | 65.875 |
| Contr_l + CrossE_L + PR_L | GIN | −0.79 0.00 | ± | −0.79 0.00 | ± | −0.79 0.00 | ± | −0.79 0.00 | ± | 46.75 |
| Contr_l + CrossE_L + PR_L | MPNN | −0.79 0.00 | ± | −0.79 0.00 | ± | −0.79 0.00 | ± | −0.79 0.00 | ± | 46.75 |
| Contr_l + CrossE_L + PR_L | PAGNN | −0.79 0.00 | ± | −0.79 0.00 | ± | −0.79 0.00 | ± | −0.79 0.00 | ± | 46.75 |
| Contr_l + CrossE_L + PR_L | SAGE | −0.79 0.01 | ± | −0.79 0.00 | ± | −0.79 0.00 | ± | −0.79 0.00 | ± | 46.75 |
| Contr_l + CrossE_L + PR_L + Triplet_L | ALL | −0.79 0.00 | ± | −0.79 0.00 | ± | −0.80 0.01 | ± | −0.79 0.00 | ± | 66.25 |

Table 41. Results for Selfcluster (↑) (continued)

| Loss Type | Model | Cora ↓ Citeseer | | Cora ↓ Bitcoin | | Citeseer ↓ Cora | | Citeseer ↓ Bitcoin | | Average Rank |
|---|---|---|---|---|---|---|---|---|---|---|
| Contr_l + CrossE_L + PR_L + Triplet_L | GAT | −0.80 0.00 | ± | −0.80 0.00 | ± | −0.80 0.00 | ± | −0.80 0.00 | ± | 137.875 |
| Contr_l + CrossE_L + PR_L + Triplet_L | GCN | −0.80 0.00 | ± | −0.80 0.00 | ± | −0.80 0.00 | ± | −0.80 0.01 | ± | 137.875 |
| Contr_l + CrossE_L + PR_L + Triplet_L | GIN | −0.80 0.00 | ± | −0.79 0.00 | ± | −0.80 0.00 | ± | −0.79 0.00 | ± | 85.375 |
| Contr_l + CrossE_L + PR_L + Triplet_L | MPNN | −0.79 0.00 | ± | −0.79 0.00 | ± | −0.80 0.00 | ± | −0.79 0.00 | ± | 66.25 |
| Contr_l + CrossE_L + PR_L + Triplet_L | PAGNN | −0.80 0.01 | ± | −0.79 0.00 | ± | −0.79 0.00 | ± | −0.79 0.00 | ± | 65.875 |
| Contr_l + CrossE_L + PR_L + Triplet_L | SAGE | −0.80 0.00 | ± | −0.79 0.01 | ± | −0.80 0.00 | ± | −0.80 0.00 | ± | 111.625 |
| Contr_l + CrossE_L + Triplet_L | ALL | −0.80 0.00 | ± | −0.79 0.00 | ± | −0.80 0.01 | ± | −0.79 0.00 | ± | 85.375 |
| Contr_l + CrossE_L + Triplet_L | GAT | −0.81 0.00 | ± | −0.80 0.00 | ± | −0.81 0.00 | ± | −0.80 0.00 | ± | 175.5 |
| Contr_l + CrossE_L + Triplet_L | GCN | −0.81 0.01 | ± | −0.80 0.00 | ± | −0.81 0.00 | ± | −0.80 0.00 | ± | 175.5 |
| Contr_l + CrossE_L + Triplet_L | GIN | −0.80 0.00 | ± | −0.79 0.00 | ± | −0.80 0.00 | ± | −0.80 0.00 | ± | 111.625 |
| Contr_l + CrossE_L + Triplet_L | MPNN | −0.80 0.00 | ± | −0.79 0.00 | ± | −0.80 0.00 | ± | −0.80 0.00 | ± | 111.625 |
| Contr_l + CrossE_L + Triplet_L | PAGNN | −0.80 0.00 | ± | −0.79 0.00 | ± | −0.80 0.00 | ± | −0.79 0.00 | ± | 85.375 |
| Contr_l + CrossE_L + Triplet_L | SAGE | −0.80 0.00 | ± | −0.80 0.00 | ± | −0.81 0.01 | ± | −0.80 0.00 | ± | 155.75 |
| Contr_l + PMI_L | ALL | −0.80 0.00 | ± | −0.79 0.00 | ± | −0.80 0.00 | ± | −0.79 0.00 | ± | 85.375 |
| Contr_l + PMI_L | GAT | −0.81 0.00 | ± | −0.80 0.00 | ± | −0.81 0.00 | ± | −0.80 0.00 | ± | 175.5 |
| Contr_l + PMI_L | GCN | −0.81 0.00 | ± | −0.80 0.00 | ± | −0.81 0.00 | ± | −0.80 0.00 | ± | 175.5 |

Table 41. Results for Selfcluster (↑) (continued)

| Loss Type | Model | Cora ↓ Citeseer | | Cora ↓ Bitcoin | | Citeseer ↓ Cora | | Citeseer ↓ Bitcoin | | Average Rank |
|---|---|---|---|---|---|---|---|---|---|---|
| Contr_l + PMI_L | GIN | −0.80 | ± | −0.79 | ± | −0.80 | ± | −0.79 | ± | 85.375 |
| | | 0.00 | | 0.00 | | 0.00 | | 0.00 | | |
| Contr_l + PMI_L | MPNN | −0.80 | ± | −0.80 | ± | −0.80 | ± | −0.80 | ± | 137.875 |
| | | 0.01 | | 0.00 | | 0.00 | | 0.00 | | |
| Contr_l + PMI_L | PAGNN | −0.80 | ± | −0.79 | ± | −0.79 | ± | −0.79 | ± | 65.875 |
| | | 0.00 | | 0.00 | | 0.00 | | 0.00 | | |
| Contr_l + PMI_L | SAGE | −0.81 | ± | −0.80 | ± | −0.81 | ± | −0.80 | ± | 175.5 |
| | | 0.00 | | 0.00 | | 0.00 | | 0.00 | | |
| Contr_l + PMI_L + PR_L | ALL | −0.79 | ± | −0.79 | ± | −0.79 | ± | −0.79 | ± | 46.75 |
| | | 0.00 | | 0.00 | | 0.00 | | 0.00 | | |
| Contr_l + PMI_L + PR_L | GAT | −0.81 | ± | −0.80 | ± | −0.81 | ± | −0.80 | ± | 175.5 |
| | | 0.00 | | 0.00 | | 0.01 | | 0.01 | | |
| Contr_l + PMI_L + PR_L | GCN | −0.80 | ± | −0.80 | ± | −0.80 | ± | −0.80 | ± | 137.875 |
| | | 0.01 | | 0.00 | | 0.00 | | 0.00 | | |
| Contr_l + PMI_L + PR_L | GIN | −0.80 | ± | −0.79 | ± | −0.79 | ± | −0.79 | ± | 65.875 |
| | | 0.00 | | 0.00 | | 0.01 | | 0.00 | | |
| Contr_l + PMI_L + PR_L | MPNN | −0.80 | ± | −0.80 | ± | −0.80 | ± | −0.79 | ± | 111.625 |
| | | 0.00 | | 0.00 | | 0.00 | | 0.00 | | |
| Contr_l + PMI_L + PR_L | PAGNN | −0.80 | ± | −0.79 | ± | −0.79 | ± | −0.79 | ± | 65.875 |
| | | 0.00 | | 0.00 | | 0.00 | | 0.00 | | |
| Contr_l + PMI_L + PR_L | SAGE | −0.81 | ± | −0.80 | ± | −0.81 | ± | −0.80 | ± | 175.5 |
| | | 0.00 | | 0.00 | | 0.01 | | 0.00 | | |
| Contr_l + PMI_L + PR_L + Triplet_L | ALL | −0.80 | ± | −0.79 | ± | −0.79 | ± | −0.79 | ± | 65.875 |
| | | 0.00 | | 0.00 | | 0.01 | | 0.00 | | |
| Contr_l + PMI_L + PR_L + Triplet_L | GAT | −0.80 | ± | −0.80 | ± | −0.81 | ± | −0.80 | ± | 155.75 |
| | | 0.00 | | 0.00 | | 0.01 | | 0.00 | | |
| Contr_l + PMI_L + PR_L + Triplet_L | GCN | −0.81 | ± | −0.80 | ± | −0.81 | ± | −0.80 | ± | 175.5 |
| | | 0.00 | | 0.00 | | 0.01 | | 0.00 | | |
| Contr_l + PMI_L + PR_L + Triplet_L | GIN | −0.80 | ± | −0.79 | ± | −0.80 | ± | −0.79 | ± | 85.375 |
| | | 0.00 | | 0.00 | | 0.00 | | 0.00 | | |
| Contr_l + PMI_L + PR_L + Triplet_L | MPNN | −0.80 | ± | −0.80 | ± | −0.80 | ± | −0.79 | ± | 111.625 |
| | | 0.01 | | 0.00 | | 0.00 | | 0.00 | | |

Table 41. Results for Selfcluster (↑) (continued)

| Loss Type | Model | Cora ↓ Citeseer | | Cora ↓ Bitcoin | | Citeseer ↓ Cora | | Citeseer ↓ Bitcoin | | Average Rank |
|-----------|-------|------|---|------|---|------|---|------|---|--------------|
| Contr_l + PMI_L + PR_L + Triplet_L | PAGNN | −0.80 ± 0.00 | | −0.79 ± 0.00 | | −0.80 ± 0.00 | | −0.79 ± 0.00 | | 85.375 |
| Contr_l + PMI_L + PR_L + Triplet_L | SAGE | −0.80 ± 0.00 | | −0.80 ± 0.00 | | −0.80 ± 0.00 | | −0.80 ± 0.00 | | 137.875 |
| Contr_l + PR_L | ALL | −0.79 ± 0.00 | | −0.79 ± 0.00 | | −0.79 ± 0.00 | | −0.79 ± 0.00 | | 46.75 |
| Contr_l + PR_L | GAT | −0.79 ± 0.01 | | −0.79 ± 0.00 | | −0.79 ± 0.00 | | −0.79 ± 0.00 | | 46.75 |
| Contr_l + PR_L | GCN | −0.79 ± 0.01 | | −0.79 ± 0.01 | | −0.79 ± 0.01 | | −0.79 ± 0.00 | | 46.75 |
| Contr_l + PR_L | GIN | −0.79 ± 0.00 | | −0.79 ± 0.00 | | −0.79 ± 0.01 | | −0.79 ± 0.00 | | 46.75 |
| Contr_l + PR_L | MPNN | −0.79 ± 0.00 | | −0.79 ± 0.00 | | −0.79 ± 0.00 | | −0.79 ± 0.00 | | 46.75 |
| Contr_l + PR_L | PAGNN | −0.79 ± 0.00 | | −0.79 ± 0.00 | | −0.79 ± 0.00 | | −0.79 ± 0.00 | | 46.75 |
| Contr_l + PR_L | SAGE | −0.80 ± 0.00 | | −0.79 ± 0.00 | | −0.80 ± 0.00 | | −0.79 ± 0.00 | | 85.375 |
| Contr_l + PR_L + Triplet_L | ALL | −0.79 ± 0.00 | | −0.79 ± 0.00 | | −0.79 ± 0.00 | | −0.79 ± 0.00 | | 46.75 |
| Contr_l + PR_L + Triplet_L | GAT | −0.80 ± 0.00 | | −0.80 ± 0.00 | | −0.80 ± 0.00 | | −0.80 ± 0.01 | | 137.875 |
| Contr_l + PR_L + Triplet_L | GCN | −0.80 ± 0.00 | | −0.80 ± 0.00 | | −0.80 ± 0.00 | | −0.79 ± 0.01 | | 111.625 |
| Contr_l + PR_L + Triplet_L | GIN | −0.80 ± 0.00 | | −0.79 ± 0.00 | | −0.80 ± 0.00 | | −0.79 ± 0.00 | | 85.375 |
| Contr_l + PR_L + Triplet_L | MPNN | −0.79 ± 0.01 | | −0.79 ± 0.00 | | −0.80 ± 0.01 | | −0.79 ± 0.00 | | 66.25 |
| Contr_l + PR_L + Triplet_L | PAGNN | −0.79 ± 0.00 | | −0.79 ± 0.00 | | −0.80 ± 0.01 | | −0.79 ± 0.00 | | 66.25 |
| Contr_l + PR_L + Triplet_L | SAGE | −0.80 ± 0.00 | | −0.79 ± 0.00 | | −0.80 ± 0.00 | | −0.80 ± 0.01 | | 111.625 |

Table 41. Results for Selfcluster (↑) (continued)

| Loss Type | Model | Cora ↓ Citeseer | | Cora ↓ Bitcoin | | Citeseer ↓ Cora | | Citeseer ↓ Bitcoin | | Average Rank |
|---|---|---|---|---|---|---|---|---|---|---|
| Contr_l + Triplet_L | ALL | −0.80 | ± | −0.79 | ± | −0.80 | ± | −0.79 | ± | 85.375 |
| | | 0.00 | | 0.00 | | 0.01 | | 0.00 | | |
| Contr_l + Triplet_L | GAT | −0.81 | ± | −0.80 | ± | −0.81 | ± | −0.80 | ± | 175.5 |
| | | 0.00 | | 0.00 | | 0.00 | | 0.00 | | |
| Contr_l + Triplet_L | GCN | −0.81 | ± | −0.80 | ± | −0.81 | ± | −0.80 | ± | 175.5 |
| | | 0.00 | | 0.00 | | 0.00 | | 0.00 | | |
| Contr_l + Triplet_L | GIN | −0.80 | ± | −0.79 | ± | −0.80 | ± | −0.79 | ± | 85.375 |
| | | 0.00 | | 0.00 | | 0.00 | | 0.00 | | |
| Contr_l + Triplet_L | MPNN | −0.80 | ± | −0.79 | ± | −0.80 | ± | −0.80 | ± | 111.625 |
| | | 0.00 | | 0.00 | | 0.01 | | 0.00 | | |
| Contr_l + Triplet_L | PAGNN | −0.80 | ± | −0.79 | ± | −0.80 | ± | −0.79 | ± | 85.375 |
| | | 0.00 | | 0.00 | | 0.00 | | 0.00 | | |
| Contr_l + Triplet_L | SAGE | −0.80 | ± | −0.80 | ± | −0.81 | ± | −0.80 | ± | 155.75 |
| | | 0.01 | | 0.00 | | 0.01 | | 0.00 | | |
| CrossE_L | ALL | −0.79 | ± | −0.79 | ± | −0.79 | ± | −0.79 | ± | 46.75 |
| | | 0.00 | | 0.00 | | 0.00 | | 0.00 | | |
| CrossE_L | GAT | −0.79 | ± | −0.79 | ± | −0.79 | ± | −0.79 | ± | 46.75 |
| | | 0.00 | | 0.00 | | 0.00 | | 0.00 | | |
| CrossE_L | GCN | −0.79 | ± | −0.79 | ± | −0.79 | ± | −0.79 | ± | 46.75 |
| | | 0.00 | | 0.00 | | 0.00 | | 0.00 | | |
| CrossE_L | GIN | −0.79 | ± | −0.79 | ± | −0.79 | ± | −0.79 | ± | 46.75 |
| | | 0.00 | | 0.00 | | 0.00 | | 0.00 | | |
| CrossE_L | MPNN | −0.79 | ± | −0.79 | ± | −0.79 | ± | −0.79 | ± | 46.75 |
| | | 0.00 | | 0.00 | | 0.00 | | 0.00 | | |
| CrossE_L | PAGNN | −0.79 | ± | −0.79 | ± | −0.79 | ± | −0.79 | ± | 46.75 |
| | | 0.00 | | 0.00 | | 0.00 | | 0.00 | | |
| CrossE_L | SAGE | −0.79 | ± | −0.79 | ± | −0.79 | ± | −0.79 | ± | 46.75 |
| | | 0.00 | | 0.00 | | 0.00 | | 0.00 | | |
| CrossE_L + PMI_L | ALL | −0.80 | ± | −0.79 | ± | −0.80 | ± | −0.79 | ± | 85.375 |
| | | 0.00 | | 0.00 | | 0.01 | | 0.00 | | |
| CrossE_L + PMI_L | GAT | −0.81 | ± | −0.80 | ± | −0.81 | ± | −0.80 | ± | 175.5 |
| | | 0.00 | | 0.00 | | 0.00 | | 0.00 | | |

Table 41. Results for Selfcluster (↑) (continued)

| Loss Type | Model | Cora ↓ Citeseer | | Cora ↓ Bitcoin | | Citeseer ↓ Cora | | Citeseer ↓ Bitcoin | | Average Rank |
|---|---|---|---|---|---|---|---|---|---|---|
| CrossE_L + PMI_L | GCN | −0.81 | ± | −0.80 | ± | −0.81 | ± | −0.80 | ± | 175.5 |
| | | 0.01 | | 0.00 | | 0.00 | | 0.00 | | |
| CrossE_L + PMI_L | GIN | −0.80 | ± | −0.79 | ± | −0.80 | ± | −0.79 | ± | 85.375 |
| | | 0.00 | | 0.00 | | 0.00 | | 0.00 | | |
| CrossE_L + PMI_L | MPNN | −0.80 | ± | −0.80 | ± | −0.80 | ± | −0.80 | ± | 137.875 |
| | | 0.00 | | 0.00 | | 0.00 | | 0.00 | | |
| CrossE_L + PMI_L | PAGNN | −0.80 | ± | −0.79 | ± | −0.79 | ± | −0.79 | ± | 65.875 |
| | | 0.00 | | 0.00 | | 0.00 | | 0.00 | | |
| CrossE_L + PMI_L | SAGE | −0.81 | ± | −0.80 | ± | −0.81 | ± | −0.80 | ± | 175.5 |
| | | 0.00 | | 0.00 | | 0.00 | | 0.00 | | |
| CrossE_L + PMI_L + PR_L | ALL | −0.79 | ± | −0.79 | ± | −0.79 | ± | −0.79 | ± | 46.75 |
| | | 0.00 | | 0.00 | | 0.00 | | 0.00 | | |
| CrossE_L + PMI_L + PR_L | GAT | −0.81 | ± | −0.80 | ± | −0.80 | ± | −0.79 | ± | 131.375 |
| | | 0.00 | | 0.00 | | 0.01 | | 0.00 | | |
| CrossE_L + PMI_L + PR_L | GCN | −0.81 | ± | −0.80 | ± | −0.81 | ± | −0.80 | ± | 175.5 |
| | | 0.00 | | 0.00 | | 0.00 | | 0.00 | | |
| CrossE_L + PMI_L + PR_L | GIN | −0.80 | ± | −0.79 | ± | −0.79 | ± | −0.79 | ± | 65.875 |
| | | 0.00 | | 0.00 | | 0.01 | | 0.00 | | |
| CrossE_L + PMI_L + PR_L | MPNN | −0.80 | ± | −0.80 | ± | −0.80 | ± | −0.80 | ± | 137.875 |
| | | 0.01 | | 0.00 | | 0.00 | | 0.00 | | |
| CrossE_L + PMI_L + PR_L | PAGNN | −0.80 | ± | −0.79 | ± | −0.79 | ± | −0.79 | ± | 65.875 |
| | | 0.00 | | 0.00 | | 0.00 | | 0.00 | | |
| CrossE_L + PMI_L + PR_L | SAGE | −0.81 | ± | −0.80 | ± | −0.81 | ± | −0.80 | ± | 175.5 |
| | | 0.00 | | 0.00 | | 0.00 | | 0.00 | | |
| CrossE_L + PMI_L + PR_L + Triplet_L | ALL | −0.80 | ± | −0.79 | ± | −0.80 | ± | −0.79 | ± | 85.375 |
| | | 0.00 | | 0.00 | | 0.01 | | 0.00 | | |
| CrossE_L + PMI_L + PR_L + Triplet_L | GAT | −0.81 | ± | −0.80 | ± | −0.80 | ± | −0.80 | ± | 157.625 |
| | | 0.00 | | 0.00 | | 0.01 | | 0.00 | | |
| CrossE_L + PMI_L + PR_L + Triplet_L | GCN | −0.81 | ± | −0.80 | ± | −0.80 | ± | −0.80 | ± | 157.625 |
| | | 0.00 | | 0.00 | | 0.01 | | 0.00 | | |
| CrossE_L + PMI_L + PR_L + Triplet_L | GIN | −0.80 | ± | −0.79 | ± | −0.80 | ± | −0.79 | ± | 85.375 |
| | | 0.00 | | 0.00 | | 0.00 | | 0.00 | | |

Table 41.  Results for Selfcluster (↑) (continued)

| Loss Type | Model | Cora ↓ Citeseer | | Cora ↓ Bitcoin | | Citeseer ↓ Cora | | Citeseer ↓ Bitcoin | | Average Rank |
|---|---|---|---|---|---|---|---|---|---|---|
| CrossE_L + PMI_L + PR_L + Triplet_L | MPNN | −0.80 0.00 | ± | −0.80 0.00 | ± | −0.80 0.00 | ± | −0.79 0.00 | ± | 111.625 |
| CrossE_L + PMI_L + PR_L + Triplet_L | PAGNN | −0.80 0.00 | ± | −0.79 0.00 | ± | −0.79 0.00 | ± | −0.79 0.00 | ± | 65.875 |
| CrossE_L + PMI_L + PR_L + Triplet_L | SAGE | −0.81 0.01 | ± | −0.80 0.00 | ± | −0.81 0.00 | ± | −0.80 0.00 | ± | 175.5 |
| CrossE_L + PMI_L + Triplet_L | ALL | −0.80 0.00 | ± | −0.79 0.00 | ± | −0.80 0.00 | ± | −0.79 0.01 | ± | 85.375 |
| CrossE_L + PMI_L + Triplet_L | GAT | −0.81 0.00 | ± | −0.80 0.00 | ± | −0.81 0.00 | ± | −0.80 0.00 | ± | 175.5 |
| CrossE_L + PMI_L + Triplet_L | GCN | −0.81 0.00 | ± | −0.80 0.00 | ± | −0.81 0.00 | ± | −0.80 0.00 | ± | 175.5 |
| CrossE_L + PMI_L + Triplet_L | GIN | −0.80 0.00 | ± | −0.79 0.00 | ± | −0.80 0.00 | ± | −0.79 0.00 | ± | 85.375 |
| CrossE_L + PMI_L + Triplet_L | MPNN | −0.80 0.00 | ± | −0.80 0.00 | ± | −0.80 0.00 | ± | −0.80 0.00 | ± | 137.875 |
| CrossE_L + PMI_L + Triplet_L | PAGNN | −0.80 0.00 | ± | −0.79 0.00 | ± | −0.79 0.00 | ± | −0.79 0.00 | ± | 65.875 |
| CrossE_L + PMI_L + Triplet_L | SAGE | −0.81 0.00 | ± | −0.80 0.00 | ± | −0.81 0.00 | ± | −0.80 0.00 | ± | 175.5 |
| CrossE_L + PR_L | ALL | −0.79 0.00 | ± | −0.79 0.00 | ± | −0.79 0.00 | ± | −0.79 0.00 | ± | 46.75 |
| CrossE_L + PR_L | GAT | −0.79 0.00 | ± | −0.79 0.00 | ± | −0.79 0.00 | ± | −0.79 0.00 | ± | 46.75 |
| CrossE_L + PR_L | GCN | −0.79 0.00 | ± | −0.79 0.00 | ± | −0.79 0.00 | ± | −0.79 0.00 | ± | 46.75 |
| CrossE_L + PR_L | GIN | −0.79 0.00 | ± | −0.79 0.00 | ± | −0.79 0.00 | ± | −0.79 0.00 | ± | 46.75 |
| CrossE_L + PR_L | MPNN | −0.79 0.00 | ± | −0.79 0.00 | ± | −0.79 0.00 | ± | −0.79 0.00 | ± | 46.75 |
| CrossE_L + PR_L | PAGNN | −0.79 0.00 | ± | −0.79 0.00 | ± | −0.79 0.00 | ± | −0.79 0.00 | ± | 46.75 |

Table 41. Results for Selfcluster (↑) (continued)

| Loss Type | Model | Cora ↓ Citeseer | | Cora ↓ Bitcoin | | Citeseer ↓ Cora | | Citeseer ↓ Bitcoin | | Average Rank |
|-----------|-------|------|------|------|------|------|------|------|------|--------------|
| CrossE_L + PR_L | SAGE | −0.79 | ± | −0.79 | ± | −0.79 | ± | −0.79 | ± | 46.75 |
| | | 0.00 | | 0.00 | | 0.00 | | 0.00 | | |
| CrossE_L + PR_L + Triplet_L | ALL | −0.79 | ± | −0.79 | ± | −0.79 | ± | −0.79 | ± | 46.75 |
| | | 0.00 | | 0.00 | | 0.00 | | 0.00 | | |
| CrossE_L + PR_L + Triplet_L | GAT | −0.79 | ± | −0.79 | ± | −0.79 | ± | −0.79 | ± | 46.75 |
| | | 0.00 | | 0.00 | | 0.01 | | 0.00 | | |
| CrossE_L + PR_L + Triplet_L | GCN | −0.80 | ± | −0.80 | ± | −0.80 | ± | −0.79 | ± | 111.625 |
| | | 0.00 | | 0.00 | | 0.01 | | 0.00 | | |
| CrossE_L + PR_L + Triplet_L | GIN | −0.80 | ± | −0.79 | ± | −0.80 | ± | −0.79 | ± | 85.375 |
| | | 0.00 | | 0.00 | | 0.00 | | 0.00 | | |
| CrossE_L + PR_L + Triplet_L | MPNN | −0.79 | ± | −0.79 | ± | −0.79 | ± | −0.79 | ± | 46.75 |
| | | 0.00 | | 0.00 | | 0.00 | | 0.00 | | |
| CrossE_L + PR_L + Triplet_L | PAGNN | −0.79 | ± | −0.79 | ± | −0.80 | ± | −0.79 | ± | 66.25 |
| | | 0.00 | | 0.00 | | 0.00 | | 0.00 | | |
| CrossE_L + PR_L + Triplet_L | SAGE | −0.80 | ± | −0.79 | ± | −0.80 | ± | −0.79 | ± | 85.375 |
| | | 0.00 | | 0.00 | | 0.00 | | 0.01 | | |
| CrossE_L + Triplet_L | ALL | −0.80 | ± | −0.79 | ± | −0.80 | ± | −0.79 | ± | 85.375 |
| | | 0.00 | | 0.00 | | 0.01 | | 0.00 | | |
| CrossE_L + Triplet_L | GAT | −0.81 | ± | −0.80 | ± | −0.81 | ± | −0.80 | ± | 175.5 |
| | | 0.00 | | 0.00 | | 0.01 | | 0.00 | | |
| CrossE_L + Triplet_L | GCN | −0.81 | ± | −0.80 | ± | −0.81 | ± | −0.80 | ± | 175.5 |
| | | 0.00 | | 0.00 | | 0.00 | | 0.00 | | |
| CrossE_L + Triplet_L | GIN | −0.80 | ± | −0.79 | ± | −0.80 | ± | −0.79 | ± | 85.375 |
| | | 0.00 | | 0.00 | | 0.00 | | 0.00 | | |
| CrossE_L + Triplet_L | MPNN | −0.80 | ± | −0.80 | ± | −0.80 | ± | −0.80 | ± | 137.875 |
| | | 0.00 | | 0.00 | | 0.00 | | 0.00 | | |
| CrossE_L + Triplet_L | PAGNN | −0.80 | ± | −0.79 | ± | −0.80 | ± | −0.79 | ± | 85.375 |
| | | 0.00 | | 0.00 | | 0.00 | | 0.00 | | |
| CrossE_L + Triplet_L | SAGE | −0.81 | ± | −0.80 | ± | −0.81 | ± | −0.80 | ± | 175.5 |
| | | 0.00 | | 0.00 | | 0.00 | | 0.00 | | |
| PMI_L | ALL | −0.80 | ± | −0.79 | ± | −0.80 | ± | −0.79 | ± | 85.375 |
| | | 0.00 | | 0.00 | | 0.01 | | 0.00 | | |

Table 41. Results for Selfcluster (↑) (continued)

| Loss Type | Model | Cora ↓ Citeseer | | Cora ↓ Bitcoin | | Citeseer ↓ Cora | | Citeseer ↓ Bitcoin | | Average Rank |
|---|---|---|---|---|---|---|---|---|---|---|
| PMI_L | GAT | −0.81 | ± | −0.80 | ± | −0.81 | ± | −0.80 | ± | 175.5 |
| | | 0.01 | | 0.00 | | 0.00 | | 0.00 | | |
| PMI_L | GCN | −0.81 | ± | −0.80 | ± | −0.81 | ± | −0.80 | ± | 175.5 |
| | | 0.00 | | 0.00 | | 0.00 | | 0.00 | | |
| PMI_L | GIN | −0.80 | ± | −0.79 | ± | −0.80 | ± | −0.79 | ± | 85.375 |
| | | 0.00 | | 0.00 | | 0.00 | | 0.00 | | |
| PMI_L | MPNN | −0.80 | ± | −0.80 | ± | −0.80 | ± | −0.80 | ± | 137.875 |
| | | 0.01 | | 0.00 | | 0.00 | | 0.00 | | |
| PMI_L | PAGNN | −0.80 | ± | −0.79 | ± | −0.79 | ± | −0.79 | ± | 65.875 |
| | | 0.00 | | 0.00 | | 0.00 | | 0.00 | | |
| PMI_L | SAGE | −0.81 | ± | −0.80 | ± | −0.81 | ± | −0.80 | ± | 175.5 |
| | | 0.00 | | 0.00 | | 0.00 | | 0.00 | | |
| PMI_L + PR_L | ALL | −0.79 | ± | −0.79 | ± | −0.79 | ± | −0.79 | ± | 46.75 |
| | | 0.00 | | 0.00 | | 0.00 | | 0.00 | | |
| PMI_L + PR_L | GAT | −0.81 | ± | −0.80 | ± | −0.81 | ± | −0.80 | ± | 175.5 |
| | | 0.01 | | 0.00 | | 0.00 | | 0.00 | | |
| PMI_L + PR_L | GCN | −0.81 | ± | −0.80 | ± | −0.80 | ± | −0.80 | ± | 157.625 |
| | | 0.01 | | 0.00 | | 0.01 | | 0.00 | | |
| PMI_L + PR_L | GIN | −0.80 | ± | −0.79 | ± | −0.80 | ± | −0.79 | ± | 85.375 |
| | | 0.00 | | 0.00 | | 0.01 | | 0.00 | | |
| PMI_L + PR_L | MPNN | −0.80 | ± | −0.80 | ± | −0.80 | ± | −0.79 | ± | 111.625 |
| | | 0.00 | | 0.00 | | 0.00 | | 0.00 | | |
| PMI_L + PR_L | PAGNN | −0.80 | ± | −0.79 | ± | −0.79 | ± | −0.79 | ± | 65.875 |
| | | 0.00 | | 0.00 | | 0.00 | | 0.00 | | |
| PMI_L + PR_L | SAGE | −0.81 | ± | −0.80 | ± | −0.81 | ± | −0.80 | ± | 175.5 |
| | | 0.00 | | 0.00 | | 0.00 | | 0.00 | | |
| PMI_L + PR_L + Triplet_L | ALL | −0.80 | ± | −0.79 | ± | −0.79 | ± | −0.79 | ± | 65.875 |
| | | 0.00 | | 0.00 | | 0.01 | | 0.00 | | |
| PMI_L + PR_L + Triplet_L | GAT | −0.81 | ± | −0.80 | ± | −0.81 | ± | −0.80 | ± | 175.5 |
| | | 0.01 | | 0.00 | | 0.01 | | 0.00 | | |
| PMI_L + PR_L + Triplet_L | GCN | −0.81 | ± | −0.80 | ± | −0.81 | ± | −0.80 | ± | 175.5 |
| | | 0.00 | | 0.00 | | 0.00 | | 0.00 | | |

Table 41. Results for Selfcluster (↑) (continued)

| Loss Type | Model | Cora ↓ Citeseer | | Cora ↓ Bitcoin | | Citeseer ↓ Cora | | Citeseer ↓ Bitcoin | | Average Rank |
|---|---|---|---|---|---|---|---|---|---|---|
| PMI_L + PR_L + Triplet_L | GIN | −0.80 | ± 0.00 | −0.79 | ± 0.00 | −0.80 | ± 0.00 | −0.79 | ± 0.00 | 85.375 |
| PMI_L + PR_L + Triplet_L | MPNN | −0.80 | ± 0.01 | −0.80 | ± 0.00 | −0.80 | ± 0.00 | −0.79 | ± 0.00 | 111.625 |
| PMI_L + PR_L + Triplet_L | PAGNN | −0.80 | ± 0.00 | −0.79 | ± 0.00 | −0.79 | ± 0.01 | −0.79 | ± 0.00 | 65.875 |
| PMI_L + PR_L + Triplet_L | SAGE | −0.81 | ± 0.01 | −0.80 | ± 0.00 | −0.81 | ± 0.00 | −0.80 | ± 0.00 | 175.5 |
| PMI_L + Triplet_L | ALL | −0.80 | ± 0.00 | −0.79 | ± 0.00 | −0.80 | ± 0.00 | −0.79 | ± 0.01 | 85.375 |
| PMI_L + Triplet_L | GAT | −0.81 | ± 0.00 | −0.80 | ± 0.00 | −0.81 | ± 0.00 | −0.80 | ± 0.00 | 175.5 |
| PMI_L + Triplet_L | GCN | −0.81 | ± 0.01 | −0.80 | ± 0.00 | −0.81 | ± 0.00 | −0.80 | ± 0.00 | 175.5 |
| PMI_L + Triplet_L | GIN | −0.80 | ± 0.00 | −0.79 | ± 0.00 | −0.80 | ± 0.00 | −0.79 | ± 0.00 | 85.375 |
| PMI_L + Triplet_L | MPNN | −0.80 | ± 0.00 | −0.80 | ± 0.00 | −0.80 | ± 0.01 | −0.80 | ± 0.00 | 137.875 |
| PMI_L + Triplet_L | PAGNN | −0.80 | ± 0.00 | −0.79 | ± 0.00 | −0.79 | ± 0.00 | −0.79 | ± 0.00 | 65.875 |
| PMI_L + Triplet_L | SAGE | −0.81 | ± 0.00 | −0.80 | ± 0.00 | −0.81 | ± 0.00 | −0.80 | ± 0.00 | 175.5 |
| PR_L | ALL | −0.79 | ± 0.00 | −0.79 | ± 0.00 | −0.79 | ± 0.00 | −0.79 | ± 0.00 | 46.75 |
| PR_L | GAT | −0.79 | ± 0.00 | −0.79 | ± 0.00 | −0.79 | ± 0.00 | −0.79 | ± 0.00 | 46.75 |
| PR_L | GCN | −0.79 | ± 0.00 | −0.79 | ± 0.00 | −0.79 | ± 0.00 | −0.79 | ± 0.00 | 46.75 |
| PR_L | GIN | −0.79 | ± 0.00 | −0.79 | ± 0.00 | −0.79 | ± 0.00 | −0.79 | ± 0.00 | 46.75 |
| PR_L | MPNN | −0.79 | ± 0.00 | −0.79 | ± 0.00 | −0.79 | ± 0.00 | −0.79 | ± 0.00 | 46.75 |

Table 41. Results for Selfcluster (↑) (continued)

| Loss Type | Model | Cora ↓ Citeseer | | Cora ↓ Bitcoin | | Citeseer ↓ Cora | | Citeseer ↓ Bitcoin | | Average Rank |
|---|---|---|---|---|---|---|---|---|---|---|
| PR_L | PAGNN | −0.79 | ± | −0.79 | ± | −0.79 | ± | −0.79 | ± | 46.75 |
| | | 0.00 | | 0.00 | | 0.00 | | 0.00 | | |
| PR_L | SAGE | −0.79 | ± | −0.79 | ± | −0.79 | ± | −0.79 | ± | 46.75 |
| | | 0.00 | | 0.00 | | 0.00 | | 0.00 | | |
| PR_L + Triplet_L | ALL | −0.79 | ± | −0.79 | ± | −0.79 | ± | −0.79 | ± | 46.75 |
| | | 0.00 | | 0.00 | | 0.00 | | 0.00 | | |
| PR_L + Triplet_L | GAT | −0.79 | ± | −0.79 | ± | −0.79 | ± | −0.79 | ± | 46.75 |
| | | 0.01 | | 0.00 | | 0.00 | | 0.00 | | |
| PR_L + Triplet_L | GCN | −0.79 | ± | −0.79 | ± | −0.79 | ± | −0.79 | ± | 46.75 |
| | | 0.00 | | 0.00 | | 0.00 | | 0.00 | | |
| PR_L + Triplet_L | GIN | −0.79 | ± | −0.79 | ± | −0.80 | ± | −0.79 | ± | 66.25 |
| | | 0.00 | | 0.00 | | 0.01 | | 0.00 | | |
| PR_L + Triplet_L | MPNN | −0.79 | ± | −0.79 | ± | −0.79 | ± | −0.79 | ± | 46.75 |
| | | 0.00 | | 0.00 | | 0.00 | | 0.00 | | |
| PR_L + Triplet_L | PAGNN | −0.79 | ± | −0.79 | ± | −0.79 | ± | −0.79 | ± | 46.75 |
| | | 0.00 | | 0.00 | | 0.00 | | 0.00 | | |
| PR_L + Triplet_L | SAGE | −0.79 | ± | −0.79 | ± | −0.79 | ± | −0.79 | ± | 46.75 |
| | | 0.00 | | 0.00 | | 0.00 | | 0.00 | | |
| Triplet_L | ALL | −0.80 | ± | −0.79 | ± | −0.80 | ± | −0.79 | ± | 85.375 |
| | | 0.00 | | 0.00 | | 0.00 | | 0.00 | | |
| Triplet_L | GAT | −0.81 | ± | −0.80 | ± | −0.80 | ± | −0.80 | ± | 157.625 |
| | | 0.00 | | 0.00 | | 0.01 | | 0.00 | | |
| Triplet_L | GCN | −0.81 | ± | −0.80 | ± | −0.81 | ± | −0.80 | ± | 175.5 |
| | | 0.00 | | 0.00 | | 0.01 | | 0.00 | | |
| Triplet_L | GIN | −0.80 | ± | −0.79 | ± | −0.80 | ± | −0.79 | ± | 85.375 |
| | | 0.00 | | 0.00 | | 0.00 | | 0.00 | | |
| Triplet_L | MPNN | −0.80 | ± | −0.80 | ± | −0.80 | ± | −0.80 | ± | 137.875 |
| | | 0.01 | | 0.00 | | 0.00 | | 0.00 | | |
| Triplet_L | PAGNN | −0.80 | ± | −0.79 | ± | −0.80 | ± | −0.79 | ± | 85.375 |
| | | 0.00 | | 0.00 | | 0.00 | | 0.00 | | |
| Triplet_L | SAGE | −0.81 | ± | −0.80 | ± | −0.81 | ± | −0.80 | ± | 175.5 |
| | | 0.01 | | 0.00 | | 0.00 | | 0.00 | | |

Table 42. Rankme Performance (↑): This table presents models (Loss function and GNN) ranked by their average performance in terms of rankme. Top-ranked results are highlighted in red, second-ranked in blue, and third-ranked in green.

| Loss Type | Model | Cora ↓ Citeseer | Cora ↓ Bitcoin | Citeseer ↓ Cora | Citeseer ↓ Bitcoin | Average Rank |
|---|---|---|---|---|---|---|
| Contr_l | ALL | 283.67 ± 11.67 | 238.56 ± 3.92 | 283.31 ± 13.65 | 254.37 ± 6.99 | 169.5 |
| Contr_l | GAT | 442.87 ± 3.78 | 430.22 ± 3.73 | 434.63 ± 2.59 | 435.30 ± 0.62 | 26.25 |
| Contr_l | GCN | 419.76 ± 2.39 | 412.33 ± 1.65 | 411.61 ± 2.67 | 418.01 ± 1.62 | 49.25 |
| Contr_l | GIN | 388.78 ± 6.01 | 347.58 ± 2.29 | 376.58 ± 8.08 | 365.96 ± 5.33 | 101.75 |
| Contr_l | MPNN | 400.19 ± 7.29 | 393.92 ± 4.00 | 389.83 ± 8.38 | 385.24 ± 4.75 | 86.75 |
| Contr_l | PAGNN | 353.58 ± 3.85 | 236.95 ± 1.54 | 332.42 ± 19.73 | 263.83 ± 2.68 | 150.5 |
| Contr_l | SAGE | 394.06 ± 4.91 | 405.84 ± 3.11 | 389.36 ± 5.53 | 402.49 ± 4.53 | 81.0 |
| Contr_l + CrossE_L | ALL | 289.58 ± 17.25 | 291.63 ± 7.35 | 251.26 ± 18.55 | 206.96 ± 4.48 | 173.0 |
| Contr_l + CrossE_L | GAT | 437.82 ± 4.39 | 422.83 ± 1.01 | 428.88 ± 2.79 | 426.13 ± 0.45 | 32.5 |
| Contr_l + CrossE_L | GCN | 414.83 ± 0.80 | 406.21 ± 2.10 | 411.21 ± 4.43 | 416.44 ± 1.88 | 60.0 |
| Contr_l + CrossE_L | GIN | 369.97 ± 11.62 | 303.78 ± 4.13 | 360.68 ± 5.63 | 349.36 ± 4.13 | 122.75 |
| Contr_l + CrossE_L | MPNN | 397.39 ± 14.40 | 390.27 ± 4.30 | 380.15 ± 10.86 | 381.26 ± 3.90 | 91.75 |
| Contr_l + CrossE_L | PAGNN | 349.89 ± 3.19 | 211.20 ± 3.24 | 320.30 ± 7.64 | 238.64 ± 3.60 | 159.0 |
| Contr_l + CrossE_L | SAGE | 382.98 ± 3.78 | 387.91 ± 1.95 | 364.20 ± 12.10 | 364.38 ± 4.55 | 107.75 |
| Contr_l + CrossE_L + PMI_L | ALL | 330.78 ± 6.46 | 248.98 ± 4.72 | 353.16 ± 6.99 | 311.63 ± 3.11 | 143.75 |

Table 42. Results for Rankme (↑) (continued)

| Loss Type | Model | Cora ↓ Citeseer | Cora ↓ Bitcoin | Citeseer ↓ Cora | Citeseer ↓ Bitcoin | Average Rank |
|---|---|---|---|---|---|---|
| Contr_l + CrossE_L + PMI_L | GAT | 454.81 ± 1.30 | 440.97 ± 1.18 | 443.67 ± 2.59 | 442.78 ± 1.80 | 7.5 |
| Contr_l + CrossE_L + PMI_L | GCN | 422.43 ± 1.73 | 419.81 ± 2.02 | 406.94 ± 4.56 | 418.99 ± 1.18 | 47.25 |
| Contr_l + CrossE_L + PMI_L | GIN | 388.59 ± 15.95 | 292.49 ± 5.02 | 365.23 ± 6.08 | 322.75 ± 6.83 | 120.0 |
| Contr_l + CrossE_L + PMI_L | MPNN | 419.22 ± 3.18 | 408.35 ± 3.24 | 395.59 ± 4.85 | 410.32 ± 1.92 | 64.75 |
| Contr_l + CrossE_L + PMI_L | PAGNN | 344.54 ± 4.16 | 194.32 ± 0.83 | 314.51 ± 8.20 | 201.35 ± 1.98 | 171.25 |
| Contr_l + CrossE_L + PMI_L | SAGE | 449.73 ± 2.99 | 444.07 ± 1.04 | 437.97 ± 3.32 | 443.06 ± 3.17 | 12.5 |
| Contr_l + CrossE_L + PMI_L + PR_L | ALL | 280.78 ± 11.21 | 109.94 ± 1.80 | 342.49 ± 4.61 | 284.14 ± 1.76 | 166.5 |
| Contr_l + CrossE_L + PMI_L + PR_L | GAT | 451.58 ± 4.08 | 438.50 ± 2.23 | 444.09 ± 2.35 | 441.02 ± 3.30 | 13.5 |
| Contr_l + CrossE_L + PMI_L + PR_L | GCN | 423.02 ± 1.71 | 421.24 ± 1.80 | 405.36 ± 8.41 | 413.14 ± 2.36 | 51.25 |
| Contr_l + CrossE_L + PMI_L + PR_L | GIN | 387.54 ± 4.73 | 291.19 ± 7.74 | 349.05 ± 23.41 | 261.93 ± 6.81 | 133.5 |
| Contr_l + CrossE_L + PMI_L + PR_L | MPNN | 414.51 ± 2.69 | 405.35 ± 3.60 | 389.16 ± 3.55 | 361.30 ± 6.43 | 85.25 |
| Contr_l + CrossE_L + PMI_L + PR_L | PAGNN | 350.75 ± 3.29 | 208.93 ± 1.44 | 324.82 ± 4.55 | 202.30 ± 1.87 | 161.25 |
| Contr_l + CrossE_L + PMI_L + PR_L | SAGE | 451.80 ± 0.98 | 447.45 ± 1.65 | 439.60 ± 3.29 | 444.03 ± 2.14 | 8.5 |
| Contr_l + CrossE_L + PMI_L + PR_L + Triplet_L | ALL | 326.89 ± 11.26 | 235.33 ± 5.55 | 329.97 ± 12.83 | 275.76 ± 13.42 | 156.75 |
| Contr_l + CrossE_L + PMI_L + PR_L + Triplet_L | GAT | 452.80 ± 1.16 | 437.72 ± 1.51 | 438.83 ± 3.64 | 440.44 ± 1.64 | 15.5 |

Table 42. Results for Rankme (↑) (continued)

| Loss Type | Model | Cora ↓ Citeseer | Cora ↓ Bitcoin | Citeseer ↓ Cora | Citeseer ↓ Bitcoin | Average Rank |
|---|---|---|---|---|---|---|
| Contr_l + CrossE_L + PMI_L + PR_L + Triplet_L | GCN | 424.20 ± 2.82 | 417.84 ± 1.02 | 406.98 ± 4.15 | 416.46 ± 3.12 | 48.75 |
| Contr_l + CrossE_L + PMI_L + PR_L + Triplet_L | GIN | 390.76 ± 7.27 | 290.04 ± 5.75 | 377.70 ± 7.13 | 326.13 ± 5.85 | 114.25 |
| Contr_l + CrossE_L + PMI_L + PR_L + Triplet_L | MPNN | 414.92 ± 3.58 | 398.94 ± 2.32 | 395.99 ± 2.89 | 376.14 ± 4.29 | 81.25 |
| Contr_l + CrossE_L + PMI_L + PR_L + Triplet_L | PAGNN | 343.82 ± 3.52 | 197.34 ± 1.83 | 336.28 ± 5.05 | 211.04 ± 1.00 | 164.25 |
| Contr_l + CrossE_L + PMI_L + PR_L + Triplet_L | SAGE | 436.61 ± 8.80 | 436.63 ± 1.62 | 423.74 ± 2.48 | 433.35 ± 2.49 | 28.5 |
| Contr_l + CrossE_L + PMI_L + Triplet_L | ALL | 358.45 ± 13.81 | 330.67 ± 4.02 | 354.10 ± 1.99 | 365.68 ± 6.67 | 120.0 |
| Contr_l + CrossE_L + PMI_L + Triplet_L | GAT | 452.88 ± 1.35 | 441.27 ± 1.12 | 443.59 ± 2.49 | 440.70 ± 0.97 | 10.5 |
| Contr_l + CrossE_L + PMI_L + Triplet_L | GCN | 424.24 ± 3.32 | 418.85 ± 2.31 | 407.48 ± 2.14 | 418.77 ± 3.50 | 45.0 |
| Contr_l + CrossE_L + PMI_L + Triplet_L | GIN | 385.99 ± 17.13 | 322.96 ± 4.45 | 372.59 ± 10.82 | 339.58 ± 2.61 | 112.0 |
| Contr_l + CrossE_L + PMI_L + Triplet_L | MPNN | 416.77 ± 1.87 | 405.44 ± 1.89 | 397.61 ± 4.33 | 404.39 ± 5.03 | 68.75 |
| Contr_l + CrossE_L + PMI_L + Triplet_L | PAGNN | 348.60 ± 12.15 | 207.45 ± 1.61 | 327.56 ± 3.38 | 217.73 ± 2.49 | 161.25 |
| Contr_l + CrossE_L + PMI_L + Triplet_L | SAGE | 435.67 ± 7.30 | 428.89 ± 2.23 | 423.00 ± 4.79 | 435.89 ± 2.18 | 31.25 |
| Contr_l + CrossE_L + PR_L | ALL | 141.43 ± 10.78 | 60.42 ± 1.08 | 207.46 ± 17.64 | 88.98 ± 0.76 | 200.5 |
| Contr_l + CrossE_L + PR_L | GAT | 402.52 ± 27.52 | 408.52 ± 3.20 | 353.08 ± 30.09 | 359.64 ± 5.61 | 96.5 |

Table 42. Results for Rankme (↑) (continued)

| Loss Type | Model | Cora ↓ Citeseer | Cora ↓ Bitcoin | Citeseer ↓ Cora | Citeseer ↓ Bitcoin | Average Rank |
|---|---|---|---|---|---|---|
| Contr_l + CrossE_L + PR_L | GCN | 399.05 ± 9.62 | 395.73 ± 7.47 | 379.00 ± 7.70 | 383.15 ± 3.61 | 90.5 |
| Contr_l + CrossE_L + PR_L | GIN | 248.48 ± 28.36 | 208.26 ± 3.05 | 309.19 ± 16.18 | 260.95 ± 4.70 | 172.0 |
| Contr_l + CrossE_L + PR_L | MPNN | 304.40 ± 23.09 | 262.44 ± 2.41 | 309.06 ± 28.85 | 267.27 ± 5.98 | 162.5 |
| Contr_l + CrossE_L + PR_L | PAGNN | 274.14 ± 11.56 | 185.33 ± 1.41 | 262.05 ± 12.86 | 162.71 ± 0.64 | 188.25 |
| Contr_l + CrossE_L + PR_L | SAGE | 370.73 ± 16.95 | 382.60 ± 4.51 | 372.92 ± 19.38 | 388.32 ± 6.43 | 103.5 |
| Contr_l + CrossE_L + PR_L + Triplet_L | ALL | 241.49 ± 13.00 | 150.77 ± 6.74 | 269.51 ± 19.56 | 240.32 ± 13.38 | 185.0 |
| Contr_l + CrossE_L + PR_L + Triplet_L | GAT | 428.09 ± 5.22 | 409.18 ± 3.58 | 419.36 ± 3.83 | 423.03 ± 2.02 | 43.5 |
| Contr_l + CrossE_L + PR_L + Triplet_L | GCN | 412.88 ± 3.02 | 411.27 ± 3.10 | 392.72 ± 7.99 | 403.89 ± 7.58 | 71.25 |
| Contr_l + CrossE_L + PR_L + Triplet_L | GIN | 347.48 ± 10.14 | 270.43 ± 5.84 | 346.38 ± 7.20 | 325.47 ± 4.79 | 139.0 |
| Contr_l + CrossE_L + PR_L + Triplet_L | MPNN | 371.17 ± 5.60 | 337.01 ± 5.68 | 367.45 ± 4.76 | 322.01 ± 4.51 | 120.0 |
| Contr_l + CrossE_L + PR_L + Triplet_L | PAGNN | 300.31 ± 6.05 | 176.48 ± 1.27 | 302.36 ± 4.53 | 198.13 ± 1.17 | 181.0 |
| Contr_l + CrossE_L + PR_L + Triplet_L | SAGE | 379.71 ± 6.99 | 385.88 ± 3.17 | 385.57 ± 6.92 | 393.23 ± 3.08 | 96.75 |
| Contr_l + CrossE_L + Triplet_L | ALL | 297.80 ± 28.33 | 265.50 ± 7.05 | 274.38 ± 13.50 | 227.71 ± 5.23 | 169.75 |
| Contr_l + CrossE_L + Triplet_L | GAT | 441.68 ± 3.48 | 431.28 ± 2.52 | 435.93 ± 1.98 | 429.32 ± 1.88 | 27.0 |
| Contr_l + CrossE_L + Triplet_L | GCN | 419.73 ± 4.20 | 410.88 ± 3.80 | 409.18 ± 5.10 | 413.26 ± 2.28 | 54.875 |
| Contr_l + CrossE_L + Triplet_L | GIN | 384.69 ± 6.97 | 344.34 ± 4.23 | 372.68 ± 8.60 | 362.94 ± 2.17 | 107.25 |

Table 42. Results for Rankme (↑) (continued)

| Loss Type | Model | Cora ↓ Citeseer | Cora ↓ Bitcoin | Citeseer ↓ Cora | Citeseer ↓ Bitcoin | Average Rank |
|---|---|---|---|---|---|---|
| Contr_l + CrossE_L + Triplet_L | MPNN | 400.76 ± 6.08 | 369.91 ± 4.16 | 386.20 ± 4.84 | 383.94 ± 3.16 | 90.75 |
| Contr_l + CrossE_L + Triplet_L | PAGNN | 367.65 ± 6.84 | 249.62 ± 2.49 | 348.37 ± 9.04 | 285.57 ± 4.21 | 139.75 |
| Contr_l + CrossE_L + Triplet_L | SAGE | 396.31 ± 4.30 | 401.19 ± 3.32 | 392.93 ± 3.19 | 406.80 ± 4.10 | 79.5 |
| Contr_l + PMI_L | ALL | 328.44 ± 11.37 | 233.60 ± 3.80 | 337.54 ± 12.59 | 320.98 ± 4.48 | 149.5 |
| Contr_l + PMI_L | GAT | 453.86 ± 1.70 | 440.47 ± 1.76 | 445.12 ± 1.78 | 441.63 ± 0.88 | 8.5 |
| Contr_l + PMI_L | GCN | 424.08 ± 1.56 | 417.61 ± 2.72 | 407.46 ± 3.77 | 418.71 ± 2.04 | 47.5 |
| Contr_l + PMI_L | GIN | 387.31 ± 9.27 | 323.83 ± 2.82 | 368.59 ± 11.48 | 323.08 ± 6.60 | 115.5 |
| Contr_l + PMI_L | MPNN | 414.58 ± 3.26 | 402.21 ± 2.53 | 400.17 ± 2.83 | 408.22 ± 3.42 | 71.25 |
| Contr_l + PMI_L | PAGNN | 345.19 ± 5.58 | 193.42 ± 1.28 | 310.31 ± 7.67 | 192.23 ± 0.99 | 173.0 |
| Contr_l + PMI_L | SAGE | 448.46 ± 1.03 | 443.69 ± 1.48 | 431.12 ± 4.63 | 444.90 ± 1.33 | 14.75 |
| Contr_l + PMI_L + PR_L | ALL | 297.46 ± 11.00 | 133.07 ± 4.51 | 342.84 ± 4.62 | 277.51 ± 3.25 | 166.0 |
| Contr_l + PMI_L + PR_L | GAT | 452.53 ± 2.10 | 439.73 ± 1.89 | 433.55 ± 6.98 | 419.69 ± 6.30 | 22.5 |
| Contr_l + PMI_L + PR_L | GCN | 418.58 ± 5.47 | 414.20 ± 3.74 | 401.76 ± 6.79 | 410.28 ± 4.84 | 59.25 |
| Contr_l + PMI_L + PR_L | GIN | 381.97 ± 10.19 | 271.27 ± 6.21 | 345.86 ± 22.05 | 280.17 ± 6.17 | 137.0 |
| Contr_l + PMI_L + PR_L | MPNN | 416.09 ± 2.80 | 399.45 ± 2.48 | 388.13 ± 4.90 | 336.51 ± 3.89 | 88.25 |
| Contr_l + PMI_L + PR_L | PAGNN | 342.18 ± 7.61 | 192.03 ± 1.37 | 321.17 ± 12.47 | 216.46 ± 2.33 | 169.25 |

Table 42. Results for Rankme (↑) (continued)

| Loss Type | Model | Cora ↓ Citeseer | Cora ↓ Bitcoin | Citeseer ↓ Cora | Citeseer ↓ Bitcoin | Average Rank |
|---|---|---|---|---|---|---|
| Contr_l + PMI_L + PR_L | SAGE | 450.57 ± 2.80 | 448.37 ± 1.03 | 429.49 ± 8.09 | 428.11 ± 3.74 | 19.0 |
| Contr_l + PMI_L + ALL PR_L + Triplet_L | ALL | 318.07 ± 14.72 | 210.19 ± 5.84 | 298.67 ± 9.84 | 247.56 ± 5.84 | 168.25 |
| Contr_l + PMI_L + PR_L + Triplet_L | GAT | 443.36 ± 5.94 | 433.17 ± 3.89 | 431.77 ± 4.41 | 425.79 ± 3.37 | 28.0 |
| Contr_l + PMI_L + PR_L + Triplet_L | GCN | 421.82 ± 3.57 | 414.05 ± 2.16 | 405.54 ± 4.13 | 419.44 ± 0.64 | 51.5 |
| Contr_l + PMI_L + PR_L + Triplet_L | GIN | 391.05 ± 7.73 | 312.66 ± 5.73 | 364.33 ± 13.11 | 328.86 ± 6.10 | 115.0 |
| Contr_l + PMI_L + PR_L + Triplet_L | MPNN | 416.91 ± 2.36 | 402.69 ± 2.39 | 392.88 ± 6.30 | 341.76 ± 4.20 | 82.75 |
| Contr_l + PMI_L + PR_L + Triplet_L | PAGNN | 350.46 ± 7.47 | 205.92 ± 1.50 | 343.86 ± 6.29 | 224.34 ± 1.31 | 154.75 |
| Contr_l + PMI_L + PR_L + Triplet_L | SAGE | 415.50 ± 8.02 | 426.43 ± 0.92 | 407.03 ± 8.10 | 409.53 ± 4.36 | 54.75 |
| Contr_l + PR_L | ALL | 147.73 ± 9.58 | 59.87 ± 1.71 | 202.37 ± 9.39 | 116.65 ± 2.21 | 200.25 |
| Contr_l + PR_L | GAT | 384.60 ± 34.50 | 408.67 ± 2.66 | 356.00 ± 11.13 | 373.51 ± 8.58 | 99.75 |
| Contr_l + PR_L | GCN | 399.09 ± 7.69 | 402.36 ± 1.88 | 375.11 ± 20.23 | 401.11 ± 6.13 | 85.25 |
| Contr_l + PR_L | GIN | 256.56 ± 13.87 | 191.94 ± 3.72 | 306.43 ± 22.04 | 282.85 ± 2.94 | 173.25 |
| Contr_l + PR_L | MPNN | 273.97 ± 21.27 | 201.26 ± 5.95 | 302.06 ± 19.86 | 283.48 ± 2.05 | 169.75 |
| Contr_l + PR_L | PAGNN | 268.38 ± 16.64 | 183.25 ± 0.82 | 261.77 ± 11.77 | 159.05 ± 1.40 | 190.0 |
| Contr_l + PR_L | SAGE | 368.94 ± 10.85 | 374.66 ± 3.40 | 363.61 ± 4.33 | 380.64 ± 8.92 | 111.25 |
| Contr_l + PR_L + Triplet_L | ALL | 234.38 ± 4.99 | 132.42 ± 6.60 | 270.00 ± 19.59 | 249.51 ± 3.88 | 186.0 |

Table 42. Results for Rankme (↑) (continued)

| Loss Type | Model | Cora ↓ Citeseer | Cora ↓ Bitcoin | Citeseer ↓ Cora | Citeseer ↓ Bitcoin | Average Rank |
|---|---|---|---|---|---|---|
| Contr_l + PR_L + Triplet_L | GAT | 425.22 ± 3.44 | 412.51 ± 2.94 | 419.39 ± 6.74 | 426.24 ± 2.11 | 41.0 |
| Contr_l + PR_L + Triplet_L | GCN | 414.41 ± 4.99 | 407.63 ± 4.13 | 398.16 ± 5.90 | 400.74 ± 4.74 | 72.0 |
| Contr_l + PR_L + Triplet_L | GIN | 348.48 ± 15.04 | 327.28 ± 2.64 | 344.28 ± 13.07 | 313.95 ± 3.68 | 134.0 |
| Contr_l + PR_L + Triplet_L | MPNN | 383.45 ± 8.30 | 326.48 ± 9.98 | 365.04 ± 10.88 | 327.79 ± 3.30 | 117.25 |
| Contr_l + PR_L + Triplet_L | PAGNN | 312.71 ± 12.01 | 191.78 ± 0.93 | 311.35 ± 20.84 | 237.73 ± 1.71 | 172.0 |
| Contr_l + PR_L + Triplet_L | SAGE | 377.91 ± 1.83 | 389.35 ± 4.68 | 381.85 ± 10.84 | 397.65 ± 4.23 | 97.0 |
| Contr_l + Triplet_L | ALL | 293.61 ± 15.70 | 232.86 ± 3.73 | 284.30 ± 27.59 | 268.02 ± 6.30 | 168.25 |
| Contr_l + Triplet_L | GAT | 442.92 ± 3.11 | 425.95 ± 2.37 | 435.57 ± 2.94 | 433.31 ± 1.79 | 27.25 |
| Contr_l + Triplet_L | GCN | 418.22 ± 3.47 | 410.70 ± 3.54 | 411.28 ± 3.59 | 415.76 ± 1.88 | 54.5 |
| Contr_l + Triplet_L | GIN | 386.80 ± 3.49 | 345.37 ± 1.93 | 371.91 ± 8.89 | 361.47 ± 2.31 | 106.75 |
| Contr_l + Triplet_L | MPNN | 404.33 ± 4.94 | 382.89 ± 2.51 | 391.17 ± 9.71 | 390.51 ± 3.77 | 86.0 |
| Contr_l + Triplet_L | PAGNN | 365.67 ± 8.55 | 238.04 ± 2.88 | 338.76 ± 8.13 | 280.25 ± 6.37 | 145.5 |
| Contr_l + Triplet_L | SAGE | 393.70 ± 6.09 | 395.24 ± 3.35 | 390.53 ± 5.33 | 400.48 ± 3.11 | 85.75 |
| CrossE_L | ALL | 56.06 ± 21.34 | 88.79 ± 1.80 | 37.46 ± 14.74 | 44.93 ± 1.21 | 202.75 |
| CrossE_L | GAT | 5.34±3.10 | 1.02±0.02 | 3.95±2.12 | 6.21±0.19 | 209.0 |
| CrossE_L | GCN | 22.91 ± 7.78 | 44.59 ± 1.29 | 12.95 ± 2.91 | 21.13 ± 1.00 | 206.0 |
| CrossE_L | GIN | 2.08±1.84 | 1.58±0.01 | 3.25±1.27 | 1.41±0.00 | 209.75 |

Table 42. Results for Rankme (↑) (continued)

| Loss Type | Model | Cora ↓ Citeseer | Cora ↓ Bitcoin | Citeseer ↓ Cora | Citeseer ↓ Bitcoin | Average Rank |
|---|---|---|---|---|---|---|
| CrossE_L | MPNN | 27.41 ± 3.91 | 11.24 ± 0.95 | 21.14 ± 2.74 | 9.45±0.54 | 207.0 |
| CrossE_L | PAGNN | 50.88 ± 2.22 | 71.61 ± 0.15 | 33.43 ± 4.52 | 15.25 ± 0.32 | 204.25 |
| CrossE_L | SAGE | 8.64±1.99 | 14.84 ± 0.04 | 3.80±2.89 | 19.56 ± 0.14 | 207.5 |
| CrossE_L + PMI_L | ALL | 366.72 ± 7.95 | 277.79 ± 2.71 | 358.24 ± 4.09 | 326.74 ± 3.02 | 130.5 |
| CrossE_L + PMI_L | GAT | 453.79 ± 1.32 | 441.52 ± 0.81 | 444.10 ± 2.24 | 442.22 ± 1.05 | 7.75 |
| CrossE_L + PMI_L | GCN | 422.48 ± 3.86 | 415.37 ± 2.83 | 407.95 ± 4.21 | 416.71 ± 1.97 | 50.5 |
| CrossE_L + PMI_L | GIN | 391.29 ± 3.73 | 306.24 ± 4.94 | 367.77 ± 12.13 | 311.15 ± 3.18 | 117.5 |
| CrossE_L + PMI_L | MPNN | 413.09 ± 2.97 | 400.22 ± 3.62 | 397.75 ± 1.12 | 408.94 ± 2.39 | 74.25 |
| CrossE_L + PMI_L | PAGNN | 350.98 ± 6.54 | 199.06 ± 1.53 | 318.54 ± 5.62 | 193.42 ± 2.47 | 166.0 |
| CrossE_L + PMI_L | SAGE | 448.32 ± 2.20 | 445.05 ± 1.81 | 443.52 ± 4.62 | **448.33 ± 0.84** | 9.75 |
| CrossE_L + PMI_L + PR_L | ALL | 302.72 ± 8.80 | 152.35 ± 3.78 | 338.84 ± 4.86 | 274.62 ± 2.27 | 165.5 |
| CrossE_L + PMI_L + PR_L | GAT | 453.40 ± 0.70 | 440.42 ± 1.16 | 430.05 ± 17.69 | 413.26 ± 7.91 | 26.875 |
| CrossE_L + PMI_L + PR_L | GCN | 424.11 ± 2.34 | 418.60 ± 3.54 | 403.74 ± 6.44 | 419.43 ± 2.65 | 47.25 |
| CrossE_L + PMI_L + PR_L | GIN | 381.71 ± 6.74 | 297.02 ± 2.73 | 337.75 ± 23.48 | 309.77 ± 6.22 | 134.0 |
| CrossE_L + PMI_L + PR_L | MPNN | 414.87 ± 1.67 | 405.25 ± 1.91 | 386.17 ± 7.75 | 401.45 ± 4.42 | 77.5 |
| CrossE_L + PMI_L + PR_L | PAGNN | 348.91 ± 9.44 | 205.31 ± 0.87 | 321.83 ± 4.35 | 204.88 ± 1.31 | 163.75 |

Table 42. Results for Rankme (↑) (continued)

| Loss Type | Model | Cora ↓ Citeseer | Cora ↓ Bitcoin | Citeseer ↓ Cora | Citeseer ↓ Bitcoin | Average Rank |
|---|---|---|---|---|---|---|
| CrossE_L + PMI_L + PR_L | SAGE | 452.50 ± 2.08 | 448.63 ± 2.04 | 443.14 ± 4.74 | 449.12 ± 1.08 | 6.125 |
| CrossE_L + PMI_L + PR_L + Triplet_L | ALL | 342.11 ± 12.55 | 260.89 ± 3.83 | 329.80 ± 5.09 | 284.10 ± 8.10 | 151.75 |
| CrossE_L + PMI_L + PR_L + Triplet_L | GAT | 452.51 ± 4.27 | 439.49 ± 1.66 | 439.39 ± 4.05 | 439.48 ± 1.04 | 15.25 |
| CrossE_L + PMI_L + PR_L + Triplet_L | GCN | 424.82 ± 3.35 | 415.81 ± 2.99 | 407.52 ± 4.08 | 418.61 ± 1.41 | 46.75 |
| CrossE_L + PMI_L + PR_L + Triplet_L | GIN | 384.81 ± 9.70 | 297.35 ± 2.75 | 370.85 ± 12.63 | 328.48 ± 5.64 | 117.25 |
| CrossE_L + PMI_L + PR_L + Triplet_L | MPNN | 414.03 ± 2.30 | 400.89 ± 3.08 | 396.83 ± 2.02 | 391.22 ± 4.61 | 79.0 |
| CrossE_L + PMI_L + PR_L + Triplet_L | PAGNN | 349.33 ± 6.35 | 211.90 ± 1.41 | 330.85 ± 3.05 | 226.39 ± 1.11 | 157.0 |
| CrossE_L + PMI_L + PR_L + Triplet_L | SAGE | 439.71 ± 4.47 | 432.53 ± 5.26 | 424.14 ± 4.93 | 431.74 ± 2.53 | 29.25 |
| CrossE_L + PMI_L + Triplet_L | ALL | 366.91 ± 8.38 | 350.33 ± 2.79 | 350.28 ± 4.78 | 372.35 ± 2.50 | 117.5 |
| CrossE_L + PMI_L + Triplet_L | GAT | 454.02 ± 0.73 | 441.37 ± 2.70 | 444.31 ± 1.22 | 442.13 ± 1.07 | 7.25 |
| CrossE_L + PMI_L + Triplet_L | GCN | 423.30 ± 2.40 | 421.75 ± 1.94 | 408.73 ± 3.64 | 417.52 ± 3.07 | 44.75 |
| CrossE_L + PMI_L + Triplet_L | GIN | 394.42 ± 4.76 | 321.09 ± 4.95 | 374.34 ± 10.08 | 340.92 ± 4.53 | 107.25 |
| CrossE_L + PMI_L + Triplet_L | MPNN | 415.37 ± 3.35 | 410.17 ± 4.15 | 398.20 ± 1.69 | 403.35 ± 4.72 | 66.5 |
| CrossE_L + PMI_L + Triplet_L | PAGNN | 350.72 ± 8.94 | 204.80 ± 0.85 | 327.09 ± 6.58 | 191.06 ± 1.62 | 164.25 |
| CrossE_L + PMI_L + Triplet_L | SAGE | 435.98 ± 3.89 | 435.66 ± 2.83 | 423.12 ± 4.12 | 437.08 ± 2.28 | 28.0 |
| CrossE_L + PR_L | ALL | 125.37 ± 16.50 | 46.80 ± 0.99 | 189.19 ± 25.11 | 65.70 ± 1.09 | 202.5 |

Continued on next page

Table 42. Results for Rankme (↑) (continued)

| Loss Type | Model | Cora ↓ Citeseer | Cora ↓ Bitcoin | Citeseer ↓ Cora | Citeseer ↓ Bitcoin | Average Rank |
|---|---|---|---|---|---|---|
| CrossE_L + PR_L | GAT | 308.00 ± 39.67 | 307.18 ± 6.77 | 268.57 ± 12.46 | 292.74 ± 6.85 | 157.0 |
| CrossE_L + PR_L | GCN | 376.96 ± 19.44 | 357.23 ± 12.75 | 329.30 ± 9.88 | 352.16 ± 17.37 | 123.5 |
| CrossE_L + PR_L | GIN | 208.61 ± 22.46 | 174.44 ± 1.96 | 254.43 ± 50.23 | 184.63 ± 5.08 | 192.75 |
| CrossE_L + PR_L | MPNN | 279.10 ± 58.69 | 168.02 ± 4.30 | 286.10 ± 18.30 | 237.33 ± 3.80 | 180.5 |
| CrossE_L + PR_L | PAGNN | 247.57 ± 30.15 | 167.72 ± 0.79 | 257.38 ± 18.61 | 162.22 ± 0.59 | 192.25 |
| CrossE_L + PR_L | SAGE | 369.63 ± 19.45 | 361.93 ± 7.89 | 364.92 ± 24.41 | 355.79 ± 4.98 | 114.25 |
| CrossE_L + PR_L + Triplet_L | ALL | 239.66 ± 27.67 | 96.95 ± 2.13 | 231.35 ± 12.79 | 219.18 ± 1.82 | 191.5 |
| CrossE_L + PR_L + Triplet_L | GAT | 424.42 ± 5.26 | 418.20 ± 2.35 | 409.42 ± 11.81 | 417.65 ± 1.89 | 44.25 |
| CrossE_L + PR_L + Triplet_L | GCN | 404.97 ± 4.89 | 402.09 ± 2.54 | 390.99 ± 10.16 | 401.92 ± 6.69 | 79.25 |
| CrossE_L + PR_L + Triplet_L | GIN | 330.20 ± 9.22 | 277.19 ± 4.64 | 340.89 ± 7.22 | 310.59 ± 2.16 | 145.75 |
| CrossE_L + PR_L + Triplet_L | MPNN | 340.64 ± 14.29 | 301.36 ± 4.82 | 350.14 ± 12.36 | 315.84 ± 3.32 | 138.5 |
| CrossE_L + PR_L + Triplet_L | PAGNN | 282.18 ± 19.53 | 190.13 ± 0.50 | 294.25 ± 16.94 | 187.11 ± 1.59 | 183.5 |
| CrossE_L + PR_L + Triplet_L | SAGE | 388.77 ± 2.48 | 391.54 ± 3.52 | 382.60 ± 4.93 | 391.47 ± 1.79 | 91.5 |
| CrossE_L + Triplet_L | ALL | 327.51 ± 14.87 | 322.69 ± 4.26 | 309.20 ± 17.38 | 311.34 ± 6.19 | 148.25 |
| CrossE_L + Triplet_L | GAT | 448.14 ± 1.48 | 433.47 ± 0.74 | 437.91 ± 1.85 | 437.44 ± 1.08 | 21.5 |
| CrossE_L + Triplet_L | GCN | 423.61 ± 3.14 | 417.50 ± 2.29 | 411.59 ± 4.05 | 415.84 ± 2.05 | 47.25 |

Table 42. Results for Rankme (↑) (continued)

| Loss Type | Model | Cora ↓ Citeseer | | Cora ↓ Bitcoin | | Citeseer ↓ Cora | | Citeseer ↓ Bitcoin | | Average Rank |
|---|---|---|---|---|---|---|---|---|---|---|
| CrossE_L + Triplet_L | GIN | 384.80 | ± 6.86 | 337.66 | ± 2.79 | 370.71 | ± 3.73 | 352.96 | ± 1.98 | 109.75 |
| CrossE_L + Triplet_L | MPNN | 414.09 | ± 2.99 | 398.81 | ± 3.88 | 397.53 | ± 5.96 | 399.69 | ± 2.77 | 78.5 |
| CrossE_L + Triplet_L | PAGNN | 372.53 | ± 8.79 | 261.35 | ± 2.53 | 353.46 | ± 9.05 | 305.57 | ± 1.86 | 135.25 |
| CrossE_L + Triplet_L | SAGE | 410.08 | ± 2.46 | 409.64 | ± 2.15 | 410.57 | ± 5.06 | 411.30 | ± 2.26 | 62.25 |
| PMI_L | ALL | 364.09 | ± 3.09 | 274.61 | ± 3.04 | 362.93 | ± 6.94 | 334.14 | ± 5.09 | 129.5 |
| PMI_L | GAT | 453.97 ± 1.38 | | 442.60 | ± 1.32 | 444.74 ± 1.47 | | 442.89 | ± 1.32 | 5.75 |
| PMI_L | GCN | 420.80 | ± 4.25 | 417.08 | ± 3.75 | 407.90 | ± 1.83 | 421.65 | ± 3.67 | 47.5 |
| PMI_L | GIN | 388.74 | ± 7.70 | 303.87 | ± 4.17 | 369.00 | ± 5.84 | 309.07 | ± 4.54 | 120.25 |
| PMI_L | MPNN | 415.53 | ± 3.97 | 410.95 | ± 2.63 | 394.99 | ± 3.07 | 407.41 | ± 3.22 | 66.25 |
| PMI_L | PAGNN | 346.00 | ± 3.61 | 200.91 | ± 1.76 | 317.26 | ± 4.79 | 202.45 | ± 1.31 | 167.5 |
| PMI_L | SAGE | 447.39 | ± 2.10 | 447.26 | ± 1.73 | 440.20 | ± 4.65 | 443.02 | ± 1.42 | 11.75 |
| PMI_L + PR_L | ALL | 318.24 | ± 13.86 | 146.85 | ± 2.39 | 328.45 | ± 23.24 | 277.85 | ± 1.34 | 167.25 |
| PMI_L + PR_L | GAT | 450.29 | ± 1.73 | 433.33 | ± 2.24 | 432.78 | ± 8.55 | 429.30 | ± 3.29 | 24.25 |
| PMI_L + PR_L | GCN | 422.87 | ± 3.79 | 419.61 | ± 2.93 | 406.48 | ± 10.02 | 422.32 | ± 1.84 | 46.0 |
| PMI_L + PR_L | GIN | 386.83 | ± 3.32 | 304.27 | ± 4.84 | 351.19 | ± 28.47 | 333.57 | ± 4.59 | 121.0 |
| PMI_L + PR_L | MPNN | 414.75 | ± 3.44 | 406.55 | ± 2.53 | 382.99 | ± 4.25 | 329.58 | ± 7.14 | 88.5 |

Table 42.  Results for Rankme (↑) (continued)

| Loss Type | Model | Cora ↓ Citeseer | Cora ↓ Bitcoin | Citeseer ↓ Cora | Citeseer ↓ Bitcoin | Average Rank |
|---|---|---|---|---|---|---|
| PMI_L + PR_L | PAGNN | 348.02 ± 6.02 | 198.34 ± 1.22 | 324.36 ± 5.74 | 200.18 ± 1.49 | 167.0 |
| PMI_L + PR_L | SAGE | 450.06 ± 3.00 | 445.57 ± 0.85 | 443.14 ± 4.11 | 448.21 ± 0.71 | 9.125 |
| PMI_L + PR_L + Triplet_L | ALL | 324.43 ± 18.27 | 198.86 ± 4.10 | 323.40 ± 4.54 | 273.70 ± 8.81 | 163.5 |
| PMI_L + PR_L + Triplet_L | GAT | 450.45 ± 2.28 | 439.53 ± 0.57 | 438.43 ± 5.87 | 437.45 ± 2.36 | 17.25 |
| PMI_L + PR_L + Triplet_L | GCN | 419.96 ± 3.00 | 415.45 ± 2.92 | 408.56 ± 5.79 | 416.38 ± 3.78 | 51.75 |
| PMI_L + PR_L + Triplet_L | GIN | 390.31 ± 5.47 | 305.14 ± 3.52 | 366.46 ± 11.23 | 315.37 ± 4.73 | 118.0 |
| PMI_L + PR_L + Triplet_L | MPNN | 415.22 ± 2.79 | 406.19 ± 1.11 | 389.91 ± 3.68 | 351.51 ± 8.05 | 83.625 |
| PMI_L + PR_L + Triplet_L | PAGNN | 348.76 ± 3.70 | 193.75 ± 1.36 | 333.55 ± 8.28 | 231.82 ± 1.62 | 160.75 |
| PMI_L + PR_L + Triplet_L | SAGE | 434.07 ± 6.61 | 437.08 ± 2.40 | 427.75 ± 5.91 | 434.27 ± 1.29 | 28.0 |
| PMI_L + Triplet_L | ALL | 367.36 ± 7.15 | 333.12 ± 3.54 | 363.95 ± 6.46 | 380.02 ± 2.93 | 114.75 |
| PMI_L + Triplet_L | GAT | 452.83 ± 0.80 | 440.77 ± 1.11 | 445.49 ± 1.23 | 442.39 ± 2.32 | 8.25 |
| PMI_L + Triplet_L | GCN | 419.46 ± 3.45 | 414.25 ± 2.86 | 408.72 ± 3.16 | 419.79 ± 1.57 | 49.0 |
| PMI_L + Triplet_L | GIN | 382.20 ± 9.66 | 319.93 ± 1.27 | 373.77 ± 11.47 | 344.33 ± 3.78 | 113.0 |
| PMI_L + Triplet_L | MPNN | 416.40 ± 3.94 | 407.06 ± 2.46 | 399.52 ± 4.46 | 413.87 ± 2.78 | 63.5 |
| PMI_L + Triplet_L | PAGNN | 345.75 ± 8.26 | 195.71 ± 1.51 | 321.63 ± 3.55 | 209.74 ± 1.62 | 167.25 |
| PMI_L + Triplet_L | SAGE | 438.20 ± 4.12 | 432.47 ± 2.90 | 425.38 ± 3.93 | 436.11 ± 1.01 | 28.25 |

Table 42. Results for Rankme (↑) (continued)

| Loss Type | Model | Cora ↓ Citeseer | Cora ↓ Bitcoin | Citeseer ↓ Cora | Citeseer ↓ Bitcoin | Average Rank |
|---|---|---|---|---|---|---|
| PR_L | ALL | 105.31 ± 8.68 | 35.91 ± 0.45 | 156.29 ± 12.08 | 60.42 ± 1.07 | 203.75 |
| PR_L | GAT | 304.15 ± 17.32 | 289.35 ± 9.44 | 275.84 ± 26.11 | 309.09 ± 5.83 | 159.25 |
| PR_L | GCN | 345.41 ± 20.45 | 355.56 ± 12.83 | 336.69 ± 21.57 | 372.30 ± 20.60 | 126.5 |
| PR_L | GIN | 193.71 ± 27.88 | 148.80 ± 2.13 | 235.28 ± 34.98 | 201.62 ± 2.46 | 193.25 |
| PR_L | MPNN | 292.42 ± 40.67 | 217.53 ± 7.95 | 279.48 ± 17.32 | 217.95 ± 2.85 | 174.25 |
| PR_L | PAGNN | 243.95 ± 33.40 | 149.94 ± 0.68 | 268.06 ± 10.28 | 153.50 ± 0.35 | 193.25 |
| PR_L | SAGE | 369.89 ± 28.03 | 385.39 ± 2.70 | 337.67 ± 26.77 | 389.14 ± 5.85 | 114.5 |
| PR_L + Triplet_L | ALL | 144.21 ± 12.64 | 71.41 ± 1.19 | 200.90 ± 11.08 | 80.59 ± 1.00 | 200.75 |
| PR_L + Triplet_L | GAT | 399.89 ± 34.16 | 406.80 ± 4.66 | 355.73 ± 14.14 | 366.75 ± 3.51 | 95.5 |
| PR_L + Triplet_L | GCN | 393.67 ± 3.65 | 387.13 ± 5.45 | 374.75 ± 10.49 | 389.71 ± 4.56 | 93.75 |
| PR_L + Triplet_L | GIN | 258.37 ± 27.75 | 189.15 ± 4.08 | 316.31 ± 21.39 | 255.13 ± 2.77 | 175.25 |
| PR_L + Triplet_L | MPNN | 308.85 ± 26.51 | 240.30 ± 3.00 | 282.58 ± 21.66 | 251.74 ± 4.95 | 167.0 |
| PR_L + Triplet_L | PAGNN | 262.10 ± 14.17 | 172.70 ± 1.26 | 256.42 ± 10.34 | 161.34 ± 1.54 | 191.25 |
| PR_L + Triplet_L | SAGE | 382.10 ± 12.43 | 376.09 ± 3.29 | 369.52 ± 14.26 | 381.11 ± 7.87 | 104.25 |
| Triplet_L | ALL | 340.29 ± 14.12 | 291.79 ± 2.46 | 339.35 ± 13.71 | 358.96 ± 2.30 | 136.0 |
| Triplet_L | GAT | 448.73 ± 1.62 | 432.52 ± 2.31 | 438.89 ± 2.41 | 437.24 ± 2.63 | 21.0 |

Table 42. Results for Rankme (↑) (continued)

| Loss Type | Model | Cora ↓ Citeseer | Cora ↓ Bitcoin | Citeseer ↓ Cora | Citeseer ↓ Bitcoin | Average Rank |
|-----------|-------|-----------------|----------------|-----------------|--------------------|--------------|
| Triplet_L | GCN | 425.69 ± 1.75 | 419.75 ± 2.58 | 411.60 ± 4.31 | 416.91 ± 3.72 | 41.75 |
| Triplet_L | GIN | 396.59 ± 2.66 | 352.21 ± 1.40 | 382.08 ± 6.46 | 377.69 ± 2.28 | 96.0 |
| Triplet_L | MPNN | 415.22 ± 5.42 | 404.13 ± 4.24 | 401.35 ± 6.25 | 403.21 ± 2.25 | 70.375 |
| Triplet_L | PAGNN | 380.62 ± 8.39 | 277.88 ± 2.84 | 364.38 ± 5.44 | 323.09 ± 0.98 | 125.75 |
| Triplet_L | SAGE | 417.21 ± 6.44 | 417.66 ± 1.24 | 419.56 ± 5.11 | 427.12 ± 1.91 | 43.5 |

**Additional Notes**

For overall analysis see main manuscript. For full filtered top 10 results table see Supplementary Information 1.